\documentclass[a4paper,nobind,dvipsnames]{ociamthesis} 

\let\include\input

\usepackage[utf8]{inputenc} %
\usepackage[T1]{fontenc}    %

\usepackage{xr} %

\usepackage[dvipsnames]{xcolor}
\usepackage{amsmath}

\usepackage{accents}
\usepackage[export]{adjustbox}
\usepackage{algorithm}
\usepackage{amsfonts} %

\usepackage{amssymb}
\usepackage{amsthm}
\usepackage{booktabs} %
\usepackage{calc}
\usepackage{cancel} %
\usepackage{caption} 
\usepackage{color}
\usepackage{colortbl}
\usepackage{comment}
\usepackage{duckuments}
\usepackage{enumitem}
\usepackage{etoolbox}
\usepackage{fancyhdr}
\usepackage{fancyvrb}
\usepackage{float}
\usepackage{graphbox}
\usepackage{graphicx} %
\usepackage{hyperref}       %
\usepackage{listings}
\usepackage{longtable}
\usepackage{makecell}
\usepackage{mathcommand}
\usepackage{mathrsfs}
\usepackage{mathtools} %
\usepackage{microtype}      %
\usepackage{multicol}
\usepackage{multirow}
\usepackage{natbib} %
\usepackage{nicefrac}       %
\usepackage{pifont}
\usepackage{placeins} %
\usepackage{pmboxdraw}
\usepackage{rotating}
\usepackage{setspace}
\usepackage{subcaption}
\usepackage{tablefootnote}
\usepackage{tabularx}
\usepackage{textcomp}
\usepackage{thm-restate}
\usepackage{thmtools}
\usepackage{tikz} %
\usepackage{url}            %
\usepackage{wrapfig}
\usepackage{xcolor}
\usepackage{xfrac}
\usepackage{xspace}
\usepackage{nccmath}

\usepackage{array}
\usepackage[capitalize,noabbrev]{cleveref} %
\usepackage[colorinlistoftodos,textsize=tiny]{todonotes}
\usepackage[many]{tcolorbox}
\usepackage{dsfont} 
\usepackage{bbold} %
\usepackage{bm}

\usepackage{bibentry}

\tcbuselibrary{breakable}
\tcbuselibrary{theorems}

\def\sqrtexplained#1{%
  \begingroup
    \sbox0{$#1$}
    \def\underbrace##1_##2{##1}
    \sbox2{$#1$}
    \dimen0=\wd0 \advance\dimen0-\wd2
    \mathrlap{\sqrt{\phantom{\displaystyle#1}\kern\dimen0 }}
    \hphantom{\sqrt{\vphantom{\displaystyle#1}}}
  \endgroup
  #1}

\setlist[enumerate,1]{noitemsep,topsep=0pt,parsep=0pt,partopsep=0pt}
\setlist[itemize,1]{noitemsep,topsep=0pt,parsep=0pt,partopsep=0pt}
\setlist[description,1]{leftmargin=1em, labelindent=0.5em}

\usetikzlibrary{backgrounds}	%
\usetikzlibrary{fit}					%

\graphicspath{{./misc/figures/}}

\newenvironment{mypapercitation}{\begin{quote}}{\end{quote}}

\excludecomment{draft}
\excludecomment{nextversion}
\excludecomment{fullpaper}

\setcitestyle{square}

\usepackage[super]{nth}

\newcommand*\Circled[1]{\tikz[baseline=(char.base)]{
		\node[shape=circle,draw,inner sep=2pt] (char) {#1};}}
\newcommand*\RRectangle[1]{\tikz[baseline=(char.base)]{
		\node[
            shape=rectangle,
            rounded corners=2pt,
            draw,
            inner sep=2pt,
            minimum height=\baselineskip
        ] (char) {\footnotesize #1};}}
\theoremstyle{plain}
\newtheorem{theorem}{Theorem}[chapter]
\newtheorem{proposition}[theorem]{Proposition}
\newtheorem{lemma}[theorem]{Lemma}
\newtheorem{corollary}[theorem]{Corollary}
\theoremstyle{definition}
\newtheorem{definition}{Definition}[chapter]
\newtheorem{assumption}{Assumption}[chapter]
\newtheorem{observation}{Observation}[chapter]
\newtheorem{notation}[theorem]{Notation}
\theoremstyle{remark}
\newtheorem{remark}{Remark}[chapter]
\newtheorem{example}[remark]{Example}

\newtheorem{insight}[remark]{Insight}

\newcommand{\shorturl}[1]{\href{https://#1}{\nolinkurl{#1}}}

\newlength{\figureheight}

\creflabelformat{equation}{#2\textup{#1}#3}

\crefname{assumption}{assumption}{assumptions}

\usepackage{titlesec}

\titlespacing*\section{0pt}{14pt plus 4pt minus 2pt}{3pt plus 2pt minus 2pt}
\titlespacing*\subsection{0pt}{10pt plus 4pt minus 2pt}{3pt plus 2pt minus 2pt}
\titlespacing*\subsubsection{0pt}{10pt plus 4pt minus 2pt}{3pt plus 2pt minus 2pt}

\titlespacing{\paragraph}{0pt}{0.5em}{.5em}[]

\definecolor{MyLightGray}{gray}{0.925}
\definecolor{MyDarkGray}{gray}{0.55}
\definecolor{myblue}{rgb}{0.00, 0.45, 0.70}
\definecolor{mygreen}{rgb}{0.01, 0.62, 0.45}
\definecolor{myred}{rgb}{0.84, 0.37, 0.00}

\PassOptionsToPackage{hyphens}{url}\usepackage{hyperref}
\definecolor{blue}{RGB}{0,114,178}
\definecolor{red}{RGB}{213,80,20}
\definecolor{green}{RGB}{0,158,115}
\definecolor{purple}{RGB}{204,121,167}
\definecolor{orange}{RGB}{230,159, 0}
\definecolor{pink}{RGB}{204,121,167}

\definecolor{ourmethod}{gray}{0.93}

\definecolor{sns0}{HTML}{1f77b4}
\definecolor{sns1}{HTML}{ff7f0e}
\definecolor{sns2}{HTML}{2ca02c}
\definecolor{sns3}{HTML}{d62728}
\definecolor{sns4}{HTML}{9467bd}
\definecolor{sns5}{HTML}{8c564b}

\definecolor{sns-orange}{HTML}{ff7f0e}
\definecolor{sns-ambiguous}{HTML}{00528C}
\definecolor{sns-nonambiguous}{HTML}{2CA9FF}
\definecolor{sns-blue}{HTML}{1f77b4}
\definecolor{solarized@base03}{HTML}{002B36}
\definecolor{solarized@base02}{HTML}{073642}
\definecolor{solarized@base01}{HTML}{586e75}
\definecolor{solarized@base00}{HTML}{657b83}
\definecolor{solarized@base0}{HTML}{839496}
\definecolor{solarized@base1}{HTML}{93a1a1}
\definecolor{solarized@base2}{HTML}{EEE8D5}
\definecolor{solarized@base3}{HTML}{FDF6E3}
\definecolor{solarized@yellow}{HTML}{B58900}
\definecolor{solarized@orange}{HTML}{CB4B16}
\definecolor{solarized@red}{HTML}{DC322F}
\definecolor{solarized@magenta}{HTML}{D33682}
\definecolor{solarized@violet}{HTML}{6C71C4}
\definecolor{solarized@blue}{HTML}{268BD2}
\definecolor{solarized@cyan}{HTML}{2AA198}
\definecolor{solarized@green}{HTML}{859900}

\lstdefinestyle{mystyle}{
    backgroundcolor=\color{solarized@base3},
    rulesepcolor=\color{solarized@base03},
    numberstyle=\tiny\color{solarized@base01},
    keywordstyle=\color{solarized@green},
    stringstyle=\color{solarized@cyan}\ttfamily,
    identifierstyle=\color{solarized@blue},
    commentstyle=\color{solarized@magenta},
    emphstyle=\color{solarized@red},
    basicstyle=\ttfamily\scriptsize\color{solarized@base0},
    breakatwhitespace=false,         
    breaklines=true,                 
    captionpos=b,                    
    keepspaces=true,                 
    numbers=none,                    
    numbersep=5pt,                  
    showspaces=false,                
    showstringspaces=false,
    showtabs=false,                  
    tabsize=2
}

\crefname{lstlisting}{listing}{listings}
\Crefname{lstlisting}{Listing}{Listings}

\crefformat{chapter}{\S#2#1#3}
\crefrangeformat{chapter}{\S#3#1#4 to~#5#2#6}
\crefmultiformat{chapter}{\S#2#1#3}{ and \S#2#1#3}{, \S#2#1#3}{ and \S#2#1#3}
\crefrangemultiformat{chapter}{\S#3#1#4 to~#5#2#6}{ and \S#3#1#4 to~#5#2#6}{, \S#3#1#4 to~#5#2#6}{ and \S#3#1#4 to~#5#2#6}
\crefformat{section}{\S#2#1#3}
\crefrangeformat{section}{\S#3#1#4 to~#5#2#6}
\crefmultiformat{section}{\S#2#1#3}{ and \S#2#1#3}{, \S#2#1#3}{ and \S#2#1#3}
\crefrangemultiformat{section}{\S#3#1#4 to~#5#2#6}{ and \S#3#1#4 to~#5#2#6}{, \S#3#1#4 to~#5#2#6}{ and \S#3#1#4 to~#5#2#6}
\crefformat{subsubsubappendix}{\S#2#1#3}
\crefrangeformat{subsubsubappendix}{\S#3#1#4 to~#5#2#6}
\crefmultiformat{subsubsubappendix}{\S#2#1#3}{ and \S#2#1#3}{, \S#2#1#3}{ and \S#2#1#3}
\crefrangemultiformat{subsubsubappendix}{\S#3#1#4 to~#5#2#6}{ and \S#3#1#4 to~#5#2#6}{, \S#3#1#4 to~#5#2#6}{ and \S#3#1#4 to~#5#2#6}

\VerbatimFootnotes

\hypersetup{
    colorlinks,
    linkcolor={solarized@base02},
    citecolor={solarized@yellow},
    urlcolor={solarized@blue}
}

\allowdisplaybreaks
\renewcommand{\arraystretch}{1.5}

\providecommand\given{\MidSymbol[\vert]}

\newcommand\MidSymbol[1][]{%
\nonscript\:#1
\allowbreak
\nonscript\:
\mathopen{}}

\DeclareMathOperator{\opVar}{\mathrm{Var}}

\DeclarePairedDelimiterXPP{\Var}[2]{\opVar_{#1}}{[}{]}{}{%
    \renewcommand\given{\MidSymbol[\delimsize\vert]}
    \ifblank{#2}{\:\cdot\:}{#2}
}

\DeclarePairedDelimiterXPP{\implicitVar}[1]{\opVar}{[}{]}{}{%
    \renewcommand\given{\MidSymbol[\delimsize\vert]}
    \ifblank{#1}{\:\cdot\:}{#1}
}

\DeclareMathOperator{\opExpectation}{\mathbb{E}}

\DeclarePairedDelimiterXPP{\implicitE}[1]{\opExpectation}{[}{]}{}{%
    \renewcommand\given{\MidSymbol[\delimsize\vert]}
    \ifblank{#1}{\:\cdot\:}{#1}
}

\DeclarePairedDelimiterXPP{\E}[2]{\opExpectation_{#1}}{[}{]}{}{%
    \renewcommand\given{\MidSymbol[\delimsize\vert]}
    \ifblank{#2}{\:\cdot\:}{#2}
}

\newcommand{\simpleE}[1]{\opExpectation_{#1}} %

\DeclareMathOperator{\opCovariance}{\mathrm{Cov}}

\DeclarePairedDelimiterXPP{\implicitCov}[1]{\opCovariance}{[}{]}{}{%
    \renewcommand\given{\MidSymbol[\delimsize\vert]}
    \ifblank{#1}{\:\cdot\:}{#1}
}

\DeclarePairedDelimiterXPP{\Cov}[2]{\opCovariance_{#1}}{[}{]}{}{%
    \renewcommand\given{\MidSymbol[\delimsize\vert]}
    \ifblank{#2}{\:\cdot\:}{#2}
}

\DeclarePairedDelimiterXPP{\emCov}[2]{\widehat{\opCovariance}_{#1}}{[}{]}{}{%
    \renewcommand\given{\MidSymbol[\delimsize\vert]}
    \ifblank{#2}{\:\cdot\:}{#2}
}

\DeclarePairedDelimiterXPP{\indicator}[1]{\mathbb{1}}{\{}{\}}{}{%
    \ifblank{#1}{\:\cdot\:}{#1}
}

\DeclareMathOperator{\opInformationContent}{H}
\DeclarePairedDelimiterXPP{\ICof}[1]{\opInformationContent}{(}{)}{}{%
    \ifblank{#1}{\:\cdot\:}{#1}
}

\DeclareMathOperator{\opEntropy}{H}
\DeclarePairedDelimiterXPP{\Hof}[1]{\opEntropy}{[}{]}{}{%
    \renewcommand\given{\MidSymbol[\delimsize\vert]}
    \ifblank{#1}{\:\cdot\:}{#1}
}
\DeclarePairedDelimiterXPP{\xHof}[1]{\opEntropy}{(}{)}{}{%
    \ifblank{#1}{\:\cdot\:}{#1}
}

\DeclareMathOperator{\opMI}{I}
\DeclarePairedDelimiterXPP{\MIof}[1]{\opMI}{[}{]}{}{%
    \renewcommand\given{\MidSymbol[\delimsize\vert]}
    \ifblank{#1}{\:\cdot\:}{#1}
}

\DeclareMathOperator{\opTC}{TC}
\DeclarePairedDelimiterXPP{\TCof}[1]{\opTC}{[}{]}{}{%
    \renewcommand\given{\MidSymbol[\delimsize\vert]}
    \ifblank{#1}{\:\cdot\:}{#1}
}

\DeclarePairedDelimiterXPP{\CrossEntropy}[2]{\opEntropy}{(}{)}{}{%
    \ifblank{#1#2}{\:\cdot\: \MidSymbol[\delimsize\Vert] \:\cdot\:}{#1 \MidSymbol[\delimsize\Vert] #2}
}

\DeclareMathOperator{\opKale}{D_\mathrm{KL}}
\DeclarePairedDelimiterXPP{\Kale}[2]{\opKale}{(}{)}{}{%
    \ifblank{#1#2}{\:\cdot\: \MidSymbol[\delimsize\Vert] \:\cdot\:}{#1 \MidSymbol[\delimsize\Vert] #2}
}

\DeclareMathOperator{\opp}{p}
\DeclarePairedDelimiterXPP{\pof}[1]{\opp}{(}{)}{}{%
    \renewcommand\given{\MidSymbol[\delimsize\vert]}
    \ifblank{#1}{\:\cdot\:}{#1}
}

\DeclarePairedDelimiterXPP{\hpof}[1]{\hat{\opp}}{(}{)}{}{%
    \renewcommand\given{\MidSymbol[\delimsize\vert]}
    \ifblank{#1}{\:\cdot\:}{#1}
}

\DeclarePairedDelimiterXPP{\fpcof}[2]{\opp{#1}}{(}{)}{}{%
    \renewcommand\given{\MidSymbol[\delimsize\vert]}
    \ifblank{#2}{\:\cdot\:}{#2}
}

\DeclarePairedDelimiterXPP{\pcof}[2]{\opp_{#1}}{(}{)}{}{%
    \renewcommand\given{\MidSymbol[\delimsize\vert]}
    \ifblank{#2}{\:\cdot\:}{#2}
}

\DeclareMathOperator{\opP}{\mathbb{P}}
\DeclarePairedDelimiterXPP{\pfor}[1]{\opP}{\lbrack}{\rbrack}{}{%
    \renewcommand\given{\MidSymbol[\delimsize\vert]}
    \ifblank{#1}{\:\cdot\:}{#1}
}

\DeclarePairedDelimiterXPP{\pcfor}[2]{\opp_{#1}}{\lbrack}{\rbrack}{}{%
    \renewcommand\given{\MidSymbol[\delimsize\vert]}
    \ifblank{#2}{\:\cdot\:}{#2}
}

\DeclarePairedDelimiterXPP{\hpcof}[2]{\hat{\opp}_{#1}}{(}{)}{}{%
    \renewcommand\given{\MidSymbol[\delimsize\vert]}
    \ifblank{#2}{\:\cdot\:}{#2}
}

\DeclareMathOperator{\opq}{q}
\DeclarePairedDelimiterXPP{\qof}[1]{\opq}{(}{)}{}{%
    \renewcommand\given{\MidSymbol[\delimsize\vert]}
    \ifblank{#1}{\:\cdot\:}{#1}
}

\DeclarePairedDelimiterXPP{\qcof}[2]{\opq_{#1}}{(}{)}{}{%
    \renewcommand\given{\MidSymbol[\delimsize\vert]}
    \ifblank{#2}{\:\cdot\:}{#2}
}

\DeclarePairedDelimiterXPP{\qcpof}[2]{\opq'_{#1}}{(}{)}{}{%
    \renewcommand\given{\MidSymbol[\delimsize\vert]}
    \ifblank{#2}{\:\cdot\:}{#2}
}

\DeclarePairedDelimiterXPP{\varHof}[2]{\opEntropy_{\ifblank{#1}{\:\cdot\:}{#1}}}{[}{]}{}{%
    \renewcommand\given{\MidSymbol[\delimsize\vert]}
    \ifblank{#2}{\:\cdot\:}{#2}
}

\DeclarePairedDelimiterXPP{\xvarHof}[2]{\opEntropy_{\ifblank{#1}{\:\cdot\:}{#1}}}{(}{)}{}{%
    \renewcommand\given{\MidSymbol[\delimsize\vert]}
    \ifblank{#2}{\:\cdot\:}{#2}
}

\DeclarePairedDelimiterXPP{\varMIof}[2]{\opMI_{\ifblank{#1}{\:\cdot\:}{#1}}}{[}{]}{}{%
    \renewcommand\given{\MidSymbol[\delimsize\vert]}
    \ifblank{#2}{\:\cdot\:}{#2}
}

\DeclareMathOperator{\opmus}{\mu^*}
\DeclarePairedDelimiterXPP{\IMof}[1]{\opmus}{[}{]}{}{%
    \renewcommand\given{\MidSymbol[\delimsize\vert]}
    \ifblank{#1}{\:\cdot\:}{#1}
}

\DeclarePairedDelimiterXPP{\aICof}[1]{\hat \opInformationContent}{(}{)}{}{%
    \ifblank{#1}{\:\cdot\:}{#1}
}

\DeclarePairedDelimiterXPP{\aHof}[1]{\hat \opEntropy}{[}{]}{}{%
    \renewcommand\given{\MidSymbol[\delimsize\vert]}
    \ifblank{#1}{\:\cdot\:}{#1}
}
\DeclarePairedDelimiterXPP{\axHof}[1]{\hat \opEntropy}{(}{)}{}{%
    \ifblank{#1}{\:\cdot\:}{#1}
}

\DeclarePairedDelimiterXPP{\aMIof}[1]{\hat \opMI}{[}{]}{}{%
    \renewcommand\given{\MidSymbol[\delimsize\vert]}
    \ifblank{#1}{\:\cdot\:}{#1}
}

\DeclarePairedDelimiterXPP{\aCrossEntropy}[2]{\hat \opEntropy}{(}{)}{}{%
    \ifblank{#1#2}{\:\cdot\: \MidSymbol[\delimsize\Vert] \:\cdot\:}{#1 \MidSymbol[\delimsize\Vert] #2}
}

\DeclarePairedDelimiterXPP{\aKale}[2]{\hat \opKale}{(}{)}{}{%
    \ifblank{#1#2}{\:\cdot\: \MidSymbol[\delimsize\Vert] \:\cdot\:}{#1 \MidSymbol[\delimsize\Vert] #2}
}

\DeclarePairedDelimiterXPP{\HofHessian}[1]{\opEntropy''}{[}{]}{}{%
    \renewcommand\given{\MidSymbol[\delimsize\vert]}
    \ifblank{#1}{\:\cdot\:}{#1}
}

\DeclarePairedDelimiterXPP{\specialHofHessian}[2]{\opEntropy''#1}{[}{]}{}{%
    \renewcommand\given{\MidSymbol[\delimsize\vert]}
    \ifblank{#2}{\:\cdot\:}{#2}
}

\DeclarePairedDelimiterXPP{\HofJacobian}[1]{\opEntropy'}{[}{]}{}{%
    \renewcommand\given{\MidSymbol[\delimsize\vert]}
    \ifblank{#1}{\:\cdot\:}{#1}
}

\DeclarePairedDelimiterXPP{\specialHofJacobian}[2]{\opEntropy'#1}{[}{]}{}{%
    \renewcommand\given{\MidSymbol[\delimsize\vert]}
    \ifblank{#2}{\:\cdot\:}{#2}
}

\newcommand{\FisherInfo}[1]{\HofHessian{#1}} 
\newcommand{\specialFisherInfo}[2]{\specialHofHessian{#1}{#2}}

\newcommand{\andreas}[1]{}
\newcommand{\yarin}[1]{}
\newcommand{\editor}[1]{}

\newtcolorbox{importantresult}{colback=solarized@yellow!5!white,
colframe=solarized@yellow,parbox, left=0.5mm, right=0.5mm,top=0.5mm,bottom=0.5mm}

\newtcolorbox{whiteresult}{colback=solarized@violet!2!white,
colframe=solarized@violet,parbox, left=0.5mm, right=0.5mm,top=0.5mm,bottom=0.5mm}

\newtcolorbox{mainresult}{colback=solarized@violet!5!white,
colframe=solarized@violet,parbox, left=0.5mm, right=0.5mm,top=0.5mm,bottom=0.5mm}

\newtcolorbox{importantresult_noparbox}{breakable,colback=solarized@yellow!5!white,
colframe=solarized@yellow,parbox=false, left=0.5mm, right=0.5mm,top=0.5mm,bottom=0.5mm}

\newtcbtheorem[number within=chapter,crefname={$\square$}{$\square$}]{insightbox}{$\square$}{colback=solarized@green!5!white,
colframe=solarized@green,    
colbacktitle=solarized@base2,coltitle=black,
fonttitle=\bfseries,
separator sign dash,
label separator={},
parbox,left=2mm,right=2mm}{}

\tcbset{
    llmmodel/.style={
        oversize=0.2\textwidth,
        colback=white,
        colbacktitle={#1},
        colframe={#1!50!black},
        fonttitle=\bfseries,
    },
    llmmsg/.style 2 args={
        colback={#1},
        colframe={#1!50!black},
        listing only,
        minted language=markdown,
        minted options={autogobble,breaklines},
        width={\textwidth},
        bottom=0mm,
        size=small,
        breakable,
        title={#2},
        title after break={#2},
    },
}

\excludecomment{gpt}
\excludecomment{chatgpt}

\begin{draft}
\tcbuselibrary{minted}

\newcommand{\chatbottype}{}
\newcommand{\chatbotcolor}{}

\renewenvironment{gpt}{
    \renewcommand{\chatbottype}{GPT-4}
    \renewcommand{\chatbotcolor}{solarized@violet}
}{}

\newtcblisting{chatuser}
{
    llmmodel=solarized@base03,coltitle=solarized@base3,
    llmmsg={solarized@yellow!20!white}{User},
    left skip={0.02\textwidth},
}

\newtcblisting{chatbot}
{
    breakable,
    llmmodel=solarized@base3,coltitle=\chatbotcolor,
    llmmsg={solarized@blue!20!white}{\chatbottype},
    right skip={0.02\textwidth},
}

\newtcblisting{chatedited}
{
    llmmodel=solarized@base3,coltitle=solarized@green,
    llmmsg={solarized@blue!20!white}{Edited \&\ Revised},
    right skip={0.02\textwidth},
}
\end{draft}

\begin{draft}
\renewcommand{\andreas}[1]{\todo[author=AK, color=gray,inline]{#1}}
\end{draft}

\usetikzlibrary{arrows.meta,positioning,shapes,calc}

\usetikzlibrary{fadings}

\tikzfading[name=fade bottom-right to top-left,
  bottom color=transparent!66, top color=transparent!100, shading angle=45]

\tikzfading[name=fade right to left,
    left color=transparent!100, right color=transparent!66]
  
\newcommand\independent{\protect\mathpalette{\protect\independenT}{\perp}}
\def\independenT#1#2{\mathrel{\rlap{$#1#2$}\mkern2mu{#1#2}}}

\DeclareMathOperator*{\argmax}{arg\,max}
\DeclareMathOperator*{\argmin}{arg\,min}
\newcommand{\defeq}{\triangleq}

\newcommand{\Lof}{\Hof}

\DeclarePairedDelimiterXPP{\ptof}[1]{\opp_{\mathrm{true}}}{(}{)}{}{%
    \renewcommand\given{\MidSymbol[\delimsize\vert]}
    \ifblank{#1}{\:\cdot\:}{#1}
}
\DeclarePairedDelimiterXPP{\pevalof}[1]{\opp_{\mathrm{eval}}}{(}{)}{}{%
    \renewcommand\given{\MidSymbol[\delimsize\vert]}
    \ifblank{#1}{\:\cdot\:}{#1}
}

\DeclarePairedDelimiterXPP{\ppof}[1]{\opp'}{(}{)}{}{%
    \renewcommand\given{\MidSymbol[\delimsize\vert]}
    \ifblank{#1}{\:\cdot\:}{#1}
}

\newmathcommand{\batchvar}{\mathtt{K}}
\newtextcommand{\batchvar}{$\batchvar$\xspace}

\newmathcommand{\poolsize}{\mathtt{M}}
\newtextcommand{\poolsize}{$\poolsize$\xspace}

\newmathcommand{\trainsize}{\mathtt{N}}
\newtextcommand{\trainsize}{$\trainsize$\xspace}

\newmathcommand{\evalsize}{\mathtt{E}}
\newtextcommand{\evalsize}{$\evalsize$\xspace}

\newmathcommand{\testsize}{\mathtt{T}}
\newtextcommand{\testsize}{$\testsize$\xspace}

\newmathcommand{\numclasses}{\mathtt{C}}
\newtextcommand{\numclasses}{$\numclasses$\xspace}

\DeclareMathOperator{\tr}{tr}
\DeclareMathOperator{\diag}{diag}
\DeclareMathOperator{\softmax}{softmax}

\DeclareMathOperator{\Dirichlet}{Dirichlet}
\DeclareMathOperator{\Multinomial}{Multinomial}

\newcommand{\N}{\mathcal{N}}
\newcommand{\normaldist}[2]{\N(#1,\,#2)}
\newcommand{\normaldistpdf}[3]{\N(#1;\,#2,\,#3)}

\newmathcommand{\targetdim}{\mathrm{D}}
\newtextcommand{\targetdim}{$\batchvar$\xspace}

\newcommand{\pdataof}[1]{\hpcof{\mathrm{true}}{#1}}
\newcommand{\ptrainof}[1]{\hpcof{\mathrm{train}}{#1}}

\newcommand{\typeeval}{\mathrm{eval}}
\newcommand{\typetest}{\mathrm{test}}
\newcommand{\typetrain}{\mathrm{train}}
\newcommand{\typeacq}{\mathrm{acq}}
\newcommand{\typepool}{\mathrm{pool}}

\newcommand{\Dany}{\mathcal{D}}
\newcommand{\Dtrain}{\Dany^{\typetrain}}
\newcommand{\Dacq}{\Dany^{\typeacq}}
\newcommand{\Dtest}{\Dany^{\typetest}}
\newcommand{\Dpool}{\Dany^{\typepool}}
\newcommand{\Deval}{\Dany^{\typeeval}}

\newlength\bigB
\newlength\smlT

\newcommand{\BBBAL}{B\raisebox{\heightof{B}-\height}{\scriptsize 3}AL\xspace}

\newcommand{\y}{y}
\newcommand{\Y}{Y}
\newcommand{\x}{\boldsymbol{x}}
\newcommand{\X}{\boldsymbol{X}}
\newcommand{\z}{\boldsymbol{z}}

\newcommand{\pigof}[1]{\mathrm{PIG}({#1})}
\newcommand{\jpigof}[1]{\mathrm{JPIG}({#1})}
\newcommand{\epigof}[1]{\mathrm{EPIG}({#1})}
\newcommand{\eigof}[1]{\mathrm{EIG}({#1})}
\newcommand{\baldof}[1]{\mathrm{BALD}({#1})}

\title{\vspace{20pt}Advancing Deep Active Learning \&\ Data Subset Selection: Unifying Principles with Information-Theory Intuitions}

\author{Andreas Kirsch}
\college{Exeter College}

\degree{Doctor of Philosophy}
\degreedate{Trinity 2023}

\newcommand{\changelocaltocdepth}[1]{%
  \addtocontents{toc}{\protect\setcounter{tocdepth}{#1}}%
  \setcounter{tocdepth}{#1}%
}

\nobibliography*

\begin{document} 

\setlength{\textbaselineskip}{20pt plus1pt minus1pt}

\setcounter{secnumdepth}{3}
\setcounter{tocdepth}{1}

\begin{romanpages}

	\maketitle

	\begin{dedication}
		To my family and friends.\\
		\vspace{1em}
		In particular:\\
		my parents, for their support;\\
		Armin, for his advice; and\\
		Alex, in his memory.\\
	\end{dedication}

	\begin{abstract}

\vspace{20pt}

At its core, this thesis aims to enhance the practicality of deep learning by improving the label and training efficiency of deep learning models.
To this end, we investigate data subset selection techniques, specifically active learning and active sampling, grounded in information-theoretic principles.
Active learning improves label efficiency, while active sampling enhances training efficiency. 

Supervised deep learning models often require extensive training with labeled data.
Label acquisition can be expensive and time-consuming, and training large models is resource-intensive, hindering the adoption outside academic research and ``big tech.'' 

Existing methods for data subset selection in deep learning often rely on heuristics or lack a principled information-theoretic foundation.
In contrast, this thesis examines several objectives for data subset selection and their applications within deep learning, striving for a more principled approach inspired by information theory.

We begin by disentangling epistemic and aleatoric uncertainty in single forward-pass deep neural networks, which provides helpful intuitions and insights into different forms of uncertainty and their relevance for data subset selection.
We then propose and investigate various approaches for active learning and data subset selection in (Bayesian) deep learning.
Finally, we relate various existing and proposed approaches to approximations of information quantities in weight or prediction space.

Underpinning this work is a principled and practical notation for information-theoretic quantities that includes both random variables and observed outcomes.
This thesis demonstrates the benefits of working from a unified perspective and highlights the potential impact of our contributions to the practical application of deep learning.

\begin{gpt}
    \begin{chatuser}
        Great! Taking a high-level view, can you consider what a good PhD thesis abstract should contain and how it is structured (think step by step) and then criticize what I have written (and your edits). After this, please make high-level suggestions, determine what is unclear and what additional information you need, and ask me to clarify. If you believe that you have sufficient information, edit the text accordingly to your feedback as the last step.
    \end{chatuser}
    \begin{chatbot}
A good PhD thesis abstract should contain the following elements:

1. Motivation: Explain the problem and its importance.
2. Objectives: Clearly state the main research questions or goals.
3. Approach: Describe the methodology or techniques used to address the problem.
4. Results: Summarize the main findings or contributions of the thesis.
5. Conclusion: Briefly discuss the implications of the results and the potential impact of the work.

Critique of the provided abstract (and the edited version):

- The motivation is clear, but it could be slightly more concise.
- The objectives are well-stated, but they could be more explicitly connected to the motivation.
- The approach is well-described, but it might be helpful to provide more context on the existing methods and how the proposed approach differs from them.
- The results are well-summarized, but it would be helpful to provide more concrete examples or quantitative evidence of the improvements achieved.
- The conclusion is somewhat implicit and could be more explicitly stated.

High-level suggestions:

- Make the motivation more concise and directly connect it to the objectives.
- Provide more context on existing methods and how the proposed approach differs from them.
- Include more concrete examples or quantitative evidence of the improvements achieved.
- Explicitly state the conclusion and the potential impact of the work.

Please provide more information on the following:

1. Can you provide more context on the existing methods for active learning and active sampling in deep learning, and how your approach differs from them?
2. Can you provide any quantitative evidence or specific examples of the improvements achieved by your approach?

Based on your response, I will revise the abstract accordingly.
    \end{chatbot}
\begin{chatuser}
1. Within Bayesian active learning, existing methods (before my contributions) were not investigating batch acquisition in a principled fashion (though Bayesian optimal experimental design had some prior art as you might be aware of). My main contribution for that is BatchBALD. (Do you know it?)
Within regular deep neural networks, most methods focus on an embedding space view (via last-layer scores, which end up being connected to the network's embeddings). Some methods try to sample by finding a diverse set of embeddings (e.g. BADGE via using k-MEANS++), others use k-DPPs directly. 
One of my contributions is to look at this prior art and show how it can be expressed as weight-space based method that also approximates the expected information gain (or the expected predictive information gain), as BAIT (from the Gone Fishing paper by Jordan Ash does). Other methods (SIMILAR and PRISM) are based on submodular functions that use analogies to mutual information etc but are not actually principled in an information-theoretic way. I'm showing that using the log det, which is submodular, with score-based similarity matrices, is approximating information quantities. In regards to active sampling, our contribution (rholoss, predictive information gain), measures a point-wise quantity, which is based on the possible loss reduction and takes into account the irreducible/aleatoric loss of data samples. Other methods focus on the total loss, and thus might prefer noisy samples that will always have high loss for active sampling.

2. The empirical evidence is on various datasets, MNIST, CINIC-10, CIFAR-10, Repeated-MNIST, ImageNet. We also propose DirtyMNIST as dataset that contains more ambiguous MNIST-like samples as well. For the unifying theory, we provides lots of small proofs in the appendix.

Does this clarify your questions? Think step by step. Please let me know then if you have any follow-up questions.
\end{chatuser}
\end{gpt}
\begin{gpt}
    \begin{chatuser}
        Thank you! I have edited the abstract and merged your suggestions. Please take a look and let me know (think step by step), how it is different, how it has improved, what weaknesses remain and any new weaknesses. Then suggest how these could be improved. 
        
        [...]
    \end{chatuser}
\begin{chatbot}
The edited abstract has the following improvements:

1. It provides a clear motivation and objective, stating the aim to make deep learning more practical by increasing label and training efficiency.
2. It emphasizes the limitations of existing methods, which often rely on heuristics or lack a principled information-theoretic foundation.
3. It presents a concise overview of the specific contributions, such as BatchBALD, EPIG, and the reducible hold-out loss, along with their applications and differences from existing methods.
4. It highlights the empirical validation of the methods on various real-world and synthetic datasets, demonstrating their effectiveness in increasing label and training efficiency.
5. It underscores the benefits of working from a unified perspective, using a principled and practical notation for information-theoretic quantities.

However, there are still some potential weaknesses:

1. The abstract could benefit from a more explicit conclusion, summarizing the main findings and their implications for the field of deep learning.
2. While the abstract mentions various methods and their applications, it might be helpful to provide a brief explanation of the underlying intuitions or principles for readers who are not familiar with the specific terminology.

To address these weaknesses, consider the following suggestions:

1. Add a concluding sentence or two that summarizes the main findings and their implications for the field of deep learning, emphasizing the potential impact of the contributions.
2. Provide a brief explanation of the underlying intuitions or principles for each method, such as BatchBALD, EPIG, and the reducible hold-out loss, to help readers understand their significance and differences from existing methods.

Please let me know if you would like me to make these changes or if you have any further questions or concerns.
\end{chatbot}
\end{gpt}

	\end{abstract}

	\begin{acknowledgements}

\vspace{20pt}
\setlength{\baselineskip}{\singlebaselineskip}

To my family and friends: thank you for your unwavering support throughout my DPhil journey. In particular, I am grateful to my parents, who have supported me in so many ways over the years; to my cousin, who shared with me the ups and downs of pursuing a doctorate at the same time in Romania; and my old friends in my hometown in Germany: Michi, Daniel, and Marina. I hope to see you all soon.

Special thanks go to my old friend, Armin Krupp, who directed me towards the exciting world of machine learning at the University of Oxford and the blog of an incoming associate professor, Yarin Gal. Your positivity and support have been a driving force in my academic and professional journey. And just as much am I grateful to Daniel and Martin for sticking with me since our time together in \href{https://www.tum.de/en/}{Munich}.

I also owe much to the many friends who have made my time in Oxford worthwhile: especially, Alessandro, Kevin, Alessio, Maja, Jonas, Charlotte, Lorika, Bastiaan, Charles, Prateek, Luisa, Jan, Clemens, Fabian, Ana, Rob, Brenda, Neil, Theo, Kayla, Anna, Aleksandra, Fiona, Serban, Stefan, Patricia, Olmo, Carli, Simone, Mar, John, Ed, Josh, Shad, Sam \&\ Tim; your friendship has been invaluable, whether on (coffee) walks, at shared dinners and co-working sessions, in Italy, singing karaoke, surviving the pandemic, hiking up the tallest mountains or in the Cotswold, visiting me, advising me, partying or going to formals together. The same goes for the \href{https://www.society-cafe.com/}{Society Cafe} regulars \&\ baristas over the last five years: Nadia, Dom, Laura, Em, Amina, Will, Johnny, Marie, Kat, Eleanor, Ricardo, Shae, Ruth, Lauren \&\ Noor; thank you for being so much fun every day; and to my DnD group: Lasya, Cyril, Mihir, Nico, Ned \&\ Luca; my experience would not have been the same without you and Gogol.

I am very grateful to the \href{https://aims.robots.ox.ac.uk/}{AIMS CDT program} and its incredible administrator, Wendy Poole, for providing funding and support throughout my studies and generally being a force for good. I am also very lucky to be a student of \href{https://www.exeter.ox.ac.uk/}{Exeter College} which I thank from my heart for their support and nurturing environment---an environment which starkly contrasts with \href{https://x.com/BlackHC/status/1398191082987479041}{my earlier experiences} at \href{https://oxfordwhisperer.uk/}{Kellogg College}.

I am also grateful to my collaborators, especially my joint co-authors, for their support and feedback: \href{https://www.robots.ox.ac.uk/~twgr/}{Tom Rainforth} for his mentorship and invaluable advice on research and writing---meeting in University Parks during the pandemic with a small whiteboard to discuss information theory and transductive acquisition functions, which directly led to the current \S\ref{chapter:pobal} \citep{kirsch2021test,epig}, is something I will always remember; my external collaborators, including Dustin Tran, Jasper Snoek, Frédéric Branchaud-Charron, Parmida Atighehchian, and Uri Shalit; and my \href{https://oatml.cs.ox.ac.uk/}{OATML} collaborators, Joost van Amersfoort, Jishnu Mukhoti, Sebastian Farquhar, Freddie Bickford Smith, Muhammed Razzak, Andrew Jesson, Jannik Kossen, Clare Lyle, Jan Brauner, S\"oren Mindermann, Panagiotis Tigas, and Aidan Gomez---your expertise, encouragement, and patience have been invaluable to me and this thesis impossible without you. My thanks go to the other members of OATML, too, for their support and feedback: especially Lisa Schut, Tim Rudner, Milad Alizadeh, and Lewis Smith. 

While I have strived for research integrity as a guiding principle, it has tested my resilience and highlighted the delicate balance between being critical and patient in scholarly conversations as you can imagine. Being referred to as an `\href{https://web.archive.org/web/20221110142356/https://twitter.com/andrewgwils/status/1550120752099180548}{unhinged reviewer 2}' in the summer of 2022 by an NYU professor (and former PhD colleague of my supervisor) for engaging in \href{https://web.archive.org/web/20220608103712/https://blog.blackhc.net/2022/06/bayesian-model-selection-marginal-likehood-generalization/}{scholarly discourse} and \href{https://web.archive.org/web/20220720213534/https://twitter.com/BlackHC/status/1549845173512929280}{highlighting inconsistencies}, while difficult, taught me about assertiveness and perseverance.
These experiences have enriched my understanding of the academic landscape, both on how to face criticism and in making thoughtful contributions\footnote{\S\ref{chapter:dldd} \citep{kirsch2023does} is evidence of this I hope.}. 
The encouragement and kindness of the broader academic community despite this have played a pivotal role in my growth as a scholar, and I'm grateful for the opportunity to learn this during my DPhil.

Of course, I also want to thank my supervisor, \href{https://www.cs.ox.ac.uk/people/yarin.gal/website/}{Yarin Gal}, for supporting my DPhil journey, helping me refine my time- and project-management skills, and providing feedback on most of my projects. 
He set up and guided my starter project, BatchBALD (\Cref{chapter:batchbald}), initiated many fruitful collaborations, and his input significantly contributed to the development of \citet{mukhoti2021deterministic} (\S\ref{chapter:ddu}) during the pandemic in early 2021. Finally, he encouraged me to start drafting this thesis in the summer of 2022 when he felt I was ready. 
As my DPhil and our collaboration evolved, I learned to work independently, finally submitting my last paper (\Cref{chapter:bbbal}) as a sole author and taking full ownership of my work. 
This final year was not without challenges, though.
When faced with artificial time constraints\footnote{I was away on an internship for three months at end of 2022 after completing 20\% of my thesis draft, and while my supervisor agreed to suspensions for such time spent away for others before and after, he refused in my case to the surprise of the department---but see the above.}, I completed the remaining 80\% of this thesis within an accelerated three-week time frame.
During my burn-out afterwards, I discovered that my supervisor had contributed as a co-author to a paper investigating a similar transductive information-theoretic quantity\footnote{It is based on a similar overarching goal and also leads to a similar approximation, but in a totally different setting and with a novel subsampling strategy.}, as a research project I worked on with him for a year\footnote{Starting with an \href{https://jamboard.google.com/d/1WvmPjhlf2b68eQXdLXf-SJKLIczdqyQyCAyqrpRlL5k/viewer?f=0}{initial pitch} in the summer of 2020, an early \href{https://arxiv.org/abs/2101.03552}{report} and \href{https://arxiv.org/abs/2106.11719}{workshop paper} in 2021, it eventually led to a \href{https://proceedings.mlr.press/v206/bickfordsmith23a.html}{conference paper} in 2023, see \S\ref{chapter:pobal}.} without letting me know---\href{https://blog.blackhc.net/2022/01/al-for-large-nlp-models/}{yet} \href{https://arxiv.org/abs/2212.07769}{again}---and then also found \href{https://github.com/anonymous35780/solaris-2023-iclr/blob/main/solaris/models/consistent_mc_dropout.py}{unattributed
code of mine}\footnote{It took some pushing to get my code properly attributed: \href{https://github.com/genedisco/genedisco/issues/8}{\#1}, \href{https://github.com/genedisco/genedisco/issues/9}{\#2}, \href{https://github.com/genedisco/genedisco/issues/10}{\#3}.} and a \href{https://colab.research.google.com/drive/1joNsZ6KFeJaiR3trMp1DIfPjcexq5b2G}{potentially crucial bug}\footnote{The code accidentally used the harmonic mean instead of the arithmetic mean in the entropy computation---in the end, six months later, the corrected code slightly improved the results, or at least wasn't performing worse.} in the paper.
I raised these issues with Yarin, and due to my conflicted judgement, also separately with the conference chairs, who determined it was concurrent work and promised they would reach out to the authors for my code to be acknowledged.
Yarin was understandably not happy with the situation, which had caused the co-authors in the lab, him, and me anxiety and stress, and I left the lab before my thesis defence when he asked me to---two months before that paper's camera-ready deadline---after I declared I would write a reproducibility report if the issues I had raised were not fixed to my satisfaction before the conference.
I absolutely believe Yarin's assurance that the above was unintentional: he thought about the paper's theory from a different perspective, not making the connection between the two research projects. In hindsight, the connection seems manifest\footnote{The connection became more evident once I read \href{https://openreview.net/pdf?id=mBkUeW8rpD6#page=3}{the paper} and \href{https://twitter.com/BlackHC/status/1655315495673798656}{earlier work}, and the implementation of our earlier method as an ablation ends up \href{https://github.com/amehrjou/DiscoBAX/blob/84f01283bc7f6ab5f66b5ea2a63632b401cc0402/discobax/methods/bax_acquisition/bax_sampling.py#L99}{a one-line addition}.}. 
Despite earlier assurances, it required some \href{https://twitter.com/BlackHC/status/1688223904660164608}{progressively less subtle nudging} to have a \href{https://arxiv.org/abs/2312.04064}{corrected paper} uploaded and the \href{https://github.com/amehrjou/DiscoBAX}{fixed code} published, some time after the actual conference\footnote{Academic commitments and personal circumstances of one of the three first-authors have likely played a part in some of the delay.}.
Supervision experiences are \href{https://twitter.com/BlackHC/status/1696456368545280380}{highly} \href{https://twitter.com/BlackHC/status/1696667972880801838}{individual}, and Yarin has been very supportive of his students' well-being throughout the years and tried to work things out. He suggested a ``two-year cool-off period [of] not reading and commenting on lab papers for a while to avoid me doing this on a `low' day which leads to triggering me and acting irrationally'' and has kindly agreed to write reference letters that ``only comment on my analytical skills and mention that I had some personal issues that have led to high anxiety affecting my performance.''
I believe in many parts thanks to him, I have become an independent and self-aware researcher who cares about academic integrity and conduct---and, indeed, while it might not have ended as I had imagined or would have liked, Yarin was the most instrumental factor in my finishing this DPhil in 2023 and passing my defence overall.

Finally, I want to thank all those who have provided feedback on my thesis, especially my assessors, \href{https://research.google/people/kevin-p-murphy/}{Kevin Murphy} and \href{https://www.robots.ox.ac.uk/~mosb/}{Mike Osborne}, and the many anonymous conference reviewers who have helped shape my work.

While I have failed in my research at times, this thesis stands as a testament to the collective efforts, support, and encouragement of the remarkable individuals who have been part of this adventure. My deepest and sincerest thanks to you all.

\begin{flushright}
    \vspace{1em}
    \emph{Andreas Kirsch, Oxford, 2023 \footnote{I also want to thank Yarin for iterating on the acknowledgements with me in early 2024.}}
\end{flushright}

	\end{acknowledgements}

	\flushbottom

	\tableofcontents

	\mtcaddchapter
\end{romanpages}

\flushbottom

\newcommand{\ddutitle}{Single Forward-Pass Aleatoric and Epistemic Uncertainty}
\newcommand{\pittitle}{A Practical \&\ Unified Notation for Information Quantities}
\newcommand{\baldtitle}{Diverse Batch Acquisition for Bayesian Active Learning}
\newcommand{\stochacqtitle}{Stochastic Batch Acquisition for Deep Active Learning}
\newcommand{\obititle}{Marginal and Joint Cross-Entropies \&\ Predictives}
\newcommand{\pobaltitle}{Prediction- \&\ Distribution-Aware Bayesian Active Learning}
\newcommand{\rholosstitle}{Prioritized Data Selection during Training}
\newcommand{\fimtitle}{Unifying Approaches in Active Learning and Active Sampling}
\newcommand{\bbbaltitle}{Black-Box Batch Active Learning for Regression}
\newcommand{\dlddtitle}{Deep Learning on a Data Diet}
\newcommand{\ddudmtitle}{Dirichlet Model of a Deep Ensemble's Softmax Predictions}
\newcommand{\jiangtitle}{A Note on ``Assessing Generalization of SGD via Disagreement''}
\newcommand{\causalbaldtitle}{Causal-BALD: Deep Bayesian Active Learning of Outcomes to Infer Treatment-Effects from Observational Data}
\newcommand{\conclusiontitle}{Conclusion}

{
	\begin{savequote}[75mm]
	When you can do nothing, what can you do?
	\end{savequote}
	\chapter{Preliminaries}
	\label{chapter:preliminaries}
	\section{Introduction}
\label{glue:sec:introduction}

Over a decade ago, the deep learning revolution began, making significant strides in various research fields, including areas once considered exclusive domains of human ingenuity and creativity.
Despite its success, deep learning still faces challenges that hinder its deployment in more practical, everyday settings. 
This thesis aims to address some of these challenges and make deep learning more accessible by reducing the costs of gathering and labeling data, speeding up training, and doing so in a principled fashion.

Deep learning models have yet to gain a strong foothold outside big tech, maybe due to issues such as lack of interpretability, robustness, and generalization guarantees. 
However, history tells us that such concerns are unlikely to deter the use of a new technology.
So, what does?

More practical concerns, such as the time-consuming and expensive processes of gathering and labeling high-quality data and training models, may also limit the deployment of deep learning.
Research often overlooks the data pipeline, which frequently is the bottleneck that constrains model performance.

The primary objective of this thesis is to address these practical challenges and make deep learning more accessible by exploring how to reduce the cost of gathering and labeling data and how to speed up training. At the same time while maintaining a principled approach allows for a better understanding of trade-offs and connections between different methods.

In particular, this thesis focuses on \emph{data subset selection} in the broader sense of active learning and active sampling for deep learning models, especially through the principled application of information theory.
\emph{Active learning}, like semi-supervised and unsupervised approaches, increases label efficiency.
It does so by selecting the most informative samples to label from a \emph{pool set} of unlabeled data, potentially significantly reducing the number of required labels for good model performance.
Similarly, \emph{active sampling} improves training efficiency by filtering the available training data to focus on the most informative samples for the model during training.
The main research questions in this thesis are:
\begin{enumerate}
\item How can uncertainty quantification, specifically aleatoric and epistemic uncertainty, be better understood and applied in the context of active learning and active sampling?

\item How can a deeper understanding of the theoretical foundations of active learning and active sampling contribute to the progress of the field and improve the practical application of these techniques?

\item How can the cost of gathering and labeling data be reduced, and how can training be sped up in a principled fashion?

\item What are the connections between different active learning and active sampling approaches, and how can information theory be used to unify these approaches?
\end{enumerate}
To answer these questions, we rely on \emph{information theory} which helps quantify the information content of random variables and events, offers valuable insights and intuitions, and provides a principled framework for reasoning about uncertainty.
Information diagrams (\emph{I-diagrams}, see e.g.\ \Cref{practical_it_notation:glue:background:idiagram}) are particularly useful for building intuition, as they resemble Venn diagrams and stem from the realization that information-theoretic operations behave similarly to set operations \citep{yeung1991new,lang2022information}.
In a Bayesian setting, where model parameters are treated as random variables with associated uncertainty, information theory can be used to express and examine different concepts of 'informativeness' that are useful for data subset selection.
This thesis examines objectives and extensions that follow from these intuitions and demonstrates that these objectives can be successfully applied in data subset selection.

In the next two sections of this chapter, we lay out the background and the structure of the thesis:
\Cref{glue:sec:background} introduces necessary background material and points towards relevant literature for the thesis overall; and
\Cref{glue:sec:outline} provides an overview of the thesis and its structure.

\begin{gpt}
    \begin{chatuser}
        I'm a DPhil student in machine learning (Bayesian deep learning and active learning) at the University of Oxford.
I'm currently writing my DPhil thesis.
Taking a high-level view, can you consider what a good PhD thesis introduction section should contain and how it is structured (think step by step) and then criticize what I have written.
After this, please make high-level suggestions, determine what is unclear and what additional information you need, and ask me to clarify.
If you believe that you have sufficient information, redraft the text accordingly as the last step.

        [...]
    \end{chatuser}
    \begin{chatbot}
A little over a decade ago, the deep learning revolution began, and since then, it has made significant strides in a wide range of research fields, including areas once thought to be exclusive domains of human ingenuity and creativity.
Despite its success, deep learning still faces challenges that hinder its deployment, particularly in more practical, everyday settings.
This thesis aims to address these challenges and make deep learning more accessible by focusing on reducing the costs of gathering and labeling data, speeding up training, and doing so in a principled fashion.

The introduction section of a good PhD thesis should contain the following elements, structured step by step:

1. A brief overview of the current state of the research field, including recent advancements and challenges.

2. A clear statement of the problem or gap in the current research that the thesis seeks to address.

3. An explanation of the significance and practical implications of addressing this problem, as well as the potential benefits that could arise from the proposed research.

4. A clear statement of the research questions or objectives that the thesis will explore.

5. An outline of the methodology and theoretical foundations that will be employed in the research, with a brief explanation of why they are appropriate for addressing the stated problem.

6. A short summary of the organization and structure of the thesis, including a brief description of each chapter or section.

Critique:

Your introduction provides a good overview of the deep learning revolution and some of the challenges it faces. However, it would benefit from a more structured presentation of the information and a clearer focus on the specific research questions and objectives of your thesis. Also, an outline of the organization and structure of the thesis is currently missing and should be added.

Suggestions:

1. Begin with a brief overview of the deep learning field, its successes, and its challenges, as you have done. However, consider presenting this information more concisely to make the introduction more engaging and easier to follow.

2. Clearly state the problem or gap in the current research that your thesis seeks to address: the practical issues of gathering and labeling data, and speeding up training.

3. Explain the significance of addressing this problem, particularly in terms of making deep learning more accessible and useful in practical settings.

4. Clearly state the research questions or objectives of your thesis: how to reduce the cost of gathering and labeling data, how to speed up training, and how to do so in a principled fashion.

5. Briefly outline the methodology and theoretical foundations you will employ (active learning, active sampling, and information theory) and explain why they are appropriate for addressing the stated problem.

6. Provide a short summary of the organization and structure of the thesis, including a brief description of each chapter or section.

Additional Information Needed:

- An outline of the organization and structure of the thesis, including a brief description of each chapter or section.

Please clarify the organization and structure of your thesis so that I can help you redraft the introduction accordingly.
    \end{chatbot}
\end{gpt}

{
	{
    \graphicspath{{./glue/figures/}}
    \newcommand{\Model}{\mathcal{M}}

\newcommand{\theAF}{a}
\newcommand{\BALDAF}{\theAF_{\mathtt{BALD}}}
\newcommand{\BatchBALDAF}{\theAF_{\mathtt{BatchBALD}}}

\newcommand{\w}{\pmb{\omega}}
\newcommand{\W}{\pmb{\Omega}}

\newcommand{\xeval}{\x^\typeeval}
\newcommand{\xtest}{\x^\typetest}
\newcommand{\xtrain}{\x^\typetrain}
\newcommand{\xacq}{\x^\typeacq}
\newcommand{\xpool}{\x^\typepool}

\newcommand{\xacqstar}{\x^{\typeacq, *}}

\newcommand{\Xeval}{\X^\typeeval}
\newcommand{\Xtest}{\X^\typetest}
\newcommand{\Xtrain}{\X^\typetrain}
\newcommand{\Xacq}{\X^\typeacq}
\newcommand{\Xpool}{\X^\typepool}

\newcommand{\yeval}{\y^\typeeval}
\newcommand{\ytest}{\y^\typetest}
\newcommand{\ytrain}{\y^\typetrain}
\newcommand{\yacq}{\y^\typeacq}
\newcommand{\ypool}{\y^\typepool}

\newcommand{\Ytest}{\Y^\typetest}
\newcommand{\Ytrain}{\Y^\typetrain}
\newcommand{\Yeval}{\Y^\typeeval}
\newcommand{\Yacq}{\Y^\typeacq}
\newcommand{\Ypool}{\Y^\typepool}

\newcommand{\Xrate}{\boldsymbol{\mathcal{\X}}}
\newcommand{\Yrate}{\mathcal{\Y}}

\newcommand{\xevalset}{{\x^\typeeval_{1..\evalsize}}}
\newcommand{\xtestset}{{\x^\typetest_{1..\testsize}}}
\newcommand{\xtrainset}{{\x^\typetrain_{1..\trainsize}}}
\newcommand{\xacqset}{{\x^\typeacq_{1..\batchvar}}}
\newcommand{\xpoolset}{{\x^\typepool_{1..\poolsize}}}

\newcommand{\xacqsetstar}{{\x^{\typeacq, *}_{1..\batchvar}}}

\newcommand{\Xevalset}{{\X^\typeeval_{1..\evalsize}}}
\newcommand{\Xtestset}{{\X^\typetest_{1..\testsize}}}
\newcommand{\Xtrainset}{{\X^\typetrain_{1..\trainsize}}}
\newcommand{\Xacqset}{{\X^\typeacq_{1..\batchvar}}}
\newcommand{\Xpoolset}{{\X^\typepool_{1..\poolsize}}}

\newcommand{\yevalset}{{\y^\typeeval_{1..\evalsize}}}
\newcommand{\ytestset}{{\y^\typetest_{1..\testsize}}}
\newcommand{\ytrainset}{{\y^\typetrain_{1..\trainsize}}}
\newcommand{\yacqset}{{\y^\typeacq_{1..\batchvar}}}
\newcommand{\ypoolset}{{\y^\typepool_{1..\poolsize}}}

\newcommand{\Ytestset}{{\Y^\typetest_{1..\testsize}}}
\newcommand{\Ytrainset}{{\Y^\typetrain_{1..\trainsize}}}
\newcommand{\Yevalset}{{\Y^\typeeval_{1..\evalsize}}}
\newcommand{\Yacqset}{{\Y^\typeacq_{1..\batchvar}}}
\newcommand{\Ypoolset}{{\Y^\typepool_{1..\poolsize}}}

\newcommand{\xset}{{\x_{1..n}}}
\newcommand{\Xset}{{\X_{1..n}}}

\newcommand{\xyset}{(\x_i,\y_i)_{i=1}^{n}}

\newcommand{\xsetc}[1]{\x_{1..#1}}
\newcommand{\Xsetc}[1]{\X_{1..#1}}

\newcommand{\yset}{{\y_{1..n}}}
\newcommand{\Yset}{{\Y_{1..n}}}

\newcommand{\rholoss}{$\rho$-Loss\xspace}

    \section{Background \&\ Literature Review}
    \label{glue:sec:background}

    In this section, we introduce concepts that are relevant for the whole thesis.
    We revisit general background in information theory, Bayesian neural networks, active learning, and active sampling (in this order), and highlight important literature.
    Related work and background that are only relevant for their respective chapters are kept there: 
    we hope that this aids the reader by keeping relevant information close to where it is needed.
    The next chapters generally follow the consistent notation introduced below, and we point out when we deviate from that in the specific chapters.

    \subsection{Information Theory}
    \label{glue:background:information_theory}
    {
    Information theory has provided insights for deep learning: information bottlenecks explain objectives both for supervised and unsupervised learning of high-dimensional data \citep{shwartz2017opening,kirsch2020unpacking,jonsson2020convergence};\yarin{add citations to all} similarly, information theory has inspired Bayesian experiment design, Bayesian optimization, and active learning as well as motivated research into submodularity in general \citep{lindley1956measure, foster2019variational}.
    \andreas{add citations}

    \yarin{perhaps start the para with:
    But IT can be opaque to ML researchers, leading to misunderstandings and misuse [give examples]
    Then say that a practical notation can accelerate research, here we propose a new practical notation, we give examples in a bunch of domains }
    A practical notation conveys valuable intuitions and concisely expresses new ideas.
    The currently employed notation in information theory, however, can be ambiguous for more complex expressions found in applied settings and often deviates between published works because
    researchers are from different backgrounds such as statistics, computer science, information engineering, which all use information theory.
    For example, $H(X,Y)$ is sometimes used to denote the \emph{cross-entropy} between $X$ and $Y$, which conflicts with common notation of the joint entropy $H(X, Y)$ for $X$ and $Y$, or it is not clarified that $\Hof{X \given Y}$ as conditional entropy of $X$ given $Y$ is an expectation over $Y$.
    Here, we present a disambiguated and consistent notation while striving to stay close to known notation when possible. 
      
    \yarin{this can be used for the BG section in the thesis / journal paper
    
    split this section into what's used to date, and what you're proposing; reading this, I assume that all the stuff that you discuss is already existing lit. Start from a proper bg sect, explain all the different notations used atm and the abuse of notation and source of confusion. Then have your own contribution section which explains what you're suggesting to do differently (which can be brief) followed by all the stuff we can deduce from the unified notation }
    For a general introduction to information theory, we refer to \citet{cover2005elements} and \citet{yeung2008information}.
    In the following section, we introduce our practical and unified notation.
    In \Cref{chapter:practical_it_notation}, we extend this notation to also include observed outcomes and point-wise mutual information, information gain and information-theoretic surprise.
    We start with notation that is explicit about the underlying probability distribution $\pof{}$.
    Note that we do not require $\opq$ to be normalized at this stage as it allows for greater notational flexibility.
        
    \begin{definition}
    Let Shannon's information content $\ICof{}$, cross-entropy $\CrossEntropy{}{}$, entropy $\xHof{}$, and KL divergence $\Kale{}{}$ (Kullback-Leibler divergence) be defined for a probability distribution $\opp$ and non-negative function $\opq$ for a random variable \(X\) and non-negative real number \(\rho\) as:
    \begin{align}
        \ICof{\rho} &\defeq -\ln{\rho} \\
        \CrossEntropy{\pof{X}}{\qof{X}} &\defeq \simpleE{\pof{x}}{\ICof{\qof{x}}}\\
        \xHof{\pof{X}} &\defeq \CrossEntropy{\pof{X}}{\pof{X}} \\
        \Kale{\pof{X}}{\qof{X}} &\defeq \CrossEntropy{\pof{X}}{\qof{X}} - \xHof{\pof{X}},
    \end{align}
    where we fix $\ICof{0} \defeq -\infty$ with $0 \cdot \ICof{0} = 0$ as usual.
    \end{definition}
    \citet{shannon1948mathematical} introduced the information content as negative logarithm due to its additivity for independent messages: \(\ICof{\pof{x,y}} = \ICof{\pof{x}} + \ICof{\pof{y}}\) for independent random variables \(X\) and \(Y\).
    \begin{proposition}
    \label{practical_it_notation:prop:xe_kale_rules}
    For a random variable \(X\) with probability distributions \(\opp\), \(\opp_1\) and \(\opp_2\), and non-negative functions \(\opq\), \(\opq_1\) and \(\opq_2\) and \(\alpha \in [0,1]\):
    \begin{align}
        &\CrossEntropy{\opp}{\alpha \opq} = \CrossEntropy{\opp}{\opq} + \ICof{\alpha}, \\
        &\CrossEntropy{\opp}{\opq^\alpha} = \alpha \CrossEntropy{\opp}{\opq} \\
        &\CrossEntropy{\opp}{\opq_1 \opq_2} = \CrossEntropy{\opp}{ \opq_1} + \CrossEntropy{\opp}{ \opq_2}, \\
        &\CrossEntropy{\alpha \opp_1 + (1 - \alpha) \opp_2}{\opq} = \\
        & \quad = \alpha \CrossEntropy{\opp_1}{\opq} + (1 - \alpha) \CrossEntropy{\opp_2}{\opq} \\
        & \quad = \CrossEntropy{\opp_1}{\opq^\alpha } + \CrossEntropy{\opp_2}{\opq^{1 - \alpha}},
    \end{align}
    where we have left out ``\((X)\)'' everywhere for brevity.
    \end{proposition}
    \begin{proof}
    The statements follow from the linearity of the expectation and the additivity of the logarithm for products.
    \end{proof}
    This can be extended to show that cross-entropies are linear in their left-hand argument and log-linear in their right-hand argument.

    When we want to emphasize that we approximate the true distribution \(\opp\) using a different distribution \(\opq\) and the true probability distribution $\opp$ is understood, we use the notation $\varHof{\opq}{}$ for $\CrossEntropy{\pof{}}{\qof{}}$ following notation in \citet{kirsch2020unpacking} and \citet{xu2020theory}:
    \begin{definition}
    When the true probability distribution $\opp$ is understood from context, we will use the following shorthand notation:
    \begin{align}
        \Hof{X} &\defeq \xHof{\pof{X}} \\
        \varHof{\opq}{X} &\defeq \CrossEntropy{\pof{X}}{\qof{X}}.
    \end{align}
    When we have a parameterized distribution $\opq_\theta$ with parameters $\theta$, we will write $\varHof{\theta}{}$ instead of $\varHof{\opq_\theta}{}$ when the context is clear. 
    \end{definition}
    Approximating a possibly intractable distribution with a parameterized one is common when performing variational inference.
    The main motivation for this notation is that when $\opq$ is a density,  $\int q(x) \, dx = 1$, we have $\varHof{\opq}{} \ge \Hof{}$ with equality when \(\opq = \opp\).
    Concretely, we have the following useful identities:
    \begin{proposition}
    We have the following lower-bounds for the cross-entropy and KL, with \(Z_q \defeq \int q(x) \, dx\):
    \begin{align}
        \CrossEntropy{\pof{X}}{\qof{X}} &\ge \xHof{\pof{X}} + \ICof{Z_q}, \\
        \Kale{\pof{X}}{\qof{X}} &\ge \ICof{Z_q},
    \end{align}
    with equality exactly when \(\opq/Z_q = \opp\) for \(Z_q \defeq \int q(x) \, dx\).
    \end{proposition}
    \begin{proof}
    The statements follow from Jensen's inequality and the convexity of \(\ICof{}\).
    \end{proof}
    This also implies the non-negativity of the KL for densities when we substitute $Z_q=1$ in above statements. We repeat the result as it is often used: 
    \begin{corollary}
    When $q$ is a probability distribution, we have:
    \begin{align}
        \CrossEntropy{\pof{X}}{\qof{X}} &\ge \xHof{\pof{X}}, \label{practical_it_notation:eq:cross_entropy_vs_entropy} \\
        \Kale{\pof{X}}{\qof{X}} &\ge 0,
    \end{align}  
    with equality exactly when $\opq = \opp$.\andreas{equality under distribution, e.g. for almost all}
    \end{corollary}
    Note that for continuous distributions, above equality \(\opp = \opq\) only has to hold almost everywhere.
    
    Above definitions are trivially extended to joints of random variables by substituting the random variable of the product space. Similarly, the conditional entropy is defined by taking the expectation over both \(X\) and \(Y\).
    For example:
    \begin{proposition}
    Given random variables \(X\) and \(Y\), we have:
    \begin{align}
        \Hof{X, Y} &= \simpleE{\pof{x, y}}{\ICof{\pof{x, y}}}; \\
        \Hof{X \given Y} &= \simpleE{\pof{x, y}}{\ICof{\pof{x \given y}}}.
    \end{align}
    \end{proposition}
    In particular, note that \(\Hof{X \given Y}\) is an expectation over \(X\) \emph{and} \(Y\).
    
    For cross-entropies and KL divergences, we expand the definitions similarly. In particular, we have the following equality for cross-entropies, which follows from these definitions:
    \begin{align}
    &\CrossEntropy{\pof{X \given Y}}{\qof{X \given Y}} = \simpleE{\pof{x,y}}{\ICof{\qof{x \given y}}} \quad = \CrossEntropy{\pof{X, Y}}{\qof{X \given Y}}.
    \end{align}
    The last idiosyncrasy only applies to cross-entropies. For KL divergences we have:
    \begin{align}
    &\Kale{\pof{X \given Y}}{\qof{X \given Y}} = \CrossEntropy{\pof{X, Y}}{\qof{X \given Y}} - \xHof{\pof{X \given Y}} \\
    &\Kale{\pof{X, Y}}{\qof{X \given Y}} = \CrossEntropy{\pof{X, Y}}{\qof{X \given Y}} - \xHof{\pof{X, Y}}.
    \end{align}
    The second terms on each right-hand side are usually not equal \(\xHof{\pof{X \given Y}} \not= \xHof{\pof{X, Y}}\), and the two expressions are thus not equivalent.
    The reader might wonder when we are interested in \(\Kale{\pof{X, Y}}{\qof{X \given Y}}\). It can arise when performing symbolic manipulations, so we mention it explicitly here.

    The mutual information and point-wise mutual information \citep{fano1962transmission,church1990word} are defined as:
    \begin{definition}
    For random variables $X$ and $Y$ and outcomes $x$ and $y$ respectively, the point-wise mutual information $\MIof{x; y}$ and the mutual information $\MIof{X; Y}$ are:
    \begin{align}
        & \MIof{x; y} \defeq \Hof{x} - \Hof{x \given y} = \ICof{\frac{\pof{x}\pof{y}}{\pof{x, y}}} \\
        & \MIof{X; Y} \defeq \Hof{X} - \Hof{X \given Y} = \simpleE{\pof{x, y} } \MIof{x; y}.
    \end{align}
    \end{definition}
    This is similarly extended to $\MIof{X ; Y \given Z} = \Hof{X \given Z} - \Hof{X \given Y, Z}$ or $\MIof{X_1, X_2 ; Y} = \Hof{X_1, X_2} - \Hof{X_1, X_2 \given Y}$ and so on.
      
    \citet{mackay2003information} has an elegant visualization for information quantities, which we reproduce in \Cref{practical_it_notation:fig:mackay_81}. 

    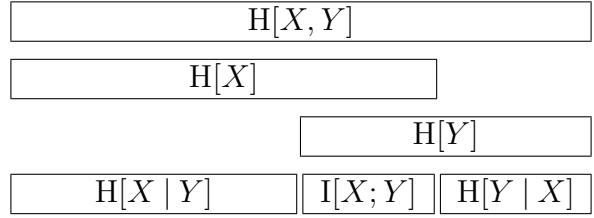
\begin{figure}[t]
        \begin{center}
        \setlength{\unitlength}{1in}
        \begin{picture}(3,1.10)(0,-0.2)
        \put(0,0.7){\framebox(3,0.20){\(\Hof{X,Y}\)}}
        \put(0,0.4){\framebox(2.2,0.20){\(\Hof{X}\)}}
        \put(1.5,0.1){\framebox(1.5,0.20){\(\Hof{Y}\)}}
        \put(1.5125,-0.2){\framebox(0.675,0.20){\(\MIof{X;Y}\)}}
        \put(0,-0.2){\framebox(1.475,0.20){\(\Hof{X \given Y}\)}}
        \put(2.225,-0.2){\framebox(0.775,0.20){\(\Hof{Y \given X}\)}}
        \end{picture}
        \end{center}
        \caption{\emph{Reproduction of Figure 8.1 from \citet{mackay2003information} using the new suggested notation:} The relationship between joint entropy, 
            marginal entropy, conditional entropy and mutual information.
        \label{practical_it_notation:fig:mackay_81}
        }%
    \end{figure}

    \subsubsection{Information Diagrams (I-Diagrams)}
    \label{practical_it_notation:glue:background:information diagrams}

    \begin{figure}
        \centering
        \tikzset{every picture/.style={line width=0.75pt}} %

        \begin{tikzpicture}[
            infoquantity/.style={line width=1.5pt},
            infoquantitylabel/.style={font=\large}
            ]
            \begin{scope}[blend group=darken]
                \draw[infoquantity, fill=solarized@red!50!white]   ( 90:1.2*3/2) circle (3);
                \draw[infoquantity, fill=solarized@green!50!white] (210:1.2*3/2) circle (3);
                \draw[infoquantity, fill=solarized@blue!50!white]  (330:1.2*3/2) circle (3);
            \end{scope}
            \path  
                ( 90:2*3/2) node {$\displaystyle \Hof{X \given Y, Z}$} 
                ([turn]0:2.5) node [infoquantitylabel] {$\displaystyle \Hof{X}$};
            \path (210:2*3/2) node {$\displaystyle \Hof{Y \given X, Z}$}
                ([turn]0:2.5) node [infoquantitylabel] {$\displaystyle \Hof{Y}$};
            \path (330:2*3/2) node {$\displaystyle \Hof{Z \given X, Y}$}
                ([turn]0:2.5) node [infoquantitylabel] {$\displaystyle \Hof{Z}$};
            \node {$\displaystyle \MIof{X;Y;Z}$};
            \node at ( 30:1.5*3/2) [rotate=30] {$\displaystyle \MIof{X; Z \given  Y}$};
            \node at (150:1.5*3/2) [rotate=-30] {$\displaystyle \MIof{X; Y \given  Z}$};
            \node at (270:1.5*3/2-0.5) [rotate=0] {$\displaystyle \MIof{Y; Z \given  Y}$};
        \end{tikzpicture}
        \caption{Example of an I-Diagram for three random variables $X, Y, Z$. All seven atomic quantities are depicted as well as the overall entropies $\Hof{X}, \Hof{Y}, \Hof{Z}$.} %
        \label{practical_it_notation:glue:background:idiagram}
    \end{figure}
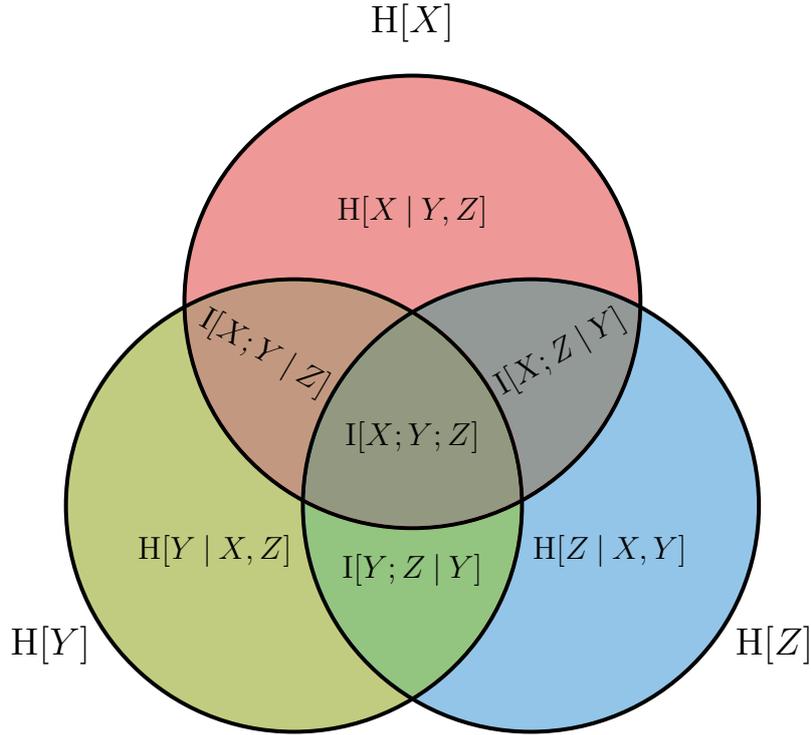

    Similarly, \citet{yeung1991new} introduces \emph{information diagrams (I-diagrams)} which provide another useful intuitive approach: they show that the intuitions that map information quantities to set expressions can be made principled using a specially defined signed information measure. 
    Note that interaction information \citep{mcgill1954multivariate} follows as canonical generalization of the mutual information to multiple variables from that work. %
    \citet{lang2022information} further generalize these results.

    Information diagrams, like the one depicted in \cref{practical_it_notation:glue:background:idiagram}, visualize the relationship between information quantities: \citet{yeung1991new} shows that we can define a signed measure $\opmus$ such that these well-known quantities map to abstract sets and are consistent with set operations.
    \begin{align*}
        \IMof{A} &= \Hof{A} \\
        \IMof{\cup_i A_i} &= \Hof{A_1, \ldots, A_n} \\
        \IMof{\cup_i A_i - \cup_i B_i} &= \Hof{A_1, \ldots, A_n \given B_1, \ldots, B_n} \\
        \IMof{\cap_i A_i} &= \MIof{A_1;\ldots; A_n} \\
        \IMof{\cap_i A_i - \cup_i B_i} &= \MIof{A_1;\ldots; A_n \given B_1,\dots,B_n}
    \end{align*}

    In other words, equalities can be read off directly from I-diagrams: an information quantity is the sum of its parts in the corresponding I-diagram. This is similar to Venn diagrams. The sets used in I-diagrams are just abstract symbolic objects, however.

    An important distinction between I-diagrams and Venn diagrams is that while we can always read off inequalities in Venn diagrams, this is not true for I-diagrams in general because mutual information terms in more than two variables can be negative. In Venn diagrams, a set is always larger or equal any subset.

    However, if we show that all information quantities are non-negative, we can treat an I-diagram like a Venn diagram and read off both equalities and inequalities from it.
    Nevertheless, caution is warranted sometimes.
    As the signed measure can be negative, $\IMof{X \cap Y} = 0$ does \emph{not} imply $X \cap Y = \emptyset$: deducing that a mutual information term is $0$ does not imply that one can simply remove the corresponding area in the I-diagram. There could be $Z$ with $\IMof{(X \cap Y) \cap Z} < 0$, such that $\IMof{X \cap Y} = \IMof{X \cap Y \cap Z} + \IMof{X \cap Y - Z} = 0$ but $X \cap Y \neq \emptyset$.
    This also means that we cannot drop the term from expressions when performing symbolic manipulation,  and we cannot remove the area from an I-diagram without loss of generality:
    \begin{importantresult}
    While a mutual information term of two random variables equalling zero implies those two random variables are independent, the mutual information of those two random variables with a third might not be zero, and thus does not allow us to draw them as disjoint areas.
    \end{importantresult}

    The only time when one can safely remove an area from the diagram is for \emph{atomic} quantities, which are quantities which reference all the available random variables \citep{yeung1991new}. For example, when we only have three variables $X, Y, Z$, $\MIof{X ;Y ; Z}$ and $\MIof{X ; Y \given Z}$ are atomic quantities, like in \Cref{practical_it_notation:glue:background:idiagram}. We can safely remove atomic quantities from I-diagrams when they are $0$ because there are no other random variables left that could lead to the issues described above.

    Continuing the example, for $\MIof{X;Y} = \IMof{X \cap Y} = 0$, having $0 = \MIof{X ;Y ;Z} = \IMof{X \cap Y \cap Z}$ would imply $X \cap Y \cap Z = \emptyset$, and we could remove that area from the diagram without loss of generality. Moreover, the atomic quantity $\MIof{X ; Y \given Z} = \IMof{X \cap Y - Z} = 0$ and could be removed from the diagram as well in that case. 

    We only use I-diagrams for the three variable case, but they supply us with tools to easily come up with equalities and inequalities for information quantities. In the general case with multiple variables, they can be difficult to draw, but for Markov chains they can be of great use \citep{yeung1991new}.
}

    \subsection{Bayesian Neural Networks (BNNs)}
    \label{glue:related_work:BNNs}

    In this thesis, we focus on Bayesian neural networks (BNNs) \citep{neal1995bayesian, mackay1992bayesian}. Unlike regular neural networks, BNNs maintain a distribution over their weights instead of point estimates.
    This allows for better uncertainty quantification and for disentangling different types of uncertainties \citep{kendall2017uncertainties}. Compared to other Bayesian approaches, BNNs scale well to high-dimensional inputs, such as images while remaining close to neural networks conceptually, allowing to apply advances in deep learning in Bayesian settings \citep{sharma2023bayesian, gal2017deep}.
    Hence, BNNs have become a powerful alternative to traditional neural networks.

    \paragraph{Challenges for BNNs.}
    However, performing exact inference in BNNs is intractable for any reasonably sized model, so we resort to using a variational approximation.
    The intractability of exact Bayesian inference in deep learning has led to the development of approximate inference methods \citep{hinton1993keeping, hernandez2015probabilistic, blundell2015weight, gal2015dropout}.
    Improvements in approximate inference \citep{blundell2015weight,gal2015dropout} have enabled their usage for high dimensional data such as image for Bayesian active learning of images \citep{gal2017deep}.
    Similar to \citet{gal2017deep}, we will mainly use MC dropout \citep{gal2015dropout}, which is easy to implement, scales well to large models and datasets, and is straightforward to optimize.
    Deep Ensembles \citep{lakshminarayanan2016simple} can also be considered as an alternative to BNNs.

    \paragraph{Probabilistic Model.}
    We consider supervised learning of a probabilistic predictive model, $\pof{\y, \w \given \x}$, where $X$ is an input, $Y$ is a label, and $\W$ are the model parameters of our model $\Model$ with prior distribution $\pof{\w}$:
    \begin{align}
        \pof{\y, \w \given \x} = \pof{\y \given \x, \w} \pof{\w}.
        \label{pobal:eq:marg_pred_posterior}
    \end{align}
    Importantly, we assume that the model's predictions are independent given $\W$, such that we can write:
    \begin{align}
        \pof{\y_1, \ldots, \y_n \given \x_1, \ldots, \x_n, \w} = \prod_{i=1}^n \pof{\y_i \given \x_i, \w}.
        \label{pobal:eq:joint_pred_posterior}
    \end{align}
    We assume classification tasks, and thus $\Y \in [\numclasses] \defeq \{ 1, \ldots, \numclasses\}$. The exceptions are \Cref{chapter:bbbal,chapter:causalbald}.

    \paragraph{Datasets.} 
    Further, we assume we have datasets $\Dtrain$ and $\Dtest$, with $\Dtest \sim \pdataof{\y, \x},$ the `true' underlying data distribution. We use $\Dany$ to represent additional data that we might condition on.
    We will define additional datasets in the respective chapters and as we go along (for example, the unlabeled pool set in active learning as $\Dpool$). 
    
    \paragraph{Bayesian Model Averaging and Bayesian Inference.}
    We are interested in the \emph{BMA (Bayesian model averaging)} prediction given the Bayesian posterior $\pof{\w \given \Dtrain}$.
    Bayesian model averaging (BMA) is performed by marginalizing over $\W$ to obtain the \emph{predictive distribution} $\pof{\y \given \x, \Dtrain}$:
    \begin{align}
        \pof{\y \given \x, \Dtrain} = \E{\pof{\w \given \Dtrain}}{\pof{\y \given \x, \w}},
    \end{align}
    where we use \emph{Bayesian inference} to obtain a posterior $\pof{\w \given \Dtrain}$ for data $\Dtrain$ by:
    \begin{align}
        \pof{\w \given \Dtrain} \propto \pof{\Dtrain \given \w} \pof{\w}.
    \end{align}
    $\pof{\Dtrain \given \w}$ is the likelihood of the data given the parameters $\W$, and $\pof{\w \given \Dtrain}$ is the new posterior distribution over $\W$.

    \paragraph{Variational Inference.}
    As mentioned, Bayesian inference is often intractable, and instead, we use variational inference methods to approximate the posterior $\pof{\w \given \Dtrain}$ via a simpler distribution $\qof{\w} \approx \pof{\w \given \Dtrain}$.
    Usually, this is phrased as an optimization problem where we try to find the distribution $\opq$ from a variational family of potential distributions that minimizes a given divergence measure between $\qof{\w}$ and $\pof{\w \given \Dtrain}$:
    \begin{align}
        \qof{\w} = \argmin_{\qof{\w}} \mathrm{D}(\qof{\w} \mathbin{\vert \vert} \pof{\w \given \Dtrain}).
    \end{align}
    We will use the KL divergence $\Kale{\qof{\w}}{\pof{\w \given \Dtrain}}$ as divergence measure in this thesis, but other divergence measures have been proposed and are used in practice, as well.
    Instead of above KL divergence which is usually intractable, the \emph{ELBO (Evidence Lower Bound)} is commonly used to optimize the variational approximation instead. By expanding above KL divergence, we obtain:
    \begin{align}
        &\Kale{\qof{\w}}{\pof{\w \given \Dtrain}} \\ 
        &\quad = \simpleE{\qof{\w}} [-{\underbrace{\log \pof{\ytrainset \given \xtrainset, \w}}_\text{log likelihood}}]
        + \underbrace{\Kale{\qof{\w}}{\pof{\w}}}_\text{prior regularization} + \underbrace{\log \pof{\Dtrain}}_\text{model evidence} \ge 0,
    \end{align}
    The model evidence is independent of $\qof{\w}$. It can be ignored during optimization. Rearranging, leaves us with the ELBO, which we then maximize:
    \begin{align}
        \simpleE{\qof{\w}} [{\log \pof{\ytrainset \given \xtrainset, \w}}]
        - {\Kale{\qof{\w}}{\pof{\w}}} \le \log \pof{\Dtrain}.
    \end{align}
    We can use the local reparameterization trick and Monte-Carlo (MC) dropout to obtain an implicit $\qof{\w}$ distribution we can draw samples from in a deep learning context \citep{gal2015dropout}.
    The posterior BMA using likelihood $\pof{\Dtrain \given \w}$ and prior $\pof{\w}$ is then approximated by:
    \begin{align}
        \pof{\y \given \x, \Dtrain} \approx \E{\qof{\w \given \Dtrain}}{\pof{\y \given \x, \w}}.
    \end{align}
    We often omit making this last step explicit in our deductions and use $\pof{\w \given \Dtrain}$ when, in practice, we would substitute $\qof{\w}$.

    \paragraph{Stochastic Parameters and Model Choices.}
    Importantly, we often use of Bayesian models in the narrow sense that we only require some \emph{stochastic} parameters $\W$ with a distribution $\pof{\w}$.
    This choice of $\pof{\w}$ covers (deep) ensembles \citep{dietterich2000ensemble, lakshminarayanan2016simple}, neural networks with stochasticity in a subset of parameters \citep{sharma2023bayesian}, as well as models with additional stochastic inputs \citep{osb2021epistemic} or randomized training data through subsampling of the training set, e.g., bagging \citep{breiman1996bagging}. Similar views have been put forward by \citet{He2020Bayesian, wilson2020bayesian} for deep ensembles, and for random forests \citep{shaker2020aleatoric}.

    \subsection{Uncertainty Quantification}
    \label{glue:background:uncertainty_quantification}
    {
        \graphicspath{{./ddu/figs/}{./ddu/}}

\renewcommand{\textuparrow}{$\uparrow$}
\renewcommand{\textdownarrow}{$\downarrow$}
\newcommand{\cmark}{\ding{51}}%
\newcommand{\xmark}{\ding{55}}%

\newcommand{\tHof}[1]{\varHof{\theta}{#1}}
\newcommand{\tpof}[1]{\pcof{\theta}{#1}}
\newcommand{\tqof}[1]{\qcof{\theta}{#1}}

        Uncertainty quantification is a broad field, but two types of uncertainty are often of interest in machine learning: \emph{epistemic uncertainty}, which is inherent to the model, caused by a lack of training data, and hence reducible with more data, and \emph{aleatoric uncertainty}, caused by inherent noise or ambiguity in data, and hence irreducible with more data \citep{der2009aleatory}.
        Disentangling these two is critical for tasks such as:
        \begin{itemize}
            \item \textbf{Active Learning.} In active learning, we want to avoid inputs with high aleatoric and low epistemic uncertainty, as these will not be informative for the model \citep{gal2017deep}.
            \item \textbf{Out-of-Distribution Detection.} In OoD detection \citep{hendrycks2016baseline}, we want to avoid flagging ambiguous in-distribution (iD) examples as OoD.
            \item \textbf{Deferral of Uncertain Predictions.} To defer predictions \citep{filos2019systematic}, we seek inputs with either high epistemic uncertainty or high aleatoric uncertainty, for different purposes: in the former case, we want to defer to an expert for a decision, while in the latter case, we want to defer to the data source for a better measurement.
        \end{itemize}        
        These tasks and distinctions matter in particular for noisy and ambiguous datasets found in safety-critical applications like autonomous driving \citep{huang2020autonomous} and medical diagnosis \citep{esteva2017dermatologist, filos2019systematic}.

        While there are many ways to approximate these uncertainties using metrics, we will focus on three in the next chapters: \emph{mutual information}, \emph{entropy}, and \emph{feature-space density}. 

        In this subsection, we will define epistemic and aleatoric uncertainty in more detail and give a brief overview of these three metrics and relevant concepts:
        \begin{description}
        \item[Epistemic Uncertainty.]
        For point $\x$, epistemic uncertainty is a quantity which is high for a previously unseen $\x$, and decreases when $\x$ is added to the training set and the model is updated \citep{kendall2017uncertainties}. This conforms with using mutual information in Bayesian models and deep ensembles \citep{smith2018understanding} and feature-space density in deterministic models as surrogates for epistemic uncertainty \citep{postels2020quantifying} as we will examine in chapter \Cref{chapter:ddu}.

        \item[Aleatoric Uncertainty.]
        For a point $\x$, aleatoric uncertainty is a quantity which is high for ambiguous or noisy samples \citep{kendall2017uncertainties}. For example, in classification, aleatoric uncertainty will be high when multiple labels were to be observed at $\x$. Crucially, aleatoric uncertainty does not decrease with more data because it is inherent to the data. As we gather more data, it can actually increase as we might have undersampled the data distribution.
        Note that aleatoric uncertainty is only meaningful in-distribution,
        as, by definition, it quantifies the level of noise or ambiguity which might be observed. More practically speaking, if the probability of observing $\x$ under the data generating distribution is zero, the conditional probability $\pof{\y \given \x}=\frac{\pof{\x, \y}}{\pof{\x}}$ is undefined, and hence, neither is the respective entropy as a measure of aleatoric uncertainty.

        \item[Bayesian Models.]
        To measure uncertainty, principled approaches exist for Bayesian models \citep{Gal2016Uncertainty}. 
        Given a Bayesian model $\pof{\y, \w \given \x}$, its predictive entropy $\Hof{Y \given \x, \Dtrain}$ upper-bounds the epistemic uncertainty, where epistemic uncertainty is quantified as the mutual information $\MIof{\Y; \W \given \x, \Dtrain}$ between parameters $\W$ and output $\Y$ for a given input $\x$ \citep{Gal2016Uncertainty, smith2018understanding}:
        \begin{align}
            \underbrace{\Hof{\Y \given \x, \Dtrain}}_\text{predictive} &= \underbrace{\MIof{\Y; \W \given\x, \Dtrain}}_\text{epistemic} + %
            \underbrace{\Hof{Y \given \x, \W, \Dtrain}}_\text{aleatoric (for iD $\x$)} \\
            \Leftrightarrow \MIof{\Y; \W \given\x, \Dtrain} &= \Hof{\Y \given \x, \Dtrain} - \Hof{Y \given \x, \W, \Dtrain}.
            \label{glue:eq:BALD}
        \end{align}
        Predictive entropy will be high for both iD ambiguous samples (high aleatoric uncertainty) and for OoD samples (high epistemic uncertainty). Hence, predictive entropy is a good measure for informativeness for active learning or as a metric for OoD detection only when used with curated datasets that do not contain ambiguous samples.
        Note that aleatoric uncertainty is only meaningful in-distribution
        because it quantifies the level of noise or ambiguity which might be observed for input $\x$.

        Looking at the two terms in equation \eqref{glue:eq:BALD}, for the mutual information to be high, the left term has to be high and the right term low. The left term is the entropy of the model prediction, which is high when the model's prediction is uncertain. The right term is an expectation of the entropy of the model prediction over the posterior of the model parameters and is low when the model is overall certain for each draw of model parameters from the posterior. Both can only happen when the model has many possible ways to explain the data, which means that the different models induced by different parameter samples are disagreeing among themselves.

        This intuitively satisfies the definition of epistemic uncertainty above, as adding a point to the training set ought to decrease the epistemic uncertainty as there will be less disagreement among the models induced by different parameter samples after training with the point.

        \item[Deep Ensembles.]
        The predictions of a deep ensemble \citep{lakshminarayanan2016simple} are the average of the outputs of an ensemble of neural networks. The total uncertainty of the prediction is then estimated as the entropy of the averaged softmax outputs. This can be viewed as approximating the BMA over the distribution of all possibly trained models \citep{wilson2020bayesian}, as each ensemble member, producing a softmax output $\pof{\y \given \x, \w}$, can be considered to be drawn from some distribution $\pof{\w}$ of the possibly trained model parameters $\W$, which is induced by the push forward of the weight initialization under stochastic optimization. As a result, \Cref{glue:eq:BALD} can also be applied to Deep Ensembles to disentangle aleatoric and epistemic uncertainty from predictive uncertainty.

        Despite the high computational overhead at training and test time, Deep Ensembles along with recent extensions \citep{smith2018understanding,wen2020batchensemble, dusenberry2020efficient} form the state-of-the-art in uncertainty quantification in deep learning.

        In practice, both mutual information $\MIof{Y; \W \given \x, \Dtrain}$ and predictive entropy $\Hof{Y \given \x, \Dtrain}$ are used in the literature for active learning and to detect OoD data, but predictive entropy will be high whenever either epistemic uncertainty is high, or when aleatoric uncertainty is high: it upper-bounds the mutual information.
        This can help separate iD and OoD data better for curated iD datasets, offering an explanation for previous empirical findings of predictive entropy outperforming mutual information \citep{malinin2018predictive}. With ambiguous iD samples, it can lead to confounding, however, which we analyze in chapter \Cref{chapter:ddu}.
        
        \item[Deterministic Models.]
        A single deterministic model, in the sense that we use a single parameter point estimate, produces a softmax distribution $\pof{\y \given \x, \w}$, and commonly either the \emph{softmax confidence} $\max_c \pof{\y=c \given \x, \w}$ or the \emph{softmax entropy} $\Hof{Y \given \x, \w}$ are used as a measure of uncertainty \citep{hendrycks2016baseline}. In active learning, the softmax entropy and confidence are often used as a baseline measure of informativeness. Note that the confidence is known as variation ratios in active learning \citet{freeman1965elementary}. In both settings, they do not perform as well as deep ensembles \citep{beluch2018power}. 

        Popular approaches to improve these metrics include pre-processing of inputs and post-hoc calibration methods \citep{liang2017enhancing, guo2017calibration}, alternative objective functions \citep{lee2017training, devries2018learning}, and exposure to outliers \citep{hendrycks2018deep}. However, these methods are known to suffer from shortcomings:  they can fail under distribution shift \citep{ovadia2019can}, require significant changes to the training setup, or assume the availability of OoD samples during training.
        
        \item[Feature-Space Distances] \textbf{\&\ Feature-Space Density.}
        Feature-space distances \citep{lee2018simple, van2020uncertainty, liu2020simple} and feature-space density \citep{postels2020quantifying,Settles2010} based on the training set offer a different approach for estimating uncertainty in deterministic models: satisfying the definition of epistemic uncertainty above, they decrease when previously unseen samples are added to the training set.
        This is the case for feature-space distance and density methods because they estimate distance or density, respectively, of the training data in feature space. %
        A previously unseen point with high distance (low density), once added to the training data, will have low distance (high density).
        Hence, they can be used as a proxy for epistemic uncertainty, under important assumptions about the feature space as detailed below. We will examine this in more detail in \S\ref{chapter:ddu}.
        None of these methods, however, is competitive with Deep Ensembles, in uncertainty quantification, potentially for the reasons discussed next.
        
        \item[Feature Collapse.]
        \citet{van2020uncertainty} argue that feature collapse is why distance and density estimation in the feature space may fail to capture epistemic uncertainty:
        feature extractors might map the features of OoD inputs into iD regions in the feature space \citep{van2021improving}, making it difficult for the later stages to distinguish between iD and OoD samples.
        
        \item[Smoothness \&\ Sensitivity.]
        To prevent feature collapse, smoothness and sensitivity can be encouraged by subjecting the feature extractor $f_\theta$, with parameters $\theta$ to a \emph{bi-Lipschitz constraint}:
        \begin{equation*}
            \label{glue:eq:bi-lipschitz}
            K_L \; d_I(\x_1, \x_2) \leq d_F(f_\theta(\x_1), f_\theta(\x_2)) \leq K_U \; d_I(\x_1, \x_2),
        \end{equation*}
        for all inputs, $\x_1$ and  $\x_2$, where $d_I$ and $d_F$ denote metrics for the input and feature space respectively, and $K_L$ and $K_U$ the lower and upper Lipschitz constants \citep{liu2020simple}. 
        The lower bound ensures \emph{sensitivity} to distances in the input space,
        and the upper bound ensures \emph{smoothness} in the features, preventing them from becoming too sensitive to input variations, which, otherwise, can lead to poor generalization and loss of robustness \citep{van2020uncertainty}.
        Methods of encouraging bi-Lipschitzness include: 
        \textbf{i)} gradient penalty, by applying a two-sided penalty to the L2 norm of the Jacobian \citep{gulrajani2017improved}, and
        \textbf{ii)} spectral normalization \citep{miyato2018spectral} in models with residual connections, like ResNets \citep{he2016deep}.
        \citep{smith2021convolutional} provides an in-depth analysis which supports that spectral normalization leads to bi-Lipschitzness.
        Compared to the Jacobian gradient penalty \citep{van2020uncertainty}, spectral normalization is significantly faster and has more stable training dynamics.
    \end{description}
    }
    \subsection{Active Learning} 
    \label{glue:background:active_learning}

    One key problem in deep learning is data efficiency. While excellent performance can be obtained with modern approaches, these are often data-hungry, rendering the deployment of deep learning in the real-world challenging for many tasks.
    Active learning \citep{atlas1989training,cohn1996active} is a powerful technique for improving labelling efficiency;  
    it has a rich history in the machine learning community, with its origins dating back to seminal works such as \citet{atlas1989training,lindley1956measure,fedorov1972theory,mackay1992information}. A comprehensive survey of early active learning methods can be found in \citet{Settles2010}, while more recent surveys of contemporary deep learning methods can be found in  \citet{ren2021survey} and \citet{zhan2022comparative}.

    Instead of a priori collecting and labelling a large dataset, which often comes at a significant expense, 
    active learning provides a mechanism for effective training of machine learning models in settings where unlabeled data is plentiful, but labelling is expensive by carefully selecting which data points to acquire labels for, using information from previously acquired data to establish the points whose labels will be most informative for training.
    After each acquisition step, the newly labeled points are added to the training set, and the model is retrained.
    This process is repeated until a suitable level of accuracy is achieved.
    The goal of active learning is to minimize the amount of data that needs to be labeled.
    Active learning has made real-world impact in manufacturing \citep{tong2001active}, robotics \citep{calinon2007learning}, recommender systems  \citep{adomavicius2005toward}, medical imaging \citep{hoi2006batch}, and NLP \citep{siddhant2018deep}, motivating further exploration of this fascinating topic.

    \paragraph{Origins.}
    The conceptual origins of active learning can be traced back to Bayesian-optimal experiment design \citep{chaloner1995bayesian,lindley1956measure,rainforth2023modern}. %
    In machine learning, the non-Bayesian approaches started in the sequential (stream-based) setting as \emph{selective sampling} \citep{atlas1989training} before being referred to more generally as \emph{active learning} \citep{atlas1989training}.
    Bayesian methods and active learning were connected early on by e.g.\ \citet{mackay1992information} with objectives that are still highly relevant today---as we will see, these objectives are especially relevant for this thesis.
    An excellent literature review of the original non-Bayesian active learning paradigms can be found in \citet{Settles2010}.

    \paragraph{Bayesian Optimization.} 
    Active learning is also closely related to \emph{Bayesian Optimization (BO)}, which is a well-established methodology for global optimization of black-box functions \citep{Mockus_1974,Jones_1998}, especially when the function evaluations are expensive. 
    Typical applications of BO span from machine learning and statistics to engineering and experimental design \citep{Snoek_2012, Shahriari_2016, Saleh_2022} under various constraints \citep{Nguyen_2017,Siivola_2020}. 

    BO is based on building a probabilistic surrogate model of the black-box function, often a Gaussian process \citep{williams2006gaussian}, and using an acquisition function to drive the search towards the regions of the input space that are likely to improve upon the current best solution \citep{Shahriari_2016}. The queries are sequentially selected and have to balance exploration and exploitation \citep{Srinivas_2010}. 

    Many BO methods utilize the expected improvement heuristic \citep{Hernandez-Lobato_2014,Hernandez-Lobato_2015,Hernandez-Lobato_2016} or rely on Thompson sampling (TS) \citep{thompson1933likelihood,Russo2013LearningTO}.

    \paragraph{Batch Bayesian Optimization.} 
    The classical BO setting is sequential, where the next query directly depends on the outcomes of the previous evaluations. However, this does not take into account that often experiments can be conducted in parallel, such as in large-scale computing environments or in laboratory experiments where measurements may come from different sources and may introduce significant waiting times otherwise \citep{Folch_2022}. 
    
    To address this, \emph{batch Bayesian optimization (BBO)} methods have been developed, where multiple decisions are made simultaneously and evaluated in parallel, providing a significant acceleration over the sequential setting \citep{Groves_2018, Chowdhury_2019, Oh_2021}.
    Many BBO methods have been proposed in the literature, including methods based on the expected improvement heuristic \citep{Shah_2015,Gonzalez_2016,alvi2019asynchronous}, Thompson sampling \citep{Kandasamy_2018}, and upper confidence bound \citep{Contal_2013,Daxberger_2017}. 

    \paragraph{Semi-Supervised Learning.}
    A related approach to active learning is semi-supervised learning (also sometimes referred to as weakly-supervised), in which the labeled data is commonly assumed to be
    fixed, and the unlabeled data is used for unsupervised learning
    \citep{kingma2014semi,rasmus2015semi}. \citet{wang2017cost, sener2017active,
    sinha2019variational} explore combining it with active learning.

    \paragraph{Pool-Based Active Learning.}
    Active learning is most often done in a \emph{pool}-based setting \citep{lewis1994sequential}, wherein we start with a large reservoir of unlabeled data points, known as the \emph{pool set} $\Dpool$, from which we sequentially choose points to label, after which they are removed from $\Dpool$ and added to the \emph{training dataset} $\Dtrain$ together with their acquired label. 
    In this thesis, we focus on such settings.
    The steps of an active learning loop in a pool-based setting are depicted in \Cref{glue:related_work:active_learning_loop}.
    The main challenge in pool-based batch active learning is the choice of the acquisition function.

    \paragraph{Acquisition Functions.}
    The mechanism by which we choose points to label is known as an acquisition strategy and most commonly corresponds to choosing the data point which maximizes a prespecified \emph{acquisition function} that reflects the utility of acquiring a label for that point.
    (Similarly, as we will see, when we perform batch acquisition, we will want to find the batch of points that maximize a joint utility score.)
    We will define acquisition functions as functions that return a score for an individual sample $\theAF \left ( \x, \Model \right )$ given the current model $\Model$.
    We then acquire labels for the sample that maximizes the score:
    \begin{align}
            \xacqstar \defeq \argmax_{ \xacq \in \Dpool} {\theAF \left ( \xacq, \Model \right )}
            .
    \end{align}
    There are several simple acquisition functions which are often used as baselines \citep{gal2017deep}:
    \begin{description}
        \item[Entropy.] The predictive entropy is used as acquisition score. It is defined as:
        \begin{align}
            \theAF_{\mathrm{Entropy}}(\x; \Model) &\defeq \Hof{\Y \given \x}.
        \end{align}
        It is non-negative and measures the total uncertainty that model assigns to an input $\x$.
        
        \item[Variation Ratio.] Also known as ``\emph{least confidence}'', the variation-ratio is the complement of the most-confident class prediction and thus selects samples with the lowest confidence for acquisition:
        \begin{align}
            \theAF_{\mathrm{Variation-Ratios}}(\x; \Model) &\defeq 1 - \max_\y \pof{\y \given \x}.
        \end{align}
        This scoring function is non-negative and a score of $0$ means that the sample is uninformative: a score of $0$ means that the respective prediction is one-hot, and then that the expected information gain is also $0$, as can be easily verified. %

        \item[Standard Deviation.] The standard deviation score function measures the sum of the class probability deviations and is closely related to the BALD scores (\Cref{jiang:prop:approxmivarsum}, \citet{smith2018understanding}):
        \begin{align}
            \theAF_{\mathrm{Std-Dev}}(\x; \Model) &\defeq \sum_\y \sqrt{\Var{\pof{\w}}{\pof{\y \given \x, \w}}}.
        \end{align}
        This scoring function is also non-negative, and zero variance for the predictions implies a zero expected information gain and thus an uninformative sample. %
    \end{description}

    \paragraph{Information-Theoretic Acquisition Functions.}
    Strategies for constructing such an acquisition function can be based on principled information-theoretic considerations that allow formalizing the notion of the information that will be gained for labelling any given point.
    These approaches usually require a probabilistic model $\pof{\y \given \x, \w}$ for label $\y$ given input $\x$, where $\w$ represents a realization of stochastic model parameters $\W$; a particularly common choice of model is a Bayesian neural network, wherein $\W$ represents the weights and biases.

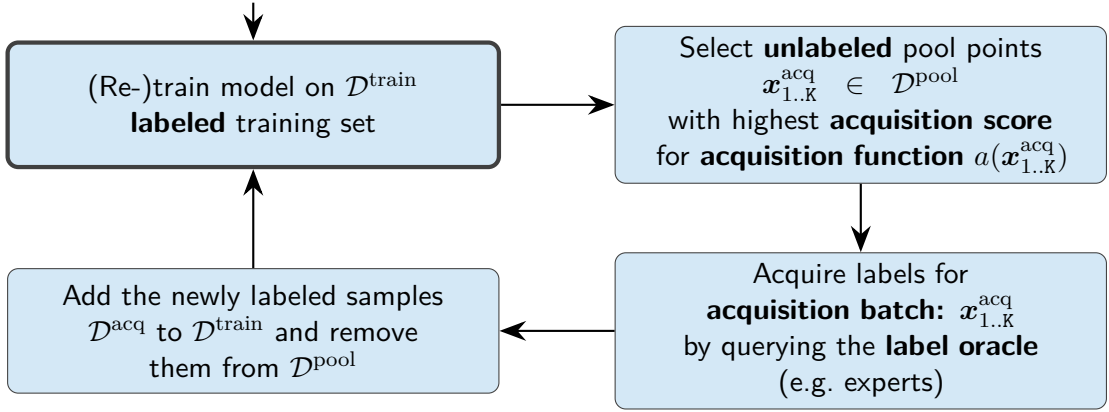
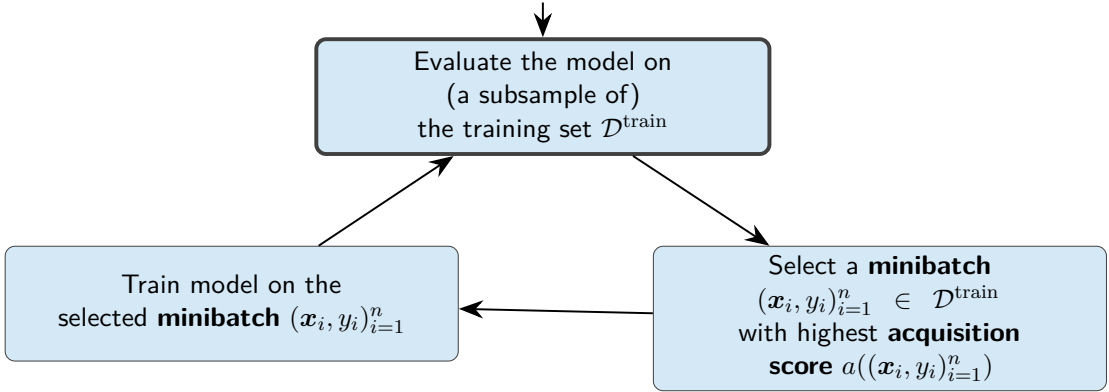
\begin{figure}
    \centering
    \begin{subfigure}[t]{\linewidth}
        \centering
        \resizebox{0.95\textwidth}{!}{
        \begin{tikzpicture}[
            on grid,
            node distance = 3cm and 8cm,
            auto,
            block/.style = {
                rectangle,
                rounded corners,
                draw=black!75,
                fill=solarized@blue!20,
                text width=15em,
                align=center,
                minimum height=4em,
                font=\small\sffamily
            },
            line/.style = {
                draw,
                thick,
                -{Stealth[length=10pt]}
            }
        ]

            \node [block, ultra thick] (init) {(Re-)train model on $\Dtrain$\\\textbf{labeled} training set};
            \node [block, right=of init] (select) {Select \textbf{unlabeled} pool points\\$\xacqset \in \Dpool$\\
            with highest \textbf{acquisition score}\\
            for \textbf{acquisition function} $\theAF(\xacqset)$};
            \node [block, below=of select] (acquire) {Acquire labels for\\\textbf{acquisition batch:} $\xacqset$\\by querying the \textbf{label oracle}\\(e.g.\ experts)};
            \node [block, below=of init] (update) {Add the newly labeled samples\\$\Dacq$ to $\Dtrain$ and remove\\them from $\Dpool$};

            \draw [line] (init.east) -- (select.west);
            \draw [line] (select.south) -- (acquire.north);
            \draw [line] (acquire.west) -- (update.east);
            \draw [line] (update.north) -- (init.south);

            \node [above=of init, yshift=-1.5cm] (start) {};
            \draw [line] (start.south) to (init.north);
        \end{tikzpicture}}
        \subcaption{Active Learning Loop}
        \label{glue:related_work:active_learning_loop}
    \end{subfigure}
    \begin{subfigure}[t]{\linewidth}
        \centering
        \resizebox{0.95\textwidth}{!}{
        \begin{tikzpicture}[
            node distance = 2cm,
            auto,
            block/.style = {
                rectangle,
                rounded corners,
                draw=black!75,
                fill=solarized@blue!20,
                text width=15em,
                align=center,
                minimum height=4em,
                font=\small\sffamily
            },
            line/.style = {
                draw,
                thick,
                -{Stealth[length=10pt]}
            }
        ]

            \node [block, ultra thick, anchor=north] (evaluate) at (0,1.5cm) {Evaluate the model on\\
            (a subsample of)\\
            the training set $\Dtrain$};
            \node [block, anchor=150] (select) at (3cm,-1.5cm) {Select a \textbf{minibatch} \\$\xyset \in \Dtrain$\\
            with highest \textbf{acquisition score} $\theAF(\xyset)$};
            \node [block, anchor=30] (train) at (-3cm,-1.5cm) {Train model on the\\selected \textbf{minibatch} $\xyset$};

            \draw [line] (evaluate) -- (select);
            \draw [line] (select) -- (train);
            \draw [line] (train) -- (evaluate);

            \node [above=of evaluate, yshift=-1.5cm] (start) {};
            \draw [line] (start.south) to (evaluate.north);
        \end{tikzpicture}}
        \subcaption{Active Sampling Loop}
        \label{glue:related_work:active_sampling_loop}
    \end{subfigure}
    \caption{\emph{(Batch) Active Learning and Active Sampling Loops.}
    Both active learning and active sampling loops share the same basic structure but differ in the way they are used to train the model. Active learning is used to train a model on a small dataset, while active sampling is used to train a model on a larger dataset. Active learning acquires labels using an expert and adds them to the dataset, while active sampling has access to the labels. Both use an acquisition function to score to select the most informative samples individually or in batches, which is more common in deep learning applications.
    }
\end{figure}

\subsubsection{Bayesian Active Learning}
\label{glue:background:bayesian_active_learning}

    The Bayesian active learning setup consists of an unlabeled dataset $\Dpool$, the current training set $\Dtrain$, a Bayesian model $\Model$ with model parameters $\w \sim \pof{\w \given \Dtrain}$, and output predictions $\pof{\y \given \x, \w, \Dtrain}$ for data point $\x$ and prediction $\y \in [C]$ in the classification case.
    The conditioning of $\w$ on $\Dtrain$ expresses that the model has been trained with $\Dtrain$.
    Furthermore, an oracle can provide us with the correct label $\tilde{\y}$ for a data point in the unlabeled pool $\x \in \Dpool$.
    The goal is to obtain a certain level of prediction accuracy with the least amount of oracle queries.
    At each acquisition step, a batch of data points $\xacqsetstar = \left \{ \xacqstar_1, \ldots, \xacqstar_\batchvar \right \}$ is selected using an acquisition function $\theAF$ which scores a candidate batch of unlabeled data points $\xacqset = \left \{ \x_1, \ldots, \x_\batchvar \right \} \subseteq \Dpool$ using the current model parameters $\pof{\w \given \Dtrain}$:
    \begin{equation}
        \xacqsetstar = \argmax_{ \xacqset \subseteq \Dpool} {\theAF \left ( \xacqset, \pof{\w \given \Dtrain} \right )}
        .
    \end{equation}
    There are a number of intuitive choices for the acquisition function. We focus on the model uncertainty, which is also known as the expected information gain or BALD \citep{houlsby2011bayesian}, and has proven itself in the context of deep learning \citep{gal2017deep, shen2017deep, janz2017actively}.
    BALD scores points based on how well their label would inform us about the true model parameter distribution.
    
    \subsubsection{BALD \&\ Expected Information Gain}
    \label{glue:background:BALD}
    BALD (\emph{Bayesian Active Learning by Disagreement}) \citep{houlsby2011bayesian} uses an acquisition function that estimates the mutual information between the model predictions and the model parameters.
    It is also referred to as the \emph{Expected Information Gain} \citep{lindley1956measure} or the \emph{Total Information Gain} \citep{mackay1992bayesian}.

    The Expected Information Gain (EIG) for $\W$ under $\pof{\y \given \x, \w}$ was originally introduced in \emph{Bayesian-optimal experiment design (BOED)} \citep{lindley1956measure,chaloner1995,rainforth2023modern} to quantify the utility of data and has a long history \citep{fedorov1972theory}.
    The framework of Bayesian experimental design has many applications outside active learning, and in these applications the model parameters are commonly the quantity of interest---Bayesian optimization \citep{Hennig_2012,Hernandez-Lobato_2014,villemonteix2009informational} being a notable exception.
    The EIG in the parameters is thus often a natural acquisition function in BOED:
    \begin{align}
    \eigof{\x} \defeq \E{\pof{\y \given \x,\Dtrain}}{\xHof{\pof{\W \given \Dtrain}} - \xHof{\pof{\W \given y,\x, \Dtrain}}}, \label{glue:eq:original_EIG}
    \end{align}
    which is equal to the mutual information $\MIof{\W; \Y \given \x}$ between the parameters and the label given the data point $\x$ \citep{cavagnaro2010adaptive}.
    The first term in the expectation is the entropy of the prior distribution over $\W$ and the second term is the entropy of the posterior distribution over $\W$ given the label of $\x$.
    Intuitively, it measures the expected reduction in uncertainty about $\W$ after observing the label of $\x$, where
    \mbox{$\pof{y \given \x,\Dtrain} 
    = \E{\pof{y \given \x,\w}}{\w \given \Dtrain}$} is the marginal predictive distribution of the model.

    Computing the EIG using above expansion can be difficult as we need to compute the conditional entropy $\Hof{\W \given \Y, \x}$---yet this is what many recent approaches effectively attempt to do via the Fisher information. (We do not need to compute the entropy of the parameters $\Hof{\W}$, as it does not depend on the data $\x$ and can be ignored.) We will examine this further in \Cref{chapter:fim}.

    While the EIG focuses on the formulation of the mutual information term as a reduction in model posterior uncertainty given a potential sample, the EIG is also equal to the conditional mutual information between the parameters and the label, $\MIof{\Y; \W \given \x, \Dtrain}$.
    This equivalent formulation, which focuses on the predictive disagreement, was popularized as BALD in deep active learning by \citep{gal2017deep} as \emph{BALD (Bayesian Active Learning by Disagreement)}~\citep{houlsby2011bayesian} when used with Bayesian neural networks \citep{neal1995bayesian}.
    BALD can be much easier to evaluate by sampling from $\W$ without the need for additional Bayesian inference.
    Originally introduced outside the context of deep learning, the only requirement on the model is that it is Bayesian, but notably BALD is often used even when inference is not precisely Bayesian, for example when using Monte Carlo dropout in a neural network \citep{gal2015dropout}.
    Concretely, BALD is defined following \eqref{glue:eq:BALD} as 
    \begin{align}
        \label{batchbald:eq:BALD}
        \baldof{\x} \defeq \MIof{\Y; \W \given \x, \Dtrain} = \Hof{\Y \given \x, \Dtrain}- \Hof{\Y \given \x, \W, \Dtrain},
    \end{align}
    which we already introduced as a means of measuring epistemic uncertainty.

    Overall, both formulations express how strongly the model predictions for a given data point and the model parameters are tied, implying that finding out about the true label of data points with high mutual information will also inform us about the true model parameters.
    Another way to see how BALD captures predictive disagreement is to see that above definition is equivalent to the expected KL divergence between the marginal distribution and the individual distribution, which measures the average disagreement of the posterior predictions:
    \begin{align}
        \MIof{\Y; \W \given \x, \Dtrain} &= {\Kale{\pof{\Y \given \x, \W}}{\pof{Y \given x}}} \\ &= \E{\pof{\w \given \Dtrain}}{\Kale{\pof{\Y \given \x, \w}}{\pof{Y \given \x}}}.
    \end{align}
    
    \subsection{Batch Active Learning}
    \label{glue:background:batch_active_learning}

    As two chapters (\Cref{chapter:batchbald,chapter:stochacq}) of the thesis are concerned with batch active learning, we examine some relevant work here. 

    Researchers in active learning \citep{atlas1989training,Settles2010} have identified the importance of \textit{batch} acquisition as well as the failures of top-\batchvar acquisition using straightforward extensions of single-sample methods in a range of settings including support vector machines \citep{campbellQuery2000,schohnLess2000,brinkerIncorporating2003,guo2008discriminative}, GMMs \citep{azimi2012batch}, and neural networks \citep{sener2017active,ash2020deep,baykal2021lowregret}.
    Many of these methods aim to introduce structured diversity to batch acquisition that accounts for the \textit{interaction} of the points acquired in the learning process.

    Maintaining diversity when acquiring a batch of data has been attempted using constrained optimization \citep{guo2008discriminative} and in Gaussian Mixture Models \citep{azimi2012batch}. In active learning of molecular data, the lack of diversity in batches of data points acquired using the BALD objective has been noted by \citet{janz2017actively}, who propose to resolve it by limiting the number of MC dropout samples and relying on noisy estimates.

    In deep learning, another practical aspect that favors batch acquisition over individual point acquisition is that retraining a model can take a substantial amount of time.

    One way to extend the formalism of acquisition functions to the batch acquisition case is by scoring batches of samples instead of individual samples: $\theAF \left ( \xset \right )$. In batch active learning, we then want to maximize over possible subsets (of size \batchvar):
    \begin{align}
        \xacqsetstar = \argmax_{ \xacqset \subseteq \Dpool} {\theAF \left ( \xacqset, \Model \right )}
        .
    \end{align}
    
    \paragraph{Top-\batchvar BALD.}
    BALD was originally intended for acquiring individual data points and immediately retraining the model.
    Applications of BALD \citep{gal2015dropout, janz2017actively} usually acquire the top $\batchvar$ samples.
    This can be expressed as summing over individual scores:
    \begin{align}
        \BALDAF \left ( \xacqset, \pof{\w \given \Dtrain} \right ) 
        &= \sum_{i=1}^{\batchvar} \baldof{\xacq_i} \\
        &= \sum_{i=1}^{\batchvar} \MIof{\W ; \Yacq_i \given \xacq_i, \Dtrain},
    \end{align}
    and finding the optimal batch for this acquisition function using a greedy algorithm, which reduces to picking the top-$\batchvar$ highest-scoring data points.

    \subsection{Active Sampling}
    \label{glue:background:active_sampling}

    Another key problem in deep learning is training efficiency. While large models can exhibit great performance, training them can be time-consuming and expensive. 
    Active sampling is concerned with improving model performance as quickly as possible by selecting the samples to train on from within a larger training set of labeled samples. Unlike in active learning, labels are available, and the goal of active sampling is to select the most informative samples at each step to train on. 

    Active sampling selects labeled samples to train on \emph{during} training.
    Like in active learning, candidate batches of labeled samples can be scored via an acquisition function.
    This results in an active sampling loop which is similar to the active learning loop.
    However, whereas in active learning the model is usually reset and retrained between iterations \citep{ash2019warmstarting}, the model weights are, of course, not reset after each iteration.
    The active sampling loop is depicted in \Cref{glue:related_work:active_sampling_loop}. 

    Given the conceptual similarities between active learning and active sampling, we will consider how to unify objectives in \Cref{chapter:fim}.
    Recent work has shown that active learning methods can even outperform many active sampling methods without label information in certain circumstances \citep{park2022active}.

    \paragraph{Origins.} 
    It is not clear where the name active sampling originates from. The author is aware of it being used in meetings with collaborators from Google. A possible reference is \citet{abernethy2020active}, which selects samples to train on based on fairness metrics.

    \paragraph{Data Subset Selection Methods.}
    Active sampling also encompasses data pruning \citep{paul2021deep,siddiqui2022metadata} and core-set selection methods \citep{campbell2017automated,bayesian_campbell_2018,mirzasoleiman2019coresets,borsos2020coresets,data_borsos_2021,probabilistic_zhou_2023} as well as other data subset selection approaches \citep{gradient_aljundi_2019,glister_killamsetty_2020,gradmatch_killamsetty_2021,prism_kaushal_2021}: those could be seen as `single-step' active sampling methods.
    \citet{deepcore_guo_2022} contain a good overview of current data subset selection methods.

    \paragraph{Information Gain.}
    A natural information-theoretic choice is the \emph{information gain} $\MIof{\W; \y \given \x}$ \citep{sun2022informationtheoretic}, which unlike the expected information gain, takes into account the available label. We discuss this notation in \Cref{chapter:practical_it_notation} in more detail and examine the information gain in \Cref{chapter:fim}. 
}

}
\section{Thesis Outline}
\label{glue:sec:outline}
In this thesis, we examine objectives and extensions that follow from information-theoretic intuitions. Similar objectives can be adapted with success for both active learning and active sampling. 
The thesis is structured as follows:
\begin{description}
    \item[\Cref{chapter:practical_it_notation}] \textbf{\pittitle.}
    We define information quantities using a notation that allows us to take into account outcomes as well. 
    Usually, information quantities are only explicitly considered between unobserved random variables. The exception to that is the point-wise mutual information, which is well-known in NLP. Our notation unifies the mutual information and the point-wise mutual information (and the information gain and information surprise).
    This chapter is entirely based on \citet{kirsch2021practical}:
    \begin{mypapercitation}
        \bibentry{kirsch2021practical}.
    \end{mypapercitation}

    \item[\Cref{chapter:ddu}] \textbf{\ddutitle.}
    Epistemic uncertainty is important as an informativeness score in active learning. 
    We examine aleatoric and epistemic uncertainty in more detail, and based on several simple but crucial high-level observations, we propose a new baseline for uncertainty quantification using single forward-pass deep neural networks, e.g.\ deterministic neural networks instead of deep ensembles or similar.
    We draw a connection between feature-space density and informativeness (epistemic uncertainty) and show that with proper inductive biases, our simple approach can quantify epistemic uncertainty well with competitive results in active learning (and out-of-distribution detection) without having to be Bayesian.
    This chapter redrafts \citet{mukhoti2021deterministic}:
    \begin{mypapercitation}
        \bibentry{mukhoti2021deterministic},
    \end{mypapercitation}
    with additional figures, explanations, and results (\Cref{ddu:sec:active_learning_vs_ood_detection,glue:fig:ood_landscape,ddu:pro:ensemble_softmax_rmse,ddu:sec:predictive_entropy_not_epistemic_uncertainty_experiments}).

    \item[\Cref{chapter:batchbald}] \textbf{\baldtitle.}
    Active learning is often extended to batch active learning by greedily selecting the top-\batchvar samples that individually have the highest informativeness score. 
    We note that this is not a principled approach and can lead to worse active learning performance as it does not take dependencies and redundancies between the samples into account.
    We derive BatchBALD, a principled extension of the BALD acquisition score for batch acquisition, which avoids this issue. 
    The employed techniques and insights are not limited to BALD and are used throughout the thesis. 
    This chapter extends \citet{kirsch2019batchbald}:
    \begin{mypapercitation}
        \bibentry{kirsch2019batchbald},
    \end{mypapercitation}
    with additional figures (\Cref{batchbald:mnist_bald_scores_intuition2,batchbald:consistent_mc_dropout_is_necessary}).

    \item[\Cref{chapter:stochacq}] \textbf{\stochacqtitle.}
    BatchBALD has difficulties in scaling to higher batch acquisition sizes, both due to the computational cost and the scores becoming too uniform at higher batch acquisition sizes.
    Instead, we take a different approach, and we examine how progressive acquisitions in active learning change the EIG scores and their effect on top-\batchvar batch acquisitions. 
    Based on this, we examine a simple stochastic baseline that avoids the pathologies of top-\batchvar acquisition by adding Gumbel noise to the individual scores.
    This very simple baseline is surprisingly strong and also much cheaper to compute than the principled extension of the EIG to batch acquisition in \Cref{chapter:batchbald}.
    This chapter is entirely based on \citet{kirsch2021stochastic}:
    \begin{mypapercitation}
        \bibentry{kirsch2021stochastic}.
    \end{mypapercitation}

    \item[\Cref{chapter:obi}] \textbf{\obititle.}
    Focusing on joint predictives rather than marginal predictives can highlight the potential of Bayesian deep learning in real-world applications.
    In this chapter, we discuss online Bayesian inference, which allows making predictions while taking into account additional data without retraining, and propose new challenging evaluation settings using active learning and active sampling.
    By examining marginal and joint predictives, their respective cross-entropies, and their role in offline and online learning, we highlight previously unidentified gaps in current research and emphasize the need for better approximate joint predictives.
    This chapter builds on insights from \citet{wen2021predictions} and \citet{osb2022evaluating}, and we suggest further experiments to explore the feasibility of current BDL inference techniques in high-dimensional parameter spaces.
    This chapter is entirely based on \citet{kirsch2022marginal}:
    \begin{mypapercitation}
        \bibentry{kirsch2022marginal}.
    \end{mypapercitation}

    \item[\Cref{chapter:pobal}] \textbf{\pobaltitle.}
    The EIG does not take the target distribution of inputs into account at all. In other words (and if the target distribution matches the pool set distribution), it does not make use of the unlabeled data to guide the sample selection. 
    To account for this, we derive the \emph{expected predictive information gain (EPIG)}, an information quantity, which takes the data distribution into account. We show that it equivalently measures the expected reduction in generalization loss on the target distribution. We also examine and compare to the \emph{joint expected predictive information gain (JEPIG)}, a related quantity, which we connect to Bayesian model selection.
    This chapter extends \citet{kirsch2021evalbald} and \citet{epig}:
    \begin{mypapercitation}
        \bibentry{kirsch2021evalbald},
    \end{mypapercitation}
    \begin{mypapercitation}
        \bibentry{epig},
    \end{mypapercitation}
    with additional and reworked sections, examples, and figures (\Cref{pobal:example:gp,pobal:sec:two_alternatives,pobal:glue:mackayterms,EPIG:sec:limit,EPIG:sec:jepig_bayesian_model_selection,EPIG:prop:relationjpigepig}).

    \item[\Cref{chapter:rholoss}] \textbf{\rholosstitle.}
    For active sampling (label-aware active learning), we examine an adaptation of the previous two acquisition functions that take the labels into account: the \emph{(joint) predictive information gain---(J)PIG}.
    Training on web-scale data can take months. But most computation and time is wasted on {redundant} and noisy points that are already learned or not learnable. To accelerate training, based on JPIG, we perform a series of approximations and introduce the \textit{Reducible Holdout Loss Selection} (RHO-LOSS) in a non-Bayesian setting using two non-Bayesian (deterministic) models for active sampling. RHO-LOSS is a simple but principled technique which selects approximately those points for training that most reduce the model's generalization loss and trains in far fewer steps than prior art, improves accuracy, and speeds up training on a wide range of datasets, hyperparameters, and architectures (MLPs, CNNs, and BERT).
    On a large web-scraped image dataset (\mbox{Clothing-1M}), RHO-LOSS trains in 18x fewer steps and reaches 2\% higher final accuracy than uniform data shuffling. 
    This chapter is based on \citet{mindermann2021prioritized}:
    \begin{mypapercitation}
        \bibentry{mindermann2021prioritized},
    \end{mypapercitation}
    with a reworked theory section (\Cref{{rholoss:sec:theory}}).

    \item[\Cref{chapter:fim}] \textbf{\fimtitle.}
    Many approaches in data subset selection use Fisher information, Hessians, similarity matrices based on gradients, or the gradient length to estimate how informative data is for a model's training. 
    Are these different approaches connected, and if so, how?
    We revisit the fundamentals of Bayesian optimal experiment design and show that these recently proposed methods can be understood as approximations to the information-theoretic quantities examined in this thesis:
    EIG/BALD and (J)EPIG for active learning, and the information gain (IG) and (J)PIG for active sampling.
    We develop a comprehensive set of approximations using Fisher information and the observed information and derive a unified framework that connects above seemingly disparate literature.
    Although Bayesian methods are often seen as separate from non-Bayesian ones, the sometimes fuzzy notion of ``informativeness'' expressed in various non-Bayesian objectives leads to the same couple of information quantities, which were, in principle, already known by \citet{lindley1956measure} and \citet{mackay1992information}.
    This chapter is entirely based on \citet{kirsch2022unifying}:
    \begin{mypapercitation}
        \bibentry{kirsch2022unifying}.
    \end{mypapercitation}

    \item[\Cref{chapter:bbbal}] \textbf{\bbbaltitle.}
    \Cref{chapter:fim} show that different active learning methods approximate the same information-theoretic quantities using two different perspectives:
    \begin{itemize}
        \item Bayesian methods (including deep ensembles) often use (sampled) predictions over the parameter distribution to approximate information quantities, while
        \item non-Bayesian methods often use the weight space (using the score, i.e.\ log loss gradients) to approximate the same information quantities.
    \end{itemize}
    The difference between these two perspectives can be viewed as being between \emph{white-box approaches}, which are limited to differentiable parametric models (weight space) and \emph{black-box approaches}, which only use model predictions (prediction space).
    White-box methods score unlabeled points using acquisition functions based on model embeddings or first- and second-order derivatives.
    We utilize recent kernel-based approaches and turn a wide range of existing state-of-the-art white-box batch active learning methods (BADGE, BAIT, LCMD) into black-box approaches.
    We demonstrate the effectiveness of our approach through extensive experimental evaluations on regression datasets, achieving surprisingly strong performance compared to white-box approaches for deep learning models.
    This chapter is entirely based on \citet{kirsch2023blackbox}:
    \begin{mypapercitation}
        \bibentry{kirsch2023blackbox}.
    \end{mypapercitation}

    \item[Appendices] In the appendix, we provide additional details on the information-theoretic quantities and the approximations used in this thesis. We also include additional works that are related to the topics in this thesis and reproducibility analyses of several existing works.
    \begin{description}
        \item[\Cref{chapter:causalbald}] \textbf{\causalbaldtitle.}
        As a different application of combining information-theoretic intuitions and active learning, we develop acquisition functions for estimating the conditional average treatment effects from observational data for causal active learning. Unlike the previous applications, it does not apply to classification tasks but regression tasks. 
        This chapter is entirely based on \citet{jesson2021causal}:
        \begin{mypapercitation}
            \bibentry{jesson2021causal}.
        \end{mypapercitation}
        \item[\Cref{chapter:reproducibility}] \textbf{Reproducibility Analysis.}
        We also include reproducibility analyses of several papers that provide additional relevant insights (which are connected to themes in this thesis):
        \begin{description}
            \item[\Cref{chapter:dldd}] \textbf{\dlddtitle.}
            We reproduce parts of \citet{paul2021deep}. This chapter is entirely based on \citet{kirsch2023does}:
            \begin{mypapercitation}
                \bibentry{kirsch2023does}.
            \end{mypapercitation}
            \item[\Cref{chapter:jiang}] \textbf{\jiangtitle}
            We examine details of \citet{jiang2021assessing}. This chapter is entirely based on \citet{kirsch2022note}:
            \begin{mypapercitation}
                \bibentry{kirsch2022note}.
            \end{mypapercitation}
            \item[\Cref{ddu:app:theory}] \textbf{\ddudmtitle}
            We examine how well Dirichlet distributions can model the predictions of deep ensembles and their members to approximate information quantities we care about and to sample from the posterior predictive distribution. This chapter is entirely based on \citet{mukhoti2021deterministic}:
            \begin{mypapercitation}
                \bibentry{mukhoti2021deterministic},
            \end{mypapercitation}
            and extends it with additional results.
        \end{description}
    \end{description}
\end{description}

	{
		\let\chapter\section
		\let\section\subsection
		\let\subsection\subsubsection
		\let\subsubsection\paragraph
		\let\paragraph\textbf
		\chapter{Contributions to Joint Work}

The following declaration of contributions summarizes my contributions to the papers that are used as a basis for my thesis. Several of the papers are joint-authored with me as joint first-author. Prof.\ Yarin Gal supervised all my research projects except for \citet{kirsch2023does} and \citet{kirsch2023blackbox}.

\subsection*{A Practical \&\ Unified Notation for Information Quantities with Observed Outcomes \citep{kirsch2021practical}}
\begin{mypapercitation}
    \bibentry{kirsch2021practical}
\end{mypapercitation}

\paragraph{Contributions.}
I developed the paper and idea.

\subsection*{Deep Deterministic Uncertainty: A Simple Baseline \citep{mukhoti2021deterministic}}
\begin{mypapercitation}
    \bibentry{mukhoti2021deterministic}
\end{mypapercitation}

\paragraph{Contributions.}
Jishnu and I co-authored the paper. Jishnu and I wrote the paper together. I contributed significantly to the theory, in particular the observations, and suggested using density scores for OoD detection (using GDA) instead of scores based on the predictive distribution (entropy). Yarin and I developed the idea of the Dirty-MNIST dataset. Jishnu implemented and ran all experiments and trained all the models while Joost and I provided code reviews and feedback. 

\subsection*{BatchBALD: Efficient and Diverse Batch Acquisition for Deep Bayesian Active Learning \citep{kirsch2019batchbald}}
\begin{mypapercitation}
    \bibentry{kirsch2019batchbald}
\end{mypapercitation}

\paragraph{Contributions.}
Joost and I are joint first-authors. I developed the idea and algorithm and implemented and ran the experiments. Joost developed the Repeated-MNIST experimental setting and came up with the acquisition size-time plot. Joost and I co-wrote the paper. I wrote the proofs and created the plots. 

\subsection*{Stochastic Batch Acquisition: A Simple Baseline for Deep Active Learning \citep{kirsch2021stochastic}}
\begin{mypapercitation}
    \bibentry{kirsch2021stochastic}
\end{mypapercitation}

\paragraph{Contributions.}
Sebastian and I are joint first-authors. I implemented and ran the initial experiments and wrote the workshop submission while Sebastian was on an internship. The initial idea of stochastic acquisition was independently and then jointly developed. Sebastian designed the rank correlation experiment, and we jointly rewrote the paper with him leading and editing the paper. I created the plots and ran experiments. Frederic, Parmida, and Andrew ran experiments.

\subsection*{Marginal and Joint Cross-Entropies \&\ Predictives for Online Bayesian Inference, Active Learning, and Active Sampling \citep{kirsch2022marginal}}
\begin{mypapercitation}
    \bibentry{kirsch2022marginal}
\end{mypapercitation}

\paragraph{Contributions.}
I was the lead author on the workshop submission and the preprint.
Jannik supported the project's experiment design and provided feedback and helped edit the paper.

\subsection*{Test Distribution--Aware Active Learning: A Principled Approach Against Distribution Shift and Outliers \citep{kirsch2021evalbald}}
\begin{mypapercitation}
    \bibentry{kirsch2021evalbald}
\end{mypapercitation}

\paragraph{Contributions.}
I was the lead author on the workshop submission, which focused on JEPIG. Tom Rainforth helped redraft the paper for the second arxiv submission which focused on JEPIG, and we jointly developed the EPIG term.

\subsubsection*{Prediction-{{Oriented Bayesian Active Learning}} \citep{epig}}
\begin{mypapercitation}
    \bibentry{epig}
\end{mypapercitation}

\paragraph{Contributions.}
Freddie and I are joint first-authors.
I helped Freddie with the experiment design and provided feedback in discussion and encouragement throughout the project. The paper was significantly and majorly redrafted by Freddie, Tom, and Adam based on the earlier arXiv preprint.

\subsection*{Prioritized Training on Points that are Learnable, Worth Learning, and Not Yet Learnt \citep{mindermann2021prioritized}}
\begin{mypapercitation}
    \bibentry{mindermann2021prioritized}
\end{mypapercitation}

\paragraph{Contributions.}
While not being a joint first author, I contributed to the theory and helped draft the information theory subsection of the workshop submission. I wrote the comparison to EPIG in the appendix of the workshop submission. I independently developed the idea of a label-aware version of JEPIG before we realized that it was equivalent to what is now the unapproximated RHO-LOSS and joined the project. 

\subsection*{Unifying Approaches in Active Learning and Active Sampling via Fisher Information and Information-Theoretic Quantities \citep{kirsch2022unifying}}
\begin{mypapercitation}
    \bibentry{kirsch2022unifying}
\end{mypapercitation}

\paragraph{Contributions.}
I developed the paper and idea.

\subsection*{Black-Box Batch Active Learning for Regression \citep{kirsch2023blackbox}}
\begin{mypapercitation}
    \bibentry{kirsch2023blackbox}
\end{mypapercitation}

\paragraph{Contributions.}
I am the single author of the paper.

\subsection*{Does ``Deep Learning on a Data Diet'' reproduce? Overall yes, but GraNd at Initialization does not \citep{kirsch2023does}}
\begin{mypapercitation}
    \bibentry{kirsch2023does}
\end{mypapercitation}

\paragraph{Contributions.}
I am the single author of the paper and received helpful feedback from the first author and senior author of the original paper, \citet{paul2021deep}. 

\subsection*{A Note on ``Assessing Generalization of SGD via Disagreement'' \citep{kirsch2022note}}
\begin{mypapercitation}
    \bibentry{kirsch2022note}
\end{mypapercitation}

\paragraph{Contributions.}
I developed the paper and idea.

\subsection*{Causal-BALD: Deep Bayesian Active Learning of Outcomes to Infer Treatment-Effects from Observational Data \citep{jesson2021causal}}
\begin{mypapercitation}
    \bibentry{jesson2021causal}
\end{mypapercitation}

\paragraph{Contributions.}
While not being a joint first author, I contributed to the theory and proofs of $\rho$-BALD in particular.

	}
}

{
	
	\begin{savequote}[75mm]
		The limits of my language mean the limits of my world.
		\qauthor{Ludwig Wittgenstein}
	\end{savequote}
	\chapter{\pittitle}
	\label{chapter:practical_it_notation}
In \Cref{glue:background:information_theory}, we have introduced well-known information-theoretic quantities using a more consistent notation.
Now, we further canonically extend the definitions to tie random variables to specific observed outcomes, e.g.\ $X = x$. We will use this extensively in the following chapters. We refer to $X$ when we have $X=x$ in an expression as \emph{tied random variable} as it is \emph{tied} to an outcome. If we mix \emph{(untied) random variables} and \emph{tied random variables}, we define $\Hof{}$ as an operator which takes an expectation of Shannon's information content for the given expression over the (untied) random variables conditioned on the tied outcomes. For example, $\Hof{X, Y=y \given Z, W=w} = \simpleE{\pof{X, Z \given y, w}} \ICof{\pof{x, y \given z, w}}$ following this notation. We generally shorten \(Y=y\) to \(y\) when the connection is clear from context---except for the datasets $\Dpool, \Dtrain,$ etc., which are sets of outcomes (either only containing inputs $\x$ when unlabeled or containing both inputs $\x$ and targets $\y$).
Similarly, we have $\CrossEntropy{\pof{X \given y}}{\qof{X \given y}} = \simpleE{\pof{x \given y}} \ICof{\qof{x \given y}}$.
\begin{mainresult}
As a memory hook for the reader, lower-case letters are always used for tied random variables and upper-case letters for (untied) random variables over which we take an expectation. This makes it easy to differentiate between the two cases and write down the actual expressions.
\end{mainresult}

Importantly, the definitions above maintain the identities $\Hof{X, Y} = \simpleE{\pof{x}} \Hof{x, Y} = \simpleE{\pof{y}} \Hof{X, y}$, which is the motivation behind these extensions. \Cref{practical_it_notation:fig:mixed_entropy_diagram} provides an overview over the quantities for two random variables \(X\) and \(Y\) when \(Y=y\) is observed. We define everything in detail below and provide intuitions.

\begin{definition}
  Given random variables $X$ and $Y$ and outcome $y$, we define:
  \begin{align}
    \Hof{y} &\defeq \ICof{\pof{y}} \\
    \Hof{X, y} &\defeq \simpleE{\pof{x \given y}} \Hof{x, y} = \simpleE{\pof{x \given y}} \ICof{\pof{x, y}} \\
    \Hof{X \given y} &\defeq \simpleE{\pof{x \given y}} \Hof{x \given y} = \simpleE{\pof{x \given y}} \ICof{\pof{x \given y}} \\
    \Hof{y \given X} &\defeq \simpleE{\pof{x \given y}} \Hof{y \given x} = \simpleE{\pof{x \given y}} \ICof{\pof{y \given x}},
  \end{align}
  where we have shortened $Y=y$ to $y$.
\end{definition}
Again, \(\Hof{y}, \Hof{X, y}\) are shorthands for \(\xHof{\pof{y}}\), \(\xHof{\pof{X, y}}\), and so on. 

The intuition from information theory behind these definitions is that, e.g., $\Hof{X, y}$ measures the average length of transmitting $X$ and $Y$ together when $Y=y$ unbeknownst to the sender and receiver, and
\(\Hof{y \given X}\) measures how much additional information needs to be transferred on average for the receiver to learn \(y\) when it already knows $X \given y$.

\begin{figure}
  \begin{center}
  \setlength{\unitlength}{1in}
  \begin{picture}(3,1.70)(0,-0.8)
  \put(0,0.7){\framebox(3,0.20){\(\Hof{X, y}\)}}
  \put(1.5,0.4){\framebox(1.5,0.20){\(\Hof{y}\)}}
  \put(0,0.1){\framebox(2.0,0.20){\(\Hof{X}\)}}
  \put(1.5125,-0.2){\framebox(0.475,0.20){\(\MIof{X ; y}\)}}
  \put(1.5125,-0.5){\framebox(0.675,0.20){\(\MIof{y ; X}\)}}
  \put(0,-0.35){\framebox(1.475,0.20){\(\Hof{X \given y}\)}}
  \put(2.225,-0.35){\framebox(0.775,0.20){\(\Hof{y\given X}\)}}
  \put(0,-0.8){\framebox(2.2,0.20){\(\simpleE{\pof{x \given y}}\Hof{x}\)}}
  \end{picture}
  
  \end{center}
  {%
  \caption{\emph{The relationship between joint entropy \(\Hof{X, y}\), 
  entropies \(\Hof{X}, \Hof{y}\), conditional entropies \(\Hof{X \given y}, \Hof{y \given X}\), information gain \(\MIof{X ; y}\) and surprise \(\MIof{y ; X}\) when \(Y=y\) is observed.}
  We include \(\simpleE{\pof{x \given y}}\Hof{x}\) to visualize \Cref{practical_it_notation:proposition:observed_conditional_entropies}.
  The figure follows Figure 8.1 in \citet{mackay2003information}.}
  \label{practical_it_notation:fig:mixed_entropy_diagram}
  }%
\end{figure}

From above definition, we also have \(\Hof{x, y} = \ICof{\pof{x, y}}\) and \(\Hof{x \given y} = \ICof{\pof{x \given y}}\).
Beware, however, that while we have \(\Hof{X \given y} = \Hof{X, y} - \Hof{y}\), for \(\Hof{y \given X}\), there is no such equality for \(\Hof{y \given X}\):
\begin{proposition}
  \label{practical_it_notation:proposition:observed_conditional_entropies}
  Given random variables $X$ and $Y$ and outcome $y$, we generally have:
  \begin{align}
    \Hof{X \given y} &= \Hof{X, y} - \Hof{y} \\
    \Hof{y \given X} &= \Hof{X, y} - \simpleE{\pof{x \given y}} \Hof{x} \notag \\
    & \mathrel{\textcolor{red}{\not=}} \Hof{X, y} - \Hof{X},
  \end{align}
\end{proposition} 
\begin{proof}
  \(\Hof{X \given y} = \Hof{X, y} - \Hof{y}\) follows immediately from the definitions.
  \(\Hof{y \given X} \not=\Hof{X, y} - \Hof{X}\) follows because, generally, \(\simpleE{\pof{x \given y}} \Hof{x} \not= \Hof{X}\) when \(\pof{x \given y} \not= \pof{x}\). 
  E.g., for \(X\) and \(Y\) only taking binary values, $0$ or $1$, let\footnote{See also \shorturl{https://colab.research.google.com/drive/1HvLXUMQYcxMGZ4S_a00xddGmfz0IHaR3}.} \(\pof{x, y}=\tfrac{1}{3} \mathbb{1}
_{\{x=0=y\}}\), then \(\simpleE{\pof{x \given y}} \Hof{x} = \log{\left(\frac{3}{2} \right)} \not= \log{\left(\frac{3 \sqrt[3]{2}}{2} \right)} = \Hof{X}\).
\end{proof}

There are two common, sensible quantities we can define when we want to consider the information overlap between a random variable and an outcome: the \emph{information gain}, also known as \emph{specific information} and the \emph{surprise} \citep{deweese1999measure, butts2003much}. These two quantities are usually defined separately in the cognitive sciences and neuroscience \citep{williams2011information}; however, we can unify them after relaxing the symmetry of the mutual information as done above: %
\begin{definition}
  Given random variables $X$ and $Y$ and outcome $y$ for $Y$, we define the \emph{information gain} $\MIof{X; y}$ and the \emph{surprise} $\MIof{y; X}$ as:
  \begin{align}
    & \MIof{X ; y} \defeq \Hof{X} - \Hof{X \given y} \\
    & \MIof{y ; X} \defeq \Hof{y} - \Hof{y \given X}.
  \end{align}
\end{definition}
This unifying definition is novel to the best of our knowledge. It works by breaking the symmetry that otherwise exists for the regular and point-wise mutual information.

Note that the surprise can also be expressed as \(\MIof{y ; X} = \Kale{\pof{X \given y}}{\pof{X}}\). For example, this is done in \citet{bellemare2016unifying}---even though the paper mistakenly calls this surprise an information gain when it is not (in our sense). 

We enumerate a few equivalent ways of writing the mutual information and surprise---the information gain has no such equivalences. This can be helpful to spot these quantities in the wild. 
\begin{proposition}
  We have
  \begin{align}
    \MIof{X; Y} &= \Kale{\pof{X, Y}}{\pof{X} \pof{Y}} \\
    \MIof{y ; X} &= \simpleE{\pof{x \given y}} \MIof{y ; x} \\
                 &= \E{\pof{x \given y}}{\Hof{x}} - \Hof{X \given y} \\
                 &= \Kale{\pof{X \given y}}{\pof{X}}.
  \end{align}
\end{proposition}

The information gain \(\MIof{X ; y}\) for $X$ given $y$ measures the reduction in uncertainty about $\Hof{X}$ when we observe $y$. \(\Hof{X}\) is the uncertainty about the true \(X\) that we want to learn: the entropy quantifies the amount of additional information that we need to transmit to fix \(X\), and similarly \(\Hof{X \given y}\) quantities the additional information we need to transmit to fix \(X\) once $y$ is known both to the sender and the receiver \citep{lindley1956measure}.
On the other hand, the surprise \(\MIof{y ; X}\) of $y$ for $X$ measures 
how much the posterior $X \given y$ lies in areas where $\pof{x}$ was small before observing $y$ \citep{deweese1999measure}. 

An important difference between the two is that the information gain can be \emph{chained} while the surprise cannot: 
\begin{proposition}
  Given random variables $X$, $Y_1$, and \(Y_2\) and outcomes $y_1$ and $y_2$ for $Y_1$ and \(Y_2\), respectively, we have:
  \begin{align}
    \MIof{X ; y_1, y_2} & = \MIof{X ; y_1} + \MIof{X ; y_2 \given y_1} \\
    \MIof{y_1, y_2 ; X} & \mathrel{\textcolor{red}{\not=}} \MIof{y_1 ; X} + \MIof{y_2 ; X \given y_1}.
  \end{align}   
\end{proposition}
\begin{proof}
We have
  \begin{align*}
    &\MIof{X ; y_1, y_2} = \Hof{X} - \Hof{X \given y_1, y_2} \\
    & \quad = \Hof{X} - \Hof{X \given y_1} + \Hof{X \given y_1} - \Hof{X \given y_1, y_2} \\ 
    & \quad = \MIof{X ; y_1} + \MIof{X ; y_2 \given y_1},
  \end{align*}
while
  \begin{align*}
    \MIof{y_1, y_2 ; X} &= \simpleE{\pof{x \given y_1, y_2}} \MIof{y_1, y_2; x} \\
    & = \underbrace{\simpleE{\pof{x \given y_1, y_2}} \MIof{y_1; x}}_{\quad \displaystyle\mathrel{\color{red}\not=}\simpleE{\pof{x \given y_1}} \MIof{y_1; x} = \MIof{y_1 ; X}} \\
    & \quad + \underbrace{\simpleE{\pof{x \given y_1, y_2}} \MIof{y_2 ; x \given y_1}}_{\displaystyle=\MIof{y_2; X \given y_1}}.
  \end{align*}
That is, generally, \(\simpleE{\pof{x \given y_1, y_2}} \MIof{y_1; x} \not= \MIof{y_1 ; X}\).
To conclude the proof, we instantiate \(\pof{x \given y_1, y_2} \not=\pof{x \given y_1}\): 
for \(X\), \(Y_1\), and \(Y_2\) taking binary values \(0, 1\) only, let \(\pof{y_1}=\tfrac{1}{2}, \, \pof{x, y_2 \given y_1=0}=\tfrac{1}{4}, \,\pof{x \given y_2=0, y_1=1} = \tfrac{1}{2}, \,\pof{x=0 \given y_2=1, y_1=1} = 1\). 
Then \(\simpleE{\pof{x \given y_1, y_2}} \MIof{y_1; x} = \log{\left(\frac{2 \sqrt{3} \sqrt[4]{5}}{5} \right)} \not= \log{\left(\frac{6}{5} \right)} = \MIof{y_1 ; X}\) for \(y_1=1, y_2=1\) as the reader can easily verify\footnote{See also \shorturl{https://colab.research.google.com/drive/1gn6oQohRMqXKEhyCogiVDcx1VZFkShaQ}.}.
\end{proof}
However, both quantities do chain in their (untied) random variables:
\begin{proposition}
  Given random variables $X_1$, $X_2$, $Y$, and outcome $y$ for $Y$:
  \begin{align}
    & \MIof{X_1, X_2 ; y} = \MIof{X_1 ; y} + \MIof{X_2 ; y \given X_1} \\
    & \MIof{y; X_1, X_2 } = \MIof{y ; X_1} + \MIof{y ; X_2 \given X_1}.
  \end{align}
\end{proposition}
\begin{proof}
  We have
  \begin{align*}
    & \MIof{X_1 ; y} + \MIof{X_2 ; y \given X_1} = \\
    & \quad = \Hof{X_1} - \Hof{X_1 \given y} + \Hof{X_2 \given X_1} + \Hof{X_2 \given X_1, y} \\
    & \quad = \underbrace{\Hof{X_1} + \Hof{X_2 \given X_1}}_{=\Hof{X_1, X_2}} -
       (\underbrace{\Hof{X_1 \given y} + \Hof{X_2 \given X_1, y}}_{=\Hof{X_1, X_2 \given y}}) \\
    & \quad = \MIof{X_1, X_2 ; y}.
  \end{align*}
  Similarly, we have
  \begin{align*}
    & \MIof{y ; X_1} + \MIof{y ; X_2 \given X_1} = \\
    & \quad = \Hof{y} - \Hof{y \given X_1} + \Hof{y \given X_1} - \Hof{y \given X_1, X_2} \\
    & \quad = \Hof{y} - \Hof{y \given X_1, X_2} \\
    & \quad = \MIof{y ; X_1, X_2}.
  \end{align*}
\end{proof}

These extensions of the mutual information are canonical as they permute with taking expectations over tied variables to obtain the regular (untied) quantities:
\begin{proposition}
  For random variables $X$ and $Y$:
  \begin{align}
    \MIof{X ; Y} &= \simpleE{\pof{y}} \MIof{X ; y} = \simpleE{\pof{y}} \MIof{y ; X} = \simpleE{\pof{x, y}} \MIof{x, y}.
  \end{align}
\end{proposition}
\begin{proof}
  Follows immediately from substituting the definitions.
\end{proof}
Likewise, when all random variables are tied to a specific outcome, the quantities behave as expected:
\begin{proposition}
  For random variables $X$, \(Y\), \(Y_1\) and \(Y_2\):
  \begin{align}
    \MIof{X; Y} &= \MIof{Y; X}, \text{and} \\
    \MIof{x ; y} &= \MIof{y ; x}; \\
    \MIof{X; Y_1, Y_2} &= \MIof{X; Y_1} + \MIof{X; Y_1 \given Y_2}, \text{and} \\
    \MIof{x; y_1, y_2} &= \MIof{x; y_1} + \MIof{x; y_2 \given y_1}.
  \end{align}
\end{proposition}
\begin{proof}
  The only interesting equality is \(\MIof{x; y_1, y_2} = \MIof{x; y_1} + \MIof{x; y_2 \given y_1}\):
  \begin{align*}
    \MIof{x; y_1} + \MIof{x; y_2 \given y_1} &= \ICof{\frac{\pof{x} \, \pof{y_1}}{\pof{x, y_1}} \, \frac{\pof{x, y_1} \, \pof{y_1, y_2} \, \pof{y_1}}{\pof{y_1} \, \pof{y_1} \pof{x, y_1, y_2}}} \\
    &= \ICof{\frac{\pof{x} \, \pof{y_1, y_2}}{\pof{x, y_1, y_2}}} \\
    &= \MIof{x; y_1, y_2}.
  \end{align*}
\end{proof}

We can extend this to triple mutual information terms by adopting the extension $\MIof{X; Y; Z} = \MIof{X ; Y} - \MIof{X ; Y \given Z }$ \citep{yeung2008information} for outcomes as well: $\MIof{X ; Y; z} = \MIof{X ; Y} - \MIof{X ; Y \given z}$, which also works for higher-order terms.\andreas{appendix again...}

Overall, for the reader, there will be little surprise when working with the fully point-wise information-theoretic quantities, that is, when all random variables are observed. But the mixed ones require more care. We refer the reader back to \Cref{practical_it_notation:fig:mixed_entropy_diagram} to recall the relationships which also provide intuitions for the inequalities we will examine next.

\textbf{Inequalities.} We review some well-known inequalities first:
\begin{proposition}
  For random variables \(X\) and \(Y\), we have:
  \begin{align}
    \MIof{X ; Y} &\ge 0 \\
    \Hof{X} &\ge \Hof{X \given Y}, \\
  \intertext{and if \(X\) is a discrete random variable, we also have:}
    \Hof{X} &\ge 0 \\
    \MIof{X ; Y} &\le \Hof{X}.
  \end{align}
\end{proposition}
\begin{proof}
  The first two statements follow from:
  \begin{align}
    \Hof{X} - \Hof{X \given Y} &= \MIof{X ; Y} \notag \\
    &= \Kale{\pof{X, Y}}{\pof{X}\pof{Y}} \notag \\
    &\ge 0.
  \end{align}
  The third statement follows from the monotony of the expectation and \(\pof{x} \le 1\) for all \(x\).
\end{proof}
For mixed outcomes we find similar inequalities:\andreas{mention independence}
\begin{proposition}
  For random variables \(X\) and \(Y\) with outcome \(y\), we have:
  \begin{align}
    \MIof{y ; X} &\ge 0 \\
    \Hof{y} &\ge \Hof{y \given X} \\    
    \simpleE{\pof{x \given y}{\Hof{x}}} &\ge \Hof{X \given y}, \\
    \intertext{and if \(Y\) is a discrete random variable, we also have:}
      \Hof{y \given X}, \Hof{y} &\ge 0 \\
      \MIof{y ; X} &\le \Hof{y},
    \intertext{and if \(X\) is also a discrete random variable, we gain:}
      \MIof{y ; X} &\le \simpleE{\pof{x \given y}}{\Hof{x}}.
  \end{align}
\end{proposition}
\begin{proof}
  Again, the first two statements follow from:
  \begin{align}
    \Hof{y} - \Hof{y \given X} &= \MIof{y ; X} \notag \\
    &= \simpleE{\pof{x \given y}}{\MIof{y ; x}} \notag \\
    &= \E{\pof{x \given y}}{\Hof{x} - \Hof{x \given y}} \label{practical_it_notation:eq:proof_inequality_Hy}\\
    &= \Kale{\pof{X \given y}}{\pof{X}} \notag \\
    &\ge 0.
  \end{align}
  The third statement follows from \cref{practical_it_notation:eq:proof_inequality_Hy} above as \(0 \le \E{\pof{x \given y}}{\Hof{x} - \Hof{x \given y}} = \simpleE{\pof{x \given y}}{\Hof{x} - \Hof{X \given y}}\).
  The fourth statement follows from \(\pof{y \given x} \le 1\) when \(Y\) is a discrete random variable, and thus \(\Hof{y \given X} \ge 0\) due to the monotony of the expectation.
  The fifth statement follows from the fourth statement and \(\MIof{y ; X} = \Hof{y} - \Hof{y \given X} \le \Hof{y}\).
  Finally, if \(X\) is a discrete random variable as well, we also have \(\Hof{X \given y} \ge 0\), and thus
  \begin{align*}
    \MIof{y ; X} = \E{\pof{x \given y}}{\Hof{x} - \Hof{X \given y}} \le \simpleE{\pof{x \given y}}{\Hof{x}}.
  \end{align*}
\end{proof}
Note that there are no such bounds for \(\MIof{X ; y}\), \(\Hof{X \given y}\) and \(\Hof{y \given X}\).

\begin{corollary}
  We have $\MIof{y ; X} = 0$ exactly when $\pof{x \given y} = \pof{x}$ for all $x$ for given $y$. 
\end{corollary}
\begin{proof}
  This follows from $0 = \MIof{y ; X} = \Kale{\pof{X \given y}}{\pof{X}}$ exactly when $\pof{x \given y} = \pof{x}$.
\end{proof}

In particular, there is a misleading intuition that the information gain \(\MIof{X ; y} = \Hof{X} - \Hof{X \given y}\) ought to be non-negative for any $y$. This is not true. This intuition may exist because in many cases when we look at posterior distributions, we only model the mean and assume a fixed variance. The uncertainty around the mean does indeed reduce with additional observations; however, the uncertainty around the variance might not. The reader is invited to experiment with a normal distribution with known mean and compute the information gain on the variance depending on new observations. 

In a sense, the information-theoretic surprise is much better behaved than the information gain because we can bound it in various ways, which does not seem possible for the information gain. The (expected) information gain is a more useful quantity though for active sampling, active learning and Bayesian optimal experimental design. Thus, unified notation that includes and differentiates between both quantities is beneficial.

\section{Example Application: Stirling's Approximation for Binomial Coefficients}
\label{practical_it_notation:sec:stirling_binomial_coefficient}
\begin{figure}
  \begin{center}
  \setlength{\unitlength}{1in}
  \begin{picture}(3,0.9)(0,0.0)
  \put(0,0.7){\framebox(3,0.20){\(\Hof{B, r}\)}}
  \put(1.5,0.4){\framebox(1.5,0.20){\(\Hof{r}\)}}
  \put(0,0.4){\framebox(1.475,0.20){\(\Hof{B \given r}\)}}
  \put(1.5,0.3){\vector(1,0){1.475}}
  \put(0,0.0){\framebox(3,0.20){\(\simpleE{\pof{b \given r}}{\Hof{b}}\)}}
  \end{picture}
  
  \end{center}
  {%
  \caption{\emph{The relationship between the information quantities used in \S\ref{practical_it_notation:sec:stirling_binomial_coefficient}.} \(B\) is the joint of the binomial random variables, \(R\) is the number of successes in \(B\) with observed outcome \(r\). The arrow below \(\Hof{r}\) symbolizes that we minimize \(\Hof{r}\) by optimizing the success probability \(\rho\) to close the gap between \(\simpleE{\pof{b \given r}}{\Hof{b}}\) and \(\Hof{B \given r}\).
  }
  \label{practical_it_notation:fig:stirling_approx_for_binomial_coeffs}
  }%
\end{figure}
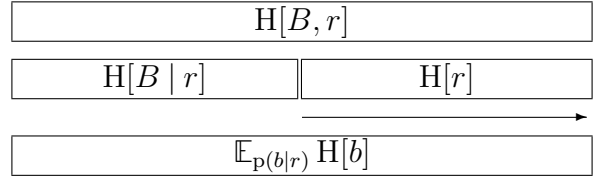

In \citet{mackay2003information} on page 2, the following simple approximation for a binomial coefficient is introduced:
\begin{equation}
    \log \binom{N}{r} \simeq(N-r) \log \frac{N}{N-r}+r \log \frac{N}{r}. \label{practical_it_notation:eq:stirling_approx}
\end{equation}
We will derive this result using the proposed extension to observed outcomes as it allows for an intuitive deduction. Moreover, we will see that this allows us to use other tools from probability theory to estimate the approximation error.

\textbf{Setup.} %
Let $B_1, \ldots, B_N$ be $N$ Bernoulli random variables with success probability $p$, and let $B$ be the joint of these random variables.

Further, let $R$ be the random variable that counts the number of successes in $B$. $R$ follows a Binomial distribution with success probability $\rho$ and $N$ trials.

\textbf{Main Idea.} %
For a given outcome $r$ of $R$, we have:
\begin{equation}
    \Hof{B, r} = \Hof{B \given r} + \Hof{r} \ge \Hof{B \given r},
\end{equation}
as $\Hof{\cdot}$ is non-negative for discrete random variables.
We will examine this inequality to obtain the approximation in \cref{practical_it_notation:eq:stirling_approx}.

Note that $\Hof{B \given r}$ is the additional number of bits needed to encode $B$ when the number of successes is already known. Similarly, $\Hof{B, r}$ is the number of bits needed to encode both $B$ and $R$ under the circumstance that $R=r$. 

\textbf{Determining $\Hof{B, r}$.} %
$R$ is fully determined by $B$, and thus we have $\Hof{B, R}=\Hof{B}$ and hence\footnote{This also follows immediately from $\Hof{R \given B} = 0 \implies \forall r: \Hof{r \given B} = 0$.}:
\begin{equation}
    \Hof{B, r} = \simpleE{\pof{b \given r}}{\Hof{b}}.
\end{equation}
$\simpleE{\pof{b \given r}}{\Hof{b}}$ is the expected number of bits needed to transmit the outcome $b$ of $B$ when $r$ is given.
When we encode $B$, we do not know $r$ upfront, so we need to transmit $N$ Bernoulli outcomes.
Hence, we need to transmit $r$ successes and $N-r$ failures. Given the success probability $\rho$, the optimal message length for this is:
\begin{align}
    &\simpleE{\pof{b \given r}}{\Hof{b}} = r \, \ICof{\rho} + (N-r)\, \ICof{1-\rho} \\
    & \quad = - r \log \rho - (N - r) \log (1-\rho).
\end{align}
All this is visualized in \Cref{practical_it_notation:fig:stirling_approx_for_binomial_coeffs}.

\textbf{Alternative Argument.}
We can also look at the terms $\Hof{B \given r} + \Hof{r}$ separately. We have
\begin{equation}
    \Hof{r} = -\log \pof{r} = - \log \left ( \binom{N}{r} \, \rho^r \,(1-\rho)^{N-r}\right ),
\end{equation}
and
\begin{equation}
    \Hof{B \given r} = -\simpleE{\pof{b \given r}} \log \pof{b \given r} = 
    \log \binom{N}{r}.
\end{equation}
The former follows from $R$ being binomial distributed.
For the latter, we observe that we need to encode $B$ while knowing $r$ already. Given $r$, $\pof{b \given r} = \text{const}$ for all valid $b$. There are $\binom{N}{r}$ possible $b$ for fixed $r$.
Hence, we can simply create a table with all possible configurations with $r$ successes. There are $\binom{N}{r}$ many. We then encode the index into this table. 

Each configuration with $r$ successes has an equal probability of happening, so we have a uniform discrete distribution with entropy $\log \binom{N}{r}$ and obtain the same result.

\textbf{Determining $\rho$.}
We already have
\begin{align}
\Hof{B \given r} + \Hof{r} &=  -r \log \rho - (N - r) \log (1-\rho) \notag \\
&\ge
\log \binom{N}{r} = \Hof{B \given r}. \label{practical_it_notation:eq:main_result}
\end{align}
How do we make this inequality as tight as possible?

We need to minimize the gap $\Hof{r}$ which creates the inequality in the first place, and $\Hof{r}=-\log \pof{r}$ is minimized exactly when $\pof{r}$ becomes maximal.

Hence, we choose the success probability $\rho$ to do so: the maximum likelihood solution $\argmax_p \pof{r \given \rho}$ is $\rho = \frac{r}{N}$.
The Binomial distribution of $R$ then has its mode, mean, and median at $r$.

Altogether, after substituting $\rho = \frac{r}{N}$ and rearranging, we see that the wanted approximation is actually an inequality:
\begin{align}
    \log \binom{N}{r} &\le -r \log \rho - (N - r) \log (1-\rho) \\
& = r \log \frac{N}{r} + (N - r) \log \frac{N}{N-r}. 
\end{align}

\textbf{Approximation Error $\Hof{r}$.} %
The approximation error is just $\Hof{r}$ as we can read off from \cref{practical_it_notation:eq:main_result}. We can easily upper-bound it with $\Hof{r} \le \log N$: 
First, $\Hof{R} \le \log N$ as the uniform distribution with entropy $\log N$ is the maximum entropy distribution in this case (discrete random variable with finite support).
Second, $\Hof{R}$ is the expectation over different $\Hof{R=r'}$. We have chosen $\rho=\tfrac{r}{N}$ such that $r$ is the mean of binomial distribution and has maximal probability mass. This means it has minimal information content. Hence, $\Hof{r} \le \log N$ by contraposition as otherwise $\log N < \Hof{r} \le \Hof{R}$.

}

{
	\begin{savequote}[75mm]
		Simplicity is the ultimate sophistication.
		\qauthor{Leonardo da Vinci}
	\end{savequote}
	
	\chapter{\ddutitle}
	\label{chapter:ddu}

\label{ddu:sec:intro}

In this chapter, we delve deeper into aleatoric and epistemic uncertainty and apply the insights gained to single forward-pass neural networks to disentangle these uncertainties.

Uncertainty quantification has garnered interest for such approaches because most well-known methods of uncertainty quantification in deep learning \citep{blundell2015weight, gal2015dropout, lakshminarayanan2016simple, wen2020batchensemble, dusenberry2020efficient} require multiple forward passes at test time.
Among these methods, deep ensembles have generally exhibited superior performance in uncertainty prediction \citep{ovadia2019can}.
However, their substantial memory and computational requirements during training and test time impede their adoption in real-life (e.g.\ mobile applications).
As a result, there has been a growing interest in uncertainty quantification using deterministic single forward-pass neural networks, which offer a smaller footprint and reduced latency.
We empirically validate our findings using active learning and out-of-distribution (OoD) detection on computer vision datasets.

To clarify, since OoD detection is not a well-defined term, we will investigate OoD detection for `related distributions', such as CIFAR-10 versus SVHN or CIFAR-100, following the definition in \citet{farquharWhat2022}.
Another term for this is `near OoD' \citep{winkens2020contrastive}.
Curiously, active learning and OoD detection are rarely evaluated together; thus, we will explore some of their nuances (recall \Cref{glue:background:uncertainty_quantification}) and contrast them.

\subsection*{Active Learning $\not=$ OoD Detection}
\label{ddu:sec:active_learning_vs_ood_detection}

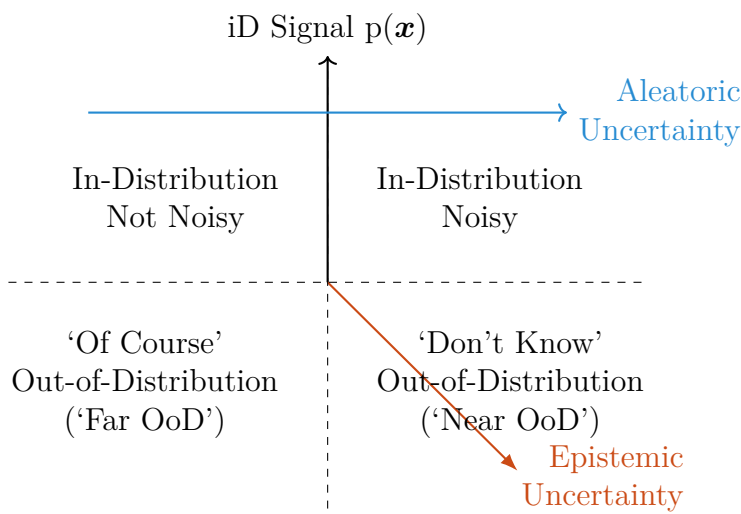
\begin{figure}[t]
    \centering
    \begin{tikzpicture}
        \fill[solarized@orange, opacity=0, path fading=fade bottom-right to top-left] (4.2, 0) -- (0, -3) -- (0, 0) -- cycle;

        \fill[solarized@blue, opacity=0, path fading=fade right to left] (0, 0) -- (4.2, 3) -- (-4.2, 3) -- cycle;

        \draw[thick,->] (0, 0) -- (0, 3) node[anchor=south,align=center] {iD Signal $\pof{\x}$};
        \draw[thick,->,color=solarized@blue] (-4.2*0.75, 3*0.75) -- (4.2*0.75, 3*0.75) node[anchor=west,align=right] {Aleatoric\\Uncertainty};

        \draw[-Latex, thick, color=solarized@orange] (0, 0) -- (2.5, -2.5) node[pos=1, anchor=west, xshift=-1mm, yshift=-1mm, align=right] {Epistemic\\Uncertainty};

        \node[anchor=south east, align=center] (IDN) at (-0.5, 0.5) {In-Distribution\\Not Noisy};
        \node[anchor=south west, align=center] (IDY) at (0.5, 0.5) {In-Distribution\\Noisy};
        \node[anchor=north east, align=center] (OOD) at (-0.5, -0.5) {`Of Course'\\Out-of-Distribution\\(`Far OoD')};
        \node[anchor=north west, align=center,] (UNK) at (0.5, -0.5) {`Don't Know'\\Out-of-Distribution\\(`Near OoD')};

        \draw[dashed] (0, -3) -- (0, 3);
        \draw[dashed] (-4.2, 0) -- (4.2, 0);
    \end{tikzpicture}
    \caption{
        \emph{The four quadrants of the OoD detection landscape.}
        Active learning focuses on the bottom-right quadrant, as data points in this area will be informative for the model. OoD detection is concerned with the lower half, as it aims to identify out-of-distribution data points (the training set is considered in-distribution). The `Don't Know' quadrant is referred to as near OoD, and the `Of Course' quadrant as far OoD.
        Consequently, feature-space density provides a suitable signal for active learning when the available pool data does not contain actual OoD data or outliers, as these will be confounded with informative in-distribution points. 
        Equivalently, epistemic uncertainty as an active learning signal will only provide a reliable OoD signal for near OoD but not for far OoD.
    }
    \label{glue:fig:ood_landscape}
\end{figure}

We can conceptually differentiate active learning and OoD detection as follows:
\begin{itemize}
    \item Active learning seeks to identify the most informative samples for labeling, while
    \item OoD detection aims to recognize samples that are not in-distribution, regardless of their informativeness.
\end{itemize}
This subtle yet crucial distinction implies that the informativeness signal in active learning will generally be useful for OoD detection (assuming the training data defines the in-distribution), \emph{but not the other way around}:
a sample that is `obviously' OoD will not be informative for active learning.
See  \Cref{glue:fig:ood_landscape} for a visualization of this concept.
In OoD literature, the terms `near OoD' and `far OoD' \citep{winkens2020contrastive} are often used to differentiate between points that are close to the in-distribution, leading to \emph{conflicting} predictions, and those that are far away, making their detection \emph{unanimous}.

Near-OoD points will likely be highly informative (high epistemic uncertainty) as the model might disagree about their interpretation, while far-OoD points will have low informativeness (low epistemic uncertainty) as the model will confidently detect them as OoD. 
This becomes an issue when using only epistemic uncertainty for OoD detection\footnote{It is unclear whether combining epistemic uncertainty with other signals has been explored in depth.}, as reported by \citet{xia2022usefulness}.
Methods that employ outlier exposure are particularly prone to this problem, as epistemic uncertainty will also be low for outliers used during training.
However, this issue can arise in general as the available training data increases: as the model parameters concentrate, epistemic uncertainty will decrease, and the model will become more confident in its predictions, causing more OoD data to become `far OoD.'

As evident from this discussion, intriguing conceptual questions remain, yet active learning and OoD detection are rarely examined together. This chapter serves as an exception.

\subsection*{Relevant Literature}
There are several single forward-pass uncertainty approaches in the literature that of particular relevance for this chapter. We will focus on methods that use feature-space distances and density \citep{Settles2010,lee2018simple, van2020uncertainty, liu2020simple,postels2020quantifying}. 

\paragraph{Mahalanobis Distance.}
Among these approaches, \citet{lee2018simple} uses Mahalanobis distances to quantify uncertainty by fitting a class-wise Gaussian distribution (with shared covariance matrices) on the feature space of a pre-trained ResNet encoder.
They do not consider the structure of the underlying feature-space however, which might explain why their competitive results require input perturbations, the ensembling of OoD metrics over multiple layers, and fine-tuning on OoD hold-out data.

\paragraph{DUQ \&\ SNGP.} Two recent works in single forward-pass uncertainty, DUQ \citep{van2020uncertainty} and SNGP \citep{liu2020simple}, propose distance-aware output layers, in the form of RBFs (radial basis functions) or GPs (Gaussian processes), and introduce additional inductive biases in the feature extractor using a Jacobian penalty \citep{gulrajani2017improved} or spectral normalization \citep{miyato2018spectral}, respectively, which encourage smoothness and sensitivity in the latent space.
These methods perform well and are almost competitive with deep ensembles on OoD benchmarks.
However, they require training to be changed substantially, and introduce additional hyperparameters due to the specialized output layers used at training.
Furthermore, DUQ and SNGP cannot disentangle aleatoric and epistemic uncertainty.
Particularly, in DUQ, the feature representation of an ambiguous data point, high on aleatoric uncertainty, will be in between two centroids, but due to the exponential decay of the RBF it will seem far from both and thus have uncertainty similar to epistemically uncertain data points that are far from all centroids.
In SNGP, the predictive variance is computed using a mean-field approximation of the softmax likelihood, which cannot be disentangled. The variance can also be computed using MC samples of the softmax likelihood which, in theory, can allow disentangling uncertainties (see \cref{glue:eq:BALD}), but requires modelling the covariance between the classes, which is not the case in SNGP. We provide a more extensive review of related work in \S\ref{ddu:sec:related work}.

\subsection*{Outline}

Concretely, we will focus on the following research questions:
\begin{enumerate}
    \item Are complex methods to estimate uncertainty, like in DUQ and SNGP, necessary beyond feature-space regularization that encourages bi-Lipschitzness?
    \item What are the conceptual challenges of different uncertainty metrics, and what instructive insights can be learned from these?
    \item How can we disentangle aleatoric and epistemic uncertainty with single forward-pass neural networks, as DUQ and SNGP do not address this directly?
\end{enumerate}
We make some simple but crucial observations that help answer our research questions in this chapter:
\begin{enumerate}
    \item Entropy is arguably the wrong proxy for epistemic uncertainty, despite its frequent use for active learning and OoD detection.
Specifically, we find that:
    \begin{enumerate}
        \item The predictive entropy as a metric confounds aleatoric and epistemic uncertainty (\Cref{ddu:fig:intro_softmax_ent}).
This can be an issue in active learning in particular.
Yet, this issue is often not visible for standard benchmark datasets without aleatoric noise.
To examine this failure in more detail, we introduce a new dataset, Dirty-MNIST, which showcases the issue more clearly than artificially curated datasets like MNIST or CIFAR-10.
\emph{Dirty-MNIST} is an expanded version of MNIST \citep{lecun1998gradient} with additional ambiguous digits (Ambiguous-MNIST) having multiple plausible labels and thus higher aleatoric uncertainty (\Cref{ddu:fig:intro_sample_viz}).
        \item The softmax entropy of a deterministic model trained with maximum likelihood, while being high for ambiguous points (i.e., with high aleatoric uncertainty), might not be consistent for points with high epistemic uncertainty, i.e., the softmax entropy for an OoD sample might be low, high, or anything in between for different models trained on the same data (\Cref{ddu:fig:intro_softmax_ent}).
    \end{enumerate}
    \item To disentangle aleatoric and epistemic uncertainty, feature-space density can be used to estimate epistemic uncertainty, and entropy for aleatoric uncertainty.
However, feature-space regularization \citep{liu2020simple} is crucial\footnote{\citep{pearce2021understanding} argue for softmax confidence and entropy in their paper, yet feature-space density performs better in their experiments, too.}.
Without such regularization, feature-space density alone might not separate iD from OoD data, possibly explaining the limited empirical success of previous approaches which attempt to use feature-space density \citep{postels2020quantifying}.
This can be seen in \Cref{ddu:fig:intro_gmm} where the feature-space density of a VGG-16 or LeNet model is not able to differentiate iD Dirty-MNIST from OoD FashionMNIST, while a ResNet-18 with spectral normalization can do so better.
    \item Objectives for density estimation and classification might have different optima (except on unambiguous, well-separable datasets), and using a single mixture model (e.g., a GMM) leads to suboptimal performance due to this \emph{objective mismatch} \citep{murphy2012machine}.
    Hence, one should separately estimate the feature-space density for epistemic uncertainty and predictive entropy for aleatoric uncertainty.
\end{enumerate}
Based on these observations, we examine an approach we call `\emph{Deep Deterministic Uncertainty (DDU)}', which uses Gaussian Discriminant Analysis (GDA) for feature-space density on a trained model \&\ the original softmax layer for estimating aleatoric uncertainty and making classification predictions.
Using DDU, we empirically investigate whether complex methods to estimate uncertainty, like in DUQ and SNGP, are necessary beyond feature-space regularization that encourages bi-Lipschitzness. When we use spectral normalization like SNGP does, the short answer is an empirical no.
    
As we only perform GDA after training, the original softmax layer is trained using cross-entropy as a proper scoring rule \citep{gneiting2007strictly} and can be temperature-scaled to provide good in-distribution calibration and aleatoric uncertainty.
DDU outperforms regular softmax neural networks, as illustrated in \Cref{ddu:fig:intro_histograms}. Furthermore, DDU is competitive with deep ensembles \citep{lakshminarayanan2016simple} and outperforms SNGP and DUQ \citep{van2020uncertainty, liu2020simple}, with no changes to the model architecture beyond spectral normalization, in several OoD benchmarks and active learning settings. Using DeepLab-v3+ \citep{chen2017rethinking} on Pascal VOC 2012 \citep{everingham2010pascal}, we also show that DDU improves upon two classic uncertainty methods---MC Dropout \citep{gal2015dropout} and deep ensembles---on the task of semantic segmentation, while being significantly faster to compute. 

\begin{figure}[t]
    \centering
    \begin{subfigure}{\linewidth}
        \includegraphics[width=\linewidth]{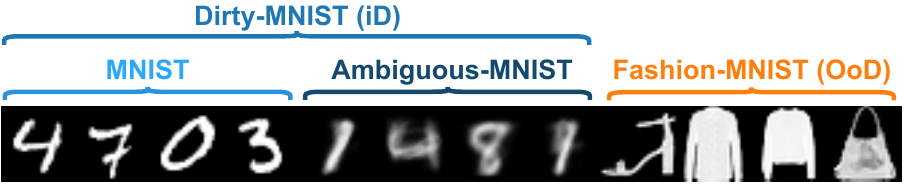}%
        \caption{{\color{sns-blue} Dirty-MNIST (iD, = {\color{sns-nonambiguous}MNIST} + {\color{sns-ambiguous}Ambiguous-MNIST})} and {\color{sns-orange} FashionMNIST (OoD)}}
        \label{ddu:fig:intro_sample_viz}
    \end{subfigure}
    \begin{subfigure}{\linewidth}
        \begin{subfigure}{0.33\linewidth}
            \centering
            \includegraphics[width=\linewidth]{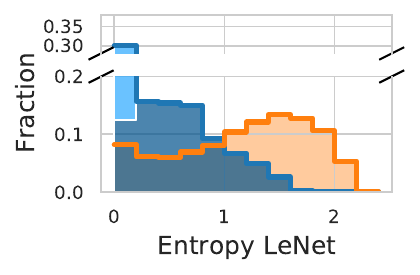}%
        \end{subfigure}\hfill
        \begin{subfigure}{0.33\linewidth}
            \centering
            \includegraphics[width=\linewidth]{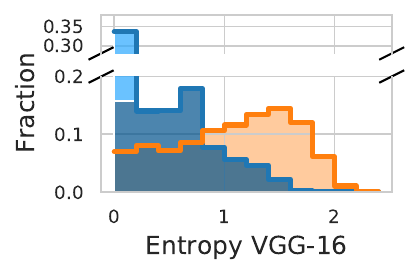}%
        \end{subfigure}\hfill
        \begin{subfigure}{0.33\linewidth}
            \centering
            \includegraphics[width=\linewidth]{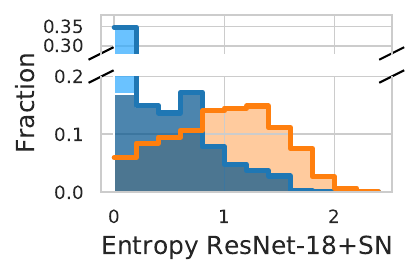}%
        \end{subfigure}\hfill 
        \subcaption{{Softmax entropy}}\label{ddu:fig:intro_softmax_ent}
    \end{subfigure}
    \begin{subfigure}{\linewidth}
        \begin{subfigure}{0.33\linewidth}
            \centering
            \includegraphics[width=\linewidth]{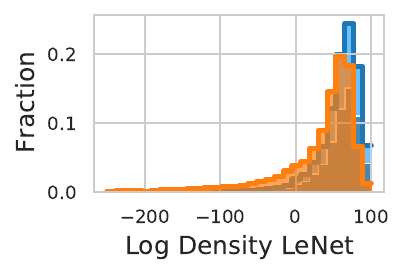}%
        \end{subfigure}\hfill
        \begin{subfigure}{0.33\linewidth}
            \centering
            \includegraphics[width=\linewidth]{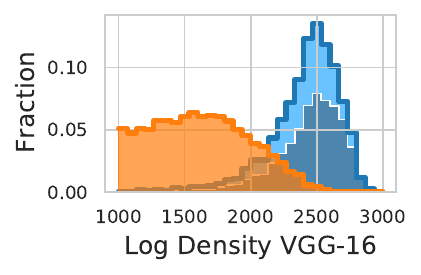}%
        \end{subfigure}\hfill
        \begin{subfigure}{0.33\linewidth}
            \centering
            \includegraphics[width=\linewidth]{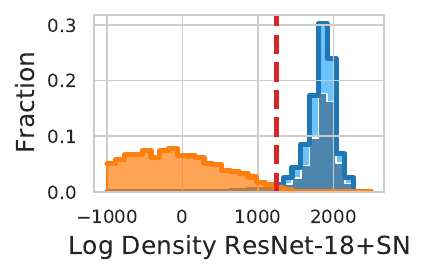}%
        \end{subfigure}%
        \subcaption{{Feature-space density}}\label{ddu:fig:intro_gmm}
    \end{subfigure}%
    \caption{
    \emph{
    Disentangling aleatoric and epistemic uncertainty on {\color{sns-blue}Dirty-MNIST (iD)} and {\color{sns-orange} FashionMNIST (OoD)} \textbf{\subref{ddu:fig:intro_sample_viz}} requires using \emph{softmax entropy} \textbf{\subref{ddu:fig:intro_softmax_ent}} and \emph{feature-space density (GMM)} \textbf{\subref{ddu:fig:intro_gmm}} with a well-regularized feature space (\emph{ResNet-18+SN} vs \emph{LeNet} \&\ \emph{VGG-16} without smoothness \&\ sensitivity).}
    \textbf{\subref{ddu:fig:intro_softmax_ent}:}
    Softmax entropy captures aleatoric uncertainty for iD data (Dirty-MNIST), thereby separating {\color{sns-nonambiguous}unambiguous MNIST samples} and {\color{sns-ambiguous}Ambiguous-MNIST samples} (stacked histogram). However, iD and OoD are confounded: softmax entropy has arbitrary values for OoD, indistinguishable from iD.
    \textbf{\subref{ddu:fig:intro_gmm}:}
    With a well-regularized feature space (DDU with ResNet-18+SN), iD and OoD densities do not overlap, capturing epistemic uncertainty. However, without such feature space (LeNet \&\ VGG-16), feature density suffers from \emph{feature collapse}: iD and OoD densities overlap. Generally, feature-space density confounds unambiguous and ambiguous iD samples as their densities overlap.
    }
    \label{ddu:fig:intro_histograms}
\end{figure}

\section{Entropy $\not=$ Epistemic Uncertainty}

In this section, we observe that:
\begin{itemize}
\item using entropy for OoD detection is inherently problematic as it cannot distinguish between aleatoric uncertainty of ambiguous iD samples and the epistemic uncertainty of near OoD samples; and
\item the softmax entropy of a single model is even more problematic as it is unreliable \emph{specifically} for samples with high epistemic uncertainty, i.e.,\ near OoD samples.
\end{itemize}
Both observations are tied to the very reason why a deep ensembles' mutual information captures epistemic uncertainty well and can be used to detect adversarial examples and near-OoD data, too \citep{smith2018understanding}. 
To exemplify the issues, we will introduce \emph{Dirty-MNIST} as a dataset with a long tail of ambiguous samples. 
We will conclude with an empirical analysis of the relationship between the softmax entropy and the predictive entropy (of a respective deep ensembles).

\begin{figure}[t]
    \centering
    \begin{subfigure}{0.33\linewidth}
        \centering
        \includegraphics[width=\linewidth]{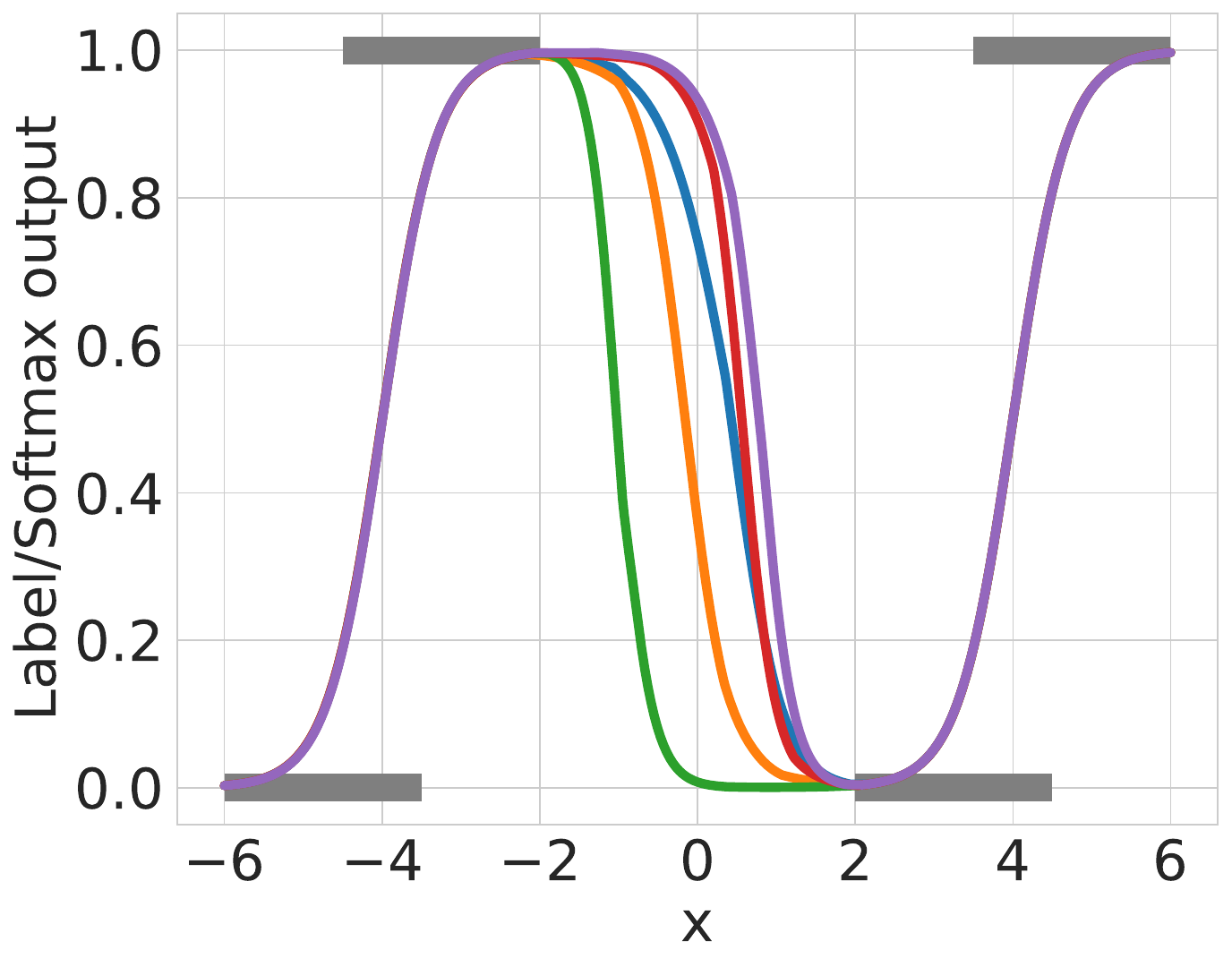}%
        \caption{Softmax Output}
        \label{ddu:subfig:softmax_output}
    \end{subfigure}\hfill
    \begin{subfigure}{0.33\linewidth}
        \centering
        \includegraphics[width=\linewidth]{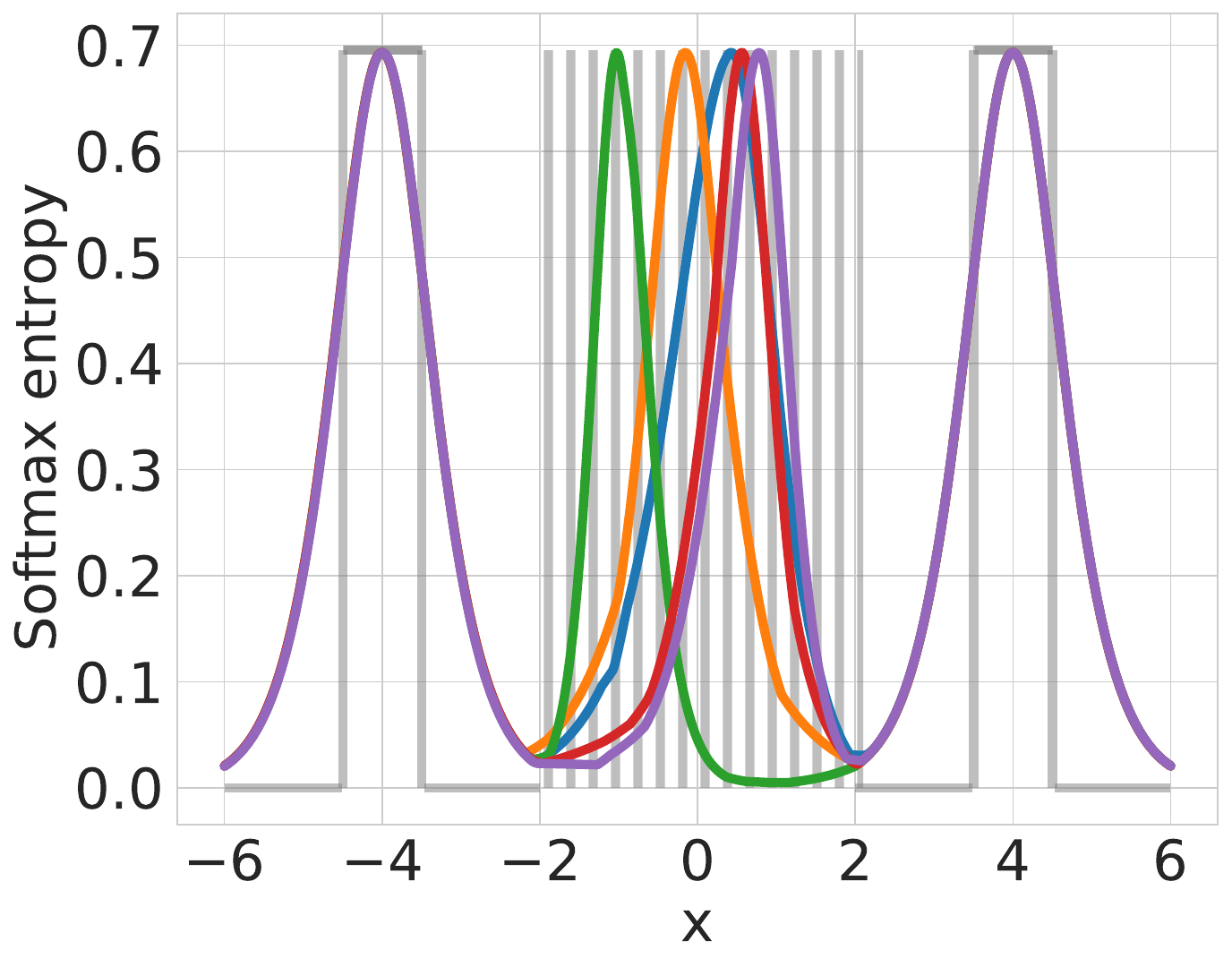}%
        \caption{Softmax Entropy}
        \label{ddu:subfig:softmax_entropy}
    \end{subfigure}\hfill
    \begin{subfigure}{0.33\linewidth}
        \centering
        \includegraphics[width=\linewidth]{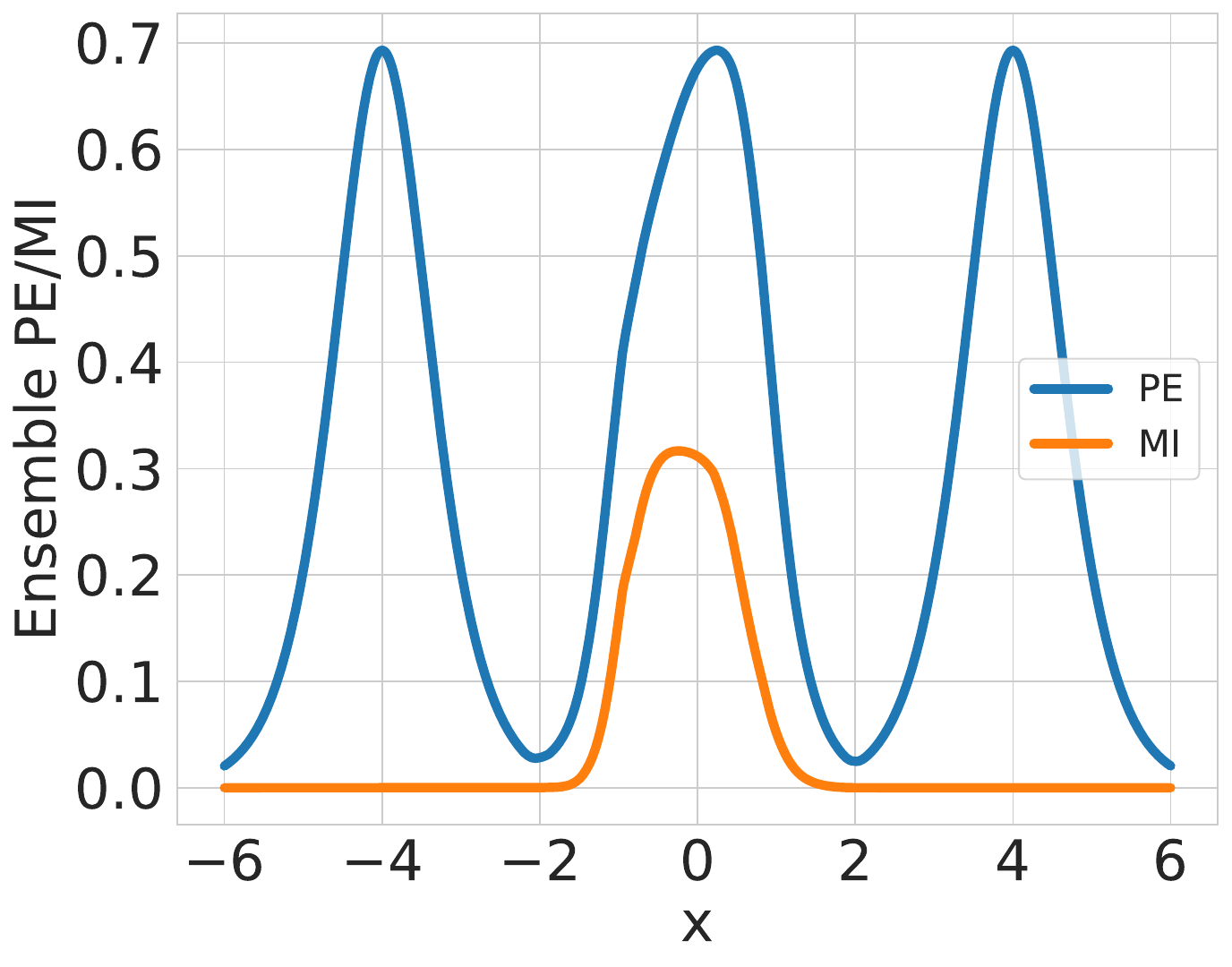}%
        \caption{5-Ensemble}
        \label{ddu:subfig:pe_mi}
    \end{subfigure}%
    \caption{
    \emph{Softmax outputs \&\ entropies for 5 softmax models along with the predictive entropy (PE) and mutual information (MI) for the resulting 5-Ensemble.}
    \subref{ddu:subfig:softmax_output} and \subref{ddu:subfig:softmax_entropy} show that the softmax entropy is only reliably high for ambiguous iD points ($\pm$3.5--4.5),
    whereas it can be low or high for OoD points (-2--2). The different colors are the different ensemble components. Similarly, \subref{ddu:subfig:pe_mi} shows that the MI of the ensemble is only high for OoD,
    whereas the PE is high for both OoD and for regions of ambiguity. See \S\ref{ddu:app:5_ensemble}.
    }
    \label{ddu:fig:lewis_vis}
\end{figure}

\subsubsection{5-Ensemble Visualization}
\label{ddu:app:5_ensemble}

We start with a visualization of a 5-ensemble (with five deterministic softmax networks) to see how softmax entropy fails to capture epistemic uncertainty precisely because the mutual information (MI) of an ensemble does. This is illustrated in \Cref{ddu:fig:lewis_vis} and provides intuition for the second point that softmax entropy is unreliable for samples with high epistemic uncertainty.
We train the networks on 1-dimensional data with binary labels 0 and 1.
The data is shown in \Cref{ddu:subfig:softmax_output}.
From \Cref{ddu:subfig:softmax_output} and \Cref{ddu:subfig:softmax_entropy}, we find that the softmax entropy is high in regions of ambiguity where the label can be both 0 and 1 (i.e.\ x between -4.5 and -3.5, and between 3.5 and 4.5). 
This indicates that softmax entropy can capture aleatoric uncertainty. Furthermore, in the x interval $(-2, 2)$, we find that the deterministic softmax networks disagree in their predictions (see \Cref{ddu:subfig:softmax_output}) and have softmax entropies which can be high, low or anywhere in between (see \Cref{ddu:subfig:softmax_entropy}) following our claim in \S\ref{ddu:sec:motivation}. 
In fact, this disagreement is the very reason why the MI of the ensemble is high in the interval $(-2, 2)$, thereby reliably capturing epistemic uncertainty. 
Finally, the predictive entropy (PE) of the ensemble is high both in the OoD interval $(-2, 2)$ as well as at points of ambiguity (i.e.\ at -4 and 4). This indicates that the PE of a deep ensemble captures both epistemic and aleatoric uncertainty well. From these visualizations, we draw the conclusion that the softmax entropy of a deterministic softmax model cannot capture epistemic uncertainty precisely because the MI of a deep ensemble can.

\subsection{Dirty-MNIST}

To show that entropy is inappropriate for OoD detection, we train a LeNet \citep{lecun1998gradient}, a VGG-16 \citep{simonyan2014very} and a ResNet-18 with spectral normalization, ResNet+SN\footnote{\citet{liu2020simple} show that spectral normalization regularizes the latent space in a way that is beneficial for OoD detection, so we also include a model trained on this recent approach.}\citep{he2016deep,miyato2018spectral} on \emph{Dirty-MNIST}, a modified version of MNIST \citep{lecun1998gradient} with additional ambiguous digits (Ambiguous-MNIST), depicted in \Cref{ddu:fig:intro_sample_viz}, %
which we introduce below.

Dirty-MNIST poses a challenge for using entropy for OoD detection as it confounds aleatoric and epistemic uncertainty:
\Cref{ddu:fig:intro_softmax_ent} shows that the softmax entropy of a deterministic model is unable to distinguish between iD (Dirty-MNIST) and FashionMNIST samples \citep{xiao2017fashion} as near OoD: the entropy for the latter heavily overlaps with the entropy for Ambiguous-MNIST samples.
With the ambiguous data having various levels of aleatoric uncertainty, Dirty-MNIST is more representative of real-world datasets compared to well-cleaned curated datasets, like MNIST and CIFAR-10, commonly used for benchmarking \citep{krizhevsky2009learning}. 

\subsubsection{Ambiguous-MNIST}
\label{ddu:app:dirty_mnist}
Each sample in Ambiguous-MNIST is constructed by decoding a linear combination of latent representations of 2 different MNIST digits from a pre-trained VAE \citep{kingma2013auto}. Every decoded image is assigned several labels sampled from the softmax probabilities of an off-the-shelf MNIST neural network ensemble, with points filtered based on an ensemble's MI to remove ``junk'' images and then stratified class-wise based on their softmax entropy. 
All off-the-shelf MNIST neural networks were then discarded, and new models were trained to generate \Cref{ddu:fig:intro_histograms}---and as can be seen, the ambiguous points we generate indeed have high entropy regardless of the model architecture used.
We create 60K such training and 10K test images to construct Ambiguous-MNIST. Finally, the Dirty-MNIST dataset in this experiment contains MNIST and Ambiguous-MNIST samples in a 1:1 ratio (thus, in total 120K training and 20K test samples). In \Cref{ddu:fig:ambiguous_mnist_samples}, we provide some samples from Ambiguous-MNIST. This provides intuition for the first point that entropy for OoD detection is inherently problematic as it cannot distinguish between aleatoric uncertainty of ambiguous iD samples and the epistemic uncertainty of near OoD samples.

\begin{figure}[t]
    \centering
    \begin{subfigure}{0.5\linewidth}
        \centering
        \includegraphics[width=\linewidth]{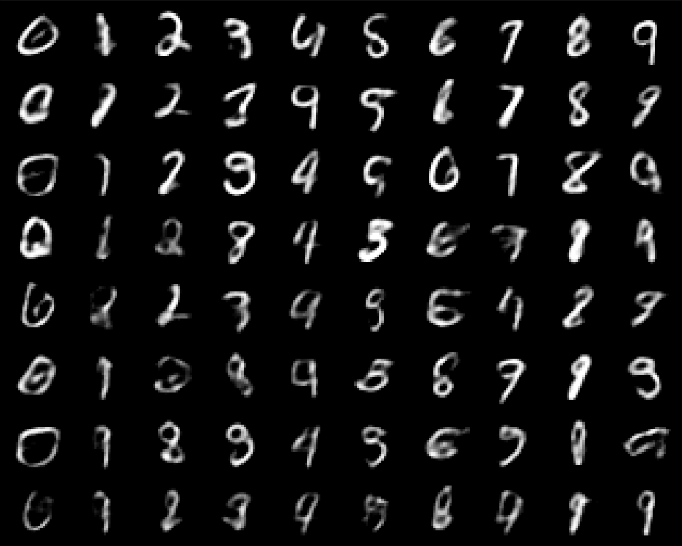}
    \end{subfigure}
    \caption{
    \emph{Samples from Ambiguous-MNIST.}
    }
    \label{ddu:fig:ambiguous_mnist_samples}
\end{figure}

\subsection{Potential Pitfalls of Predictive and Softmax Entropy}

Now let us discuss potential pitfalls of predictive entropy in general and softmax entropy of deterministic models in particular in more detail.

\subsubsection{Potential Pitfalls of \emph{Predictive Entropy}}
Conceptually, \emph{predictive entropy confounds epistemic and aleatoric uncertainty}. 
Since ensembling can also be interpreted as Bayesian Model Averaging \citep{He2020Bayesian, wilson2020bayesian}, with each ensemble member approximating a sample from a posterior, eq.~\eqref{glue:eq:BALD} can be applied to ensembles to disentangle epistemic and aleatoric uncertainty.
Both mutual information $\MIof{Y; \omega \given \x, \Dtrain}$ and predictive entropy $\Hof{Y \given \x, \Dtrain}$ could be used to detect OoD samples. 
However, previous empirical findings show predictive entropy outperforming mutual information \citep{malinin2018predictive}.
Indeed, much of the recent literature only focuses on predictive entropy for OoD detection.
\label{ddu:app:entropy_confidence_lit_review}
\Cref{ddu:table:confidence_entropy_lit_review} shows a selection of recently published papers which use entropy or confidence as OoD score. Only two papers examine using mutual information with deep ensembles as OoD score at all. None of the papers examines the possible confounding of aleatoric and epistemic uncertainty when using predictive entropy or confidence, or the consistency issues of softmax entropy (and softmax confidence), detailed in \S\ref{ddu:sec:motivation}. This list is not exhaustive, of course.
We explain these findings using the following (obvious) observation:
\begin{importantresult}
    \begin{observation}
    When we \emph{already know} that \emph{either} aleatoric \emph{or} epistemic uncertainty is \emph{low} for an iD sample, predictive entropy is an appropriate measure of the other uncertainty type.
    \end{observation}
\end{importantresult}
Thus, predictive entropy, as an upper-bound of mutual information, can separate iD and OoD data better when datasets are curated and have low aleatoric uncertainty.
However, as seen in eq.~\eqref{glue:eq:BALD}, predictive entropy can be high for both iD ambiguous samples (high aleatoric) as well as near OoD samples (high epistemic) (see \Cref{ddu:fig:lewis_vis}) and might \emph{not} be an effective measure for OoD detection when used with datasets that are not curated with ambiguous samples, like Dirty-MNIST, as seen in our active learning results. 

\begin{table}[!t]
    \centering
    \caption{\emph{A sample of recently published papers and OoD metrics.} Many recently published papers only use Predictive Entropy or Predictive Confidence  (for deep ensembles) or Softmax Confidence (for deterministic models) as OoD scores without addressing the possible confounding of aleatoric and epistemic uncertainty, that is ambiguous iD samples with OoD samples. Only two papers examine using mutual information with deep ensembles as OoD score at all.}
    \label{ddu:table:confidence_entropy_lit_review}
    \resizebox{\linewidth}{!}{%
    \renewcommand{\arraystretch}{1.3} 
    \begin{tabular}{m{30em}rrrr}
    \toprule
    \textbf{Title [Citation]} & \makecell[r]{\textbf{Softmax} \\ \textbf{Confidence}} & \makecell[r]{\textbf{Predictive} \\ \textbf{Confidence}} & \makecell[r]{\textbf{Predictive} \\ \textbf{Entropy}} & \makecell[r]{\textbf{Mutual} \\ \textbf{Information}} \\
    \midrule
    \makecell*[{{m{30em}}}]{A Baseline for Detecting Misclassified and Out-of-Distribution Examples in Neural Networks\\\citep{hendrycks2016baseline}} & \cmark & \xmark & \xmark & \xmark \\
    \makecell*[{{m{30em}}}]{Deep Anomaly Detection with Outlier Exposure\\\citep{hendrycks2018deep}} & \cmark & \xmark & \xmark & \xmark \\
    \makecell*[{{m{30em}}}]{Enhancing The Reliability of Out-of-distribution Image Detection in Neural Networks\\\citep{liang2017enhancing}} & \cmark & \xmark & \xmark & \xmark \\
    \makecell*[{{m{30em}}}]{Training Confidence-calibrated Classifiers for Detecting Out-of-Distribution Samples\\\citep{lee2017training}} & \cmark & \xmark & \xmark & \xmark \\
    \makecell*[{{m{30em}}}]{Learning Confidence for Out-of-Distribution Detection in Neural Networks\\\citep{devries2018learning}} & \cmark & \xmark & \xmark & \xmark \\
    \makecell*[{{m{30em}}}]{Simple and Scalable Predictive Uncertainty Estimation using Deep Ensembles\\\citep{lakshminarayanan2016simple}} & \xmark & \cmark & \cmark & \xmark \\
    \makecell*[{{m{30em}}}]{Predictive Uncertainty Estimation via Prior Networks\\\citep{malinin2018predictive}} & \xmark & \cmark & \cmark & \cmark \\
    \makecell*[{{m{30em}}}]{Ensemble Distribution Distillation\\\citep{malinin2019ensemble}} & \xmark & \xmark & \cmark & \cmark \\
    \makecell*[{{m{30em}}}]{Generalized ODIN: Detecting Out-of-Distribution Image Without Learning From Out-of-Distribution Data\\\citep{hsu2020generalized}} & \cmark & \xmark & \xmark & \xmark \\
    \makecell*[{{m{30em}}}]{Being Bayesian, Even Just a Bit, Fixes Overconfidence in ReLU Networks\\\citep{Kristiadi2020BeingBE}} & \xmark & \cmark & \xmark & \xmark \\
    \bottomrule
    \end{tabular}
    }
\end{table}

\subsubsection{Potential Pitfalls of \emph{Softmax Entropy}}
\emph{The softmax entropy for deterministic models trained with maximum likelihood can be inconsistent}.
As we have noted in \Cref{glue:background:uncertainty_quantification}, \cref{glue:eq:BALD} can be used with deep ensembles, as each ensemble member can be considered a sample from \emph{some} distribution $\pof{\omega \given \Dtrain}$ over model parameters $\omega \subset \Omega$ (e.g.\ a uniform distribution over $K$ trained ensemble members $\omega_1, ..., \omega_K$):
\begin{align}
    \underbrace{\Hof{\Y \given \x, \Dtrain}}_\text{predictive} = \underbrace{\MIof{\Y; \W \given\x, \Dtrain}}_\text{epistemic} + %
    \underbrace{\Hof{Y \given \x, \W, \Dtrain}}_\text{aleatoric (for iD $\x$)}.
    \tag{\ref{glue:eq:BALD}}
\end{align}
Note that the mutual information $\MIof{Y; \omega \given \x, \Dtrain}$ isolates epistemic from aleatoric uncertainty for deep ensembles as well, whereas the predictive entropy $\Hof{Y \given \x, \Dtrain}$ (often used with deep ensembles) measures predictive uncertainty, which will be high whenever either epistemic or aleatoric uncertainties are high. 

\paragraph{Rank Inconsistency.}
Crucially, the mechanism underlying deep ensemble uncertainty that can push epistemic uncertainty to be high on OoD data is the function disagreement between different ensemble members, i.e., arbitrary and disagreeing predictive extrapolations of the softmax models composing the ensemble due to a lack of relevant training data\footnote{Something similar has also recently been reported in the context of ensemble calibration \citep{jordan2023calibrated}.}:
Deep ensembles demonstrating high epistemic uncertainty (mutual information) on OoD data entails that at least two ensemble members must \emph{extrapolate differently} (`arbitrary extrapolations') on that data, since the predictive and aleatoric terms in \cref{glue:eq:BALD} must cancel out 
(leading the ``aleatoric'' term in eq.~\eqref{glue:eq:BALD} to vanish \citep{smith2018understanding}):
This is because for OoD data points \cref{glue:eq:BALD} guarantees that whenever the epistemic uncertainty is high, the predictive entropy must be high as well. 
The following simple qualitative result captures this intuition:
\begin{restatable}{proposition}{ensemblesoftmax}
\label{ddu:pro:ensemble_softmax}
Let $\x_1$ and $\x_2$ be points such that $\x_1$ has \textbf{higher} epistemic uncertainty than $\x_2$ under the ensemble:
\begin{align}
\MIof{Y_1; \omega \given x_1, \Dtrain} > \MIof{Y_2; \omega \given x_2, \Dtrain} + \delta,
\end{align}
$\delta \ge 0$.
Further, assume both have similar predictive entropy 
\begin{align}
    | \Hof{Y_1 \given x_1, \Dtrain} - \Hof{Y_2 \given x_2, \Dtrain} | \le \epsilon,
\end{align}
$\epsilon \ge 0$.
Then, there exist sets of ensemble members $\Omega'$ (assuming a countable ensemble),
with $\pof{\Omega' \given \Dtrain} > 0$, such that for all softmax models $\omega' \in \Omega'$ the softmax entropy of $\x_1$ is \textbf{lower} than the softmax entropy of $\x_2$: 
\begin{align}
    \Hof{Y_1 \given \x_1, \omega'} < \Hof{Y_2 \given \x_2, \omega'} - (\delta - \epsilon).
\end{align}
\end{restatable}

\begin{proof}
From \cref{glue:eq:BALD}, we obtain:
\begin{align}
    &| \Hof{Y_1 \given x_1, \Dtrain} - \Hof{Y_2 \given x_2, \Dtrain} | \le \epsilon \\
    \Leftrightarrow &
    |\MIof{Y_1; \omega \given x_1, \Dtrain} + \E{\pof{\omega \given \Dtrain}}{\Hof{Y_1 \given x_1, \omega}} \notag \\
    &\quad -\MIof{Y_2; \omega \given x_2, \Dtrain} - \E{\pof{\omega \given \Dtrain}}{\Hof{Y_2 \given x_2, \omega}}| \le \epsilon,
\end{align}
and hence we have:
\begin{align}
    & \E{\pof{\omega \given \Dtrain}}{\Hof{Y_1 \given x_1, \omega}} - \E{\pof{\omega \given \Dtrain}}{\Hof{Y_2 \given x_2, \omega}} \notag \\
    & + \underbrace{(\MIof{Y_1; \omega \given x_1, \Dtrain} - \MIof{Y_2; \omega \given x_2, \Dtrain})}_{> \delta} \le \epsilon.
\end{align}
We can rearrange the terms:
\begin{equation}
    \E{\pof{\omega \given \Dtrain}}{\Hof{Y_1 \given x_1, \omega}} <
    \E{\pof{\omega \given \Dtrain}}{\Hof{Y_2 \given x_2, \omega}} - (\delta - \epsilon).
\end{equation}
Now, the statement follows by contraposition: if ${\Hof{Y_1 \given \x_1, \omega}} \ge {\Hof{Y_2 \given \x_2, \omega}} - (\delta - \epsilon)$ for all $\omega$, the monotonicity of the expectation would yield $\E{\pof{\omega \given \Dtrain}}{\Hof{Y_1 \given \x_1, \omega}} \ge \E{\pof{\omega \given \Dtrain}}{\Hof{Y_2 \given \x_2, \omega}} - (\delta - \epsilon).$
Thus, there is a set $\Omega'$ with $\pof{\Omega' \given \Dtrain} > 0$, such that
\begin{equation}
    {\Hof{Y_1 \given x_1, \omega}} < {\Hof{Y_2 \given x_2, \omega}} - (\delta - \epsilon),
\end{equation}
for all $\omega \in \Omega'$.
\end{proof}
If a sample is assigned higher epistemic uncertainty (in the form of mutual information) by a deep ensemble than another sample, it will necessarily be assigned lower softmax entropy by at least one of the ensemble's members. 
As a result, a priori, we cannot know whether a softmax model preserves the order or not, and \emph{the empirical observation that the mutual information of an ensemble can quantify epistemic uncertainty well implies that the softmax entropy of a deterministic model might not}.
We see this in \Cref{ddu:fig:intro_softmax_ent}, \ref{ddu:fig:lewis_vis} and in \Cref{ddu:sec:predictive_entropy_not_epistemic_uncertainty_experiments} where softmax entropy for OoD samples can be high, low or anywhere in between. While \emph{not all} model architectures might behave like this, when the mutual information of a deep ensemble works well empirically, \Cref{ddu:pro:ensemble_softmax} holds.

This directly impacts the quality of the OoD detection, as we will verify in \Cref{ddu:sec:experiments,ddu:sec:predictive_entropy_not_epistemic_uncertainty_experiments}.
For OoD detection, the changes in the score ranks can induce additional false positive or false negative---the AUROC, for example, directly measures the probability that an iD point has higher score than an OoD point. 
For active learning, the changes in the order can lead to less uninformative samples being selected---on the other hand, the additional noise this introduces can also be beneficial as we investigate in \Cref{chapter:stochacq}.

\paragraph{Bias-Variance Trade-Off.}
We can take a different perspective and view the softmax entropy of a single model as a (biased) estimator of the predictive entropy of the ensemble. 
What is the root mean squared error of this estimator?

\newcommand{\RMSEof}[2]{\mathrm{RMSE}_{#1}(\medmath{#2})}
\begin{proposition}
    \label{ddu:pro:ensemble_softmax_rmse}
    The root mean squared error of the softmax entropy of a single model as an estimator of the predictive entropy of the ensemble:
    \begin{align}
        \medmath{\RMSEof{\w}{\Hof{\Y \given \x, \W, \Dtrain}, \Hof{\Y \given \x, \Dtrain}} = \E{\pof{\w \given \Dtrain}}{\left(\Hof{\Y \given \x, \w} - \Hof{\Y \given \x, \Dtrain}\right)^2}^{1/2}}
    \end{align}
    decomposes into a bias-variance trade-off with bias $\MIof{\Y; \w \given \x, \Dtrain}$ and variance $\Var{\pof{\w\given\Dtrain}}{\Hof{\Y \given \x, \w}}$.
\end{proposition}
\begin{proof}
    We will drop conditioning on $\Dtrain$ for this proof.
    \begin{enumerate}
        \item We use that $\MIof{\Y; \w \given \x, \Dtrain} \defeq \Hof{\Y \given \x} - \Hof{\Y \given \x, \w, \Dtrain}$ (from \Cref{chapter:practical_it_notation}), 
        \item As $\Hof{\Y \given \x}$ is independent of $\W$, we have:
            \begin{align}
                \Var{\w}{\Hof{\Y \given \x, \w}} = \Var{\w}{\MIof{\Y; \w \given \x}}.
            \end{align}
        \item Expanding the variance, we obtain:
            \begin{align}
                \Var{\w}{\MIof{\Y; \w \given \x}} 
                &= \E{\w}{\MIof{\Y; \w \given \x}^2} - \E{\w}{\MIof{\Y; \w \given \x}}^2 \\
                &= \E{\w}{\MIof{\Y; \w \given \x}^2} - \MIof{\Y; \W \given \x}^2.
            \end{align}
        \item We substitute the definition of $\MIof{\Y;\w\given\x}$ in the first term and use the equality of the variances above. Rearranging:
        \begin{align}
            \E{\w}{(\Hof{\Y \given \x} - \Hof{\Y \given \x, \w, \Dtrain})^2} = \Var{\w}{\Hof{\Y \given \x, \w}} + \MIof{\Y; \W \given \x}^2. 
        \end{align}
        \item Taking the square root yields the result:
        \begin{align}
            &\RMSEof{\w}{\Hof{\Y \given \x, \W}, \Hof{\Y \given \x}} \\
            &\quad = \E{\w}{(\Hof{\Y \given \x} - \Hof{\Y \given \x, \w, \Dtrain})^2}^1/2 \\
            &\quad = \sqrtexplained{\underbrace{\Var{\w}{\Hof{\Y \given \x, \w}}}_{\text{Variance}} + {\underbrace{\MIof{\Y; \W \given \x}}_{\text{Bias}}}^2}.
        \end{align}
    \end{enumerate}
\end{proof}

Hence, the expected deviation from the predictive entropy becomes the largest for high mutual information/epistemic uncertainty. 
This is in line with the empirical observations that the softmax entropy of a deterministic model might not be a good estimator of the predictive entropy of the ensemble from \Cref{ddu:sec:predictive_entropy_not_epistemic_uncertainty_experiments} below.

Given that we can view an ensemble member as a single deterministic model and vice versa, these two propositions provide an intuitive explanation for why single deterministic models can report inconsistent and widely varying predictive entropies and confidence scores for OoD samples for which a deep ensemble would report high epistemic uncertainty (expected information gain) and high predictive entropy.

\subsection{Qualitative Empirical Validation}
\label{ddu:sec:predictive_entropy_not_epistemic_uncertainty_experiments}

While the 5-ensemble visualization and qualitative and quantitative statements provide us with an intuition for why ensemble members and thus deterministic models cannot provide epistemic uncertainty reliably through their softmax entropies, to gain further insights, we empirically analyze the relationship between softmax entropies and predictive entropies more precisely next.

To do so we train deep ensembles of 25 members on CIFAR-10 using different model architectures (VGG-16, Wide-ResNet-28-10/+SN) and visualize the relationship between the softmax entropies of the ensemble members, and the mutual information (epistemic uncertainty/EIG/BALD) and the predictive entropies of the overall ensemble on two OoD datasets (CIFAR-100 and SVHN). Details for the experiment setup can be found in \Cref{ddu:sec:experiments}.

\begin{figure}[t]
    \begin{subfigure}[t]{0.5\linewidth}
        \centering
        \includegraphics[width=\linewidth]{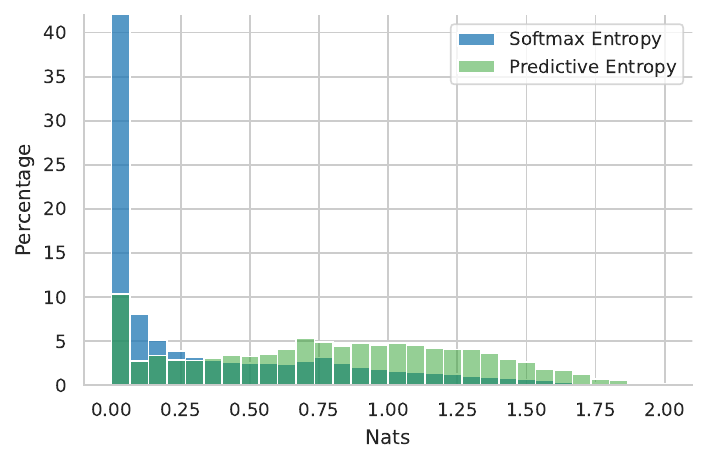}
        \caption{VGG16: CIFAR-100 (OoD)}
    \end{subfigure}\hfill
    \begin{subfigure}[t]{0.5\linewidth}
        \centering
        \includegraphics[width=\linewidth]{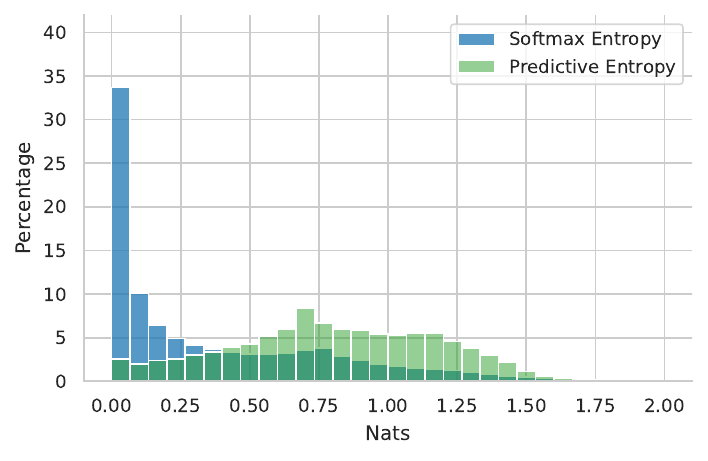}
        \caption*{SVHN (OoD)}
    \end{subfigure}
    \begin{subfigure}[t]{0.5\linewidth}
        \centering
        \includegraphics[width=\linewidth]{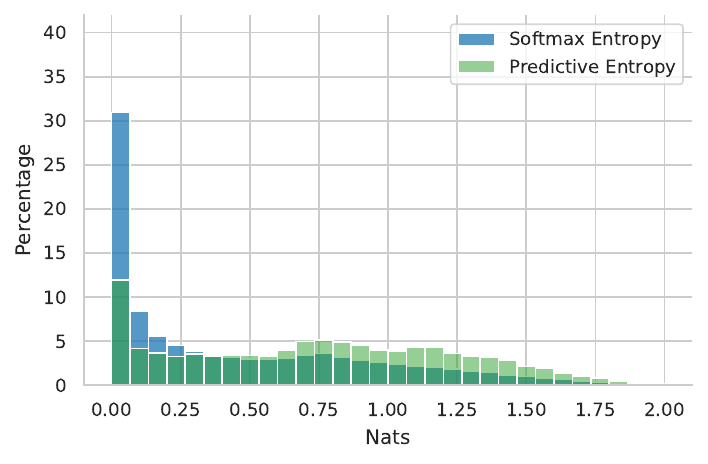}
        \caption{WideResNet-28-10: CIFAR-100 (OoD)}
    \end{subfigure}\hfill
    \begin{subfigure}[t]{0.5\linewidth}
        \centering
        \includegraphics[width=\linewidth]{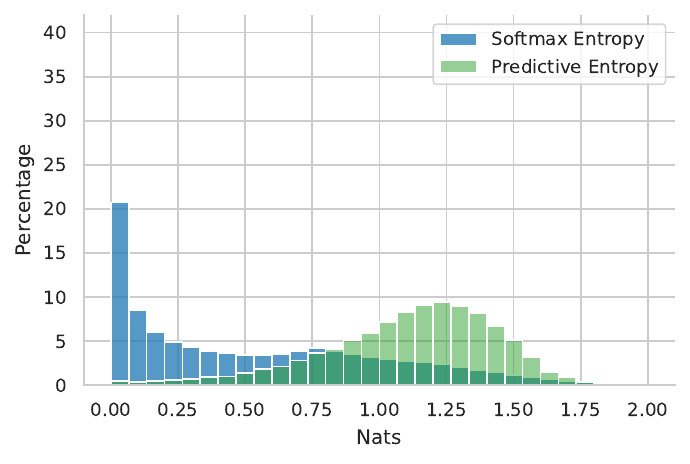}
        \caption*{SVHN (OoD)}
    \end{subfigure}
    \begin{subfigure}[t]{0.5\linewidth}
        \centering
        \includegraphics[width=\linewidth]{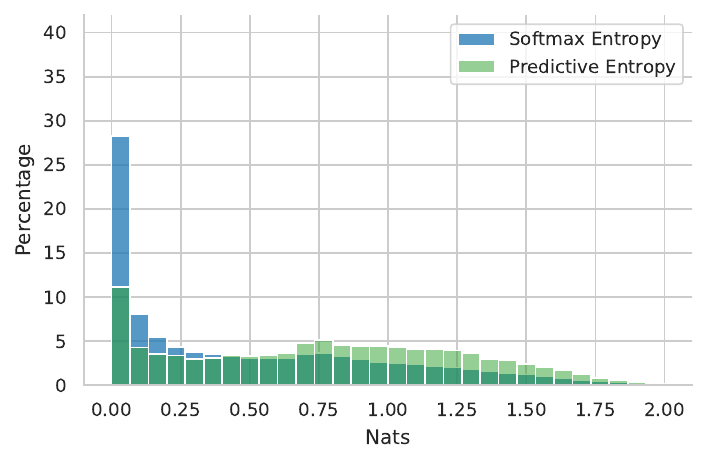}
        \caption{WideResNet-28-10+SN: CIFAR-100 (OoD)}
    \end{subfigure}\hfill
    \begin{subfigure}[t]{0.5\linewidth}
        \centering
        \includegraphics[width=\linewidth]{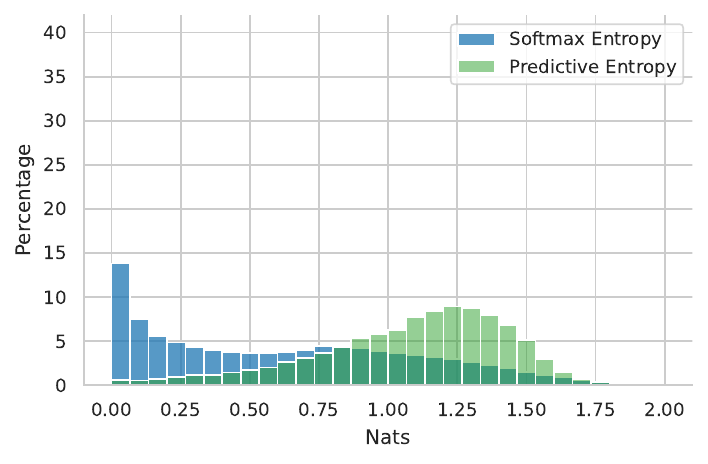}
        \caption*{SVHN (OoD)}
    \end{subfigure}
    \caption{
        \emph{Softmax entropy vs.\ predictive entropy trained on CIFAR-10 (iD) using different model architectures (25 models each).} We see that predictive and softmax entropy are distributed very differently. (\texttt{+SN} refers to models trained with spectral norm and small modifications to the architecture described in \Cref{ddu:app:implementation}.)
    }
    \label{ddu:fig:real_pe_vs_se}
\end{figure}

\begin{sidewaysfigure}
    \begin{subfigure}[t]{0.33\linewidth}
        \centering
        \includegraphics[width=\linewidth]{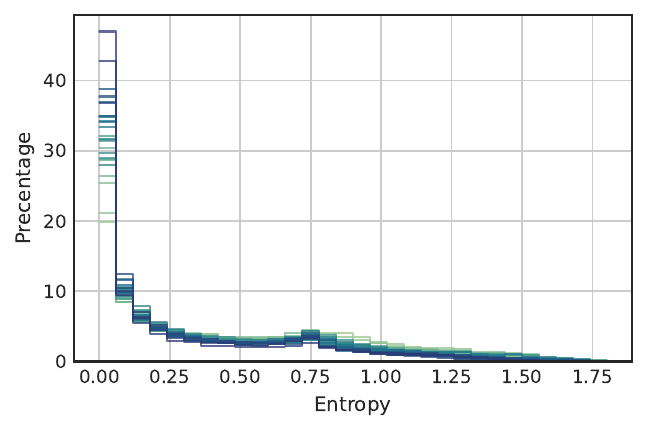}
        \caption{SVHN: Overlaid Histograms}
        \label{ddu:subfig:histograms:real_pe_vs_se:vgg16_svhn}
    \end{subfigure}\hfill
    \begin{subfigure}[t]{0.66\linewidth}
        \centering
        \includegraphics[width=\linewidth]{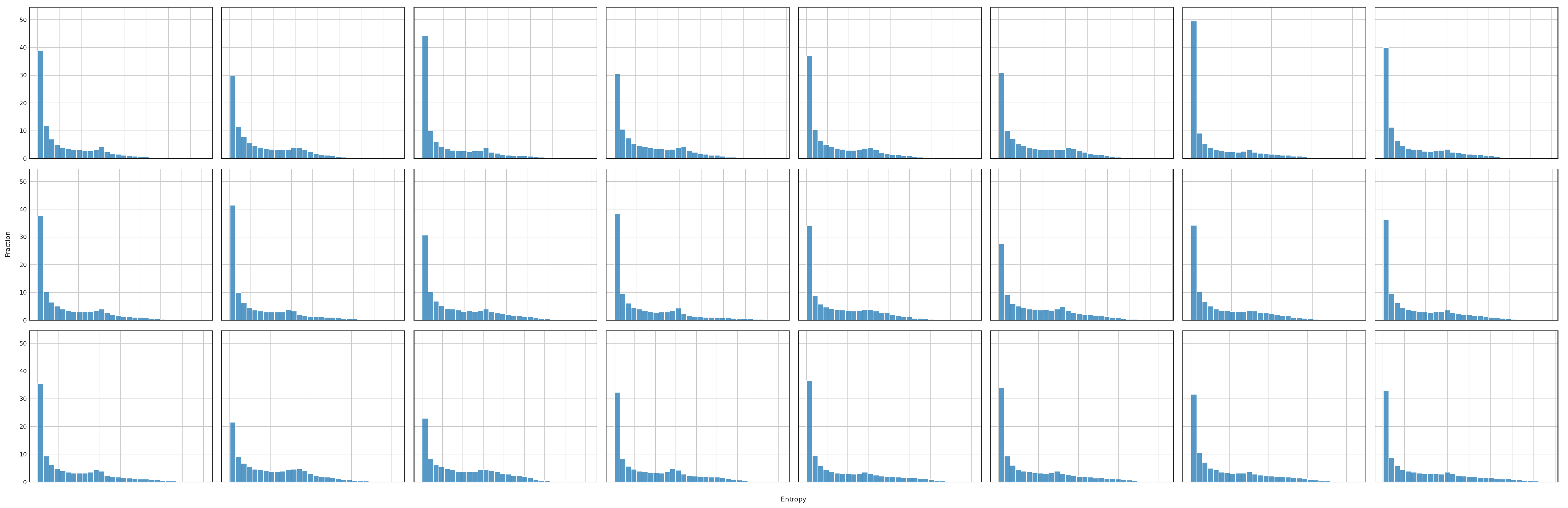}
        \caption*{Individual Histograms}
    \end{subfigure}
    \begin{subfigure}[t]{0.33\linewidth}
        \centering
        \includegraphics[width=\linewidth]{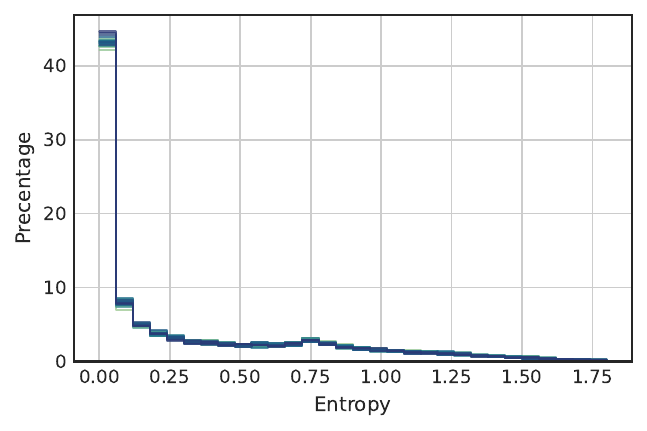}
        \caption{CIFAR-100: Overlaid Histograms}
        \label{ddu:subfig:histograms:real_pe_vs_se:vgg16_cifar100}
    \end{subfigure}\hfill
    \begin{subfigure}[t]{0.66\linewidth}
        \centering
        \includegraphics[width=\linewidth]{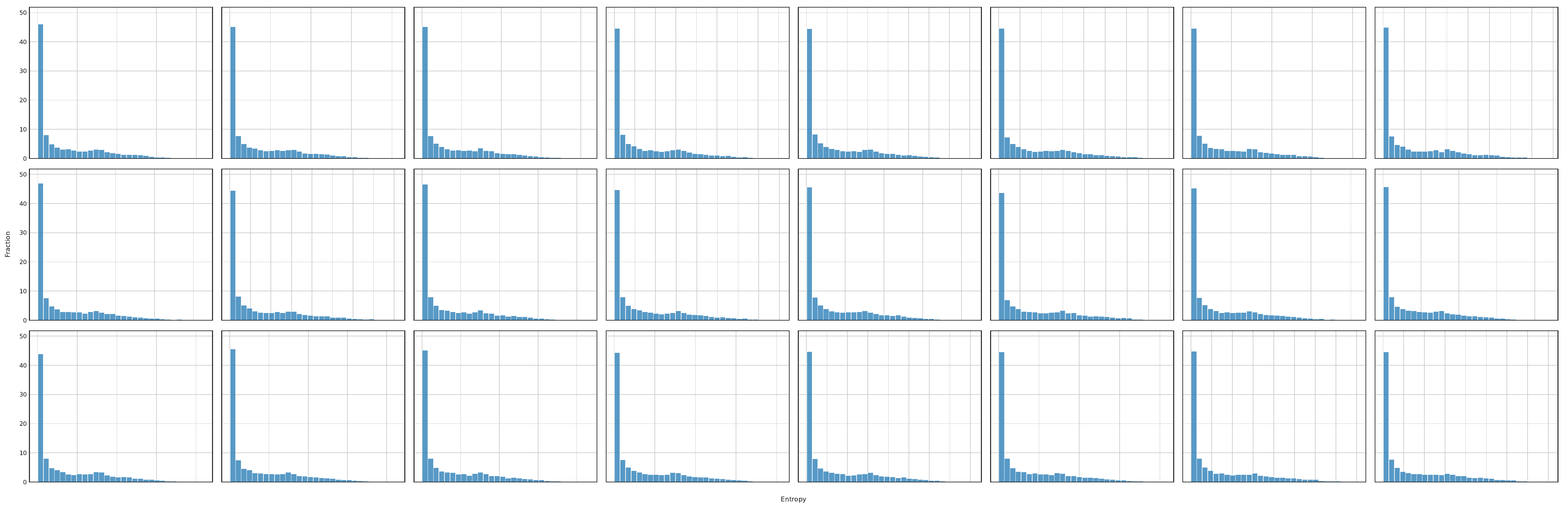}
        \caption*{Individual Histograms}
    \end{subfigure}
    \caption{
        \emph{VGG-16: Empirical distribution of softmax entropies of the ensemble members (25 models, 24 are shown).} We see that the empirical distribution of softmax entropies can vary a lot between different models of the same architecture.
        \textbf{\subref{ddu:subfig:histograms:real_pe_vs_se:vgg16_svhn}} the empirical distribution of softmax entropies for SVHN as OoD displays a lot of variance; while \textbf{\subref{ddu:subfig:histograms:real_pe_vs_se:vgg16_cifar100}} the empirical distribution of softmax entropies for CIFAR-100 as OoD display very little variation. \texttt{+SN} refers to models trained with spectral norm and small modifications to the architecture described in \Cref{ddu:app:implementation}.
    }
    \label{ddu:subfig:empirical_softmax_entropy:vgg16}
\end{sidewaysfigure}

\begin{sidewaysfigure}
    \begin{subfigure}[t]{0.33\linewidth}
        \centering
        \includegraphics[width=\linewidth]{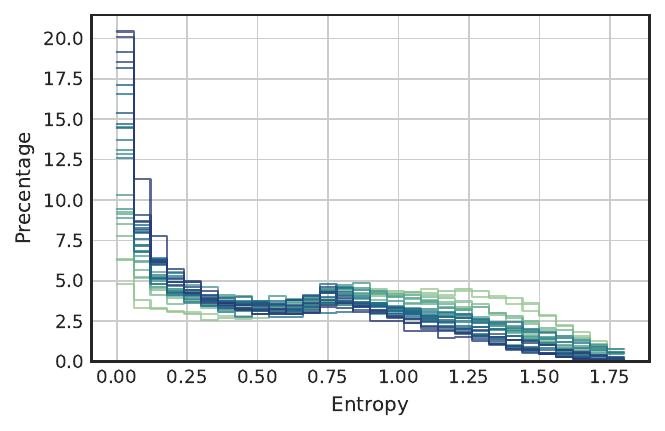}
        \caption{SVHN: Overlaid Histograms}
        \label{ddu:subfig:histograms:real_pe_vs_se:wide_resnet_sn_mod_svhn}
    \end{subfigure}\hfill
    \begin{subfigure}[t]{0.66\linewidth}
        \centering
        \includegraphics[width=\linewidth]{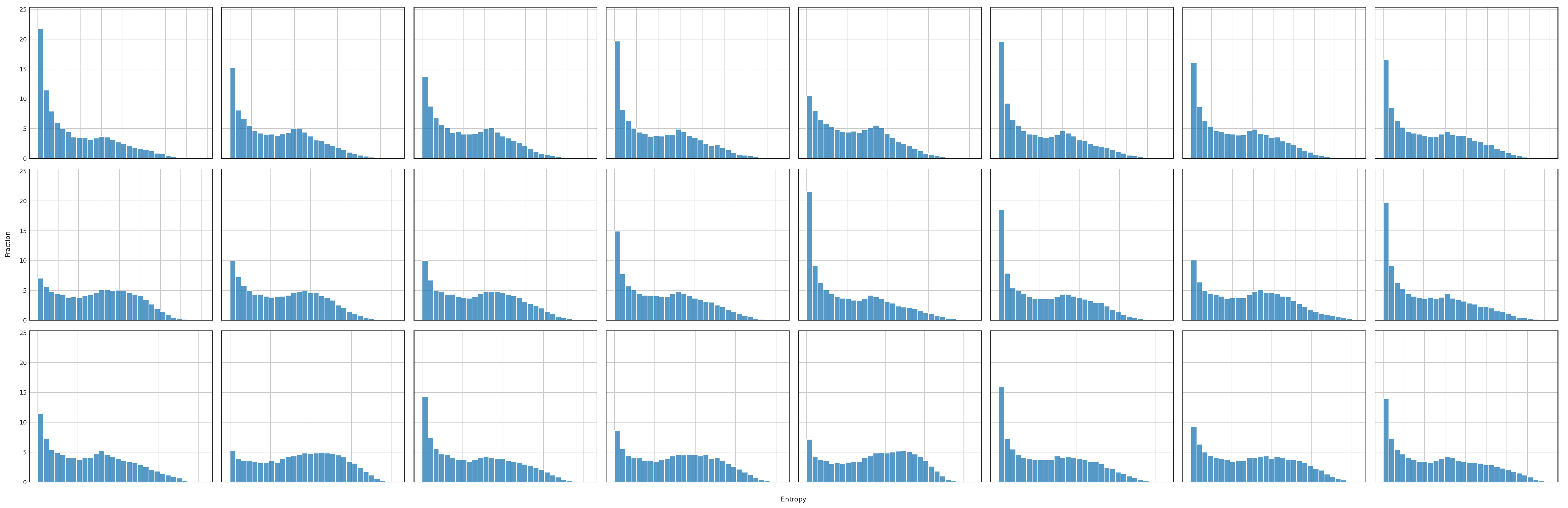}
        \caption*{Individual Histograms}
    \end{subfigure}
    \begin{subfigure}[t]{0.33\linewidth}
        \centering
        \includegraphics[width=\linewidth]{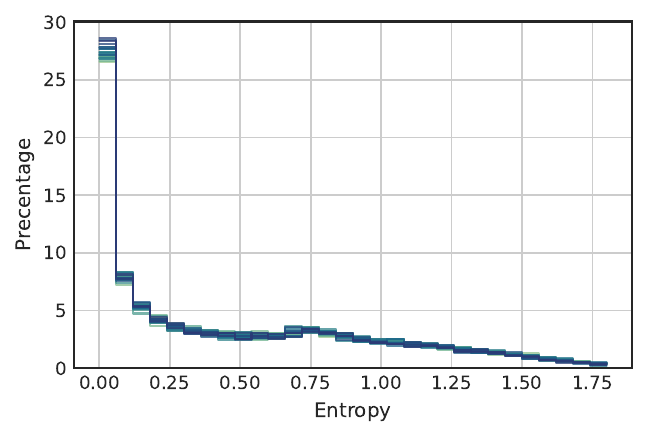}
        \caption{CIFAR-100: Overlaid Histograms}
        \label{ddu:subfig:histograms:real_pe_vs_se:wide_resnet_sn_mod_cifar100}
    \end{subfigure}\hfill
    \begin{subfigure}[t]{0.66\linewidth}
        \centering
        \includegraphics[width=\linewidth]{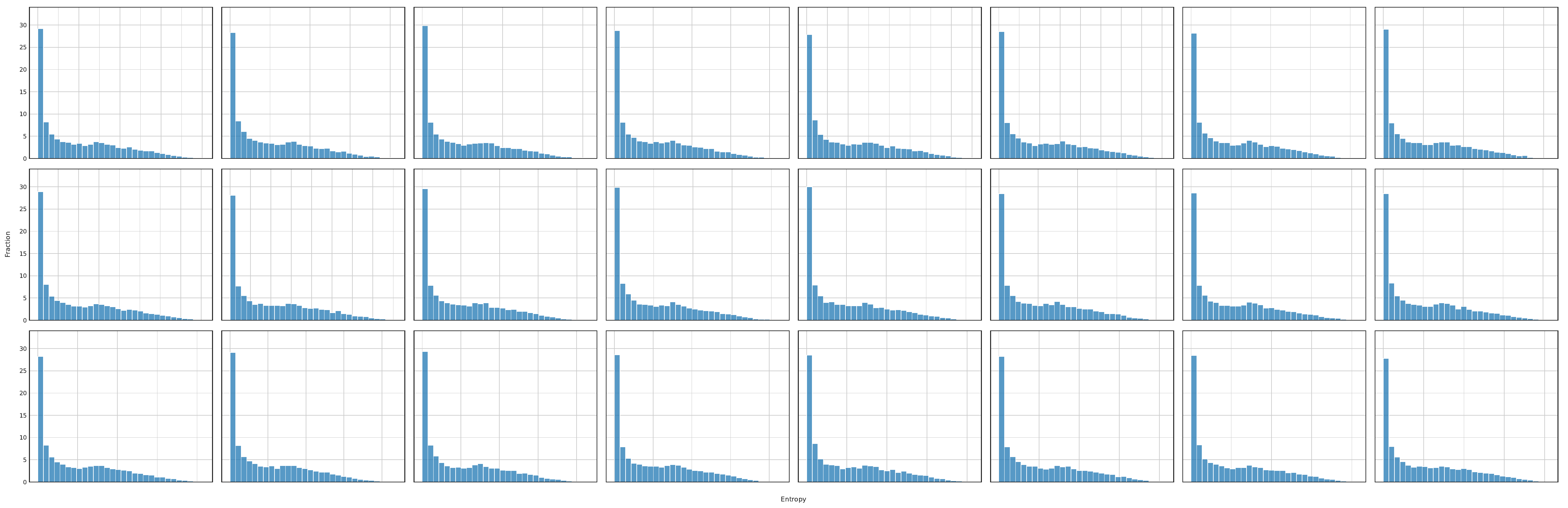}
        \caption*{Individual Histograms}
    \end{subfigure}
    \caption{
        \emph{Wide-ResNet-28-10+SN: Empirical distribution of softmax entropies of the ensemble members (25 models, 24 are shown).} We see that the empirical distribution of softmax entropies can vary a lot between different models of the same architecture.
        \textbf{\subref{ddu:subfig:histograms:real_pe_vs_se:wide_resnet_sn_mod_svhn}} the empirical distribution of softmax entropies for SVHN as OoD displays a lot of variation between models; while \textbf{\subref{ddu:subfig:histograms:real_pe_vs_se:wide_resnet_sn_mod_cifar100}} the empirical distribution of softmax entropies for CIFAR-100 as OoD display very little variation. (\texttt{+SN} refers to models trained with spectral norm and small modifications to the architecture described in \Cref{ddu:app:implementation}.)
    }
    \label{ddu:subfig:empirical_softmax_entropy:wide_resnet_sn_mod}
\end{sidewaysfigure}

\begin{figure}[t]
    \begin{subfigure}[t]{0.5\linewidth}
        \centering
        \includegraphics[width=\linewidth]{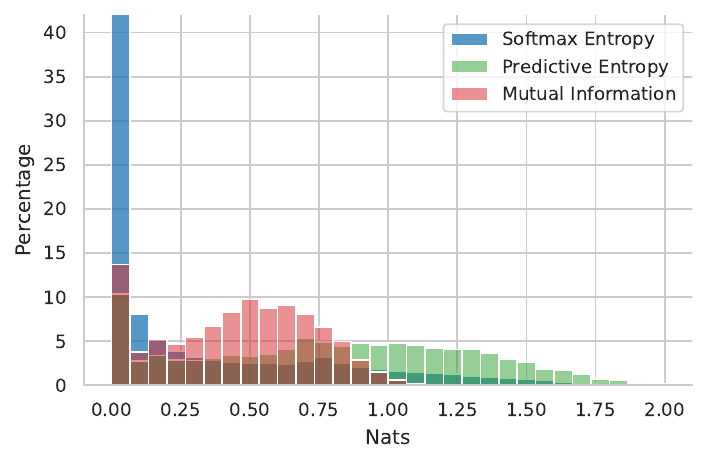}
        \caption{VGG16: CIFAR-100 (OoD)}
    \end{subfigure}\hfill
    \begin{subfigure}[t]{0.5\linewidth}
        \centering
        \includegraphics[width=\linewidth]{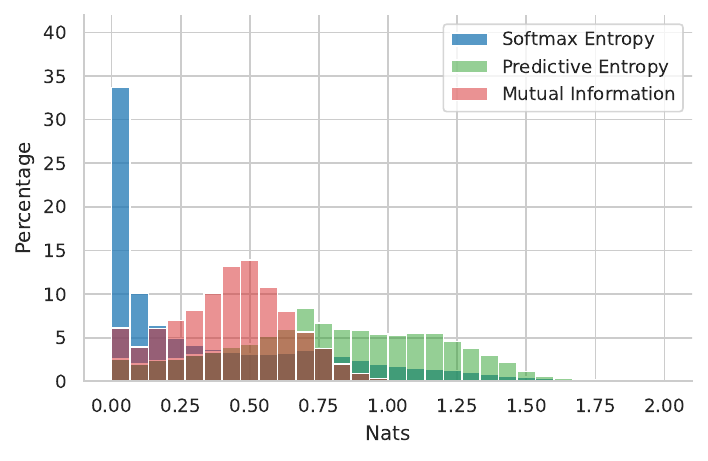}
        \caption*{SVHN (OoD)}
    \end{subfigure}
    \begin{subfigure}[t]{0.5\linewidth}
        \centering
        \includegraphics[width=\linewidth]{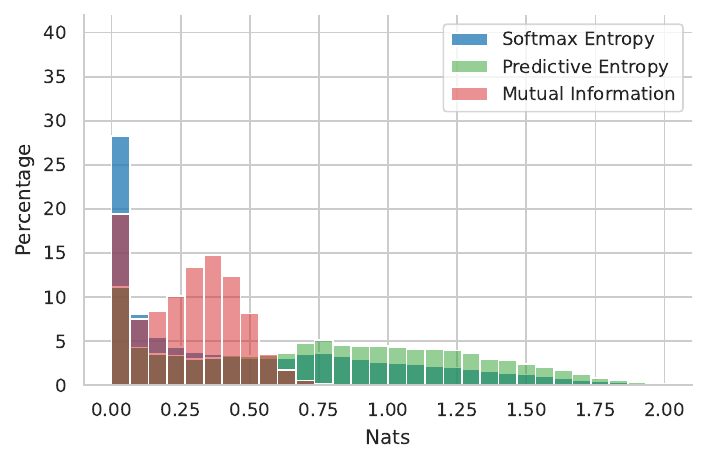}
        \caption{WideResNet-28-10+SN: CIFAR-100 (OoD)}
    \end{subfigure}\hfill
    \begin{subfigure}[t]{0.5\linewidth}
        \centering
        \includegraphics[width=\linewidth]{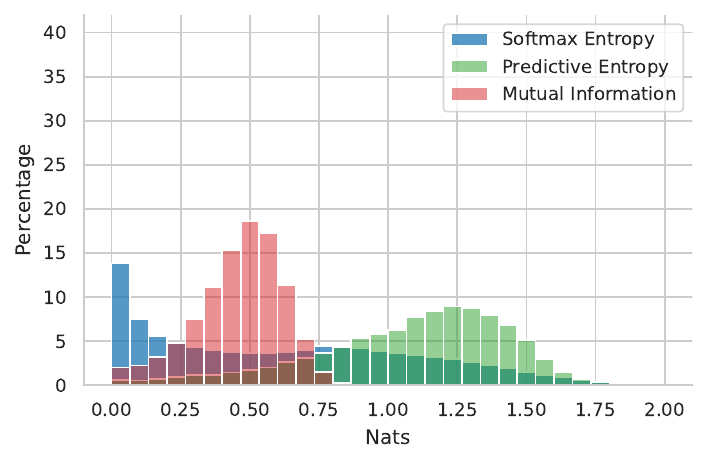}
        \caption*{SVHN (OoD)}
    \end{subfigure}
    \caption{
        \emph{Mutual Information (Epistemic Uncertainty).} Trained on CIFAR-10 (iD) using different model architectures (25 models each). The mutual information is better behaved than the softmax entropies, but less broad than the predictive entropy. (\texttt{+SN} refers to models trained with spectral norm and small modifications to the architecture described in \Cref{ddu:app:implementation}.)
    }
    \label{ddu:fig:real_pe_vs_se_vs_eig}
\end{figure}

\begin{figure}[t]
    \begin{subfigure}[t]{0.5\linewidth}
        \centering
        \includegraphics[width=\linewidth]{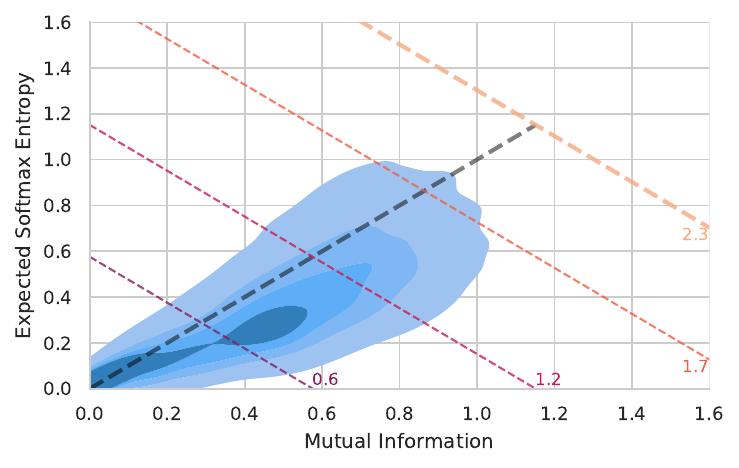}
        \caption{VGG16: CIFAR-100 (OoD)}
    \end{subfigure}\hfill
    \begin{subfigure}[t]{0.5\linewidth}
        \centering
        \includegraphics[width=\linewidth]{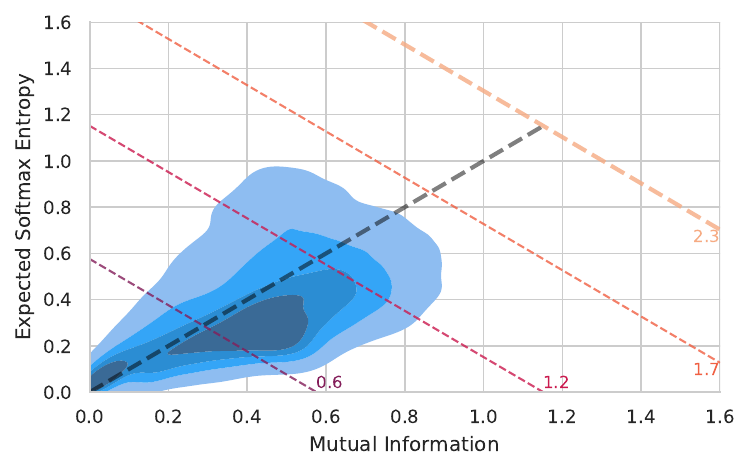}
        \caption*{SVHN (OoD)}
    \end{subfigure}
    \begin{subfigure}[t]{0.5\linewidth}
        \centering
        \includegraphics[width=\linewidth]{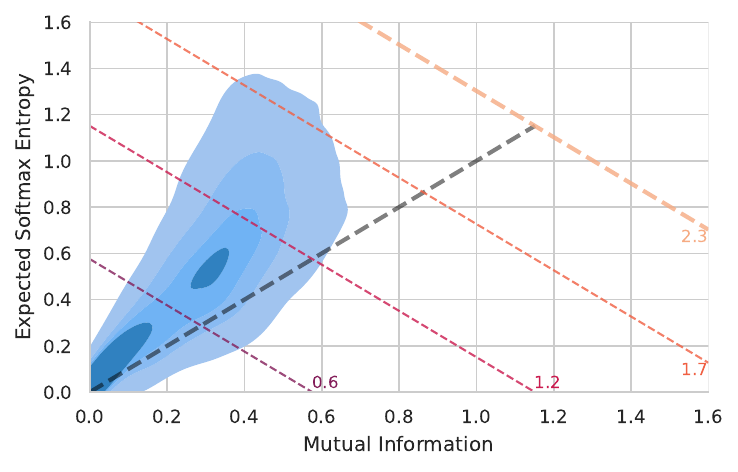}
        \caption{WideResNet-28-10+SN: CIFAR-100 (OoD)}
    \end{subfigure}\hfill
    \begin{subfigure}[t]{0.5\linewidth}
        \centering
        \includegraphics[width=\linewidth]{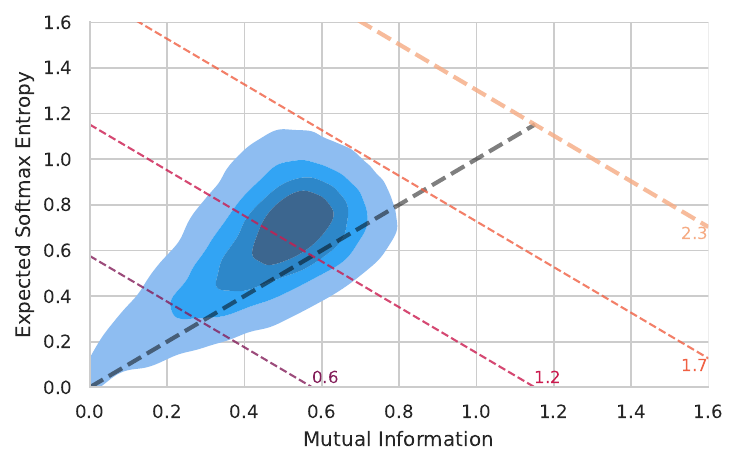}
        \caption*{SVHN (OoD)}
    \end{subfigure}
    \caption{
        \emph{Mutual Information (Epistemic Uncertainty) vs Expected Softmax Entropy (Aleatoric Uncertainty) vs Predictive Entropy (Total Uncertainty).} Trained on CIFAR-10 (iD) using different model architectures (25 models each). The predictive entropy is shown via its iso-lines (anti-diagonals). Darker is denser.(\texttt{+SN} refers to models trained with spectral norm and small modifications to the architecture described in \Cref{ddu:app:implementation}.)
    }
    \label{ddu:fig:eig_vs_ese}
\end{figure}

\begin{figure}[t]
    \begin{subfigure}[t]{\linewidth}
    \begin{subfigure}[t]{0.25\linewidth}
        \centering
        \includegraphics[width=\linewidth]{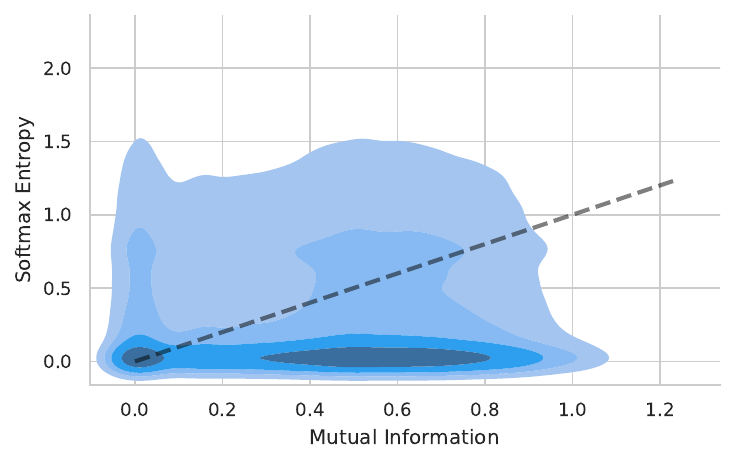}
        \caption*{VGG16: CIFAR-100}
    \end{subfigure}\hfill
    \begin{subfigure}[t]{0.25\linewidth}
        \centering
        \includegraphics[width=\linewidth]{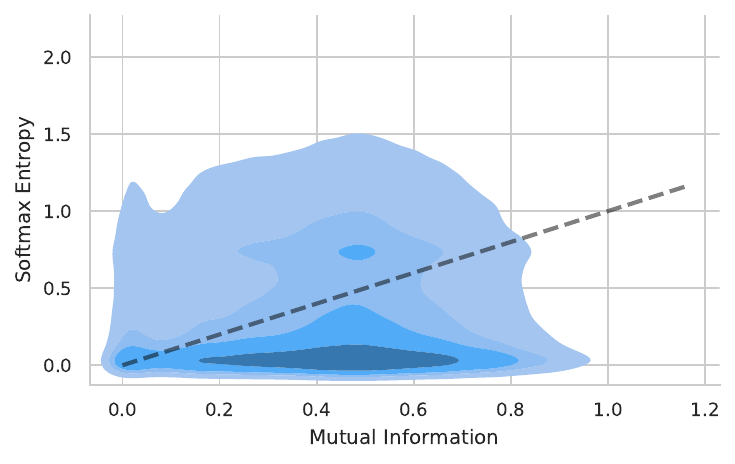}
        \caption*{SVHN}
    \end{subfigure}\hfill
    \begin{subfigure}[t]{0.25\linewidth}
        \centering
        \includegraphics[width=\linewidth]{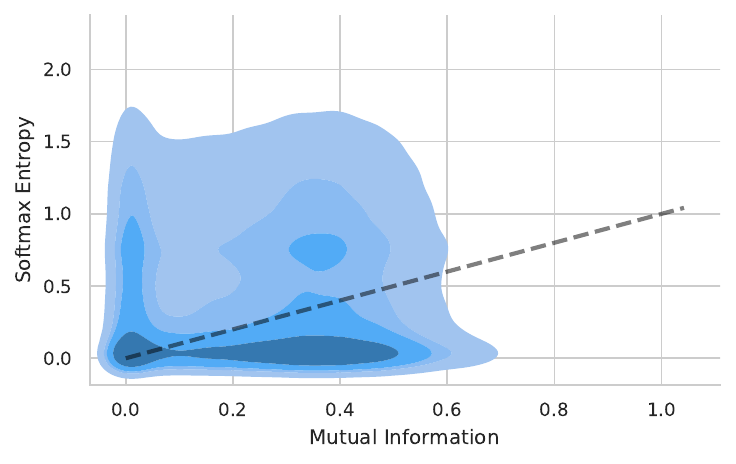}
        \caption*{WRN+SN: CIFAR-100}
    \end{subfigure}\hfill
    \begin{subfigure}[t]{0.25\linewidth}
        \centering
        \includegraphics[width=\linewidth]{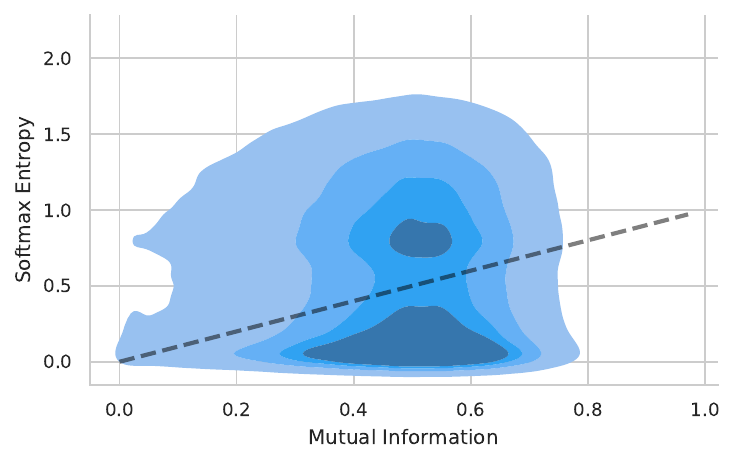}
        \caption*{SVHN}
    \end{subfigure}
        \caption{Mutual Information vs Softmax Entropy}
    \end{subfigure}
    \begin{subfigure}[t]{\linewidth}
        \begin{subfigure}[t]{0.5\linewidth}
            \centering
            \includegraphics[width=\linewidth]{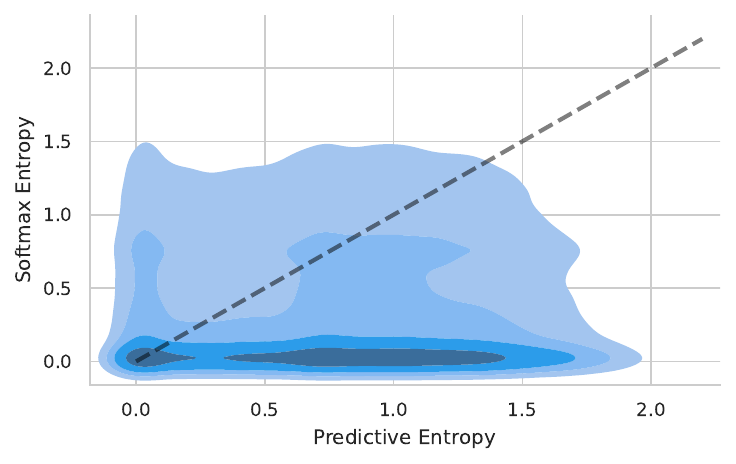}
            \caption*{VGG16: CIFAR-100 (OoD)}
        \end{subfigure}\hfill
        \begin{subfigure}[t]{0.5\linewidth}
            \centering
            \includegraphics[width=\linewidth]{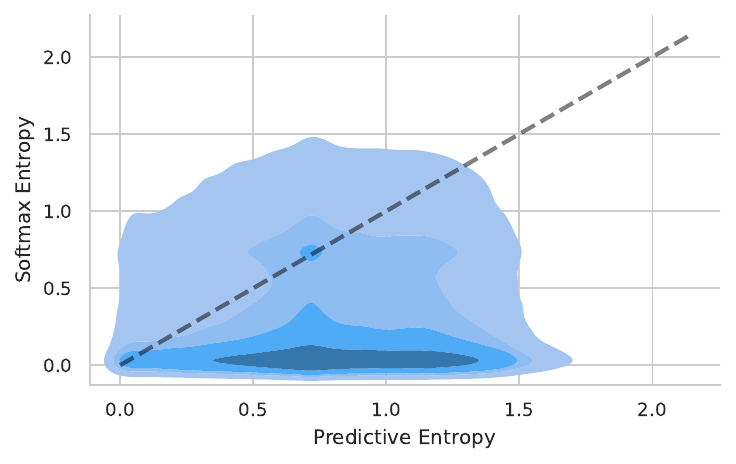}
            \caption*{SVHN (OoD)}
        \end{subfigure}
        \begin{subfigure}[t]{0.5\linewidth}
            \centering
            \includegraphics[width=\linewidth]{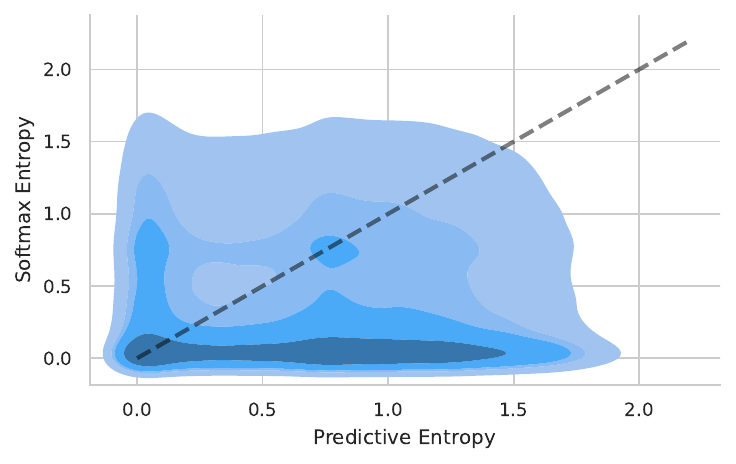}
            \caption*{WideResNet-28-10+SN: CIFAR-100 (OoD)}
        \end{subfigure}\hfill
        \begin{subfigure}[t]{0.5\linewidth}
            \centering
            \includegraphics[width=\linewidth]{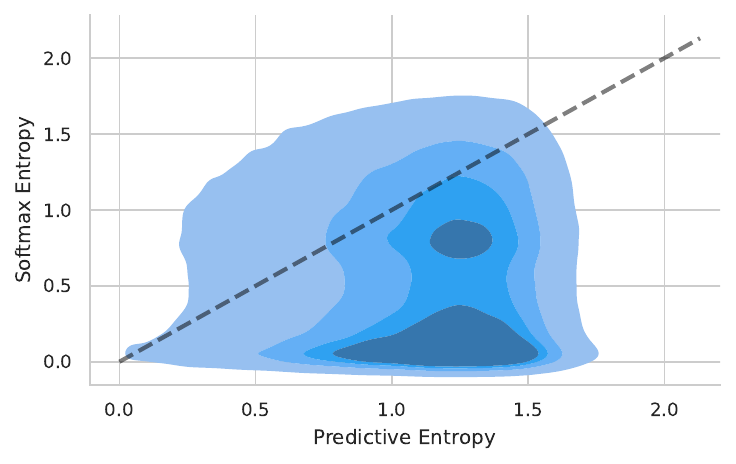}
            \caption*{SVHN (OoD)}
        \end{subfigure}
        \caption{Predictive Entropy vs Softmax Entropy}
    \end{subfigure}
    \caption{
        \emph{Predictive Entropy (Total Uncertainty) \&\ Mutual Information (Epistemic Uncertainty) vs Expected Softmax Entropy (Aleatoric Uncertainty)} Trained on CIFAR-10 (iD) using different model architectures (25 models each). Darker is denser.(\texttt{+SN} refers to models trained with spectral norm and small modifications to the architecture described in \Cref{ddu:app:implementation}.)
    }
    \label{ddu:fig:eig_and_pe_vs_se}
\end{figure}

\paragraph{Softmax Entropy vs Predictive Entropy.}
\Cref{ddu:fig:real_pe_vs_se} shows that the softmax entropy of the ensemble members is not a good estimator of the predictive entropy of the ensemble. Not even the histograms look similar. 
\Cref{ddu:subfig:empirical_softmax_entropy:vgg16,ddu:subfig:empirical_softmax_entropy:wide_resnet_sn_mod} show histograms of the different ensemble members.
We see that on SVHN there is a lot more variation in the softmax entropy distribution of the ensemble members than on CIFAR-100. This is also reflected in the predictive entropy of the ensemble, which is often higher on SVHN than on CIFAR-100.
\Cref{ddu:fig:real_pe_vs_se_vs_eig}, which also includes the mutual information (EIG) of the ensemble, validates this: 
the mutual information near 0 is much lower on CIFAR-100 than on SVHN.
The model expresses less disagreement on CIFAR-100.

\Cref{ddu:fig:eig_vs_ese} provides a more complete picture of all three information quantities we care about (mutual information, predictive entropy, and expected softmax entropy). The model architectures perform very differently on the two datasets: the VGG-16 model has smaller predictive entropies overall which pushes the expected softmax entropies below the mutual information for many OoD samples.

\Cref{ddu:fig:eig_and_pe_vs_se} shows density plots for predictive entropy versus softmax entropy (and mutual information versus softmax entropy) across all ensemble members. We see that most of the softmax entropies are very low even when the same sample has very high predictive entropy or mutual information. 
This implies that even small aleatoric uncertainty in iD samples will lead the model to confound them with OoD samples.

\begin{figure}[t]
    \begin{subfigure}[t]{0.5\linewidth}
        \centering
        \includegraphics[width=\linewidth]{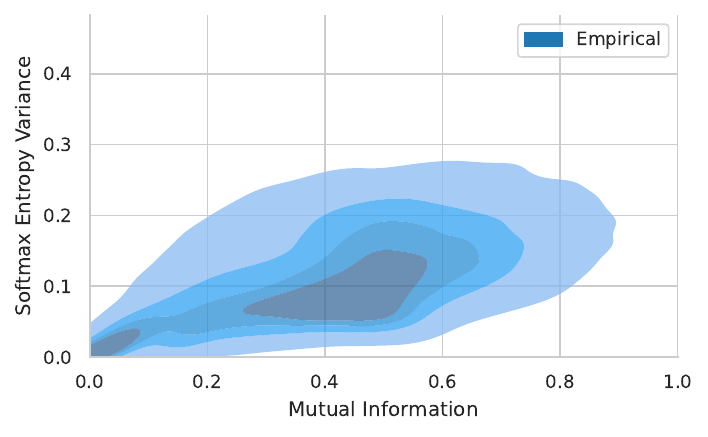}
        \caption{SVHN (OoD): VGG-16}
    \end{subfigure}\hfill
    \begin{subfigure}[t]{0.5\linewidth}
        \centering
        \includegraphics[width=\linewidth]{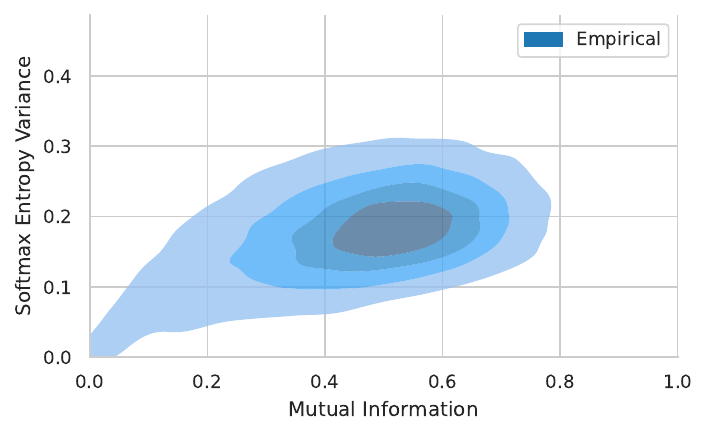}
        \caption*{Wide-ResNet-28-10+SN}
    \end{subfigure}
    \begin{subfigure}[t]{0.5\linewidth}
        \centering
        \includegraphics[width=\linewidth]{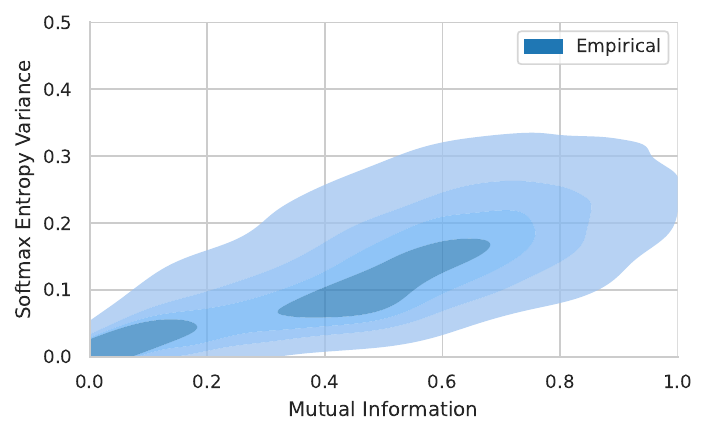}
        \caption{CIFAR-100 (OoD): VGG-16}
    \end{subfigure}\hfill
    \begin{subfigure}[t]{0.5\linewidth}
        \centering
        \includegraphics[width=\linewidth]{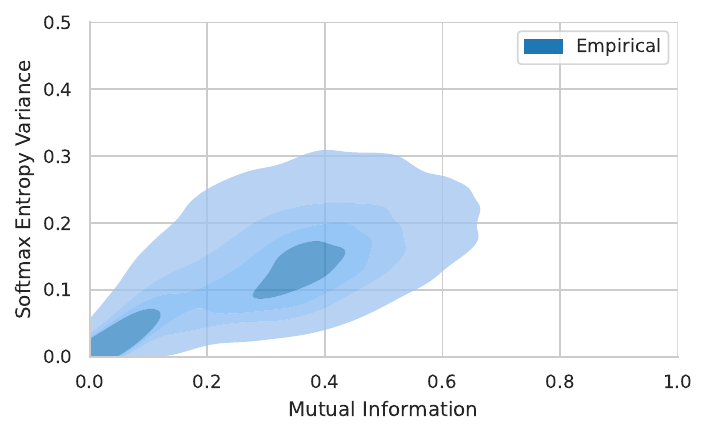}
        \caption*{Wide-ResNet-28-10+SN}
    \end{subfigure}
    \caption{
        \emph{Mutual Information (Epistemic Uncertainty) vs Softmax Entropy Variance (ESV).} Trained on CIFAR-10 (iD) using different model architectures (25 models each). For both SVHN and CIFAR-100 as OoD dataset, we see that the softmax variance first increases as the mutual information increases and then decreases. Overall, compared to the mutual information, it is quite small. (\texttt{+SN} refers to models trained with spectral norm and small modifications to the architecture described in \Cref{ddu:app:implementation}).
    }
    \label{ddu:fig:real_eig_vs_esv}
\end{figure}

\begin{figure}[t]
    \begin{subfigure}[t]{0.5\linewidth}
        \centering
        \includegraphics[width=\linewidth]{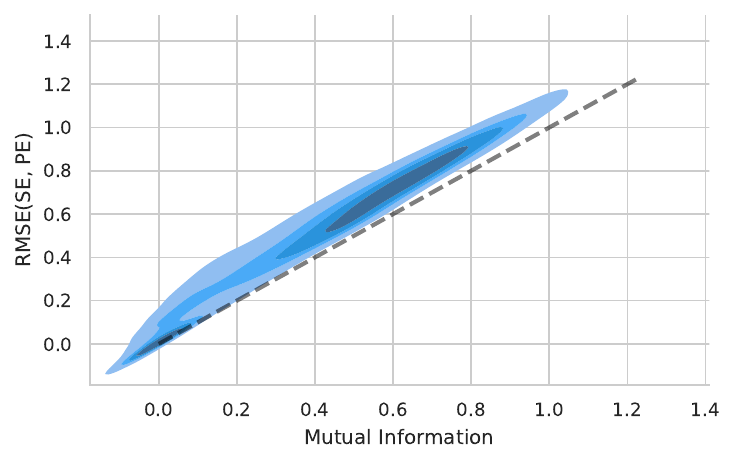}
        \caption{SVHN (OoD): VGG-16}
    \end{subfigure}\hfill
    \begin{subfigure}[t]{0.5\linewidth}
        \centering
        \includegraphics[width=\linewidth]{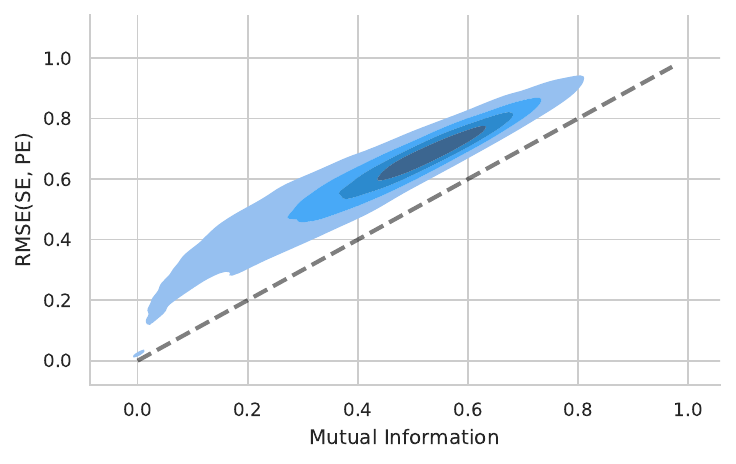}
        \caption*{Wide-ResNet-28-10+SN}
    \end{subfigure}
    \begin{subfigure}[t]{0.5\linewidth}
        \centering
        \includegraphics[width=\linewidth]{figs/thesis/de_val_split_vgg16_cifar100_test_contour_eig_vs_real_rmse.pdf}
        \caption{CIFAR-100 (OoD): VGG-16}
    \end{subfigure}\hfill
    \begin{subfigure}[t]{0.5\linewidth}
        \centering
        \includegraphics[width=\linewidth]{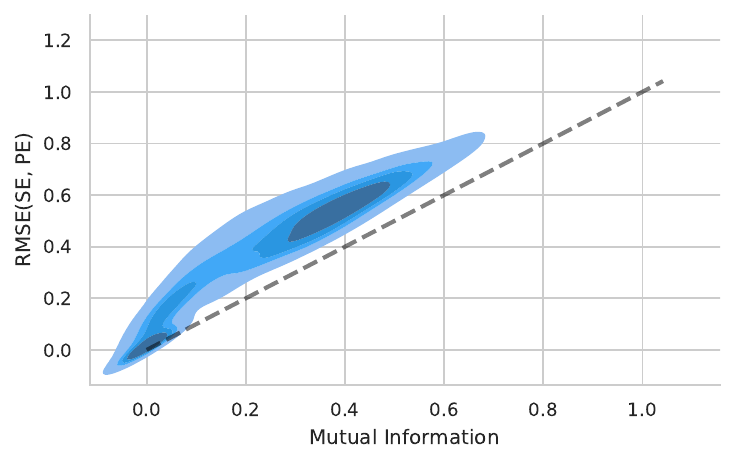}
        \caption*{Wide-ResNet-28-10+SN}
    \end{subfigure}
    \caption{
        \emph{Mutual Information (Epistemic Uncertainty) vs (Entropy) Root Mean Squared Error (RMSE, \Cref{ddu:pro:ensemble_softmax_rmse}).} Trained on CIFAR-10 (iD) using different model architectures (25 models each). For both SVHN and CIFAR-100 as OoD dataset, we see that the RMSE increases as the mutual information (the bias) increases. The bias seems to be dominating the error compared to the softmax variance. (\texttt{+SN} refers to models trained with spectral norm and small modifications to the architecture described in \Cref{ddu:app:implementation}).
    }
    \label{ddu:fig:real_eig_vs_real_rmse}
\end{figure}

\paragraph{Softmax Entropy Variance.}
\Cref{ddu:fig:real_eig_vs_esv} show density plots for the mutual information vs the variance of the softmax entropy of the ensemble.
As the mutual information increases, the variance of the softmax entropy first increases and then decreases slightly.
Overall, it is mostly quite low compared to the mutual information.
\Cref{ddu:fig:real_eig_vs_real_rmse} validates this. It shows a histogram of the RMSE following \Cref{ddu:pro:ensemble_softmax_rmse} vs the mutual information. Indeed, the error is dominated by its bias, the mutual information.

\subsection{Discussion}

Both through quantitative and qualitative statements as well as through empirical validation, we see that neither the predictive entropy of deep ensembles nor the softmax entropy of deterministic models is appropriate for measuring epistemic uncertainty and OoD detection tasks. This holds in particularly for real world datasets that contain more ambiguous data than the curated datasets that are employed for benchmarking.
\Cref{ddu:pro:ensemble_softmax} shows that the softmax entropy does not provide a stable ranking of points with high epistemic uncertainty, and that indeed, the predictive entropy of deep ensembles captures epistemic uncertainty through the variance of the softmax entropies of deterministic models.
\Cref{ddu:pro:ensemble_softmax_rmse} shows that when we view softmax entropies as estimators of the respective predictive entropies of an ensemble and decompose the error using a bias-variance trade-off, the mutual information (epistemic uncertainty) is the bias of this estimator, and the variance of the softmax entropy is the variance of the estimator.
Additionally, we have empirically analyzed several model architectures and found that the softmax entropy varies considerably across the ensemble members and is neither a good measure of epistemic uncertainty (with the mutual information as proxy) nor of predictive entropy, which could be surprising given that we usually do not differentiate between softmax and predictive entropy.

While our criticism of softmax entropy seems generally valid, mutual information (expected information gain/epistemic uncertainty) is not necessarily a good measure for \emph{far OoD detection} as we have argued in \S\ref{ddu:sec:active_learning_vs_ood_detection}. Predictive entropy can be high for both far OoD points and near OoD points, which it can confound with ambiguous iD points, however. 

Is there a way to ensure that the model will have high uncertainty for OoD points in general (and thus minimize the amount of possible far OoD points)?
Yes, feature-space regularization, whether implicit and explicit, can do just that: in \Cref{glue:background:uncertainty_quantification}, we introduced bi-Lipschitzness as a concept that can encourage OoD points to be separated from iD points, thus allowing different ensemble members to potentially (hopefully) express higher disagreement on these points.

Importantly, note that outlier exposure, a popular method to improve OoD detection performance, trains the model on held-out ``OoD'' data. This breaks the equivalence ``\emph{OoD data $\Longleftrightarrow$ high epistemic uncertainty}'' that underlies using epistemic uncertainty for OoD detection:
training using outlier exposure transforms epistemic uncertainty into aleatoric uncertainty, which can be captured by the softmax entropy of deterministic models as well as the predictive entropy of a deep ensemble---\emph{even though it confounds ambiguous iD samples with OoD samples}. However, is such data still truly OoD when we start training or fine-tuning on it, or are we simply moving the goal posts?

\section{Aleatoric \&\ Epistemic Uncertainty}
\label{ddu:sec:motivation}

The failure of softmax entropy to capture epistemic uncertainty motivates us to study feature-space density as an alternative for single-forward pass approaches. 
Indeed, feature-space density is a well-known acquisition function in active learning \citep{Settles2010}.

\paragraph{Epistemic Uncertainty via Feature-Space Density.}
In \S\ref{glue:background:uncertainty_quantification}, we noted that feature-space density conceptually fulfills the requirement of epistemic uncertainty as being a reducible uncertainty. Here, we will investigate this further---or rather, to be more precise---we will investigate the \emph{negative} feature-space density as a proxy for epistemic uncertainty.

\andreas{Note and I've made this mistake a few times already in the last few days: epistemic uncertainty is not about the causal mechanism but about the outputs here. It's about the predictions. If we have high feature-space density on the training set and the model can fit different labels well, then the epistemic uncertainty will go down.}

With a well-regularized feature space using spectral normalization, we find that simply performing GDA (Gaussian Discriminant Analysis) \emph{after training} as feature-space density estimator can reliably capture epistemic uncertainty. However, unlike \citet{lee2018simple}, which does not place any constraints on the feature space, training on ``OoD'' hold-out data, feature ensembling, and input pre-processing are not needed to obtain good performance (see \Cref{ddu:table:ood_wrn}). This results in a conceptually simpler method. Moreover, we find that using a separate covariance matrix for each class improves OoD detection performance as compared to a shared covariance matrix.

Crucially, feature-space density cannot express aleatoric uncertainty: an iD sample ought to have a high density regardless of whether it is ambiguous (with high aleatoric uncertainty) or not, as is the case with Dirty-MNIST in \Cref{ddu:fig:intro_gmm}. %
However, softmax entropy is suitable for that. 
Hence, we can use the softmax entropy to predict the aleatoric uncertainty on iD samples with low epistemic uncertainty, for which is it meaningfully defined:
\begin{importantresult}
\begin{observation}
The softmax entropy of a deterministic model together with its feature-space density can disentangle epistemic and aleatoric uncertainty while either alone cannot.
\end{observation}
\end{importantresult}

\paragraph{Sensitivity \&\ Smoothness.}
As discussed in \S\ref{glue:background:uncertainty_quantification}, feature extractors that do not fulfill a sensitivity constraint
can suffer from feature collapse: they might map OoD samples to iD regions of the feature space. Thus, we encourage these properties using residual connections and spectral normalization.
The effects of these properties on feature collapse are visible in \Cref{ddu:fig:intro_histograms}. In the case of feature collapse, we must have \textit{some} OoD inputs for which the features are mapped on top of the features of iD inputs. 
The distances of these OoD features to each class centroid must be equal to the distances of the corresponding iD inputs to class centroids, and hence the density for these OoD inputs must be equal to the density of the iD inputs. 
If the density histograms for given iD and OoD samples do not overlap, no feature collapse can be present for those samples.
We see no overlapping densities in \Cref{ddu:fig:intro_gmm}(right most), therefore we indeed have no feature collapse between Dirty-MNIST and FashionMNIST.
Similarly, smoothness constraints are necessary to encourage generalization when using feature-space density as a proxy for epistemic uncertainty \citep{van2020uncertainty}.

\paragraph{Gaussian \&\ Linear Discriminant Analysis.}
Given feature vector $\z$ and class label $\y$, we  model the class-conditional probabilities $\qof{\z \given \y}$ of feature vectors given class labels using Gaussian Discriminant Analysis (GDA), which is a generative classifier $\qof{\z, \y}$ based on a Gaussian mixture model (GMM) $\qof{\z \given \y} \, \qof{\y}$, where $\qof{\y}$ is the class prior and $\qof{\z \given \y}$ is a multivariate Gaussian density. GDA is closely related to Linear Discriminant Analysis (LDA) \citep{murphy2012machine}.
To predict the class label $\qof{\y \given \z}$ of a new sample $\z$, we use Bayes' theorem:
\begin{align}
    \qof{y \given \z} \propto \qof{\z \given y}\,\qof{y}
\end{align}

In GDA, each class is modeled by a Gaussian mixture component with its own covariance matrix. The class-conditional probabilities are given by $\qof{\z \given \y=c} = \normaldistpdf{\z}{\mu_c}{\Sigma_c}$, where $\mu_c$ and $\Sigma_c$ are the mean and covariance matrix for class $c$, respectively. 

On the other hand, LDA models the class-conditional probabilities using a single multivariate Gaussian distribution per class, with a \emph{shared} covariance matrix $\Sigma$ among all classes. The class-conditional probabilities in LDA are given by $\qof{\z \given \y=c} \sim \normaldistpdf{\z}{\mu_c}{\Sigma}$.

Thus, the difference between GDA and LDA lies in their modeling of the covariance matrix. While GDA allows for a separate covariance matrix for each class, LDA assumes a shared covariance matrix among all classes. This distinction influences the flexibility and complexity of the models: GDA is more flexible but potentially more prone to overfitting, while LDA is more constrained but may be more robust in some scenarios.

\paragraph{Scope.}
We only use GDA for estimating the feature-space density as it is straight-forward to implement and does not require performing expectation maximization or variational inference like other density estimators. Normalizing flows \citep{dinh2015nice} or other more complex density estimators might provide even better density estimates, of course. Yet despite its simplicity, GDA is already sufficient to outperform other more complex approaches and obtain good results as we report in \Cref{ddu:sec:experiments}.

Our focus is on obtaining a well-regularized feature space using spectral normalization in model architectures with residual connections, following \citet{liu2020simple}. 
Note that unsupervised methods using contrastive learning \citep{winkens2020contrastive} might also obtain such a feature space by training on very large datasets, but training on them can be very expensive \citep{sun2017revisiting}.
Generally, as the amount of training data available grows and feature extractors improve, the quality of feature representations might improve as well. The underlying motivation of this chapter is that simple approaches will remain more applicable than more complex ones as our empirical results suggest.

\begin{figure*}[!t]
    \centering
    \begin{subfigure}{\linewidth}
        \centering
        \begin{subfigure}[t]{0.33\linewidth}
            \centering
            \includegraphics[width=\linewidth]{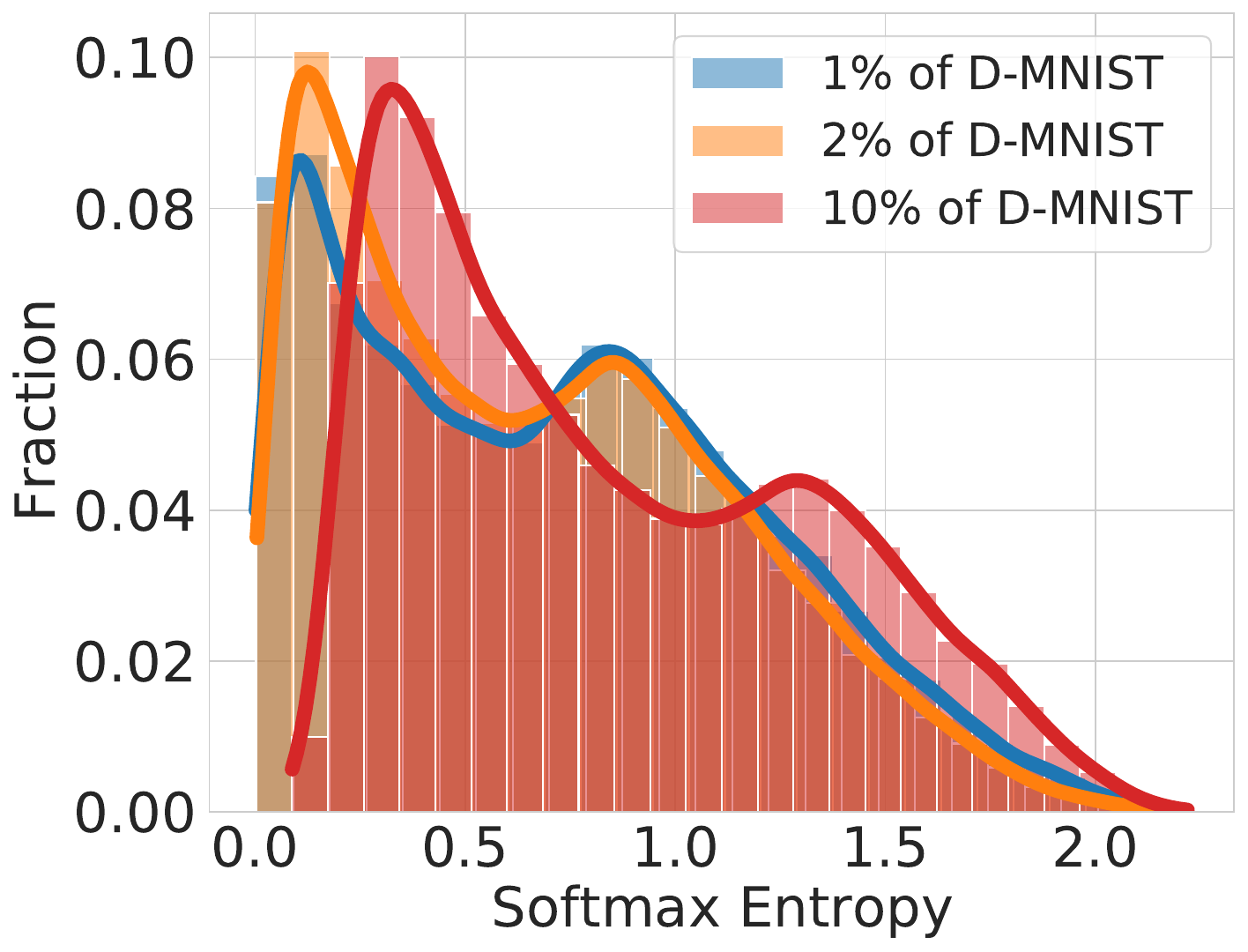}
            \label{ddu:subfig:softmax_entropy_dmnist}
        \end{subfigure}\hfill
        \begin{subfigure}[t]{0.33\linewidth}
            \centering
            \includegraphics[width=\linewidth]{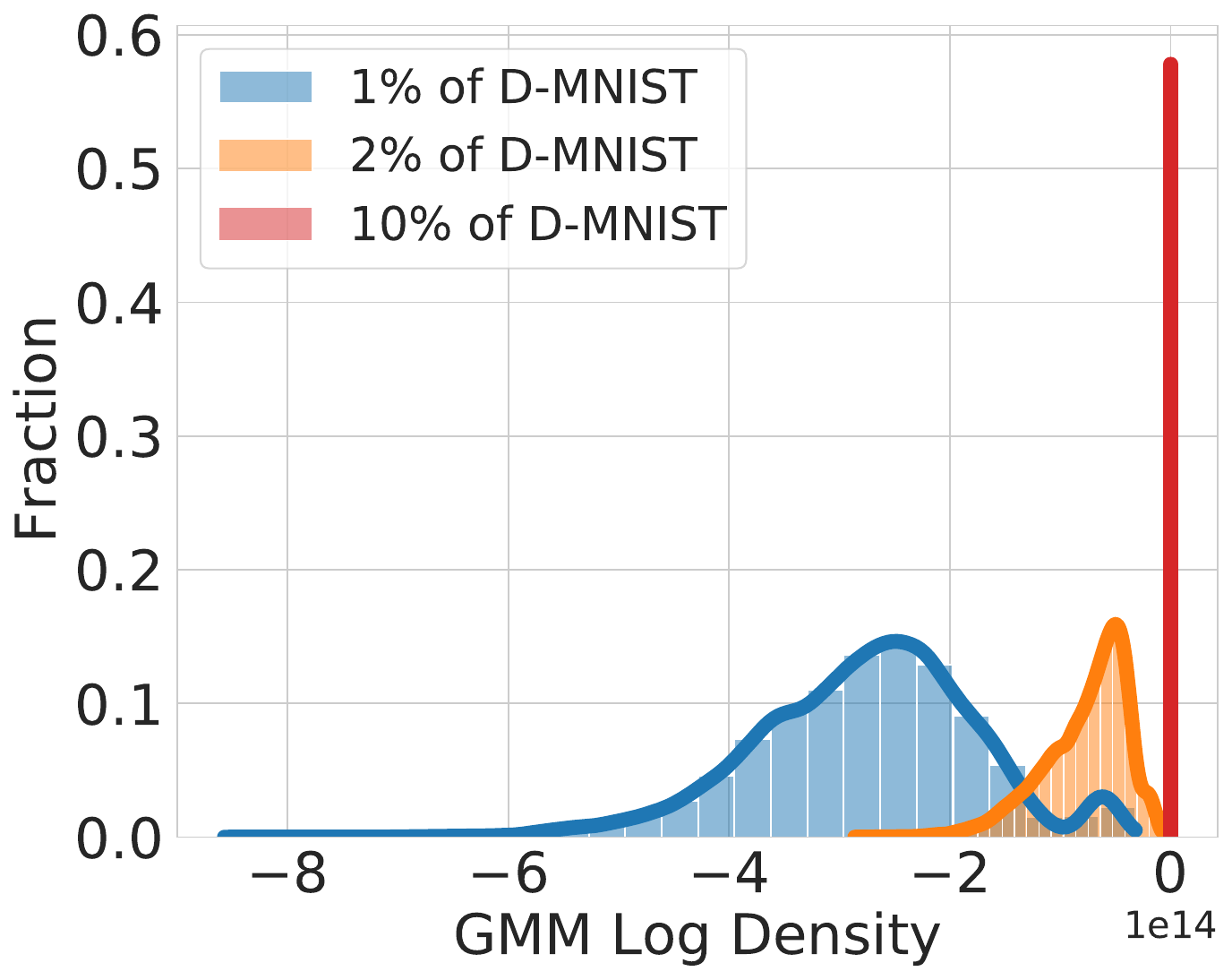}%
            \caption{Dirty-MNIST}
            \label{ddu:subfig:ddu_density_dmnist}
        \end{subfigure}\hfill
        \begin{subfigure}[t]{0.33\linewidth}
            \centering
            \includegraphics[width=\linewidth]{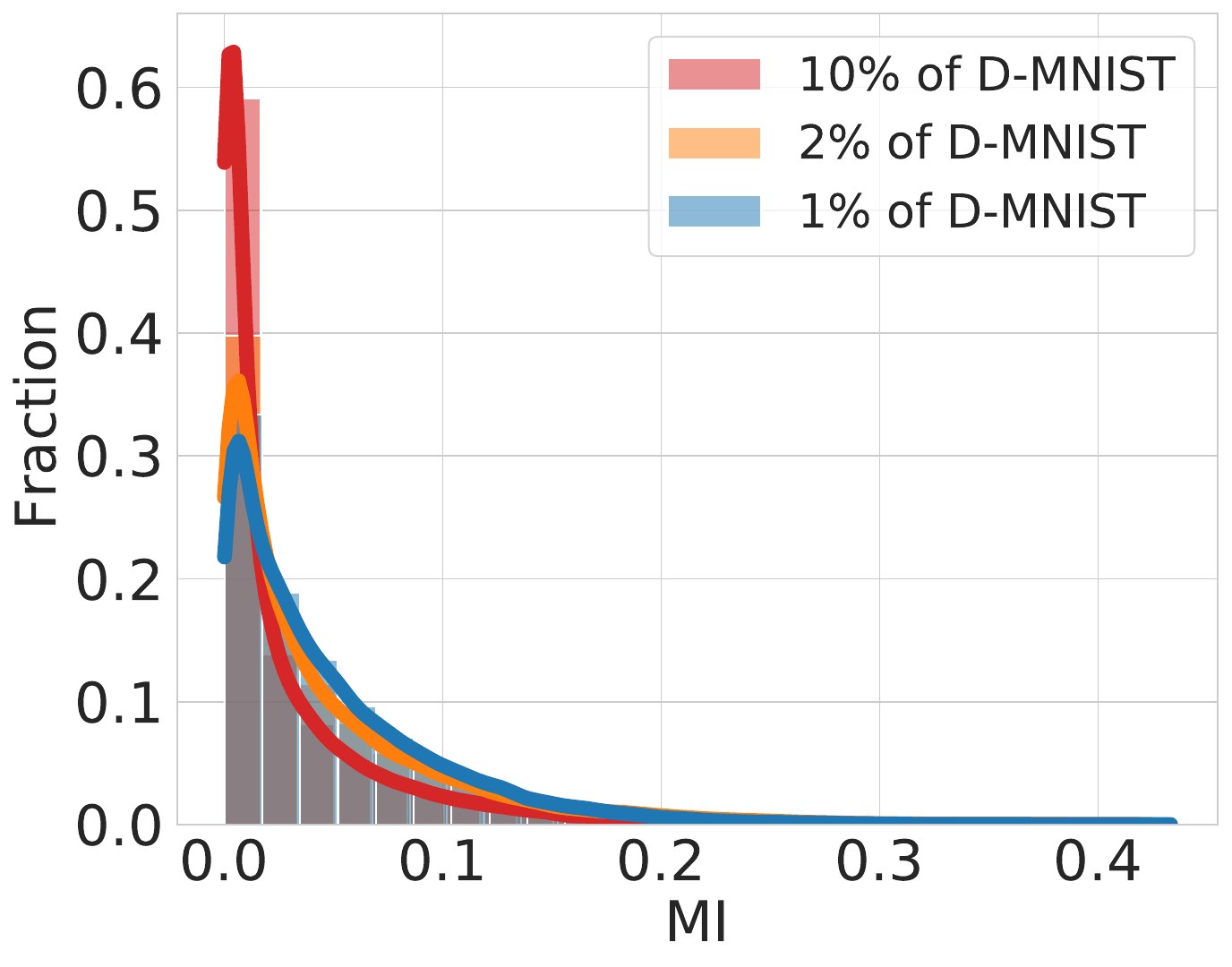}
            \label{ddu:subfig:ensemble_mi_dmnist}
        \end{subfigure}
    \end{subfigure}
    \begin{subfigure}{\linewidth}
        \centering
        \begin{subfigure}[b]{0.33\linewidth}
            \centering
            \includegraphics[width=\linewidth]{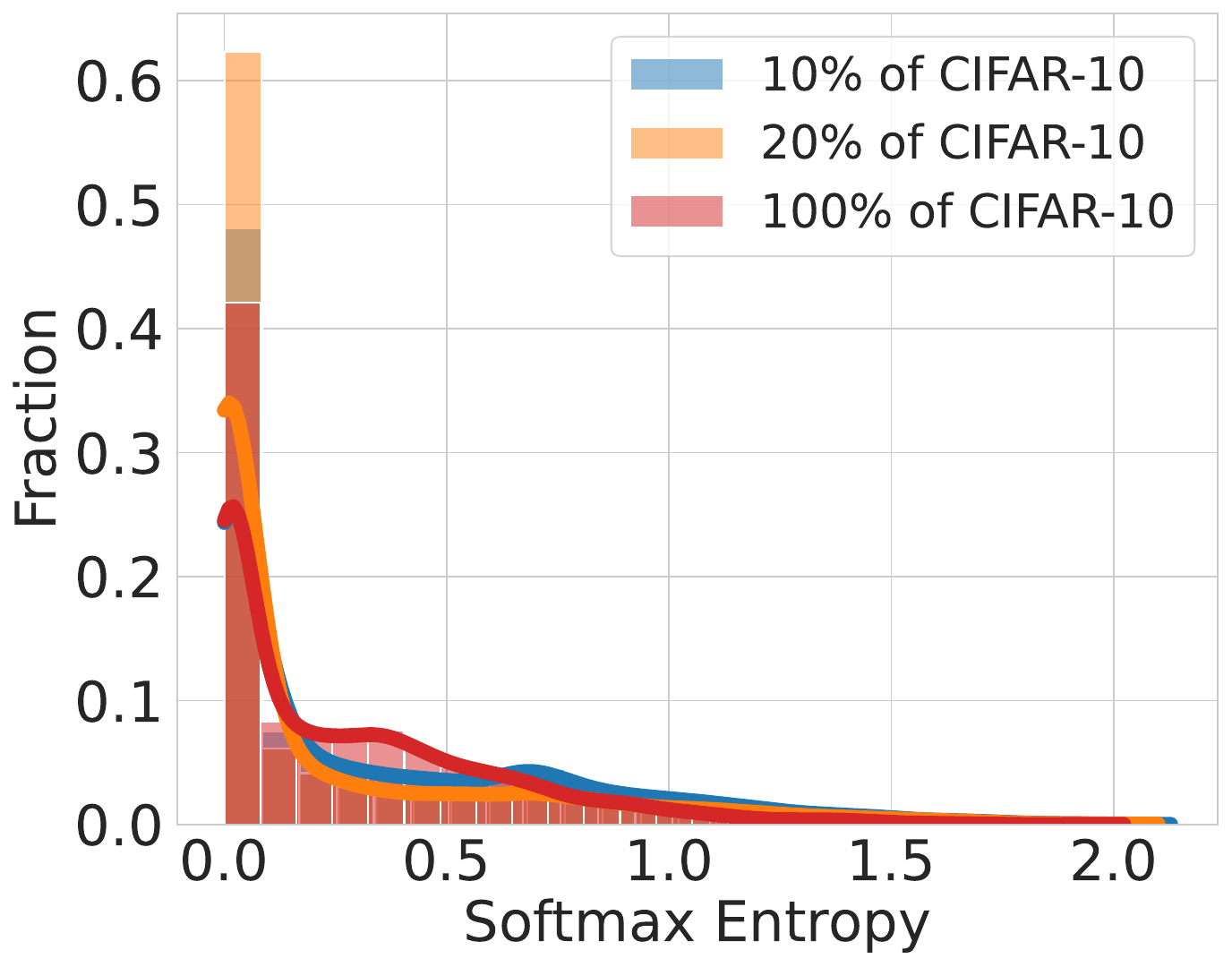}
            \label{ddu:subfig:softmax_entropy_cifar10}
        \end{subfigure}\hfill
        \begin{subfigure}[b]{0.33\linewidth}
            \centering
            \includegraphics[width=\linewidth]{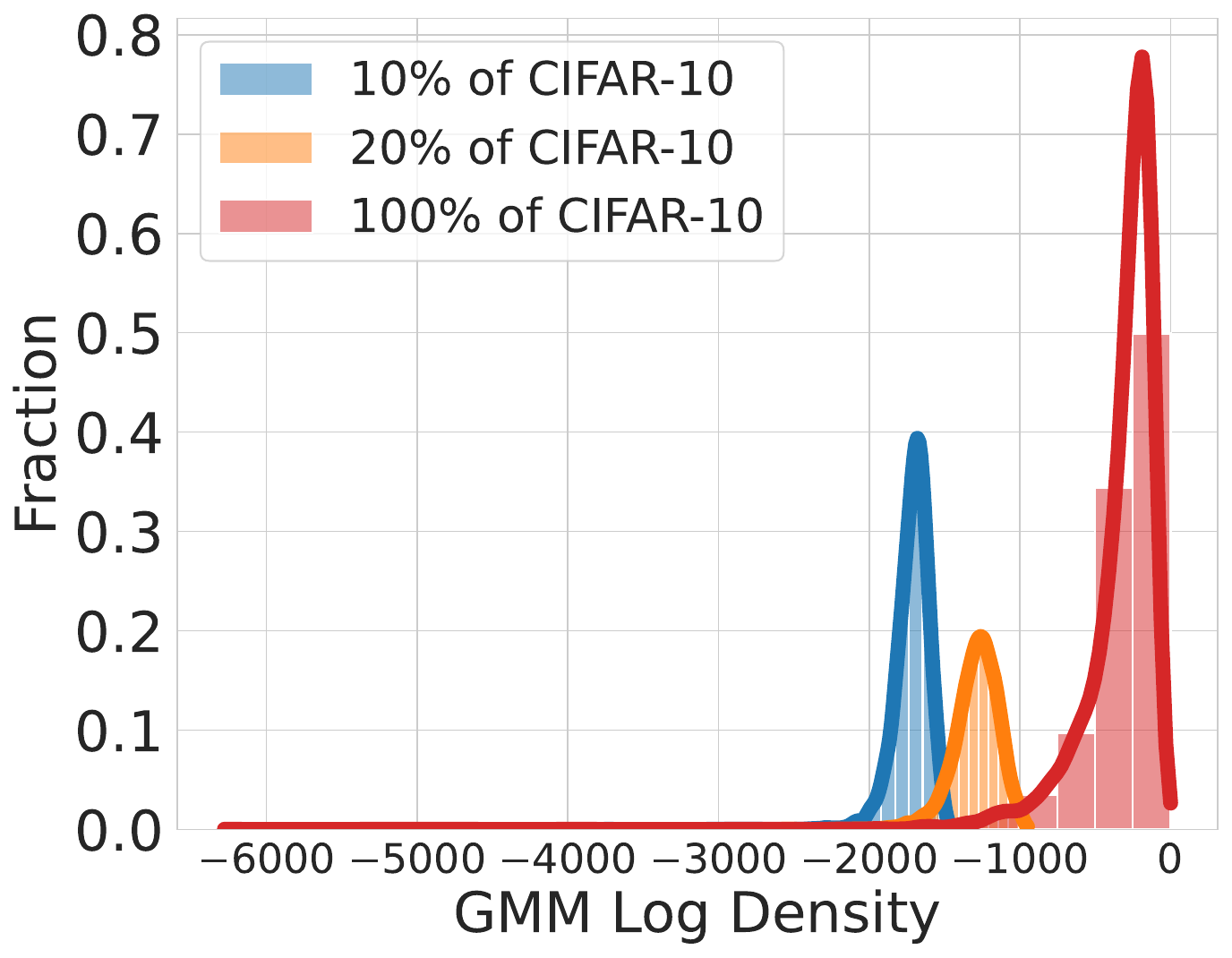}
            \caption{CIFAR-10}
            \label{ddu:subfig:ddu_density_cifar10}
        \end{subfigure}\hfill
        \begin{subfigure}[b]{0.33\linewidth}
            \centering
            \includegraphics[width=\linewidth]{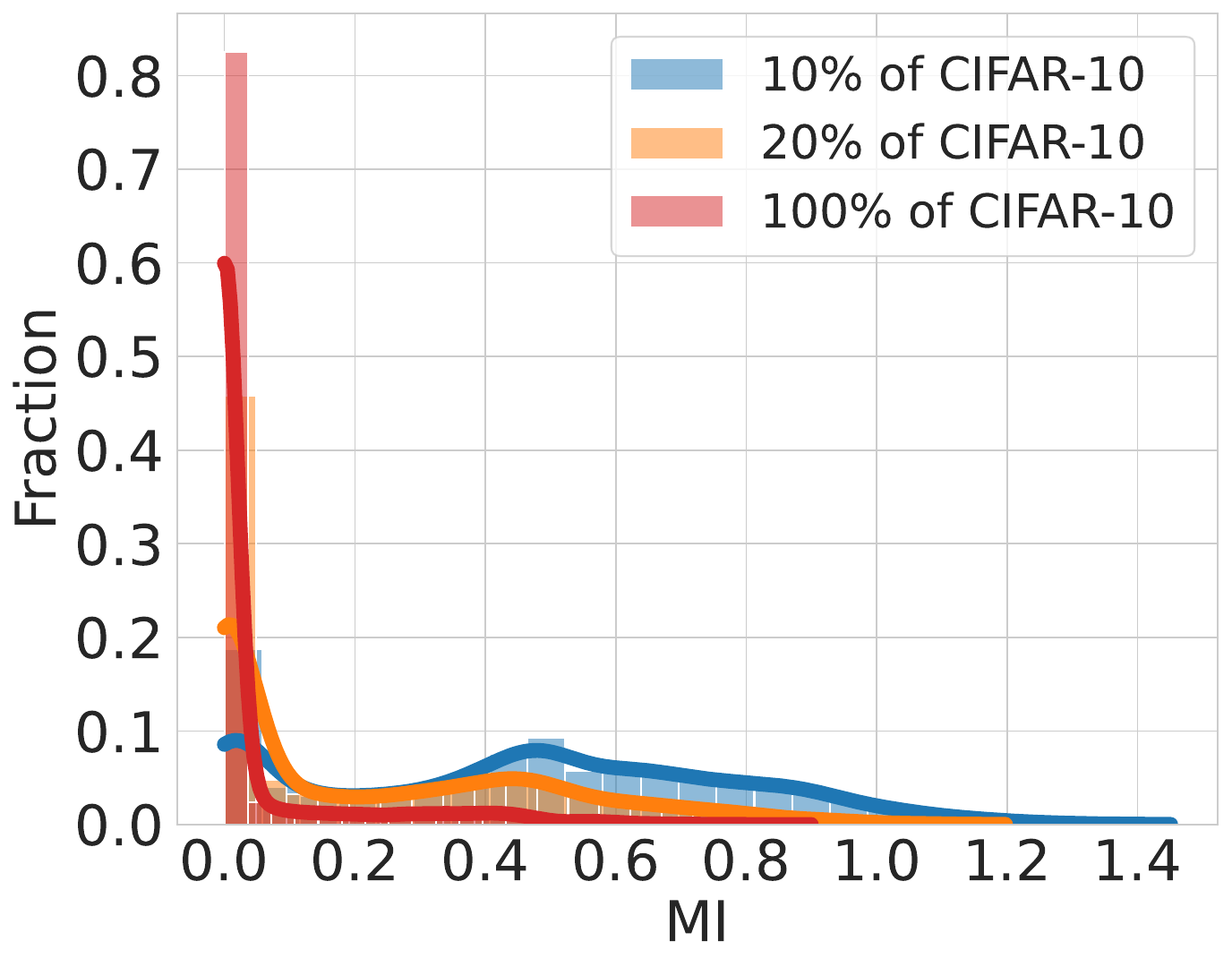}
            \label{ddu:subfig:ensemble_mi_cifar10}
        \end{subfigure}
    \end{subfigure}
    \caption{
    \emph{Comparison of epistemic and aleatoric uncertainty captured by ResNet-18+SN on increasingly large subsets of Dirty-MNIST and CIFAR-10. Feature density captures epistemic uncertainty which reduces when the model is trained on increasingly large subsets of training data, whereas softmax entropy (SE) does not. For comparison, we also plot a deep-ensemble's epistemic uncertainty, through mutual information (MI) for the same settings. For more details, see \Cref{ddu:table:kendall_viz}.}}
    \label{ddu:fig:kendall_viz_2}
\end{figure*}
\subsection{Reducible Feature-Space Density \&\ Irreducible Entropy}

\label{ddu:app:kendall_viz}

\begin{table}[t!]
    \centering
    \caption{
    \emph{
        Average softmax entropy (SE) and feature-space density of the test set for models trained on different amounts of the training set (Dirty-MNIST and CIFAR-10) behave consistently with aleatoric and epistemic uncertainty.
    }
    Aleatoric uncertainty for individual samples does not change much as more data is added to the training set while epistemic uncertainty decreases as more data is added. This is also consistent with Table 3 in \citet{kendall2017uncertainties}. Finally, we observe a consistent strong positive correlation between the negative log feature space density and the mutual information (MI) of a deep ensemble trained on the same subsets of data for both Dirty-MNIST and CIFAR-10. However, the correlation between softmax entropy and MI is not consistent.}
    \label{ddu:table:kendall_viz}
    \resizebox{\linewidth}{!}{%
    \renewcommand{\arraystretch}{1.2} 
    \begin{tabular}{cccccc}
    \toprule
    \textbf{Training Set} & \textbf{Avg Test SE ($\boldmath\approx$)} & \textbf{Avg Test Log GMM Density (\textuparrow)} & \textbf{Avg Test MI} & \textbf{Correlation(SE || MI)} & \textbf{Correlation(-Log GMM Density || MI)} \\
    \midrule
    1\% of D-MNIST & $0.7407$ & $-2.7268e+14$ & $0.0476$ & \multirow{3}{*}{$-0.79897$} & \multirow{3}{*}{$0.8132$} \\ 
    2\% of D-MNIST & $0.6580$ & $-7.8633e+13$ & $0.0447$ && \\ 
    10\% of D-MNIST & $0.8295$ & $-1279.1753$ & $0.0286$ && \\
    \midrule
    10\% of CIFAR-10 & $0.3189$ & $-1715.3516$ & $0.4573$ & \multirow{3}{*}{$0.5663$} & \multirow{3}{*}{$0.9556$} \\
    20\% of CIFAR-10 & $0.2305$ & $-1290.1726$ & $0.2247$ && \\
    100\% of CIFAR-10 & $0.2747$ & $-324.8040$ & $0.0479$ && \\
    \bottomrule
    \end{tabular}%
  }
\end{table}

To empirically verify the connection between feature-space density and epistemic uncertainty on the one hand and the connection between softmax entropy and aleatoric uncertainty on the other hand, we train ResNet-18+SN models on increasingly large subsets of Dirty-MNIST and CIFAR-10 and evaluate the epistemic and aleatoric uncertainty on the corresponding test sets using the feature-space density and softmax entropy, respectively. Moreover, we also train a 5-ensemble on the same subsets of data and use the ensemble's mutual information as a baseline measure of epistemic uncertainty.

In \Cref{ddu:fig:kendall_viz_2,ddu:table:kendall_viz}, we see that with larger training sets, the average feature-space density increases which is consistent with the epistemic uncertainty decreasing as more data is available as reducible uncertainty. This is also evident from the consistent strong positive correlation between the negative log density and mutual information of the ensemble.
On the other hand, the softmax entropy stays roughly the same which is consistent with aleatoric uncertainty as irreducible uncertainty, which is independent of the training data. Importantly, all of this is also consistent with the experiments comparing epistemic and aleatoric uncertainty on increasing training set sizes in Table 3 of \citep{kendall2017uncertainties}.

\section{Objective Mismatch}

So far, we have seen that the feature-space density of a model can be used as a proxy to quantify the epistemic uncertainty of the model, and that the softmax entropy of the model can be used as a proxy to quantify the aleatoric uncertainty of the model.

Why did we use softmax entropy to estimate aleatoric uncertainty? Why not use the predictive probability $\qof{\y \given \z}$ of the GDA model that we use to estimate the feature-space density $\pof{\z}$?
It is not a matter of convenience but rather a matter of potentially conflicting objectives:

In this section, we show that the feature-space density and softmax entropy are optimal for the respective uncertainty quantification tasks, and that this is because of an \emph{objective mismatch} between the two tasks.
The predictive probability induced by a feature-density estimator will generally not be well-calibrated as there is an objective mismatch. %
This was overlooked in previous research on uncertainty quantification for deterministic models: \citet{lee2018simple,liu2020simple,van2020uncertainty,he2016deep, postels2020quantifying}.
Specifically, a mixture model $\qof{\y, \z} = \sum_\y \qof{\z \given \y} \qof{\y}$, using one component per class, cannot be optimal for both feature-space density and predictive distribution estimation as there is an \emph{objective mismatch} \citep[Ex. 4.20, p. 145]{murphy2012machine}:
\begin{restatable}{proposition}{objectivemismatch}
\label{ddu:pro:objectivemismatch}
For an input $\x$, let $\z=f_\theta(\x)$ denote its feature representation in a feature extractor $f_\theta$ with parameters $\theta$. Then the following hold:
\begin{enumerate}[leftmargin=*]
    \item a discriminative classifier $\tqof{\y \given \z}$, e.g. a softmax layer, is well-calibrated in its predictions when it maximizes the conditional log-likelihood $\tqof{\y \given \z}$;
    \item a feature-space density estimator $\tqof{\z}$ is optimal when it maximizes the marginalized log-likelihood $\log \qof{\z}$;
    \item a mixture model $\tqof{\y, \z} = \sum_\y \tqof{\z \given \y} \qof{\y}$ might not be able to maximize both objectives, conditional log-likelihood and marginalized log-likelihood, at the same time. In the specific instance that a GMM with one component per class does maximize both, the resulting model must be a GDA (\emph{but the opposite does not hold}).
\end{enumerate}
\end{restatable}
Following the notation form \S\ref{glue:background:information_theory}, $\tqof{\cdot}$ denotes a probability distribution parameterized by $\theta$. 
As cross-entropies upper-bound the respective entropies, we have:
\begin{math}
    \tHof{Y, Z} \ge \Hof{Y, Z},
    \tHof{Z} \ge \Hof{Z}, \text{and}
    \tHof{Y \given Z} \ge \Hof{Y \given Z}.
\end{math}
\newcommand{\trivialminimizer}{\fpcof{^*}{y, z}}
\begin{proof}
We prove the statements in order. The first two are trivial.
\begin{enumerate}[wide, labelwidth=!, labelindent=0pt]
    \item 
    The conditional log-likelihood is a strictly proper scoring rule \citep{gneiting2007strictly}. The optimization objective can be rewritten as
    \begin{align}
        \label{ddu:eq:discriminative_classifier}
        \max_\theta \implicitE{\log \tpof{y \given z}} = -\min_\theta \tHof{Y \given Z} \le -\Hof{Y \given Z}.
    \end{align}
    An optimal discriminative classifier $\tqof{\y \given \z}$ with equality above would thus capture the true (empirical) distribution everywhere: $\tqof{\y \given \z} = \pdataof{\y \given \z}$. %
    \item
    For density estimation $\tqof{\z}$, we maximize the log-likelihood $\E{}{\log \tqof{\z}}$ using the empirical data distribution. We can rewrite this as
    \begin{align}
        \label{ddu:eq:density_estimator}
        \max_\theta \simpleE{\pdataof{y,z}}{\log \tpof{z}} = -\min_\theta \tHof{Z} \le -\Hof{Z}.
    \end{align}
    When we have equality, we have $\tpof{\z} = \pdataof{\z}$.
    \item
    Using $\tHof{Y, Z} = \tHof{Y \given Z} + \tHof{Z}$, we can relate the objectives from \Cref{ddu:eq:discriminative_classifier} and \eqref{ddu:eq:density_estimator} to each other. 
    First, we characterize a shared optimum, and then we show that both objectives are generally not minimized at the same time.
    For both objectives to be minimized, we have $\nabla \tHof{Y \given Z} = 0$ and $\nabla \tHof{Z} = 0$, and we obtain
    \begin{align}
        \nabla \tHof{Y, Z} = \nabla \tHof{Y \given Z} + \nabla \tHof{Z} = 0.
    \end{align}
    From this we conclude that minimizing both objectives also minimizes $\tHof{Y, Z}$, and that generally the objectives trade-off with each other at stationary points $\theta$ of $\tHof{Y, Z}$:
    \begin{align}
        \label{ddu:eq:classifiers_trade_off}
        \nabla \tHof{Y \given Z} = -\nabla \tHof{Z} \quad \text{when $\nabla \tHof{Y, Z} = 0$}.
    \end{align}
    This tells us that to construct a case where the optima do not coincide, discriminative classification needs to be opposed better density estimation. 
\end{enumerate}
\end{proof}

\paragraph{Existence.}
Let us also show that the cases described above can occur.
Specifically, when we have a GMM with one component per class, minimizing $\tHof{Y, Z}$ on an empirical data distribution is equivalent to Gaussian Discriminant Analysis, as is easy to check, and minimizing $\tHof{Z}$ is equivalent to fitting a density estimator, following \Cref{ddu:eq:density_estimator}.
The difference is that using a GMM as a density estimator does not constrain the component assignment for a sample, unlike in GDA.
Overall, we see that \emph{both objectives can be minimized at the same time exactly when the feature representations of different classes are perfectly separated}, such that a GMM fit as density estimator would assign each class's feature representations to a single component.

By the above, we can construct simple examples for both cases: if the samples of different classes are not separated in feature-space, optima for the objectives will not coincide. For example, if samples were drawn from the same Gaussian and labeled randomly. On the other hand, when the features of different classes all lie in well-separated clusters, GDA can minimize all objectives at the same time.

Given that perfect separation is impossible with ambiguous data for a GMM, a shared optimum will be rare with noisy real-world data, but only then would GDA be optimal.
In all other cases, GDA does not optimize both objectives, and neither can any other GMM with one component per class.

\Cref{ddu:eq:classifiers_trade_off} tells us that a GMM fit using Expectation Maximization is likely a better density estimator than GDA, and a softmax layer is a better classifier, as optimizing the density objective $\Hof{Z}$ or softmax objective $\tHof{Y \given Z}$ using gradient descent will move away from the  optimum of the GDA objective

\subsection{Toy Example} %
\label{ddu:app:objective_mismatch_toy_example}

\begin{figure}[t!]
    \begin{subfigure}{\linewidth}
        \centering
        \includegraphics[width=\linewidth]{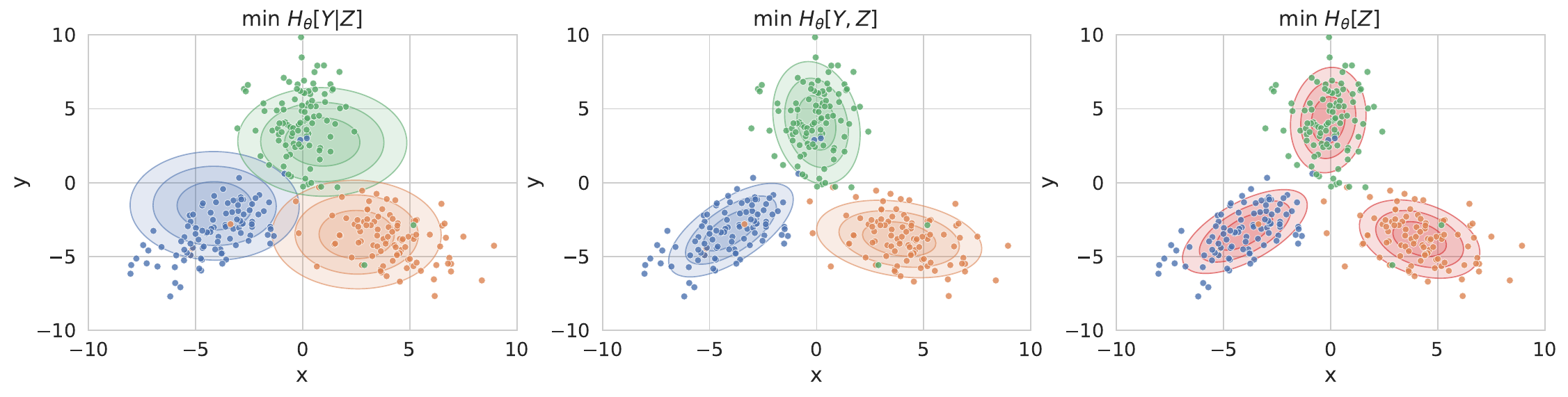}%
        \caption{
        \emph{Density.} Contours at 68.26\%, 95.44\%, and 99.7\%.}
        \label{ddu:fig:objective_mismatch_density}
    \end{subfigure}
    \hfill
    \begin{subfigure}{\linewidth}
        \centering
        \includegraphics[width=\linewidth]{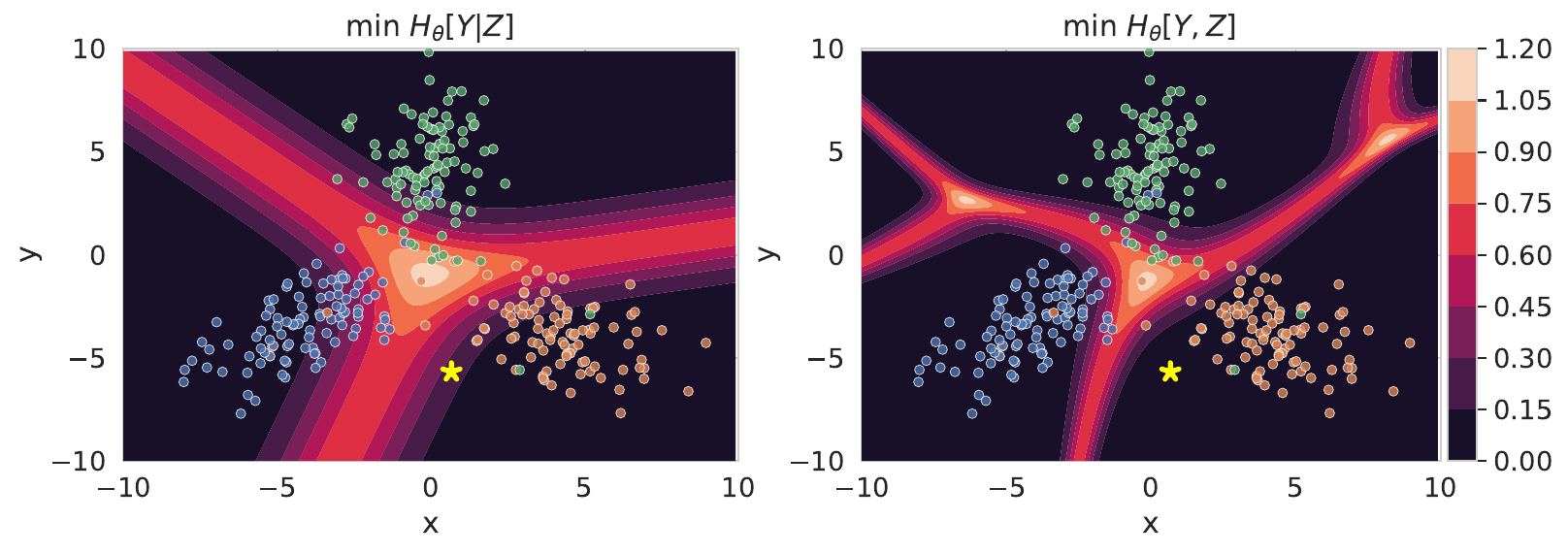}%
        \caption{\emph{Entropy.} Darker is lower.}
        \label{ddu:fig:objective_mismatch_entropy}
    \end{subfigure}
    \caption{\emph{3-component GMM fitted to a synthetic dataset with 3 different classes (differently colored) with 4\% label noise using different objectives}.
    \textbf{\subref{ddu:fig:objective_mismatch_density}:} The optima for conditional log-likelihood $\tHof{Y \given Z}$, joint log-likelihood $\tHof{Y, Z}$, and marginalized log-likelihood $\tHof{Z}$ all differ. Hence, the best calibrated model ($\tHof{Y \given Z}$) will not provide the best density estimate ($\tHof{Z}$), and vice-versa. 
    \textbf{\subref{ddu:fig:objective_mismatch_entropy}:} A mixture model that optimizes $\tHof{Y, Z}$ (GDA) does not have calibrated decision boundaries for aleatoric uncertainty: the ambiguous sample (due to label noise) marked by the yellow star has no aleatoric uncertainty under the GDA model. See \S\ref{ddu:app:objective_mismatch_toy_example} for details.
    }
    \label{ddu:fig:objective_mismatch}
\end{figure}

\begin{table}[t!]
    \centering
    \caption{\emph{Realized objective scores (columns) for different optimization objectives (rows) for the synthetic 2D toy example depicted in \Cref{ddu:fig:objective_mismatch}.} Smaller is better. We see that each objective minimizes its own score while being suboptimal in regard to the other two objectives (when it is possible to compute the scores). This empirically further validates \Cref{ddu:pro:objectivemismatch}.
    }
    \label{ddu:table:objective_mismatch}
    \scriptsize
    \begin{tabular}{lrrr}
        \toprule
        Loss $\rightarrow$ & {\boldmath$\tHof{Y \given Z}$} (\textdownarrow) & {\boldmath$\tHof{Y, Z}$} (\textdownarrow) & {\boldmath$\tHof{Z}$} (\textdownarrow) \\ 
        {Objective $\downarrow$} & & & \\
        \midrule
        {\boldmath$\min \tHof{Y \given Z}$} & \textbf{0.1794} & 5.4924 & 5.2995 \\ 
        {\boldmath$\min \tHof{Y, Z}$} & 0.2165 & \textbf{4.9744} & 4.7580 \\ 
        {\boldmath$\min \tHof{Z}$} & n/a & n/a & \textbf{4.7073} \\
        \bottomrule
    \end{tabular}%
\end{table}

To explain \Cref{ddu:pro:objectivemismatch} in an intuitive way, we focus on a simple synthetic 2D dataset with three classes and 4\% label noise and fit GMMs using the different objectives. To construct, the dataset we sample ``latents'' $\z$ from three different Gaussians (each representing a different class $\y$) with 4\% label noise. Following the construction in the proof, this will lead the objectives to have different optima. Note that label noise and non-separability of features are common issues in real-world datasets.

In \Cref{ddu:table:objective_mismatch} and \Cref{ddu:fig:objective_mismatch_density}, we see that each solution minimizes its own objective best. The regular GMM (which optimizes the density) provides the best density model (best fit according to the entropy), while the LDA (like a softmax linear layer) provides the best NLL for the labels. The GDA provides a density model that is almost as good as the GMM.
Let us discuss the different objectives in \Cref{ddu:fig:objective_mismatch} and the resulting scores in more detail:
\begin{description}
\item[\boldmath${\min \tHof{Y \given Z}}$.] A softmax linear layer is equivalent to an LDA (Linear Discriminant Analysis) with conditional likelihood as detailed in \citet{murphy2012machine}. We optimize an LDA with the usual objective "$\min -1/N \sum \log \qof{\y \given \z}$", i.e.\ the cross-entropy of $\qof{\y \given \z}$ or (average) negative log-likelihood (NLL). 
Because we optimize only $\qof{\y \given \z}$, $\qof{\z}$ does not affect the objective and is thus not optimized. Indeed, the components do not actually cover the latents well, as can be seen in the first density plot of \Cref{ddu:fig:objective_mismatch_density}. However, it does provide the lowest NLL.

\item[\boldmath$\min \tHof{Y, Z}$.] We optimize a GDA for the combined objective "$\min -1/N \sum \log \qof{\y, \z}$", i.e. the cross-entropy of $\qof{\y, \z}$. We use the shorthand "$\min \tHof{Y \given Z}$" for this.

\item[\boldmath$\min \tHof{Z}$.] We optimize a GMM for the objective "$\min -1/N \sum \log \qof{\z}$", i.e. the cross-entropy of $\qof{\z}$. We use the shorthand "$\min \tHof{Z}$" for this.
Scores for $\tHof{Y \given Z}$ and $\tHof{Y,Z}$ for the third objective $\min \tHof{Z}$ are not provided in \Cref{ddu:table:objective_mismatch} as it does not depend on $Y$, and hence the different components do not actually model the different classes. Hence, we also use a single color to visualize the components for this objective in \Cref{ddu:fig:objective_mismatch_density}.
\end{description}

\paragraph{Entropy Plot.} 
Looking at the entropy plots in \Cref{ddu:fig:objective_mismatch_entropy}, we first notice that the LDA solution optimized for $\min \tHof{Y \given Z}$ has a wide decision boundary. This is due to the overlap of the Gaussian components, which is necessary to provide the right aleatoric uncertainty.

Optimizing the negative log-likelihood $-\log \pof{\y \given \z}$ is a proper scoring rule, and hence is optimized for calibrated predictions.

Compared to this, the GDA solution (optimized for $\min \tHof{Y, Z}$) has a much narrower decision boundary and cannot capture aleatoric uncertainty as well. This is reflected in the higher NLL. Moreover, unlike for LDA, GDA decision boundaries behave differently than one would naively expect due to the untied covariance matrices. They can be curved, and the decisions change far away from the data \citep{murphy2012machine}.

To show the difference between the two objectives we have marked an ambiguous point near $(0, -5)$ with  a yellow star. Under the first objective $\min \tHof{Y, Z}$, the point has high aleatoric uncertainty (high entropy), as seen in the left entropy plot while under the second objective ($\min \tHof{Y, Z}$) the point is only assigned very low entropy. The GDA optimized for the second objective thus is overconfident.

As explained above, we do not show an entropy plot of $Y \mid Z$ for the third objective $\min \tHof{Z}$ in \Cref{ddu:fig:objective_mismatch_entropy} because the objective does not depend on $Y$, and there are thus no class predictions.

Intuitively, for aleatoric uncertainty, the Gaussian components need to overlap to express high aleatoric uncertainty (uncertain labelling). At the same time, this necessarily provides looser density estimates. On the other hand, the GDA density is much tighter, but this comes at the cost of NLL for classification because it cannot express aleatoric uncertainty that well. \Cref{ddu:fig:objective_mismatch} visualizes how the objectives trade-off between each other, and why we use the softmax layer trained for $\pof{\y \given \z}$ for classification and aleatoric uncertainty, and GDA as density model for $\qof{\z}$.

\subsection{Discussion}
We have shown that the objectives for a GMM with one component per class are obviously not equivalent, and that the optima of these objectives do not need to coincide.
Hence, importantly, the above statement tells us that we can expect better performance by \emph{using both a discriminative classifier (e.g., softmax layer) to capture aleatoric uncertainty for iD samples and a separate feature-density estimator to capture epistemic uncertainty even on a model trained using conditional log-likelihood, i.e.\ the usual cross-entropy objective}.
As noted in the previous section \Cref{ddu:sec:motivation}, we focus on the GDA objective instead of the GMM objective as it is easier to compute, even though it also suffers from an objective mismatch. However, both in our toy example (\Cref{ddu:table:objective_mismatch}), where the difference between the GDA objective to the softmax objective is larger than the difference to the GMM objective for the relevant metrics, and in our experiments (\Cref{ddu:table:objective_mismatch_real}), we found that the GDA objective is sufficient for good performance.

\section{Deep Deterministic Uncertainty}
\label{ddu:section:methods}
\label{ddu:app:algorithm}

Based on the previous sections, we propose the following method:
\begin{mainresult}
\emph{Deep Deterministic Uncertainty (DDU)} is a simple baseline method for uncertainty quantification for deterministic neural networks. It uses
\emph{a deterministic neural network with an appropriately regularized feature-space, using spectral normalization \citep{liu2020simple}}, which
can disentangle aleatoric and epistemic uncertainty. It estimates:
\begin{enumerate}
    \item aleatoric uncertainty using softmax entropy, and
    \item epistemic uncertainty by fitting a GDA after training.
\end{enumerate}
There is no need for any additional precessing steps: no hold-out ``OoD'' data, feature ensembling, or input pre-processing, unlike in \citet{lee2018simple}.
\end{mainresult}

\begin{algorithm}[t!]
    \caption{Deep Deterministic Uncertainty}
    \label{ddu:algorithm:fancy}
    \includegraphics{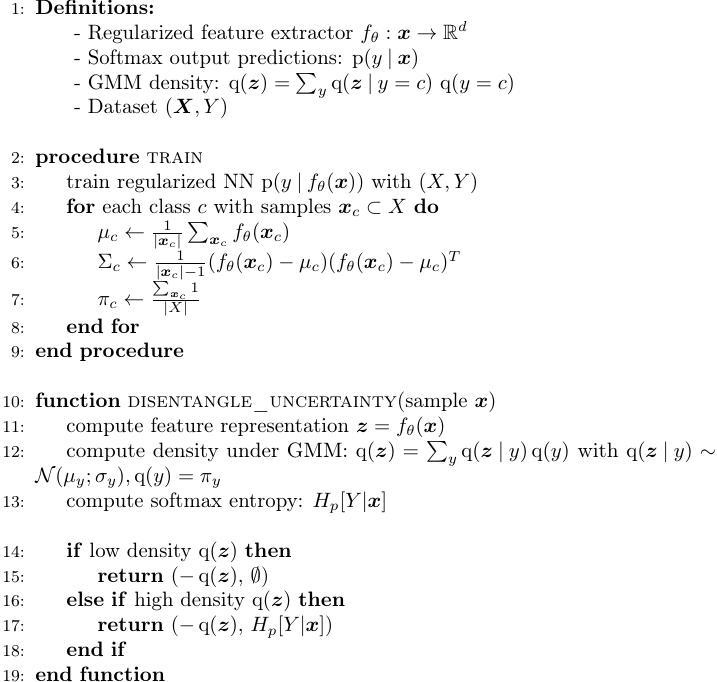}
\end{algorithm}

\paragraph{Ensuring Sensitivity \&\ Smoothness.} %
We ensure sensitivity and smoothness using spectral normalization in models with residual connections. In addition, we make minor changes to the standard residual block to further encourage sensitivity 
without sacrificing accuracy (see details in \S\ref{ddu:app:more_model_architecture}).

\paragraph{Disentangling Epistemic \&\ Aleatoric Uncertainty.} %
To quantify epistemic uncertainty, we fit a feature-space density estimator after training. We use GDA, a GMM $\qof{\y, \z}$ with a single Gaussian component per class, and fit each class component by computing the empirical mean and covariance, per class, of the feature vectors $z=f_\theta(\x)$, which are the outputs of the last convolutional layer of the model computed on the training samples $\x$. \emph{Note that we do not require OoD data to fit these and unlike \citet{lee2018simple} we use a separate covariance matrix for each class.}
Fitting a GDA on the feature space, thus requires no further training and only requires a single forward-pass through the training set.

\paragraph{Evaluation.} At test time, we estimate the epistemic uncertainty by evaluating the marginal likelihood of the feature representation under our density $\qof{\z} = \sum_\y \qof{\z \given \y} \qof{\y}$.
To quantify aleatoric uncertainty for in-distribution samples, we use the entropy $\Hof{Y \given \x, \theta}$ of the softmax distribution $\pof{\y \given \x, \theta}$. %
Note that the softmax distribution thus obtained can be further calibrated using temperature scaling \citep{guo2017calibration}.
Thus, for a given input, a high feature-space density indicates low epistemic uncertainty (iD), at which point, we can trust the aleatoric estimate from the softmax entropy.
The sample can then be either unambiguous (low softmax entropy) or ambiguous (high softmax entropy).
Conversely, a low feature density %
indicates high epistemic uncertainty (OoD), and we cannot trust softmax predictions. The algorithm is depicted in
\Cref{ddu:algorithm:fancy}.

\subsubsection{Computational Complexity}
\label{ddu:app:complexity}

\newcommand{\landauO}{\mathcal{O}}

Let $N$ be the number of samples; $D$, the feature space dimensionality; and $C$, the number of classes; with $\approx N/C$ samples per class (balanced). For fitting the GMM via GDA: computing the covariance matrix per class requires $\landauO(C(N/C)D^2)$ = $\landauO(ND^2)$ complexity. Computing the inverse and determinant of the covariance matrices via the Cholesky decomposition requires $\landauO(D^3)$ per class. Thus, the total computational cost for GDA is $\landauO(ND^2 + CD^3)$. Evaluating density of a single point: distance from class means requires $\landauO(CD)$, and matrix vector multiplications requires $\landauO(CD^2)$. Hence, the total cost for evaluating density on a single point is $\landauO(CD^2).$

\subsection{Implementation}
\label{ddu:app:implementation}

\begin{lstlisting}[float=tp, language=Python, caption=Deep Deterministic Uncertainty Pseudo-Code, label=ddu:fig:algorithm, mathescape=true]
    # instantiate models
    model = create_sensitive_smooth_model()
    gda = create_gda()
    
    # train
    training_samples, training_labels = load_training_set()
    model.fit(training_samples, training_labels)
    
    training_features = model.features(training_samples)
    gda.fit(training_features, training_labels)
    
    # test
    test_features = model.features(test_sample)
    
    epistemic_uncertainty = -gda.log_density(test_features)
    
    is_ood = epistemic_uncertainty <= ood_threshold
    if not is_ood:
      predictions = model.softmax_layer(test_features)
      aleatoric_uncertainty = entropy(predictions)
      return epistemic_uncertainty, aleatoric_uncertainty
     
    # aleatoric uncertainty is only valid for iD
    return epistemic_uncertainty, None
\end{lstlisting}

Here, we describe our reference implementation in more detail.
A simple Python pseudocode using a `scikit-learn'-like API \citep{sklearn_api} is shown in \Cref{ddu:fig:algorithm}.
Note that in order to compute thresholds for low and high density or entropy, we simply use the training set containing iD data. We consider all points having density lower than the $99\%$ quantile as OoD. 

\paragraph{Increasing sensitivity.}
\label{ddu:app:more_model_architecture}

Using residual connections to enforce sensitivity works well in practice when the layer is defined as $\x' = \x + f(\x)$.
However, there are several places in the network where additional spatial downsampling is done in $f(\cdot)$ (through a strided convolution), and in order to compute the residual operation $\x$ needs to be downsampled as well.
These downsampling operations are crucial for managing memory consumption and generalization.
The way this is traditionally done in ResNets is by introducing an additional function $g(\cdot)$ on the residual branch (obtaining $\x' = g(\x) + f(\x)$) which is a strided 1x1 convolution.
In practice, the stride is set to 2 pixels, which leads to the output of $g(\cdot)$ only being dependent on the top-left pixel of each 2x2 patch, which reduces sensitivity.
We overcome this issue by making an architectural change that improves uncertainty quality without sacrificing accuracy.
We use a strided average pooling operation instead of a 1x1 convolution in $g(\cdot)$. This makes the output of $g(\cdot)$ dependent on all input pixels.
Additionally, we use leaky ReLU activation functions, which are equivalent to ReLU activations when the input is larger than 0, but below 0 they compute $p * \x$ with $p = 0.01$ in practice.
These further improve sensitivity as all negative activations still propagate in the network.

\section{Empirical Validation}
\label{ddu:sec:experiments}

We evaluate DDU on active learning and OoD detection tasks:
\begin{description}
    \item[Visualizations.]
We detail toy experiments on how DDU can disentangle epistemic and aleatoric uncertainty and the effect of feature-space regularization in, depicted in \Cref{ddu:fig:intro_histograms}, and on the well-known Two Moons toy dataset in \S\ref{ddu:app:2_moons}.

\item[Active Learning.]
For active learning \citep{cohn1996active}, we evaluate DDU using MNIST, CIFAR-10 and an ambiguous version of MNIST (Dirty-MNIST).

\item[OoD Detection.]
Understanding the caveats detailed in \S\ref{ddu:sec:active_learning_vs_ood_detection}, we can use OoD detection to evaluate using DDU for estimating epistemic uncertainty and aleatoric uncertainty such that it will be meaningful for active learning as well: we will focus on `near OoD' datasets such that good OoD detection performance is a good proxy for good epistemic uncertainty performance, and we do not use `far OoD' datasets as they would not be informative for active learning performance. 

Thus, we evaluate DDU's quality of epistemic uncertainty estimation on several OoD detection settings for:
\begin{itemize}
    \item image classification on CIFAR-10 vs SVHN/CIFAR-100/Tiny-ImageNet/CIFAR-10-C, CIFAR-100 vs SVHN/Tiny-ImageNet and ImageNet vs ImageNet-O dataset pairings, where we outperform other deterministic single-forward-pass methods and perform on par with deep ensembles;
    \item semantic segmentation on Pascal VOC, comparing with a deterministic model, MC Dropout (MCDO) \cite{gal2015dropout} and deep ensembles;
    \item on the real-world QUBIQ challenge in \S\ref{ddu:app:qubiq} 
\end{itemize} 

\item[Ablations.] We ablate feature space density on different model architectures, compare GDA and LDA, examine the effect of the objective mismatch on CIFAR-10, and provide results on additional baselines, with additional ablations in \Cref{ddu:app:additional_exp_results}.
\end{description}
While the focus of this thesis is on data subset selection, the bulk of the experiments for DDU is on OoD detection as these experiments are easier to run and ablate than active learning experiments.

\subsection{Visualizations}
Two toy experiments illustrate the effect of feature-space regularization on the quality of epistemic and aleatoric uncertainty estimation.
The first experiment is on a simple 2D toy dataset, and the second experiment is on the MNIST dataset.

\subsubsection{Two Moons}
\label{ddu:app:2_moons}

\begin{figure*}[!t]
    \centering
    \begin{subfigure}[b]{0.5\linewidth}
        \centering
        \includegraphics[width=\linewidth]{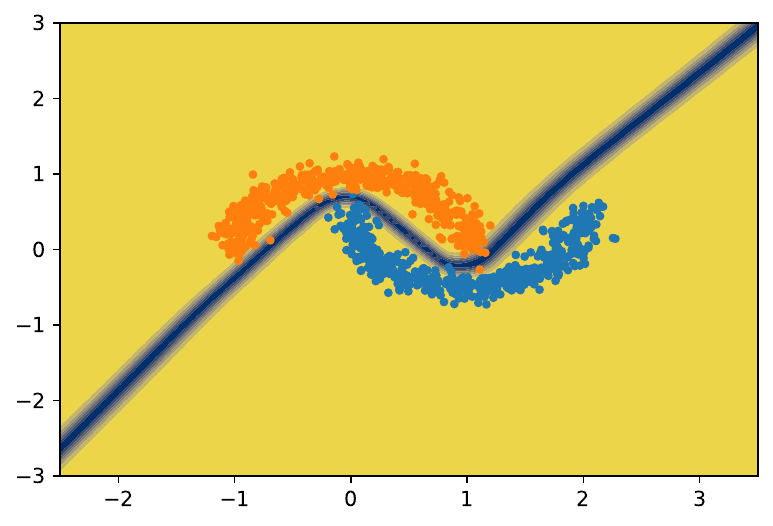}
        \caption{Softmax Entropy}
        \label{ddu:subfig:two_moons_softmax_entropy_no_sn}
    \end{subfigure}\hfill
    \begin{subfigure}[b]{0.5\linewidth}
        \centering
        \includegraphics[width=\linewidth]{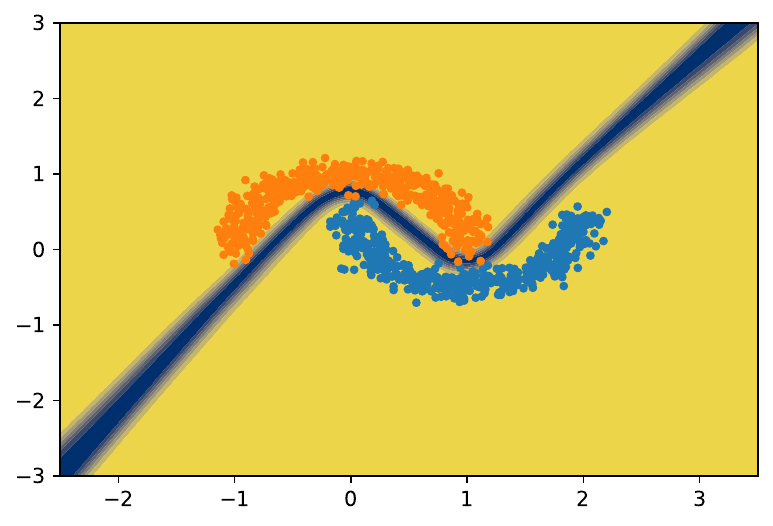}
        \caption{Ensemble Predictive Entropy}
        \label{ddu:subfig:two_moons_ensemble_predictive_entropy_no_sn}
    \end{subfigure}
    \begin{subfigure}[b]{0.5\linewidth}
        \centering
        \includegraphics[width=\linewidth]{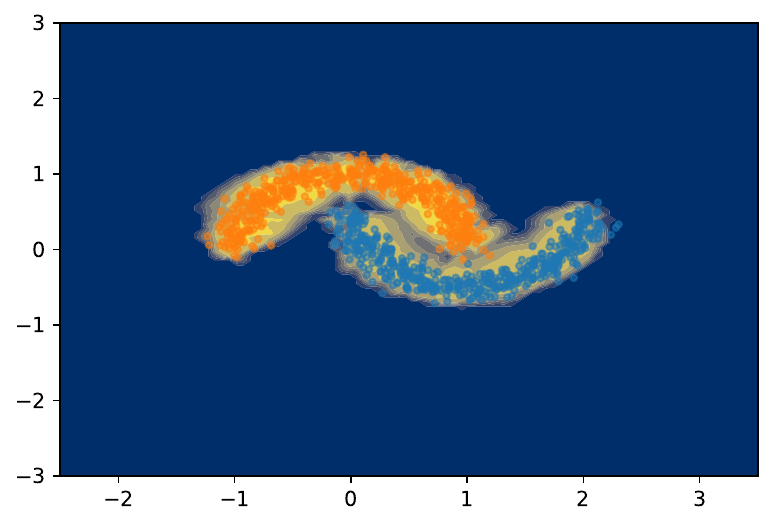}
        \caption{DDU Feature-space density}
        \label{ddu:subfig:two_moons_ddu_sn}
    \end{subfigure}\hfill
    \begin{subfigure}[b]{0.5\linewidth}
        \centering
        \includegraphics[width=\linewidth]{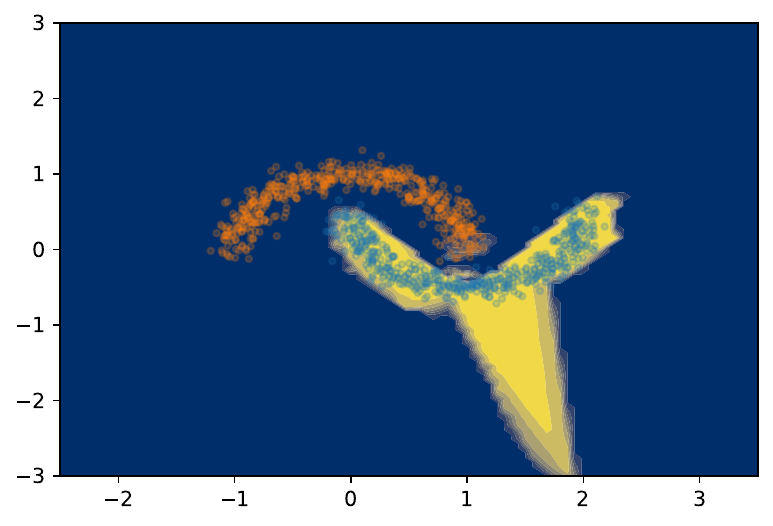}
        \caption{FC-Net Feature-space density}
        \label{ddu:subfig:two_moons_fcnet}
    \end{subfigure}
    \caption{
    \emph{Uncertainty on Two Moons dataset}. Blue indicates high uncertainty and yellow indicates low uncertainty. Both the softmax entropy of a single model and the predictive entropy of a deep ensemble are uncertain only along the decision boundary whereas the feature-space density of DDU is uncertain everywhere except on the data distribution (the ideal behavior). However, the feature density of a normal fully connected network (FC-Net) without any inductive biases can't capture uncertainty properly.
    }
    \label{ddu:fig:2-moons}
\end{figure*}

In this section, we evaluate DDU's performance on a well-known toy setup: the Two Moons dataset. We use scikit-learn's \emph{datasets} package to generate 2000 samples with a noise rate of 0.1. We use a 4-layer fully connected architecture, ResFFN-4-128 with 128 neurons in each layer and a residual connection, following \citep{liu2020simple}. As an ablation, we also train using a 4-layer fully connected architecture with 128 neurons in each layer, but \emph{without the residual connection}. We name this architecture FC-Net. The input is 2-dimensional and is projected into the 128 dimensional space using a fully connected layer. Using the ResFFN-4-128 architecture we train 3 baselines:
\begin{description}
    \item [Softmax.] We train a single softmax model and use the softmax entropy as the uncertainty metric.
    \item [3-Ensemble.] We train an ensemble of 3 softmax models and use the predictive entropy of the ensemble as the measure of uncertainty.
    \item [DDU.] We train a single softmax model applying spectral normalization on the fully connected layers and using the feature density as the measure of model confidence.
\end{description}

Each model is trained using the Adam optimizer for 150 epochs. In \Cref{ddu:fig:2-moons}, we show the uncertainty results for all the above 3 baselines. It is clear that both the softmax entropy and the predictive entropy of the ensemble is uncertain only along the decision boundary between the two classes whereas DDU is confident only on the data distribution and is not confident anywhere else. It is worth mentioning that even DUQ and SNGP perform well in this setup and deep ensembles have been known to underperform in the Two-Moons setup primarily due to the simplicity of the dataset causing all the ensemble components to generalize in the same way. Finally, also note that the feature space density of FC-Net without residual connections is not able to capture uncertainty well (\Cref{ddu:subfig:two_moons_fcnet}), thereby reaffirming our claim that proper inductive biases are indeed a necessary component to ensure that feature space density captures uncertainty reliably. %

\subsubsection{Motivational Example in \Cref{ddu:fig:intro_histograms}}

\begin{table}[t]
    \centering
    \caption{\emph{ECE for Dirty-MNIST test set and AUROC for Dirty-MNIST vs FashionMNIST as proxies for aleatoric and epistemic uncertainty quality respectively.}
    }
    \label{ddu:table:auroc_tab1}
    \scriptsize
    \renewcommand{\arraystretch}{1.2} 
    \begin{tabular}{@{}lccc@{}}
    \toprule
    \textbf{Model} & \textbf{ECE (\textdownarrow)} & \multicolumn{2}{c}{\textbf{AUROC}} \\
    \cmidrule(lr){3-4}
     & & \textbf{Softmax Entropy (\textuparrow)} & \textbf{Feature Density (\textuparrow)} \\
    \midrule
    LeNet & $2.22$ & $84.23$ & $71.41$\\
    VGG-16 & $\mathbf{2.11}$ & $84.04$ & $89.01$\\
    \textbf{ResNet-18+SN (DDU)} & $2.34$ & $83.01$ & $\mathbf{99.91}$\\
    \bottomrule
    \end{tabular}
\end{table}

As mentioned, in \Cref{ddu:fig:intro_histograms} we train a LeNet \citep{lecun1998gradient}, a VGG-16 \citep{simonyan2014very} and a ResNet-18 with spectral normalization \citep{he2016deep,miyato2018spectral} (ResNet-18+SN) on Dirty-MNIST.
\Cref{ddu:fig:intro_softmax_ent} shows that the softmax entropy of a deterministic model is unable to distinguish between iD (Dirty-MNIST) and OoD (FashionMNIST \citep{xiao2017fashion}) samples as the entropy for the latter heavily overlaps with the entropy for Ambiguous-MNIST samples.
However, the feature-space density of the model with our inductive biases in \Cref{ddu:fig:intro_gmm} captures epistemic uncertainty reliably and is able to distinguish iD from OoD samples.
The same cannot be said for LeNet or VGG in \Cref{ddu:fig:intro_gmm}, whose densities are unable to separate OoD from iD samples.
This demonstrates the importance of the inductive bias to ensure the sensitivity and smoothness of the feature space as we further argue below.
Finally, \Cref{ddu:fig:intro_softmax_ent} and \Cref{ddu:fig:intro_gmm} demonstrate that our method is able to separate aleatoric from epistemic uncertainty: 
samples with low feature density have high epistemic uncertainty, whereas those with both high feature density and high softmax entropy have high aleatoric uncertainty---note the high softmax entropy for the most ambiguous Ambiguous-MNIST samples in \Cref{ddu:fig:intro_softmax_ent}.

\paragraph{Disentangling Epistemic and Aleatoric Uncertainty}
\label{ddu:app:experiments_disentangling}

\Cref{ddu:table:auroc_tab1} gives a quantitative evaluation of the qualitative results in \S\ref{ddu:sec:intro}.
The AUROC metric reflects the quality of the epistemic uncertainty as it measures the probability that iD and OoD samples can be distinguished, and OoD samples are never seen during training while iD samples are semantically similar to training samples. The ECE metric measures the quality of the aleatoric uncertainty.
The softmax outputs capture aleatoric uncertainty well, as expected, and all 3 models obtain similar ECE scores on the Dirty-MNIST test set.
However, with an AUROC of around $84\%$ for all the 3 models, on Dirty-MNIST vs FashionMNIST, we conclude that softmax
entropy is unable to capture epistemic uncertainty well.
This is reinforced in \Cref{ddu:fig:intro_softmax_ent}, which shows a strong overlap between the softmax entropy of OoD and ambiguous iD samples.
At the same time, the feature-space densities of LeNet and VGG-16, with AUROC scores around $71\%$ and $89\%$ respectively, are unable to distinguish OoD from iD samples, indicating that simply using feature-space density without appropriate inductive biases (as seen in \citep{lee2018simple}) is not sufficient.

\emph{Only by fitting a GMM on top of a feature extractor with appropriate inductive biases (DDU) and using its feature density are we able to obtain performance far better (with AUROC of $99.9\%$) than the alternatives in the ablation study (see \Cref{ddu:table:auroc_tab1}, but this is also noticeable in \Cref{ddu:fig:intro_gmm})}.
The entropy of a softmax model can capture aleatoric uncertainty, even without additional inductive biases, but it \emph{cannot} be used to estimate epistemic uncertainty (see \S\ref{ddu:sec:motivation}). On the other hand, feature-space density can \emph{only} be used to estimate epistemic uncertainty \emph{when the feature extractor is sensitive and smooth}, as achieved by using a ResNet and spectral normalization in DDU.

\subsection{Active Learning}
\label{ddu:sec:experiments_active_learning}

\begin{figure}[!t]
    \centering
    \begin{minipage}[b]{0.35\linewidth}
        \begin{subfigure}{\linewidth}
            \centering
            \includegraphics[width=\linewidth]{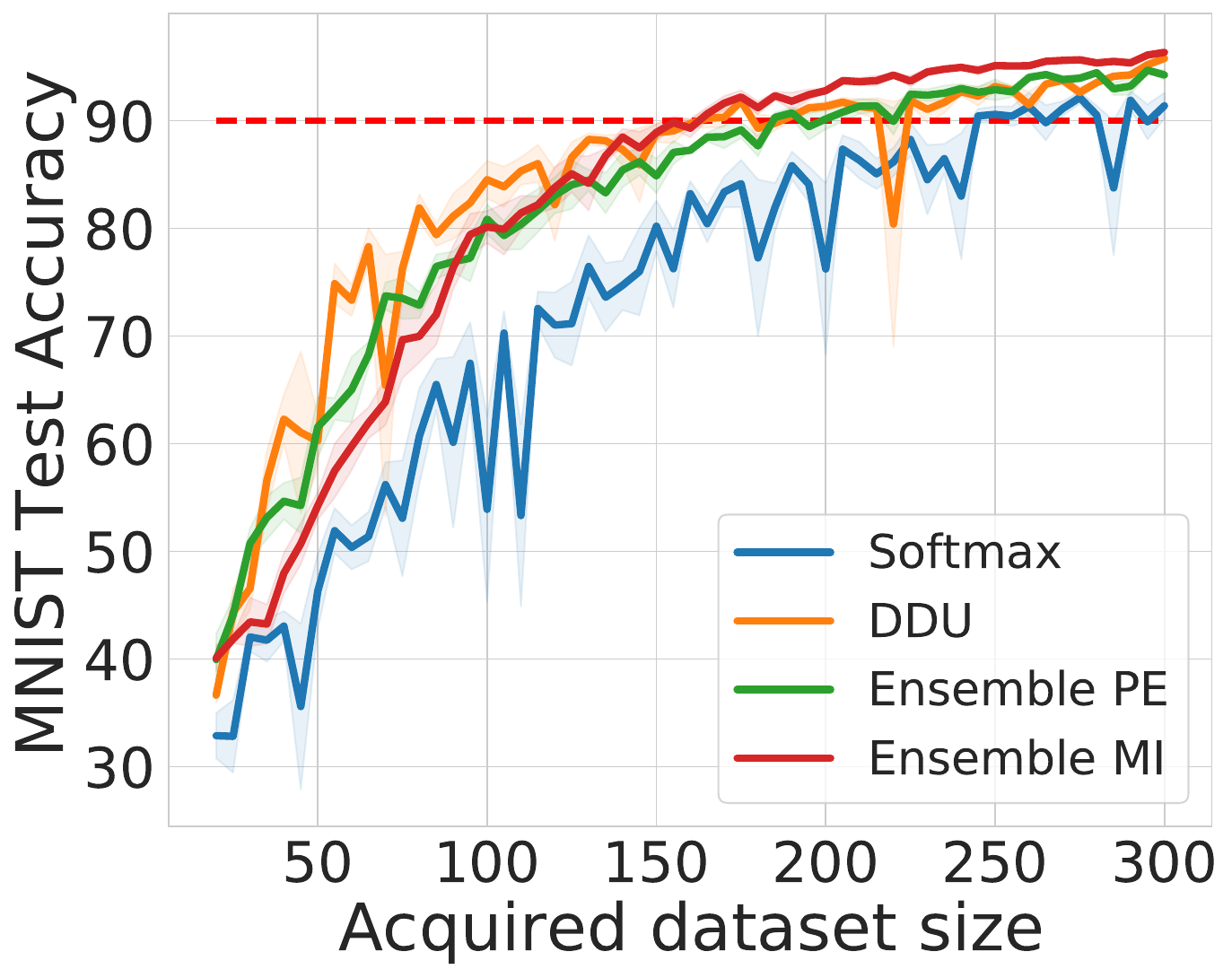}%
            \caption{MNIST}
            \label{ddu:subfig:test_accuracy_active_learning_mnist}
        \end{subfigure}
        \begin{subfigure}{\linewidth}
            \centering
            \includegraphics[width=\linewidth]{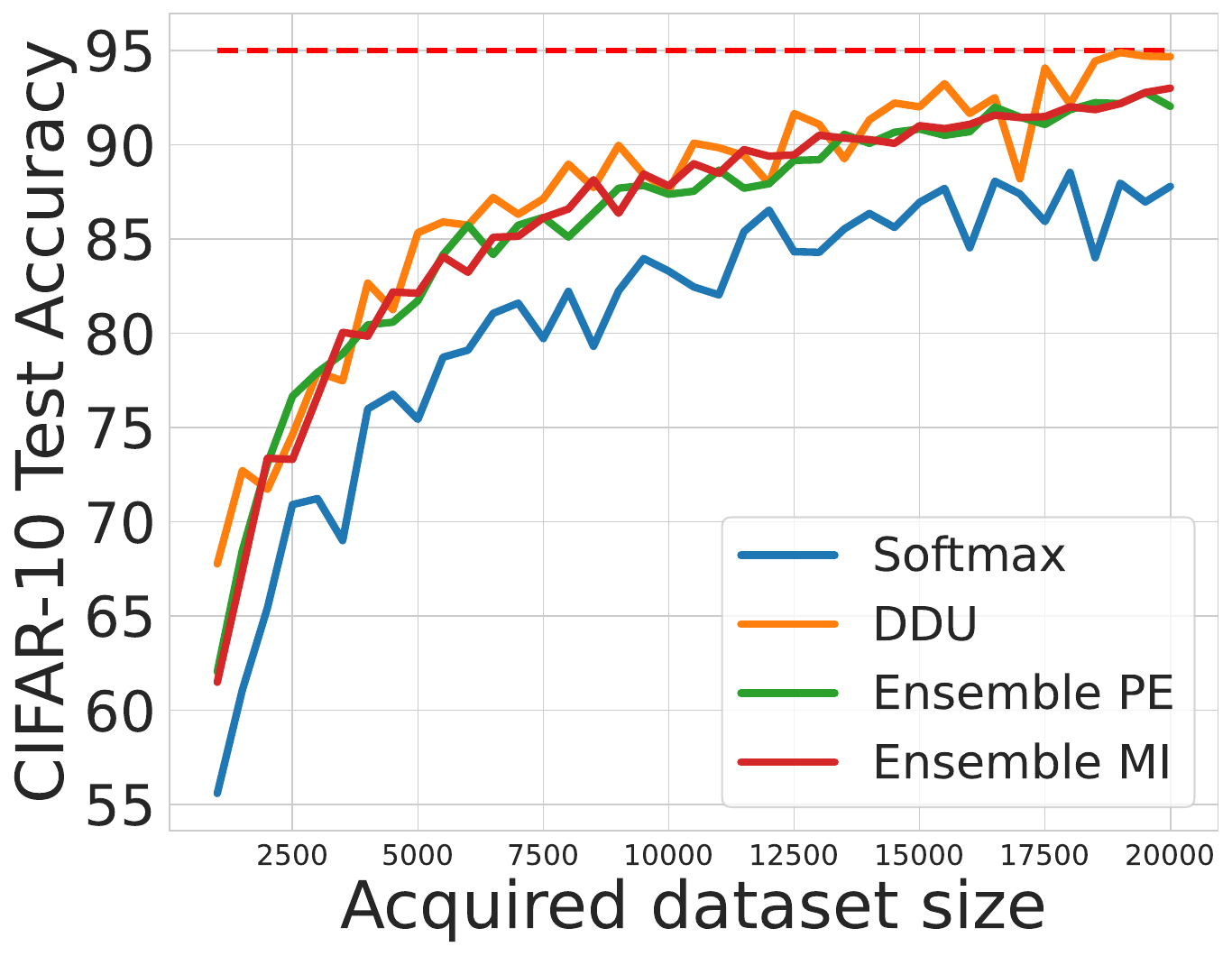}
            \caption{CIFAR-10}
            \label{ddu:subfig:test_accuracy_active_learning_cifar10}
        \end{subfigure}
    \end{minipage}\hfill
    \begin{subfigure}[b]{0.65\linewidth}
        \centering
        \includegraphics[width=\linewidth]{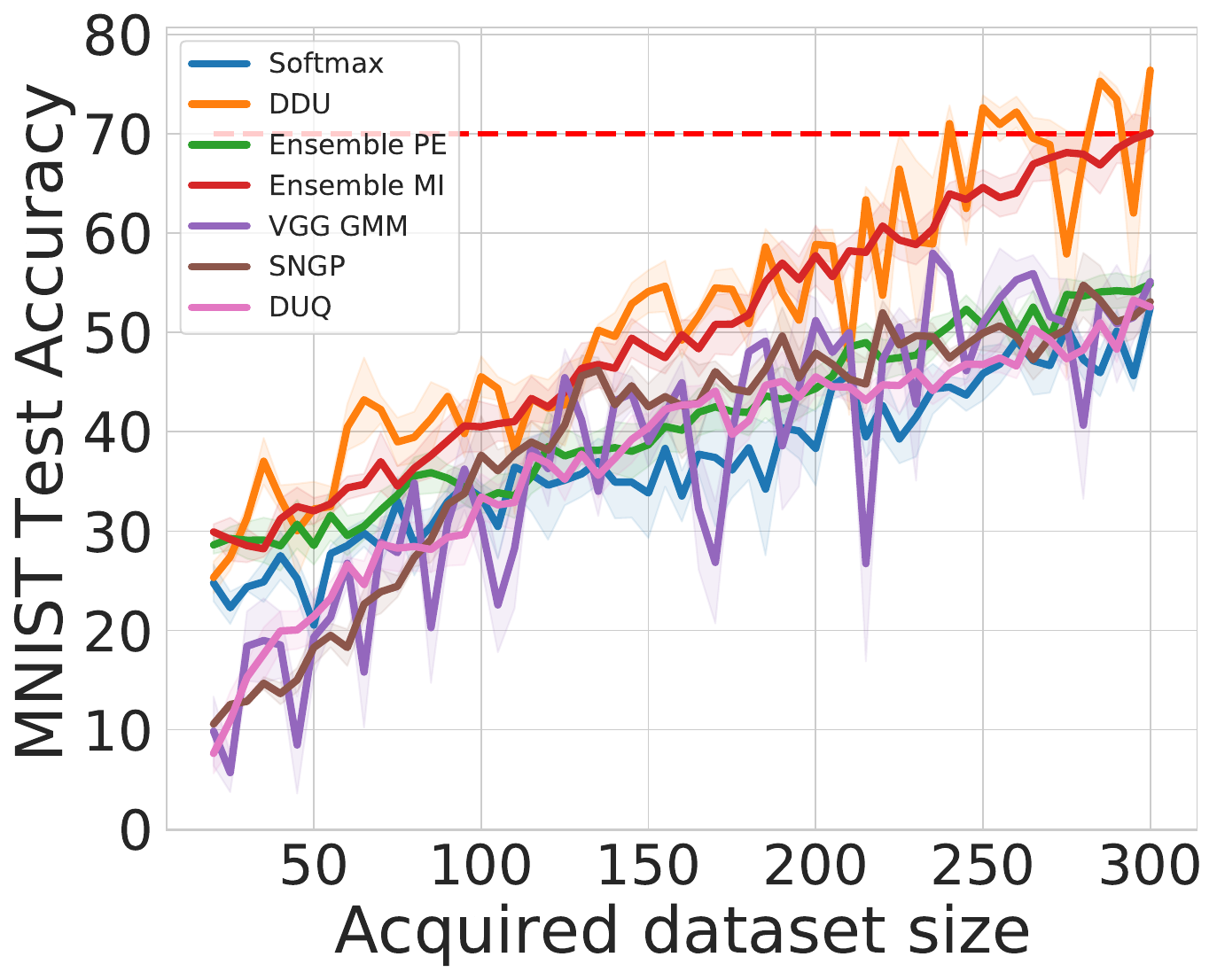}%
        \caption{Dirty-MNIST}
        \label{ddu:subfig:test_accuracy_active_learning_dirty_mnist_1000}
    \end{subfigure}
    \caption{
    \emph{Active Learning experiments.} Acquired training set size vs test accuracy. DDU performs on par with deep ensembles.
    }
    \label{ddu:fig:active_learning_mnist_dirty_mnist}
\end{figure}

We evaluate DDU on three different active learning setups: 
\begin{enumerate}
\item with clean MNIST samples in the pool set,
\item with clean CIFAR-10 samples in the pool set, and 
\item with Dirty-MNIST, having a 1:60 ratio of MNIST to Ambiguous-MNIST samples, in the pool set.
\end{enumerate}
In the first two setups, we compare three baselines as for two moons:
\begin{itemize}
    \item a ResNet-18 with softmax entropy as the acquisition function,
    \item DDU trained using a ResNet-18 with feature density as acquisition function, and 
    \item a deep ensemble of 3 ResNet-18s with the predictive entropy (PE) and mutual information (MI) of the ensemble as the acquisition functions.
\end{itemize}
For Dirty-MNIST, in addition to the above 3 approaches, we also compare to:
\begin{itemize}
    \item feature density of a VGG-16 instead of ResNet-18+SN as an ablation to see if feature density of a model without inductive biases performs well, as well as
    \item SNGP \citep{lee2020predictive} and DUQ \citep{van2020uncertainty} as additional baselines.
\end{itemize}
For MNIST and Dirty-MNIST, we start with an initial training-set size of 20 randomly chosen MNIST points, and in each iteration, acquire the 5 samples with the highest reported epistemic uncertainty. We re-train the models after each batch acquisition step using Adam \citep{kingma2014adam} for 100 epochs and choose the model with the best validation set accuracy. We stop the process when the training set size reaches 300. For CIFAR-10, we start with 1000 samples and go up to 20000 samples with an acquisition size of 500 samples in each step.

\paragraph{MNIST \&\ CIFAR-10.} In \Cref{ddu:fig:active_learning_mnist_dirty_mnist}\subref{ddu:subfig:test_accuracy_active_learning_mnist} and \Cref{ddu:fig:active_learning_mnist_dirty_mnist}\subref{ddu:subfig:test_accuracy_active_learning_cifar10}, for regular curated MNIST and CIFAR-10 in the pool set, DDU clearly outperforms the deterministic softmax baseline and is competitive with deep ensembles. For MNIST, the softmax baseline reaches 90\% test-set accuracy at a training-set size of 245. DDU reaches 90\% accuracy at a training-set size of 160, whereas deep ensemble reaches the same at 185 and 155 training samples with PE and MI as the acquisition functions respectively. Note that DDU is three times faster than a deep ensemble, which needs to train three models independently after every acquisition.

\paragraph{Dirty-MNIST.} Real-life datasets often contain observation noise and ambiguous samples.
What happens when the pool set contains a lot of such noisy samples having high aleatoric uncertainty?
In such cases, it becomes important for models to identify unseen and informative samples with high epistemic uncertainty and not with high aleatoric uncertainty.
To study this, we construct a pool set with samples from Dirty-MNIST (see \S\ref{ddu:app:dirty_mnist}). We significantly increase the proportion of ambiguous samples by using a 1:60 split of MNIST to Ambiguous-MNIST (a total of 1K MNIST and 60K Ambiguous-MNIST samples).
In \Cref{ddu:fig:active_learning_mnist_dirty_mnist}\subref{ddu:subfig:test_accuracy_active_learning_dirty_mnist_1000}, for Dirty-MNIST in the pool set, the difference in the performance of DDU and the deterministic softmax model is stark. While DDU achieves a test set accuracy of 70\% at a training set size of 240 samples, the accuracy of the softmax baseline peaks at a mere 50\%. In addition, all baselines, including SNGP, DUQ and the feature density of a VGG-16, which fail to solely capture epistemic uncertainty, are significantly outperformed by DDU and the MI baseline of the deep ensemble. However, note that DDU also performs better than deep ensembles with the PE acquisition function. The difference gets larger as the training set size grows: DDU's feature density and deep ensemble's MI solely capture epistemic uncertainty and hence, do not get confounded by iD ambiguous samples with high aleatoric uncertainty. 

\subsection{OoD Detection}
\label{ddu:sec:experiments_ood_detection}

\begin{sidewaystable}[!]
    \begin{minipage}{\linewidth}
        \centering
        \caption{\emph{OoD detection performance of different baselines using a Wide-ResNet-28-10 architecture with the CIFAR-10 vs SVHN/CIFAR-100/Tiny-ImageNet and CIFAR-100 vs SVHN/Tiny-ImageNet dataset pairs averaged over 25 runs.} \emph{SN:} Spectral Normalization, \emph{JP:} Jacobian Penalty. The best deterministic single-forward pass method and the best method overall are in bold for each metric.}
        \vspace{-2.5mm}
        \label{ddu:table:ood_wrn}
        \resizebox{\linewidth}{!}{%
        \begin{tabular}{cccccccccc}
        \toprule
        \textbf{\small Train Dataset} & \textbf{\small Method} & \textbf{\small Penalty} & {\small Aleatoric Uncertainty} &
        \textbf{\small Epistemic Uncertainty} & {\small Accuracy (\textuparrow)} & {\small ECE (\textdownarrow)} & \multicolumn{3}{c}{\textbf{\small AUROC}} \\
        \cmidrule{8-10}
        & & & & & & & \textbf{\small SVHN (\textuparrow)} & \textbf{\small CIFAR-100 (\textuparrow)} & \textbf{\small Tiny-ImageNet (\textuparrow)}\\
        \midrule
        \multirow{7}{*}{CIFAR-10} & Softmax & - & \multirow{2}{*}{Softmax Entropy} & Softmax Entropy & \multirow{2}{*}{$95.98\pm0.02$} & \multirow{2}{*}{$\mathbf{0.85\pm0.02}$} & $94.44\pm0.43$ & $89.39\pm0.06$ & $88.42\pm0.05$ \\
        &Energy-based {\scriptsize \citep{liu2020energy}} & - && Softmax Density &&& $94.56\pm0.51$ & $88.89\pm0.07$ & $88.11\pm0.06$\\
        &DUQ {\scriptsize \citep{van2020uncertainty}} & JP & Kernel Distance & Kernel Distance & $94.6\pm0.16$ & $1.55\pm0.08$ & $93.71\pm0.61$ & $85.92\pm0.35$ & $86.83\pm0.12$\\
        &SNGP {\scriptsize \citep{liu2020simple}} & SN & Predictive Entropy & Predictive Entropy & $\mathbf{96.04\pm0.09}$ & $1.8\pm0.1$ & $94.0\pm1.3$ & $91.13\pm0.15$ & $89.97\pm0.19$\\
        &\textbf{DDU (ours)} & \textbf{SN} & \textbf{Softmax Entropy} & \textbf{GDA Density} & $95.97\pm0.03$&$\mathbf{0.85\pm0.04}$&$\mathbf{97.86\pm0.19}$&$\mathbf{91.34\pm0.04}$ & $\mathbf{91.07\pm0.05}$\\
        \cmidrule{2-10}
        &5-Ensemble & \multirow{2}{*}{-} & \multirow{2}{*}{Predictive Entropy} & Predictive Entropy & \multirow{2}{*}{$\mathbf{96.59\pm0.02}$}&\multirow{2}{*}{$\mathbf{0.76\pm0.03}$}&$97.73\pm0.31$&$\mathbf{92.13\pm0.02}$ & $90.06\pm0.03$\\
        &{\scriptsize \citep{lakshminarayanan2016simple}} &&& Mutual Information &&&$97.18\pm0.19$&$91.33\pm0.03$ & $90.90\pm0.03$\\
        \midrule
        &&&&& {\small Accuracy (\textuparrow)} & {\small{ECE (\textdownarrow)}} & \multicolumn{2}{c}{\textbf{\small SVHN (\textuparrow)}} & \textbf{\small Tiny-ImageNet (\textuparrow)} \\
        \cmidrule{6-10}
        \multirow{6}{*}{CIFAR-100} & Softmax & - & \multirow{2}{*}{Softmax Entropy} & Softmax Entropy & \multirow{2}{*}{$80.26\pm0.06$}&\multirow{2}{*}{$4.62\pm0.06$}& \multicolumn{2}{c}{$77.42\pm0.57$} & $81.53\pm0.05$ \\
        &Energy-based {\scriptsize \citep{liu2020energy}} & - && Softmax Density &&&\multicolumn{2}{c}{$78\pm0.63$} & $81.33\pm0.06$ \\
        &SNGP {\scriptsize \citep{liu2020simple}} & SN & Predictive Entropy & Predictive Entropy & $80.00\pm0.11$ & $4.33\pm0.01$ & \multicolumn{2}{c}{$85.71\pm0.81$} & $78.85\pm0.43$ \\
        &\textbf{DDU (ours)} & \textbf{SN} & \textbf{Softmax Entropy} &  \textbf{GMM Density} & $\mathbf{80.98\pm0.06}$&$\mathbf{4.10\pm0.08}$& \multicolumn{2}{c}{$\mathbf{87.53\pm0.62}$}& $\mathbf{83.13\pm0.06}$ \\
        \cmidrule{2-10}
        &5-Ensemble & \multirow{2}{*}{-} & \multirow{2}{*}{Predictive Entropy} & Predictive Entropy &\multirow{2}{*}{$\mathbf{82.79\pm0.10}$}&\multirow{2}{*}{$\mathbf{3.32\pm0.09}$}& \multicolumn{2}{c}{$79.54\pm0.91$} & $82.95\pm0.09$\\
        &{\scriptsize \citep{lakshminarayanan2016simple}} &&& Mutual Information &&& \multicolumn{2}{c}{$77.00\pm1.54$} & $82.82\pm0.04$\\
        \bottomrule
        \end{tabular}}
    \end{minipage}
\end{sidewaystable}

\begin{figure}[!t]
    \centering
    \begin{subfigure}{0.50\linewidth}
        \centering
        \includegraphics[width=\linewidth]{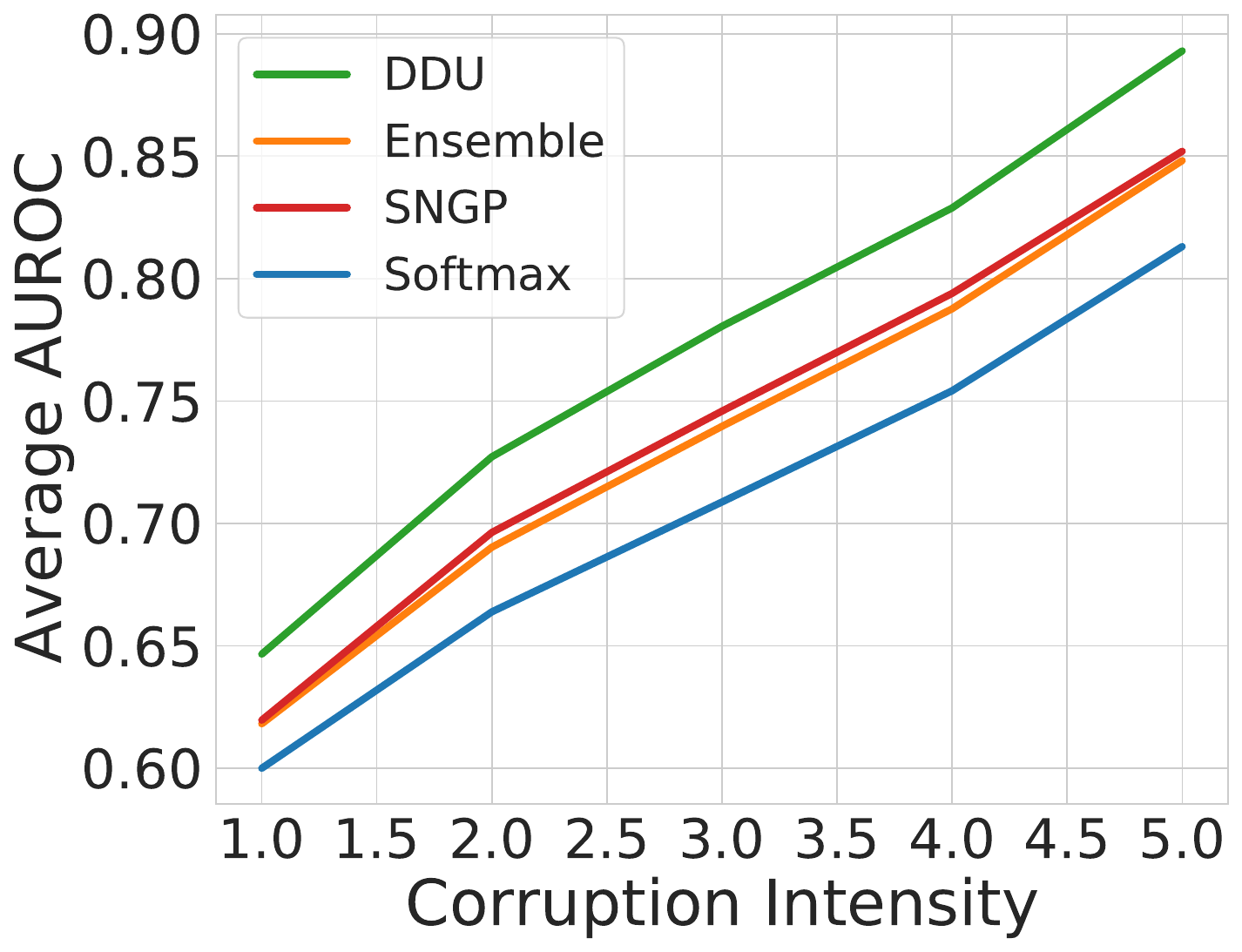}
        \caption{WRN-28-10}
        \label{ddu:subfig:cifar10_c_wide_resnet}
    \end{subfigure}\hfill
    \begin{subfigure}{0.50\linewidth}
        \centering
        \includegraphics[width=\linewidth]{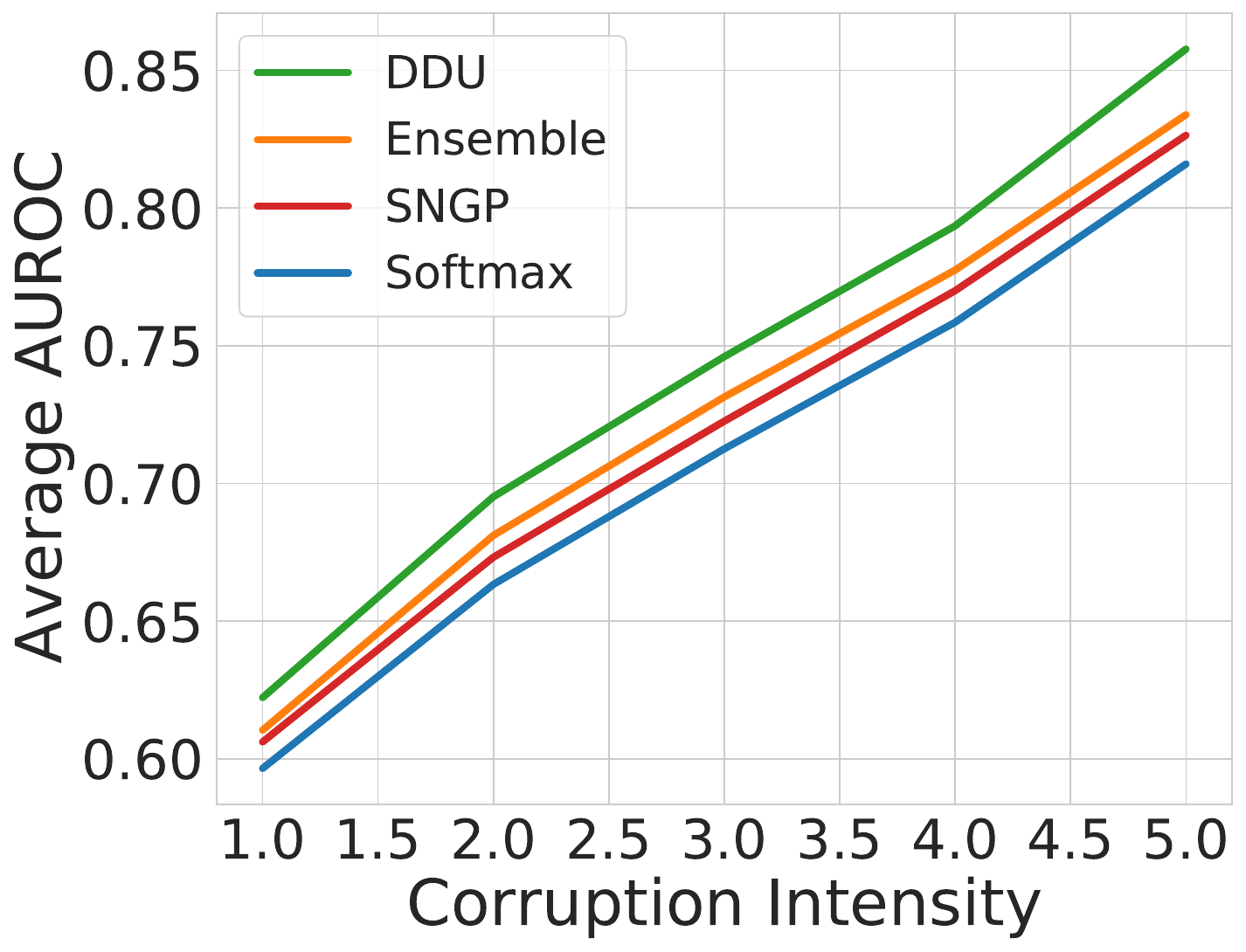}
        \caption{RN-50}
        \label{ddu:subfig:cifar10_c_resnet50}
    \end{subfigure} 
    \begin{subfigure}{0.50\linewidth}
        \centering
        \includegraphics[width=\linewidth]{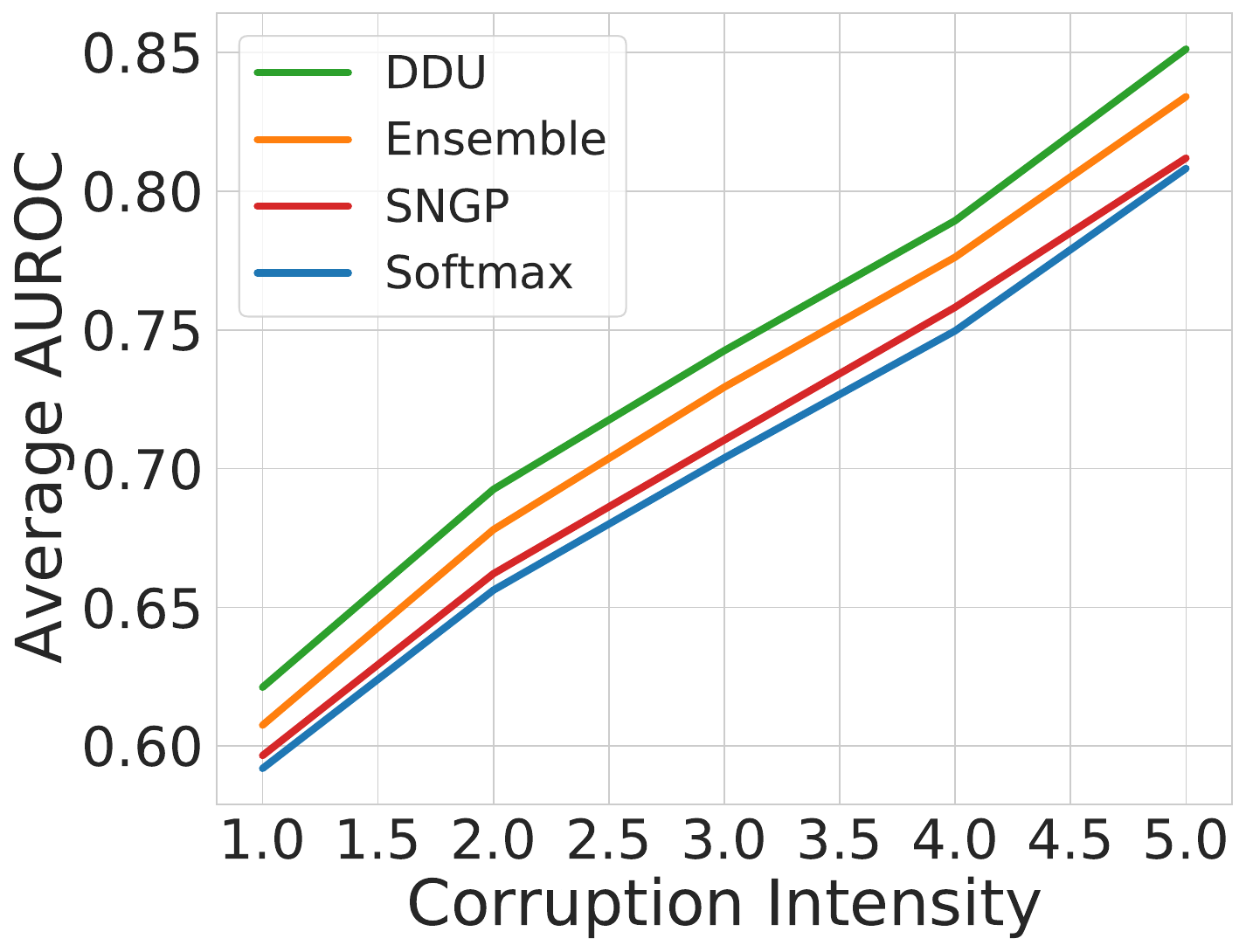}
        \caption{RN-110}
        \label{ddu:subfig:cifar10_c_resnet110}
    \end{subfigure}\hfill
    \begin{subfigure}{0.50\linewidth}
        \centering
        \includegraphics[width=\linewidth]{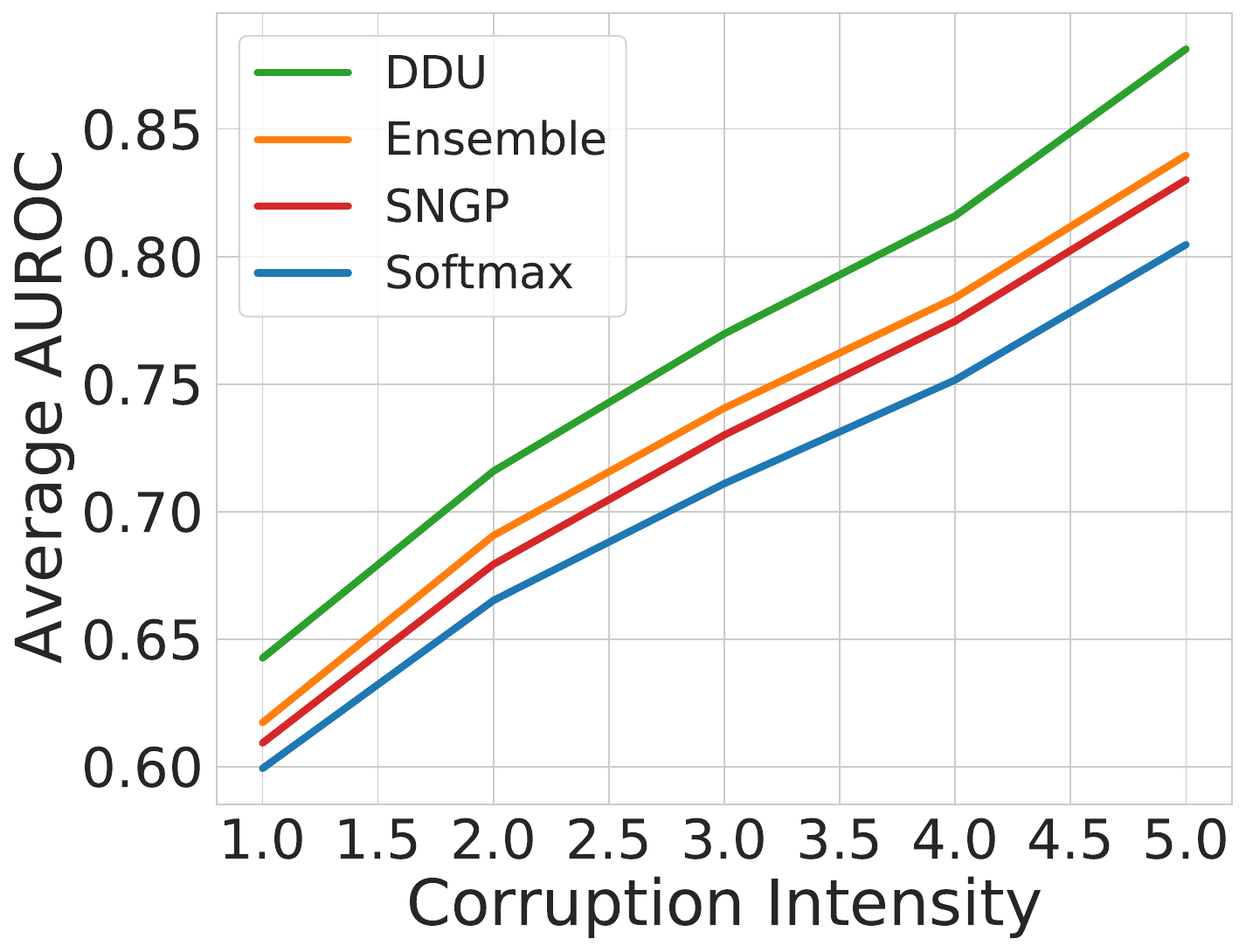}
        \caption{DN-121}
        \label{ddu:subfig:cifar10_c_densenet121}
    \end{subfigure} 
    \caption{
    AUROC vs corruption intensity averaged over all corruption types in CIFAR-10-C for 4 architectures.
    More details in \S\ref{ddu:sec:experiments_ood_detection} and more ablations in \S\ref{ddu:app:exp_details} in the appendix. 
    }
    \label{ddu:fig:cifar10_c_results}
\end{figure}

Near-OoD detection is an application of epistemic uncertainty quantification: if we do not train on OoD data, we expect near OoD data points to have higher epistemic uncertainty than iD data.

\subsubsection{Image Classification}
We evaluate CIFAR-10 vs SVHN/CIFAR-100/Tiny-ImageNet/CIFAR-10-C, CIFAR-100 vs SVHN/Tiny-ImageNet and ImageNet vs ImageNet-O as iD vs OoD dataset pairs for this experiment \citep{krizhevsky2009learning, svhn2011,deng2009imagenet,hendrycks2019benchmarking}.
We also evaluate DDU on different architectures: Wide-ResNet-28-10, Wide-ResNet-50-2, ResNet-50, ResNet-110 and DenseNet-121 \citep{zagoruyko2016wide,he2016deep,huang2017densely}.
The training setup is described in \S\ref{ddu:app:exp_details_cifar}.
In addition to using softmax entropy of a deterministic model (\emph{Softmax}) for both aleatoric and epistemic uncertainty, we also compare with the following \textbf{baselines} that do not require training or fine-tuning on OoD data:
\begin{description}
\item[{Energy-based model} \citep{liu2020energy}:] We use the softmax entropy of a deterministic model as aleatoric uncertainty and the unnormalized softmax density (the logsumexp of the logits) as epistemic uncertainty \emph{without} regularization to avoid feature collapse. We only compare with the version that does not train on OoD data.

\item[{DUQ \citep{van2020uncertainty} \&\ SNGP \citep{liu2020simple}}:] We compare with the state-of-the-art deterministic methods for uncertainty quantification including DUQ and SNGP. For SNGP, we use the exact predictive covariance computation. We measure uncertainty via the entropy of the average of the MC softmax samples. For DUQ, we use the closest kernel distance. Note that for CIFAR-100, DUQ's one-vs-all objective did not converge during training and hence, we do not include the DUQ baseline for CIFAR-100.

\item[{5-Ensemble}:] We use an ensemble of 5 networks
with the same architecture
and compute the predictive entropy of the ensemble as both epistemic and aleatoric uncertainty and mutual information as epistemic uncertainty.
\end{description}
\paragraph{Results.} \Cref{ddu:table:ood_wrn} presents the AUROC for Wide-ResNet-28-10 models on CIFAR-10 vs SVHN/CIFAR-100/Tiny-ImageNet and CIFAR-100 vs SVHN/Tiny-ImageNet along with their respective test set accuracy and ECE post temp-scaling (additional calibration scores in \S\ref{ddu:app:additional_calibration_metrics} and comparison with more baselines in \S\ref{ddu:app:ddu_additional_baselines}). The equivalent results for other architectures: ResNet-50/110 and DenseNet-121 can be found in \Cref{ddu:table:ood_resnet50}, \Cref{ddu:table:ood_resnet110} and \Cref{ddu:table:ood_densenet121} in the appendix. Note that for DDU, post-hoc calibration with temperature scaling \citep{guo2017calibration}, is simple as it does not affect the GMM density. We also plot the AUROC averaged over corruption types vs corruption intensity for CIFAR-10 vs CIFAR-10-C in \Cref{ddu:fig:cifar10_c_results}, with AUROC plots per corruption type in \Cref{ddu:fig:cifar10_c_results_wide_resnet}, \Cref{ddu:fig:cifar10_c_results_resnet50}, \Cref{ddu:fig:cifar10_c_results_resnet110} and \Cref{ddu:fig:cifar10_c_results_densenet121} of the appendix. Finally, in \Cref{ddu:table:ood_imagenet}, we present AUROC for models trained on ImageNet.

For OoD detection, \emph{DDU outperforms all other deterministic single-forward-pass methods, DUQ, SNGP and the energy-based model approach from \citep{liu2020energy}, on CIFAR-10 vs SVHN/CIFAR-100/Tiny-ImageNet, CIFAR-10 vs CIFAR-10-C and CIFAR-100 vs SVHN/Tiny-ImageNet, often performs on par with state-of-the-art deep ensembles---and even performing better in a few cases}. This holds true for all the architectures we experimented on. Similar observations can be made on ImageNet vs ImageNet-O as well.
Importantly, the great performance in OoD detection comes without compromising on the single-model test set accuracy in comparison to other deterministic methods.

\subsubsection{Semantic Segmentation}
\label{ddu:sec:experiments_semantic_seg}

\begin{sidewaysfigure}
    \centering
    \begin{subfigure}{0.13\linewidth}
        \centering
        \includegraphics[width=\linewidth]{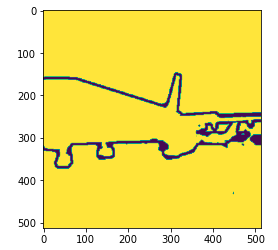}
    \end{subfigure}
    \begin{subfigure}{0.13\linewidth}
        \centering
        \includegraphics[width=\linewidth]{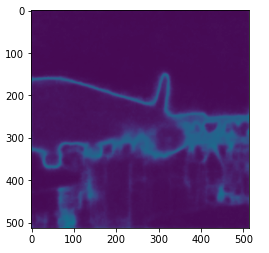}
    \end{subfigure}
    \begin{subfigure}{0.13\linewidth}
        \centering
        \includegraphics[width=\linewidth]{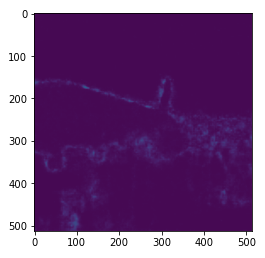}
    \end{subfigure}
    \begin{subfigure}{0.13\linewidth}
        \centering
        \includegraphics[width=\linewidth]{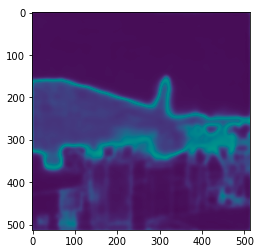}
    \end{subfigure}
    \begin{subfigure}{0.13\linewidth}
        \centering
        \includegraphics[width=\linewidth]{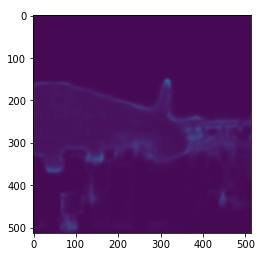}
    \end{subfigure}
    \begin{subfigure}{0.13\linewidth}
        \centering
        \includegraphics[width=\linewidth]{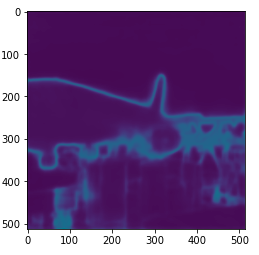}
    \end{subfigure}
    \begin{subfigure}{0.13\linewidth}
        \centering
        \includegraphics[width=\linewidth]{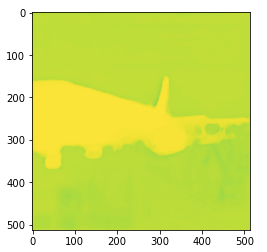}
    \end{subfigure}

    \begin{subfigure}{0.13\linewidth}
        \centering
        \includegraphics[width=\linewidth]{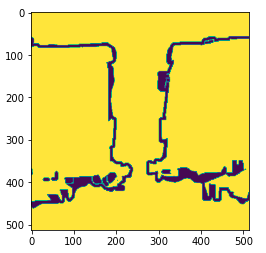}
    \end{subfigure}
    \begin{subfigure}{0.13\linewidth}
        \centering
        \includegraphics[width=\linewidth]{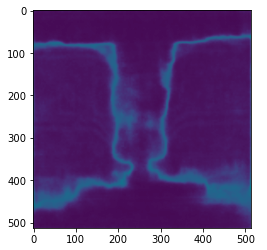}
    \end{subfigure}
    \begin{subfigure}{0.13\linewidth}
        \centering
        \includegraphics[width=\linewidth]{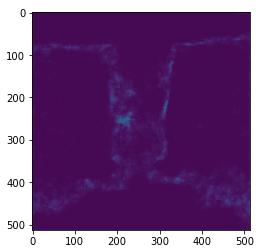}
    \end{subfigure}
    \begin{subfigure}{0.13\linewidth}
        \centering
        \includegraphics[width=\linewidth]{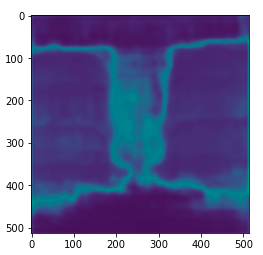}
    \end{subfigure}
    \begin{subfigure}{0.13\linewidth}
        \centering
        \includegraphics[width=\linewidth]{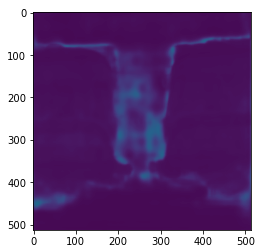}
    \end{subfigure}
    \begin{subfigure}{0.13\linewidth}
        \centering
        \includegraphics[width=\linewidth]{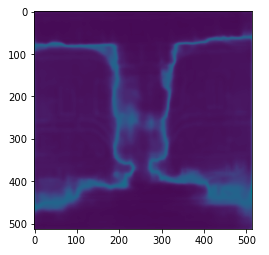}
    \end{subfigure}
    \begin{subfigure}{0.13\linewidth}
        \centering
        \includegraphics[width=\linewidth]{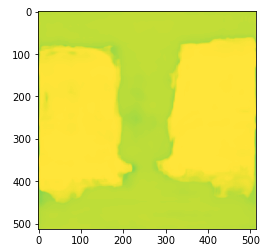}
    \end{subfigure}

    \begin{subfigure}{0.13\linewidth}
        \centering
        \includegraphics[width=\linewidth]{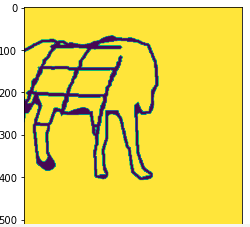}
    \end{subfigure}
    \begin{subfigure}{0.13\linewidth}
        \centering
        \includegraphics[width=\linewidth]{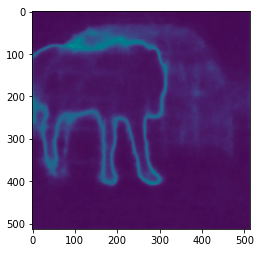}
    \end{subfigure}
    \begin{subfigure}{0.13\linewidth}
        \centering
        \includegraphics[width=\linewidth]{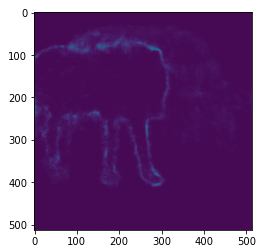}
    \end{subfigure}
    \begin{subfigure}{0.13\linewidth}
        \centering
        \includegraphics[width=\linewidth]{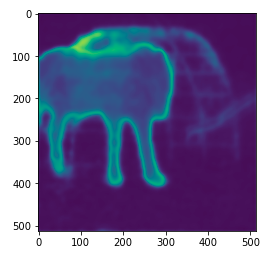}
    \end{subfigure}
    \begin{subfigure}{0.13\linewidth}
        \centering
        \includegraphics[width=\linewidth]{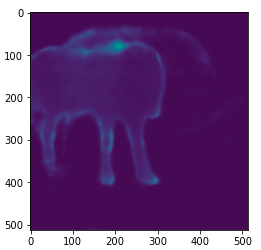}
    \end{subfigure}
    \begin{subfigure}{0.13\linewidth}
        \centering
        \includegraphics[width=\linewidth]{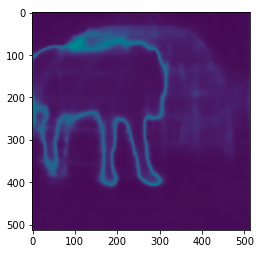}
    \end{subfigure}
    \begin{subfigure}{0.13\linewidth}
        \centering
        \includegraphics[width=\linewidth]{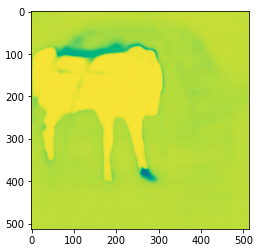}
    \end{subfigure}

    \begin{subfigure}{0.13\linewidth}
        \centering
        \includegraphics[width=\linewidth]{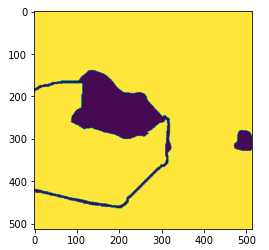}
        \caption{{\tiny Accuracy}}
        \label{ddu:subfig:seg_unc_vis_accuracy}
    \end{subfigure}
    \begin{subfigure}{0.13\linewidth}
        \centering
        \includegraphics[width=\linewidth]{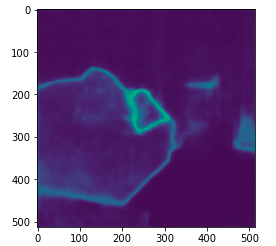}
        \caption{{\tiny MCD PE}}
        \label{ddu:subfig:seg_unc_vis_mcd_pe}
    \end{subfigure}
    \begin{subfigure}{0.13\linewidth}
        \centering
        \includegraphics[width=\linewidth]{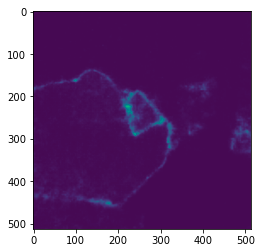}
        \caption{{\tiny MCD MI}}
        \label{ddu:subfig:seg_unc_vis_mcd_mi}
    \end{subfigure}
    \begin{subfigure}{0.13\linewidth}
        \centering
        \includegraphics[width=\linewidth]{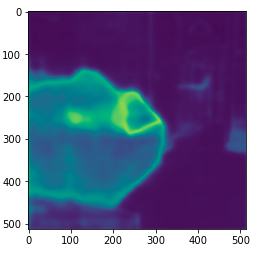}
        \caption{{\tiny En PE}}
        \label{ddu:subfig:seg_unc_vis_en_pe}
    \end{subfigure}
    \begin{subfigure}{0.13\linewidth}
        \centering
        \includegraphics[width=\linewidth]{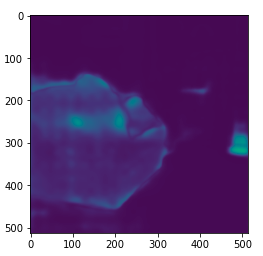}
        \caption{{\tiny En MI}}
        \label{ddu:subfig:seg_unc_vis_en_mi}
    \end{subfigure}
    \begin{subfigure}{0.13\linewidth}
        \centering
        \includegraphics[width=\linewidth]{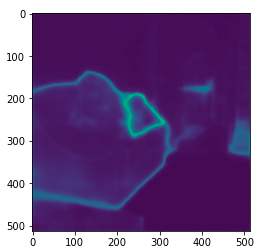}
        \caption{{\tiny Entropy}}
        \label{ddu:subfig:seg_unc_vis_entropy}
    \end{subfigure}
    \begin{subfigure}{0.13\linewidth}
        \centering
        \includegraphics[width=\linewidth]{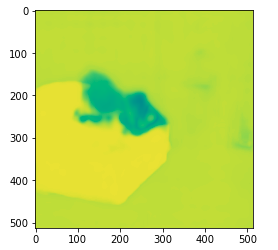}
        \caption{{\tiny Density}}
        \label{ddu:subfig:seg_unc_vis_density}
    \end{subfigure}
    \caption{
    \emph{Visualization of uncertainty baselines on four PASCAL VOC validation samples (\textbf{rows}).} \textbf{Columns}: \textbf{\subref{ddu:subfig:seg_unc_vis_accuracy}} shows pixel-wise accuracy; \textbf{\subref{ddu:subfig:seg_unc_vis_mcd_pe}, \subref{ddu:subfig:seg_unc_vis_mcd_mi}} predictive entropy (PE) and mutual information (MI) obtained for MC Dropout (MCD); \textbf{\subref{ddu:subfig:seg_unc_vis_en_pe}, \subref{ddu:subfig:seg_unc_vis_en_mi}} for deep ensembles; \textbf{\subref{ddu:subfig:seg_unc_vis_entropy}} per-pixel softmax entropy, \emph{the aleatoric estimate of DDU}; and \textbf{\subref{ddu:subfig:seg_unc_vis_density}} feature density, \emph{the epistemic component of DDU}. For all  but \subref{ddu:subfig:seg_unc_vis_density}: the brighter, the more uncertain, whereas DDU's density \subref{ddu:subfig:seg_unc_vis_density} captures certainty: hence, the brighter, the more certain.
    }
    \label{ddu:fig:seg_unc_vis}
\end{sidewaysfigure}

\begin{figure}[t]
    \centering
    \begin{subfigure}{0.33\linewidth}
        \includegraphics[width=\linewidth]{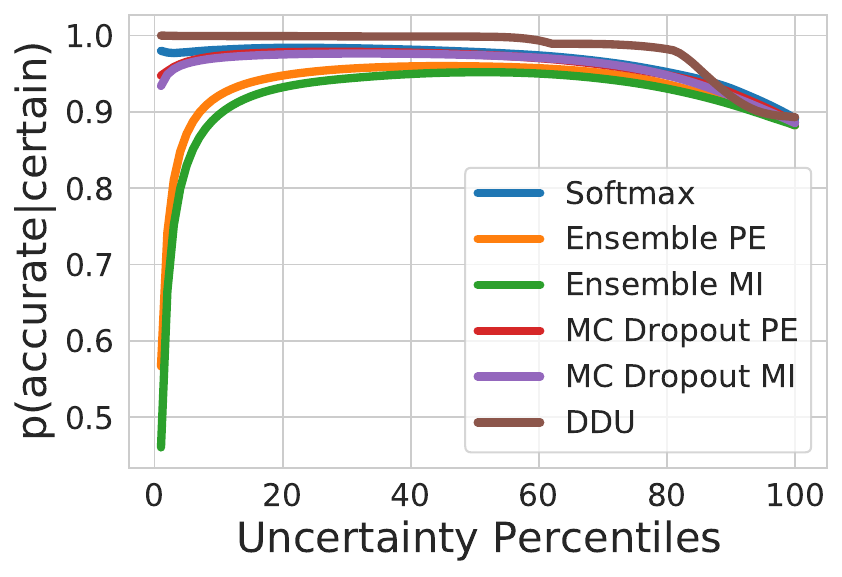}
        \label{ddu:fig:p_acc_cert}
    \end{subfigure}\hfill
    \begin{subfigure}{0.33\linewidth}
        \centering
        \includegraphics[width=\linewidth]{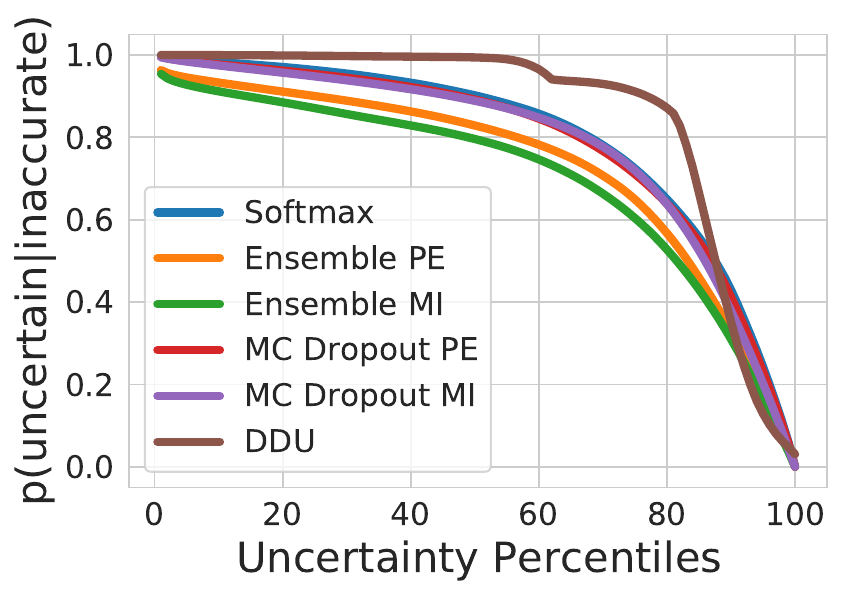}
        \label{ddu:fig:p_unc_inacc}
    \end{subfigure}\hfill
    \begin{subfigure}{0.33\linewidth}
        \centering
        \includegraphics[width=\linewidth]{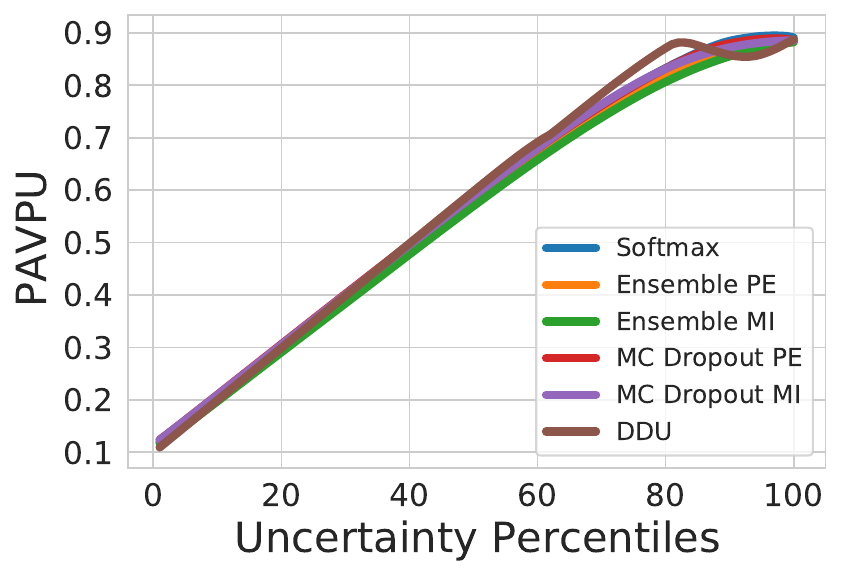}
        \label{ddu:fig:pavpu}
    \end{subfigure}%
    \caption{%
    \emph{$\pof{\mathrm{accurate} \given \mathrm{certain}}$, $\pof{\mathrm{uncertain} \given \mathrm{inaccurate}}$ and $\mathrm{PAvPU}$ evaluated on PASCAL VOC validation set.} DDU outperforms all other baselines.
    }%
    \label{ddu:fig:seg_metrics}%
\end{figure}

\begin{table}[t]
    \centering
    \caption{\emph{Pascal VOC validation set mIoU and runtime in milliseconds 
    averaged over 10 forward passes.} For MC Dropout, we perform 5 stochastic forward passes.}
    \label{ddu:tab:runtime}
    \scriptsize
    \begin{tabular}{@{}ccccc@{}}
    \toprule
    \textbf{Baseline} & Softmax & MC Dropout & Deep Ensemble & \textbf{DDU} \\ \midrule
    \textbf{mIoU} & $78.53$ & $78.61$ & $78.47$ & $78.53$ \\
    \textbf{Runtime (ms)} & $275.48\pm1.91$ & $1576.75\pm1.56$ & $875.87\pm0.79$ & $\mathbf{263.83\pm2.79}$ \\ \bottomrule
    \end{tabular}
\end{table}

In this section, we apply DDU to the task of semantic segmentation on Pascal VOC 2012 \cite{everingham2010pascal}, comparing with a vanilla softmax model, MC Dropout and deep ensembles.
Semantic segmentation \citep{long2015fully}, %
classifies every pixel of a given to one of a fixed set of classes. Since different classes can have different levels of representation in a segmentation dataset, it forms a classic example of a problem with class imbalance, thereby requiring reliable estimates of epistemic uncertainty.
Furthermore, due to the computationally heavy nature of semantic segmentation, classic uncertainty quantification approaches like MC Dropout and deep ensembles
are often impractical in real-world applications. 

\paragraph{Pixel-Independent Class-Wise Means and Covariances.} As each pixel has a corresponding prediction in semantic segmentation, it is natural to ask if the Gaussian means and covariance matrices need to be computed per pixel. To examine this, in \Cref{ddu:fig:l2_dist} of \S\ref{ddu:app:segmentation_training_setup}, we plot the L2 distances between feature space means of all pairs of classes obtained from a DeepLab-v3+ \cite{chen2017rethinking} model with a ResNet-101 backbone for two ``distant'' pixels. We observe that pixels of the same class are much closer in the feature space than pixels of different classes, irrespective of their location in the image. In spirit of a new simple baseline, we thus compute the Gaussian means and covariances per class, taking each pixel as a separate data point.

\paragraph{Architecture, Training and Evaluation Metrics.} As mentioned above, we evaluate DDU on Pascal VOC 2012 and compare to a vanilla softmax model, MC Dropout with 5 forward passes at test time, and a deep ensembles with 3 members. We use DeepLab-v3+ with a ResNet-101 backbone as the model architecture. Additional training details are in \S\ref{ddu:app:segmentation_training_setup}. Finally, to evaluate the uncertainty estimates, we use patch-based metrics proposed in \cite{mukhoti2018evaluating}: $\pof{\mathrm{accurate} \given \mathrm{certain}}$, $\pof{\mathrm{uncertainty} \given \mathrm{inaccurate}}$ and the  \emph{Patch Accuracy vs Patch Uncertainty} $\mathrm{PAvPU}$. $\pof{\mathrm{accurate} \given \mathrm{certain}}$ computes the probability of the model being accurate given that it is confident. Similarly, $\pof{\mathrm{uncertainty} \given \mathrm{inaccurate}}$ measures probability of the model being uncertain given that it is inaccurate, and $PAvPU$ computes the probability of the model being confident on accurate predictions and uncertain on inaccurate ones, so the accuracy depending on the uncertainty threshold similar to a rejection plot. Ideally, high values for these metrics indicate better uncertainty estimates in segmentation. Furthermore, note that these metrics can be computed at different thresholds of uncertainty (defining if a model is certain or not).

\paragraph{Results and Discussion.} In \Cref{ddu:fig:seg_metrics}, we present the above 3 metrics for all segmentation baselines evaluated on the Pascal VOC validation set. We also report the val set accuracy and runtime of a single forward pass in \Cref{ddu:tab:runtime}. Finally, we visualize uncertainty estimates from each baseline in \Cref{ddu:fig:seg_unc_vis}.
Firstly, from \Cref{ddu:tab:runtime}, it is clear that DDU has the runtime of a deterministic model which is significantly faster than both MC Dropout and deep ensembles. Also note that DDU's mIoU is the same as that of the vanilla softmax model.
Secondly, from \Cref{ddu:fig:seg_metrics}, we see that DDU consistently performs better on all 3 evaluation metrics compared to the other baselines. 
Finally, \Cref{ddu:fig:seg_unc_vis} qualitatively validates that DDU's feature-space density captures epistemic uncertainty while the softmax entropy captures aleatoric uncertainty.
For DDU, for the first two samples (first two rows, \Cref{ddu:subfig:seg_unc_vis_density}), the epistemic uncertainty is not high and only aleatoric uncertainty is captured along edges of objects. However, for the last sample (\nth{4} row, \Cref{ddu:subfig:seg_unc_vis_density}), the epistemic uncertainty is high for a relatively large patch on the image which is inaccurately predicted by the model as well. Note that only DDU's feature density is significantly lower for that entire region, whereas softmax entropy does not capture high uncertainty there and is only high along the edges.
These observations are in line with \citep{kendall2017uncertainties}: aleatoric uncertainty is high on edges of objects as they correspond to regions of high ambiguity and noise; on the other hand, epistemic uncertainty is high for regions of the image which are previously unseen.

\subsection{Ablations}

Additional ablations for the CIFAR-10/100 experiments are detailed in \S\ref{ddu:app:additional_exp_results}, \Cref{ddu:table:ood_2} and \ref{ddu:table:ood_3}. We highlight a few results here:

\begin{sidewaystable}[t]
    \begin{minipage}{\linewidth}
        \centering
        \caption{\emph{OoD detection performance of different baselines using ResNet-50, Wide-ResNet-50-2 and VGG-16 architectures on ImageNet vs ImageNet-O \citep{hendrycks2021natural}.} Best AUROC scores are marked in bold.}
        \label{ddu:table:ood_imagenet}
        \resizebox{\linewidth}{!}{%
        \begin{tabular}{cccccccccc}
        \toprule
        \textbf{\small Model} & \multicolumn{2}{c}{{\small Accuracy (\textuparrow)}} & \multicolumn{2}{c}{{\small ECE (\textdownarrow)}} &
        \multicolumn{5}{c}{\textbf{\small AUROC (\textuparrow)}} \\
        & {\small Deterministic} & {\small 3-Ensemble} & {\small Deterministic} & {\small 3-Ensemble} & \textbf{\small Softmax Entropy} & \textbf{\small Energy-based Model} & \textbf{\small DDU} & \textbf{\small 3-Ensemble PE} & \textbf{\small 3-Ensemble MI} \\
        \midrule
        {\small ResNet-50} & $74.8\pm0.05$ & $76.01$ & $2.08\pm0.11$ & $2.07$ & $51.42\pm0.61$ & $55.76\pm0.81$ & $\mathbf{71.29\pm0.08}$ & $60.3$ & $62.43$ \\
        {\small Wide-ResNet-50-2} & $76.75\pm0.11$ & $77.58$ & $1.18\pm0.07$ & $1.22$ & $52.71\pm0.23$ & $57.13\pm0.4$ & $\mathbf{73.12\pm0.19}$ & $60.45$ & $64.81$ \\
        \midrule
        {\small VGG-16} & $72.48\pm0.02$ & $73.54$ & $2.62\pm0.11$ & $2.59$ & $50.67\pm0.22$ & $52.04\pm0.23$ & $54.32\pm0.14$ & $58.74$ & $\mathbf{60.56}$ \\
        \bottomrule
        \end{tabular}}
    \end{minipage}
\end{sidewaystable}

\paragraph{Feature Density.}
These tables along with observations in \Cref{ddu:table:ood_imagenet}, show that \emph{the feature density of a VGG-16 (i.e.\ without residual connections and spectral normalization) is unable to beat a VGG-16 ensemble, whereas a Wide-ResNet-28-10 with spectral normalization outperforms its corresponding ensemble in almost all the cases}. %
This result further validates the importance of having a regularized feature space on the model to obtain smoothness and sensitivity. Also note that, even without spectral normalization, a Wide-ResNet-28 has residual connections built into its model architecture, which can be a contributing factor towards good performance as residual connections make the model sensitive to changes in the input space. 

\begin{sidewaystable*}[!]
    \centering
    \caption{\emph{LDA vs GDA ablation for OoD detection performance using Wide-ResNet-50-2, ResNet-50, Wide-ResNet-50-2 architectures (depending on dataset) on CIFAR-10 vs SVHN/CIFAR-100/Tiny-ImageNet, CIFAR-100 vs SVHN/Tiny-ImageNet, and ImageNet vs ImageNet-O \citep{hendrycks2021natural}.} Best AUROC (\textuparrow) scores are marked in bold. GDA performs better, except with SVHN as OoD dataset.}
    \label{ddu:table:ood_lda_vs_gda}
    \vspace{-2mm}
    \scriptsize
    \begin{tabular}{cccccccc}
    \toprule
    \textbf{Model} & \multicolumn{3}{c}{WRN-28-10} & \multicolumn{2}{c}{WRN-28-10} & WRN-50-2 & RN-50 \\
    \textbf{iD} & \multicolumn{3}{c}{CIFAR-10} & \multicolumn{2}{c}{CIFAR-100} &  \multicolumn{2}{c}{ImageNet} \\
    \cmidrule(l){2-4} \cmidrule(l){5-6} \cmidrule(l){7-8}
    \textbf{OoD} & SVHN & CIFAR-100 & Tiny-ImageNet & SVHN & Tiny-ImageNet & \multicolumn{2}{c}{ImageNet-O} \\
    \midrule
    {\small LDA (Maha \cite{lee2018simple})} & $\mathbf{98.41\pm0.09}$ & $82.90\pm0.23$ & $82.48\pm0.25$ & $\mathbf{92.53\pm0.62}$ & $68.86\pm0.13$ & $64.19\pm0.23$ & $61.68\pm0.14$ \\
    {\small GDA (DDU, ours)} & $97.86\pm0.19$ & $\mathbf{91.34\pm0.04}$ & $\mathbf{91.07\pm0.05}$ & $87.53\pm0.62$ & $\mathbf{83.13\pm0.06}$ & $\mathbf{73.12\pm0.19}$ & $\mathbf{71.29\pm0.08}$ \\
    \bottomrule
    \end{tabular}%
\end{sidewaystable*}

\paragraph{GDA vs.\ LDA.}
We also provide an ablation using LDA \citep{lee2018simple}, which uses a shared covariance matrix over all classes, instead of GDA with covariance matrices per class. The resulting AUROC for Wide-ResNet-28-10 trained on CIFAR-10/100 and for Wide-ResNet-50-2 and ResNet-50 trained on ImageNet in \Cref{ddu:table:ood_lda_vs_gda} in \S\ref{ddu:app:additional_exp_results}. LDA only outperforms GDA when using SVHN as an OoD dataset. In all other cases, GDA obtains significantly higher AUROC, thereby indicating the advantage of modeling density using individual covariance matrices per class.

\paragraph{Objective Mismatch with Wide-ResNet-28-10 on CIFAR-10.}

\begin{table}[t]
    \centering
    \caption{\emph{Objective Mismatch Ablation with WideResNet-28-10 models with and without spectral normalization on CIFAR-10.} While the GDA objective performs much better than cross-entropy objective for feature-space density/epistemic uncertainty estimation, it underperforms for aleatoric uncertainty estimation: both accuracy and in particular ECE are much worse than a regular softmax layer. Averaged over 25 runs.}
    \renewcommand{\arraystretch}{1.2} 
    \scriptsize
    \begin{tabular}{llrr}
    \toprule
        \textbf{Model} & \textbf{Prediction Source} & \textbf{Accuracy in \% (\textuparrow)} & \textbf{ECE (\textdownarrow)} \\
        \midrule
        WideResNet-28-10 & Softmax & $95.98\pm0.02$ & $2.29\pm0.02$ \\
         & GMM & $95.86\pm0.02$ & $4.13\pm0.02$ \\
        WideResNet-28-10+SN & Softmax & $95.97\pm0.03$ & $2.23\pm0.03$ \\
         & GMM & $95.88\pm0.02$ & $4.12\pm0.02$ \\
        \bottomrule
    \end{tabular}%
    \label{ddu:table:objective_mismatch_real}
\end{table}

We further validate \Cref{ddu:pro:objectivemismatch} by running an ablation on Wide-ResNet-28-10 on CIFAR-10. \Cref{ddu:table:objective_mismatch_real} shows that the feature-space density estimator indeed performs worse than the softmax layer for aleatoric uncertainty (accuracy and ECE).

\paragraph{Additional Baselines.}
\label{ddu:app:ddu_additional_baselines}

We provide an ablation with additional baselines on OoD detection for comparison with DDU. In particular, we provide comparisons with Feature Space Singularity (FSSD) \citep{huang2021feature}, Batch Ensemble (BE) \citep{wen2020batchensemble} and SWAG \citep{maddox_2019_simple} using Wide-ResNet-28-10 as additional recent baselines. \citet{huang2021feature} computes the distance to the centroid of noise samples in feature-space together with input perturbations, like in \citep{lee2018simple}. Noise samples count as `far OoD'. \citep{wen2020batchensemble} and \citep{maddox_2019_simple} are computationally cheaper ensembling methods.

We also use the Wide-ResNet-28-10 feature extractor trained using SNGP loss and fit DDU (i.e., GDA) on its feature space. Since SNGP also uses a sensitive smooth feature space with residual connections and spectral normalization, its feature space makes for a good candidate to apply DDU. In \Cref{ddu:table:ddu_additional_baselines_comparison}, we provide the AUROC scores for models trained on CIFAR-10 and CIFAR-100. Broadly, DDU outperforms all competitive baselines. Additionally, we observe a broad improvement in AUROC when DDU is applied on the SNGP feature extractor as compared to vanilla SNGP. However, DDU on a feature extractor trained using softmax loss is still superior to DDU on the SNGP feature extractor.

\begin{table}[t]
    \centering
    \caption{OoD detection ablation with WRN-28-10 model with additional baselines, FSSD \citep{huang2021feature}, Batch Ensemble (BE) \citep{wen2020batchensemble} and SWAG \citep{maddox_2019_simple} as well as using DDU with a feature extractor trained on SNGP. For comparison, we also provide performance for vanilla SNGP, deep ensemble and DDU.}
    \label{ddu:table:ddu_additional_baselines_comparison}
    \scriptsize
    \resizebox{\linewidth}{!}
    {
    \begin{tabular}{lllll}
    \toprule
    \textbf{Train Dataset} & \textbf{Method} & \multicolumn{3}{c}{\textbf{AUROC (\textuparrow)}} \\
    
    & & \textbf{\emph{SVHN}} & \textbf{\emph{CIFAR-100}} & \textbf{\emph{Tiny-ImageNet}} \\
    \midrule
    \multirow{7}{*}{CIFAR-10} & FSSD \citep{huang2021feature} & $97.24$ & $89.88$ & $90.23$ \\
                         & BE \citep{wen2020batchensemble} & $95.36$ & $87.63$ & $88.14$ \\
                         & SWAG \citep{maddox_2019_simple} & $96.37$ & $90.33$ & $90.24$ \\
                         & SNGP \citep{liu2020simple} & $94.0 \pm 1.3$ & $91.13 \pm 0.15$ & $89.97 \pm 0.19$ \\
                         & 5-Ensemble \citep{lakshminarayanan2016simple} & $97.73 \pm 0.31$ & $\mathbf{92.13 \pm 0.02}$ & $90.06 \pm 0.03$ \\
                         & SNGP + DDU & $96.47\pm0.7$ & $89.97 \pm 0.13$ & $90.3 \pm 0.12$ \\
                         & \textbf{DDU (Ours)} & $\mathbf{97.86 \pm 0.19}$ & $91.34 \pm 0.04$ & $\mathbf{91.07 \pm 0.05}$ \\
    \midrule
    \multirow{8}{*}{CIFAR-100} & & \multicolumn{2}{l}{\textbf{\emph{SVHN}}} & \textbf{\emph{Tiny-ImageNet}} \\
    \cmidrule{3-5}
                          & FSSD \citep{huang2021feature} & \multicolumn{2}{l}{$\mathbf{87.64}$} & $82.2$ \\
                          & BE \cite{wen2020batchensemble} & \multicolumn{2}{l}{$86.44$} & $78.33$ \\
                          & SWAG \cite{maddox_2019_simple} & \multicolumn{2}{l}{$81.41$} & $81.67$ \\
                          & SNGP \cite{liu2020simple} & \multicolumn{2}{l}{$85.71 \pm 0.81$} & $78.85 \pm 0.43$ \\
                          & 5-Ensemble \cite{lakshminarayanan2016simple} & \multicolumn{2}{l}{$79.54 \pm 0.91$} & $82.95 \pm 0.09$ \\
                          & SNGP + DDU & \multicolumn{2}{l}{$87.34 \pm 0.76$} & $79.62 \pm 0.36$ \\
                          & \textbf{DDU (Ours)} & \multicolumn{2}{l}{$87.53 \pm 0.62$} & $\mathbf{83.13 \pm 0.06}$ \\
    \bottomrule
    \end{tabular}
    }
\end{table}

\section{Comparison to Prior Work}
\label{ddu:sec:related work}

Several existing approaches model uncertainty using feature-space density but require fine-tuning on OoD data. This chapter has identified feature collapse and objective mismatch as possible reasons for this.

Among these approaches, we have already discussed Mahalanobis distances, DUQ, and SNGP \citep{lee2018simple,van2020uncertainty,lee2020predictive} and the important findings they provide. In this section, we contrast them to the approach presented in this chapter.
For one, the best results of \citet{lee2018simple} require input perturbations, ensembling GMM densities from multiple layers, and fine-tuning on OoD hold-out data. \citet{lee2018simple} do not discuss any constraints which the ResNet feature encoder should satisfy, and therefore, are vulnerable to feature collapse---we recall that in \Cref{ddu:fig:intro_gmm}, for example, the feature density of a LeNet and a VGG are unable to distinguish OoD from iD samples. 
Our method also improves upon \citet{van2020uncertainty} and \citet{liu2020simple} by alleviating the need for additional hyperparameters:
DDU only needs minimal changes from the standard softmax setup to outperform DUQ and SNGP on uncertainty benchmarks, and our GMM parameters are optimized for the already trained model using the training set. 
The insights in \S\ref{ddu:sec:motivation} might also explain why \citet{liu2020simple} found that an ablation that uses \emph{the softmax entropy instead of the feature-space density} of a deterministic network with bi-Lipschitz constraints underperforms.

Among other related works---there are many, and we can only highlight very few here---\citet{postels2020quantifying}, \citet{liu2020energy}, and \citep{winkens2020contrastive} are the most relevant:
\citep{postels2020quantifying} propose a density-based estimation of aleatoric and epistemic uncertainty. Similar to \citep{lee2018simple}, they do not constrain their pre-trained ResNet encoder.
They do discuss feature collapse though, noting that they do not address this problem. 
They also do not consider the objective mismatch that arises (see \Cref{ddu:pro:objectivemismatch} below) and use a single estimator for both epistemic and aleatoric uncertainty. Consequently, they report worse epistemic uncertainty: 74\% AUROC on CIFAR-10 vs SVHN, which we show to considerably fall behind modern approaches for uncertainty estimation in deep learning in \S\ref{ddu:sec:experiments}.
Indeed, \citet{postels2020quantifying} report that in their case deeper layers provide better aleatoric uncertainty while shallower layers provide better epistemic uncertainty. This might be a result of the objective mismatch and not regularizing the feature space using appropriate inductive biases 
Likewise, \citet{liu2020energy} compute an unnormalized density based on the softmax logits without taking into account the need for inductive biases to ensure smoothness and sensitivity of the feature space.
Finally, \citet{winkens2020contrastive} use contrastive training on the feature extractor before estimating the feature-space density.
Our method is orthogonal to this work as we restrict ourselves to the supervised setting and show that the inductive biases that encourage bi-Lipschitzness \citep{van2020uncertainty, liu2020simple} are sufficient for the feature-space density to more reliably capture epistemic uncertainty.

Overall, compared to other methods that require held-out ``OoD data'' for outlier exposure, DDU does not require training or fine-tuning with OoD data in any form.

\section{Discussion}
\label{ddu:sec:conclusion}
We began this chapter by looking at uncertainty quantification from a conceptual view, comparing how active learning and OoD detection use uncertainty quantification. 
Having identified potential pitfalls when using the predictive entropy of deep ensembles or the softmax entropy of deterministic models as proxy for epistemic uncertainty---these potential pitfalls likely apply in general when using the predictive distribution to measure epistemic uncertainty---we investigated using feature-space density as a proxy for epistemic uncertainty and the entropy of the predictive distribution for aleatoric uncertainty, and looked at a possible objective mismatch in detail.

Based on these insights and detailed experiments, a simple method, DDU, was proposed that can obtain reliable epistemic and aleatoric uncertainty estimates for single-pass, deterministic models.
By fitting a GDA to estimate feature-space density after training an off-the-shelf neural network with appropriate inductive biases: residual connections and spectral normalization \citep{lee2018simple, liu2020simple},
our method was able to outperform state-of-the-art deterministic single-pass uncertainty methods in active learning and OoD detection, while performing as well as deep ensembles in several settings.
Hence, DDU provides a very simple method that presents an alternative to deep ensembles without requiring the complexities or computational cost of deep ensembles while still providing reliable uncertainty quantification. 

This is crucial in applications like active learning, which require a reliable estimate of epistemic uncertainty, but reliable uncertainty quantification is also an important requirement to make deep neural nets safe for deployment. Thus, we hope the insights from this chapter can help increase safety, reliability and trust in AI in the future.

}

{
	\graphicspath{{./batchbald/}}

\newcommand{\indep}{\perp \!\!\! \perp }

\newcommand{\theReals}{\mathbb{R}}

\newcommand{\Yi}{\Y_i}
\newcommand{\Yiv}{\Y_{1:i}}
\newcommand{\Yjv}{\Y_{1:j}}

\newcommand{\yi}{\y_i}
\newcommand{\yiv}{\y_{1:i}}
\newcommand{\yjv}{\y_{1:j}}
\newcommand{\yhat}{\hat{y}}

\newcommand{\sxbs}{\x_{1}^{*}, \ldots, \x_{\numB}^{*}}

\newcommand{\ybv}{\y_{1:b}}
\newcommand{\ybs}{\y_1, \ldots, \y_b}
\newcommand{\ybi}{\y_{b+1}}
\newcommand{\ybis}{\y_1, \ldots, \y_{b+1}}
\newcommand{\ybiv}{\y_{1:b+1}}

\newcommand{\xbi}{\x_{b+1}}
\newcommand{\xbv}{\x_{1:b}}
\newcommand{\xbiv}{\x_{1:b+1}}
\newcommand{\xbs}{\x_{1}, \ldots, \x_{b}}
\newcommand{\xbis}{\x_1, \ldots, \x_{b+1}}

\newcommand{\yn}{\y_n}
\newcommand{\ynv}{\y_{1:n}}
\newcommand{\yns}{\y_1, \ldots, \y_n}
\newcommand{\yni}{\y_{n+1}}
\newcommand{\ynis}{\y_1, \ldots, \y_{n+1}}
\newcommand{\yniv}{\y_{1:n+1}}

\newcommand{\ynd}{y_{n-1}}
\newcommand{\ynds}{y_1, \ldots, y_{n-1}}
\newcommand{\yndv}{y_{1:n-1}}

\newcommand{\xn}{\x_{n}}
\newcommand{\xnd}{\x_{n-1}}
\newcommand{\xni}{\x_{n+1}}
\newcommand{\xnv}{\x_{1:n}}
\newcommand{\xns}{\x_1, \ldots, \x_n}
\newcommand{\xniv}{\x_{1:n+1}}
\newcommand{\xnis}{\x_1, \ldots, \x_{n+1}}

\newcommand{\xndv}{\x_{1:n-1}}
\newcommand{\xnds}{\x_1, \ldots, \x_{n-1}}

\newcommand{\ii}{i}
\newcommand{\jj}{j}

\newcommand{\numK}{k}
\newcommand{\numC}{\numclasses}
\newcommand{\numB}{\batchvar}
\newcommand{\numM}{m}
\newcommand{\numN}{n}

\newcommand{\chainedE}[2]{\simpleE{#2}{#1}}

\newcommand{\pw}{\pof{\w}}

\newcommand{\wj}{\hat{\pmb{\omega}}_\jj}

\newcommand{\pyxh}{\pof{\ybv \given \xbv}}
\newcommand{\pyxha}{\pof{\ybiv \given \xbiv}}
\newcommand{\qyxa}{\qprobc{\ybi}{\xbi}}

\newcommand{\pyxaw}{\pof{\ybi \given \xbi \w}}
\newcommand{\pyxhw}{\pof{\ybv \given \xbv \w}}

\newcommand{\si}{s}
\newcommand{\ybvrs}{\hat{\y}_{1:\numN, s}}

\newcommand{\ybvars}{\hat{\y}_{1:\numN+1, s}}
\newcommand{\yar}{\hat{\y}_{\numN+1}}

\newcommand{\yr}{\hat{\y}}
\newcommand{\ynr}{\yr_{\numN}}
\newcommand{\ynvr}{\yr_{1:\numN}}
\newcommand{\yndvr}{\yr_{1:\numN-1}}
\newcommand{\pwfuchsia}{{\leavevmode\color{Sepia}\pof{\w}}}
	\begin{savequote}[75mm]
		Batch learning refined,\\
		With correlations in mind,\\
		Yields diverse knowledge
	\end{savequote}
	\chapter{\baldtitle}
	\label{chapter:batchbald}

In practical active learning applications, instead of single data points, batches of data
points are acquired during each acquisition step to reduce the number of times
the model is retrained and expert-time is requested. Reasons for this are that model retraining becomes a
computational bottleneck for larger models and expert time is
expensive. 
Consider, for example, the effort that goes into commissioning a medical specialist to label
 a single MRI scan, then waiting until the model is retrained, and then commissioning a new
medical specialist to label the next MRI scan, and the extra amount of time this takes.

In \citet{gal2017deep}, \emph{batch acquisition}, i.e. the acquisition of
multiple points, takes the top $\numB$ points with the highest BALD acquisition
score. This naive approach leads to acquiring points that are individually very
informative, but not necessarily so jointly. See
\Cref{batchbald:acquisition_example} for such a batch acquisition of BALD in which it performs
poorly whereas scoring points jointly (`BatchBALD') can find \emph{batches} of
informative data points. Similarly, \Cref{batchbald:mnist_bald_scores_intuition2} provides additional anecdotal evidence that the naive approach to batch acquisition can lead to poor acquisitions.
\Cref{batchbald:rmnist_graph} shows how a dataset
consisting of repeated MNIST digits (with added Gaussian noise) leads BALD to
perform worse than random acquisition while BatchBALD sustains good performance.
\begin{figure}[t]
	\begin{minipage}[t]{0.49\textwidth}
		\centering
		\includegraphics[width=0.8\linewidth]{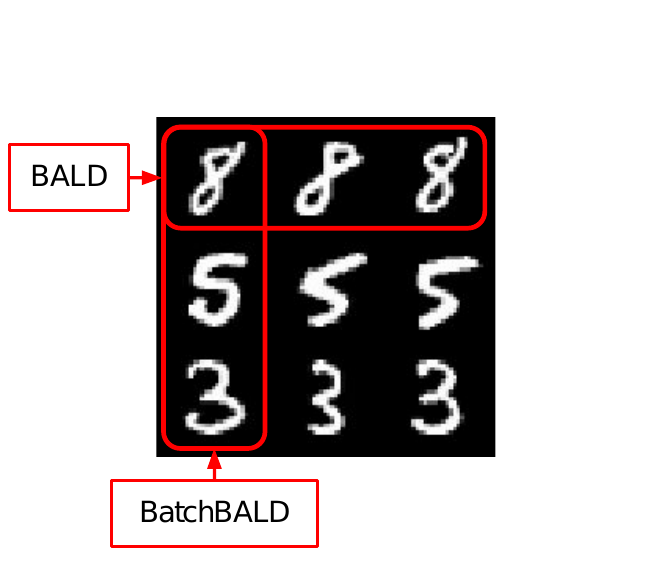}
		\caption{\emph{Idealized acquisitions of BALD and BatchBALD.}
		If a dataset
		were to contain many (near) replicas for each data point, then BALD would select
		all replicas of a single informative data point at the expense of other
		informative data points, wasting data efficiency. }
		\label{batchbald:acquisition_example}
	\end{minipage}
	\hfill
	\begin{minipage}[t]{0.49\textwidth}
		\includegraphics[width=0.95\linewidth]{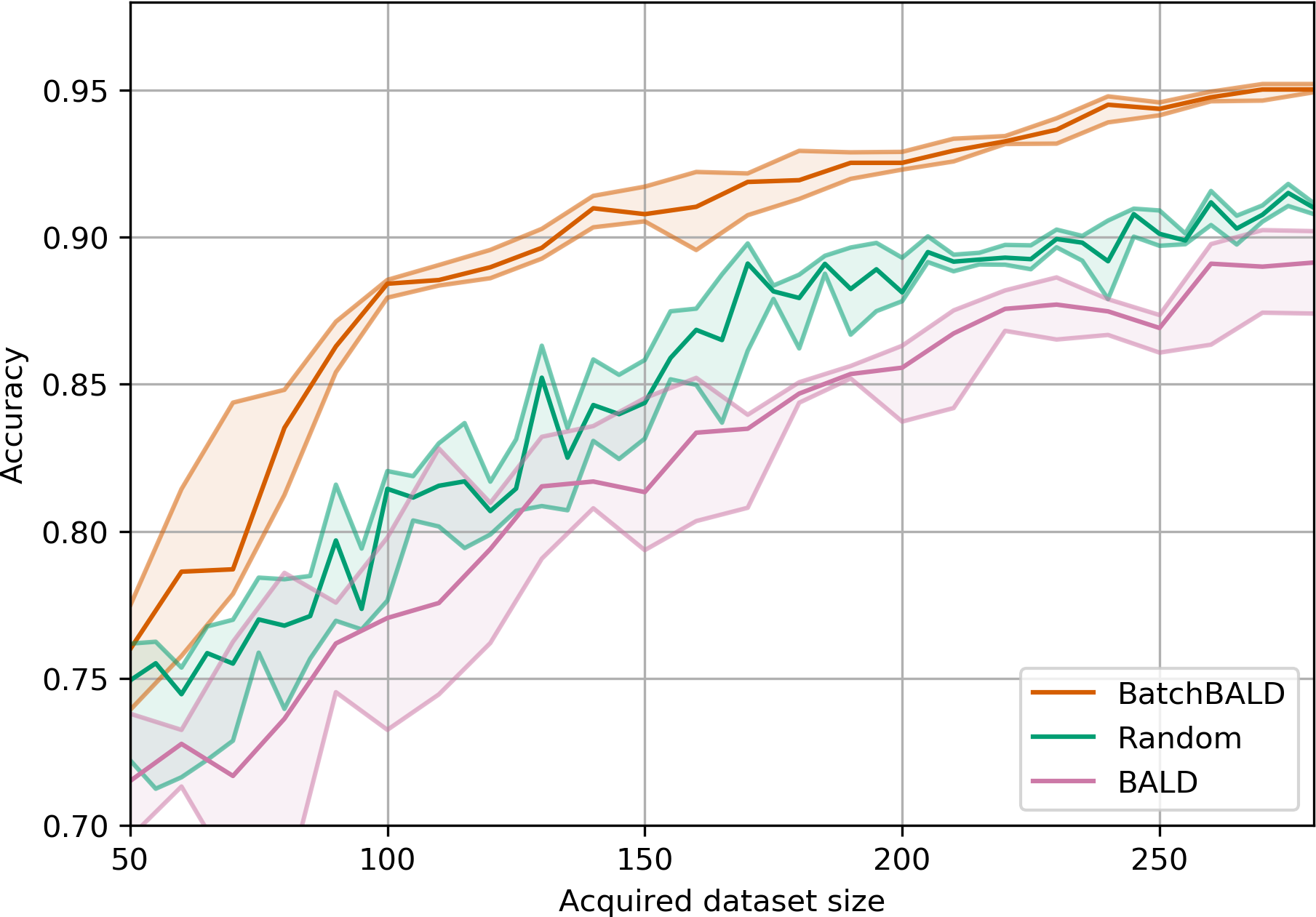}
		\caption{\emph{Performance on \emph{Repeated MNIST} with acquisition size 10}.
		See section \ref{batchbald:repeated_mnist} for further details. BatchBALD
		outperforms BALD while BALD performs worse than random
		acquisition due to the replications in the dataset.}
		\label{batchbald:rmnist_graph}
	\end{minipage}
\end{figure}

\begin{figure}[t]
	\centering
	\includegraphics[width=\linewidth]{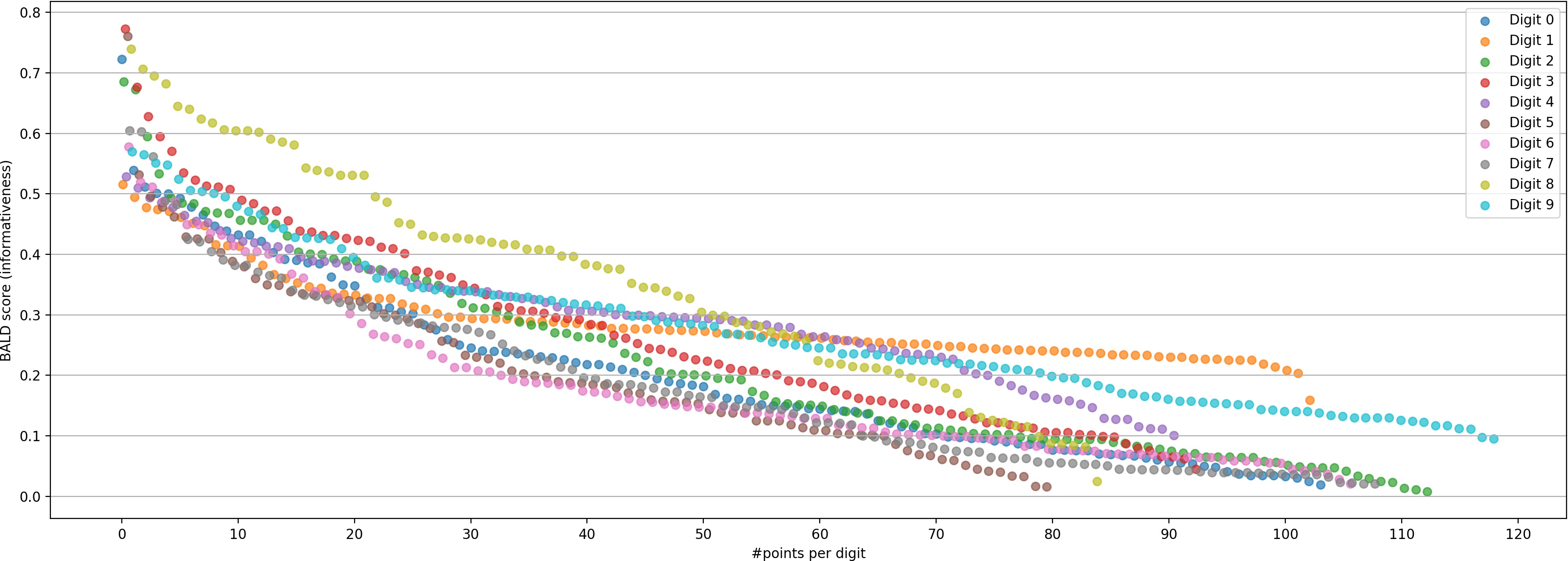}
	\caption{
	\emph{BALD scores for 1000 randomly-chosen points from the MNIST dataset (handwritten digits).} The points are color-coded by digit label and sorted by score. The model used for scoring has been trained to 90\% accuracy first. 
	If we were to pick the top scoring points (e.g. scores above 0.6), most of them would be 8s, even though we can assume that after acquiring the first couple of them our model would consider them less informative than other available data. Points are slightly jittered on the x-axis by digit label to avoid overlaps.
	}
	\label{batchbald:mnist_bald_scores_intuition2}
\end{figure}

Naively finding the best batch to acquire requires enumerating all possible
subsets within the available data, which is intractable as the number of
potential subsets grows exponentially with the acquisition size $\numB$ and the
size of available points to choose from. Instead, we develop a greedy algorithm
that selects a batch in linear time, and show that it is at worst a $1 -
\nicefrac{1}{e}$ approximation to the optimal choice for our acquisition
function. 

The main contributions of this chapter are:
\begin{enumerate}
	\item \emph{BatchBALD}, a data-efficient active learning method that acquires \emph{sets} of high-dimensional image
	data, leading to improved data efficiency and reduced total run time, section \ref{batchbald:batchbald};
	\item a greedy algorithm to select a batch of points efficiently, section \ref{batchbald:batchbald_algorithm}; and
    \item an estimator for the acquisition function that scales to larger acquisition sizes and to datasets with
	many classes, section \ref{batchbald:batchbald_derivation}.
\end{enumerate}
We provide two open-source
implementations of BatchBALD in \shorturl{https://github.com/BlackHC/BatchBALD} and \shorturl{https://github.com/BlackHC/BatchBALD_redux}.

\section{BatchBALD}
\label{batchbald:batchbald}

We propose \emph{BatchBALD} as an extension of BALD whereby we jointly score points by estimating the mutual information between a \emph{joint of multiple data points} and the model parameters:\footnote{
	We use the notation $\MIof{\x, \y;  \z \given c}$ to denote the mutual information between the \emph{joint of the random variables} $\x, \y$ and the random variable $\z$ conditioned on $c$.
	}
\begin{align}
	\BatchBALDAF \left ( \xacqset, \pof{\w \given \Dtrain} \right ) = \MIof{\Yacqset ;  \W \given \xacqset, \Dtrain}.
\end{align}
This builds on the insight that independent selection of a batch of data points
leads to data inefficiency as correlations between data points in an acquisition
batch are not taken into account.

To understand how to compute the mutual information between a set of points and the
model parameters, we express $\Xacqset$, and $\Yacqset$ through joint random
variables in a product probability space and use the definition of the mutual information for two random variables:
\begin{align}
    \MIof{\Yacqset; \W \given \xacqset, \Dtrain} = \Hof{\Yacqset \given \xacqset, \Dtrain} - \Hof{\Yacqset \given \xacqset, \W, \Dtrain}.
	\label{batchbald:eq:batchbald}
\end{align}
Intuitively, the mutual information between two random variables can be seen as the intersection of their information content.
In fact, \citet{yeung1991new} shows that a signed measure $\mu^*$ can be defined for discrete random variables $X$, $Y$, such that
$\MIof{X; Y} = \IMof*{X \cap Y}$, $\Hof{X,Y} = \IMof*{X \cup Y}$, $\Hof{X \given Y} = \IMof*{X \setminus Y}$, and so on,
where we identify random variables with their counterparts in information space.

Using this perspective, BALD can be viewed as the sum of individual intersections $\sum_\ii \IMof*{\Yi \cap \W}$,
which double counts overlaps between the $\Y_i$. Naively extending BALD to the mutual information between $\Yacqset \given \xacqset$ and $\W$,
which is equivalent to $\IMof*{\bigcap_\ii \Y_i \cap \W}$, can lead to selecting \emph{similar} data points instead of diverse ones under maximization.
BatchBALD, on the other hand, takes overlaps into account by computing $\IMof*{\bigcup_\ii \Y_i \cap \W}$
and is more likely to acquire a more diverse cover under maximization:
\begin{align}
	& \medmath{\MIof{\Yacqset ;  \W \given \xacqset, \Dtrain} = \Hof{\Yacqset \given \xacqset, \Dtrain}- \chainedE{\Hof{\Yacqset \given \xacqset, \w, \Dtrain}}{\pof{\w \given \Dtrain}}} \\
	= & \IMof*{\bigcup_\ii \yi} - \IMof*{\bigcup_\ii \Yi \setminus \W} = \IMof*{\bigcup_\ii \Yi \cap \W}
\end{align}
This is depicted in Figure \ref{batchbald:im_intuition} and also motivates that $\BatchBALDAF \le \BALDAF$, which we prove in Appendix \ref{batchbald:bald_approximates_batchbald}.
For acquisition size 1, BatchBALD and BALD are equivalent.

\begin{figure}[t]
	\begin{subfigure}[t]{.49\textwidth}%
		\centering
		\includegraphics[width=0.65\linewidth]{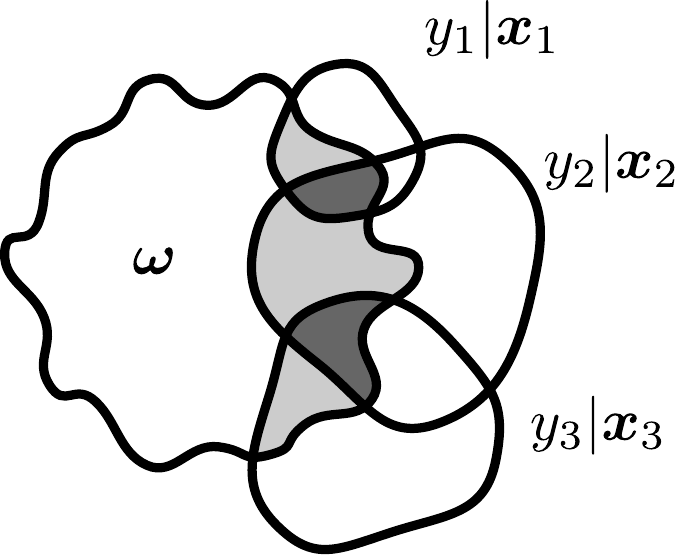} %
	\end{subfigure}%
	\begin{subfigure}[t]{.49\textwidth}%
		\centering
		\includegraphics[width=0.65\linewidth]{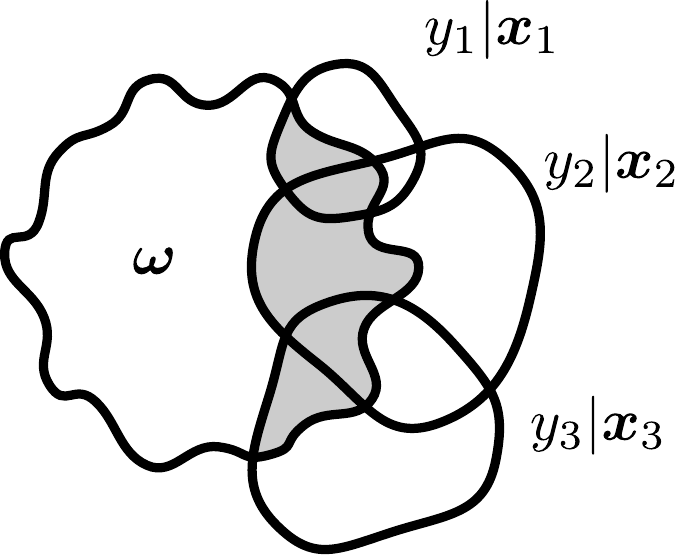}
	\end{subfigure}
	\begin{subfigure}[b]{.49\textwidth}%
		$$\sum_{\ii} \MIof{\Y_i; \W \given \x_i, \Dtrain} = \sum_\ii \IMof*{\Y_i \cap \W}$$
		\caption{\textbf{BALD}}
		\label{batchbald:BALD_intuition}
	\end{subfigure}%
	\begin{subfigure}[b]{.49\textwidth}%
		$$\MIof{\Yset ; \W \given \xset, \Dtrain} = \IMof*{\bigcup_\ii \Y_i \cap \W}$$
		\caption{\textbf{BatchBALD}}
		\label{batchbald:BatchBALD_intuition}
	\end{subfigure}%
	\caption{\emph{Intuition behind \emph{BALD} and \emph{BatchBALD} using I-diagrams \citep{yeung1991new}.}
	\emph{BALD} overestimates the joint mutual information. \emph{BatchBALD}, however, takes the overlap between variables into account and will strive to acquire a better cover of $\W$.
	Areas contributing to the respective score are shown in gray, and areas that are double-counted in dark gray.}
	\label{batchbald:im_intuition}
\end{figure}

\subsection{Greedy Approximation Algorithm for BatchBALD}
\label{batchbald:batchbald_algorithm}

\begin{algorithm}[t]
	\caption{Greedy BatchBALD $1 - \nicefrac{1}{e}$-approximate algorithm \label{batchbald:algo:greedy_batchbald}}
	\includegraphics{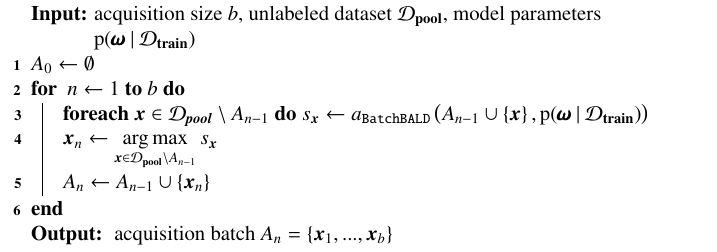}
\end{algorithm}

To avoid the combinatorial explosion that arises from jointly scoring subsets of
points, we introduce a greedy approximation for computing BatchBALD, depicted in
Algorithm \ref{batchbald:algo:greedy_batchbald}. In Appendix \ref{batchbald:submodular_proof}, we
prove that $\BatchBALDAF$ is submodular, which means the greedy algorithm is
$1-\nicefrac{1}{e}$-approximate \citep{cuong2013active,
krause2008near,nemhauser1978analysis}.

\subsection{Approximating $\BatchBALDAF$ via \emph{Consistent} Monte-Carlo Sampling}
\label{batchbald:batchbald_derivation}

\begin{figure}[t]
	\centering
	\includegraphics[width=\linewidth]{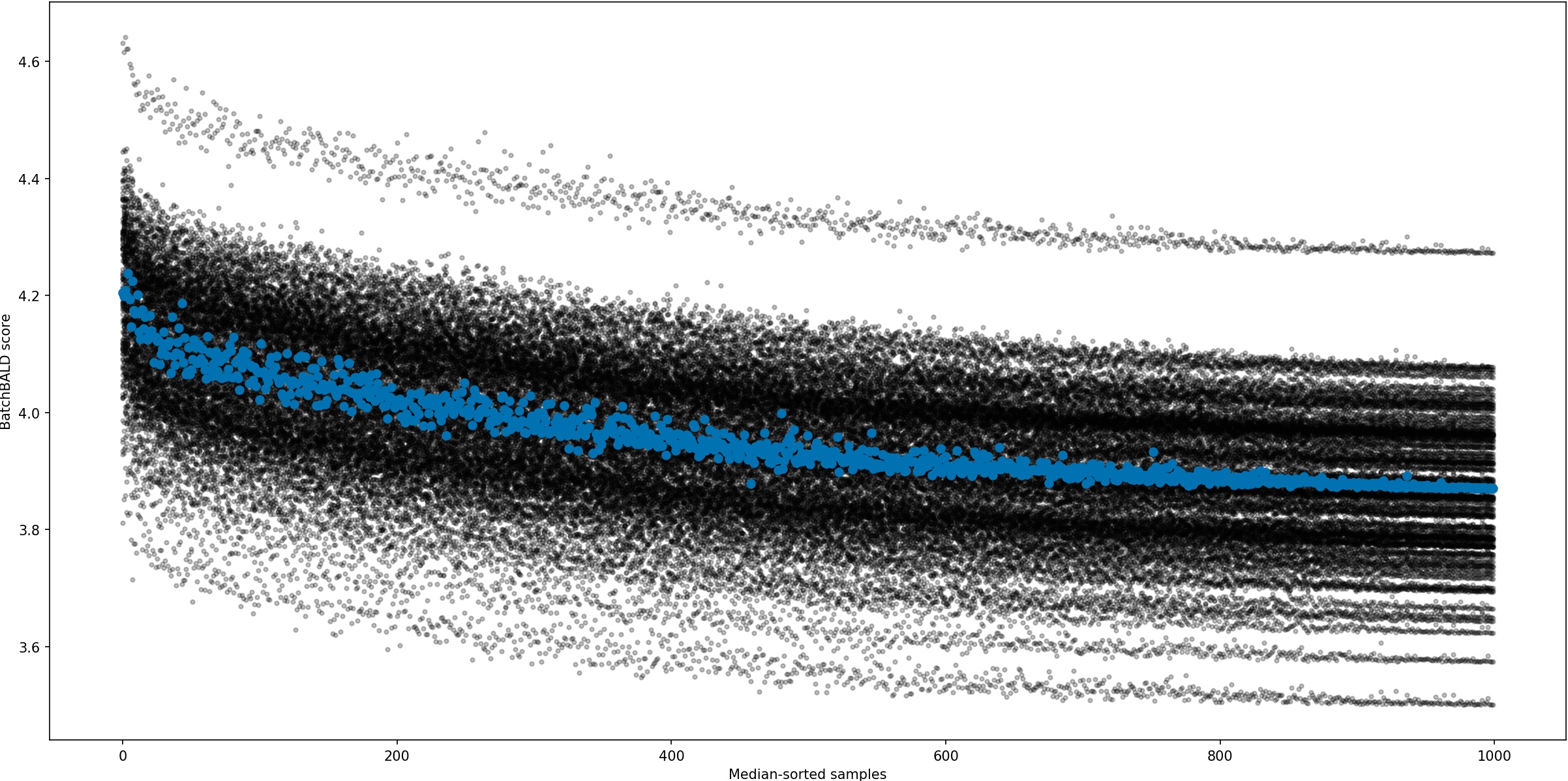}
	\caption{
		\emph{Why \textbf{consistent MC dropout} is necessary: BatchBALD scores for different sets of 100 sampled model parameters.} This shows the BatchBALD scores for 1000 randomly selected points from the pool set while selecting the 10th point in a batch for an MNIST model that has already reached 90\% accuracy. The scores for a single set of 100 model parameters (randomly chosen) are shown in blue. The BatchBALD estimates show strong banding with the score differences between different sets of sampled parameters being larger than the differences between different data points within a specific set of model parameters. Without consistent sampling, the $\argmax$ would essentially be randomly sampled and not be informative.
	}
	\label{batchbald:consistent_mc_dropout_is_necessary}
\end{figure}

{
	\renewcommand{\pw}{\pwfuchsia}

For brevity, we leave out conditioning on $\xset$, and $\Dtrain$, and $\pw$ denotes $\pof{\w \given \Dtrain}$ in this section.
$\BatchBALDAF$ is then written as:
\begin{align}
	\label{batchbald:eq:BB_simple}
    \BatchBALDAF \left ( \left \{ \xset \right \}, \pw \right ) &= \Hof{\Yset} - \Hof{\Yset \given \W} \\
	&= \Hof{\Yset} - \E{\pw}{\Hof{\Yset \given \w}}.
\end{align}
Because the $\Y_i$ are independent when conditioned on $\w$, computing the right
term of equation \eqref{batchbald:eq:BB_simple} is simplified as the conditional joint
entropy decomposes into a sum. We can approximate the expectation using a
Monte-Carlo estimator with $\numK$ samples from our model parameter distribution $\wj \sim \pw$:
\begin{equation}
    \E{\pw}{\Hof{\Yset \given \w}} = \sum_{\ii=1}^{\numN}\E{\pw}{\Hof{\Y_i \given \w}} \approx \frac{1}{\numK} \sum_{\ii=1}^{\numN} \sum_{\jj=1}^{\numK} \Hof{\Y_i \given \wj}.
\end{equation}
\begin{mainresult}
	Crucially, the samples have to stay fixed across different pool samples. This is \emph{not} how Monte-Carlo dropout \citep{gal2015dropout} is usually implemented. We call this \emph{consistent} Monte-Carlo dropout. See \Cref{batchbald:consistent_mc_dropout_is_necessary} for why this is necessary (based on real data).
\end{mainresult}
Computing the left term of equation \eqref{batchbald:eq:BB_simple} is difficult because
the unconditioned joint probability does not factorize. Applying the equality
$\pof{\y} = \E{\pw}{\pof{\y \given \w}}$, and, using sampled $\wj$, we
compute the entropy by summing over all possible configurations $\ynvr$ of
$\ynv$:
\begin{align}
    \Hof{\Yset} &= \E{\pof{\yset}}{-\log{\pof{\yset}}} \\
    &=
    	\simpleE
		{\pw}
		{
			\E{\pof{\yset \given \w}}{-\log{
    			\E{\pw}{\pof{\yset \given \w}}
    		}}
		 } \\
	&\approx -\sum_{\ynvr} \left ( \frac{1}{\numK} \sum_{\jj=1}^{\numK} \pof{\ynvr \given \wj} \right ) \log \left ( {
		\frac{1}{\numK} \sum_{\jj=1}^{\numK} \pof{\ynvr \given \wj} } \right ). \label{batchbald:eq:bb_approx1}
\end{align}

\subsection{Efficient Estimation}
In each iteration of the algorithm, $\xnds$ stay fixed
while $\xn$ varies over $\Dpool \setminus A_{n-1}$. We can reduce the required
computations by factorizing $\pof{\ynv \given \w}$ into $\pof{\yndv \given \w}
\pof{\yn \given \w}$. We store $ \pof{\yndvr \given \wj}$ in a matrix
$\hat{P}_{1:\numN-1}$ of shape $\numC^{\numN-1} \times \numK$ and $\pof{\yn \given \wj}$
in a matrix $\hat{P}_{\numN}$ of shape $\numC \times \numK$. The sum $\sum_{\jj=1}^{\numK}
\pof{\ynvr \given \wj}$ in \eqref{batchbald:eq:bb_approx1} can be then be turned into a matrix product:
\begin{equation}
	\frac{1}{\numK} \sum_{\jj=1}^{\numK} \pof{\ynvr \given \wj} =
	\frac{1}{\numK} \sum_{\jj=1}^{\numK} \pof{\yndvr \given \wj} \pof{\ynr \given \wj} =  \left ( \frac{1}{\numK} \hat{P}_{1:\numN-1} \hat{P}_{\numN}^T \right )_{\yndvr, \ynr}.
\end{equation}
This can be further sped up by using batch matrix multiplication to compute the
joint entropy for different $\xn$. $\hat{P}_{1:\numN-1}$ only has to be
computed once, and we can recursively compute $\hat{P}_{1:\numN}$ using
$\hat{P}_{1:\numN-1}$ and $\hat{P}_{\numN}$, which allows us to sample $\pof{\y \given \wj}$
 for each $\x \in \Dpool$ only once at the beginning of the algorithm.

\newcommand{\cDpool}{{| \Dpool |}}

For larger acquisition sizes, we use $\numM$ MC samples of $\yndv$ as enumerating all
possible configurations becomes infeasible. See Appendix \ref{batchbald:batchbald_mc_approx} for details.

Monte-Carlo sampling bounds the time complexity of the full BatchBALD algorithm to $\mathcal{O} (\numB \numC \cdot \min \{ \numC^\numB, \numM \} \cdot \cDpool \cdot \numK)$ compared to $\mathcal{O} (\numC^\numB \cdot \cDpool^\numB \cdot \numK)$ for naively finding the exact optimal batch and
$\mathcal{O} ( \left ( \numB + \numK \right ) \cdot \cDpool)$ for BALD\footnote{
	$\numB$ is the acquisition size, $\numC$ is the number of classes,
	$\numK$ is the number of MC dropout samples, and
	$\numM$ is the number of sampled configurations of $\yndv$.
	}.
}

\section{Empirical Validation}
\begin{figure}[t]
	\begin{subfigure}[t]{.49\textwidth}%
		\centering
		\includegraphics[width=0.95\linewidth]{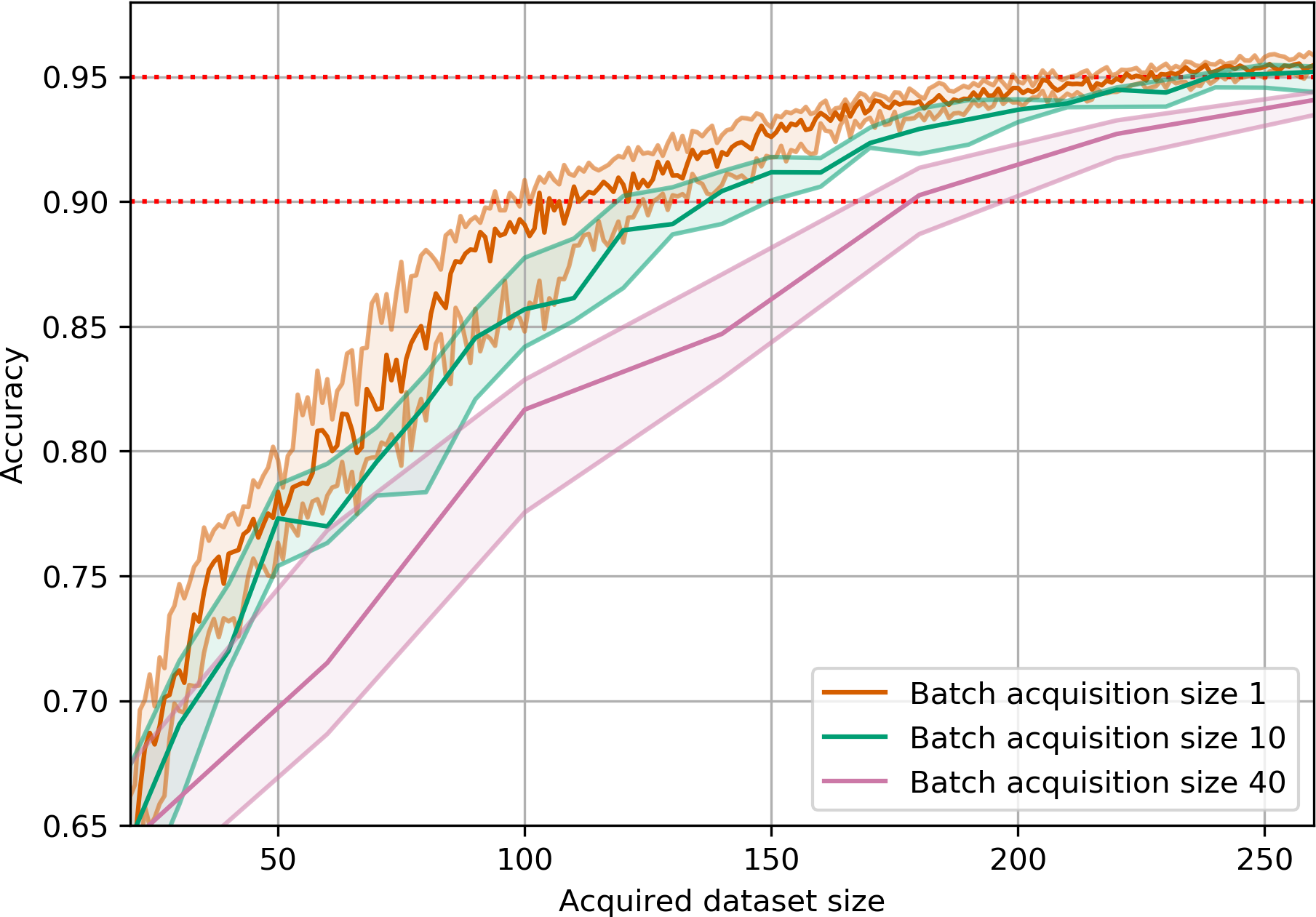} %
		\caption{\textbf{BALD}}
		\label{batchbald:MNIST_BALD}
	\end{subfigure}%
	\hfill
	\begin{subfigure}[t]{.49\textwidth}%
		\centering
		\includegraphics[width=0.95\linewidth]{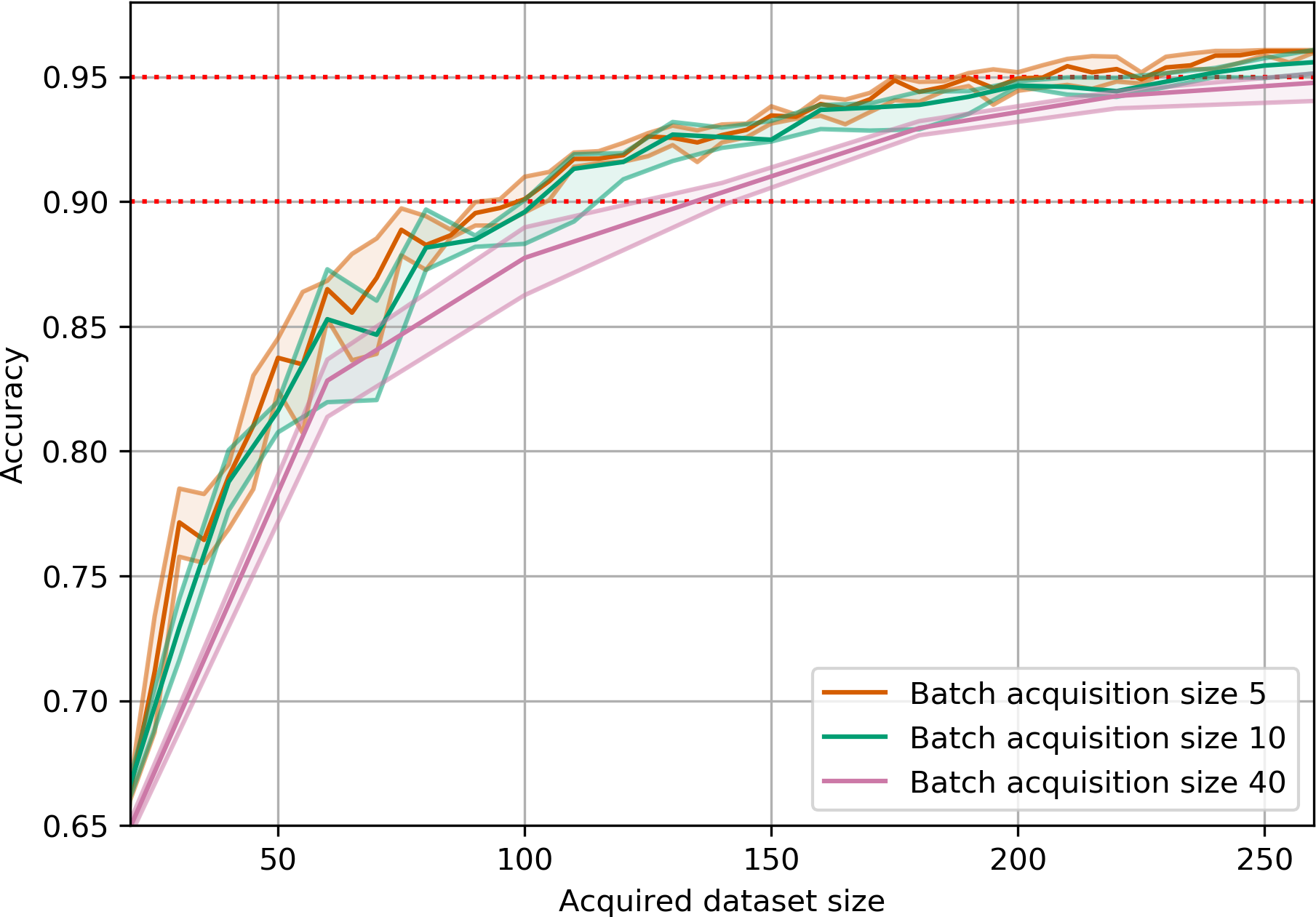}
		\caption{\textbf{BatchBALD}}
		\label{batchbald:MNIST_BatchBALD}
	\end{subfigure}%
	\caption{\emph{Performance on \emph{MNIST} for increasing acquisition sizes.}
	BALD's performance drops drastically as the acquisition size increases.
	BatchBALD maintains strong performance even with increasing acquisition size.}
\end{figure}

In our experiments, we start by showing how a naive application of the BALD
algorithm to an image dataset can lead to poor results in a dataset with many
(near) duplicate data points and show that BatchBALD solves this
problem in a grounded way while obtaining favorable results (Figure
\ref{batchbald:rmnist_graph}).
We then illustrate BatchBALD's effectiveness on standard active learning datasets: MNIST and
EMNIST. EMNIST \citep{cohen2017emnist} is an extension of MNIST that also
includes letters, for a total of 47 classes and has twice as large a training
set. See Appendix \ref{batchbald:emnist_visualisation} for examples of the dataset. We
show that BatchBALD provides a substantial performance improvement in these
scenarios, too, and has more diverse acquisitions. Finally, we look at BatchBALD
in the setting of transfer learning, where we fine-tune a large pretrained model
on a more difficult dataset called CINIC-10 \citep{darlow2018cinic}, which is a
combination of CIFAR-10 and down-scaled ImageNet.

In our experiments, we repeatedly go through active learning loops. One active
learning loop consists of training the model on the available labeled data and
subsequently acquiring new data points using a chosen acquisition function. As the
labeled dataset is small in the beginning, it is important to avoid
overfitting. We do this by using early stopping after 3 epochs of declining
accuracy on the validation set. We pick the model with the highest validation
accuracy. Throughout our experiments, we use the Adam \citep{kingma2014adam}
optimizer with learning rate 0.001 and betas 0.9/0.999. All our results report the median
of 6 trials, with lower and upper quartiles. We use these quartiles to draw
the filled error bars on our figures.

We reinitialize the model after each acquisition, similar to
\citet{gal2017deep}: empirically, we found this helps the model improve even when very small
batches are acquired. It also decorrelates subsequent acquisitions as
final model performance is dependent on a particular
initialization \citep{frankle2018lottery}.

When computing $\pof{\y \given \x, \w, \Dtrain}$, it is important to keep the dropout
masks in MC dropout consistent while sampling from the model. This is
necessary to capture dependencies between the inputs for BatchBALD, and it makes
the scores for different points more comparable by removing this source of
noise. We do not keep the
masks fixed when computing BALD scores because its performance usually benefits
from the added noise. We also do not need to keep these masks fixed for training
and evaluating the model.

In all our experiments, we either compute joint entropies exactly by enumerating
all configurations, or we estimate them using 10,000 MC samples, picking
whichever method is faster. In practice, we compute joint entropies exactly for roughly the
first 4 data points in an acquisition batch and use MC sampling thereafter.
\begin{figure}[tbp]
	\begin{minipage}[t]{0.49\textwidth}
		\centering
		\includegraphics[width=0.95\linewidth]{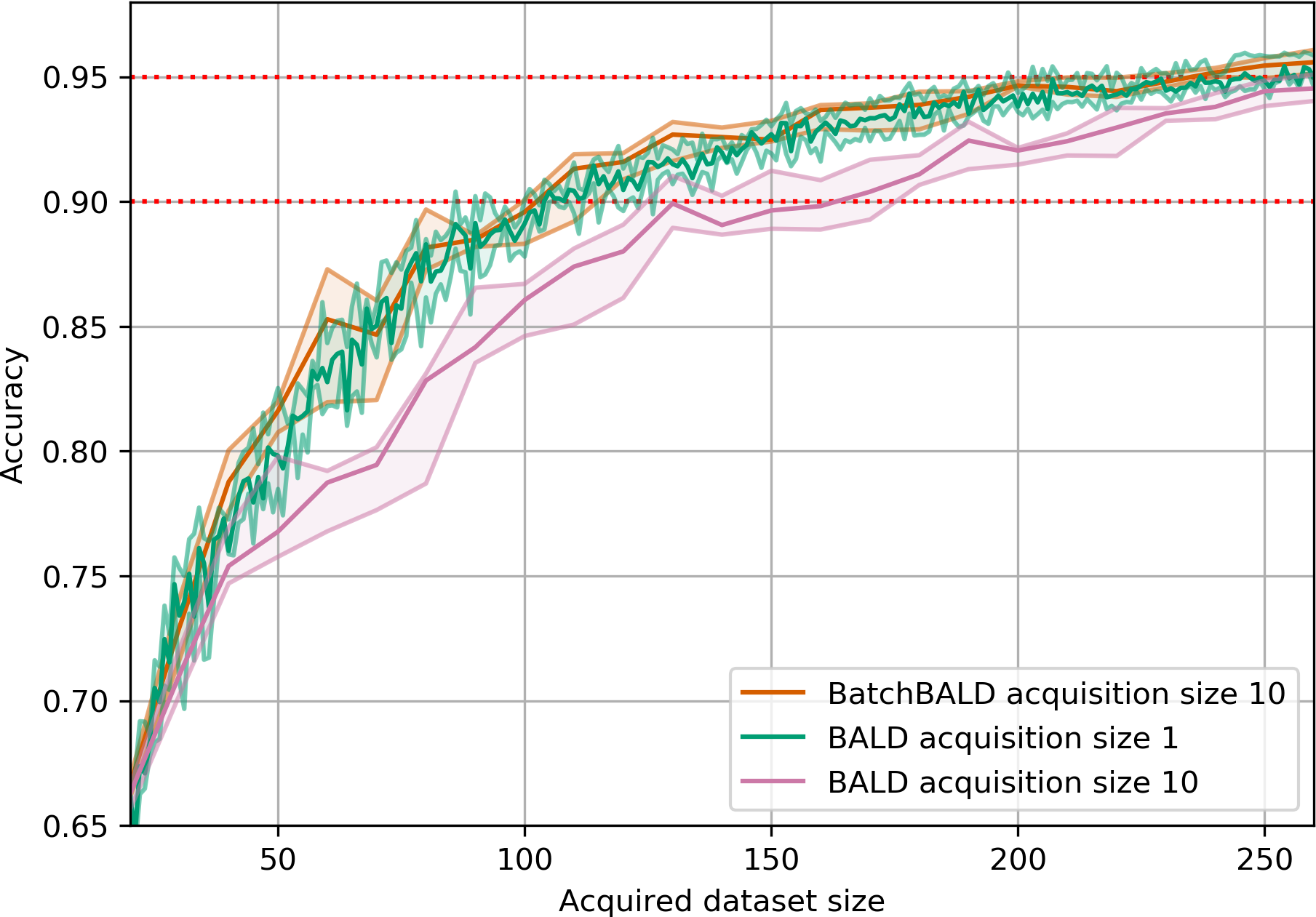}%
		\caption{\emph{Performance on \emph{MNIST}.}
		BatchBALD outperforms BALD with acquisition size 10 and performs close to
		the optimum of acquisition size 1.
		}
		\label{batchbald:MNIST_BALD_MULTIBALD}
	\end{minipage}
	\hfill
	\begin{minipage}[t]{0.49\textwidth}
		\includegraphics[width=0.95\linewidth]{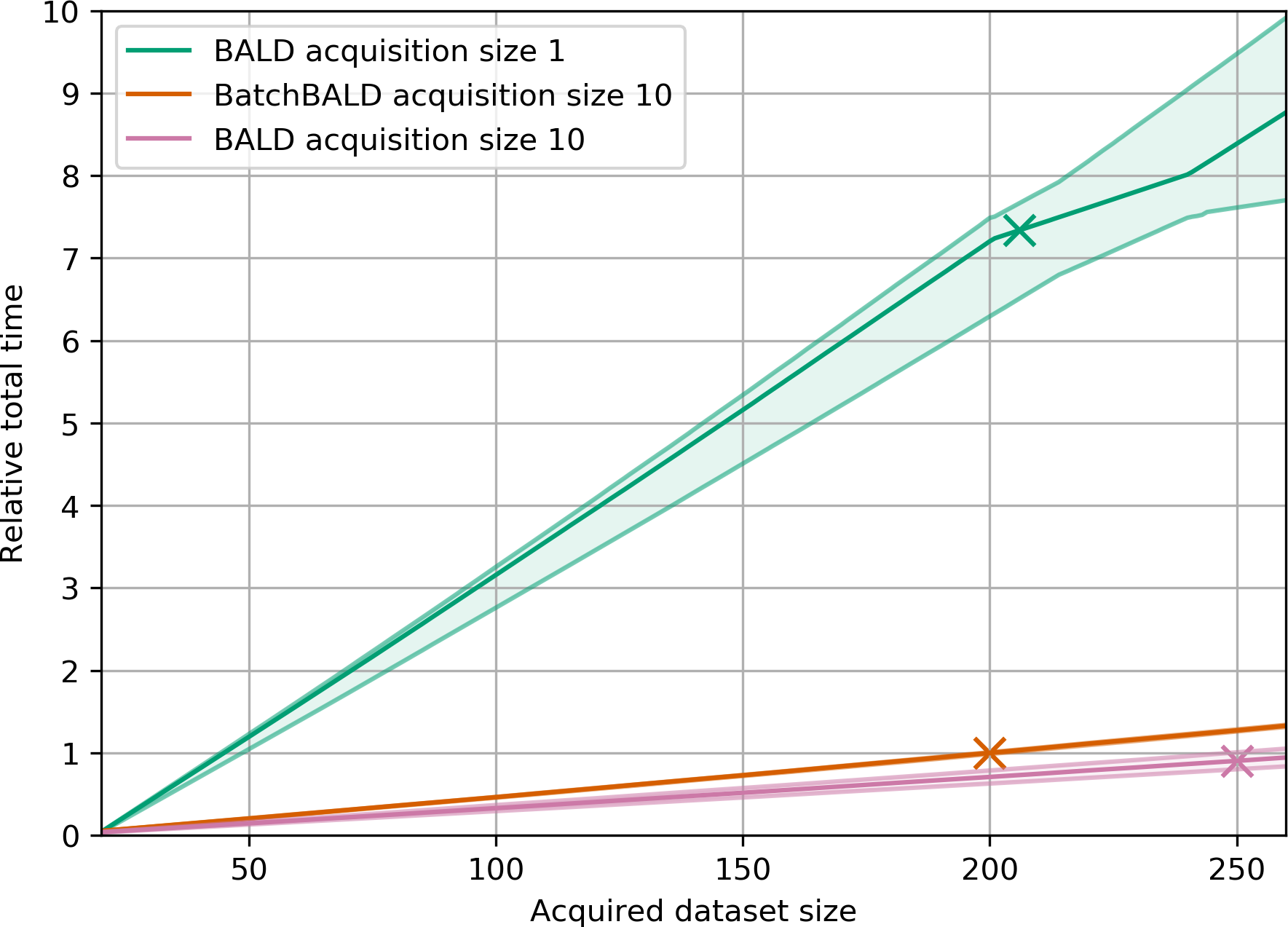}
	   \caption{\emph{Relative total time on \emph{MNIST}.}
	   Normalized to training BatchBALD with acquisition size 10 to 95\% accuracy. The stars mark when 95\%
	   accuracy is reached for each method.
	   }
	   \label{batchbald:time_mnist}
	\end{minipage}
\end{figure}

\subsection{Repeated-MNIST}
\label{batchbald:repeated_mnist}
As demonstrated in the introduction, naively applying BALD to a dataset that
contains many (near) replicated data points leads to poor performance. We show
how this manifests in practice by taking the MNIST dataset and replicating each
data point in the training set twice (obtaining a training set that is three
times larger than the original). After normalizing the dataset, we add isotropic
Gaussian noise with a standard deviation of 0.1 to simulate slight differences
between the duplicated data points in the training set. All results are obtained
using an acquisition size of 10 and 10 MC dropout samples. The initial dataset
was constructed by taking a balanced set of 20 data points\footnote{These
initial data points were chosen by running BALD 6 times with the initial dataset
picked randomly and choosing the set of the median model. They were subsequently
held fixed.}, two of each class, similar to \citep{gal2017deep}.

Our model consists of two blocks of [convolution, dropout, max-pooling, relu],
with 32 and 64 5x5 convolution filters. These blocks are followed by a two-layer
MLP that includes dropout between the layers and has 128 and 10 hidden units.
The dropout probability is 0.5 in all three locations. This architecture
achieves 99\% accuracy with 10 MC dropout samples during test time on the full MNIST dataset.

The results can be seen in Figure \ref{batchbald:rmnist_graph}. In this illustrative
scenario, BALD performs poorly, and even randomly acquiring points performs
better. However, BatchBALD is able to cope with the replication perfectly. In
Appendix \ref{batchbald:ablation_rmnist}, we look at varying the repetition number and
show that as we increase the number of repetitions BALD gradually performs
worse. In Appendix \ref{batchbald:rmnist_comparison_graph}, we also compare with Variation
Ratios \citep{freeman1965elementary}, and Mean STD \citep{kendall2015bayesian}
which perform on par with random acquisition.

\subsection{MNIST}

\begin{table}[t]
	\centering
	\caption{\emph{Number of required data points on \emph{MNIST} until 90\% and 95\% accuracy are reached.} 25\%-, 50\%-
	and 75\%-quartiles for the number of required data points when available.}
	\label{batchbald:tb:mnist}
	\small
	\begin{tabular}{l l l}
	\toprule
	\textbf{Accuracy}		  & \textbf{90\%} & \textbf{95\%} \\
	\midrule
	\emph{BatchBALD} & 70 / 90 / 110     & 190 / 200 / 230                \\
	\emph{BALD} \footnotemark & 120 / 120 / 170   & 250 / 250 / \textgreater{}300  \\
	\emph{BALD} \citep{gal2017deep} &   145   &  335 \\
	\bottomrule
	\end{tabular}
\end{table}
\footnotetext{reimplementation using reported experimental setup}

For the second experiment, we follow the setup of \citet{gal2017deep} and perform
active learning on the MNIST dataset using
100 MC dropout samples. We use the same model architecture and initial dataset
as described in section \ref{batchbald:repeated_mnist}.
Due to differences in model architecture, hyperparameters and model retraining, we
significantly outperform the original results in \citet{gal2017deep} as shown in table \ref{batchbald:tb:mnist}.

We first look at BALD for increasing acquisition size in Figure
\ref{batchbald:MNIST_BALD}. As we increase the acquisition size from the ideal of
acquiring points individually and fully retraining after each point
(acquisition size 1) to 40, there is a substantial performance drop.

BatchBALD, in Figure \ref{batchbald:MNIST_BatchBALD}, is able to maintain performance when
doubling the acquisition size from 5 to 10. Performance drops only slightly at
40, possibly due to estimator noise.

The results for acquisition size 10 for both BALD and BatchBALD are compared in
Figure \ref{batchbald:MNIST_BALD_MULTIBALD}.
BatchBALD outperforms BALD.
Indeed, BatchBALD with acquisition size 10 performs close to the ideal with acquisition size
1.
The total run time of training these three models until 95\%
accuracy is visualized in Figure \ref{batchbald:time_mnist}, where we see that BatchBALD with acquisition size 10
is much faster than BALD with acquisition size 1, and only marginally slower
than BALD with acquisition size 10.

\subsection{EMNIST}
In this experiment, we show that BatchBALD also provides a significant
improvement when we consider the more difficult EMNIST dataset
\citep{cohen2017emnist} in the \emph{Balanced} setup, which consists of 47
classes, comprising letters and digits. The training set consists of 112,800
28x28 images balanced by class, of which the last 18,800 images constitute the validation set.
We do not use an initial dataset and instead perform
the initial acquisition step with the randomly initialized model and use 10 MC dropout samples.

We use a similar model architecture as before, but with added capacity. Three
blocks of [convolution, dropout, max-pooling, relu], with 32, 64 and 128 3x3
convolution filters, and 2x2 max pooling. These blocks are followed by a
two-layer MLP with 512 and 47 hidden units, with again a dropout layer in
between. We use dropout probability 0.5 throughout the model.

The results for acquisition size 5 can be seen in Figure \ref{batchbald:EMNIST_BALD_MULTIBALD}. BatchBALD
outperforms both random acquisition and BALD while BALD is unable to beat
random acquisition. 
Figure \ref{batchbald:entropy_labels_emnist} gives some insight into why BatchBALD performs better than
BALD. The entropy of the categorical distribution of acquired class labels is consistently
higher, meaning that BatchBALD acquires a more diverse set of data points. In
Figure \ref{batchbald:histogram_labels_emnist}, the classes on the x-axis are sorted by
number of data points that were acquired of that class. We see that BALD
undersamples classes while BatchBALD is more consistent.

\begin{figure}[t]
	\begin{minipage}[t]{0.49\textwidth}
		\centering
		\includegraphics[width=0.95\linewidth]{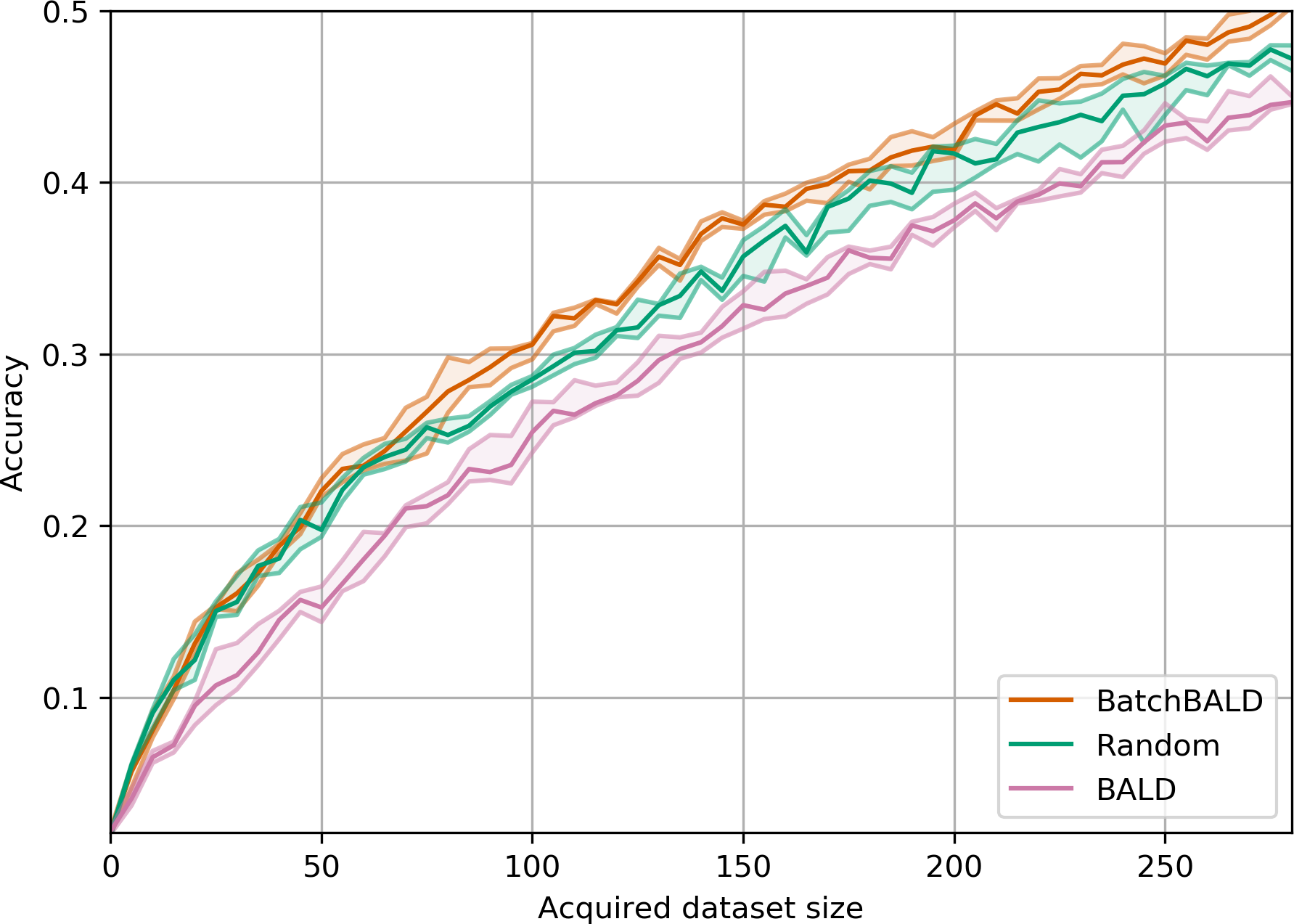}
		\caption{\emph{Performance on \emph{EMNIST}.}
		BatchBALD consistently outperforms both random acquisition and BALD
		while BALD is unable to beat random acquisition.
		}
		\label{batchbald:EMNIST_BALD_MULTIBALD}
	\end{minipage}\hfill
	\begin{minipage}[t]{0.49\textwidth}
		\centering
        \includegraphics[width=0.95\linewidth]{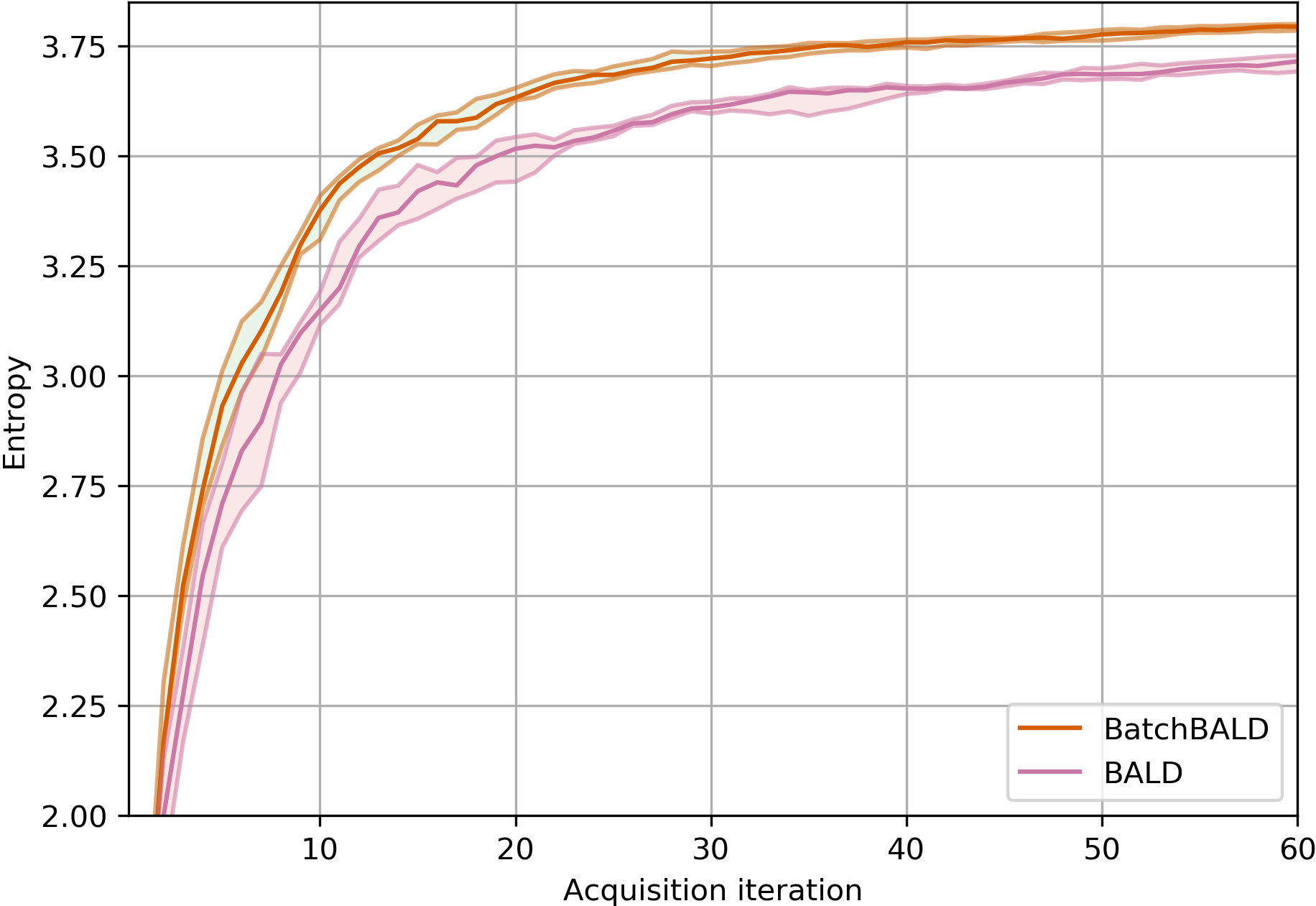}
		\caption{\emph{Entropy of acquired class labels over acquisition steps on \emph{EMNIST}.}
		BatchBALD steadily acquires a more diverse set of data points.
		}
        \label{batchbald:entropy_labels_emnist}
	\end{minipage}
\end{figure}

\begin{figure}[t]%
	\centering
	\includegraphics[width=0.99\linewidth]{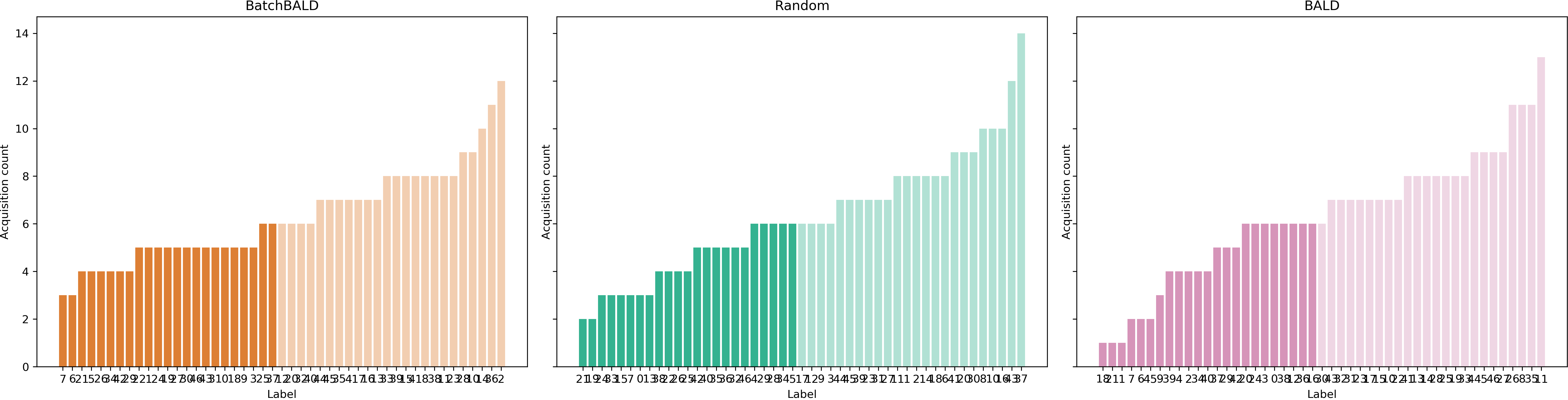}
	\caption{\emph{Histogram of acquired class labels on \emph{EMNIST}.} BatchBALD left and
	BALD right. Classes are sorted by number of acquisitions.
	Several EMNIST classes are underrepresented in BALD and random acquisition while BatchBALD acquires classes
	more uniformly. The histograms were created from all acquired points at the
	end of an active learning loop}
    \label{batchbald:histogram_labels_emnist}
\end{figure}

\begin{figure}[t]
	\centering
	\includegraphics[width=0.4655\linewidth]{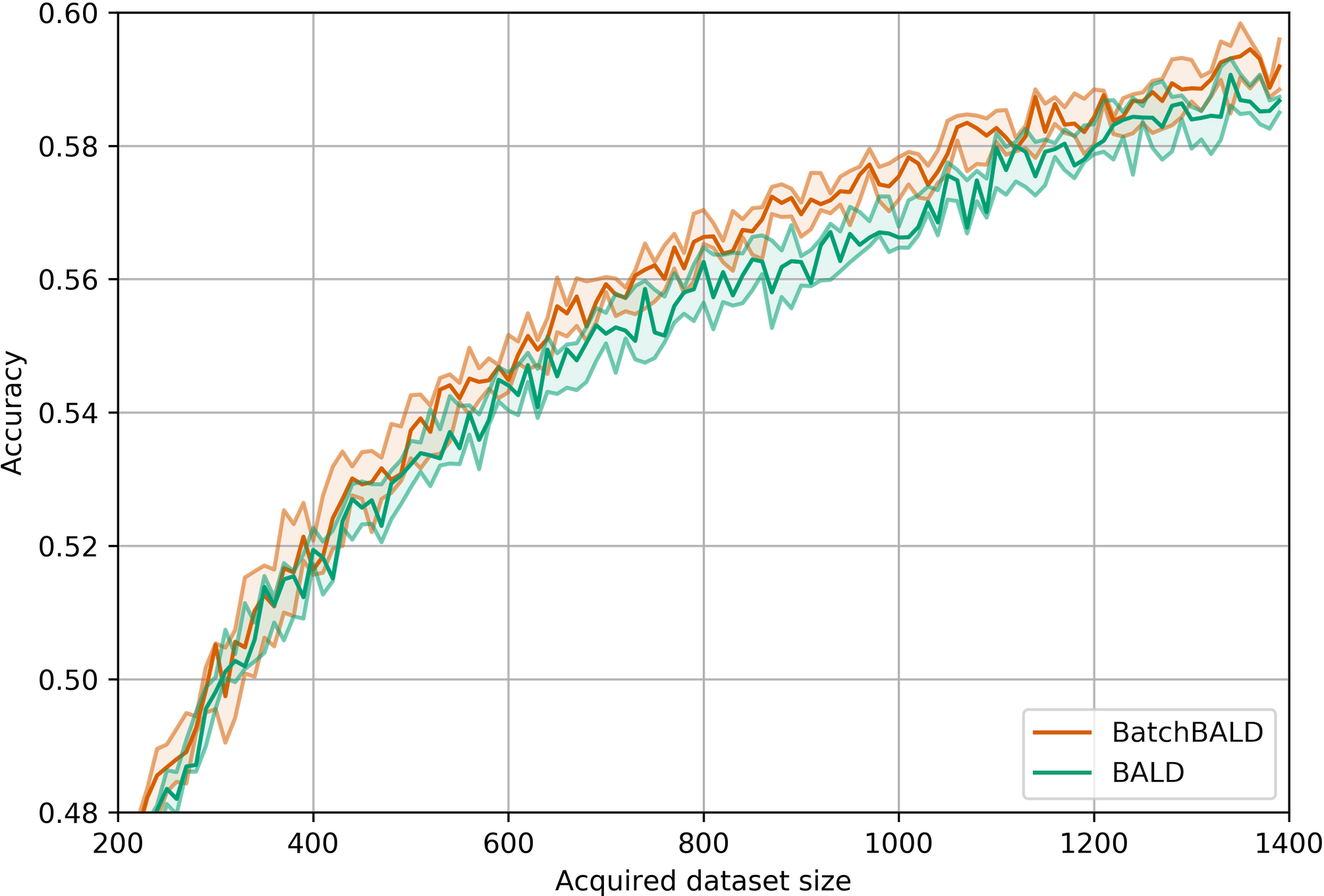}
	\caption{\emph{Performance on \emph{CINIC-10}.}
	BatchBALD outperforms BALD from 500 acquired samples onwards.}
    \label{batchbald:cinic10_results}
\end{figure}

\subsection{CINIC-10}
CINIC-10 is an interesting dataset because it is large (270k data points) and
its data comes from two different sources: CIFAR-10 and ImageNet. To get strong
performance on the test set it is important to obtain data from both sets.
Instead of training a very deep model from scratch on a small dataset, we opt to
run this experiment in a transfer learning setting, where we use a pretrained
model and acquire data only to fine-tune the original model. This is common
practice and suitable in cases where there is plenty of data available in an auxiliary domain, but it is expensive to label data for the domain of interest.

For the CINIC-10 experiment, we use 160k training samples for the unlabeled
pool, 20k validation samples, and the other 90k as test samples. We use an
ImageNet pretrained VGG-16, provided by PyTorch \citep{paszke2017automatic}, with a
dropout layer before a 512 hidden unit (instead of 4096) fully connected layer.
We use 50 MC dropout samples, acquisition size 10 and repeat the experiment for
6 trials. The results are in Figure \ref{batchbald:cinic10_results}, with the 59\% mark reached at
1170 for BatchBALD and 1330 for BALD (median).

\section{Discussion}
We have introduced a new batch acquisition function, BatchBALD, for Deep Bayesian
Active Learning, and a greedy algorithm that selects good candidate batches compared to
the intractable optimal solution. Acquisitions show
increased diversity of data points and improved performance over BALD and other
methods.

While our method comes with additional computational cost during acquisition,
BatchBALD is able to significantly reduce the number of data points that need to
be labeled and the number of times the model has to be retrained, potentially
saving considerable costs and filling an important gap in practical Deep
Bayesian Active Learning.

This proposed method of batch acquisition has some weaknesses and limitations, which we discuss here.

\begin{itemize}	
	\item \textbf{Unbalanced Datasets.} BALD and BatchBALD do not work well when the test
set is unbalanced as they aim to learn well about all classes and do not follow
the density of the dataset. However, if the test set is balanced, but the
training set is not, we expect BatchBALD to perform well.

\item \textbf{Unlabeled Data.} BatchBALD does not take into account any information
from the unlabeled dataset. However, BatchBALD uses the underlying Bayesian
model for estimating uncertainty for unlabeled data points, and semi-supervised
learning could improve these estimates by providing more information about the
underlying structure of the feature space. We leave a semi-supervised extension
of BatchBALD to future work.

\item \textbf{Noisy Estimator.} A significant amount of noise is introduced by
MC-dropout's variational approximation to training BNNs. Sampling of the joint
entropies introduces additional noise. The quality of larger acquisition batches
would be improved by reducing this noise. We examine this further in \S\ref{chapter:stochacq} and \ref{chapter:obi}.
\end{itemize}

}

{
	\graphicspath{{./stochacq/}}

\newcommand{\Ipool}{{\mathcal{I}^\text{pool}}}
\newcommand{\Itrain}{{\mathcal{I}^\text{train}}}

\newcommand{\Dbatch}[1]{\mathcal{D}^\text{acq}(#1)}
\newcommand{\yDeval}{\mathcal{D}^\text{eval}(\yevalset)}

\newcommand{\w}{\omega}
\newcommand{\W}{\Omega}

\newcommand{\Ypred}{Y}
\newcommand{\Ytrue}{\hat{Y}}
\newcommand{\ypred}{y}
\newcommand{\ytrue}{\hat{y}}

\newcommandPIE{\xeval}{x^{\text{eval}}#2#3#1}
\newcommandPIE{\xtest}{x^{\text{test}}#2#3#1}
\newcommandPIE{\xtrain}{x^{\text{train}}#2#3#1}
\newcommandPIE{\xbatch}{x^{\text{acq}}#2#3#1}
\newcommandPIE{\xpool}{x^{\text{pool}}#2#3#1}

\newcommandPIE{\Xeval}{X^{\text{eval}}#2#3#1}
\newcommandPIE{\Xtest}{X^{\text{test}}#2#3#1}
\newcommandPIE{\Xtrain}{X^{\text{train}}#2#3#1}
\newcommandPIE{\Xbatch}{X^{\text{acq}}#2#3#1}
\newcommandPIE{\Xpool}{X^{\text{pool}}#2#3#1}

\newcommandPIE{\yeval}{y^{\text{eval}}#2#3#1}
\newcommandPIE{\ytest}{y^{\text{test}}#2#3#1}
\newcommandPIE{\ytrain}{y^{\text{train}}#2#3#1}
\newcommandPIE{\ybatch}{y^{\text{acq}}#2#3#1}
\newcommandPIE{\ypool}{y^{\text{pool}}#2#3#1}

\newcommandPIE{\Ytest}{Y^{\text{test}}#2#3#1}
\newcommandPIE{\Ytrain}{Y^{\text{train}}#2#3#1}
\newcommandPIE{\Yeval}{Y^{\text{eval}}#2#3#1}
\newcommandPIE{\Ybatch}{Y^{\text{acq}}#2#3#1}
\newcommandPIE{\Ypool}{Y^{\text{pool}}#2#3#1}

\newcommand{\xevalset}{\{x^\text{eval}_i\}_i}
\newcommand{\xtestset}{\{x^\text{test}_i\}_i}
\newcommand{\xtrainset}{\{x^\text{train}_i\}_i}
\newcommand{\xbatchset}{\{x^\text{acq}_i\}_i}
\newcommand{\xpoolset}{\{x^\text{pool}_i\}_i}

\newcommand{\Xevalset}{\{X^\text{eval}_i\}_i}
\newcommand{\Xtestset}{\{X^\text{test}_i\}_i}
\newcommand{\Xtrainset}{\{X^\text{train}_i\}_i}
\newcommand{\Xbatchset}{\{X^\text{acq}_i\}_i}
\newcommand{\Xpoolset}{\{X^\text{pool}_i\}_i}

\newcommand{\yevalset}{\{y^\text{eval}_i\}_i}
\newcommand{\ytestset}{\{y^\text{test}_i\}_i}
\newcommand{\ytrainset}{\{y^\text{train}_i\}_i}
\newcommand{\ybatchset}{\{y^\text{acq}_i\}_i}
\newcommand{\ypoolset}{\{y^\text{pool}_i\}_i}

\newcommand{\Ytestset}{\{Y^\text{test}_i\}_i}
\newcommand{\Ytrainset}{\{Y^\text{train}_i\}_i}
\newcommand{\Yevalset}{\{Y^\text{eval}_i\}_i}
\newcommand{\Ybatchset}{\{Y^\text{acq}_i\}_i}
\newcommand{\Ypoolset}{\{Y^\text{pool}_i\}_i}

\newcommand{\xset}{\{x_i\}_i}
\newcommand{\Xset}{\{X_i\}_i}

\newcommand{\yset}{\{y_i\}_i}
\newcommand{\Yset}{\{Y_i\}_i}

\newcommand{\xevalsetfull}{\{x^\text{eval}_i\}_{i \in \{1, \dots, |\Deval|\}}}
\newcommand{\xtestsetfull}{\{x^\text{test}_i\}_{i \in \{1, \dots, |\Dtest|\}}}
\newcommand{\xtrainsetfull}{\{x^\text{train}_i\}_{i \in \Itrain}}
\newcommand{\xpoolsetfull}{\{x_i\}_{i \in \Ipool}}

\newcommand{\Xevalsetfull}{\{X^\text{eval}_i\}_{i \in \{1, \dots, |\Deval|\}}}
\newcommand{\Xtestsetfull}{\{X^\text{test}_i\}_{i \in \{1, \dots, |\Dtest|\}}}
\newcommand{\Xtrainsetfull}{\{X^\text{train}_i\}_{i \in \Itrain}}
\newcommand{\Xpoolsetfull}{\{X^\text{pool}_i\}_{i \in \Ipool}}

\newcommand{\yevalsetfull}{\{y^\text{eval}_i\}_{i \in \{1, \dots, |\Deval|\}}}
\newcommand{\ytestsetfull}{\{y^\text{test}_i\}_{i \in \{1, \dots, |\Dtest|\}}}
\newcommand{\ytrainsetfull}{\{y^\text{train}_i\}_{i \in \{1, \dots, |\Dtrain|\}}}

\newcommand{\Ytestsetfull}{\{Y^\text{test}_i\}_{i \in \{1, \dots, |\Dtest|\}}}
\newcommand{\Ytrainsetfull}{\{Y^\text{train}_i\}_{i \in \{1, \dots, |\Dtrain|\}}}
\newcommand{\Yevalsetfull}{\{Y^\text{eval}_i\}_{i \in \{1, \dots, |\Deval|\}}}
\newcommand{\Ypoolsetfull}{\{Y^\text{pool}_i\}_{i \in \{1, \dots, n\}}}

\newcommand{\xsetfull}{\{x_i\}_{i \in I}}
\newcommand{\Xsetfull}{\{X_i\}_{i \in I}}

\newcommand{\ysetfull}{\{y_i\}_{i \in I}}
\newcommand{\Ysetfull}{\{Y_i\}_{i \in I}}

\newcommand{\gumbel}[2]{\text{Gumbel}(#1; #2)}

\newcommand{\toref}[1]{{\color{red}REF #1 }}
\newcommand{\figref}[1]{\hyperref[#1]{Fig.\ \ref*{#1}}}

\newcommand{\seb}[1]{{\leavevmode\color{orange}{\footnotesize SF}{\tiny says}[[#1]]}}
\newcommand{\fred}[1]{{\leavevmode\color{teal}{\footnotesize FB}{\tiny says}[[#1]]}}

\renewcommand{\fred}[1]{}
\renewcommand{\seb}[1]{}
	\begin{savequote}[75mm]
		I have no mind of my own or mouth to speak, yet I reply when you call. In valleys and halls, my voice may be found, but only as an answer without its own sound. What am I?
	\end{savequote}
	\chapter{\stochacqtitle}
	\label{chapter:stochacq}

While many acquisition schemes are designed to acquire labels one at a time \citep{houlsby2011bayesian,gal2017deep}, we have already highlighted the importance of \textit{batch acquisition} in \Cref{glue:background:active_learning,chapter:batchbald}.
Unfortunately, existing batch acquisition schemes are computationally expensive (\Cref{stochacq:tab:runtime}).
Intuitively, this is because batch acquisition schemes face combinatorial complexity when accounting for the interactions between possible acquisition points.
Recent works \citep{ash2020deep, ash2021gone} trade off a principled motivation with various approximations to remain tractable.
A commonly used, though extreme, heuristic is to take the top-\batchvar highest scoring points from an acquisition scheme designed to select a single point.

This chapter introduces a simple baseline for batch active learning that can be competitive with methods that cost orders of magnitude more across a wide range of experimental contexts.
The presented method is motivated by noticing that single-acquisition score methods such as BALD \citep{houlsby2011bayesian} act as a noisy proxy for future acquisition scores (\Cref{stochacq:fig:bald_full_rank_correlations}).
This observation leads us to stochastically acquire points following a distribution determined by the single-acquisition scores.
Importantly, such a simple approach can match the prior state of the art for batch acquisition despite being very simple.
Moreover, this acquisition scheme has a time complexity of only $\mathcal{O}(\poolsize \log \batchvar)$ in the pool size $\poolsize$ and acquisition size \batchvar, just like top-\batchvar acquisition. %

We show empirically that the presented stochastic strategy performs as well as or better than top-\batchvar acquisition with almost identical computational cost on several commonly used acquisition scores, making it a strictly-better batch strategy.
Strikingly, the empirical comparisons between this stochastic strategy and the evaluated more complex methods cast doubt on whether they function as well as claimed.
Concretely, in this chapter, we:
\begin{itemize}
    \item examine a family of three computationally cheap stochastic batch acquisition strategies;
    \item demonstrate that these strategies are preferable to the commonly used top-\batchvar acquisition heuristic; and
    \item identify the failure of existing SotA batch acquisition strategies to outperform this vastly cheaper and more heuristic strategy.
\end{itemize}
In \S\ref{stochacq:sec:problem_setting}, we present active learning notation and commonly used acquisition functions. We propose stochastic extensions in \S\ref{stochacq:sec:method}, relate them to previous work in \S\ref{stochacq:sec:related_work}, and validate them empirically in \S\ref{stochacq:sec:experiments} on various datasets, showing that these extensions are competitive with some much more complex active learning approaches despite being orders of magnitude computationally cheaper. Finally, we validate the underlying theoretical motivation in \S\ref{stochacq:sec:investigation} and discuss limitations in \S\ref{stochacq:sec:discussion}.

\begin{table*}[t]
    \caption{
    \emph{Acquisition runtime (in seconds, 5 trials, $\pm$ s.d.)}.
    The examined stochastic acquisition methods are as fast as top-\batchvar, and \textbf{orders of magnitude} faster than BADGE or BatchBALD.
    Synthetic pool set with $M=10,000$ pool points with 10 classes.
    BatchBALD and BALD with 20 parameter samples.
    }
    \centering
    \small
    \begin{tabular}{ccccc}
    \toprule
        \batchvar & Top-\batchvar & \textbf{Stochastic} & BADGE & BatchBALD  \\
        \midrule
        10 & $0.2\pm0.0$ & $0.2\pm0.0$ & $9.2\pm0.3$ & $566.0\pm17.4$ \\
        100 & $0.2\pm0.0$ & $0.2\pm0.0$ & $82.1\pm2.5$ & $5,363.6\pm95.4$ \\
        500 & $0.2\pm0.0$ & $0.2\pm0.0$ & $409.3\pm3.7$ & $29,984.1\pm598.7$\\
        \bottomrule
    \end{tabular}
    \label{stochacq:tab:runtime}
\end{table*}

\section{Problem Setting}
\label{stochacq:sec:problem_setting}

The stochastic approach we examine applies to batch acquisition for active learning in a pool-based setting \citep{Settles2010} where we have access to a large unlabeled \emph{pool} set, but we can only label a small subset of the points.
The challenge of active learning is to use what we already know to pick which points to label in the most efficient way, and generally, we want to avoid labelling points similar to those already labeled. %

\paragraph{Notation.} Following \citet{farquhar2020statistical}, and unlike in the rest of the thesis, we formulate active learning over \emph{indices} instead over data points. This simplifies the notation.
The large, initially fully unlabeled, pool set containing \poolsize input points is 
\begin{equation}
    \Dpool=\xpoolsetfull,
\end{equation}
where $\Ipool = \{1, \ldots, \poolsize\}$ is the initial full index set.
We initialize a training dataset with $\trainsize_0$ randomly selected points from $\Dpool$ by acquiring their labels, $\y_i$,
\begin{equation}
    \Dtrain=\{ ( x_i, y_i) \}_{i \in \Itrain},
\end{equation}
where $\Itrain$ is the index set of $\Dtrain$, \emph{initially} containing $\trainsize_0$ indices between $1$ and \poolsize.
A model of the predictive distribution, $\pof{\ypred \given \x}$, can then be trained on $\Dtrain$.

\paragraph{Active Learning.} At each acquisition step, we select additional points for which to acquire labels.
Although many methods acquire one point at a time \citep{houlsby2011bayesian, gal2017deep}, one can alternatively acquire a whole batch of \batchvar examples.
An acquisition function $a$ takes $\Itrain$ and $\Ipool$ and returns \batchvar indices from $\Ipool$ to be added to $\Itrain$.
We then label those \batchvar data points and add them to $\Itrain$ while making them unavailable from the pool set.
That is,
\begin{align}
    & \Itrain \leftarrow \Itrain \cup a(\Itrain, \Ipool), \\
    & \Ipool \leftarrow \Ipool \setminus \Itrain.
\end{align}
A common way to construct the acquisition function is to define some scoring function, $s$, and then select the point(s) that score the highest.

We consider the following scoring functions:
\begin{description}
    \item[BALD.]  
For each candidate pool index, $i$, the BALD score is
\begin{align}
    s_{\textup{BALD}}(i; \Itrain) &\defeq \MIof{\Ypred; \W \given X=x_i, \Dtrain} \nonumber\\
    &= \Hof{\Ypred \given X=x_i, \Dtrain} - \E{\pof{\w \given \Dtrain}}{\Hof{\Ypred \given X=x_i, \w, \Dtrain}}.
\end{align}
\item[Entropy.] Entropy does not require Bayesian models, unlike BALD, and performs worse for data with high observation noise as we have noted in \Cref{chapter:ddu}.
It is identical to the first term of the BALD score
\begin{align}
    s_{\textup{entropy}}(i; \Itrain) &\defeq \Hof{\Ypred \given X=x_i, \Dtrain}.
\end{align}
\end{description}
See \Cref{glue:background:active_learning} for other acquisition functions.

\paragraph{Acquisition Functions.} These scoring functions were introduced for single-point acquisition:
\begin{align}
    a_s(\Itrain) \defeq \argmax_{i \in \Ipool} s(i; \Itrain).
\end{align}
For deep learning in particular, single-point acquisition is computationally expensive due to retraining the model for every acquired sample. Moreover, it also means that labelling can only happen sequentially instead of in bulk. Thus, single-point acquisition functions were expanded to multi-point acquisition via acquisition batches in batch active learning.
The most naive batch acquisition function selects the highest \batchvar scoring points
\begin{align}
    & a^\text{batch}_{s}(\Itrain; \batchvar) \defeq \argmax_{I \subseteq \Ipool, |I| = \batchvar} \sum_{i \in I} s(i; \Itrain).
\end{align}
Maximizing this sum is equivalent to taking the top-\batchvar scoring points, which cannot account for the interactions between points in an acquisition batch because individual points are scored independently.
Some acquisition functions are explicitly designed for batch acquisition, e.g.\ BatchBALD from \S\ref{chapter:batchbald} or BADGE from \citet{ash2020deep}.
They try to account for the interaction between points, which can improve performance relative to simply selecting the top-\batchvar scoring points.
However, existing methods are computationally expensive. For example, BatchBALD rarely scales to acquisition sizes of more than 5--10 points as noted in \S\ref{chapter:batchbald}; see \Cref{stochacq:tab:runtime}.

\paragraph{BADGE.} \citet{ash2020deep} propose \emph{Batch Active learning by Diverse Gradient Embeddings}: it motivates its batch selection approach using a k-Determinantal Point Process \citep{kulesza2011k} based on the (inner product) similarity matrix of the scores (gradients of the log loss) using hard pseudo-labels (the highest probability class according to the model's prediction) for each pool sample. In \Cref{chapter:fim} we provide a more detailed analysis. In practice, they use the initialization step of k-MEANS++ with Euclidian distances between the scores to select an acquisition batch. BADGE is also computationally expensive.

\section{Method}
\label{stochacq:sec:method}
\begin{table}[t]%
    \centering%
    \caption{\emph{Summary of stochastic acquisition variants.} Perturbing the scores $s_i$  themselves with $\epsilon_i \sim \gumbel{0}{\beta^{-1}}$ i.i.d.\ yields a softmax distribution. Log-scores result in a power distribution, with assumptions that are reasonable for active learning. Using the score-ranking, $r_i$ finally is a robustifying assumption. $\beta$ is included for completeness; we use $\beta\defeq 1$ in our experiments---except for the ablation in \S\ref{stochacq:sec:finetuning_beta}.}
    \small
    \begin{tabular}{lll}
        \toprule
        Perturbation & Distribution & Probability mass \\
        \midrule
        $s_i + \epsilon_i$ & Softmax & $\propto \exp{\beta s_i}$ \\
        $\log s_i + \epsilon_i$ & Power & $\propto s_i^\beta$ \\ %
        $- \log r_i + \epsilon_i$ & Soft-rank & $\propto r_i^{-\beta}$ \\
        \bottomrule
    \end{tabular}
    \label{stochacq:tbl:distributions}
\end{table}

\begin{figure*}[t]
    \centering
    \begin{minipage}[b]{.49\textwidth}
        \centering %
        \includegraphics[width=\linewidth]{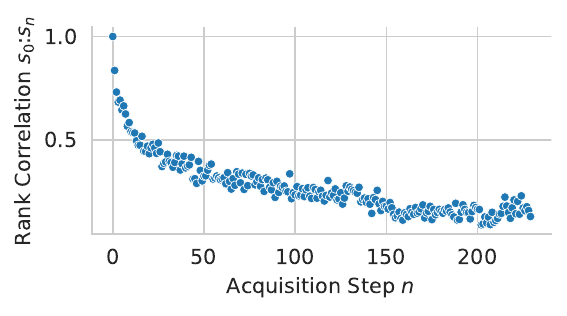} %
        \caption{
            \emph{Early acquisition scores are only a loose proxy for later scores.}
            Specifically, the Spearman rank-correlation between acquisition scores on the first and $n$'th time-step falls with $n$.
            While top-\batchvar acquisition incorrectly implicitly assumes the rank-correlation remains $1$, stochastic acquisitions do not.
            BNN trained on MNIST at initial 20 points and 73\% initial accuracy, score ranks over test set.
        } %
        \label{stochacq:fig:bald_full_rank_correlations}
    \end{minipage}%
    \hfill
    \begin{minipage}[b]{.49\textwidth}
        \centering %
        \includegraphics[width=\linewidth]{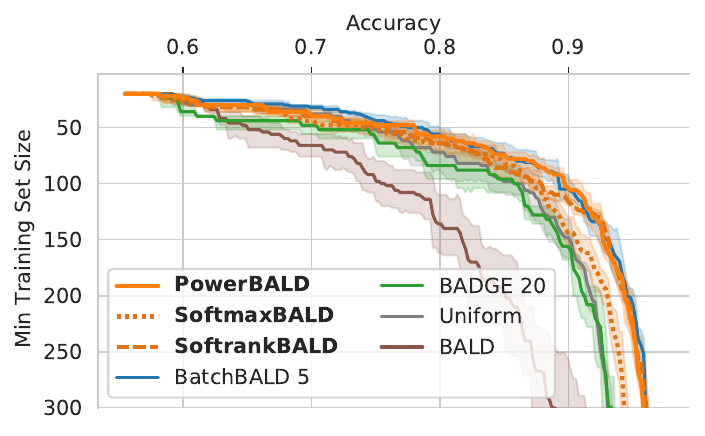} %
        \caption{
        \emph{Performance on Repeated-MNIST with 4 repetitions (5 trials).} \textbf{Up and to the right is better ($\nearrow$).}
        PowerBALD outperforms (top-\batchvar) BALD and BADGE and is on par with BatchBALD.
        This is despite being orders of magnitude faster.
        Acquisition sizes: BatchBALD--5, BADGE--20, others--10. See \Cref{stochacq:appfig:rmnist_badge_ablation} in the appendix for an ablation study of BADGE's acquisition size.
        }
        \label{stochacq:fig:rmnist}
    \end{minipage}
\end{figure*}

We observe that selecting the top-\batchvar points at acquisition step $t$ amounts to the assumption that the informativeness of these points is independent of each other.
Imagine adding the top-\batchvar points at a given acquisition step $t$ to the training set one at a time.
Each time, you retrain the model.
Of course, the acquisition scores for the models trained with these additional points will be different from the first set of scores.
After all, the purpose of active learning is to add the \emph{most informative} points: those that will update the model the most.
Yet selecting a top-\batchvar batch in one step implicitly assumes that the score ranking will not change due to these points.
This is clearly wrong.
We provide empirical confirmation that, in fact, the ranking of acquisition scores at step $t$ and $t+\batchvar$ is decreasingly correlated as \batchvar grows; see \Cref{stochacq:fig:bald_full_rank_correlations}.
Moreover, this effect is the strongest for the most informative points; see \S\ref{stochacq:subsec:rank_correlation} for more details.

Instead, this chapter uses stochastic sampling to acknowledge the uncertainty within the batch acquisition step using a simple noise process model governing how scores change.
We examine three simple stochastic extensions of single-sample scoring functions $s(i; \Itrain)$ that make slightly different assumptions.
These methods are compatible with conventional active learning frameworks that typically take the top-\batchvar highest scoring samples.
For example, it is straightforward to adapt entropy, BALD, and other scoring functions for use with these extensions. 

These stochastic acquisition distributions assume that future scores differ from the current score by a perturbation.
We model the noise distribution of this perturbation as the addition of Gumbel-distributed noise $\epsilon_i \sim \gumbel{0}{1}$, which is used frequently for modelling extrema.
The choice of a Gumbel distribution for the noise is one of mathematical convenience, in the spirit of providing a simple baseline.
For example,
the maximum of sets of many other standard distributions, such as the Gaussian distribution, is not analytically tractable.

Taking the highest-scoring points from this perturbed distribution is equivalent to sampling from a softmax distribution\footnote{Also known as Boltzmann/Gibbs distribution.} without replacement with a `coldness' parameter $\beta \ge 0$, which represents the expected rate at which the scores change as more data is acquired.
This follows from the Gumbel-Max trick \citep{gumbel1954statistical, maddison2014sampling} and, more specifically, the Gumbel-Top-\batchvar trick \citep{kool2019stochastic}.
We provide a short proof in appendix \ref{stochacq:app:proof_gumbel_top_k}.
Expanding on \citet{maddison2014sampling}:
\begin{restatable}{proposition}{gumbeltopk}
\label{stochacq:prop:gumbel_top_k}
For scores $s_i$, $i \in \{1, \ldots, n\}$, and $k \le n$ and $\beta > 0$, if we draw $\epsilon_i \sim \gumbel{0}{\beta^{-1}}$ independently, then $\arg \mathrm{top}_k \{s_i + \epsilon_i\}_i$ is an (ordered) sample without replacement from the categorical distribution $\mathrm{Categorical}(\nicefrac{\exp(\beta \, s_i)}{\sum_j \exp(\beta \, s_j)}, i \in \{1, \ldots, n\})$. 
\end{restatable}

In the spirit of providing a simple and surprisingly effective baseline without hyperparameters, we fix $\beta \defeq 1$.
For $\beta \to \infty$, this distribution will converge towards top-\batchvar acquisition. 
Whereas for $\beta \to 0$, it will converge towards uniform acquisition.
We examine ablations of $\beta$ in \S\ref{stochacq:sec:finetuning_beta}.

We apply the perturbation to three quantities in the three sampling schemes: the scores themselves, the log scores, and the rank of the scores.
Perturbing the log scores assumes that scores are non-negative and uninformative points should be avoided.
Perturbing the ranks can be seen as a robustifying assumption that requires the relative scores to be reliable but allows the absolute scores to be unreliable.
We summarize the three versions with their associated sampling distributions are in \Cref{stochacq:tbl:distributions}.

\paragraph{Soft-Rank Acquisition.} This first variant only relies on the rank order of the scores and makes no assumptions on whether the acquisition scores are meaningful beyond that. It thus uses the \emph{least} amount of information from the acquisition scores. 
It only requires the \textit{relative score order} to be useful and ignores the \textit{absolute score values}.
If the absolute scores provide useful information, we would expect this method to perform worse than the variants below, which make use of the score values. As we will see, this is indeed sometimes the case.

Ranking the scores $s(i ;\Itrain)$ with descending ranks $\{r_i\}_{i \in \Ipool}$ such that $s(r_i;\Itrain) \ge s(r_j;\Itrain)$ for $r_i \le r_j$ and smallest rank being $1$, we sample index $i$ with probability $\pcof{\text{softrank}}{i} \propto r_i^{-\beta}$ with coldness $\beta$. This is invariant to the actual scores.
We can draw $\epsilon_i \sim \gumbel{0}{\beta^{-1}}$ and create a perturbed `rank'
\begin{align}
    s^\text{softrank}(i; \Itrain):= -\log r_i + \epsilon_i.
\end{align}
Taking the top-\batchvar samples is now equivalent to sampling without replacement from the rank distribution $\pcof{\text{softrank}}{i}$.

\paragraph{Softmax Acquisition.} %
The next simplest variant uses the actual scores instead of the ranks. Again, it perturbs the scores by a Gumbel-distributed random variable $\epsilon_i \sim \gumbel{0}{\beta^{-1}}$ 
\begin{align}
    s^\text{softmax}(i; \Itrain):= s(i; \Itrain) + \epsilon_i.
\end{align}
However, this makes no assumptions about the semantics of the absolute values of the scores: the softmax function is invariant to constants shifts. 
Hence, the sampling distribution will only depend on the \emph{relative scores} and not their absolute value.

\paragraph{Power Acquisition.} For many scoring functions, the scores are non-negative, and a score close to zero means that the sample is not informative in the sense that we do not expect it will improve the model---we do not want to sample it.
This is the case with commonly used score functions such as BALD and entropy.
BALD measures the expected information gain. When it is zero for a sample, we do not expect anything to be gained from acquiring a label for that sample. Similarly, entropy is upper-bounding BALD, and the same consideration applies. 
This assumption also holds ideally for other scoring functions that are easily transformed to be non-negative (see \Cref{glue:background:active_learning}).
To take this into account, the last variant models the future log scores as perturbations of the current log score with Gumbel-distributed noise
\begin{align}
    s^\text{power}(i; \Itrain):= \log s(i; \Itrain) + \epsilon_i.
\end{align}
By \Cref{stochacq:prop:gumbel_top_k}, this is equivalent to sampling from a power distribution
\begin{align}
    \pcof{power}{i} \propto \left (\frac{1}{s(i ; \Itrain)} \right )^{-\beta}.
    \label{stochacq:eq:power_sampling}
\end{align}
This may be seen by noting that $\exp(\beta \log s(i;\Itrain)) = s(i;\Itrain)^\beta$.
Importantly, as scores $\to 0$, the (perturbed) log scores $\to -\infty$ and will have probability mass $\to 0$ assigned. This variant takes the absolute scores into account and avoids data points with score $0$.

\paragraph{Summary.} 
Given the above considerations, when using BALD, entropy, and other appropriate scoring functions, power acquisition is the most sensible. Thus, we expect it to work best. Indeed, we find this to be the case in the toy experiment on Repeated-MNIST (see \S\ref{chapter:batchbald}) depicted in \Cref{stochacq:fig:rmnist}.
However, even soft-rank acquisition works well in practice, suggesting that the choice of score perturbation is not critical for its effectiveness; see also appendix \S\ref{stochacq:sec:power_softmax_softrank_comparison} for a more in-depth comparison.
In the rest of this chapter, we focus on power acquisition---we include results for all methods in \S\ref{stochacq:appsec:experiments}.

\section{Related Work}
\label{stochacq:sec:related_work}

In most cases, the computational complexity scales poorly with the acquisition size (\batchvar) or pool size ($\poolsize$), for example because of the estimation of joint mutual information (\S\ref{chapter:batchbald}), the $\mathcal{O}(\batchvar \poolsize)$ complexity of using a k-means++ initialization scheme \citep{ash2020deep}, which approximates k-DPP-based batch active learning \citep{biyik2019batch}, or the $\mathcal{O}(\poolsize^2 \log \poolsize)$ complexity of methods based on \batchvar-center coresets \citep{sener2017active} (although heuristics and continuous relaxations can improve this somewhat).
In contrast, we examine simple and efficient stochastic strategies for adapting well-known single-sample acquisition functions to the batch setting. 
The proposed stochastic strategies are based on observing that acquisition scores would change as new points are added to the acquisition batch and modelling this difference for additional batch samples in the most naive way, using Gumbel noise.
The presented stochastic extensions have the same complexity $\mathcal{O}(\poolsize \log\batchvar)$ as naive top-\batchvar batch acquisition, yet outperform it, and they can perform on par with above more complex methods.

For multi-armed bandits, it has been shown that Thompson sampling from the posterior is effective for choosing informative batches \citep{kalkanli2021batched}. Compared to using the Bayesian model average of the posterior, this can be seen as noising the BMA acquisition scores.
Similarly, in reinforcement learning, stochastic prioritization has been employed as \textit{prioritized replay} \citep{schaul2016prioritized} which may be effective for reasons analogous to those motivating the approach examined in this chapter.

While stochastic sampling has not been extensively explored for acquisition in deep active learning, most recently it has been used as an auxiliary step in diversity-based active learning methods that rely on clustering as main mechanism \citep{ash2020deep, citovsky2021batch}. 
In \Cref{chapter:batchbald} we noted that additional noise in the acquisition scores seems to benefit top-\batchvar batch acquisition in our experiments but did not investigate further.
\citet{fredlund2010bayesian} suggest modeling single-point acquisition as sampling from a ``\emph{query density}'' modulated by the (unknown) sample density $\pof{x}$ and analyze a binary classification toy problem. They model the query density using a parameterized model.
\citet{farquhar2020statistical} propose stochastic acquisition as part of debiasing actively learned estimators.

Most relevant to this chapter, and building on \citet{fredlund2010bayesian,farquhar2020statistical}, \citet{zhan2022asymptotic} propose a stochastic acquisition scheme that is asymptotically optimal. They normalize the acquisition scores via the softmax function to obtain a query density function for unlabeled samples and draw an acquisition batch from it, similar to SoftmaxEntropy. Their method aims to achieve asymptotic optimality for active learning processes by mitigating the impact of bias. 
However, in this chapter, we propose multiple stochastic acquisition strategies based on score-based or rank-based distributions and apply these strategies to several single-sample acquisition functions, such as BALD and entropy (and standard deviation, variation ratios, see \Cref{stochacq:fig:rmnist_other_acquisition_functions}). We focus on active learning in a (Bayesian) deep learning setting and not in a classical machine learning setting. As such the empirical results and other proposed strategies can be seen as complementary to their work.

Thus, while stochastic sampling is generally well-known within acquisition functions, entirely simple stochastic sampling have not been investigated as alternatives to naive top-\batchvar acquisition in (Bayesian) deep active learning and compared to more complex approaches in various settings.

\section{Empirical Validation}

\label{stochacq:sec:experiments}

In this section, we empirically verify that the presented stochastic acquisition methods (a) outperform top-\batchvar acquisition and (b) are competitive with specially designed batch acquisition schemes like BADGE~\citep{ash2020deep} and BatchBALD (\S\ref{chapter:batchbald}); and are vastly cheaper than these more complicated methods. 

To demonstrate the seriousness of the possible weakness of recent batch acquisition methods, we use a range of datasets.
These experiments show that the performance of the stochastic extensions is not dependent on the specific characteristics of any particular dataset.
Our experiments include computer vision, natural language processing (NLP), and causal inference (in \S\ref{stochacq:sec:finetuning_beta}).
We show that stochastic acquisition helps avoid selecting redundant samples on Repeated-MNIST~(\S\ref{chapter:batchbald}), examine performance in active learning for computer vision on EMNIST~\citep{cohen2017emnist}, MIO-TCD~\citep{luo2018mio}, Synbols \citep{lacoste2020synbols}, and CLINC-150~\citep{larson-etal-2019-evaluation} for intent classification in NLP. MIO-TCD is especially close to real-world datasets in size and quality. In appendix \ref{stochacq:sec:synbols}, we further investigate edges cases using the Synbols dataset under different types of biases and aleatoric uncertainty.

Here, we consider both BALD and predictive entropy as scoring functions. We examine other scoring functions on Repeated-MNIST in appendix \ref{stochacq:appsec:exp_repeated_mnist_other_acq_functions} and observe similar results.
For the sake of legible figures, we focus on power acquisition in this section, as it fits BALD and entropy best: the scores are non-negative, and zero scores imply uninformative samples.
We show that all three methods (power, softmax, softrank) perform similarly in appendix \ref{stochacq:sec:power_softmax_softrank_comparison}.

We are not always able to compare to BADGE and BatchBALD because of computational limitations of those methods.
BatchBALD is computationally infeasible for large acquisition sizes ($>10$) because of time constraints, cf.\ \Cref{stochacq:tab:runtime}. When possible, we use BatchBALD with acquisition size 5 as baseline.
Similarly, BADGE runs out of memory for large dataset sizes, such as EMNIST `ByMerge' with 814,255 examples.

Figures interpolate linearly between available points, and we show 95\% confidence intervals.

\paragraph{Experimental Setup \&\ Compute.} We document the experimental setup and model architectures in detail in appendix \ref{stochacq:sec:exp_setup}. Our experiments used about 25,000 compute hours on Titan RTX GPUs.

\begin{figure}[t]
    \captionsetup[subfigure]{aboveskip=0pt,belowskip=-1pt}
    \begin{subfigure}[t]{0.49\linewidth} %
        \centering %
        \includegraphics[width=\linewidth]{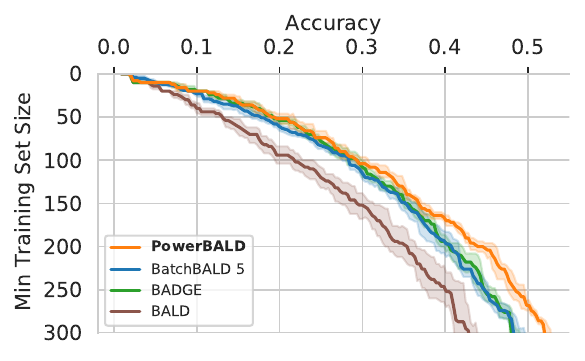} %
        \caption{EMNIST (Balanced) (5 trials)} %
        \label{stochacq:fig:emnist_balanced} %
    \end{subfigure} %
    \hfill
    \begin{subfigure}[t]{0.49\linewidth} %
        \centering %
        \includegraphics[width=\linewidth]{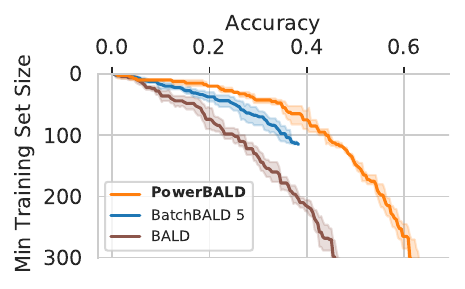} %
        \caption{EMNIST (ByMerge) (5 trials)} %
        \label{stochacq:fig:emnist} %
    \end{subfigure} %
    \hfill %
    \begin{subfigure}[t]{0.49\linewidth} %
        \centering %
        \includegraphics[width=\linewidth]{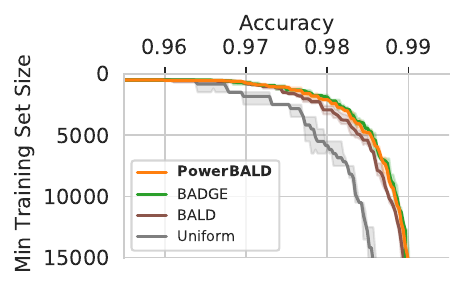} %
        \caption{MIO-TCD (3 trials)} %
        \label{stochacq:fig:miotcd} %
    \end{subfigure} %
    \hfill %
    \begin{subfigure}[t]{0.49\linewidth} %
        \centering %
        \includegraphics[width=\linewidth]{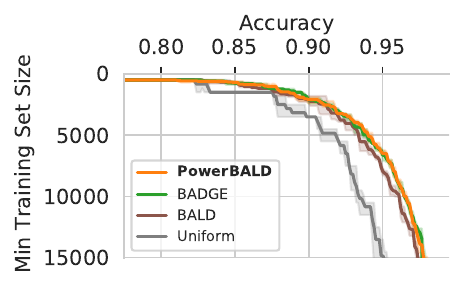} %
        \caption{
        Synbols with minority groups (10 trials)
        } %
        \label{stochacq:fig:synbols_underrepresented} %
    \end{subfigure} %
    \caption{
        \emph{Performance on various datasets.} BatchBALD took infeasibly long on these datasets \&\ acquisition sizes.
        \textbf{(\subref{stochacq:fig:emnist_balanced})} \emph{EMNIST `Balanced':} On 132k samples, PowerBALD (acq.\ size 10) outperforms BatchBALD (acq.\ size 5) and BADGE (acq.\ size 40).
        \textbf{(\subref{stochacq:fig:emnist})} \emph{EMNIST `ByMerge':} On 814k samples, PowerBALD (acq.\ size 10) outperforms BatchBALD (acq.\ size 5). BADGE (not shown) OOM'ed, and BatchBALD took $>12$ days for 115 acquisitions.
        \textbf{(\subref{stochacq:fig:miotcd})} \emph{MIO-TCD:} PowerBALD performs better than BALD and on par with BADGE (all acq.\ size $100$).
        \textbf{(\subref{stochacq:fig:synbols_underrepresented})} \emph{Synbols with minority groups:} PowerBALD performs on par with BADGE (all acq.\ size $100$). %
    }
\end{figure}

\paragraph{Runtime Measurements.}\label{stochacq:subsec:runtime} %
We emphasize that the stochastic acquisition strategies are much more computationally efficient compared to specialized batch-acquisition approaches like BADGE and BatchBALD.
Runtimes, shown in \Cref{stochacq:tab:runtime}, are essentially identical for top-\batchvar and the stochastic versions.
Both are orders of magnitude faster than BADGE and BatchBALD even for small batches.
Unlike those methods, stochastic acquisition scales \textit{linearly} in pool size and \emph{logarithmically} in acquisition size.
Runtime numbers do not include the cost of retraining models (identical in each case). %
The runtimes for top-\batchvar and stochastic acquisition appear constant over \batchvar because the execution time is dominated by fixed-cost memory operations.
The synthetic dataset used for benchmarking has 4,096 features, 10 classes, and 10,000 pool points.

\paragraph{Repeated-MNIST.} %
Repeated-MNIST (\Cref{chapter:batchbald}) duplicates MNIST a specified number of times and adds Gaussian noise to prevent perfect duplicates.
Redundant data are incredibly common in industrial applications but are usually removed from standard benchmark datasets.
The controlled redundancies in the dataset allow us to showcase pathologies in batch acquisition methods.
We use an acquisition size of 10 and 4 dataset repetitions.

\Cref{stochacq:fig:rmnist} shows that PowerBALD outperforms top-\batchvar BALD.
While much cheaper computationally, cf. \Cref{stochacq:tab:runtime}, PowerBALD also outperforms BADGE and even performs on par with BatchBALD. For BatchBALD, we use an acquisition size of 5, and for BADGE of 20.
Note that BatchBALD performs better for smaller acquisition sizes while BADGE (counterintuitively) performs better for larger ones; see \Cref{stochacq:appfig:rmnist_badge_ablation} in the appendix for an ablation. BatchBALD, BALD, and the stochastic variants all become equivalent for acquisition size 1 when individual points are sampled, which performs best (\Cref{chapter:batchbald}).

\paragraph{Computer Vision: EMNIST.} %
EMNIST \citep{cohen2017emnist} contains handwritten digits and letters and comes with several splits: 
we examine the `Balanced` split with 131,600 samples in \Cref{stochacq:fig:emnist_balanced}\footnote{This result exactly reproduces BatchBALD's trajectory in Figure 7 from \S\ref{chapter:batchbald}.} and the `ByMerge` split with 814,255 samples in \Cref{stochacq:fig:emnist}. 
Both have 47 classes.
We use an acquisition size of 5 for BatchBALD, of 40 for BADGE, and of 10 otherwise. 

We see that the stochastic methods outperform BatchBALD on it and both BADGE and BatchBALD on `Balanced' (\Cref{stochacq:fig:emnist_balanced}). They do not have any issues with the huge pool set in `ByMerge` (\Cref{stochacq:fig:emnist}). In the appendix, \Cref{stochacq:appfig:emnist_bald,stochacq:appfig:emnist_balanced_bald} show results for all three stochastic extensions, and \Cref{stochacq:appfig:emnist_balanced_badge_ablation} shows an ablation of different acquisition batch sizes for BADGE.
For `ByMerge', BADGE ran out of memory on our machines, and BatchBALD took more than 12 days for 115 acquisitions when we halted execution.

\paragraph{Computer Vision: MIO-TCD.} %
\label{stochacq:sec:miotcd} %
The Miovision Traffic Camera Dataset (MIO-TCD)~\citep{luo2018mio} is a vehicle classification and localization dataset with 648,959 images designed to exhibit realistic data characteristics like class imbalance, duplicate data, compression artifacts, varying resolution (between 100 and 2,000 pixels), and uninformative examples; see \Cref{stochacq:fig:mio_tcd_examples} in the appendix.
As depicted in \Cref{stochacq:fig:miotcd}, PowerBALD performs better than BALD and essentially matches BADGE despite being much cheaper to compute. We use an acquisition size of 100 for all methods.

\paragraph{Computer Vision: Synbols.}
\label{stochacq:sec:synbols_main}
Synbols \citep{lacoste2020synbols} is a character dataset generator which can demonstrate the behavior of batch active learning under various edge cases \citep{lacoste2020synbols, branchaud2021can}.
In \Cref{stochacq:fig:synbols_underrepresented}, we evaluate PowerBALD on a dataset with minority character types and colors.
PowerBALD outperforms BALD and matches BADGE.
Further details as well as an examination of the `spurious correlation' and `missing symbols' edge cases \citep{lacoste2020synbols, branchaud2021can} can be found in appendix \ref{stochacq:sec:synbols}.

\paragraph{Natural Language Processing: CLINC-150.} %
\label{stochacq:sec:cinic150} %
We perform intent classification on CLINC-150~\citep{larson-etal-2019-evaluation}, which contains 150 intent classes plus an out-of-scope class. This setting captures data seen in production for chatbots. We fine-tune a pretrained DistilBERT model from HuggingFace~\citep{dosovitskiy2020image} on CLINC-150 for 5 epochs with Adam as optimizer. 
In appendix \ref{stochacq:appsec:clinc}, we see that PowerEntropy shows strong performance. This demonstrates that our technique is domain independent and can be easily reused for other tasks.

\paragraph{Summary.} %
We have verified that stochastic acquisition functions outperform top-\batchvar batch acquisition in several settings and perform on par with more complex methods such as BADGE or BatchBALD.
Moreover, we refer the reader to \citet{jesson2021causal}, \citet{murray2021depth}, \citet{tigas2022interventions}, \citet{holmes2022efficiently} for additional works that use the proposed stochastic acquisition functions from this chapter and provide further empirical validation.

\section{Further Investigations}
\label{stochacq:sec:investigation}

In this section, we validate our assumptions about the underlying score dynamics by examining the score rank correlations across acquisitions.
We further hypothesize about when top-\batchvar acquisition is the most detrimental to active learning.

\begin{figure}[t]
    \begin{minipage}[t]{.66\linewidth} %
        \captionsetup[subfigure]{aboveskip=-1pt,belowskip=-1pt}
        \begin{subfigure}[t]{0.49\linewidth} %
            \centering %
            \includegraphics[width=\textwidth]{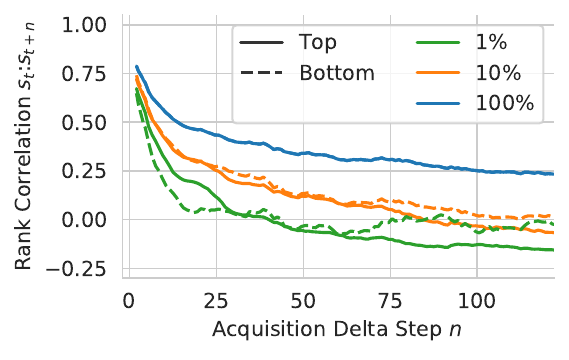} %
            \caption{$t=0$} %
            \label{stochacq:subfig:bald_rank_corr_0} %
        \end{subfigure}
        \hfill
        \begin{subfigure}[t]{0.49\linewidth} %
            \centering %
            \includegraphics[width=\textwidth]{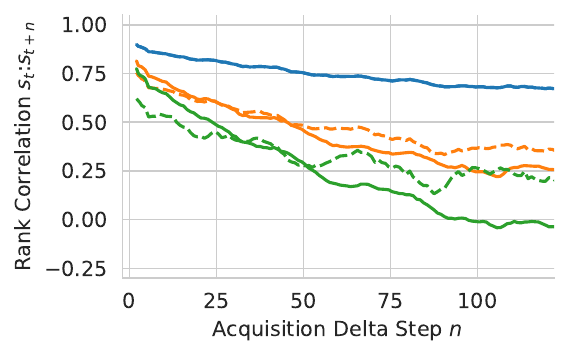} %
            \caption{$t=100$} %
            \label{stochacq:subfig:bald_rank_corr_100} %
        \end{subfigure}
        \caption{
            \emph{Rank correlations for BALD scores on MNIST between the initial scores and future scores of the top- or bottom-scoring 1\%, 10\% and 100\% of test points (smoothed with a size-10 Parzen window).}
            The rank order decorrelates faster for the most informative samples and in the early stages of training.
            The top-1\% scorers' ranks \textit{anti-correlate} after roughly 40 (100) acquisitions unlike the bottom-1\%. 
            Later in training, the acquisition scores stay more strongly correlated. This suggests \emph{the acquisition size could be increased later in training}.
        }
        \label{stochacq:fig:bald_full_rank_correlations_detail}
    \end{minipage}%
    \hfill %
    \begin{minipage}[t]{.33\linewidth} %
        \centering %
        \includegraphics[width=\linewidth]{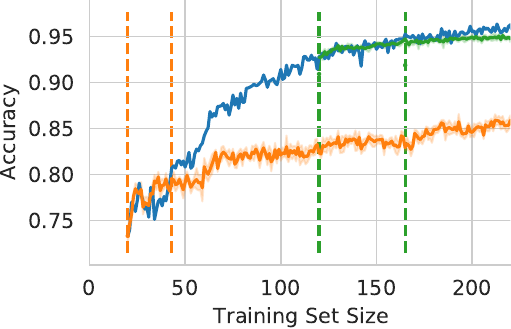} %
        \caption{ %
        \emph{Top-\batchvar acquisition hurts less later in training (BALD on MNIST).} At $t\in\{20, 100\}$ (\textcolor{sns0}{blue}), we keep acquiring samples using the BALD scores from those two steps. At $t=20$ (\textcolor{sns1}{orange}), the model performs well for $\approx20$ acquisitions; at $t=120$ (\textcolor{sns2}{green}), for $\approx50$; see \S\ref{stochacq:subsec:query_size_analysis}. %
        } %
        \label{stochacq:fig:BALD_predetermined_acquisitions}
    \end{minipage}%
\end{figure}

\paragraph{Rank Correlations Across Acquisitions.} %
\label{stochacq:subsec:rank_correlation} %
Our method is based on assuming: (1) the acquisition scores $s_t$ at step $t$ are a proxy for scores $s_{t'}$ at step $t' > t$; (2) the larger $t'-t$ is, the worse a proxy $s_t$ is for $s_t'$; (3) this effect is the largest for the most informative points.

We demonstrate these empirically by examining the Spearman rank correlation between scores during acquisition.
Specifically, we train a model for $n$ steps using BALD as single-point acquisition function.
We compare the rank order at each step to the starting rank order at step $t$.

\Cref{stochacq:fig:bald_full_rank_correlations} shows that acquisition scores become less correlated as more points are acquired.
\Cref{stochacq:fig:bald_full_rank_correlations_detail}\subref{stochacq:subfig:bald_rank_corr_0} shows this in more detail for the top and bottom 1\%, 10\% or 100\% of scorers of the test set across acquisitions starting at step $t=0$ for a model initialized with 20 points.
The top-10\% scoring points (\textcolor{sns2}{solid green}) quickly become uncorrelated across acquisitions and even become \textit{anti-correlated}.
In contrast, the points overall (\textcolor{sns0}{solid blue}) correlate well over time (although they have a much weaker training signal on average).
This result supports all three of our hypotheses.

At the same time, we see that as training progresses, and we converge towards the best model, the order of scores becomes more stable across acquisitions.
In \Cref{stochacq:subfig:bald_rank_corr_100} the model begins with 120 points ($t=100$), rather than 20 ($t=0$).
Here, the most informative points are less likely to change their rank---even the top-1\% ranks do not become \textit{anti-correlated}, only uncorrelated.
Thus, we hypothesize that further in training, we might be able to choose larger \batchvar.

\paragraph{Increasing Top-\batchvar Analysis.} %
\label{stochacq:subsec:query_size_analysis} %
Another way to investigate the effect of top-\batchvar selection is to freeze the acquisition scores during training and then continue single-point `active learning' as if those were the correct scores. Comparing this to the performance of regular active learning with updated single-point scores allows us to examine how well earlier scores perform as proxies for later scores.
We perform this toy experiment on MNIST, showing that freezing scores early on greatly harms performance while doing it later has only a small effect (\Cref{stochacq:fig:BALD_predetermined_acquisitions}).
For frozen scores at a training set size of 20 (73\% accuracy, $t=0$), the accuracy matches single-acquisition BALD up to a training set size of roughly 40 (\textcolor{sns1}{dashed orange lines}) before diverging to a lower level.
But when freezing the scores of a more accurate model, at a training set size of 120 labels (93\% accuracy, $t=100$),  selecting the next fifty points according to those frozen scores performs indistinguishably from step-by-step acquisition (\textcolor{sns2}{dashed green lines}).
This result shows that top-\batchvar acquisition hurts less later in training but can negatively affect performance at the beginning of training.

These observations lead us to ask whether we could dynamically change the acquisition size: with smaller acquisition batches at the beginning and larger ones towards the end of active learning. We leave the exploration of this for future work.

\subsection{Ablation: Changing $\beta$}
\label{stochacq:sec:finetuning_beta}

\begin{figure*}[t]
    \centering
    \begin{minipage}[t]{\linewidth} %
        \captionsetup[subfigure]{aboveskip=-1pt,belowskip=-1pt} %
        \begin{subfigure}[t]{0.49\linewidth} %
            \centering %
            \includegraphics[width=\linewidth]{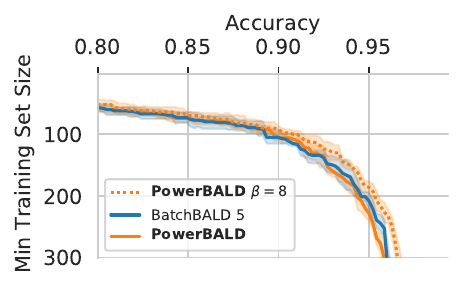} %
            \caption{
                Repeated-MNISTx4 
            } %
            \label{stochacq:fig:rmnist_finetuned} %
        \end{subfigure}
        \hfill %
        \begin{subfigure}[t]{0.49\linewidth} %
            \centering
            \includegraphics[width=\linewidth]{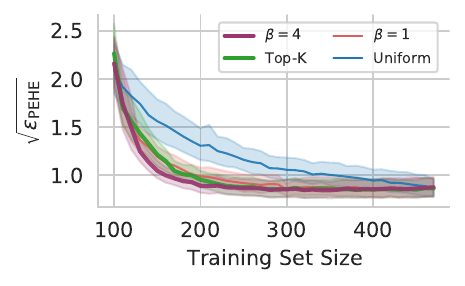} %
            \caption{
                IHDP
            } %
            \label{stochacq:fig:ihdp_causal} %
        \end{subfigure} %
        \caption{
            \emph{Effect of changing $\beta$.} 
            \textbf{(\subref{stochacq:fig:rmnist_finetuned})} \emph{Repeated-MNISTx4 (5 trials):} PowerBALD outperforms BatchBALD for $\beta=8$.
            \textbf{(\subref{stochacq:fig:ihdp_causal})} \emph{IHDP (400 trials):} At high temperature ($\beta = 0.1$), CausalBALD with power acquisition is like random acquisition. As the temperature decreases, the performance improves (lower $\sqrt{\epsilon_{\mathrm{PEHE}}}$), surpassing top-\batchvar acquisition. 
        } %
    \end{minipage} %
\end{figure*}

So far, we have set $\beta=1$ in the spirit of providing a simple baseline without additional hyperparameters.
The results above show that this already works well and matches the performance of much more expensive methods, raising questions about their value.
In addition, however, tuning $\beta$ may be able to further improve performance.
Next, we show that other values of $\beta$ can yield even higher performance on Repeated-MNIST and when estimating causal treatment effects; we provide additional results in appendix \ref{stochacq:appsec:tuning_beta}.

\paragraph{Repeated-MNIST.} In \Cref{stochacq:fig:rmnist_finetuned}, we see that for PowerBALD the best-performing value, $\beta=8$, outperforms BatchBALD.

\paragraph{Causal Treatment Effects: Infant Health Development Programme.} 
Active learning for Conditional Average Treatment Effect (CATE) estimation \cite{heckman1997matching, heckman1998matching, hahn1998role, abrevaya2015estimating} on data from the Infant Health and Development Program (IHDP) estimates the causal effect of treatments on an infant's health from observational data.
Statistical estimands of the CATE are obtainable from observational data under certain assumptions.  
\citet{jesson2021causal} show how to use active learning to acquire data for label-efficient estimation.
Among other subtleties, this prioritizes the data for which matched treated/untreated pairs are available.

We follow the experiments of \citet{jesson2021causal} on both synthetic data and the semisynthetic IHDP dataset \citep{hill2011bayesian}, a commonly used benchmark for causal effects estimation.
In \Cref{stochacq:fig:ihdp_causal} we show that power acquisition performs significantly better than both top-\batchvar and uniform acquisition, using an acquisition size of 10 in all cases with further.
We provide additional results on semisynthetic data in appendix \ref{stochacq:app:causal_synth_temp}.
Note that methods such as BADGE and BatchBALD are not well-defined for causal-effect estimation, while our approach remains applicable and is effective when fine-tuning $\beta$: BatchBALD and BADGE are specifically designed for active learning given (classification) predictions, which is not the same as estimating causal effects.

Performance on these tasks is measured using the expected \emph{Precision in Estimation of Heterogeneous Effect (PEHE)} \citep{hill2011bayesian} such that $\sqrt{\epsilon_{\mathrm{PEHE}}} = \sqrt{\mathbb{E}[(\widetilde{\tau}(\mathbf{X}) - \tau(\mathbf{X}))^2]}$ \citep{shalit2017estimating} where $\widetilde{\tau}$ is the estimated CATE and $\tau$ is CATE (i.e.\ a form of RMSE).

\paragraph{Limitations.} Although we highlight the possibility for future work to adapt $\beta$ to specific datasets or score functions, our aim is not to offer a practical recipe for this to practitioners.
Our focus is on showing how even the simplest form of stochastic acquisition already raises questions for some recent more complex methods.

\section{Discussion}
\label{stochacq:sec:discussion}
We have demonstrated a surprisingly effective and efficient baseline for batch acquisition in active learning.
The presented stochastic extensions are orders of magnitude faster than sophisticated batch-acquisition strategies like BADGE and BatchBALD while retaining comparable performance in many settings.
Compared to the flawed top-\batchvar batch acquisition heuristic, it is never worse: we see no reason to continue using top-\batchvar acquisition.

Importantly, this chapter raises serious questions about these current methods.
If they fail to outperform such a simple baseline in a wide range of settings, do they model the interaction between points sufficiently well?
If so, are the scores themselves unreliable?

At the same time, this presented framework opens doors for improved methods.
Although our stochastic model is put forward for its computational and mathematical simplicity, future work could explore more sophisticated modelling of the predicted score changes that take the current model and dataset into account.
In its simplest form, this might mean adapting the temperature of the acquisition distribution to the dataset or estimating it online.
Our experiments also highlight that the acquisition size could be dynamic, with larger batch sizes acceptable later in training.

\FloatBarrier

}

{
	\graphicspath{{./obi/plots/}}

\newcommand{\Yhat}{\hat{Y}}
\newcommand{\yhat}{\hat{y}}

\newcommand{\Ypred}{\Yhat}
\newcommand{\ypred}{\yhat}
\newcommand{\Ytrue}{Y}
\newcommand{\ytrue}{y}
\newcommand{\levelthreshold}{q}

\newcommand{\Dataset}{\mathcal{D}}

\newcommand{\Dunlabeled}{{\Dataset_\text{unlabeled}}}
\newcommand{\pwof}[1]{\pcof{}{#1}}
\newcommand{\qwof}[1]{\qcof{}{#1}}
\newcommand{\qpwof}[1]{\qcpof{}{#1}}

\newcommand{\w}{\omega}
\newcommand{\W}{\Omega}

\newcommand{\xeW}{\varHof{\hat \opp_\text{train}\Vert\opq}}
\newcommand{\xeWprime}{\varHof{\hat \opp_\text{train}\Vert\opq'}}
\newcommand{\HofW}{\varHof{\opq}}
\newcommand{\MIofW}{\varMIof{\opq}}
\newcommand{\xmiW}{\varMIof{\hat \opp_\text{train}\Vert\opq}}
\newcommand{\TCW}{\opTC_{\opq}}

\newcommand{\forreviewers}[1]{}
\newcommand{\jannik}[1]{}

\begin{draft}
\renewcommand{\forreviewers}[1]{{\leavevmode\color{purple}{\footnotesize NOTE} {\tiny please: }#1}}
\renewcommand{\jannik}[1]{{\leavevmode\color{teal}{\footnotesize JK} {\tiny says: }#1}}
\end{draft}

	\begin{savequote}[75mm]
		The whole universe is contained in a single flower.
		\qauthor{Toshiro Kawase}
	\end{savequote}
	\chapter{\obititle}
	\label{chapter:obi}

Beyond deep ensembles \citep{lakshminarayanan2016simple}, more principled methods of deep learning that attempt to be (approximately) Bayesian, commonly referred to as (approximate) 
\emph{Bayesian Neural Networks (BNN)}, have arguably not lived up to their full potential \citep{ovadia2019can, beluch2018power}.
This might be because the focus in their evaluation has been on marginal predictions $\qwof{\y \given \x}$, where they can only provide \emph{marginal}\textsuperscript{:D} improvements over unprincipled regular NNs.

Yet the strength of a Bayesian approach for deep learning might not solely lie in marginal predictions but in joint predictions and in allowing for online learning via online Bayesian inference. \emph{Online Bayesian inference (OBI)} refers to incorporating additional data into the posterior predictive \emph{without} explicitly retraining in the common sense, i.e.\ by computing gradients and optimizing the model parameters further\footnote{\citet{maddox2021conditioning} have referred to this approach as `Online Variational Conditioning' in the context of Gaussian processes. However, we would argue that in the context of this chapter OBI is the better term because there is nothing variational about it. Indeed, OBI applies Bayesian inference to an `imperfect' variational intermediate posterior instead of performing Bayesian inference end-to-end from the original prior.}. 
This could offer important performance benefits for applications that would otherwise require repeated retraining like active learning and could have important implications for how we could use large supervised models in production: currently, they are seen as strictly static; however, online Bayesian inference would allow them to dynamically adapt to new data on the fly.

Generally, the difference between an approximate BNN and a regular NN is that the former assumes a distribution $\qof{\w}$ over the model parameters $\w$, where $\qof{\w}$ approximates the Bayesian posterior $\pof{\w \given \Dtrain}$, which is the optimal distribution given prior information $\pof{\w}$ and training data $\Dtrain$:
$\qof{\w} \approx \pof{\w \given \Dtrain}.$

\paragraph{Online Bayesian Inference.} To incorporate new data $\{\y_i, \x_i\} \sim \pdataof{\y, \x}^n$, via online Bayesian inference, we simply apply Bayes' theorem:
for a test point $\x$, the predictive $\qwof{\y \given \x, \y_{n}, \x_{n}, \ldots, \y_1, \x_1}$ is proportional to its joint predictive. We obtain:
\begin{align}
    \qwof{y \given x, y_{n}, x_{n}, \ldots, y_1, x_1} &=
    \frac{\qwof{y, y_{n}, \ldots, y_1 \given x, x_{n}, \ldots, x_1}}{\qwof{y_{n} \ldots, y_1 \given x_{n}, \ldots, x_1}} 
    \label{obi:eq:online_bayesian_inference}
    \\
    &\propto \qwof{y, y_{n}, \ldots, y_1 \given x, x_{n}, \ldots, x_1}. 
\end{align}
We can thus use a joint predictive $\qwof{\y, \y_1, \ldots, \y_n \given \x, \x_1, \ldots, \x_n}$ to incorporate fixed$\{\y_i, \linebreak[0] \x_i\}^n$ and make predictions for $\x$ without explicit retraining. %

Hence, for online Bayesian inference, we require joint predictives, which only Bayesian methods can give us:\footnote{For consistent joint predictives, adhering to the chain rule of probability, a model needs to adhere to the ``Bayesian update rule'', i.e.~Bayes' theorem, and thus is Bayesian. See also \cref{obi:eq:prob_chain_rule} below.} through BNNs in the parametric case or through Gaussian processes in the non-parameteric case, for example. This strongly contrasts with marginal predictives which can also be modelled by regular NNs.

Can we perform online Bayesian inference well for high-dimensional inputs and parameters using current approximate BNNs?
The quality of the resulting predictions crucially depends on the joint predictives. 
However, computing joint predictives can be challenging. For example, in Bayesian literature, the joint predictive of all samples in the training set marginalized over the prior distribution is just the well-known marginal likelihood, which can be used for model selection \citep{mackay2003information, lyle2020bayesian, fern2020marginal}, and is known to be difficult to estimate in high-dimensional parameter spaces \citep{lotfi2022bayesian}. 
Similar challenges can be expected for performing online Bayesian inference using approximate posterior distributions.

Based on recent works by \citep{wen2021predictions, osb2022evaluating}, we examine the joint predictives for online Bayesian inference using naive prior sampling. However, we find negative results when using MC dropout \citep{gal2016dropout}.

\paragraph{Relevant Literature.} %
\citet{osb2022evaluating,wen2021predictions,osb2021neural,osb2021epistemic} mention the importance of joint predictives in the context of combinatorial decision problems, sequential predictions and multi-armed bandits in low dimensions. Compared to these previous works, we explore important connections to supervised learning, e.g.~in active learning \citep{atlas1989training, Settles2010} and Bayesian optimal experimental design \citep{lindley1956measure, foster2022a}, and focus on online Bayesian inference in high dimensions using deep neural networks. This is also an important difference to \citet{maddox2021conditioning} which examines online Bayesian inference in the context of Gaussian processes.

In addition to these works, we clarify that both marginal and joint cross-entropies have their use, and it is not the case that one is always preferable over the other. Simply put, we argue that they capture different quantities that are separately useful in offline and online learning, and we combine them to evaluate the performance of online Bayesian inference using approximate BNNs. We provide more details in \S\ref{obi:sec:related_work}.

\paragraph{Marginal Cross-Entropy.} As will become evident, the marginal cross-entropy for a fixed predictive model captures the expected performance (\emph{under log loss}) when the model does not adapt to data at test time.
For supervised tasks, the marginal cross-entropy is what is commonly referred to as cross-entropy loss and represents common practice: we obtain a fixed set of parameters by training the model on a training set, and we re-use these parameters at test time without any further updates.
The performance of the model does not change as it observes more test data, and there is no feedback loop of any sort.
In this \emph{offline learning} setting, the marginal cross-entropy is the right choice to estimate performance.

\paragraph{Joint Cross-Entropy.} On the other hand, the joint cross-entropy for a predictive model captures the performance in a \emph{sequential learning} setting, where sequential model updates take place.
Here, the parameter distribution $\qof{\w}$ serves as a prior for online Bayesian inference. This fits the context used in \citet{wen2021predictions} in which the model makes a prediction for the next step, observes the outcome, and then updates the model (agent). 

\paragraph{Applications \&\ Experiments.} We will also see that the joint predictive is important for data selection in active learning and active sampling.
This chapter also connects several recent works \citep{wen2021predictions,osb2022evaluating} with active learning and active sampling and present new realistic and challenging experimental settings.
Most importantly, we examine online Bayesian inference within these contexts as it allows us to avoid \emph{retraining across acquisitions}.

\paragraph{Notation.} 
This chapter mentions many cross-entropies. We will use a more concise notation to save space:
\begin{align}
    \varHof{\opp \Vert \opq}{\cdot} \defeq \CrossEntropy{\pof{}}{\qof{}} = \E{\pof{\cdot}}{- \log \qof{\cdot}}.
\end{align}
For example:
\begin{math}
    \varHof{\opp \Vert \opq}{Y \given x} = \E{\pof{y \given x}}{- \log \qof{y \given x}}.
\end{math}

\paragraph{Background.}
\label{obi:sec:background}

The setting in this chapter deviates slightly from the one introduced in \Cref{glue:related_work:BNNs}:
\emph{The important difference in this chapter to \Cref{glue:related_work:BNNs} is that we investigate how $\opq$ compares to $\opp$ and do not ignore the difference.}
That is, we assume an underlying parametric predictive model $\pof{\y \given \x, \w}$ for input samples $\x$ with targets or labels $\y$
with a prior parameter distribution $\pof{\w}$ over $\w$, i.e.~our model is Bayesian.
As noted previously, capturing the true posterior distribution $\pof{\w \given \Dtrain}$ is infeasible, and we assume we have an approximate distribution $\qof{\w} \approx \pof{\w \given \Dtrain}$ that we use instead of the true posterior. 
For example, $\qof{\w}$ could be based on a deep ensemble \citep{lakshminarayanan2016simple} as a mixture of Dirac delta distributions positioned at the parameters of individually trained ensemble members, or it could be an MC dropout model that is trained using variational inference \citep{gal2015dropout}. We use $\qwof{\y \given \x}$ to denote the predictions after marginalizing over $\qwof{\w}$: $\qwof{\y \given \x} = \E{\qof{\w}}{\pof{\y \given \x, \w}}.$ Note that the underlying discriminative model (or the likelihood function) $\pof{\y \given \x, \w}$ remains the same---we only exchange the distribution over its parameters $\w$. 

\section{Marginal and Joint Cross-Entropy}
\label{obi:sec:margial_and_joint_cross-entropy}
We begin by contrasting marginal and joint predictive cross-entropies and revisiting how they are useful for offline and online learning separately.

\paragraph{Marginal Cross-Entropy.} Given an underlying, possibly empirical, data distribution $\ptrainof{\x,\y}$, the \emph{marginal cross-entropy} is:
\begin{align}
    \xeW{Y \given X} &= \E{\ptrainof{x, y}}{-\log \E{\qof{\w}}{\pof{y\given x, \w}}} \notag \\
    &= \E{\ptrainof{x, y}}{-\log \qwof{y\given x}},
\end{align}
where we use $\xeW{Y \given X}$ to denote the cross-entropy.
This cross-entropy is the population loss when $\qof{\w}$ is not updated after seeing new samples. Each sample $\x, \y$ is treated independently. Hence, the marginal cross-entropy captures the expected performance in an \emph{offline learning} setting.

\paragraph{Joint Cross-Entropy.} On the other hand, given an initial parameter distribution $\qof{\w}$ above, the joint cross-entropy measures how well the parameter distribution can \emph{adapt} to new data $\Dataset$. 
\jannik{perhaps you want to show the joint cross-entropy equation first once, before doing (4)-(7)?}

To show how this connects to joint cross-entropies, we can look at the joint cross-entropy of specific samples $\Dataset = \{\y_i, \x_i\}_{i=1}^n$ (without taking an expectation).
The joint cross-entropy for these specific samples is just the sum of (negative) log marginal likelihoods using the chain rule, where each $\y_i$ is conditioned on all `previous' observations $\Dataset_{<i}$:
\begin{align}
    \xeW{y_1, \ldots, y_n \given x_1, \ldots, x_n} &= -\log \qwof{y_1, \ldots, y_n \given x_1, \ldots, x_n} \\
    &= -\log \prod_i \qwof{y_i \given x_i, y_{i-1}, x_{i-1}, \ldots, y_1, x_1} \\
    &= -\sum_i \log \qwof{y_i \given x_i, \Dataset_{<i}} \label{obi:eq:prob_chain_rule} \\
    &= \sum_i \xeW{y_i \given x_i, \Dataset_{<i}},
\end{align}
where $\Dataset_{<i}$ denotes ``$\y_{i-1}, \x_{i-1}, \ldots, \y_1, \x_1$'', and the marginal predictive is:
\begin{math}
    \qwof{y_i \given x_i, \Dataset_{<i}} = \E{\qof{\w \given \Dataset_{<i}}}{\pof{y_i \given x_i, \w}}.
\end{math}
Semantically, we compute the following in each iteration of the sum: we \emph{update} the parameter posterior $\qof{\w \given \Dataset_{<i}}$, compute the losses for our predictions at outcomes $\y_i$ for $\x_i$, and then include $\y_i, \x_i$ in our observed data.
We will denote this as the \emph{online learning setting}.

When we are interested in the expected loss given arbitrary data, we can compute the following joint cross-entropy:
\newcommand{\OLL}[1]{OLL(#1)}
\begin{align}
    \OLL{n} &\defeq \E{(x_i, y_i)_{i=1}^n \sim \pdataof{x_i, y_i}^n}{-\log \qwof{y_1, \ldots, y_n \given x_1, \ldots x_n}} \\
    &= \E{(x_i, y_i)_{i=1}^n \sim \pdataof{x_i, y_i}^n}{\xeW{y_1, \ldots, y_n \given x_1, \ldots x_n}} \\
    &= \xeW{Y_1, \ldots, Y_n \given X_1, \ldots X_n},
\end{align}
where $OLL$ stands for ``online learning loss''.

\paragraph{Connection to the Conditional Cross-Entropy Rate.} %
As an aside, if we let $n \to \infty$, we also have $\OLL{n} \to \infty$. This is not helpful, so instead we can look at the average: $\frac{1}{n} \OLL{n}$. In the limit, this average is just the \emph{cross-entropy rate}:
\begin{align}
    \xeW{\mathcal{Y} \given \mathcal{X}} \defeq \lim_{n \to \infty} \frac{1}{n} \xeW{Y_1, \ldots, Y_n \given X_1, \ldots X_n},
\end{align}
which we define analogously to the entropy rate in \citet{cover2005elements}.\footnote{Cf.~the entropy rate, which is:
\begin{math}
    \Hof{\mathcal{X}} = \lim_{n \to \infty} \frac{1}{n} \Hof{X_1, \ldots X_n}.
\end{math}
}

\paragraph{Summary.} The marginal cross-entropy is useful for offline learning as it predicts the performance of a fixed model on the data distribution. The joint cross-entropy is useful for online learning as it predicts the performance of a model as it adapts to additional data. Hence, it is important for sequential decision-making.

\begin{draft}
\section{Connection to Active Learning and Active Sampling}
\label{obi:sec:active_learning_and_active_sampling}

Marginal and joint cross-entropies and the related predictives play an important role in active learning and active sampling as we will examine here.
In \emph{active learning} \citep{atlas1989training, Settles2010}, we do not know the outcome (label) for every sample and have a budget for the additional labels we can acquire for training from an unlabeled pool set. On the other hand, in \emph{active sampling} (also data subset selection or coreset selection) \citep{campbell2017automated,mirzasoleiman2019coresets, borsos2020coresets}, we have access to the labels, but we assume that we have a budget for the samples we can use to train the model on, and our goal is to pick the best subset of labeled samples.

In both cases, we have a budget of $n$ additional samples we can condition on, and we are usually interested in maximizing the performance of the model as end-goal, which is equivalent to finding $\x_1,\ldots,\x_n$ which minimize the following cross-entropy (assuming the usual cross-entropy loss):
\begin{align}
    \xeW{Y \given X, y_n, x_n, \ldots, y_1, x_1} \text{ or } \xeW{Y \given X, Y_n, x_n, \ldots, Y_1, x_1},
\end{align}
depending on whether we have access to the labels (active sampling) or not (active learning).

These objectives depend on a joint predictive via \cref{obi:eq:online_bayesian_inference} but in expectation over $X$ and $Y$ sampled from the data distribution. As such it is a hybrid between the joint and marginal cross-entropy.

\paragraph{Transductive Active Sampling.} %
If we have access to the labels $\y_1, \ldots, \y_n$, this leads to active sampling approaches, where we want to find the best samples to train on to increase model performance.
Using
\begin{align}
    \xmiW{Y ; y_n, \ldots, y_1 \given X, x_n, \ldots, x_1} = \xeW{Y \given X} - \xeW{Y \given X, y_n, x_n, \ldots, y_1, x_1},
\end{align}
where $\xmiW{}$ is a ``cross mutual information'', \emph{minimizing} above cross-conditional entropy is equivalent to \emph{maximizing} the cross-mutual information:
\begin{align}
    \argmin_{\{x_i\}_{1..n}} \xeW{Y \given X, y_n, x_n, \ldots, y_1, x_1} = \argmax_{\{x_i\}_{1..n}} \xmiW{Y ; y_n, \ldots, x_1 \given X, x_n, \ldots, x_1},
\end{align}
because $\xeW{Y \given X}$ is constant (independent of the $\x_i$).

Approximations of this term lead to the \emph{reducible hold-out loss} in \Cref{chapter:rholoss}. Note that this is a transductive approach \citep{yu2006active} because we examine performance in regard to samples $(\x, \y)$ from the data distribution.

\paragraph{Transductive Active Learning.} %
If we do not have access to the labels, the Bayesian-optimal approach is to take the expectation using labels drawn from the joint predictive of the model \citep{lindley1956measure}. This leads to minimizing conditional entropies or, equivalently, maximizing mutual information terms \citep{mackay1992information, McCallum1998EmployingEA, yu2006active, wang2021beyond, kirsch2021test}. Assuming we have access to no labels beyond the current training set at all, we have the following objectives:
\begin{align}
    \argmin_{\{x_i\}_{1..n}} \HofW{Y \given X, Y_n, x_n, \ldots, Y_1, x_1} = \argmax_{\{x_i\}_{1..n}} \MIofW{Y ; Y_n, \ldots, Y_1 \given X, x_n, \ldots, x_1}.
\end{align}
The latter is exactly the Expected Predictive Information Gain (EPIG) acquisition function from \Cref{chapter:pobal}\footnote{EPIG is formulated using an unlabeled evaluation set that specifies the target domain. Here, we use the pool set itself as an evaluation set to simplify the exposition.}.

\paragraph{Active Learning.} %
Many active learning approaches, especially the ones that are not transductive, maximize the \emph{expected information gain}, also called BALD \citep{houlsby2011bayesian}, as an acquisition function---or, equivalently, minimize the model uncertainty.  
The corresponding information-theoretic expression is $\MIofW{\W; Y \given \x}$, and in the batch case, we have:
\begin{align}
    &\argmax_{\{x_i\}_{1..n}} \MIofW{\W ; Y_n, \ldots, Y_1 \given x_n, \ldots, x_1} \\
    &\quad =
    \argmax_{\{x_i\}_{1..n}} \HofW{Y_n, \ldots, Y_1 \given x_n, \ldots, x_1} - \HofW{Y_n, \ldots, Y_1 \given x_n, \ldots, x_1, \W} \notag \\
    &\quad = \argmax_{\{x_i\}_{1..n}} \HofW{Y_n, \ldots, Y_1 \given x_n, \ldots, x_1} - \sum_i \HofW{Y_i, \given x_i, \W}, 
\end{align}
where the last step follows from the fact that predictions factorize conditioned on the model parameters (\Cref{chapter:batchbald}).
The corresponding active sampling acquisition function $\MIofW{\W ; \y_n, \ldots, \y_1 \given \x_n, \ldots, \x_1}$, which quantifies the information gain, is examined further in \citet{kirsch2021practical}.

\paragraph{Summary.} Above objectives depend on the joint predictive in some way to compute acquisition scores without explicit retraining. However, it is common to update models after acquiring a set of labels to take into account the newly acquired data.
\end{draft}

\section{Online Bayesian Inference}
\label{obi:subsec:joint_entropy_data_adaptation}

To see what we mean by incorporating new data, assume we have sampled $n$ additional points $\x_i, \y_i \sim \pdataof{\x_i, \y_i}$. 
Traditionally, we would now update the posterior approximation $\qof{w}$ to take this new data into account for our predictions at future test points. However, this can be prohibitively expensive---especially in applications that require frequent retraining. Instead, online Bayesian inference allows Bayesian models to adapt their predictions without explicitly updating the posterior approximation.

Following \cref{obi:eq:online_bayesian_inference}, for a test point $\x$, the predictive $\qwof{\y \given \x, \y_{n}, \x_{n}, \ldots, \y_1, \x_1}$ is proportional to the joint predictive:
\begin{align}
    \qwof{y \given x, y_{n}, x_{n}, \ldots, y_1, x_1} &=
    \frac{\qwof{y, y_{n}, \ldots, y_1 \given x, x_{n}, \ldots, x_1}}{\qwof{y_{n} \ldots, y_1 \given x_{n}, \ldots, x_1}} \\
    &\propto \qwof{y, y_{n}, \ldots, y_1 \given x, x_{n}, \ldots, x_1},
\end{align}
since the normalization constant $\qwof{\y_{n} \ldots, \y_1 \given \x_{n}, \ldots, \x_1}$ is independent of $\y$ and $\x$.
Hence, this allows us to make predictions that take into account new data \emph{without explicit retraining} by simply computing the joint predictive of the test point and newly observed data.

We refer to this as \emph{online Bayesian inference (OBI)}. While this inference is precisely Bayesian, $\qof{\w}$ is commonly only an approximate posterior, and thus the quality of this inference depends on the properties of the approximation and how we estimate the joint predictive. 

The simplest approach to estimate the joint predictive is via sampling, which applies to e.g.~Monte-Carlo dropout, deep ensembles, and deep ensembles with prior functions \citep{gal2015dropout, lakshminarayanan2016simple, osb2018randomized}, by factorizing the joint:
\begin{align}
    \qwof{y, y_{n}, \ldots, y_1 \given x, x_{n}, \ldots, x_1} &= \E{\qof{\w}}{\pof{y, y_{n}, \ldots, y_1 \given x, x_{n}, \ldots, x_1, \w}} \\
    &= \E*{\qof{\w}}{\pof{y \given x, \w} \, \prod_{i=1}^n \pof{y_{i} \given x_{i}, \w}}.
\end{align}
Thus, if we draw fixed parameter samples $\w_j \sim \qof{\w}$, we can pre-compute $\prod_{i=1}^n \pof{\y_{i} \given \x_{i}, \w_j}$ for each $j$ and estimate the joint predictive. 

Finally, we can view $\prod_{i=1}^n \pof{\y_{i} \given \x_{i}, \w}$ as unnormalized importance weights:
\begin{align}
    \qof{\w} \, \prod_{i=1}^n \pof{y_{i} \given x_{i}, \w} \propto \qof{\w \given y_{n}, x_{n}, \ldots, y_1, x_1},
\end{align}
and hence, overall:
\begin{align}
    \E*{\qof{\w}}{\pof{y \given x, \w} \, \prod_{i=1}^n \pof{y_{i} \given x_{i}, \w}} \propto \E{\qof{\w \given y_{n}, x_{n}, \ldots, y_1, x_1}}{\pof{y \given x, \w}}. \label{obi:eq:implicit_posterior_samples}
\end{align}
To evaluate the performance, we can use these predictions to compute the following marginal cross-entropy which incorporates the additional samples using OBI:
\begin{align}
    \xeW{Y \given X, y_n, x_n, \ldots, y_1, x_1},
\end{align}
where $X, Y$ are sampled from the data distribution. Comparing this entropy with the performance of a fully retrained model allows us to obtain a practical estimate for the quality of the approximate posterior $\qof{\w}$.

\section{New Evaluations \&\ Applications}
\label{obi:sec:evaluations_applications}

We suggest new experimental settings that allow us to evaluate the quality of the joint predictive and compare the settings to ones suggested in prior work.

\subsection{Performance in Active Learning and Active Sampling Methods}
A conceptually simple set of downstream tasks is to evaluate the performance of the joint predictives in different active learning or active sampling settings using different approximate model architectures (e.g.~based on Epistemic Neural Networks \citet{osb2021epistemic} as abstraction). \citet{wang2021beyond} show that performance for transductive active learning is correlated to the quality of the joint predictive. We recall that we noticed the same in \Cref{chapter:batchbald} for batch active learning with non-transductive acquisition functions, i.e. BatchBALD's performance heavily on the number of MC dropout samples which informs the quality of the joint predictive.

We suggest, similar to Repeated-MNIST (\Cref{chapter:batchbald}), to duplicate the underlying pool sets or to simply \emph{allow the same sample to be selected multiple times}. Importantly, approximate BNNs that do not provide good joint predictives will greedily select the same sample over and over again or degrade to uninformed data acquisitions. This avoids an issue pointed out by \citet{wang2021beyond} and \citet{osb2022evaluating} as we explain next.

\paragraph{Connection to Total Correlation.} %
\citet{wang2021beyond} argue that the joint cross-entropy is dominated by the sum of the individual marginal cross-entropy scores. This is equivalent to saying that the total correlation between samples is negligible since the difference between the joint cross-entropy and its individual marginal cross-entropies for specific $\y_i, \x_i$ is just the \emph{total correlation}:
\begin{align}
    \TCW[y_1, \ldots, y_n \given x_1, \ldots, x_n] \defeq \sum_i \HofW{y_i \given x_i} - \HofW{y_1, \ldots, y_n \given x_1, \ldots, x_n}.
\end{align}
The total correlation measures the amount of information shared between the samples.

For random batches, the total correlation is, indeed, likely going to be negligible because most random batches are not very informative overall, and importantly, on curated datasets, they are most likely uncorrelated as it is unlikely that observing $\y_i$, $\x_i$ in the batch informs prediction for $\y_j$, $\x_j$ for most $j \not= i$ as curated datasets are usually as diverse as possible.

Only with increasing redundancy in the dataset, e.g. by duplicating samples like in Repeated-MNIST (\Cref{chapter:batchbald}), random batches will become more correlated on average and the total correlation larger. %

This setup is similar to the dyadic sampling proposed by \citet{osb2022evaluating} which repeatedly samples $\y^j_i$ for $\x_i$, with $i \in \{1,2\}$ and evaluates the joint predictive. 
However, this setting in essence only measures the ability of the approximate model to perform Bayesian updates on two fixed training samples at a time.
Hence, we suggest that a better adaption to evaluate joint predictives using active learning is to duplicate the dataset or to simply allow the same sample to be selected multiple times.

\subsection{Performance of Online Bayesian Inference} %
As a practical ``ground truth'', we can compare the performance of retrained models after acquiring additional samples with the performance of OBI as explained in \cref{obi:subsec:joint_entropy_data_adaptation}.

A particular challenging scenario for OBI is to use acquisition sequences $(\x_i, \y_i)_{i=1}^T$ that were collected using active learning or active sampling on the dataset. We evaluate OBI on models trained at different $\Dtrain_{t} = \{\y_i,\x_i\}_{i=1}^t$ with increasing subsets of online data $\{\y_i, \x_i\}_{i=t+1}^T$ from these acquisition sequences. This scenario is particularly challenging for OBI because the sequence of acquisition is selected to result in large changes in the predictives and posterior distributions.

The average performance difference between OBI and fully retrained models across different training acquisition sequences will tell us how good a given joint predictive is for ``meaningful'' online learning.
The expectation is that for most approximate BNNs, OBI will quickly suffer from degraded performance compared to the retrained models.

That is, we compare the performance of OBI as we acquire new samples $\x_i, \y_i$ to an approximate BNN retrained with the same additional data for increasing $n$:
\begin{align}
    \xeW{Y \given X, y_n, x_n, \ldots, y_1, x_1} - \xeWprime{Y \given X}, 
    \label{obi:eq:joint_chain_error}
\end{align}
where $\qcpof{}{\w} \approx \pof{\w \given y_n, x_n, \ldots, y_1, x_1, 
\Dtrain}$ is the parameter distribution of an (approximate) BNN after retraining with the additional $\y_n, \x_n, \ldots, \y_1, \x_1$. %

Ideally, we would compare to the predictions from the correct updated posterior distribution; however, this is infeasible in most practical scenarios.
Instead, when we use an approximate BNN $\qcpof{}{\w}$ that is similar to the one used for $\qof{\w}$, we can measure the practical degradation between OBI and retraining. Here, the ideal would be for OBI to behave exactly like a fully retrained model---even if the latter does not match exact Bayesian inference---as such an approach would be \emph{self-consistent}.
Note that with exact Bayesian inference, we would have $\qof{\w} = \qcpof{}{\w}$, and the above would be zero. %

\paragraph{Comparison to Dyadic Sampling.} Unlike \citet{osb2022evaluating} which focuses on selecting labels for dyadic samples repeatedly, this experiment setting is both more practical and more insightful: active learning picks samples that are the most informative and are supposed to update the posterior the most. This is because, for informative samples, we would expect the changes in model predictions to be the largest. Hence, one could expect that these samples pose the most significant challenge to approximate BNNs and their joint predictives. Ideally, we would hope that OBI would keep up with retrained model, but this might prove to be challenging in high-dimensional scenarios.

\paragraph{Sample Selection Bias.} %
The suggested evaluation is orthogonal to any sample selection bias that is added through the data acquisition process itself as we use the same training data at each step for both OBI and retraining in \cref{obi:eq:joint_chain_error}.
Specifically, \citet{farquhar2020statistical} observed that active learning introduces a bias by sampling from the data distribution using an acquisition function and not uniformly.

\subsection{Application: Active Learning with Online Bayesian Inference}
 
In many settings, retraining models for when only few new samples are added is prohibitively expensive. This motivates batch active learning, where batches are acquired instead of individual samples. Expanding on this, when equipped with models that perform well under OBI, one could avoid retraining models when acquiring new data by using OBI. Only when OBI degrades, fully retraining will become necessary.
We can evaluate this both for individual acquisition and for batch acquisition.

\section{Empirical Validation}
\label{obi:sec:experiments}

\begin{figure}[tb]
    \centering
    \includegraphics[width=\textwidth]{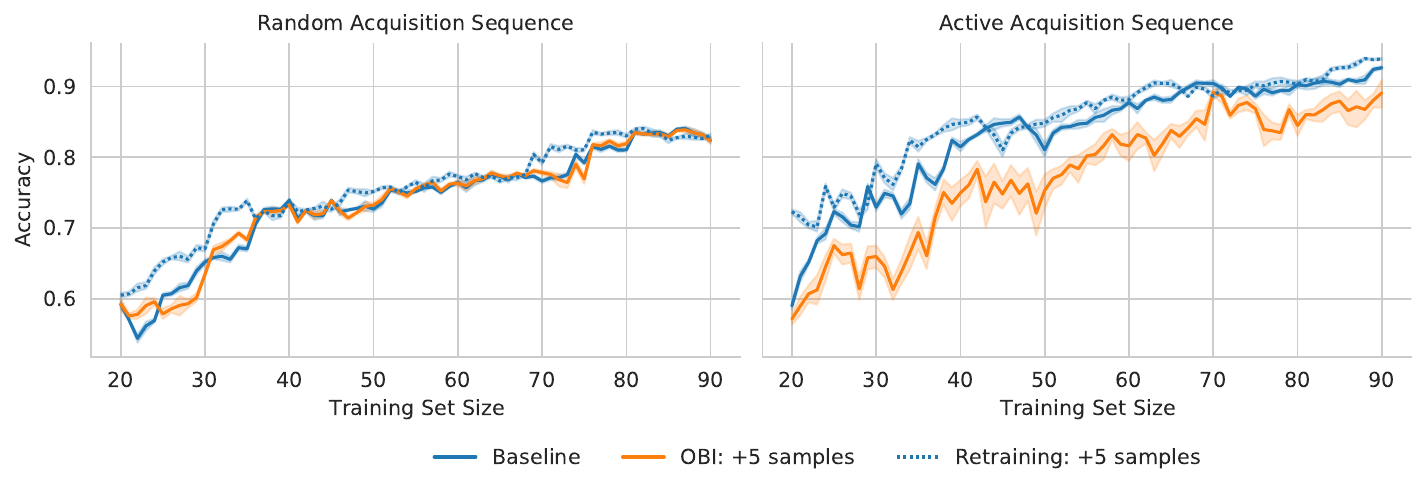}
    \includegraphics[width=\textwidth]{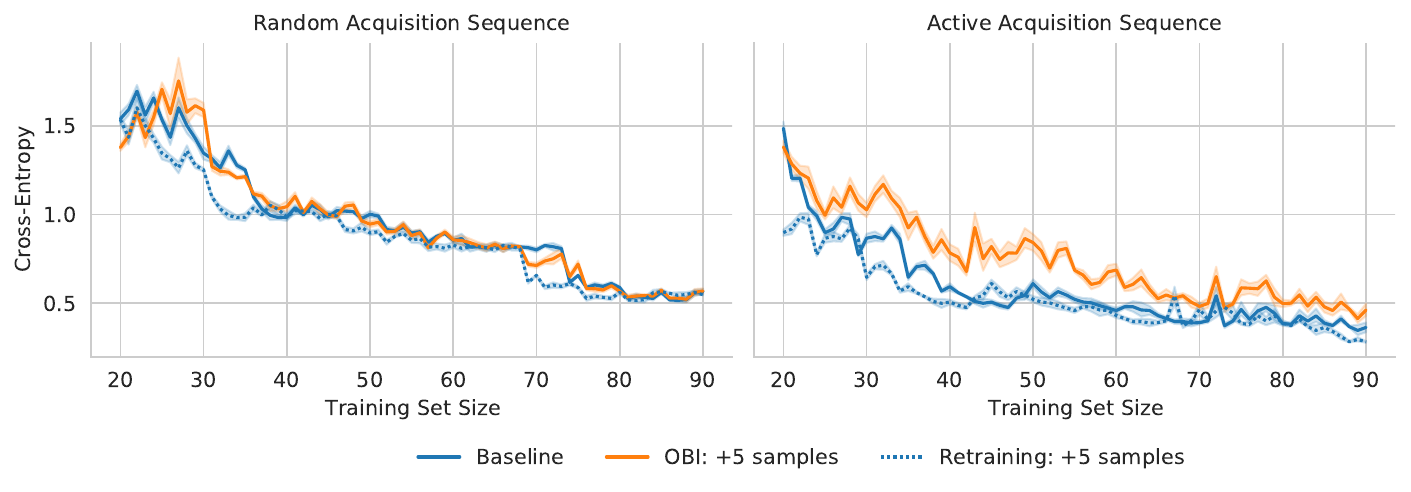}
    \caption{\emph{Comparison between online Bayesian inference and retraining for 5 additional samples on MNIST.} We compare the dynamics between using a random acquisition sequence and using an \emph{active} acquisition sequence, drawn using active sampling. While OBI does not appear to be significantly worse or better on the random acquisition sequence, it markedly deteriorates when using the active acquisition sequence. OBI struggles with the informative samples that change the posterior a lot: the active acquisition sequence reaches 90\% accuracy with just 70 samples, unlike the random one.}
    \label{obi:fig:obi_mnist_comparison}
\end{figure}

\begin{table}[tb]
    \caption{\emph{Comparison between online Bayesian inference and retraining for 5 additional samples on MNIST.} OBI performs worse than the baseline (i.e.\ not taking into account new samples at all) when using an active acquisition sequence. On the random acquisition sequence, it only performs just as well as not updating at all. (Mean for 5 model trials and 5 OBI trials for each.)}
    \centering
    \resizebox{\linewidth}{!}{%
    \begin{tabular}{lrrrr}
    \toprule
    {Baseline \textbf{vs}} & \multicolumn{2}{l}{OBI: +5 samples} & \multicolumn{2}{l}{Retraining: +5 samples} \\
    {Avg} & $\Delta$ Cross-Entropy (\textdownarrow) &  $\Delta$ Accuracy (\textuparrow) & $\Delta$ Cross-Entropy (\textdownarrow) & $\Delta$ Accuracy (\textuparrow) \\
    \midrule
    Active Acquisition Sequence &     0.16 & -5.7\% &    -0.062 &  2.0\% \\
    Random Acquisition Sequence &     $0.00$ &  $0.0\%$ &    -0.075 &  1.8\% \\
    \bottomrule
    \end{tabular}}
    \label{obi:tab:obi_mnist_comparison}    
\end{table}

Following the newly suggested experimental settings, we want to run comparisons using different kinds of approximate BNNs in active learning and active sampling settings.
Moreover, given fixed acquisition sequences, determined using active learning and active sampling, we want to evaluate the difference between OBI and fully retrained models.

An initial experiment shows that OBI in high dimensions might not be feasible using simple Monte Carlo approximations of the expectations in parameter space. This is likely because of the much higher dimensionality of the problems we consider---especially in comparison to \citet{osb2021neural}.

Specifically, we use an acquisition sequence created using active sampling which achieves ~90\% accuracy on MNIST with just 70 samples and use the same model and training setup as \citet{kirsch2021practical}\footnote{The preprint version contains information gain experiments.}. After every new data point (starting from 20), we evaluate a retrained model using OBI with 5 additional data points and compare it to a fully retrained model with the same 5 additional data points as well as to a model that is not retrained at all. We train the models (5 trials) for each training set size and (bootstrap) sample 10000 MC dropout samples for OBI 5 times out of 20000 MC dropout samples using consistent MC dropout (\Cref{chapter:batchbald}) for each model (5 sub-trials) to reduce the variance.

Ideally, when using OBI, we should recover the same performance as if we fully retrained the models using the additional data. However, as is visible in \Cref{obi:fig:obi_mnist_comparison}, this is not the case when using the challenging acquisition sequence from active sampling.

\Cref{obi:tab:obi_mnist_comparison} shows the average performance difference when using OBI with additional samples from the acquisition sequence and when fully retraining. In all cases, OBI performs worse than retraining. On the random acquisition sequence, it performs only as well as not updating at all, while on the active acquisition sequence, it always performs worse.

\section{Related Work}
\label{obi:sec:related_work}
The most relevant works and indeed one of the inspirations for this work are \citet{wen2021predictions} and \citet{osb2022evaluating}, which are recommended reading. We see this chapter as a contribution that provides a different and differentiated position on the benefits of marginal versus joint predictives and respective cross-entropies as performance metrics and that puts greater focus on OBI.

Moreover, our suggested experimental settings expand on these prior works and draw attention to active learning and active sampling as more realistic use-cases. Measuring the error between OBI and retrained models expands on the experiments in \citet{wen2021predictions} while evaluating performance in active learning and active sampling on highly redundant datasets (allowing to re-select previously selected points) expands on the idea of dyadic sampling from \citet{osb2022evaluating}.

Lastly, \citet{wen2021predictions} focus on the KL divergence between the exact Bayesian joint predictive and the joint predictive of an approximate Bayesian model for different numbers of samples in the joint.
Our suggested experimental settings focus on evaluating downstream tasks.

\citet{wang2021beyond} also examine the quality of joint predictives on low-dimensional datasets using a more synthetic evaluation method, the cross-normalized log-likelihood. They also evaluate the quality of joint predictives in regression settings using active learning experiments. Our evaluation settings from \S\ref{obi:sec:evaluations_applications} extend these.

\newcommand{\dkl}[1]{d_{KL}^{#1}}
\newcommand{\dce}[1]{d_{CE}^{#1}}

\section{Discussion}

In this short chapter, we have revisited the difference between marginal and joint cross-entropies and predictives, clarifying in which contexts either is appropriate: for offline learning, the marginal cross-entropy is the right choice to evaluate performance while for online learning, it is the joint cross-entropy. 
We have shown how the joint predictive plays an important role in information-theoretic acquisition functions in active learning and active sampling.

Importantly, we argue that online Bayesian inference could provide many benefits and have proposed new more practical and challenging experimental settings which expand on prior art by using active learning and active sampling.

Given the results of the presented experiment, it is an open question how much better other sampling-based approaches can be when using high-dimensional parameter spaces. Especially deep ensembles which usually provide a much smaller ``sample count'' (i.e.~number of ensemble members) might not perform well under online Bayesian inference because the hypothesis space will be exhausted faster---even when the ensemble members are diverse. 
Future research will hopefully offer further experimental evaluation following \S\ref{obi:sec:evaluations_applications}, e.g.~improving the quality of online Bayesian inference by studying higher quality posterior distributions such as those from HMC or efficient low-dimensional posterior approximations that might make parameter-space integrals tractable.
Prior research into failures of Bayesian model averaging under model misspecification might provide further insights and paths to improvements \citep{Minka2002BayesianMA}.

Using the suggested experimental settings from this chapter (or rather the original preprint), \citet{herde2022fast} suggest using last-layer Laplace. Similarly, \citet{osband2022fine} explore active learning for evaluating their proposed EpiNets with great results.

\FloatBarrier

}

{
	
	\graphicspath{{./epig/figures/}}

\newcommand{\ptest}[1]{\pcof{\mathrm{test}}{#1}}
\newcommand{\petest}[1]{\hpcof{\mathrm{test}}{#1}}
\newcommand{\ppool}[1]{\pcof{\mathrm{pool}}{#1}}
\newcommand{\peval}[1]{\pcof{\mathrm{eval}}{#1}}

\newcommand{\Dbatch}[1]{\Dany^\mathrm{acq}(#1)}
\newcommand{\yDeval}{\Dany^\mathrm{eval}(\yevalset)}

	\graphicspath{{./pobal/}{./epig/figures/}}
\makeatletter
\def\input@path{{./pobal/}}
\makeatother

\setlength{\figureheight}{4cm}

\newcommand{\basedata}[0]{\mathcal{D}_{\mathrm{base}}}
\newcommand{\pooldata}[0]{\mathcal{D}_{\mathrm{pool}}}
\newcommand{\traindata}[0]{\mathcal{D}_{\mathrm{train}}}
\newcommand{\testdata}[0]{\mathcal{D}_{\mathrm{test}}}

	\begin{savequote}[75mm]
		Strategy without tactics is the slowest route to victory. Tactics without strategy is the noise before defeat.
		\qauthor{Sun Tzu, The Art of War}
	\end{savequote}
	\chapter{\pobaltitle}
	\label{chapter:pobal}
Historically the literature in active learning and Bayesian experimental design has focused on trying to maximize the expected information gain (EIG) in the model parameters.
This yields the acquisition function typically known as BALD, having been popularized by a method called Bayesian active learning by disagreement \citep{houlsby2011bayesian}, which  has been successfully applied in settings including computer vision \citep{gal2017deep} and natural-language processing \citep{shen2017deep} as we have seen already in this thesis.

In this chapter, we highlight that BALD can be misaligned with our typical overarching goal of making accurate predictions on unseen inputs.
In machine learning, we often care about predictions more than the parameter distribution---the parameters $\W$ are not actually what we care about: they are merely a stepping stone to the predictions we will later make, and these future predictions are the ultimate quantity of interest.
Why can BALD be misaligned with this, and why is this distinction important? 
Unfortunately, BALD neglects a crucial fact: not all information about the model parameters is equally useful when it comes to making predictions.
This distinction can be surprisingly important: accurate predictions may not correspond to the most precise knowledge of the model parameters themselves.
With a non-parametric model, for instance, we can gain an infinite amount of information about the model parameters without any of it being relevant to prediction on inputs of interest.
More generally, different information about the model parameters may not be equal in regard to how it enables effective prediction for a particular target input distribution over possible inputs $\x$.
The models we deploy are often of limited capacity or misspecified, and there are trade-offs in the performance of the model depending on what we value \citep{cobb2018loss}. 
In short, BALD lacks a notion of how the model will be used and so fails to ensure that the data acquired is relevant to our particular predictive task.
As a result, BALD is liable to acquire labels that are informative with respect to the model parameters but not with respect to the predictions of interest.

\paragraph{Relevance.}
This has considerable practical implications.
Real-world datasets are often messy, with inputs that vary widely in their relevance to a given task.
Large pools of audio, images and text commonly fit this description \citep{ardila2020common,gemmeke2017audio,mahajan2018exploring,radford2021learning,raffel2020exploring,sun2017revisiting}.
We may have data from different sources of varying fidelity and relevance to our task. Indeed, if we have a large unlabeled dataset generated by scraping the internet, for example, we might have a significant variance in how closely related individual data points are to the task we actually care about---if the task is to detect hate speech in text, social-media posts are much more useful than articles about set theory.
Likewise, consider the outputs of large-scale experiments: if we want to predict the behavior of plasma in a new fusion-reactor configuration, results from similar configurations will likely be more pertinent than those from completely different ones.
In an extreme case, the task distribution might consist of samples that cannot even be directly labeled. This might apply in protein folding, where complex human proteins might make up the task distribution of interest, which would be too costly to crystallize, while the pool set might contain simpler proteins which could be more easily crystallized and their three-dimensional structure learned.
Here, we will show that BALD can be actively counterproductive in cases like these, picking out the most obscure and least relevant inputs, potentially.

\begin{figure}[t]
    \centering
    \begin{subfigure}[t]{\linewidth}
        \centering
        \includegraphics[height=\figureheight]{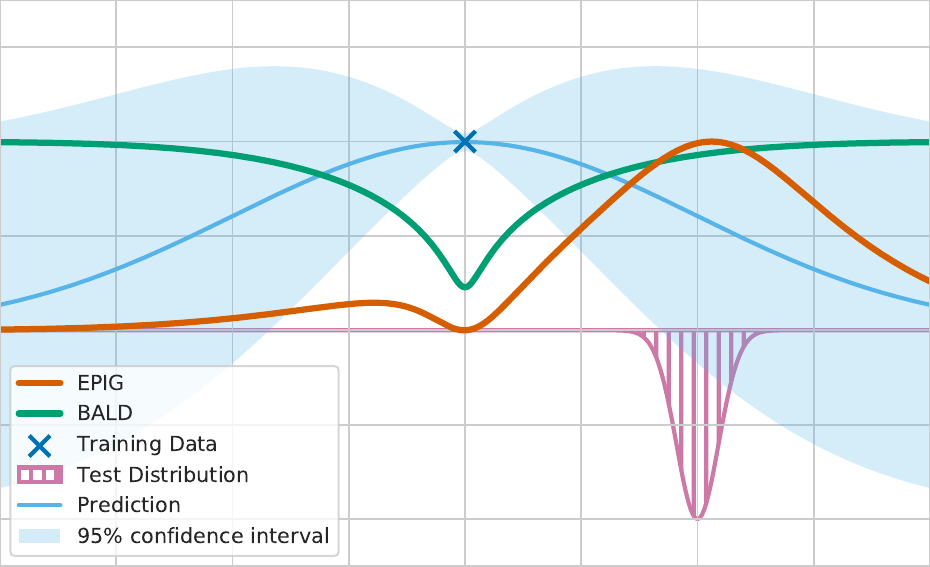}
        \caption{1D Toy Example}
        \label{pobal:fig:heatmaps:1d}
    \end{subfigure}
    \begin{subfigure}[t]{\linewidth}
        \includegraphics[width=\linewidth,trim={0.2cm 0 0.2cm 0}]{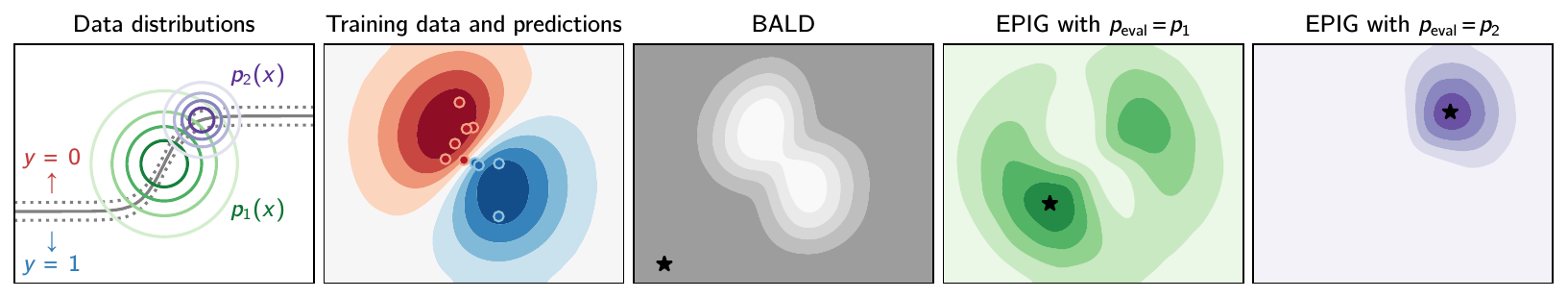}
        \caption{2D Toy Example}
        \label{pobal:fig:heatmaps:2d}
    \end{subfigure}
    \caption{
        \emph{The expected predictive information gain (EPIG) can differ dramatically from the expected information gain in the model parameters (BALD).}
        \textbf{\subref{pobal:fig:heatmaps:1d}:}
        We consider a simple 1D regression task with a Gaussian process model. Larger values of the acquisition function indicate a preference for those points.
        While BALD is predominantly concerned with labeling inputs far away from the data it has already seen, EPIG takes into account the target input distribution (the test distribution).
        \textbf{\subref{pobal:fig:heatmaps:2d}:} BALD increases (darker shading) as we move away from the existing data, yielding a distant acquisition (star) when maximized.
        It seeks a global reduction in parameter uncertainty, regardless of any input distribution.
        In contrast, EPIG is maximized only in regions of relatively high density under the target input distribution, $\peval{\xeval}$.
        It seeks a reduction in parameter uncertainty only insofar as it reduces predictive uncertainty on samples from $\peval{\xeval}$.
        See \Cref{pobal:sec:2d_data} for details.
    }
    \label{pobal:fig:heatmaps}
\end{figure}

\paragraph{EPIG.}
To address BALD's shortcomings, we propose the expected predictive information gain (EPIG), an alternative acquisition function.
We derive EPIG by returning to the foundational framework of Bayesian experimental design \citep{lindley1956measure}, from which BALD itself is derived.
Whereas BALD is the EIG in the model parameters, EPIG is the EIG in the model's predictions: it measures how much information the label of a candidate input is expected to provide about the label of a random target input.
While BALD favors global reductions in parameter uncertainty, EPIG favors only information that reduces downstream predictive uncertainty 
(\Cref{pobal:fig:heatmaps}).
Thus, EPIG allows us to directly seek improvements in predictive performance.

The randomness of the target input in EPIG is critical.
We do not aim for predictive information gain on a particular input or set of inputs.
Instead, the gain is in expectation with respect to a target input distribution.
This can be chosen to be the same distribution that the pool of unlabeled inputs is drawn from, or it can be a distinct distribution that reflects a downstream task of interest.

We find that EPIG often produces notable gains in final predictive performance over BALD across a range of datasets and models.
EPIG's gains are largest when the pool of unlabeled inputs contains a high proportion of irrelevant inputs with respect to the target input distribution.
But its advantage still holds when the pool is directly drawn from this distribution.
As such, it can provide a simple and effective drop-in replacement for BALD in many settings.

\paragraph{JEPIG.}
We also provide a preliminary examination of an alternative target input distribution acquisition function, which we name the joint expected predictive information gain (JEPIG).
Like the EIG, JEPIG is still based around targeting information gain in the model parameters, but it discounts information that is not relevant to model's prediction for the target input distribution.
Specifically, it discards from the EIG any information that will remain unknown after we would have acquired the labels for the target input distribution, on the basis that this information would not be helpful for prediction.

\section{Shortfalls of BALD}

\label{pobal:sec:bald_shortfalls}

In this section, we highlight that BALD can be poorly suited to the prediction-oriented settings that constitute much of machine learning.
We explain that this stems from the mismatch that can exist between parameter uncertainty and predictive uncertainty.
Importantly, we find that BALD does not take the input distribution into account at all---it, thus, cannot target the predictive uncertainty of the inputs we actually want to make predictions on.

\subsection{No Focus on Predictions}

In statistics, it is common for the model parameters to be valued in their own right \citep{beck1977parameter,blei2003latent,fisher1925statistical}.
But in many machine-learning contexts, particularly the supervised settings where BALD is typically applied, the parameters are only valued insofar as they serve a prediction-oriented goal.
We often, for example, seek the parameters that maximize the model's predictive performance on a test data distribution \citep{hastie2009elements}.
This frequentist notion of success often remains our motivation even if we use a Bayesian approach to data acquisition and/or learning \citep{komaki1996asymptotic,snelson2005sparse}.

\subsection{No Notion of an Input Distribution}

\label{pobal:sec:input_dist}

In order to reason about what information is relevant to prediction, we need some notion of the inputs on which we want to make predictions.
Without this we have no mechanism to ensure the model we learn is well-suited to the task we care about.
Our model could be highly effective on inputs from one region of input space but useless for typical samples from an input distribution of interest.

Appreciating the need to account for which inputs might arise at test time, it becomes clear why BALD can be problematic.
BALD focuses on the model parameters in isolation, with no explicit connection to prediction.
As such, it does not account for the distribution over inputs.

\subsubsection{Failures under Real-World Data}

BALD can be particularly problematic in the very settings that often motivate active learning: those where we have access to a large pool of unlabeled inputs whose relevance to some task of interest varies widely.
In contrast with the carefully curated datasets often used in basic research, real-world data is often drawn from many sources of varying fidelity and relation to the task.
Pools of web-scraped audio, images and text are canonical examples of this.
Active learning ought to help deal with the mess by identifying only the most useful inputs to label.
But BALD can in fact be worse than random acquisition in these settings, targeting obscure data that is not helpful for prediction.

The experiment presented in \Cref{pobal:fig:bald_pool_size} highlights this flaw.
As we increase the size of the pool that BALD is maximized over, inputs of greater obscurity become more likely to be included in the pool, and BALD produces worse and worse predictive accuracy.
This result is corroborated by the work of \citet{karamcheti2021mind}.
Focusing on visual-question-answering tasks, they found that BALD failed to outperform random acquisition when using uncurated pools, and that a substantial amount of curation was required before this shortfall could be overturned.

\subsubsection{Failures \emph{without} Distribution Shift}

\begin{figure}[t]
    \begin{subfigure}[t]{0.49\linewidth}
        \centering
        \includegraphics[height=0.95\figureheight,trim={0 0 0 0}]{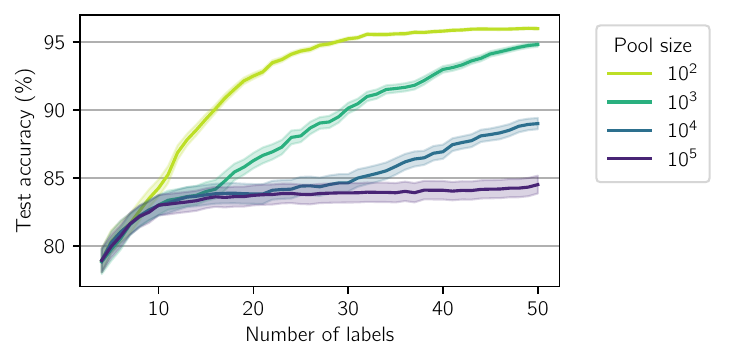}
        \caption{Pool Size Ablation}
        \label{pobal:fig:bald_pool_size}
    \end{subfigure}\hfill
    \begin{subfigure}[t]{0.49\linewidth}
        \centering
        \includegraphics[height=0.95\figureheight,trim={0 0 0 0}]{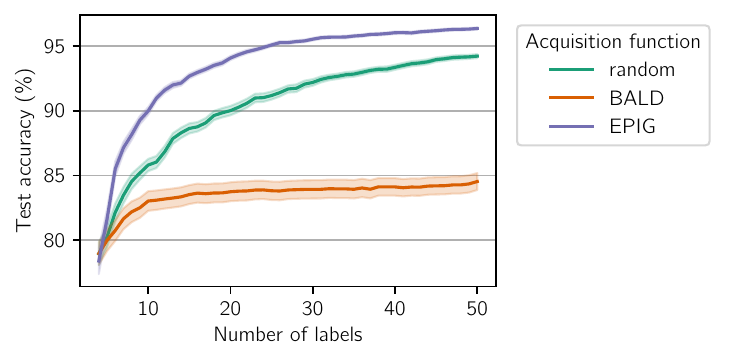}
        \caption{EPIG vs BALD vs Random}
        \label{pobal:fig:2d_data}
    \end{subfigure}    
    \caption{
        \emph{Shortfalls of BALD, and how we fix them.}
        See \Cref{pobal:fig:heatmaps} for intuition and \Cref{pobal:sec:2d_data} for details.
        \textbf{\subref{pobal:fig:bald_pool_size}}
        BALD can fail catastrophically on big pools.
        A bigger pool typically contains more inputs with low density under the data-generating distribution.
        Often these inputs are of low relevance if the aim is to maximize expected predictive performance.
        BALD can nevertheless favor these inputs.
        \textbf{\subref{pobal:fig:2d_data}}
        In contrast with BALD, EPIG deals effectively with a big pool ($10^5$ unlabeled inputs).
        BALD is overwhelming counterproductive even relative to random acquisition.
    }
\end{figure}

It might be tempting to just think of this problem with BALD as being analogous to the issues caused by train-test input-distribution shifts elsewhere in machine learning.
But the problem is more deep-rooted than this: BALD has no notion of any input distribution in the first place.
This is why increasing the size of the pool can induce failures as in \Cref{pobal:fig:bald_pool_size}, without any distribution shift or changes to the distribution that the pool inputs are drawn from.
Distribution shift can cause additional problems for BALD, as the results in \Cref{pobal:sec:experiments} show.
But it is by no means a necessary condition for failure to occur.

\subsection{Not All Information is Equal}

In some models, such as linear models, parameters and predictions are tightly coupled, but more generally the coupling can be loose.
This means that a reduction in parameter uncertainty typically might not yield a wholesale reduction in predictive uncertainty \citep{chaloner1995bayesian}.
Deep neural networks, for instance, can have substantial redundancy in their parameters \citep{belkin2019reconciling}, while Bayesian non-parametric models can be thought of as having an infinite number of parameters \citep{hjort2010bayesian}.
When the coupling is loose, parameter uncertainty can be reduced without a corresponding reduction in predictive uncertainty on inputs of interest.
In fact, it is possible to gain an infinite amount of parameter information while seeing an arbitrarily small reduction in predictive uncertainty.

\begin{example}[Infinite Information Gain]
\label{pobal:example:gp}
{
    \newcommand{\ff}{\mathbf{f}}
    \newcommand{\FF}{\mathbf{F}}
Consider the supervised-learning problem depicted in \Cref{pobal:fig:heatmaps:1d}, where $x\in\mathbb{R}$ is an input, $y\in\mathbb{R}$ is a label, and we use a model consisting of a Gaussian likelihood function, $p(y|x,\ff) = \normaldist{\ff(x)}{1}$, and a zero-mean Gaussian-process prior, $\ff \sim \mathrm{GP}(0,k)$, with covariance function $k(x,x') = \exp(-\nicefrac{(x-x')^2}{2l^2})$, where $l$ is the length scale of the model \citep{williams2006gaussian}, e.g.\ $l=2$.
Suppose we are interested in making predictions $\Yeval \given \Xeval$ in the region, e.g.\ $\xeval \sim \normaldist{2}{(1/4)^2}$, and consider we are observing only odd integers $x$: $x_1, x_2, \dots, x_M$ for some $M \in \mathbb{N}^+$ with predictions $Y_1, Y_2, \dots, Y_M$. We only allow odd integers to avoid direct overlap with the region of interest as we decrease as the length scale (to make this example easier to reason about).

We will vary the length scale and the number of points $M$ to show that BALD can become arbitrarily large while the predictive uncertainty of interest, $\Hof{\Yeval \given \Xeval, \ldots}$, will change arbitrarily little. 

BALD is the EIG between the predictions and GP functions $\ff \sim GP(0,k)$ (as a non-parametric model):
\begin{align}
    &\MIof{\FF ; Y_1, Y_2, \dots, Y_M; \given x_1, x_2, \dots, x_M} = \ldots\\
    &\quad = \Hof{Y_1, Y_2, \dots, Y_M \given x_1, x_2, \dots, x_M} - \Hof{Y_1, Y_2, \dots, Y_M \given x_1, x_2, \dots, x_M, \FF}. \notag
\end{align}
The first term is the (joint) entropy of the predictions (incl. the Gaussian likelihood), and the second term is the (joint) entropy of the Gaussian likelihood itself.
The entropy of the multivariate Gaussian distribution with covariance $\Sigma \in \mathbb{R}^{M\times M}$ is \citep{cover2005elements}:
\begin{align}
    \Hof{\normaldist{\mu}{\Sigma}} &= \frac{1}{2} \log \left( \left(2\pi\right)^{M} \det \Sigma \right).
\end{align}
For the two terms, we have:
\begin{align}
    \Hof{Y_1, Y_2, \dots, Y_M \given x_1, x_2, \dots, x_M} &= \frac{1}{2} \log \left( \left(2\pi\right)^{M} \det \left( (k(x_i, x_j))_{ij} + \mathrm{Id}_M \right) \right), \\
    \Hof{Y_1, Y_2, \dots, Y_M \given x_1, x_2, \dots, x_M, \FF} &= \frac{1}{2} \log \left( \left(2\pi\right)^{M} \det \left( \mathrm{Id}_M \right) \right).
\end{align}
We see that the $\log \left(2\pi\right)^{M}$ constants cancel out in both terms, and we end up with:
\begin{align}
    &\MIof{Y_1, Y_2, \dots, Y_M; \FF \given x_1, x_2, \dots, x_M} \\
    &\quad = \frac{1}{2} \log \det \left( (k(x_i, x_j))_{ij} + \mathrm{Id}_M \right) - \frac{1}{2} \underbrace{\log \det \mathrm{Id}_M}_{=\log 1=0} \notag \\
    &\quad = \frac{1}{2} \log \det \left( (k(x_i, x_j))_{ij} + \mathrm{Id}_M \right).
\end{align}
Now we can use the length scale: for $l \to 0$, we have $k(x_i, x_j) \to \indicator{i=j}$---the kernel matrix becomes an identity matrix---and thus, $\det \left( (k(x_i, x_j))_{ij} + \mathrm{Id}_M \right) \to \det \left( 2 \mathrm{Id}_M \right) \to 2^M$.
Similarly, we are interested in the reduction in the predictive uncertainty, $\Hof{\Yeval \given \Xeval, \ldots}$:
\begin{align}
    \Hof{\Yeval \given \Xeval} - \Hof{\Yeval \given \Xeval, Y_1, x_1, Y_2, x_2, \dots, Y_M, x_M}.
\end{align}
With our eyes now well-trained for information quantities, we can see that this just the EIG of the predictive $\Yeval \given \Xeval$ with respect to the samples $Y_1, Y_2, \dots, Y_M \given x_1, x_2, \dots, x_M$:
\begin{align}
    \MIof{\Yeval; Y_1, Y_2, \dots, Y_M \given \Xeval, x_1, x_2, \dots, x_M}.
\end{align}
This is the expected predictive information gain, which we will introduce in \Cref{pobal:sec:method}.
We can compute this quantity via $\Hof{A \given B}=\Hof{A, B} - \Hof{B}$. Let us fix $\xeval$. The constant $\log$ terms cancel out again, and we have:
\begin{align}
    &\MIof{\Yeval; Y_1, Y_2, \dots, Y_M \given \xeval, x_1, x_2, \dots, x_M} \notag \\
    &\quad = \Hof{\Yeval \given \xeval} + \Hof{Y_1, Y_2, \dots, Y_M \given x_1, x_2, \dots, x_M}\\
    &\quad\quad- \Hof{\Yeval, Y_1, Y_2, \dots, Y_M \given \xeval, x_1, x_2, \dots, x_M}\notag \\
    &\quad = \frac{1}{2} \log \det \left( (k(\xeval, \xeval)) + \mathrm{Id}_M \right)
    + \frac{1}{2} \log \det \left( (k(x_i, x_j))_{ij} + \mathrm{Id}_M \right)\\
    &\quad\quad - \frac{1}{2} \log \det \left( (k(x_i, x_j))_{i,j=\{*, 1..M\}} + \mathrm{Id}_M \right) \notag \\
    &\quad \overset{l \to 0}{\to} \frac{1}{2} \log 2 + \frac{1}{2} M \log 2 - \frac{1}{2} (M+1) \log 2\\
    &\quad = 0.
\end{align}
Then, taking the expectation over $\xeval$, we have:
\begin{align}
    &\MIof{\Yeval; Y_1, Y_2, \dots, Y_M \given \Xeval, x_1, x_2, \dots, x_M} \\
    &\quad = \E{\pof{\xeval}}{\MIof{\Yeval; Y_1, Y_2, \dots, Y_M \given \xeval, x_1, x_2, \dots, x_M}} \notag\\
    &\quad \overset{l \to 0}{\to} \E{\pof{\xeval}}{0} = 0.
\end{align}
Altogether, as $l \to 0$ and for $M \to \infty$, BALD diverges to infinity while the EIG in the prediction of interest converges to zero:
\begin{align}
    \MIof{\FF; Y_1, Y_2, \dots, Y_M \given x_1, x_2, \dots, x_M} &\overset{l \to 0}{\to} \frac{M}{2} \log 2 \overset{M \to \infty}{\to} \infty,\\
    \MIof{\Yeval; Y_1, Y_2, \dots, Y_M \given \Xeval, x_1, x_2, \dots, x_M} &\overset{l \to 0}{\to} 0 \implies \Hof{\Yeval \given \Xeval, \ldots} \approx const.
\end{align}

This example is a concrete demonstration that a high BALD score need not coincide with any reduction in the predictive uncertainty of interest. %
If the aim is to predict, then maximizing BALD is not guaranteed to help to any extent whatsoever.
}
\end{example}

\section{Distribution-Aware Acquisition Functions}

Motivated by BALD's weakness in prediction-oriented settings and the need for principled approaches,
we return to the framework of Bayesian experimental design that underlies BALD, and derive two acquisition functions that we call the expected predictive information gain (EPIG) and the joint expected predictive information gain (JEPIG).
Whereas BALD targets a reduction in parameter uncertainty, EPIG and JEPIG directly target a reduction in predictive uncertainty on inputs of interest.

\subsection{Why not use Filtering Heuristics?}
\label{pobal:sec:filtering}

We might suppose we could just discard irrelevant data before deploying BALD.
But this filtering process would require us to be able to determine each input's relevance at the outset of training, which is impractical in many cases.
Even if we have access to a target input distribution, this on its own can be insufficient for judging relevance to a task of interest.
A candidate input could have relatively low density under the target distribution but nevertheless share high-level features with a target input, such that the two inputs' labels are highly mutually informative.
With high-dimensional inputs, it can also be surprisingly difficult to identify unrepresentative inputs purely through their density~\citep{nalisnick2018deep}.
Rather than trying to design an auxiliary process to mitigate BALD's problematic behavior, we seek an acquisition function that can automatically determine what is relevant.

\subsection{Two Alternative Expected Information Gains}
\label{pobal:sec:two_alternatives}

To be precise, when we talk about EIG and BALD, these are the \emph{parameter EIG}, which does not focus on predictions and is not distribution-aware. 
But we can also look at different EIGs which focus on the predictions: target EIG or \emph{predictive EIG}\footnote{Any naming scheme is going to be confusing and ambiguous given different prior art, so we might sadly add to this confusion.}.

\paragraph{Evaluation Distribution.}
To reason about the predictions we are interested in, we need an explicit notion of the predictions we want to make with our model.
We therefore introduce a random target input, $\xeval \sim \peval{\xeval}$, for unlabeled \emph{evaluation samples} from the target input distribution and define our goal to be the confident prediction\footnote{We still have to hope that by virtue of providing correct labels for training, confident predictions will also be likely correct predictions.} of $\yeval \given \xeval$. 

Similar to other unlabeled samples in active learning, these samples could be either provided in a stream-based setting, where we can repeatedly draw new unlabeled samples i.i.d., or in a pool-based setting, where we only have a fixed reservoir of evaluation samples, which we call the \emph{evaluation set} \(\Deval\).
In the pool-based setting, we denote the \emph{empirical} distribution of the evaluation set by \(\peval{\xeval}\), whereas in a stream-based setting, we could set \(\peval{\xeval} \defeq \ptest{\xeval}\). This allows us to abstract away the exact setting in the following exposition.

\paragraph{Predictive EIG.}
Let us fix a single target input $\xeval \sim \peval{\xeval}$ with prediction $\Yeval \given \xeval$, and consider the expected information gain given a single pool sample $\xacq \in \Dpool$ with prediction $\Yacq \given \xacq$.
The corresponding predictive EIG is, of course:
\begin{align}
    \MIof{\Yeval ; \Yacq \given \xeval, \xacq}.
\end{align}

The case when we want to take the whole distribution $\peval{\xeval}$ over $\xeval$ into account is slightly more complicated.
There are two alternatives that we can examine:
\begin{itemize}
    \item the \emph{(mean) marginal predictive EIG},
    \begin{align}
        \MIof{\Yeval; \Yacq \given \Xeval, \xacq},
    \end{align}
    where we take an expectation over $\Xeval$, hence `mean'; and 
    \item the \emph{joint predictive EIG},
    \begin{align}
        \MIof{\Yevalset; \Yacq \given \xevalset, \xacq},
    \end{align}
    for an empirical evaluation set $\Deval=\xevalset$. (We will look at the general case below.)
\end{itemize}
We will see that the marginal predictive EIG is equivalent to the expected reduction in generalization loss if we were to acquire $\xacq$, while the joint predictive EIG is equivalent to performing Bayesian model selection on the pool set.

In the next sections, we will explore both using slightly different names, the joint predictive EIG as joint expected predictive information gain, and the (marginal) predictive EIG as expected predictive information gain. 
Sadly, there are several naming schemes. EPIG and JEPIG are mentioned in already published papers that spun out of this thesis, and while above names seem more fitting in retrospect, they also do not exactly match the ones from \citet{mackay1992information}, who referred to them as `mean marginal information gain' and `joint information gain', respectively, and left out the rather important term `predictive'. The parameter EIG was referred to as `total information gain' in the same paper, which also makes sense:
\Cref{pobal:glue:mackayterms} shows that the (parameter) EIG upper-bounds both EPIG and JEPIG. Hence, it is the total information gain, while EPIG and JEPIG only focus on the part of the information gain that is relevant for the predictions.

\begin{figure}[t]
    \centering
    \tikzset{every picture/.style={line width=0.75pt}} %

    \begin{tikzpicture}[
        infoquantity/.style={line width=1pt},
        infoquantitylabel/.style={font=\large}
        ]

        \draw[infoquantity] (0,-3/2) circle (3/2);
        \draw[infoquantity] (-45:2.5/2) circle (2.5/2);
        \draw[infoquantity] (270-45:2.5/2) circle (2.5/2);

        \begin{scope}
            \begin{scope}
                \clip (-45:2.5/2) circle (2.5/2);
                \fill[infoquantity,solarized@blue!50!white] (270-45:2.5/2) circle (2.5/2);
            \end{scope}
            \begin{scope}
                \clip (0,-3/2) circle (3/2);
                \clip (-45:2.5/2) circle (2.5/2);
                \fill[
                fill=solarized@red!50!white,
                fill opacity=0.5] (-45:2.5/2) circle (2.5/2)
                (0,-3/2) circle (3/2);
            \end{scope}
            \begin{scope}
                \clip (-45:2.5/2) circle (2.5/2);
                \draw[infoquantity,solarized@blue] (270-45:2.5/2) circle (2.5/2);
            \end{scope}
            \begin{scope}
                \clip (0,-3/2) circle (3/2);
                \clip (-45:2.5/2) circle (2.5/2);
                \draw[line width=4pt,draw=solarized@red] (-45:2.5/2) circle (2.5/2)
                (0,-3/2) circle (3/2);
            \end{scope}
        \end{scope}

        \draw (0, -1.5/2) +(0,3/2) node [anchor=south, infoquantitylabel] {$\displaystyle \MIof{\Yeval; \Yacq \given \Xeval, \xacq}$ /
        $\displaystyle \MIof{\Yevalset; \Yacq \given \xevalset, \xacq}$} [-{Stealth}] -- (0, -1/2);

        \draw (0, 0) +(180-235.12:4.92/2+0.75) node [anchor=north west, infoquantitylabel] {$\displaystyle \MIof{\W; \Yacq \given \xacq}$} [-{Stealth}] -- (180-235.12:4.92/2);

        \path (0, -3/2) +(180+30:3.5/2) node [anchor=north east] {$\displaystyle \Hof{\W}$};
        \path (-45:2.5/2) +(30:3/2) node [anchor=south west] {$\displaystyle \Hof{\Yacq \given \xacq}$};
        \path (270-45:2.5/2) +(150:3/2) node [anchor=south east] {$\displaystyle \Hof{\Yeval \given \Xeval}$/$\displaystyle \Hof{\Yevalset \given \xevalset}$};
    \end{tikzpicture}
    \caption{%
    \emph{Visualizing both EPIG $\MIof{\Yevalset; \Yacq \given \xevalset, \xacq}$ \&\ JEPIG
    $\MIof{\Yeval; \Yacq \given \Xeval, \xacq}$ vs.\ EIG $\MIof{\W; \Yacq \given \xacq}$ in the same I-Diagram.}
    MacKay's `total information gain' for the EIG is a fitting term because we can immediately read off that it upper-bounds both EPIG and JEPIG.
    } %
    \label{pobal:glue:mackayterms}
\end{figure}

\section{Expected Predictive Information Gain}
\label{pobal:sec:method}

\begin{figure}[t]
    \begin{insightbox}{Expected Reduction in Generalization Loss}{pobal:insight:gen_loss}
        The same samples maximize EPIG and minimize the expected generalization loss \citep{roy2001toward}.
        EPIG measures the expected reduction in the uncertainty in a model's predictions on target points $\xeval \sim \peval{\xeval}$:
        \begin{align}
            &\MIof{\Yeval; \Yacqset \given \Xeval, \xacqset, \Dtrain} = \notag \\
            &\quad \Hof{\Yeval \given \Xeval, \Dtrain} - \Hof{\Yeval \given \Xeval, \Yacqset, \xacqset, \Dtrain}. 
        \end{align}
        Importantly, this objective is equivalent to minimizing the \emph{expected generalization loss} under the model's predictions:
        \begin{align}
            &\Hof{\Yeval \given \Xeval, \Yacqset, \xacqset, \Dtrain} = \label{pobal:eq:gen_loss} \notag \\
            &\quad = \simpleE{\peval{\xeval}}  \simpleE{\pof{\yacqset \given \xacqset, \Dtrain}} \simpleE{\pof{\yeval \given \xeval, \yacqset, \xacqset, \Dtrain}}\mathcal{L}(\xeval, \yeval),\\
            &\mathcal{L}(\xeval, \yeval) \defeq -\log \pof{\yeval \given \xeval, \yacqset, \xacqset, \Dtrain},
        \end{align}
        where we have marginalized over the model parameters
        \begin{align}
            &\w \sim \pof{\w \given \yacqset, \xacqset, \Dtrain}.   
        \end{align}
    \end{insightbox}
\end{figure}

To derive EPIG, we first consider the  information gain (IG) in $\yeval$ that results from conditioning on new data, $(\x, \y)$ and fix $\xeval \sim \pof{\xeval}$:
\begin{align}
    &\MIof{\Yeval ; \y \given \xeval, \x, \Dtrain} 
    = \Hof{\Yeval \given \xeval, \Dtrain} - \Hof{\Yeval \given \xeval, \y, \x, \Dtrain} \\
    &\quad = \xHof{\pof{\yeval \given \xeval, \Dtrain}} - \xHof{\pof{\yeval \given \xeval, \y, \x, \Dtrain}}
\end{align}
where $\pof{\yeval \given \xeval, \y, \x, \Dtrain}=\E{\pof{\w \given \y, \x, \Dtrain}}{\pof{\yeval \given \xeval, \y, \x, \w}}$.
Note that this is a function of $\xeval$ as well as $\x$ and $\y$.
Next we take an expectation over both the random target input, $\xeval$, and the unknown label, $\y$:
\begin{align}
\epigof{\x} &= \E{\peval{\xeval} \pof{\y \given \x, \Dtrain}}{\MIof{\Yeval ; \y \given \xeval, \x, \Dtrain}} \\
&= \MIof{\Yeval ; \Y \given \Xeval, \x, \Dtrain}
,
\end{align}
where we were able to use our concise \&\ unified notation for great benefit.
Thus, we see that EPIG is the expected reduction in predictive uncertainty at a randomly sampled target input, $\xeval$.

We can also take a frequentist perspective and relate it to expected reduction in prediction uncertainty, which is equivalent to the expected reduction in generalization loss (over $\peval{\xeval}$) as we detail in \RRectangle{\textbf{\cref{pobal:insight:gen_loss} Expected Reduction in Generalization Loss}}.

There are other interpretations too. For example, we could write EPIG as an expected KL divergence. More interesting is that we can write EPIG as an expected information gain between $(\Xeval, \Yeval)$ and $\Y \given \x$:
\begin{align}
    \MIof{(\Xeval, \Yeval) ; \Y \given \x, \Dtrain} = \underbrace{\MIof{\Xeval ; \Y \given \x, \Dtrain}}_{=0} + \MIof{\Yeval ; \Y \given \Xeval, \x, \Dtrain}.
\end{align}
The first term on the right-hand side is 0 because $\Xeval$ and $\Y \given \x$ are independent. We can also write EPIG as difference between two EIGs:
\begin{align}
    &\MIof{\Yeval ; \Y \given \Xeval, \x, \Dtrain} \\
    &\quad =
    \MIof{\Yeval ; \Y \given \Xeval, \x, \Dtrain} + \underbrace{\MIof{\Yeval ; \Y \given \Xeval, \x, \W, \Dtrain}}_{=0} \\
    &\quad = \MIof{\Yeval ; \Y ; \W \given \Xeval, \x, \Dtrain} \\
    &\quad = \MIof{\Yeval ; \W \given \Xeval, \Dtrain} - \MIof{\Yeval ; \W \given \Xeval, \Y, \x, \Dtrain}.
\end{align}
$\MIof{\Yeval ; \Y \given \Xeval, \x, \W, \Dtrain}$ is zero because $\Yeval$ and $\Y$ are independent given $\W$.
Thus, we can also view EPIG as the expected reduction in epistemic uncertainty for evaluation samples, $\Yeval$, given the model parameters, $\W$.

\subsection{Sampling Target Inputs}
\label{pobal:sec:sampling_x_star}

EPIG involves an expectation with respect to a target input distribution of evaluation samples, $\peval{\xeval}$.
In practice, we estimate this expectation by Monte Carlo and so require a sampling mechanism.

In many active-learning settings an input distribution is implied by the existence of a pool of unlabeled inputs.
There are cases where we know (or are happy to assume) the pool has been sampled from $\peval{\xeval}$.
Alternatively we might be forced to assume this is the case: perhaps we know the pool is not sampled from $\peval{\xeval}$ but lack access to anything better.
In these cases we can simply subsample from the pool to obtain samples of $\xeval$.
Empirically we find that this can work well relative to acquisition with BALD (\Cref{pobal:sec:experiments}).

Another important case is where we have access to samples from $\peval{\xeval}$, but we cannot label them.
Limits on the ability to acquire labels might arise due to privacy-related and other ethical concerns, geographical restrictions, the complexity of the labelling process for some inputs, or the presence of commercially sensitive information in some inputs.
At the same time there might be a pool of inputs for which we have no labelling restrictions.
In a case like this we can estimate EPIG using samples from $\peval{\xeval}$ while using only the pool as a source of candidates for labelling.
Thus, we can target information gain in predictions on samples from $\peval{\xeval}$ without labelling those samples themselves.

A further scenario that we might encounter is a classification problem where the pool is representative of the target class-conditional input distribution but not the target marginal class distribution: that is, $\ppool{\xeval \given \yeval}=\peval{\xeval \given \yeval}$ but $\ppool{\yeval}\neq\peval{\yeval}$.
The pool might, for example, consist of uncurated web-scraped inputs from many more classes than those we care about. 
In this scenario it can often be the case that we know or can reasonably approximate the distribution over classes that we are targeting, $\peval{\yeval}$.
With this we can approximately sample from $\peval{\xeval}$ using a combination of our model, $\pof{\y \given \x}$, and the pool.
We first note that
\begin{align}
\ppool{\xeval \given \yeval} = \frac{\ppool{\xeval} \ppool{\yeval \given \xeval}}{\int \ppool{\x}\ppool{\Y=\yeval \given x} dx}
.
\end{align}
Then, using the fact that $\ppool{\xeval \given \yeval} = \peval{\xeval \given \yeval}$, we get
\begin{align}
\peval{\xeval} &= \sum_{\yeval} \peval{\yeval} \peval{\xeval \given \yeval} \\
&= \ppool{\xeval} \sum_{\yeval} \frac{\peval{\yeval} \ppool{\yeval \given \xeval}}{\int p_{\mathrm{pool}}(x)p_{\mathrm{pool}}(y=\yeval|x) dx} \\
&\approx \ppool{\xeval} \sum_{\yeval} \frac{\peval{\yeval} \pof{\yeval \given \xeval}}{\frac{1}{N} \sum_{x\in\pooldata} \pof{y=\yeval \given x}} \\
&= \ppool{\xeval} \, w(\xeval)
,
\end{align}
where we have approximated $\ppool{\yeval \given \xeval}$ with our model.
Now we can approximately sample from $\peval{\xeval}$ by subsampling inputs from the pool using a categorical distribution with probabilities $w(\xeval)/N$.

\subsection{Estimation}
\label{pobal:sec:estimation}

The best way to estimate EPIG depends on the task and model of interest.
In the empirical evaluations in this chapter we focus on classification problems and use models whose marginal and joint predictive distributions are not known in closed form.
This leads us to use $\epigof{\x} \approx$
\begin{align}
\label{pobal:eq:epig_fast_estimator}
\frac{1}{M} \sum_{j=1}^{M} \Kale{\hpof{\Y,\Yeval \given \x,\xeval_j}}{\hpof{\Y \given \x}\,\hpof{\Yeval \given \xeval_j}}
,
\end{align}
where $\xeval_j \sim \peval{\xeval}$, and $\hat{\opp}$ denotes Monte Carlo approximations of the respective predictive distributions (see also \Cref{pobal:eq:marg_pred_posterior,pobal:eq:joint_pred_posterior}).
Classification is an instance of where the required expectation over $\y$ and $\yeval$ can be computed analytically, such that our only required estimation is from marginalizations over $\Theta$.

If we cannot integrate over $y$ and $\yeval$ analytically, we can revert to nested Monte Carlo estimation \citep{rainforth2018nesting}.
For this we first note that, using \Cref{pobal:eq:joint_pred_posterior}, we can sample $y,\yeval \sim \pof{y,\yeval \given x,\xeval}$ exactly by drawing a $\theta$ and then a $y$ and $\yeval$ conditioned on this $\theta$.
By also drawing samples for $\theta$, we can then construct the estimator $\epigof{\x} \approx$
\begin{align}
\label{pobal:eq:epig_sample_estimator}
\frac{1}{M} \sum_{j=1}^{M} \log \frac{
        K \sum_{i=1}^K \pof{\y_{j} \given \x,\theta_i}\pof{\yeval_{j} \given \xeval_{j},\theta_i}
    }{
        \sum_{k=1}^K \pof{\y_{j} \given \x,\theta_k} \sum_{k=1}^K \pof{\yeval_{j} \given \xeval_{j},\theta_k}
    }
,
\end{align}
where $\y_{j},\yeval_{j}\sim \pof{\y,\yeval \given \x,\xeval_{j}}$, $\theta_i \sim \pof{\theta}$, and $\xeval_{j} \sim \peval{\xeval}$.
Subject to some weak assumptions on $\opp$, this estimator converges as $K,M\to\infty$ \citep{rainforth2018nesting}.

\newcommand{\landauO}{\ensuremath{\mathcal{O}}}
The EPIG estimators in \Cref{pobal:eq:epig_fast_estimator,pobal:eq:epig_sample_estimator} each have a total computational cost of $\landauO(MK)$.
This is comparable to BALD estimation for regression problems.
But it can be more expensive than BALD estimation for classification problems: BALD can be collapsed to a non-nested Monte Carlo estimation for an $\landauO(K)$ cost, but EPIG cannot.

Other possible estimation schemes include a variational approach inspired by \citet{foster2019variational}.
This is too expensive to be practically applicable in the settings we consider but could be useful elsewhere.
See \Cref{pobal:sec:epig_estimation} for details.

\subsection{Empirical Validation}
\label{pobal:sec:experiments}

For consistency with existing work on active learning for prediction, our empirical evaluation of EPIG focuses on classification problems.
Code for reproducing our results is available at \shorturl{github.com/fbickfordsmith/epig}.

\subsection{Synthetic Data (\Cref{pobal:fig:heatmaps,pobal:fig:bald_pool_size,pobal:fig:2d_data})}
\label{pobal:sec:2d_data}

First we demonstrate the difference between BALD and EPIG in a setting that is easy to visualize and understand: binary classification with two-dimensional inputs.

\paragraph{Data.}

The first input distribution of interest, denoted $p_1(x)$ in \Cref{pobal:fig:heatmaps}, is a bivariate Student's $t$ distribution with $\nu=5$ degrees of freedom, location $\mu=[0,0]$ and scale matrix $\Sigma=0.8I$.
The second distribution, denoted $p_2(x)$ in \Cref{pobal:fig:heatmaps}, is a scaled and shifted version of the first, with parameters $\nu=5$, $\mu\approx[0.8,0.9]$ and $\Sigma=0.4I$.
This serves to illustrate in \Cref{pobal:fig:heatmaps} how EPIG's value changes with the target input distribution; it is not used elsewhere.
The conditional label distribution is defined as $p(\y=1|x)=\Phi(20 (\tanh(2x_{[1]}) - x_{[2]}))$, where $x_{[i]}$ denotes the component of input $x$ in dimension $i$, and $\Phi$ is the cumulative distribution function of the standard normal distribution.
For the training data in \Cref{pobal:fig:heatmaps}, we sample ten input-label pairs, $\traindata=\{(x_i,\y_i)\}_{i=1}^{10}$, where $x_i, y_i \sim p(y|x)p_1(x)$.
Likewise, we sample $\testdata=\{(x_i,\y_i)\}_{i=1}^{10,000}$ for evaluating the model's performance in active learning.

\paragraph{Model and Training.}

We use a model with a probit likelihood function, $\pof{\y=1 \given x,\theta}=\Phi(\theta(x))$, where $\Phi$ is defined as above, and a Gaussian-process prior, $\theta \sim \mathrm{GP}(0,k)$, where $k(x,x')=10\cdot\exp\left(-\sfrac{\|x-x'\|^2}{2}\right)$.
The posterior over latent-function values cannot be computed exactly, so we optimize an approximation to it using variational inference \citep{hensman2015scalable}.
To do this we run 10,000 steps of full-batch gradient descent using a learning rate of 0.005 and a momentum factor of 0.95.

\paragraph{Active Learning.}

We initialize the training dataset, $\traindata$, with two randomly sampled inputs from each class.
Thereafter, we run the active-learning loop described in \Cref{glue:background:active_learning} until a budget of 50 labels is used up.
We acquire data using three acquisition functions: random, BALD and EPIG.
Random acquisition involves sampling uniformly from the pool without replacement.
We estimate BALD using \Cref{pobal:eq:bald_classification_estimator} and EPIG using \Cref{pobal:eq:epig_fast_estimator}, in both cases drawing 5,000 samples from the model's approximate posterior.
For EPIG we sample $\xeval\sim p_1(\xeval)$.
After each time the model is trained, we evaluate its predictive accuracy on $\testdata$ as defined above.
Using a different random-number-generator seed each time, we run active learning with each acquisition function 100 times.
We report the test accuracy (mean $\pm$ standard error) as a function of the size of $\mathcal{D}_\mathrm{train}$.

\paragraph{Discussion}

\Cref{pobal:fig:2d_data} shows a striking gap between BALD and EPIG in active learning.
\Cref{pobal:fig:heatmaps,pobal:fig:bald_pool_size} provide some intuition about the underlying cause of this disparity: BALD has a tendency to acquire labels at the extrema of the input space, regardless of their relevance to the predictive task of interest.

\begin{figure*}[t]
\centering
\includegraphics[height=\figureheight,trim={0 0 0 0}]{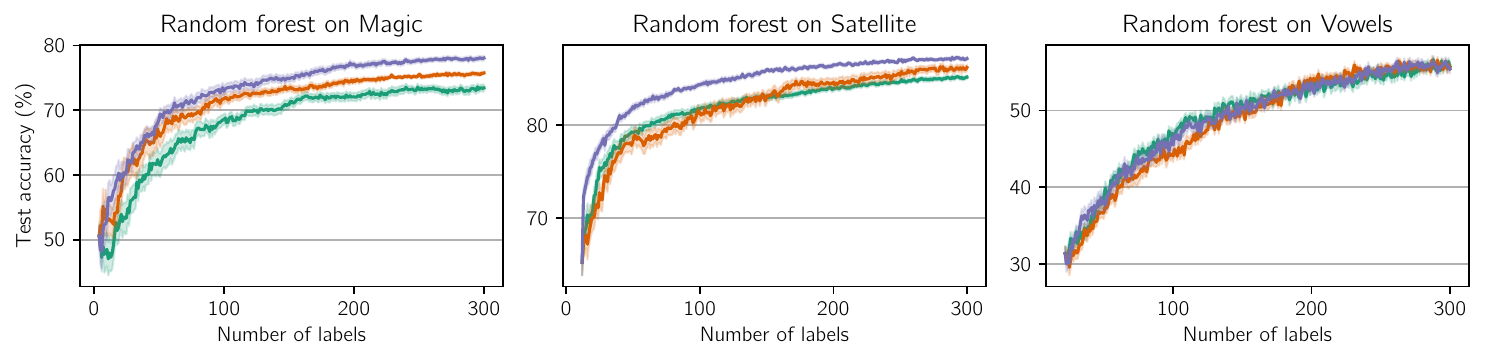}
\includegraphics[height=\figureheight,trim={0 0 0 0}]{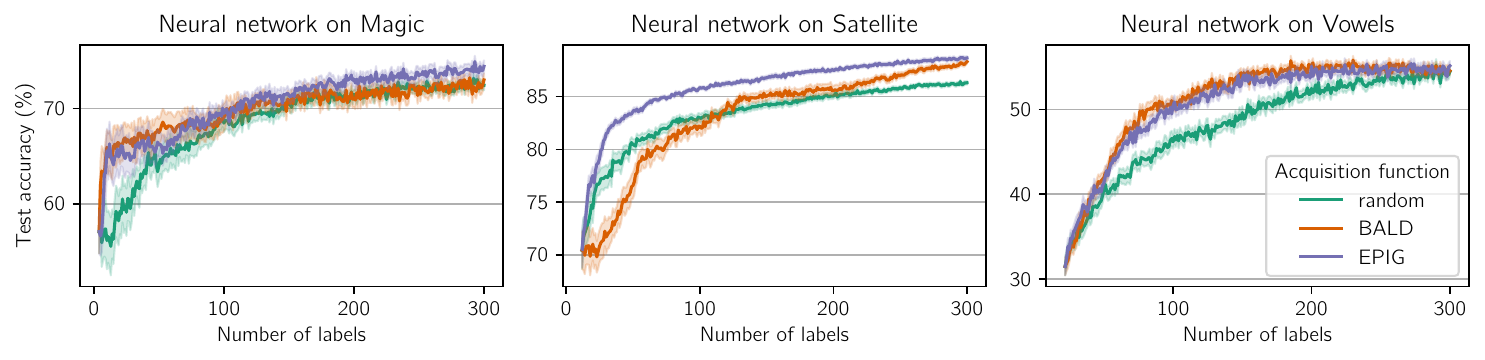}
\caption{
EPIG outperforms or matches BALD across three standard classification problems from the UCI machine-learning repository (Magic, Satellite and Vowels) and two models (random forest and neural network).
See \Cref{pobal:sec:uci} for details.
}
\label{pobal:fig:uci}
\end{figure*}

\subsection{UCI Data (\Cref{pobal:fig:uci})}
\label{pobal:sec:uci}

Next we compare BALD and EPIG in a broader range of settings.
We use problems drawn from a repository maintained at UC Irvine (UCI; \citealp{Dua:2019}), which has been widely used as a data source in past work on Bayesian methods \citep{gal2015dropout,lakshminarayanan2016simple,sun2018differentiable,zhang2018noisy}.
The problems we use vary in terms of the number of classes, the input dimension and any divergence between the pool and target data distributions.
We assume knowledge of $\peval{\xeval}$ when estimating EPIG but note that this assumption has little significance if $\ppool{\x}$ and $\peval{\xeval}$ match, which is true for two out of the three problems.

\paragraph{Data.}

We use three classification datasets from the UCI repository, each with a different number of classes, $C$, and input dimension, $D$: Magic ($C=2$, $D=11$), Satellite ($C=6$, $D=36$) and Vowels ($C=11$, $D=10$).
The inputs are telescope readings in Magic, satellite images in Satellite and speech recordings in Vowels.
Magic is interesting because it serves as a natural instance of a mismatch between pool and target distributions (see \Cref{pobal:sec:uci_appendix}).

\paragraph{Models and Training.}

We use two different models.
The first is a random forest \citep{breiman2001random}.
To emphasize that EPIG works with an off-the-shelf setup, we use the Scikit-learn \citep{pedregosa2011scikitlearn} implementation with its default parameters.
The second model is a dropout-enabled fully connected neural network with three hidden layers and a softmax output layer.
A dropout rate of 0.1 is used during both training and testing.
Training the neural network consists of running up to 50,000 steps of full-batch gradient descent using a learning rate of 0.1.
We use a loss function consisting of the negative log likelihood (NLL) of the training data combined with an $l_2$ regularizer (with coefficient 0.0001) on the model parameters.
To mitigate overfitting we use early stopping: we track the model's NLL on a small validation set (approximately 20\% of the size of the training-label budget) and stop training if this NLL does not decrease for more than 10,000 consecutive steps.
We then restore the model parameters to the configuration that achieved the lowest validation-set NLL.

\paragraph{Active Learning.}

We use largely the same setup as described in \Cref{pobal:sec:2d_data}.
Here the label budget is 300, and we run active learning 20 times with different seeds.
We use the same BALD and EPIG estimators as before, treating each tree in the random forest as a different $\theta$ value, and treating each stochastic forward pass through the neural network (we compute 100 of them) as corresponding to a different $\theta$ value.
To estimate EPIG we sample $\xeval$ from a set of inputs designed to be representative of $\peval{\xeval}$.

\paragraph{Discussion.}

\Cref{pobal:fig:uci} shows EPIG performing convincingly better than BALD in some cases while matching it in others.
These results provide broader validation of EPIG, complementing the results in \Cref{pobal:fig:2d_data}.

\begin{figure*}[t]
\centering
\includegraphics[height=\figureheight,trim={0 0 0 0}]{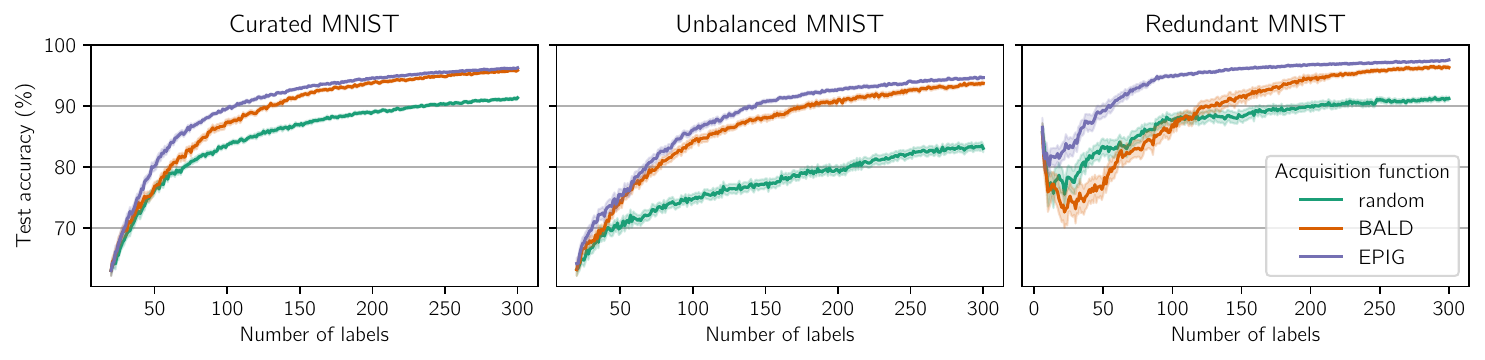}
\caption{
EPIG outperforms BALD across three image-classification settings.
Curated MNIST reflects the data often used in academic research.
The pool and target input distributions, $\ppool{\x}$ and $\peval{\xeval}$ match; the marginal class distributions, $\ppool{\y}$ and $\peval{\yeval}$, are uniform.
Unbalanced MNIST is a step closer to real-world data.
While $\peval{\yeval}$ remains uniform, $\ppool{\y}$ is non-uniform: the pool contains more inputs from some classes than others.
Redundant MNIST simulates a separate practical problem.
Whereas $\peval{\yeval}$ only has nonzero mass on two classes of interest, $\ppool{\y}$ has substantial mass across all classes.
See \Cref{pobal:sec:mnist} for details.
}
\label{pobal:fig:mnist_standard}
\end{figure*}

\begin{figure*}[h!]
\centering
\includegraphics[height=\figureheight,trim={0 0 0 0}]{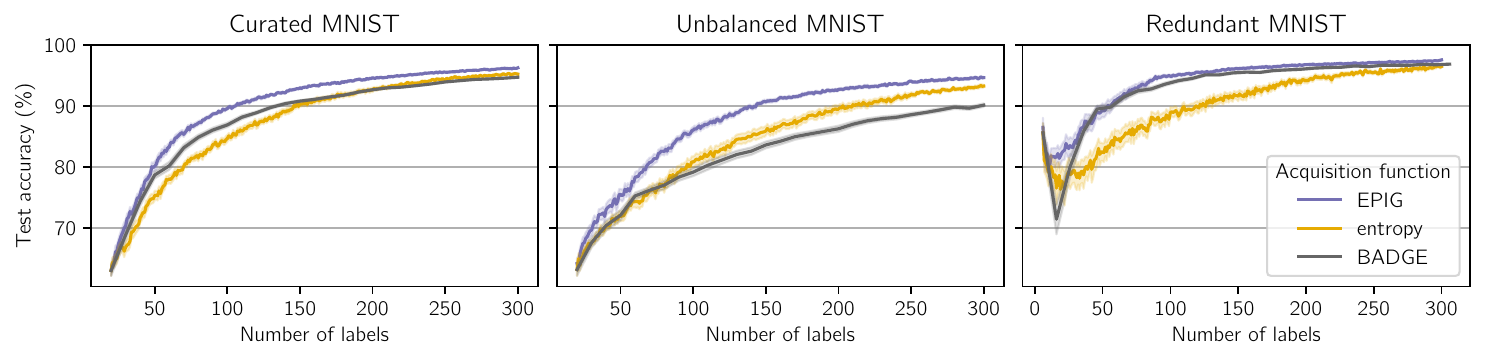}
\caption{
EPIG outperforms two acquisition functions popularly used as baselines in the active-learning literature.
The first is the model's predictive entropy, $\xHof{\pof{\y \given \y}}$ \citep{settles2008analysis}.
The second is BADGE \citep{ash2020deep}.
Calculating BADGE involves computing a gradient-based embedding for each candidate input in the pool and then applying $k$-means++ initialization \citep{arthur2006k} in embedding space to select a diverse batch of inputs for labelling.
We acquire 10 labels at a time with BADGE.
}
\label{pobal:fig:mnist_extra_baselines}
\end{figure*}

\subsection{MNIST Data (\Cref{pobal:fig:mnist_standard,pobal:fig:mnist_ablation,pobal:fig:mnist_extra_baselines})}
\label{pobal:sec:mnist}

Finally, we evaluate BALD and EPIG in settings intended to capture challenges that occur when applying deep neural networks to high-dimensional inputs.
Our starting point is the MNIST dataset \citep{lecun1998gradient}, in which each input is an image of a handwritten number between 0 and 9.
This dataset has been widely used in related work on Bayesian active learning with deep neural networks \citep{beluch2018power,gal2017deep,jeon2020thompsonbald,lee2019baldvae,tran2019bayesian}.
We construct three settings based on this dataset, each corresponding to a different practical scenario: Curated MNIST, Unbalanced MNIST and Redundant MNIST.

As well as investigating how BALD and EPIG perform across these settings, we seek to understand the effect on EPIG of varying the amount of knowledge we have of the target data distribution, $\peval{\xeval}$.
To this end we assume we know this for one set of runs (\Cref{pobal:fig:mnist_standard}) and then relax this assumption for another set (\Cref{pobal:fig:mnist_ablation}).

\paragraph{Data.}

Curated MNIST is intended to reflect the data often used in academic machine-learning research.
The pool and target class distributions, $\ppool{\y}$ and $\peval{\yeval}$, are both uniform over all 10 classes.
In terms of class distributions, this effectively represents a worst-case scenario for active learning relative to random acquisition.
Given matching class-conditional input distributions, namely $\ppool{\xeval \given \yeval}=\peval{\xeval \given \yeval}$, uniformly sampling from the pool input distribution, $\ppool{\x}$, is equivalent to uniformly sampling from the target input distribution, $\peval{\xeval}$.
Thus, random acquisition is a strong baseline in this setting.

Unbalanced MNIST is a step closer to real-world data.
We might expect $\peval{\yeval}$ to be uniform---that is, the task of interest might involve classifying examples in equal proportion from each class---but it could be difficult to curate a pool that is similarly uniform in its class distribution.
To reflect this we consider a case with a non-uniform $\ppool(\y)$: classes 0 to 4 each have probability $\sfrac{1}{55}$ and classes 5 to 9 each have probability $\sfrac{10}{55}$.

Redundant MNIST captures a separate practical problem that occurs, for instance, when using web-scraped data.
The pool might contain inputs from many more classes than we want to focus on in the predictive task of interest.
To simulate this we suppose that the task involves classifying just images of 1s and 7s, occurring in equal proportion---that is, $\peval{\yeval}$ places probability mass of $\sfrac{1}{2}$ on class 1, $\sfrac{1}{2}$ on class 7, and 0 on all other classes---while $\ppool{y}$ is uniform over all 10 classes.
If the acquisition function selects an input from a class other than 1 and 7, the labelling function produces a ``neither'' label.
Thus, we have three-way classification during training: 1 vs 7 vs neither.

\paragraph{Model and Training.}

For both runs we use the same dropout-enabled convolutional neural network we already used in \Cref{chapter:batchbald}.
The dropout rate here is 0.5.
Training is similar to as described in \Cref{pobal:sec:uci}, except that the learning rate is 0.01 and early stopping triggers after 5,000 consecutive steps of non-decreasing validation-set NLL.

\begin{figure*}[t]
\centering
\includegraphics[height=\figureheight,trim={0 0 0 0}]{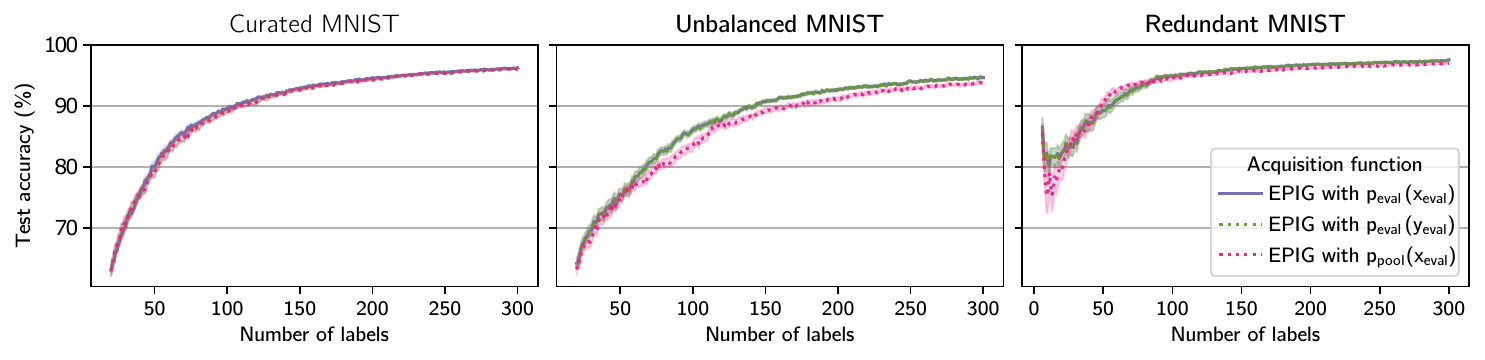}
\caption{
Even without knowledge of the target input distribution, $\peval{\xeval}$, EPIG retains its strong performance on Curated MNIST, Unbalanced MNIST and Redundant MNIST.
``EPIG with $\peval{\xeval}$'' assumes exact samples from $\peval{\xeval}$, as in \Cref{pobal:fig:mnist_standard}.
``EPIG with $\peval{\yeval}$'' corresponds to using the approximate-sampling scheme outlined in \Cref{pobal:sec:sampling_x_star}, using knowledge of $\peval{\yeval}$.
``EPIG with $\ppool{\xeval}$'' corresponds to using samples from the pool as a proxy for $\peval{\xeval}$.
See \Cref{pobal:sec:mnist} for details.
}
\label{pobal:fig:mnist_ablation}
\end{figure*}

\paragraph{Active Learning.}

Initially we retain the setup described in \Cref{pobal:sec:uci}, with $\peval{\xeval}$ known (\Cref{pobal:fig:mnist_standard}).
Then we investigate the sensitivity of EPIG to removing full access to $\peval{\xeval}$, focusing on two different settings (\Cref{pobal:fig:mnist_ablation}).
In one setting, we assume knowledge of $\peval{\yeval}$ and use the resampling technique discussed in \Cref{pobal:sec:sampling_x_star}.
In the other, we simply sample target inputs from the pool: $\xeval \sim \ppool(\xeval)$.

\paragraph{Discussion.}

\Cref{pobal:fig:mnist_standard} shows EPIG again outperforming BALD and random across all three dataset variants when given access to $\peval{\xeval}$.
(EPIG additionally beats predictive entropy \citep{settles2008analysis} and BADGE \citep{ash2020deep}, acquisition functions commonly studied in the active-learning literature, as shown in \Cref{pobal:sec:app:extra_results}.)
EPIG's advantage over BALD is appreciable on Curated MNIST and Unbalanced MNIST.
But it is emphatic on Redundant MNIST.
This suggests EPIG is particularly useful when working with highly diverse pools.

\Cref{pobal:fig:mnist_ablation} shows the even more impressive result that EPIG retains its strong performance even when no access to $\peval{\xeval}$ is assumed.
We thus see that EPIG provides a good degree of robustness in its performance to the level of knowledge about the target data distribution.

\section{Joint Expected Predictive Information Gain}
\label{EPIG:sec:epigbald}

\begin{figure}[t!]
    \begin{insightbox}{Expected Reduction of Epistemic Uncertainty}{pobal:insight:epistemic_uncertainty}
    By rewriting the joint expected predictive information gain (JEPIG) as a triple mutual information which takes into account the model parameters we can rephrase it as a difference of two (parameter) EIG terms (dropping $\Dtrain$ for clarity):
    \begin{align}
        & \MIof{\Yevalset; \Yacqset \given \xacqset, \xevalset}
        \notag \\
        & \quad = \MIof{\Yevalset; \Yacqset \given \xacqset, \xevalset} - \underbrace{\MIof{\Yevalset; \Yacqset \given \xacqset, \xevalset, \W}}_{=0}\\
        & \quad = \MIof{\Yevalset; \Yacqset ; \W \given \xacqset, \xevalset} \\
        & \quad  = \underbrace{\MIof{\Yacqset ; \W \given \xacqset}}_{\Circled{5}} - \underbrace{\MIof{\Yacqset ; \W \given \xacqset, \Yevalset,\xevalset}}_{\Circled{6}}.         \label{EPIG:eq:EREU}
        \end{align}
    Since BALD is known to measure epistemic uncertainty \citep{smith2018understanding}, we can take another look at these two EIGs:
    Intuitively, the first EIG \Circled{5} is large when the model has high epistemic uncertainty about its prediction at $\xacqset$, and learning the true label would thus be informative for the model parameters,
    while the second EIG \Circled{6} captures the epistemic uncertainty about the model's prediction at $\xacqset$ assuming we had obtained labels for evaluation samples.
    This second term is small when $\x$ is similar to drawn evaluation samples and the model can explain it well given the pseudo-labels. 
    Thus, JEPIG is large when the first term is large and the second term is small, so learning $\xacqset$ is informative for the model, and $\xacqset$ is similar to evaluation samples. 
    In reverse, JEPIG is small, when knowing about evaluation samples makes no difference for the epistemic uncertainty of the acquisition candidates, which means that they are unrelated.
\end{insightbox}
\end{figure}

In this section, we examine \emph{joint expected predictive information gain (JEPIG)},
\begin{align}
    \MIof{\Yevalset; \Yacq \given \xevalset, \xacq, \Dtrain},
\end{align} in more detail. 
As explained in \RRectangle{\textbf{\cref{pobal:insight:epistemic_uncertainty} Expected Reduction of Epistemic Uncertainty}}, JEPIG also has an intuitive explanation as expected reduction in epistemic uncertainty when taking into account evaluation samples.

First, we will show that we can take the limit of the number of evaluation set samples to model $\peval{\xeval}$. Then, we connect the criterion to Bayesian model selection and the marginal log likelihood. Finally, we contrast JEPIG with BALD and EPIG and present a practical reason why JEPIG might be easier to compute in the batch acquisition setting.
We report some early results on the performance of JEPIG in \Cref{EPIG:sec:experiments}.

\subsection{JEPIG for $\evalsize \to \infty$}
\label{EPIG:sec:limit}

If we do not have a finite number of evaluation samples but a distribution, we might want to take the limit to take into account the full distribution. For countable many $\xeval$, we can easily show that the limit exist. 
\begin{proposition}
    The following limit exists for countable $\peval{\xeval}$:
    \begin{align}
        \lim_{\substack{\evalsize \to \infty}} \MIof{\Yevalset; \Yacq \given \xevalset, \xacq}.
    \end{align}
\end{proposition}
\begin{proof}
    Fix $\evalsize > 1$, and fix the order of $\xevalset$. JEPIG for fixed $\evalsize$ is upper-bounded by the predictive entropy of the acquisition batch:
    \begin{align}
        \MIof{\Yevalset; \Yacq \given \xevalset, \xacq} \le \Hof{\Yacq \given \xacq},
    \end{align}
    and it is non-decreasing:
    \begin{align}
        &\MIof{\Yevalset; \Yacq \given \xevalset, \xacq} \notag \\
        &\quad = \MIof{\Y^\typeeval_{1..\evalsize-1}; \Yacq \given \x^\typeeval_{1..\evalsize-1}, \xacq}
        + \underbrace{\MIof{\Y^\typeeval_\evalsize; \Yacq \given \x^\typeeval_\evalsize, \Y^\typeeval_{1..\evalsize-1}, \x^\typeeval_{1..\evalsize-1}, \xacq}}_{\ge 0}. \label{eq:jepig_limit_split}
    \end{align}
    Hence, the limit exists.
    However, it is not clear if the order of the evaluation samples matters.

    To show that the limits are equal for arbitrary orderings of the evaluation, let's first fix a natural ordering $(1,2,\ldots)$ and define:
    \begin{align}
        I(n) \defeq \MIof{\Y^\typeeval_{1..n}; \Yacq \given \x^\typeeval_{1..n}, \xacq}.
    \end{align} 
    Let $(a(1), a(2), \ldots)$ denote another arbitrary ordering, and let $I^a(n)$ denote the respective truncated JEPIG term:
    \begin{align}
        I^a(n) \defeq \MIof{\Y^\typeeval_{a(1)..a(n)}; \Yacq \given \x^\typeeval_{a(1)..a(n)}, \xacq}.
    \end{align}
    Finally, let's denote the limits for $n \to \infty$ as $I$ and $I^a$, respectively, which exist according to the argument above.
    We can now show that $I = I^a$ for all orderings $a$.

    Fix $\epsilon \ge 0$. Then, $\exists n \, \forall k \ge n: |I^a(k) - I^a(k)| \le \epsilon$ because of the limit.
    For any $n$, we choose an $M$ such that $\{1, \ldots, M\} \supseteq \{a(1), \ldots, a(n)\}$---we can simply take $M \defeq \max(a(1), \ldots, a(n))$.
    Then, following the same argument as in \cref{eq:jepig_limit_split}:
    \begin{align}
        I(K) \ge I^a(n) \quad \forall K \ge M.
    \end{align}
    Further, $\forall K \ge M$, we can choose an $N(K)$, such that $\{a(1), \ldots, a(N)\} \supseteq \{1, \ldots, K\}$---as $a$ is an ordering, it is bijective, and we can set $N(K) \defeq \max(a^{-1}(1), \ldots, a^{-1}(K))$---and we obtain:
    \begin{align}
        I^a \ge I^a(N(K)) \ge I(K).
    \end{align}
    Hence, we can sandwich $I(K)$ between $I^a(n)$ and $I^a$:
    \begin{align}
        I^a \ge I(K) \ge I^a(n).
    \end{align}
    This concludes the proof because for all $K \ge M$:
    \begin{align}
        | I^a - I(K) | \le | I^a - I^a(n) | \le \epsilon,
    \end{align}
    and we obtain the limit $I(K) \to I^a$.
    As already $I(K) \to I$, we have $I^a = I$.
\end{proof}
We conjecture that the limit also exists for continuous $\peval{\xeval}$.

\subsection{JEPIG as Bayesian Model Selection}
\label{EPIG:sec:jepig_bayesian_model_selection}

JEPIG can also be viewed as a reduction in expected marginal log likelihood for the evaluation set: how much does acquiring a label for an acquisition sample improve the model's ability to adapt to the evaluation set? We can view this as Bayesian model selection:

In \emph{Bayesian model selection}, for a set of models ${M_i}$ and a random variable $\Model$ that represents the chosen model, we are interested in inferring $\pof{\Model \given \Dtrain}$. This quantifies the model that is best for our purposes.
Conventionally, the marginal likelihood $\pof{\Dtrain \given \Model}$ is maximized, which assumes a uniform prior distribution over $\Model$ \citep{mackay2003information}.
The marginal likelihood does not actually measure the performance of the fully trained model but its generalization `speed' \citep{lyle2020bayesian}.
To measure the performance of a model, we can use cross-validation, which can be cast into a Bayesian setting  \citep{fong2020marginal}. 
This was also explored in \citep{lotfi2022bayesian} (but see also \citet{kirsch2022clml} for a critical review).
We explore aspects of this further in \Cref{chapter:obi}.

When $\Deval \subseteq \Dpool$, maximizing JEPIG performs Bayesian model selection towards selecting the best acquisition batch in expectation: we precisely perform Bayesian model selection in the above sense by trying to find the acquisition batch that will help the model best adapt to future data points (assuming we aim to acquire all pool points eventually.)
Obviously, this is not true when the evaluation set is disjoint from the pool. 

\subsection{JEPIG \&\ EPIG vs.\ BALD}
\label{EPIG:sec:jepig_bald}

From \Cref{pobal:glue:mackayterms}, we already know that BALD upper-bounds both JEPIG and EPIG. Let us examine this in more detail.
From \cref{EPIG:eq:EREU} from the intuition box for JEPIG, we also see that we have
\begin{align}
& \MIof{\Yevalset; \Yacqset ; \W \given \xacqset, \xevalset, \Dtrain} \notag \\
& \quad \le \MIof{\Yacqset ; \W \given \xacqset, \Dtrain},
\end{align}
that is, JEPIG is upper-bounded by BALD---a two-term mutual information is always non-negative---and JEPIG is equivalent to BALD exactly when \(\MIof{\Yacqset ; \W \given \xacqset, \Yevalset,\xevalset, \Dtrain}\) is zero. 
This is the case when either there is no uncertainty about the parameters left, or when the distribution of \(\Yacqset \given \xacqset\) is fully determined by conditioning the posterior on \(\yevalset \given \xevalset \sim \pof{\yevalset \given \xevalset, \Dtrain}\).
When \citet{mackay1992information} analyzed JEPIG, they stated that it was equivalent to BALD because in the specific context of that work, using Bayesian linear regression with simple homoscedastic noise, it was.

BALD and JEPIG will trivially be equal when $\xacqset \subseteq \xevalset$. In reverse, for non-parametric models, this will trivially not be the case when $\xacqset$ is not `near' the evaluation set. Intuitively, JEPIG is different from BALD exactly when BALD fails: for outlier pool samples which are not similar to the test--time distribution, in which case JEPIG will tend towards 0 as the two terms cancel out.

Finally, a qualitative result:
\begin{restatable}{proposition}{relationjpigepig}
    \label{EPIG:prop:relationjpigepig}
    EPIG lower-bounds `averaged' JEPIG:
    \begin{align}
        \MIof{\Yeval; \Yacq \given \Xeval, \xacq, \Dtrain} \le \nicefrac{1}{\evalsize} \MIof{\Yevalset; \Yacq \given \xevalset, \xacq} + c_\typeeval,
    \end{align} 
    up to an additive constant ($c_\typeeval$) that only depends on the evaluation samples and is independent of $\xacq$. The inequality gap is the total correlation:
    \begin{align}
        \nicefrac{1}{\evalsize} \TCof{\Yeval_1;\ldots;\Yeval_\evalsize \given \xevalset, \Yacq, \xacq}
    \end{align}
    We have equality when it is zero, that is when the predictions on the evaluation set are independent (given the acquisition samples).
\end{restatable}
See \Cref{EPIG:appsec:epig_and_jepig} for the formal proof. This result might seem unintuitive given that we have the opposite for BatchBALD in \Cref{chapter:batchbald}. However, it is important to notice that the redundancy that JEPIG removes is in the evaluation set. In the case of BatchBALD, the redundancy is in the acquisition set. Hence, the different direction of the inequality sign.

{

\textbf{Batch Acquisition using EPIG.}
A practical reason for examining JEPIG is batch active learning with EPIG.
While the expectation in EPIG can be evaluated for individual acquisition, the batch setting is computationally more complex. Following \S\ref{chapter:batchbald}, for batch acquisition, we need to maximize
\begin{math}
    \MIof{\Yeval ; \Yacqset \given \Xeval, \xacqset, \Dtrain}.
\end{math}
Unlike the expected information gain, this term is not submodular. However, as the global subset problem is not feasible, we will examine the case of using greedy selection nonetheless.
Computing \(\MIof{\Yeval ; \Yacqset \given \Xeval, \xacqset, \Dtrain}\) requires estimating a joint density in \(i+1\) many variables for each $\xeval$ sample for the \(i\)-th element in the acquisition batch:
\begin{math}
    \pof{\yeval, \yacqset \given \xeval, \xacqset}.
\end{math}
Overall, for an acquisition batch of size \(B\), this requires $O(\lvert \Deval \rvert \, \lvert \Dpool \rvert^B)$ many joint densities with 2 to $B+1$ many variables compared to $O(\lvert \Dpool \rvert^B)$ for BatchBALD with 1 to $B$ many variables.
Unfortunately, BatchBALD has already been found to be computationally intractable for larger acquisition batch sizes in practice as noted in \S\ref{chapter:batchbald} and \ref{chapter:stochacq}, and here we consider an additional variable by default, while we also have to evaluate this term for many $\xeval$. This does not spark joy.

\textbf{Evaluation of JEPIG.}
We could maximize $\MIof{\Yevalset; \Yacq \given \xevalset, \xacq, \Dtrain}$ via
\begin{align}
    &\MIof{\Yevalset; \Yacq \given \xevalset, \xacq, \Dtrain} = \notag \\
    & \quad =
    \Hof{\Yacq \given \xacq, \Dtrain} \notag \\
    & \quad \quad - \Hof{\Yacq \given \xacq, \Yevalset, \xevalset, \Dtrain}.
\end{align}
Note that the second term is expensive to evaluate: 
\begin{align}
    & \Hof{\Yacqset \given \xacqset, \Yevalset, \xevalset, \Dtrain} = \notag \\
    & \quad = \simpleE{\pof{\yevalset \given \xevalset, \Dtrain}} \Hof{\Yacqset \given \xacqset, \yevalset, \xevalset, \Dtrain}. \label{EPIG:eq:pig_pseudolabel_conditional}
\end{align}
That is, we need to train a model on $\Dtrain$ then sample joint label predictions $\yevalset$ for the evaluation set $\xevalset$ using the model, and then, for each such sampled joint prediction, we evaluate the conditional entropy $\Hof{\Yacqset \given \xacqset, \W'}$ using new $\w' \sim \pof{\w \given \yevalset, \xevalset, \Dtrain}$, which requires additional training using the labeled training set augmented with the sampled labels \(\Dtrain \cup \{\yeval_i, \xeval_i\}_i\).

In appendix \S\ref{EPIG:sec:practical_approximation}, we present a computationally tractable approximation for this term using `\emph{self-distillation}`, where we train a new model on predictions for evaluation samples from the current model trained on $\Dtrain$. On a high-level, this yields a model that fulfills the intuitions presented in \RRectangle{\textbf{\cref{pobal:insight:epistemic_uncertainty} Expected Reduction of Epistemic Uncertainty}}. We refer to the appendix for more details.
}
\FloatBarrier

\subsection{Empirical Validation}
\label{EPIG:sec:experiments}
\begin{figure}[t]
    \begin{minipage}[t]{0.48\linewidth}
        \centering
        \includegraphics[width=\linewidth]{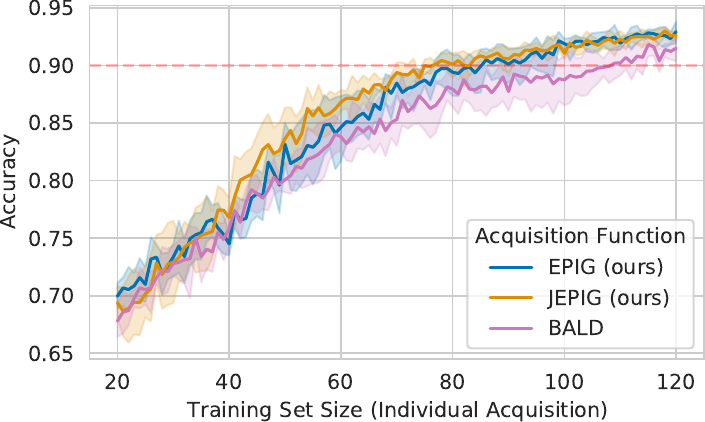}
        \caption{\emph{EPIG vs JEPIG vs BALD with Bayesian Neural Networks on MNIST.}
        JEPIG performs better under than MC Dropout than EPIG.
        }
        \label{EPIG:fig:ablation_epig_vs_epigbald}   
    \end{minipage}\hfill
    \begin{minipage}[t]{0.48\linewidth}
        \centering
        \includegraphics[width=\linewidth]{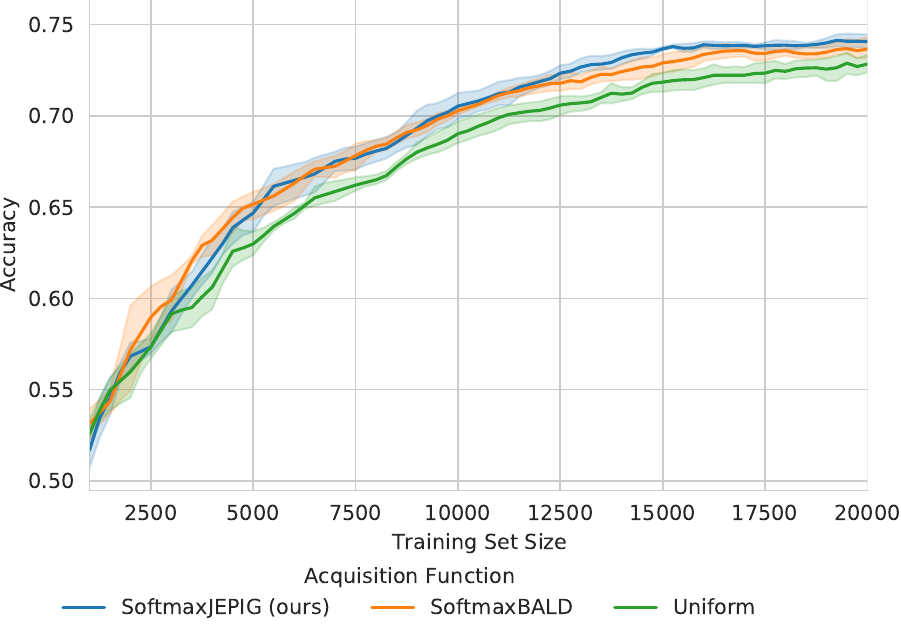}
        \caption{\emph{BALD vs JEPIG on CIFAR-10.} JEPIG outperforms BALD. 5 trials each. With batch acquisition size 250, and initial training size 1000. Median accuracy after smoothing with a Parzen window filter over 30 acquisition steps to denoise.}
        \label{EPIG:fig:normal_active_learning_cifar10}
    \end{minipage}
\end{figure}

\begin{figure}[t]
    \centering
    \includegraphics[width=\linewidth]{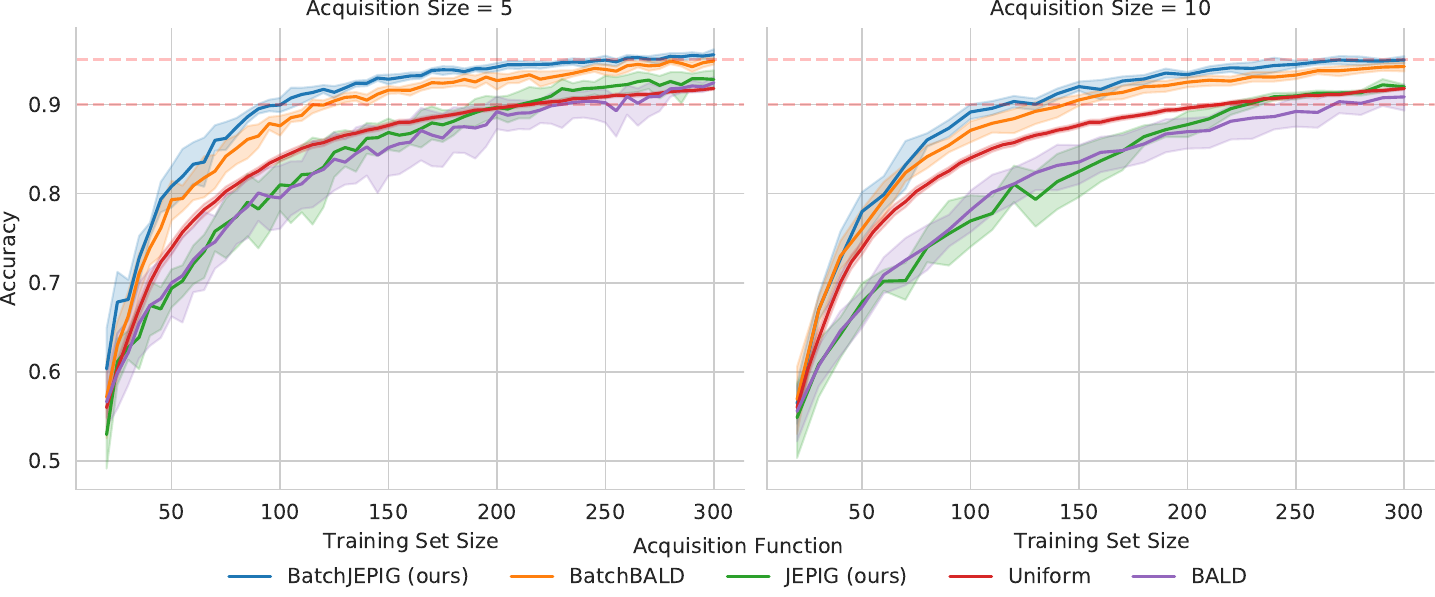}
    \caption{\emph{(Batch)BALD vs (Batch)JEPIG on Repeated-MNIST (MNISTx2).} JEPIG outperforms BALD. %
    }
    \label{EPIG:fig:normal_active_learning}
\end{figure}

\begin{figure*}[th]
    \centering
    \begin{subfigure}{0.48\linewidth}
        \includegraphics[width=\linewidth]{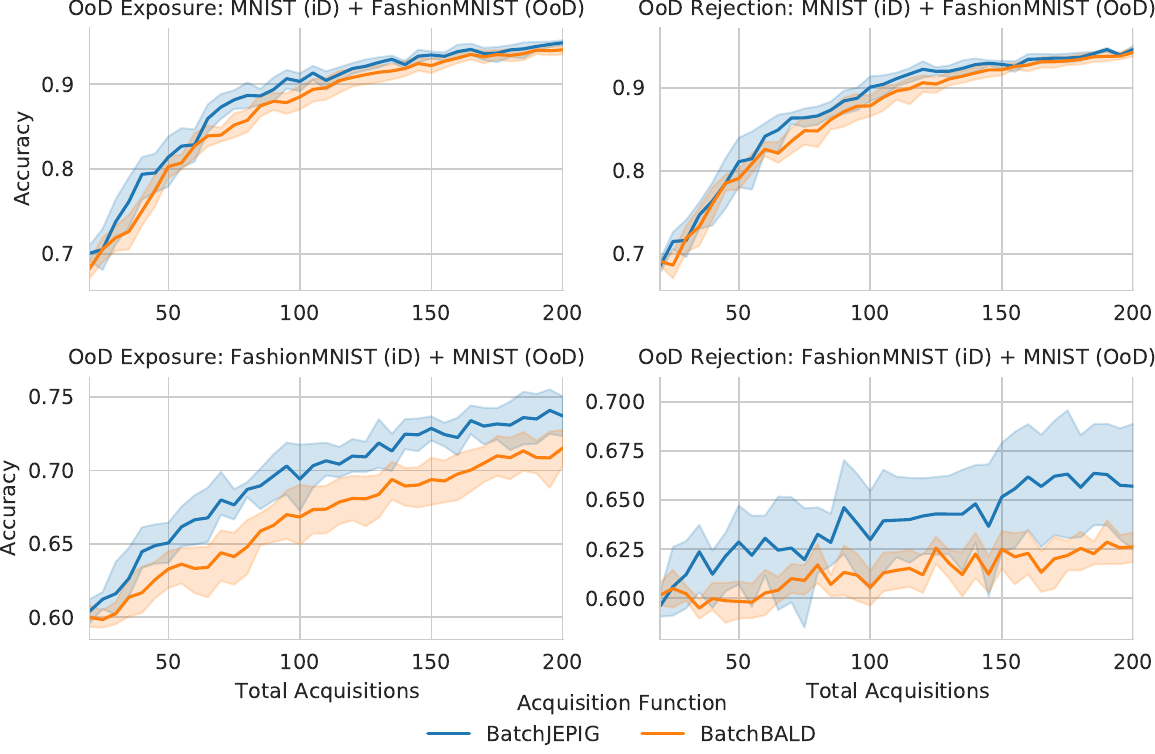}
        \subcaption{Accuracy}
        \label{EPIG:subfig:ood_mnist_and_fmnist accuracy}
    \end{subfigure}
    \begin{subfigure}{0.48\linewidth}
        \includegraphics[width=\linewidth]{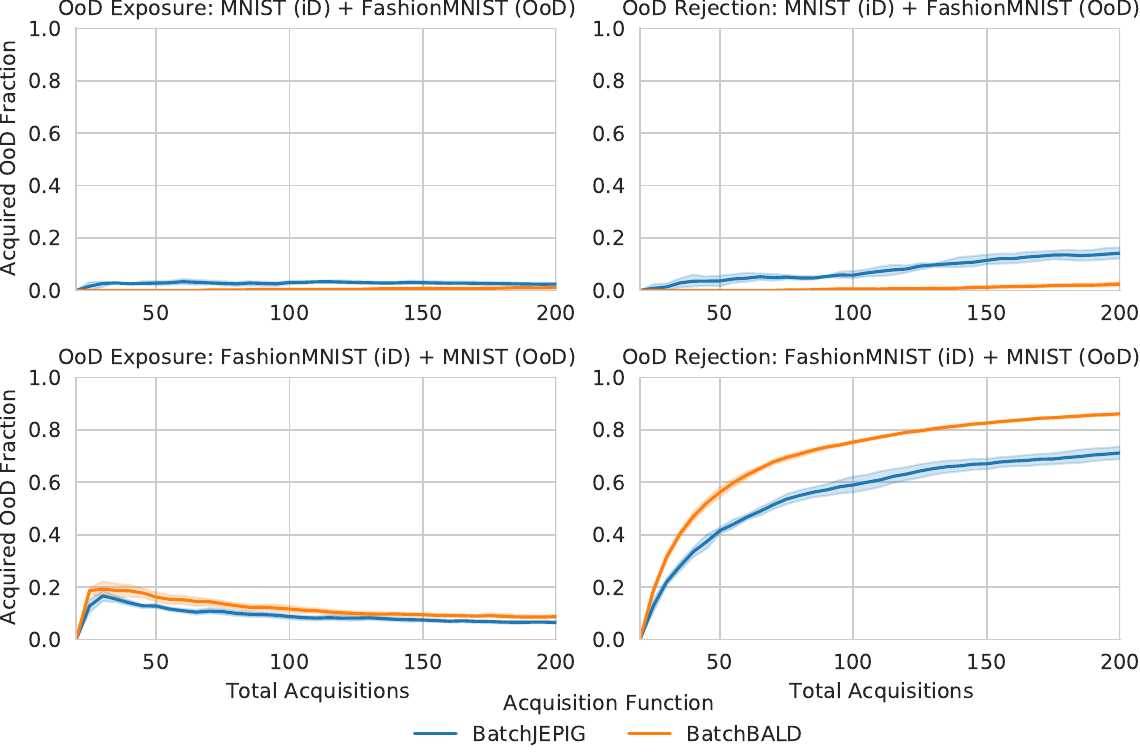}
        \subcaption{Acquired OoD Ratio}
        \label{EPIG:subfig:ood_mnist_and_fmnist ood fraction}
    \end{subfigure}
    \caption{
        \emph{MNIST and FashionMNIST pairings with OoD rejection or exposure.} JEPIG performs better than BALD. 5 trials.
        }
    \label{EPIG:fig:ood_mnist_and_fmnist}
\end{figure*}

\begin{figure*}[th]
    \centering
    \begin{subfigure}{0.49\linewidth}
        \includegraphics[width=\linewidth]{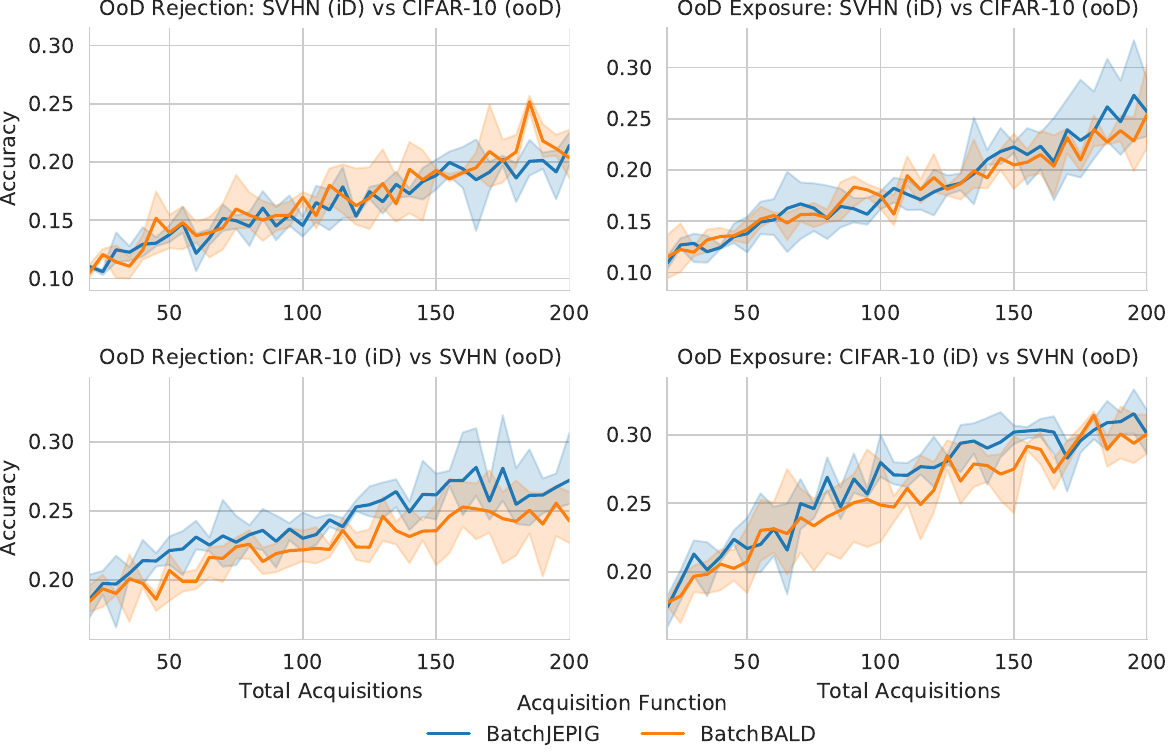}
        \subcaption{Accuracy}
    \end{subfigure}
    \begin{subfigure}{0.49\linewidth}
        \includegraphics[width=\linewidth]{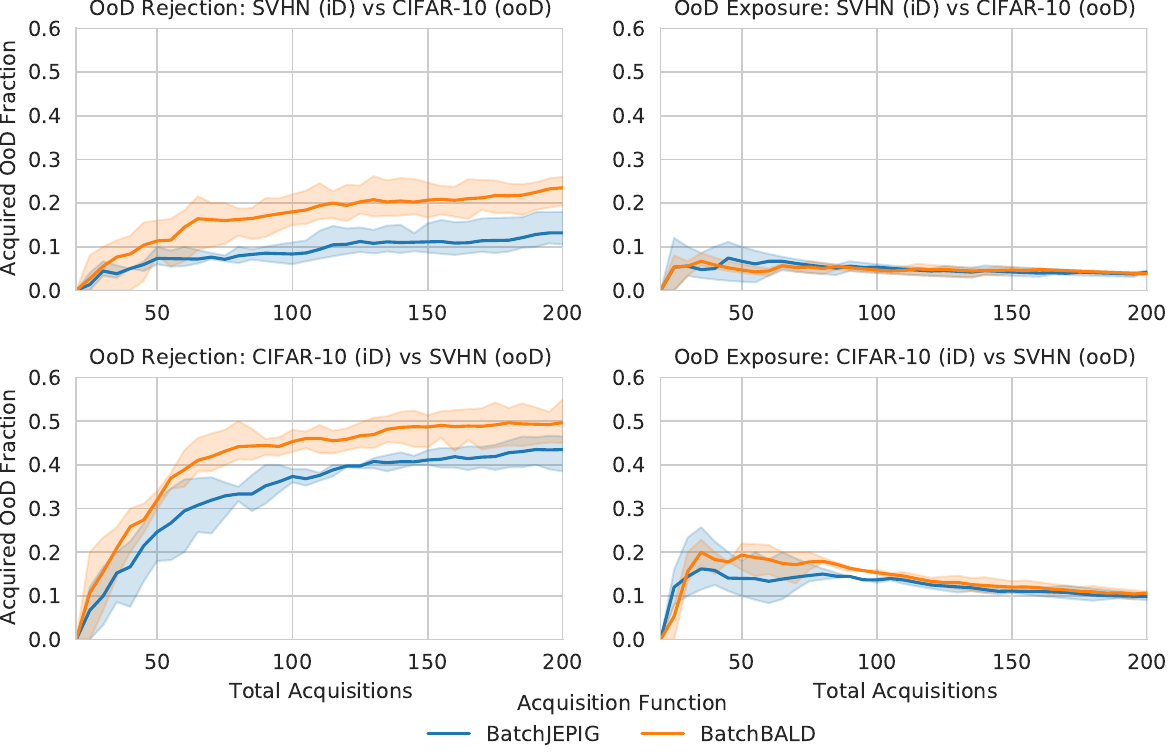}
        \subcaption{Acquired OoD Ratio}
    \end{subfigure}
    \caption{
        \emph{CIFAR-10 and SVHN pairings with OoD rejection or exposure.} JEPIG performs better than BALD. 5 trials. Acquisition size 5. Initial training size 5.
        }
    \label{EPIG:fig:ood_cifar10_and_svhn}
\end{figure*}

\begin{figure}[th]
    \centering
    \includegraphics[width=\linewidth]{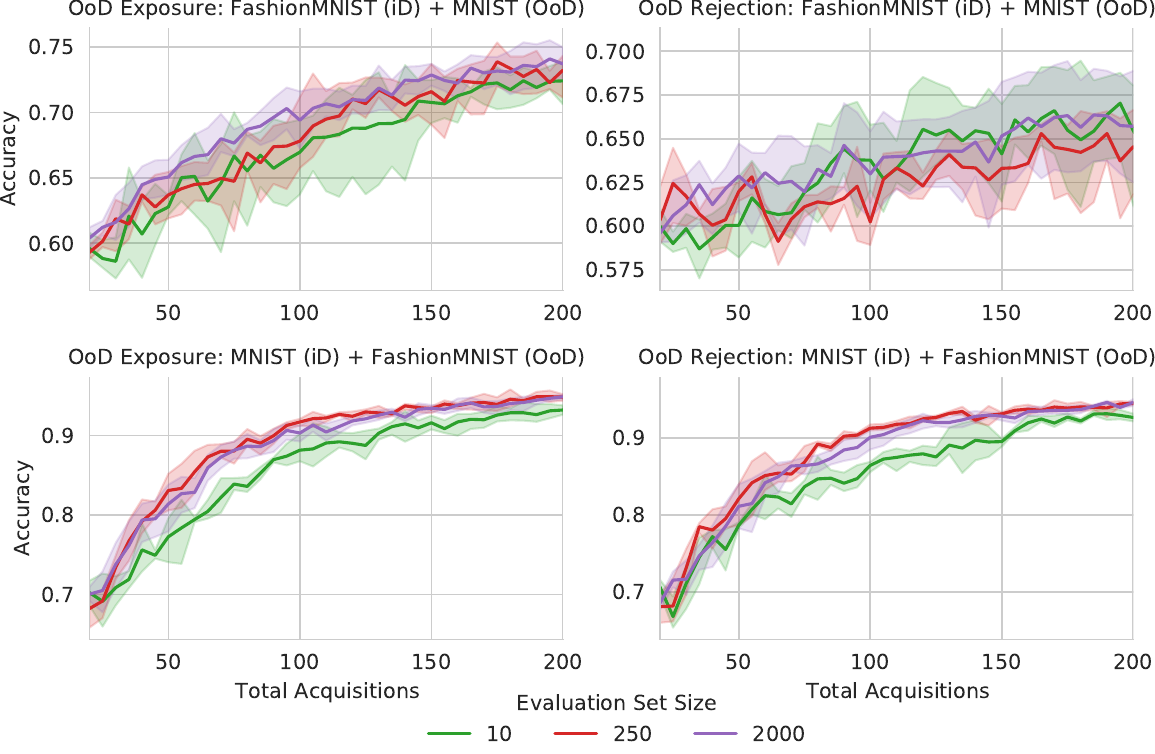}
    \caption{\emph{Evaluation Set Size Ablation}. MNIST and FashionMNIST pairings with OoD rejection or exposure. A larger evaluation set performs better. 5 trials.}
    \label{EPIG:subfig:ood_mnist_and_fmnist eval_set_size}
\end{figure}

We evaluate the performance of JEPIG using a form of self-distillation described in appendix \S\ref{EPIG:sec:practical_approximation} in regular active learning and under distribution shift.
Moreover, we provide an ablation with different evaluation set sizes. We use approximate BNNs based on MC dropout, see \Cref{glue:related_work:BNNs}.

\textbf{Setup.} We compare EPIG and JEPIG with different acquisition sizes to BALD, either using the top-k  of individual scores \citep{gal2015dropout}, the batch version BatchBALD (\S\ref{chapter:batchbald})---which is equivalent to the previous for individual acquisition---or SoftmaxBALD (\Cref{chapter:stochacq}) for larger acquisition batch sizes. SoftmaxBALD samples without replacement from the pool set using the Softmax of the acquisition scores with temperature 8.

On MNIST and MNISTx2 (Repeated-MNIST), we use a LeNet-5 model \citep{lecun1998gradient}, which we train as described in \S\ref{chapter:batchbald}. 

For CIFAR-10 \citep{krizhevsky2009learning}, we use a ResNet18 model \citep{he2016deep} which was modified as described in \S\ref{chapter:batchbald} to add MC dropout to the classifier head and also follows the described training regime. We train with an acquisition batch size of 250 and an initial training set size of 1000.
We use MC dropout models with 100 dropout samples when computing the acquisition scores.

\textbf{Performance on Regular Active Learning.}
We evaluate whether ignoring the test--time input distribution has a detrimental effect on BALD even when the pool set distribution matches the test distribution.

For this, we compare BALD and JEPIG in \cref{EPIG:fig:normal_active_learning} on MNISTx2 and EPIG and JEPIG in \cref{EPIG:fig:ablation_epig_vs_epigbald} on MNIST. In the regular active learning setup, there is no distribution shift between the pool set and test set, so we use the whole unlabeled pool set as evaluation set. 

Both in the top-k and the batch variant, EPIG and JEPIG outperform BALD on MNISTx2 (and also MNIST, not shown). 
On CIFAR-10, JEPIG also outperforms SoftmaxBALD, as depicted in \cref{EPIG:fig:normal_active_learning_cifar10}.

However, why does JEPIG outperform EPIG? We hypothesize that this is because EPIG takes an expectation over the evaluation set using individual points which by itself might be myopic. It might work well for simple models, but taking the whole evaluation set into account and retraining the model might allow for learning better abstraction in deep models, which might not be the case for EPIG.

\textbf{Performance on Active Learning under Distribution Shift with MNIST and FashionMNIST.}
We want to evaluate how BALD and JEPIG behave under distribution shift, that is when pool set and test distribution do not match.
For this, we add junk out-of-distribution data to the pool set. In this experiment, the pool set contains MNIST and FashionMNIST \citep{xiao2017fashion} while the test set contains one or the other.
We deal with an acquisition function attempting to acquire OoD data in two different modes: \emph{OoD rejection} rejects OoD data from the batch and does not acquire it; while \emph{OoD exposure} acquires OoD data with uniform targets, similar to outlier exposure methods in OoD detection \citep{hendrycks2018deep}. We use an evaluation set with 2000 unlabeled samples.

JEPIG outperforms BALD on in all combinations, see \cref{EPIG:subfig:ood_mnist_and_fmnist accuracy}. In all cases but one, JEPIG acquires fewer junk/OoD samples, see \cref{EPIG:subfig:ood_mnist_and_fmnist ood fraction}. The ablation in \cref{EPIG:subfig:ood_mnist_and_fmnist eval_set_size} shows that larger evaluation sets are beneficial. Note that the evaluation set is unlabeled and thus does not count towards sample acquisitions.

For CIFAR-10 and SVHN \citep{svhn2011}, JEPIG outperforms BALD under distribution shift in all but one combination and selects fewer OoD samples, see \cref{EPIG:fig:ood_cifar10_and_svhn}. We use an evaluation set with 1000 unlabeled samples.

\FloatBarrier

\section{Related Work}

The idea of using the EIG to quantify the utility of data was introduced by \citet{lindley1956measure} and has a long history of use in experimental design \citep{chaloner1995bayesian,rainforth2023modern}.
The framework of Bayesian experimental design has many applications outside active learning, and in these applications the model parameters are commonly the quantity of interest---Bayesian optimization \citep{Hennig_2012,Hernandez-Lobato_2014,villemonteix2009informational} being a notable exception.
The EIG in the parameters is thus often a natural acquisition function.

The EIG in the parameters was originally suggested as an acquisition function for active learning by \citet{mackay1992evidence,mackay1992information}, who called it the total information gain.
Despite its shortfalls, some of which were also briefly discussed by \citet{freund1997selective} and \citet{mackay1992information} for special cases such as Bayesian linear regression, it has been widely used in cases where the model's predictions, not the model parameters, are the ultimate objects of interest \citep{atighehchian2020bayesian,beluch2018power,gal2017deep,houlsby2011bayesian,jeon2020thompsonbald,lee2019baldvae,munjal2022towards,pinsler2019bayesian,shen2017deep,siddhant2018deep,tran2019bayesian}.

Maximizing the information gathered about a quantity other than the model parameters has been proposed a number of times as an approach to active learning.
The predictive information as mutual information between the past and future was introduced by \citet{bialek1999predictive} and has been used to increase sample efficiency in reinforcement learning \citep{lee2020predictive}.
Perhaps most relevant to this chapter, \citet{mackay1992evidence,mackay1992information} also discussed two other acquisition functions, namely the mean marginal information gain and the joint information gain, which correspond to EPIG and JEPIG, respectively.

The mean marginal information gain measures the average information gain in the predictions made on a fixed set of inputs based on a Gaussian approximation of the posterior over the model parameters.
Though the mean marginal information gain has since received surprisingly little attention in the literature, it was discussed by \citet{huszar2013scoring} and later used by \citet{wang2021beyond} to evaluate the quality of predictive-posterior correlations.
\citet{wang2021beyond} extended the mean marginal information gain to the batch setting following insights from BatchBALD for regression tasks and evaluated it in a transductive active learning setting. \citet{wang2021beyond} only examine the case where pool and target input distribution are identical.
The transductive approach to active learning \citep{vapnik1982estimation,yu2006active}
seeks to maximize the performance on a fixed set of inputs---in contrast with the input distribution considered by EPIG. 

The joint information gain has received even less attention than the mean marginal information gain in the literature. This might be because MacKay showed that it is equivalent to BALD when assuming constant aleatoric noise (with a sufficiently large number of evaluation samples).
Constant aleatoric noise is also a common and convenient choice for Gaussian Processes, which could explain why it was not considered by works in Bayesian optimization either.
However, for deep neural networks, constant aleatoric noise is not a common assumption, and indeed the main benefit of using BALD over entropy is that it performs well when aleatoric uncertainty varies across samples because it estimates epistemic uncertainty and not aleatoric uncertainty (see also \Cref{chapter:ddu}). Moreover, our detailed re-examination of the relationship between BALD and JEPIG reaches a more varied conclusion than \citet{mackay1992information} by specifically taking into account the support of pool and evaluation samples.

Aside from the work of \citet{mackay1992evidence,mackay1992information}, there are numerous prediction-oriented methods \citep{afrabandpey2019human,chapelle2005active,cohn1993neural,cohn1996active,daee2017knowledge,donmez2008optimizing,evans2015estimating,filstroff2021targeted,krause2008near,neiswanger2021bayesian,seo2000gaussian,sundin2018improving,sundin2019active,tan2021diversity,yu2006active,zhao2021bayesian,zhao2021efficient,zhao2021uncertainty,zhu2003combining}.
Many of these, with notable examples including the work of \citet{cohn1996active}, \citet{krause2008near} and \citet{neiswanger2021bayesian}, are tied to a particular model class or approximation scheme and so lack EPIG's generality.

There is an additional limitation associated with techniques based on the idea, due to \citet{roy2001toward}, of measuring the expected loss reduction that would result from updating the model on a given input-label pair.
These techniques often require updating the model within the computation of the acquisition function, which can be extremely expensive.
Despite a strong conceptual connection to the acquisition function proposed by \citet{roy2001toward}, EPIG allows a significantly lower computational cost: its information-theoretic formulation allows us to derive an estimator that does not require nested model updating.

Concurrently to \citet{kirsch2021powerevaluationbald} (which this chapter derives from), \citet{neiswanger2021bayesian} introduce Bayesian Algorithm Execution (BAX), which goes beyond traditional Bayesian optimization by steering the acquisition for estimating computable properties of black-box functions.
This is achieved without the need to modify the underlying algorithm. 
Instead, BAX uses an information-theoretic approach to guide the selection of queries, based on the mutual information between the potential queries and the computable properties.
This is approximated with samples from execution paths of the algorithm computing the property of interest. 
BAX expands the applicability of Bayesian experimental design principles to a wider array of practical problems. 
Within this context, EPIG and JEPIG can be seen as a special case of BAX, where the computable property of interest is the model's predictive performance on a set or distribution of inputs. However, the methods in this chapter specifically make us of the Bayesian model's predictive joint distribution to reduce the computational cost.

Unlike much active learning literature whose experiment setting implicitly assumes that that pool and test samples are drawn from the same distribution, EPIG and JEPIG support the setting in which pool and test distribution do \emph{not} match. This is not the case for other diversity-based active learning methods such as BADGE \citep{ash2020deep}, for example, which implicitly use the empirical pool set distribution to select diverse samples via clustering.

\section{Discussion}

We have demonstrated that BALD, a widely used acquisition function for Bayesian active learning, can be suboptimal.
While much of machine learning focuses on prediction, BALD targets information gain in a model's parameters in isolation and so can seek labels that have limited relevance to the predictions of interest.

Motivated by this, we have proposed EPIG, an acquisition function that targets information gain in terms of predictions.
Our results show EPIG outperforming BALD across a number of data settings (low- and high-dimensional inputs, varying degrees of divergence between the pool and target data distributions, and varying degrees of knowledge of the target distribution) and across multiple different models.
This suggests EPIG can serve as a compelling drop-in replacement for BALD, with particular scope for performance gains when using large, diverse pools of unlabeled data.

\label{EPIG:sec:limitations}

While EPIG can be evaluated efficiently in the individual acquisition case, it becomes very costly for batch acquisition (when using the same estimators for classification).
Our initial investigations of JEPIG with self-distillation shows it to be an approximation that is also well motivated and potentially faster to evaluate in the batch setting.
However, it requires training an additional model. While this could be sped up by using warm-starting \citep{ash2019warmstarting}, it is still a significant cost.
At the same time, two (Batch)BALD terms need to be computed for JEPIG which can also be slow and which does not scale well beyond small acquisition batch sizes.
Thus, EPIG and JEPIG represent a trade-off between the time it takes to compute the expectation over evaluation samples and training another model. In the case of individual acquisition, EPIG is strongly favored.
In \S\ref{chapter:stochacq}, we discuss a simple stochastic extension to avoid computing BatchBALD terms. We do not examine this approach in detail for the acquisition functions presented in this chapter.

Importantly, unlike BALD, EPIG is not submodular, and thus greedy acquisition is not guaranteed to obtain $1-\tfrac{1}{e}$ optimality for batch acquisition, even though we have not experienced any degradation in comparison with BALD/BatchBALD empirically in our initial experiments. Indeed, JEPIG performs on par or better than BALD in the regular active learning case without distribution shift. 
At the same time, while neither BALD nor JEPIG are adaptive submodular, and no statements about global optimality have been presented so far \citep{golovin2017adaptive, foster2022a}, this has not been an issue in practice.
\citet{ash2021gone} recently introduced a new forward-backward strategy for a weight-space version of JEPIG. We leave an evaluation using the estimators we have used to future work.

Finally, we present two scenarios where EPIG might not perform better than BALD:
Firstly, if a task's performance is dominated by a more general ``sub-task'', there might be many samples highly informative for this general sub-task. For image data, this could be the case when feature learning is of particular importance, and convolution kernels can be learned from image data, no matter the label or actual task. In this regime, both BALD and EPIG will perform similarly, yet will likely outperform random acquisition.
Secondly, if the model's architecture and its inductive biases have specifically been evolved for a task (and dataset), there will be a high overlap between the model parameters and the task's target input distribution as model architectures which converge faster are preferred as research outputs. BALD and EPIG might perform similarly in this case, too.
That is, prediction-oriented acquisition functions might show their strength when the task performance is highly specific and dependent on specific samples in the pool set, and the model's architecture is over-parameterized and not yet adapted to the task.
We leave investigation of this for the future.

}

{
	\begin{savequote}[75mm]
		I am always in front of you but can never be seen. What am I?
	\end{savequote}
	\chapter{\rholosstitle}
	\label{chapter:rholoss}

	\graphicspath{{./epig/figures/}}

\newcommand{\ptest}[1]{\pcof{\mathrm{test}}{#1}}
\newcommand{\petest}[1]{\hpcof{\mathrm{test}}{#1}}
\newcommand{\ppool}[1]{\pcof{\mathrm{pool}}{#1}}
\newcommand{\peval}[1]{\pcof{\mathrm{eval}}{#1}}

\newcommand{\Dbatch}[1]{\Dany^\mathrm{acq}(#1)}
\newcommand{\yDeval}{\Dany^\mathrm{eval}(\yevalset)}

	\graphicspath{{./rholoss/}}

\newcommand{\xval}{\xevalset}
\newcommand{\yval}{\yevalset}

\newcommand{\redloss}{\mathcal{L}_\mathrm{red}}
\newcommand{\irrloss}{\mathcal{L}_\mathrm{irr}}
\newcommand{\loss}{\mathcal{L}_\mathrm{t}}

\newcommand{\xsubset}{\x^\mathrm{subset}}
\newcommand{\ysubset}{\y^\mathrm{subset}}

\newcommand{\xevali}{\x^\mathrm{eval}}
\newcommand{\Xevali}{\X^\mathrm{eval}}
\newcommand{\yevali}{\y^\mathrm{eval}}
\newcommand{\Yevali}{\Y^\mathrm{eval}}
\newcommand{\xbatch}{\x^\mathrm{batch}}
\newcommand{\Xbatch}{\X^\mathrm{batch}}
\newcommand{\ybatch}{\y^\mathrm{batch}}
\newcommand{\Ybatch}{\Y^\mathrm{batch}}

\newcommand{\fix}{\marginpar{FIX}}
\newcommand{\new}{\marginpar{NEW}}

\newtcolorbox{mybox}[1]{%
    tikznode boxed title,
    enhanced,
    arc=0mm,
    interior style={white},
    attach boxed title to top center= {yshift=-\tcboxedtitleheight/2},
    fonttitle=\bfseries,
    colbacktitle=white,coltitle=black,
    boxed title style={size=normal,colframe=white,boxrule=0pt},
    title={#1}}

\label{rholoss:sec:intro}

Now, let us take a look at active sampling. We will apply information-theoretic ideas to it. Active sampling is important because
state-of-the-art models such as GPT-3 \citep{brown2020language}, CLIP \citep{radford2021learning}, and ViT \citep{dosovitskiy2020image} achieve remarkable results by training on vast amounts of web-scraped data, yet despite intense parallelization, training such a model takes weeks or months \citep{radford2021learning, chowdhery2022palm}.
Even practitioners who work with smaller models face slow development cycles, due to numerous iterations of algorithm design and hyperparameter selection. 
As a result, the total time required for training is a core constraint in the development of such deep learning models. 

If it further sped up training, practitioners with sufficient resources would use much larger batches and distribute them across many more machines \citep{anil2018large}. However, this has rapidly diminishing returns \citep{lecun2012efficient}, to a point where adding machines does not reduce training time \citep{mccandlish2018empirical, anil2018large}---see e.g. GPT-3 and PaLM \citep{chowdhery2022palm}.

Additional machines can, however, still speed up training by filtering out less useful samples \citep{alain2015variance}.
Many web-scraped samples are \textcolor{myblue}{\textit{noisy}}, i.e. their label is incorrect or inherently ambiguous. For example, the text associated with a web-scraped image is rarely an accurate description of the image. 
Other samples are learned quickly and are then \textcolor{myred}{\textit{redundant}}. Redundant samples are commonly part of object classes that are over-represented in web-scraped data \citep{tian2021divide}, and they can often be left out without losing performance. 
Given that web-scraped data is plentiful---often enough to finish training in a single epoch \citep{komatsuzaki2019one, brown2020language}---one can afford to skip less useful points. 

However, there is no consensus on which data points are the most useful. Some works, including curriculum learning, suggest prioritizing \textit{easy} points with low \textcolor{myblue}{label noise} before training on all points equally \citep{bengio2009curriculum}. While 
this approach may improve convergence and generalization, it lacks a mechanism to skip points that are \textcolor{myred}{already learned} (\textcolor{myred}{redundant}). Other works instead suggest training on points that are \textit{hard} for the model, thereby avoiding \textcolor{myred}{redundant} points, whose loss cannot be further reduced. Online batch selection methods \citep{loshchilov2015online, katharopoulos2018not, jiang2019accelerating} do so by selecting points with high loss or high gradient norm.

We show two failure modes of prioritizing hard examples. Firstly, in real-world noisy datasets, high loss examples 
may be \textcolor{myblue}{mislabeled or ambiguous}.
Indeed, in controlled experiments, points selected by high loss or gradient norm are overwhelmingly those with \textcolor{myblue}{noise-corrupted} labels. Our results show that this failure mode degrades performance severely. More subtly, we show that some samples are hard because they are \textcolor{mygreen}{outliers}---points with unusual features that are \textcolor{mygreen}{less likely to appear at test time}. For the aim of reducing test loss, such points are \textcolor{mygreen}{less \textit{worth learning}}. 

To overcome these limitations, we introduce the \emph{reducible holdout loss selection} (RHO-LOSS) in this chapter. We propose a selection function grounded in probabilistic and information-theoretic modelling that quantifies by how much each point would reduce the loss on unseen data if we were to train on it, \textit{without actually training on it}. 
We show that optimal points for reducing holdout loss are \textcolor{myblue}{non-noisy}, \textcolor{myred}{non-redundant}, and \textcolor{mygreen}{task-relevant}. 
To approximate optimal selection, we derive an efficient and easy-to-implement selection function: the reducible holdout loss.

We explore RHO-LOSS in extensive experiments on 7 datasets. We evaluate the reduction in required training steps compared to uniform sampling and state-of-the-art batch selection methods. 
Our evaluation includes Clothing-1M, the main large benchmark with noisy, web-scraped labels, matching our main application.
RHO-LOSS reaches target accuracy in $18$x fewer steps than uniform selection and achieves $2\%$ higher final accuracy (\Cref{rholoss:fig:clothing-1M}). 
Further, RHO-LOSS consistently outperforms prior art and speeds up training across datasets, modalities, architectures, and hyperparameter choices. Explaining this, we show that methods selecting ``hard'' points prioritize noisy and less relevant examples. In contrast, RHO-LOSS chooses \textcolor{myblue}{low-noise}, \textcolor{mygreen}{task-relevant}, \textcolor{myred}{non-redundant} points---points that are \textcolor{myblue}{learnable}, \textcolor{mygreen}{worth learning}, and \textcolor{myred}{not yet learned}.

\section{Active Sampling: Online Batch Selection}
\label{rholoss:sec:background}

\begin{figure}
    \centering
    \includegraphics{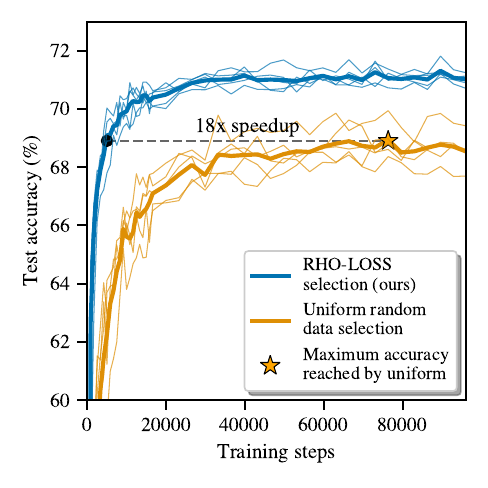}
    \caption{
    \textbf{Speedup on large-scale classification of \mbox{web-scraped} data (Clothing-1M)}. 
    RHO-LOSS trains all architectures with fewer gradient steps than standard uniform data selection (i.e. shuffling), helping reduce training time. Thin lines: ResNet-50, MobileNet v2, DenseNet121, Inception v3, GoogleNet, mean across seeds. Bold lines: mean across all architectures.}
    \label{rholoss:fig:clothing-1M}
\end{figure}

Unlike the setting we use in most of the other chapters, which we introduced in \Cref{glue:related_work:BNNs}, active sampling follows a different paradigm by virtue of being an online algorithm and using labeled data. We will try to use most of the notation from \Cref{glue:related_work:BNNs} and \Cref{glue:background:active_learning}; however, note that \emph{the labels are available in the pool set and evaluation set in this chapter}.

We consider a model $\pof{y\given \x, \theta}$ with parameters $\theta$ that we want to train using stochastic gradient descent (SGD) on a subset of samples from the available \emph{labeled} data $\Dpool = \{(\x_i, \y_i)\}_{i=1}^\poolsize$ using the cross-entropy loss.
At each training step $t$, we load a batch $b_t$ of size $\batchvar$ from $\Dpool$. 
In online batch selection \citep{loshchilov2015online}, we uniformly pre-sample a larger batch $B_t$ of size $\batchvar' \gg \batchvar$.
Then, we construct a smaller batch $b_t$ that consists of the top-ranking $\batchvar$ points in $B_t$ ranked by a label-aware selection function $S(\x_i, \y_i)$. 
We perform a gradient step to minimize a mini-batch loss $\Hof{\y_i \given \x_i, \theta}$ summed over $i\in b_t$. 
The next large batch $B_{t+1}$ is then pre-sampled from $\Dpool$ without replacement of previously sampled points (points are only replaced at the start of the next epoch). 

\section{(Joint) Predictive Information Gain \&\ Reducible Holdout Loss Selection}
\label{rholoss:sec:theory}

Previous online batch selection methods, such as loss or gradient norm selection, aim to select points that, if we were to train on them, would minimize the \textit{training set} loss. \citep{loshchilov2015online,katharopoulos2018not, kawaguchi2020ordered, alain2015variance}. Instead, we aim to select points that minimize the loss on a \textit{holdout set}.
In spirit of this thesis, we will refer to this holdout set as a \emph{labeled} evaluation set\footnote{Supposedly, we should then also rename the loss to REV-LOSS in this thesis, but let us not.} $\Deval$. 
It would be too expensive to naively train on every candidate point and evaluate the holdout loss each time. 

In this section, we show how to (approximately) find the points that would most reduce the holdout loss if we were to train the current model on them (without actually training on them). For simplicity, we first assume only one point \mbox{$(\x,\y)\in B_t$} is selected for training  at each time step $t$ (we discuss selection of multiple points below). $\pof{\y' \given \x'; \Dtrain}$ is the predictive distribution of the current model, where $\Dtrain$ is the sequence of data the model was trained on before training step $t$. $\Deval = \{(\xeval_i, \yeval_i)\}_{i=1}^{\evalsize}$, written as $\xval$ and $\yval$, respectively, for brevity, is a \emph{labeled} evaluation set drawn from the same data-generating distribution $\ptof{\x',\y'}$ as is the set of available training data $\Dpool$---this is the holdout set. 

\paragraph{Reduction in Holdout Loss.}
We aim to acquire the point $(\x, \y)\in B_t$ that, if we were to train on it, would minimize the cross-entropy loss on the holdout set, our \emph{labeled} evaluation set: 
\begin{flalign} 
    &\argmin_{(x,y) \in B_t} \CrossEntropy{\peval{\xeval,\yeval}}{\pof{\yeval \given \xeval; (\y, \x), \Dtrain}} \label{rholoss:PSI1}.
\end{flalign}
Thinking back to \Cref{chapter:pobal}, we see that in the active learning case, we were looking for a reduction in expected loss, while here we are looking for a reduction in the holdout loss.

\paragraph{(Joint) Predictive Information Gain.}
Using the notation from \Cref{chapter:practical_it_notation}, the cross-entropy loss over the empirical $\Deval$ is just:
\begin{align}
    &\CrossEntropy{\peval{\xeval,\yeval}}{\pof{\yeval \given \xeval, (\y, \x), \Dtrain}} \\
    &\quad = \E{\peval{\xeval, \yeval}}{\Hof{\yevalset \given \xevalset, (\y, \x), \Dtrain}}. 
\end{align}
Thus, we can  write the minimization above as a maximization of a (point-wise) mutual information analogously to derivations in \cref{chapter:pobal}:
\begin{align}
    &\argmin \E{\peval{\xeval, \yeval}}{\Hof{\yeval \given \xeval, (\y, \x), \Dtrain}} \\
    &\quad = \argmax \E{\peval{\xeval, \yeval}}{\MIof{\yeval; \y \given \xeval, \x, \Dtrain}}.
\end{align}
This is the predictive information gain:
\begin{definition}
    The \emph{predictive information gain} (PIG) of a point $(\x, \y)$ is defined as:
    \begin{align}
        \pigof{\x, \y} = \E{\peval{\xeval, \yeval}}{\MIof{\yeval \given \xeval, (\y, \x), \Dtrain}}.
    \end{align}
\end{definition}
The motivation for it is the same as for EPIG in \Cref{pobal:sec:method} with the crucial different label information is available here, making batch evaluation easier.

Similarly, we can define a joint predictive information gain.
\begin{definition}
    The \emph{joint predictive information gain} (JPIG) of a point $(\x, \y)$ is defined as:
    \begin{align}
        &\jpigof{\x, \y} = \MIof{\yevalset \given \xevalset, (\y, \x), \Dtrain} \\
        &\quad = \Hof{\yevalset \given \xevalset \Dtrain} - \Hof{\yevalset \given \xevalset, (\y, \x), \Dtrain} \\
        &\quad = -\log \pof{\yevalset \given \xevalset, \Dtrain} - (-\log \pof{\yevalset \given \xevalset, (\y, \x), \Dtrain}).
    \end{align}
\end{definition}
The motivation for JPIG is also the same as for JEPIG in \Cref{EPIG:sec:jepig_bayesian_model_selection}, again with the crucial difference that label information is available here. JPIG is even closer to the minimization of the log marginal likelihood, which is usually the proxy goal of Bayesian model selection.

In \Cref{chapter:obi}, we will take a different perspective on these quantities and examine them as marginal and joint cross-entropies (or marginal and joint cross-mutual information), which is more sensible, perhaps. However, we will still use the notation of PIG and JPIG for the rest of this chapter and this thesis otherwise, as it ties the different information quantities together more nicely\footnote{Although we could also follow a different naming scheme and call them T(E)IG, J(E)IG, and M(E)IG, which would be more similar to \citet{mackay1992information}. Naming things is not easy.}.

\paragraph{Deriving a Tractable Selection Function.}
We now derive a tractable expression for the term in Eq.~\eqref{rholoss:PSI1} that does not require us to train on each candidate point $(\x, \y)\in B_t$ and then evaluate the loss on $\Deval$. To make our claims precise and our assumptions transparent, we use the language of Bayesian probability theory. We treat model parameters as a random variable with prior $\pof{\theta}$ and infer a posterior $\pof{\theta \given \Dtrain}$ using the already-seen training data $\Dtrain$. The model has a predictive distribution $\pof{\y \given \x,\Dtrain} =~\int_\theta
\pof{\y \given \x,\theta}\pof{\theta \given \Dtrain}d\theta$. When using a point estimate of $\theta$, the predictive distribution can be written as an integral with respect to a Dirac delta.

We have already motivated that \cref{rholoss:PSI1} is equivalent to maximizing the PIG objective. As a first step, we will switch to using the JPIG objective (\textbf{Approximation 0}), which is more closely related to the log marginal likelihood and model selection. We will show that it leads to a tractable selection function: 
We simply rewrite JPIG using the symmetry of the point-wise mutual information as:
\begin{align}
    \MIof{\yevalset \given \xevalset, (\y, \x), \Dtrain} = \Hof{\y \given \x, \Dtrain} - \Hof{\y \given \x, \yevalset, \xevalset, \Dtrain}. \label{rholoss:RHOLOSS_wo_approximation}
\end{align}

As exact Bayesian inference (conditioning on $\Dtrain$ or $\Deval$) is intractable in neural networks \citep{blundell2015weight}, we fit the models with SGD instead (\textbf{Approximation 1}). We study the impact of this approximation in \Cref{rholoss:sec:impact_of_approximations}. The first term, $\Lof{\y \given \x, \Dtrain}$, is then the \textit{training loss} on the point $(\x,\y)$ using the current model trained on $\Dtrain$. The second term, $\Lof{\y \given \x, \Deval, \Dtrain}$, is the loss of a model trained on $\Dtrain$ and $\Deval$.

Although the selection function in Eq.~\eqref{rholoss:RHOLOSS_wo_approximation} is tractable, it is still somewhat expensive to compute, as both terms must be updated after each acquisition of a new point. However, we can approximate the second term with a model trained only on the holdout dataset, \mbox{$\Lof{\y \given \x, \Deval, \Dtrain} \approx \Lof{\y \given \x, \Deval}$} (\textbf{Approximation 2}).
This approximation saves a lot of compute: it is now sufficient to compute the term once before the first epoch of training. Later on, we show that this approximation empirically does not hurt performance on any tested dataset and even has additional benefits (\Cref{rholoss:sec:impact_of_approximations} and Appendix~\ref{rholoss:app:update_irreducible_2}).
We term $\Lof{\y \given \x, \Deval}$ the \textit{irreducible holdout loss} (IL) since it is the remaining loss on point $(\x,\y) \in  \Dpool$ after training on the holdout set (\emph{labeled} evaluation set) $\Deval$; in the limit of $\Deval$ being large, it would be the lowest loss that the model can achieve without training on $(\x,\y)$. Accordingly, we name our approximation of Eq.~\eqref{rholoss:RHOLOSS_wo_approximation} the \textit{reducible holdout loss}---the difference between the training loss and the irreducible holdout loss (IL).

Our method still requires us to train a model on a holdout set (\emph{labeled} evaluation set), but a final approximation greatly reduces that cost. We can efficiently compute the IL with an ``irreducible loss model'' (IL model) that is smaller than the target model and has low accuracy (\textbf{Approximation 3}). We show 
this and explain it in Sections~\ref{rholoss:sec:impact_of_approximations}, \ref{rholoss:sec:cheap_irlo}, and \ref{rholoss:sec:properties}. Counterintuitively, the reducible holdout loss can therefore be negative. Additionally, one IL model can be reused for many target model runs, amortizing its cost (\Cref{rholoss:sec:cheap_irlo}). For example, we trained all 40 seeds of 5 target architectures in \Cref{rholoss:fig:clothing-1M} using a single ResNet18 IL model. Further, this model trained for 37x fewer steps than each target model (reaching only 62\% accuracy). \Cref{rholoss:sec:related} details further possible efficiency improvements. 

In summary, selecting a point that minimizes the holdout loss in Eq.~\eqref{rholoss:PSI1}, for a model trained on $\Dtrain$, can be approximated with the following easy-to-compute objective:

\begin{mybox}{Reducible holdout loss selection (RHO-LOSS)}
\begin{flalign} \label{rholoss:eq:rholoss}
    \hspace{-2.5mm} \argmax_{(\x,\y) \in B_t} \:\:\: \overbrace{\underbrace{\Lof{\y \given \x, \Dtrain}}_{\text{training loss}}  ~~~-\underbrace{\Lof{\y \given \x, \Deval}}_{\text{irreducible holdout loss (IL)}}}^{\text{reducible holdout loss}}
\end{flalign}
\end{mybox}

Although we required additional data $\Deval$, this is not essential for large (\Cref{rholoss:sec:experiments}.0) nor small (\Cref{rholoss:sec:cheap_irlo}) datasets.

\begin{algorithm}[t]
	\caption{Reducible holdout loss selection (RHO-LOSS)}
    \label{rholoss:alg:RHOLOSS}

    \includegraphics{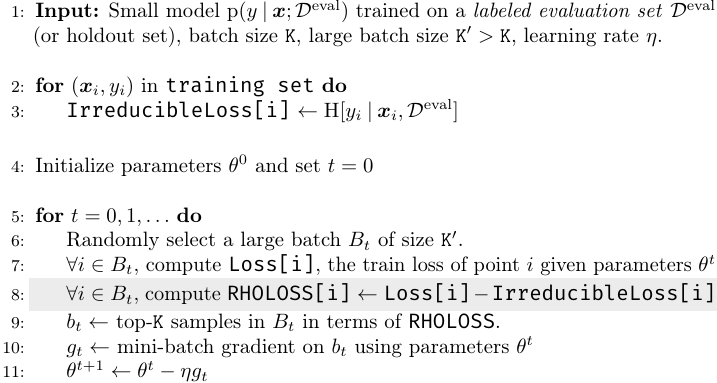}
\end{algorithm}

\paragraph{Understanding the Reducible Loss.} We now provide intuition on why reducible holdout loss selection (RHO-LOSS) avoids 
\textcolor{myred}{redundant}, \textcolor{myblue}{noisy}, and \textcolor{mygreen}{less relevant} points.
\mbox{\textbf{i) Redundant points.}} We call a training point redundant when the model has already learned it, i.e. its training loss cannot be further reduced. Since \textcolor{myred}{redundant} points have \textcolor{myred}{low training loss}, and the reducible loss is always less than the training loss (Eq.~(\ref{rholoss:eq:rholoss})), such points have low reducible loss and are not selected. 
And if the model forgets them, they are revisited in the next epoch. 
\textbf{ii) Noisy points.} While 
prior methods select based on high training loss (or gradient norm), not all points with high loss are informative---some may have an \textcolor{myblue}{ambiguous or incorrect} (i.e. \textcolor{myblue}{noisy}) label. The labels of such points cannot be predicted using the holdout set (\emph{labeled} evaluation set)~\citep{chen2019understanding}. Such points have \textcolor{myblue}{high IL} and, consequently, low reducible loss. These \textcolor{myblue}{noisy} points are less likely to be selected compared to equivalent points with less noise. \textbf{iii) Less relevant points.} Loss-based selection has an additional pitfall. The training loss is likely higher for \textcolor{mygreen}{outliers} in input space---values of $\x$ far from most of the training data, in regions with \textcolor{mygreen}{low input density} under $\ptof{\x}$.
Points with low $\ptof{\x}$ should not be prioritized, all else equal. Consider an `outlier' $(\x, \y)$ and a non-outlier $(\x', \y')$, with $\ptof{\x}< \ptof{\x'}$ but \textit{equal} training loss $\Lof{\y|\x,\Dtrain}=\Lof{y'|\x',\Dtrain}$. As the holdout set (\emph{labeled} evaluation set) $\Deval$ is also drawn from $\text{p}_{\text{true}}$, $\Deval$ will contain fewer points from the region around $\x$ in input space compared to the region around $\x'$. Thus, training on $(\x, \y)$ is likely to reduce the holdout loss (Eq.~(\ref{rholoss:PSI1})) less, and so we prefer to train on the non-outlier $(\x', \y')$. In the specific sense described, $(\x, \y)$ is thus \textcolor{mygreen}{less \textit{relevant}} to the holdout set (\emph{labeled} evaluation set). 
As desired, RHO-LOSS deprioritizes $(\x, \y)$: since $\Deval$ contains few points from the region around $\x$, the IL of $(\x, \y)$ will be large.

In short, RHO-LOSS \textit{deprioritizes} points that are \textcolor{myred}{redundant} (low training loss), \textcolor{myblue}{noisy} (high IL), or \textcolor{mygreen}{less relevant} to the holdout set (high IL). That is, RHO-LOSS \textit{prioritizes} points that are \textcolor{myred}{not yet learned}, \textcolor{myblue}{learnable}, and \textcolor{mygreen}{worth learning}. We provide empirical evidence for these claims in \Cref{rholoss:sec:properties}. See Algorithm~\ref{rholoss:alg:RHOLOSS} for the implementation of RHO-LOSS.

\paragraph{Batch Selection.}
We showed which point is optimal when selecting a single point $(\x,\y)$. 
When selecting an entire batch $b_t$, we select the points with the top-$\batchvar$ scores from the randomly pre-sampled set $B_t$. 
This is nearly optimal when assuming that 
each point has little effect on the score of other points, which is often used as a simplifying assumption in active learning. This assumption is much more reasonable in our case than in active learning because model predictions are not changed much by a single gradient step on one mini-batch.

\paragraph{Simple Batch Selection.}

For large-scale neural network training, practitioners with sufficient resources would use many more machines if it further sped up training \citep{anil2018large}. However, as more workers are added in synchronous or asynchronous gradient descent, the returns diminish to a point where adding more workers does not further improve wall clock time \citep{anil2018large, mccandlish2018empirical}. For example, there are rapidly diminishing returns for using larger batch sizes or distributing a given batch across more workers, for multiple reasons \citep{mccandlish2018empirical, keskar2016large}. The same holds for distributing the model across more workers along its width or depth dimension \citep{rasley2020deepspeed, shoeybi2019megatron, huang2019gpipe}. However, we can circumvent these diminishing returns by adding a new dimension of parallelization, namely, for data selection.

Since parallel \textit{forward} passes do not suffer from such diminishing returns, one can use extra workers to evaluate training losses in parallel \citep{alain2015variance}. The theoretical runtime speedup can be understood as follows. The cost per training step of computing the selection function on $B_t$ is $\frac{\batchvar'}{3\batchvar}$ times as much as the cost of the forward-backward pass needed to train on $b_t$ since a forward pass requires at least $3$x less computation than a forward-backward pass \citep{jouppi2017datacenter}. One can reduce the time for the selection phase almost arbitrarily by adding more workers that compute training losses using a copy of the model being trained. The limit is reached when the time for selection is dominated by the communication of parameter updates to workers. More sophisticated parallelization strategies allow reducing the time overhead even further (\Cref{rholoss:sec:related}). To avoid assumptions about the particular strategy used, we report experiment results in terms of the required number of training epochs.

\section{Empirical Validation}\label{rholoss:sec:experiments}

We evaluate our selection method on several datasets (both in controlled environments and real-world conditions) and show significant speedups compared to prior art, in the process shedding light on the properties of different selection functions. 

Recall that our setting assumes training time is a bottleneck, but data is abundant---more than we can train on (see \citet{bottou2004large}). This is common e.g. for web-scraped data where state-of-the-art performance is often reached in less than half of one epoch \citep{komatsuzaki2019one, brown2020language}. As data is abundant, we can set aside a holdout set (\emph{labeled} evaluation set) for training the IL model with little to no downside. For the large Clothing-1M dataset, we implement RHO-LOSS by training the IL model on 10\% of the training data, while all baselines are trained on the full 100\% of the training data. 
For the smaller datasets, we simulate abundance of data by reserving a holdout set and training \textit{all} methods only on the remaining data. 
However, RHO-LOSS also works on small datasets without additional data by double-using the training set (\Cref{rholoss:sec:cheap_irlo}).

\paragraph{Datasets.}
We evaluate on 7 datasets: 
1) QMNIST \citep{qmnist-2019} extends MNIST \citep{lecun1998gradient} with 50k extra images which we use as the holdout set (\emph{labeled} evaluation set).
2) On CIFAR-10 \citep{krizhevsky2009learning} we train on half of the training set and use the other half as a holdout to train the irreducible loss (IL) model. 3) CIFAR-100: same as CIFAR-10. 4) CINIC-10 \citep{darlow2018cinic} has 4.5x more images than CIFAR-100 and includes a validation set (which we use as holdout set) and a test set with 90k images each. 5) Clothing-1M \citep{xiao2015learning}, which contains over 1 million 256x256-resolution clothing images from 14 classes. The dataset is fully web-scraped---a key application area of this chapter---and is the most widely accepted benchmark for image recognition with noisy labels \citep{algan2021image}. 
We use the whole training set for training and reuse 10\% of it to train the IL model. We further evaluate on two NLP datasets from GLUE \citep{wang2018glue}: 6) CoLA (grammatical acceptability) and 7) SST-2 (sentiment). We split their training sets as for CIFAR.

\begin{figure*}
    \centering
    \hspace*{-1.5mm}\includegraphics[width=0.95\linewidth]{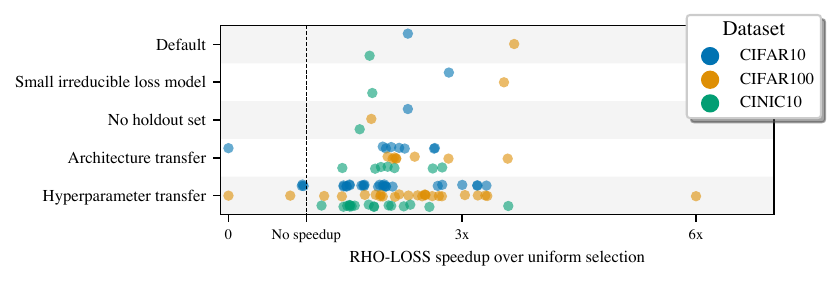}
    \caption{\textbf{The irreducible loss model can be small, trained with no holdout data, and reused across target architectures and hyperparameters.} Here, we use clean datasets, where speedups are smallest. The x-axis shows speedup, i.e. after how many fewer epochs RHO-LOSS exceeds the highest accuracy uniform selection achieves within 100 epochs. 
    Row 1 uses a ResNet18 as irreducible loss model. All other rows instead use a small, cheap CNN.
    Each dot shows an experiment with a given combination of irreducible loss model and target model (mean across 2-3 seeds for all but the last row).}
    \label{rholoss:fig5}
\end{figure*}

\paragraph{Baselines.}
Aside from uniform sampling (without replacement, i.e. random shuffling), we also compare to selection functions that have achieved competitive performance in online batch selection recently: the (training) loss, as implemented by \citet{kawaguchi2020ordered}, gradient norm, and gradient norm with importance sampling (called \textit{gradient norm IS} in our figures), as implemented by \citet{katharopoulos2018not}.
We also compare to the core-set method Selection-via-Proxy (SVP) that selects data offline before training \citep{coleman2019selection}. We report results using maximum entropy SVP and select with the best-performing model, ResNet18. 
We further compare to four baselines from active learning, shown in Appendix~\ref{rholoss:app:active_learning} as they assume labels are unobserved. Finally, we include selection using the negative IL (see Eq.~\ref{rholoss:eq:rholoss}) to test if it is sufficient to only skip noisy and less relevant but not redundant points. %

\paragraph{Models and Hyperparameters.}
To show our method needs no tuning, we use the PyTorch default hyperparameters (with the AdamW optimizer \citep{loshchilov2017decoupled}) and $\frac{\batchvar}{\batchvar'}=0.1$. 
We test many additional hyperparameter settings in Figs.~\ref{rholoss:fig5} (row 5) and \ref{rholoss:fig:percent_train}. We test various architectures in Figs.~\ref{rholoss:fig:clothing-1M} and \ref{rholoss:fig5} (row 4). 
In all other figures, we use a 3 layer MLP for experiments on QMNIST, a ResNet-18 adapted for small images for CIFAR-10/CIFAR-100/CINIC-10, and a ResNet-50 for Clothing-1M. All models for Clothing-1M are pre-trained on ImageNet (standard for this dataset \cite{algan2021image}) and the IL model is \textit{always} a ResNet-18. For the NLP datasets, we use a pre-trained ALBERT v2 \citep{lan2019albert}.
We always use the IL model checkpoint with the lowest validation loss (not the highest accuracy); this performs best. Details in Appendix~\ref{rholoss:app:exp_details}.

\paragraph{Evaluation.} We measure speedup in terms of the number of epochs needed to reach a given test accuracy. We measure epochs needed, rather than wall clock time, as our focus is on evaluating a new selection function, not an entire training pipeline. 
Wall clock time depends primarily on the hardware used and implementation details that are beyond our scope. 
Most importantly, data selection is amenable to parallelization beyond standard data parallelism as discussed in \Cref{rholoss:sec:theory}.

\subsection{Impact of Approximations}\label{rholoss:sec:impact_of_approximations}

In \Cref{rholoss:sec:theory}, we introduced a function for selecting exactly the points that most reduce the model's loss on a holdout set (\emph{labeled} evaluation set). To make this selection function efficient for deep neural networks, we made several approximations. Here, we study how these approximations affect the points selected, by successively introducing one approximation after the other.

Because the exact selection function (Eq.~\eqref{rholoss:RHOLOSS_wo_approximation}) is intractable, we start with a close (and expensive) approximation as the gold standard (Approximation 0). To make Approximation 0 feasible, the experiments are conducted on an easy dataset---QMNIST (with 10\% uniform label noise and data duplication to mimic the properties of web-scraped data). We then successively introduce the Approximations 1, 2, and 3 described in \Cref{rholoss:sec:theory}. To assess the impact of each approximation, we train a model without and with the approximations, 
and then compute the rank correlation (Spearman's correlation coefficient) of the selection function evaluated on each batch $B_t$. Across the first epoch, we present the mean of the rank correlations.
Since each approximation selects different data, the corresponding models become more different over time; this divergence likely causes some of the observed differences in the points they select.
See Appendix~\ref{rholoss:sec:approximation_appendix} for details.

\begin{table}
\caption{
Spearman's rank correlation of rankings of data points by selection functions that are increasingly less faithful approximations of Eq.~\eqref{rholoss:RHOLOSS_wo_approximation}, compared to the most faithful approximation. Approximations added from left to right. Mean across 3 seeds.}
\begin{tabular}{@{}lrrrrl@{}}
\toprule
                  & Non-     & Not       & Not updating & Small  \\
                  & Bayesian & converged & IL model    & IL model \\ \midrule
Rank correlation   & 0.75 & 0.76 & 0.63 & 0.51 \\
\bottomrule
\end{tabular}
\label{rholoss:table:overlap}
\end{table}

\textit{Approximation 0.} To get as close as possible to the Bayesian inference/conditioning used in Eq.~\eqref{rholoss:RHOLOSS_wo_approximation}, we use a deep ensemble of 5 neural networks and train them \textit{to convergence} after every time step $t$ on the acquired dataset $b_t \cup \Dtrain$ \cite{wilson2020bayesian}.

\textit{Approximation 1: SGD instead of Bayesian inference/conditioning.} Approximation 0 is a close approximation of Eq.~\eqref{rholoss:RHOLOSS_wo_approximation}, but training an ensemble to convergence at every step $t$ is far too expensive in practice. Starting from this gold-standard, we introduce two stronger approximations (1a and 1b) to move to standard neural network fitting with AdamW. 1a) First, we replace the ensemble with a single model, while still training to convergence at each time step. The Spearman's coefficient between this approximation and Approximation~0 is 0.75, suggesting similar points are selected (``Non-Bayesian'' in \Cref{rholoss:table:overlap}).
1b) Next, we only take one gradient step on each new batch $b_t$. The Spearman's coefficient, when comparing this to Approximation 0, is 0.76 (``Not Converged'' in Table \ref{rholoss:table:overlap}).%

\textit{Approximation 2. Not updating the IL model on the acquired data $\Dtrain$.} Second, we save compute by approximating $\Lof{\y \given \x, \Dtrain, \Deval}$ with $\Lof{\y \given \x, \Deval}$. The points selected are still similar to Approximation~0 (Spearman's coefficient 0.63, ``Not updating IL model'' in \Cref{rholoss:table:overlap}). This approximation also performs well on other datasets (Appendix \ref{rholoss:app:update_irreducible_2}).

\textit{Approximation 3: Small IL model.} Lastly, we use a model with 256 hidden units instead of 512 (4x fewer parameters) as the IL model and see again that similar points are selected (Spearman's coefficient 0.51). We study cheaper IL models in other forms and datasets in the next section.

\begin{figure*}
    \centering
    \includegraphics[width=\textwidth]{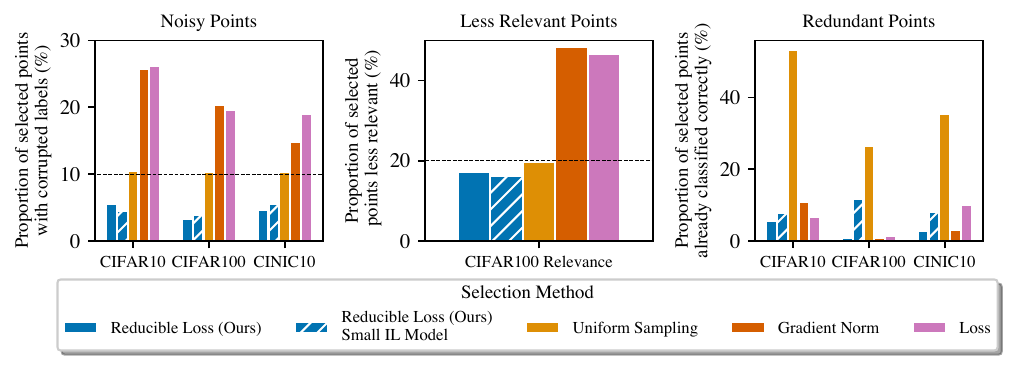}
    \caption{Properties of RHO-LOSS and other methods. 
    RHO-LOSS prioritizes points that are \textcolor{myblue}{non-noisy}, \textcolor{mygreen}{task-relevant}, and \textcolor{myred}{non-redundant}---even when the irreducible loss (IL) model is a small CNN. In contrast, loss and gradient norm prioritize noisy and less relevant points (while also avoiding redundant points). 
    \textbf{Left.} Proportion of selected points with corrupted labels. We added $10\%$ uniform label noise, i.e., we randomly switched each point's label with $10\%$ probability.
    \textbf{Middle.} Proportion of selected points from low relevance classes on CIFAR100 Relevance dataset.
    \textbf{Right.} Proportion of selected points that are already classified correctly, which is a proxy for redundancy.
    Mean over 150 epochs of training and 2-3 seeds. 
    }
    \label{rholoss:fig:properties}
\end{figure*}

\subsection{Cheap Irreducible Loss Models \&\ Robustness}
\label{rholoss:sec:cheap_irlo}%

RHO-LOSS requires training an IL model on a holdout set (\emph{labeled} evaluation set), which poses additional costs. Here, we show how to minimize these costs and amortize them across many training runs of target models. The same experiments also  show the robustness of RHO-LOSS across architectures and hyperparameter settings. To fit our computational budget, we perform these experiments on moderate-sized clean benchmark datasets although RHO-LOSS speeds up training more on noisy or redundant web-scraped data (see \Cref{rholoss:sec:speedups}).

\paragraph{Irreducible Loss Models: Small \&\ Cheap.}

In our default setting (\Cref{rholoss:fig5}, row 1), both the target model and IL model have the same architecture (ResNet-18). In rows 2 and below, we instead used a small CNN similar to LeNet as the IL model \citep{lecun1989backpropagation}. It has 21x fewer parameters and requires 29x fewer FLOP per forward pass than the ResNet-18. \textbf{The smaller IL model accelerates training as much or more than the larger model}, even though its final accuracy is far lower than the target ResNet-18 (11.5\% lower on CIFAR-10, 7\% on CIFAR-100, and 8.1\% on CINIC-10). We examine in \Cref{rholoss:sec:properties} why this useful result holds.

\paragraph{Irreducible Loss Models: No Holdout Data.}
Web-scraped datasets are often so large that even a small fraction of the overall data can be sufficient to train the IL model. E.g., in our experiments on Clothing-1M (\Cref{rholoss:fig:clothing-1M}), the holdout set (\emph{labeled} evaluation set) is only 10\% as large as the main train set. Additionally, we can train the IL model without any holdout data (\Cref{rholoss:fig5}, row 3). We split the training set $\Dpool$ into two halves and train an IL model on each half (still using small IL models). Each model computes the IL for the half of $\Dpool$ that it was not trained on. Training two IL models costs no additional compute since each model is trained on half as much data compared to the default settings. 

\paragraph{Irreducible Loss Models: Reuse for Different Target Architectures.}
We find that a single small CNN IL model accelerates the training of 7 target architectures (\Cref{rholoss:fig5}, row 4): VGG11 (with batch norm), GoogleNet, Resnet34, Resnet50, Densenet121, MobileNet-v2, Inception-v3. RHO-LOSS does not accelerate training on CIFAR-10 for VGG11, which is also the architecture on which uniform training performs the worst; i.e. RHO-LOSS empirically does not “miss” a good architecture. Not only is RHO-LOSS robust to architectures choice, a single IL model can also be \textit{reused by many practitioners} who use different architectures (as we did in \Cref{rholoss:fig:clothing-1M}).

\paragraph{Irreducible Loss Models: Reuse for Hyperparameter Sweeps.}
We find that a single small CNN accelerates the training of ResNet-18 target models across a hyperparameter grid search (\Cref{rholoss:fig5}, last row). We vary the batch size (160, 320, 960), learning rate (0.0001, 0.001, 0.01), and weight decay coefficient (0.001, 0.01, 0.1). RHO-LOSS speeds up training compared to uniform on nearly all target hyperparameters. The few settings in which it doesn't speed up training are also settings in which uniform training performs very poorly ($<30\%$ accuracy on CIFAR-100, $<80\%$ on CIFAR-10). 

\begin{table*}[ht]
\caption{Epochs required to reach a given target test accuracy (final accuracy in parentheses).  Figs.~\ref{rholoss:fig:walltime}~and~\ref{rholoss:fig:walltime_nlp} (Appendix) show all training curves.
Some datasets have 10\% uniform label noise added. 
Results averaged across 2-4 seeds. Best performance in \textbf{bold}.  RHO-LOSS performs best in both epochs required and final accuracy. {\small \emph{NR}} indicates that the target accuracy was not reached. $^*$On CIFAR10/100, CoLA, and SST-2, only half of the data is used for training (\Cref{rholoss:sec:experiments}.0).}
\begin{adjustbox}{width=\textwidth}
\begin{tabular}{c|c|ccccc|c|c}
\hline
\multirow{2}{*}{Dataset} & \multirow{2}{*}{Target Acc} & \multicolumn{7}{c}{Number of epochs method needs to reach target accuracy $\downarrow$ (Final accuracy in parentheses)} \\
 &  & Train Loss & Grad Norm & Grad Norm IS & SVP & Irred Loss & Uniform & RHO-LOSS \\ \hline

 Clothing-1M & 60.0\% & {\color{MyDarkGray}8} & {\color{MyDarkGray}13} & {\color{MyDarkGray}2} & {\small \color{MyDarkGray} \emph{NR}} & {\small \color{MyDarkGray} \emph{NR}} & 2 & \textbf{1} \\
  & 69.0\% & {\small \color{MyDarkGray} \emph{NR}} {\color{MyDarkGray} (65\%)} & {\small \color{MyDarkGray} \emph{NR}} {\color{MyDarkGray} (64\%)} & {\color{MyDarkGray}9} {\color{MyDarkGray} (70\%)} & {\small \color{MyDarkGray} \emph{NR}} {\color{MyDarkGray} (55\%)} & {\small \color{MyDarkGray} \emph{NR}} {\color{MyDarkGray} (48\%)} & 30 (70\%) & \textbf{2 (72\%)} \\

  \rowcolor{MyLightGray} CIFAR10$^*$ & 80.0\% & {\color{MyDarkGray}81} & {\small \color{MyDarkGray} \emph{NR}} & {\color{MyDarkGray}57} & {\small \color{MyDarkGray} \emph{NR}} & {\small \color{MyDarkGray} \emph{NR}} & 79 & \textbf{39} \\
  \rowcolor{MyLightGray} & 87.5\% & {\color{MyDarkGray}129} {\color{MyDarkGray} (90\%)} & {\small \color{MyDarkGray} \emph{NR}} {\color{MyDarkGray} (61\%)} & {\color{MyDarkGray}139} {\color{MyDarkGray} (89\%)} & {\small \color{MyDarkGray} \emph{NR}} {\color{MyDarkGray} (55\%)} & {\small \color{MyDarkGray} \emph{NR}} {\color{MyDarkGray} (60\%)} & {\small \emph{NR}} (87\%) & \textbf{65 (91\%)} \\

  CIFAR10$^*$ & 75.0\% & {\small \color{MyDarkGray} \emph{NR}} & {\small \color{MyDarkGray} \emph{NR}} & {\color{MyDarkGray}57} & {\small \color{MyDarkGray} \emph{NR}} & {\small \color{MyDarkGray} \emph{NR}} & 62 & \textbf{27} \\
  (Label Noise) & 85.0\% & {\small \color{MyDarkGray} \emph{NR}} {\color{MyDarkGray} (28\%)} & {\small \color{MyDarkGray} \emph{NR}} {\color{MyDarkGray} (23\%)} & {\small \color{MyDarkGray} \emph{NR}} {\color{MyDarkGray} (84\%)} & {\small \color{MyDarkGray} \emph{NR}} {\color{MyDarkGray} (48\%)} & {\small \color{MyDarkGray} \emph{NR}} {\color{MyDarkGray} (62\%)} & {\small \emph{NR}} (85\%) & \textbf{49 (91\%)} \\

  \rowcolor{MyLightGray} CIFAR100$^*$ & 40.0\% & {\color{MyDarkGray}138} & {\color{MyDarkGray}139} & {\color{MyDarkGray}71} & {\small \color{MyDarkGray} \emph{NR}} & {\color{MyDarkGray}93} & 65 & \textbf{48} \\
  \rowcolor{MyLightGray} & 52.5\% & {\small \color{MyDarkGray} \emph{NR}} {\color{MyDarkGray} (42\%)} & {\small \color{MyDarkGray} \emph{NR}} {\color{MyDarkGray} (42\%)} & {\color{MyDarkGray}132} {\color{MyDarkGray} (55\%)} & {\small \color{MyDarkGray} \emph{NR}} {\color{MyDarkGray} (18\%)} & {\small \color{MyDarkGray} \emph{NR}} {\color{MyDarkGray} (43\%)} & 133 (54\%) & \textbf{77 (61\%)} \\

  CIFAR100$^*$ & 40.0\% & {\small \color{MyDarkGray} \emph{NR}} & {\small \color{MyDarkGray} \emph{NR}} & {\color{MyDarkGray}94} & {\small \color{MyDarkGray} \emph{NR}} & {\color{MyDarkGray}89} & 79 & \textbf{49} \\
  (Label Noise) & 47.5\% & {\small \color{MyDarkGray} \emph{NR}} {\color{MyDarkGray} (4\%)} & {\small \color{MyDarkGray} \emph{NR}} {\color{MyDarkGray} (4\%)} & {\color{MyDarkGray}142} {\color{MyDarkGray} (48\%)} & {\small \color{MyDarkGray} \emph{NR}} {\color{MyDarkGray} (14\%)} & {\small \color{MyDarkGray} \emph{NR}} {\color{MyDarkGray} (43\%)} & 116 (50\%) & \textbf{65 (60\%)} \\

  \rowcolor{MyLightGray} CINIC10 & 70.0\% & {\small \color{MyDarkGray} \emph{NR}} & {\small \color{MyDarkGray} \emph{NR}} & {\color{MyDarkGray}34} & {\small \color{MyDarkGray} \emph{NR}} & {\small \color{MyDarkGray} \emph{NR}} & 38 & \textbf{27} \\
  \rowcolor{MyLightGray} & 77.5\% & {\small \color{MyDarkGray} \emph{NR}} {\color{MyDarkGray} (36\%)} & {\small \color{MyDarkGray} \emph{NR}} {\color{MyDarkGray} (50\%)} & {\color{MyDarkGray}64} {\color{MyDarkGray} (82\%)} & {\small \color{MyDarkGray} \emph{NR}} {\color{MyDarkGray} (39\%)} & {\small \color{MyDarkGray} \emph{NR}} {\color{MyDarkGray} (60\%)} & 97 (80\%) & \textbf{38 (83\%)} \\

  CINIC10 & 60.0\% & {\small \color{MyDarkGray} \emph{NR}} & {\small \color{MyDarkGray} \emph{NR}} & {\color{MyDarkGray}22} & {\small \color{MyDarkGray} \emph{NR}} & {\color{MyDarkGray}30} & 24 & \textbf{13 }\\
  (Label Noise) & 67.5\% & {\small \color{MyDarkGray} \emph{NR}} {\color{MyDarkGray} (16\%)} & {\small \color{MyDarkGray} \emph{NR}} {\color{MyDarkGray} (16\%)} & {\color{MyDarkGray}35} {\color{MyDarkGray} (79\%)} & {\small \color{MyDarkGray} \emph{NR}} {\color{MyDarkGray} (39\%)} & {\small \color{MyDarkGray} \emph{NR}} {\color{MyDarkGray} (64\%)} & 38 (78\%) & \textbf{17 (82\%)} \\

  \rowcolor{MyLightGray} SST2$^*$ & 82.5\% & {\small \color{MyDarkGray}8} & {\small \color{MyDarkGray}2} & {\color{MyDarkGray}3} & {\small \color{MyDarkGray} \emph{NR}} & {\small \color{MyDarkGray}7} & \textbf{1} & \textbf{1} \\
  \rowcolor{MyLightGray} & 90.0\% & {\small \color{MyDarkGray} \emph{NR}} {\color{MyDarkGray} (87\%)} & {\small \color{MyDarkGray}4} {\color{MyDarkGray} (91\%)} & {\color{MyDarkGray} \emph{NR}} {\color{MyDarkGray} (89.7\%)} & {\small \color{MyDarkGray} \emph{NR}} {\color{MyDarkGray} (66\%)} & {\small \color{MyDarkGray} \emph{NR}} {\color{MyDarkGray} (83\%)} & 6 (90\%) & \textbf{3 (92\%)} \\

  CoLA$^*$ & 75.0\% & {\small \color{MyDarkGray}8} & {\small \color{MyDarkGray}6} & {\color{MyDarkGray}16} & {\small \color{MyDarkGray} \emph{NR}} & {\color{MyDarkGray} \emph{NR}} & 34 & \textbf{3}\\
  & 80.0\% & {\small \color{MyDarkGray} \emph{NR}} {\color{MyDarkGray} (78\%)} & {\small \color{MyDarkGray} \emph{NR}} {\color{MyDarkGray} (79\%)} & {\color{MyDarkGray} \emph{NR}} {\color{MyDarkGray} (78\%)} & {\small \color{MyDarkGray} \emph{NR}} {\color{MyDarkGray} (62\%)} & {\small \color{MyDarkGray} \emph{NR}} {\color{MyDarkGray} (69\%)} & \emph{NR} (76\%) & \textbf{39 (80\%)} \\

 \hline \end{tabular}
  \label{rholoss:tab:speedups}
\end{adjustbox}
\end{table*}

\subsection{Properties of RHO-LOSS \&\ Other Selection Functions}
\label{rholoss:sec:properties}

We established that RHO-LOSS can accelerate the training of various target architectures with a single IL model, even if the IL model is smaller and has considerably lower accuracy than the target models (\Cref{rholoss:sec:cheap_irlo}). This suggests robustness to target-IL architecture mismatches.

To understand this robustness, we investigate the properties of points selected by RHO-LOSS, when the target and IL model architectures are identical, and when they differ. In both cases, we find that RHO-LOSS prioritizes points that are \textcolor{myblue}{non-noisy}, \textcolor{mygreen}{task-relevant}, and \textcolor{myred}{not redundant}. We also investigate the properties of points selected by prior art. 

\paragraph{Noisy Points.} We investigate how often different methods select noisy points by uniformly corrupting the labels for 10\% of points and tracking what proportion of selected points are corrupted. RHO-LOSS deprioritizes noisy points for both IL models (\Cref{rholoss:fig:properties}). We observe a failure mode of the widely-used loss and gradient norm selection functions: they 
select far more noisy points than uniform. These methods also severely drop in accuracy when the noise follows the class confusion matrix \citep{rolnick2017deep} and when we add ambiguous images \citep{mukhoti2021deterministic} (Appendix~\ref{rholoss:app:noise}).

Together, this suggests that noisy points have high loss (and gradient norm), but also high IL and thus low reducible loss. Their IL is high even when the IL model is small as noisy labels cannot be predicted well using the holdout set (\emph{labeled} evaluation set).

\paragraph{Relevant Points.} We study how often less relevant points are selected by creating the CIFAR100 Relevance dataset, in which 80\% of the data comes from 20\% of the classes. 
This mimics natural distributions of NLP and vision data where most data comes from few object classes, topics, or words \citep{baayen2001word,tian2021divide}. 
Concretely, we retain all examples from 20 randomly chosen ``high relevance'' classes but only 6\% of the examples from other, ``low relevance'' classes. Intuitively, since the high relevance classes have higher $\ptof{\x}$ and are 17x more likely to appear at test time, improving their accuracy improves the test accuracy much more than improving the accuracy of less relevant classes. 

The loss and gradient norm methods select more points than uniform selection from the low relevance classes (\Cref{rholoss:fig:properties}).
In contrast, RHO-LOSS selects somewhat fewer low relevance points, suggesting these classes have high IL.
Since the less relevant classes are less abundant in the holdout set (\emph{labeled} evaluation set), both the small and large IL models have higher loss on them.

\paragraph{Redundant Points.} To investigate whether methods select redundant points, we track the percentage of selected points that are already classified correctly. This is only a proxy for redundancy; points that are classified correctly but with low confidence are not fully redundant, since their loss can be further reduced. We control for the different accuracy reached by each method by averaging only over epochs in which test accuracy is lower than the final accuracy reached by the weakest performing method. \Cref{rholoss:fig:properties} shows that all methods select fewer redundant points than uniform sampling.

\subsection{Speedup}
\label{rholoss:sec:speedups}
Finally, we evaluate how much different selection methods speed up training. 
Recall that the main application area for this chapter is large web-scraped datasets, known for high levels of noise and redundancy. Clothing-1M is such a dataset (\Cref{rholoss:sec:experiments}.0). We also include smaller, clean benchmarks from vision (CIFAR-10, CIFAR-100, CINIC-10) and NLP (CoLA, SST-2). 
Finally, we study if selection functions are robust to the controlled addition of label noise.

\paragraph{Speedup on Clean Data.}
RHO-LOSS reaches target accuracies in fewer epochs than uniform selection on all datasets (\Cref{rholoss:tab:speedups}). It also outperforms state-of-the-art methods by a clear margin in terms of speed and final accuracy. On the challenging CoLA language understanding dataset, the speedup over uniform selection exceeds $10$x.
In \Cref{rholoss:tab:double_irlomo} (Appendix~\ref{rholoss:app:steps_needed}), we find similar speedups when using no holdout data. 

\paragraph{Speedup on Noisy Data.}
When adding $10$\% label noise, batch selection with RHO-LOSS achieves greater speedups while, as hypothesized, prior art degrades (\Cref{rholoss:tab:speedups}). Notably, on noisier data, the speedup over uniform selection grows. 

\paragraph{Speedup on Large Web-Scraped Data.} 
On Clothing-1M, loss-based and gradient norm-based selection fail to match uniform selection, suggesting they are not robust to noise. In contrast, RHO-LOSS reaches the highest accuracy that uniform selection achieves during 50 epochs in just 2 epochs and improves final accuracy (72\% vs 70\%). Notably, this was possible even though the IL model we used has low accuracy (62.2\%) and was trained on ca. $10$x fewer data. RHO-LOSS also used 2.7x fewer FLOPs to reach the peak accuracy of uniform selection, including the cost of training the IL model (which could be amortized) and despite our implementation being designed to save time, not compute. While \Cref{rholoss:tab:speedups} shows results for a Resnet-50, \Cref{rholoss:fig:clothing-1M} includes several additional architectures, with an average speedup of $18$x.

\section{Related Work}
\label{rholoss:sec:related}

\paragraph{Time-Efficient Data Selection.}
Forward passes for selection can be accelerated using low-precision hardware or parallelization. While backward passes typically require high precision, forward passes can tolerate lower precision \citep{jouppi2017datacenter, jiang2019accelerating}, especially as we only need the loss (not the activations which would be needed for backpropagation). A forward pass by default requires roughly $3$x less time than a forward-backward pass, but this speedup can be increased to a factor around $10$x when using the low-precision cores available in modern GPUs and TPUs \citep{jouppi2017datacenter, jiang2019accelerating}. Further, prior work uses a set of workers that perform forward passes on $B_t$ or on the entire dataset asynchronously while the master process trains on recently selected data \citep{alain2015variance}. 

\paragraph{Compute-Efficient Data Selection.}
While we limit our scope to comparing selection functions, and we compute them naively, this choice is inefficient in practice. Selection can be made cheaper by reusing losses computed in previous epochs \citep{loshchilov2015online, jiang2019accelerating} or training a small model to predict them \citep{katharopoulos2017biased, zhang2019autoassist, coleman2019selection}. Alternatively, core set methods perform selection once before training \citep{mirzasoleiman2019coresets, borsos2020coresets}, although typically with more expensive selection functions.

\paragraph{Data Selection Functions.}
RHO-LOSS is best understood as an alternative to existing selection functions, which can be categorized by the properties of points they select and whether they use information about labels. ``Hard’’ points are selected both by high loss \citep{loshchilov2015online,kawaguchi2020ordered,jiang2019accelerating} and high prediction uncertainty \citep{Settles2010, li2006confidence, coleman2019selection}. However, prediction uncertainty does not require labels and can thus be used for active learning.
Despite this, they both suffer from the same problem: high loss and high uncertainty can be caused by noisy (in particular, ambiguous) labels.
This also applies to selection of points whose labels are easily forgotten during training \citep{toneva2018empirical}. Noisy points are avoided by our negative IL baseline and similar methods \citep{pleiss2020identifying, chen2019understanding}.
Points that reduce (expected) holdout loss are also selected for other applications \citep{ killamsetty2020glister, ren2018learning}, although using much more computation. 

\paragraph{Variance Reduction Methods.}
Online batch selection is also used to reduce the variance of the gradient estimator computed by SGD \citep{katharopoulos2018not, katharopoulos2017biased,johnson2018training,alain2015variance}. 
Such methods typically use importance sampling---points with high (approximate) gradient norm are sampled with high probability but then down-weighted in the gradient calculation to de-bias the gradient estimate. 
Without de-biasing, methods like RHO-LOSS also create selection bias. However, bias can improve test performance, both in theory and practice \citep{farquhar2020statistical, kawaguchi2020ordered}.

\section{Discussion}

To reduce excessive training times, we introduce a theoretically grounded selection function that enables substantial speedups on clean data and even larger speedups on noisy and web-scraped data. 
By illuminating three properties of optimal selection, we hope to motivate new directions in batch selection. 
However, our selection function should be combined with methods in \Cref{rholoss:sec:related} for cheap and fast selection with maximal speedups.

}

{
	\begin{savequote}[75mm]
		If you understand, things are just as they are. If you do not understand, things are just as they are.		
	\end{savequote}
	\chapter{\fimtitle}
	\label{chapter:fim}
	\graphicspath{{./fim/figs/}}

\newcommand{\Yhat}{\hat{\Y}}
\newcommand{\yhat}{\hat{\y}}

\newcommand{\Ypred}{\Y}
\newcommand{\ypred}{\y}
\newcommand{\Ytrue}{\Y}
\newcommand{\ytrue}{\y}

\newcommand{\logits}{{\hat \z}}

\newcommand{\xacq}{{\x^\text{acq}}}
\newcommand{\Yacq}{{\Y^\text{acq}}}
\newcommand{\yacq}{{\y^\text{acq}}}
\newcommand{\xacqs}{{\{\x^\text{acq}_i\}}}
\newcommand{\yacqs}{{\{\y^\text{acq}_i\}}}
\newcommand{\Yacqs}{{\{\Y^\text{acq}_i\}}}

\newcommand{\yacqsstar}{{\{\y^\text{acq,*}_i\}}}

\newcommand{\xs}{{\{\x_i\}}}
\newcommand{\ys}{{\{\y_i\}}}
\newcommand{\Ys}{{\{\Y_i\}}}

\newcommand{\Xeval}{{X^\text{eval}}}
\newcommand{\xeval}{{\x^\text{eval}}}
\newcommand{\Yeval}{{\Y^\text{eval}}}
\newcommand{\yeval}{{\y^\text{eval}}}

\newcommand{\xevals}{{\{\x^\text{eval}_i\}}}
\newcommand{\yevals}{{\{\y^\text{eval}_i\}}}
\newcommand{\Yevals}{{\{\Y^\text{eval}_i\}}}

\newcommand{\xtrain}{{\x^\text{train}}}
\newcommand{\ytrain}{{\y^\text{train}}}
\newcommand{\Ytrain}{{\Y^\text{train}}}

\newcommand{\xtrains}{{\{\x^\text{train}_i\}}}

\newcommand{\w}{\omega}
\newcommand{\W}{\Omega}
\newcommand{\wstar}{\w^*}

\newcommand{\realnum}{\mathbb{R}}

\newcommand{\verteq}{\rotatebox{90}{$\,=$}}
\newcommand{\equalto}[2]{\underset{\displaystyle\overset{\mkern4mu\verteq}{#2}}{#1}}

\newcommand{\encoder}[1]{{\hat f(#1)}}
\newcommand{\realencoder}[1]{{f (#1)}}
\newcommand{\embedding}{\z}

\newcommand{\similarityMatrix}[2]{{S_{#1}[#2]}}
\newcommand{\HofJacobianData}[1]{{\hat{\opEntropy}'[#1]}}
\newcommand{\HofJacobianDataShort}{\hat{\opEntropy}'}
\label{fim:sec:introduction}

The topic of this chapter is the explicit connection between Bayesian active learning and active sampling, and the connection between these and other recent approaches in data subset selection: amongst them, BADGE \citep{ash2020deep}, BAIT \citep{ash2021gone}, PRISM\footnotemark \citep{kothawade2022prism}, SIMILAR\footnotemark[\value{footnote}] \citep{kothawade2021similar}, and GraNd \citep{paul2021deep}.
Specifically, we connect the acquisition functions used to select informative samples in these approaches to information-theoretic quantities (short: \emph{information quantities}) that are known from Bayesian optimal experiment design \citep{lindley1956measure, mackay1992information}. 

By examining how Fisher information and second-order posterior approximations (Gaussian approximations) can be used for estimating information quantities,
we develop a unifying perspective and relate these recent methods to information quantities used in Bayesian active learning (and introduced in previous chapters): for active learning, the expected information gain/(Batch)BALD, which we examined in \Cref{glue:background:BALD,chapter:batchbald}, the (joint) expected predictive information gain from \Cref{chapter:pobal}; and, for active sampling, the information gain, mentioned in \Cref{glue:background:active_sampling}, and the (joint) predictive information gain from \Cref{chapter:rholoss}.

These connections point us towards possible failure modes of above methods and potential extensions in principled ways.
Reciprocaly, they also point towards new extensions of what we have introduced in the previous chapters---very exciting!

We examine well-known approximations that lead to last-layer approaches, find a potential estimation bias when using similarity matrices (kernels) in active learning, compare trace and log determinant approximations in regard to batch acquisition pathologies, and trade off weight- and prediction-space methods in principle. 

\subsection*{Limitations}

It is important to note the limitations of this chapter. We string together different research areas and disparate literature, while focusing on providing an information-theoretic perspective. 
For a perspective that is focused on kernel methods and Gaussian Process approximations of neural networks, \citet{holzmuller2022framework} extensively covers somve of the mentioned works in active learning above in great detail, which provides a useful addition to this chapter.

\paragraph{Hierarchy of Approximations.}
Although our results employ a hierarchy of approximations, we do not examine the error terms in detail. This is in line with how these approximations are used in deep learning, where the approximations often only provide motivation for useful mechanisms.
However, we try to identify where these approximations might break, enumerate their limitations, and raise several (empirical) research questions for the future.

\paragraph{Log Loss.} While many active learning and active sampling methods are motivated independently of the underlying loss, we will remain focused on log losses, such as the common cross-entropy loss or squared error loss (Gaussian error), as these log losses can be viewed through an information-theoretic and probabilistic lens.

\subsection*{Chapter Structure} %
In \S\ref{fim:sec:laplace_approximation}, we look at second-order posterior approximations (Gaussian approximations), which we use to revisit Fisher information, its properties, special cases, and approximations in \S\ref{fim:sec:fisher_information}.
Our contribution here is to summarize results and provide a consistent notation that simplifies reasoning about information quantities, observed information, and Fisher information. 

In \S\ref{fim:sec:iq_approximations}, we approximate the information quantities mentioned above using observed information and Fisher information.
We provide a comprehensive overview to understand the differences and similarities and make it easier to spot applications of these approximations in the literature.
We pay special attention to the limitations: 
for example, we will see that some approximations that use the trace of the Fisher information do not take redundancies between samples into account.
They exhibit the same pathologies as other methods that, in essence, score points individually (\Cref{chapter:stochacq}).
In \S\ref{fim:sec:similarity_matrices}, we expand our approach to approximations that use similarity matrices of log-loss gradients. %
Our contributions are a comprehensive overview of the approximations and the connection to similarity matrices.

In \S\ref{fim:sec:unification}, we show that (Batch-)BALD and EPIG on the one hand; and BADGE, BAIT, PRISM\footnotemark[\value{footnote}], and SIMILAR\footnotemark[\value{footnote}]\footnotetext{using log determinant objectives} on the other hand can be seen as optimizing the same objectives.
The difference is that (Batch-)BALD ( \Cref{chapter:batchbald}; \citet{houlsby2011bayesian}) and EPIG (\Cref{chapter:pobal}) operate in prediction space, while Fisher information-based methods operate in weight space:
we show that an approximation of EPIG, a transductive active learning objective, using Fisher information, matches the BAIT objective \citep{ash2021gone}. Similarly, we show how BADGE \citep{ash2020deep} approximates the EIG, using the connection to similarity matrices.
Finally, we find that \emph{submodularity}-based approaches \citep{iyer2021submodular} such as SIMILAR \citep{kothawade2021similar} and PRISM \citep{kothawade2022prism}, which report their best results using the log determinant of similarity matrices, approximate information quantities when they perform best. 
We also show that gradient-length-based methods like EGL \citep{settles2007multiple} and GraNd \citep{paul2021deep} can be connected to information quantities.

\section{Setting}
\label{fim:sec:background}

This section some additional notation, concepts, and probabilistic model that we use in this chapter. 

\paragraph{Transductive Acquisition Functions.} When an acquisition function in data subset selection uses (additional) data $\Deval$, unlabeled or labeled, to guide acquisitions, we refer to the objective as a \emph{transductive} objective \citep{yu2006active, wang2021beyond}.

\paragraph{Active Learning.} To increase label efficiency, instead of labeling data indiscriminately, active learning iteratively selects and acquires labels for the \emph{most informative} unlabeled data from a \emph{pool set} $\Dpool$ according to some \emph{acquisition function}. An acquisition function scores the informativeness of an unlabeled candidate sample $\xacq$, and the sample that maximizes this score is selected for labeling. After each acquisition step, the model is retrained to take the newly labeled data into account. Labels can be acquired individually or in batches (\emph{batch acquisition}, see below). The \emph{expected information gain (EIG)}
\begin{align}
    \MIof{\W; \Yacq \given \xacq} \tag{EIG/BALD}
\end{align} and \emph{(joint) expected predictive information gain (JEPIG and EPIG, respectively)}
\begin{align}
    \MIof{\Yevals ; \Yacq \given \xevals, \xacq}, \label{fim:eq:background_jepig} \tag{JEPIG} \\
    \MIof{\Yeval ; \Yacq \given \Xeval, \xacq} \label{fim:eq:background_epig} \tag{EPIG}
\end{align}
are examples of such acquisition functions.
EPIG and JEPIG are transductive acquisition functions as they depend on $\Xeval, \xevals,$ respectively.

\paragraph{Active Sampling.} To increase training efficiency, instead of training with all samples, active sampling (sometimes also called \emph{data pruning}) \citep{paul2021deep} selects the most informative sample $(\xacq, \yacq)$ from the training set to train on next. This can be done statically before training the model, in which case this is also referred to as core-set selection, or dynamically, in which case it is also referred to as curriculum learning. The \emph{information gain (IG)}\andreas{add citations for AS?!}
\begin{align}
    \MIof{\Omega; \yacq \given \xacq}, \tag{IG}
\end{align}
and \emph{(joint) predictive information gain (JPIG or PIG, respectively)} \begin{align}
    &\MIof{\yevals; \yacq \given \xevals, \xacq} \tag{JPIG}, \\
    &\MIof{\Yeval; \yacq \given \Xeval, \xacq} \quad ( = \simpleE{\pdataof{\xeval, \yeval}}{\MIof{\yeval; \yacq \given \xeval, \xacq}}) \tag{PIG}
\end{align}
are examples of such acquisition functions.
PIG and JPIG are transductive acquisition functions.

\paragraph{Submodular Acquisition Functions.} Choosing the subset $\xacqs$ naively is intractable due to the exponential number of possible acquisition batches. Instead of maximizing the acquisition function on all possible batches, we can often use submodularity \citep{nemhauser1978analysis}. 
A set function $f$ is submodular when:
\begin{align}
    f(A \cup B) \leq f(A) + f(B) - f(A \cap B) \tag{submodular}.
\end{align}
An acquisition batch $\xacqs$ can be constructed greedily by selecting the samples that increase the acquisition function the most one-by-one. This greedy algorithm is guaranteed to find a $1-\tfrac{1}{e}$-optimal acquisition batch for monotone submodular acquisition functions. %

Although the EIG is (monotone) submodular, leading to efficient batch acquisition (\Cref{chapter:batchbald}), the other information quantities (IG, EPIG, JEPIG, PIG) are usually not submodular. %
We examine the details of this and compare to the relevant literature in \S\ref{fim:sec:unification}.

\paragraph{Taxonomy of Information Quantities.} %
\Cref{fim:table:taxonomy_info_quantities} shows the information quantities along three dimensions: active learning vs active sampling, non-transductive vs transductive, and taking the expectation vs the joint over evaluation samples for transductive information quantities.

\begin{table}[t!]
    \caption{\emph{Taxonomy of Information Quantities for Data Subset Selection.} In general, information quantities can be split into ones for active sampling or active learning, into non-transductive and transductive ones, and  in the transductive case, into taking an expectation or the joint over (additional) evaluation samples. Here, we show the information quantities for individual acquisition. For batch acquisition, $\Yacqs, \yacqs, \xacqs$ can be substituted.}
    \label{fim:table:taxonomy_info_quantities}
    \centering
    \renewcommand{\arraystretch}{1.3}
    \resizebox{\linewidth}{!}{%
    \begin{tabular}{llrlrl}
    \toprule
                                  &             & \multicolumn{2}{c}{Active Learning} & \multicolumn{2}{c}{Active Sampling} \\
    \midrule
    \multicolumn{2}{l}{Non-Transductive} & EIG/BALD & $\MIof{\W; \Yacq \given \xacq}$ & IG & $\MIof{\Omega; \yacq \given \xacq}$ \\
    \multirow{2}{*}{\shortstack{Transductive \\(using $\Deval$)}} & Expectation  & EPIG & $\simpleE{\pdataof{\xeval}} \MIof{\Yeval ; \Yacq \given \xeval, \xacq}$ & PIG & $\simpleE{\pdataof{\yeval,\xeval}}\MIof{\yeval; \yacq \given \xeval, \xacq}$ \\
                                  & Joint       & JEPIG & $ \MIof{\Yevals ; \Yacq \given \xevals, \xacq} $ & JPIG &$\MIof{\yevals; \yacq \given \xevals, \xacq}$ \\
    \bottomrule  
    \end{tabular}}
\end{table}

\section{Second-Order Posterior Approximation}
\label{fim:sec:laplace_approximation}

Laplace approximations are a standard tool in Bayesian statistics and machine learning \citep{daxberger2021laplace, immer2021improving}. In this section, we review the Laplace approximation and introduce it as a special case of a more flexible second-order posterior approximation, a \emph{Gaussian approximation}. It is central to approximating information quantities using observed information, defined in this section, and Fisher information, defined in \S\ref{fim:sec:fisher_information}.

Our goal is to approximate the posterior $\pof{\w \given \Dany, \Dtrain}$ using a (multivariate) Gaussian distribution, where $\Dany = \{(\x_i, \y_i)\}_{i=1}^N$ are additional (new) samples, and we start with $\pof{\w \given \Dtrain}$ as the ``prior'' distribution---we will drop $\Dtrain$ and use $\pof{\w}$ when possible, to shorten the notation.

To begin, we complete the square of a second-order Taylor approximation around the log-parameter likelihood for a fixed $\wstar$:
\begin{align}
    &\log \pof{\w} \notag \\
    &\quad \approx \log \pof{\wstar} + \nabla_\w [\log \pof{\wstar}] (\w - \wstar) + \frac{1}{2} (\w - \wstar)^T \nabla_\w^2 [\log \pof{\wstar}] (\w - \wstar) \\
    &\quad = \frac{1}{2} (\w - (\wstar - \nabla_\w^2[\log \pof{\wstar}]^{-1} \nabla_\w [\log \pof{\wstar}])^T \notag \\
    &\quad \quad \nabla_\w^2 [\log \pof{\wstar}] \notag \\
    &\quad \quad (\w - (\wstar - \nabla_\w^2 [\log \pof{\wstar}]^{-1} \nabla_\w [\log \pof{\wstar}])) \notag \\
    &\quad \quad + \ldots.
\end{align}
Importantly, we can express this more concisely by extending the notation of \(\Hof{\cdot}\) to its derivatives:
\begin{importantresult}
\begin{notation}
    We write $\HofJacobian{\cdot}$ for the Jacobian and $\HofHessian{\cdot}$ for the Hessian of $\Hof{\cdot}$:
    \begin{align}
        \HofJacobian{\cdot} &\defeq -\nabla_\w \log \pof{\cdot}, \\
        \HofHessian{\cdot} &\defeq -\nabla_\w^2 \log \pof{\cdot}.
    \end{align}
\end{notation}
\end{importantresult}
This notation will be helpful throughout this chapter, as both observed information and Fisher information can be expressed in terms of the Hessian of the negative log-parameter likelihood. The Jacobian of the entropy is also known as score function.

Then, we can write:
\begin{align}
    \Hof{\w} &\approx \Hof{\wstar} + \HofJacobian{\wstar} (\w - \wstar) + \frac{1}{2} (\w - \wstar)^T \, \HofHessian{\wstar} \, (\w - \wstar) \\
    &= \frac{1}{2} (\w - (\wstar - \HofHessian{\wstar}^{-1} \HofJacobian{\wstar})^T \, \HofHessian{\wstar} \, (\w - (\wstar - \HofHessian{\wstar}^{-1} \HofJacobian{\wstar})) \notag \\
    &\quad + \ldots.
\end{align}
Comparing this to the information content of a multivariate Gaussian distribution:
\begin{align}
    \Hof{\normaldistpdf{w}{\mu}{\Sigma}} = \frac{1}{2} (\w - \mu)^T \, \Sigma^{-1} \, (\w - \mu) + \ldots,
\end{align}
we obtain the Gaussian approximation, which we will apply throughout this chapter:
\begin{proposition}
    \label{fim:prop:param_approximation}
    The \emph{Gaussian approximation} of the distribution $\pof{\w}$ of $\W$ around some $\wstar$ is given by:
    \begin{align}
        \W \overset{\approx}{\sim} \normaldist{\wstar - \HofHessian{\wstar}^{-1} \HofJacobian{\wstar}}{\HofHessian{\wstar}^{-1}},
    \end{align}   
    where $\HofHessian{\wstar}$ must be positive-definite.
    If $\wstar$ is also a (global) minimizer of $\Hof{\w}$ (that is, $\HofJacobian{\wstar} = 0$), we obtain the \emph{Laplace approximation}:
    \begin{align}
        \W \overset{\approx}{\sim} \normaldist{\wstar}{\HofHessian{\wstar}^{-1}}.
    \end{align}
\end{proposition}
\paragraph{Approximation Quality.} However, this approximation can be arbitrarily bad depending on $\pof{\w}$ and $\wstar$. Given enough data, it is often argued that $\pof{\w}$ will concentrate around the maximum a posteriori (MAP) estimate, giving rise to the Laplace approximation. In statistics, the Bernstein-von Mises theorem is often used to motivate this, but insufficient data to reach concentration of parameters and multimodality in over-parameterized models \citep{long2021multimodal} can be an issue for deep active learning and active sampling.

\paragraph{Flat Minimum Intuition.} A positive definite Hessian implies that the information content (point-wise entropy) is convex around $\wstar$ and, equivalently, that the (log) posterior is concave around $\wstar$. The latter provides an intuition for the Gaussian approximation:
the Hessian measures curvature, and the ``flatter'' the Hessian, e.g., the smaller the largest eigenvalue or the smaller the determinant, the less the loss changes when $\wstar$ is perturbed. This leads to the search for flat minima as a way to improve generalization \citep{hinton1993keeping,hochreiter1994simplifying,smith2017bayesian}.

\begin{notation}
To further shorten the notation, we write $\HofHessian{\Dany \given \wstar}$ instead of $\HofHessian{\ys \given \xs, \wstar}$.
\end{notation}

\paragraph{Posterior Approximation of $\Omega \given \Dany$.} %
While the Laplace approximation is centered on a (global) minimizer, the Gaussian approximation can be used for a (potentially low-quality) posterior approximation in general.
We can expand $\Hof{\wstar \given \Dany}$ using Bayes' theorem and the additivity of the logarithm. That is, we have:
\begin{align}
    \Hof{\wstar \given \Dany} = \Hof{\Dany \given \wstar} + \Hof{\wstar} - \Hof{\Dany},
\end{align}
and then, as $\Hof{\Dany}$ is independent of $\w$:
\begin{align}
    \HofJacobian{\wstar \given \Dany} &= \HofJacobian{\Dany \given \wstar} + \HofJacobian{\wstar} + 0
    = \HofJacobian{\Dany \given \wstar} + \HofJacobian{\wstar}, \\
    \HofHessian{\wstar \given \Dany} &= \HofHessian{\Dany \given \wstar} + \HofHessian{\wstar}.
\end{align}

\begin{proposition}
    \label{fim:prop:observed_information_additive}
    The \emph{observed information} $\HofHessian{\ys \given \xs, \wstar}$ is additive:
    \begin{align}
        \HofHessian{\ys \given \xs, \wstar} &= \sum_i \HofHessian{\y_i \given \x_i, \wstar} 
        = \sum_i - \nabla^2_{\w} \log \pof{\y_i \given \x_i, \wstar}.
    \end{align}
\end{proposition}
Note that the observed information has the opposite sign compared to other works because it simplifies the exposition.

\paragraph{Uninformative Prior.} For a Gaussian prior $\pof{\w} \sim \normaldist{\mu}{\Sigma}$, we have $\HofHessian{\wstar} = \Sigma^{-1}$ and
\begin{math}
    \HofHessian{\wstar \given \Dany} = \HofHessian{\Dany \given \wstar} + \Sigma^{-1}.
\end{math}
For an uninformative prior with ``infinite prior variance'' $\Sigma^{-1} \to 0$, we have $\HofHessian{\wstar} = 0$, and $\HofHessian{\wstar \given \Dany} = \HofHessian{\Dany \given \wstar}$.

\begin{importantresult}
\begin{proposition}
    \label{fim:prop:entropy_appox}
    The entropy of the second-order approximation of $\pof{\w}$ around $\wstar$ is 
    \begin{align}
        \Hof{\W} \approx -\tfrac{1}{2}\log \det \HofHessian{\wstar} + C_k,
    \end{align}
    where $C_k = \tfrac k 2 \log 2\pi e$ is a constant (independent of $\Dany$ and $\wstar$) and $k$ is the number of dimensions of $\w$.
\end{proposition}
\end{importantresult}
While \Cref{fim:prop:entropy_appox} is straightforward, it is the main result for this section as it will allow us to approximate all the mentioned information quantities in \S\ref{fim:sec:iq_approximations} and \S\ref{fim:sec:similarity_matrices}.
\section{Fisher Information}
\label{fim:sec:fisher_information}

Fisher information plays a central role in the approximations of information quantities because, unlike the observed information, it is always positive semi-definite.
We use Fisher information to unify various acquisition functions in \S\ref{fim:sec:unification}.
The following section revisits Fisher information, its properties, special cases, and common approximations.
All proofs are given in \S\ref{fim:appsec:fisher_information}.

In particular, we look at two special cases with more favorable properties: following \citet{kunstner2019limitations}, when we can write our model as $\pof{\y \given \logits=\encoder{\x; \w}}$, where $\encoder{\x; \w}$ are the logits, and $\pof{\y \given \logits}$ is a distribution from the exponential family, Fisher information is independent of $\y$, which has useful consequences as we shall see; and following \citet{chaudhuri2015convergence}, when we have a \emph{Generalized Linear Model (GLM)}, observed information also is independent of $\y$.
The results for the GLM are often applied as an approximation known as \emph{Generalized Gauss-Newton approximation (GGN)}.
Together with numerical approximations, such as a diagonal approximation or low-rank factorizations, observed information and Fisher information can then be efficiently approximated for large deep neural networks \citep{daxberger2021laplace}.

\begin{definition}
    \label{fim:def:fisher_info}
    The \emph{Fisher information} $\FisherInfo{\Y \given \x, \wstar}$ is the expectation over observed information using the model's own predictions $\pof{\y\given\x,\wstar}$ for a given $\x$ at $\wstar$:
    \begin{align}
        \FisherInfo{\Y \given \x, \wstar} = \E{\pof{\y \given \x, \wstar}}{\HofHessian{\y \given \x, \wstar}}.
    \end{align}
\end{definition}
This notation of the Fisher information is consistent with the notation for information quantities we have used far (\Cref{chapter:practical_it_notation,glue:background:information_theory}), but extended to the observed information: the Fisher is but an expectation over the observed information, and the observed information is the Hessian of the negative log-likelihood.

\begin{restatable}{proposition}{fimadditive}
    Like observed information, Fisher information is additive:
    \begin{align}
        \FisherInfo{\Ys \given \xs, \wstar} = \sum_i \FisherInfo{\{Y_i\} \given x_i, \wstar}.
    \end{align}
\end{restatable}
There are two other equivalent definitions of Fisher information:
\begin{restatable}{proposition}{fisherinformationequivalences}
    \label{fim:prop:fisher_information_equivalences}
    Fisher information is equivalent to:
    \begin{align}
        \FisherInfo{Y \given x, \wstar} &= \E{\pof{y \given x, \wstar}}{\HofJacobian{y \given x, \wstar}^T \, \HofJacobian{y \given x, \wstar}} = \implicitCov{\HofJacobian{Y \given x, \wstar}}. 
    \end{align}
\end{restatable}

\paragraph{Special Case: Exponential Family.} %
\citet{kunstner2019limitations} show in their appendix that if we split a discriminative model into prelogits $\encoder{\x; \w}$ and a predictor $\pof{\y \given \logits=\encoder{\x; \w}}$, Fisher information does not depend on $\y$ when $\pof{\y \given \logits}$ is a distribution from an exponential family (independent of $\w$). This covers a normal distribution for regression parameterized by mean and variance predictions or a categorical distribution via the softmax function.
The following statements and proofs follow \citet{kunstner2019limitations}:
\begin{restatable}{proposition}{fimexpfamily}
    The Fisher information $\FisherInfo{Y \given \x, \wstar}$ for a model $\pof{\y \given \logits=\encoder{\x; \wstar}}$ is equivalent to:
    \begin{align}
        \FisherInfo{Y \given x, \wstar} = \nabla_\w \encoder{x; \wstar}^T \E{\pof{y \given x, \wstar}}{\nabla_\logits^2 \Hof{y \given \logits=\encoder{x ; \wstar}}} \nabla_\w \encoder{x; \wstar},
    \end{align}
    where $\nabla_\logits^2 \Hof{\y \given \logits=\encoder{\x ; \wstar}}$ is short for $\left. \nabla_\logits^2 \Hof{\y \given \logits}\right |_{\logits=\encoder{\x ; \wstar}}$.
\end{restatable}
\yarin{I think sending the draft to Philip Henning will be good for references about some early work on those topics}

\begin{proposition}
    \label{fim:prop:special_case_exp_family_fim}
    The Fisher information $\FisherInfo{Y \given \x, \wstar}$ of a model of the form $\pof{\y \given \logits=\encoder{\x; \wstar}}$  is independent of $\y$, where $\pof{\y \given \logits}$ is a distribution from an exponential family, i.e., $\log \pof{\y \given \logits} = \logits^T T(\y) - A(\logits) + \log h(\y)$:
    \begin{align}
        \FisherInfo{Y \given x, \wstar} = \nabla_\w \encoder{x; \wstar}^T \, {\nabla_\logits^2 A(\logits=\encoder{x ; \wstar})} \, \nabla_\w \encoder{x; \wstar}.
    \end{align}
\end{proposition}
\emph{It is crucial that the exponential distribution not depend on $\w$.}
This simplifies computing Fisher information: no expectation over $\y$s is needed anymore. The full outer product may not be needed explicitly either.

As examples, we will consider two common parameteric distributions from the exponential family:

\paragraph{Gaussian Distribution.} When $\pof{\y \given \logits} = \normaldistpdf{\y}{\logits}{1}$, we have $\HofHessian{\y \given \logits} = 1$ for all $\y, \logits$, and thus
\begin{align}
    \FisherInfo{Y \given x, \wstar} = \nabla_\w \encoder{x; \wstar}^T \, \nabla_\w \encoder{x; \wstar}.
\end{align}

\paragraph{Categorical Distribution.} When $\pof{\y \given \logits} = \softmax(\logits)_\y$, we have $\HofHessian{\y \given \logits} = \diag(\pi) - \pi\,\pi^T,$ with $\pi_\y=\pof{\y \given \logits}$, and thus:
\begin{align}
    \FisherInfo{Y \given x, \wstar} = \nabla_\w \encoder{x; \wstar}^T \, (\diag(\pi) - \pi\,\pi^T) \, \nabla_\w \encoder{x; \wstar}.
\end{align}

\paragraph{Special Case: Generalized Linear Models.} %
\citet{chaudhuri2015convergence} require that observed information is independent of $\y$, which we will also use later. This holds for Generalized Linear Models:
\begin{importantresult}
\begin{definition}
    A \emph{generalized linear model (GLM)} is a model $\pof{\y \given \logits=\encoder{\x ; \w}}$ such that $\log \pof{\y \given \logits} = \logits^T T(\y) - A(\logits) + \log h(\y)$ is a distribution of the exponential family, independent of $\w$, and $\encoder{\x ; \w} = \w^T \, \x$ is linear in the parameters $\w$. 
\end{definition}
\begin{restatable}{proposition}{glmhessian}
    \label{fim:prop:glm_hessian}
    The observed information $\HofHessian{\y \given \x, \wstar}$ of a GLM is independent of $\y$.
    \begin{align}
        \HofHessian{y \given x, \wstar} &= \nabla_\w \encoder{x ; \wstar}^T \, \nabla_\logits^2 \Hof{y \given \logits=\encoder{x ; \wstar}} \, \nabla_\w \encoder{x ; \wstar} \\
        &= \nabla_\w \encoder{x ; \wstar}^T \, \nabla_\logits^2 A(w^T x) \, \nabla_\w \encoder{x ; \wstar}.
    \end{align}
\end{restatable}
\begin{restatable}{proposition}{glmhessianfim}
    \label{fim:prop:glm_hessian_fim}
    For a model such that the observed information $\HofHessian{\y \given \x, \wstar}$ is independent of $\y$, we have:
    \begin{align}
        \FisherInfo{Y \given x, \wstar} = \HofHessian{y^*\given x, \wstar}
    \end{align}
    for any $\y^*$, and also trivially:
    \begin{align}
        \FisherInfo{Y \given x, \wstar} = \E{\pof{y \given x}}{\HofHessian{y \given x, \wstar}}.
    \end{align}
\end{restatable}
\end{importantresult}
Note that the expectation is over $\pof{\y \given \x}$ and not $\pof{\y \given \x, \wstar}$, and \begin{math}
        \E{\pof{\ys \given \xs}}{\HofHessian{\ys \given \xs, \wstar}} = \FisherInfo{\Ys \given \xs, \wstar}
\end{math} is additive then.
\begin{restatable}{proposition}{propkroneckerproductfisher}
    For a GLM, when $\encoder{\x ; \w}: \realnum^D \to \realnum^\numclasses$, where $\numclasses$ is the number of classes (outputs), $D$ is the number of input dimensions, $\w \in \realnum^{D \times \numclasses}$, and
    assuming the parameters are flattened into a single vector for the Jacobian, we have $\nabla_\w \encoder{\x ; \w} = \mathrm{Id}_\numclasses \otimes \x^T \in \realnum^{\numclasses \times (\numclasses \cdot D)}$, where $\otimes$ denotes the Kronecker product, and:
    \begin{align}
        \nabla_\w \encoder{x ; \wstar}^T \, \nabla_\logits^2 A(\w^T x) \, \nabla_\w \encoder{x ; \wstar} = \nabla_\logits^2 A(w^T x) \otimes x \, x^T.
    \end{align}
\end{restatable}
This property is useful for computing the Fisher information of a GLM in practice \citep{ash2021gone}.

\begin{importantresult_noparbox}
\paragraph{$\mathbf{\pof{\y \given \x, \wstar}}$ vs $\mathbf{\pof{\y \given \x}}$.} %
Having a GLM solves an important issue we will encounter in \S\ref{fim:sec:iq_approximations}: approximating the EIG requires taking an expectation over $\pof{\y \given \x}$ and not $\pof{\y \given \x, \wstar}$. One can approximate $\pof{\y \given \x} \approx \pof{\y \given \x, \wstar}$, which can be justified in the limit, but this is probably not a good approximation in the cases interesting for active learning and active sampling. With a GLM, this is not a problem.

\paragraph{Generalized Gauss-Newton Approximation.} %
In the case of an exponential family but not a GLM, the equality in \Cref{fim:prop:glm_hessian} is often used as an approximation for the observed information---we simply use the respective Fisher information as an approximation of observed information (via \Cref{fim:prop:special_case_exp_family_fim}):
\begin{align}
    \HofHessian{y \given x, \wstar} \approx \FisherInfo{Y \given x, \wstar} =  \nabla_\w \encoder{x ; \wstar}^T \, \nabla_\logits^2 A(w^T x) \, \nabla_\w \encoder{x ; \wstar}.
\end{align}
This is known as \emph{Generalized Gauss-Newton (GGN) approximation} \citep{kunstner2019limitations, immer2021improving}. This approximation has the advantage that it is always positive semi-definite unlike the true Hessian.

\paragraph{Last-Layer Approaches.} GLMs are often used in deep active learning \citep{ash2020deep,ash2021gone,kothawade2022prism,kothawade2021similar}. If we split the model into $\pof{\y \given \x, \w} = \pof{\y \given \embedding = \w^T \, \realencoder{\x}}$, where $\embedding = \realencoder{\x}$ are the embeddings and treat the encoder $\realencoder{\x}$ as fixed, we obtain a GLM based on the weights of the last layer, which uses the embeddings as input.
\end{importantresult_noparbox}

Armed with this knowledge, we can now derive approximations for the information quantities of interest using observed information and Fisher information and consider their properties. The GGN approximation and last-layer approaches feature heavily in the literature to make computing these approximations more tractable as they reduce computational requirements and memory usage.

\section{Approximating Information Quantities}
\label{fim:sec:iq_approximations}

We now derive approximations and proxy objectives for information quantities. We base them on observed information and Fisher information introduced in the previous sections.
These approximations help us connect the information quantities to existing the literature in non-Bayesian data subset selection in \S\ref{fim:sec:unification}. 

In particular, we derive approximations for EIG and EPIG as they show the qualitative differences between non-transductive and transductive objectives, and compare the approximations of the IG and EIG: importantly, there is no difference between the latter when we use a GLM or the GGN approximation. 
This covers two of the three dimensions in \Cref{fim:table:taxonomy_info_quantities}.
We examine JEPIG and the other quantities in the appendix in \S\ref{fim:appsec:approximate_iq}.
We find that the trace approximations of the EPIG and JEPIG objective matches, suggesting that using the trace approximations might be too loose an approximation to capture important qualities of EPIG (\Cref{chapter:pobal}). 
Additional derivations and details can also be found in \S\ref{fim:appsec:approximate_iq}.
All this leads to \Cref{fim:fig:fim_comparison}, which relates the different approximations to each other and shows that they follow simple patterns.

\subsection{Approximate Expected Information Gain}
The expected information gain is a popular acquisition function in Bayesian optimal experimental design \citep{lindley1956measure} and in active learning, where it is also known as BALD \citep{houlsby2011bayesian, gal2017deep}.

We can approximate the EIG $\MIof{\W; \Yacqs \given \xacqs}$ of acquisition candidates $\xacqs$ using Gaussian approximations:
\begin{align}
    &\MIof{\W; \Yacqs \given \xacqs} \notag \\
    &\quad = \Hof{\W} - \Hof{\W \given \Yacqs, \xacqs} \label{fim:eq:EIG_vs_Conditional_Entropy} \\
    &\quad = \Hof{\W} - \E{\pof{\yacqs \given \xacqs}}{\Hof{\W \given \yacqs, \xacqs}} \\
    &\quad \approx -\tfrac{1}{2}\log \det \HofHessian{\wstar} - \E{\pof{\yacqs \given \xacqs}}{- \tfrac{1}{2}\log \det \HofHessian{\w \given \yacqs, \xacqs}} \label{fim:eq:eig_approx_constant_cancellation} \\
    &\quad = \tfrac{1}{2}\E{\pof{\yacqs \given \xacqs}}{\log \det \left ( (\HofHessian{\yacqs \given \xacqs, \wstar} + \HofHessian{\wstar})\HofHessian{\wstar}^{-1} \right)}\\
    &\quad = \tfrac{1}{2}\E{\pof{\yacqs \given \xacqs}}{\log \det \left ( \HofHessian{\yacqs \given \xacqs, \wstar} \, \HofHessian{\wstar}^{-1} + Id \right )}. \label{fim:eq:eig_log_det}
\end{align}
using \Cref{fim:prop:entropy_appox} twice, where the constant $C_k$ cancels out in \cref{fim:eq:eig_approx_constant_cancellation} as we subtract two entropy terms.

\paragraph{Generalized Linear Model.} %
When we have a GLM, we can use \Cref{fim:prop:glm_hessian_fim} to obtain:
\begin{align}
    &\MIof{\W; \Yacqs \given \xacqs} \\
    &\quad \approx \ldots =  \tfrac{1}{2}\E{\pof{\yacqs \given \xacqs}}{\log \det \left ( \HofHessian{\yacqs \given \xacqs, \wstar} \, \HofHessian{\wstar}^{-1} + Id \right )} \\
    &\quad = \tfrac{1}{2} {\log \det \left ( \FisherInfo{\Yacqs \given \xacqs, \wstar} \, \HofHessian{\wstar}^{-1} + Id \right )}.
\intertext{We can upper-bound the log determinant and obtain:}
    &\quad \le \tfrac{1}{2} \tr \left ( \FisherInfo{\Yacqs \given \xacqs, \wstar} \, \HofHessian{\wstar}^{-1} \right ) \\
    &\quad = \tfrac{1}{2} \sum_i \tr \left ( \FisherInfo{\Yacqs \given \xacq_i, \wstar} \, \HofHessian{\wstar}^{-1} \right ).
\end{align}
where we have used the following inequality (proof in \S\ref{fim:appsec:eig_general_case}):
\begin{restatable}{lemma}{logdettrinequality}
    \label{fim:lemma:log_det_tr_inequality}
    For symmetric, positive semi-definite matrices $A$, we have (with equality iff $A=0$):
    \begin{align}
        \log \det (A + Id) \le \tr(A).
    \end{align}
\end{restatable}

\paragraph{General Case \&\ Exponential Family.} %
For the general case, we need to make a strong approximation: 
\begin{align}
    \pof{\yacqs \given \xacqs} \approx \pof{\yacqs \given \xacqs, \wstar},
\end{align}
which might hold for a mostly converged posterior but probably not in cases with little data.
This turns the approximation into an upper bound.
Alternatively, we could use the GGN approximation when we have an exponential family for the same result (but not an upper bound).
See \S\ref{fim:appsec:eig_general_case} for the derivation.

\begin{importantresult}
\begin{proposition}[EIG]
    \label{fim:prop:approximate_eig}
    The expected information gain can be approximately upper bounded via:
    \begin{align}
        &\MIof{\W; \Yacqs \given \xacqs, \Dtrain} \notag \\
        &\quad
        \overset{\approx}{\le}
        \tfrac{1}{2} {\log \det \left ( \sum_i \FisherInfo{\Yacq_i \given \xacq_i, \wstar} \, \HofHessian{\wstar \given \Dtrain}^{-1} + Id \right )}  \label{fim:eq:eig_log_det_approx} \\
        &\quad \le \tfrac{1}{2} \sum_i \tr \left ( \FisherInfo{\Yacq_i \given \xacq_i, \wstar} \HofHessian{\wstar \given \Dtrain}^{-1} \right ). \label{fim:eq:eig_trace_approx} 
    \intertext{Furthermore, we have the following proxy objective:}
        &\argmax_{\xacqs} \MIof{\W; \Yacqs \given \xacqs, \Dtrain} = \argmax_{\xacqs} -\Hof{\W \given \Yacqs, \xacqs, \Dtrain}, 
    \intertext{with} 
        &-\Hof{\W \given \Yacqs, \xacqs, \Dtrain} 
        \notag \\
        &\quad 
        \overset{\approx}{\le}
        \tfrac{1}{2} \log \det \left ( \sum_i \FisherInfo{\Yacq_i \given \xacq_i, \wstar} + \HofHessian{\wstar \given \Dtrain} \right ) - C_k. \label{fim:eq:approx_conditional_entropy}
    \end{align}
\end{proposition}
\end{importantresult}
The second statement follows from \Cref{fim:eq:EIG_vs_Conditional_Entropy}, since $\Hof{\W \given \Dtrain}$ is constant and provides a proxy objective when we are only interested in optimizing the EIG. In \S\ref{fim:sec:unification}, we connect it to the expected gradient length approach in active learning and show that an ablation in \citet{ash2021gone} examines the wrong objective.

\paragraph{Batch Acquisition Pathologies.} Importantly, this approximation of the EIG using the trace is additive, whereas the one using the log determinant is not. This means that the trace approximation ignores the dependencies between the samples and can only lead to naive top-k batch acquisition; see \Cref{chapterchapter:batchbald,chapter:stochacq} for details of the pathologies of top-k batch acquisition. 
\andreas{How well does the log determinant capture these dependencies?}

\andreas{can we verify this through experiments? is this a worthy/interesting research question?}

\subsection{Approximate Information Gain}

Following the same steps, we can also approximate the information gain, which is useful for active sampling: %
\begin{importantresult}
\begin{proposition}[IG]
    \label{fim:prop:approximate_ig}
    The \emph{information gain} $\MIof{\W ; \yacqs \given \xacqs, \Dtrain} = \Hof{\W \given \Dtrain} - \Hof{\W \given \yacqs, \xacqs, \Dtrain}$ can be approximately upper bounded via:
    \begin{align}
        &\MIof{\W ; \yacqs \given \xacqs, \Dtrain} &
        \notag \\
        &\quad \approx
        \tfrac{1}{2} {\log \det \left ( \HofHessian{\yacqs \given \xacqs, \wstar} \, \HofHessian{\wstar \given \Dtrain}^{-1} + Id \right )} \\
        &\quad \le \tfrac{1}{2} \sum_i \tr \left ( \HofHessian{\yacq_i \given \xacq_i, \wstar} \, \HofHessian{\wstar \given \Dtrain}^{-1} \right ).
\intertext{
    Furthermore, we have the following proxy objective:
}
        &\argmax_{\xacqs} \MIof{\W ; \yacqs \given \xacqs, \Dtrain} = \argmax_{\xacqs} -\Hof{\W \given \yacqs, \xacqs, \Dtrain} 
\intertext{with}
        &\medmath{-\Hof{\W \given \yacqs, \xacqs, \Dtrain}
        \approx
        \tfrac{1}{2} \log \det \left ( \HofHessian{\yacqs \given \xacqs, \wstar} + \HofHessian{\wstar \given \Dtrain} \right ) - C_k}. 
    \end{align}
\end{proposition}
\end{importantresult}

\paragraph{Comparison to EIG.} %
Importantly, when we have a GLM or use the GGN approximation, this approximation of the IG is equal to the one of the EIG. This tells us that active learning on a GLM with the EIG approximation will work as well as if we had access to the labels. Equivalently, active sampling via IG with the GGN approximation will not work better than the respective active learning approach.

\subsection{Approximate (Joint) Expected Predictive Information Gain}

In transductive active learning, we have access to an (empirical) distribution $\pdataof{\xeval}$, e.g., the pool set, and want to find the $\xacqs$ that maximize the \emph{expected predictive information gain} from \Cref{chapter:pobal}).
The approximations here will help us connect BAIT \citep{ash2021gone} to EPIG.
For simplicity, we consider the non-batch case here. The batch case can be handled analogously. The EPIG objective is defined as:
\begin{align}
    \argmax_\xacq \MIof{\Yeval; \Yacq \given \Xeval, \xacq} = \argmax_\xacq \simpleE{\pdataof{\xeval}} \MIof{\Yeval; \Yacq \given \xeval, \xacq},
\end{align}
We expand the objective as follows:
\begin{align}
    \MIof{\Yeval; \Yacq \given \Xeval, \xacq} &= \MIof{\W; \Yeval  \given \Xeval} - \MIof{\W ; \Yeval \given \Xeval, \Yacq, \xacq}, \label{fim:eq:epig_bald_decomposition}
\end{align}
where $\MIof{\W; \Yeval \given \Xeval}$ can be removed from the objective because it is independent of $\xacq$.
Thus, optimizing EPIG is equivalent to \emph{minimizing} $\MIof{\W ; \Yeval \given \Xeval, \Yacq, \xacq}$:
\begin{align}
    \argmax_\xacq \MIof{\Yeval; \Yacq \given \Xeval, \xacq} = 
    \argmin_\xacq \MIof{\W ; \Yeval \given \Xeval, \Yacq, \xacq}.
    \label{fim:eq:transductive_active_learning}
\end{align}
Following \Cref{fim:prop:approximate_eig}, this can be approximated by:
\begin{align}
    &\MIof{\W ; \Yeval \given \Xeval, \Yacq, \xacq} \notag \\
    &\quad \approx \tfrac{1}{2} \simpleE{\pof{\yeval, \yacq \given \xeval, \xacq} \, \pdataof{\xeval}} \\
    &\quad \quad \left \lbrack {\log \det
    \left (
    \HofHessian{\yeval \given \xeval, \wstar} \, ( \HofHessian{\yacq \given \xacq, \wstar} + \HofHessian{\wstar} )^{-1} + Id
    \right )} \right \rbrack.
\end{align}

\paragraph{Generalized Linear Model.} %
For a generalized linear model, we can drop the expectation and obtain:
\begin{align}
    &\MIof{\W ; \Yeval \given \Xeval, \Yacq, \xacq} \notag \\
    &\quad \approx 
    \medmath{\tfrac{1}{2} 
    \E{\pdataof{\xeval}} {\log \det
    \left (
    \FisherInfo{\Yeval \given \xeval, \wstar} \, ( \FisherInfo{\Yacq \given \xacq, \wstar} + \HofHessian{\wstar} )^{-1} + Id
    \right )}} \label{fim:eq:epig_trace_hook} \\
    &\quad \le \tfrac{1}{2} \medmath{\log \det
    \left (
    \E{\pdataof{\xeval}} {\FisherInfo{\Yeval \given \xeval, \wstar}} \, ( \FisherInfo{\Yeval \given \xacq, \wstar} + \HofHessian{\wstar} )^{-1} + Id
    \right )} \label{fim:eq:epig_deduction_jepig_hook} \\
    &\quad \le \tfrac{1}{2} {\tr
    \left (
    \E*{\pdataof{\xeval}} {\FisherInfo{\Yeval \given \xeval, \wstar}} \, ( \FisherInfo{\Yacq \given \xacq, \wstar} + \HofHessian{\wstar} )^{-1}
    \right )},
\end{align} 
where we have used the concavity of the log determinant and \Cref{fim:lemma:log_det_tr_inequality}.

\paragraph{General Case \&\ Exponential Family.} %
To our knowledge, there is no rigorous way to obtain a similar result in the general case as the Fisher information for an acquisition candidate now lies within an inverted term. Of course, the GGN approximation can be applied when we have an exponential family, which leads to the GLM result above as an approximation. See \S\ref{fim:appsec:epig_general_case} for more details.

\begin{importantresult}
\begin{proposition}[EPIG]
    \label{fim:prop:epig_fisher_approximation}
    For a generalized linear model (or with the GGN approximation), we have:
    \begin{align}
        \medmath{\argmax_\xacqs \MIof{\Yeval; \Yacqs \given \Xeval, \xacqs, \Dtrain} = 
        \argmin_\xacqs \MIof{\W ; \Yeval \given \Xeval, \Yacqs, \xacqs, \Dtrain}},
    \end{align}
    with
    \begin{align}
        &\MIof{\W ; \Yeval \given \Xeval, \Yacqs, \xacqs, \Dtrain} \notag \\
        &\quad 
        \approx
        \E{\pdataof{\xeval}} {\log \det
        \left (
        \FisherInfo{\Yeval \given \xeval, \wstar} \, ( \FisherInfo{\Yacqs \given \xacqs, \wstar} + \HofHessian{\wstar} )^{-1} + Id
        \right )} \\
        &\quad \le
        \medmath{\tfrac{1}{2} {\log \det
        \left (
        \E{\pdataof{\xeval}} {\FisherInfo{\Yeval \given \xeval, \wstar}} \, ( \FisherInfo{\Yacqs \given \xacqs, \wstar} + \HofHessian{\wstar \given \Dtrain} )^{-1} + Id
        \right )}} \label{fim:eq:epig_fisher_approximation_logdet} \\
        &\quad \le \tfrac{1}{2} {\tr
        \left (
        \E*{\pdataof{\xeval}} {\FisherInfo{\Yeval \given \xeval, \wstar}} \, ( \FisherInfo{\Yacqs \given \xacqs, \wstar} + \HofHessian{\wstar \given \Dtrain} )^{-1}
        \right )}. \label{fim:eq:epig_fisher_approximation_trace}
    \end{align}
\end{proposition}
\end{importantresult}
\paragraph{Batch Acquisition Pathologies.} Unlike for the EIG, the trace approximation of EPIG is not additive in $\xacqs$, and we cannot conclude that it suffers from batch acquisition pathologies like the trace approximation of the EIG.

\paragraph{Approximations for JEPIG, PIG and JPIG.} %
We can follow the same derivation for JEPIG: %
\begin{align}
    &\MIof{\W ; \Yevals \given \xevals, \Yacq, \xacq} \\
    &\quad \approx \medmath{\tfrac{1}{2} {\log \det
    \left (
        \E{\pof{\yevals, \yacq \given \xevals, \xacq}}{\HofHessian{\yevals \given \xevals, \wstar}} \, ( \HofHessian{\yacq \given \xacq, \wstar} + \HofHessian{\wstar} )^{-1} + Id
    \right )}}. \notag
\end{align}
Then, applying the steps after \cref{fim:eq:epig_deduction_jepig_hook}, we can devise similar approximations. PIG and JPIG follow the same pattern.
Details can be found in \S\ref{fim:appsubsec:pig} and \S\ref{fim:appsubsec:jepig_jpig}.
But how do all these approximations relate to each other?

\subsection{Comparison of the Different Information Quantity Approximations}
\begin{figure}[t!]
    
    \begin{subfigure}{\linewidth}
        \begin{whiteresult}
        \centering
        \includegraphics[width=\linewidth]{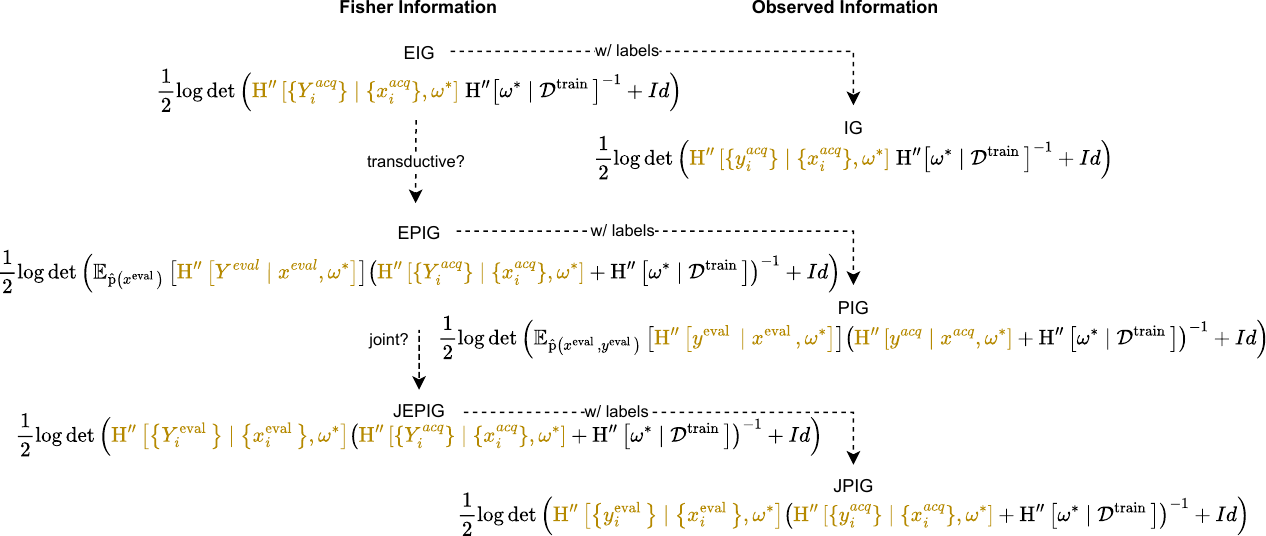}  
        \subcaption{$\log \det$ Proxy Objectives}     
        \end{whiteresult}
    \end{subfigure}
    \begin{subfigure}{\linewidth}
        \begin{whiteresult}
        \centering
        \includegraphics[width=\linewidth]{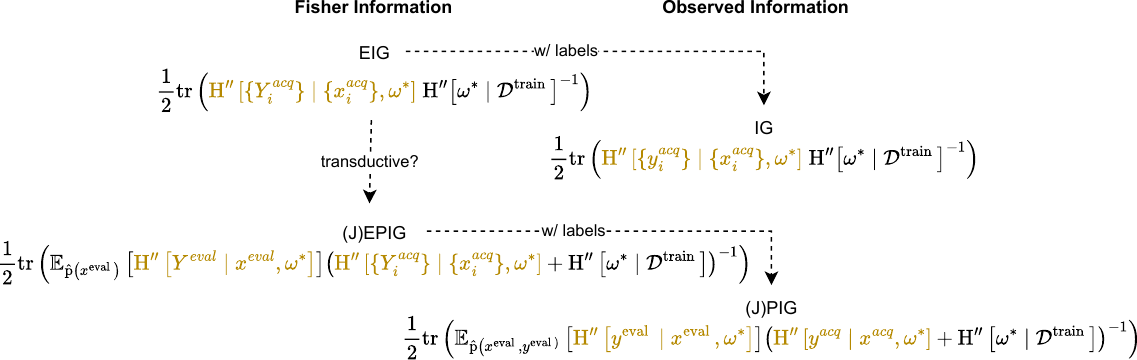}       
        \subcaption{Matrix Trace Proxy Objectives}
        \end{whiteresult}
    \end{subfigure}

    \caption{\emph{Comparison of the Approximations/Proxy Objectives.}
    The difference between active learning and active sampling objectives is in using the Fisher information, which is label independent, or the observed information, which uses label information. JEPIG and EPIG have equivalent proxy objectives when using the matrix trace; see \S\ref{fim:appsubsec:jepig_jpig}.
    The notation makes it obvious that in the GLM case, or when the GGN approximation is used, active learning and active sampling approximations match as $\HofHessian{\y \given \x, \wstar} =\; {(\text{resp.} \approx)} \HofHessian{\Y \given \x, \wstar}.$
    }
    \label{fim:fig:fim_comparison}
\end{figure}

\Cref{fim:fig:fim_comparison} compares the different information quantity approximations for both the log-determinant and trace approximations.
We empirically compare the approximations with prediction-space methods in \S\ref{fim:appsec:empirical_comparison}.
Importantly, the trace approximations of (E)PIG match those of J(E)PIG up to a constant factor (unlike the log-determinant approximations); see \S\ref{fim:appsubsec:jepig_jpig} for details.

In \Cref{chapter:fim}, we argued that JEPIG converges to BALD in the data limit of the evaluation set---when there are no outliers in the pool set--- while EPIG does not.
The trace approximation is too strong to preserve this difference. Does this difference matter in practice? We leave this for future work.

Crucially, for GLMs or when using the GGN approximation, the respective active learning and active sampling objectives (EIG and IG, etc.) are equivalent as Fisher information and observed information are the same. In contrast, in the general case, the approximations for EPIG and JEPIG do not have a principled derivation. 

\section{Similarity Matrices and One-Sample Approximations}
\label{fim:sec:similarity_matrices}

Many data subset selection methods \citep{iyer2021submodular,kothawade2022prism,kothawade2021similar, ash2020deep} use similarity matrices of the loss Jacobians $\HofJacobian{\hat{\y} \given \x}$ (gradient kernels), where $\hat{\y}$ is usually a hypothesized pseudo-label: often the $\argmax$ prediction of the model for $\x$.
Here, we connect such similarity matrices to the Fisher information and the approximations of information quantities from \S\ref{fim:sec:iq_approximations}. The proofs are given in \S\ref{fim:appsec:similarity_matrices}. Together with \S\ref{fim:sec:iq_approximations}, this section provides a unified framework for understanding the approximations of information quantities using Fisher information and lays the foundation for the next section, which will connect the cited works in \S\ref{fim:sec:introduction} to the approximations of information quantities.

\paragraph{Connection to Fisher Information.} %
Crucially, given $\Dany = \{ (\y_i,\x_i)_i \}$, if we let
\begin{align}
    \HofJacobianData{\Dany \given \wstar} \defeq \begin{pmatrix}
    \vdots \\ \HofJacobian{y_i \given x_i, \wstar} \\ \vdots
    \end{pmatrix}
\end{align}
be a ``data matrix'' of the Jacobians,
then $\HofJacobianData{\Dany \given \wstar} \HofJacobianData{\Dany \given \wstar}^T$ gives the similarity matrix $\similarityMatrix{}{\Dany \given \wstar}$ using the Euclidean inner product:
\begin{align}
    \similarityMatrix{}{\Dany \given \w}_{ij} &\defeq \langle \HofJacobian{\y_i \given \x_i, \wstar}, \HofJacobian{\y_j \given \x_j, \wstar} \rangle = \HofJacobianData{\Dany \given \wstar} \HofJacobianData{\Dany \given \wstar}^T.
\end{align}
Sampling $\ys \sim \pof{\ys \given \xs, \wstar}$, the ``flipped'' product $\HofJacobianData{\Dany \given \wstar}^T \HofJacobianData{\Dany \given \wstar}$ yields a \emph{one-sample estimate} of the Fisher information $\FisherInfo{\Ys \given \xs, \wstar}$:
\begin{align}
    \FisherInfo{\Ys \given \xs, \wstar} &= \sum_i \FisherInfo{Y_i \given x_i, \wstar} = \simpleE{\pof{\ys\given \xs,\wstar}}{\sum_i \HofJacobian{y_i \given x_i, \wstar}^T \HofJacobian{y_i \given x_i, \wstar}} \\
    &=\simpleE{\pof{\ys\given \xs,\wstar}}{ \HofJacobianData{\Dany \given \wstar}^T \HofJacobianData{\Dany \given \wstar} }. 
\end{align}
\paragraph{Hard Pseudo-Labels.} Importantly, using the $\argmax$ class for $\y_i$, we only obtain a biased estimate \citep[\S{B}]{kunstner2019limitations}.

\paragraph{Connection to the Expected Information Gain.} %
When we define an inner product $\langle \cdot, \cdot \rangle_{\HofHessian{\wstar \given \Dtrain}}$ using the Hessian, we can connect the similarity matrix, which uses this inner product:
\begin{align}
    \similarityMatrix{\HofHessian{\wstar \given \Dtrain}}{ \Dany \given \wstar} &\defeq \HofJacobianData{\Dany \given \wstar}\HofHessian{\wstar \given \Dtrain}^{-1} \HofJacobianData{\Dany \given \wstar}^T,
\end{align}
to our information gain approximations.

Specifically, we apply the matrix-determinant lemma $\det (A B + M) = \det M \det (Id + B M^{-1} A)$ to obtain:
\begin{importantresult}
\begin{restatable}{proposition}{propsimilaritybasedIG}
    \label{fim:prop:similarity_based_IG}
    Given $\Dtrain$, $\xacqs$ and (sampled) $\yacqs$, we have for the EIG:
    \begin{align}
        \MIof{\W; \Yacqs \given \xacqs, \Dtrain} &\overset{\approx}{\le}
        \tfrac{1}{2} {\log \det \left ( \similarityMatrix{\HofHessian{\wstar \given \Dtrain}}{ \Dacq \given \wstar}  + Id \right )} \label{fim:eq:similarity_based_IG} \\
        &\le \tfrac{1}{2} {\tr \similarityMatrix{\HofHessian{\wstar \given \Dtrain}}{ \Dacq \given \wstar}}
    \end{align}
\end{restatable}

\begin{restatable}{proposition}{similaruninformativeposterior}
    \label{fim:prop:similar_uninformative_posterior}
    Assuming an uninformative posterior $\HofHessian{\wstar \given \Dtrain} = \lambda Id$ for $\lambda \to 0$, and given $\Dtrain$, $\xacqs$, and (sampled) $\yacqs$, we have for the EIG (before taking $\lambda \to 0$):
    \begin{align}
        \MIof{\W; \Yacqs \given \xacqs, \Dtrain} &\overset{\approx}{\le} \tfrac{1}{2} \log \det \left (  \similarityMatrix{}{ \Dacq \given \wstar}  + \lambda Id \right ) - \tfrac{\lvert \Dacq \rvert}{2} \log \lambda.
    \end{align}
    As the second term is independent of $\xacqs$, we can use the following proxy objective in the limit:
    \begin{align}
        \frac{1}{2} \log \det \left (  \similarityMatrix{}{ \Dacq \given \wstar} \right ).
    \end{align}
\end{restatable}
\end{importantresult}

\paragraph{Connection to Other Approximate Information Quantities.} Interestingly, we can use the above to obtain approximations of the predictive information gains (EPIG and JEPIG) because the terms that would tend towards $-\infty$ cancel out; see \S\ref{fim:appsec:similarity_matrices} for details. For EPIG, we have:
\begin{importantresult}
\begin{proposition}
    Given $\Deval, \Dtrain$, $\xacqs$ and (sampled) $\yacqs$, we have for the EPIG:
    \begin{align}
        &\MIof{\Yeval; \Yacqs \given \Xeval, \xacqs, \Dtrain} \notag \\
        &\approx \medmath{\tfrac{1}{2} \log \det \left ( \similarityMatrix{\HofHessian{\wstar \given \Dtrain}}{ \Deval \given \wstar}  + Id \right )
        -
        \tfrac{1}{2} {\log \det \left ( \similarityMatrix{\HofHessian{\wstar \given \Dtrain}}{ \Dacq, \Deval \given \wstar} + Id \right )}} \notag \\
        &\quad \quad 
        \medmath{+ \tfrac{1}{2} {\log \det \left ( \similarityMatrix{\HofHessian{\wstar \given \Dtrain}}{ \Dacq \given \wstar}  + Id \right )}}.
    \intertext{For an uninformative prior, we have:}
        &\approx \tfrac{1}{2} \medmath{\log \det \left (  \similarityMatrix{}{ \Deval \given \wstar} \right )
        -
        \tfrac{1}{2} \log \det \left (  \similarityMatrix{}{ \Dacq, \Deval \given \wstar} \right )
        +
        \tfrac{1}{2} \log \det \left (  \similarityMatrix{}{ \Dacq \given \wstar} \right )}.
    \end{align}
    We can drop the terms that only depend on $\Deval$ when we are interested in proxy objectives for optimization.
\end{proposition}
\end{importantresult}

These results help connect the objectives of PRISM and SIMILAR to the EIG and EPIG in the next section.

\section{Information Quantities in Prior Literature}
\label{fim:sec:unification}

Now, we can connect approaches in non-Bayesian literature to information quantities.
Additional proofs are given in \S\ref{fim:appsec:unification}.

\subsection{BAIT, ActiveSetSelect, and (J)EPIG}

BAIT in ``Gone Fishing'' \citep{ash2021gone}, ActiveSetSelect in ``Convergence Rates of Active Learning for Maximum
Likelihood Estimation'' \citep{chaudhuri2015convergence}, and (J)EPIG (\Cref{chapter:pobal}) approximate the same objective.

\citet{ash2021gone} introduce the \emph{BAIT} objective for deep active learning:
\begin{align}
    \medmath{\argmin_\xacqs \tr \left (\left (\FisherInfo{\Yacqs \given \xacqs, \wstar} + \FisherInfo{\Ytrain \given \xtrain, \wstar} + \lambda I \right )^{-1} \, \FisherInfo{\Yevals \given \xevals, \wstar} \right )}, \tag{BAIT}
\end{align}
where $\lambda$ is a hyperparameter\footnote{This is the BAIT objective as computed in Algorithm 1 in \citet{ash2021gone} and in the published implementation \shorturl{https://github.com/JordanAsh/badge/blob/master/query_strategies/bait_sampling.py}.}.

BAIT is based on a similar objective for GLMs from \citet{chaudhuri2015convergence}.
While \citet{ash2021gone} apply this objective to DNNs, they only use the last layer to approximate the Fisher information. The last layer, with appropriate activation functions and losses, constitutes a GLM as seen in \S\ref{fim:sec:fisher_information}.

Following \Cref{fim:prop:epig_fisher_approximation}, we immediately see that \citet{ash2021gone} perform transductive active learning (using the pool set as an evaluation set) and approximate a proxy objective for (J)EPIG:
\begin{importantresult}
\begin{proposition}
    Both \citet{ash2021gone} and \citet{chaudhuri2015convergence} perform transductive active learning, approximating (J)EPIG (\Cref{chapter:pobal}) using a last-layer approach (or GLM):
    \begin{align}
        &\argmax_\xacq \MIof{\Yeval; \Yacqs \given \Xeval, \xacqs} \\
        &\quad \approx \argmin_\xacq
        \tr
         ( \FisherInfo{\Yevals \given \xevals, \wstar} \,  (\FisherInfo{\Yacqs \given \xacqs, \wstar} + \HofHessian{\wstar \given \Dtrain} )^{-1}  ), \notag
    \end{align}
    with $\HofHessian{\wstar \given \Dtrain} = \HofHessian{\Dtrain \given \wstar} + \HofHessian{\wstar}$ and $\HofHessian{\wstar} = \lambda \, Id$.
\end{proposition}
\end{importantresult}
\begin{proof}
    This follows immediately for GLM (last-layer approaches) when we expand $\HofHessian{\wstar \given \Dtrain}$.
    \citet{chaudhuri2015convergence} in particular uses an uninformative prior, that is $\lambda = 0$.
    Comparing the resulting objectives yields the statement.
\end{proof}

Thus, \citet{ash2021gone} and we in \Cref{chapter:pobal} employ the same underlying acquisition function, albeit using very different approaches: \citet{ash2021gone} use a last-layer Fisher information matrix, whereas we use approximate BNNs and sample joint predictions in \Cref{chapter:pobal}. 

\paragraph{Research Questions.} %
EPIG is not submodular, and the greedy selection of an acquisition batch does not come with any optimality guarantees. While we sidesteps this in \Cref{chapter:pobal} by focusing on individual acquisitions mainly, \citet{ash2021gone} propose a heuristic that empirically performs better: They greedily select additional acquisition candidates in forward pass (twice the intended batch acquisition size) and then greedily remove the least informative samples from the batch in a backward pass. 
\emph{Would this heuristic also prove beneficial for all the other information quantities that are not submodular?}

While \citet{ash2021gone} state that they only use the last-layer approach for performance reasons, following \S\ref{fim:sec:fisher_information}, it does not seem that this approach translates beyond a last-layer approach for DNNs in a principled fashion (see \S\ref{fim:appsec:epig_general_case}). 
\emph{Is there a principled approach for the general case that goes beyond last-layer active learning when using Fisher information without the GGN approximation?}

\citet{ash2021gone} ablate trace and determinantal approaches, similar to comparing \cref{fim:eq:epig_fisher_approximation_trace} and \cref{fim:eq:epig_fisher_approximation_logdet}, yet they do not include $+Id$ in the log determinant expression, which leads them to examine the EIG in their ablation\footnote{\citet{ash2021gone} accidentally writes $\argmax$ instead of $\argmin$ in \S5.1 in their paper, but c.f. Algorithm 1 with the trace objective. Algorithm 2 \&\ 3 in \S{B} in the appendix use the correct final objective.}:
\begin{align}
    &\argmin_\xacq \log \det ( \FisherInfo{\Yevals \given \xevals, \wstar} \,  (\sum_i \FisherInfo{\Yacq_i \given \xacq_i, \wstar} + \HofHessian{\wstar \given \Dtrain} )^{-1} ) \notag \\
    &\quad = \medmath{\argmin_\xacq \log \det \FisherInfo{\Yevals \given \xevals, \wstar}  - \log \det (\sum_i \FisherInfo{\Yacq_i \given \xacq_i, \wstar} + \HofHessian{\wstar \given \Dtrain} )} \\
    &\quad = \argmax_\xacq \log \det (\sum_i \FisherInfo{\Yacq_i \given \xacq_i, \wstar} + \HofHessian{\wstar \given \Dtrain} ) + C,
\end{align}
and the last term matches the EIG in \cref{fim:eq:epig_fisher_approximation_logdet} up constant terms independent of $\xacq$. 
Thus, the ablations in \citet{ash2021gone} only compares EIG and EPIG.
\emph{Could comparing \cref{fim:eq:epig_fisher_approximation_trace} and \cref{fim:eq:epig_fisher_approximation_logdet} provide more insightful results about the trade-offs between trace and determinant approximations?}

\subsection{BADGE and BatchBALD}

BADGE \citep{ash2020deep} performs batch acquisition using a similarity matrix:
Using the concepts of \S\ref{fim:sec:similarity_matrices}, BADGE uses hard pseudo-labels 
together with last-layer gradient embeddings for the similarity matrix $\similarityMatrix{}{\Dacq \given \wstar}$. 
The authors sample from a k-DPP \citep{kulesza2011k} based on this similarity matrix to select a diverse batch of samples for acquisition. 
However, to further speed up acquisitions, BADGE uses k-MEANS++ \citep{arthur2006k, ostrovsky2013effectiveness} instead of a k-DPP: it uses the Jacobians $\HofJacobian{\y \given \x, \wstar}$ of the data matrix directly and samples a diverse batch based on the Euclidean distance between these Jacobians.
However, sampling from k-DPPs does not pick the most informative batch overall, and the ablations in \citet{ash2020deep} show that k-MEANS++ outperforms k-DPP.
Finally, the paper only motivates using gradient embeddings with hard pseudo-labels through intuitions: the gradient length captures information about the model's uncertainty, and diverse update directions capture information about the model's diversity \citep{ash2020deep}. The paper makes no explicit connection to information theory.

Following \Cref{fim:prop:similarity_based_IG}, since BADGE can be seen as using a last-layer approach for the similarity matrix $\similarityMatrix{}{\Dacq \given \wstar}$ with hard pseudo-labels, BADGE approximates $\MIof{\W;\Yacqs \given \xacqs, \Dtrain}$ with an uninformative posterior distribution:
\begin{importantresult}
    \begin{proposition}
        BADGE maximizes an approximation of the EIG with an uninformative prior.
    \end{proposition}
\end{importantresult}

\paragraph{Comparison to BatchBALD.} Similarly, BatchBALD (\Cref{chapter:batchbald}) approximates the EIG in the batch acquisition case but by using prediction-space samples. Moreover, BatchBALD uses a greedy approach to select batch candidates instead of sampling via a k-DPP or k-MEANS++.

As the EIG is submodular, determining the acquisition batch is a submodular optimization problem and, therefore, can be solved by greedy selection with $1-\tfrac{1}{e}$ optimality \citep{nemhauser1978analysis}.

\paragraph{Research Questions.} Hard pseudo-labels lead to biased estimates. \emph{Would one-sample estimates perform better?} \emph{And could greedy batch selection work better than sampling via a k-DPP?} This would be closer to the batch acquisition strategy followed by BatchBALD.

\subsection{SIMILAR and PRISM}
Based on \citet{iyer2021submodular}, \citet{kothawade2021similar} and \citet{kothawade2022prism} investigate \emph{submodular active learning} for DNNs: they take an \emph{information function} $f$, which is a non-negative, montone/non-decreasing, submodular function (and then is also subadditive as a consequence):
\begin{align}
    f(A) \geq 0 \tag{non-negative}, \\
    f(A) \leq f(B) \text{ for } A \subseteq B, \tag{monotone} \\
    f(A \cup B) \leq f(A) + f(B) - f(A \cap B), \tag{submodular} \\
    f(A \cup B) \leq f(A) + f(B) \tag{subadditive}
\end{align} 
for all $A, B \subseteq \Dpool$ and define a ``\emph{submodular conditional gain}`` and an ''\emph{submodular (conditional) mutual information}'' as
\begin{align}
    H_f(A \given B) &\defeq f(A \cup B) - f(B) \\
    I_f(A; B) &\defeq f(A \cup B) - f(A) - f(B).
\end{align}
For $f(\xs) = \Hof{\Ys \given \xs}$, this simply yields the regular information quantities.
Hence,
\citet{kothawade2021similar} and \citet{kothawade2022prism} examine other information functions and submodular quantities in the context of active learning: amongst them set covers, graph cuts, facility location, and log determinants (LogDet) of similarity matrices.
Like BADGE \citep{ash2020deep}, the similarity matrix uses hard pseudo-labels.
Like BatchBALD (\Cref{chapter:batchbald}), they use a greedy approach for acquisition \citep{nemhauser1978analysis}.

Using our results, we immediately see that the LogDet objective, which we can write as $\log \det \similarityMatrix{}{\Dacq \given \wstar}$, exactly matches the EIG approximation in \S\ref{fim:prop:similar_uninformative_posterior}; furthermore, in \S\ref{fim:appsec:unify_similar}, we show that the LogDetMI objective matches an approximation of JEPIG (and similarly, derive the LogDetCMI objective as well):
\begin{importantresult}
\begin{proposition}
The LogDet objective
\begin{math}
    \log \det \similarityMatrix{}{\Dacq \given \wstar}
\end{math}
is an approximation of the EIG and the LogDetMI objective
\begin{align}
    \medmath{\log \det \similarityMatrix{}{\Dacq \given \wstar} - \log \det (
        \similarityMatrix{}{\Dacq \given \wstar} 
        -
        \similarityMatrix{}{\Dacq ; \Deval \given \wstar} 
        \similarityMatrix{}{\Deval \given \wstar} ^{-1}
        \similarityMatrix{}{\Deval ; \Dacq \given \wstar} 
    )} %
\end{align}
is an approximation of a proxy objective for EPIG, where we use $\similarityMatrix{}{\Dany_1 ; \Dany_2 \given \wstar}$ to denote the (non-symmetric) similarity matrix between $\Dany_1$ and $\Dany_2$.
\end{proposition}
\end{importantresult}

Notably, the experimental results for the LogDet-based quantities are reported as among the best in \citet{kothawade2021similar} and \citet{kothawade2022prism}.
As such, since the LogDet quantities approximate Shannon's information quantities (which are not explicitly examined in those works), the promising experimental results compared to other submodular information functions support the hypothesis that approximating Shannon's information quantities works well in active learning and active sampling.

\paragraph{Research Questions.} Similar to BADGE, the scores are biased by using hard pseudo-labels. \emph{Could one-sample estimates perform better?} Furthermore, as LogDetMI is not submodular, \emph{could the approach from BAIT of expanding and shrinking the acquisition batch in a forward and backward pass improve performance here as well?}

\subsection{Expected Gradient Length}

The \emph{Expected Gradient Length (EGL)} \citep{settles2007multiple, Settles2010} is an acquisition function in active learning and is usually defined for non-Bayesian models.
Originally, it was an expectation over the gradient norm. In more recent literature \citep{huang2016active}, it is introduced using the squared gradient norm:
\begin{align}
    \simpleE{\pof{\yacq \given \xacq, \wstar}}{\left \Vert \HofJacobian{\yacq \given \xacq, \wstar} \right \Vert^2}. \tag{EGL}
\end{align}

Using a diagonal approximation of Fisher information, we show in \S\ref{fim:appsec:unify_egl}:
\begin{importantresult}
\begin{restatable}{proposition}{propegleigconnection}
    The EIG for a candidate sample $\xacq$ approximately lower-bounds the EGL:
    \begin{align}
        2 \MIof{\W; \Yacq \given \xacq} &\overset{\approx}{\le} \simpleE{\pof{\yacq \given \xacq, \wstar}}{\left \Vert \HofJacobian{\yacq \given \xacq, \wstar} \right \Vert^2} + \text{const.}
    \end{align}
\end{restatable}
\end{importantresult}
\subsection{Deep Learning on a Data Diet}

In active sampling, \citet{paul2021deep} use the gradient length of given labeled samples $\x, \y$ (averaged over multiple training runs) as an acquisition function to select the most informative samples from the training set to speed up training:
\begin{align}
    \simpleE{\qof{\w}}{\left \Vert \HofJacobian{\yacq \given \xacq, \w} \right \Vert^2}, \tag{GraNd}
\end{align}
which they call the \emph{gradient norm score (GraNd)}. The expectation is taken over the model parameters at initialization or after training for a few epochs---as this is not easily expressed using a posterior distribution, we use $\qof{w}$ to denote the distribution.

\begin{importantresult}
\begin{restatable}{proposition}{propdatadietigconnection}
    The IG for a candidate sample $\xacq$ approximately lower-bounds the gradient norm score (GraNd) at $\wstar$ up to a second-order term:
    \begin{align}
        2 \MIof{\W; \yacq \given \xacq} &\overset{\approx}{\le} \E{\qof{\w}}{\left \Vert \HofJacobian{\yacq \given \xacq, \w} \right \Vert^2} - \E{\qof{\w}}{\tr \left ( \frac{\nabla_\w^2 \pof{y \given x, \w}}{\pof{y \given x, \w}} \right )} + \text{const.}
    \end{align}
\end{restatable}
\end{importantresult}
The second term might not be negligible. Hence, GraNd (the first term on the left) might deviate from the information gain. \emph{How does the information gain compares to GraNd in practice?}

We provide a reproducibility analysis in \Cref{chapter:dldd}, which helped discover a now fixed bug in \citet{paul2021deep}.

\section{Discussion}
We have examined Fisher information and Gaussian approximations and have derived weight-space approximations for various information quantities. This has allowed us to connect these information quantities to objectives already used in the literature.
Moreover, we can make the following concluding points:

\paragraph{Last-Layer Approaches.} Methods that only use last-layer Fisher information or similar perform data subset selection on embeddings only, despite feature learning being arguably the most important strength of deep neural networks. However, these approaches can find great use with large pre-trained models, which are only fine-tuned on new data domains \citep{tran2022plex}. 

\paragraph{Bias in Hard Pseudo-Labels.} Several methods (BADGE, PRISM, SIMILAR) use hard pseudo-labels for computing the similarity matrices, leading to biased estimates. Can one-sample estimates perform better? Can the equivalent approximations that do not make use of similarity matrices perform better? %

\paragraph{Trace vs $\log \det$ Approximations.} We have presented a hierarchy of approximations and bounds. Ablating these approximations along multiple dimensions, including whether to use the trace or $\log \det$ approximation and whether to use a GLM, the GGN approximation or the full Fisher information, could provide interesting insights into what is attainable.

\paragraph{Batch Acquisition Pathologies.} Approaches that use the matrix trace instead of the log determinant can end up being additive for batch candidates $\xacqs$ and, therefore, by definition, cannot take redundancies between batch candidates into account, leading to failures detailed in \Cref{chapter:batchbald,chapter:stochacq}. This is an issue for trace approximations of the (E)IG. Does the trace approximation of (JE)PIG handle batch acquisition pathologies better?
Similarly, most information quantities are not submodular, yet we use a greedy algorithm to select the acquisition batches. 
Is the heuristic proposed by \citet{ash2021gone} generally beneficial?

\paragraph{Weight vs. Prediction Space.} BADGE, BAIT, and the LogDet objectives of PRISM and SIMILAR \citep{ash2020deep, ash2021gone, kothawade2021similar, kothawade2022prism} approximate information quantities in weight space, while (Batch)BALD, (J)EPIG, and (J)PIG (\Cref{chapter:batchbald,chapter:pobal,chapter:rholoss}) approximate the information quantities in prediction space. 
Both approaches have their limitations: 

Weight-space approaches can suffer from the Gaussian approximation being of low quality: the Laplace approximation only captures the posterior distribution well once it concentrates sufficiently. 
However, this is unlikely to happen in a low-data regime.

Prediction-space approaches can suffer from a combinatorial explosion as the batch acquisition size increases because prediction configurations have to be enumerated or sampled to approximate the information quantities. In addition, many parameter samples might be needed to obtain low-variance estimates.

Importantly, prediction space approaches also require drawing samples from the posterior distribution but do not estimate the information quantities using the posterior distribution, unlike weight space approaches.

\paragraph{Informativeness Scores.} Taking a step back, we have seen that a Bayesian perspective using information quantities connects seemingly disparate literature. Although Bayesian methods are often seen as separate from (non-Bayesian) active learning and active sampling, the sometimes fuzzy notion of ``informativeness'' expressed through various different objectives in non-Bayesian settings collapses to the same couple of information quantities, which were, in principle, already known by \citet{lindley1956measure} and \citet{mackay1992information}.

\andreas{Could we maybe somehow approximate the BALD symmetry using the Delta method instead in a more tractable way? \citet{ash2021gone} gets away with it because they only use last-layer Fisher information. They say it is to keep their approach practical, but we can argue that it simply makes no sense otherwise!
THIS IS SOMETHING TO TALK TO TIM ABOUT!}

\FloatBarrier

}

{
	\begin{savequote}[75mm]
		Old paths tread anew,\\
		In light of insights accrued,\\
		Publications grew.
	\end{savequote}
	\chapter{\bbbaltitle}
	\label{chapter:bbbal}
	\graphicspath{{./b3al/}}
\makeatletter
\def\input@path{{./b3al/}}
\makeatother

\newcolumntype{C}[1]{>{\centering\arraybackslash}p{#1}}
\newcolumntype{R}[1]{>{\raggedleft\arraybackslash}p{#1}}

\newcommand{\Yhat}{\hat{Y}}
\newcommand{\yhat}{\hat{y}}

\newcommand{\Ypred}{Y}
\newcommand{\ypred}{y}
\newcommand{\Ytrue}{Y}
\newcommand{\ytrue}{y}

\newcommand{\logits}{{\hat z}}

\newcommand{\xs}{{\{\x_i\}}}
\newcommand{\ys}{{\{\y_i\}}}
\newcommand{\Ys}{{\{\Y_i\}}}

\newcommand{\xpools}{{\{\x^\text{pool}_i\}}}

\newcommand{\xacqs}{\xacqset} %
\newcommand{\yacqs}{\yacqset} %
\newcommand{\Yacqs}{\Yacqset} %

\newcommand{\xyacqs}{{\{(\xacq_i,\yacq_i)\}^\typeacq_{i=1..\batchvar}}}

\newcommand{\yacqsstar}{{\{\y^{\typeacq\text{*}}_i\}}}

\newcommand{\xevals}{\xevalset} %
\newcommand{\yevals}{\yevalset} %
\newcommand{\Yevals}{\Yevalset} %

\newcommand{\xytrains}{{\{(\xtrain_i,\ytrain_i)\}^\typetrain_{i=1..\batchvar}}}
\newcommand{\xtrains}{\xtrainset} %

\newcommand{\wstar}{\w^*}

\newcommand{\acqf}{\mathcal{A}}

\newcommand{\noiseobs}{\sigma_N}
\newcommand{\Sigmapred}[2]{\noiseobs(#1; #2)}

\newcommand{\mupred}[2]{{\mu}(#1; #2)}
\newcommand{\mupredshort}[2]{{\mu}^{#1}_{#2}}
\newcommand{\gradmupred}[2]{\nabla_{\w} \mu(#1; #2)}
\newcommand{\mushort}[2]{\mu_{#1}^{#2}}
\newcommand{\cmushort}[2]{\bar{\mu}_{#1}^{#2}}
\newcommand{\gradmushort}[2]{{\nabla_{\w} \mu_{#1}^{#2}}}

\newcommand{\predkernel}[1]{{k_\mathrm{pred}(#1)}}
\newcommand{\emOmega}{\hat{\W}}
\newcommand{\empredkernel}[1]{{k_{\widehat{\mathrm{pred}}}(#1)}}
\newcommand{\gradkernel}[1]{k_\mathrm{grad}(#1)}
\newcommand{\xpostgradkernel}[2]{{k_{\mathrm{grad}\to \mathrm{post}(#1)}(#2)}}
\newcommand{\postgradkernel}[1]{{k_{\mathrm{grad}\to \mathrm{post}(\Dtrain)}(#1)}}

\newcommand{\realnum}{\mathbb{R}}

\newcommand{\verteq}{\rotatebox{90}{$\,=$}}
\newcommand{\equalto}[2]{\underset{\displaystyle\overset{\mkern4mu\verteq}{#2}}{#1}}

By selectively acquiring labels for a subset of available unlabeled data, active learning \citep{atlas1989training} is suited for situations where the acquisition of labels is costly or time-consuming, such as in medical imaging or natural language processing. 
However, in deep learning, many recent batch active learning methods have focused on \emph{white-box approaches} that rely on the model being parametric and differentiable and which use first or second-order derivatives (e.g.\ model embeddings)\footnote{Model embeddings can also be seen as first-order derivatives of the model score under regression in regard to the last layer. 
}.

This can present a limitation in real-world scenarios where model internals or gradients might not be accessible---or might be expensive to access.
This is particularly true in the case of `foundation models` \citep{bommasani2021opportunities} and large language models such as GPT-3 \citep{brown2020language}, for example, when accessed via a third party.  
More generally, a lack of differentiability might hinder application of white-box batch active learning approaches to non-differentiable models.

To address these limitations, we examine \emph{black-box batch active learning (\BBBAL)} for regression which is compatible with a wider range of machine learning models. By \emph{black-box}, we mean that our approach only relies on model predictions and does not require access to model internals or gradients.
Our approach is rooted in Bayesian principles and only requires model predictions from a small (bootstrapped) ensemble.
Specifically, we utilize an \emph{(empirical) predictive covariance kernel} based on sampled predictions.
We show that the well-known gradient kernel \citep{kothawade2021similar,kothawade2022prism,ash2020deep,ash2021gone} can be seen as an approximation of this predictive covariance kernel.

The proposed approach extends to non-differentiable models through a Bayesian view on the hypothesis space formulation of active learning, based on the ideas behind query-by-committee \citep{seung1992query}. 
This enables us to use batch active learning methods, such as BAIT \citep{ash2021gone} and BADGE \citep{ash2020deep} in a black-box setting with non-differentiable models, such as random forests or gradient-boosted trees. 

We evaluate black-box batch active learning on a diverse set of regression datasets. 
Unlike the above \emph{white-box} parametric active learning methods which scale in the number of (last-layer) model parameters or the embedding size, our method scales in the number of drawn predictions, and we show that we can already obtain excellent results with a small ensemble.
Our results demonstrate the label efficiency of \BBBAL for various machine learning models. For deep learning models, \BBBAL even performs better than the corresponding state-of-the-art white-box methods in many cases.

We focus on regression as classification using the same approach would require Laplace approximations, Monte Carlo sampling, or expectation propagation \citep{williams2006gaussian,hernandez2011robust}. This complicates a fair comparison as translating existing classification methods in active learning is not straightforward.
For the same reasons, we also do not consider proxy-based active learning methods \citep{coleman2019selection}, which constitutes an orthogonal direction to our investigation.

In summary, by leveraging the strengths of kernel-based methods and Bayesian principles, our approach improves the labeling efficiency of a range of differentiable and non-differentiable machine-learning models with surprisingly strong performance.

The rest of the chapter is organized as follows: in \S\ref{b3al:sec:related_work}, we discuss related work in active learning and kernel-based methods. 
In \S\ref{b3al:sec:methodology}, we describe the relevant background and provide a detailed description of \BBBAL. 
In \S\ref{b3al:sec:results}, we detail the experimental setup and provide the results of our experimental evaluation. Finally, \S\ref{b3al:sec:conclusion} concludes with a discussion and directions for future research.

\section{Related Work}
\label{b3al:sec:related_work}

\paragraph{Differentiable Models}. %
Many acquisition functions approximate well-known information-theoretic quantities \citep{mackay1992information}, 
often by approximating Fisher information implicitly or explicitly. 
This can be computationally expensive, particularly in deep learning where the number of model parameters can be large---even when using last-layer approximations or assuming a generalized linear model (\Cref{chapter:fim}).
BADGE \citep{ash2020deep} and BAIT \citep{ash2021gone} approximate the Fisher information using last-layer loss gradients or the Hessian, respectively, but still have a computational cost scaling with the number of last layer weights. This also applies to methods using similarity matrices (kernels) based on loss gradients of last-layer weights such as SIMILAR \citep{kothawade2022prism} and PRISM \citep{kothawade2022prism}, to only name a few. 
Importantly, all of these approaches require differentiable models.

\paragraph{Non-Differentiable Models.} %
\emph{Query-by-committee (QbC, \citet{seung1992query})} measures the disagreement between different model instances to identify informative samples and has been applied to regression \citep{krogh1994neural,burbidge2007active}. 
\citet{kee2018query} extend QbC to the batch setting with a diversity term based on the distance of data points in input space. 
\citet{nguyen2012efficient} show batch active learning for random forests. They train an ensemble of random forests and evaluate the joint entropy of the predictions of the ensemble for batch active learning, which can be seen as a special case of BatchBALD in regression.

\paragraph{BALD.} %
This chapter most closely aligns with the BALD-family of Bayesian active learning acquisition functions \citep{houlsby2011bayesian}, which focus on classification tasks, however. 
The crucial insight of BALD is applying the symmetry of mutual information to compute the expected information gain in prediction space instead of in parameter space.
As a result, BALD is a \emph{black-box technique} that only leverages model predictions.
The estimators utilized by \citet{gal2017deep} and also the ones for BatchBALD from \Cref{chapter:batchbald} enumerate over all classes, leading to a trade-off between combinatorial explosion and Monte-Carlo sampling, which can result in degraded quality estimates as acquisition batch sizes increase.
\citet{houlsby2011bayesian}, \citet{gal2017deep} have not applied BALD to regression tasks.

\paragraph{Kernel-Based Methods.} \citet{holzmuller2022framework} examine the previously mentioned methods and unify them using gradient-based kernels. 
Specifically, they express BALD \citep{houlsby2011bayesian}, BatchBALD (\Cref{chapter:batchbald}), BAIT \citep{ash2021gone}, BADGE \citep{ash2020deep}, ACS-FW \citep{pinsler2019bayesian}, and Core-Set \citep{sener2017active}/FF-Active \citep{geifman2017deep} using kernel-based methods for regression tasks.
They also propose a new method, LCMD (largest cluster maximum distance). %

\paragraph{Comparison to This Chapter.} %
This chapter extends the work of \citet{houlsby2011bayesian} and \citet{holzmuller2022framework} by combining the prediction-based approach with a kernel-based formulation. This trivially enables batch active learning on regression tasks using black-box predictions for a wide range of existing batch active learning methods. 

\section{Methodology}
\label{b3al:sec:methodology}

In this chapter, we focus on regression, which is a common task in machine learning. We assume that the target $\y$ is real-valued ($\in \realnum$) with homoscedastic Gaussian noise:
\begin{align}
    \Y \given \x, \w &\sim \normaldist{\mupred{\x}{\w}}{\noiseobs^2}. \label{b3al:eq:gaussian_noise}
\end{align}
Equivalently, $\Y \given \x, \w \sim \mupred{\x}{\w}+\varepsilon$ with $\varepsilon \sim \normaldist{0}{\noiseobs^2}$.
As usual, we assume that the noise is independent for different inputs $\x$ and  parameters $\w$.
Homoscedastic noise is a special case of the general heteroscedastic setting: the noise variance is simply a constant. Our approach could easily be extended to heteroscedastic noise by substituting a function $\Sigmapred{\x}{\w}$ for $\noiseobs$, but for this work we limit ourselves to the simplest case.

\subsection{Kernel-based Methods \&\ Information Theory}
\label{b3al:sec:kernelbasedintro}

We build on \citet{holzmuller2022framework} which expresses contemporary batch active learning methods using kernel methods. Specifically, active learning is performed using kernel-based surrogates for deep neural network models.
In this paper, we focus on surrogates using empirical covariance kernels instead of gradient kernels. 
While a full treatment of \citet{holzmuller2022framework} is beyond the scope of this chapter, we briefly review some key ideas here. We refer the reader to the extensive paper for more details.

\paragraph{Gaussian Processes.} Gaussian Processes are one way to introduce kernel-based methods. A simple way to think about Gaussian Processes \citep{williams2006gaussian,lazaro2010marginalized,Rudner2022fsvi} is as a Bayesian linear regression model with an implicit, potentially infinite-dimensional feature space (depending on the covariance kernel) that uses the kernel trick to abstract away the feature map which maps the input space to the feature space.

\paragraph{Multivariate Gaussian Distribution.} The distinctive property of a Gaussian Process is that all predictions are jointly Gaussian distributed.
We can then write the joint distribution for a univariate regression model as:
\begin{align}
    &\Y_1, \ldots, \Y_n \given \x_1, \ldots, \x_n \sim 
    \normaldist{\mathbf{0}}{\Cov{}{\mu(\x_1), \ldots, \mu(\x_n)} + \noiseobs^2 \mathbf{I}}, \label{b3al:eq:gp_joint} 
\end{align}
where $\mu(\x){}$ are the observation-noise free predictions as random variables, and $\Cov{}{\mu(\x_1), \linebreak[0] \ldots, \mu(\x_n)}$ is the covariance matrix of the predictions.
The covariance matrix is defined via the kernel function $k(\x, \x')$:
\begin{align}
    \Cov{}{\mu(\x_1), \ldots, \mu(\x_n)} = 
    \begin{bmatrix}
        k(\x_i, \x_j)
    \end{bmatrix}_{i,j=1}^{n,n}.
\end{align}
The kernel function $k(\x, \x')$ can be chosen almost arbitrarily as long as it is positive semi-definite, e.g.\ see \citet[Ch. 4]{williams2006gaussian}.  The linear kernel $k(\x, \x') = \x \cdot \x'$ and the radial basis function kernel $k(\x, \x') = \exp(-\frac{1}{2} \lvert {\x-\x'} \rvert^2)$ are common examples, as is the gradient kernel, which we examine next.

\newmathcommand{\xHessian}[1]{\nabla^2_{\w} #1}

\paragraph{Fisher Information \&\ Linearization.} When using neural networks for regression, the gradient kernel 
\begin{align}
    \gradkernel{\x, \x' \given \wstar} &\triangleq 
    \gradmupred{\x}{\wstar} \xHessian{[-\log \pof{\wstar}]}^{-1} \gradmupred{\x'}{\wstar}^\top \\ 
    &= \langle \gradmupred{\x}{\wstar}, \gradmupred{\x'}{\wstar} \rangle_{\xHessian{[-\log \pof{\wstar}]}^{-1}}
\end{align} 
is the canonical choice, where $\wstar$ is a \emph{maximum likelihood} or \emph{maximum a posteriori estimate (MLE, MAP)} and $\xHessian{[-\log \pof{\wstar}]}$ is the Hessian of the prior at $\wstar$. 
Note that $\gradmupred{\x}{\wstar}$ is a \emph{row} vector.
Commonly, the prior is a Gaussian distribution with an identity covariance matrix, and thus $\xHessian{[-\log \pof{\wstar}]} = \mathbf{I}$. 

The significance of this kernel lies in its relationship with the Fisher information matrix at $\wstar$ \citep{immer2020disentangling,immer2021improving}, as we have seen in \Cref{chapter:fim}, or equivalently, with the linearization of the loss function around $\wstar$ \citep{holzmuller2022framework}. This leads to a Gaussian approximation, which results in a Gaussian predictive posterior distribution when combined with a Gaussian likelihood. The use of the finite-dimensional gradient kernel thus results in an implicit Bayesian linear regression in the context of regression models.

\paragraph{Posterior Gradient Kernel.} %
We can use the well-known properties of multivariate normal distributions to marginalize or condition the joint distribution in \eqref{b3al:eq:gp_joint}. Following \citet{holzmuller2022framework}, this allows us to explicitly obtain the posterior gradient kernel given additional $\x_1, \ldots, \x_n$ as:
\begin{align}
    \label{b3al:eq:gp_postgradkernel} 
    &\xpostgradkernel{\x_1,\ldots,\x_n}{\x, \x' \given \wstar} \\
    &\quad \triangleq \medmath{\nabla_{\w} \mupred{\x}{\wstar}
    \left ( \noiseobs^{-2}  \begin{pmatrix}
        \gradmupred{\x_1}{\wstar} \\
        \vdots \\
        \gradmupred{\x_n}{\wstar}
    \end{pmatrix} \begin{pmatrix}
        \gradmupred{\x_1}{\wstar} \\
        \vdots \\
        \gradmupred{\x_n}{\wstar}
    \end{pmatrix}^\top + \xHessian{[-\log \pof{\wstar}]} \right )^{-1} 
    \nabla_{\w} \mupred{\x'}{\wstar}^\top.} \notag
\end{align}
The factor $\noiseobs^{-2}$ originates from implicitly conditioning on $\Y_i \given \x_i$, which include observation noise. %

Importantly for active learning, the multivariate normal distribution is the maximum entropy distribution for a given covariance matrix, and is thus an upper-bound for the entropy of any distribution with the same covariance matrix.
The entropy is given by the log-determinant of the covariance matrix:
\begin{align}
    \Hof{\Y_1, \ldots, \Y_n \given \x_1, \ldots, \x_n} &= \frac{1}{2} \log \det (\Cov{}{\mu(\x_1), \ldots, \mu(\x_n)} + \sigma^2 \mathbf{I}) + C_n,
\end{align}
where $C_n \triangleq \frac{n}{2} \log(2 \, \pi \, e)$ is a constant that only depends on the number of samples $n$.
Connecting kernel-based methods to information-theoretic quantities like the expected information gain, the respective acquisition scores are upper-bounds on the expected information gain, as we have examined in the previous chapter.

\paragraph{Expected Information Gain for Regression.} In \Cref{chapter:batchbald}, for acquisition of samples $\xacqs$ for classification tasks, the EIG was defined via BatchBALD as:
\begin{align}
    \MIof{\Yacqset; \W \given \xacqset, \Dtrain} = \Hof{\Yacqset \given \xacqset, \Dtrain} - \Hof{\Yacqset \given \xacqset, \W, \Dtrain}.
    \tag{\ref{batchbald:eq:batchbald}}
\end{align}
using the mutual information between the parameters $\w$ and the predictions $\Ys$ on an acquisition candidate batch $\xs$.
The conditional entropy term is precisely the entropy of a normal distribution given our model assumptions for the observation noise and can be computed exactly. In particular, for our regression task, as we assume fixed aleatoric uncertainty, the conditional entropy is constant in $\x$ and can even be dropped from the objective. BALD and the EIG for models with fixed aleatoric are equivalent to entropy maximization, which is a common objective in BOED \citep{Hernandez-Lobato_2016}.
To estimate BALD, we thus only need to compute the joint entropy, which can be upper-bounded by the corresponding entropy of the multivariate normal distribution with the same covariance matrix, yielding an upper-bound overall.

\subsection{Black-Box Batch Active Learning}

We more formally introduce the empirical covariance kernel and compare it to the gradient kernel commonly used for active learning  with deep learning models in parameter-space. For non-differentiable models, we show how it can also be derived using a Bayesian model perspective on the hypothesis space.

In addition to being illustrative, this section allows us to connect prediction-based kernels to the kernels used by \citet{holzmuller2022framework}, which in turns connects them to various SotA active learning methods.

\subsubsection{Predictive Covariance Kernel}

To perform black-box batch active learning, we directly use the \emph{predictive covariance} of $\Y_{i} | \x_i$ and $\Y_{j} | \x_j$:
\begin{align}
    \Cov{\W}{Y_i; \Y_j \given \x_i, \x_j} &= \Cov{\W}{\mushort{\x_i}{\w}; \mushort{\x_j}{\w}} + \noiseobs^2 \indicator{i=j},
\end{align}
where we have abbreviated $\mupred{\x}{\w}$ with $\mushort{\x}{\w}$, and used the law of total covariance and the fact that the noise is uncorrelated between samples.

\begin{mainresult}
We define the \emph{predictive covariance kernel} $\predkernel{\x_i, \x_j}$ as the covariance of the predicted means:
\begin{align}
    \predkernel{\x_i; \x_j} \triangleq \Cov{\W}{\mushort{\x_i}{\w}; \mushort{\x_j}{\w}}.
\end{align}
Compared to \S\ref{b3al:sec:kernelbasedintro}, we do not define the covariance via the kernel but the kernel via the covariance.
\end{mainresult}
This is also simply known as the \emph{covariance kernel} in the literature \citep{shawe2004kernel}. We use the prefix \emph{predictive} to make clear that we look at the covariance of the predictions. The resulting Gram matrix is equal the covariance matrix of the predictions and positive definite (for positive $\noiseobs$ and otherwise positive semi-definite), and thus the kernel is a valid kernel.

\subsubsection{Empirical Predictive Covariance Kernel}

For $K$ sampled model parameters $\w_1, \ldots, \w_K \sim \pof{\w}$---for example, the members of a deep ensemble---the \emph{empirical predictive covariance kernel} $\empredkernel{\x_i; \x_j}$ is the empirical estimate:
\begin{align}
    \empredkernel{\x_i; \x_j} \triangleq \emCov{\W}{\mushort{\x_i}{\w}; \mushort{\x_j}{\w}}
    &= \frac{1}{K} \sum_{k=1}^K \left (\mushort{\x_i}{\w_k} - \frac{1}{K} \sum_{l=1}^K \mushort{\x_i}{\w_l} \right )^\top \left (\mushort{\x_j}{\w_k} - \frac{1}{K} \sum_{l=1}^K \mushort{\x_j}{\w_l} \right )
    \\
    &= \left \langle \frac{1}{\sqrt{K}} (\cmushort{\x_i}{\w_1}, \ldots, \cmushort{\x_i}{\w_K}), 
    \frac{1}{\sqrt{K}} (\cmushort{\x_j}{\w_1}, \ldots, \cmushort{\x_j}{\w_K}) \right \rangle, \label{b3al:eq:empirical_pred_kernel}
\end{align}
with centered predictions $\cmushort{\x}{\w_k} \triangleq \mushort{\x}{\w_k} - \frac{1}{K} \sum_{l=1}^K \mushort{\x}{\w_l}$. As we can write this kernel as an inner product, it also immediately follows that the empirical predictive covariance kernel is a valid kernel and positive semi-definite.

\subsubsection{Differentiable Models}

Similar to \citet[\S C.1]{holzmuller2022framework}, we show that the posterior gradient kernel is a first-order approximation of the (predictive) covariance kernel.
This section explicitly conditions on $\Dtrain$.
The result is simple but instructive:
\begin{importantresult}
\begin{proposition}
    The \emph{posterior} gradient kernel $\postgradkernel{\x_i; \x_j \given \wstar}$ is an approximation of the predictive covariance kernel $\predkernel{\x_i; \x_j}$.
\end{proposition}
\end{importantresult}
\begin{proof}
    We use a first-order Taylor expansion of the mean function $\mupred{\x}{\w}$ around $\wstar$:
    \begin{align}
        \mupred{\x}{\w} \approx \mupred{\x}{\wstar} + \gradmupred{\x}{\wstar}\, \underbrace{(\w - \wstar)}_{\triangleq \Delta \w}.
    \end{align}
    Choose $\wstar = \E{\w \sim \pof{\w \given \Dtrain}}{\w}$ (BMA). Then we have $\E{\pof{w \given \Dtrain}}{\mupred{\x}{\w}} = \mupred{\x}{\wstar}$. We then have:
    \begin{align}
        \predkernel{\x_i; \x_j} &= \Cov{\w \sim \pof{\w \given \Dtrain}}{\mupred{\x_i}{\w}; \mupred{\x_j}{\w}} \\
        &\approx \E{\wstar + \Delta \w \sim \pof{w \given \Dtrain}}{\langle \gradmushort{\x_i}{\wstar}\, \Delta \w, \gradmushort{\x_j}{\wstar}\, \Delta \w \rangle} \\
        &= \gradmushort{\x_i}{\wstar} \, \E{\wstar + \Delta \w \sim \pof{w \given \Dtrain}}{ \Delta \w \Delta \w^\top} \, \gradmushort{\x_j}{\wstar}^\top \\
        &= \gradmupred{\x_i}{\wstar} \, \Cov{}{\W \given \Dtrain} \, \gradmupred{\x_j}{\wstar}^\top \\
        &\approx \postgradkernel{\x_i; \x_j \given \wstar}.
    \end{align}
    The intermediate expectation is the model covariance $\Cov{}{\W \given \Dtrain}$ as $\wstar$ is the BMA.
    For the last step, we use the Gauss-Newton approximation again \citep{immer2021improving} and approximate the inverse of the covariance using the Hessian of the negative log likelihood at $\wstar$:
    \begin{align}
        &\Cov{}{\W \given \Dtrain}^{-1} \approx \xHessian{[- \log \pof{\wstar \given \Dtrain}]} \\
        &\quad = \xHessian{[- \log \pof{\Dtrain \given \wstar} - \log \pof{\wstar}]} \\
        &\quad = \sigma^{-2} \textstyle \sum_i \gradmupred{\xtrain_i}{\wstar}^\top \gradmupred{\xtrain_i}{\wstar} \\
        &\quad \quad - \xHessian{\log \pof{\wstar}}, 
    \end{align}
    where we have first used Bayes' theorem and that $\pof{\Dtrain}$ vanishes under differentiation---it is constant in $\w$.
    Secondly, the Hessian of the negative log likelihood is just the outer product of the gradients divided by the noise variance in the homoscedastic regression case.
    $\xHessian{[-\log \pof{\wstar}]}$ is the prior term. 
    This matches \eqref{b3al:eq:gp_postgradkernel}.
\end{proof}

\subsubsection{Non-Differentiable Models}

\newmathcommand{\bmcparam}{{\pmb{\alpha}}}

\newcommand{\bmch}{\psi}
\newcommand{\Bmch}{\Psi}
\newcommand{\bmccatprob}{\pmb{q}}

\newcommand{\bmcmupred}[2]{{\tilde{\mu}}(#1; #2)}
\newcommand{\bmcgradmupred}[2]{{\nabla_{\bmch} \bmcmupred{#1}{#2}}}

\newmathcommand{\bmcpredkernel}[1]{k_{\mathrm{pred}, \bmch}(#1)}
\newmathcommand{\bmcgradkernel}[1]{k_{\mathrm{grad}, \bmccatprob}(#1)}
\newmathcommand{\bmcpostgradkernel}[1]{{k_{\mathrm{grad}, \bmch \to \mathrm{post}(\Dtrain)}(#1)}}

How can we apply the above result to non-differentiable models?
In the following, we use a Bayesian view on the hypothesis space to show that we can connect the empirical predictive covariance kernel to a gradient kernel here, too.
With $\emOmega \triangleq (\w_1, \ldots, \w_K)$ fixed---e.g. these could be the different parameters of the members of a deep ensemble---we introduce a latent $\Bmch$ to represent the `true' hypothesis $\w_\bmch \in \emOmega$ from this empirical hypothesis space $\emOmega$, which we want to identify. This is similar to QbC \citep{seung1992query}. In essence, the latent $\Bmch$ takes on the role of $\W$ from the previous section, and we are interested in learning the `true' $\Bmch$ from additional data. We, thus, examine the kernels for $\Bmch$, as opposed to $\W$.

Specifically, we model $\Bmch$ using a one-hot categorical distribution, that is a multinomial distribution from which we draw one sample: $\Bmch \sim \Multinomial(\bmccatprob, 1)$, with $\bmccatprob \in S^{K-1}$ parameterizing the distribution, where $S^{K-1}$ denotes the $K-1$ simplex in $\realnum^K$. Then, $\bmccatprob_k = \pof{\Bmch = e_k}$, where $e_k$ denotes the $k$-th unit vector; and $\sum_{k=1}^K \bmccatprob_k = 1$.
For the predictive $\bmcmupred{\x}{\Bmch}$, we have:
\begin{align}
    \bmcmupred{\x}{\Bmch} &\triangleq \mupred{\x}{\w_\Bmch} = \langle \mupred{\x}{\cdot}, \Bmch \rangle,
\end{align}
where we use $\w_\bmch$ to denote the $\w_k$ when we have $\bmch=e_k$ in slight abuse of notation, and $\mupred{\x}{\cdot} \in \realnum^K$ is a column vector of the predictions $\mupred{\x}{\w_k}$ for $\x$ for all $\w_k$. This follows from $\bmch$ being a one-hot vector.

We now examine this model and its kernels. The BMA of $\bmcmupred{\x}{\Bmch}$ matches the previous empirical mean, and, if we choose $\bmccatprob$ to have an uninformative\footnote{If we had additional information about the $\emOmega$---for example, if we had validation losses---we could use that to inform $\bmccatprob$.} uniform distribution over the hypotheses ($\bmccatprob_k \triangleq \frac{1}{K}$), we obtain:
\begin{align}
    \bmcmupred{\x}{\bmccatprob} &\triangleq \E{\pof{\bmch}}{\mupred{\x}{\w_\bmch}}
    = \langle \mupred{\x}{\cdot}, \bmccatprob \rangle \label{b3al:eq:empirical_bma_linearization} = \sum_{\bmch=1}^K \bmccatprob_\bmch \mupred{\x}{\w_\bmch} =  \sum_{\bmch=1}^K \frac{1}{K} \mupred{\x}{\w_\bmch}. 
\end{align}
What is the predictive covariance kernel of this model? And what is the posterior gradient kernel for $\bmccatprob$?
\begin{importantresult}
    \begin{proposition}
        \
        \begin{enumerate}[leftmargin=*]
            \item The predictive covariance kernel $\bmcpredkernel{\x_i, \x_j}$ for $\emOmega$ using uniform $\bmccatprob$ is equal to the empirical predictive covariance kernel $\empredkernel{\x_i; \x_j}$.
            \item The `posterior' gradient kernel $\bmcpostgradkernel{\x_i ; \x_j}$ for $\emOmega$ \emph{in respect to} $\Bmch$ using uniform $\bmccatprob$ is equal to the empirical predictive covariance kernel $\empredkernel{\x_i; \x_j}$.
        \end{enumerate}
    \end{proposition}
\end{importantresult}
\begin{proof}
    Like for the previous differentiable model, the BMA of the model parameters $\Bmch$ is just $\bmccatprob$: $\E{}{\Bmch} = \bmccatprob$. 
    The first statement immediately follows:
    \begin{align}
        \bmcpredkernel{\x_i, \x_j} &= \Cov{\bmch}{\bmcmupred{\x_i}{\bmch}; \bmcmupred{\x_j}{\bmch}} 
        = \E{\pof{\bmch}}{ \cmushort{\x_i}{\w_\bmch} \, \cmushort{\x_j}{\w_\bmch}}
        \\
        &= {\textstyle \frac{1}{K}} \sum_\bmch \cmushort{\x_i}{\w_\bmch} \cmushort{\x_j}{\w_\bmch} = \empredkernel{\x_i; \x_j}.
    \end{align}
    For the second statement, we will show that we can express the predictive covariance kernel as a linearization around $\Bmch$.
    We can read off a linearization for $\bmcgradmupred{\x_i}{\bmch}$ from the inner product in \Cref{b3al:eq:empirical_bma_linearization}:
    \begin{align}
        \bmcgradmupred{\x_i}{\bmch} = \mupred{\x}{\cdot}^\top,
    \end{align}
    This allows us to write the predictive covariance kernel as a linearization around $\bmccatprob$:
    \begin{align}
        \bmcpredkernel{\x_i, \x_j} &= \Cov{\bmch \sim \pof{\bmch}}{\bmcmupred{\x_i}{\bmch}; \bmcmupred{\x_j}{\bmch}} \\
        &= \E{\bmccatprob + \Delta \bmch \sim \pof{\bmch}}{\bmcgradmupred{\x_i}{\bmch} \Delta \bmch, \bmcgradmupred{\x_j}{\bmch} \Delta \bmch} \\
        &= \bmcgradmupred{\x_i}{\bmccatprob} \, \Cov{}{\Bmch} \, \bmcgradmupred{\x_i}{\bmccatprob}^\top\\
        &=\bmcpostgradkernel{\x_i ; \x_j}.
    \end{align}
\end{proof}
The above gradient kernel is only the posterior gradient kernel in the sense that we have sampled $\w_\bmch$ from the non-differentiable model after inference on training data. The samples themselves are drawn uniformly.

The covariance of the multinomial $\Bmch$ is:
\begin{math}
    \Cov{}{\Bmch} = \diag(\bmccatprob) - \bmccatprob \bmccatprob^\top.
\end{math}    
Thus, substituting, we can verify that the posterior gradient kernel is indeed equal to the predictive covariance kernel:
\begin{align}
    \bmcpostgradkernel{\x_i ; \x_j} &= \bmcgradmupred{\x_i}{\bmccatprob} \, (\diag(\bmccatprob) - \bmccatprob \bmccatprob^\top) \, \bmcgradmupred{\x_i}{\bmccatprob}^\top\\
    &=
    \mupred{\x_i}{\cdot}^\top \diag(\bmccatprob) \, \mupred{\x_j}{\cdot} - (\mupred{\x_i}{\cdot}^\top \, \bmccatprob) \, (\bmccatprob^\top \, \mupred{\x_j}{\cdot}) \\
    &= \frac{1}{K} \sum_\bmch \mupred{\x_i}{\w_\bmch} \, \mupred{\x_j}{\w_\bmch}^\top \\
    &\quad - \left (\frac{1}{K} \sum_\bmch \mupred{\x_i}{\w_\bmch} \right ) \, \left (\frac{1}{K} \sum_\bmch \mupred{\x_j}{\w_\bmch}\right ) \notag \\
    &= \empredkernel{\x_i; \x_j}.
\end{align}
This demonstrates that a straightforward Bayesian model can be constructed on top of a non-differentiable ensemble model. Bayesian inference in this context aims to identify the most suitable member of the ensemble. Given the limited number of samples and likelihood of model misspecification, it is likely that none of the members accurately represents the true model. However, for active learning purposes, the main focus is solely on quantifying the degree of disagreement among the ensemble members.

A similar Bayesian model using Bayesian Model Combination (BMC) could be set up which allows for arbitrary convex mixtures of the ensemble members. 
This would entail using a Dirichlet distribution $\Bmch \sim \Dirichlet(\bmcparam)$ instead of the multinomial distribution. Assuming an uninformative prior ($\bmcparam_k \triangleq \nicefrac{1}{K}$), this leads to the same results up to a constant factor of $1+\sum_k \bmcparam_k = 2$. This is pleasing because it does not matter whether we use a multinomial or Dirichlet distribution, that is: whether we take a hypothesis space view with a `true' hypothesis or accept that our model is likely misspecified, and we are dealing with a mixture of models, the results are the same up to a constant factor.

\paragraph{Application to DNNs, BNNs, and Other Models.} %

The proposed approach has relevance due to its versatility, as it can be applied to a wide range of models that can be consistently queried for prediction, including deep ensembles \citep{lakshminarayanan2016simple}, Bayesian neural networks (BNNs) \citep{blundell2015weight,gal2015dropout}, and non-differentiable models. The kernel used in this approach is simple to implement and scales in the number of empirical predictions per sample, rather than in the parameter space, as seen in other methods such as \citet{ash2021gone}. 

\begin{figure}[t]
    \centering
    \begin{subfigure}[]{0.90\linewidth}
        \centering
        \includegraphics[width=0.50\linewidth]{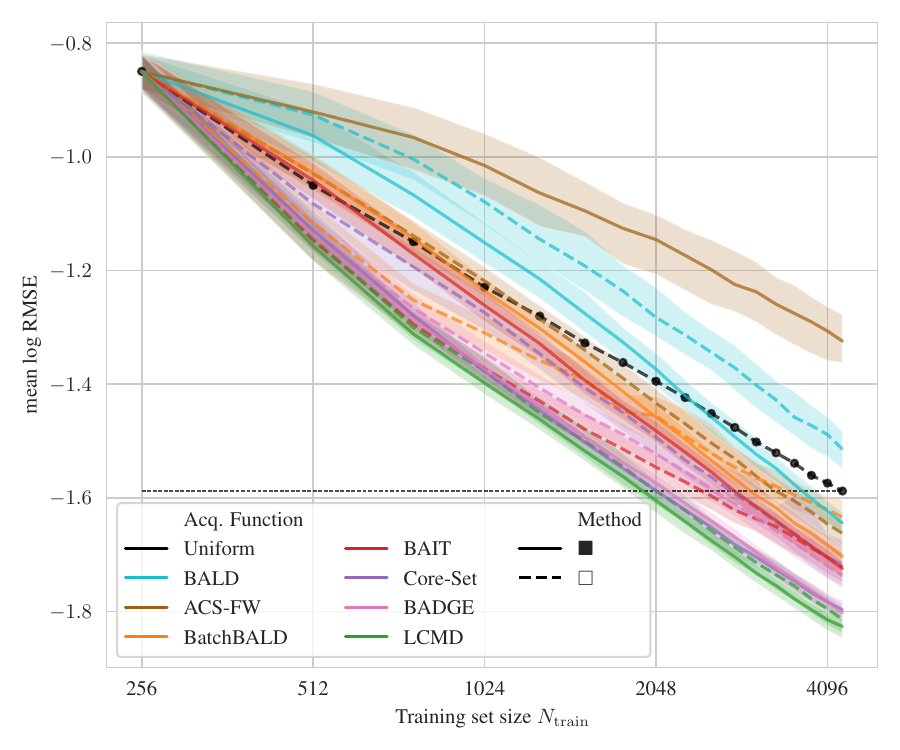}\hfill
        \includegraphics[width=0.50\linewidth]{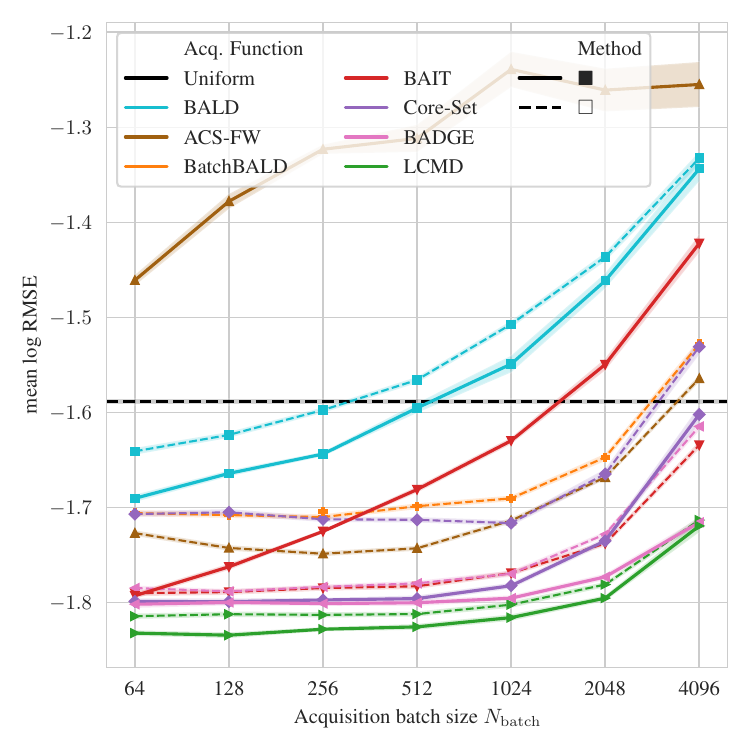} 
        \caption{Deep Neural Networks.} %
        \label{b3al:fig:main:relu}
    \end{subfigure}
    \begin{subfigure}[]{0.90\linewidth} %
        \centering
        \includegraphics[width=0.50\linewidth]{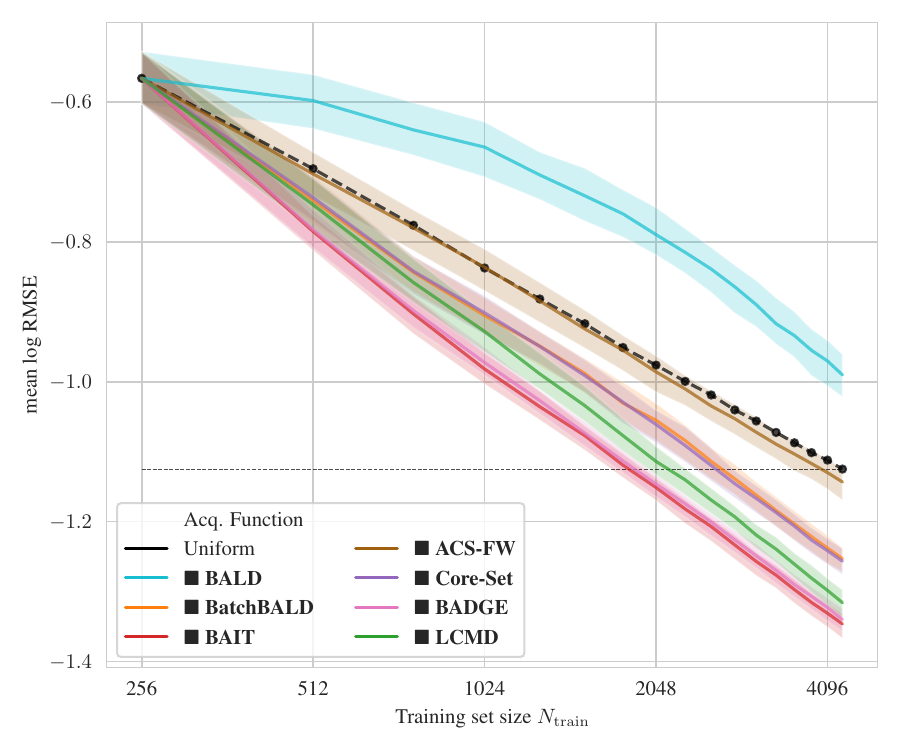}\hfill
        \includegraphics[width=0.50\linewidth]{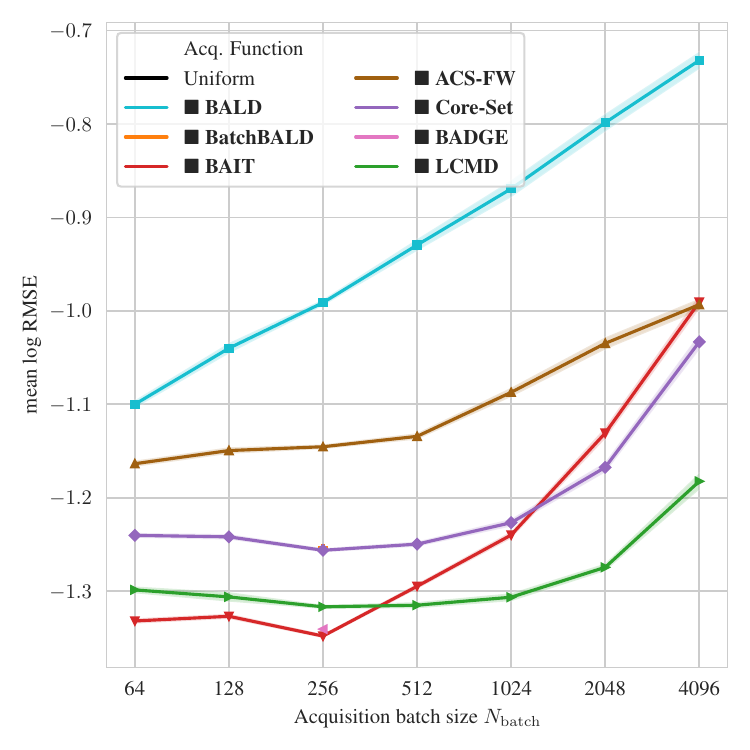} 
        \caption{Random Forests.} %
        \label{b3al:fig:main:rf}
    \end{subfigure} %
    \caption{\emph{Mean logarithmic RMSE over 15 regression datasets.}
    \textbf{(\subref{b3al:fig:main:relu})} For DNNs, we see that black-box $\blacksquare$ methods work as well as white-box $\square$ methods, and in most cases better, except for ACS-FW and BAIT.
    \textbf{(\subref{b3al:fig:main:rf})} For random forests (100 estimators) with the default hyperparameters from scikit-learn \citep{scikit-learn}, we see that black-box methods perform better than the uniform baseline, except for BALD, which uses top-\batchvar acquisition.
    In the appendix, see \Cref{b3al:tab:avg_performance_bb_vs_wb_relu} for average performance metrics and \Cref{b3al:fig:details:relu} and \ref{b3al:fig:details:rf} for plots with additional error metrics. Averaged over 20 trials.}%
    \label{b3al:fig:main:relu_rf}
\end{figure}

\begin{figure}[t]
    \centering
    \begin{subfigure}[]{0.90\linewidth}
        \centering
        \includegraphics[width=0.50\linewidth]{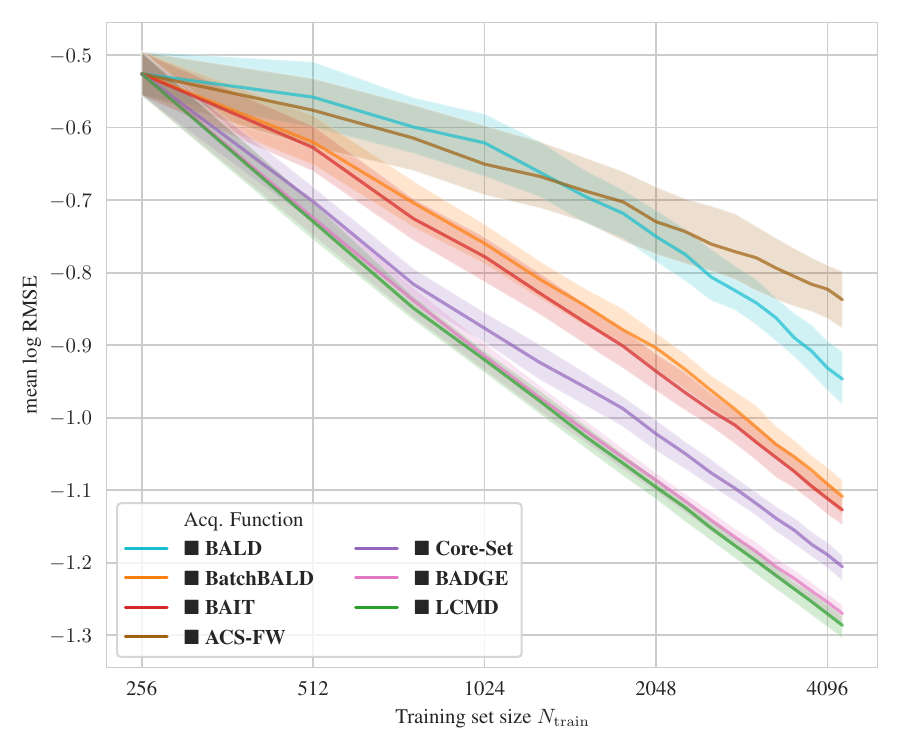}\hfill
        \includegraphics[width=0.50\linewidth]{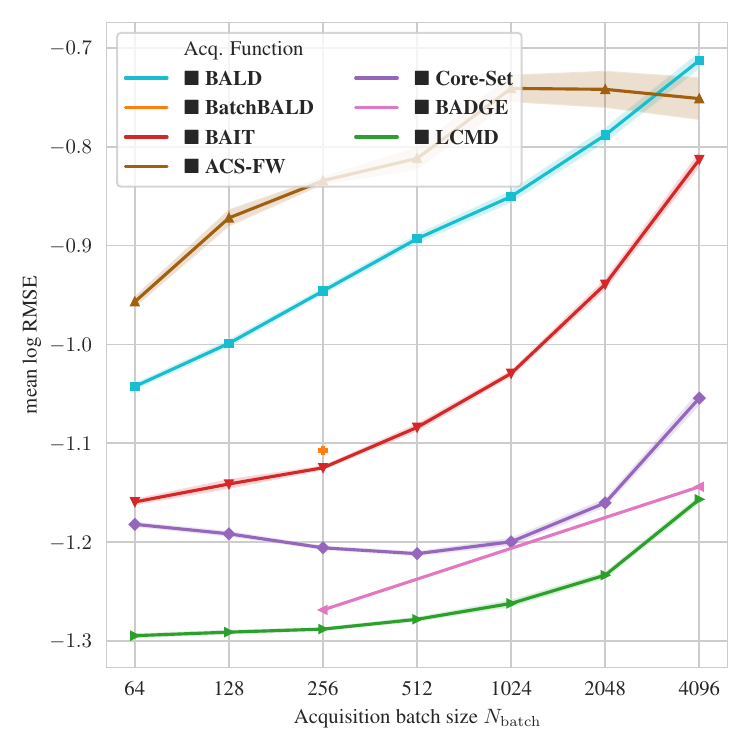} 
        \caption{Random Forests (Bagging).}
        \label{b3al:fig:main:bagging_rf}  
    \end{subfigure}
    \begin{subfigure}[]{0.90\linewidth}
        \centering
        \includegraphics[width=0.50\linewidth]{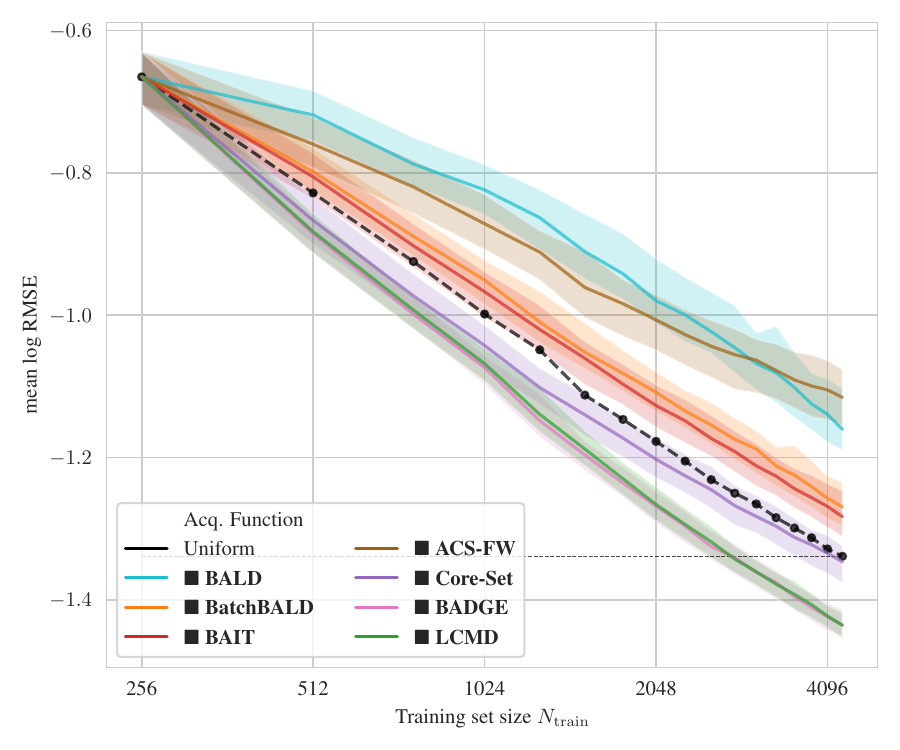}\hfill
        \includegraphics[width=0.50\linewidth]{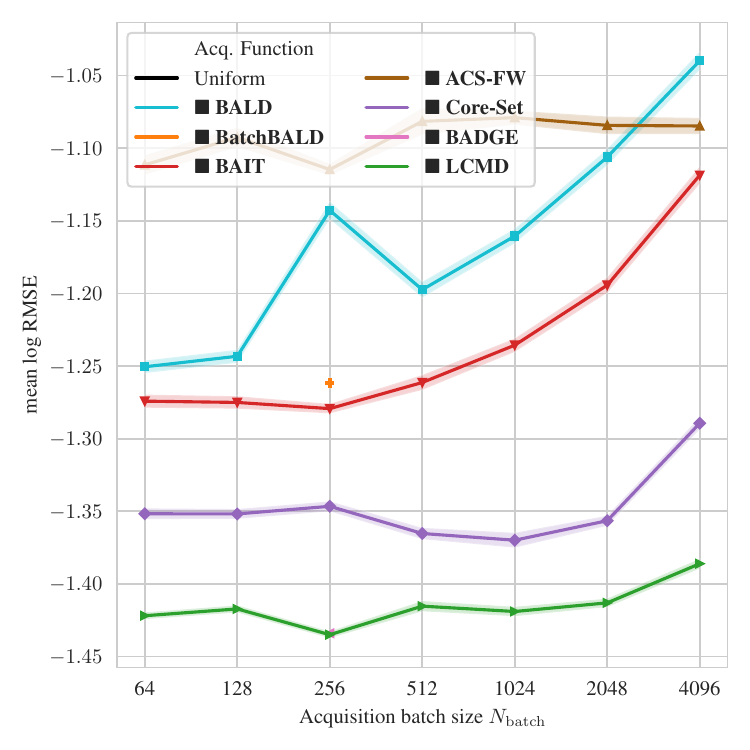}
        \caption{Gradient-Boosted Trees.} %
        \label{b3al:fig:main:ve-cat}
    \end{subfigure}
    \caption{\emph{Mean logarithmic RMSE over 15 regression datasets (cont'd).}
    For random forests using bagging \textbf{(\subref{b3al:fig:main:bagging_rf})} with 10 bootstrapped training sets, and
    for gradient-boosted trees \citep{dorogush2018catboost} \textbf{(\subref{b3al:fig:main:ve-cat})} with a \emph{virtual ensemble} of 20 members, we see that only a few of the black-box methods perform better than the uniform baseline: LCMD, BADGE and CoreSet. We hypothesize that the virtual ensembles and a bagged ensemble of random forests do not express as much predictive disagreement which leads to worse performance for active learning. In the appendix, see \Cref{b3al:tab:avg_performance_bb_vs_wb_relu} for average performance metrics and \Cref{b3al:fig:details:bagging_rf} and \ref{b3al:fig:details:ve-cat} for plots with additional error metrics. Averaged over 20 trials.}%
    \label{b3al:fig:main:bagging_rf_ve-cat}%
\end{figure}

\begin{figure}[t]
    \centering
    \includegraphics[width=\linewidth,trim=0 0 0 0.2cm,clip]{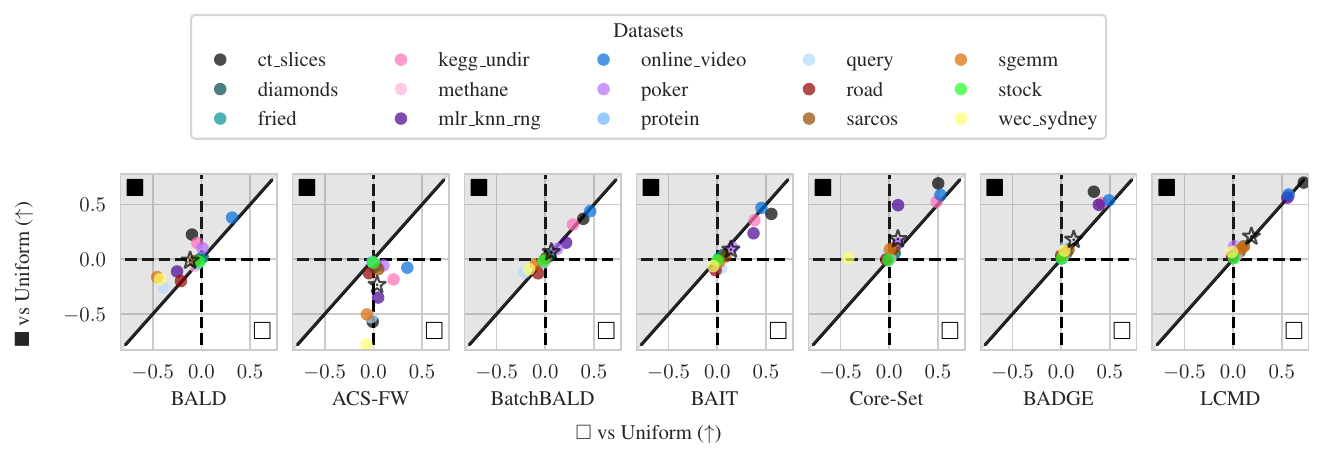}
    \caption{\emph{Average Logarithmic RMSE by regression datasets for DNNs: $\blacksquare$ vs $\square$ (vs Uniform).}
    Across acquisition functions, the performance of black-box methods is highly correlated with the performance of white-box methods, even though black-box methods make fewer assumptions about the model. 
    We plot the improvement of the white-box $\square$ method over the uniform baseline on the x-axis (so for datasets with markers right of the dashed vertical lines, the white-box method performs better than uniform) and the improvement of the black-box $\blacksquare$ method over the uniform baseline on the y-axis (so for datasets with markers above the dashed horizontal lines, the black-box method performs better than uniform).
    For datasets with markers in the \emph{$\blacksquare$ diagonal half}, the black-box method performs better than the white-box method.
    The average over all datasets is marked with a star $\star$.
    Surprisingly, on average over all acquisition rounds, the black-box methods perform slightly better than the white-box methods for all but ACS-FW and BAIT.
    In the appendix, see \Cref{b3al:fig:correlation_between_bb_vs_wb_methods_final} for the final acquisition round and \Cref{b3al:table:data_sets_source} for details on the datasets. Averaged over 20 trials.
    }
    \label{b3al:fig:correlation_between_bb_vs_wb_methods}
\end{figure}

\section{Empirical Validation}
\label{b3al:sec:results}

We follow the evaluation from \citet{holzmuller2022framework} and use their framework  to ease comparison.
This allows us to directly compare to several SotA methods in a regression setting, respectively their kernel-based analogues. 
Specifically, we compare to the following popular deep active learning methods:
BALD \citep{houlsby2011bayesian}, BatchBALD (\Cref{chapter:batchbald}), BAIT \citep{ash2021gone}, BADGE \citep{ash2020deep}, ACS-FW \citep{pinsler2019bayesian}, Core-Set \citep{sener2017active}/FF-Active \citep{geifman2017deep}, and LCMD \citep{holzmuller2022framework}.
We also compare to the random selection baseline (`Uniform'). 
We use 15 large tabular datasets from the UCI Machine Learning Repository \citep{Dua:2019} and the OpenML benchmark suite \citep{vanschoren2014openml} for our experiments: $sgemm$ (SGEMM GPU kernel performance); $wec\_sydney$ (Wave Energy Converters); $ct\_slices$ (Relative location of CT slices on axial axis);
$kegg\_undir$ (KEGG Metabolic Reaction Network - Undirected);
$online\_video$ (Online Video Characteristics and Transcoding Time);
$query$ (Query Analytics Workloads);
$poker$ (Poker Hand);
$road$ (3D Road Network - North Jutland, Denmark);
$mlr\_knn\_rng$;
$fried$;
$diamonds$;
$methane$;
$stock$ (BNG stock);
$protein$ (physicochemical-protein);
$sarcos$ (SARCOS data).
See \Cref{b3al:table:data_sets_source} in the appendix for more details.

\paragraph{Experimental Setup.} %
We use the same experimental setup and hyperparameters as \citet{holzmuller2022framework}.
We report the logarithmic RMSE averaged over 20 trials for each dataset and method.
For ensembles, we compute the performance for each ensemble member separately, enabling a fair comparison to the non-ensemble methods. Performance differences can thus be attributed to the acquisition function, rather than the ensemble. 
We used A100 GPUs with 40 GB of GPU memory.

\paragraph{Ensemble Size.} %
For deep learning, we use a small ensemble of 10 models, which is sufficient to achieve good performance. This ensemble size can be considered  `small' in the regression setting \citep{lazaro2010marginalized, zhang2013exploiting}, whereas in Bayesian Optimal Experiment Design much higher sample counts are regularly used \citep{foster2021deep}. In many cases, training an ensemble of regression models is still fast and could be considered cheap compared to the cost of acquiring additional labels.
For non-differentiable models, we experiment with random forests \citep{breiman2001random}, using the different trees as ensemble members, and a virtual ensemble of gradient-boosted decision trees \citep{prokhorenkova2017catboost}. 
For random forests, we use the implementation provided in scikit-learn \citep{scikit-learn} with default hyperparameters, that is using 100 trees per forest. We use the predictions from each tree as a virtual ensemble member. We do not perform any hyperparameter tuning.
We also report results for random forests with bagging, where we train a real ensemble of 10 random forests.
For gradient-boosted decision trees, we use a virtual ensemble of up to 20 members with early stopping using a validation set\footnote{If the virtual ensemble creation fails because there are no sufficiently many trees due to early stopping, we halve the ensemble size and retry. This was only needed for the \emph{poker} dataset.}. We use the implementation in CatBoost \citep{dorogush2018catboost}.
We do not perform any hyperparameter tuning.

\paragraph{Black-Box vs White-Box Deep Active Learning.} In \Cref{b3al:fig:main:relu} and \ref{b3al:fig:correlation_between_bb_vs_wb_methods}, we see that \BBBAL is competitive with white-box active learning, when using BALD, BatchBALD, BAIT, BADGE, and Core-Set. On average, \BBBAL outperforms the white-box methods on the 15 datasets we analyzed (excluding ACS-FW and BAIT).  
We hypothesize that this is due to the implicit Fisher information approximation in the white-box methods we have examined (\Cref{chapter:fim}), which is not as accurate in the low data regime as the more explicit approximation in \BBBAL via ensembling.
On the other hand, it seems that the black-box methods suffer from the low feature space dimensionality, as they are much closer to BALD.

\paragraph{Why can Black-Box Methods Outperform White-Box Methods?} %
Following \S\ref{b3al:sec:methodology},
white-box and black-box methods are based on kernels, which can be seen as different \emph{approximations} of the predictive covariance kernel. White-box methods implicitly assume that the predictive covariance kernel is well approximated by the Fisher information kernel and the gradient kernel (\Cref{chapter:fim}). However, \citet{long2021multimodal} demonstrated that this assumption does not always hold, particularly in low data regimes, where a Gaussian might not approximate the parameter distribution well. Instead, \citet{long2021multimodal} suggests using a multimodal distribution.
In these situations, methods that employ ensembling, such as \BBBAL, to approximate the predictive covariance kernel are more robust. The different ensemble members can reside in different modes of the parameter distribution, allowing black-box methods to outperform their white-box counterparts.

\paragraph{Non-Differentiable Models.} %
In \Cref{b3al:fig:main:rf}, \ref{b3al:fig:main:bagging_rf}, and \ref{b3al:fig:main:ve-cat}, we observe that \BBBAL is effective for non-differentiable models, including random forests and gradient-boosted decision trees. %
BALD for non-differentiable models can be considered equivalent to QbC \citep{seung1992query}, while BatchBALD for non-differentiable models can be viewed as equivalent to QbC with batch acquisition \citep{nguyen2012efficient}.
For random forests, all methods except BALD (using top-k selection) outperform uniform acquisition. However, for random forests with bagging and gradient-boosted decision trees, \BBBAL surpasses random acquisition only when employing LMCD and BADGE. This may be attributed to the reduced disagreement within a virtual ensemble for gradient-boosted decision trees and between distinct random forests. In particular, random forests with bagging appear to support this explanation, as a single random forest seems to exhibit more disagreement among its individual trees than an ensemble of random forests with bagging does between different forests. This is evident in the superior overall active learning performance of the single random forest compared to the ensemble of random forests with bagging, c.f.\ \Cref{b3al:fig:main:rf} and \ref{b3al:fig:main:bagging_rf}.

\section{Discussion}
\label{b3al:sec:conclusion}
In this chapter, we have demonstrated the effectiveness of a simple extension to kernel-based methods that utilizes empirical predictions rather than gradient kernels.
This modification enables black-box batch active learning with good performance.
Importantly, \BBBAL also generalizes to non-differentiable models, an area that has received limited attention as of late.

This result also partially answer one of the research questions from the previous chapter (\Cref{chapter:fim}): how do prediction-based methods compare to parameter-based ones?
We find that for regression the prediction-based methods are competitive with the parameter-based methods in batch active learning.

The main limitation of our proposed approach lies in the acquisition of a sufficient amount of empirical predictions. 
This could be a challenge, particularly when using deep ensembles with larger models or non-differentiable models that cannot be parallelized efficiently. 
Our experiments using virtual ensembles indicate that the diversity of the ensemble members plays a crucial role in determining the performance.

Likewise, the main limitation of this chapter and the empirical comparisons is that we only consider regression tasks.
Extending the results to classification is an important direction for future work.

}

\begin{savequote}[75mm]
	What gets sharper the more you use it?
\end{savequote}
\chapter{\conclusiontitle}
\label{chapter:conclusion}
\label{glue:conclusion}

In this thesis, we set out to address several research questions focused on active learning and data subset selection in deep learning, with an emphasis on information-theory intuitions:
\begin{enumerate}
    \item How can uncertainty quantification, specifically aleatoric and epistemic uncertainty, be better understood and applied in the context of active learning and active sampling?
    \item How can a deeper understanding of the theoretical foundations of active learning and active sampling contribute to the progress of the field and improve the practical application of these techniques?
    \item How can the cost of gathering and labeling data be reduced, and how can training be sped up in a principled fashion?
    \item What are the connections between different active learning and active sampling approaches, and how can information theory be used to unify these approaches?
\end{enumerate}
Our main findings and contributions include the unifying perspective provided by information-theory intuitions, the discovery that many existing methods can be explained using the same framework, and the effectiveness of simple methods in various applications.

Regarding the first research question, we have demonstrated that by examining aleatoric and epistemic uncertainty in more detail, we can propose new baselines for uncertainty quantification using single forward-pass deep neural networks (\Cref{chapter:ddu}). This approach allows us to quantify epistemic uncertainty well and achieve competitive results in active learning without having to be Bayesian.

In addressing the second research question, this thesis contributes to a deeper understanding of the theoretical foundations of active learning and active sampling by providing a unified framework based on information theory (\Cref{chapter:fim,chapter:practical_it_notation}). This framework helps researchers better understand the trade-offs and connections between different approaches, ultimately improving the practical application of these techniques.

To answer the third research question, we have explored various methods for reducing the cost of gathering and labeling data (\Cref{chapter:ddu,chapter:batchbald,chapter:stochacq,chapter:pobal,chapter:bbbal}) and speeding up training in a principled fashion (\Cref{chapter:rholoss,chapter:dldd}). While we have provided a principled approach for batch acquisition in \Cref{chapter:batchbald}, it is too slow at larger acquisition batch sizes and can suffer from issues related to the estimation of joint predictives for larger batch acquisition sizes. 
On the other hand, 
our conceptually simpler and effective methods, such as stochastic batch acquisition (\Cref{chapter:stochacq}) and single forward-pass deterministic methods (\Cref{chapter:ddu}), have shown promising results and have the potential to benefit practitioners in a wide range of applications.

Finally, in addressing the fourth research question, we have identified connections between different active learning and active sampling approaches by using information theory to unify these methods (\Cref{chapter:fim,chapter:bbbal}).
This unifying perspective allows us to better understand the relationships between various techniques and access the principled framework for reasoning about uncertainty and informativeness that information theory provides.

Nonetheless, there are some obvious limitations to our research. One of the main challenges is improving joint predictions for multiple samples (\Cref{chapter:obi}). Additionally, a more satisfying approach for classification tasks and beyond is needed when using kernel-based methods. 
Finally, there are various related areas such as active testing, active inference, and exploration in reinforcement learning that are yet to be connected in more detail to active learning and active sampling.

Future research directions could include developing more effective methods for joint predictions and batch selection, exploring other alternative approaches for classification, and investigating the quality of the chosen approximations. 
We detail several other specific research questions in various chapters: in particular in \Cref{chapter:obi,chapter:fim}.

In conclusion, this thesis has advanced our understanding of active learning and data subset selection in deep learning by providing a unifying perspective based on information-theory intuitions. We hope this thesis has also contributed to the reader's understanding and that our contributions will inspire further research and help practitioners adopt effective techniques in their work.

\changelocaltocdepth{2}

\startappendices

	\begin{savequote}[75mm]
		Can you imagine what I would do if I could do all I can?
		\qauthor{Sun Tzu, The Art of War}
	\end{savequote}
	\chapter{\causalbaldtitle}
	\label{chapter:causalbald}
	{
		\graphicspath{{./causalbald/}}

\renewcommand{\E}{\mathop{\mathbb{E}}}
\newcommand{\cov}{\mathop{\mathrm{Cov}}}
\newcommand{\var}{\mathop{\mathrm{Var}}}
\newcommand{\mi}{\mathrm{I}}
\newcommand{\ent}{\mathrm{H}}

\newcommand{\D}{\mathcal{D}}

\newcommand{\tb}{\mathbf{t}}

\newcommand{\tf}{\mathrm{t}}
\newcommand{\tcf}{\tf^{\prime}}
\newcommand{\T}{\mathrm{T}}

\newcommand{\yzero}{\y^0}
\newcommand{\Yzero}{\Y^0}
\newcommand{\yone}{\y^1}
\newcommand{\Yone}{\Y^1}
\newcommand{\Ydot}{\dot{\Y}}
\newcommand{\yt}{\y^\tf}
\newcommand{\ytcf}{\y^{\tcf}}
\newcommand{\Yt}{\Y^\tf}
\newcommand{\Ytcf}{\Y^{\tcf}}

\newcommand{\muh}{\widehat{\mu}}
\newcommand{\tauhw}{\widehat{\tau}_{\w}}

\newcommand{\w}{\bm{\omega}}
\newcommand{\W}{\bm{\Omega}}

\newcommand{\bftab}{\fontseries{b}\selectfont}
\newcommand{\panos}[1]{{\color{purple}\texttt{#1}}}
\newcommand{\etal}{\textit{et al.}}

\newcommand{\method}{Causal-BALD}
\newcommand{\mubald}{$\mu$BALD}
\newcommand{\rhobald}{$\rho$BALD}
\newcommand{\mupibald}{$\mu\pi$BALD}
\newcommand{\murhobald}{$\mu\rho$BALD}
\newcommand{\taubald}{$\tau$BALD}

\label{causalbald:sec:intro}
How will a patient's health be affected by taking a medication \citep{Criado-Perez2020invisible}? 
How will a user's question be answered by a search recommendation \citep{noble2018algorithms}?
Insight into these questions can be gained by learning about personalized treatment effects.
Estimating personalized treatment effects from observational data is essential in situations where experimental designs are infeasible, unethical or expensive.
Observational data represent a population of individuals described by a set of pre-treatment covariates (age, blood pressure, socioeconomic status), an assigned treatment (medication, no medication), and a post-treatment outcome (severity of migraines).
An ideal personalized treatment effect is the difference between the post-treatment outcome had the individual been treated, and the outcome had they not been treated. But, it is impossible to observe both outcomes for an individual, so the difference must instead be computed between populations.
Therefore, in the common setting of binary treatments, data is partitioned into the \emph{treatment group} (individuals that received the treatment) and the \emph{control group} (individuals who did not).
The personalized treatment effect is then given by the expected difference in outcomes between treated and controlled individuals who share the same (or similar) measured covariates (difference between solid lines in the middle pane of \cref{causalbald:fig:observational_data}).

Increasingly, pre-treatment covariates are being assembled from high-dimensional, heterogeneous measurements such as medical images and electronic health records \citep{sudlow2015uk}. 
Deep learning methods have been shown capable of learning personalized treatment effects from such data \citep{shalit2017estimating, shi2019adapting,jesson2021quantifying}.
However, a key problem in deep learning is data efficiency.
While modern methods are capable of impressive performance, they need a significant amount of labeled data.
Acquiring labeled data can be expensive, often requiring specialist knowledge or an invasive procedure to determine the outcome. 
Therefore, it is desirable to minimize the amount of labeled data needed to obtain a well-performing model.
Active learning provides a principled framework to address this concern \citep{cohn1996active}.
In active learning for treatment effects \citep{deng2011active, sundin2019active}, a model is trained on available labeled data consisting of covariates, assigned treatments, and acquired outcomes.
The model predictions are then used to select the most informative examples from a set of data consisting of only covariates and treatment indicators.
Outcomes are then acquired, e.g. by performing a biopsy, for the selected patients and the model is retrained and evaluated.
This process is repeated until either a satisfactory performance level is achieved, or the labeling budget is exhausted.

\begin{figure}[ht]
    \centering
    \includegraphics[width=0.40\textwidth]{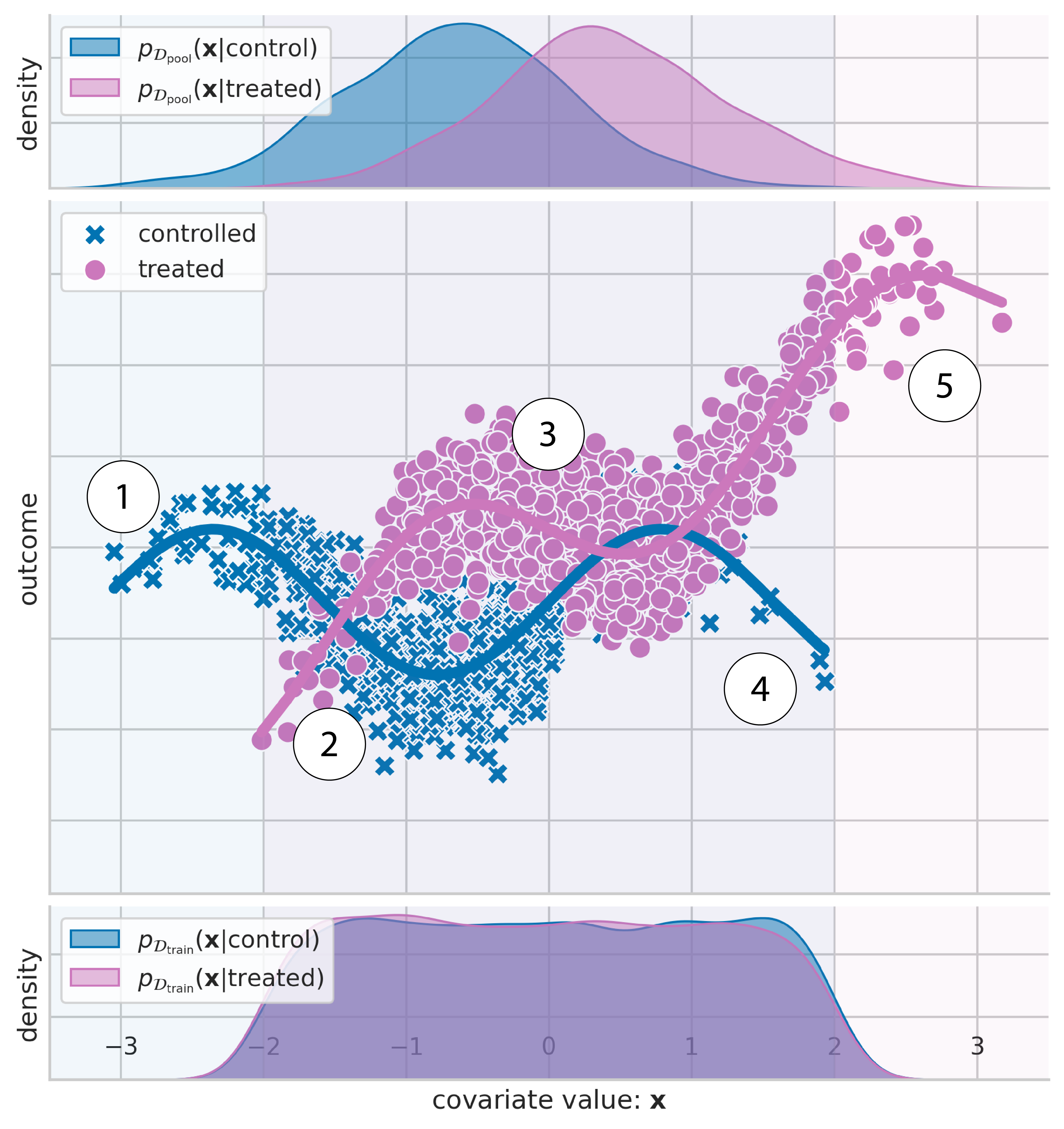}
    \caption{
        Observational data. 
        Top: data density of treatment (right) and control (left) groups. 
        Middle: observed outcome response for treatment (circles) and control (x's) groups.
        Bottom: data density for active learned training set after a number of acquisition steps.
    }
    \label{causalbald:fig:observational_data}
    \end{figure}

At first sight this might seem simple; however, active learning induces bias resulting in divergence between the distribution of the acquired training data and the distribution of the pool set data \citep{farquhar2020statistical}.
In the context of learning causal effects, such bias has important positive and negative consequences.
For example, while random acquisition active learning results in an unbiased sample of the training data, it can lead to over-allocation of resources to the mode of the data at the expense of learning about underrepresented data. 
Conversely, while biasing acquisitions toward lower density regions of the pool data can be desirable, it can also lead to acquisitions for which we cannot know the treatment effect, which could in turn lead to uninformed, potentially harmful, personalized decisions.

To gain insight into how biasing the acquisition of training data can be beneficial for learning treatment effects, consider a key difference between experimental and observational data: the treatment assignment mechanism is not available for observational data.
This means that there may be unobserved variables that affect treatment assignment (an untestable condition), but also that the relative proportion of individuals treated to those controlled can vary across different subpopulations of the data. 
The later point is illustrated in \cref{causalbald:fig:observational_data}, where there are relatively equal proportions of treated and controlled examples for data in region 3, but the proportions become less balanced as we move to either the left or the right.
In extreme cases, say if a group described by some covariate values were systematically excluded from treatment, the treatment effect for that group \emph{cannot be known} \citep{petersen2012diagnosing}.
This is illustrated in \cref{causalbald:fig:observational_data} by region 1, where there are only controlled examples, and by region 5, where there are only treated examples.
In the language of causal inference, the necessity of seeing both treated and untreated examples for each subpopulation corresponds to satisfaction of the overlap (or positivity) assumption (see \ref{causalbald:asm:overlap}).
Regions 2 and 4 of \cref{causalbald:fig:observational_data} are interesting as while either the treated or control group are underrepresented, there may still be sufficient coverage to learn treatment effects.

We propose that the acquisition of unlabeled data should focus on exploring all regions with sufficient overlap, but not areas with no overlap.
The bottom pane of \cref{causalbald:fig:observational_data} imagines what a resulting training set distribution could look like at an intermediate active learning step.
It is not trivial to design such acquisition functions: naively applying active learning acquisition functions results in suboptimal and sample inefficient acquisitions of training examples, as we show below. 
To this end, we develop an epistemic uncertainty aware method for active learning of personalized treatment effects from high dimensional observational data. We demonstrate the performance of the proposed acquisition strategies on a synthetic and semisynthetic datasets.

\section{Background}

\label{causalbald:sec:background}

\subsection{Estimation of Personalized Treatment-Effects}
Personalized treatment-effect estimation seeks to know the effect of a treatment $\T \in \mathcal{T}$ on the outcome $\Y \in \mathcal{Y}$ for individuals described by covariates $\X \in \mathcal{X}$. 
In this chapter, we consider the random variable (r.v.) $\T$ to be binary ($\mathcal{T} = \{0, 1\}$), the r.v. $\Y$ to be part of a bounded set $\mathcal{Y}$, and $\X$ to be a multi-variate r.v. of dimension $d$ ($\mathcal{X} = \mathbb{R}^d$).
Under the Neyman-Rubin causal model \citep{neyman1923applications, rubin1974estimating}, the individual treatment effect (ITE) for a person $u$ is defined as the difference in potential outcomes $\Yone(u) - \Yzero(u)$, where the r.v. $\Yone$ represents the potential outcome were they \emph{treated}, and the r.v. $\Yzero$ represents the potential outcome were they \emph{controlled} (not treated).
Realizations of the random variables $\X$, $\T$, $\Y$, $\Yzero$, and $\Yone$ are denoted by $\x$, $\tf$, $\y$, $\yzero$, and $\yone$, respectively.

The ITE is a fundamentally unidentifiable quantity, so instead we look at the expected difference in potential outcomes for individuals described by $\X$, or the Conditional Average Treatment Effect (CATE): $\tau(\x) \equiv \E[\Yone - \Yzero \mid \X = \x]$ \cite{abrevaya2015estimating}.
The CATE is identifiable from an observational dataset $\D = \left\{ (\x_i, \tf_i, \y_i)\right\}_{i=1}^n$ of samples $(\x_i, \tf_i, \y_i)$ from the joint empirical distribution $P_{\D}(\X, \T, \Yzero, \Yone)$, under the following three assumptions:
\begin{assumption}
    (Consistency) $\y = \tf \yt + (1 - \tf) \y^{1 - \tf}$, i.e. an individual's observed outcome $\y$ given assigned treatment $\tf$ is identical to their potential outcome $\yt$.
    \label{causalbald:asm:consistency}
\end{assumption}
\begin{assumption}
    (Unconfoundedness) $(\Yzero, \Yone) \independent \T \mid \X$.
    \label{causalbald:asm:confounded}
\end{assumption}
\begin{assumption}
    (Overlap) $0 < \pi_{\tf}(\x) < 1: \forall \tf \in \mathcal{T}$,
    \label{causalbald:asm:overlap}
\end{assumption}
where $\pi_{\tf}(\x) \equiv \mathrm{P}(\T = \tf \mid \X = \x)$ is the \textbf{propensity for treatment} for individuals described by covariates $\X = \x$ \cite{rubin1974estimating}.
When these assumptions are satisfied, $\widehat{\tau}(\x) \equiv \E[Y \mid \T = 1, \X = \x] - \E[Y \mid \T = 0, \X = \x]$ is an identifiable, unbiased estimator of $\tau(\x)$.

A variety of parametric \citep{robins2000marginal,tian2014simple,shalit2017estimating} and non-parametric estimators \citep{hill2011bayesian,xie2012estimating,alaa2017bayesian,gao2020minimax} have been proposed for CATE. 
Here, we focus on parametric estimators for compactness. 
Parametric CATE estimators assume that outcomes $\y$ are generated according to a likelihood $p_{\w}(\y \mid \x, \tf)$, given measured covariates $\x$, observed treatment $\tf$, and model parameters $\w$. 
For continuous outcomes, a Gaussian likelihood can be used: $\mathcal{N}(\y \mid \muh_{\w}(\x, \tf), \widehat{\sigma}_{\w}(\x, \tf))$. 
For discrete outcomes, a Bernoulli likelihood can be used: $\mathrm{Bern}(\y \mid \muh_{\w}(\x, \tf))$. 
In both cases, $\muh_{\w}(\x, \tf)$ is a parametric estimator of $\E[Y \mid \T = \tf, \X = \x]$, which leads to: $\widehat{\tau}_{\w}(\x) \equiv \muh_{\w}(\x, 1) - \muh_{\w}(\x, 0)$, a parametric CATE estimator.

Bayesian inference over the model parameters $\w$ treated as stochastic instances of the random variable $\W \in \mathcal{W}$ has been shown by \citet{jesson2020identifying} to yield models capable of quantifying when assumption \ref{causalbald:asm:overlap} (overlap) does not hold, or when there is insufficient knowledge about the treatment effect $\tau(\x)$ because the observed value $\x$ lies far from the support of $P_{\D}(\X, \T, \Yzero, \Yone)$. 
Such methods seek to enable sampling from the posterior distribution of the model parameters given the data, $p(\W \mid \D)$. 
Each sample, $\w \sim p(\W \mid \D)$ induces a unique CATE function $\widehat{\tau}_{\w}(\x)$. 
Epistemic uncertainty is a measure of ``disagreement'' between the functions at a given value $\x$ \citep{kendall2017uncertainties}. 
\citet{jesson2020identifying} propose $\var_{\w \sim p(\W \mid \D)}(\muh_{\w}(\x, 1) - \muh_{\w}(\x, 0))$ as a measure of epistemic uncertainty in the CATE.

\subsection{Active Learning}

We use a slightly different setup here than in the other chapters.
Formally, an active learning setup consists of an unlabeled dataset $\Dpool = \{\x_i\}_{i = 1}^{n_{\mathrm{pool}}}$, a labeled training set $\Dtrain = \{\x_i, \y_i\}_{i = 1}^{n_{\mathrm{train}}}$, and a predictive model with likelihood $p_{\w}(\y \mid \x)$ parameterized by $\w \sim p(\W \mid \Dtrain)$. 
It is further assumed that an oracle exists to provide outcomes $\y$ for any data point in $\Dpool$.
After model training, a batch of data $\{ \x^*_i \}_{i=1}^{b}$ is selected from $\Dpool$ using an acquisition function $a$ according to the informativeness of the batch.

We depart from the standard active learning setting and include the treatment:
for active learning of treatment effects, we define $\Dpool = \{\x_i, \tf_i\}_{i = 1}^{n_{\mathrm{pool}}}$, a labeled training set $\Dtrain = \{\x_i, \tf_i, \y_i\}_{i = 1}^{n_{\mathrm{train}}}$, and a predictive model with likelihood $p_{\w}(\y \mid \x, \tf)$ parameterized by $\w \sim p(\W \mid \Dtrain)$.
The acquisition function takes as input $\Dpool$ and returns a batch of data $\{\x_i, \tf_i\}_{i=1}^{b}$ which are labeled using an oracle and added to $\Dtrain$. 
We focus on examining when there is only access to the observed treatments $\{\tf_i\}_{i = 1}^{n_{\mathrm{pool}}}$: scenarios where treatment assignment is not possible.

An intuitive way to define informativeness is using the estimated uncertainty of our model.
In general, we can distinguish two sources of uncertainty: epistemic and aleatoric uncertainty \citep{der2009aleatory,kendall2017uncertainties}.
Epistemic (or model) uncertainty, arises from uncertainty in the model parameters.
This is for example caused by the model not having seen similar data points before, and therefore is unclear what the correct label would be.
As before, we focus on using epistemic uncertainty to identify informative points to acquire the label for.

\begin{figure*}[ht!]
     \centering
     \begin{subfigure}[]{0.28\textwidth}
         \centering
         \includegraphics[width=\textwidth]{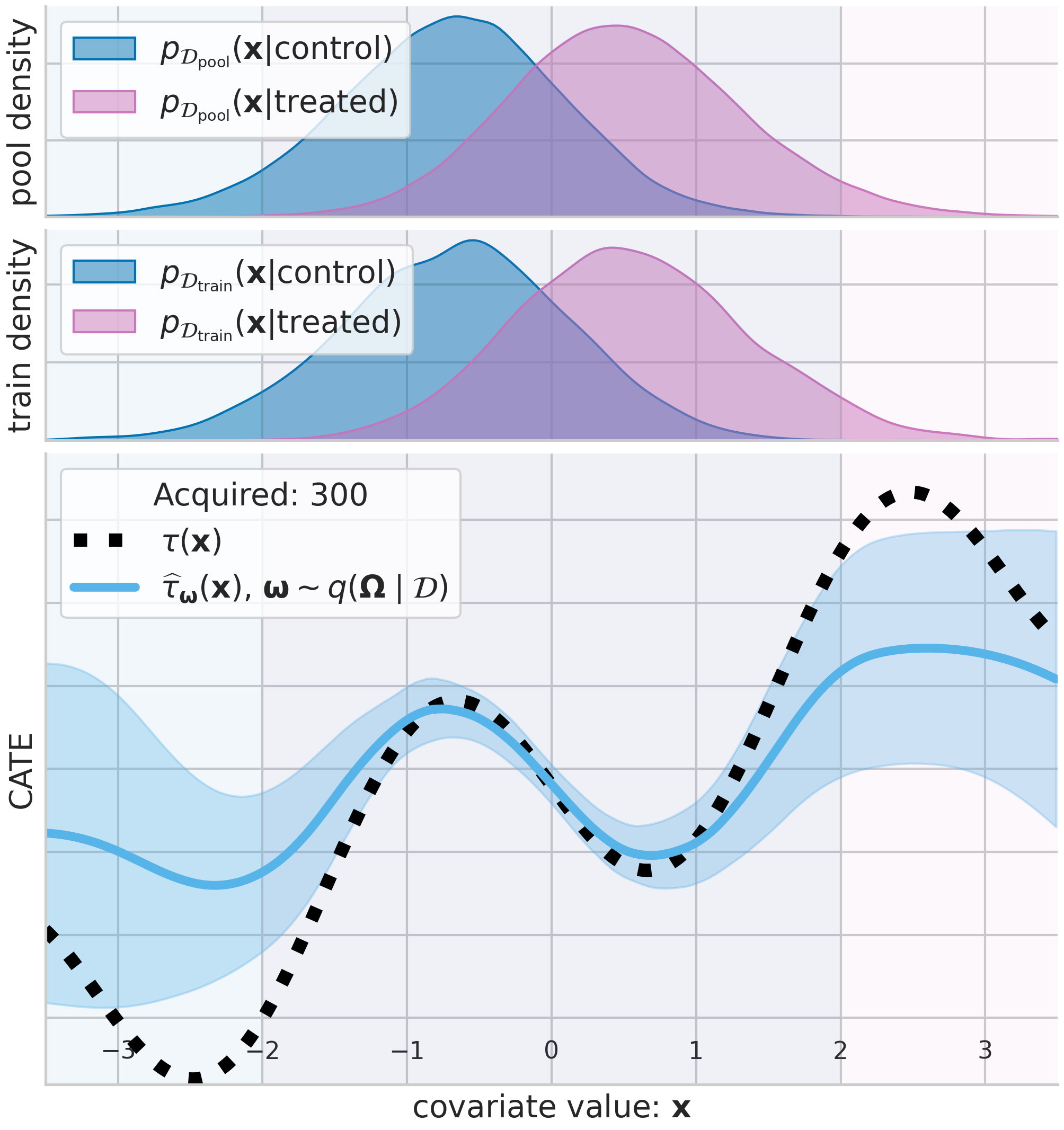}
         \caption{Random}
         \label{causalbald:fig:bias_random}
     \end{subfigure}
     \hfill
     \begin{subfigure}[]{0.28\textwidth}
         \centering
         \includegraphics[width=\textwidth]{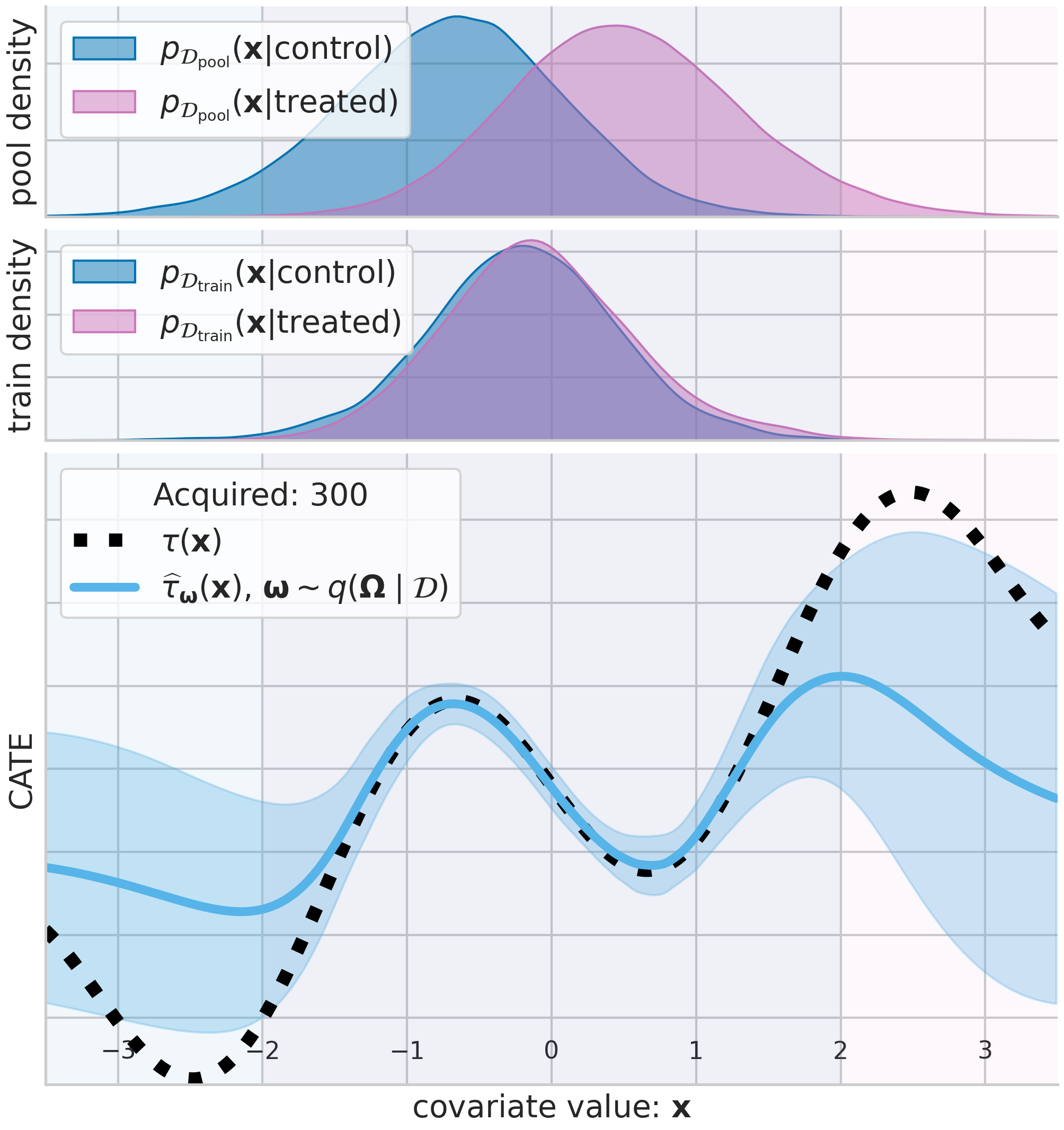}
         \caption{Propensity}
         \label{causalbald:fig:bias_pi}
     \end{subfigure}
     \hfill
     \begin{subfigure}[]{0.28\textwidth}
         \centering
         \includegraphics[width=\textwidth]{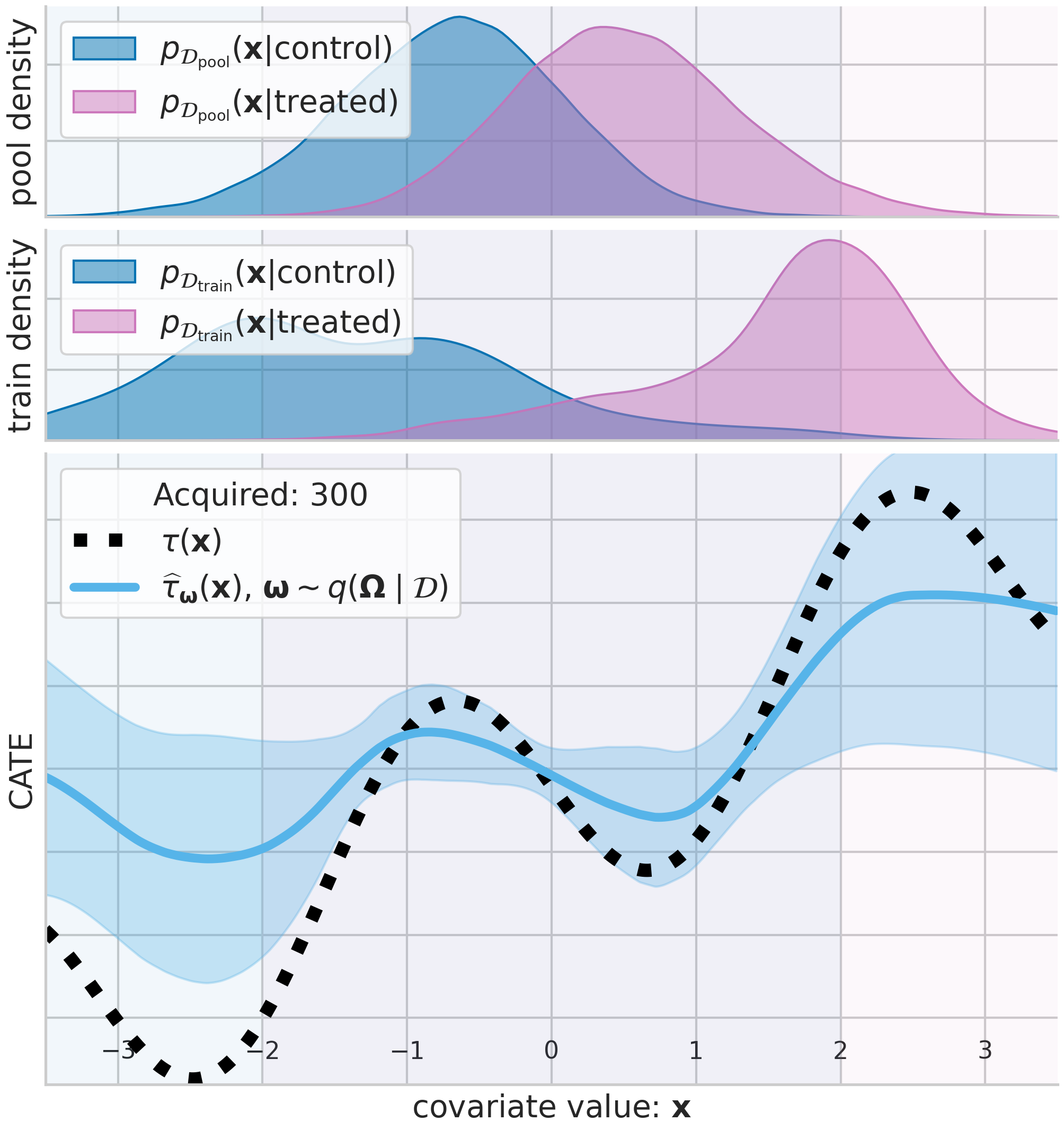}
         \caption{$\tau$BALD}
         \label{causalbald:fig:bias_tau}
     \end{subfigure}
    \caption{Naive acquisition functions: How the training set is biased and how this effects the CATE function with a fixed budget of 300 acquired points.}
    \label{causalbald:fig:naive_acquisitions}
    \end{figure*}

\section{Methods}
\label{causalbald:sec:methods}
In this section, we introduce several acquisition functions, analyze how they bias the acquisition of training data, and show the resulting CATE functions learned from such training data.
We are interested in acquisition functions conditioned on realizations of both $\x$ and $\tf$:
\begin{equation*}
        a(\Dpool, p(\W \mid \Dtrain)) =
        \argmax_{\{\x_i, \tf_i\}_{i=1}^{b} \subseteq \Dpool} \mi(\bullet \mid \{ \x_i, \tf_i \}, \Dtrain),
\end{equation*}
where $\mi(\bullet \mid \x, \tf, \Dtrain)$ is a measure of disagreement between parametric function predictions given $\x$ and $\tf$ over samples $\w \sim p(\W \mid \D)$. 
We make \cref{causalbald:asm:consistency,causalbald:asm:confounded} (consistency, and unconfoundedness).
We relax \cref{causalbald:asm:overlap} (overlap) by allowing for its violation over subsets of the support of $\Dpool$.
We present all theorems, proofs, and detailed assumptions in \cref{causalbald:sec:theoretical_results}.
\subsection{Naive Acquisition Functions, Training Data Bias, and the Effect on the Estimated CATE Function.}

To motivate Causal--BALD, we first look at a set of naive acquisition functions. The simplest function selects data points uniformly at random from $\Dpool$ and adds them to $\Dtrain$.
In \cref{causalbald:fig:bias_random} we have acquired 300 such examples from a synthetic dataset and trained a deep-kernel Gaussian process \citep{van2021improving} on those labeled examples.
Comparing the top two panes, we see that $\Dtrain$ (middle) contains an unbiased sample of the data in $\Dpool$ (top).
However, in the bottom pane we see that while the CATE estimator is accurate (and certain) near the modes of $\Dpool$, it becomes less accurate as we move to lower density regions.
In this way random acquisition reflects the biases inherent in $\Dpool$ and over-allocates resources to the modes of the distribution.
If the mode were to coincide with a region of non-overlap, the function would most frequently acquire uninformative examples.

Next, we look at using propensity scores to bias acquisition toward regions where the overlap assumption is satisfied.
\begin{definition}
    Propensity based acquisition
    \begin{equation}
        \mi(\widehat{\pi}_{\tf} \mid \x, \tf, \Dtrain) \equiv 1 - \widehat{\pi}_{\tf}(\x)
        \label{causalbald:eq:propensity-acquisition}
    \end{equation}
\end{definition}
Intuitively, this function prefers points where the propensity for observing the counterfactual is high.
We are considering the setup where $\Dpool$ contains observations of both $\X$ and $\T$, so it is straightforward to train an estimator for the propensity, $\widehat{\pi}_{\tf}(\x)$.
\Cref{causalbald:fig:bias_pi} shows that while propensity score acquisition matches the treated and control densities in $\Dtrain$, it still biases acquisition towards the modes of $\Dpool$.

\begin{figure*}[ht!]
     \centering
     \begin{subfigure}[]{0.28\textwidth}
         \centering
         \includegraphics[width=\textwidth]{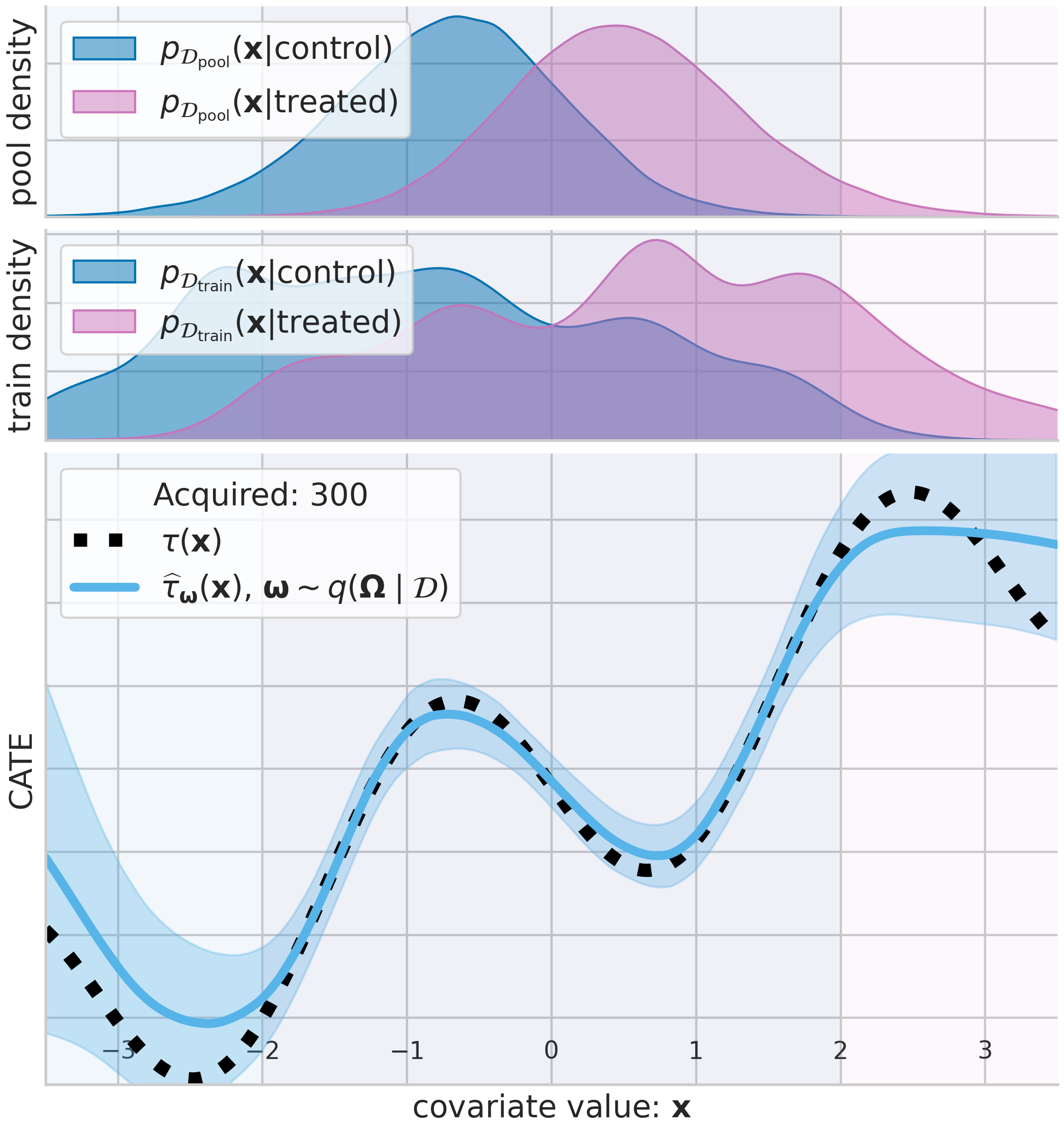}
         \caption{$\mu$BALD}
         \label{causalbald:fig:bias_mu}
     \end{subfigure}
     \hfill
     \begin{subfigure}[]{0.28\textwidth}
         \centering
         \includegraphics[width=\textwidth]{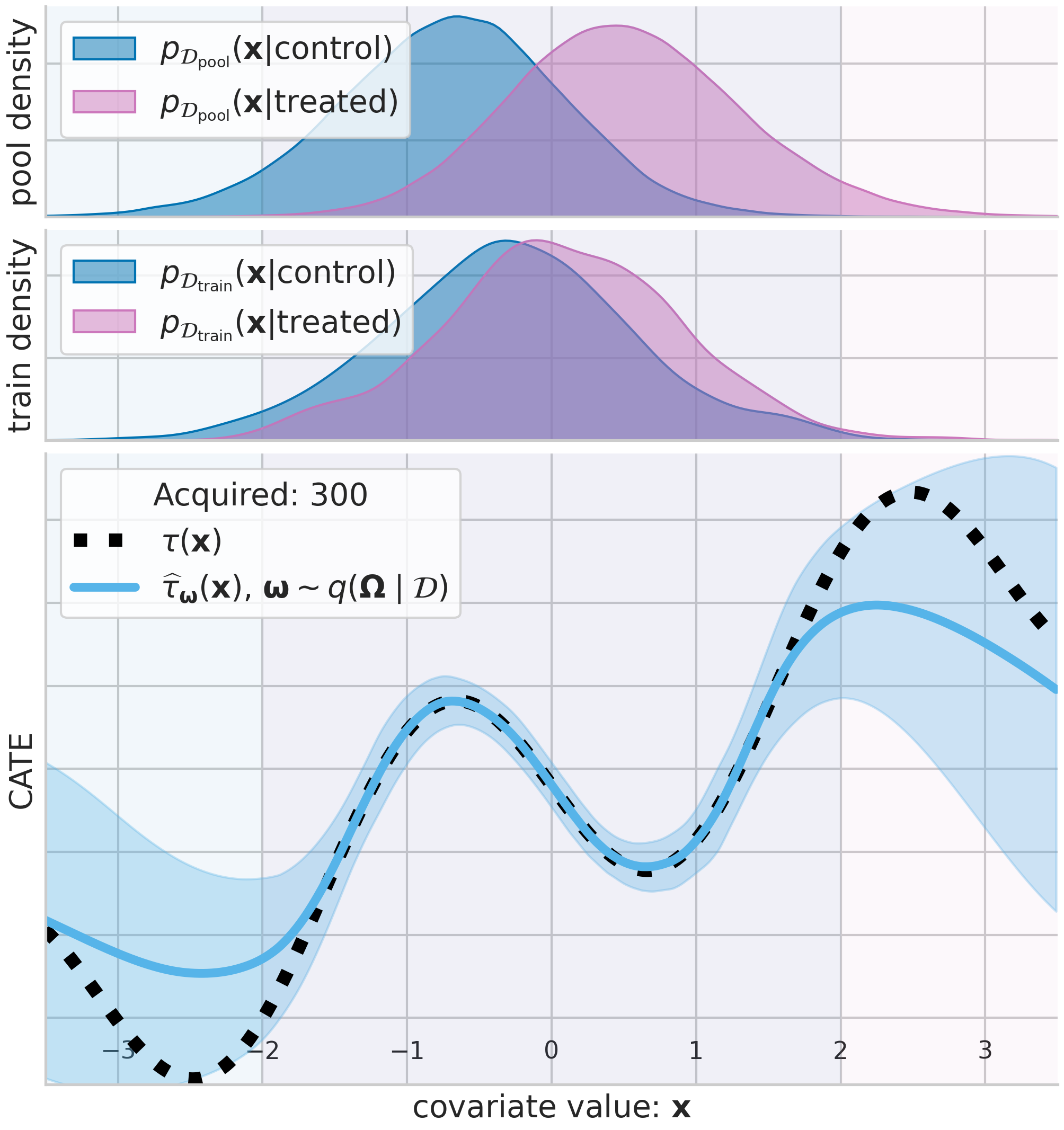}
         \caption{$\rho$BALD}
         \label{causalbald:fig:bias_rho}
     \end{subfigure}
     \hfill
     \begin{subfigure}[]{0.28\textwidth}
         \centering
         \includegraphics[width=\textwidth]{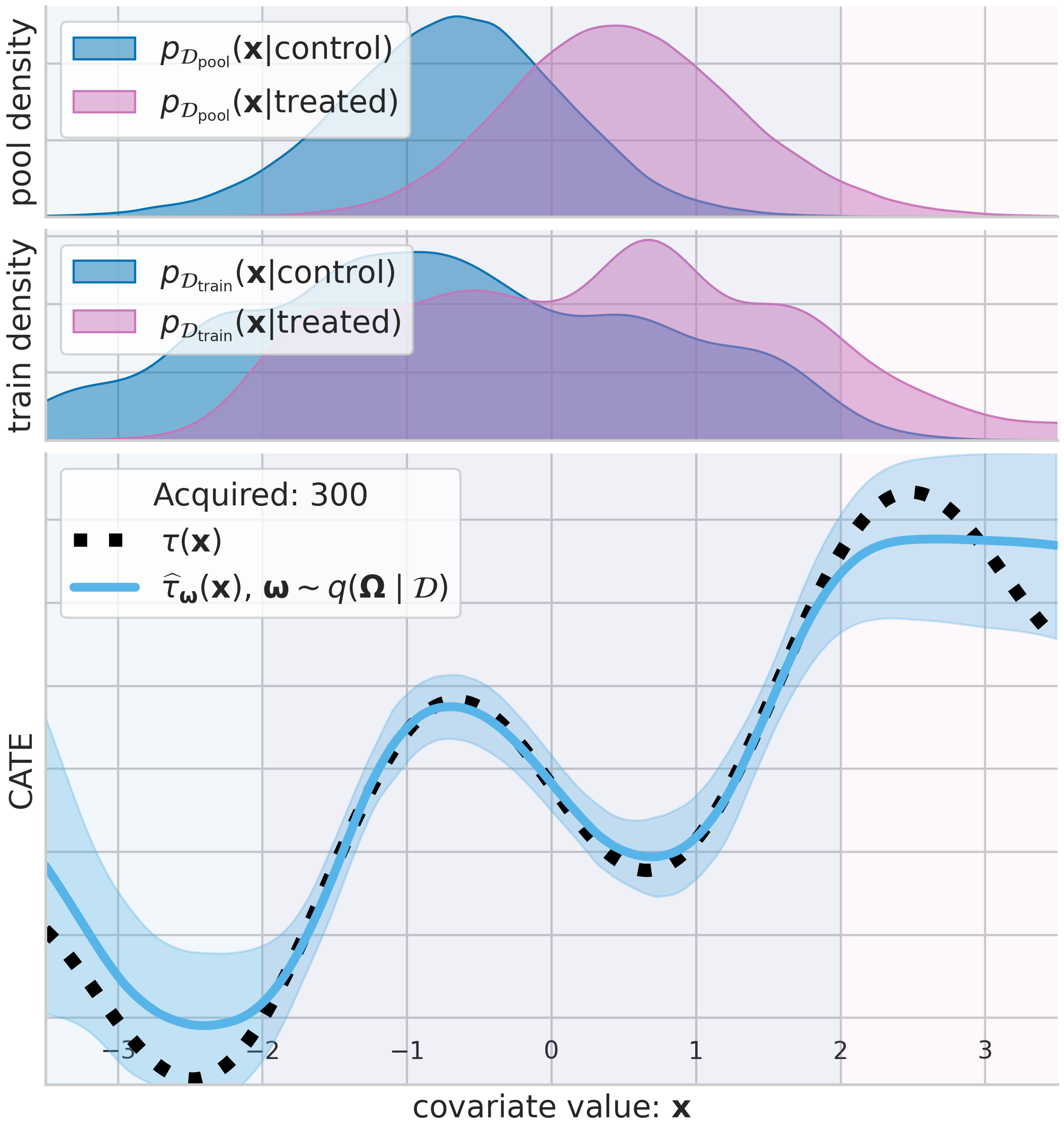}
         \caption{$\mu\rho$BALD}
         \label{causalbald:fig:bias_causal}
     \end{subfigure}
    \caption{Causal--BALD acquisition functions: How the training set is biased and how this effects the CATE function with a fixed budget of 300 acquired points.}
    \end{figure*}

BALD aims to acquire data $(\x, \tf)$ that maximally reduce uncertainty in the model parameters $\W$ used to predict the treatment effect.
The most direct way to apply BALD is to use our uncertainty over the predicted treatment effect, expressed using the following information theoretic quantity:
\begin{definition}
    $\tau$BALD
    \begin{equation*}
        \mi(\Yone - \Yzero; \W \mid \x, \tf, \Dtrain) 
        \approx \var_{\w} \left( \muh_{\w}(\x, 1) - \muh_{\w}(\x, 0) \right)
    \end{equation*}
\end{definition}
Building off the result in \citep{jesson2020identifying}, we show how the LHS measure about the \emph{unobservable potential outcomes} can be estimated by the variance over $\W$ of the \emph{identifiable difference in expected outcomes} in Theorem \ref{causalbald:th:tau-BALD} of the appendix.
A similar result has been proposed for non-parametric models \cite{alaa2018bayesian}.
Intuitively, this measure represents the information gain for $\W$ if we could observe the difference in potential outcomes $\Yone - \Yzero$ for a given measurement $\x$ and $\Dtrain$.

However, labels for the random variable $\Yone - \Yzero$ are never observed so $\tau$BALD represents an irreducible measure of uncertainty.
That is, $\tau$BALD will be high if it is uncertain about the label given the unobserved treatment $\tcf$, regardless of its certainty about the label given the observed treatment $\tf$, which makes $\tau$BALD highest for low-density regions and regions with no overlap.
\Cref{causalbald:fig:bias_tau} illustrates these consequences.
We see the acquisition biases the training data away from the modes of the $\Dpool$, where  we cannot know the treatment effect (no overlap).
In datasets where we do not have full overlap, it leads to uninformative acquisitions.

\subsection{Causal--BALD}
In the previous section we looked at naive methods that either considered overlap, or considered information gain. In this section we develop a measure that take into account both factors when choosing a new training data point.

First, we focus only on reducible uncertainty:
\begin{definition}
    $\mu$BALD
    \begin{equation}
        \mi(\Yt; \W \mid \x, \tf, \Dtrain)
        \approx \var_{\w \sim p(\W \mid \Dtrain)} \left( \muh_{\w}(\x, \tf) \right).
        \label{causalbald:eq:mu-bald-estimate}
    \end{equation}
\end{definition}
This measure represents the information gain for the model parameters $\W$ if we obtain a label for the observed potential outcome $\Yt$ given a data point ($\x, \tf$) and $\Dtrain$.
Proof is given in Theorem \ref{causalbald:th:mu-BALD} of the appendix.

$\mu$BALD only contains observable quantities; however, it does not take into account our belief about the counterfactual outcome.
As illustrated in \cref{causalbald:fig:bias_mu}, this approach can prefer acquiring $(\x, \tf)$ when we are also very uncertain about $(\x, \tcf)$, even if $(\x, \tcf)$ is not in $\Dpool$.
Since we can neither reduce uncertainty over such $(\x, \tcf)$ nor know the treatment effect, acquisition would not be data efficient.

Next, we can take an information theoretic approach to combining knowledge about a data point's information gain and overlap.
Let $\muh_{\w}(\x, \tf)$ be an instance of the random variable $\muh_{\W}^{\tf} \in \mathbb{R}$ corresponding to the expected outcome conditioned on $\tf$. 
Further, let $\widehat{\tau}_{\w}(\x)$ be an instance of the random variable $\widehat{\tau}_{\W} = \muh_{\W}^{1} - \muh_{\W}^{0}$ corresponding to the CATE. 
Then,
\begin{definition}
    $\rho$BALD
    \begin{equation}
        \begin{split}
            \mi(\Yt; \widehat{\tau}_{\W} \mid \x, \tf, \Dtrain) \gtrapprox
            \frac{1}{2}\log{\left(\frac{\var_{\w}(\widehat{\tau}_{\w}(\x))}{\var_{\w}(\muh_{\w}(\x, \tcf))}\right)}
        \end{split}
        \label{causalbald:eq:rho_def}
    \end{equation}
\end{definition}
This measure represents the information gain for the CATE $\tau_{\W}$ if we observe the outcome $\Y$ for a data point ($\x$, $\tf$) and the data we have trained on $\Dtrain$. 
Proof for this result is given in Theorem \ref{causalbald:th:rho-BALD}.

In contrast to $\mu$-BALD, this measure accounts for overlap in two ways.
First, $\rho$--BALD will be scaled by the inverse of the variance of the expected counterfactual outcome $\muh_{\w}(\x, \tcf)$.
This will bias acquisition towards examples for which we are certain about counterfactual outcome, and so we can assume that overlap is satisfied for observed $(\x, \tf)$.
Second, $\rho$--BALD is discounted by $\cov_{\w}(\muh_{\w}(\x, \tf), \muh_{\w}(\x, \tcf))$.
This is an interesting concept that we will leave for future discussion.

In \cref{causalbald:fig:bias_rho} we see that $\rho$--BALD has matched the distributions of the treated and control groups in a similar manner to propensity acquisition in \cref{causalbald:fig:bias_pi}.
Further, we see that the CATE estimator is more accurate over the support of the data.
However, there is a shortcoming of $\rho$--BALD that results in it under exploring low density regions of $\Dpool$, which we comment on in \cref{causalbald:a:rho_failure}. 

To combine the positive attributes of $\mu$--BALD and $\rho$--BALD, while mitigating their shortcomings, we introduce $\mu\rho$BALD.
\begin{definition}
    $\mu\rho$BALD
    \begin{equation*}
        \begin{split}
            \mi(\mu \rho \mid \x, \tf, \Dtrain) 
            &\equiv  \var_{\w}\left( \muh_{\w}(\x, \tf) \right) \frac{\var_{\w}(\widehat{\tau}_{\w}(\x))}{\var_{\w}(\muh_{\w}(\x, \tcf))}.
        \end{split}
    \end{equation*}
\end{definition}
Here, we scale Equation \ref{causalbald:eq:rho_def}, which has equivalent expression $\frac{\var_{\w}(\widehat{\tau}_{\w}(\x))}{\var_{\w}(\muh_{\w}(\x, \tcf))}$ by our measure for $\mu$BALD such that in the cases where the ratio may be equal, there is a preference for data points the current model is more uncertain about.
We can see in \cref{causalbald:fig:bias_causal} that the acquisition of training data examples is more uniformly distributed over the support of the pool data where overlap is satisfied.
Furthermore, the accuracy of the CATE estimator is the highest over that region.

\section{Related Work}
\label{causalbald:sec:related_word}
\Citet{deng2011active} propose the use of Active Learning for recruiting patients to assign treatments that will reduce the uncertainty of an Individual Treatment Effect model. However, their setting is different from ours---we assume that suggesting treatments are too risky or even potentially lethal---and instead we acquire patients for the purpose of revealing their outcome (e.g. by having a biopsy). Additionally, although their method uses the predictive uncertainty to identify which patients to recruit, it does not disentangle the sources of uncertainty and as such treatments with high outcome variance will be recruited as well. 
Closer to our proposal is the work from \citet{sundin2019active}, where the authors propose the use of a Gaussian process (GP) to model the individual treatment effect and use the expected information gain over the S-type error rate, defined as the error in predicting the sign of the CATE, as their acquisition function.
We compare to this in our experiment by limiting the access to counterfactual observations ($\boldsymbol\gamma$ baseline) and adapting it to Deep Ensembles \citep{lakshminarayanan2016simple} and DUE \citep{van2021improving} (more details about the adaptation is provided in Appendix \ref{causalbald:a:sundin}).
\section{Empirical Validation}
\label{causalbald:sec:experiments}

In this section we evaluate our acquisition objectives on synthetic and semi-synthetic datasets.

\noindent\textbf{Models}
Our objectives rely on methods that are capable of modeling the uncertainty and handling high-dimensional data modalities. 
DUE \citep{van2021improving} is an instance of Deep Kernel Learning \citep{wilson2016deep}, where a deep feature extractor is used to transform the inputs over which a Gaussian process' (GP) kernel is defined. 
In particular, DUE uses a variational inducing point approximation \citep{hensman2015scalable} and a constrained feature extractor which contains residual connections and spectral normalisation to enable reliable uncertainty. 
It was previously shown to obtain SotA results on IHDP \citep{van2021improving}.
In DUE, we distinguish between the model parameters $\theta$ and the variational parameters $\w$, and we are Bayesian only over the $\w$ parameters. Since DUE is a GP, we obtain a full Gaussian posterior over our outputs from which we can use the mean and covariance directly. 
When necessary, sampling is very efficient and only requires a single forward pass in the deep model. We describe all hyper parameters in Appendix \ref{causalbald:a:architecture}.

\noindent\textbf{Baselines}
We compare against the following baselines:
\begin{description}
    \item[Random.] This acquisition function selects points uniformly at random.
    \item[Propensity.] An acquisition function based on the propensity score (Eq.~\ref{causalbald:eq:propensity-acquisition}). We train a propensity model on the combination of the train and pool dataset which we then use acquire points based on their propensity score. Please note that this is a valid assumption as training a propensity model does not require outcomes.
    \item[$\boldsymbol\gamma$ (S-type error rate)~\citep{sundin2019active}.]This acquisition function is the S-type error rate based method proposed by~\citet{sundin2019active}. We have adapted the acquisition function to use with Bayesian Deep Neural Networks. The objective is defined as $\mi(\boldsymbol\gamma;\W\mid\x,\Dtrain)$, where $\gamma(\x)=\text{probit}^{-1}(-\frac{|\mathbf{E}_{p(\tau\mid \x, \Dtrain)}[\tau]|}{\sqrt{\text{Var}(\tau|\x, \Dtrain)}})$ and $\text{probit}^{-1}(\cdot)$ is the cumulative distribution function of normal distribution. In contrast to the original formulation, we do not assume access to counterfactual observations at training time.
\end{description}

\noindent\textbf{Datasets}
Starting from the hypothesis that different objectives can target different types of imbalances and overlap ratios we construct a \textbf{synthetic} dataset~\citep{kallus2019interval} demonstrating the different biases. Additionally, we study the performance of our acquisition functions on the \textbf{IHDP} dataset~\citep{hill2011bayesian, shalit2017estimating}, a standard benchmark in causal treatment effect literature, and finally we demonstrate that our method is suitable for high dimensional datasets on \textbf{CMNIST}~\citep{jesson2021quantifying}, an MNIST~\citep{lecun1998mnist} based dataset adapted for causal treatment effect studies.
Detailed descriptions of each dataset are given in \cref{causalbald:a:datasets}.

\begin{figure*}[!ht]
  \centering
  \begin{subfigure}[b]{\linewidth}
    \includegraphics[width=.9\linewidth]{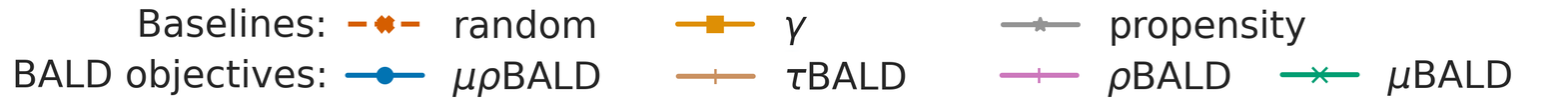}
  \end{subfigure}
  \\~\\
  \begin{subfigure}[b]{0.32\linewidth}
    \centering
    \includegraphics[width=\linewidth]{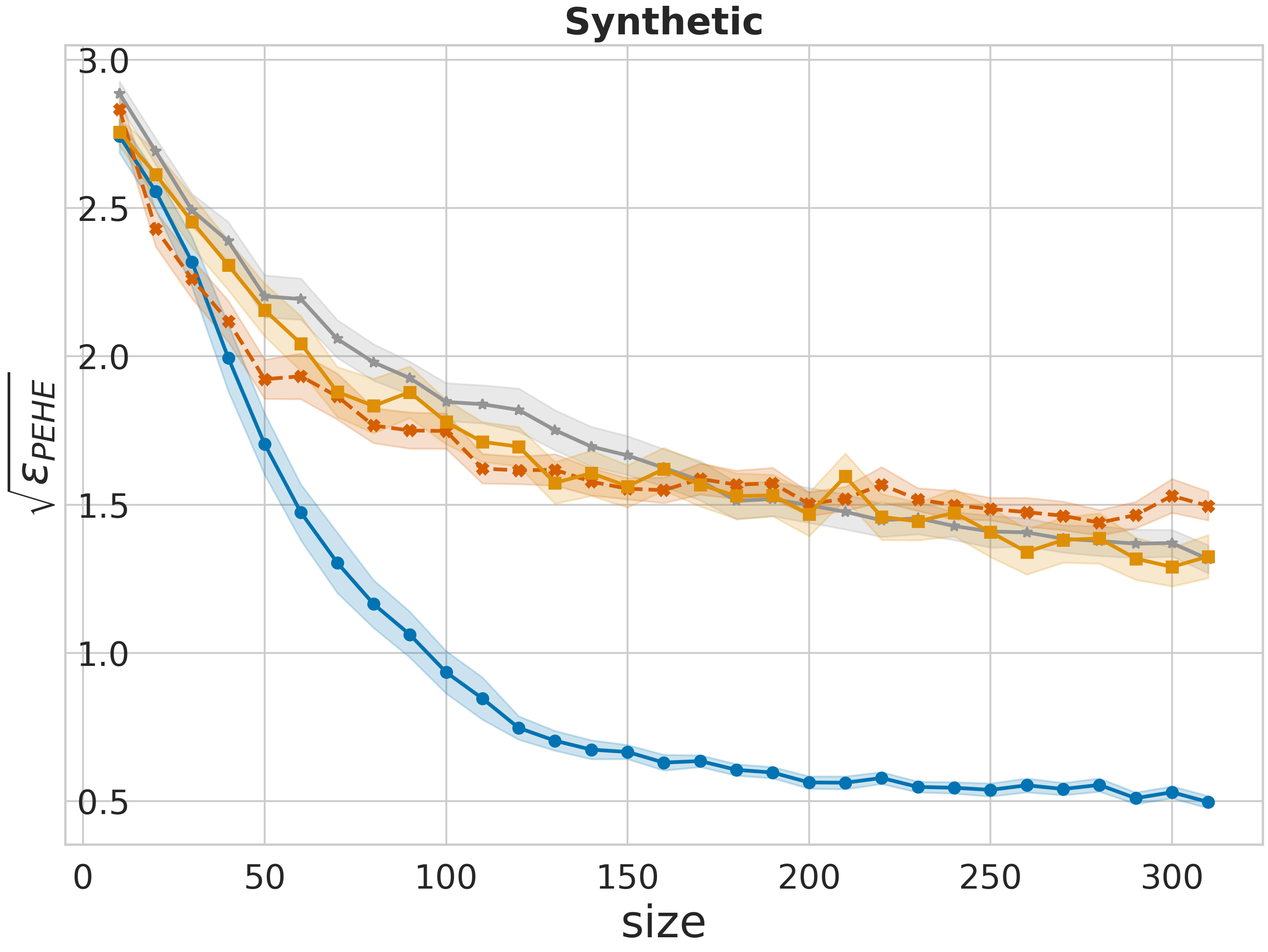}
  \end{subfigure}
  \begin{subfigure}[b]{0.32\linewidth}
    \centering
    \includegraphics[width=\linewidth]{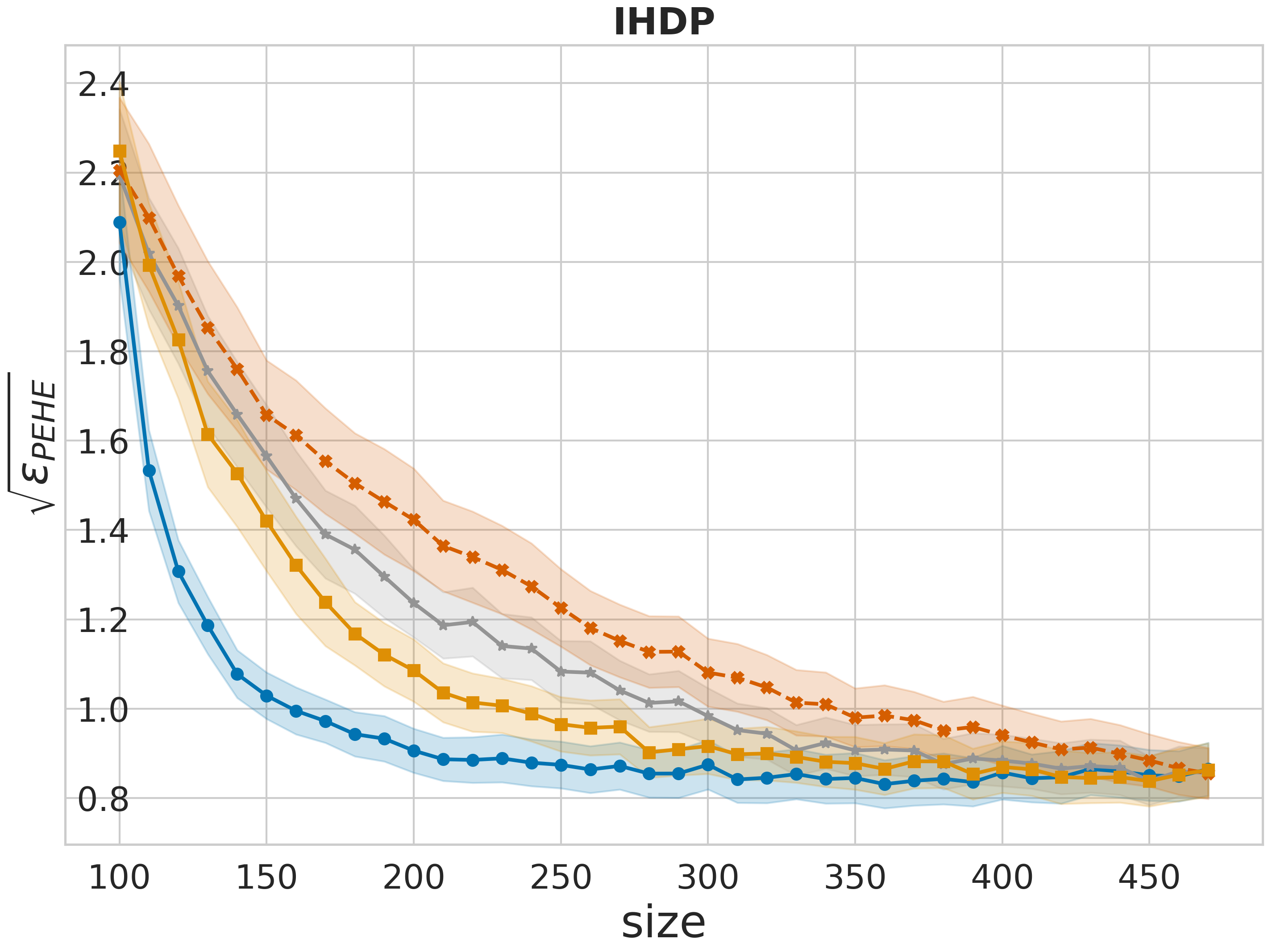}
  \end{subfigure}
  \begin{subfigure}[b]{0.32\linewidth}
    \centering
    \includegraphics[width=\linewidth]{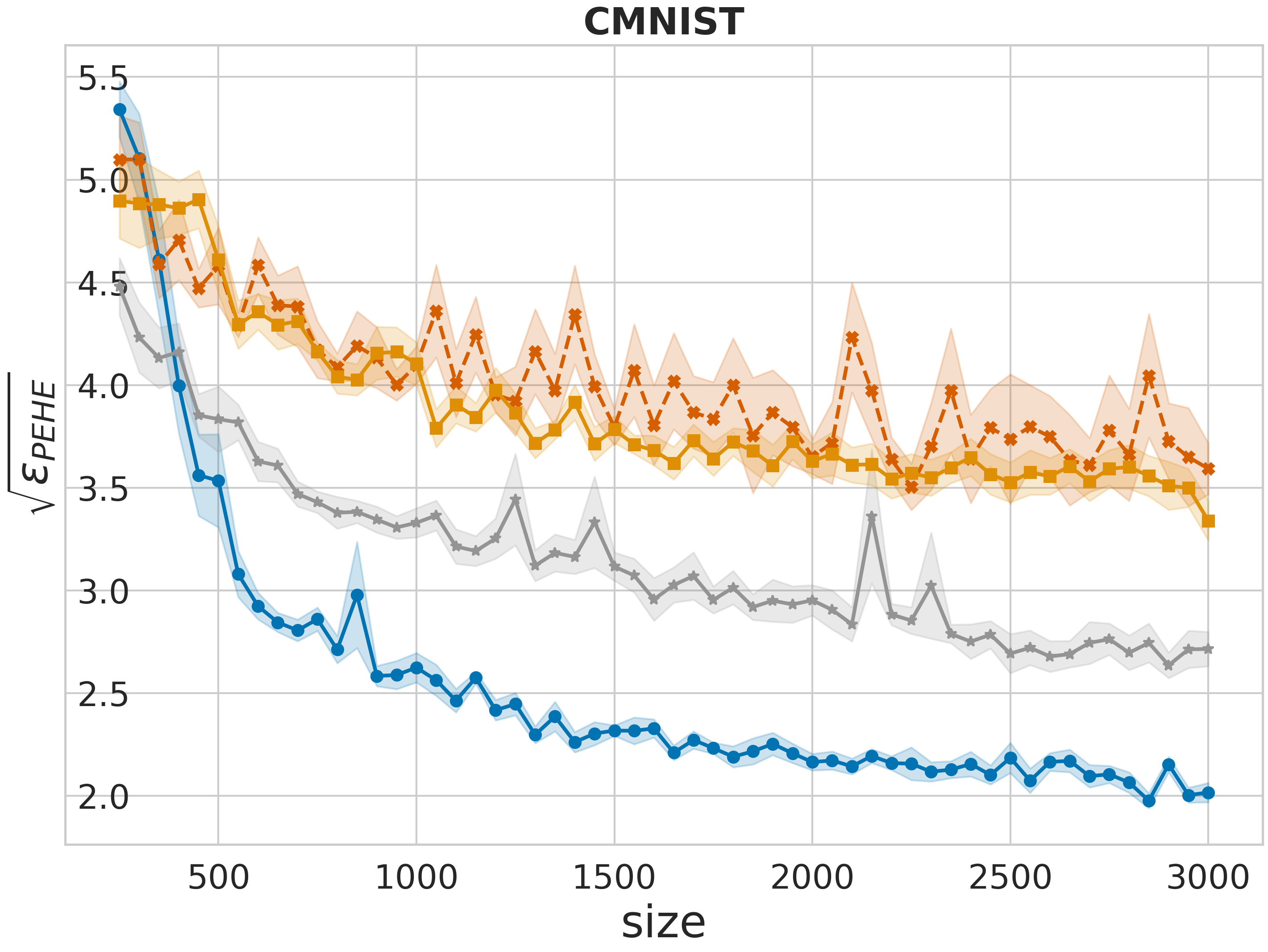}
  \end{subfigure}
  \begin{subfigure}[b]{0.32\linewidth}
    \centering
    \includegraphics[width=\linewidth]{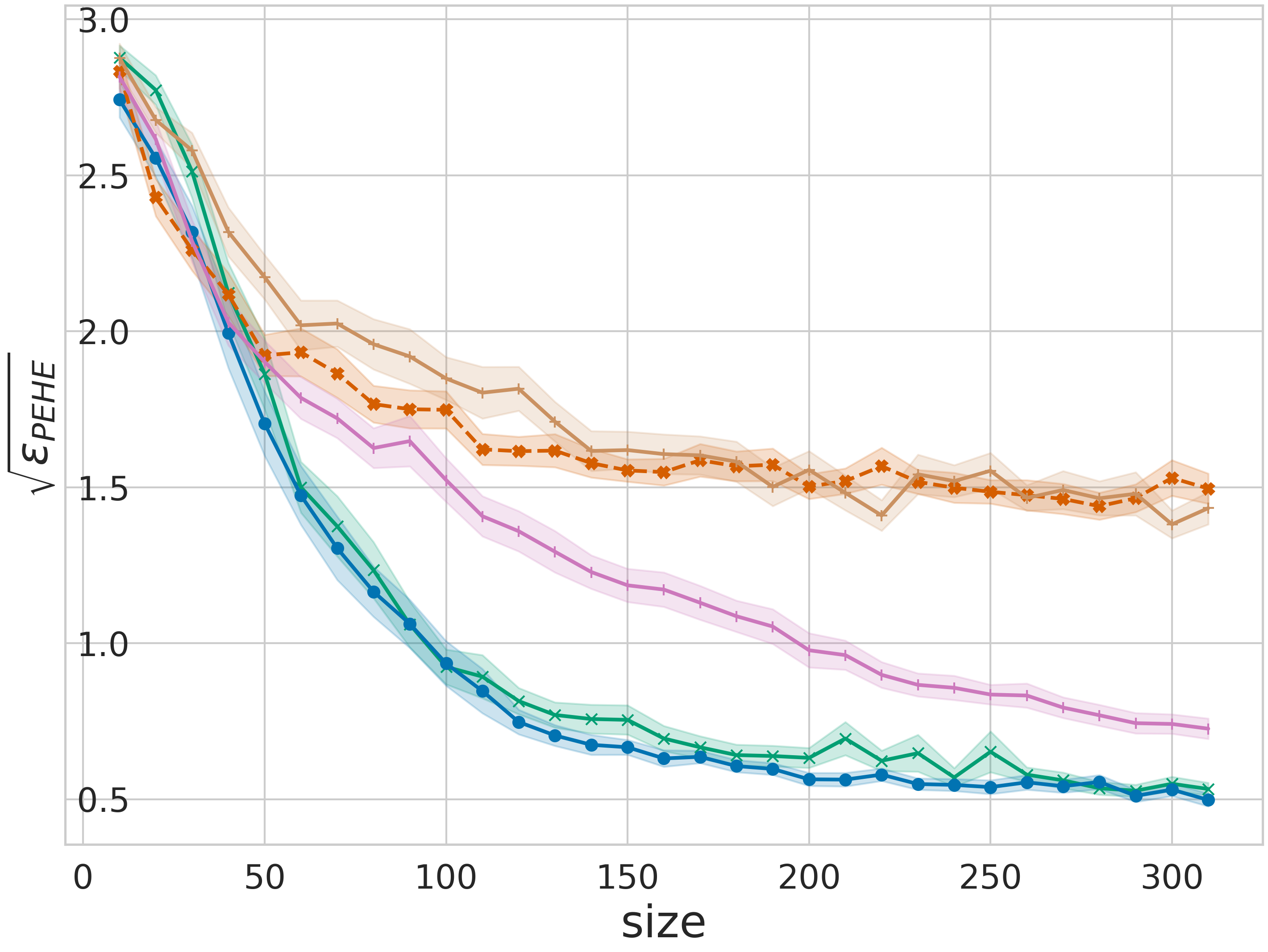}
  \end{subfigure}
  \begin{subfigure}[b]{0.32\linewidth}
    \centering
    \includegraphics[width=\linewidth]{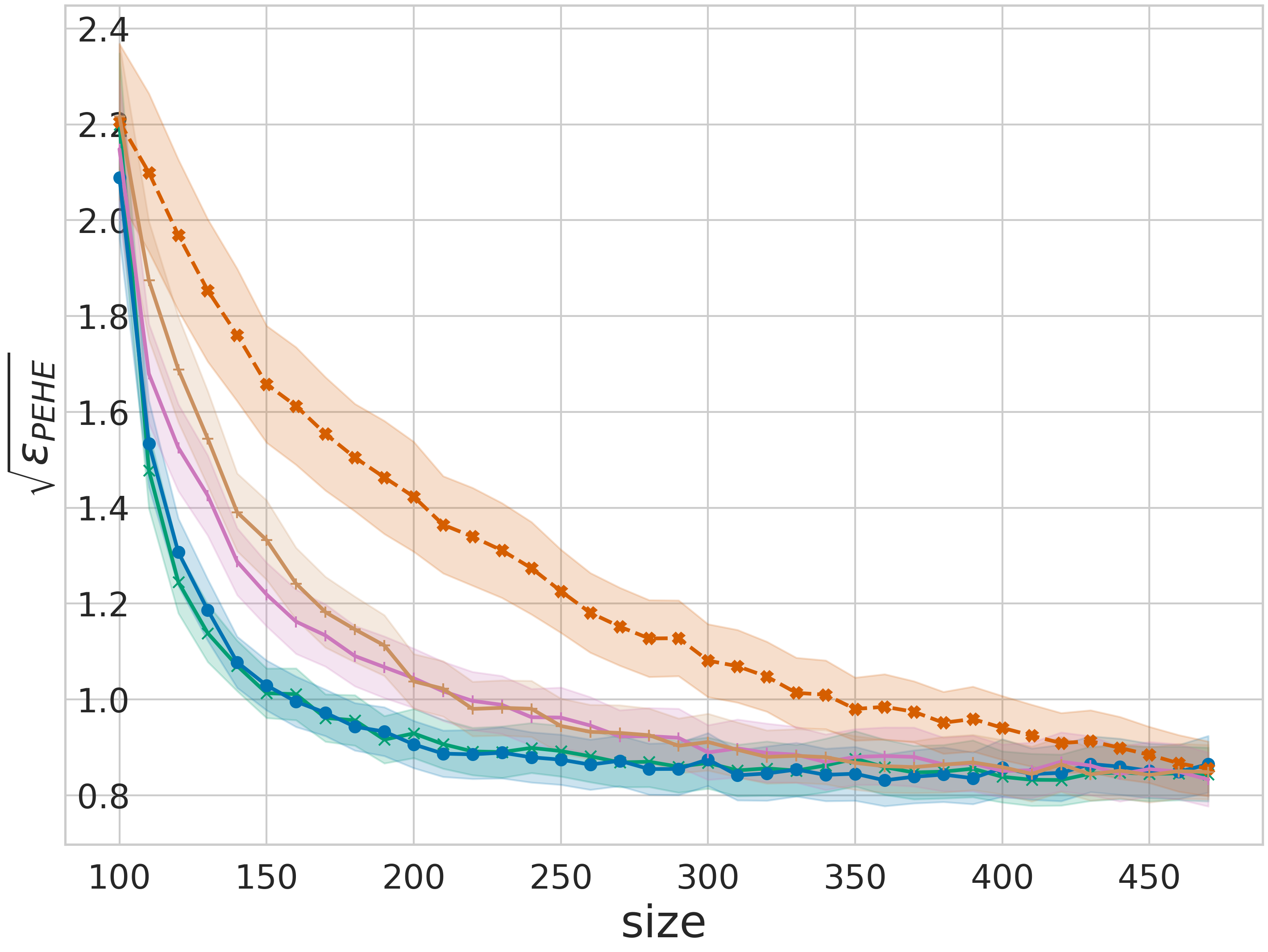}
  \end{subfigure}
  \begin{subfigure}[b]{0.32\linewidth}
    \centering
    \includegraphics[width=\linewidth]{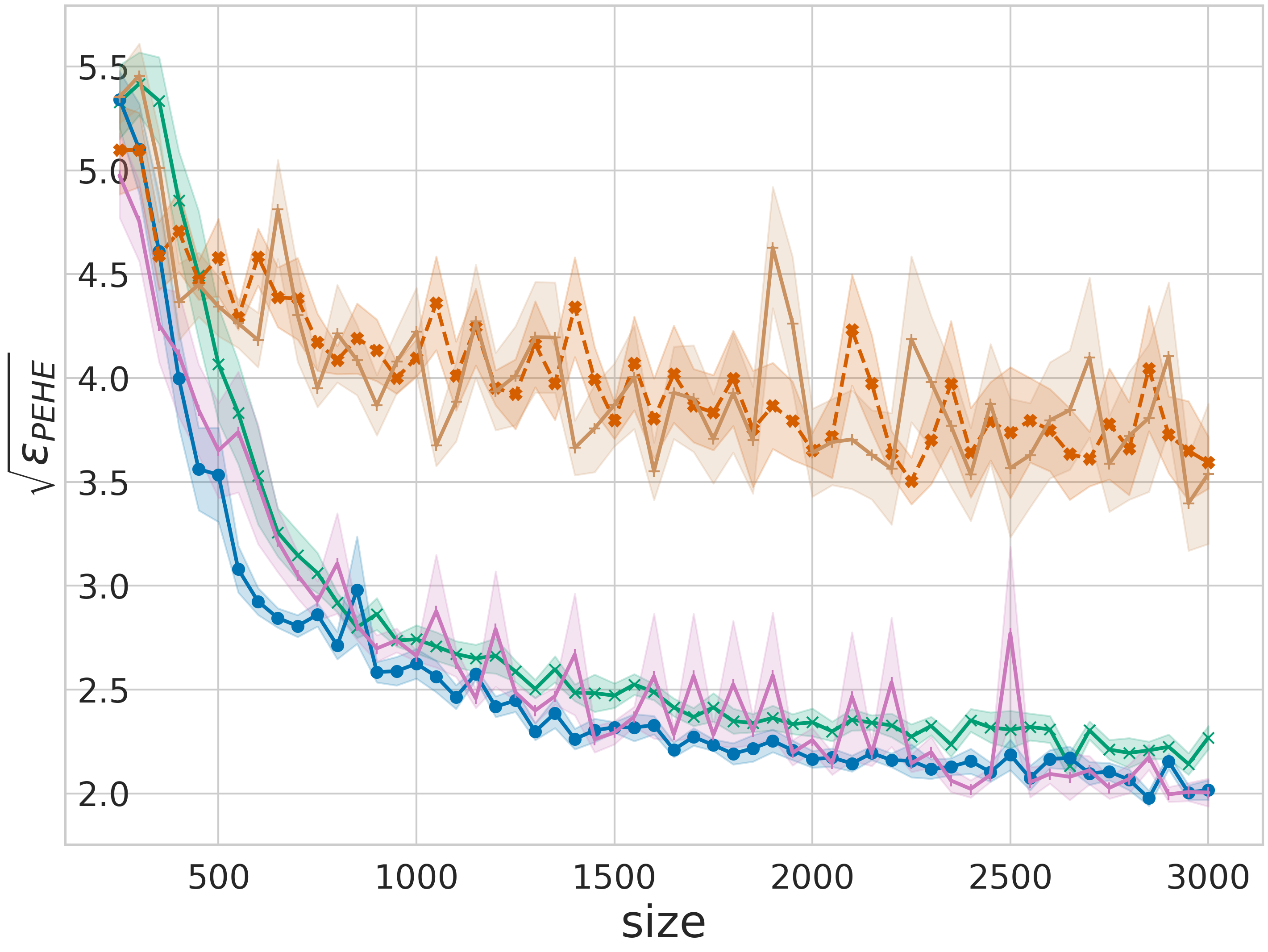}
  \end{subfigure}
  \caption{
    $\sqrt{\epsilon_{PEHE}}$ performance (shaded standard error) for DUE models. \textbf{(left to right)} \textbf{synthetic} (40 seeds), and \textbf{IHDP} (200 seeds). We observe that BALD objectives outperform the \textbf{random}, \textbf{$\boldsymbol\gamma$} and \textbf{propensity} acquisition functions significantly, suggesting that epistemic uncertainty aware methods that target reducible uncertainty can be more sample efficient.
  }
  \label{causalbald:fig:main_results}
\end{figure*}

\subsection{Results}
\label{causalbald:sec:results}

For each of the acquisition objectives, dataset, and model we present the mean and standard error of empirical square root of precision in estimation of heterogenous effect (PEHE)~\footnote{$\sqrt{\epsilon_{PEHE}}=\sqrt{ \frac{1}{N} \sum_{\x}{(\hat{\tau}(\x)-\tau(\x))^2}}$}. We summarize each active learning setup in ~\cref{causalbald:table:acquisition_settings}. 

In \cref{causalbald:fig:main_results}, we see that epistemic uncertainty aware ~\murhobald~ outperforms the baselines, random, propensity and S-Type error rate ($\boldsymbol\gamma$). As we analysed in section \ref{causalbald:sec:methods}, this is expected as our acquisition objectives target the type of uncertainty that can be reduced -- that is the epistemic uncertainty for which we have overlap between treatment and control. Additionally,~\murhobald~shows superior performance over the other objectives in the high dimensional dataset CMNIST verifying our qualitative analysis in Figure~\ref{causalbald:fig:bias_causal}.

\section{Discussion}
\label{causalbald:sec:conclusion}
We have introduced a new acquisition function for active learning of individual-level causal-treatment effects from high dimensional observational data, based on Bayesian Active Learning by Disagreement~\cite{houlsby2011bayesian}. We derive our proposed method from an information theoretic perspective and compared with various acquisition functions that do not take into consideration epistemic uncertainty (like random or propensity based) or they target uncertainties that cannot be reduced in the observational setting (i.e. when we do not have access to counterfactual observations). We show that our methods significantly outperform the baselines while also studying the various properties of each of our proposed objectives in both a quantitative and a qualitative analysis, potentially impacting areas like healthcare where sample efficiency in acquisition of new examples imply improved safety and reductions in costs.

		\section{Details}
		\let\section\subsection
		\let\subsection\subsubsection
		\let\subsubsection\paragraph
		\let\paragraph\textbf
		\section{Theoretical Results}
\label{causalbald:sec:theoretical_results}
\subsection{$\tau$-BALD}
\begin{theorem}
    Under the following assumptions:
    \begin{enumerate}
        \item Unconfoundedness $(\Yzero, \Yone) \independent \T \mid \X$;
        \item Consistency $\Y \mid \T = \Yt$;
        \item $\Yone$ and $\Yzero$, when conditioned on realizations $\x$ of the r.v. $\X$ and $\tf$ of the r.v. $\T$, are independent-normally distributed or joint-normally distributed r.v.s.
        \item $\muh_{\w}(\x, \tf)$ is a consistent estimator of $\E[\Y \mid \T=\tf, \X=\x]$
    \end{enumerate}
    the information gain for $\W$ if we could observe a label for the difference in potential outcomes $\Yone - \Yzero$ given measured covariates $\x$, treatment $\tf$ and a dataset of observations $\Dtrain=\{\x_i, \tf_i, \y_i\}_{i=1}^{n}$ is approximated as
    \begin{equation}
        \mi(\Yone - \Yzero; \W \mid \x, \tf, \Dtrain)
        \approx \var_{\w \sim p(\W \mid \Dtrain)} \left( \muh_{\w}(\x, 1) - \muh_{\w}(\x, 0) \right)
    \end{equation}
    \begin{proof}
        \begin{subequations}
            \begin{align}
                \mi(\Yone - \Yzero; \W \mid \x, \Dtrain)
                &= \ent(\Yone - \Yzero \mid \x, \Dtrain) - \E_{p(\W \mid \Dtrain)} \left[ \ent(\Yone - \Yzero \mid \x, \w) \right] \label{causalbald:eq:tau_bald} \\
                &\approx \var(\Yone - \Yzero \mid \x, \Dtrain) - \E_{p(\W \mid \Dtrain)} \left[ \var(\Yone - \Yzero \mid \x, \w) \right] \label{causalbald:eq:tau_var_1} \\
                &= \var_{p(\W \mid \Dtrain)} \left( \E[\Yone - \Yzero \mid \x, \w] \right) \label{causalbald:eq:tau_var_2} \\
                &= \var_{p(\W \mid \Dtrain)} \left( \muh_{\w}(\x, 1) - \muh_{\w}(\x, 0) \right) \label{causalbald:eq:tau_var_3}
            \end{align}
        \end{subequations}
        In \eqref{causalbald:eq:tau_bald} we adapt the result of \citet{houlsby2011bayesian} and express the information gain as the mutual information between the observable difference in potential outcomes $\Yone - \Yzero$ and the parameters $\W$; given observed covariates $\x$, treatment $\tf$, and training data $\Dtrain = \left\{ (\x_i, \tf_i, \y_i)\right\}_{i=1}^{n_\mathrm{train}}$.
        In \eqref{causalbald:eq:tau_var_1} we apply lemma \ref{causalbald:l:variance_entropy} to the r.h.s terms of \eqref{causalbald:eq:tau_bald}.
        We then use the result in \citet{jesson2020identifying} and move from \eqref{causalbald:eq:tau_var_1} to \eqref{causalbald:eq:tau_var_2} by application of the law of total variance. 
        Finally, under the consistency and unconfoundedness assumptions we express the information gain in terms of the identifiable difference in expected outcomes $\muh_{\w}(\x, 1) - \muh_{\w}(\x, 0)$.
    \end{proof}
    \label{causalbald:th:tau-BALD}
\end{theorem}

\begin{lemma}
    Under the following assumptions:
    \begin{enumerate}
        \item $\Yone, \Yzero$ are independent-normally distributed or joint-normally distributed r.v.s;
        \item With $A = \var(\Yone - \Yzero)$: let $|A - 1| \leq 1$ and $A \neq 0$. 
        That is to say, the predictive variance must be greater than 0 and less than or equal to 2;
    \end{enumerate}
    \begin{equation}
        \ent(\Yone - \Yzero) \approx \var(\Yone - \Yzero)
    \end{equation}
    \begin{proof}
        By assumption 1, $\Yone - \Yzero$ is also a normally distributed random variable.
        By corollary \ref{causalbald:c:ent_normal},
        \begin{equation}
            \ent(\Yone - \Yzero) = \frac{1}{2} + \frac{1}{2}\log(2\pi \var(\Yone - \Yzero))
        \end{equation}
        So given assumption 2, the first order Taylor polynomial of $\ent(\Yone - \Yzero)$ is
        \begin{equation}
            \begin{split}
                \frac{1}{2} + \frac{1}{2}\log(2\pi \var(\Yone - \Yzero)) 
                &\approx \frac{1}{2} + \frac{1}{2}(2\pi \var(\Yone - \Yzero) - 1) \\
                &= \frac{1}{2} + \pi \var(\Yone - \Yzero) - \frac{1}{2} \\
                &= \pi \var(\Yone - \Yzero) \\
                &\propto \var(\Yone - \Yzero)
            \end{split}
        \end{equation}
    \end{proof}
    \label{causalbald:l:variance_entropy}
\end{lemma}

\begin{corollary}
    The entropy of a normally distributed random variable with variance $\sigma^2$ is $\frac{1}{2} + \frac{1}{2}\log(2\pi \sigma^2)$
    \label{causalbald:c:ent_normal}
\end{corollary}

\subsection{$\mu$-BALD}
\label{causalbald:a:mu-BALD}

\begin{theorem}
    Under the following assumptions:
    \begin{enumerate}
        \item Unconfoundedness $(\Yzero, \Yone) \independent \T \mid \X$,
        \item Consistency $\Y \mid \T = \Yt$,
        \item $\Y$ conditioned on $\x$ and $\tf$ is a normally distributed random variable,
        \item $\muh_{\w}(\x, \tf)$ is a consistent estimator of $\E[\Y \mid \T=\tf, \X=\x]$,
    \end{enumerate}
    the information gain for $\W$ when we observe a label for the potential outcome $\Yt$ given measured covariates $\x$, treatment $\tf$ and a dataset of observations $\Dtrain=\{\x_i, \tf_i, \y_i\}_{i=1}^{n}$ can be approximated as
    is
    \begin{equation}
        \mi(\Yt; \W \mid \x, \tf, \Dtrain)
        \approx \frac{1}{2}\log{\left(\frac{\var(\Y \mid \x, \tf, \Dtrain)}{\E_{\w}[\var( \Y \mid \x, \tf, \w)]} \right)},
        \label{causalbald:eq:mu-ratio}
    \end{equation}
    or
    \begin{equation}
        \mi(\Yt; \W \mid \x, \tf, \Dtrain)
        \approx \var_{\w \sim p(\W \mid \Dtrain)} \left( \muh_{\w}(\x, \tf) \right).
        \label{causalbald:eq:mu-estimate}
    \end{equation}
    Equation \eqref{causalbald:eq:mu-ratio} expresses the information gain as the logarithm of a ratio between predictive and aleatoric uncertainty in the outcome.
    Whereas, equation \eqref{causalbald:eq:mu-ratio} expresses the information gain as a direct estimate of the epistemic uncertainty.
    \begin{proof}
        \begin{subequations}
            \begin{align}
                \mi(\Yt; \W \mid \x, \tf, \Dtrain) 
                &= \ent(\Y \mid \x, \tf, \Dtrain) - \E_{p(\W \mid \Dtrain)} \left[ \ent(\Y \mid \x, \tf, \w) \right]
                \label{causalbald:eq:mu_ratio_a} \\
                &= \frac{1}{2} \log{ \left( 2\pi \var(\Y \mid \x, \tf, \Dtrain) \right) }  - \E_{p(\W \mid \Dtrain)} \frac{1}{2} \log{ \left( 2\pi \var(\Y \mid \w, \x, \tf) \right) }
                \label{causalbald:eq:mu_ratio_b} \\
                &\geq \frac{1}{2} \log{ \left( 2\pi \var(\Y \mid \x, \tf, \Dtrain) \right) }  -  \frac{1}{2} \log{ \left( 2\pi \E_{p(\W \mid \Dtrain)} \var(\Y \mid \w, \x, \tf) \right) }
                \label{causalbald:eq:mu_ratio_c} \\
                &= \frac{1}{2}\log{\left(\frac{\var(\Y \mid \x, \tf, \Dtrain)}{\E_{\w}[\var( \Y \mid \x, \tf, \w)]} \right)}
                \label{causalbald:eq:mu_ratio_d}
            \end{align}
        \end{subequations}
        \begin{subequations}
            \begin{align}
                \mi(\Yt; \W \mid \x, \tf, \Dtrain) 
                &= \ent(\Y \mid \x, \tf, \Dtrain) - \E_{p(\W \mid \Dtrain)} \left[ \ent(\Y \mid \x, \tf, \w) \right] \label{causalbald:eq:mu_bald} \\
                &\approx \var[\Y \mid \x, \tf, \Dtrain] - \E_{p(\W \mid \Dtrain)} \left[ \var[\Y \mid \x, \tf, \w] \right] \label{causalbald:eq:mu_var_1} \\
                &= \var_{\w \sim p(\W \mid \Dtrain)} \left( \muh_{\w}(\x, \tf) \right) \label{causalbald:eq:mu_var_2}
            \end{align}
        \end{subequations}
        In \eqref{causalbald:eq:mu_bald} we express the information gain as the mutual information between the observed potential outcome $\Yt$ and the parameters $\W$; given observed covariates $\x$, treatment $\tf$, and training data $\Dtrain$.
        By consistency, we can drop the superscript on the potential outcome.
        In \eqref{causalbald:eq:mu_var_1} we approximate the r.h.s terms of \eqref{causalbald:eq:mu_bald} by application of Lemma \ref{causalbald:l:variance_entropy}.
        Finally, we can move from \eqref{causalbald:eq:mu_var_1} to \eqref{causalbald:eq:mu_var_2} by application of the law of total variance.
    \end{proof}
    Note that for discrete or categorical $\Y$, it is straightforward to evaluate Equation \eqref{causalbald:eq:mu_bald} directly.
    \label{causalbald:th:mu-BALD}
\end{theorem}

\subsection{$\rho$-BALD}
\label{causalbald:a:rho-BALD}
\begin{theorem}
    Under the following assumptions
    \begin{enumerate}
        \item $\{\muh_{\w}(\x, \tf): \tf \in \{0, 1\}\}$ are instances of the independent-normally distributed or joint-normally distributed random variables $\{\muh_{\W}^{\tf} = \E[\Y \mid \W, \T=\tf, \x]: \tf \in \{0, 1\}\}$,
        \item $\var_{\w \sim p(\W \mid \Dtrain)}(\muh_{\w}(\x, \tcf)) > 0$ .
    \end{enumerate}
    Let $\widehat{\tau}_{\w}(\x)$ be a realization of the random variable $\widehat{\tau}_{\W} = \muh_{\W}^{1} - \muh_{\W}^{0}$.
    The information gain for $\widehat{\tau}_{\W}$ if we observe the label for the potential outcome $\Yt$ given measured covariates $\x$, treatment $\tf$ and a dataset of observations $\Dtrain=\{\x_i, \tf_i, \y_i\}_{i=1}^{n}$ is approximately
    \begin{equation}
        \begin{split}
            \mi(\Yt; \widehat{\tau}_{\W} \mid \x, \tf, \Dtrain) 
            &\approx \frac{\var_{\w}(\widehat{\tau}_{\w}(\x))}{\var_{\w}(\muh_{\w}(\x, \tcf))}, \\
            &= \frac{\var_{\w} \left( \muh_{\w}(\x, \tf) \right) - 2\mathrm{Cov}_{\w}(\muh_{\w}(\x, \tf), \muh_{\w}(\x, \tcf))}{\var_{\w} \left( \muh_{\w}(\x, \tcf) \right)} + 1,
        \end{split}
    \end{equation}
    where for binary $\T=t$, $\tcf = (1 - \tf)$.
    \begin{proof}
        \begin{subequations}
            \begin{align}
                \mi(\Yt; \widehat{\tau}_{\W} \mid \x, \tf, \D)
                &= \ent(\widehat{\tau}_{\W} \mid \x, \tf, \D) - \ent(\widehat{\tau}_{\W} \mid \Yt, \x, \tf, \D) 
                \label{causalbald:eq:rho_proof_a} \\
                &= \ent(\widehat{\tau}_{\W} \mid \x, \tf, \D) - \E_{\yt \sim p(\Yt | \x, \tf, \D)} \ent(\widehat{\tau}_{\W} \mid \yt, \x, \tf) 
                \label{causalbald:eq:rho_proof_b} \\
                &= \frac{1}{2}\log(2\pi \var(\widehat{\tau}_{\W})) - \E_{\yt \sim p(\Yt | \x, \tf, \D)} \left[ \frac{1}{2}\log(2\pi \var(\widehat{\tau}_{\W} \mid \yt)) \right]
                \label{causalbald:eq:rho_proof_c} \\
                &\geq \frac{1}{2}\log(2\pi \var(\widehat{\tau}_{\W})) - \frac{1}{2} \log(2\pi \E \left[ \var(\widehat{\tau}_{\W} \mid \yt) \right]) 
                \label{causalbald:eq:rho_proof_d} \\
                &= \frac{1}{2}\log{ \left( \frac{ \var(\widehat{\tau}_{\W})}{\E \left[ \var(\widehat{\tau}_{\W} \mid \yt) \right]} \right) }
                \label{causalbald:eq:rho_proof_e}, \\
                \intertext{and we can further expand the fraction to}
                \frac{\var(\widehat{\tau}_{\W} \mid \x, \tf, \D)}{\E \left[ \var(\widehat{\tau}_{\W} \mid \yt) \right]}
                \label{causalbald:eq:rho_proof_f} &=\\
                &= \frac{\var(\widehat{\tau}_{\W} \mid \x, \tf, \D)}{\var_{\w \sim p(\W \mid \D)}(\muh_{\w}(\x, \tcf))}
                \label{causalbald:eq:rho_proof_g} \\
                &= \frac{\var_{\w \sim p(\W \mid \D)}(\widehat{\tau}_{\w}(\x) \mid \tf)}{\var_{\w}(\muh_{\w}(\x, \tcf))}
                \label{causalbald:eq:rho_proof_h} \\
                &= \frac{\var_{\w}(\muh_{\w}(\x, 1) - \muh_{\w}(\x, 0) \mid \tf)}{\var_{\w}(\muh_{\w}(\x, \tcf))}
                \label{causalbald:eq:rho_proof_i} \\
                &= \frac{\var_{\w }(\muh_{\w}(\x, \tf) - \muh_{\w}(\x, \tcf))}{\var_{\w}(\muh_{\w}(\x, \tcf))}
                \label{causalbald:eq:rho_proof_j} \\
                &= \frac{\var_{\w}(\muh_{\w}(\x, \tf)) + \var_{\w}(\muh_{\w}(\x, \tcf)) - 2 \cov_{\w}(\muh_{\w}(\x, \tf), \muh_{\w}(\x, \tcf))}{\var_{\w}(\muh_{\w}(\x, \tcf))}
                \label{causalbald:eq:rho_proof_k} \\
                &= \frac{\var_{\w}(\muh_{\w}(\x, \tf)) - 2 \cov_{\w}(\muh_{\w}(\x, \tf), \muh_{\w}(\x, \tcf))}{\var_{\w}(\muh_{\w}(\x, \tcf))} + 1,
                \label{causalbald:eq:rho_proof_l}
            \end{align}
            \label{causalbald:eq:rho_proof}
        \end{subequations}
        where \eqref{causalbald:eq:rho_proof_a} by definition of mutual information;
        \eqref{causalbald:eq:rho_proof_a}-\eqref{causalbald:eq:rho_proof_b} from the result of \citet{houlsby2011bayesian};
        \eqref{causalbald:eq:rho_proof_b}-\eqref{causalbald:eq:rho_proof_c} by Assumption 1. and Corollary \ref{causalbald:c:ent_normal};
        \eqref{causalbald:eq:rho_proof_c}-\eqref{causalbald:eq:rho_proof_d} by Jensen's inequality;
        \eqref{causalbald:eq:rho_proof_d}-\eqref{causalbald:eq:rho_proof_e} by the logarithmic quotient identity;
        \eqref{causalbald:eq:rho_proof_f}-\eqref{causalbald:eq:rho_proof_g} by Lemma \ref{causalbald:l:counterfactual_variance};
        \eqref{causalbald:eq:rho_proof_g}-\eqref{causalbald:eq:rho_proof_h} by definition of the variance.
        \eqref{causalbald:eq:rho_proof_h}-\eqref{causalbald:eq:rho_proof_i} by definition of $\widehat{\tau}_{\w}$;
        \eqref{causalbald:eq:rho_proof_i}-\eqref{causalbald:eq:rho_proof_j} by symmetry of the variance of the difference of two random variables;
        \eqref{causalbald:eq:rho_proof_j}-\eqref{causalbald:eq:rho_proof_k} by the definition of the variance of the difference of two random variables; and
        \eqref{causalbald:eq:rho_proof_k}-\eqref{causalbald:eq:rho_proof_l} by cancelling terms.
    \end{proof}
    \label{causalbald:th:rho-BALD}
\end{theorem}
\newpage
\begin{lemma}
    Under the following assumptions
    \begin{enumerate}
        \item Consistency $\Y \mid \T = \Yt$;
        \item Unconfoundedness $(\Yzero, \Yone) \independent \T \mid \X$;
    \end{enumerate}
    \begin{equation}
        \E_{\yt \sim p(\Yt | \x, \tf, \D)} \left[ \var(\widehat{\tau}_{\W} \mid \yt) \right] \approx \E_{\yt \sim p(\Yt | \x, \tf, \D)} \left[ \var_{\w \sim p(\W \mid \Dtrain)}(\muh_{\w}(\x, \tcf)) \right],
    \end{equation}
    where for binary $\T=\tf$, $\tcf = (1 - \tf)$.
    \begin{proof}
            \begin{align}
                \E&_{\yt \sim p(\Yt | \x, \tf, \D)} \left[ \var(\widehat{\tau}_{\W} \mid \yt) \right]
                = \E_{p(\yt)} \left[ \E_{p(\w)} \left[ \left(\tauhw - \E_{p(\w)} \left[ \tauhw \mid \yt \right] \right)^2 \mid \yt \right] \right],
                \label{causalbald:eq:counter_var_a} \\
                &= \E_{p(\yt)} \left[ \E_{p(\w)} \left[ \left( \E[\Yone - \Yzero \mid \x, \w]  - \E_{p(\w)} \left[ \E[\Yone - \Yzero \mid \x, \w]  \mid \yt \right] \right)^2 \mid \yt \right] \right],
                \label{causalbald:eq:counter_var_b} \\
                &= \medmath{\E_{p(\yt)} \left[ \E_{p(\w)} \left[ \left( \E[\Yone \mid \x, \w] - \E[\Yzero \mid \x, \w]  - \E_{p(\w)} \left[ \E[\Yone\mid \x, \w]  \mid \yt \right] + \E_{p(\w)} \left[ \E[\Yzero \mid \x, \w]  \mid \yt \right] \right)^2 \mid \yt \right] \right]},
                \label{causalbald:eq:counter_var_c} \\
                &= \medmath{\E_{p(\yt)} \left[ \E_{p(\w)} \left[ \left( \left(\E[\Yone \mid \x, \w]  - \E_{p(\w)} \left[ \E[\Yone\mid \x, \w]  \mid \yt \right]\right) - \left(\E[\Yzero \mid \x, \w] - \E_{p(\w)} \left[ \E[\Yzero \mid \x, \w]  \mid \yt \right]\right) \right)^2 \mid \yt \right] \right],}
                \label{causalbald:eq:counter_var_d} \\
                &= \medmath{\E_{p(\yt)} \left[ \E_{p(\w)} \left[ \left( \left(\E[\Yt \mid \x, \w]  - \E_{p(\w)} \left[ \E[\Yt \mid \x, \w]  \mid \yt \right]\right) - \left(\E[\Ytcf \mid \x, \w] - \E_{p(\w)} \left[ \E[\Ytcf \mid \x, \w]  \mid \yt \right]\right) \right)^2 \mid \yt \right] \right]},
                \label{causalbald:eq:counter_var_e} \\
                &= \medmath{\E_{p(\yt)} \left[ \E_{p(\w \mid \yt)} \left[ \left( \left(\E_{p(\yt \mid \x, \w)}[\yt]  - \E_{p(\w  \mid \yt)} \left[ \E_{p(\yt \mid \x, \w)}[\yt] \right]\right) - \textcolor{blue}{\left(\E_{p(\ytcf \mid \x, \w)}[\ytcf] - \E_{p(\w  \mid \yt)} \left[ \E_{p(\ytcf \mid \x, \w)}[\ytcf]\right]\right)} \right)^2 \right] \right]},
                \label{causalbald:eq:counter_var_f} \\
                &= \medmath{\E_{p(\yt)} \left[ \E_{p(\w \mid \yt)} \left[ \left( \textcolor{red}{\underbrace{\left(\E_{p(\yt \mid \x, \w)}[\yt]  - \E_{p(\w  \mid \yt)} \left[ \E_{p(\yt \mid \x, \w)}[\yt] \right]\right)}_{\displaystyle \approx 0}} - \textcolor{blue}{\left(\E_{p(\ytcf \mid \x, \w)}[\ytcf] - \E_{p(\w)} \left[ \E_{p(\ytcf \mid \x, \w)}[\ytcf]\right]\right)} \right)^2 \right] \right]},
                \label{causalbald:eq:counter_var_g} \\
                &\approx \E_{p(\yt)} \left[ \E_{p(\w \mid \yt)} \left[ \left(\E_{p(\ytcf \mid \x, \w)}[\ytcf] - \E_{p(\w)} \left[ \E_{p(\ytcf \mid \x, \w)}[\ytcf]\right] \right)^2 \right] \right],
                \label{causalbald:eq:counter_var_h} \\
                &= \E_{p(\yt)} \left[ \E_{p(\w \mid \yt)} \left[ \left( \muh_{\w}(\x, \tcf) - \E_{p(\w)} \left[ \muh_{\w}(\x, \tcf) \right] \right)^2 \right] \right],
                \label{causalbald:eq:counter_var_i} \\
                &= \E_{\yt \sim p(\Yt | \x, \tf, \D)} \left[ \var_{\w \sim p(\W \mid \Dtrain)}(\muh_{\w}(\x, \tcf)) \right],
                \label{causalbald:eq:counter_var_j}
            \end{align}
            \label{causalbald:eq:counter_var}
        where \eqref{causalbald:eq:counter_var_a} by definition of variance;
        \eqref{causalbald:eq:counter_var_a}-\eqref{causalbald:eq:counter_var_b} by definition of $\widehat{\tau}_{\w}$;
        \eqref{causalbald:eq:counter_var_b}-\eqref{causalbald:eq:counter_var_c} by linearity of expectations;
        \eqref{causalbald:eq:counter_var_c}-\eqref{causalbald:eq:counter_var_d} by grouping terms;
        \eqref{causalbald:eq:counter_var_d}-\eqref{causalbald:eq:counter_var_e} by symmetry of the square;
        \eqref{causalbald:eq:counter_var_e}-\eqref{causalbald:eq:counter_var_f} by rewriting expectations in terms of densities;
        \eqref{causalbald:eq:counter_var_f}-\eqref{causalbald:eq:counter_var_g} \textcolor{blue}{the observed potential outcome does not have an effect on the expectation of the model for the counterfactual outcome};
        \eqref{causalbald:eq:counter_var_g}-\eqref{causalbald:eq:counter_var_h} \textcolor{red}{we drop the term as an approximation as we cannot estimate here how much the expected outcome is going to change---the conservative assumption is that will not change};
        \eqref{causalbald:eq:counter_var_h}-\eqref{causalbald:eq:counter_var_i} by definition of $\muh_{\w}$;
        \eqref{causalbald:eq:counter_var_i}-\eqref{causalbald:eq:counter_var_j} by definition of variance;
    \end{proof}
    \label{causalbald:l:counterfactual_variance}
\end{lemma}

\subsubsection{$\rho$-BALD Failure}
\label{causalbald:a:rho_failure}
Consider two examples in $\Dpool$, $(\x_1, \tf_1)$ and $(\x_2, \tf_2)$ where $\var_{\w}\linebreak[0](\muh_{\w}(\x_1, \tf_1)) = \var_{\w}(\muh_{\w}(\x_1, \tcf_1))$ and $\var_{\w}(\muh_{\w}(\x_2, \tf_2)) =\var_{\w}( \muh_{\w}(\x_2, \tcf_2))$. 
That is, for each point we are as uncertain about the conditional expectation given the observed treatment as we would be given the counterfactual treatment. 
Further, let $\cov_{\w}(\muh_{\w}(\x_1, \tf_1), \muh_{\w}(\x_1, \tcf_1)) = \cov_{\w}(\muh_{\w}(\x_2, \tf_2), \muh_{\w}(\x_2, \tcf_2))$ 
and $\var_{\w}(\muh_{\w}(\x_1, \tf_1)) > \var_{\w}(\muh_{\w}(\x_2, \tf_2))$.
In this scenario $\rho$--BALD would rank these two points equally, but in practice it may be preferable to choose $(\x_1, \tf_1)$ over $(\x_2, \tf_2)$ as it would more likely be a point as yet unseen by the model.
When naively acquiring multiple points per acquisition step, this method biases training data to the modes of $\Dpool$.

\subsection{$\mu\pi$BALD}

\begin{figure}[ht]
    \centering
    \includegraphics[width=0.28\textwidth]{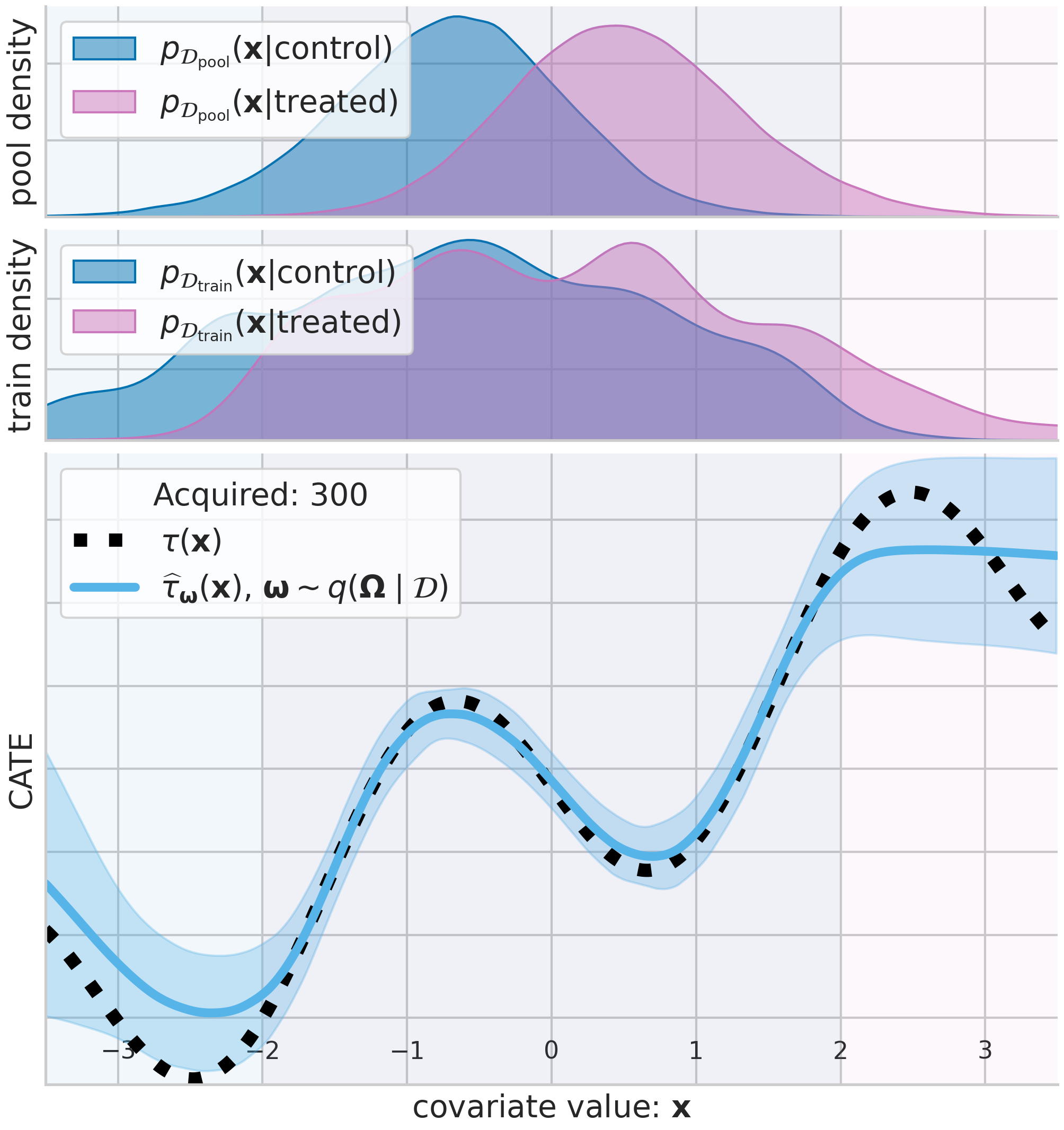}
    \caption{$\mu\pi$BALD}
    \label{causalbald:fig:bias_mupi}
\end{figure}

The most straightforward way to combine knowledge about a data point's information gain and overlap is to simply multiply \mubald \eqref{causalbald:eq:mu-bald-estimate} by the propensity acquisition term \eqref{causalbald:eq:propensity-acquisition}:
\begin{definition}
    $\mu\pi$BALD
    \begin{equation*}
        \mi(\mu \pi \mid \x, \tf, \Dtrain) \equiv (1 - \widehat{\pi}_{\tf}(\x)) \var_{\w \sim p(\W \mid \Dtrain)} \left( \muh_{\w}(\x, \tf) \right).
    \end{equation*}
\end{definition}
We can see in \cref{causalbald:fig:bias_mupi} that the acquisition of training data results in matched sampling as we saw for propensity acquisition in \cref{causalbald:fig:bias_pi}, but that the tails of the overlapping distributions extend further into the low density regions of the pool set support where overlap is satisfied.

\section{Baselines}
\subsection{S-type Error Information Gain}
\label{causalbald:a:sundin}

In their work, \citet{sundin2019active} assume that the underlying model is a Gaussian Process (GP) and also that they have access to the counterfactual outcome. Although GPs are suitable for uncertainty estimation, they do not scale up to high dimensional datasets (e.g. images). We propose to use Deep Ensembles and DUE for alleviating the capabilities issues and we modified the objective to be more suitable for our architecture.

Following the formulation from ~\citet{houlsby2011bayesian}, the acquisition strategy becomes $\argmax_{\x} \mathbb{H}[\gamma|\x,D] - \mathbb{E}_{\mathbb{H}[p(\theta|D)}[\gamma|\x,\theta]]$, where $\gamma(\x)=\text{probit}^{-1}(-\frac{|\mathbf{E}_{p(\tau\mid \x, \Dtrain)}[\tau]|}{\sqrt{\text{Var}(\tau|\x, \Dtrain)}})$, $\text{probit}^{-1}(\cdot)$ is the cumulative distribution function of normal distribution and $p(\gamma|\x,D) = \text{Bernoulli}(\gamma)$. With DUE (Deep Kernel Learning method) Deep Ensembles (samples from $p(\theta|D)$ we can compute those terms similarly to how we implemented our BALD objectives.

Below is an example of how this was implemented in PyTorch:
\begin{verbatim}
tau_mu = mu1s - mu0s
tau_var = var1s + var0s + 1e-07
gammas = torch.distributions.normal.Normal(0, 1).cdf(
    -tau_mu.abs() / tau_var.sqrt()
)
gamma = gammas.mean(-1)
predictive_entropy = dist.Bernoulli(gamma).entropy()
conditional_entropy = dist.Bernoulli(gammas).entropy().mean(-1)
# it can get negative very small number 
# because of numerical instabilities
scores = (predictive_entropy - conditional_entropy).clamp_min(1e-07)
\end{verbatim}

\section{Datasets}
\label{causalbald:a:datasets}

\subsection{Synthetic Data}
\label{causalbald:a:synthetic}

We modify the synthetic dataset presented by \citet{kallus2019interval}. Our dataset is described by the following structural causal model (SCM):
\begin{subequations}
    \begin{align}
        \x &\defeq N_{\x}, \\
        \tf &\defeq N_{\tf}, \\
        \y &\defeq (2\tf - 1)\x + (2\tf - 1) - 2 \sin(2(2\tf - 1)\x) + 2 (1 + 0.5\x) + N_{\y},
    \end{align}
\end{subequations}
where $N_{\x} \sim \mathcal{N}(0, 1)$, $N_{\tf} \sim \text{Bern}(\text{sigmoid}(2 \x + 0.5))$, and $N_{\y} \sim \mathcal{N}(0, 1)$. 

Each random realization of the simulated dataset generates 10000 pool set examples, 1000 validation examples, and 1000 test examples. In the experiments we report results over 20 random realizations. The seeds for the random number generators are $i$, $i + 1$, and $i + 2$; $\{i \in [0, 1, \dots, 19]\}$, for the training, validation, and test sets, respectively.

\subsection{IHDP Data.} 
Infant Health and Development Program (IHDP) is a semisynthetic dataset~\citep{hill2011bayesian, shalit2017estimating} commonly used in literature to study the performance of causal effect estimation methods. The dataset consists of 747 cases, out of which 139 are assigned in treatment group and 608 in control. Each unit is represented by 25 covariates describing different aspects of the infants and their mothers.

\subsection{CMNIST Data.} 
\label{causalbald:a:CMNIST}
Following the setup from~\cite{jesson2021quantifying}, we use a simulated dataset based on MNIST~\citep{lecun1998mnist}.  
CMNIST is described by the following SCM:
\begin{subequations}
    \begin{align}
        \x &\defeq N_{\x}, \\
        \phi &\defeq \left( \text{clip}\left( \frac{\mu_{N_{\x}} - \mu_{\mathrm{c}}}{\sigma_{\mathrm{c}}}; -1.4, 1.4 \right) - \text{Min}_{\mathrm{c}} \right) \frac{\text{Max}_{\mathrm{c}} - \text{Min}_{\mathrm{c}}}{1.4 - \text{-}
        1.4}\\
        \tf &\defeq N_{\tf}, \\
        \y &\defeq (2\tf - 1)\phi + (2\tf - 1) - 2 \sin(2(2\tf - 1)\phi) + 2 (1 + 0.5\phi) + N_{\y},
    \end{align}
\end{subequations}
where $N_{\tf}$ (swapping $\mathrm{\x}$ for $\phi$), and $N_{\y}$ are as described in Appendix \ref{causalbald:a:synthetic}. 
$N_{\x}$ is a sample of an MNIST image. The sampled image has a corresponding label $c \in [0, \dots, 9]$. 
$\mu_{N_{\x}}$ is the average intensity of the sampled image. $\mu_{\mathrm{c}}$ and $\sigma_{\mathrm{c}}$ are the mean and standard deviation of the average image intensities over all images with label $\mathrm{c}$ in the MNIST training set. 
In other words, $\mu_{\mathrm{c}} = \E[\mu_{N_{\x}} \mid \mathrm{c}]$ and $\sigma^2_{\mathrm{c}} = \var[\mu_{N_{\x}} \mid \mathrm{c}]$. 
To map the high dimensional images $\x$ onto a one-dimensional manifold $\phi$ with domain $[-3, 3]$ above, we first clip the standardized average image intensity on the range $(-1.4, 1.4)$. 
Each digit class has its own domain in $\phi$, so there is a linear transformation of the clipped value onto the range $[\text{Min}_{\mathrm{c}}, \text{Max}_{\mathrm{c}}]$. 
Finally, $\text{Min}_{\mathrm{c}} = -2 + \frac{4}{10} \mathrm{c}$, and $\text{Max}_{\mathrm{c}} = -2 + \frac{4}{10}(\mathrm{c} + 1)$.

For each random realization of the dataset, the MNIST training set is split into training ($n=35000$) and validation ($n=15000$) subsets using the scikit-learn function train\_test\_split(). 
The test set is generated using the MNIST test set ($n=10000$). 
The random seeds are $\{i \in [0, 1, \dots, 19]\}$ for the 20 random realizations generated.

\section{Active Learning Setup Details}
\begin{table}[ht]
    \centering
    \caption{Summary of active learning setup per dataset.}
    \label{causalbald:table:acquisition_settings}
    \small
    \begin{tabular}{llll}
    \toprule
    \textbf{Dataset} & \textbf{Warm up size} & \textbf{Acq. size} & \textbf{\# of Acq.} \\
    \midrule
    Synthetic &   0 & 10 & 31 \\
    IHDP      & 100 & 10 & 38 (max) \\
    CMNIST    & 250 & 50 & 20 \\
    \bottomrule
    \end{tabular}
\end{table}

\section{More Results}
\label{causalbald:a:more_results}

\begin{figure}[!ht]
  \centering
  \begin{subfigure}[b]{\linewidth}
    \includegraphics[width=.9\linewidth]{figures/results/legends.png}
  \end{subfigure}
  \\~\\
  \begin{subfigure}[b]{0.32\linewidth}
    \centering
    \includegraphics[width=\linewidth]{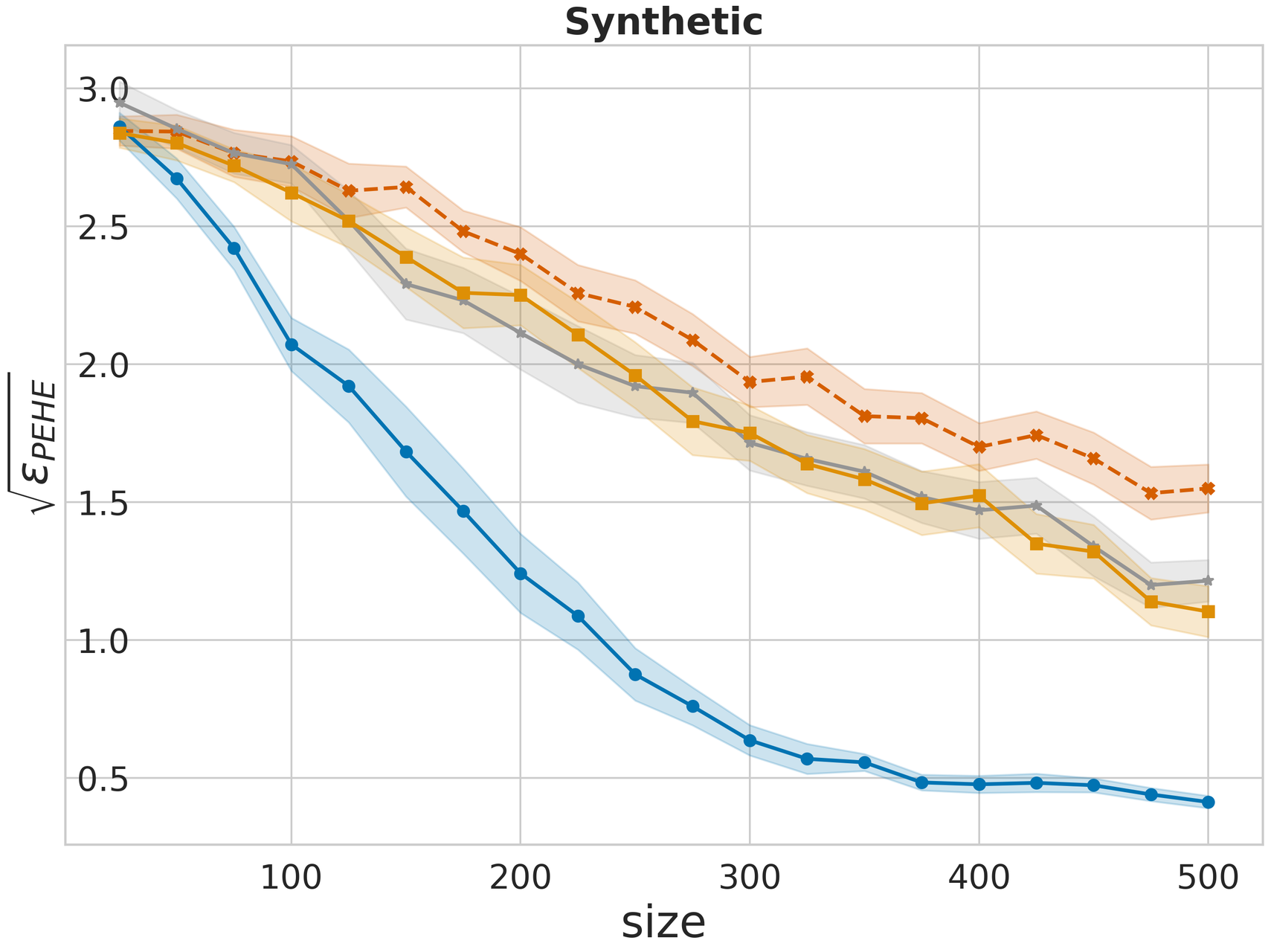}
  \end{subfigure}
  \begin{subfigure}[b]{0.32\linewidth}
    \centering
    \includegraphics[width=\linewidth]{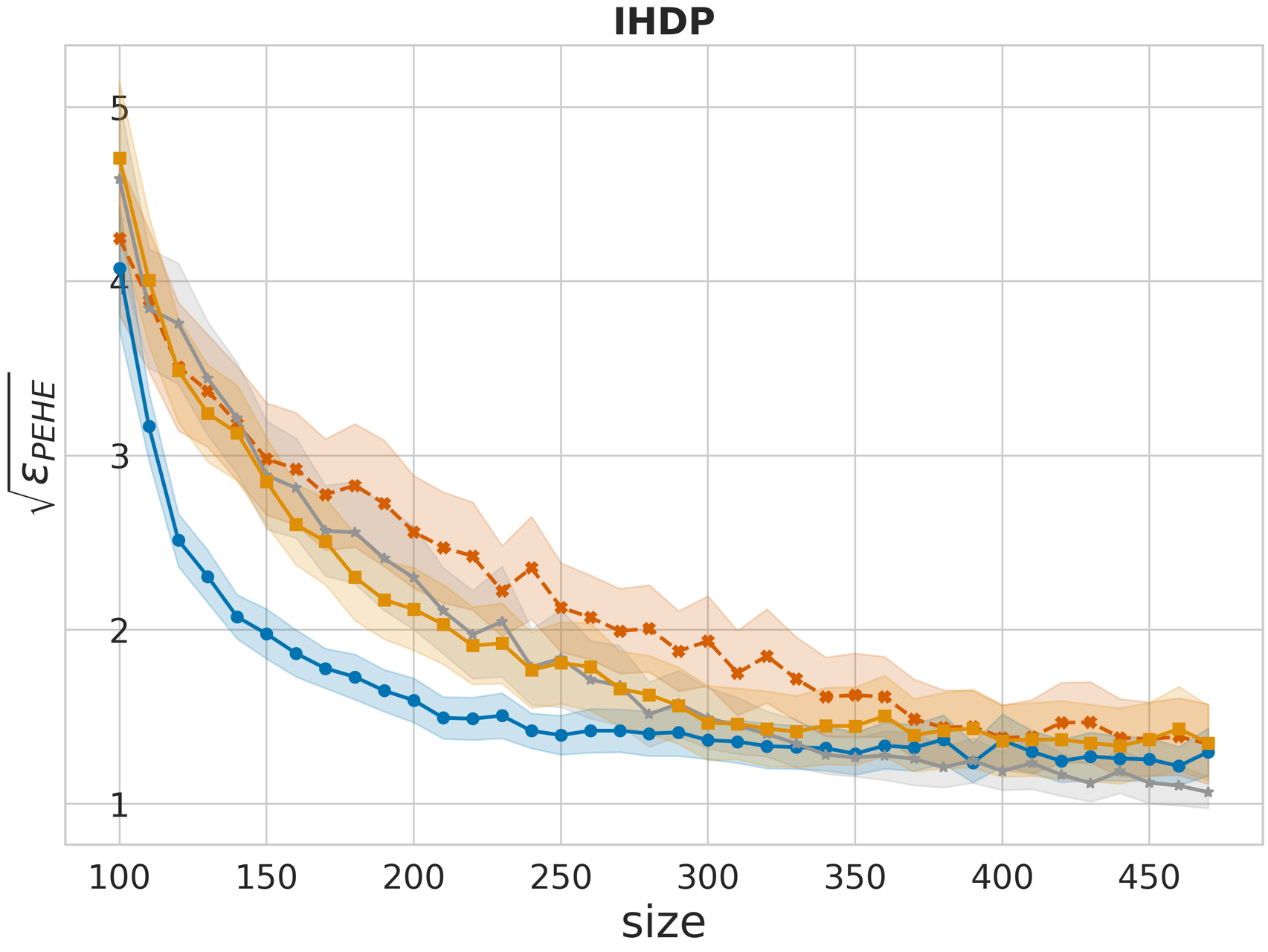}
  \end{subfigure}
  \begin{subfigure}[b]{0.32\linewidth}
    \centering
    \includegraphics[width=\linewidth]{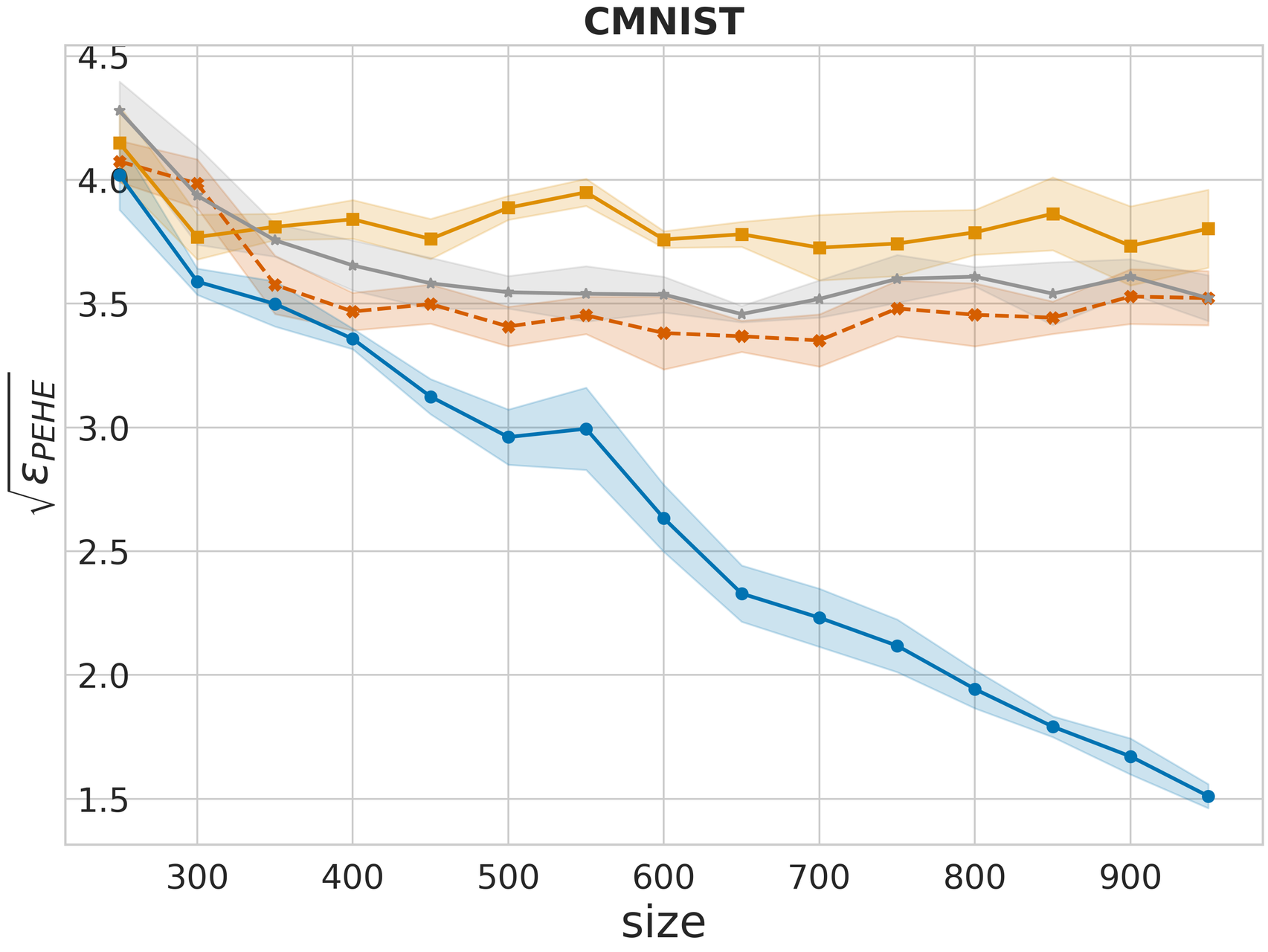}
  \end{subfigure}
  \begin{subfigure}[b]{0.32\linewidth}
    \centering
    \includegraphics[width=\linewidth]{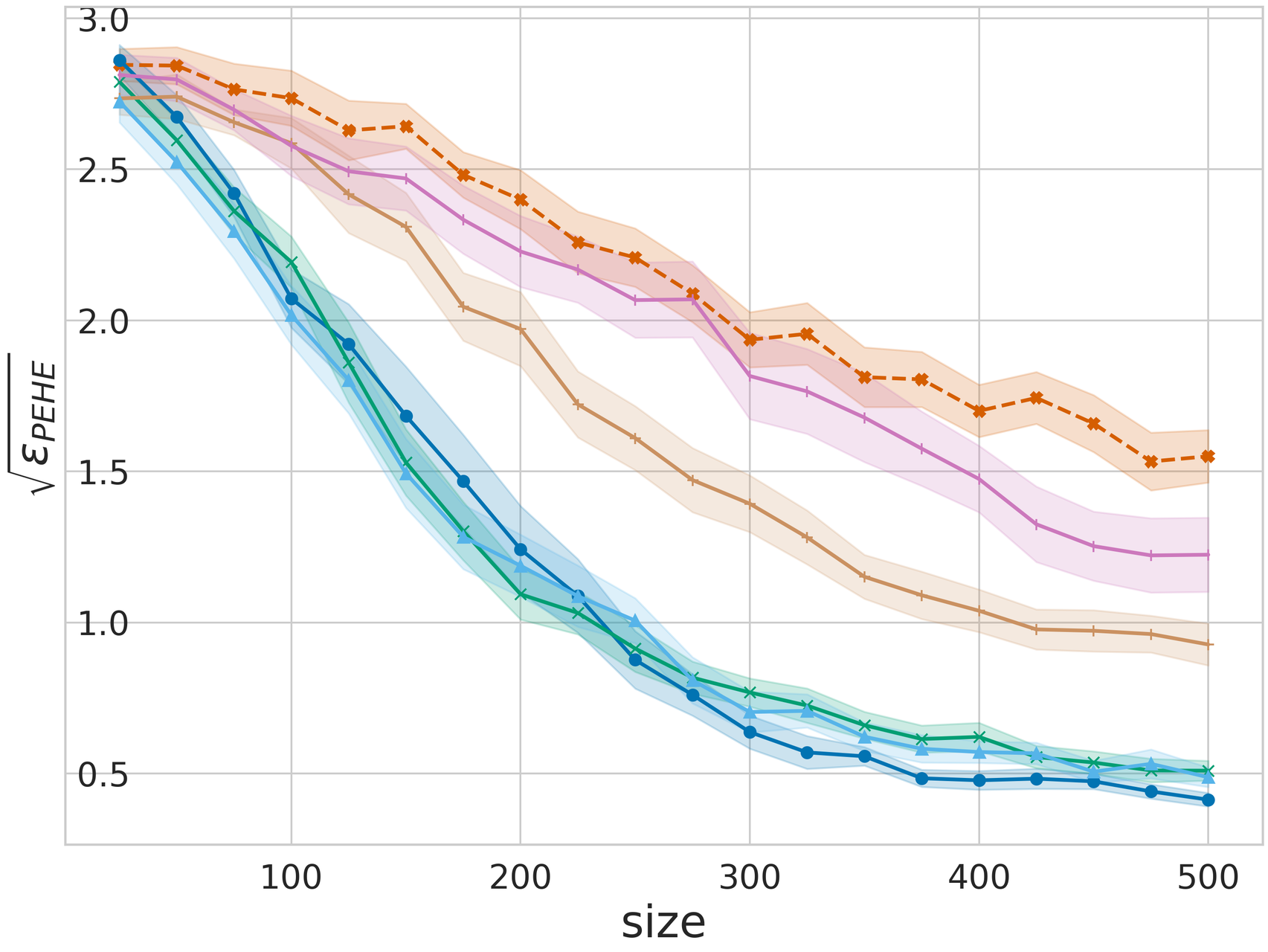}
  \end{subfigure}
  \begin{subfigure}[b]{0.32\linewidth}
    \centering
    \includegraphics[width=\linewidth]{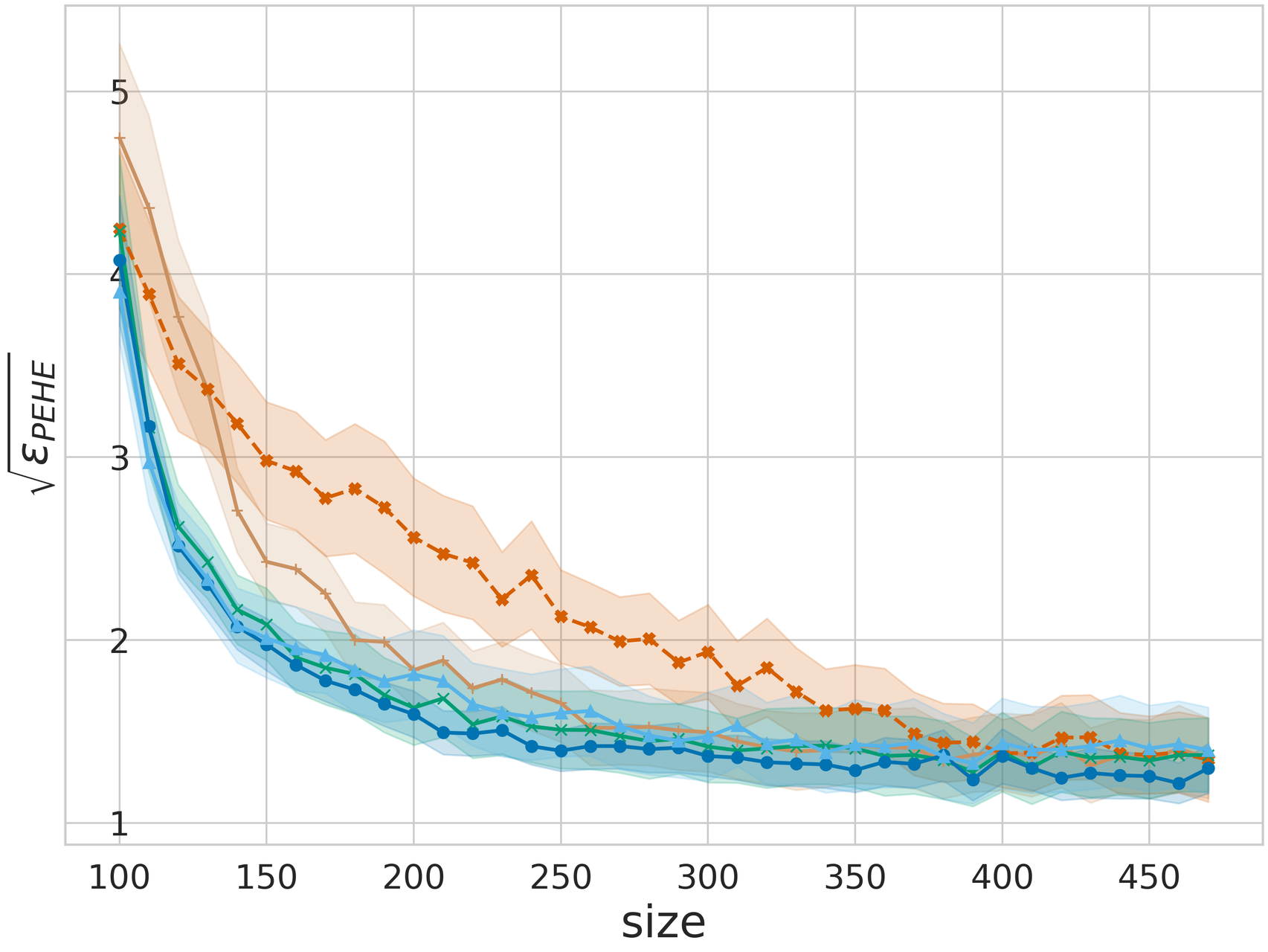}
  \end{subfigure}
  \begin{subfigure}[b]{0.32\linewidth}
    \centering
    \includegraphics[width=\linewidth]{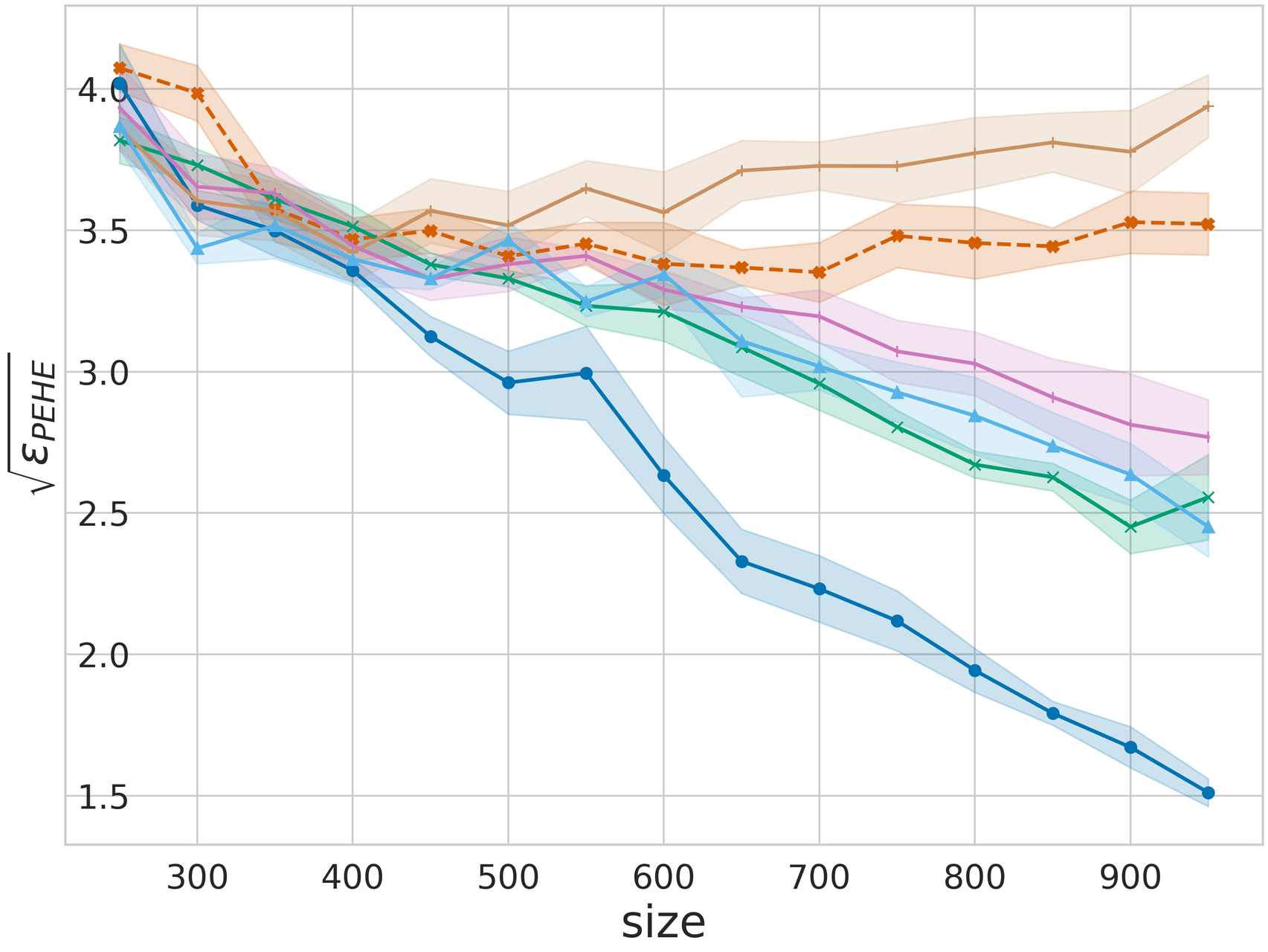}
  \end{subfigure}
  \caption{
    $\sqrt{\epsilon_{PEHE}}$ performance (shaded standard error) for Deep Ensembles based models. \textbf{(left to right)} \textbf{synthetic} (20 seeds), \textbf{IHDP} (50 seeds) and \textit{CMNIST } (5 seeds) dataset results, \textbf{(top to bottom)} comparison with baselines, comparison between BALD objectives. We observe that BALD objectives outperform the \textbf{random}, \textbf{$\boldsymbol\gamma$} and \textbf{propensity} acquisition functions significantly, suggesting that epistemic uncertainty aware methods that target reducible uncertainty can be more sample efficient.
  }
\label{causalbald:fig:ensemble_results}
\end{figure}

\begin{figure}[!ht]
  \centering
  \begin{subfigure}[b]{\linewidth}
    \includegraphics[width=.9\linewidth]{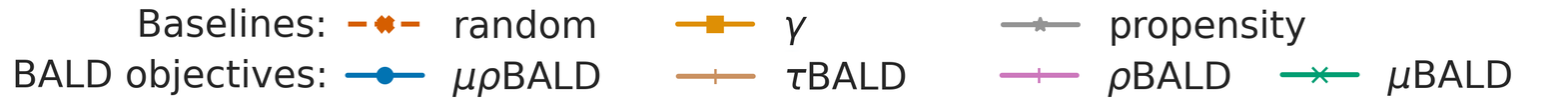}
  \end{subfigure}
  \\~\\
  \begin{subfigure}[b]{0.32\linewidth}
    \centering
    \includegraphics[width=\linewidth]{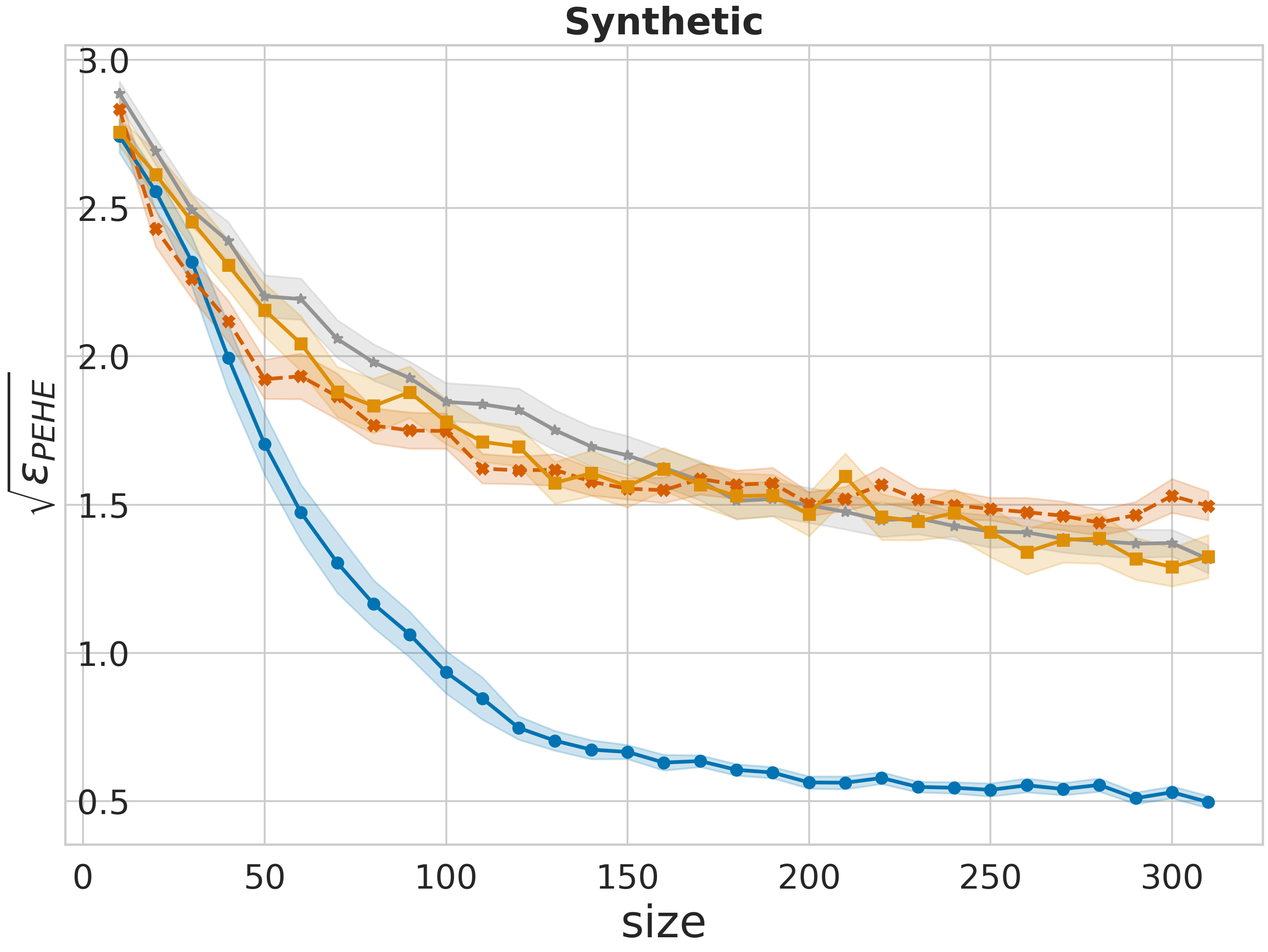}
  \end{subfigure}
  \begin{subfigure}[b]{0.32\linewidth}
    \centering
    \includegraphics[width=\linewidth]{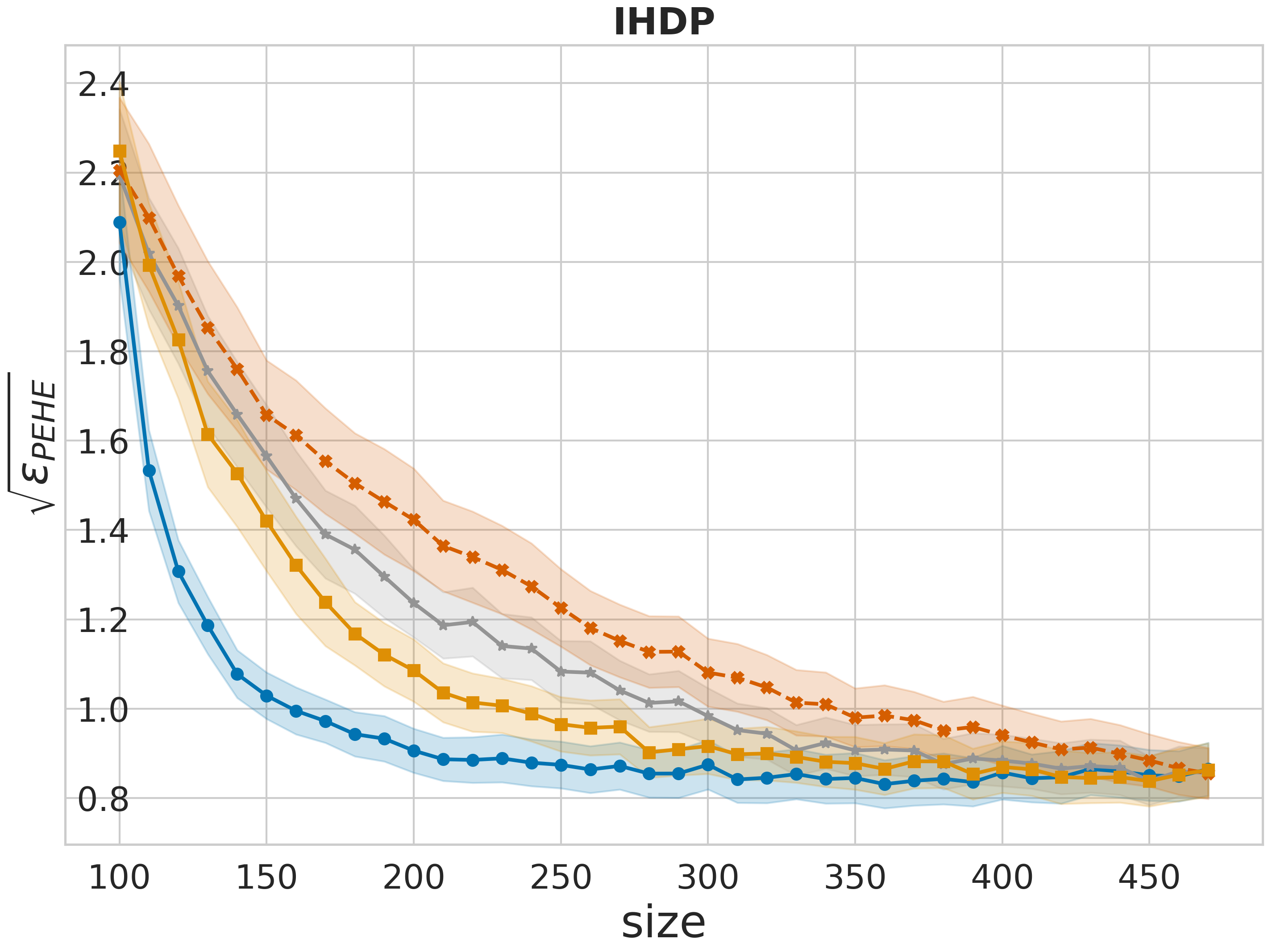}
  \end{subfigure}
  \begin{subfigure}[b]{0.32\linewidth}
    \centering
    \includegraphics[width=\linewidth]{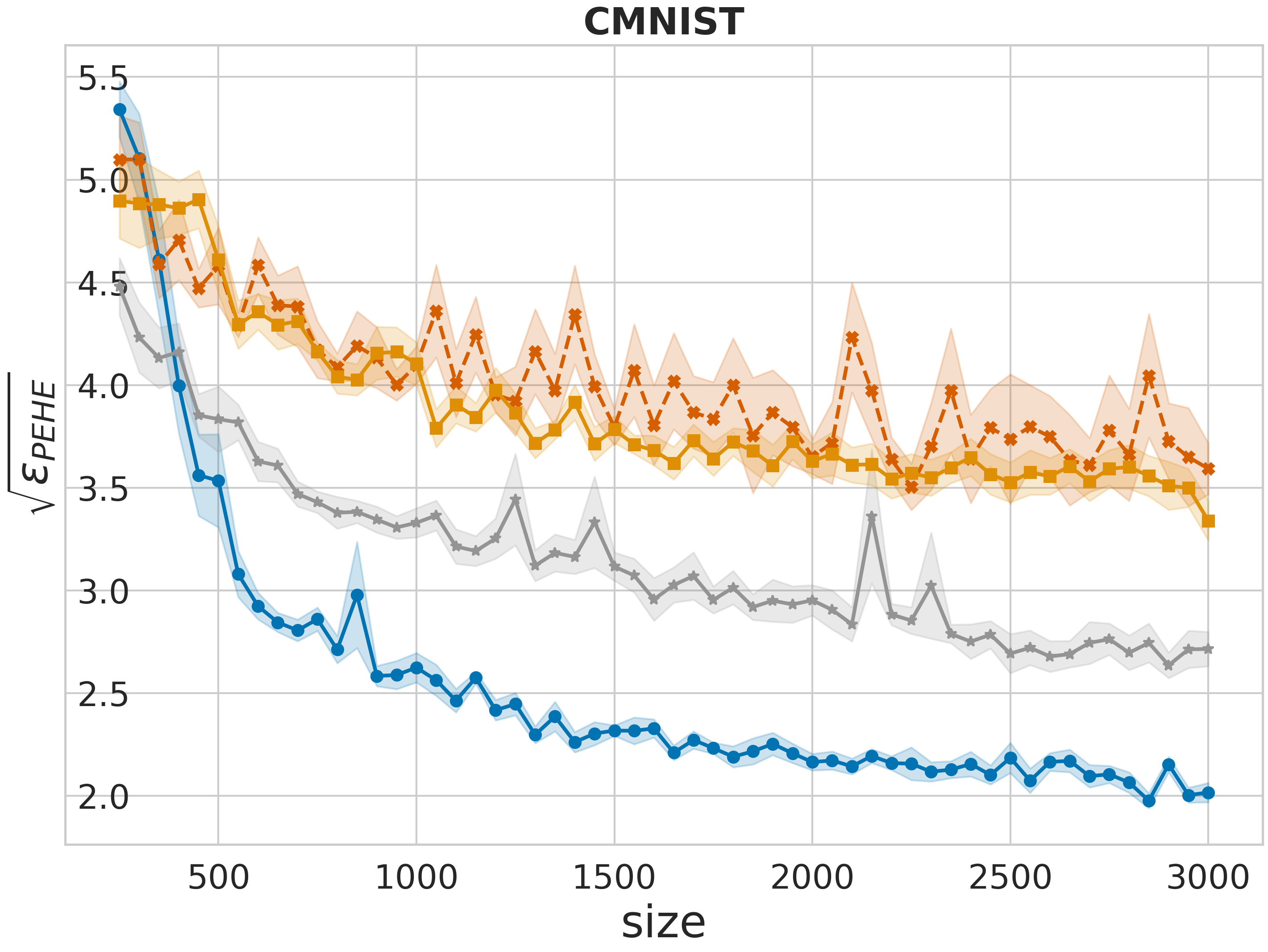}
  \end{subfigure}
  \begin{subfigure}[b]{0.32\linewidth}
    \centering
    \includegraphics[width=\linewidth]{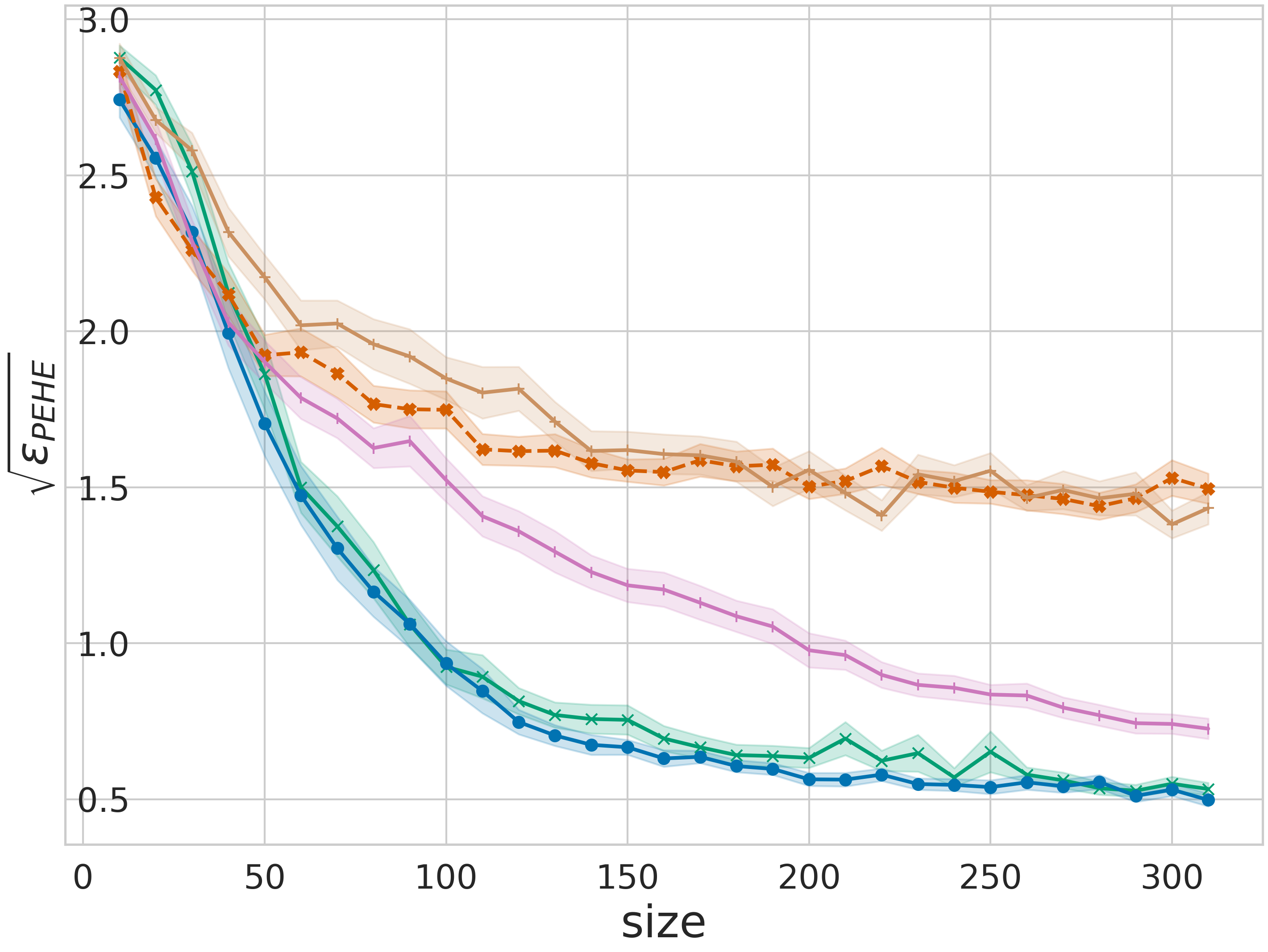}
  \end{subfigure}
  \begin{subfigure}[b]{0.32\linewidth}
    \centering
    \includegraphics[width=\linewidth]{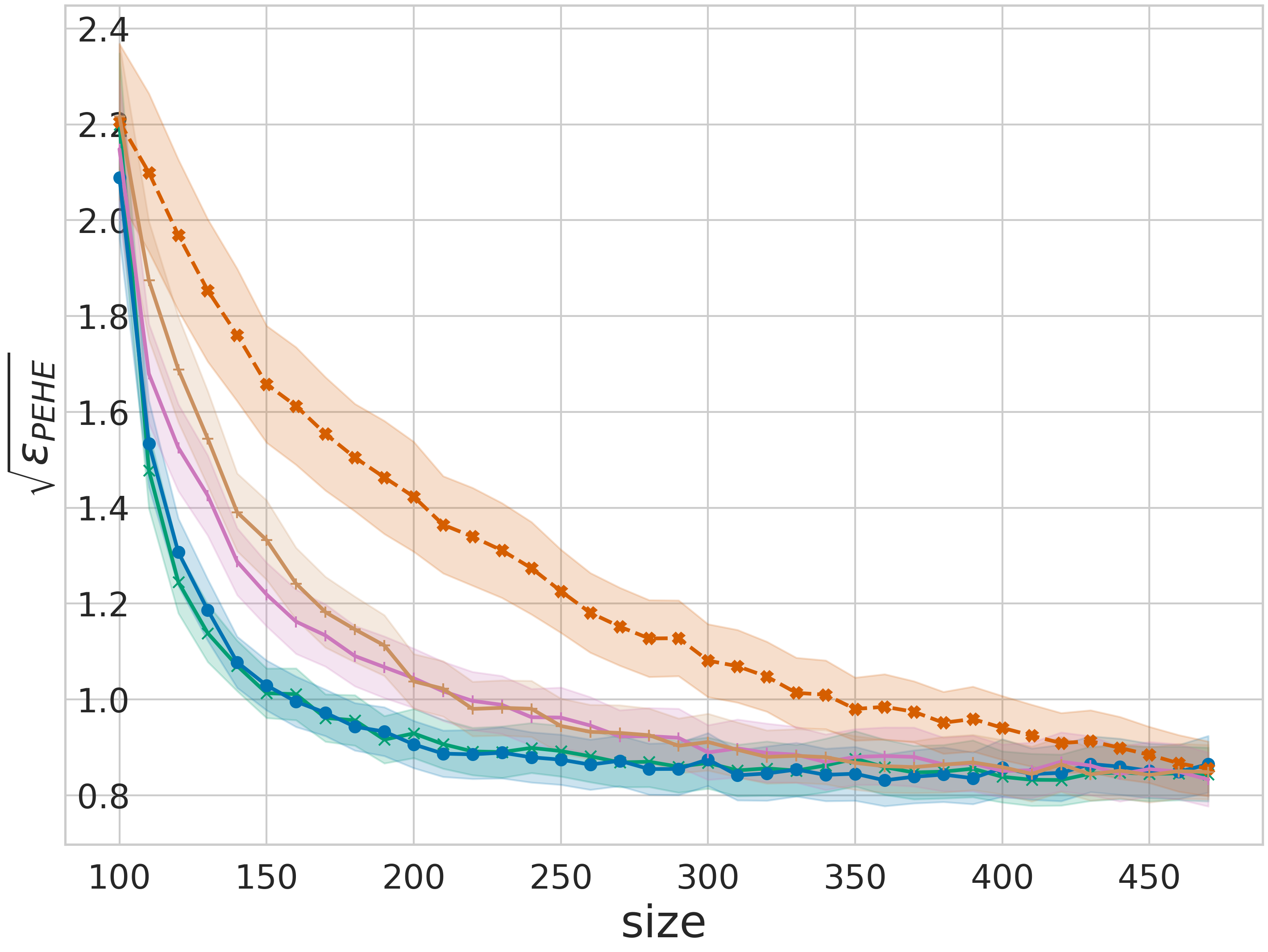}
  \end{subfigure}
  \begin{subfigure}[b]{0.32\linewidth}
    \centering
    \includegraphics[width=\linewidth]{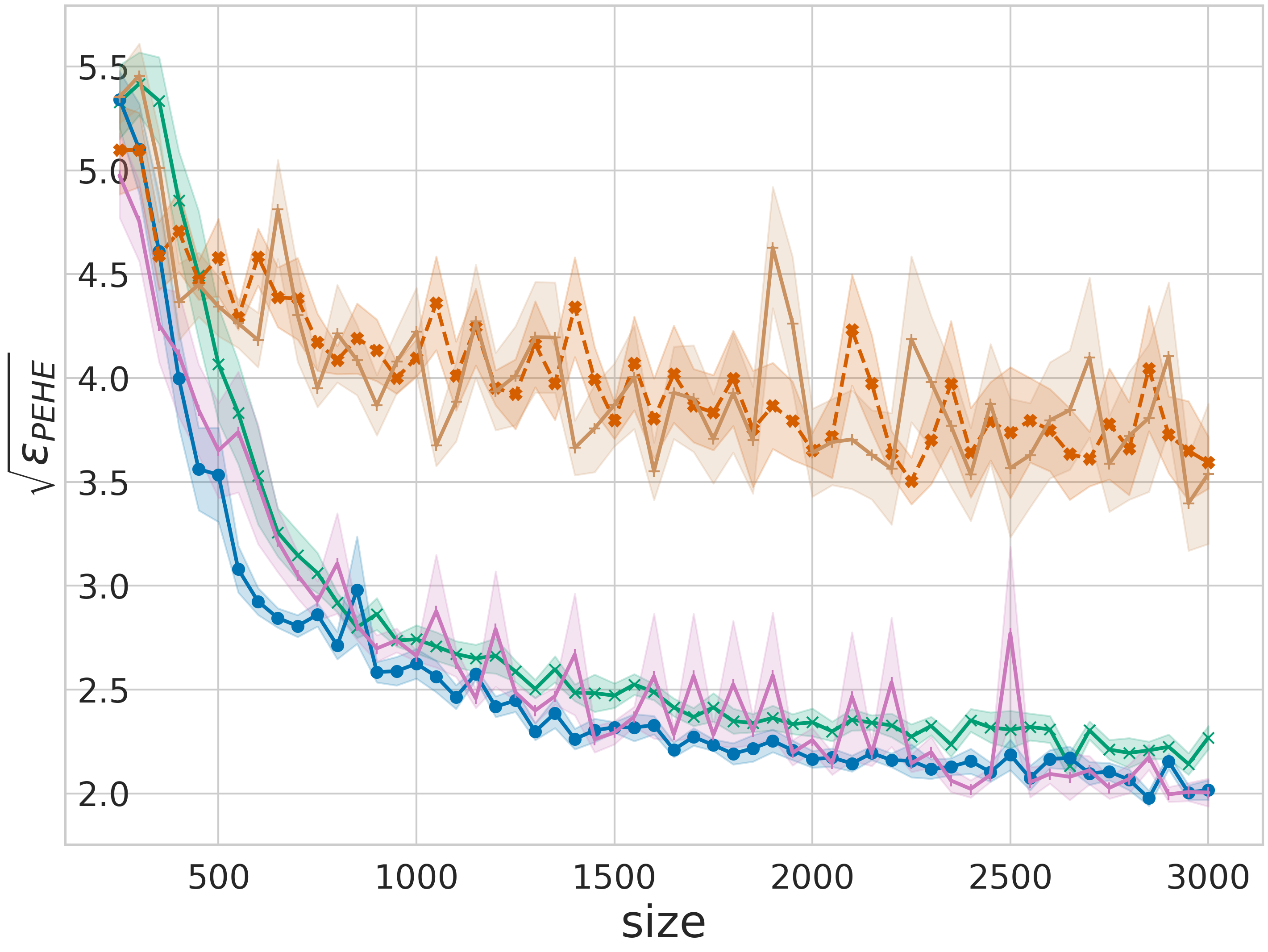}
  \end{subfigure}
  \caption{
    $\sqrt{\epsilon_{PEHE}}$ performance (shaded standard error) for DUE models. \textbf{(left to right)} \textbf{synthetic} (40 seeds), and \textbf{IHDP} (200 seeds). We observe that BALD objectives outperform the \textbf{random}, \textbf{$\boldsymbol\gamma$} and \textbf{propensity} acquisition functions significantly, suggesting that epistemic uncertainty aware methods that target reducible uncertainty can be more sample efficient.
  }
\label{causalbald:fig:due_results}
\end{figure}

\FloatBarrier

\section{Compute}
\label{causalbald:a:compute}

We used a cluster of 8 nodes with 4 GPUs each (16 RTX 2080 and 16 Titan RTX). The total GPU hours is estimated to be:

8 baselines x (.5 + 1 + 1) days per dataset x (5 ensemble components * 0.25 GPU usage + 1 DUE * 0.3 GPU usage) x 24 hours = 744 GPU hours

\section{Neural Network Architecture}
\label{causalbald:a:architecture}

\noindent\textbf{Synthetic Architecture}
\begin{Verbatim}[fontsize=\small]
===========================================================
Layer (type:depth-idx)                         Output Shape
===========================================================
Sequential                                     --
├─NeuralNetwork: 1-1                           [64, 100]
│    └─Sequential: 2-1                         [64, 100]
│    │    └─Linear: 3-1                        [64, 100]
│    │    └─ResidualDense: 3-2                 [64, 100]
│    │    │    └─PreactivationDense: 4-1       [64, 100]
│    │    │    │    └─Sequential: 5-1          [64, 100]
│    │    │    │    │    └─Activation: 6-1     [64, 100]
│    │    │    │    │    └─Linear: 6-2         [64, 100]
│    │    │    └─Identity: 4-2                 [64, 100]
│    │    └─ResidualDense: 3-3                 [64, 100]
│    │    │    └─PreactivationDense: 4-3       [64, 100]
│    │    │    │    └─Sequential: 5-2          [64, 100]
│    │    │    │    │    └─Activation: 6-3     [64, 100]
│    │    │    │    │    └─Linear: 6-4         [64, 100]
│    │    │    └─Identity: 4-4                 [64, 100]
│    │    └─ResidualDense: 3-4                 [64, 100]
│    │    │    └─PreactivationDense: 4-5       [64, 100]
│    │    │    │    └─Sequential: 5-3          [64, 100]
│    │    │    │    │    └─Activation: 6-5     [64, 100]
│    │    │    │    │    └─Linear: 6-6         [64, 100]
│    │    │    └─Identity: 4-6                 [64, 100]
│    │    └─Activation: 3-5                    [64, 100]
│    │    │    └─Sequential: 4-7               [64, 100]
│    │    │    │    └─Identity: 5-4            [64, 100]
│    │    │    │    └─LeakyReLU: 5-5           [64, 100]
│    │    │    │    └─Dropout: 5-6             [64, 100]
├─GMM: 1-2                                     [64, 5]
│    └─Linear: 2-2                             [64, 5]
│    └─Linear: 2-3                             [64, 5]
│    └─Sequential: 2-4                         [64, 5]
│    │    └─Linear: 3-6                        [64, 5]
│    │    └─Softplus: 3-7                      [64, 5]
===========================================================
Total params: 32,115
\end{Verbatim}

\noindent\textbf{IHDP Architecture}
\begin{Verbatim}[fontsize=\small]
==============================================================
Layer (type:depth-idx)                           Output Shape
==============================================================
Sequential                                       --
├─TARNet: 1-1                                    [64, 400]
│    └─NeuralNetwork: 2-1                        [64, 400]
│    │    └─Sequential: 3-1                      [64, 400]
│    │    │    └─Linear: 4-1                     [64, 400]
│    │    │    └─ResidualDense: 4-2              [64, 400]
│    │    │    │    └─PreactivationDense: 5-1    [64, 400]
│    │    │    │    │    └─Sequential: 6-1       [64, 400]
│    │    │    │    └─Identity: 5-2              [64, 400]
│    │    │    └─ResidualDense: 4-3              [64, 400]
│    │    │    │    └─PreactivationDense: 5-3    [64, 400]
│    │    │    │    │    └─Sequential: 6-2       [64, 400]
│    │    │    │    └─Identity: 5-4              [64, 400]
│    └─Sequential: 2-2                           [64, 400]
│    │    └─ResidualDense: 3-2                   [64, 400]
│    │    │    └─PreactivationDense: 4-4         [64, 400]
│    │    │    │    └─Sequential: 5-5            [64, 400]
│    │    │    │    │    └─Activation: 6-3       [64, 401]
│    │    │    │    │    └─Linear: 6-4           [64, 400]
│    │    │    └─Sequential: 4-5                 [64, 400]
│    │    │    │    └─Dropout: 5-6               [64, 401]
│    │    │    │    └─Linear: 5-7                [64, 400]
│    │    └─ResidualDense: 3-3                   [64, 400]
│    │    │    └─PreactivationDense: 4-6         [64, 400]
│    │    │    │    └─Sequential: 5-8            [64, 400]
│    │    │    │    │    └─Activation: 6-5       [64, 400]
│    │    │    │    │    └─Linear: 6-6           [64, 400]
│    │    │    └─Identity: 4-7                   [64, 400]
│    │    └─Activation: 3-4                      [64, 400]
│    │    │    └─Sequential: 4-8                 [64, 400]
│    │    │    │    └─Identity: 5-9              [64, 400]
│    │    │    │    └─ELU: 5-10                  [64, 400]
│    │    │    │    └─Dropout: 5-11              [64, 400]
├─GMM: 1-2                                       [64, 5]
│    └─Linear: 2-3                               [64, 5]
│    └─Linear: 2-4                               [64, 5]
│    └─Sequential: 2-5                           [64, 5]
│    │    └─Linear: 3-5                          [64, 5]
│    │    └─Softplus: 3-6                        [64, 5]
==============================================================
\end{Verbatim}

\noindent\textbf{CMNIST Architecture}
\begin{Verbatim}[fontsize=\small]
=============================================================================
Layer (type:depth-idx)                                       Output Shape
=============================================================================
Sequential                                                   --
├─TARNet: 1-1                                                [200, 100]
│    └─ResNet: 2-1                                           [200, 48]
│    │    └─Sequential: 3-1                                  [200, 48, 1, 1]
│    │    │    └─Conv2d: 4-1                                 [200, 12, 28, 28]
│    │    │    └─Identity: 4-2                               [200, 12, 28, 28]
│    │    │    └─ResidualConv: 4-3                           [200, 12, 28, 28]
│    │    │    │    └─Sequential: 5-1                        [200, 12, 28, 28]
│    │    │    │    │    └─PreactivationConv: 6-1            [200, 12, 28, 28]
│    │    │    │    │    └─PreactivationConv: 6-2            [200, 12, 28, 28]
│    │    │    │    └─Sequential: 5-2                        [200, 12, 28, 28]
│    │    │    │    │    └─Dropout2d: 6-3                    [200, 12, 28, 28]
│    │    │    │    │    └─Conv2d: 6-4                       [200, 12, 28, 28]
│    │    │    └─ResidualConv: 4-4                           [200, 24, 14, 14]
│    │    │    │    └─Sequential: 5-3                        [200, 24, 14, 14]
│    │    │    │    │    └─PreactivationConv: 6-5            [200, 12, 28, 28]
│    │    │    │    │    └─PreactivationConv: 6-6            [200, 24, 14, 14]
│    │    │    │    └─Sequential: 5-4                        [200, 24, 14, 14]
│    │    │    │    │    └─Dropout2d: 6-7                    [200, 12, 28, 28]
│    │    │    │    │    └─Conv2d: 6-8                       [200, 24, 14, 14]
│    │    │    └─ResidualConv: 4-5                           [200, 24, 14, 14]
│    │    │    │    └─Sequential: 5-5                        [200, 24, 14, 14]
│    │    │    │    │    └─PreactivationConv: 6-9            [200, 24, 14, 14]
│    │    │    │    │    └─PreactivationConv: 6-10           [200, 24, 14, 14]
│    │    │    │    └─Sequential: 5-6                        [200, 24, 14, 14]
│    │    │    │    │    └─Dropout2d: 6-11                   [200, 24, 14, 14]
│    │    │    │    │    └─Conv2d: 6-12                      [200, 24, 14, 14]
│    │    │    └─ResidualConv: 4-6                           [200, 48, 7, 7]
│    │    │    │    └─Sequential: 5-7                        [200, 48, 7, 7]
│    │    │    │    │    └─PreactivationConv: 6-13           [200, 24, 14, 14]
│    │    │    │    │    └─PreactivationConv: 6-14           [200, 48, 7, 7]
│    │    │    │    └─Sequential: 5-8                        [200, 48, 7, 7]
│    │    │    │    │    └─Dropout2d: 6-15                   [200, 24, 14, 14]
│    │    │    │    │    └─Conv2d: 6-16                      [200, 48, 7, 7]
│    │    │    └─ResidualConv: 4-7                           [200, 48, 7, 7]
│    │    │    │    └─Sequential: 5-9                        [200, 48, 7, 7]
│    │    │    │    │    └─PreactivationConv: 6-17           [200, 48, 7, 7]
│    │    │    │    │    └─PreactivationConv: 6-18           [200, 48, 7, 7]
│    │    │    │    └─Sequential: 5-10                       [200, 48, 7, 7]
│    │    │    │    │    └─Dropout2d: 6-19                   [200, 48, 7, 7]
│    │    │    │    │    └─Conv2d: 6-20                      [200, 48, 7, 7]
│    │    │    └─ResidualConv: 4-8                           [200, 48, 7, 7]
│    │    │    │    └─Sequential: 5-11                       [200, 48, 7, 7]
│    │    │    │    │    └─PreactivationConv: 6-21           [200, 48, 7, 7]
│    │    │    │    │    └─PreactivationConv: 6-22           [200, 48, 7, 7]
│    │    │    │    └─Sequential: 5-12                       [200, 48, 7, 7]
│    │    │    │    │    └─Dropout2d: 6-23                   [200, 48, 7, 7]
│    │    │    │    │    └─Conv2d: 6-24                      [200, 48, 7, 7]
│    │    │    └─AdaptiveAvgPool2d: 4-9                      [200, 48, 1, 1]
│    └─Sequential: 2-2                                       [200, 100]
│    │    └─ResidualDense: 3-2                               [200, 100]
│    │    │    └─PreactivationDense: 4-10                    [200, 100]
│    │    │    │    └─Sequential: 5-13                       [200, 100]
│    │    │    │    │    └─Activation: 6-25                  [200, 49]
│    │    │    │    │    └─Linear: 6-26                      [200, 100]
│    │    │    └─Sequential: 4-11                            [200, 100]
│    │    │    │    └─Dropout: 5-14                          [200, 49]
│    │    │    │    └─Linear: 5-15                           [200, 100]
│    │    └─ResidualDense: 3-3                               [200, 100]
│    │    │    └─PreactivationDense: 4-12                    [200, 100]
│    │    │    │    └─Sequential: 5-16                       [200, 100]
│    │    │    │    │    └─Activation: 6-27                  [200, 100]
│    │    │    │    │    └─Linear: 6-28                      [200, 100]
│    │    │    └─Identity: 4-13                              [200, 100]
│    │    └─Activation: 3-4                                  [200, 100]
│    │    │    └─Sequential: 4-14                            [200, 100]
│    │    │    │    └─Identity: 5-17                         [200, 100]
│    │    │    │    └─LeakyReLU: 5-18                        [200, 100]
│    │    │    │    └─Dropout: 5-19                          [200, 100]
├─GMM: 1-2                                                   [200, 5]
│    └─Linear: 2-3                                           [200, 5]
│    └─Linear: 2-4                                           [200, 5]
│    └─Sequential: 2-5                                       [200, 5]
│    │    └─Linear: 3-5                                      [200, 5]
│    │    └─Softplus: 3-6                                    [200, 5]
=============================================================================
\end{Verbatim}

\begin{figure}[!ht]
  \centering
    \includegraphics[width=.7\linewidth]{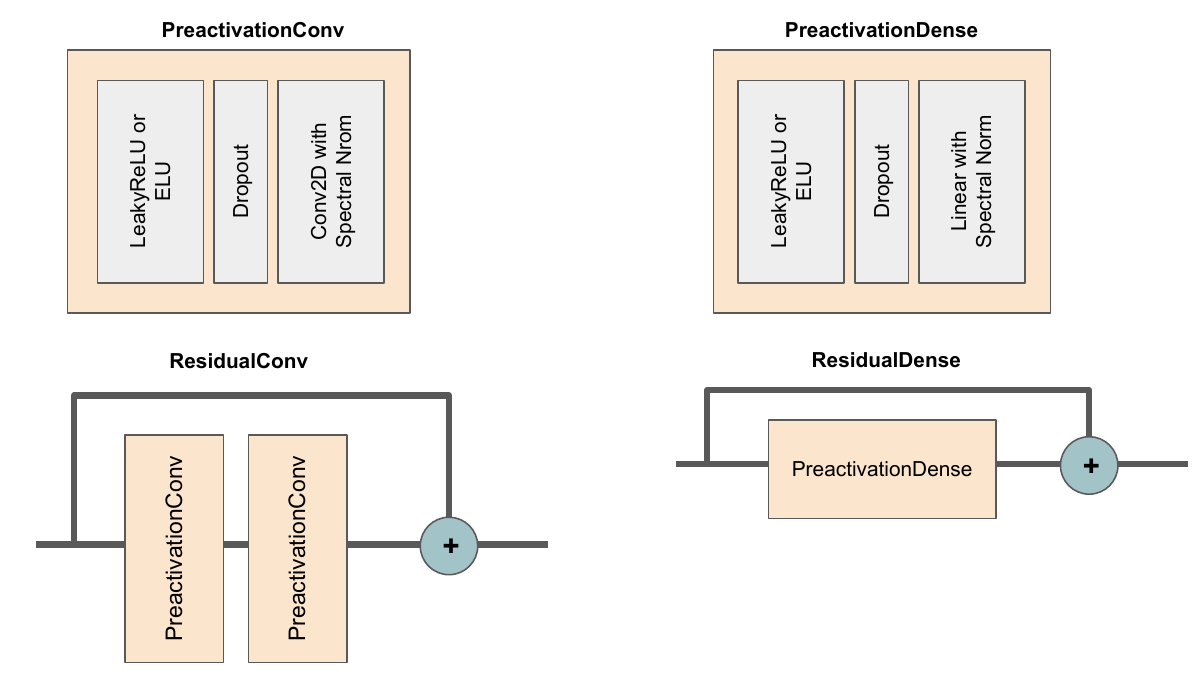}
  \caption{
\texttt{PreactivationConv} is a convolution layer with LeakyReLU (or ELU when slope is negative) activation, dropout and spectral norm applied~\citep{gouk2018regularisation,miyato2018spectral}. Similarly, \texttt{PreactivationDense} is a dense layer with BatchNorm~\citep{ioffe2015batch}, LeakyReLU (or ELU when slope is negative) activation and spectral norm applied~\citep{gouk2018regularisation,miyato2018spectral}. \texttt{ResidualConv} is the residual convolution layer, defined as
\texttt{PreactivationConv(PreactivationConv(x)) + SpectralNorm(1x1Conv(x))}
and \texttt{ResidualDense} are residual dense layers, defined as \texttt{PreactivationDense(x)+x)}.
}
\end{figure}

All experiments were trained using Adam optimizer.

\FloatBarrier

\subsection{Hyper-Parameters}
\begin{table}[h!]
\caption{Training hyper parameters for \textbf{Deep Ensemble} experiments}
\centering
\small
\begin{tabular}{llll}
\toprule
 \textbf{Parameter}&\textbf{Synthetic}&\textbf{IHDP}&\textbf{CMNIST}\\
\midrule
dim hidden & $100$ & $400$ & $100$ \\
dropout & $0.0$ & $0.15$ & $0.1$ \\
depth & $4$ & $3$ & $3$ \\
spectral norm & $12$ & $0.95$ & $24$ \\
learning rate & $0.001$ & $0.001$ & $0.001$ \\
negative slope & $0.0$ & $-1.0$ & $0.0$\\
\bottomrule
\end{tabular}
\end{table}

\begin{table}[h!]
\caption{Training hyper parameters for \textbf{DUE} experiments}
\centering
\small
\begin{tabular}{llll}
    \toprule
 \textbf{Parameter}&\textbf{Synthetic}&\textbf{IHDP}&\textbf{CMNIST}\\
\midrule
inducing points & $100$ & $100$ & $100$ \\
dim hidden & $100$ & $200$ & $200$ \\
dropout & $0.2$ & $0.1$ & $0.05$ \\
depth & $3$ & $3$ & $2$ \\
batch size & $200$ & $100$ & $64$ \\
spectral norm & $0.95$ & $0.95$ & $3.0$ \\
learning rate & $0.001$ & $0.001$ & $0.001$ \\
negative slope & $0.0$ & $-1.0$ & $-1.0$\\
\bottomrule
\end{tabular}
\end{table}

	}
	
\begin{savequote}[75mm]
	The first principle is that you must not fool yourself---and you are the easiest person to fool.
	\qauthor{Richard Feynman}
\end{savequote}
\chapter{Reproducibility Analysis}
{
	\label{chapter:reproducibility}

	\let\chapter\section
	\let\section\subsection
	\let\subsection\subsubsection
	\let\subsubsection\paragraph
	\let\paragraph\textbf
	
	{
		\graphicspath{{./dldd/}}

		\chapter{\dlddtitle}
		\label{chapter:dldd}

The senior author of `Deep Learning on a Data Diet' \citep{paul2021deep} recently gave a talk at our lab that explored this issue, presenting their novel metrics for pruning datasets. During the talk, the author of this current work suggested \emph{a correlation between the proposed GraNd score at initialization and input norms}, sparking further research into the effectiveness of these new pruning techniques. In this chapter, we delve deeper into this intriguing question, exploring the practicality and efficacy of these metrics for data pruning.

\textbf{`Deep Learning on a Data Diet'.} %
\citet{paul2021deep} introduce two novel metrics: \emph{Error L2 Norm (EL2N)} and \emph{Gradient Norm at Initialization (GraNd)}. These metrics aim to provide a more effective means of dataset pruning. It is important to emphasize that the GraNd score at initialization is calculated before any training has taken place, averaging over several randomly initialized models. This fact has been met with skepticism by reviewers\footnote{See also \shorturl{https://openreview.net/forum?id=Uj7pF-D-YvT&noteId=qwy3HouKSX}.}, but \citet{paul2021deep} specifically remark on GraNd at initialization:
\begin{quote}
\textbf{Pruning at initialization.} In all settings, GraNd scores can be used to select a training subset at initialization that achieves test accuracy significantly better than random, and in some cases, competitive with training on all the data. This is remarkable because GraNd only contains information about the gradient norm at initialization. This suggests that the geometry of the training distribution induced by a random network contains a surprising amount of information about the structure of the classification problem.
\end{quote}

\textbf{GraNd.} The GraNd score measures the magnitude of the gradient vector for a specific input sample in the context of neural network training over different parameter draws. The formula for calculating the (expected) gradient norm is:

\begin{equation}
\operatorname{GraNd}(x)=\mathbb{E}_{\theta_t}[\left\|\nabla_{\theta_t} L(f(x ; \theta_t), y)\right\|_2]
\end{equation}
 
where $\nabla_{\theta_t} L(f(\x; \theta_t), \y)$ is the gradient of the loss function $L$ with respect to the model's parameters $\theta_t$ at epoch $t$, $f(\x; \theta)$ is the model's prediction for input $\x$, and $\y$ is the true label for the input. We take an expectation over several training runs.
The gradient norm provides information about the model's sensitivity to a particular input and helps in identifying data points that have a strong influence on the learning process.

\textbf{EL2N.} The EL2N score measures the squared difference between the predicted and (one-hot) true labels for a specific input sample. The formula for calculating the EL2N score is:

\begin{equation}
\operatorname{EL2N}(x)=\mathbb{E}_{\theta_t}[\|f(x ; \theta_t)-y\|_2^2]
\end{equation}
 
where $f(\x; \theta)$ is the model's prediction for input $\x$, $\y$ is the (one-hot) true label for the input, and $\lVert \cdot \rVert_2$ denotes the Euclidean (L2) norm. The EL2N score provides insight into the model's performance on individual data points, allowing for a more targeted analysis of errors and potential improvements.

The GraNd and EL2N scores are proposed in the context of dataset pruning, where the goal is to remove less informative samples from the training data. Thus, one can create a smaller, more efficient dataset that maintains the model's overall performance while reducing training time and computational resources.

While GraNd at initialization does not require model training, it requires a model and is not cheap to compute. In contrast, the input norm of training samples is incredibly cheap to compute and would thus provide an exciting new baseline to use for data pruning experiments. We investigate this correlation in this chapter and find positive evidence for it. However, we also find that the GraNd score at initialization does not outperform random pruning, unlike the respective results of \citet{paul2021deep} for GraNd at initialization.

\begin{figure}
    \centering
    \begin{subfigure}[b]{0.5\textwidth}
        \centering
        \includegraphics[width=\textwidth]{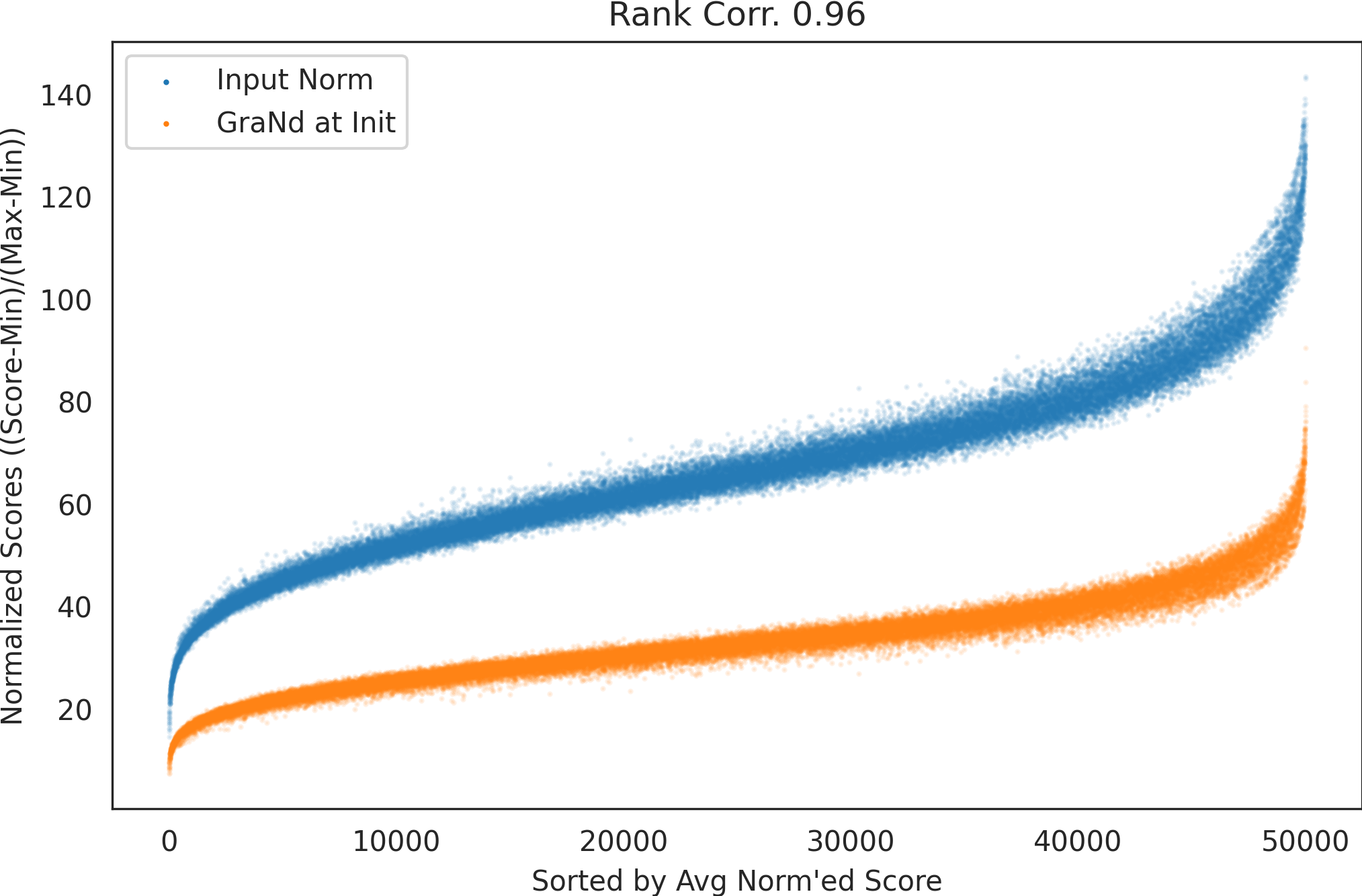}
        \caption{Original Repo (10 trials)}
        \label{dldd:fig:grand_init_vs_input_norm:original}
    \end{subfigure}\hfill
    \begin{subfigure}[b]{0.5\textwidth}
        \centering
        \includegraphics[width=\textwidth]{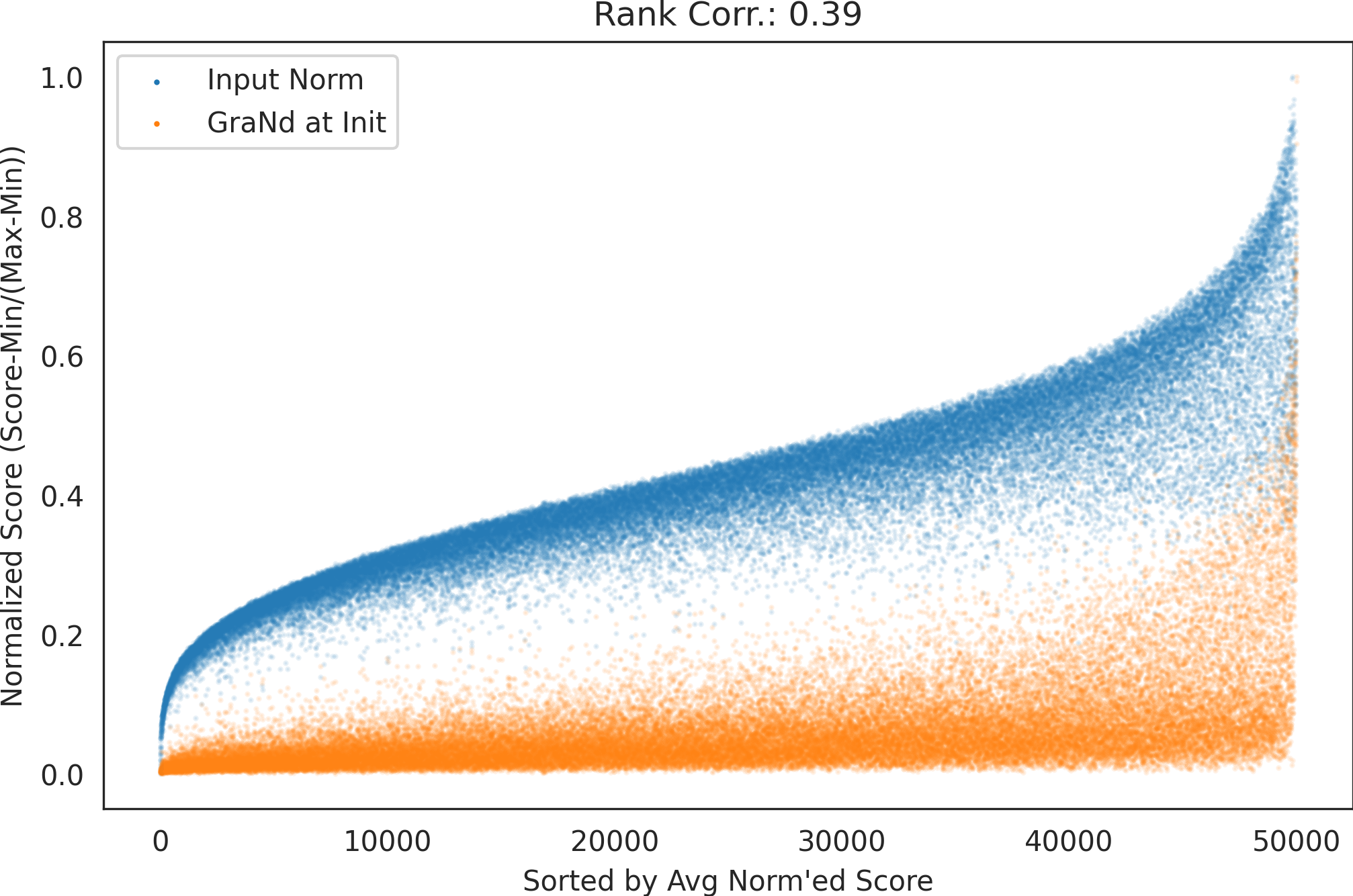}
        \caption{Hlb (10 trials)}
        \label{dldd:fig:grand_init_vs_input_norm:hlb}
    \end{subfigure}
    \begin{subfigure}[b]{0.5\textwidth}
        \centering
        \includegraphics[width=\textwidth]{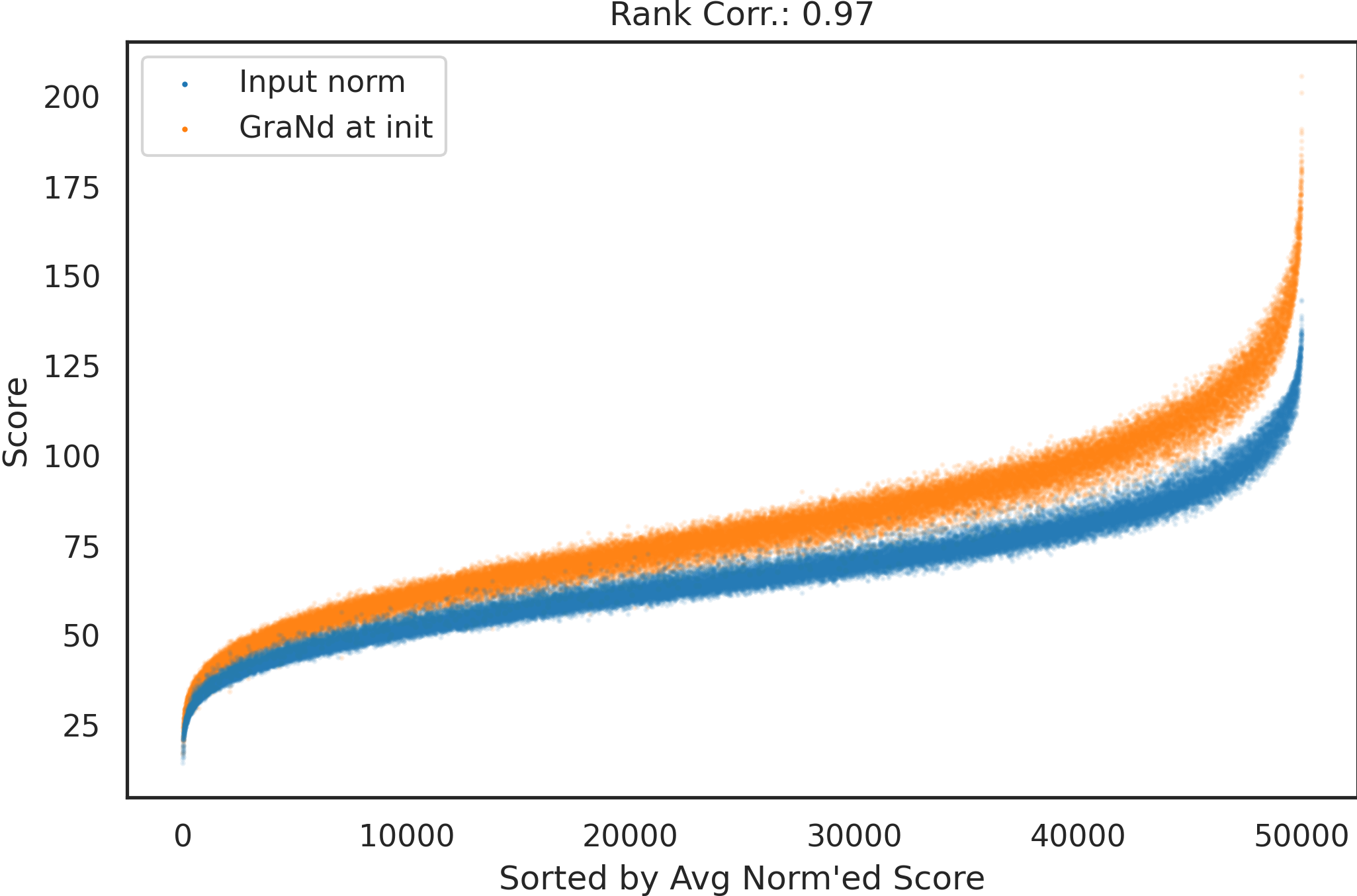}
        \caption{Minimal (120 trials)}
        \label{dldd:fig:grand_init_vs_input_norm:joost}
    \end{subfigure}\hfill
    \begin{subfigure}[b]{0.5\textwidth}
      \centering
      \includegraphics[width=\textwidth]{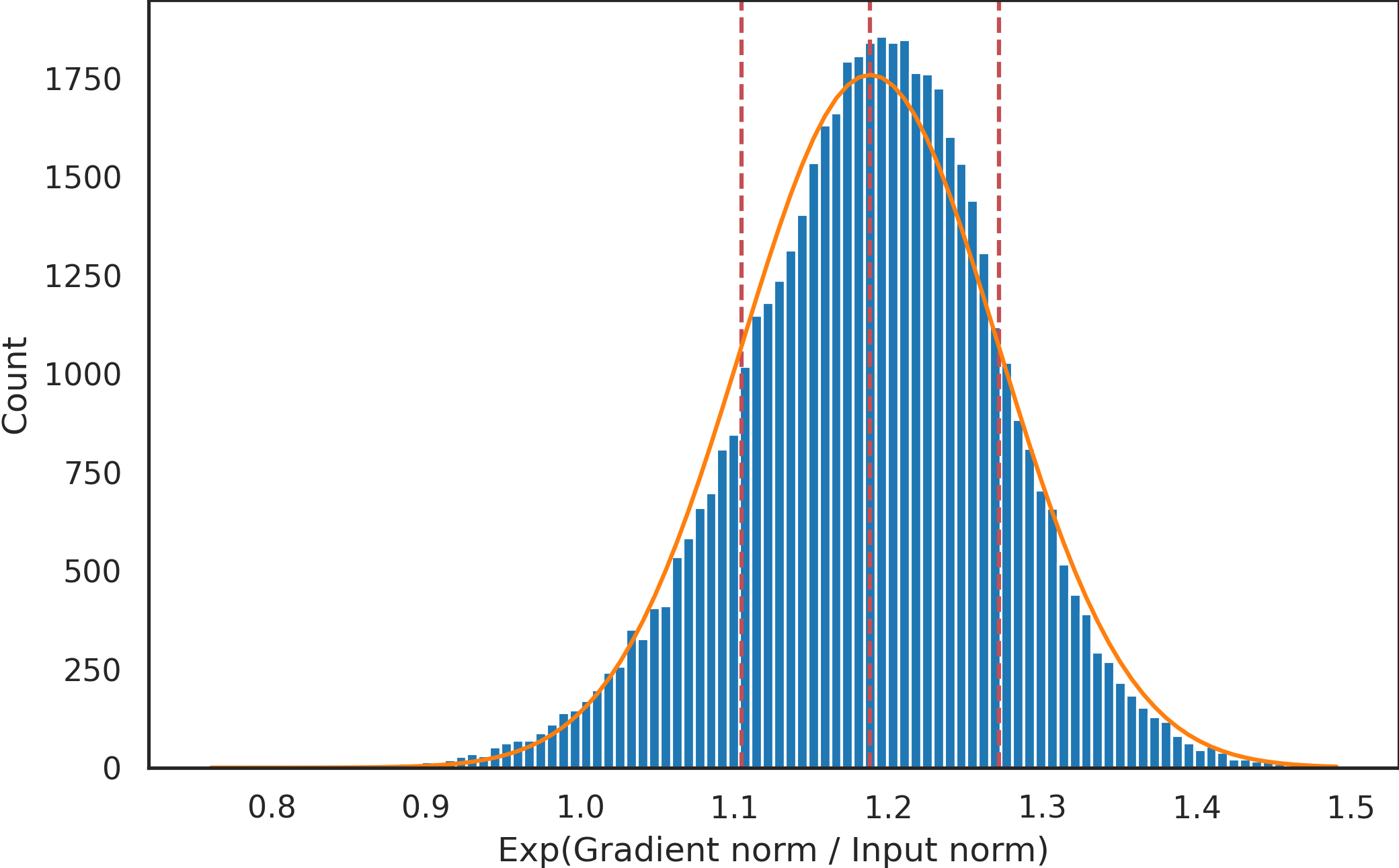}
      \caption{Minimal (120 trials)}
      \label{dldd:fig:grand_init_div_input_norm:joost}
  \end{subfigure}
    \caption{\emph{Correlation between GraNd at Initialization and Input Norm for CIFAR-10's training set.}
    \textbf{\subref{dldd:fig:grand_init_vs_input_norm:original}, \subref{dldd:fig:grand_init_vs_input_norm:hlb}, \subref{dldd:fig:grand_init_vs_input_norm:joost}}: We sort the samples by their average normalized score (i.e., the score minus its minimum divided by its range), plot the scores and compute Spearman's rank correlation on CIFAR-10's training data. The original repository and the `minimal' implementation have very high rank correlation---`hlb' has a lower but still strong rank correlation.
    \textbf{\subref{dldd:fig:grand_init_div_input_norm:joost}}: \emph{Ratio between input norm and gradient norm.} In the `minimal' implementation, the ratio between input norm and gradient norm is roughly log-normal distributed.}
    \label{dldd:fig:grand_init_vs_input_norm}
\end{figure}

\textbf{Outline.} %
In \S\ref{dldd:sec:correlation}, we begin by discussing the correlation between input norm and gradient norm at initialization.
We empirically find strong correlation between GraNd scores at initialization and input norms as we average over models.
In \S\ref{dldd:subsec:repro_fig1}, we explore the implication of this insight for dataset pruning and find that both GraNd at initialization and input norm scores do not outperform random pruning, but GraNd scores after a few epochs perform similar to EL2N scores at these later epochs.

In summary, this reproduction contributes a new insight on the relationship between input norm and gradient norm at initialization and finds a failure to reproduce one of the six contributions of \citet{paul2021deep}.

\section{Investigation}
\label{dldd:sec:reproduction}

\begin{figure}
  \centering
  \begin{subfigure}[b]{0.5\textwidth}
      \centering
      \includegraphics[width=\textwidth]{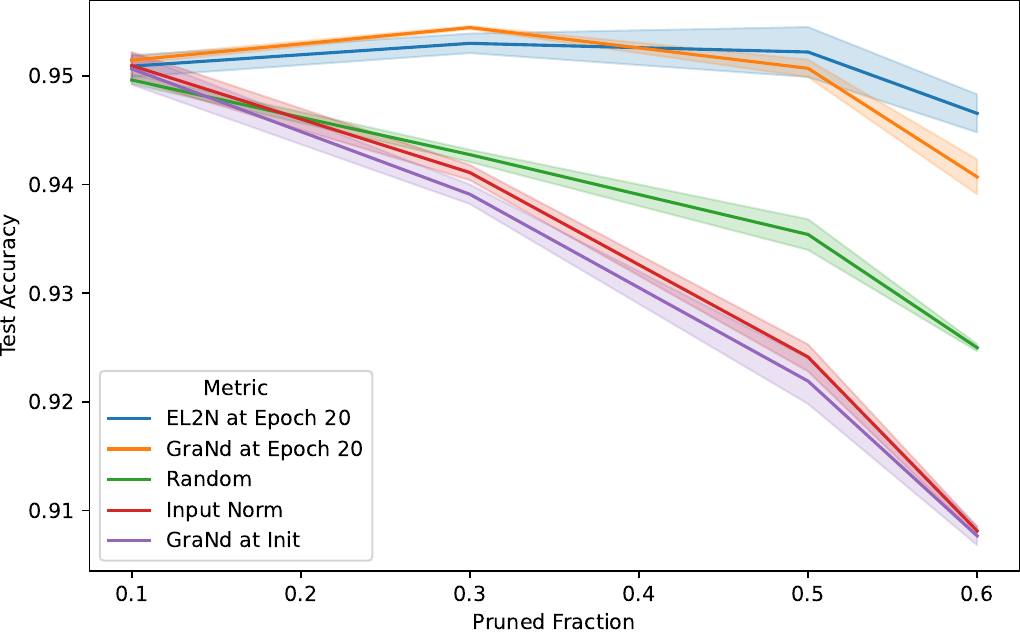}
      \caption{Original Repo (2 trials each)}
      \label{dldd:fig:fig1_repro:original}
  \end{subfigure}\hfill
  \begin{subfigure}[b]{0.5\textwidth}
    \centering
    \includegraphics[width=\textwidth]{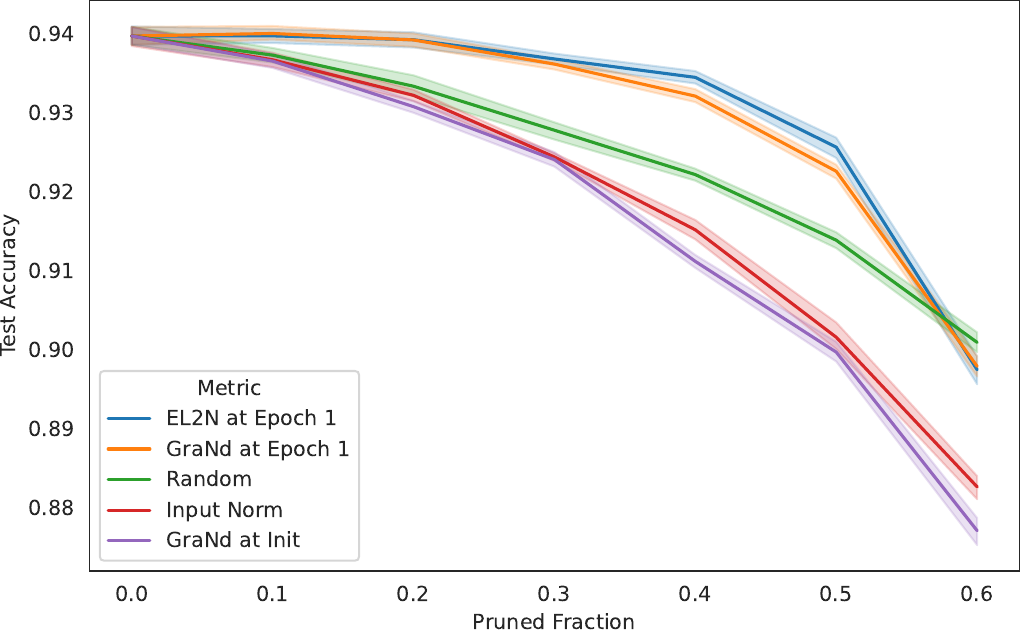}
    \caption{Hlb (10 trials)}
    \label{dldd:fig:fig1_repro:hlb}
  \end{subfigure}
  \caption{\emph{Reproduction of Figure 1 (second row) from \citet{paul2021deep}.}
  In both reproductions, GraNd at initialization performs as well as the input norm. However, it does not perform better than random pruning. Importantly, it also fails to reproduce the results from \citet{paul2021deep}. However, GraNd at epoch 20 (respectively at epoch 1 for `hlb') performs similar to EL2N and like GraNd at initialization in \citet{paul2021deep}.}
  \label{dldd:fig:fig1_repro}
\end{figure}

We investigate the correlation between input norm and GraNd at initialization and the other scores on CIFAR-10 \citep{krizhevsky2009learning} in three different ways:
First, we update the original paper repository\footnote{
  \shorturl{https://github.com/blackhc/data_diet}
} (\shorturl{https://github.com/mansheej/data_diet}), which uses JAX \citep{jax2018github}, rerun the experiments for Figure 1 (second row) in \citet{paul2021deep} for CIFAR-10, which trains for 200 epochs, using GraNd at initialization, GraNd at epoch 20, E2LN at epoch 20, Forget Score at epoch 200, and input norm. Second, we reproduce the same experiments using `hlb' \citep{Balsam_hlb-CIFAR10_2023}, which is a strongly modified version of ResNet-18 that allows to train to high accuracy in 12 epochs taking about 30 seconds total on an Nvidia RTX 4090 in PyTorch \citep{pytorchgithub} For the latter, we compare GraNd at initialization, GraNd at epoch 1 ($\approx 20/200\cdot 12$ epochs), EL2N at epoch 1, and input norm\footnote{\label{dldd:code:pytorch}\shorturl{https://github.com/blackhc/pytorch_datadiet}}.
Third, we compare the rank correlations between the different scores for those two repositories and also use another `minimal' CIFAR-10 implementation \citep{Joost_minimal_cifar10} with a standard ResNet18 architecture for CIFAR-10 to compare the rank correlations.

\subsection{Correlation between GraNd at Initialization and Input Norm}
\label{dldd:sec:correlation}

To better understand the relationship between the input norm and the gradient norm at initialization, let us consider a toy example first and then appeal to empirical evidence as is common in deep learning research:
let's examine linear softmax classification with $C$ classes (without a bias term). The model takes the form:
\begin{equation}
  f(x) = \operatorname{softmax}(W x),
\end{equation}
together with the cross-entropy loss function:
\begin{equation}
L = -\log f(x)_y.
\end{equation}
The gradient of the loss function with respect to the rows $w_j$ of the weight matrix $W$ is:
\begin{equation}
\nabla_{w_j} L = (f(x)_j - \mathbb{1}\{j = y\}) x
\end{equation}
where $\mathbb{1}\{j = \y\}$ is the indicator function that is 1 if $j = \y$ and 0 otherwise.
The squared norm of the gradient is:
\begin{equation}
\| \nabla_w L \|_2^2 = \sum_{j = 1}^C (f(x)_j - \mathbb{1}\{j = y\})^2 \|x\|_2^2.
\end{equation}
In expectation over $W$ (different initializations), the norm of the gradient is:
\begin{equation}
\mathbb{E}_W \left [\| \nabla_w L \|_2 \right ] = \mathbb{E}_W \left [\left (\sum_{j = 1}^C (f(x)_j - \mathbb{1}\{j = y\})^2 \right )^{1/2} \right ] \|x\|_2.
\end{equation}
Thus, we see that the gradient norm is a multiple of the input norm. The factor depends on $f(\x)_j$, which we could typically expect to be $1/C$.

\textbf{Empirical Evidence.} %
In \Cref{dldd:fig:grand_init_vs_input_norm}, we see that on CIFAR-10's training set, GraNd at initialization and the input norm are highly correlated. This is true for the original repository, the `hlb' and the `minimal' implementation. The `hlb' implementation has a lower but still strong correlation.

\subsection{Reproducing Figure 1 of \citet{paul2021deep} on CIFAR-10}
\label{dldd:subsec:repro_fig1}

\begin{figure}
  \centering
  \begin{subfigure}[b]{0.5\textwidth}
      \centering
      \includegraphics[width=\textwidth]{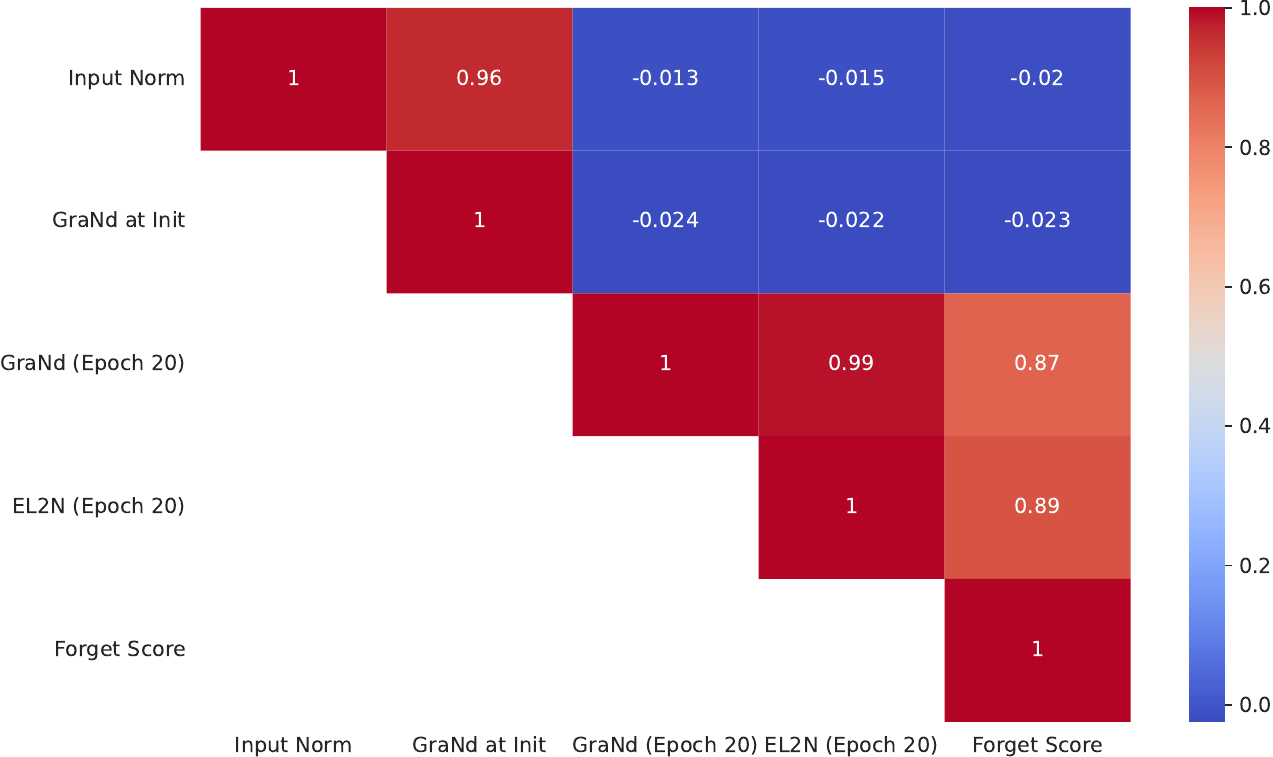}
      \caption{Original Repo (2 trials each)}
      \label{dldd:fig:heatmap:original}
  \end{subfigure}\hfill
  \begin{subfigure}[b]{0.5\textwidth}
    \centering
    \includegraphics[width=\textwidth]{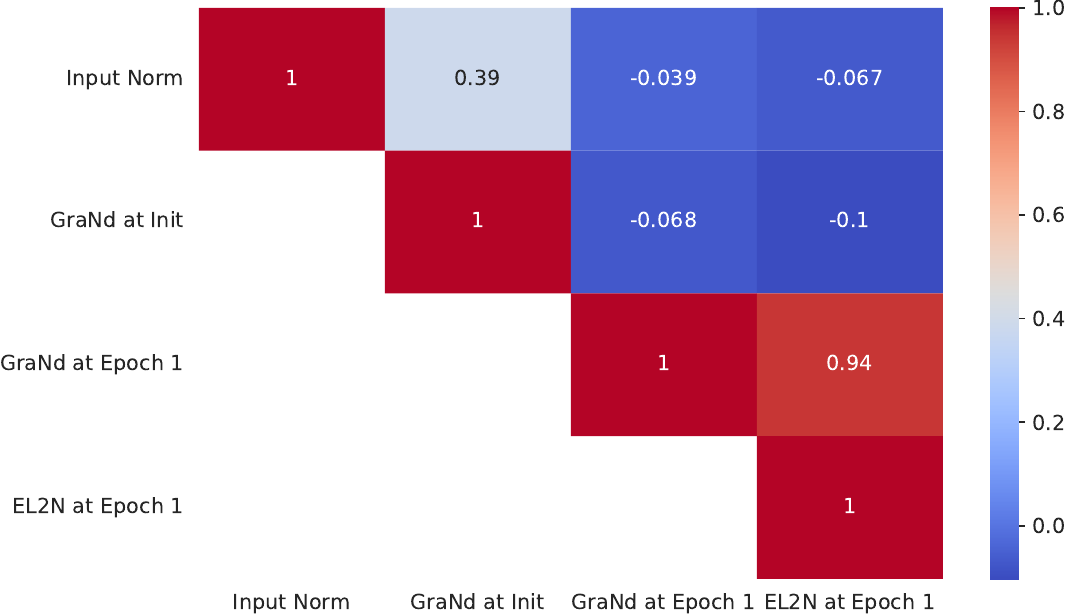}
    \caption{Hlb (10 trials)}
    \label{dldd:fig:heatmap:hlb}
  \end{subfigure}
  \caption{\emph{Rank Correlations of the Scores.} Cf. Figure 12 in the appendix of \citet{paul2021deep}.
  In both reproductions, GraNd at initialization and input norm are positively correlated, while GraNd and EL2N at later epochs are strongly correlated with each other and the Forget Score (at epoch 200).}
  \label{dldd:fig:heatmap}
\end{figure}

In \Cref{dldd:fig:fig1_repro}, we see that GraNd at initialization performs about as well as using the input norm. However, it does not reproduce the results from \citet{paul2021deep}.
It performs worse than random pruning (for `hlb'). However, GraNd at epoch 20 (respectively at epoch 1 for `hlb') performs like GraNd at initialization in \citet{paul2021deep}.
Similarly, in \Cref{dldd:fig:heatmap}, we see that GraNd at initialization and the input norm are strongly correlated as are GraNd at later epochs, EL2N and the Forget Score, with little correlation between these two groups.

\section{Discussion}

If GraNd at initialization performed as well as claimed in \citet{paul2021deep}, using the input norm would provide a new exciting baseline for data pruning because it is model independent and cheaper to compute than GraNd or other scores.
However, since only GraNd at later epochs seems to perform as expected, we cannot recommend using input norm or GraNd at initialization for data pruning.

As to the failure to reproduce the results of \citet{paul2021deep}, we could not rerun the code using the original JAX version because it is too old for our GPU. The authors of \citet{paul2021deep} were, however, able to set up a Google Cloud VM with an old image that was able to reproduce the original results using the original JAX version. On further investigation, the author of this reproduction found a bug in \texttt{flax.training.restore\_checkpoint} that was fixed in April 2021\footnote{See \shorturl{https://github.com/google/flax/commit/28fbd95500f4bf2f9924d2560062fa50e919b1a5}.}: passing a \texttt{0} step (i.e. initialization) would trigger loading the \emph{latest} checkpoint instead of the zero-th checkpoint because the internal implementation was checking 
\texttt{if step:} instead of \texttt{if step is not None:} when deciding whether to fall back to loading the latest checkpoint. This bug was fixed in April 2021, but the authors of \citet{paul2021deep} were not aware of this bug and did not rerun their experiments with newer JAX/FLAX versions. We have accordingly informed the authors of \citet{paul2021deep}.

		\section{Details}
		\begin{figure}[h]
  \centering
  \begin{subfigure}[b]{0.5\textwidth}
      \centering
      \includegraphics[width=\textwidth]{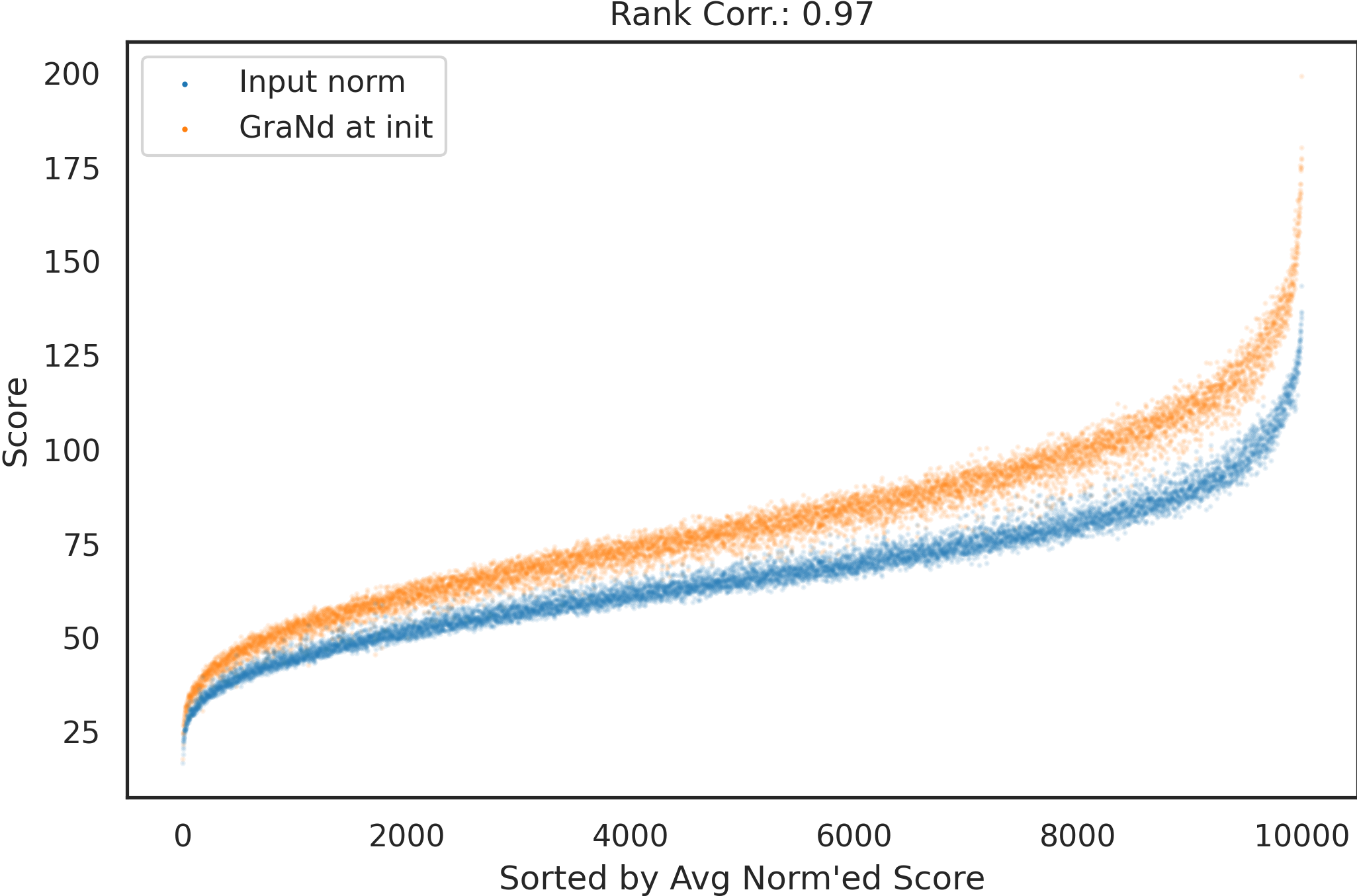}
      \caption{Minimal (1000 trials)}
      \label{fig:test_grand_init_vs_input_norm:joost}
  \end{subfigure}\hfill
  \begin{subfigure}[b]{0.5\textwidth}
    \centering
    \includegraphics[width=\textwidth]{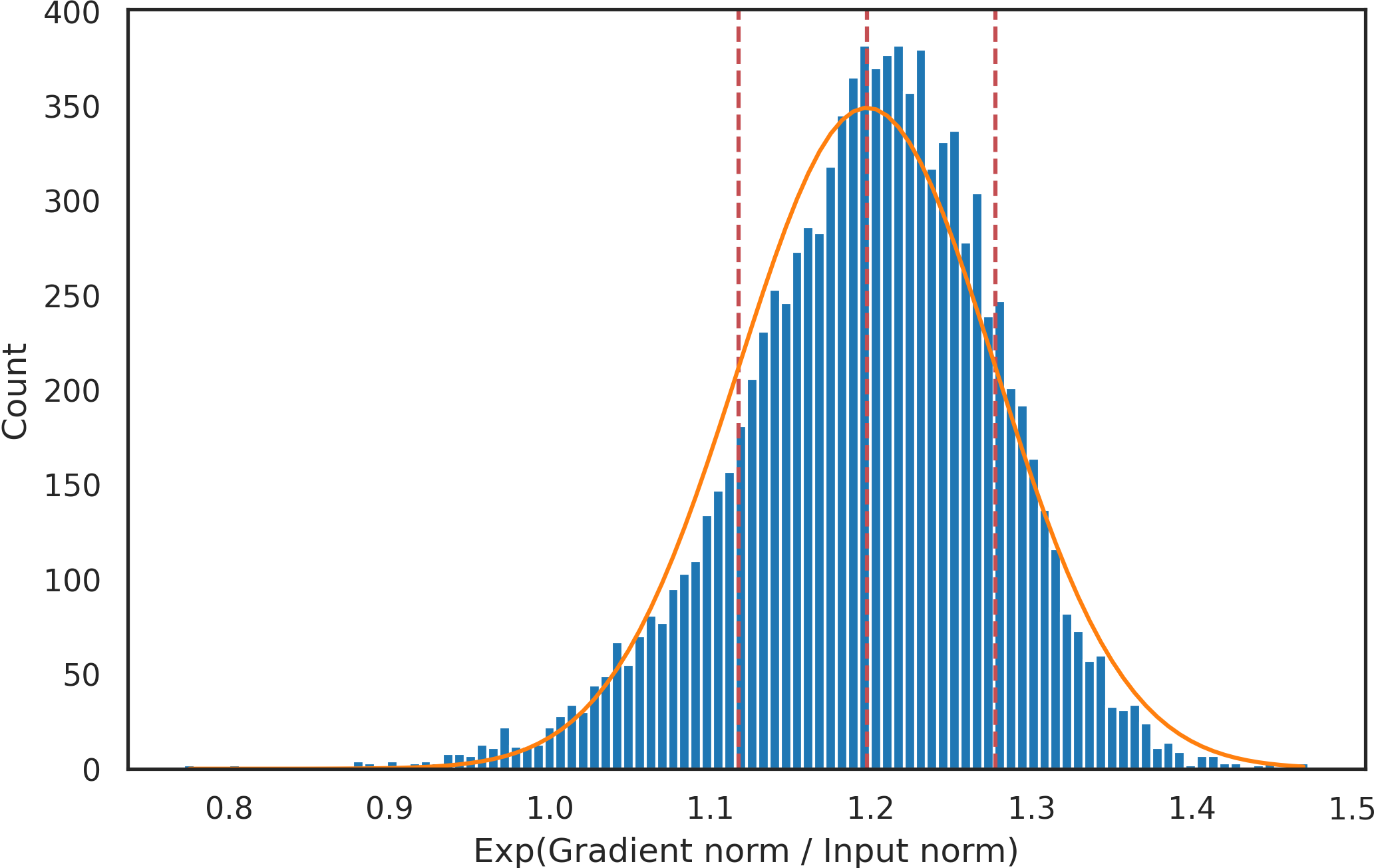}
    \caption{Minimal (1000 trials)}
    \label{fig:test_grand_init_div_input_norm:joost}
\end{subfigure}
  \caption{\emph{Correlation between GraNd at Initialization and Input Norm on the Test Set.}
  \textbf{(\subref{fig:test_grand_init_vs_input_norm:joost})}: We sort the samples by their average normalized score (i.e., the score minus its minimum divided by its range), plot the scores and compute Spearman's rank correlation on CIFAR-10's test data. The original repository and the `minimal' implementation have very high rank correlation---`hlb' has a lower but still strong rank correlation.
  \textbf{(\subref{fig:test_grand_init_div_input_norm:joost})}: \emph{Ratio between input norm and gradient norm.} In the `minimal' implementation, the ratio between input norm and gradient norm is roughly log-normal distributed}
  \label{fig:test_grand_init_vs_input_norm}
\end{figure}

	}

	{
		\graphicspath{{./jiang/}}

\newcommand{\classindex}{k}

\newcommand{\conf}{\mathrm{Conf_{Top1}}}
\newcommand{\acc}{\mathrm{Acc}}
\newcommand{\predacc}{\mathrm{PredAcc}}
\newcommand{\acctop}{\mathrm{Acc_{Top1}}}
\newcommand{\CE}{\mathrm{CE}}
\newcommand{\testerror}{\mathrm{TestError}}
\newcommand{\disrate}{\mathrm{Dis}}
\newcommand{\ECE}{ECE}
\newcommand{\CACE}{\mathrm{CACE}}
\newcommand{\ECACE}{\mathrm{ECACE}}
\newcommand{\CWCE}{\mathrm{CWCE}}

\newcommand{\Yhat}{\hat{Y}}
\newcommand{\yhat}{\hat{y}}

\newcommand{\Ypred}{\Yhat}
\newcommand{\ypred}{\yhat}
\newcommand{\Ytrue}{\Y}
\newcommand{\ytrue}{\y}
\newcommand{\levelthreshold}{q}

\newcommand{\w}{\omega}
\newcommand{\W}{\Omega}

\newcommand{\verteq}{\rotatebox{90}{$\,=$}}
\newcommand{\equalto}[2]{\overset{\displaystyle\underset{\mkern4mu\verteq}{#2}}{#1}}

\newcommand{\innerexpr}[1]{{\color{solarized@base03}#1}}

		\chapter{\jiangtitle}
		\label{chapter:jiang}

Machine learning models can cause harm when their predictions become unreliable, yet we trust them blindly.
This is not fiction but has happened in real-world applications of machine learning \citep{schneider2021algorithmic}.
Thus, there has been significant research interest in model robustness, uncertainty quantification, and bias mitigation.
In particular, finding ways to bound the test error of a trained deep neural network without access to the labels would be of great importance: one could estimate the performance of models in the wild where unlabeled data is ubiquitous, labeling is expensive, and the data often does not match the training distribution. Crucially, it would provide a signal on when to trust the output of a model and when to defer to human experts instead.

Several recent works \citep{chen2021detecting,granese2021doctor,garg2022leveraging,jiang2021assessing} look at the question of how model predictions in a non-Bayesian setting can be used to estimate model accuracy.
In this chapter, we focus on one work\footnote{An ICLR 2022 Spotlight, which has spawned additional follow-up works, e.g. \citet{baek2022agreementontheline}.} \citep{jiang2021assessing} specifically, and examine the theoretical and empirical results from a Bayesian perspective.

As we have already examined in \Cref{glue:background:uncertainty_quantification,chapter:ddu}, \emph{epistemic uncertainty} \citep{der2009aleatory} captures the uncertainty of a model about the reliability of its predictions, that is epistemic uncertainty quantifies the uncertainty of a model about its predictive distribution, while \emph{aleatoric uncertainty}
quantifies the ambiguity within the predictive distribution and the label noise, c.f.\ \citet{kendall2017uncertainties}. Epistemic uncertainty thus tells us whether we can trust a model's predictions or not. Assuming a \emph{well-specified} and \emph{well-calibrated} Bayesian model, when its predictive distribution has low epistemic uncertainty for an input, it can be trusted. But likewise, a Bayesian model's calibration ought to deteriorate as epistemic uncertainty increases for a sample: in that case, the predictions become less reliable and so does the model's calibration.

In this context, calibration is an \emph{aleatoric} metric for a model's reliability \citep{gopal2021calibration}. Calibration captures how well a model's confidence for a given prediction matches the actual frequency of that prediction in the limit of observations \emph{in distribution}: when a model is $70\%$ confident about assigning label $A$, does $A$ indeed occur with $70\%$ probability in these instances?

\citet{jiang2021assessing}, while not Bayesian, %
make the very interesting empirical and theoretical discovery that deep ensembles satisfy a `\emph{Generalization Disagreement Equality}' when they are well-calibrated according to a proposed `\emph{class-aggregated calibration}' (or a `\emph{class-wise calibration}') and empirically find that the respective calibration error generally bounds the absolute difference between the test error and `\emph{disagreement rate}.'
\citet{jiang2021assessing}'s theory builds upon \citet{nakkiran2020distributional}'s `\emph{Agreement Property}' and provides backing for an empirical connection between the \emph{test error} and \emph{disagreement rate} of two separately trained networks on the same training data. 
Yet, while \citet{nakkiran2020distributional} limit the applicability of their Agreement Property to in-distribution data, \citet{jiang2021assessing} carefully extend it: 
`\emph{our theory is general and makes no restrictions on the hypothesis class, the algorithm, the source of stochasticity, or the test distributions (which may be different from the training distribution)}' with qualified evidence: `\emph{we present preliminary observations showing that GDE is approximately satisfied even for certain distribution shifts within the PACS \citep{li2017deeper} dataset.}'

In this chapter, we present a new perspective on the theoretical results using a standard probabilistic approach for discriminative (Bayesian) models, whereas \citet{jiang2021assessing} use a hypothesis space of models that output one-hot predictions. Indeed, their theory does not require one-hot predictions, separately trained models (deep ensembles) or Bayesian models. Moreover, as remarked by the authors, their theoretical results also apply to a single model that outputs softmax probabilities. We will see that our perspective greatly simplifies the results and proofs. 

This also means that the employed notion of disagreement rate \emph{does not capture epistemic uncertainty but overall uncertainty}, similar to the predictive entropy, which is a major difference to Bayesian approaches which can evaluate epistemic uncertainty separately \citep{smith2018understanding}. %
Overall uncertainty is the sum of aleatoric and epistemic uncertainty.
For in-distribution data, epistemic uncertainty will generally be low: overall uncertainty will mainly capture aleatoric uncertainty and align well with (aleatoric) calibration measures. However, under distribution shift, epistemic uncertainty can be a confounding factor. %

Importantly, we find that the connection between the proposed calibration metrics and the gap between test error and disagreement rate exists because the introduced notion of class-aggregated calibration is so strong that this connection follows almost at once.

Moreover, the suggested approach is circular\footnote{This was added as a caveat to the camera-ready version of \citet{jiang2021assessing} after reviewing a preprint of the preprint of the paper this chapter is based on.}: calibration must be measured on the data distribution we want to evaluate. Otherwise, we cannot bound the difference between the test error and the disagreement rate and obtain a signal on how trustworthy our model is. This reintroduces the need for labels on the unlabeled dataset, limiting practicality. 
Alternatively, one would have to assume that these calibration metrics do not change for different datasets or under distribution shifts, which we show not to hold: deep ensembles are less calibrated the more the ensemble members disagree (even on in-distribution data).

Lastly, we draw connections and show that the `class-aggregated calibration error' and the `class-wise calibration error'\footnote{Which is not explicitly introduced in \citet{jiang2021assessing} but can be analogously constructed.} are equivalent to the `adaptive calibration error' and `static calibration error' introduced in \citet{nixon2019measuring} and its implementation.

\subsection*{Outline.}

We introduce the necessary background and notation in \S\ref{jiang:sec:background}. 
In \S\ref{jiang:sec:rephrasing_jiang} we rephrase the theoretical statements from \citet{jiang2021assessing} using a parameter distribution (instead of a version space) and auxiliary random variables. This allows us to simplify the theoretical statements and proofs greatly in \S\ref{jiang:sec:rederivation} and to examine the connection to \citet{nixon2019measuring}. Finally, in \S\ref{jiang:sec:empirical_example}, we provide empirical evidence that deep ensembles are less calibrated exactly when their ensemble members disagree. %

\section{Background \&\ Setting}
\label{jiang:sec:background}

In this section, we introduce additional notation, the initial Bayesian formalism, the connection to deep ensembles, and the probabilistic model. We restate the statements from \citet{jiang2021assessing} using this formalism in \S\ref{jiang:sec:rephrasing_jiang}.

\paragraph{Notation.} We use an implicit notation for expectations $\implicitE{f(X)}$ when possible. For additional clarity, we also use $\E{X}{f(X)}$ and $\simpleE{\pof{x}}{f(x)}$, which fix the random variables and distribution, respectively, when needed. 

We will use nested probabilistic expressions of the form $\implicitE{\pof{\Ypred = \Ytrue \given X}}$.
Prima facie, this seems unambiguous, but is $\pof{\Ypred = \Ytrue \given X}$ a transformed random variable of only $X$ or also of $\Ytrue$ (and $\Ypred$): what are we taking the expectation over?
This is not always unambiguous, so we disambiguate between the probability for an event defined by an expression $\pfor{\ldots} = \E{}{\indicator{\ldots}}$, where $\indicator{\ldots}$ is the indicator function\footnote{The indicator function is $1$ when the predicate `$\ldots$' is true and $0$ otherwise.},
and a probability given specific outcomes for various random variables $\pof{\ypred \given x}$, c.f.:
\begin{align}
    \pfor{\Ypred = \Ytrue \given X} &= \E{\Ypred, \Ytrue}{\mathbb{1}{\{\Ypred = \Ytrue\}} \given X}
    = \simpleE{\pof{\ypred, \ytrue \given X}}{\mathbb{1}{\{\ypred = \ytrue\}}},
\end{align}
which is a transformed random variable of $X$, while
$\pof{\Ypred = \Ytrue \given X}$ is simply a (transformed) random variable, applying the probability density on the random variables $\Ytrue$ and $X$. Put differently, $\Ytrue$ is bound within the former but not the latter: $\pfor{\ldots \given X}$ is a transformed random variable of $X$, and any random variable that appears within the $\ldots$ is bound within that expression. 

\paragraph{Probabilistic Model.} %
We assume classification with $\numclasses$ classes. For inputs \(X\) with ground-truth labels \(\Ytrue\), we have a Bayesian model with parameters \(\W\) that makes predictions \(\Ypred\): 
\begin{equation}
    \pof{\ytrue, \ypred, \w \given x} = \pof{\ytrue \given x} \, \pof{\ypred \given x, \w} \, \pof{\w}.
\end{equation}
We focus on model evaluation. (Input) samples $x$ can come either from `in-distribution data' which follows the training set or from samples under covariate shift (distribution shift).
The expected prediction over the model parameters is the \emph{marginal predictive distribution}:
\begin{equation}
    \pof{\ypred \given x} 
    = \E{\W}{\pof{\ypred \given x, \W}}.
\end{equation}

\paragraph{On \boldmath{$\pof{\w}$}.} The main emphasis in Bayesian modelling can be Bayesian inference or Bayesian model averaging \citep{wilson2020bayesian}. Here we concentrate on the model averaging perspective, and for simplicity take the model averaging to be with respect to \emph{some} distribution $\pof{\w}$. 
Hence, we will use $\pof{\w}$ as the push-forward of models initialized with different initial seeds through SGD to minimize the negative log likelihood with weight decay and a specific learning rate schedule (MLE or MAP)---the same we also did in \Cref{chapter:ddu}:

\begin{assumption}
    We assume that \(\pof{\w}\) is a distribution of possible models we obtain by training with a specific training regime on the training data with different seeds. A single \(\w\)  identifies a single trained model.
\end{assumption}

We cast deep ensembles \citep{hansen1990neural,lakshminarayanan2016simple}, which refer to training multiple models and averaging predictions, into the introduced Bayesian perspective above by viewing them as an empirical finite sample estimate of the parameter distribution $\pof{\w}$. Then, $\w_1, \ldots, \w_N \sim \pof{\w}$ drawn i.i.d.\ are the \emph{ensemble members}. 

Again, the implicit model parameter distribution $\pof{w}$ is given by the models that are obtained through training. Hence, we can view the predictions of a deep ensemble or the ensemble's prediction disagreement for specific $x$ (or over the data) as empirical estimates of the predictions or the model disagreement using the implicit model distribution, respectively. %

\paragraph{Calibration.}
A model's calibration for a given $x$ measures how well the model's \emph{top-1 (argmax) confidence}
\begin{align}
    \conf \defeq \pof{\Ypred = \argmax_\classindex \innerexpr{\pof{\Ypred = \classindex \given X}} \given X}
\end{align}
matches its \emph{top-1 accuracy}
\begin{align}
    \acctop \defeq \pof{\Ytrue = \argmax_\classindex \innerexpr{\pof{\Ypred = \classindex \given X}} \given X},
\end{align}
where we define both as transformed random variables of $X$.
The calibration error is usually defined as the absolute difference between the two:
\begin{align}
    \CE \defeq \lvert \acctop - \conf \rvert.
\end{align}
In general, we are interested in the \emph{expected calibration error (ECE)} over the data distribution \citep{guo2017calibration} where we bin samples by their top-1 confidence.
Intuitively, the ECE will be low when we can trust the model's top-1 confidence on the given data distribution.

We usually use top-1 predictions in machine learning. However, if we were to draw $\Ypred$ according to $\pof{\ypred \given x}$ instead, the (expected) accuracy would be:
\begin{align}
    \acc &\defeq
    \pfor{\Ytrue = \Ypred \given X} \\
    &= \sum_\classindex \pof{\Ytrue = \classindex \given X} \, \pof{\Ypred = \classindex \given X} \\
    &= \E{\Ytrue}{\pof{\Ypred = \Ytrue \given X} \given X}, 
\end{align}
as a random variable of $X$.
Usually we are interested in the accuracy over the whole dataset:
\begin{align}
    \pfor{\Ypred = Y} &= \E{}{\acc} = \E{X}{\pfor{\Ypred = Y \given X}} = \E{X, \Ytrue}{\pof{\Ypred = \Ytrue \given X}}.
\end{align}
For example, for binary classification with two classes A and B, if class A appears with probability 0.7 and a model predicts class A with probability 0.2 (and thus class B appears with probability 0.3, which a model predicts as 0.8), its accuracy is $0.7 \times 0.2 + 0.3 \times 0.8 = 0.38$, while the top-1 accuracy is $0.3$. Likewise, the predicted accuracy is $0.2^2 + 0.8^2=0.68$ while the top-1 predicted accuracy is $0.8$.

\section{Rephrasing \citet{jiang2021assessing} in a Probabilistic Context}
\label{jiang:sec:rephrasing_jiang}

We present the same theoretical results as \citet{jiang2021assessing} but use a Bayesian formulation instead of a hypothesis space and define the relevant quantities as (transformed) random variables. 
As such, our definitions and theorems are equivalent and follow the paper but look different. We show these equivalences in \S\ref{jiang:app:equivalent_definitions} in the appendix and prove the theorems and statements themselves in the next section. 

First, however, we note a distinctive property of \citet{jiang2021assessing}. It is assumed that each $\pof{\ypred \given x, \w}$ is always one-hot for any $\w$. In practice, this could be achieved by turning a neural network's softmax probabilities into a one-hot prediction for the $\argmax$ class. We call this the \emph{Top1-Output-Property} (TOP).
\begin{assumption}
    The Bayesian model $\pof{\ypred, \w \given x}$ satisfies TOP: $\pof{\ypred \given x, \w}$ is one-hot for all $x$ and $\w$.
\end{assumption}

\begin{definition}
The \emph{test error} and \emph{disagreement rate}, as transformed random variables of $\W$ (and $\W'$), are:
\begin{align}
    \testerror &\defeq \pfor{\Ypred \not= \Ytrue \given \W}\\
    \bigl( &= 1 - \pfor{\Ypred = \Ytrue \given \W} \\
    &= 1 - \E{X,\Ytrue} {\pof{\Ypred = \Ytrue \given X, \W}}, \\
    &= 1 - \simpleE{\pof{x,\ytrue}} {\pof{\Ypred = \ytrue \given x, \W}}) \bigr), \\
    \disrate &\defeq \pfor{\Ypred \not= \Ypred' \given \W, \W'} \\
    \bigl( &= 1 - \pfor{\Ypred = \Ypred' \given \W, \W'} \\
    &= 1 - \E{X, \Ypred} {\pof{\Ypred' = \Ypred \given X, \W'} \given \W, \W'} \\
    &= 1 - \simpleE{\pof{x, \ypred \given \W}} {\pof{\Ypred' = \ypred \given x, \W'}}\bigr),
\end{align}
where for the disagreement rate, we expand our probabilistic model to take a second model $\W'$ with prediction $\Ypred'$ into account (and which uses the same parameter distribution), so:
\begin{align}
    \pof{\ytrue, \ypred, \w, \ypred', \w' \given x} \defeq &
        \pof{\ytrue \given x} \, 
        \pof{\ypred \given x, \w} \, \pof{\w} \pof{\Ypred = \ypred' \given x, \W = \w'} \, \pof{\W=\w'}. \notag
\end{align}
\end{definition}

\citet{jiang2021assessing} then introduce the property of interest: %
\begin{definition}
    A Bayesian model $\pof{\ypred, \w \given x}$ fulfills the \emph{Generalization Disagreement Equality (GDE)} when:
    \begin{align}
        &\E{\W}{\testerror(\W)} = \E{\W, \W'}{\disrate(\W, \W')} \quad (
        \Leftrightarrow \implicitE{\testerror} = \implicitE{\disrate}).
    \end{align} 
\end{definition}
When this property holds, we seemingly do not require knowledge of the labels to estimate the (expected) test error: computing the (expected) disagreement rate is sufficient.

Two different types of calibration are then introduced, \emph{class-wise} and \emph{class-aggregated} calibration, and it is shown that they imply the GDE:
\begin{definition}
    The Bayesian model $\pof{\ypred, \w \given x}$ satisfies \emph{class-wise calibration} when for any $\levelthreshold \in [0, 1]$ and any class $\classindex \in [\numclasses]$:
    \begin{align}
        \pof{\Ytrue = \classindex \given \innerexpr{\pof{\Ypred = \classindex \given X}} = \levelthreshold} = \levelthreshold.
    \end{align}
    Similarly, the Bayesian model $\pof{\ypred, \w \given x}$ satisfies \emph{class-aggregated calibration} when for any $\levelthreshold \in [0, 1]$:
    \begin{align}
        \sum_\classindex \pof{\Ytrue = \classindex , \innerexpr{\pof{\Ypred=\classindex \given X}} = \levelthreshold} = \levelthreshold \sum_\classindex \pof{\innerexpr{\pof{\Ypred=\classindex \given X}} = \levelthreshold}
        . \label{jiang:eq:cac}
    \end{align}
\end{definition}
\begin{restatable}{theorem}{cacgdetheorem}
    \label{jiang:thm:cac_gde}
    When the Bayesian model $\pof{\ypred, \w \given x}$ satisfies class-wise or class-aggregated calibration, it also satisfies GDE.
\end{restatable}

Finally, \citet{jiang2021assessing} introduce the \emph{class-aggregated calibration error} similar to the ECE and then use it to bound the magnitude of any GDE gap:
\begin{definition}
The \emph{class-aggregated calibration error (CACE)} is the integral of the absolute difference of the two sides in \cref{jiang:eq:cac} over possible $\levelthreshold \in [0,1]$:
\begin{align}
    \CACE \defeq \int_{\levelthreshold \in [0, 1]} 
    \begin{aligned}[t]
        \bigl| &\sum_\classindex \pof{\Ytrue = \classindex , \innerexpr{\pof{\Ypred=\classindex \given X}} = \levelthreshold} - \levelthreshold \sum_\classindex \pof{\innerexpr{\pof{\Ypred=\classindex \given X}} = \levelthreshold} \bigr| d\levelthreshold.
    \end{aligned}
\end{align}
\end{definition}
\begin{restatable}{theorem}{caceboundtheorem}
    \label{jiang:thm:cace_bound}
    For any Bayesian model $\pof{\ypred, \w \given x}$, we have:
    \begin{align*}
        \lvert \implicitE{\testerror} - \implicitE{\disrate} \rvert \le \CACE.
    \end{align*}
\end{restatable}

In the following section, we simplify the definitions and prove the statements using elementary probability theory, showing that notational complexity is the main source of complexity.

\section{GDE is Class-Aggregated Calibration in Expectation}
\label{jiang:sec:rederivation}

We show that proof for \Cref{jiang:thm:cace_bound} is trivial if we use different but equivalent definitions of the class-wise and class-aggregate calibration. First though, we establish a better understanding for these definitions by examining the GDE property \(\implicitE{\testerror} = \implicitE{\disrate}\). For this, we expand the definitions of $\implicitE{\testerror}$ and $\implicitE{\disrate}$, and use random variables to our advantage.

We define a quantity which will be of intuitive use later on: the \emph{predicted accuracy}
\begin{align}
    \predacc &\defeq \E{\Ypred}{\pof{\Ypred \given X} \given X} = \sum_\classindex \pof{\Ypred = \classindex \given X} \pof{\Ypred = \classindex \given X},
\end{align}
as a random variable of $X$. It measures the expected accuracy assuming the model's predictions are correct, that is the true labels follow $\pof{\ypred \given x}$. This also assumes that we draw $\Yhat$ accordingly and do not always use the top-1 prediction.

\paragraph{Revisiting GDE.} %
On the one hand, we have:
\begin{align}
    \implicitE{\testerror} & = \E{\W}{\pfor{\Ypred \not= Y \given \W}} \\
    & = 1 - \pfor{\Ypred = Y} \\
    & = 1 - \E{X, \Ypred}{\pof{\Ytrue = \Ypred \given X}} \\
    & = 1 - \implicitE{\acc}
\end{align}
and on the other hand, we have:
\begin{align}
    \implicitE{\disrate} &= \E{\W, \W'}{\pfor{\Ypred \not= \Ypred' \given \W, \W'}} \\
    &= 1 - \E{\W, \W'}{\pfor{\Ypred = \Ypred' \given \W, \W'}} \\
    &= 1 - \pfor{\Ypred = \Ypred'} \label{jiang:eq:re_gde_soi0} \\
    &= 1 - \E{X, \Ypred}{\pof{\Ypred' = \Ypred \given X}} \label{jiang:eq:re_gde_soi1}\\
    &= 1 - \E{X, \Ypred}{\pof{\Ypred \given X}} \label{jiang:eq:re_gde_soi2} \\
    &= 1 - \implicitE{\predacc}.
\end{align}
The step from \eqref{jiang:eq:re_gde_soi0} to \eqref{jiang:eq:re_gde_soi1} is valid because $\Ypred \independent \Ypred' \given X$, and the step from \eqref{jiang:eq:re_gde_soi1} to \eqref{jiang:eq:re_gde_soi2} is valid because $\pof{\ypred' \given x} = \pof{\ypred \given x}$. Thus, we can rewrite \Cref{jiang:thm:cac_gde} as:
\begin{importantresult}
\begin{lemma}
    \label{jiang:lemma:nicer_gde}
    The model $\pof{\ypred \given x}$ satisfies GDE, when
    \begin{align}
       \implicitE{\acc} = \implicitE{\pof{\Ytrue = \Ypred\given X}} = \implicitE{\pof{\Ypred \given X}} = \implicitE{\predacc},
   \end{align} 
   i.e.\ in expectation, the accuracy of the model equals the predicted accuracy of the model, or equivalently, the error of the model equals its predicted error. 
\end{lemma}
\end{importantresult}

Crucially, while \citet{jiang2021assessing} calls $1 - \E{X, \Ypred}{\pof{\Ypred \given X}}$ the (expected) disagreement rate $\implicitE{\disrate}$, it actually is just the predicted error of the (Bayesian) model as a whole. %

Equally important, all dependencies on $\W$ have vanished. Indeed, we will not use $\W$ anymore for the remainder of this section. This reproduces the corresponding remark from \citet{jiang2021assessing}\footnote{The remark did not exist in the first preprint version.}:
\begin{importantresult}
\begin{insight}
    The theoretical statements in \citet{jiang2021assessing} can be made about any discriminative model with predictions $\pof{y \given x}$.
\end{insight}
\end{importantresult}

When is \(\E{X, \Ypred}{\pof{\Ytrue = \Ypred \given X}} = \E{X, \Ypred}{\pof{\Ypred \given X}}\)? Or in other words: when does $\pof{\Ytrue = \ypred \given x}$ equal $\pof{\Ypred = \ypred \given x}$ in expectation over $\pof{x, \ytrue, \ypred}$?

As a trivial sufficient condition, when the predictive distribution matches our data distribution---\emph{i.e.\ when the model $\pof{\ypred \given x}$ is perfectly calibrated on average for all classes---and not only for the top-1 predicted class}. $\ECE = 0$ is not sufficient because the standard calibration error only ensures that the data distribution and predictive distribution match for the top-1 predicted class  \citep{nixon2019measuring}. But class-wise calibration entails this equality. 

\paragraph{Class-Wise and Class-Aggregated Calibration.} %
To see this, we rewrite class-wise and class-aggregated calibration slightly by employing the following tautology:
\begin{align}
    \pof{\Ypred = \classindex \given \innerexpr{\pof{\Ypred = \classindex \given X}} = \levelthreshold} = \levelthreshold, \label{jiang:eq:magical_tautology}
\end{align}
which is obviously true due its self-referential nature. We provide a formal proof in \S\ref{jiang:app:additional_proofs} in the appendix.
Then we have the following equivalent definition:
\begin{restatable}{lemma}{lemmaccnice}
    \label{jiang:lemma:cc_nice}
    The model $\pof{\ypred \given x}$ satisfies \emph{class-wise calibration} when for any $\levelthreshold \in [0, 1]$ and any class $\classindex \in [\numclasses]$:
    \begin{align}
        \pof{\Ytrue = \classindex, \innerexpr{\pof{\Ypred = \classindex \given X}} = \levelthreshold} 
        =
        \pof{\Ypred = \classindex, \innerexpr{\pof{\Ypred = \classindex \given X}} = \levelthreshold}.
    \end{align}
    Similarly, the model $\pof{\ypred \given x}$ satisfies \emph{class-aggregated calibration} when for any $\levelthreshold \in [0, 1]$:
    \begin{align}
        \pof{\innerexpr{\pof{\Ypred = \Ytrue \given X}} = \levelthreshold} 
        =
        \pof{\innerexpr{\pof{\Ypred \given X}} = \levelthreshold},
    \end{align}
    and \emph{class-wise} calibration implies \emph{class-aggregate} calibration.
\end{restatable}
The straightforward proof is found in \S\ref{jiang:app:additional_proofs} in the appendix.

\citet{jiang2021assessing} mention `level sets' as intuition in their proof sketch. Here, we have been able to make this even clearer: class-aggregated calibration means that level-sets for accuracy $\pof{\Ypred=\Ytrue \given X}$ and predicted accuracy $\pof{\Ypred \given X}$---as random variables of $\Ytrue$ and $X$, and $\Ypred$ and $X$, respectively---have equal measure, that is probability, for all $\levelthreshold$. 

\begin{importantresult_noparbox} 
\paragraph{GDE.} Now, class-aggregated calibration immediately and trivially implies GDE. To see this, we use the following property of expectations:
\begin{lemma}
    \label{jiang:lemma:basic_random_variable_property}
    For a random variable $X$, a function $t(x)$, and the random variable $T = t(X)$, it holds that
    \begin{align}
        \E{T}{T} = \implicitE{T} = \E{X}{t(X)}.
    \end{align}
\end{lemma}
This basic property states that we can either compute an expectation over $T$ by integrating over $\pof{T=t}$ or by integrating $t(x)$ over $\pof{X=x}$. This is just a change of variable (push-forward of a measure).

We can use this property together with the class-aggregated calibration to see:
\begin{align}
    \equalto{\equalto{\equalto{\E{\levelthreshold \sim \pof{\Ypred=\Ytrue \given X}}{q}}{\implicitE{\pof{\Ypred=\Ytrue \given X}}}}{\E{X, \Ytrue}{\pof{\Ypred=\Ytrue \given X}}}}{\implicitE{\acc}} =
    \equalto{\equalto{\equalto{\E{\levelthreshold \sim \pof{\Ypred \given X}}{q}}{\implicitE{\pof{\Ypred \given X}}}}{\E{X, \Ypred}{\pof{\Ypred \given X}}}}{\implicitE{\predacc}},
    \label{jiang:eq:magic_step}
\end{align}
which is exactly \Cref{jiang:lemma:nicer_gde}, where we start with the equality following from class-aggregated calibration and then apply \Cref{jiang:lemma:basic_random_variable_property} along each side.
Thus, GDE is but an expectation over class-aggregated calibration; we have:
\begin{theorem}
    \label{jiang:thm:cac_gde_more_general}
    When a model $\pof{\ypred \given x}$ satisfies class-wise or class-aggregated calibration, it satisfies GDE.
\end{theorem}
\end{importantresult_noparbox}
\begin{proof}
    We can formalize the proof to be even more explicit and introduce two auxiliary random variables: 
    \begin{align}
        S \defeq \pof{\Ypred=\Ytrue \given X},
    \end{align} 
    as a transformed random variable of $\Ytrue$ and $X$, and
    \begin{align}
        T \defeq \pof{\Ypred \given X},
    \end{align}
    as a transformed random variable of $\Ypred$ and $X$. 
    Class-wise calibration implies class-aggregated calibration.
    Class-aggregated calibration then is $\pof{S = \levelthreshold} = \pof{T = \levelthreshold}$ (*).
    Writing out \cref{jiang:eq:magic_step}, we have
    \begin{align}
        \MoveEqLeft{} \implicitE{\pof{\Ypred=\Ytrue \given X}} = \E{X, \Ytrue}{S} =
        \implicitE{S} = \E{S}{S} \\
        & = \int \pof{S=q} \, q \, dq \\
        & \overset{(*)}{=} \int \pof{T=q} \, q \, dq \\
        &= \E{T}{T} = \implicitE{T} = \E{X, \Ypred}{T} = \implicitE{\pof{\Ypred \given X}},
    \end{align}
    which concludes the proof.
\end{proof}
The reader is invited to compare this derivation to the corresponding longer proof in the appendix of \citet{jiang2021assessing}.
The fully probabilistic perspective greatly simplifies the results, and the proofs are straightforward.

\paragraph{CACE.} %
Showing that $\CACE$ bounds the gap between test error and disagreement is also straightforward:
\begin{theorem}
    \label{jiang:thm:cace_bound_more_general}
    For any model $\pof{\ypred \given x}$, we have:
    \begin{align*}
        \lvert \implicitE{\testerror} - \implicitE{\disrate} \rvert \le \CACE.
    \end{align*}
\end{theorem}
\begin{proof}
First, we note that
\begin{align}
    \CACE = \int_{\levelthreshold \in [0, 1]} 
    \begin{aligned}[t]
        \bigl| &\pof{\innerexpr{\pof{\Ypred=\Ytrue \given X}} = \levelthreshold} - \pof{\innerexpr{\pof{\Ypred \given X}} = \levelthreshold} \bigr| d\levelthreshold,
    \end{aligned}
\end{align}
following the equivalences in the proof of \Cref{jiang:lemma:cc_nice}. Then using the triangle inequality for integrals and $0\le \levelthreshold \le 1$, we obtain:
\begin{align}
    & \CACE \\
    & \quad = \int_{\levelthreshold \in [0, 1]} 
    \begin{aligned}[t]
        \bigl| &\pof{\innerexpr{\pof{\Ypred=\Ytrue \given X}} = \levelthreshold} - \pof{\innerexpr{\pof{\Ypred=\Ypred \given X}} = \levelthreshold} \bigr| \, d\levelthreshold
    \end{aligned} \\
    & \quad \ge \int_{\levelthreshold \in [0, 1]} 
    \begin{aligned}[t]
        \levelthreshold \, \bigl| &\pof{\innerexpr{\pof{\Ypred=\Ytrue \given X}} = \levelthreshold} - \pof{\innerexpr{\pof{\Ypred=\Ypred \given X}} = \levelthreshold} \bigr| \, d\levelthreshold
    \end{aligned} \label{jiang:eq:cace_bound_monotonicity} \\
    & \quad \ge 
    \begin{aligned}[t]
        & \bigl| \int_{\levelthreshold \in [0, 1]} \levelthreshold \, \pof{\innerexpr{\pof{\Ypred=\Ytrue \given X}} = \levelthreshold} \, d\levelthreshold - \int_{\levelthreshold \in [0, 1]} \levelthreshold \, \pof{\innerexpr{\pof{\Ypred \given X}} = \levelthreshold} \, d\levelthreshold \bigr|. 
    \end{aligned} \label{jiang:eq:cace_bound_triangle_inequality} \\
    & \quad = \bigl| \implicitE{S} - \implicitE{T} \big| \\
    & \quad = \bigl| \implicitE{\testerror} - \implicitE{\disrate} \big|,
\end{align}
where we have used the monotonicity of integration in \eqref{jiang:eq:cace_bound_monotonicity} and the triangle inequality in \eqref{jiang:eq:cace_bound_triangle_inequality}.
\end{proof}

The bound also serves as another---even simpler---proof for \Cref{jiang:thm:cac_gde_more_general}:
\begin{importantresult}
\begin{insight}
When the Bayesian model satisfies class-wise or class-aggregated calibration, we have $\CACE=0$ and thus $\implicitE{\testerror}=\implicitE{\disrate}$, i.e.\ the model satisfies GDE.
\end{insight}
\end{importantresult}

Furthermore, note again that a Bayesian model was not necessary for the last two theorems. The model parameters $\W$ were not mentioned---except for the specific definitions of $\testerror$ and $\disrate$ which depend on $\W$ following \citet{jiang2021assessing} but which we only use in expectation. 

Moreover, we see that we can easily upper-bound $\CACE$ using the triangle inequality by $2$, narrowing the statement in \citet{jiang2021assessing} that $\CACE$ can lie anywhere in $[0, \numclasses]$:
\begin{insight}
    $\CACE \le 2.$
\end{insight}

Additionally, for completeness, we can also define the class-wise calibration error formally and show that it is bounded by $\CACE$ using the triangle inequality:
\begin{definition}
    The \emph{class-wise calibration error (CWCE)} is defined as:
    \begin{align}
        \CWCE \defeq \sum_\classindex \int_{\levelthreshold \in [0, 1]} & 
        \begin{aligned}[t]
            \bigl| &\pof{\Ytrue = \classindex, \innerexpr{\pof{\Ypred=\classindex \given X}} = \levelthreshold} - \pof{\Ypred = \classindex, \innerexpr{\pof{\Ypred = \classindex \given X}} = \levelthreshold} \bigr|.
        \end{aligned}
    \end{align}
\end{definition}
\begin{lemma}
    $\CWCE \ge \CACE \ge | \implicitE{\acc} - \implicitE{\predacc} |.$
\end{lemma}
Note that when we compute $\CACE$ empirically, we divide the dataset into several bins for different intervals of $\pof{\Ypred=\classindex \given X}$. \citet{jiang2021assessing} use 15 bins. If we were to use a single bin, we would compute $| \implicitE{\acc} - \implicitE{\predacc} |$ directly.

\begin{importantresult}
In \S\ref{jiang:app:robustness_metrics} we show that $\CWCE$ has previously been introduced as `adaptive calibration error' in \citet{nixon2019measuring} and $\CACE$ as `static calibration error' (with noteworthy differences between \citet{nixon2019measuring} and its implementation).
\end{importantresult}

\begin{figure}
    \centering
    \begin{subfigure}{\linewidth}
        \vspace{-0.4em}
        \centering
        \includegraphics[width=\linewidth,trim={5cm 0 0 11.5cm},clip]{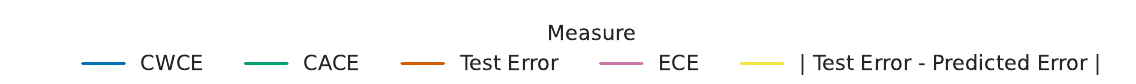} 
    \end{subfigure}
    \begin{subfigure}{\linewidth}
        \vspace{0.1em}
        \caption*{\textbf{CIFAR-10 (In-Distribution)}}
        \vspace{0.1em}
        \begin{subfigure}{0.52\linewidth}
            \centering
            \includegraphics[width=\linewidth,trim={0 1.5cm 11.5cm 0},clip]{plots/cifar10_test_rejection_plot_disrate.pdf}
        \end{subfigure}
        \begin{subfigure}{0.46\linewidth}
            \centering
            \includegraphics[width=\linewidth,trim={13cm 1.5cm 0 0},clip]{plots/cifar10_test_rejection_plot_disrate.pdf}
        \end{subfigure}
    \end{subfigure}
    \begin{subfigure}{\linewidth}
        \vspace{0.1em}
        \caption*{\textbf{CINIC-10 (Distribution Shift)}}
        \vspace{0.1em}
        \begin{subfigure}{0.52\linewidth}
            \centering
            \includegraphics[width=\linewidth,trim={0 1.5cm 11.5cm 0},clip]{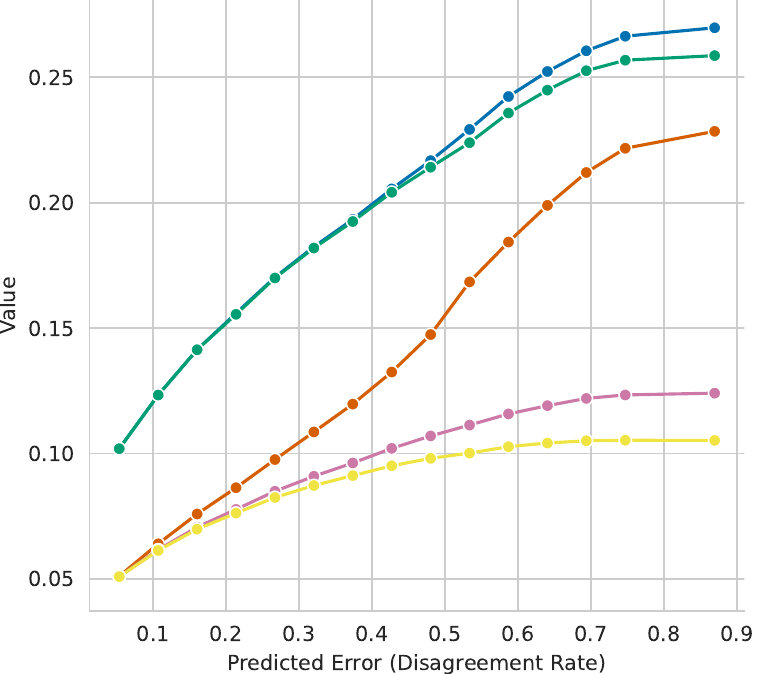}
            \subcaption{Ensemble with TOP}    
        \end{subfigure}
        \begin{subfigure}{0.46\linewidth}
            \centering
            \includegraphics[width=\linewidth,trim={13cm 1.5cm 0 0},clip]{plots/cinic10_test_rejection_plot_disrate.pdf}
            \subcaption{Ensemble without TOP (same underlying models)}    
        \end{subfigure}
    \end{subfigure}
    \caption{\emph{Rejection Plot of Calibration Metrics for Increasing Disagreement In-Distribution (CIFAR-10) and Under Distribution Shift (CINIC-10).} Different calibration metrics ($\ECE$, $\CWCE$, $\CACE$) vary across CIFAR-10 and CINIC-10 on an ensemble of 25 Wide-ResNet-28-10 model trained on CIFAR-10, depending on the rejection threshold of the predicted error (disagreement rate). Thus, calibration cannot be assumed constant for in-distribution data or under distribution shift. The test error increases almost linearly with the predicted error (disagreement rate), 
    leading to `GDE gap' $| \text{Test Error} - \text{Predicted Error} |$ becoming almost flat, providing evidence for the empirical observations in \citet{nakkiran2020distributional, jiang2021assessing}. The mean predicted error (disagreement rate) is shown on the x-axis. \textbf{(a)} shows results for an ensemble using TOP (following \citet{jiang2021assessing}), and \textbf{(b)} for a regular deep ensemble without TOP. The regular deep ensemble is better calibrated but has higher test error overall and lower test error for samples with small predicted error.
    }
    \label{jiang:fig:rejection_plot}
\end{figure}

\section{Deterioration of Calibration under Increasing Disagreement}
\label{jiang:sec:empirical_example}

Generally, we can only hope to trust model calibration for in-distribution data, while under distribution shift, the calibration ought to deteriorate. %
In our empirical falsification using models trained on CIFAR-10 and evaluated on the test sets of CIFAR-10 and CINIC-10, as a dataset with a distribution shift, we find in both cases that calibration deteriorates under increasing disagreement. 
We further examine ImageNET and PACS in \S\ref{jiang:appsec:additional_results}.
Most importantly though, calibration markedly worsens under distribution shift. 

Specifically, we examine an ensemble of 25 WideResNet models \citep{zagoruyko2016wide} trained on CIFAR-10 \citep{krizhevsky2009learning} and evaluated on CIFAR-10 and CINIC-10 test data. CINIC-10 \citep{darlow2018cinic} consists of CIFAR-10 and down-scaled ImageNet samples for the same classes, and thus includes a distribution shift. 
The training setup follows the one described in \citet{mukhoti2021deterministic}, see appendix \S\ref{jiang:appsec:experiment_setup}.

\Cref{jiang:fig:rejection_plot} shows rejection plots under increasing disagreement for in-distribution data (CIFAR-10) and under distribution shift (CINIC-10).
The rejection plots threshold the dataset on increasing levels of the predicted error (disagreement rate)---which is a measure of epistemic uncertainty when there is no expected aleatoric uncertainty in the dataset. %
We examine ECE, class-aggregated calibration error ($\CACE$), class-wise calibration error ($\CWCE$), error $\implicitE{\testerror}$, and `GDE gap', $| \implicitE{\acc} - \implicitE{\predacc} |$, as the predicted error (disagreement rate), $\implicitE{\disrate} = 1 - \implicitE{\predacc}$,  increases.
We observe that all calibration metrics, ECE, CACE and CWCE, deteriorate under increasing disagreement, both in distribution and under distribution shift, and also worsen under distribution shift overall.

We also observe the same for ImageNet \citep{deng2009imagenet} and PACS \citep{li2017deeper}, which we show in appendix \S\ref{jiang:appsec:additional_results}.

This is consistent with the experimental results of \citet{ovadia2019can} which examines dataset shifts.
However, given that the calibration metrics change with the quantity of interest, we conclude that: 

\begin{importantresult}
\begin{insight}
    The bound from \Cref{jiang:thm:cace_bound} might not have as much expressive power as hoped since the calibration metrics themselves deteriorate as the model becomes more `uncertain' about the data.
\end{insight}
\end{importantresult}

At the same time, the `GDE gap', which is the actual gap between test error and predicted error, flattens, and the test error develops an almost linear relationship with the predicted error (up to a bias). This shows that there seem to be intriguing empirical properties of deep ensemble as observed previously \citep{nakkiran2020distributional,jiang2021assessing}. However, they are not explained by the proposed calibration metrics\footnote{The simplest explanation is that very few samples have high predicted error and thus the rejection plots flatten. This is not true. For CINIC-10, the first bucket contains 50k samples, and each latter buckets adds additional $\sim$10k samples.}.

As described previously, the results are not limited to Bayesian or version-space models but also apply to any model $\pof{\ypred \given x}$, including regular deep ensembles without TOP.
In our experiment, we find that a regular deep ensemble is better calibrated than the same ensemble made to satisfy TOP. 
We hypothesize that each ensemble member's own predictive distribution is better calibrated than its one-hot outputs, yielding a better calibrated ensemble overall.

Given that all these calibration metrics require access to the labels, and we cannot assume the model to be calibrated under distribution shift, we might just as well use the labels directly to assess the test error.

\section{Related Work}
\label{jiang:sec:discussion}

Here, we discuss connections to Bayesian model disagreement and epistemic uncertainty, as well as connections to information theory. %
We expand on these points in much greater detail in appendix \S\ref{jiang:appsec:discussion}.

\paragraph{Bayesian Model Disagreement.} From a Bayesian perspective, as the epistemic uncertainty increases, we expect the model to become less reliable in its predictions.
The predicted error of the model is a measure of the model's overall uncertainty, which is the total of aleatoric and epistemic uncertainty and thus correlated with epistemic uncertainty.
Thus, we can hypothesize that as the predicted error increases, the model should become less reliable, which will be reflected in increasing calibration metrics. 
This is exactly what we have empirically validated in the previous section.

\paragraph{Connection to Information Theory.}
At first sight, \citet{jiang2021assessing} seems disconnected from information theory. However, we can draw a connection by using $\aICof{p} \defeq 1 - p$ as a linear approximation for Shannon's information content $\ICof{p}=-\log p$.
Semantically, both this approximation and Shannon's information content quantify surprise (i.e., prediction error). Both are $0$ for certain events. For unlikely events, the former tends to $1$ while the latter tends to $+\infty$.

This leads to common-sense definitions and statements from an information-theoretic point of view. We can even formulate parallel statements using information theory and see that the statements relate to total uncertainty and not epistemic uncertainty in a Bayesian sense.

\paragraph{Other Related Literature.} %
Beyond \citet{jiang2021assessing}, this note offers a perspective on \citet{granese2021doctor} and \citet{garg2022leveraging}, which are proposing related approaches.

\citet{granese2021doctor} use the predicted error (disagreement rate), referred to as $D_\alpha$, and the predicted top-1 error, $D_\beta$, to estimate when the model will be wrong. 
As noted in \S\ref{jiang:appsec:discussion}, the predicted error can be seen as an approximation of Shannon's entropy.
Thus, $D_\alpha$ is effectively using an approximation of the prediction entropy for OoD detection and rejection classification. Similarly, $D_\beta$ is the maximum class confidence. Both are well-known baselines for OoD detection \citep{hendrycks2016baseline}. There is no ablation to see how $D_\alpha$ and $D_\beta$ differ from these baselines. We leave this to future work.
The paper frames the question of whether a model's predictions will be correct as binary classification problem on top of the underlying model's output probabilities and investigate this from a theoretical point of view. They also examine using input perturbations similar to \citet{liang2017enhancing} and \citet{lee2018simple}.

\citet{garg2022leveraging} threshold the predictive entropy or maximum class confidence to estimate the test error under distribution shift. They estimate the threshold by calibrating it on in-distribution labeled data: the threshold is chosen such that the percentage of rejected in-distribution validation data approximately equals the test error on this in-distribution data. They call this approach \emph{Average Thresholded Confidence (ATC)}. They find that ATC using entropy performs better than ATC using the maximum confidence and other approaches, including GDE. Their results show that ATC also degrades under increasing distribution shifts similar to what we have seen for GDE in \S\ref{jiang:sec:empirical_example} as the choice of threshold is explicitly tied to the in-distribution\footnote{See also Figures 7, 8, and 9 in the appendix of \citet{garg2022leveraging}}.
\citet{garg2022leveraging} explicitly examine the theoretical limits when no further assumptions are made.

\section{Discussion}
\label{jiang:sec:conclusion}

We have found that the theoretical statements in \citet{jiang2021assessing} can be expressed and proven more concisely when using probabilistic notation for (Bayesian) models that output softmax probabilities.

Moreover, we empirically found the proposed calibration metrics to deteriorate under increasing disagreement for in-distribution data, and as expected, we have found the same behavior under distribution shifts.

While \citet{jiang2021assessing} are careful to qualify their results for distribution shifts, above results should give us pause: strong assumptions are still needed to conjecture about model generalization, and we need to beware of circular arguments.

\FloatBarrier

		\section{Details}
		\let\chapter\section
		\let\section\subsection
		\let\subsection\subsubsection
		\let\subsubsection\paragraph
		\let\paragraph\textbf
		
\section{Equivalent Definitions}
\label{jiang:app:equivalent_definitions}
{
\newcommand{\versionspace}{\mathcal{H}}
\newcommand{\trainingalgorithm}{\mathcal{A}}
\citet{jiang2021assessing} defines a \emph{hypothesis space} $\versionspace$. In the literature, this is also sometimes called a version space. The hypothesis space induced by a stochastic training algorithm $\trainingalgorithm$ is named $\versionspace_\trainingalgorithm$. 

We can identify each hypothesis $h: \mathcal{X} \rightarrow[\numclasses]$ with itself as parameter $\w_h = h$ and define $\pof{\w_h}$ as a uniform distribution over all parameters/hypotheses in $\versionspace_\trainingalgorithm$. This has the advantage of formalizing the distribution from which hypothesis are drawn ($h \sim \versionspace_\trainingalgorithm$), which is not made explicit in \citet{jiang2021assessing}. $h(x) = \classindex$ then becomes $\argmax_{\ypred} {\pof{\ypred \given x, \w_h}} = \classindex$. Moreover, as $\pof{\ypred, \w \given x}$ satisfies TOP, we have\footnote{We put definitions and expressions written using the notation and variables from \citet{jiang2021assessing} inside quotation marks ``'' to avoid ambiguities.} 
\begin{align}
    ``\indicator{h(x) = \classindex}'' &= \pof{\Ypred = \classindex \given x, \w_h}.
\end{align}

Thus, ``$\operatorname{TestErr}_{\mathscr{D}}(h) \triangleq \mathbb{E}_{\mathscr{D}}[\mathbb{1}[h(X) \neq Y]]$'' is equivalent to:
\begin{align}
    ``\operatorname{TestErr}_{\mathscr{D}}(h) &=  \mathbb{E}_{\mathscr{D}}[\mathbb{1}[h(X) \neq Y]]'' \\
    &= \E{X, \Ytrue}{\pof{\Ypred \not= \Ytrue \given X, \w_h}} \\
    &= \E{X}{\pfor{\Ypred \not= \Ytrue \given X, \w_h}} \\
    &= \pfor{\Ypred \not= \Ytrue \given \w_h} \\
    &= \testerror(\w_h).
\end{align}

Similarly, ``$\operatorname{Dis}_{\mathscr{D}}\left(h, h^{\prime}\right) \triangleq \mathbb{E}_{\mathscr{D}}\left[\mathbb{1}\left[h(X) \neq h^{\prime}(X)\right]\right]$'' is equivalent to:
\begin{align}
    ``\operatorname{Dis}_{\mathscr{D}}\left(h, h^{\prime}\right) &\triangleq \mathbb{E}_{\mathscr{D}}\left[\mathbb{1}\left[h(X) \neq h^{\prime}(X)\right]\right]'' \\
    &= \E{\Ypred, \Ypred'}{\pfor{\Ypred \not= \Ypred' \given \w_h, \w_{h'}}} \\
    &= \disrate(\w_h, \w_{h'}).
\end{align}

Further, ``$\tilde{h}_{k}(x) \triangleq \mathbb{E}_{\mathscr{H}_{\mathcal{A}}}[\mathbb{1}[h(x)=k]]$'' is equivalent to:
\begin{align}
    ``\tilde{h}_{k}(x) &\triangleq \mathbb{E}_{\mathscr{H}_{\mathcal{A}}}[\mathbb{1}[h(x)=k]]'' \\
    &= \E{\W}{\pof{\Ypred = \classindex \given x, \W}} \\
    &= \pof{\Ypred = \classindex \given x}. 
\end{align}

For the GDE, ``$\mathbb{E}_{h, h^{\prime} \sim \mathscr{H}_{\mathcal{A}}}\left[\operatorname{Dis}_{\mathscr{D}}\left(h, h^{\prime}\right)\right]=\mathbb{E}_{h \sim \mathscr{H}_{\mathcal{A}}}\left[\operatorname{TestErr}_{\mathscr{g}}(h)\right]$'' is equivalent to:
\begin{align}
    & ``\mathbb{E}_{h, h^{\prime} \sim \mathscr{H}_{\mathcal{A}}}\left[\operatorname{Dis}_{\mathscr{D}}\left(h, h^{\prime}\right)\right]=\mathbb{E}_{h \sim \mathscr{H}_{\mathcal{A}}}\left[\operatorname{TestErr}_{\mathscr{g}}(h)\right]'' \notag \\
    \Leftrightarrow & \E{\W, \W'}{\disrate(\W, \W')} = \E{\W}{\testerror(\W)}.
\end{align}

For the class-wise calibration, ``$p(Y=k \mid \tilde{h}_{k}(X)=q)=q$'' is equivalent to:
\begin{align}
    & ``p(Y=k \mid \tilde{h}_{k}(X)=q)=q'' \\
    \Leftrightarrow & \pof{\Ytrue = \classindex \given \innerexpr{\pof{\Ypred = \classindex \given X}} = q} = q.
\end{align}

For the class-aggregated calibration, ``$\frac{\sum_{k=0}^{K-1} p(Y=k, \tilde{h}_{k}(X)=q)}{\sum_{k=0}^{K-1} p(\tilde{h}_{k}(X)=q)}=q$'' (and note in \citet{jiang2021assessing}, class indices run from $0..K-1$) is equivalent to:
\begin{align}
    &``\frac{\sum_{k=0}^{K-1} p(Y=k, \tilde{h}_{k}(X)=q)}{\sum_{k=0}^{K-1} p(\tilde{h}_{k}(X)=q)}=q'' \\
    \Leftrightarrow &\frac{\sum_{\classindex=1}^{\numclasses} \pof{\Ytrue=k, \innerexpr{\pof{\Ypred = \classindex \given X}} = q}}{\sum_{\classindex=1}^\numclasses \pof{\innerexpr{\pof{\Ypred = \classindex \given X }} = q}} = q \\
    \Leftrightarrow & \sum_{\classindex=1}^{\numclasses} \pof{\Ytrue=k, \innerexpr{\pof{\Ypred = \classindex \given X}} = q} \notag \\
    & \quad = q \sum_{\classindex=1}^\numclasses \pof{\innerexpr{\pof{\Ypred = \classindex \given X }} = q}.
\end{align}

Finally, for the class-aggregated calibration error, the definition is equivalent to:
\begin{align}
    \MoveEqLeft{} ``\operatorname{CACE}_{\mathscr{D}}(\tilde{h}) \notag \\
    & \triangleq \int_{q \in[0,1]}\left|\frac{\sum_{k} p\left(Y=k, \tilde{h}_{k}(X)=q\right)}{\sum_{k} p\left(\tilde{h}_{k}(X)=q\right)}-q\right| \cdot \sum_{k} p\left(\tilde{h}_{k}(X)=q\right) d q \\
    & = \int_{q \in[0,1]}\bigl|\sum_{k} p\left(Y=k, \tilde{h}_{k}(X)=q\right) -q \sum_{k} p\left(\tilde{h}_{k}(X)=q\right) \bigl| d q'' \\
    & = \int_{q \in[0,1]}\bigl|\sum_{k} \pof{\Ytrue=\classindex, \innerexpr{\pof{\Ypred =\classindex \given X}}=q} -q \sum_{k} \pof{\innerexpr{\pof{\Ypred =\classindex \given X}}=q} \bigl| d q
\end{align}

\section{Comparison of $\CACE$ and $\CWCE$ with calibration metrics with `adaptive calibration error' and `static calibration error'}
\label{jiang:app:robustness_metrics}

\citet{nixon2019measuring} examine shortcomings of the ECE metric and identify a lack of class conditionality, adaptivity and the focus on the maximum probability (argmax class) as issues.
They suggest an adaptive calibration error which uses adaptive binning and averages of the calibration error separately for each class, thus equivalent to the class-wise calibration error and class-wise calibration (up to adaptive vs.\ even binning). In the paper, the static calibration error is defined as ACE with even instead of adaptive binning. However, in the widely used implementation\footnote{\shorturl{https://github.com/google-research/robustness_metrics/blob/baa47fbe38f80913590545fe7c199898f9aff349/robustness_metrics/metrics/uncertainty.py\#L1585}, added in April 2021}, SCE is defined as equivalent to the class-aggregated calibration error.

\section{Expanded Discussion}
\label{jiang:appsec:discussion}

Here, we discuss connections to Bayesian model disagreement and epistemic uncertainty, as well as connections to information theory, the bias-variance trade-off, and prior literature.

\subsection{Bayesian Model Disagreement}

From a Bayesian perspective, as the epistemic uncertainty increases, we expect the model to become less reliable in its predictions. The predicted error of the model is a measure of the overall uncertainty of the model which is the total of aleatoric and epistemic uncertainty, and thus correlated with epistemic uncertainty. Thus, we can hypothesize that as the predicted error increases, the model should become less reliable, which will be reflected in increasing calibration metrics. This is what we have empirically validated in the previous section.
We can expand on the connection to the Bayesian perspective. %
In particular, we can connect the statements of \citet{jiang2021assessing} to a well-known Bayesian measure of model disagreement.

In \S\ref{jiang:appsec:additional_results}, we also report empirical results for rejection plots based on Bayesian model disagreement instead of predicted error.

\subsection{Connection to Information Theory}
\label{jiang:sec:special_case}

At first sight, \citet{jiang2021assessing} seems  disconnected from information theory. However, we can recover statements by using $\aICof{p} \defeq 1 - p$ as a linear approximation for Shannon's information content $\ICof{p}$:
\begin{align}
    \aICof{p} = 1 - p \le -\log p = \ICof{p}.
\end{align} 
$\aICof{p}$ is just the first-order Taylor expansion of $\ICof{p} = -\log p$ around 1.
Semantically, both Shannon's information content and this approximation quantify surprise. Both are $0$ for certain events. For unlikely events, the former tends to $+\infty$ while the latter tends to $1$. 

We can define an \emph{approximate entropy} $\aHof{X}$ using $\opInformationContent'$:
\begin{align}
    \aHof{X} \defeq \implicitE{\aICof{\pof{X}}}=1 - \implicitE{\pof{x}} = 1 - \sum_x \pof{x}^2,
\end{align}
and an \emph{approximate mutual information} $\aMIof{X ; Y}$:
\begin{align}
    \aMIof{X ; Y} \defeq \aHof{X} - \aHof{X \given Y} = \aHof{X} - \simpleE{\pof{y}} \aHof{X \given y},
\end{align}
following the semantic notion of mutual information as expected information gain in \S\ref{jiang:sec:background}.

\textbf{$\aMIof{\Ypred ; \W \given x}$ as Covariance Trace.} This approximate mutual information has a surprisingly nice property, which was detailed in \citet{smith2018understanding} originally:
\begin{restatable}{proposition}{approxmivarsum}
    \label{jiang:prop:approxmivarsum}
    The approximate mutual information $\aMIof{\Ypred ; \W \given x}$ is equal the sum of the variances of $\ypred \given x, \W$ over all $\ypred$:
    \begin{align}
        \aMIof{\Ypred ; \W \given x} = \sum_{\ypred=1}^K \Var{\W}{\pof{\ypred \given x,\W}} \ge 0.
    \end{align} 
\end{restatable}
We present a proof in \S\ref{jiang:app:additional_proofs}.
The sum of variances of the predictive probabilities (or trace of the respective covariance matrix) is a common proxy for epistemic uncertainty \citep{gal2017deep}, and here the mutual information $\aMIof{\Ypred ; \W \given x}$ using $\hat \opInformationContent$ is just that. This gives evidence that these definitions are sensible and connects them to other prior Bayesian literature. 
Importantly, this also shows that $\aHof{\Ypred \given x} \ge \aHof{\Ypred \given x, \W}$.

\textbf{Connection to \citet{jiang2021assessing}.} %
As random variable of $X$ and $Y$, $\aHof{\Ypred=\Ytrue \given X}$ is the test error:
\begin{align}
    \aHof{\Ypred=\Ytrue \given X} = 1 - \pof{\Ypred=\Ytrue \given X} = \testerror.
\end{align} 
Thus, the approximate cross-entropy 
\begin{align}
    \MoveEqLeft{} \aCrossEntropy{\pof{\Ytrue \given X}}{\pof{\Ypred = \Ytrue \given X}} = \E{\pof{\Ytrue \given X}}{\aICof{\pof{\Ypred = \Ytrue \given X}}}
\end{align}
is the expected test error $\implicitE{\testerror}$.

Similarly, when TOP is fulfilled, the mutual information $\aMIof{\Ypred ; \W \given X}$ is the expected disagreement rate $\implicitE{\disrate}$. That is, when $\Ypred \given X, \W$ is one-hot, we have:
\begin{align}
    \aHof{\Ypred \given X, \W} = 1 - \simpleE{X} \underbrace{\E{\Ypred} {\pof{\Ypred \given X, \W} \given X}}_{=1} = 0,
\end{align}
and thus:
\begin{align}
    \aMIof{\Ypred ; \W \given X} &= \aHof{\Ypred \given X} - \aHof{\Ypred \given X, \W} \\
    &= \aHof{\Ypred \given X} \\
    &= 1 - \E{X, \Ypred}{\pof{\Ypred \given X}} \\
    &= \implicitE{\disrate}.
\end{align}
\begin{lemma}
    When the model $\pof{\ypred \given x, \w}$ satisfies TOP, the GDE is equivalent to:
    \begin{align}
        \aCrossEntropy{\pof{\Ytrue \given X}}{\pof{\Ypred = \Ytrue \given X}} 
        = \aMIof{\Ypred ; \W \given X}.
    \end{align}
\end{lemma}
This relates the approximate cross-entropy loss (test error) to the approximate Bayesian model disagreement.

\textbf{Without TOP.} If TOP does not hold, the \emph{actual} expected disagreement $\aMIof{\Ypred ; \W \given x}$ lower-bounds the 
``expected disagreement rate'' $\implicitE{\disrate}$, which then equals the expected \emph{predicted} error $1 - \E{X, \Ypred}{\pof{\Ypred \given X}}$ when we have GDE. We have the following general equivalence to GDE:
\begin{lemma}
    For a model $\pof{\ypred \given x}$, the GDE is equivalent to:
    \begin{align}
        \aCrossEntropy{\pof{\Ytrue \given X}}{\pof{\Ypred = \Ytrue \given X}} 
        = \aHof{\Ypred \given X} \ge \aMIof{\Ypred ; \W \given X}.
    \end{align}
\end{lemma}
The other statements and proofs translate likewise, and intuitively seem sensible from an information-theoretic perspective. We can go further and directly establish analogous properties using information theory in the next subsection.

\subsection{Information-Theoretic Version}
\label{jiang:sec:info_theoretic_version}

Here, we derive an information-theoretic version of the GDE both under the assumption of TOP and without. Importantly, we will not require a Bayesian model for any of the main statements as they hold for any model $\pof{\ypred \given x}$. We show that we can artificially introduce a connection to disagreement using TOP. %

\textbf{Information-Theoretic GDE.} We have already introduced the BALD equation \cref{jiang:eq:BALD}, which connects expected disagreement and predictive uncertainty:
\begin{align*}
    \MIof{\Ypred ; \W \given x} &= \Hof{\Ypred \given x} - \Hof{\Ypred \given x, \W}
\end{align*}
The expected disagreement is measured by the mutual information $\MIof{\Ypred ; \W \given X}$, and the prediction error is measured by the cross-entropy of the predictive distribution under the true data generating distribution $\CrossEntropy{\pof{\Ytrue \given X}}{ \pof{\Ypred = \Ytrue \given X} }$. Indeed, the test error is bounded by it \citep{kirsch2020unpacking}: 
\begin{align}
    \pof{\Ytrue \not= \Ypred} \le 1-e^{-\CrossEntropy{\pof{\Ytrue \given X}}{ \pof{\Ypred = \Ytrue \given X} }}.
\end{align}

When our model fulfills TOP, we have $\Hof{\Ypred \given X, \W} = 0$, and thus $\MIof{\Ypred ; \W \given X} = \Hof{\Ypred \given X}$. The expected disagreement then equals the predicted label uncertainty $\Hof{\Ypred \given X}$. Generally, we can define an `entropic GDE':
\begin{definition}
    A model $\pof{\ypred \given x}$ satisfies entropic GDE, when:
    \begin{align}
        \CrossEntropy{\pof{\Ytrue \given X}}{ \pof{\Ypred = \Ytrue \given X} } = \Hof{\Ypred \given X}.
    \end{align}
\end{definition}
\begin{lemma}
    When a Bayesian model $\pof{\ypred, \w \given x}$ satisfies TOP, entropic GDE is equivalent to
    \begin{align}
        \CrossEntropy{\pof{\Ytrue \given X}}{ \pof{\Ypred = \Ytrue \given X} } = \MIof{\Ypred ; \W \given X}.
    \end{align}
\end{lemma}
The latter is close to GDE, especially when comparing to the previous section.

We can formulate an entropic class-aggregated calibration by connecting $\Hof{\ypred \given x}$ with $\Hof{\ytrue \given x}$. That is, instead of using probabilities, we use Shannon's information-content:
\begin{definition}
    The model $\pof{\ypred \given x}$ satisfies \emph{entropic class-aggregated calibration} when for any $\levelthreshold \ge 0$:
    \begin{align}
        \pof{\Hof{\Ypred = \Ytrue \given X} = \levelthreshold} 
        =
        \pof{\Hof{\Ypred = \Ypred \given X} = \levelthreshold}.
    \end{align}
    Similarly, we can define the \emph{entropic class-aggregated calibration error (ECACE)}:
    \begin{align}
        \ECACE \defeq \int_{\levelthreshold \in [0, \infty)} 
        \begin{aligned}[t]
            \bigl| &\pof{\Hof{\Ypred=\Ytrue \given X} = \levelthreshold} \\
            & - \pof{\Hof{\Ypred = \Ypred \given X} = \levelthreshold} \bigr| d\levelthreshold.
        \end{aligned}
    \end{align}
\end{definition}
As $- \log{p}$ is strictly monotonic and thus invertible for non-negative $p$, entropic class-aggregated calibration and class-aggregated calibration are equivalent. $\ECACE$ and $\CACE$ are not, though.

The expectation of the transformed random variable $\Hof{\Ypred = \Ytrue \given X}$ (in $\Ytrue$ and $X$) is just the cross-entropy:
\begin{align}
    \simpleE{X, \Ytrue} \Hof{\Ypred = \Ytrue \given X} = \simpleE{\pof{x, \ytrue}} \Hof{\Ypred = \ytrue \given X} = \CrossEntropy{\pof{\Ytrue \given X}}{\pof{\Ypred = \Ytrue \given X}}.
\end{align} 

Using this notation, and analogous to \Cref{jiang:thm:cace_bound}, we can show:
\begin{theorem}
    \label{jiang:thm:ecace_bound}
    When $\Hof{\ypred \given x} = -\log \pof{\ypred \given x}$ is upper-bounded by $L$ for all $\ypred$ and $x$, we have:
    \begin{align}
        \ECACE 
        &\ge \frac{1}{L} \bigl| \CrossEntropy{\pof{\Ytrue \given X}}{\pof{\Ypred = \Ytrue \given X}} - \Hof{\Ypred \given X} \bigr|, \\
    \intertext{and when the model satisfies TOP, equivalently:}
        &= \frac{1}{L} \bigl| \CrossEntropy{\pof{\Ytrue \given X}}{\pof{\Ypred = \Ytrue \given X}} - \MIof{\Ypred ; \W \given X} \bigr|.
    \end{align}
\end{theorem}
There might be better conditions than the upper-bound above, but this bound is in the spirit of \citet{jiang2021assessing}. Indeed, the proof of \Cref{jiang:thm:cace_bound} is the same, except that we use $\levelthreshold \le L$ instead of $\levelthreshold \le 1$. 
Finally, when the model satisfies entropic class-aggregated calibration, $\ECACE = 0$, cross-entropy (or negative expected log likelihood) equals disagreement (respectively, predicted label uncertainty when TOP does not hold). Thus, we have:
\begin{theorem}
    When the model $\pof{\ypred \given x}$ satisfies \emph{entropic class-aggregated calibration}, it trivially also satisfies entropic GDE:
    \begin{align}
        \CrossEntropy{\pof{\Ytrue \given X}}{\pof{\Ypred = \Ytrue \given X}} = \Hof{\Ypred \given X} \ge \MIof{\Ypred ; \W \given X},
    \end{align}
    and when TOP holds:
    \begin{align}
        \CrossEntropy{\pof{\Ytrue \given X}}{\pof{\Ypred = \Ytrue \given X}} = \MIof{\Ypred ; \W \given X}.
    \end{align}
\end{theorem}

\textbf{Without TOP.} Again, if we do not expect one-hot predictions for our ensemble members, the analogy put forward in \citet{jiang2021assessing} breaks down because the Bayesian disagreement $\MIof{\Ypred ; \W \given X}$ only lower bounds the predicted label uncertainty $\Hof{\Ypred \given X}$ and can not be connected to $\ECACE$ the same way. But this also breaks down in the regular version in \citet{jiang2021assessing}.

\section{Additional Proofs}
\label{jiang:app:additional_proofs}
\begin{lemma}
    \label{jiang:lemma:magic_tautology}
    For a model $\pof{\ypred \given x}$, we have for all $k \in [K]$ and $q \in [0, 1]$:
    \begin{align}
        \pof{\Ypred = \classindex \given \innerexpr{\pof{\Ypred = \classindex \given X}} = \levelthreshold} = \levelthreshold,
    \end{align}
    when the left-hand side is well-defined.
\end{lemma}
\begin{proof}
    This is equivalent to
    \begin{align}
        \pof{\Ypred = \classindex, \innerexpr{\pof{\Ypred = \classindex \given X}} = \levelthreshold} = \levelthreshold \, \pof{\innerexpr{\pof{\Ypred = \classindex \given X}} = \levelthreshold},
    \end{align}
    as the conditional probability is either defined or $\pof{\innerexpr{\pof{\Ypred = \classindex \given X}} = \levelthreshold} = 0$.
    Assume the former. Let $\pof{\innerexpr{\pof{\Ypred = \classindex \given X}} = \levelthreshold} > 0$.
    Introducing the auxiliary random variable $T_k \defeq \innerexpr{\pof{\Ypred = \classindex \given X}}$ as a transformed random variable of $X$, we have
    \begin{align}
        \pof{\Ypred = k, T_k = q} = q \, \pof{T_k = q}.
    \end{align}
    We can write the probability as an expectation over an indicator function
    \begin{align}
        \MoveEqLeft{} \pof{\Ypred = k, T_k = q} \\
        & = \E{X, \Ypred} {\indicator{\Ypred=k, T_k(X) = q}} \\
        & = \E{X, \Ypred} {\indicator{\Ypred=k} \, \indicator{T_k(X) = q}} \\
        & = \E{X}{\indicator{T_k(X) = q} \, \E{\Ypred}{\indicator{\Ypred = k} \given X }} \\
        &= \E{X}{\indicator{T_k(X) = q} \, \pof{\Ypred = k \given X}}.
    \intertext{
    Importantly, if $\indicator{T_k(x) = q} = 1$ for an $x$, we have $T_k(x) = \pof{\Ypred = k \given x} = q$, and otherwise, we multiply with $0$. Thus, this is equivalent to
    }
        & = \E{X}{\indicator{T_k(X) = q} \, q} \\
        & = q \E{X}{\indicator{T_k(X) = q}} \\
        & = q \pof{T_k(X) = q}.
    \end{align}
\end{proof}

\lemmaccnice*
\begin{proof}
    Beginning from 
    \begin{align}
        & \pof{\Ytrue = \classindex \given \innerexpr{\pof{\Ypred = \classindex \given X}} = \levelthreshold} = \levelthreshold,
    \end{align}
    we expand the conditional probability to
    \begin{align}
        \Leftrightarrow
            \pof{\Ytrue = \classindex, \innerexpr{\pof{\Ypred = \classindex \given X}} = \levelthreshold} = \levelthreshold \, \pof{\innerexpr{\pof{\Ypred = \classindex \given X}} = \levelthreshold},
    \end{align}   
    and substitute \cref{jiang:eq:magical_tautology} into the outer $q$, obtaining the first equivalence
    \begin{align}
        \Leftrightarrow \pof{\Ytrue = \classindex, \innerexpr{\pof{\Ypred = \classindex \given X}} = \levelthreshold} 
        =
        \pof{\Ypred = \classindex, \innerexpr{\pof{\Ypred = \classindex \given X}} = \levelthreshold}. \label{jiang:eq:cwc_nice}
    \end{align}
    For the second equivalence, we follow the same approach. 
    Beginning from
    \begin{align}
        \sum_\classindex \pof{\Ytrue = \classindex , \innerexpr{\pof{\Ypred=\classindex \given X}} = \levelthreshold}
        = \levelthreshold \sum_\classindex \pof{\innerexpr{\pof{\Ypred=\classindex \given X}} = \levelthreshold}
        ,
    \end{align}
    we pull the outer $q$ into the sum and expand using \eqref{jiang:eq:magical_tautology}
    \begin{align}
        \Leftrightarrow \sum_\classindex \pof{\Ytrue = \classindex , \innerexpr{\pof{\Ypred=\classindex \given X}} = \levelthreshold} 
         =
        \sum_\classindex \levelthreshold \pof{\innerexpr{\pof{\Ypred=\classindex \given X}} = \levelthreshold}
         =
        \sum_\classindex \pof{\Ypred = \classindex , \innerexpr{\pof{\Ypred=\classindex \given X}} = \levelthreshold}. 
        \label{jiang:eq:cac_intermediate}
    \end{align}
    In the inner expression, $k$ is tied to $\Ytrue$ on the left-hand side and $\Ypred$ on the right-hand side, so we have
    \begin{align}
        \Leftrightarrow \sum_\classindex \pof{\Ytrue = \classindex , \innerexpr{\pof{\Ypred=\Ytrue \given X}} = \levelthreshold} 
        =
        \sum_\classindex \pof{\Ypred = \classindex , \innerexpr{\pof{\Ypred\given X}} = \levelthreshold}. 
    \end{align}
    Summing over $k$, marginalizes out $\Ytrue = k$ and $\Ypred = k$ respectively, yielding the second equivalence
    \begin{align}
        \Leftrightarrow \pof{\innerexpr{\pof{\Ypred=\Ytrue \given X}} = \levelthreshold} 
        = 
        \pof{\innerexpr{\pof{\Ypred \given X}} = \levelthreshold}. 
    \end{align}
    Finally, class-wise calibration implies class-aggregated calibration as summing over different $k$ in \eqref{jiang:eq:cwc_nice}, which is equivalent to class-wise calibration, yields \eqref{jiang:eq:cac_intermediate}, which is equivalent to class-aggregated calibration. 
\end{proof}
}

\approxmivarsum*
\begin{proof}
    We show that both sides are equal:
    \begin{align}
        \MoveEqLeft{}
        \aMIof{\Ypred ; \W \given x} = \aHof{\Ypred \given x} - \aHof{\Ypred \given x, \W} \\
        &= 
        \E{\Ypred}{1-\pof{\Ypred \given x}} 
        - 
        \E{\Ypred, \W}{1-\pof{\Ypred \given x, \W}}\\
        &= 
        \E{\Ypred, \W}{\pof{\Ypred \given x, \W}}
        - 
        \E{\Ypred}{\pof{\Ypred \given x}}\\
        &= \simpleE{\W} \E{\pof{\ypred, x, \W}}{\pof{\ypred \given x, \W}} 
        - 
        \simpleE{\pof{\ypred \given x}}{\pof{\ypred \given x}}\\
        &=  \E*{\W}{\sum_{\ypred=1}^K \pof{\ypred \given x, \W}^2} 
        -
        \sum_{\ypred=1}^K \E{\W}{\pof{\ypred \given x, \W}}^2 \\
        &=  \sum_{\ypred=1}^K \E*{\W}{\pof{\ypred \given x, \W}^2} -
        \E{\W}{\pof{\ypred \given x, \W}}^2 \\
        &= \sum_{\ypred=1}^K \Var{\W}{\pof{\ypred \given x, \W}} \\
        &\ge 0,
    \end{align}
    where we have used that $\simpleE{\pof{\ypred \given x}}{\pof{\ypred \given x}} = \sum_{\ypred=1}^K \pof{\ypred \given x}^2$.
\end{proof}

\section{Empirical Validation of Calibration Deterioration under Increasing Disagreement}
\label{jiang:appsec:experiments}

Here, we discuss additional details to allow for reproduction and present results on additional datasets. In addition to the experiments on CIFAR-10 \citep{krizhevsky2009learning} and CINIC-10 \citep{darlow2018cinic}, we report results for ImageNet \citep{deng2009imagenet} (in-distribution) using an ensemble of pretrained models and PACS \citep{li2017deeper} (distribution shift) where we fine-tune ImageNet models on PACS' `photo' domain, which is close to ImageNet as source domain, and evaluate it on PACS' `art painting', `sketch', and `cartoon' domains. We use all three domains together for distribution shift evaluation to have more samples for the rejection plots.

\subsection{Experiment Setup}
\label{jiang:appsec:experiment_setup}

We use PyTorch \citep{pytorchgithub} for all experiments.

\textbf{CIFAR-10 and CINIC-10.} We follow the training setup from \citet{mukhoti2021deterministic}: we train 25 WideResNet-28-10 models \citep{zagoruyko2016wide} for 350 epochs on CIFAR-10. We use SGD with a learning rate of $0.1$ and momentum of $0.9$. We use a learning rate schedule with a decay of 10 at 150 and 250 epochs.

\textbf{ImageNet and PACS.} We use pretrained models with various architectures (specifically: ResNet-152-D \citep{he2018bag}, BEiT-L/16 \citep{bao2021beit}, ConvNext-L \citep{liu2022}, DeiT3-L/16 \citep{touvron2020training}, and ViT-B/16 \citep{dosovitskiy2020image}) from the timm package \citep{rw2019timm} as base models. We freeze all weights except for the final linear layer, which we fine-tune on PACS' `photo` domain using Adam \citep{kingma2014adam} with learning rate $5\times10^{-3}$ and batch size 128 for 1000 steps. We then build an ensemble using these different models.

\subsection{Additional Results}
\label{jiang:appsec:additional_results}

\begin{figure}
    \centering
    \begin{subfigure}{\linewidth}
        \vspace{-0.4em}
        \centering
        \includegraphics[width=\linewidth,trim={5cm 0 0 11.5cm},clip]{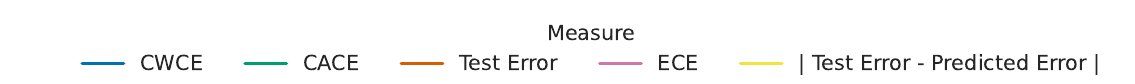} 
    \end{subfigure}
    \begin{subfigure}{\linewidth}
        \vspace{0.1em}
        \caption*{\textbf{ImageNet (In-Distribution)}}
        \vspace{0.1em}
        \begin{subfigure}{0.52\linewidth}
            \centering
            \includegraphics[width=\linewidth,trim={0 1.5cm 11.5cm 0},clip]{plots/imagenet_rejection_plot_disrate.pdf}
        \end{subfigure}
        \begin{subfigure}{0.46\linewidth}
            \centering
            \includegraphics[width=\linewidth,trim={13cm 1.5cm 0 0},clip]{plots/imagenet_rejection_plot_disrate.pdf}
        \end{subfigure}
    \end{subfigure}
    \begin{subfigure}{\linewidth}
        \vspace{0.1em}
        \caption*{\textbf{PACS (Distribution Shift)}}
        \vspace{0.1em}
        \begin{subfigure}{0.52\linewidth}
            \centering
            \includegraphics[width=\linewidth,trim={0 1.5cm 11.5cm 0},clip]{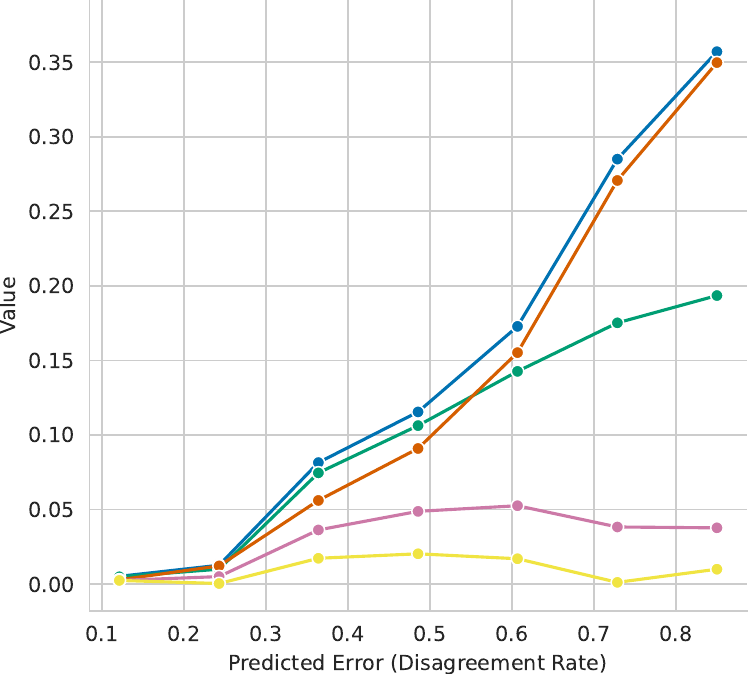}
            \subcaption{Ensemble with TOP}    
        \end{subfigure}
        \begin{subfigure}{0.46\linewidth}
            \centering
            \includegraphics[width=\linewidth,trim={13cm 1.5cm 0 0},clip]{plots/pacs_cartoon_sketch_rejection_plot_disrate.pdf}
            \subcaption{Ensemble without TOP (same underlying models)}    
        \end{subfigure}
    \end{subfigure}
    \caption{\emph{Rejection Plot of Calibration Metrics for Increasing Disagreement In-Distribution (ImageNet) and Under Distribution Shift (PACS `photo` domain $\rightarrow$ other domains).} Different calibration metrics ($\ECE$, $\CWCE$, $\CACE$) vary across ImageNet and PACS' `art painting', `cartoon', and `sketch' domains across an ensemble of 5 models trained on ImageNet and 25 models fine-tuned on PACS' `photo' domain, depending on the rejection threshold of the predicted error (disagreement rate). Again, calibration cannot be assumed constant for in-distribution data or under distribution shift. The mean predicted error (disagreement rate) is shown on the x-axis. \textbf{(a)} shows results for an ensemble using TOP (following \citet{jiang2021assessing}), and \textbf{(b)} for a regular deep ensemble without TOP. Details in \S\ref{jiang:appsec:additional_results}.
    }
    \label{jiang:appfig:rejection_plot_imagenet_pacs}
\end{figure}

In \Cref{jiang:appfig:rejection_plot_imagenet_pacs}, we see that for ImageNet and PACS, the calibration metrics behave like for CIFAR-10 and CINIC-10, matching the described behavior in the main text. We use 5 models from each of the enumerated architectures to build an ensemble of 25 models. Individual architectures also behave as expected as we ablate in \Cref{jiang:appfig:rejection_plot_imagenet_pacs_archs}.

Additionally, in \Cref{jiang:appfig:rejection_plot_infogain} and \Cref{jiang:appfig:rejection_plot_imagenet_pacs_infogain}, we also show rejection plots using the Expected Information Gain/BALD for thresholding. We observe similar trajectories. Comparing these results with \Cref{jiang:fig:rejection_plot} and \Cref{jiang:appfig:rejection_plot_imagenet_pacs}, we see that both the predicted error and the Bayesian metric behave similarly. We hypothesize that this could be because the datasets only contain few samples with high aleatoric uncertainty (e.g.\ noise), which would otherwise act as confounder \citep{mukhoti2021deterministic}. See also the discussion in \S\ref{jiang:sec:discussion}.

\begin{figure}
    \centering
    \begin{subfigure}{\linewidth}
        \centering
        \includegraphics[width=\linewidth,trim={5cm 0 0 11.5cm},clip]{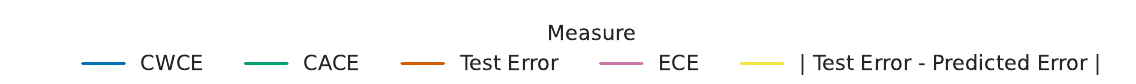} 
    \end{subfigure}
    \begin{subfigure}{\linewidth}
        \vspace{0.1em}
        \caption*{\textbf{CIFAR-10 (In-distribution)}}
        \vspace{0.1em}
        \begin{subfigure}{0.52\linewidth}
            \centering
            \includegraphics[width=\linewidth,trim={0 1.5cm 11.5cm 0},clip]{plots/cifar10_test_rejection_plot_infogain.pdf}
        \end{subfigure}
        \begin{subfigure}{0.46\linewidth}
            \centering
            \includegraphics[width=\linewidth,trim={13cm 1.5cm 0 0},clip]{plots/cifar10_test_rejection_plot_infogain.pdf}
        \end{subfigure}
    \end{subfigure}
    \begin{subfigure}{\linewidth}
        \vspace{0.1em}
        \caption*{\textbf{CINIC-10 (Distribution Shift)}}
        \vspace{0.1em}
        \begin{subfigure}{0.52\linewidth}
            \centering
            \includegraphics[width=\linewidth,trim={0 1.5cm 11.5cm 0},clip]{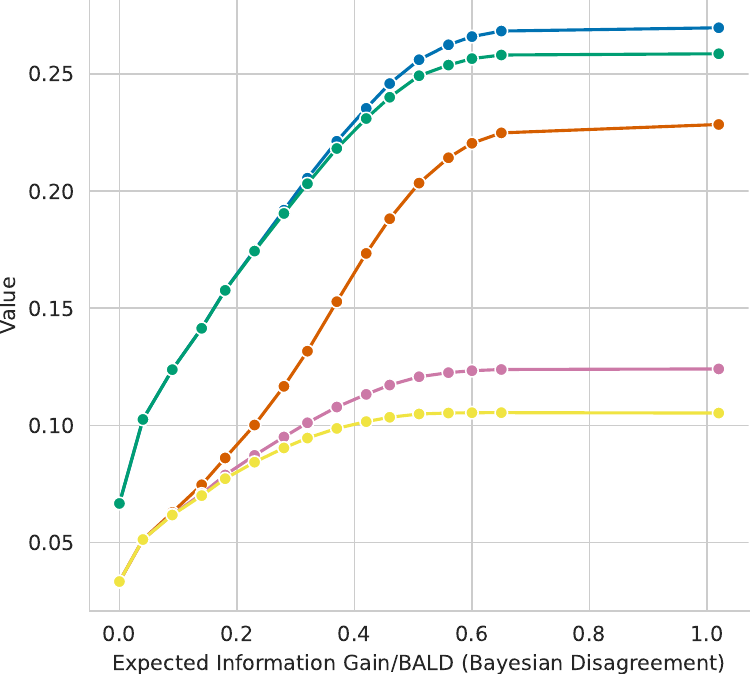}
            \subcaption{Ensemble with TOP}    
        \end{subfigure}
        \begin{subfigure}{0.46\linewidth}
            \centering
            \includegraphics[width=\linewidth,trim={13cm 1.5cm 0 0},clip]{plots/cinic10_test_rejection_plot_infogain.pdf}
            \subcaption{Ensemble without TOP (same underlying models)}    
        \end{subfigure}
    \end{subfigure}
    \caption{\emph{Rejection Plot of Calibration Metrics for Increasing Bayesian Disagreement In-Distribution (CIFAR-10) and Under Distribution Shift (CINIC-10).} Different calibration metrics ($\ECE$, $\CWCE$, $\CACE$) vary across CIFAR-10 and CINIC-10, depending on the rejection threshold of Bayesian disagreement (Expected Information Gain/BALD). The trajectory matches the one for prediction disagreement. We hypothesize this is because there are few noisy samples in the dataset which would act as a confounder for prediction disagreement otherwise. 
    Details in \S\ref{jiang:appsec:additional_results}.
    }
    \label{jiang:appfig:rejection_plot_infogain}
\end{figure}

\begin{figure}
    \centering
    \begin{subfigure}{\linewidth}
        \centering
        \includegraphics[width=\linewidth,trim={5cm 0 0 11.5cm},clip]{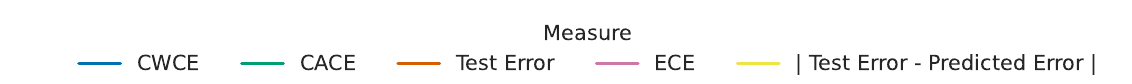} 
    \end{subfigure}
    \begin{subfigure}{\linewidth}
        \vspace{0.1em}
        \caption*{\textbf{ImageNet (In-distribution)}}
        \vspace{0.1em}
        \begin{subfigure}{0.52\linewidth}
            \centering
            \includegraphics[width=\linewidth,trim={0 1.5cm 11.5cm 0},clip]{plots/imagenet_rejection_plot_infogain.pdf}
        \end{subfigure}
        \begin{subfigure}{0.46\linewidth}
            \centering
            \includegraphics[width=\linewidth,trim={13cm 1.5cm 0 0},clip]{plots/imagenet_rejection_plot_infogain.pdf}
        \end{subfigure}
    \end{subfigure}
    \begin{subfigure}{\linewidth}
        \vspace{0.1em}
        \caption*{\textbf{PACS (Distribution Shift)}}
        \vspace{0.1em}
        \begin{subfigure}{0.52\linewidth}
            \centering
            \includegraphics[width=\linewidth,trim={0 1.5cm 11.5cm 0},clip]{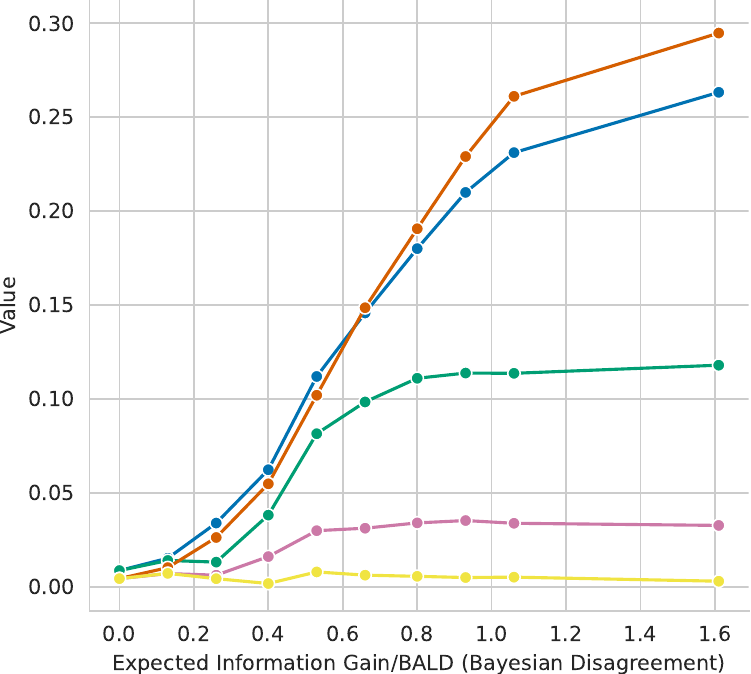}
            \subcaption{Ensemble with TOP}    
        \end{subfigure}
        \begin{subfigure}{0.46\linewidth}
            \centering
            \includegraphics[width=\linewidth,trim={13cm 1.5cm 0 0},clip]{plots/pacs_cartoon_sketch_rejection_plot_infogain.pdf}
            \subcaption{Ensemble without TOP (same underlying models)}    
        \end{subfigure}
    \end{subfigure}
    \caption{\emph{Rejection Plot of Calibration Metrics for Increasing Bayesian Disagreement In-Distribution (CIFAR-10) and Under Distribution Shift (CINIC-10).} Different calibration metrics ($\ECE$, $\CWCE$, $\CACE$) vary across CIFAR-10 and CINIC-10, depending on the rejection threshold of Bayesian disagreement (Expected Information Gain/BALD). The trajectory matches the one for prediction disagreement. We hypothesize this is because there are few noisy samples in the dataset which would act as a confounder for prediction disagreement otherwise. 
    Details in \S\ref{jiang:appsec:additional_results}.
    }
    \label{jiang:appfig:rejection_plot_imagenet_pacs_infogain}
\end{figure}

\begin{figure}
    \centering
    \begin{subfigure}{\linewidth}
        \centering
        \includegraphics[width=\linewidth,trim={5cm 0 0 11.5cm},clip]{plots/imagenet_rejection_plot_disrate.pdf} 
    \end{subfigure}
    \begin{subfigure}{\linewidth}
        \vspace{0.1em}
        \caption*{\textbf{PACS, BeiT-L/16}}
        \vspace{0.1em}
        \begin{subfigure}{0.52\linewidth}
            \centering
            \includegraphics[width=\linewidth,trim={0 1.5cm 11.5cm 0},clip]{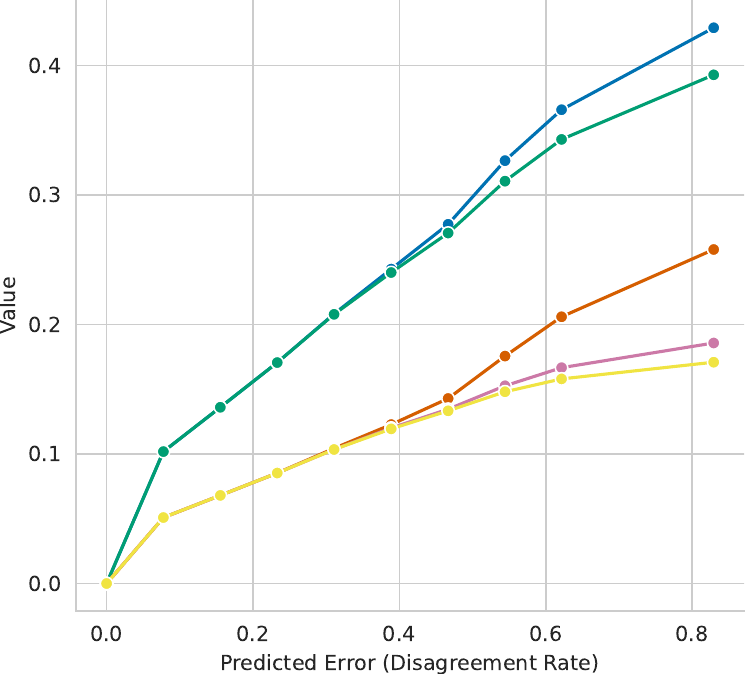}
        \end{subfigure}
        \begin{subfigure}{0.46\linewidth}
            \centering
            \includegraphics[width=\linewidth,trim={13cm 1.5cm 0 0},clip]{plots/pacs_cartoon_sketch_beit_large_patch16_224_rejection_plot_disrate.pdf}
        \end{subfigure}
    \end{subfigure}
    \begin{subfigure}{\linewidth}
        \vspace{0.1em}
        \caption*{\textbf{PACS, ResNet-152-D}}
        \vspace{0.1em}
        \begin{subfigure}{0.52\linewidth}
            \centering
            \includegraphics[width=\linewidth,trim={0 1.5cm 11.5cm 0},clip]{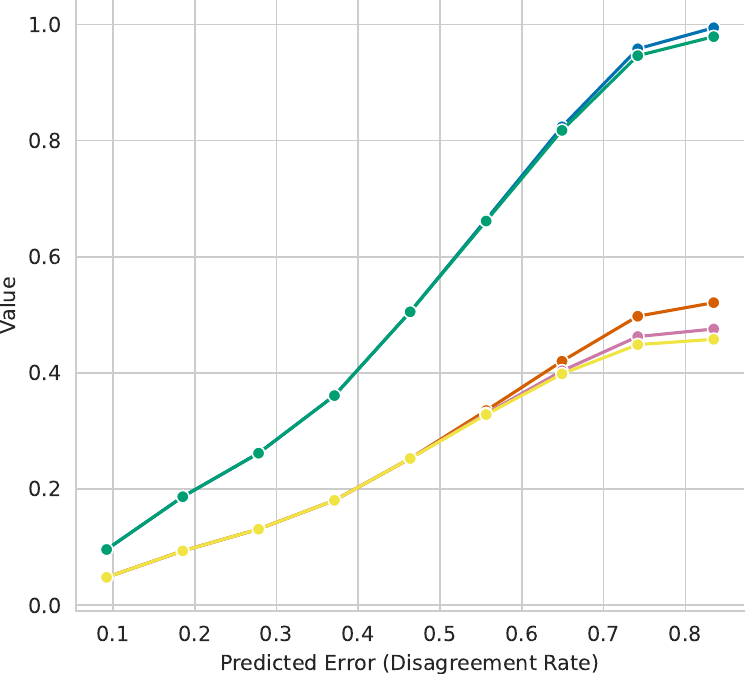}
            \subcaption{Ensemble with TOP}    
        \end{subfigure}
        \begin{subfigure}{0.46\linewidth}
            \centering
            \includegraphics[width=\linewidth,trim={13cm 1.5cm 0 0},clip]{plots/pacs_cartoon_sketch_resnet152d_rejection_plot_disrate.pdf}
            \subcaption{Ensemble without TOP (same underlying models)}    
        \end{subfigure}
    \end{subfigure}
    \caption{\emph{Rejection Plot of Calibration Metrics for Increasing Disagreement Under Distribution Shift (PACS `photo` domain $\rightarrow$ other domains) for Specific Model Architectures.} We use the same encoder weights and evaluate on an ensemble of 5 models, which were last-layer fine-tuned on PACS. We show ResNet-152-D and BeiT-L/16. Details in \S\ref{jiang:appsec:additional_results}.
    }
    \label{jiang:appfig:rejection_plot_imagenet_pacs_archs}
\end{figure}

	}

	{
		
		\chapter{\ddudmtitle}
		\label{ddu:app:theory}
		How well can these distributions capture the outputs of an ensemble?
Several works have examined distilling the uncertainty of deep ensembles into a single model \citep{malinin2019ensemble,fathullah2021ensemble,ryabinin2021scaling} or using Dirichlet distributions to model epistemic uncertainty \citep{malinin2018predictive,malinin2019reverse,hammam2022predictive}. But 
In this chapter, we qualitatively examine how well Dirichlet distributions can capture predictions of deep ensembles in computer vision. 

Concretely, we examine how well they can approximate the variance of softmax entropies when matching the predictions and epistemic uncertainty of input samples on OoD data (which is arguably more difficult than when trying to capture uncertainty of iD which will have low epistemic uncertainty). This is easier to visualize for datasets than individual predictions.

We qualitatively evaluate deep ensembles across different model architectures and qualitatively find that the modelled Dirichlet distributions provide more concentrated predictions. Hence, samples from them are also unlikely to model the actual individual predictions of a deep ensemble well. This does not invalidate any of the results in the previous chapters, but it does suggest that the Dirichlet distribution is not a good approximation of the posterior predictive distribution of the ensemble members.

\section{Methodology}

There are two interpretations of the ensemble parameter distribution $\pof{\omega \given \Dtrain}$: 
\begin{enumerate}
    \item we can view it as an empirical distribution given a specific ensemble with members $\omega_{i\in\{1,\ldots,K\}}$, or
    \item we can view it as a distribution over all possible trained models which depends on random weight initializations, the dataset, stochasticity in the minibatches, and the optimization process.
\end{enumerate}
In the latter case, any deep ensemble with $K$ members can be seen as finite Monte-Carlo sample of this posterior distribution.
The predictions of an ensemble then are an unbiased estimate of the predictive distribution $\E{\pof{\omega \given \Dtrain}}{\pof{\y \given \x,\omega}}$, and similarly the expected information gain computed using the members of the deep ensemble is just a (biased) estimator of $\MIof{Y; \Omega \given \x, \Dtrain}$.

\newcommand{\infogain}{\MIof{Y; \Omega \given x, \Dtrain}}
\newcommand{\expectedsment}{\simpleE{\omega \given \Dtrain}{\Hof{Y \given x,\omega}}}
\newcommand{\preddis}{\pof{y \given x, \Dtrain}}
\newcommand{\predent}{\Hof{Y \given x, \Dtrain}}
\newcommand{\smdis}{\pof{y \given x,\omega}}
\newcommand{\sment}{\Hof{Y \given x,\omega}}
\newcommand{\Dir}{\operatorname{Dir}}
\newcommand{\Cat}{\operatorname{Cat}}
\newcommand{\dirp}{\mathbf{p}}

\begin{figure}[t]
    \centering
    \begin{subfigure}{0.5\linewidth}
        \centering
        \includegraphics[width=\linewidth]{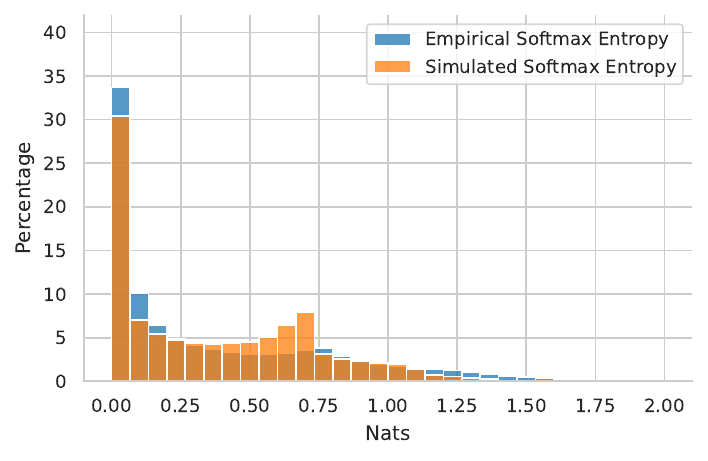}
        \caption{SVHN (OoD): VGG16}
    \end{subfigure}\hfill
    \begin{subfigure}{0.5\linewidth}
        \centering
        \includegraphics[width=\linewidth]{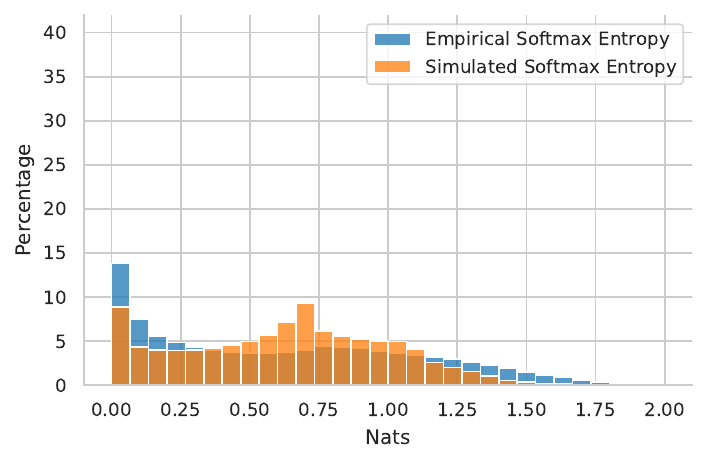}
        \caption*{WideResNet-28-10+SN}
    \end{subfigure}
    \begin{subfigure}{0.5\linewidth}
        \centering
        \includegraphics[width=\linewidth]{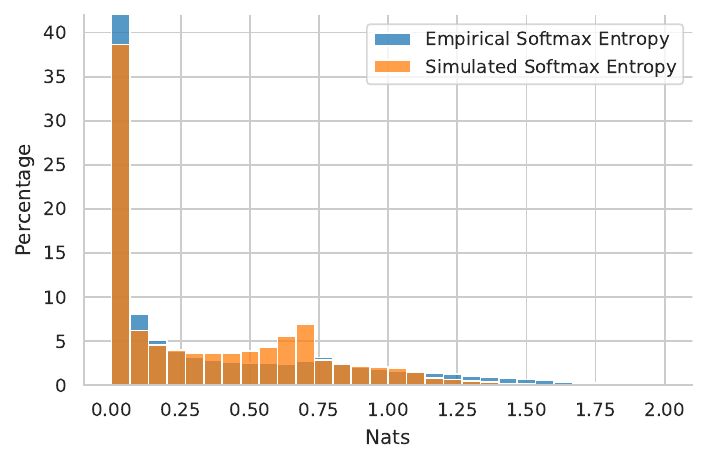}
        \caption{CIFAR-100 (OoD): VGG16}
    \end{subfigure}\hfill
    \begin{subfigure}{0.5\linewidth}
        \centering
        \includegraphics[width=\linewidth]{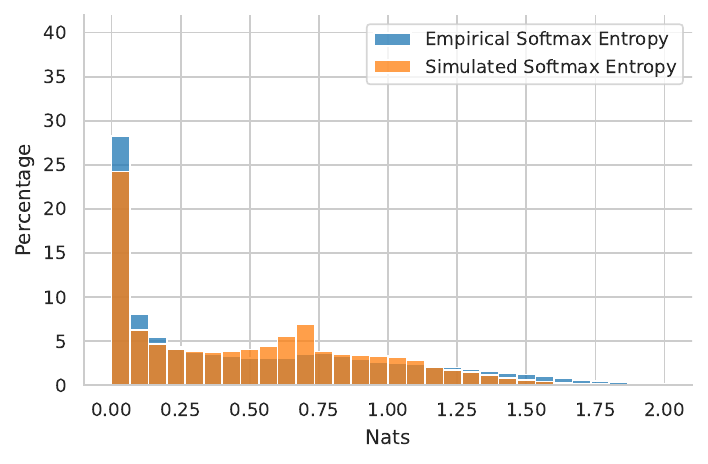}
        \caption*{WideResNet-28-10+SN}
    \end{subfigure} 
    \caption{
    \emph{Simulated vs Empirical Softmax Entropy and Predictive Entropy.} WideResNet-28-10+SN and VGG16 trained on CIFAR-10 (25 models). Although we use a simple Dirichlet model, sampling from the fitted Dirichlet distributions does approximate the empirical entropy distribution of softmax entropies well. 
    }
    \label{ddu:fig:vgg16_vs_wideresnet_sm}
\end{figure}

\begin{figure}[t]
    \centering
    \begin{subfigure}{\linewidth}
        \centering
        \includegraphics[trim=0 50 0 0,clip,width=0.8\linewidth]{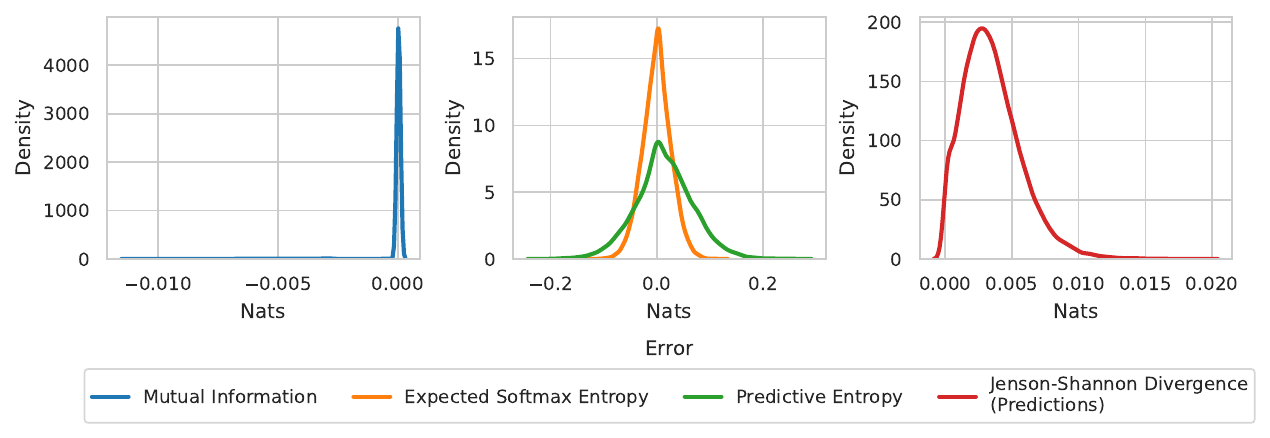}
        \caption*{SVHN: VGG16}
    \end{subfigure}\hfill
    \begin{subfigure}{\linewidth}
        \centering
        \includegraphics[trim=0 50 0 0,clip,width=0.8\linewidth]{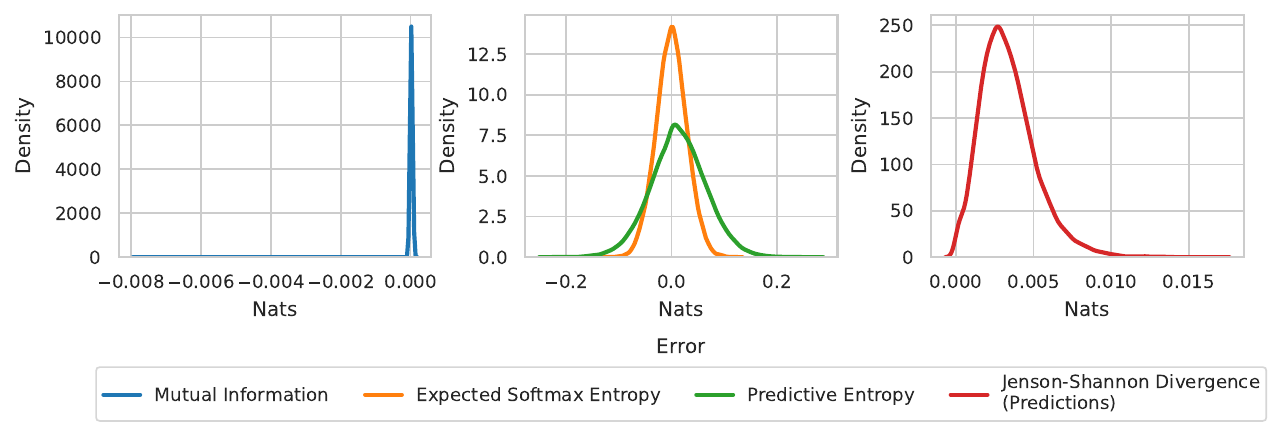}
        \caption*{WideResNet-28-10+SN}
    \end{subfigure}
    \begin{subfigure}{\linewidth}
        \centering
        \includegraphics[trim=0 50 0 0,clip,width=0.8\linewidth]{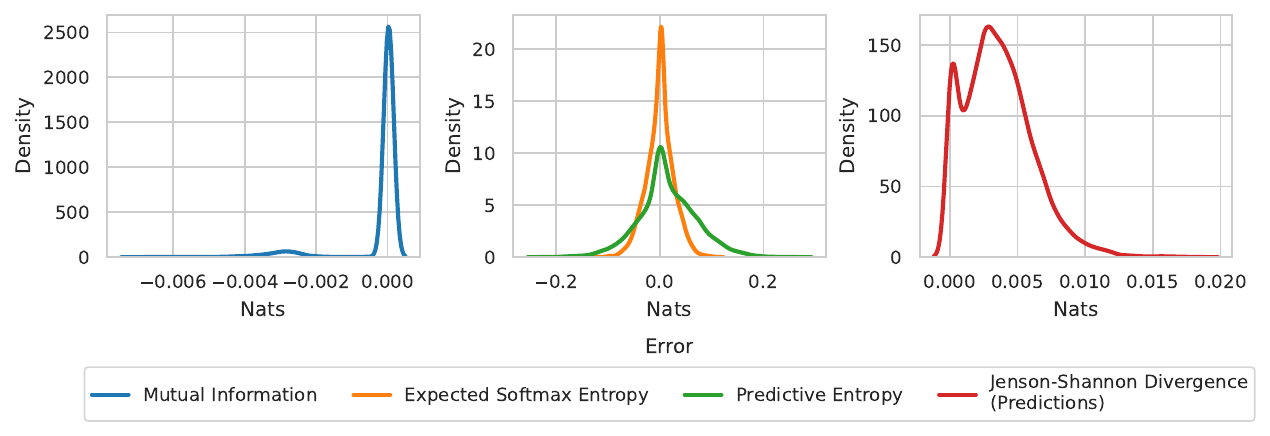}
        \caption*{CIFAR-100: VGG16}
    \end{subfigure}\hfill
    \begin{subfigure}{\linewidth}
        \centering
        \includegraphics[width=0.8\linewidth]{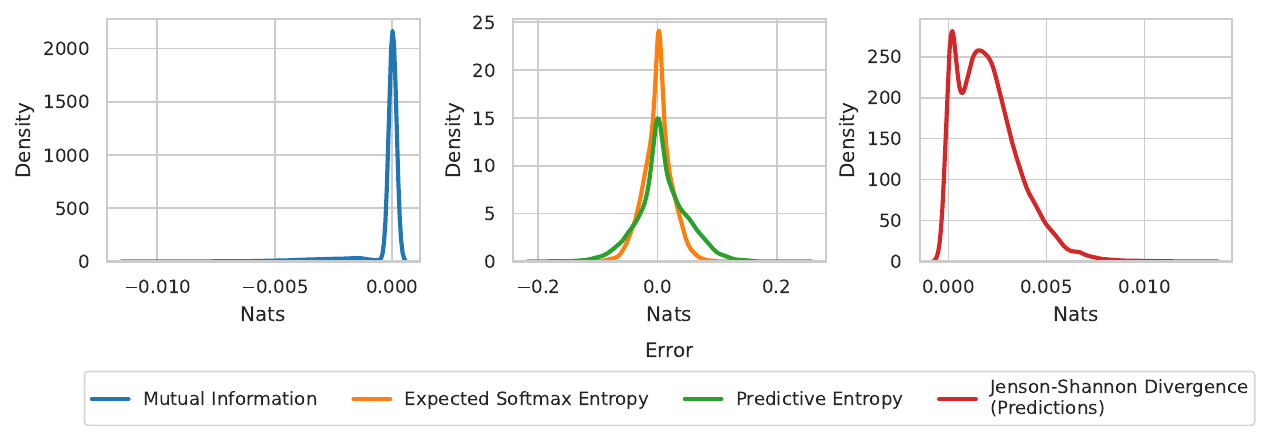}
        \caption*{WideResNet-28-10+SN}
    \end{subfigure} 
    \caption{
    \emph{Simulated Quantities (via Dirichlet Distributions) vs Empirical Quantities.} WideResNet-28-10+SN and VGG16 trained on CIFAR-10 (25 models). Although we use a simple Dirichlet model, sampling from the fitted Dirichlet distributions does approximate the empirical entropy distribution of softmax entropies well. 
    }
    \label{ddu:fig:error_signals}
\end{figure}

\paragraph{Inverse Problem.}
Based on this interpretation of deep ensembles as a distribution over model parameters, we can look at the following inverse problem: 
given \emph{some value} for the predictive distribution and epistemic uncertainty of a deep ensemble, estimate what the softmax entropies from each ensemble component must have been. That is if we observe deep ensembles to have high epistemic uncertainty on (near) OoD data, we can deduce from that what the distribution of softmax entropy of deterministic neural nets (the ensemble members) ought to look like.

That is, given a predictive distribution $\preddis$ and epistemic uncertainty $\infogain$ (expected information gain) of the deep ensemble, we can observe the softmax entropy $\sment$ as a random variable of a single deterministic model, $\omega \sim \pof{\omega \given \Dtrain}$, and estimate its variance $\Var{\omega \given \Dtrain}{\sment}$. 

Empirically, we find the real variance to be higher by a large amount for OoD samples, showing that softmax entropies do not capture epistemic uncertainty well for samples with high epistemic uncertainty. 

We will need to make several strong assumptions that limit the generality of our estimation, but we can show that our analysis models the resulting softmax entropy distributions appropriately. This will show that deterministic softmax models can have widely different entropies and confidence values.

\paragraph{Approximate Model.}
For a \emph{fixed} $\x$, we approximate the distribution over softmax probability vectors $\pof{\y \given \x,\omega}$ for different $\omega$ using a Dirichlet distribution $\dirp \sim \Dir(\alpha)$ with non-negative concentration parameters $\alpha = (\alpha_1, \ldots, \alpha_K)$ and $\alpha_0 := \sum \alpha_i$. Note that we only use the Dirichlet distribution \emph{as an analysis tool}.

Concretely, for a distribution over models $\pof{\omega \given \Dtrain}$, and a sample $\x$, we obtain $\preddis$, and $\infogain$. 
We use moment matching with these two quantities to fit a Dirichlet distribution $\dirp \sim \Dir(\alpha)$ on $\smdis$ over $\Omega$, which satisfies:
\begin{align}
    \preddis &= \frac{\alpha_i}{\alpha_0} \\
    \predent - \infogain &= \psi(\alpha_0+1) - \sum_{y=1}^K \pof{y \given x} \psi(\alpha_0 \,\pof{y \given x} +1).
\end{align}
Then, we can model the softmax distribution as given in eq.~\eqref{ddu:app:thm:smentvar}. The details and proofs can be found below in \S\ref{ddu:app:app_softmax_theory}.

\section{Qualitative Empirical Validation}

\begin{figure}[t]
    \centering
    \begin{subfigure}{0.5\linewidth}
        \centering
        \includegraphics[width=\linewidth]{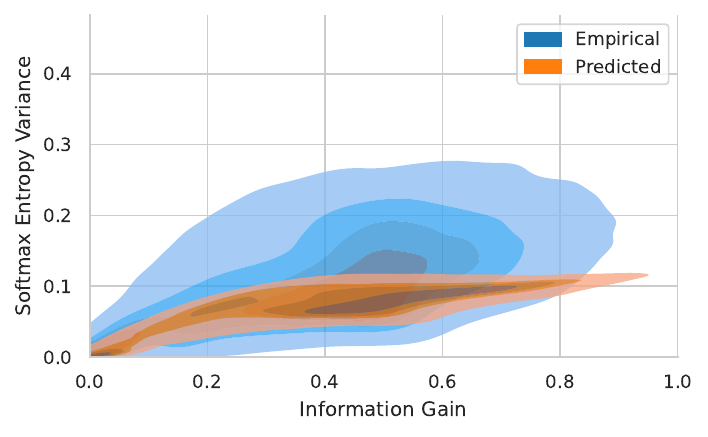}
        \caption*{VGG16}
    \end{subfigure}\hfill
    \begin{subfigure}{0.5\linewidth}
        \centering
        \includegraphics[width=\linewidth]{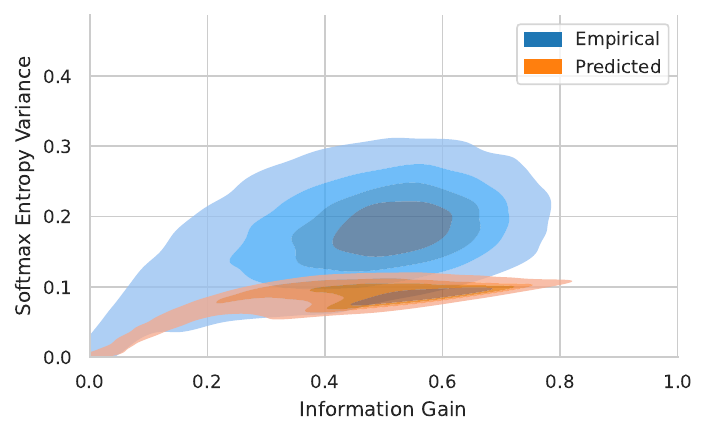}
        \caption{SVHN: WideResNet-28-10+SN}
    \end{subfigure}
    \begin{subfigure}{0.5\linewidth}
        \centering
        \includegraphics[width=\linewidth]{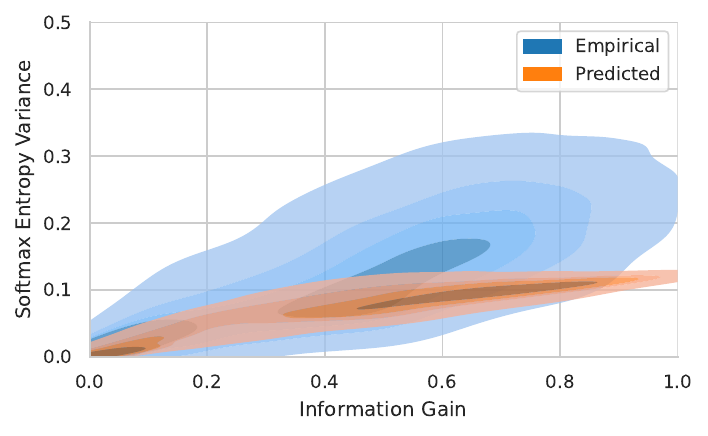}
        \caption*{VGG16}
    \end{subfigure}\hfill
    \begin{subfigure}{0.5\linewidth}
        \centering
        \includegraphics[width=\linewidth]{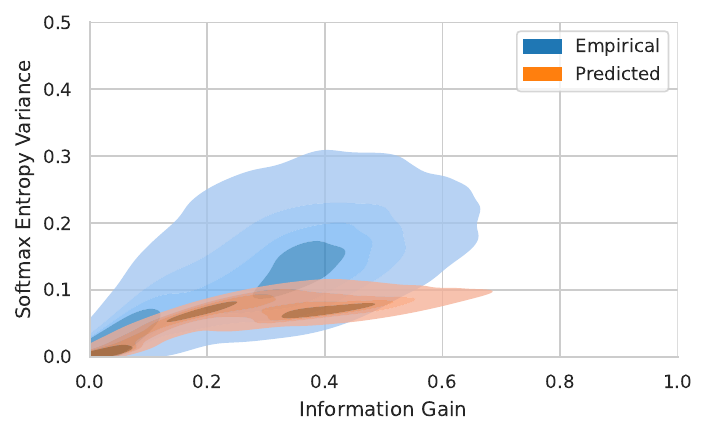}
        \caption{CIFAR-100: WideResNet-28-10+SN}
    \end{subfigure} 
    \caption{
    \emph{Simulated \&\ empirical softmax entropy vs mutual information (EIG) on WideResNet-28-10+SN and VGG16.} Although we use a simple Dirichlet model, samples from the fitted Dirichlet distributions  approximate the three major information quantities and the BMA of the ensemble predictions very well.
    }
    \label{ddu:fig:contour_eig_vs_psv_esv}
\end{figure}

\begin{figure}[t]
    \centering
    \begin{subfigure}{0.5\linewidth}
        \centering
        \includegraphics[width=\linewidth]{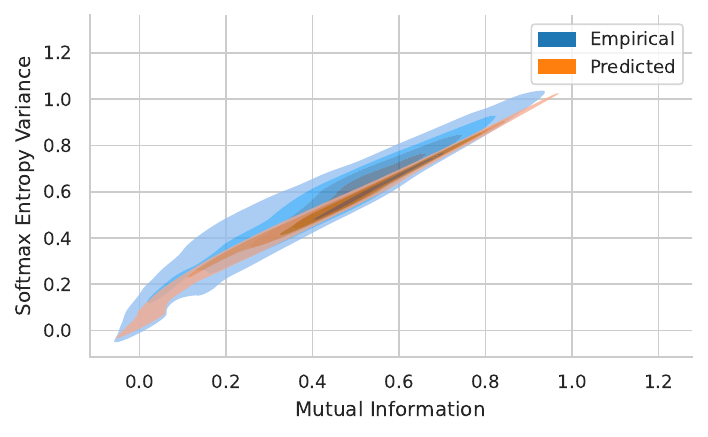}
        \caption{SVHN: VGG16}
    \end{subfigure}\hfill
    \begin{subfigure}{0.5\linewidth}
        \centering
        \includegraphics[width=\linewidth]{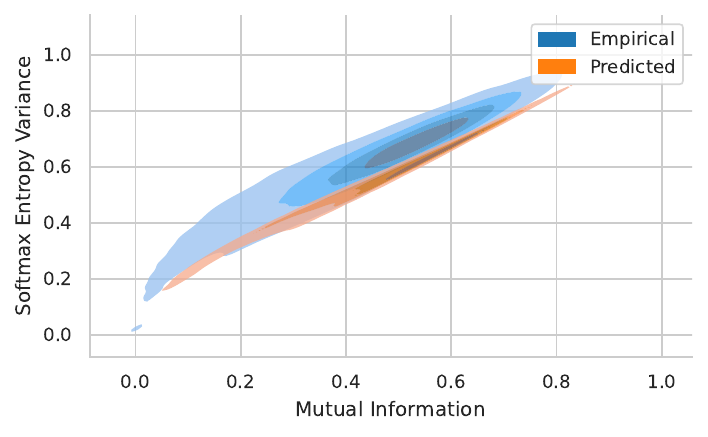}
        \caption*{WideResNet-28-10+SN}
    \end{subfigure}
    \begin{subfigure}{0.5\linewidth}
        \centering
        \includegraphics[width=\linewidth]{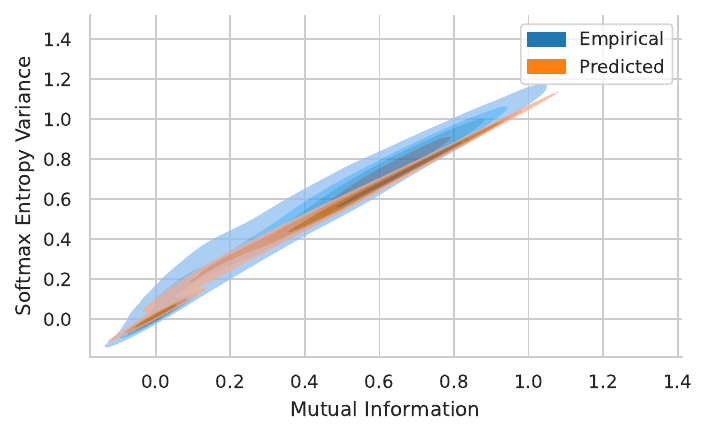}
    \end{subfigure}\hfill
    \begin{subfigure}{0.5\linewidth}
        \centering
        \includegraphics[width=\linewidth]{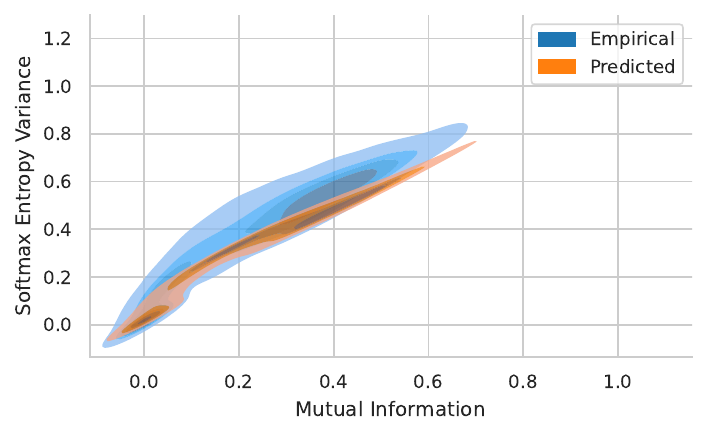}
    \end{subfigure} 
    \caption{
    \emph{RMSE for simulated \&\ empirical softmax entropy vs mutual information (EIG) on WideResNet-28-10+SN and VGG16.}
    }
    \label{ddu:fig:contour_eig_vs_rmse}
\end{figure}

We train two deep ensembles of VGG and WideResNet-28-10+SN models (25 members each) on CIFAR-10 and compute the predictive entropy, mutual information, and softmax entropies for each sample in SVHN and CIFAR-100, which we use as OoD distribution. 
Then we fit a Dirichlet distribution on the softmax entropies of the ensemble members and use the fitted distribution to simulate the softmax entropies for the OoD samples. We fit the Dirichlet distribution on the BMA prediction of the ensemble and the corresponding mutual information (epistemic uncertainty).

We have already empirically verified that softmax entropies vary considerably in \Cref{ddu:sec:predictive_entropy_not_epistemic_uncertainty_experiments}.
The distribution of actual softmax entropies and predicted softmax entropies seems to be close in the histograms of \Cref{ddu:fig:vgg16_vs_wideresnet_sm}, and the overall error terms for the three information quantities as well as the Jensen-Shannon Divergence between the simulated BMA predictions and real ones are small \Cref{ddu:fig:error_signals}.
Yet, \Cref{ddu:fig:contour_eig_vs_psv_esv,ddu:fig:contour_eig_vs_rmse} show that the distributions are quite different in terms of softmax entropy variance and RMSE (for the softmax entropies as estimates of the respective predictive entropy).

\section{Discussion}

Hence, we can conclude without evaluating the simulated predictions themselves in more detail that the Dirichlet model is not a very good fit for the empirical distribution of softmax entropies from individual members of a deep ensemble.
This does not invalidate that these approximations have great and useful value on downstream tasks, but it raises the question for future work to find distributions that can better capture the empirical distribution of softmax entropies from individual members of a deep ensemble.

\section{Details}
\label{ddu:app:app_softmax_theory}
Based on the interpretation of deep ensembles as a distribution over model parameters, we can walk backwards and, given \emph{some value} for the predictive distribution and epistemic uncertainty of a deep ensemble, estimate what the softmax entropies from each ensemble component must have been. I.e.\ if we observe deep ensembles to have high epistemic uncertainty on OoD data, we can deduce from that what the softmax entropy of deterministic neural nets (the ensemble components) must look like.
More specifically, given a predictive distribution $\preddis$ and epistemic uncertainty, that is expected information gain $\infogain$, of the infinite deep ensemble, we estimate the expected softmax entropy from a single deterministic model, considered as a sample $\omega \sim \pof{\omega \given \Dtrain}$ and model the variance. Empirically, we find the real variance to be higher by a large amount for OoD samples, showing that softmax entropies do not capture epistemic uncertainty well for samples with high epistemic uncertainty. 
We will need to make several strong assumptions that limit the generality of our estimation. 

Given the predictive distribution $\preddis$ and epistemic uncertainty $\infogain$, we can approximate the distribution over softmax probability vectors $\pof{\y \given \x,\omega}$ for different $\omega$ using its  maximum-entropy estimate: a Dirichlet distribution $\left(Y_{1}, \ldots, Y_{K}\right) \sim \Dir(\alpha)$ with non-negative concentration parameters $\alpha = (\alpha_1, \ldots, \alpha_K)$ and $\alpha_0 := \sum \alpha_i$. Note that the Dirichlet distribution is used \emph{only as an analysis tool}, and at no point do we need to actually fit Dirichlet distributions to our data.

\paragraph{Preliminaries.} 
Before we can establish our main result, we need to look more closely at Dirichlet-Multinomial distributions. Given a Dirichlet distribution $\Dir(\alpha)$ and a random variable $\dirp \sim \Dir(\alpha)$, we want to quantify the expected entropy $\simpleE{\dirp \sim \Dir(\alpha)} {\varHof{Y\sim \Cat(\dirp)}{Y}}$ and its variance $\Var{\dirp \sim \Dir(\alpha)}{\varHof{Y\sim \Cat(\dirp)}{Y}}$. For this, we need to develop more theory. In the following, $\Gamma$ denotes the Gamma function, $\psi$ denotes the Digamma function, $\psi'$ denotes the Trigamma function.
\begin{lemma}
\label{ddu:lemma:dirichlet_basics}
\begin{enumerate}[wide, labelwidth=!, labelindent=0pt]
Given a Dirichlet distribution and random variable $\dirp\sim\Dir(\alpha)$, the following hold:
\item The expectation $\E{}{\log \dirp_i}$ is given by:
\begin{align}
    \label{ddu:eq:psi_log_expectation}
    \E{}{\log \dirp_i} = \psi(\alpha_i) - \psi(\alpha_0).
\end{align}
\item 
The covariance $\implicitCov{\log \dirp_i, \log \dirp_j}$ is given by
\begin{align}
    \label{ddu:eq:psip_log_expectation}
    \implicitCov{\log \dirp_i,\log \dirp_j} = \psi'(\alpha_i) \, \delta_{ij} - \psi'(\alpha_0).
\end{align}
\item The expectation $\E{}{\dirp_i^n \dirp_j^m \log \dirp_i}$ is given by:
\begin{align}
\begin{split}
\label{ddu:eq:nm_log_expectation}
\MoveEqLeft[3] \E{}{\dirp_i^n \dirp_j^m \log \dirp_i} \\
={}& \frac{\alpha_i^{\overline{n}} \, \alpha_j^{\overline{m}}}{\alpha_0^{\overline{n+m}}} \left ( \psi(\alpha_i + n) - \psi(\alpha_0 + n + m) \right ),
\end{split}
\end{align}
where $i \not =j$, and $n^{\overline{k}}=n \, (n+1) \, \ldots \, (n+k-1)$ denotes the rising factorial.
\end{enumerate}
\end{lemma}
\begin{proof}
\begin{enumerate}[wide, labelwidth=!, labelindent=0pt]
\item
The Dirichlet distribution is members of the exponential family. Therefore, the moments of the sufficient statistics are given by the derivatives of the partition function with respect to the natural parameters.
The natural parameters of the Dirichlet distribution are just its concentration parameters $\alpha_i$. The partition function is
\begin{align}
    A(\alpha) = \sum_{i=1}^{k} \log \Gamma\left(\alpha_{i}\right)-\log \Gamma\left(\alpha_{0}\right),
\end{align}
the sufficient statistics is $T(\x) = \log \x$, and the expectation $\E{}{T}$ is given by
\begin{align}
    \E{}{T_i} = \frac{\partial A(\alpha)}{\partial \alpha_{i}}
\end{align}
as the Dirichlet distribution is a member of the exponential family.
Substituting the definitions and evaluating the partial derivative yields
\begin{align}
    \E{}{\log \dirp_i} &= \frac{\partial}{\partial \alpha_{i}} \left [ \sum_{i=1}^{k} \log \Gamma\left(\alpha_{i}\right)-\log \Gamma\left(\sum_{i=1}^{k} \alpha_i\right) \right ] \\
    & = \psi\left(\alpha_{i}\right) - \psi\left(\alpha_{0}\right) \frac{\partial}{\partial \alpha_{i}} \alpha_0,
\end{align}
where we have used that the Digamma function $\psi$ is the log derivative of the Gamma function $\psi(\x) = \frac{d}{dx} \ln \Gamma(\x)$.
This proves \eqref{ddu:eq:psi_log_expectation} as $\frac{\partial}{\partial \alpha_{i}} \alpha_0 = 1$.
\item Similarly, the covariance is obtained using a second-order partial derivative:
\begin{align}
    \implicitCov{T_i, T_j} = \frac{\partial^2 A(\alpha)}{\partial \alpha_{i} \, \partial \alpha_{i}}.
\end{align}
Again, substituting yields
\begin{align}
    \implicitCov{\log \dirp_i, \log \dirp_j} &= \frac{\partial}{\partial \alpha_{j}} \left [ \psi\left(\alpha_{i}\right) - \psi\left(\alpha_{0}\right) \right ] \\
    &= \psi'\left(\alpha_{i}\right) \delta_{ij} - \psi'\left(\alpha_{0}\right).
\end{align}
\item
We will make use of a simple reparameterization to prove the statement using \Cref{ddu:eq:psi_log_expectation}.
Expanding the expectation and substituting the density $\Dir(\dirp; \alpha)$, we obtain
\begin{align}
    & \E{}{\dirp_i^n \dirp_j^m \log \dirp_i} = \int \Dir(\dirp; \alpha) \, \dirp_i^n \, \dirp_j^m \, \log \dirp_i \, d\dirp \\
    & \quad = \int \frac{\Gamma\left(\alpha_0\right)}{\prod_{i=1}^{K} \Gamma\left(\alpha_{i}\right)} \prod_{k=1}^{K} \dirp_{k}^{\alpha_{k}-1}  \, \dirp_i^n \, \dirp_j^m \, \log \dirp_i \, d\dirp \\
    \begin{split}
    & \quad = \frac{\Gamma(\alpha_i + n) \Gamma(\alpha_j + m) \Gamma(\alpha_0 + n + m)}
    {\Gamma(\alpha_i) \Gamma(\alpha_j) \Gamma(\alpha_0)} \\
    &\quad \quad  \int \Dir(\hat\dirp; \hat\alpha) \, \hat\dirp_i^n \, \hat\dirp_j^m \, \log \hat\dirp_i \, d\hat\dirp
    \end{split} \\
    & \quad = \frac{\alpha_i^{\overline{n}} \, \alpha_j^{\overline{m}}}{\alpha_0^{\overline{n+m}}} \E{}{\log \hat\dirp_i},
\end{align}
where $\hat\dirp \sim \Dir(\hat\alpha)$ with $\hat\alpha = (\alpha_0, \dots, \alpha_i + n, \ldots, \alpha_j + m, \ldots, \alpha_K)$, and we made use of the fact that $\frac{\Gamma(z+n)}{\Gamma(z)}=z^{\overline{n}}$. Finally, we can apply \Cref{ddu:eq:psi_log_expectation} on $\hat\dirp \sim \Dir(\hat\alpha)$ to show
\begin{align}
    \quad = \frac{\alpha_i^{\overline{n}} \, \alpha_j^{\overline{m}}}{\alpha_0^{\overline{n+m}}} \left ( \psi(\alpha_i + n) - \psi(\alpha_0 + n + m) \right ).
\end{align}
\end{enumerate}
\end{proof}
With this, we can already quantify the expected entropy $\simpleE{\dirp \sim \Dir(\alpha)} {\varHof{Y\sim \Cat(\dirp)}{Y}}$:
\begin{lemma}
\label{ddu:lemma:dirichlet_expected_categorical}
Given a Dirichlet distribution and a random variable $\dirp \sim Dir(\alpha)$, the expected entropy $\simpleE{\dirp \sim \Dir(\alpha)} {\varHof{Y\sim \Cat(\dirp)}{Y}}$ of the categorical distribution $Y \sim \Cat(\dirp)$ is given by
\begin{align}
     \simpleE{\pof{\dirp \given \alpha}}{\Hof{Y \given \dirp}} = \psi(\alpha_0+1) - \sum_{y=1}^K \frac{\alpha_i}{\alpha_0} \psi(\alpha_i +1).  
\end{align}
\end{lemma}
\begin{proof}
Applying the sum rule of expectations and \Cref{ddu:eq:nm_log_expectation} from \Cref{ddu:lemma:dirichlet_basics}, we can write
\begin{align}
    & \simpleE{}{\Hof{Y \given \dirp}} = \E{}{-\sum_{i=1}^K \dirp_i \log \dirp_i} = -\sum_i \E{}{\dirp_i \log \dirp_i} \\
    & \quad \quad = - \sum_i \frac{\alpha_i}{\alpha_0} \left ( \psi(\alpha_i+1) - \psi(\alpha_0+1)\right).
\end{align}
The result follows after rearranging and making use of $\sum_i \frac{\alpha_i}{\alpha_0} = 1$.
\end{proof}
With these statements, we can answer a slightly more complex problem:
\begin{lemma}
\label{ddu:lemma:log_covariance}
Given a Dirichlet distribution and a random variable $\dirp\sim\Dir(\alpha)$,
the covariance $\implicitCov{\dirp_i^n \log \dirp_i,  \dirp_j^m \log \dirp_j}$ is given by
\begin{align}
    \MoveEqLeft[3] \implicitCov{\dirp_i^n \log \dirp_i,  \dirp_j^m \log \dirp_j} \\
    \begin{split}
        ={}& \frac{\alpha_i^{\overline{n}}\,\alpha_j^{\overline{m}}}{\alpha_0^{\overline{n+m}}}
            \left ((\psi(\alpha_i+n)-\psi(\alpha_0+n+m)) \right. \\
         &  (\psi(\alpha_j+m)-\psi(\alpha_0+n+m)) \\
         & \left. - \psi'(\alpha_0+n+m) \right)\\
         & + \frac{\alpha_i^{\overline{n}}\,\alpha_j^{\overline{m}}}{\alpha_0^{\overline{n}}\,\alpha_0^{\overline{m}}}
         (\psi(\alpha_i+n)-\psi(\alpha_0+n)) \\
         &  (\psi(\alpha_j+m)-\psi(\alpha_0+n)),
    \end{split}
\end{align}
for $i\not=j$, where $\psi$ is the Digamma function and $\psi'$ is the Trigamma function.
Similarly, the covariance $\implicitCov{\dirp_i^n \log \dirp_i,  \dirp_i^m \log \dirp_i}$ is given by
\begin{align}
    \MoveEqLeft[3] \implicitCov{\dirp_i^n \log \dirp_i,  \dirp_i^m \log \dirp_i} \\
    \begin{split}
        ={}& \frac{\alpha_i^{\overline{n+m}}}{\alpha_0^{\overline{n+m}}}
            \left((\psi(\alpha_i+n+m)-\psi(\alpha_0+n+m))^2 \right.\\
         & + \left. \psi'(\alpha_i+n+m) - \psi'(\alpha_0+n+m)\right) \\
         & + \frac{\alpha_i^{\overline{n}}\,\alpha_i^{\overline{m}}}{\alpha_0^{\overline{n}}\,\alpha_0^{\overline{m}}}
         (\psi(\alpha_i+n)-\psi(\alpha_0+n)) \\
         & \quad \quad (\psi(\alpha_i+m)-\psi(\alpha_0+n)).
    \end{split}
\end{align}
\end{lemma}
Regrettably, the equations are getting large. By abuse of notation, we introduce a convenient shorthand before proving the lemma.
\newcommand{\psishort}[2]{\overline{\E{}{\log \hat{\dirp}_{#1}^{#2}}}}
\newcommand{\psipshort}[4]{\overline{\implicitCov{\log \hat{\dirp}_{#1}^{#2}, \log \hat{\dirp}_{#3}^{#4}}}}
\begin{definition}
We will denote by 
\begin{align}
    \psishort{i}{n,m} = \psi(\alpha_i+n)-\psi(\alpha_0+n+m),
\end{align} and use $\psishort{i}{n}$ for $\psishort{i}{n,0}$.
Likewise,
\begin{align}
    \psipshort{i}{n,m}{j}{n,m} = \psi'(\alpha_i + n) \delta_{ij} - \psi'(\alpha_0 + n + m).
\end{align}
\end{definition}
This notation agrees with the proof of \Cref{ddu:eq:psi_log_expectation} and \eqref{ddu:eq:psip_log_expectation} in \Cref{ddu:lemma:dirichlet_basics}. With this, we can significantly simplify the previous statements:
\begin{corollary}
Given a Dirichlet distribution and random variable $\dirp\sim\Dir(\alpha)$,
\begin{align}
    \E{}{\dirp_i^n \dirp_j^m \log \dirp_i} &= \frac{\alpha_i^{\overline{n}} \, \alpha_j^{\overline{m}}}{\alpha_0^{\overline{n+m}}} \psishort{i}{n,m},
\end{align}
\begin{align}
\MoveEqLeft[3] \implicitCov{\dirp_i^n \log \dirp_i,  \dirp_j^m \log \dirp_j} \\
\begin{split}
={}& \frac{\alpha_i^{\overline{n}}\,\alpha_j^{\overline{m}}}{\alpha_0^{\overline{n+m}}}
\left( \psishort{i}{n,m} \psishort{j}{m,n} \right. \\
& \quad \quad \left. \psipshort{i}{n,m}{j}{n,m} \right) \\
& + \frac{\alpha_i^{\overline{n}}\,\alpha_j^{\overline{m}}}{\alpha_0^{\overline{n}}\,\alpha_0^{\overline{m}}}
\psishort{i}{n} \psishort{j}{m} \quad \text{for $i \not= j$, and}
\end{split} \\
\MoveEqLeft[3] \implicitCov{\dirp_i^n \log \dirp_i,  \dirp_i^m \log \dirp_i} \\
    \begin{split}
        ={}& \frac{\alpha_i^{\overline{n+m}}}{\alpha_0^{\overline{n+m}}}
        \left(\psishort{i}{n+m}^2 \right.\\ 
        & \left. + \psipshort{i}{n+m}{i}{n+m} \right)\\
        & + \frac{\alpha_i^{\overline{n}}\,\alpha_i^{\overline{m}}}{\alpha_0^{\overline{n}}\,\alpha_0^{\overline{m}}} \psishort{i}{n}\psishort{j}{m}.
    \end{split}
\end{align}
\end{corollary}
\begin{proof}[Proof of \Cref{ddu:lemma:log_covariance}]
This proof applies the well-know formula \textbf{(cov)} $\implicitCov{X,Y} = \E{}{X \, Y} - \E{}{X} \E{}{Y}$ once forward and once backward \textbf{(rcov)} $\E{}{X \, Y} = \implicitCov{X,Y} + \E{}{X}\E{}{Y}$ while applying \Cref{ddu:eq:nm_log_expectation} several times:
\begin{align}
    & \implicitCov{\dirp_i^n \log \dirp_i,  \dirp_j^m \log \dirp_j} \\
    \begin{split}
    & \quad \overset{\textbf{cov}}{=} \E{}{\dirp_i^n \log(\dirp_i) \, \dirp_j^m \log(\dirp_j)} \\
    & \quad \quad - \E{}{\dirp_i^n \log \dirp_i}\E{}{\dirp_j^m \log \dirp_j}   
    \end{split} \\
    \begin{split}
    & \quad = \frac{\alpha_i^{\overline{n}}\,\alpha_j^{\overline{m}}}{\alpha_0^{\overline{n+m}}} \E{}{\log (\hat\dirp_i^{i,j}) \log(\hat\dirp_j^{i,j})} \\
    & \quad \quad - \E{}{\log \hat\dirp_i^{i}}\E{}{\log \dirp_j^{j}}
    \end{split} \\ 
    \begin{split}
    & \quad \overset{\textbf{(rcov)}}{=} \frac{\alpha_i^{\overline{n}}\,\alpha_j^{\overline{m}}}{\alpha_0^{\overline{n+m}}} 
    \left ( \implicitCov{\log \hat\dirp_i^{i,j}, \log \hat\dirp_j^{i,j} } \right. \\
    & \left. \quad \quad \quad \quad + \E{}{\log \hat\dirp_i^{i,j}}\E{}{\log \hat\dirp_j^{i,j}} \right ) \\
    & \quad \quad - \frac{\alpha_i^{\overline{n}}\,\alpha_j^{\overline{m}}}{\alpha_0^{\overline{n}}\,\alpha_0^{\overline{m}}} \E{}{\log \hat\dirp_i^{i}}\E{}{\log \dirp_j^{j}},
    \end{split}
\end{align}
where $\dirp^{i,j} \sim \Dir(\alpha^{i,j})$ with $\alpha^{i,j} = (\ldots, \alpha_i + n, \ldots, \alpha_j +m,\ldots)$. $\dirp^{i/j}$ and $\alpha^{i/j}$ are defined analogously. Applying \Cref{ddu:eq:psip_log_expectation} and \Cref{ddu:eq:psi_log_expectation} from \Cref{ddu:lemma:dirichlet_basics} yields the statement.
For $i=j$, the proof follows the same pattern.
\end{proof}

\subsubsection{Variance of Softmax Entropy}

Now, we can prove the theorem that quantifies the variance of the entropy of $Y$:
\begin{theorem}
\label{ddu:app:thm:smentvar}
Given a Dirichlet distribution and a random variable $\dirp \sim Dir(\alpha)$, the variance of the entropy $\Var{\dirp \sim \Dir(\alpha)}{\varHof{Y\sim \Cat(\dirp)}{Y}}$ of the categorical distribution $Y \sim \Cat(\dirp)$ is given by
\begin{align}
    \MoveEqLeft[2] \implicitVar{\Hof{Y \given \dirp}} & \\
    \begin{split}
    ={} & \sum_i \frac{\alpha_i^{\overline{2}}}{\alpha_0^{\overline{2}}} \left(\psipshort{i}{2}{i}{2} + \psishort{i}{2}^2 \right ) \\
    +{} & \sum_{i\not=j} \frac{\alpha_i \, \alpha_j}{\alpha_0^{\overline{2}}} \left( \psipshort{i}{1}{j}{1} \right. \\
    {} & \quad \left. + \psishort{i}{1,1} \, \psishort{j}{1,1} \right) \\
    -& \sum_{i,j} \frac{\alpha_i \, \alpha_j}{\alpha_0^{2}} \psishort{i}{1} \psishort{j}{1}.
    \end{split}\label{ddu:thm:E6}
\end{align}
\end{theorem}
\begin{proof}
We start by applying the well-known formula $\implicitVar{\sum_i X_i}=\sum_{i,j} \implicitCov{X_i, X_j}$ and then apply  \Cref{ddu:lemma:log_covariance} repeatedly.
\end{proof}

	}
}

\chapter{\ddutitle}
{
	
	\section{Experimental Details}
\label{ddu:app:exp_details}

\subsection{Dirty-MNIST}
\label{ddu:app:exp_details_dirty_mnist}

We train for 50 epochs using SGD with a momentum of 0.9 and an initial learning rate of 0.1.
The learning rate drops by a factor of 10 at training epochs 25 and 40. 
Following SNGP \citep{liu2020simple}, we apply online spectral normalization with one step of a power iteration on the convolutional weights.
For 1x1 convolutions, we use the exact algorithm, and for 3x3 convolutions, the approximate algorithm from \citet{gouk2018regularisation}.
The coefficient for SN is a hyperparameter which we set to 3 using cross-validation.

\subsection{OoD Detection Training Setup}
\label{ddu:app:exp_details_cifar}
We train the softmax baselines on CIFAR-10/100 for 350 epochs using SGD as the optimizer with a momentum of 0.9, and an initial learning rate of 0.1. The learning rate drops by a factor of 10 at epochs 150 and 250. We train the 5-Ensemble baseline using this same training setup. The SNGP and DUQ models were trained using the setup of SNGP and hyperparameters mentioned in their respective papers \citep{liu2020simple, van2020uncertainty}. For models trained on ImageNet, we train for 90 epochs with SGD optimizer, an initial learning rate of 0.1 and a weight decay of 1e-4. We use a learning rate warm-up decay of 0.01 along with a step scheduler with step size of 30 and a step factor of 0.1.

\subsection{Semantic Segmentation Training Setup}
\label{ddu:app:segmentation_training_setup}

In \Cref{ddu:fig:l2_dist}, we plot the L2 distance between feature space means of different classes for a pair of randomly chosen distant pixels on the Pascal VOC 2012 val set. We observe that feature space means between pairs of different classes are more distant compared to the same class irrespective of the location of the pixel for the class. This leads us to construct a Gaussian mean and covariance matrix per class as opposed to one mean and one covariance matrix per class per pixel, thereby greatly reducing the computational load of fitting a GMM in semantic segmentation. Similar to classification, we treat each pixel in the training set as a separate sample and fit a single Gaussian mean and covariance matrix per class.

For the semantic segmentation experiment, we use a DeepLab-v3+ \citep{chen2017rethinking} model with a ResNet-101 backbone as the architecture of choice. We train each of the models on Pascal VOC for 50 epochs using SGD as the optimizer, with a momentum of 0.9 and a weight decay of $5e-4$. We set the initial learning rate to $0.007$ with a polynomial decay during the course of training. Finally, we trained with a batch size of 32 parallelized over 4 GPUs.

\begin{figure}[t!]
    \centering
    \begin{subfigure}{0.33\linewidth}
        \includegraphics[width=\linewidth]{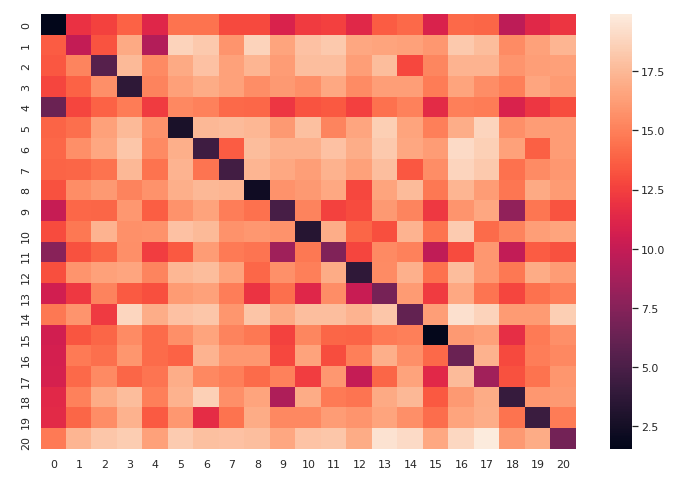}
        \label{ddu:fig:p1_matrix}
    \end{subfigure}\hfill
    \begin{subfigure}{0.33\linewidth}
        \centering
        \includegraphics[width=\linewidth]{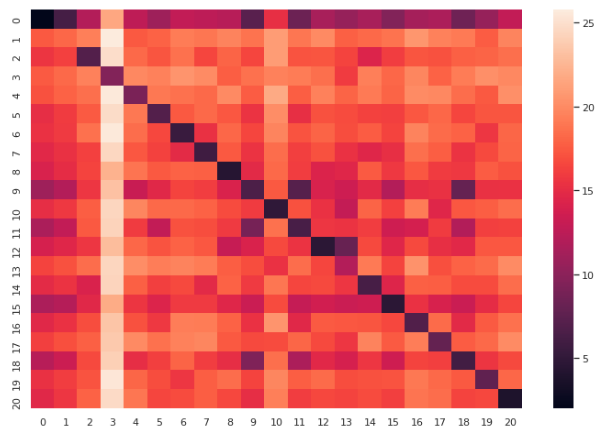}
        \label{ddu:fig:p2_matrix}
    \end{subfigure}\hfill
    \begin{subfigure}{0.33\linewidth}
        \centering
        \includegraphics[width=\linewidth]{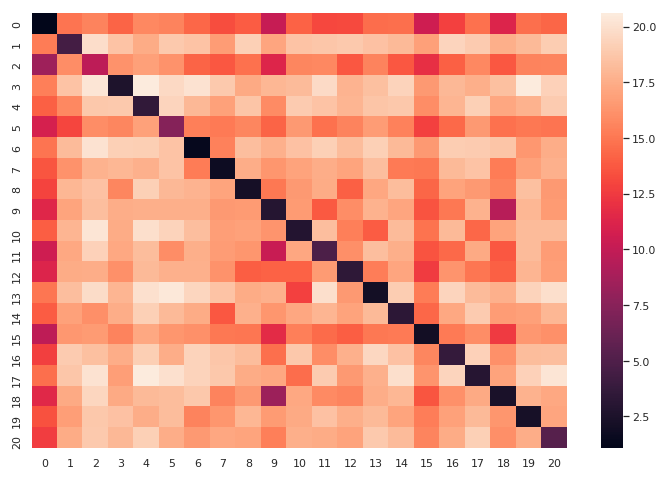}
        \label{ddu:fig:p3_matrix}
    \end{subfigure} 
    \caption{
    L2 distances between the feature space means of different classes for a pair of distant pixels on the Pascal VOC 2012 val set: (left) Pixels $(10, 255)$ and $(500, 255)$, (middle) Pixels $(234, 349)$ and $(36,22)$ and (right) Pixels $(300, 500)$ and $(400, 255)$.
    }
    \label{ddu:fig:l2_dist}
    \vspace{-1mm}
\end{figure}

\subsection{Compute Resources}
\label{ddu:app:exp_details_compute}

Each model (ResNet-18, Wide-ResNet-28-10, ResNet-50, ResNet-110, DenseNet-121 or VGG-16) used for the large scale active learning, CIFAR-10 vs SVHN/CIFAR-100/Tiny-ImageNet/CIFAR-10-C and CIFAR-100 vs SVHN/Tiny-ImageNet tasks was trained on a single Nvidia Quadro RTX 6000 GPU. Each model (LeNet, VGG-16 and ResNet-18) used to get the results in \Cref{ddu:fig:intro_histograms} and \Cref{ddu:table:auroc_tab1} was trained on a single Nvidia GeForce RTX 2060 GPU. Each model (ResNet-50, Wide-ResNet-50-2, VGG-16) trained on ImageNet was trained using 8 Nvidia Quadro RTX 6000 GPUs.

\begin{sidewaystable}[!]
\centering
\caption{\emph{OoD detection performance of different baselines using a ResNet-50 architecture with the CIFAR-10 vs SVHN/CIFAR-100/Tiny-ImageNet and CIFAR-100 vs SVHN/Tiny-ImageNet dataset pairs averaged over 25 runs.} Note: SN stands for Spectral Normalization, JP stands for Jacobian Penalty. We highlight the best deterministic and best method overall in bold for each metric.}
\label{ddu:table:ood_resnet50}
\resizebox{\linewidth}{!}{%
\begin{tabular}{cccccccccc}
\toprule
\textbf{\small Train Dataset} & \textbf{\small Method} & \textbf{\small Penalty} & {\small Aleatoric Uncertainty} &
\textbf{\small Epistemic Uncertainty} & {\small Accuracy (\textuparrow)} & {\small ECE (\textdownarrow)} & \multicolumn{3}{c}{\textbf{\small AUROC}} \\
\cmidrule{8-10}
& & & & & & & \textbf{\small SVHN (\textuparrow)} & \textbf{\small CIFAR-100 (\textuparrow)} & \textbf{\small Tiny-ImageNet (\textuparrow)}\\
\midrule
\multirow{7}{*}{CIFAR-10} & Softmax & - & \multirow{2}{*}{Softmax Entropy} & Softmax Entropy & \multirow{2}{*}{$\mathbf{95.04\pm0.05}$} & \multirow{2}{*}{$\mathbf{0.97\pm0.04}$} & $93.80\pm0.41$ & $88.91\pm0.07$ & $88.32\pm0.07$ \\
&Energy-based {\scriptsize \citep{liu2020energy}} & - && Softmax Density &&& $94.48\pm0.44$ & $88.84\pm0.08$ & $88.45\pm0.08$\\
&DUQ {\scriptsize \citep{van2020uncertainty}} & JP & Kernel Distance & Kernel Distance & $94.05\pm0.11$ & $1.71\pm0.07$ & $93.14\pm0.43$ & $83.87\pm0.27$ & $84.28\pm0.26$\\
&SNGP {\scriptsize \citep{liu2020simple}} & SN & Predictive Entropy & Predictive Entropy & $94.90\pm0.11$ & $1.01\pm0.03$ & $93.15\pm0.85$ & $89.32\pm0.10$ & $88.96\pm0.13$\\
&\textbf{DDU (ours)} & SN & Softmax Entropy & GMM Density & $94.92\pm0.06$&$1\pm0.04$&$\mathbf{94.77\pm0.35}$&$\mathbf{89.98\pm0.17}$ & $\mathbf{89.12\pm0.13}$\\
\cmidrule{2-10}
&5-Ensemble & \multirow{2}{*}{-} & \multirow{2}{*}{Predictive Entropy} & Predictive Entropy & \multirow{2}{*}{$\mathbf{96.06\pm0.04}$}&\multirow{2}{*}{$1.65\pm0.07$}&$94.75\pm0.39$&$89.87\pm0.06$ & $88.69\pm0.05$\\
&{\scriptsize \citep{lakshminarayanan2016simple}} &&& Mutual Information &&&$94.09\pm0.20$&$89.76\pm0.06$ & $89.04\pm0.03$\\
\midrule
&&&&& \textbf{\small Test Accuracy (\textuparrow)} & \textbf{\small{Test ECE (\textdownarrow)}} & \multicolumn{2}{c}{\textbf{\small SVHN (\textuparrow)}} & \textbf{\small Tiny-ImageNet (\textuparrow)} \\
\cmidrule{6-10}
\multirow{6}{*}{CIFAR-100} & Softmax & - & \multirow{2}{*}{Softmax Entropy} & Softmax Entropy & \multirow{2}{*}{$77.91\pm0.09$} & \multirow{2}{*}{$4.32\pm0.10$} & \multicolumn{2}{c}{$81.32\pm0.65$} & $79.83\pm0.07$ \\
&Energy-based {\scriptsize \citep{liu2020energy}} & - && Softmax Density &&&\multicolumn{2}{c}{$82.05\pm0.69$} & $79.61\pm0.08$ \\
&SNGP {\scriptsize \citep{liu2020simple}} & SN & Predictive Entropy & Predictive Entropy & $74.73\pm0.22$ & $7.68\pm0.13$ & \multicolumn{2}{c}{$82.50\pm2.09$} & $77.05\pm0.16$ \\
&\textbf{DDU (ours)} & SN & Softmax Entropy & GMM Density & $\mathbf{79.26\pm0.16}$&$\mathbf{4.07\pm0.06}$& \multicolumn{2}{c}{$\mathbf{87.34\pm0.64}$}& $\mathbf{82.11\pm0.20}$ \\
\cmidrule{2-10}
&5-Ensemble & \multirow{2}{*}{-} & \multirow{2}{*}{Predictive Entropy} & Predictive Entropy & \multirow{2}{*}{$\mathbf{81.06\pm0.07}$}&\multirow{2}{*}{$\mathbf{3.54\pm0.12}$}& \multicolumn{2}{c}{$83.42\pm0.89$} & $77.69\pm0.12$\\
&{\scriptsize \citep{lakshminarayanan2016simple}} &&& Mutual Information &&& \multicolumn{2}{c}{$84.24\pm0.90$} & $81.59\pm0.05$\\
\bottomrule
\end{tabular}}
\end{sidewaystable}
\begin{sidewaystable}[!]
\centering
\caption{\emph{OoD detection performance of different baselines using a ResNet-110 architecture with the CIFAR-10 vs SVHN/CIFAR-100/Tiny-ImageNet and CIFAR-100 vs SVHN/Tiny-ImageNet dataset pairs averaged over 25 runs.} Note: SN stands for Spectral Normalization, JP stands for Jacobian Penalty. We highlight the best deterministic and best method overall in bold for each metric.}
\label{ddu:table:ood_resnet110}
\resizebox{\linewidth}{!}{%
\begin{tabular}{cccccccccc}
\toprule
\textbf{\small Train Dataset} & \textbf{\small Method} & \textbf{\small Penalty} & {\small Aleatoric Uncertainty} &
\textbf{\small Epistemic Uncertainty} & {\small Accuracy (\textuparrow)} & {\small ECE (\textdownarrow)} & \multicolumn{3}{c}{\textbf{\small AUROC}} \\
\cmidrule{8-10}
& & & & & & & \textbf{\small SVHN (\textuparrow)} & \textbf{\small CIFAR-100 (\textuparrow)} & \textbf{\small Tiny-ImageNet (\textuparrow)}\\
\midrule
\multirow{7}{*}{CIFAR-10} & Softmax & - & \multirow{2}{*}{Softmax Entropy} & Softmax Entropy & \multirow{2}{*}{$\mathbf{95.08\pm0.04}$} & \multirow{2}{*}{$1.02\pm0.04$} & $93.12\pm0.44$ & $88.7\pm0.1$ & $88.07\pm0.11$ \\
&Energy-based {\scriptsize \citep{liu2020energy}} & - && Softmax Density &&& $93.67\pm0.47$ & $88.60\pm0.11$ & $88.13\pm0.11$\\
&DUQ {\scriptsize \citep{van2020uncertainty}} & JP & Kernel Distance & Kernel Distance & $94.32\pm0.17$ & $1.21\pm0.07$ & $94.02\pm0.45$ & $86.17\pm0.35$ & $85.24\pm0.21$\\
&SNGP {\scriptsize \citep{liu2020simple}} & SN & Predictive Entropy & Predictive Entropy & $94.85\pm0.09$ & $1.04\pm0.02$ & $93.17\pm0.53$ & $89.23\pm0.10$ & $88.80\pm0.12$\\
&\textbf{DDU (ours)} & SN & Softmax Entropy & GMM Density & $94.82\pm0.06$&$\mathbf{1.01\pm0.04}$&$\mathbf{95.48\pm0.30}$&$\mathbf{90.08\pm0.13}$ & $\mathbf{89.18\pm0.15}$\\
\cmidrule{2-10}
&5-Ensemble & \multirow{2}{*}{-} & \multirow{2}{*}{Predictive Entropy} & Predictive Entropy & \multirow{2}{*}{$\mathbf{96.18\pm0.05}$}& \multirow{2}{*}{$1.57\pm0.05$}&$95.07\pm0.45$&$\mathbf{90.23\pm0.04}$ & $89\pm0.03$\\
&{\scriptsize \citep{lakshminarayanan2016simple}} &&& Mutual Information &&&$94.72\pm0.34$&$89.69\pm0.05$ & $88.35\pm0.05$\\
\midrule
&&&&& \textbf{\small Test Accuracy (\textuparrow)} & \textbf{\small{Test ECE (\textdownarrow)}} & \multicolumn{2}{c}{\textbf{\small AUROC SVHN (\textuparrow)}} & \textbf{\small AUROC Tiny-ImageNet (\textuparrow)} \\
\cmidrule{6-10}
\multirow{6}{*}{CIFAR-100} & Softmax & - & \multirow{2}{*}{Softmax Entropy} & Softmax Entropy & \multirow{2}{*}{$78.65\pm0.10$} & \multirow{2}{*}{$3.93\pm0.13$} & \multicolumn{2}{c}{$82.04\pm0.57$} & $80.13\pm0.07$ \\
&Energy-based {\scriptsize \citep{liu2020energy}} & - && Softmax Density &&&\multicolumn{2}{c}{$82.78\pm0.60$} & $80.01\pm0.09$ \\
&SNGP {\scriptsize \citep{liu2020simple}} & SN & Predictive Entropy & Predictive Entropy & $76.16\pm0.27$ & $6.43\pm0.75$ & \multicolumn{2}{c}{$83.94\pm0.10$} & $78.54\pm0.28$ \\
&\textbf{DDU (ours)} & SN & Softmax Entropy & GMM Density & $\mathbf{78.89\pm0.17}$&$\mathbf{3.79\pm0.07}$& \multicolumn{2}{c}{$\mathbf{88.66\pm0.56}$}& $\mathbf{82.58\pm0.24}$ \\
\cmidrule{2-10}
&5-Ensemble & \multirow{2}{*}{-} & \multirow{2}{*}{Predictive Entropy} & Predictive Entropy & \multirow{2}{*}{$\mathbf{81.80\pm0.10}$} & \multirow{2}{*}{$\mathbf{3.67\pm0.11}$} & \multicolumn{2}{c}{$83.68\pm0.33$} & $81.12\pm0.13$\\
&{\scriptsize \citep{lakshminarayanan2016simple}} &&& Mutual Information &&& \multicolumn{2}{c}{$85.11\pm0.57$} & $81.94\pm0.06$\\
\bottomrule
\end{tabular}}
\end{sidewaystable}

\begin{sidewaystable*}[!]
\centering
\caption{\emph{OoD detection performance of different baselines using a DenseNet-121 architecture with the CIFAR-10 vs SVHN/CIFAR-100/Tiny-ImageNet and CIFAR-100 vs SVHN/Tiny-ImageNet dataset pairs averaged over 25 runs.} Note: SN stands for Spectral Normalization, JP stands for Jacobian Penalty. We highlight the best deterministic and best method overall in bold for each metric.}
\label{ddu:table:ood_densenet121}
\resizebox{\linewidth}{!}{%
\begin{tabular}{cccccccccc}
\toprule
\textbf{\small Train Dataset} & \textbf{\small Method} & \textbf{\small Penalty} & {\small Aleatoric Uncertainty} &
\textbf{\small Epistemic Uncertainty} & {\small Accuracy (\textuparrow)} & {\small ECE (\textdownarrow)} & \multicolumn{3}{c}{\textbf{\small AUROC}} \\
\cmidrule{8-10}
& & & & & & & \textbf{\small SVHN (\textuparrow)} & \textbf{\small CIFAR-100 (\textuparrow)} & \textbf{\small Tiny-ImageNet (\textuparrow)}\\
\midrule
\multirow{7}{*}{CIFAR-10} & Softmax & - & \multirow{2}{*}{Softmax Entropy} & Softmax Entropy & \multirow{2}{*}{$95.16\pm0.03$} & \multirow{2}{*}{$1.10\pm0.04$} & $94\pm0.44$ & $87.55\pm0.11$ & $86.99\pm0.12$ \\
&Energy-based {\scriptsize \citep{liu2020energy}} & - && Softmax Density &&& $94.07\pm0.54$ & $86.73\pm0.15$ & $86.43\pm0.16$\\
&DUQ {\scriptsize \citep{van2020uncertainty}} & JP & Kernel Distance & Kernel Distance & $95.02\pm0.14$ & $1.08\pm0.08$ & $94.67\pm0.41$ & $87.38\pm0.21$ & $86.72\pm0.14$\\
&SNGP {\scriptsize \citep{liu2020simple}} & SN & Predictive Entropy & Predictive Entropy & $94.31\pm0.21$ & $1.08\pm0.10$ & $94.48\pm0.34$ & $88.86\pm0.46$ & $88.40\pm0.48$\\
&\textbf{DDU (ours)} & SN & Softmax Entropy & GMM Density & $\mathbf{95.21\pm0.03}$&$\mathbf{1.05\pm0.03}$&$\mathbf{96.21\pm0.31}$&$\mathbf{90.84\pm0.06}$ & $\mathbf{89.70\pm0.06}$\\
\cmidrule{2-10}
&5-Ensemble & \multirow{2}{*}{-} & \multirow{2}{*}{Predictive Entropy} & Predictive Entropy & \multirow{2}{*}{$\mathbf{96.18\pm0.05}$}&\multirow{2}{*}{$1.07\pm0.07$}&$95.78\pm0.11$&$90.65\pm0.03$ & $89.62\pm0.06$\\
&{\scriptsize \citep{lakshminarayanan2016simple}} &&& Mutual Information &&&$95.75\pm0.10$&$90.71\pm0.04$ & $89.34\pm0.06$\\
\midrule
&&&&& \textbf{\small Test Accuracy (\textuparrow)} & \textbf{\small{Test ECE (\textdownarrow)}} & \multicolumn{2}{c}{\textbf{\small SVHN (\textuparrow)}} & \textbf{\small Tiny-ImageNet (\textuparrow)} \\
\cmidrule{6-10}
\multirow{6}{*}{CIFAR-100} & Softmax & - & \multirow{2}{*}{Softmax Entropy} & Softmax Entropy & \multirow{2}{*}{$79.02\pm0.08$} & \multirow{2}{*}{$4.11\pm0.08$} & \multicolumn{2}{c}{$85.86\pm0.42$} & $81.10\pm0.07$ \\
&Energy-based {\scriptsize \citep{liu2020energy}} & - && Softmax Density &&&\multicolumn{2}{c}{$87.09\pm0.49$} & $80.84\pm0.08$ \\
&SNGP {\scriptsize \citep{liu2020simple}} & SN & Predictive Entropy & Predictive Entropy & $79.15\pm0.15$ & $6.73\pm0.10$ & \multicolumn{2}{c}{$85.00\pm0.12$} & $79.76\pm0.15$ \\
&\textbf{DDU (ours)} & SN & Softmax Entropy & GMM Density & $\mathbf{79.15\pm0.07}$&$\mathbf{4.11\pm0.06}$& \multicolumn{2}{c}{$\mathbf{88.44\pm0.55}$}& $\mathbf{81.85\pm0.11}$ \\
\cmidrule{2-10}
&5-Ensemble & \multirow{2}{*}{-} & \multirow{2}{*}{Predictive Entropy} & Predictive Entropy & \multirow{2}{*}{$\mathbf{81.01\pm0.13}$}&\multirow{2}{*}{$4.81\pm0.05$}& \multicolumn{2}{c}{$88.32\pm0.61$} & $81.45\pm0.12$\\
&{\scriptsize \citep{lakshminarayanan2016simple}} &&& Mutual Information &&& \multicolumn{2}{c}{$88.36\pm0.17$} & $81.73\pm0.06$\\
\bottomrule
\end{tabular}}
\end{sidewaystable*}

\begin{sidewaystable*}
\centering
\caption{\emph{OoD detection performance of different ablations trained on CIFAR-10 using Wide-ResNet-28-10 and VGG-16 architectures with SVHN, CIFAR-100 and Tiny-ImageNet as OoD datasets averaged over 25 runs.} Note: SN stands for Spectral Normalization. We highlight the best deterministic and best method overall in bold for each metric.}
\label{ddu:table:ood_2}
\scriptsize
\resizebox{\linewidth}{!}{%
\begin{tabular}{cccccccccccc}
\toprule
\multicolumn{5}{c}{\textbf{Ablations}} & \textbf{Aleatoric Uncertainty} &
\textbf{Epistemic Uncertainty} & \textbf{Test Accuracy (\textuparrow)} & \textbf{Test ECE (\textdownarrow)} & \multicolumn{3}{c}{\textbf{AUROC}} \\
\cmidrule{1-5}
\cmidrule{10-12}
\textbf{Architecture} & \textbf{Ensemble} & \makecell[c]{\textbf{Residual} \\ \textbf{Connections}} & \textbf{SN} & \textbf{GMM} &&&&& \textbf{SVHN (\textuparrow)} & \textbf{CIFAR-100 (\textuparrow)} & \textbf{Tiny-ImageNet (\textuparrow)} \\
\midrule
\toprule
\multicolumn{5}{c}{\textbf{Ablations}} & \textbf{Aleatoric Uncertainty} &
\textbf{Epistemic Uncertainty} & \textbf{Test Accuracy (\textuparrow)} & \textbf{Test ECE (\textdownarrow)} & \multicolumn{3}{c}{\textbf{AUROC}} \\
\cmidrule{1-5}
\cmidrule{10-12}
\textbf{Architecture} & \textbf{Ensemble} & \makecell[c]{\textbf{Residual} \\ \textbf{Connections}} & \textbf{SN} & \textbf{GMM} &&&&& \textbf{SVHN (\textuparrow)} & \textbf{CIFAR-100 (\textuparrow)} & \textbf{Tiny-ImageNet (\textuparrow)} \\
\midrule
\multirow{10}{*}{Wide-ResNet-28-10} & \multirow{8}{*}{\xmark} & \multirow{8}{*}{\cmark} & \multirow{4}{*}{\xmark} & \multirow{2}{*}{\xmark} & \multirow{2}{*}{Softmax Entropy} & Softmax Entropy & \multirow{2}{*}{$\mathbf{95.98\pm0.02}$}&\multirow{2}{*}{$0.85\pm0.02$}&$94.44\pm0.43$&$89.39\pm0.06$&$88.42\pm0.05$ \\
   &&                         &                         &                         &                                  & Softmax Density &&&$94.56\pm0.51$&$88.89\pm0.07$&$88.11\pm0.06$ \\
                                                                                 \cmidrule{5-12}
   &&                         &                         &   \cmark                & Softmax Entropy & GMM Density   & $95.98\pm0.02$&$0.85\pm0.02$&$96.08\pm0.25$&$90.94\pm0.03$&$90.62\pm0.05$ \\
                                                       \cmidrule{4-12}
   &&                         & \multirow{4}{*}{\cmark} & \multirow{2}{*}{\xmark}  & \multirow{2}{*}{Softmax Entropy} & Softmax Entropy & \multirow{2}{*}{$95.97\pm0.03$}&\multirow{2}{*}{$0.85\pm0.04$}&$94.05\pm0.26$&$90.02\pm0.07$&$89.07\pm0.06$ \\
   &&                         &                         &                         &                                   & Softmax Density &&&$94.31\pm0.33$&$89.78\pm0.08$&$88.96\pm0.07$ \\
                                                                                 \cmidrule{5-12}
   &&                         &                         &      \cmark                   & \textbf{Softmax Entropy} &   \textbf{GMM Density}                               & $95.97\pm0.03$&$\mathbf{0.85\pm0.04}$&$\mathbf{97.86\pm0.19}$&$\mathbf{91.34\pm0.04}$&$\mathbf{91.07\pm0.05}$ \\
   \cmidrule{2-12}
   & \multirow{2}{*}{\cmark} & \multirow{2}{*}{\cmark} & \multirow{2}{*}{\xmark} & \multirow{2}{*}{\xmark} & \multirow{2}{*}{Predictive Entropy} & Predictive Entropy & \multirow{2}{*}{$\mathbf{96.59\pm0.02}$} & \multirow{2}{*}{$\mathbf{0.76\pm0.03}$} & $97.73\pm0.31$ & $\mathbf{92.13\pm0.02}$ & $90.06\pm0.03$\\
   & & & & & & Mutual Information &&& $97.18\pm0.19$ & $91.33\pm0.03$ & $90.90\pm0.03$\\
\midrule
\multirow{10}{*}{VGG-16} & \multirow{8}{*}{\xmark} & \multirow{8}{*}{\cmark} & \multirow{4}{*}{\xmark} & \multirow{2}{*}{\xmark} & \multirow{2}{*}{Softmax Entropy} & Softmax Entropy & \multirow{2}{*}{$93.63\pm0.04$}&\multirow{2}{*}{$1.64\pm0.03$}&$85.76\pm0.84$&$82.48\pm0.14$&$83.07\pm0.12$ \\
&&                         &                         &                         &                                  & Softmax Density &&&$84.24\pm1.04$&$81.91\pm0.17$&$82.82\pm0.14$ \\
                                                                             \cmidrule{5-12}
&&                         &                         &   \cmark                & Softmax Entropy & GMM Density   & $93.63\pm0.04$&$1.64\pm0.03$&$89.25\pm0.36$&$86.55\pm0.10$&$86.78\pm0.09$ \\
                                                   \cmidrule{4-12}
&&                         & \multirow{4}{*}{\cmark} & \multirow{2}{*}{\xmark}  & \multirow{2}{*}{Softmax Entropy} & Softmax Entropy & \multirow{2}{*}{$93.62\pm0.04$}&\multirow{2}{*}{$1.78\pm0.04$}&$87.54\pm0.41$&$82.71\pm0.09$&$83.33\pm0.08$ \\
&&                         &                         &                         &                                   & Softmax Density &&&$86.28\pm0.51$&$82.15\pm0.11$&$83.07\pm0.10$ \\
                                                                             \cmidrule{5-12}
&&                         &                         &      \cmark                   & Softmax Entropy &   GMM Density                               & $93.62\pm0.04$&$1.78\pm0.04$&$89.62\pm0.37$&$86.37\pm0.14$&$86.63\pm0.11$ \\
\cmidrule{2-12}
& \multirow{2}{*}{\cmark} & \multirow{2}{*}{\cmark} & \multirow{2}{*}{\xmark} & \multirow{2}{*}{\xmark} & \multirow{2}{*}{Predictive Entropy} & Predictive Entropy & \multirow{2}{*}{$94.9\pm0.05$} & \multirow{2}{*}{$2.03\pm0.03$} & $92.80\pm0.18$ & $89.01\pm0.08$ & $87.66\pm0.08$\\
& & & & & & Mutual Information &&& $91\pm0.22$ & $88.43\pm0.08$ & $88.74\pm0.05$\\
\bottomrule
\end{tabular}}
\end{sidewaystable*}

\begin{sidewaystable*}[!]
\centering
\caption{\emph{OoD detection performance of different ablations trained on CIFAR-100 using Wide-ResNet-28-10 and VGG-16 architectures with SVHN and Tiny-ImageNet as the OoD dataset averaged over 25 runs.} Note: SN stands for Spectral Normalization. We highlight the best deterministic and best method overall in bold for each metric.}
\label{ddu:table:ood_3}
\scriptsize
\resizebox{\linewidth}{!}{%
\begin{tabular}{ccccccccccc}
\toprule
\multicolumn{5}{c}{\textbf{Ablations}} & \textbf{Aleatoric Uncertainty} &
\textbf{Epistemic Uncertainty} & \textbf{Test Accuracy (\textuparrow)} & \textbf{Test ECE (\textdownarrow)} & \multicolumn{2}{c}{\textbf{AUROC}} \\
\cmidrule{1-5}
\cmidrule{10-11}
\textbf{Architecture} & \textbf{Ensemble} & \makecell[c]{\textbf{Residual} \\ \textbf{Connections}} & \textbf{SN} & \textbf{GMM} &&&&& \textbf{SVHN (\textuparrow)} & \textbf{Tiny-ImageNet (\textuparrow)} \\
\midrule
\multirow{10}{*}{Wide-ResNet-28-10} & \multirow{8}{*}{\xmark} & \multirow{8}{*}{\cmark} & \multirow{4}{*}{\xmark} & \multirow{2}{*}{\xmark} & \multirow{2}{*}{Softmax Entropy} & Softmax Entropy & \multirow{2}{*}{$80.26\pm0.06$}&\multirow{2}{*}{$4.62\pm0.06$}&$77.42\pm0.57$&$81.53\pm0.05$ \\
       &&                         &                         &                         &                                  & Softmax Density &&&$78.00\pm0.63$&$81.33\pm0.06$ \\
                                                                                     \cmidrule{5-11}
       &&                         &                         &   \cmark                & Softmax Entropy & GMM Density   & $80.26\pm0.06$&$4.62\pm0.06$&$87.54\pm0.61$&$78.13\pm0.08$ \\
                                                           \cmidrule{4-11}
       &&                         & \multirow{4}{*}{\cmark} & \multirow{2}{*}{\xmark}  & \multirow{2}{*}{Softmax Entropy} & Softmax Entropy & \multirow{2}{*}{$80.98\pm0.06$}&\multirow{2}{*}{$4.10\pm0.08$}&$85.37\pm0.36$&$82.57\pm0.03$ \\
       &&                         &                         &                         &                                   & Softmax Density &&&$86.41\pm0.38$&$82.49\pm0.04$ \\
                                                                                     \cmidrule{5-11}
       &&                         &                         &      \cmark                   & \textbf{Softmax Entropy} &   \textbf{GMM Density}                               & $\mathbf{80.98\pm0.06}$&$\mathbf{4.10\pm0.08}$&$\mathbf{87.53\pm0.62}$&$\mathbf{83.13\pm0.06}$ \\
       \cmidrule{2-11}
       & \multirow{2}{*}{\cmark} & \multirow{2}{*}{\cmark} & \multirow{2}{*}{\xmark} & \multirow{2}{*}{\xmark} & \multirow{2}{*}{Predictive Entropy} & Predictive Entropy & \multirow{2}{*}{$\mathbf{82.79\pm0.10}$} & \multirow{2}{*}{$\mathbf{3.32\pm0.09}$} & $79.54\pm0.91$&$82.95\pm0.09$ \\
       & & & & & & Mutual Information &&& $77.00\pm1.54$ & $82.82\pm0.04$ \\
\midrule
\multirow{10}{*}{VGG-16} & \multirow{8}{*}{\xmark} & \multirow{8}{*}{\cmark} & \multirow{4}{*}{\xmark} & \multirow{2}{*}{\xmark} & \multirow{2}{*}{Softmax Entropy} & Softmax Entropy & \multirow{2}{*}{$73.48\pm0.05$}&\multirow{2}{*}{$4.46\pm0.05$}&$76.73\pm0.72$&$76.43\pm0.05$ \\
       &&                         &                         &                         &                                  & Softmax Density &&&$77.70\pm0.86$&$74.68\pm0.07$ \\
                                                                                     \cmidrule{5-11}
       &&                         &                         &   \cmark                & Softmax Entropy & GMM Density   & $73.48\pm0.05$&$4.46\pm0.05$&$75.65\pm0.95$&$74.32\pm1.73$ \\
                                                           \cmidrule{4-11}
       &&                         & \multirow{4}{*}{\cmark} & \multirow{2}{*}{\xmark}  & \multirow{2}{*}{Softmax Entropy} & Softmax Entropy & \multirow{2}{*}{$73.58\pm0.06$}&\multirow{2}{*}{$4.32\pm0.06$}&$77.21\pm0.77$&$76.59\pm0.06$ \\
       &&                         &                         &                         &                                   & Softmax Density &&&$77.76\pm0.90$&$74.86\pm0.08$ \\
                                                                                     \cmidrule{5-11}
       &&                         &                         &      \cmark                   & Softmax Entropy &   GMM Density                               & $73.58\pm0.06$&$4.32\pm0.06$&$75.99\pm1.23$&$74.06\pm1.67$ \\
       \cmidrule{2-11}
       & \multirow{2}{*}{\cmark} & \multirow{2}{*}{\cmark} & \multirow{2}{*}{\xmark} & \multirow{2}{*}{\xmark} & \multirow{2}{*}{Predictive Entropy} & Predictive Entropy & \multirow{2}{*}{$77.84\pm0.11$} & \multirow{2}{*}{$5.32\pm0.10$} & $79.62\pm0.73$&$78.66\pm06$ \\
       & & & & & & Mutual Information &&& $72.07\pm0.48$ & $76.27\pm0.05$ \\
\bottomrule
\end{tabular}}
\end{sidewaystable*}
\begin{figure}[!t]
    \centering
    \begin{subfigure}{0.25\linewidth}
        \centering
        \includegraphics[width=\linewidth]{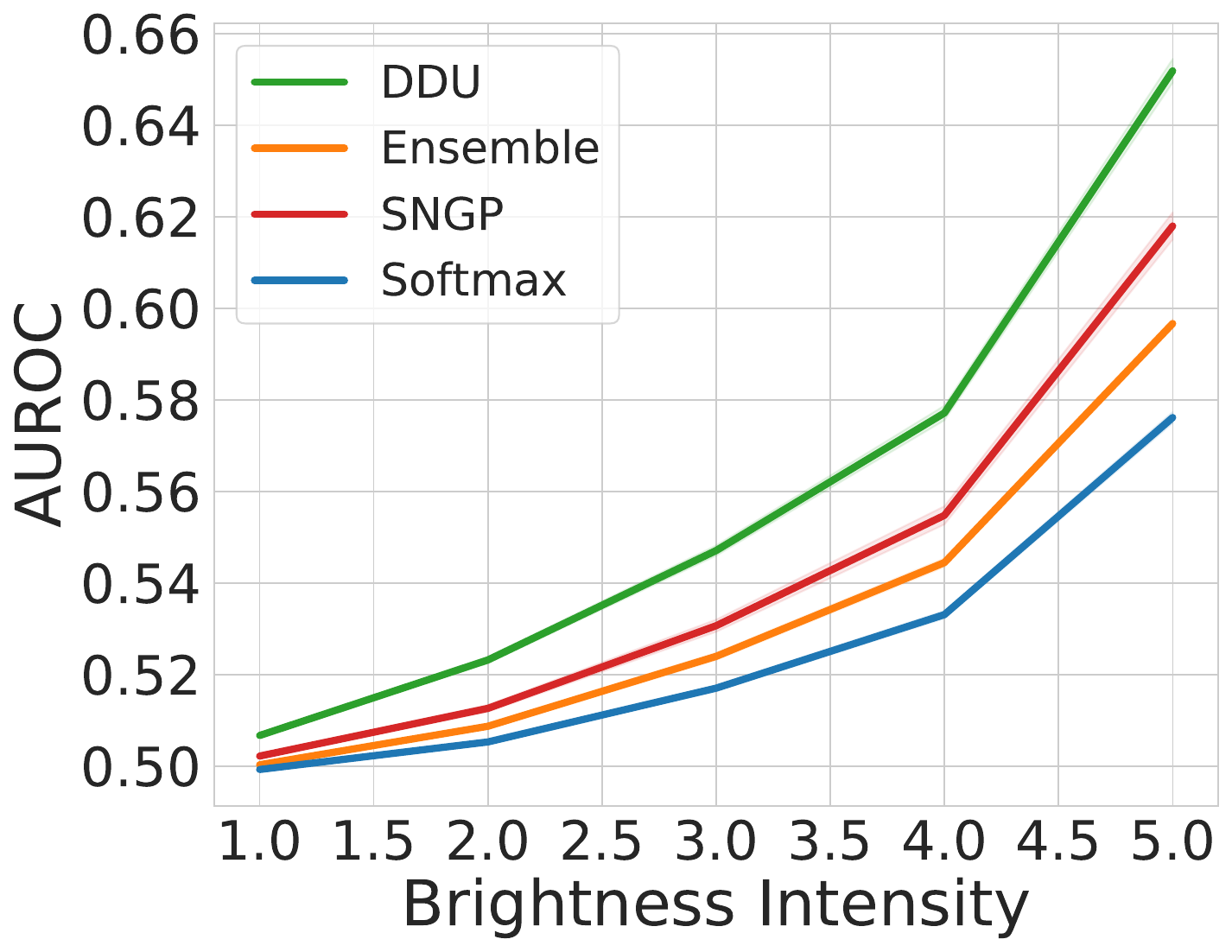}
    \end{subfigure}\hfill
    \begin{subfigure}{0.25\linewidth}
        \centering
        \includegraphics[width=\linewidth]{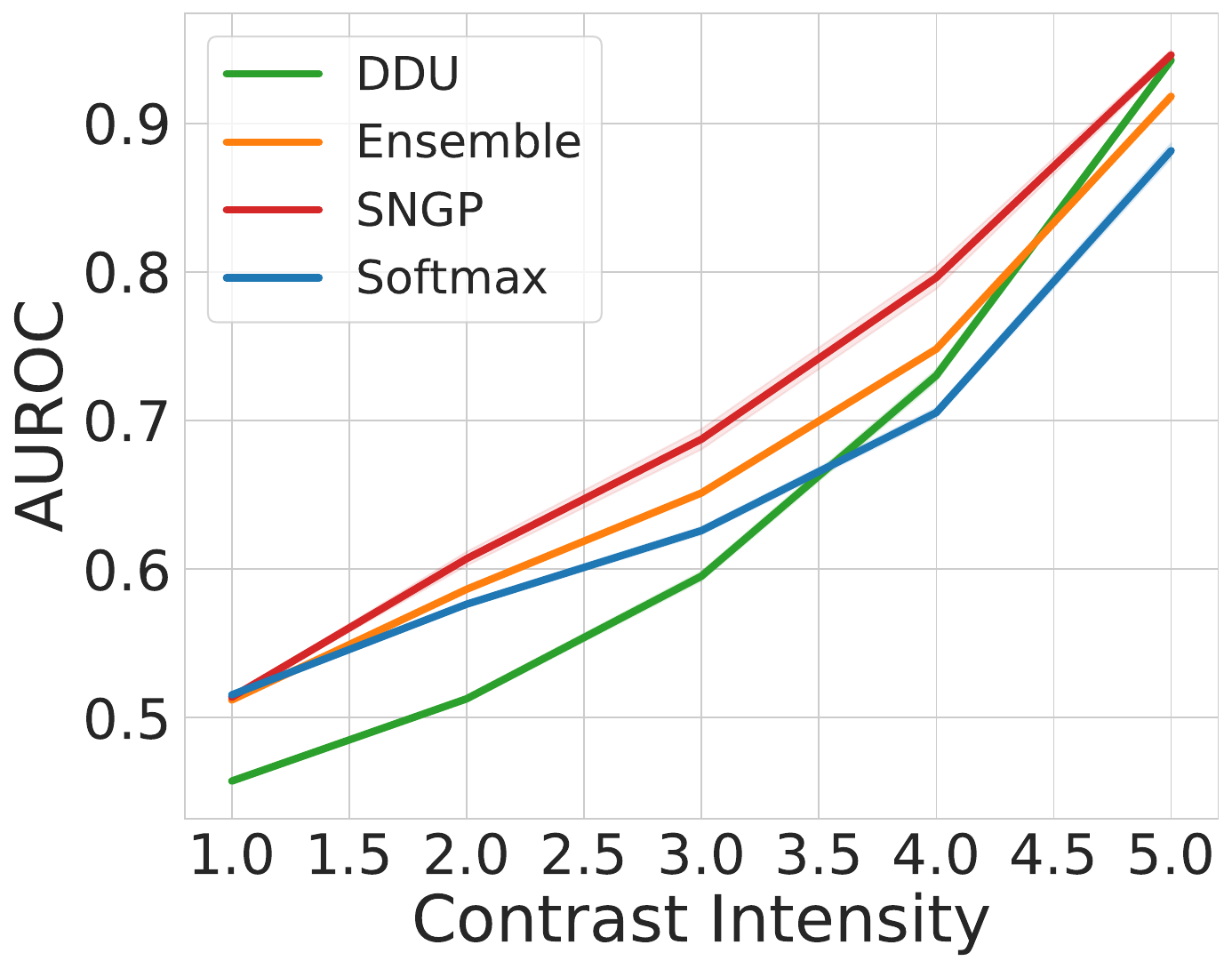}
    \end{subfigure}\hfill 
    \begin{subfigure}{0.25\linewidth}
        \centering
        \includegraphics[width=\linewidth]{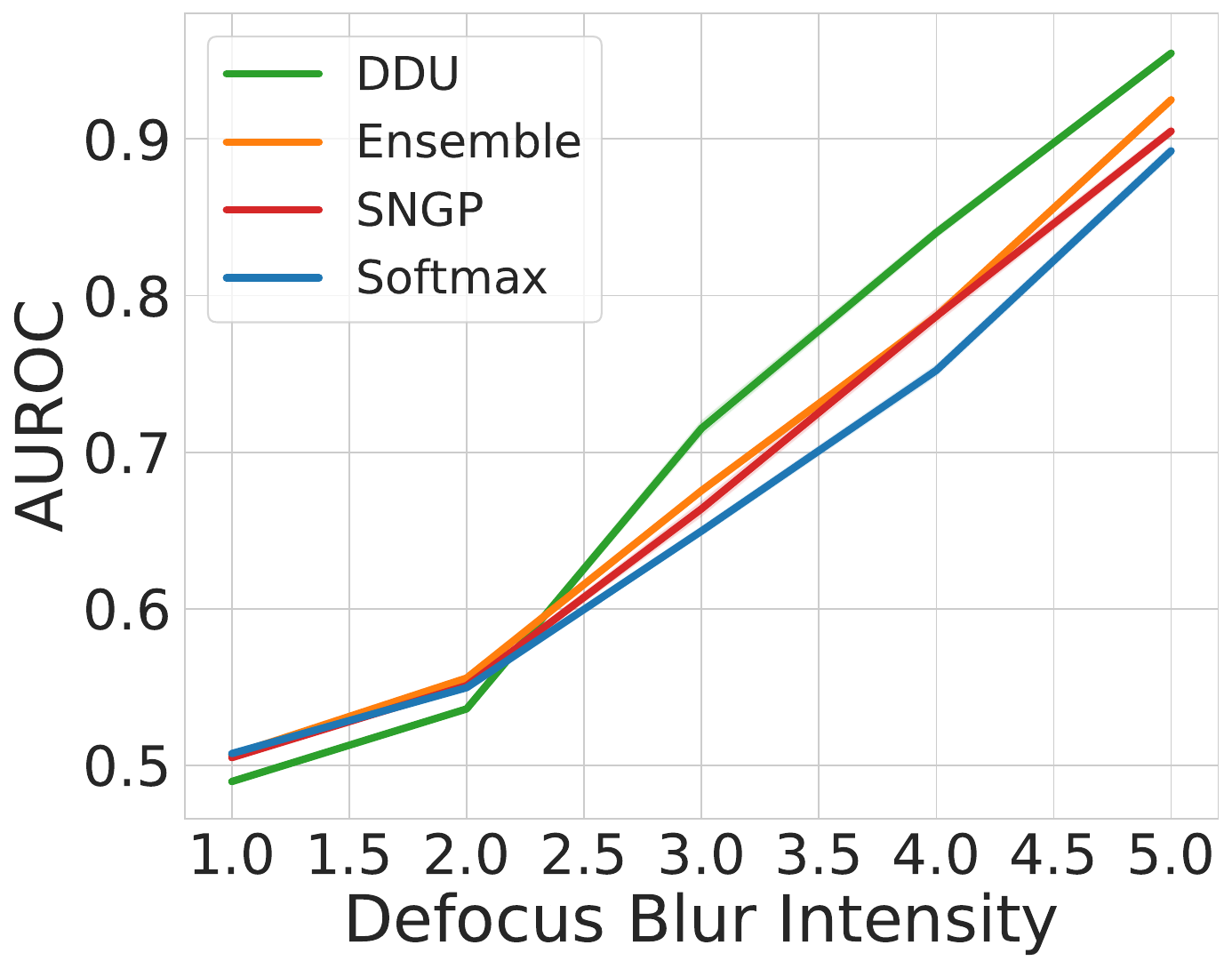}
    \end{subfigure}\hfill 
    \begin{subfigure}{0.25\linewidth}
        \centering
        \includegraphics[width=\linewidth]{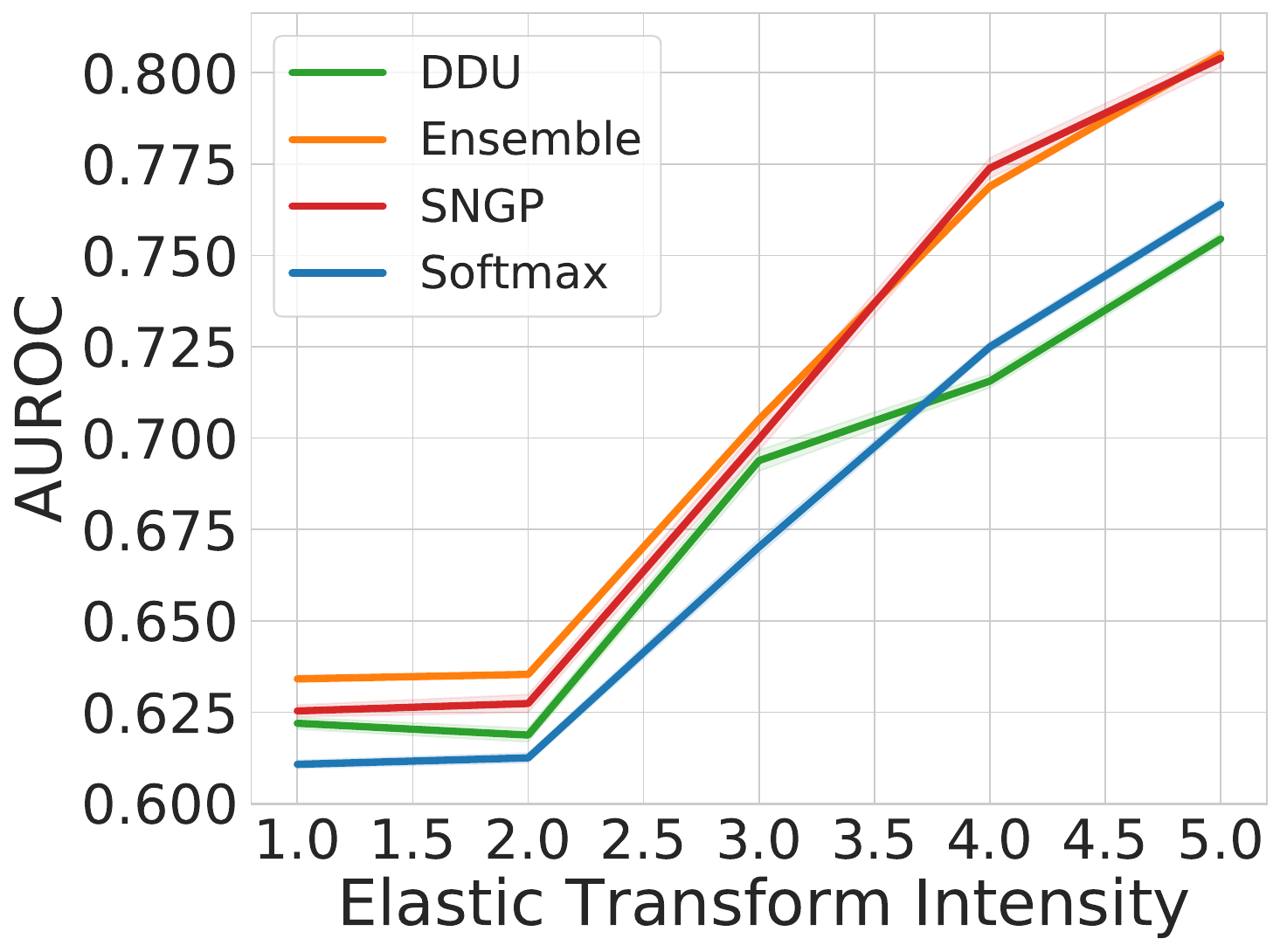}
    \end{subfigure}\hfill
    \begin{subfigure}{0.25\linewidth}
        \centering
        \includegraphics[width=\linewidth]{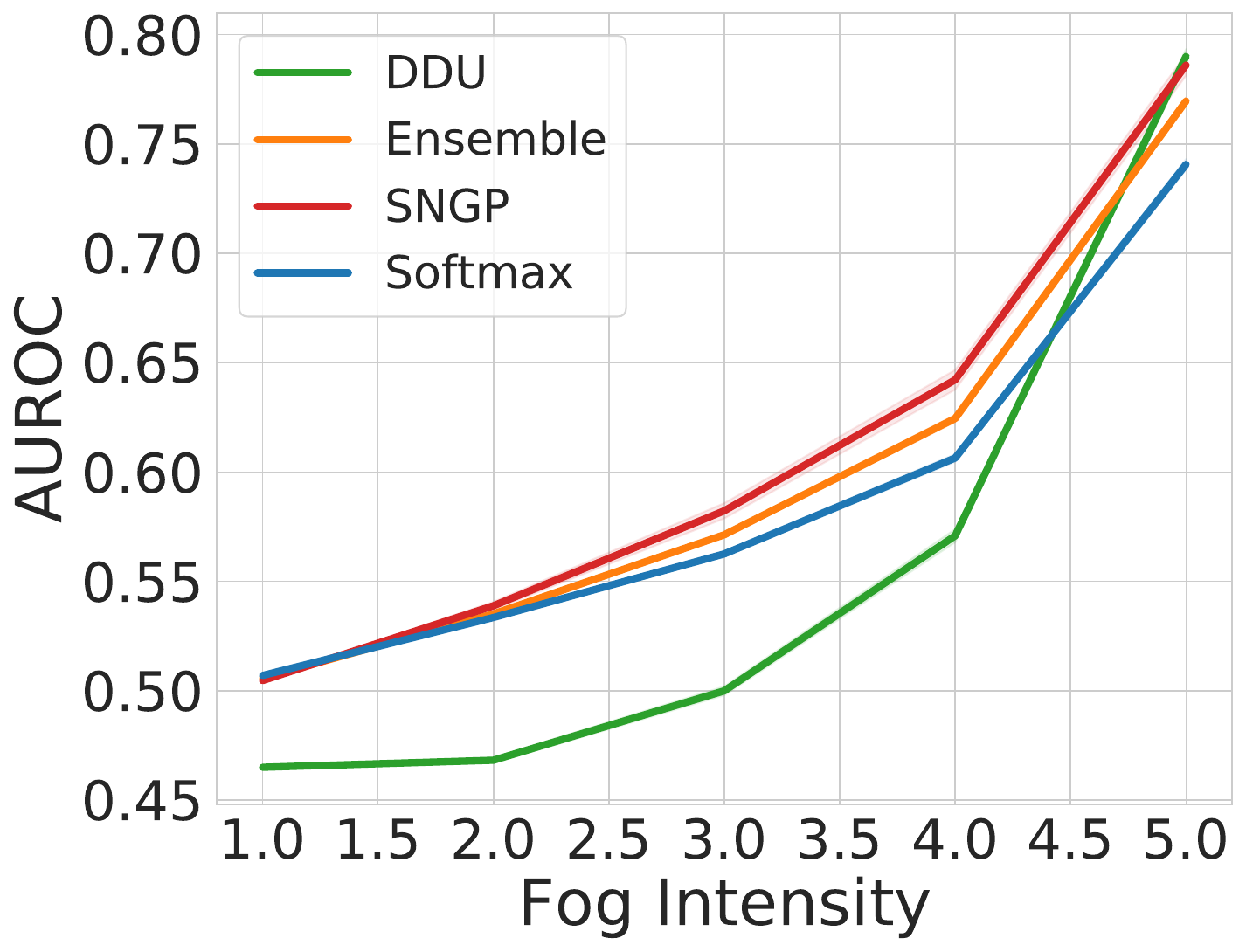}
    \end{subfigure}\hfill
    \begin{subfigure}{0.25\linewidth}
        \centering
        \includegraphics[width=\linewidth]{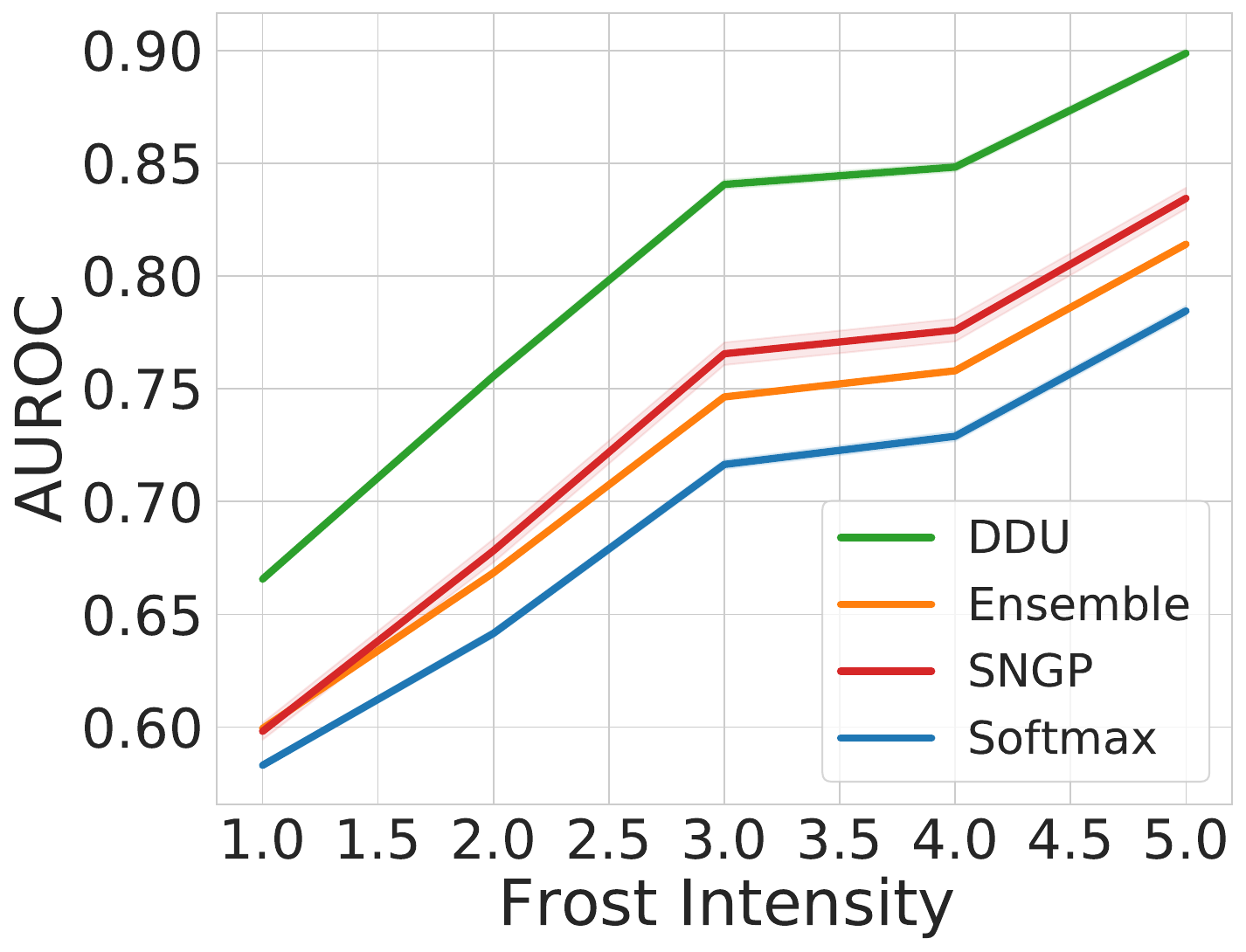}
    \end{subfigure}\hfill
    \begin{subfigure}{0.25\linewidth}
        \centering
        \includegraphics[width=\linewidth]{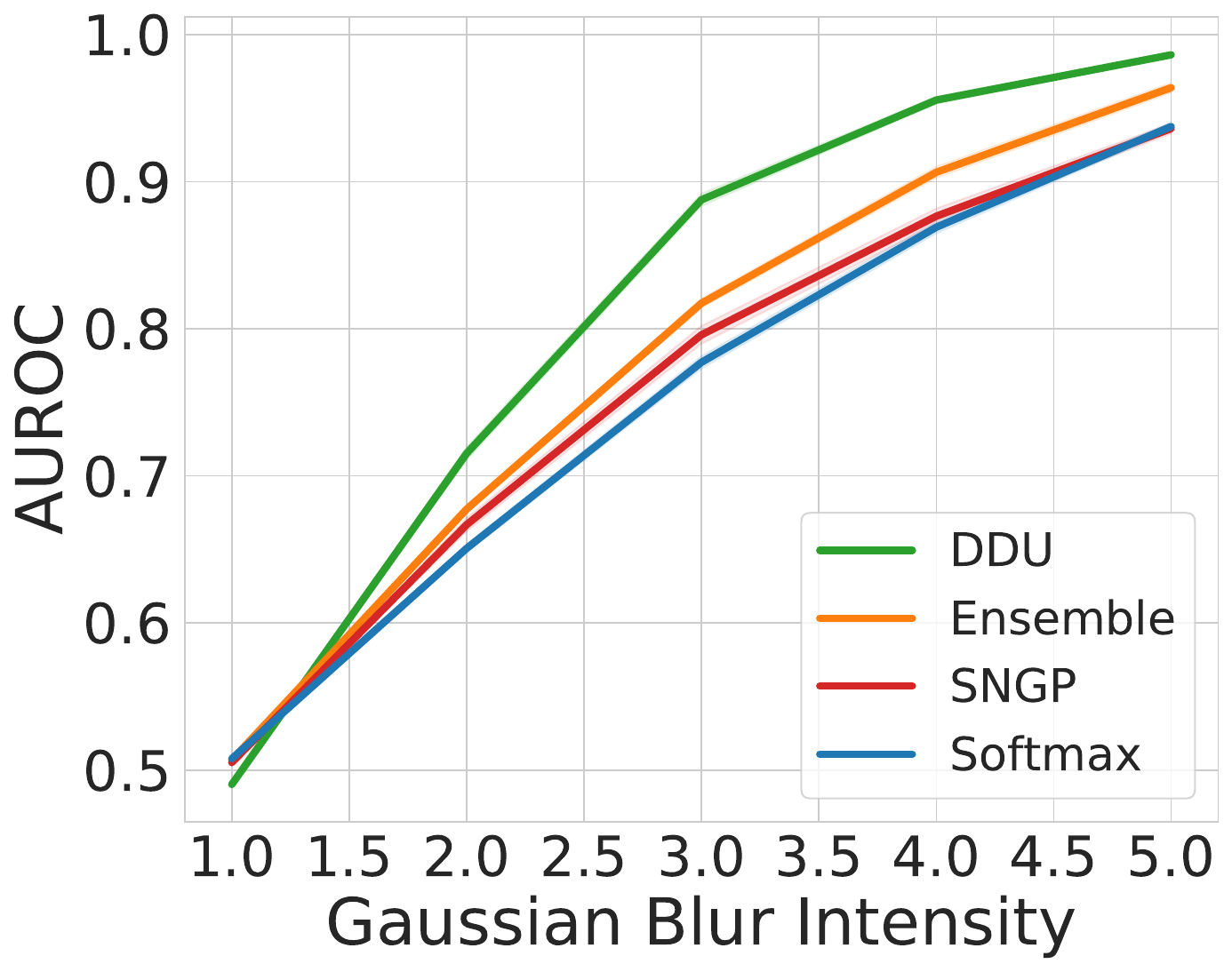}
    \end{subfigure}\hfill
    \begin{subfigure}{0.25\linewidth}
        \centering
        \includegraphics[width=\linewidth]{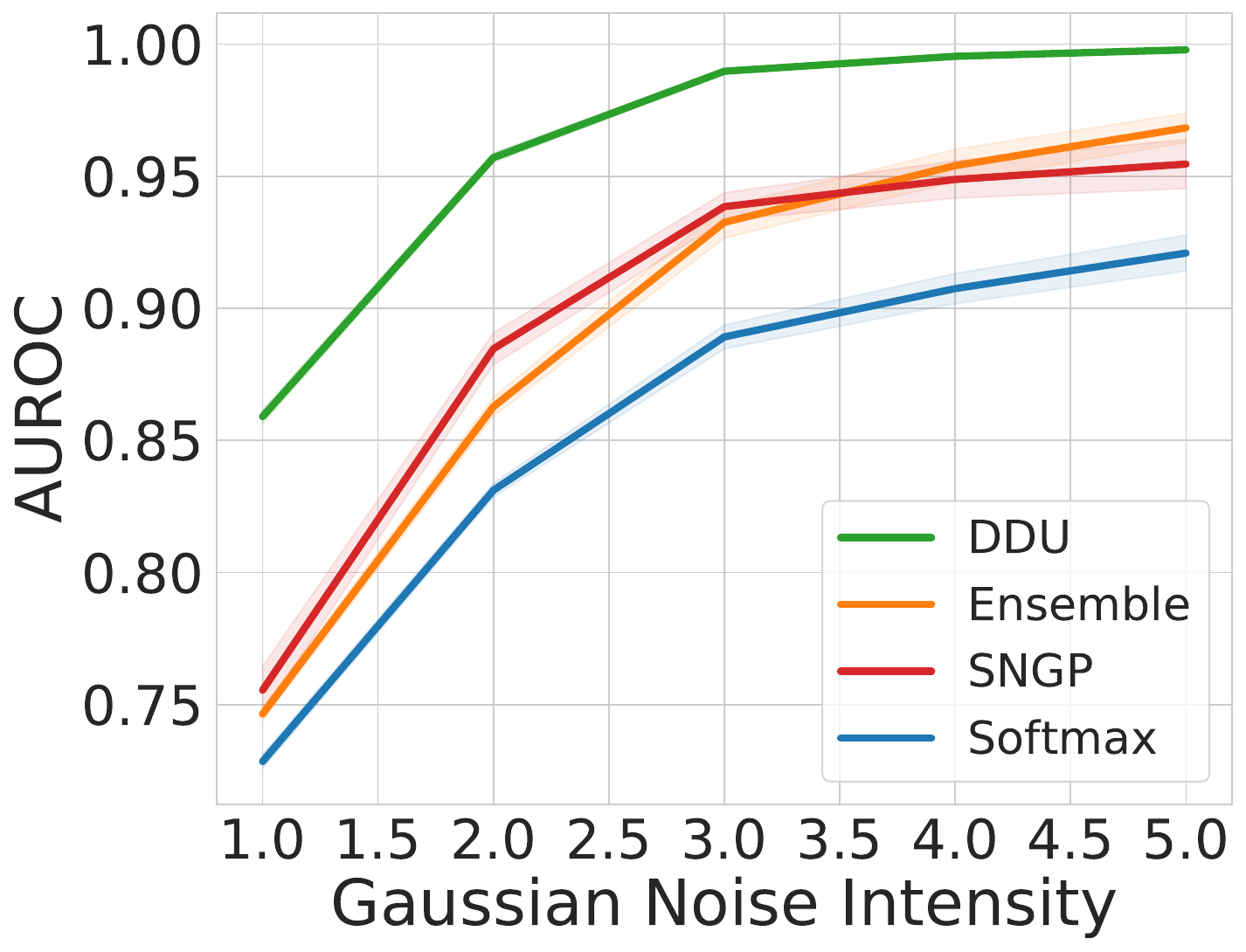}
    \end{subfigure}\hfill
    \begin{subfigure}{0.25\linewidth}
        \centering
        \includegraphics[width=\linewidth]{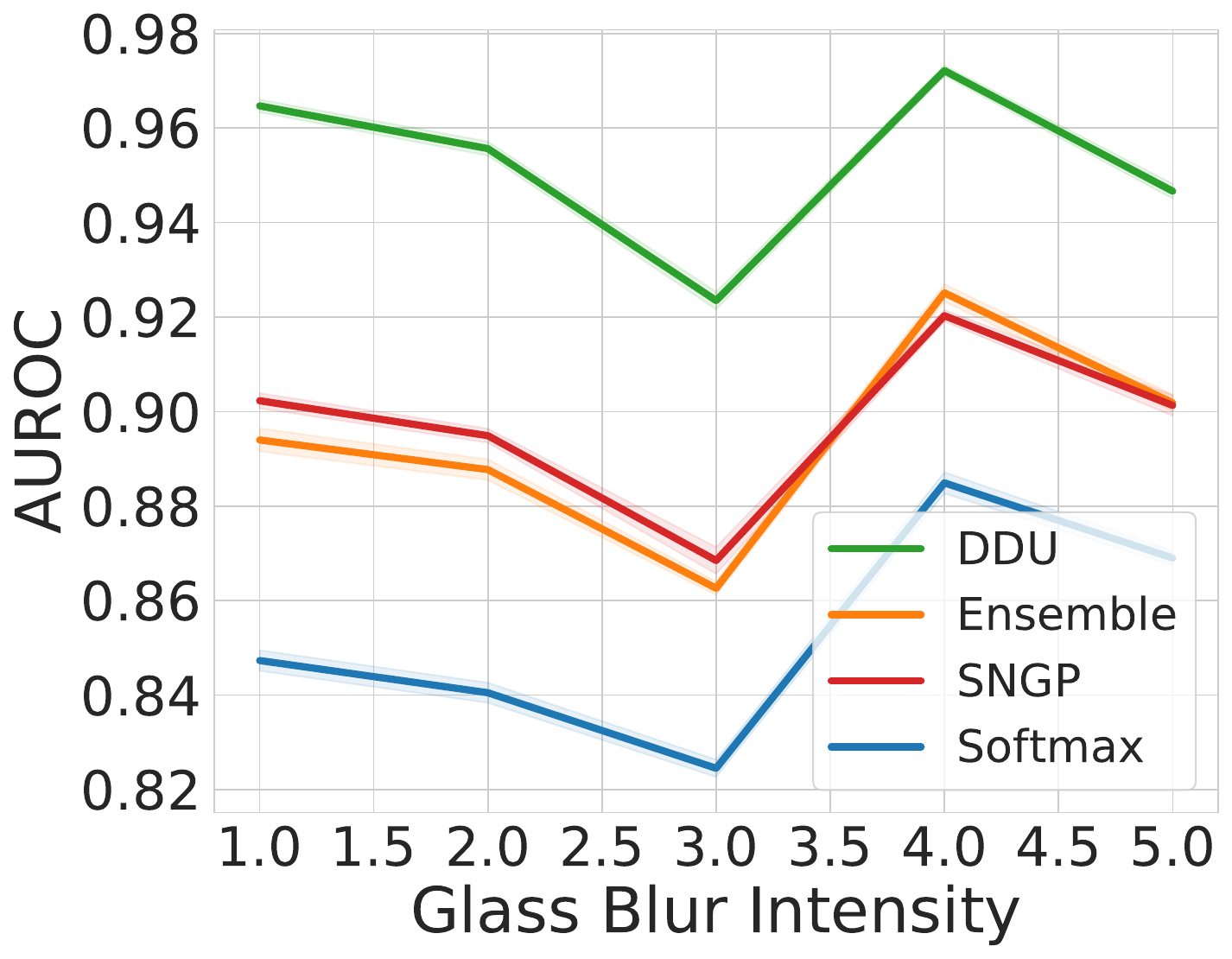}
    \end{subfigure}\hfill
    \begin{subfigure}{0.25\linewidth}
        \centering
        \includegraphics[width=\linewidth]{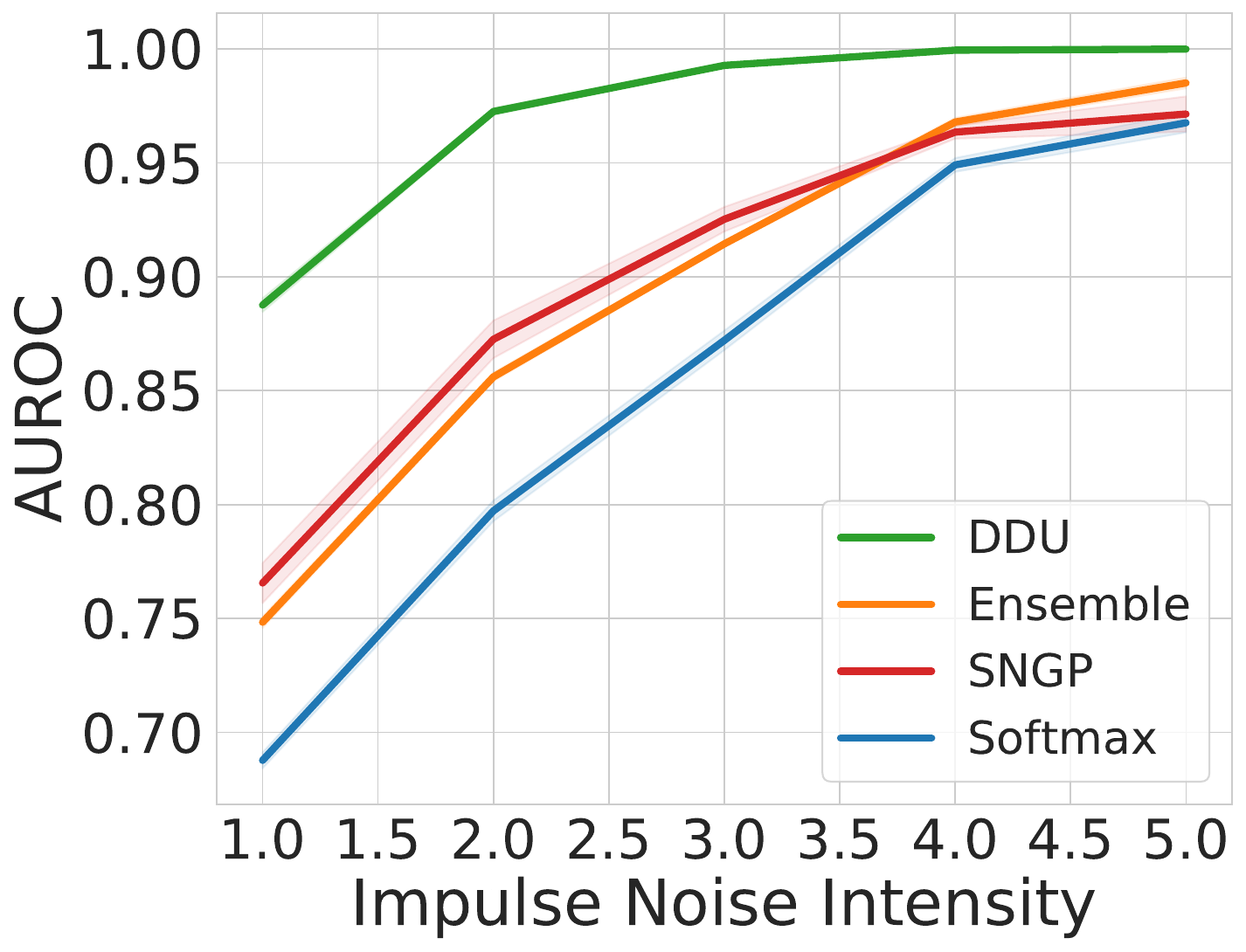}
    \end{subfigure}\hfill
    \begin{subfigure}{0.25\linewidth}
        \centering
        \includegraphics[width=\linewidth]{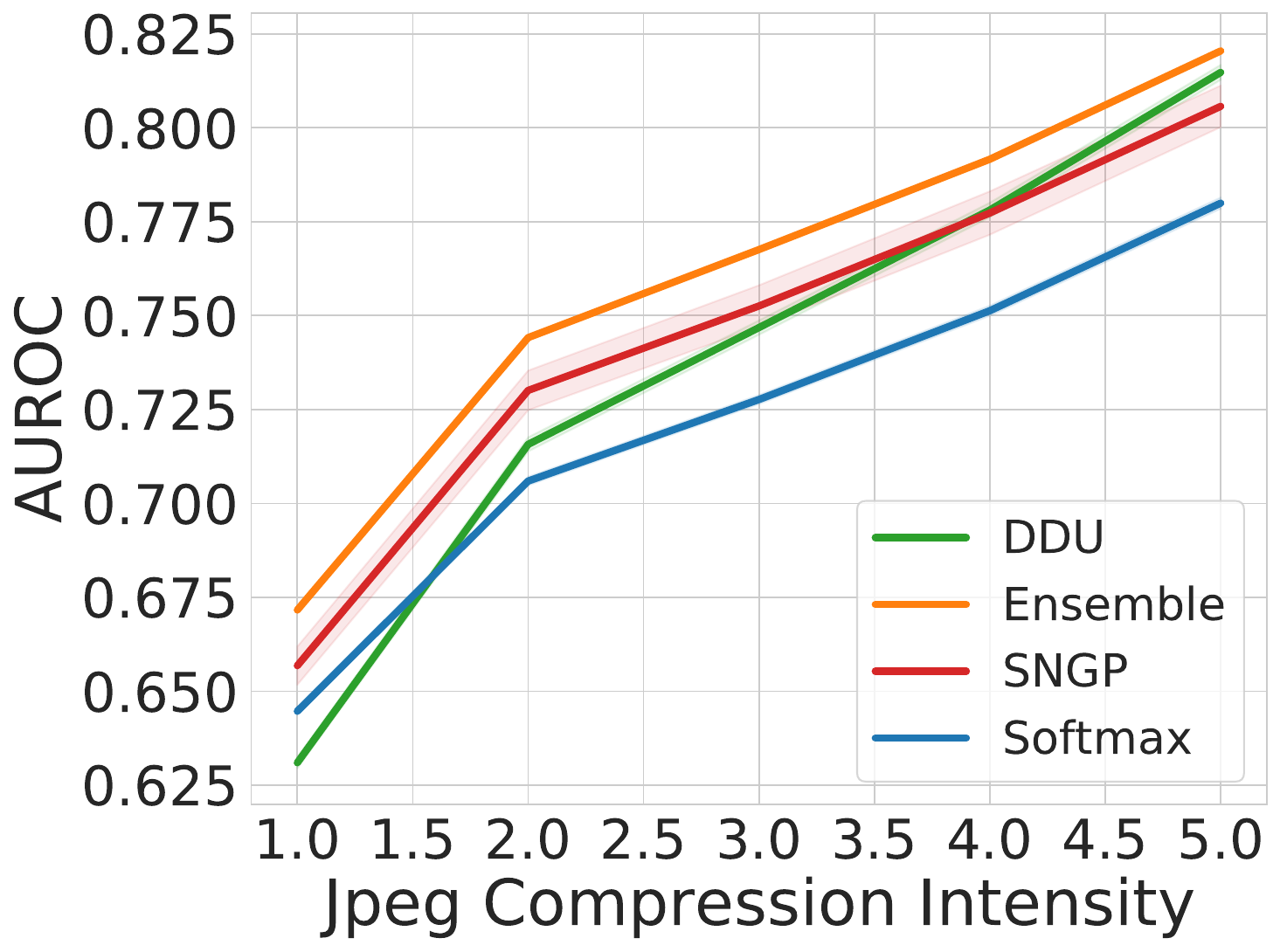}
    \end{subfigure}\hfill
    \begin{subfigure}{0.25\linewidth}
        \centering
        \includegraphics[width=\linewidth]{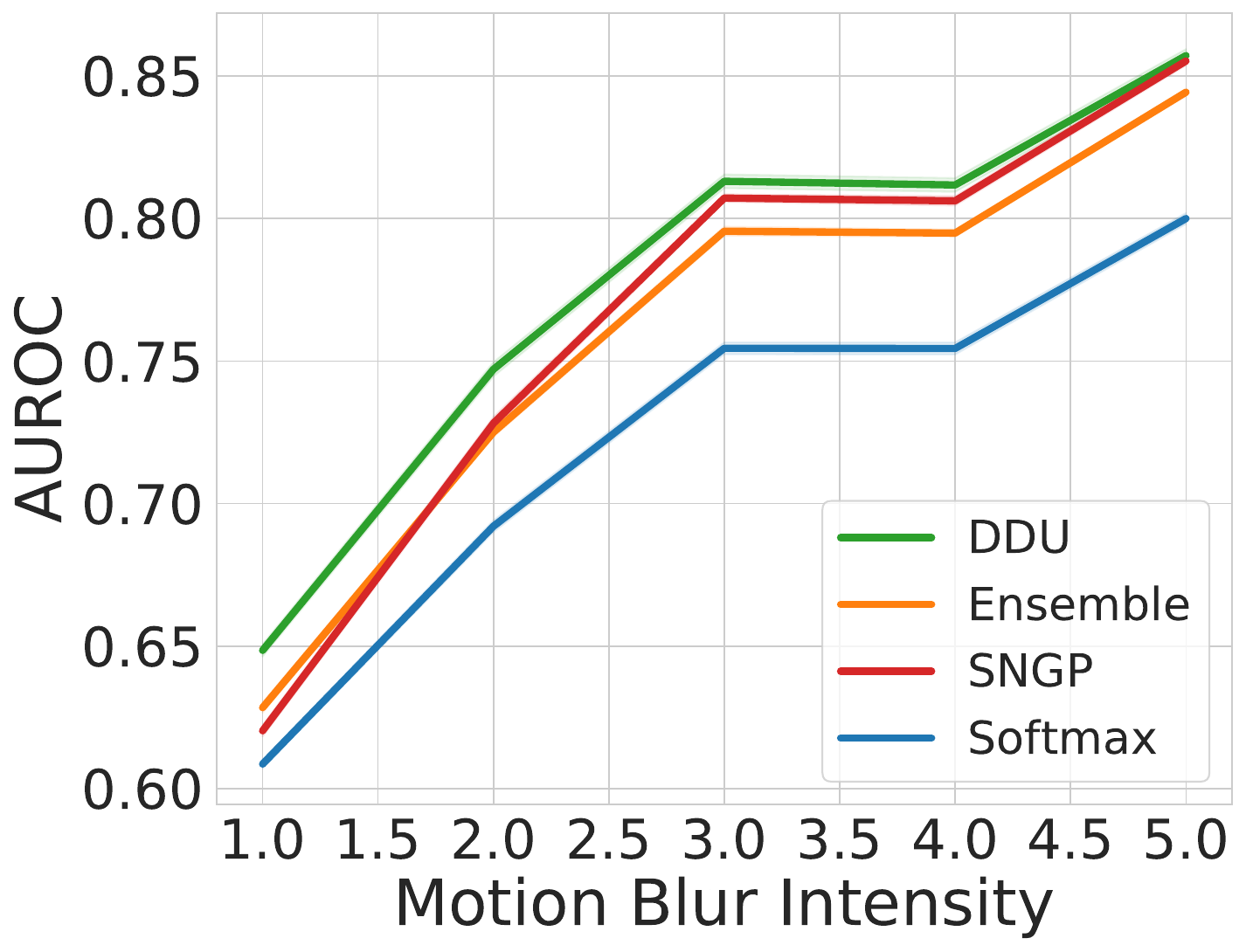}
    \end{subfigure}\hfill
    \begin{subfigure}{0.25\linewidth}
        \centering
        \includegraphics[width=\linewidth]{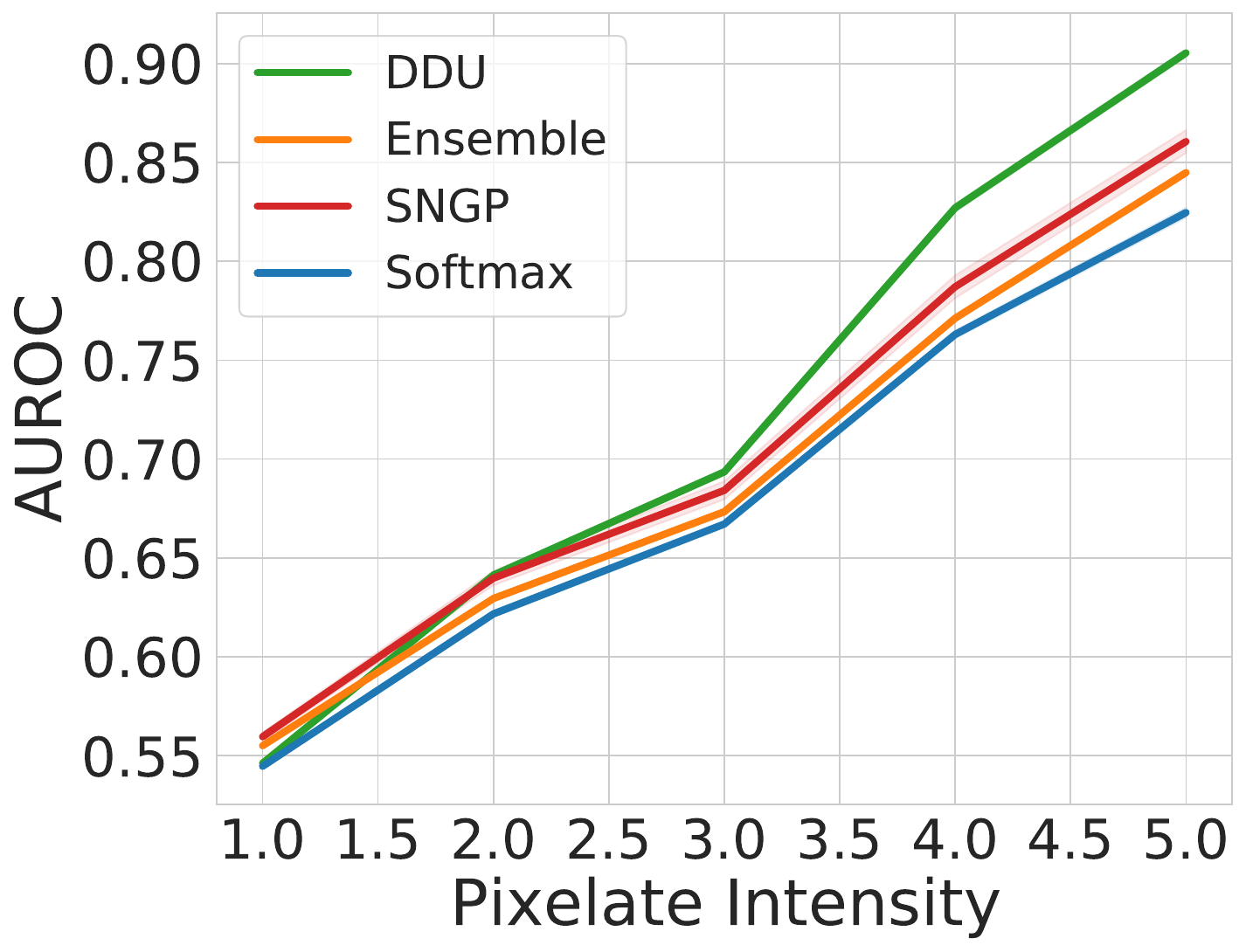}
    \end{subfigure}\hfill
    \begin{subfigure}{0.25\linewidth}
        \centering
        \includegraphics[width=\linewidth]{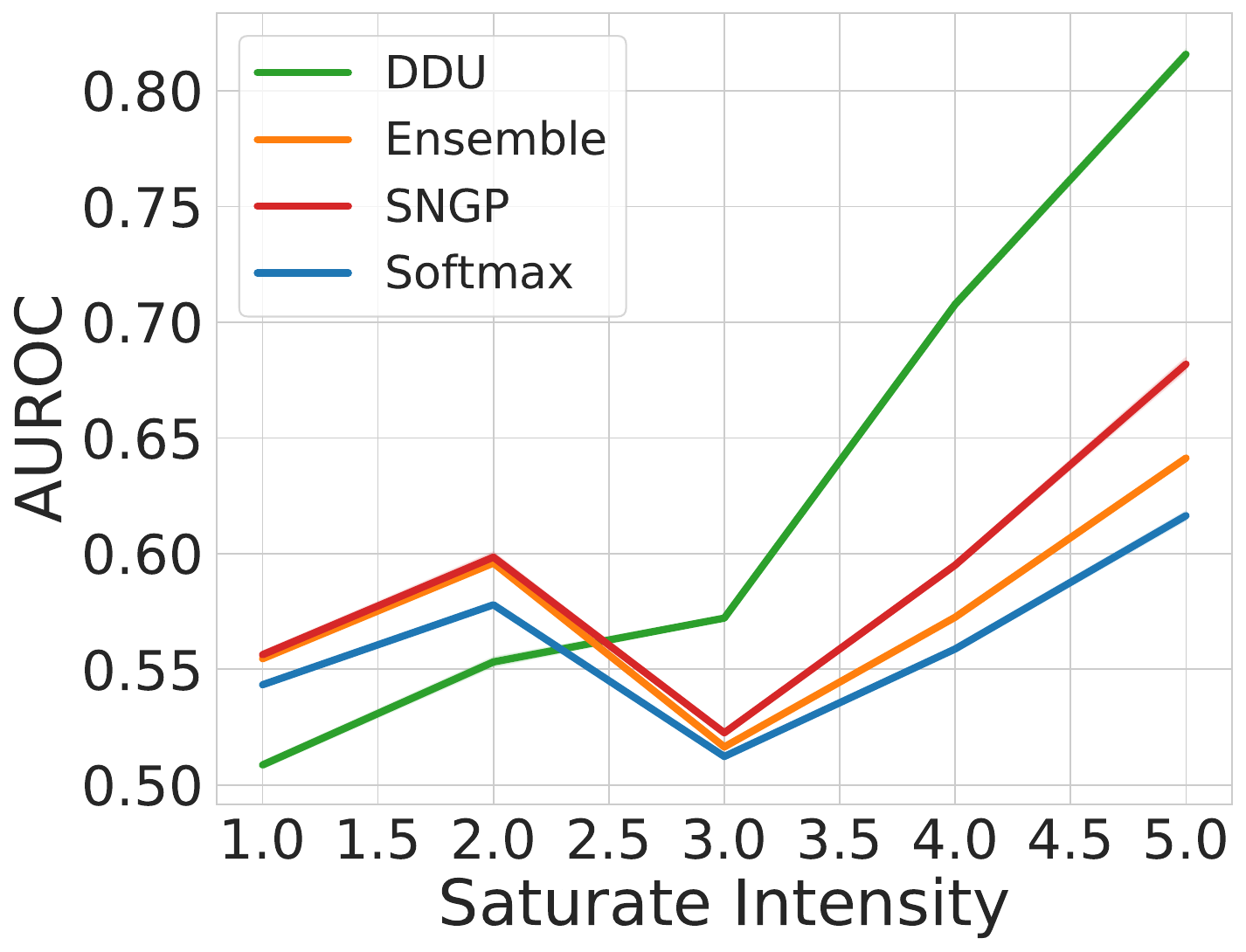}
    \end{subfigure}\hfill
    \begin{subfigure}{0.25\linewidth}
        \centering
        \includegraphics[width=\linewidth]{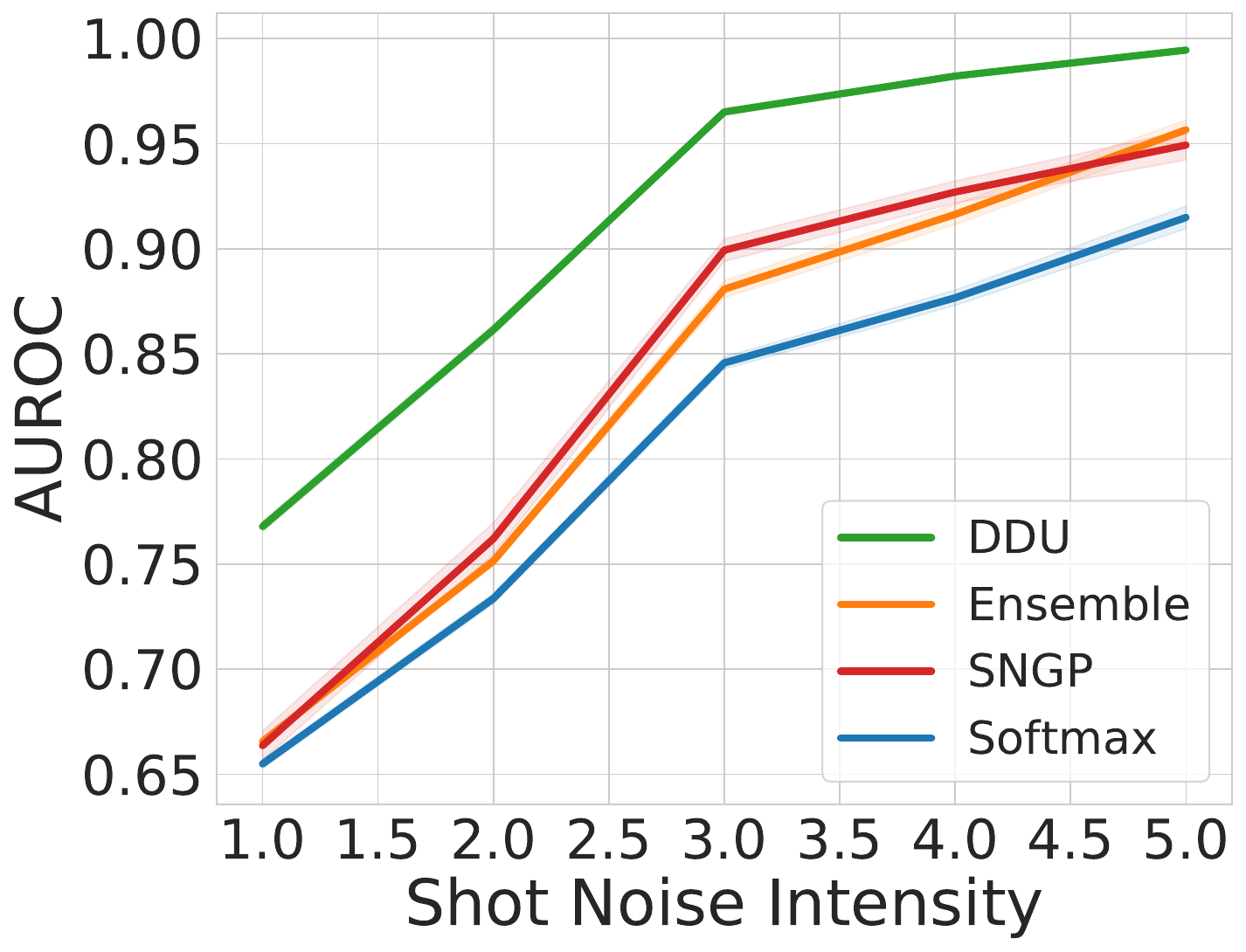}
    \end{subfigure}\hfill
    \begin{subfigure}{0.25\linewidth}
        \centering
        \includegraphics[width=\linewidth]{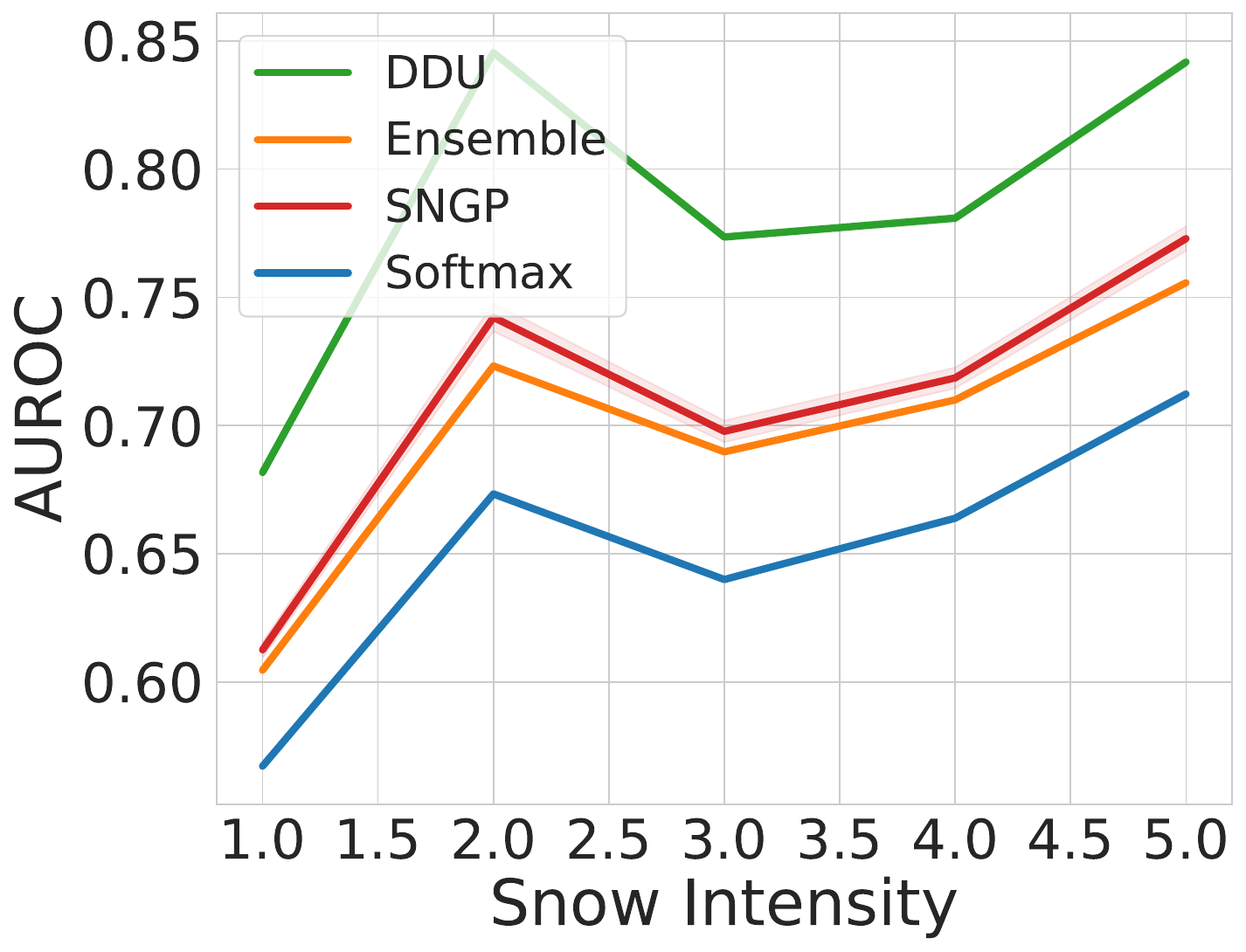}
    \end{subfigure}\hfill
    \begin{subfigure}{0.25\linewidth}
        \centering
        \includegraphics[width=\linewidth]{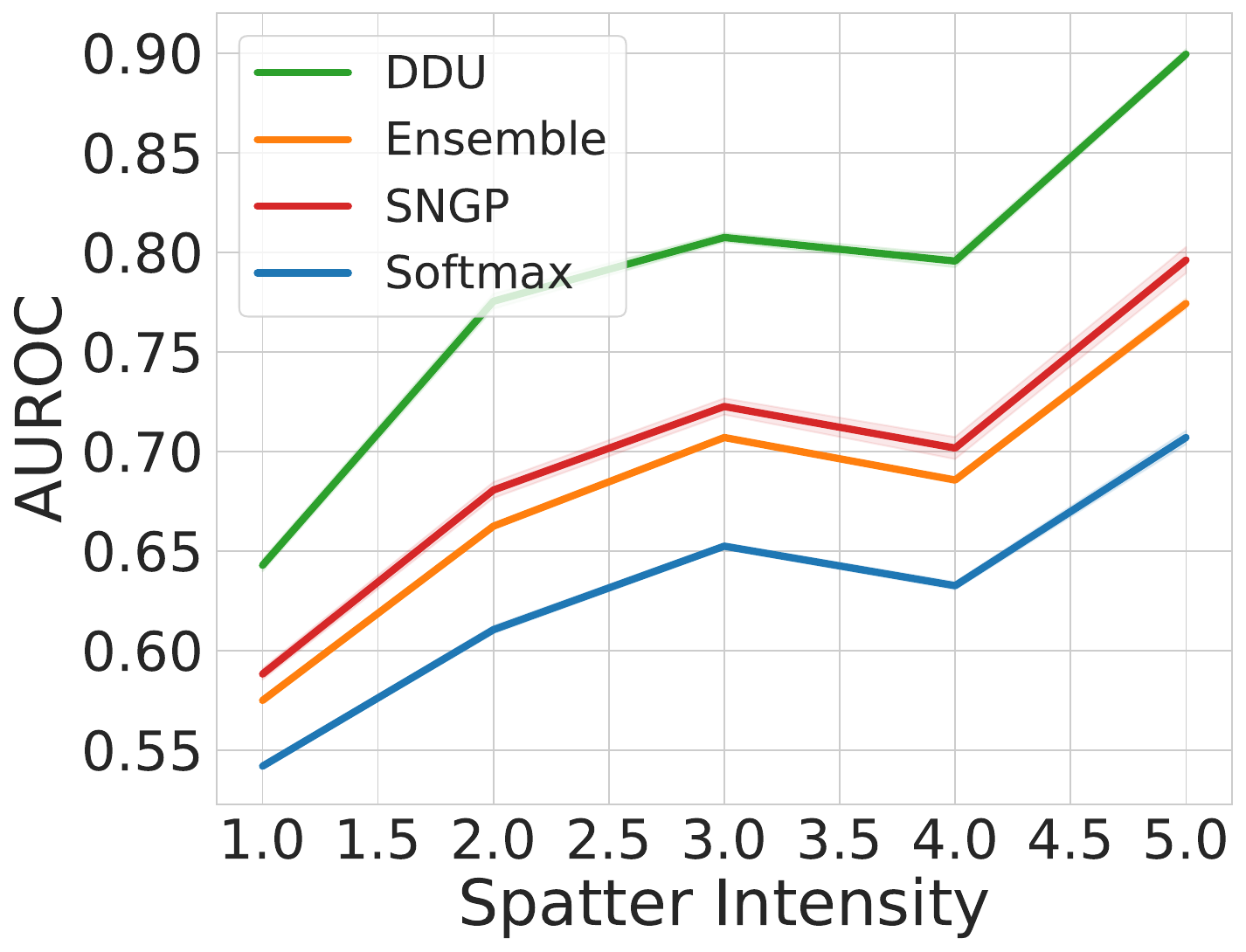}
    \end{subfigure}
    \begin{subfigure}{0.25\linewidth}
        \centering
        \includegraphics[width=\linewidth]{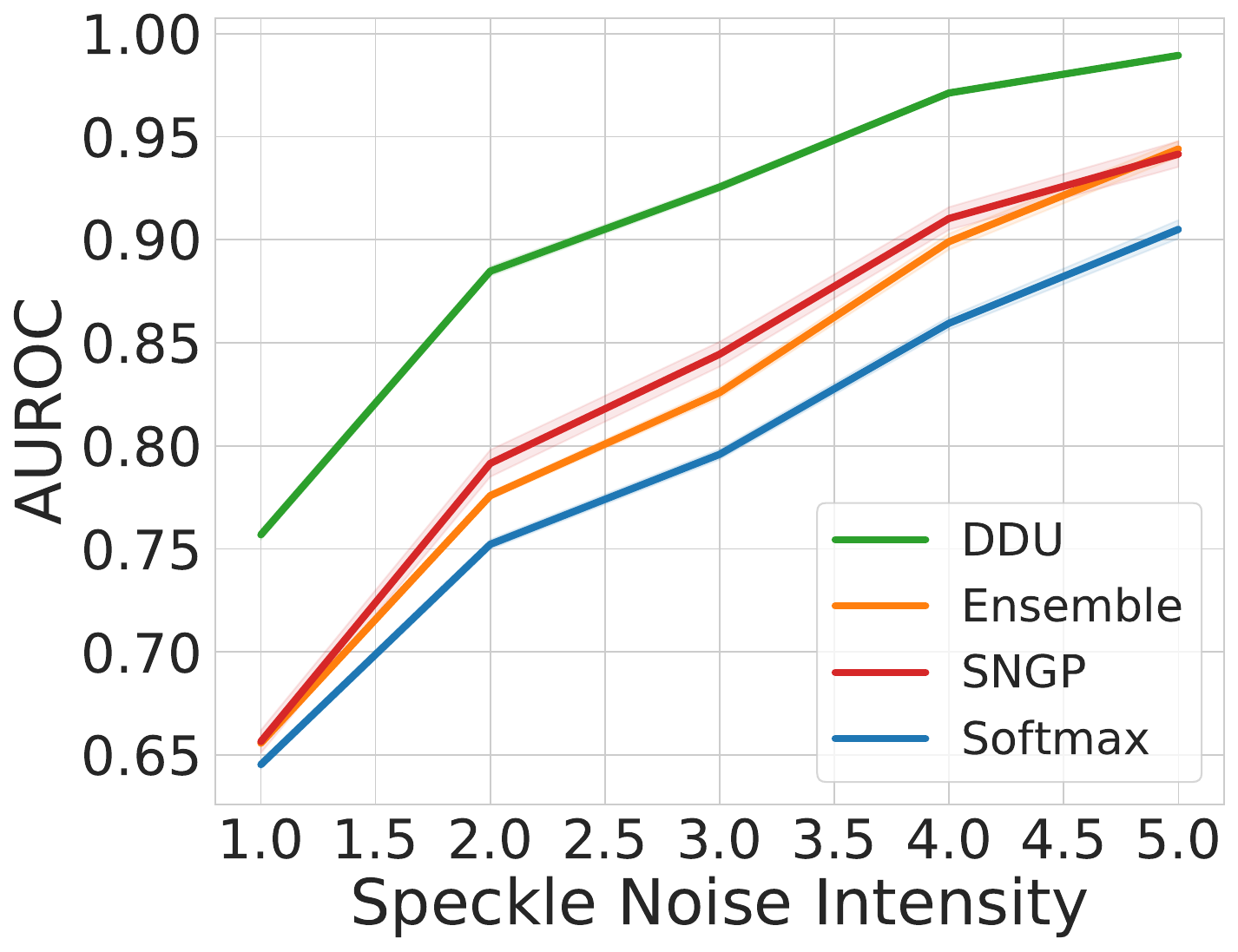}
    \end{subfigure}
    \begin{subfigure}{0.25\linewidth}
        \centering
        \includegraphics[width=\linewidth]{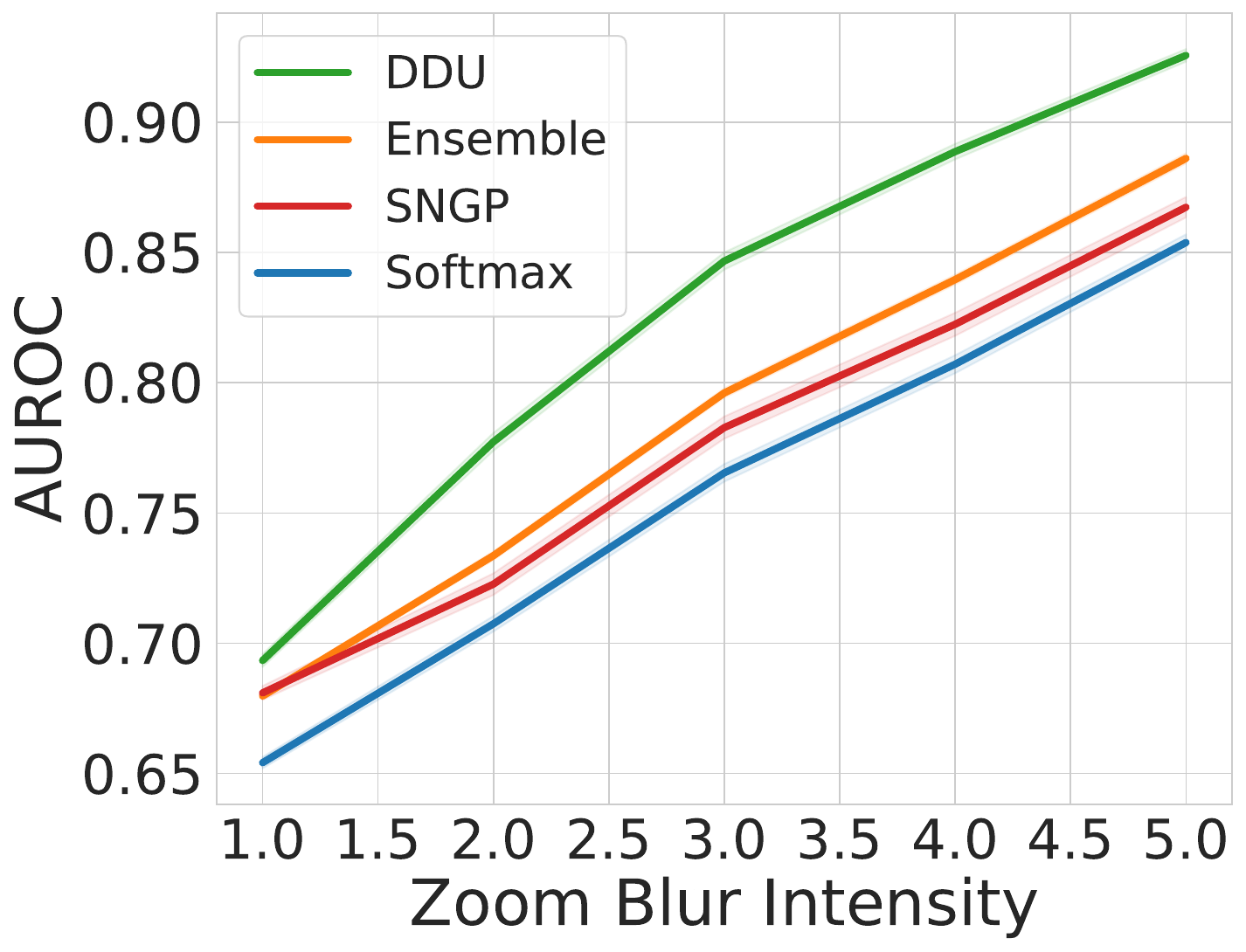}
    \end{subfigure}
    \caption{
    AUROC vs corruption intensity for all corruption types in CIFAR-10-C with Wide-ResNet-28-10 as the architecture and baselines: Softmax Entropy, Ensemble (using Predictive Entropy as uncertainty), SNGP and DDU feature density.
    }
    \label{ddu:fig:cifar10_c_results_wide_resnet}
\end{figure}

\begin{figure}[!t]
    \centering
    \begin{subfigure}{0.25\linewidth}
        \centering
        \includegraphics[width=\linewidth]{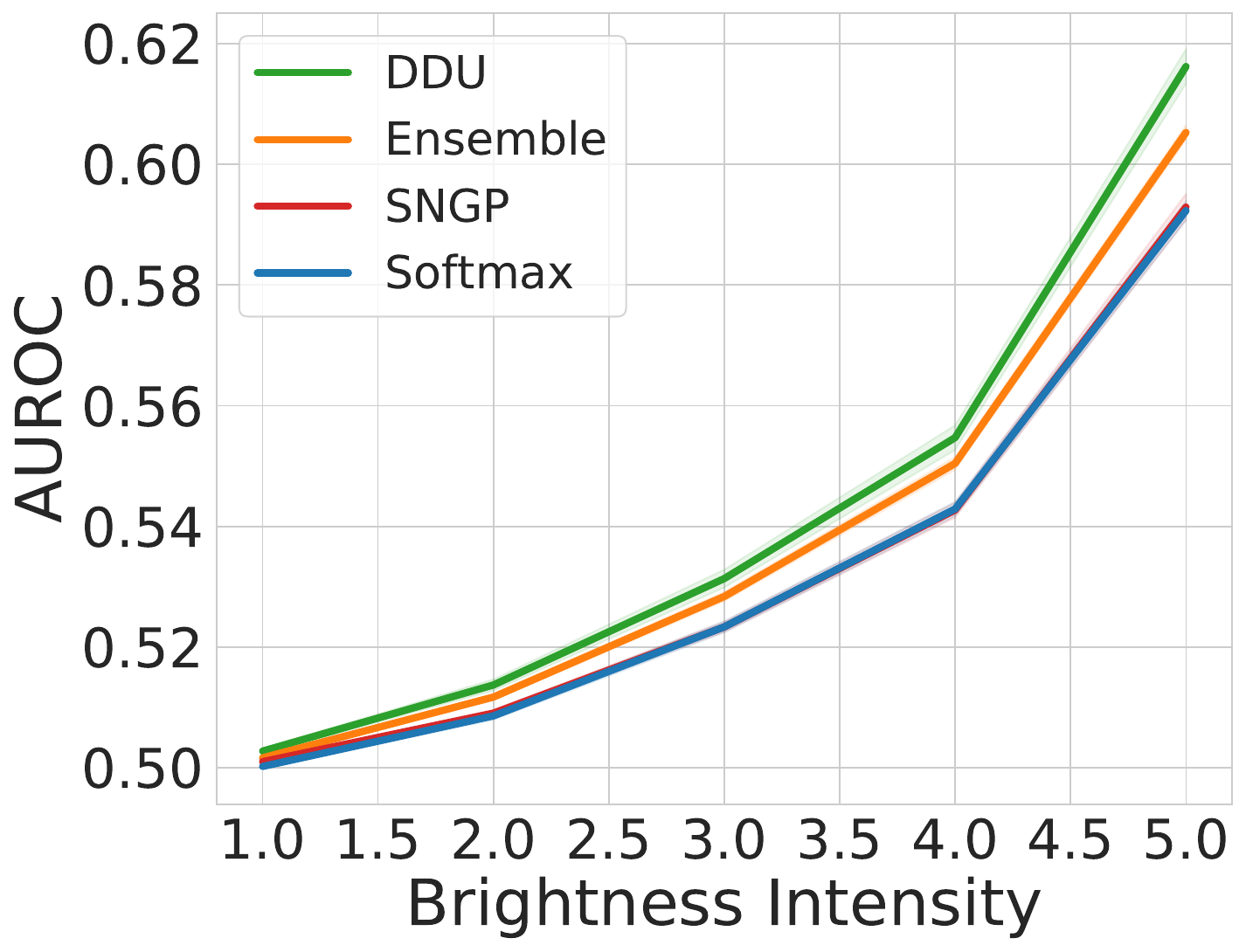}
    \end{subfigure}\hfill
    \begin{subfigure}{0.25\linewidth}
        \centering
        \includegraphics[width=\linewidth]{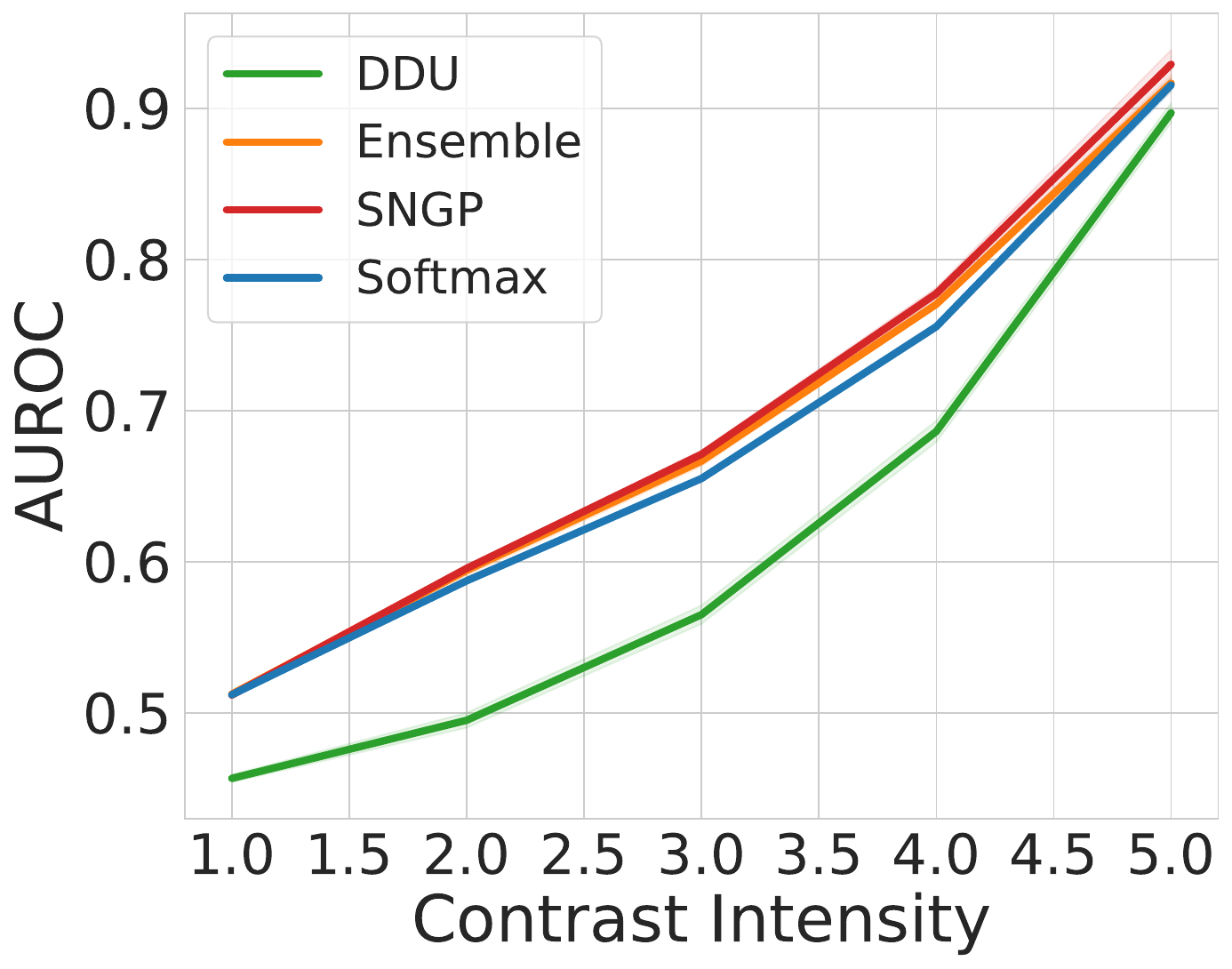}
    \end{subfigure}\hfill 
    \begin{subfigure}{0.25\linewidth}
        \centering
        \includegraphics[width=\linewidth]{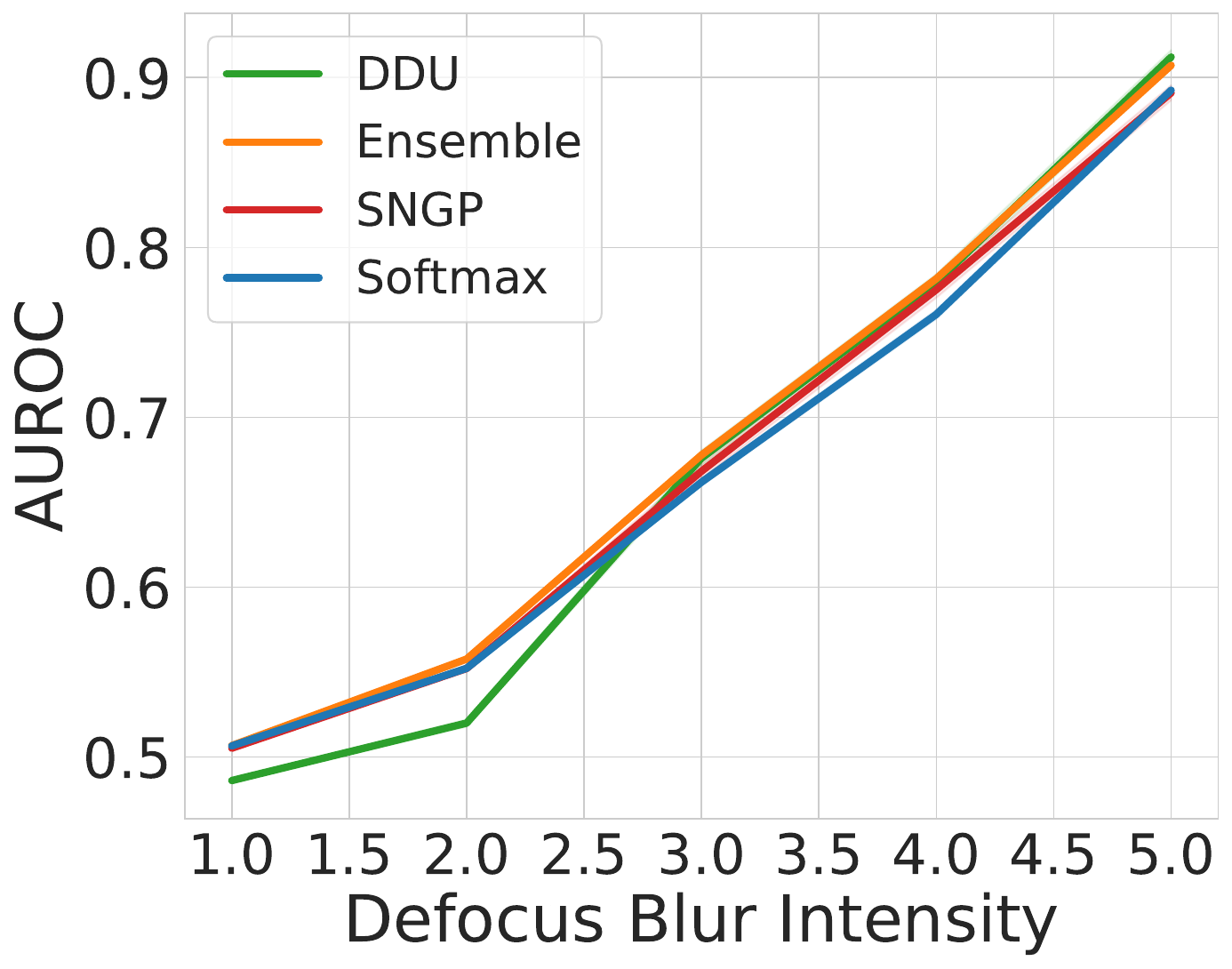}
    \end{subfigure}\hfill 
    \begin{subfigure}{0.25\linewidth}
        \centering
        \includegraphics[width=\linewidth]{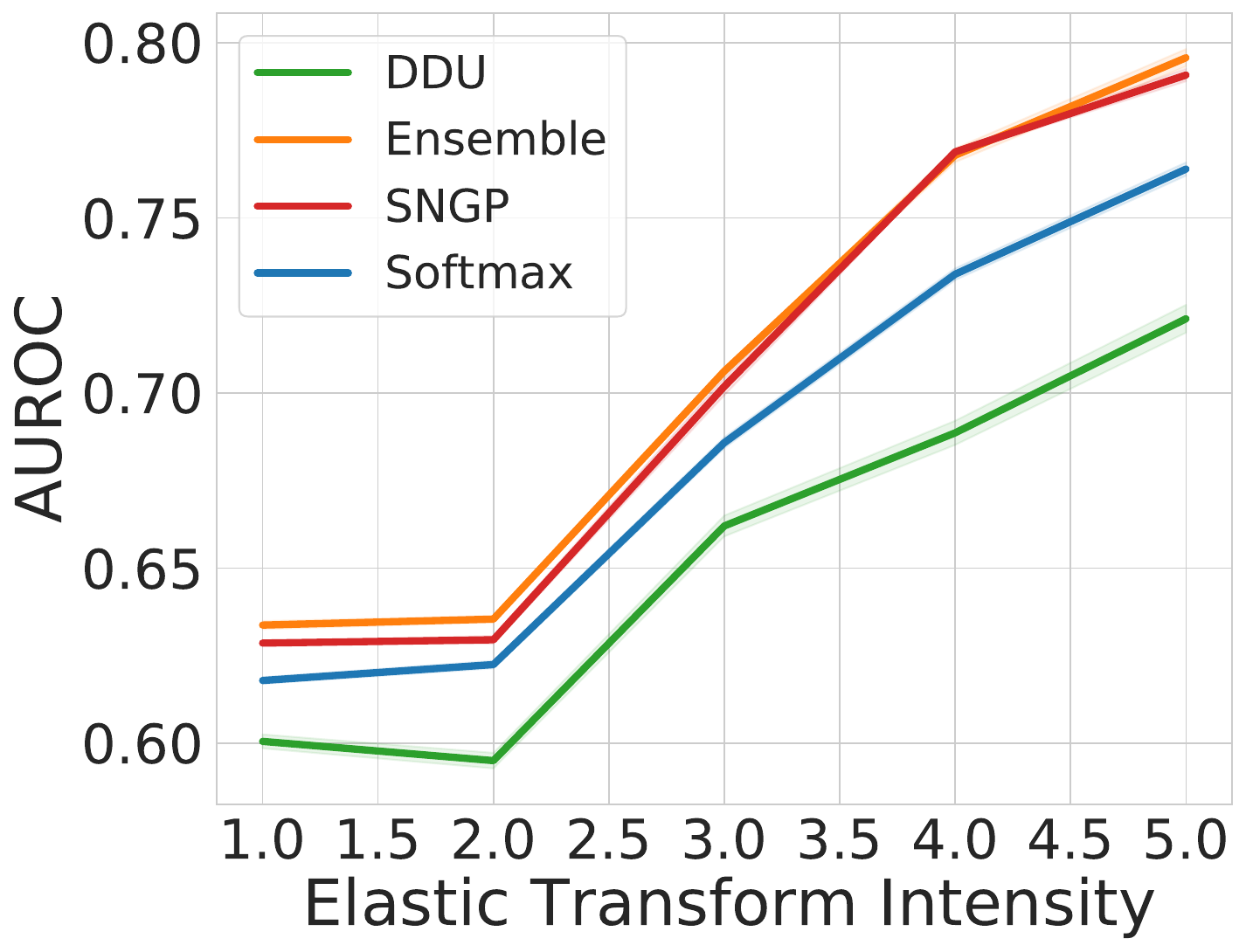}
    \end{subfigure}\hfill
    \begin{subfigure}{0.25\linewidth}
        \centering
        \includegraphics[width=\linewidth]{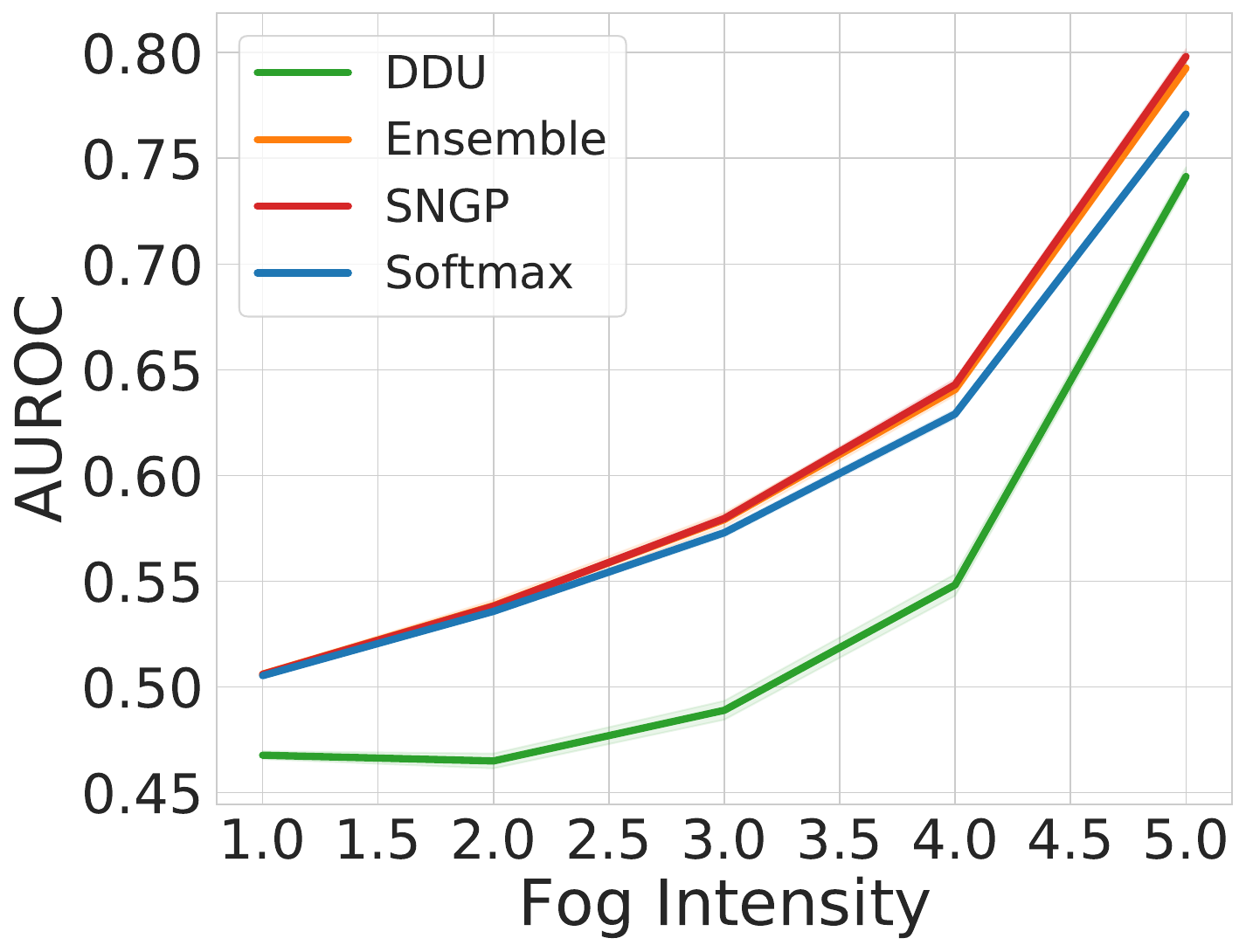}
    \end{subfigure}\hfill
    \begin{subfigure}{0.25\linewidth}
        \centering
        \includegraphics[width=\linewidth]{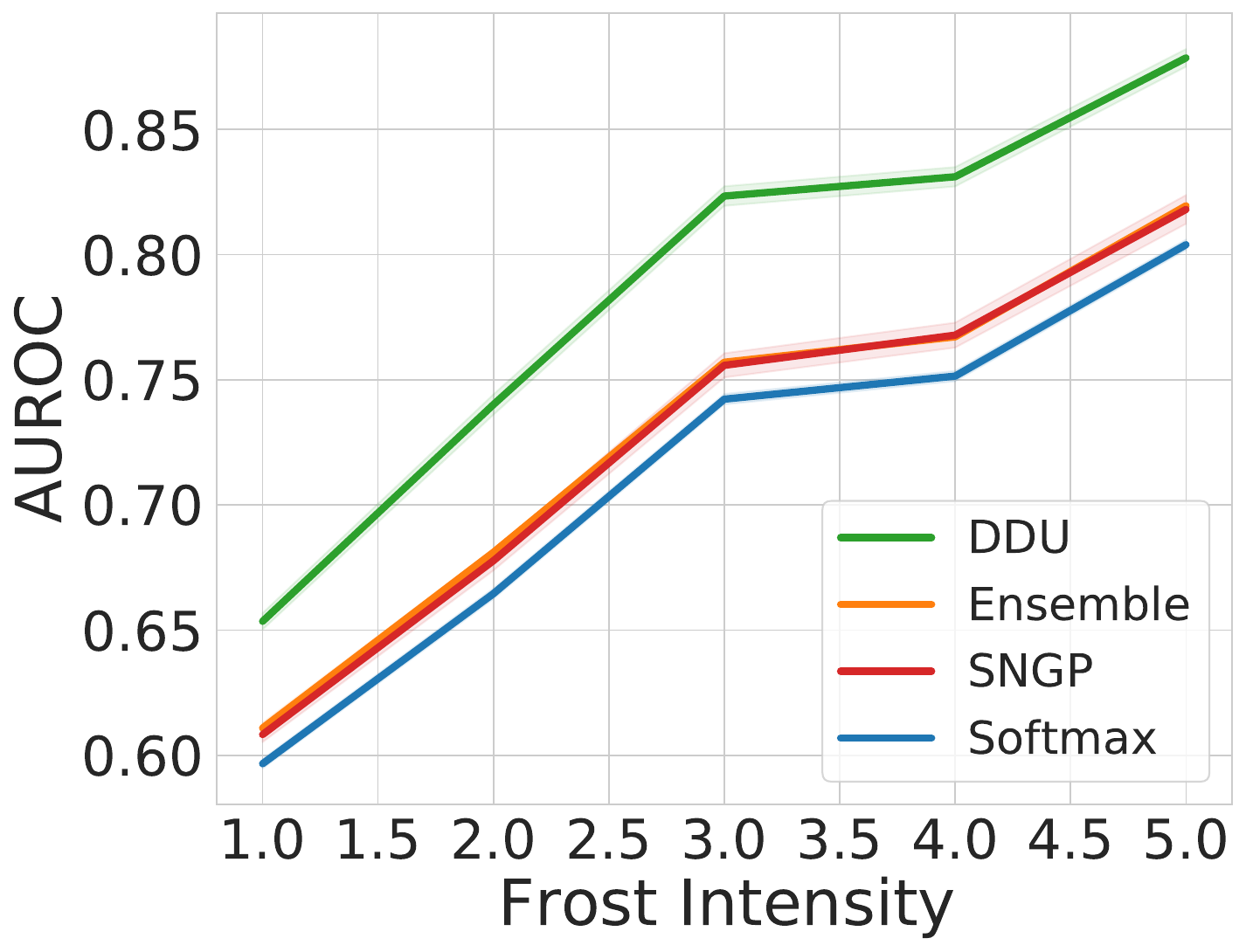}
    \end{subfigure}\hfill
    \begin{subfigure}{0.25\linewidth}
        \centering
        \includegraphics[width=\linewidth]{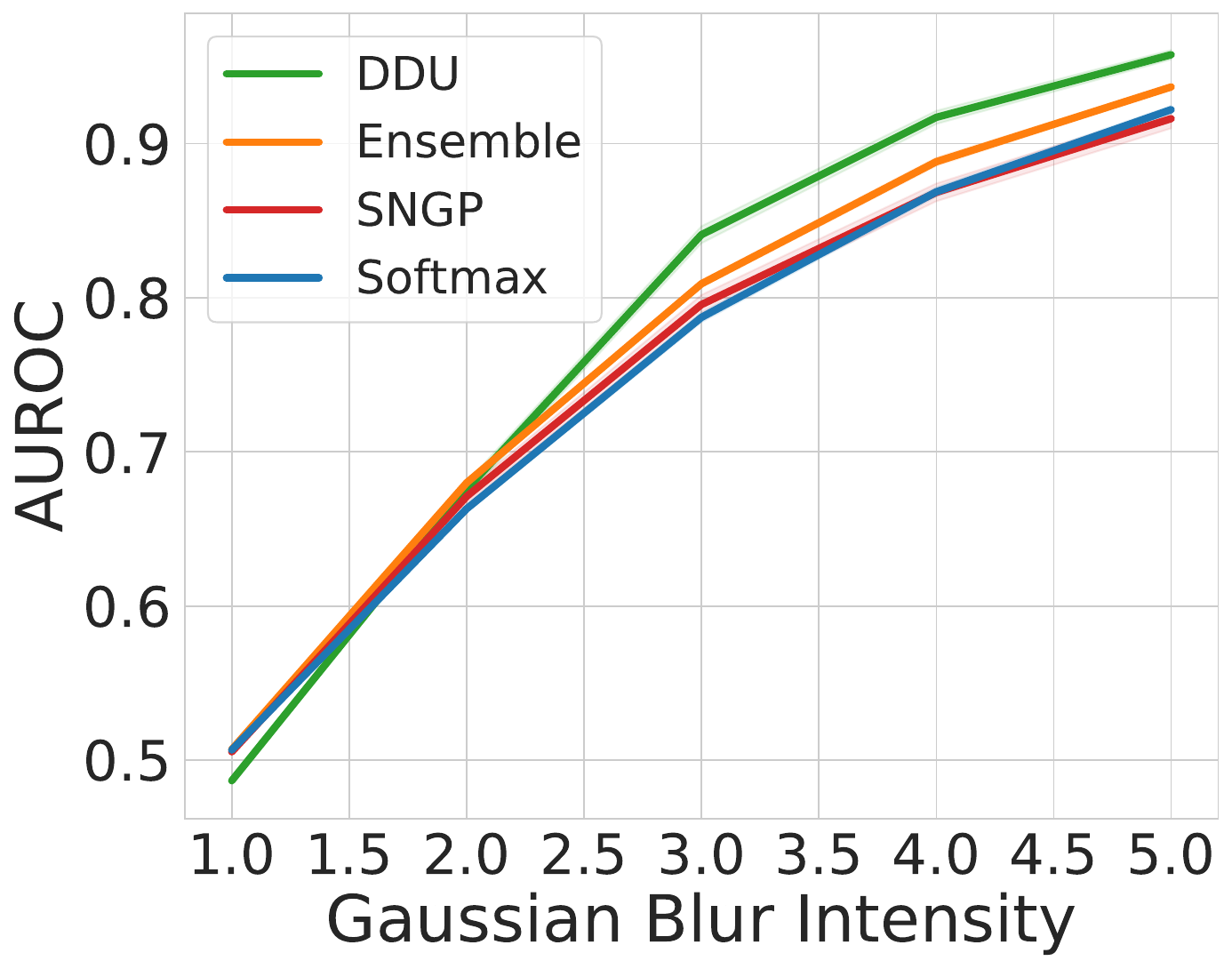}
    \end{subfigure}\hfill
    \begin{subfigure}{0.25\linewidth}
        \centering
        \includegraphics[width=\linewidth]{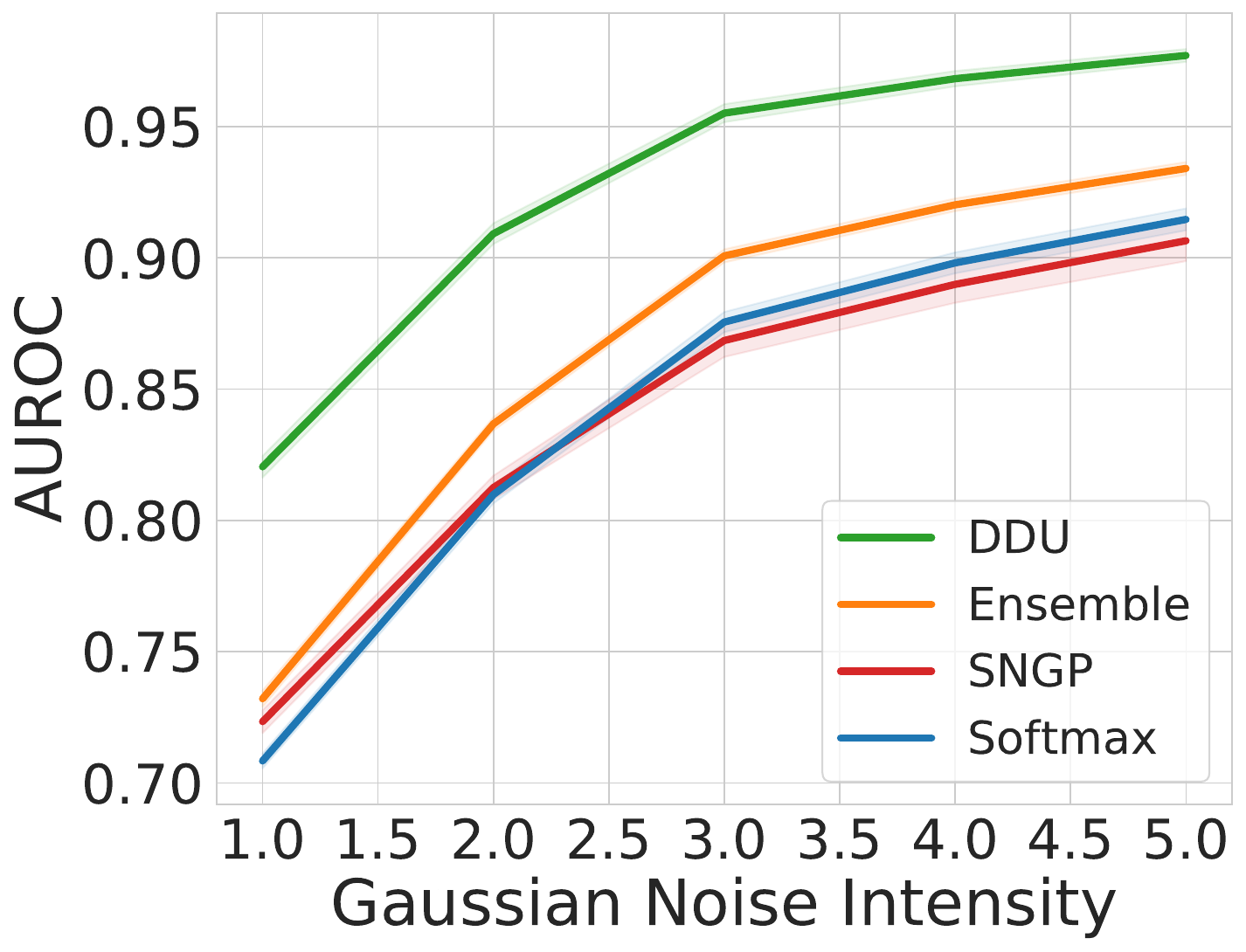}
    \end{subfigure}\hfill
    \begin{subfigure}{0.25\linewidth}
        \centering
        \includegraphics[width=\linewidth]{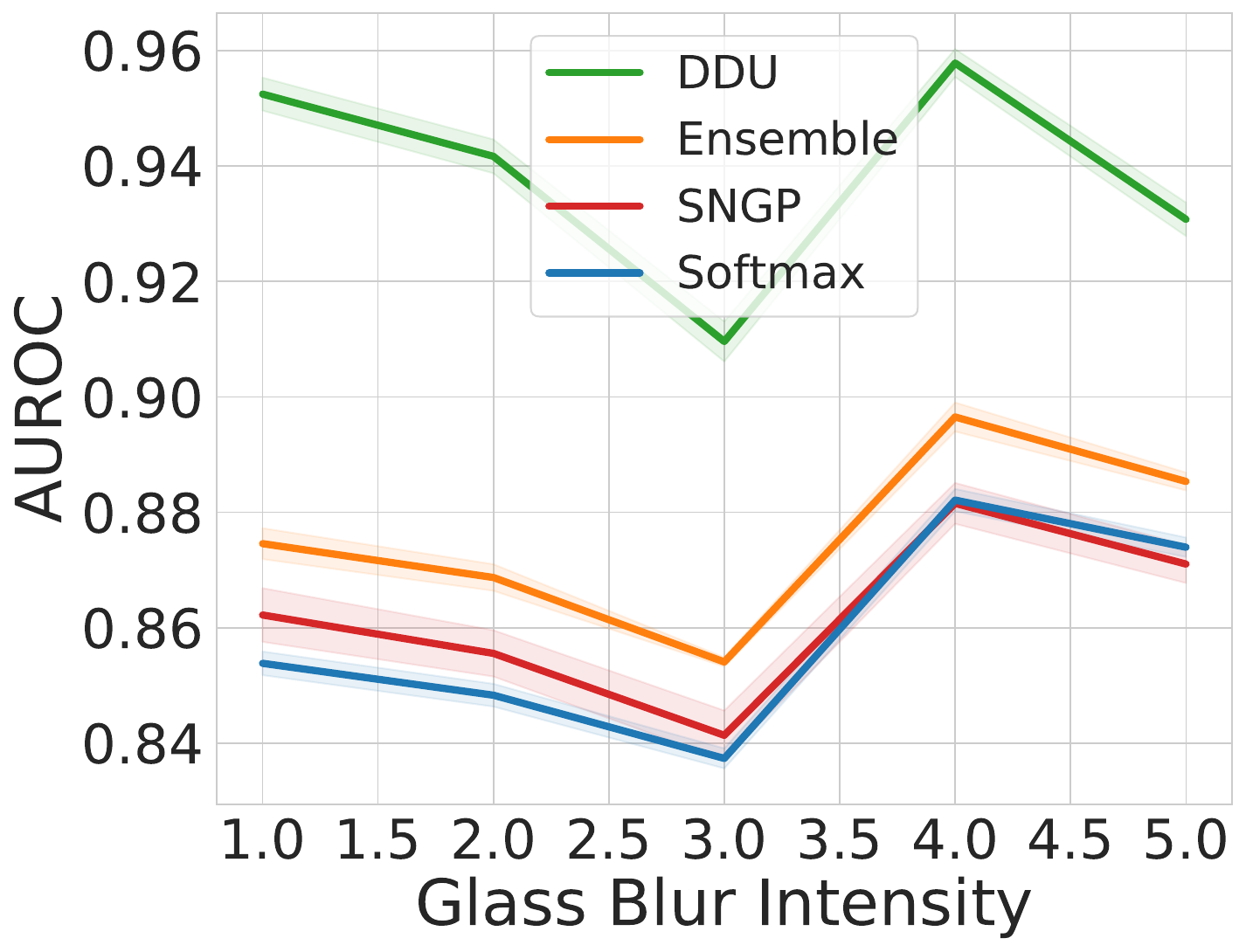}
    \end{subfigure}\hfill
    \begin{subfigure}{0.25\linewidth}
        \centering
        \includegraphics[width=\linewidth]{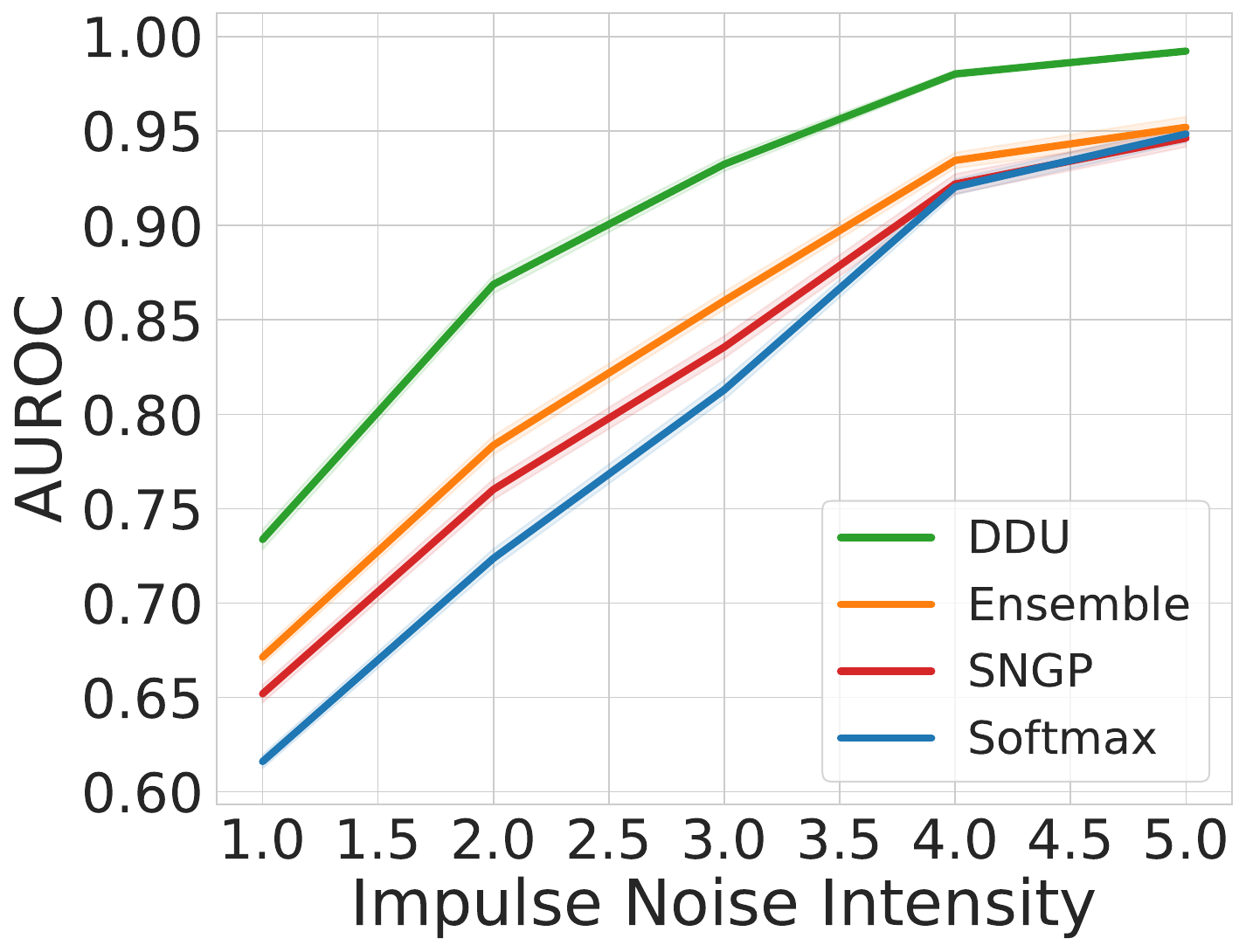}
    \end{subfigure}\hfill
    \begin{subfigure}{0.25\linewidth}
        \centering
        \includegraphics[width=\linewidth]{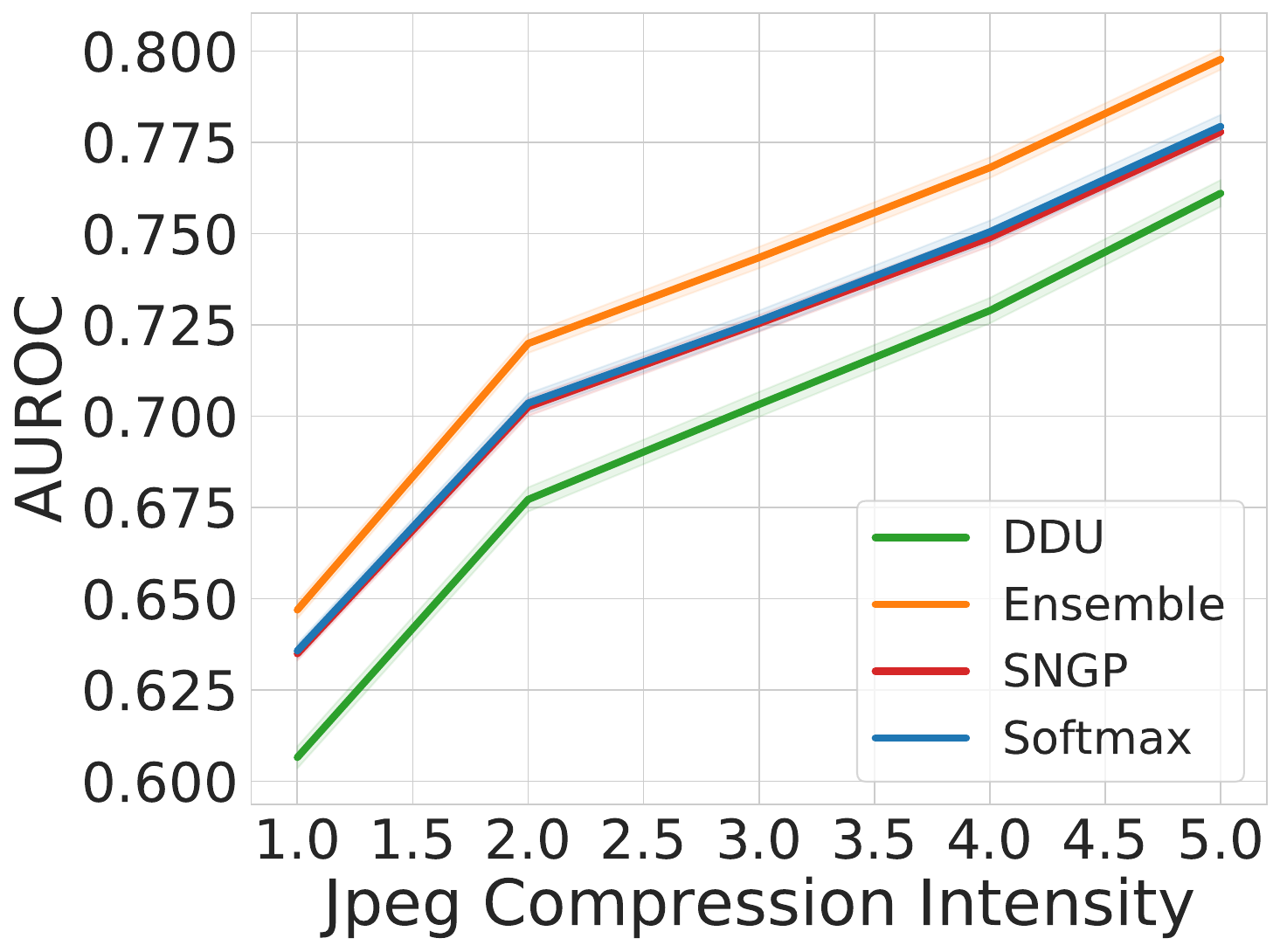}
    \end{subfigure}\hfill
    \begin{subfigure}{0.25\linewidth}
        \centering
        \includegraphics[width=\linewidth]{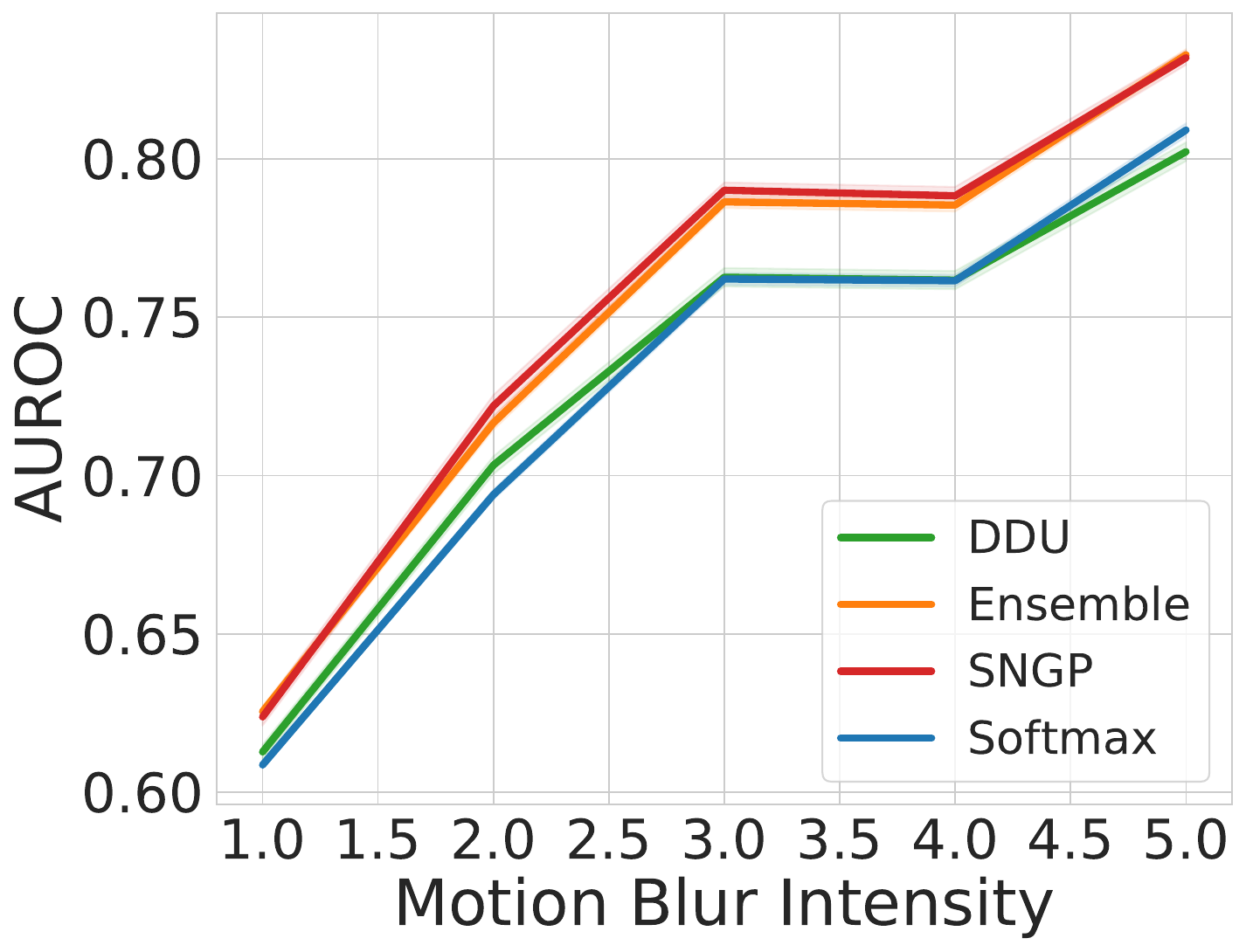}
    \end{subfigure}\hfill
    \begin{subfigure}{0.25\linewidth}
        \centering
        \includegraphics[width=\linewidth]{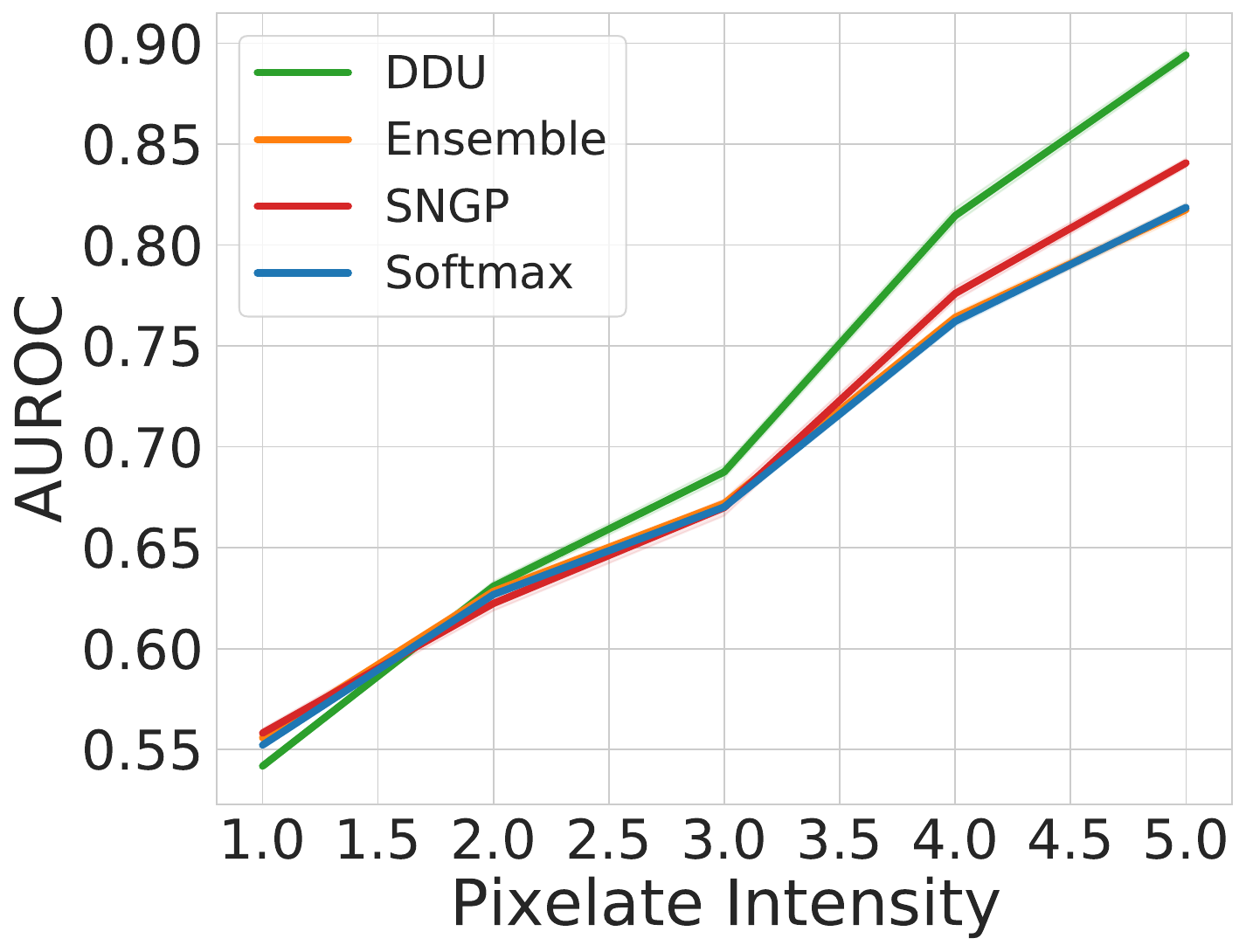}
    \end{subfigure}\hfill
    \begin{subfigure}{0.25\linewidth}
        \centering
        \includegraphics[width=\linewidth]{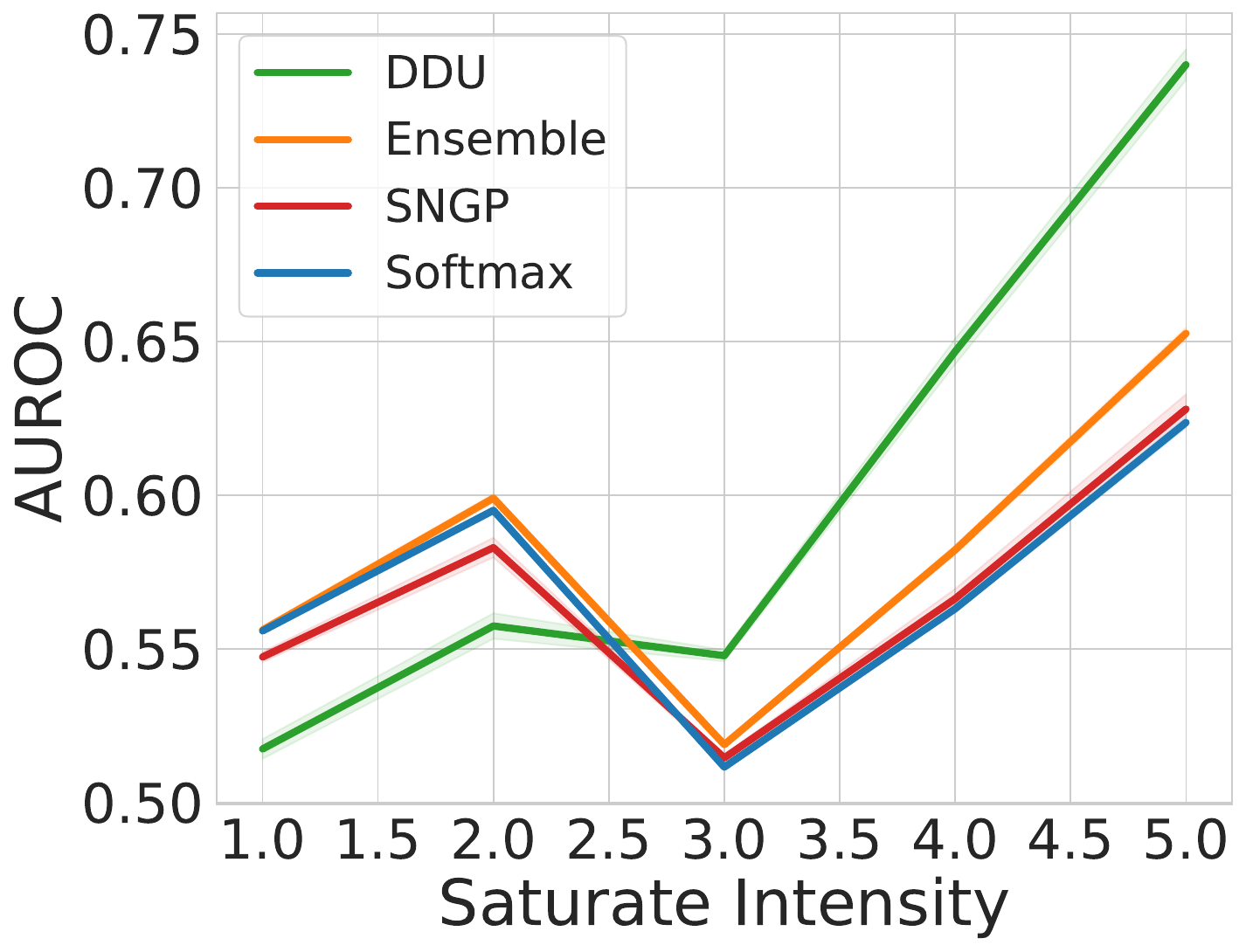}
    \end{subfigure}\hfill
    \begin{subfigure}{0.25\linewidth}
        \centering
        \includegraphics[width=\linewidth]{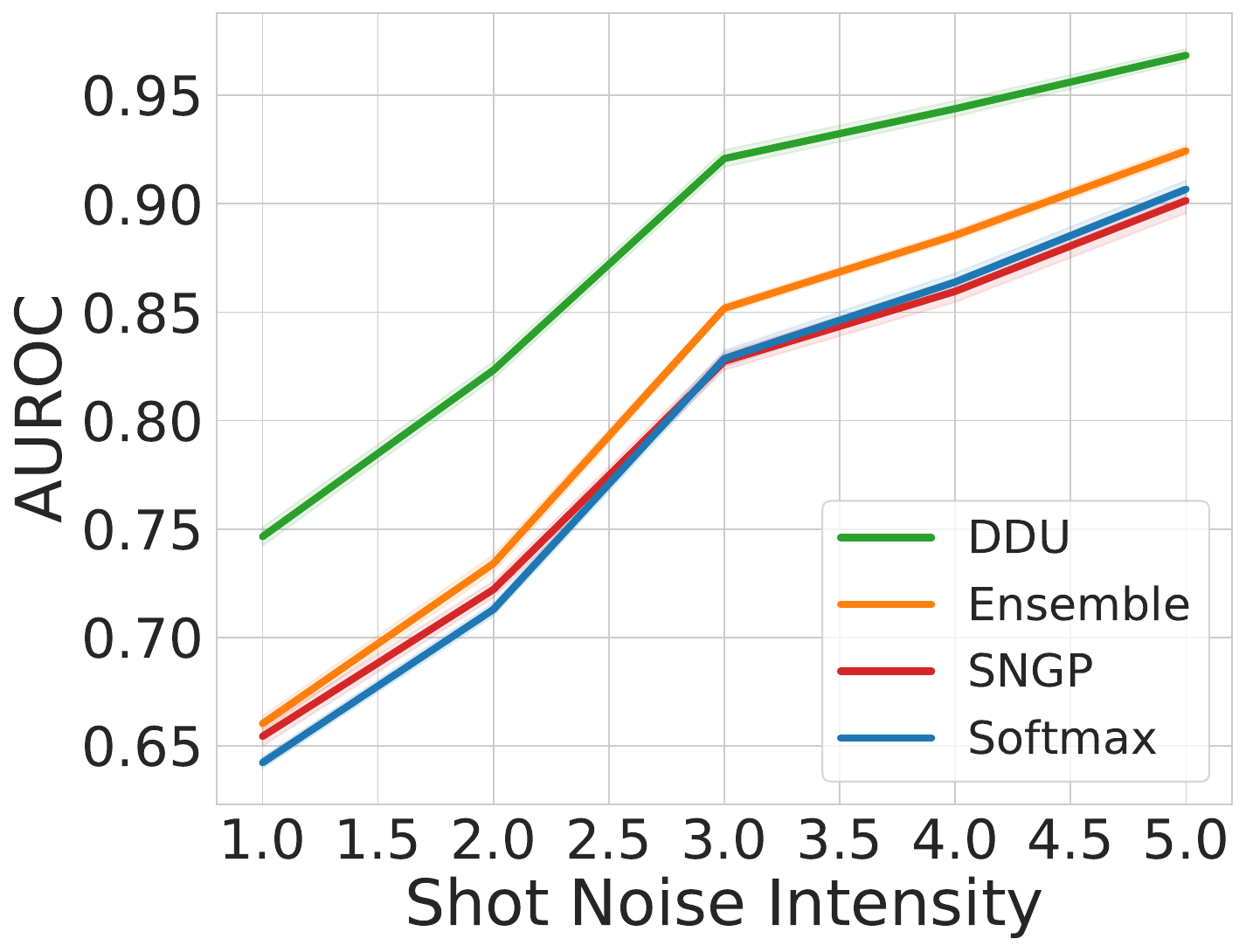}
    \end{subfigure}\hfill
    \begin{subfigure}{0.25\linewidth}
        \centering
        \includegraphics[width=\linewidth]{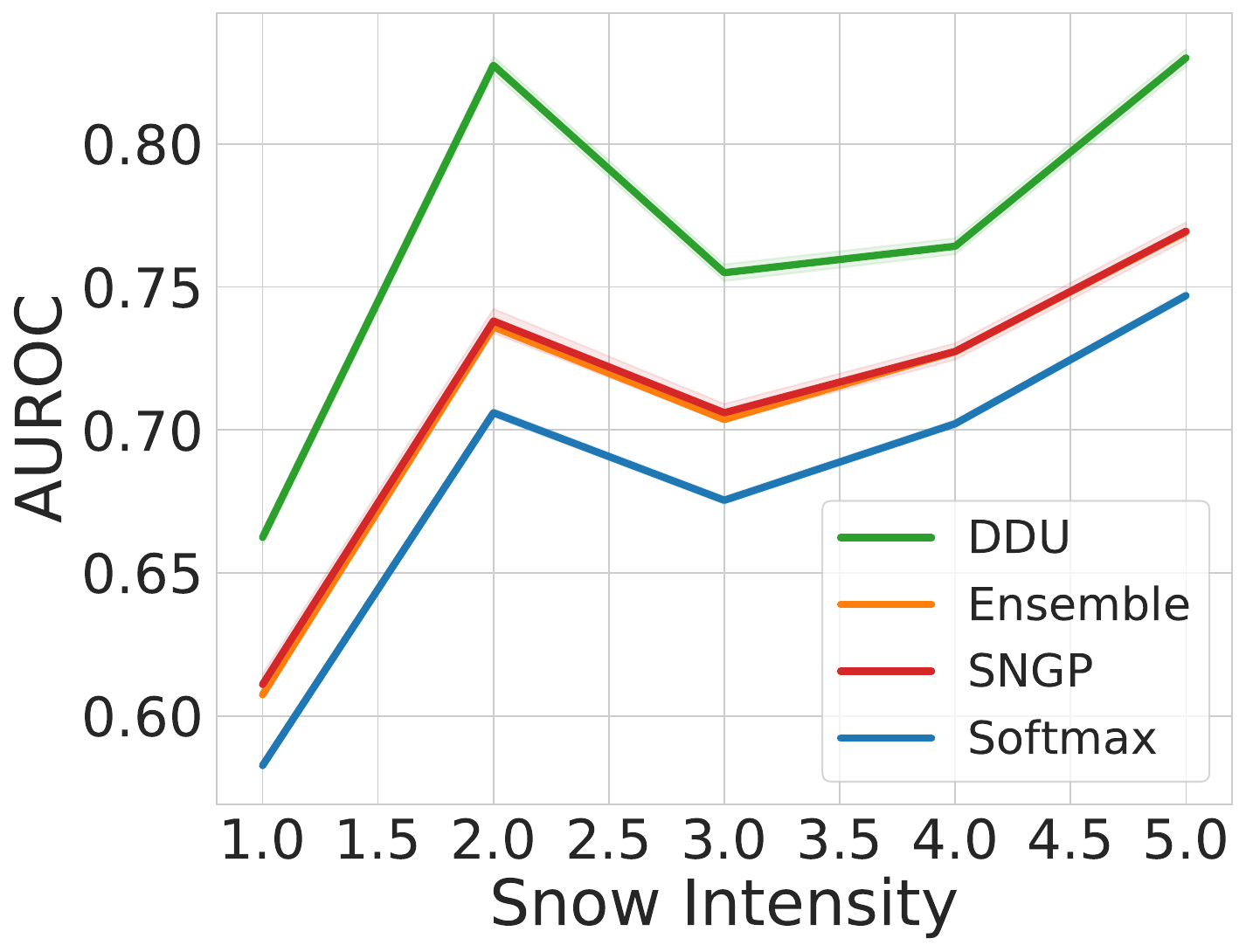}
    \end{subfigure}\hfill
    \begin{subfigure}{0.25\linewidth}
        \centering
        \includegraphics[width=\linewidth]{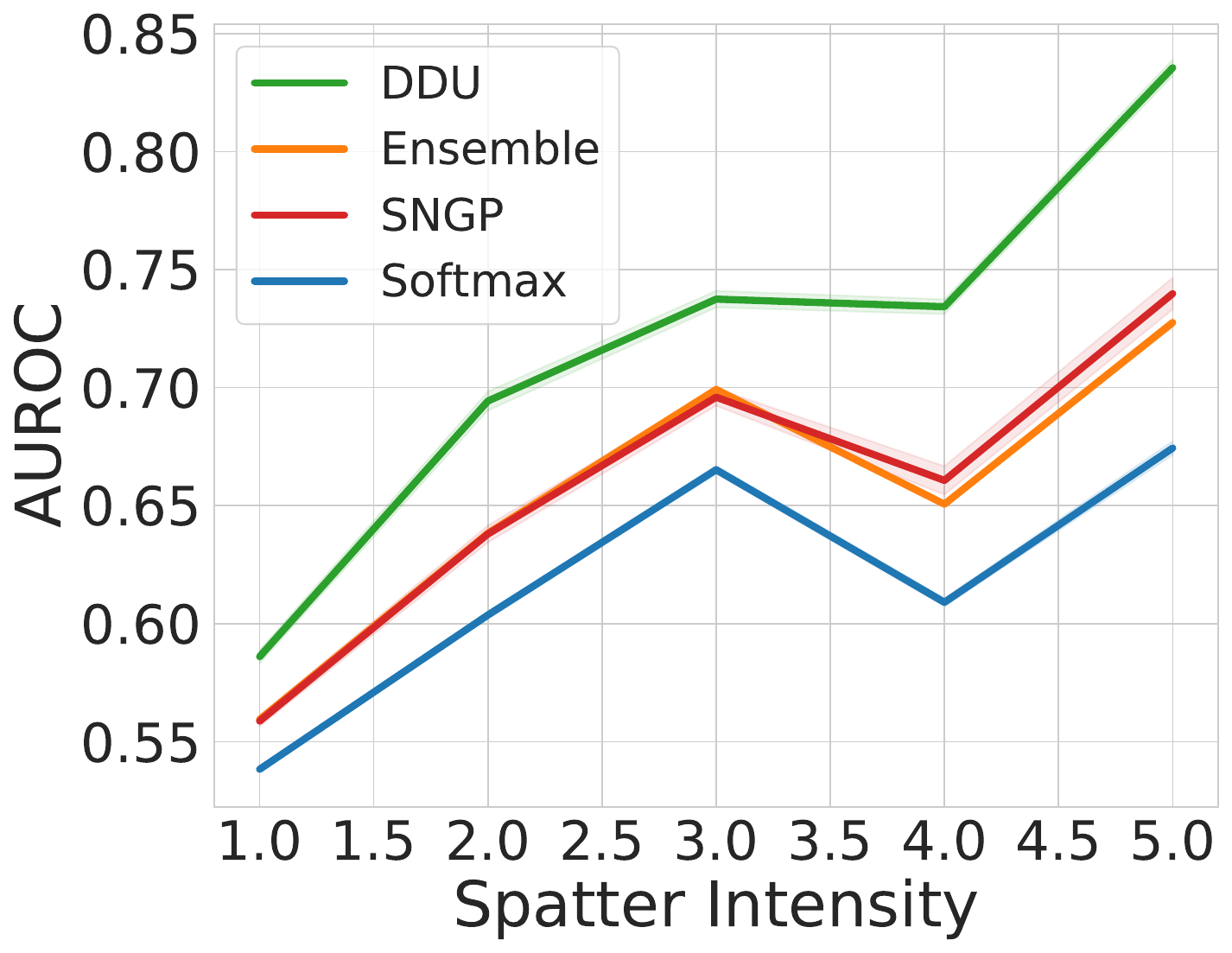}
    \end{subfigure}
    \begin{subfigure}{0.25\linewidth}
        \centering
        \includegraphics[width=\linewidth]{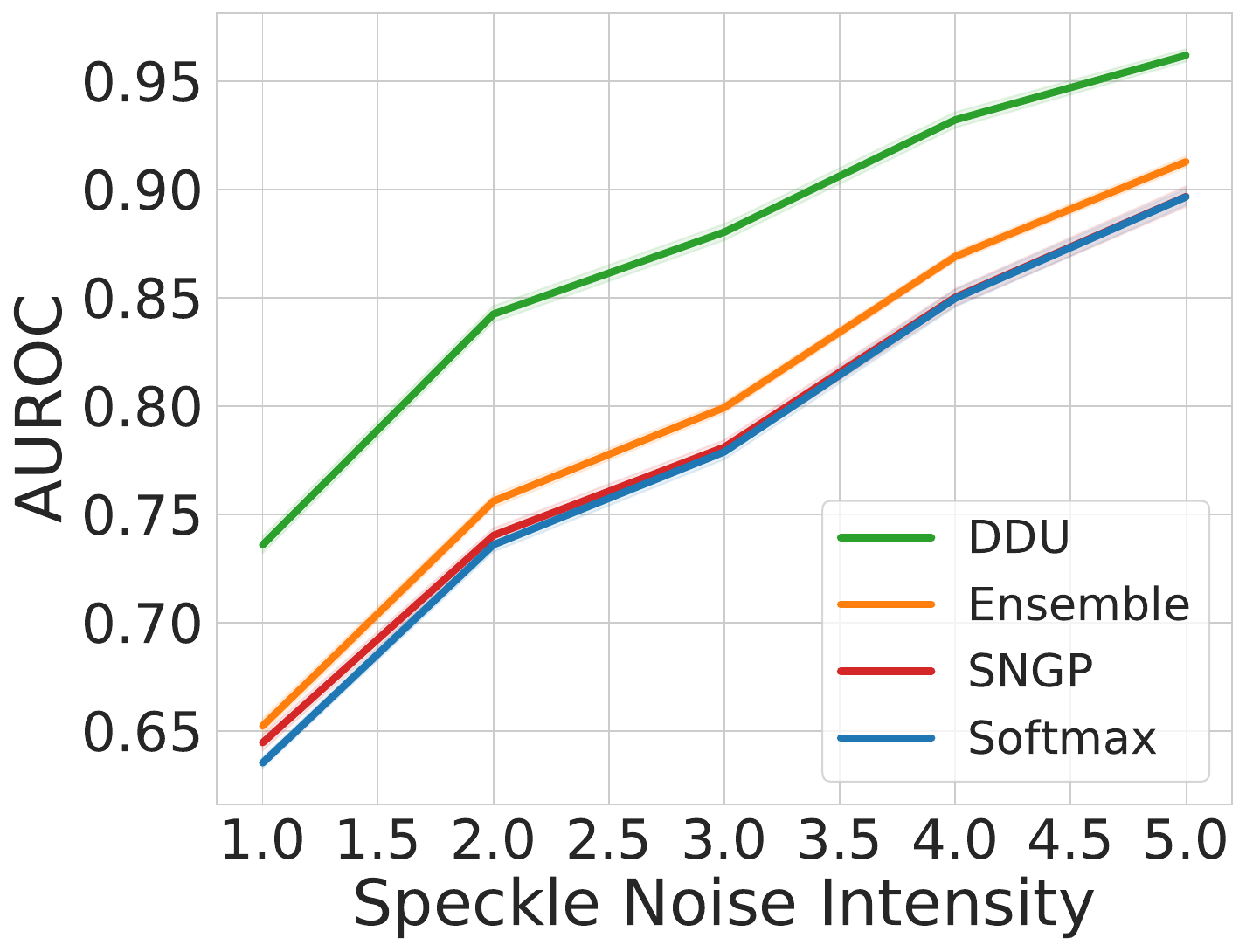}
    \end{subfigure}
    \begin{subfigure}{0.25\linewidth}
        \centering
        \includegraphics[width=\linewidth]{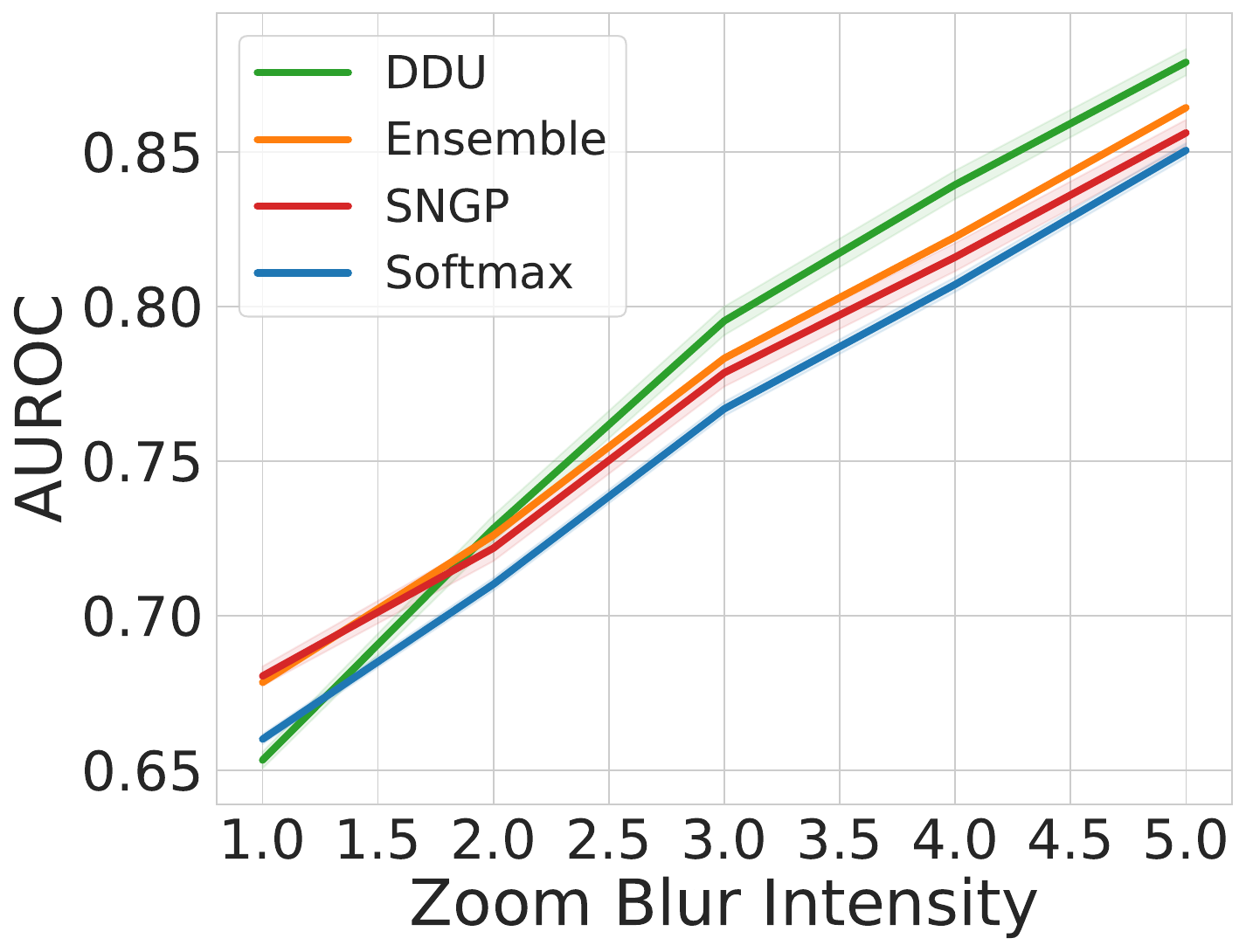}
    \end{subfigure}
    \caption{
    AUROC vs corruption intensity for all corruption types in CIFAR-10-C with ResNet-50 as the architecture and baselines: Softmax Entropy, Ensemble (using Predictive Entropy as uncertainty), SNGP and DDU feature density.
    }
    \label{ddu:fig:cifar10_c_results_resnet50}
\end{figure}

\begin{figure}[!t]
    \centering
    \begin{subfigure}{0.25\linewidth}
        \centering
        \includegraphics[width=\linewidth]{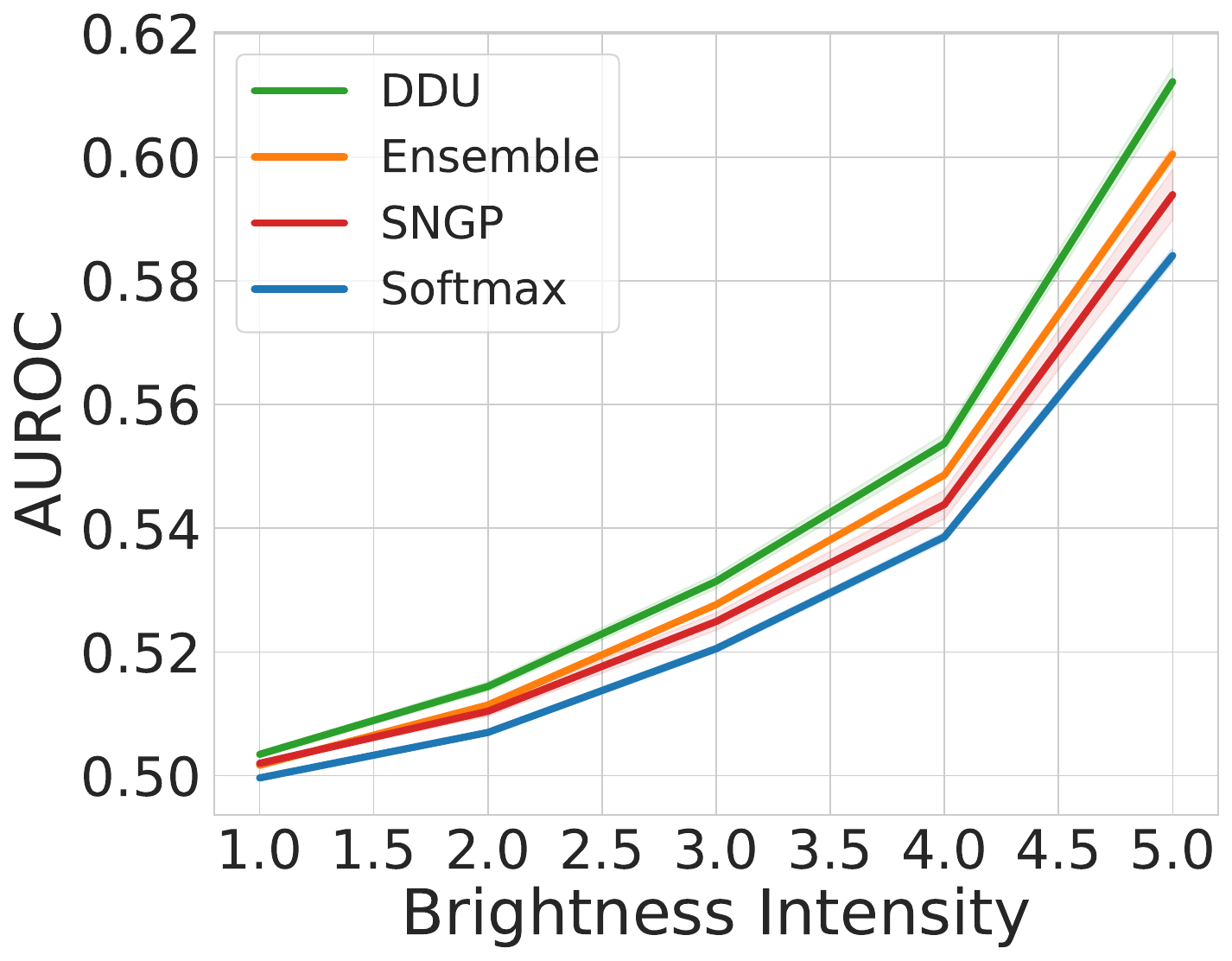}
    \end{subfigure}\hfill
    \begin{subfigure}{0.25\linewidth}
        \centering
        \includegraphics[width=\linewidth]{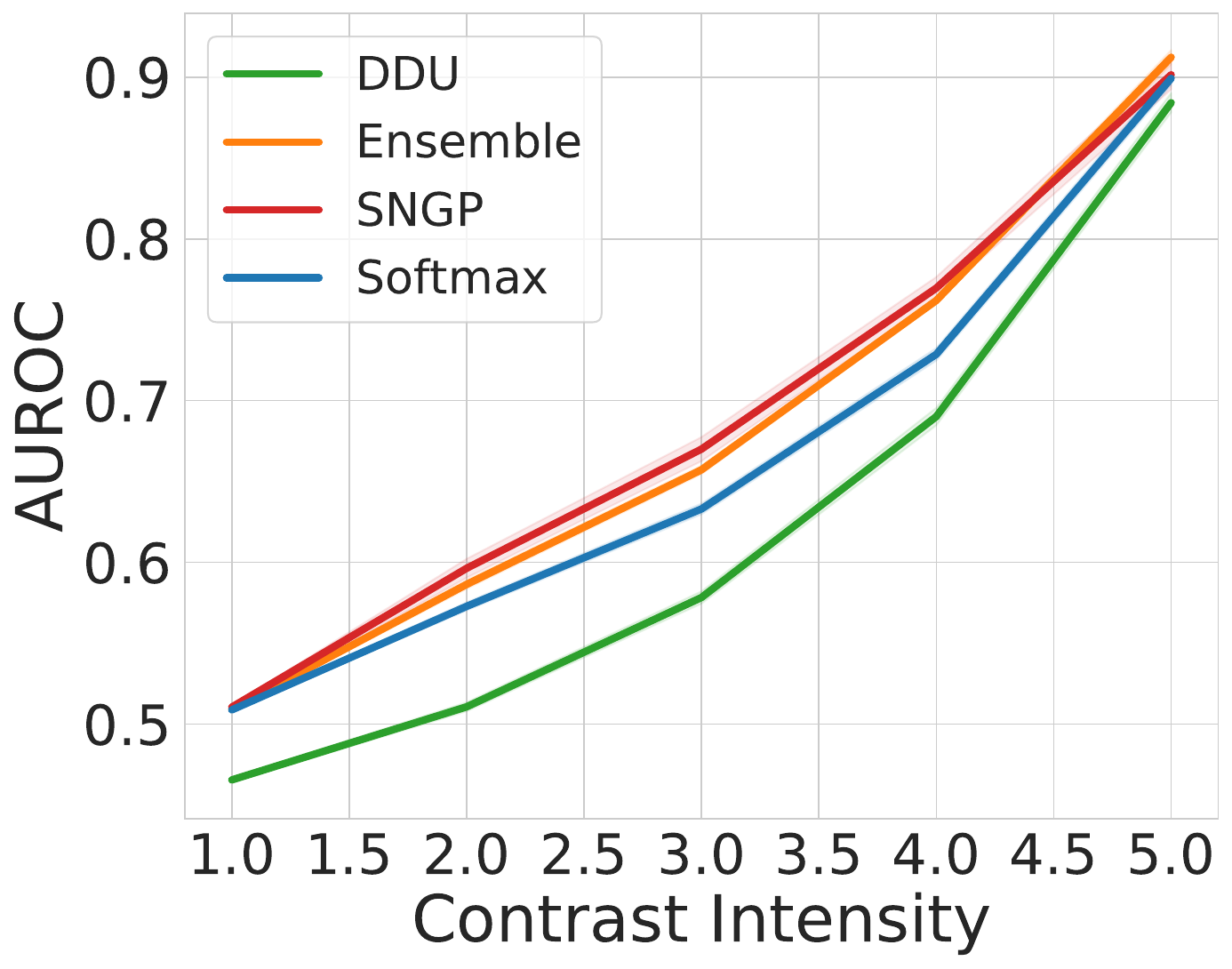}
    \end{subfigure}\hfill 
    \begin{subfigure}{0.25\linewidth}
        \centering
        \includegraphics[width=\linewidth]{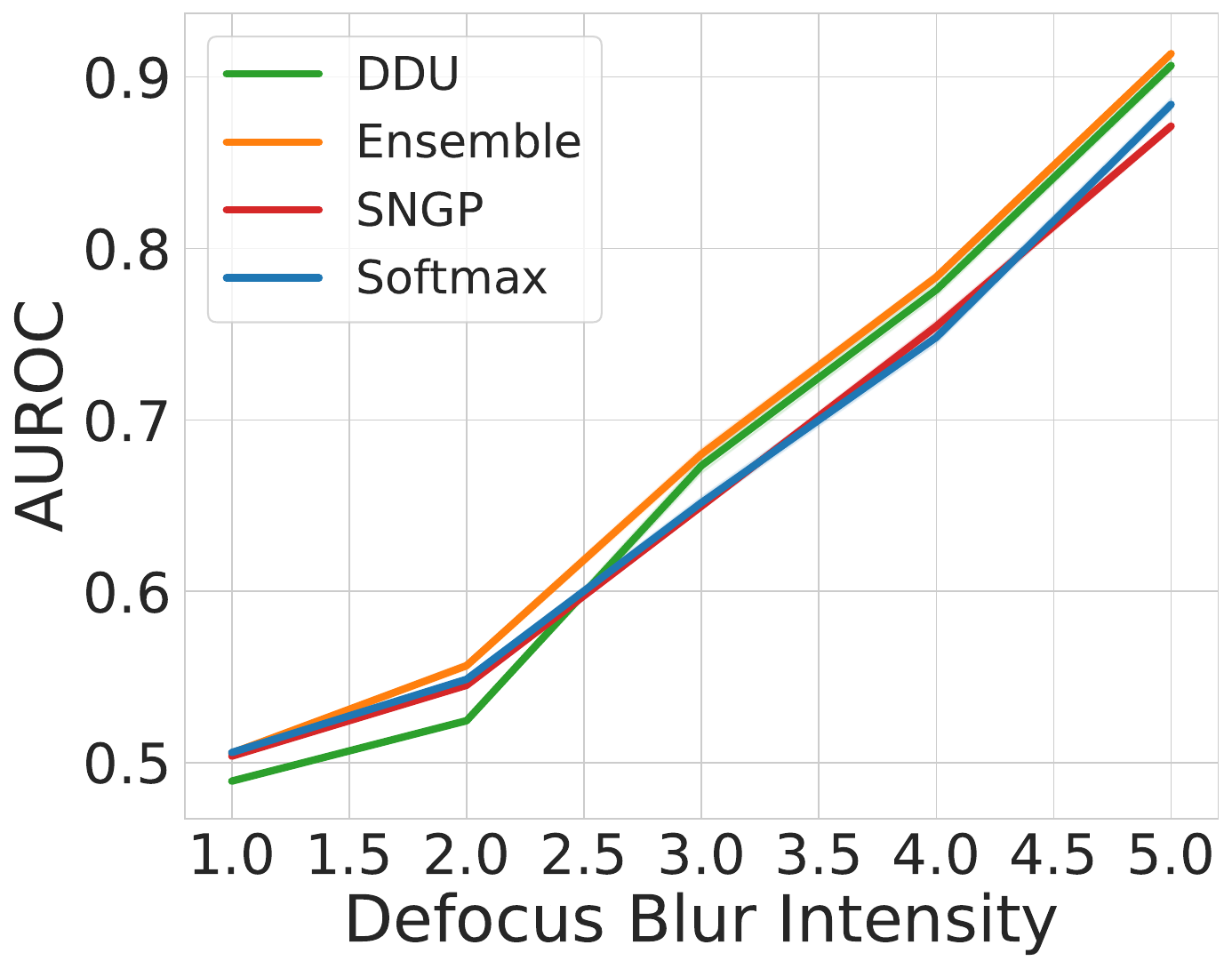}
    \end{subfigure}\hfill 
    \begin{subfigure}{0.25\linewidth}
        \centering
        \includegraphics[width=\linewidth]{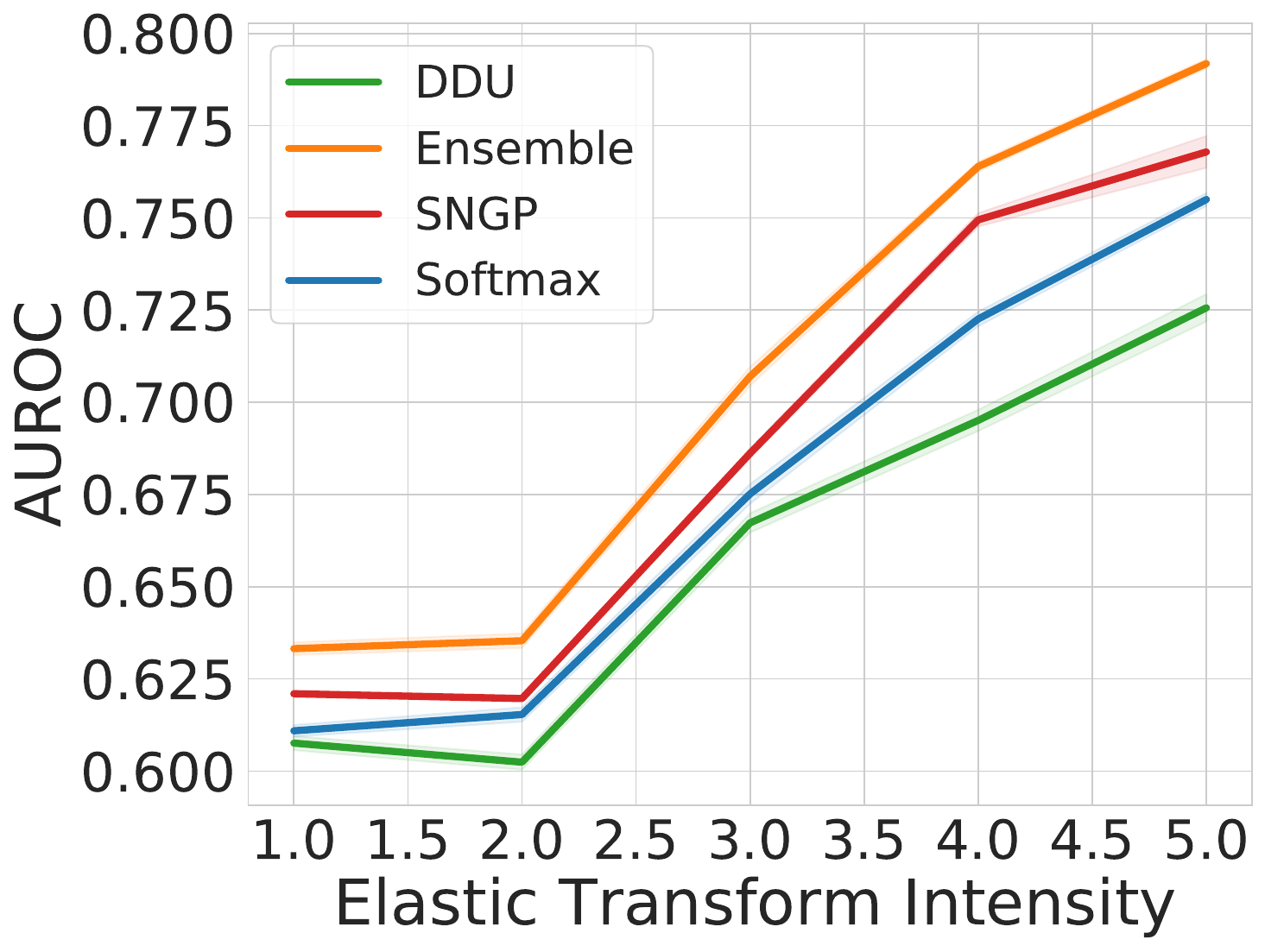}
    \end{subfigure}\hfill
    \begin{subfigure}{0.25\linewidth}
        \centering
        \includegraphics[width=\linewidth]{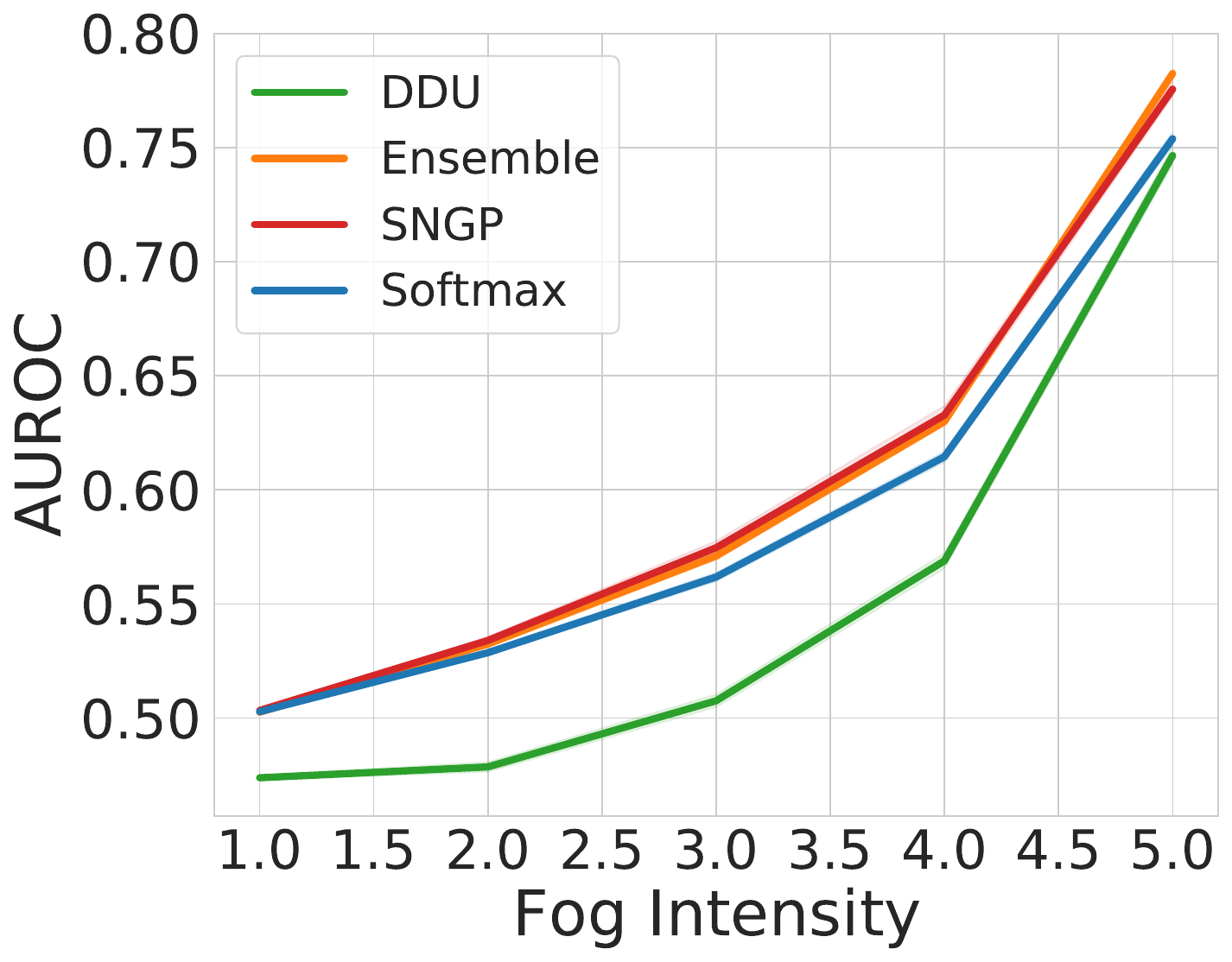}
    \end{subfigure}\hfill
    \begin{subfigure}{0.25\linewidth}
        \centering
        \includegraphics[width=\linewidth]{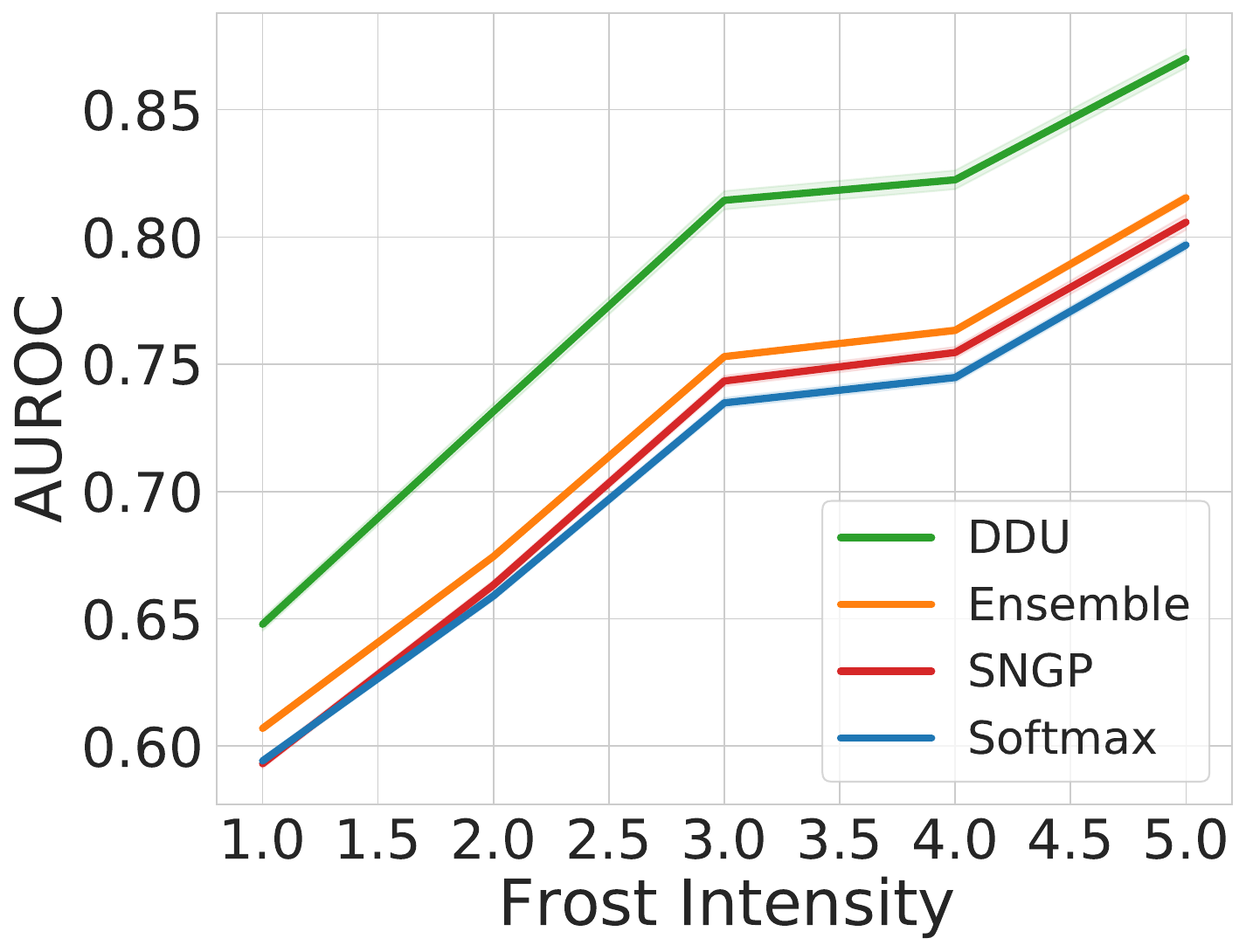}
    \end{subfigure}\hfill
    \begin{subfigure}{0.25\linewidth}
        \centering
        \includegraphics[width=\linewidth]{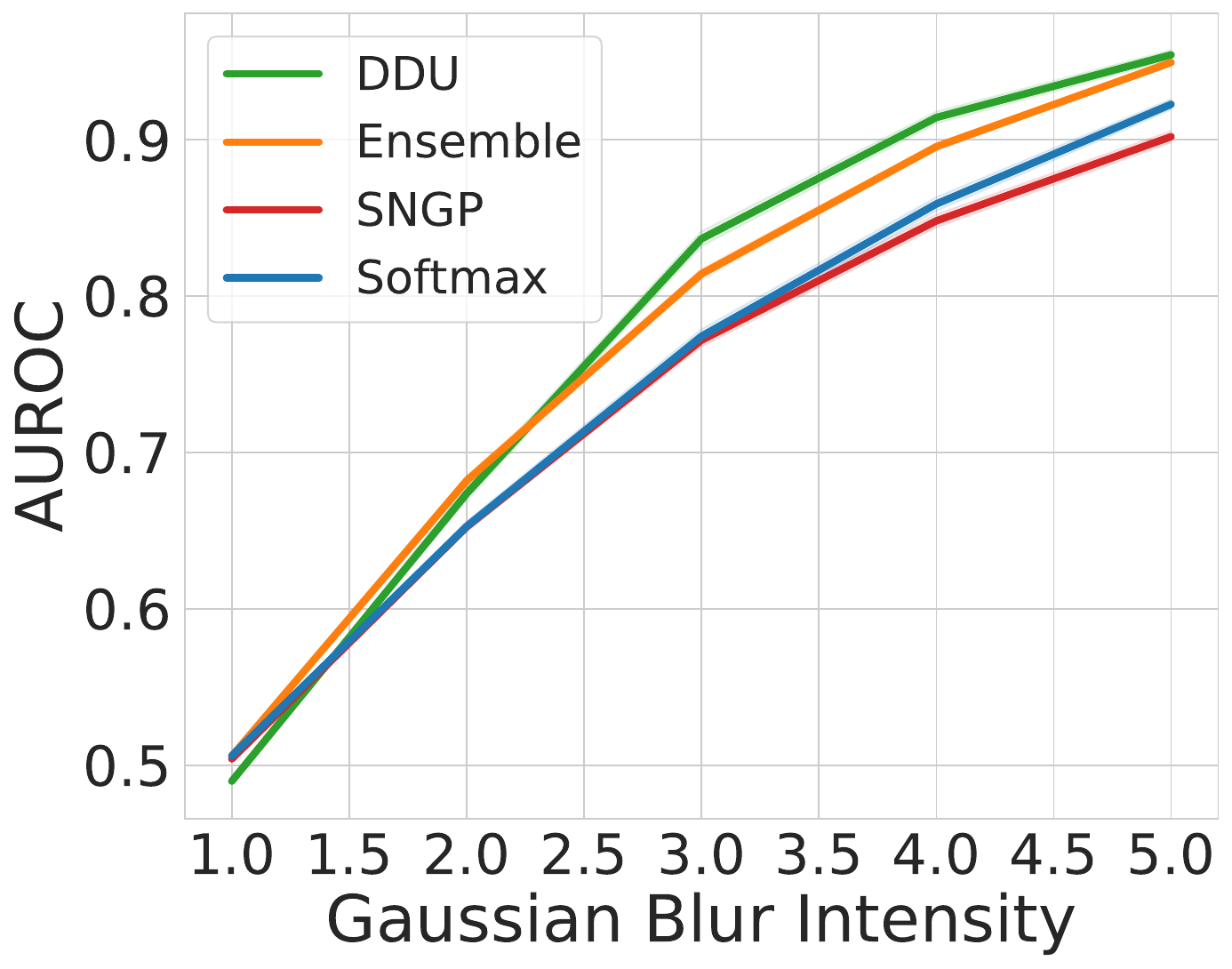}
    \end{subfigure}\hfill
    \begin{subfigure}{0.25\linewidth}
        \centering
        \includegraphics[width=\linewidth]{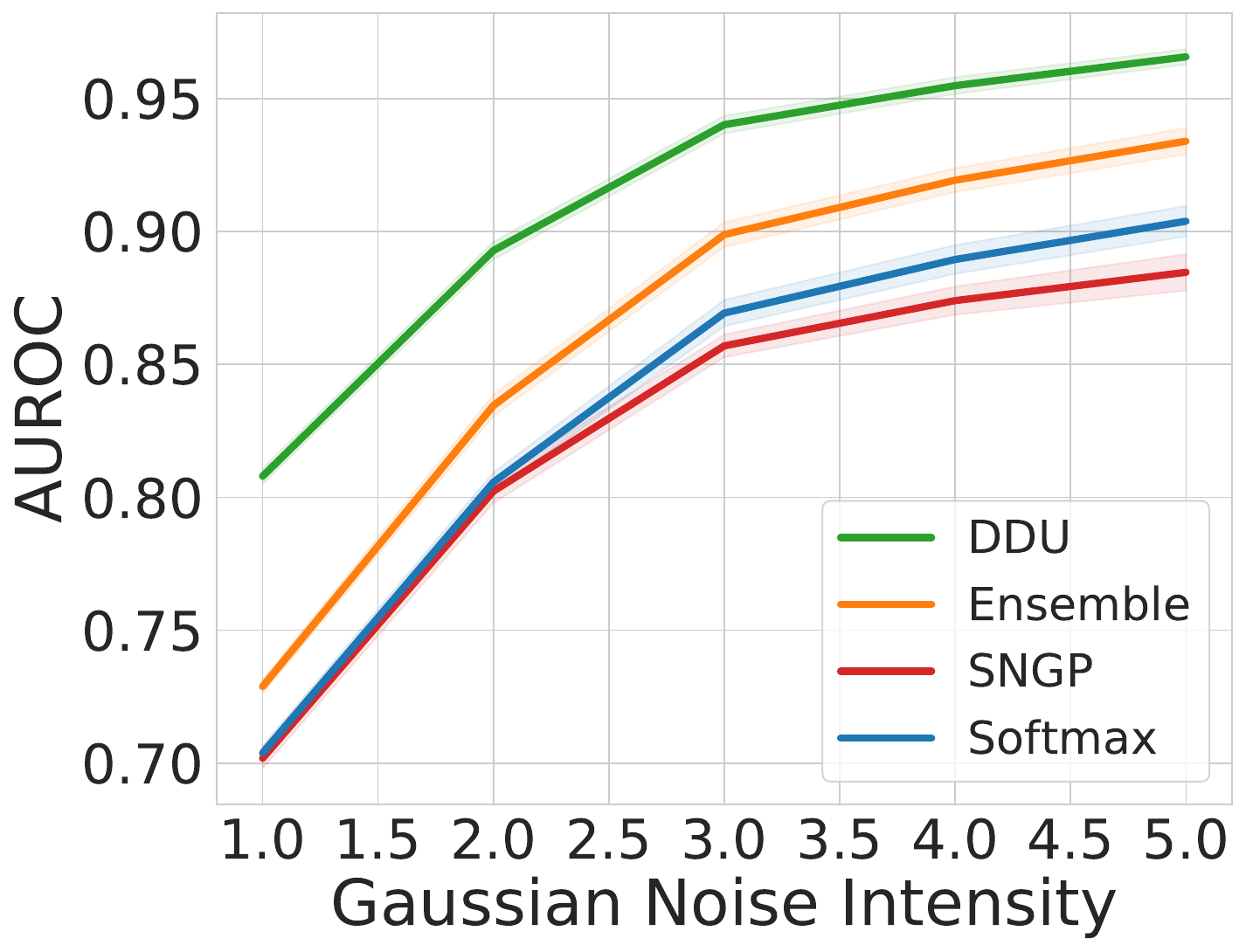}
    \end{subfigure}\hfill
    \begin{subfigure}{0.25\linewidth}
        \centering
        \includegraphics[width=\linewidth]{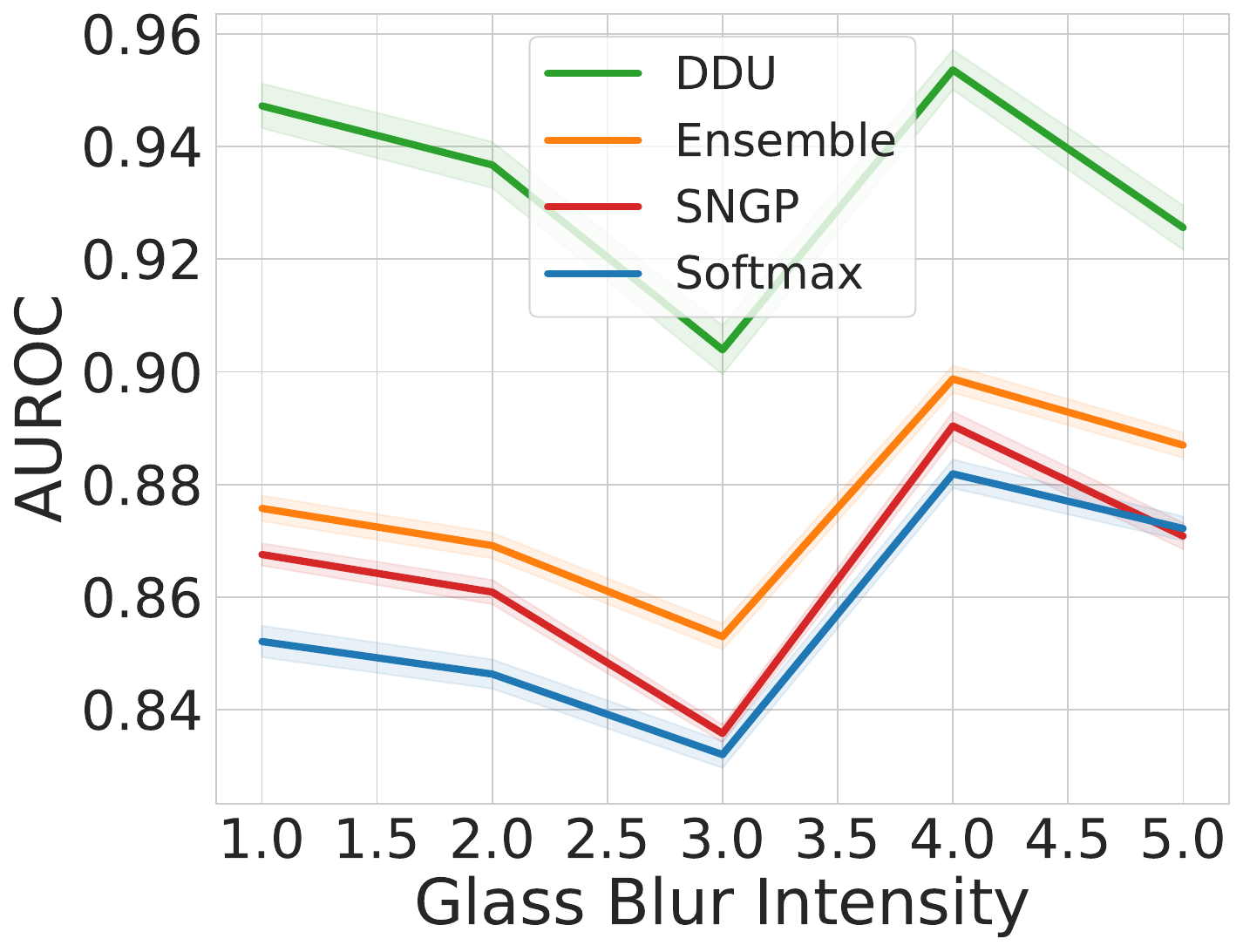}
    \end{subfigure}\hfill
    \begin{subfigure}{0.25\linewidth}
        \centering
        \includegraphics[width=\linewidth]{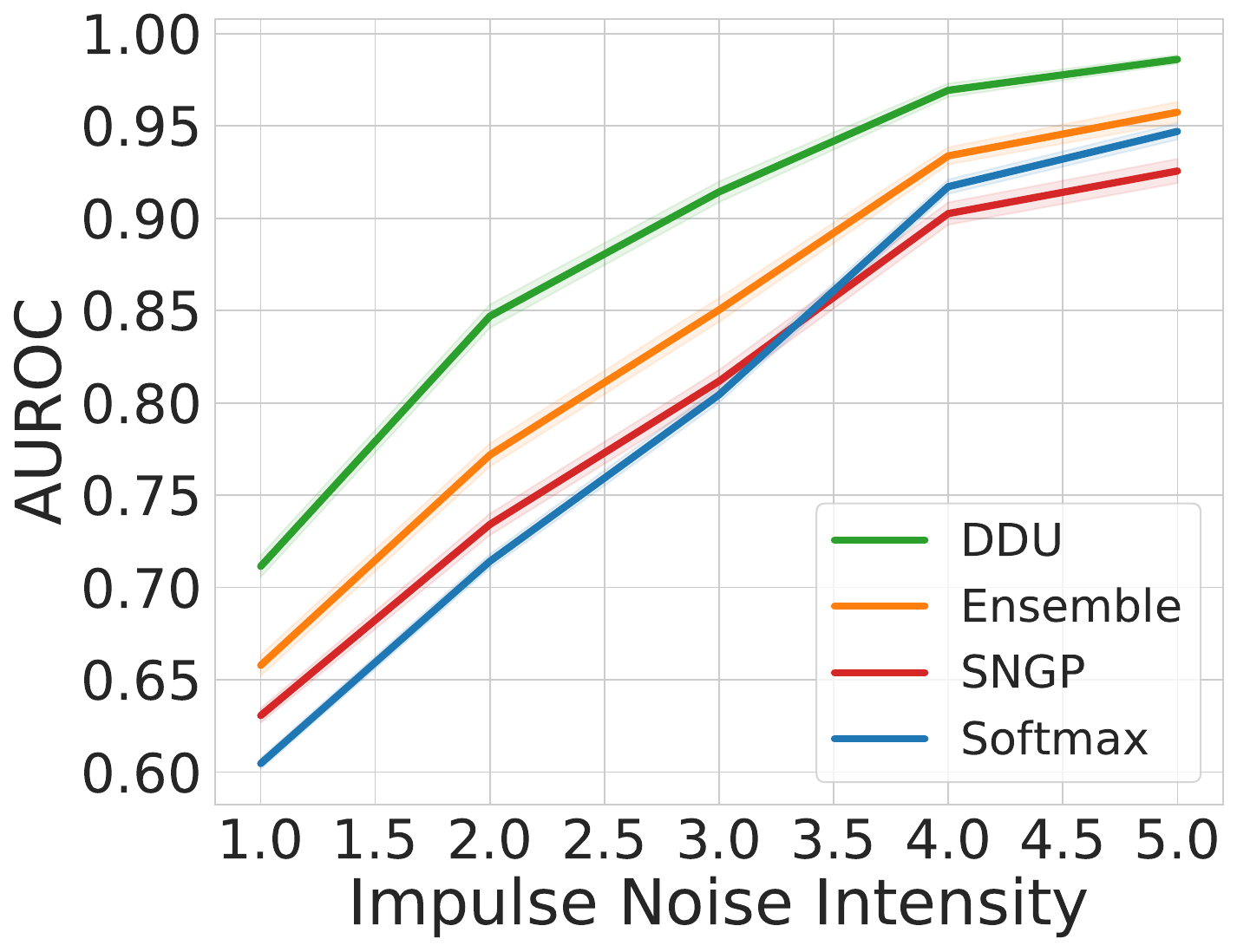}
    \end{subfigure}\hfill
    \begin{subfigure}{0.25\linewidth}
        \centering
        \includegraphics[width=\linewidth]{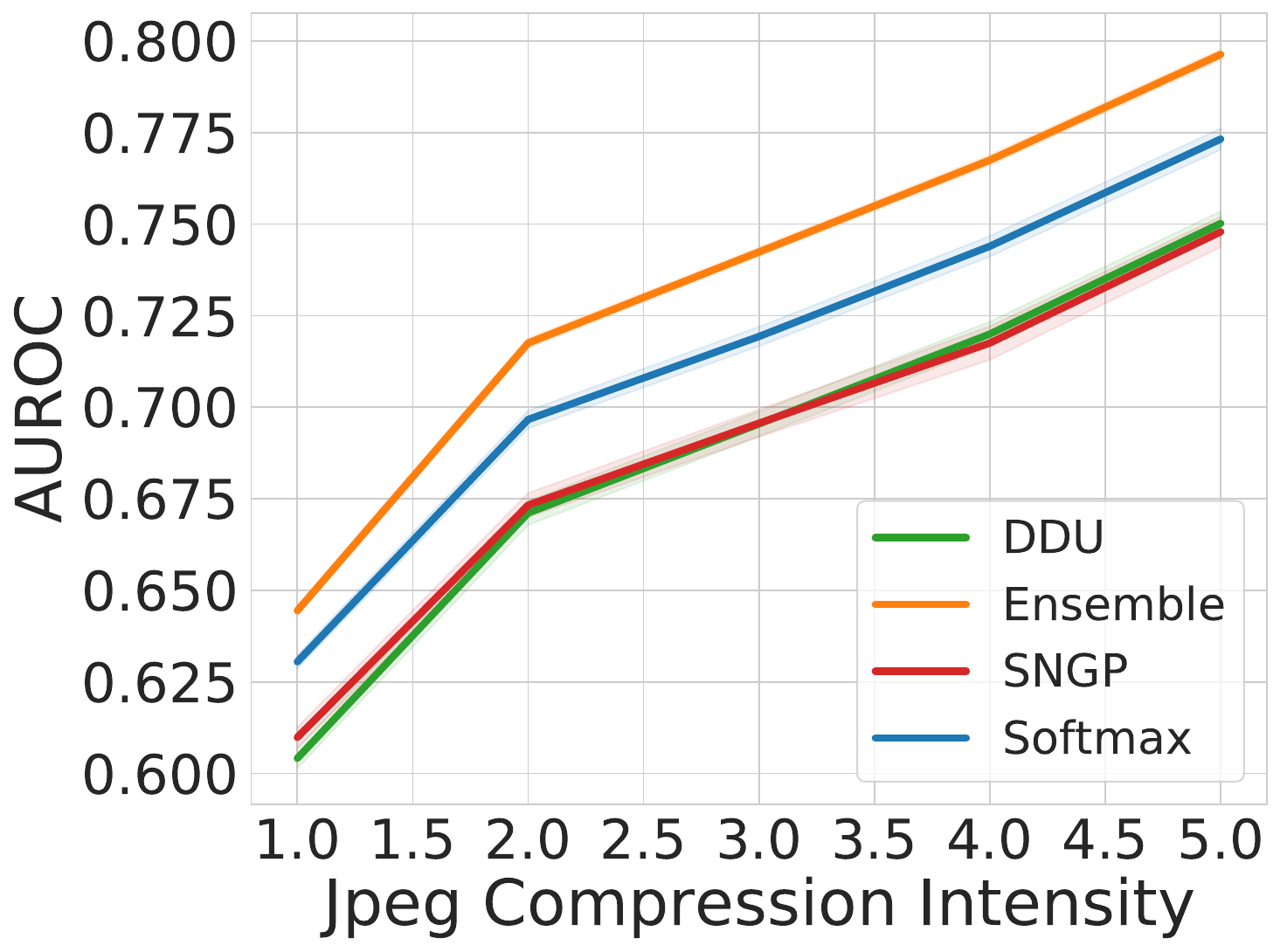}
    \end{subfigure}\hfill
    \begin{subfigure}{0.25\linewidth}
        \centering
        \includegraphics[width=\linewidth]{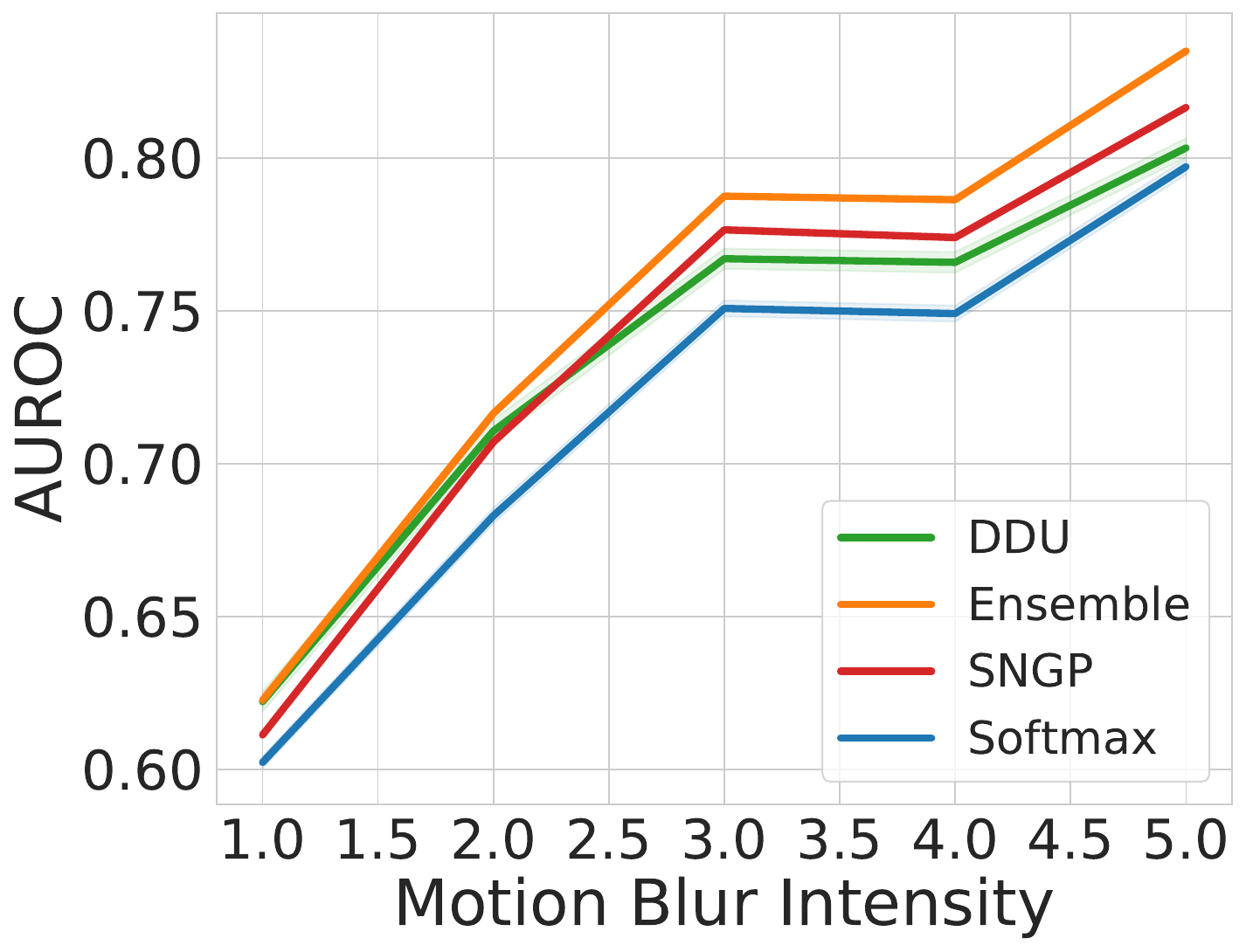}
    \end{subfigure}\hfill
    \begin{subfigure}{0.25\linewidth}
        \centering
        \includegraphics[width=\linewidth]{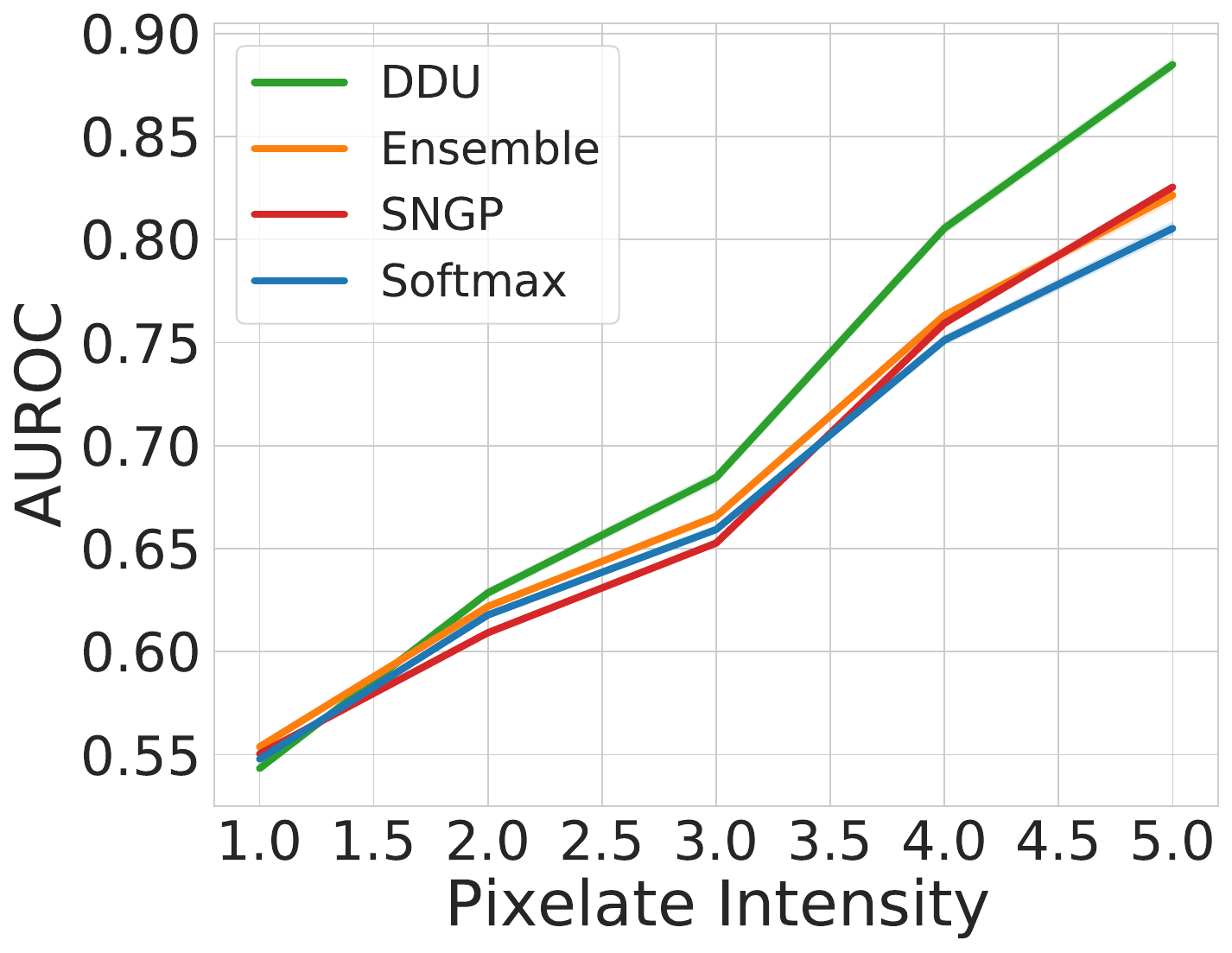}
    \end{subfigure}\hfill
    \begin{subfigure}{0.25\linewidth}
        \centering
        \includegraphics[width=\linewidth]{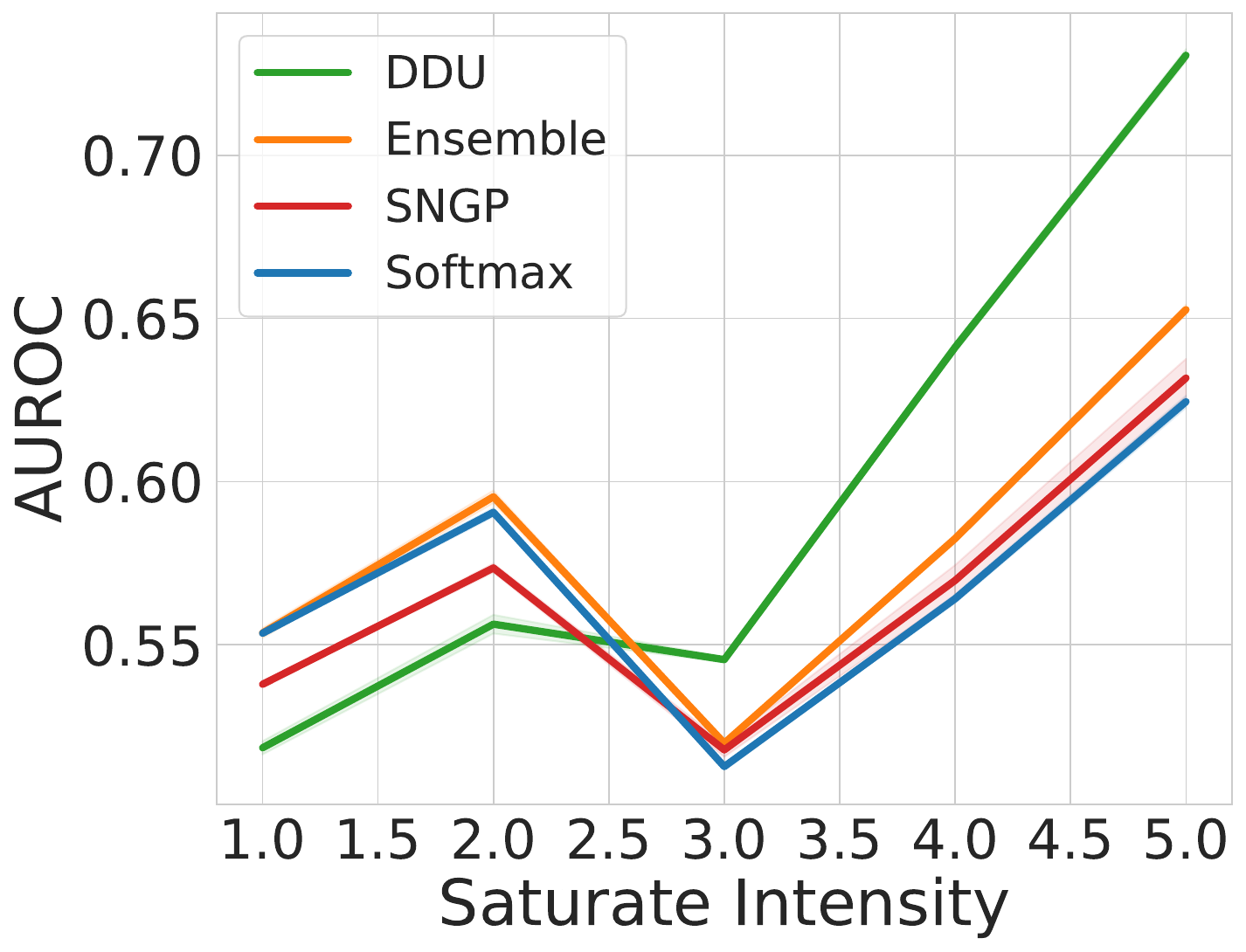}
    \end{subfigure}\hfill
    \begin{subfigure}{0.25\linewidth}
        \centering
        \includegraphics[width=\linewidth]{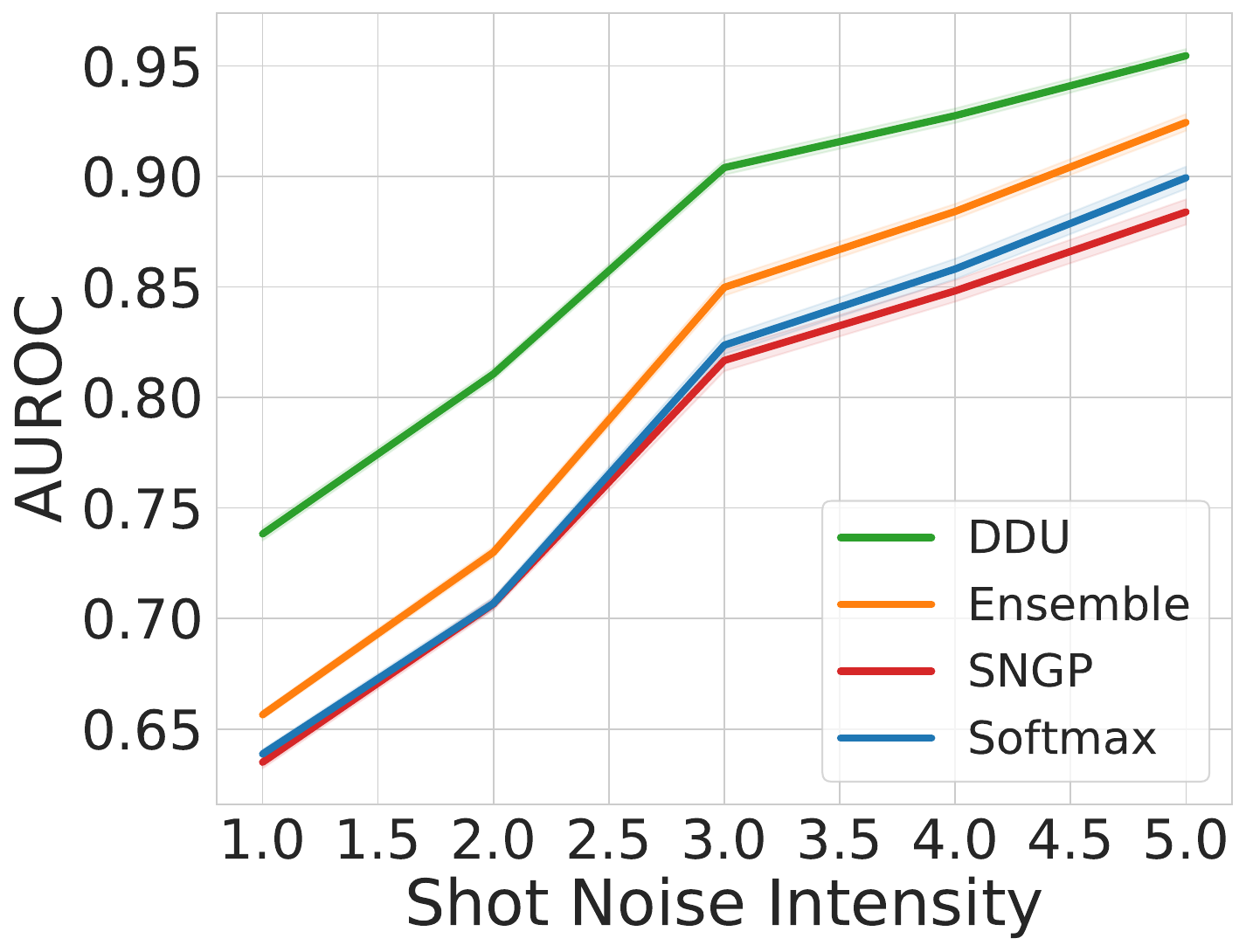}
    \end{subfigure}\hfill
    \begin{subfigure}{0.25\linewidth}
        \centering
        \includegraphics[width=\linewidth]{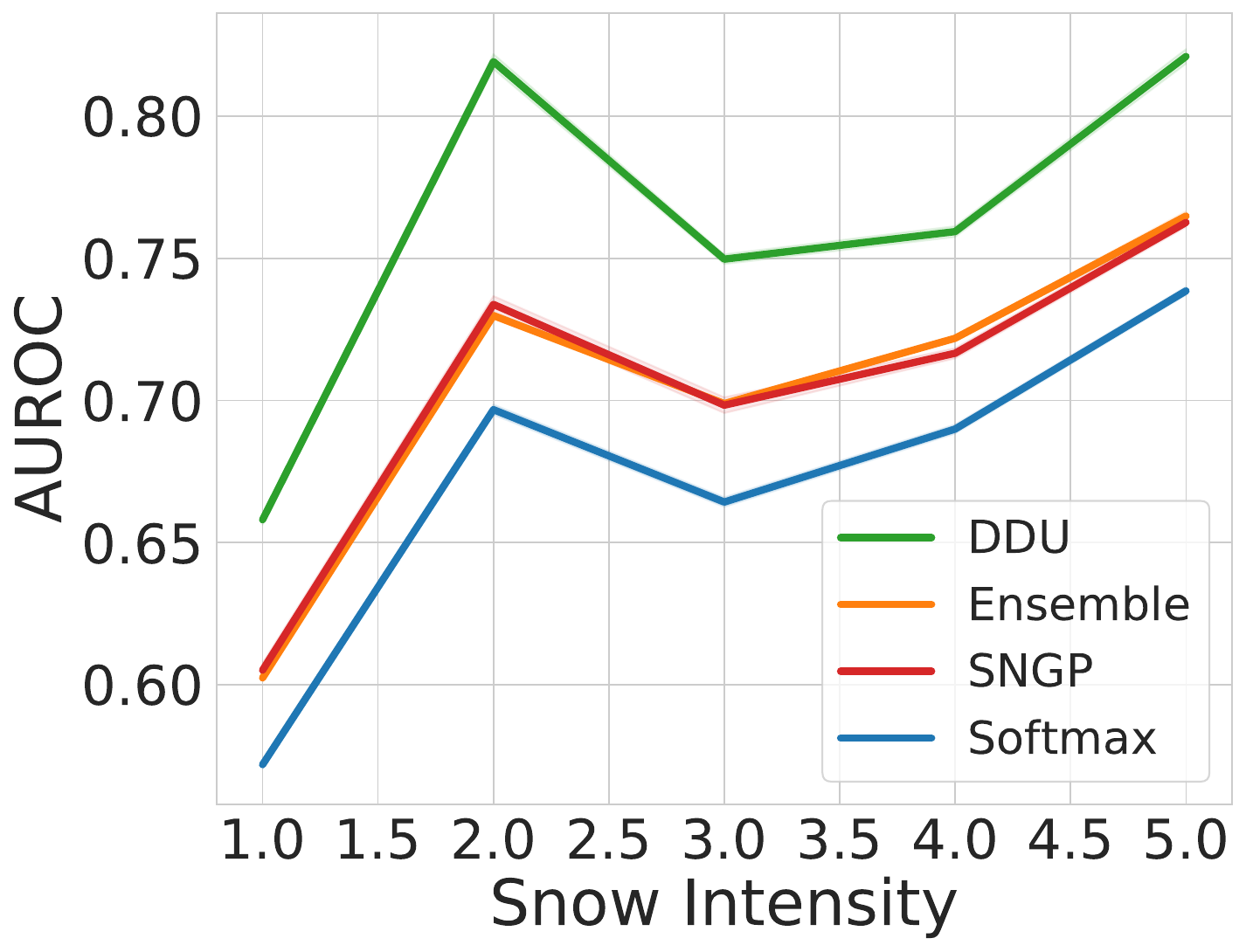}
    \end{subfigure}\hfill
    \begin{subfigure}{0.25\linewidth}
        \centering
        \includegraphics[width=\linewidth]{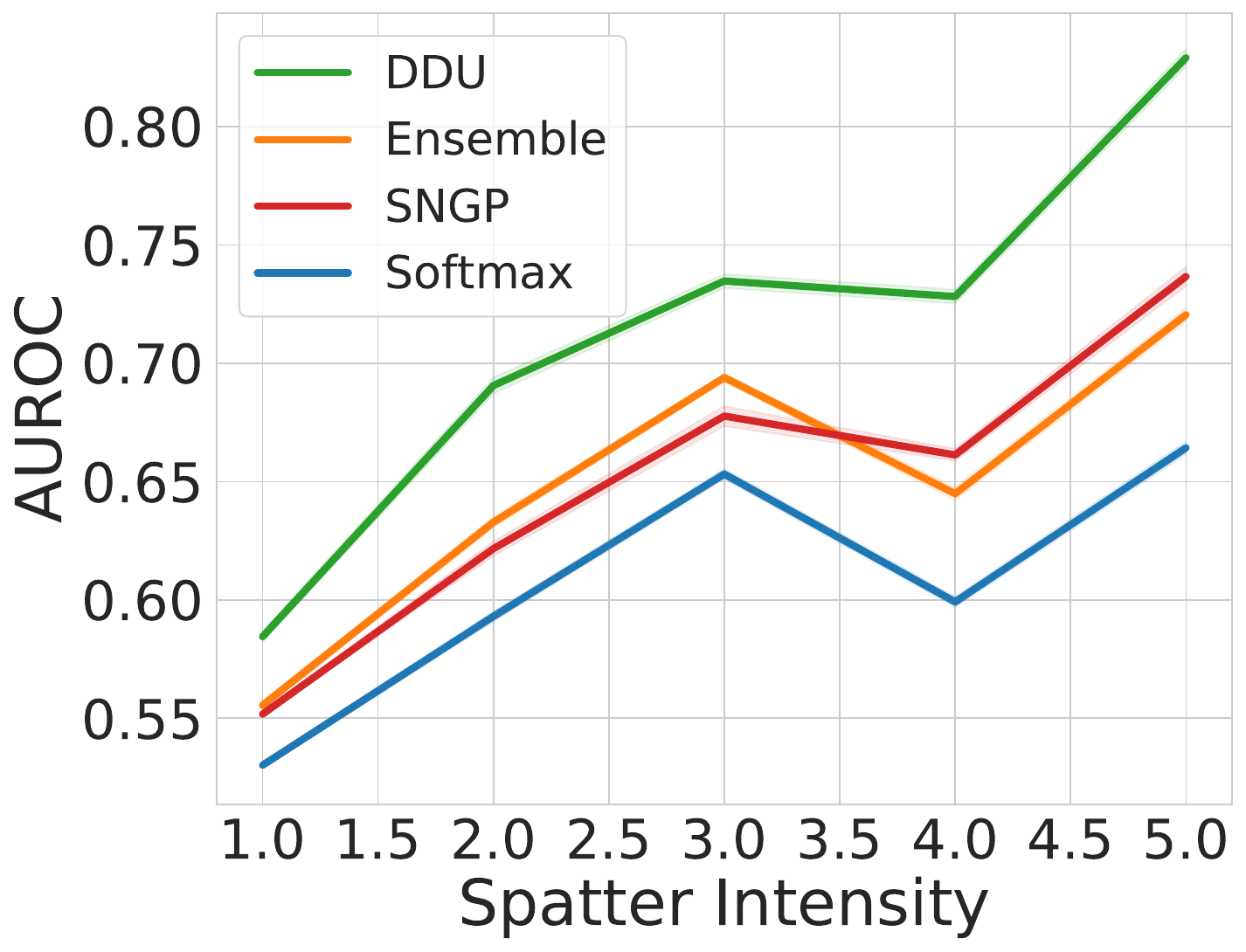}
    \end{subfigure}
    \begin{subfigure}{0.25\linewidth}
        \centering
        \includegraphics[width=\linewidth]{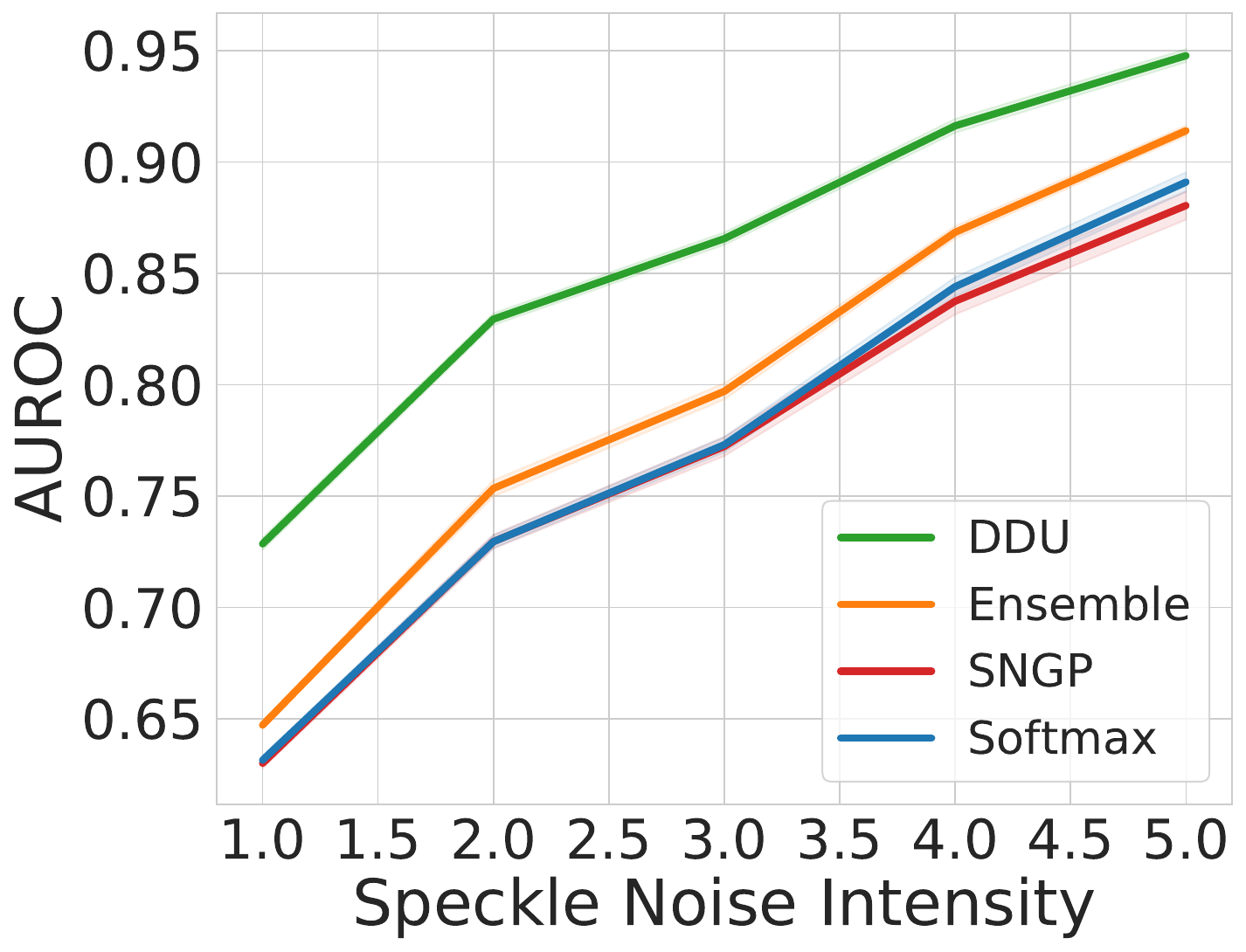}
    \end{subfigure}
    \begin{subfigure}{0.25\linewidth}
        \centering
        \includegraphics[width=\linewidth]{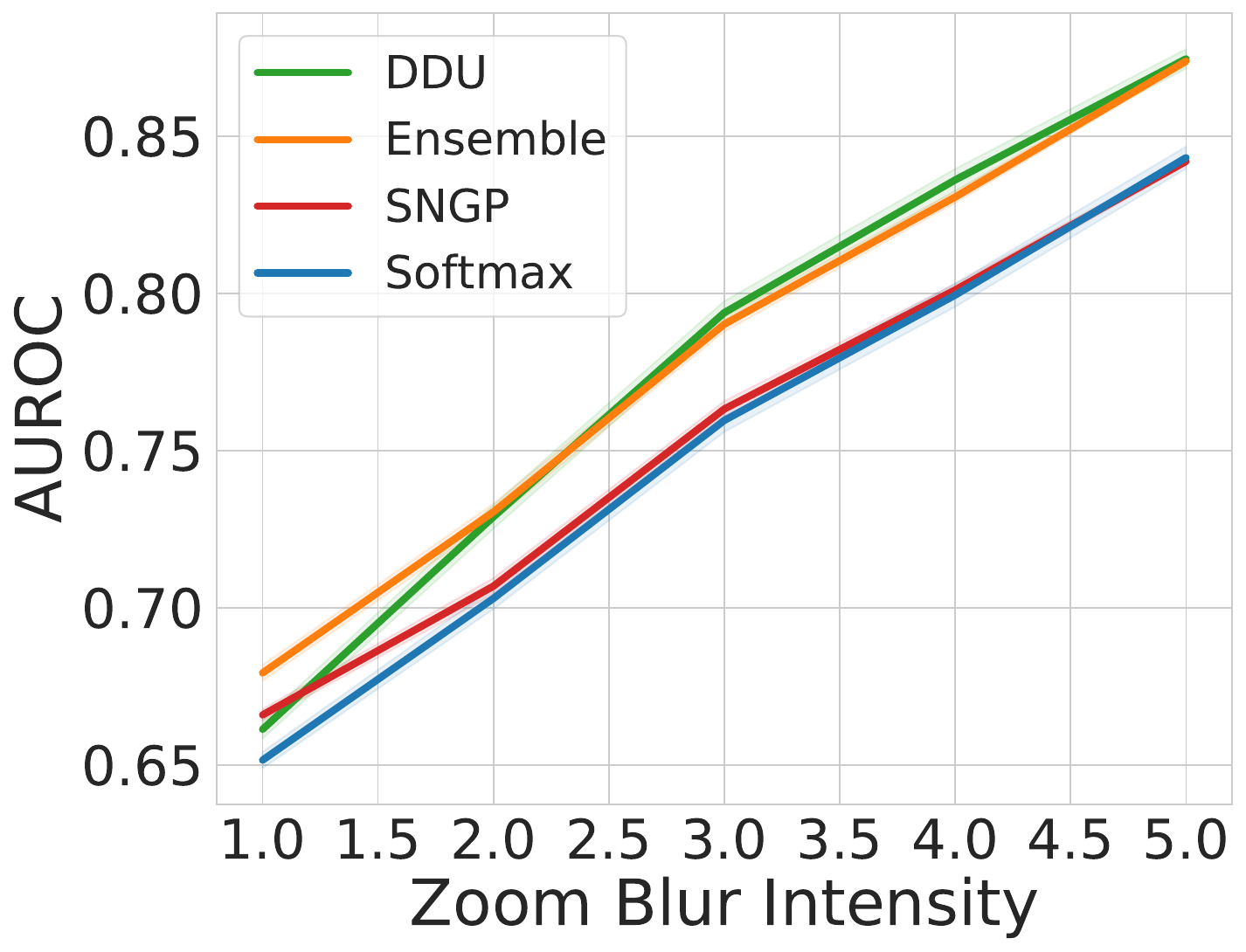}
    \end{subfigure}
    \caption{
    AUROC vs corruption intensity for all corruption types in CIFAR-10-C with ResNet-110 as the architecture and baselines: Softmax Entropy, Ensemble (using Predictive Entropy as uncertainty), SNGP and DDU feature density.
    }
    \label{ddu:fig:cifar10_c_results_resnet110}
\end{figure}

\begin{figure}[!t]
    \centering
    \begin{subfigure}{0.25\linewidth}
        \centering
        \includegraphics[width=\linewidth]{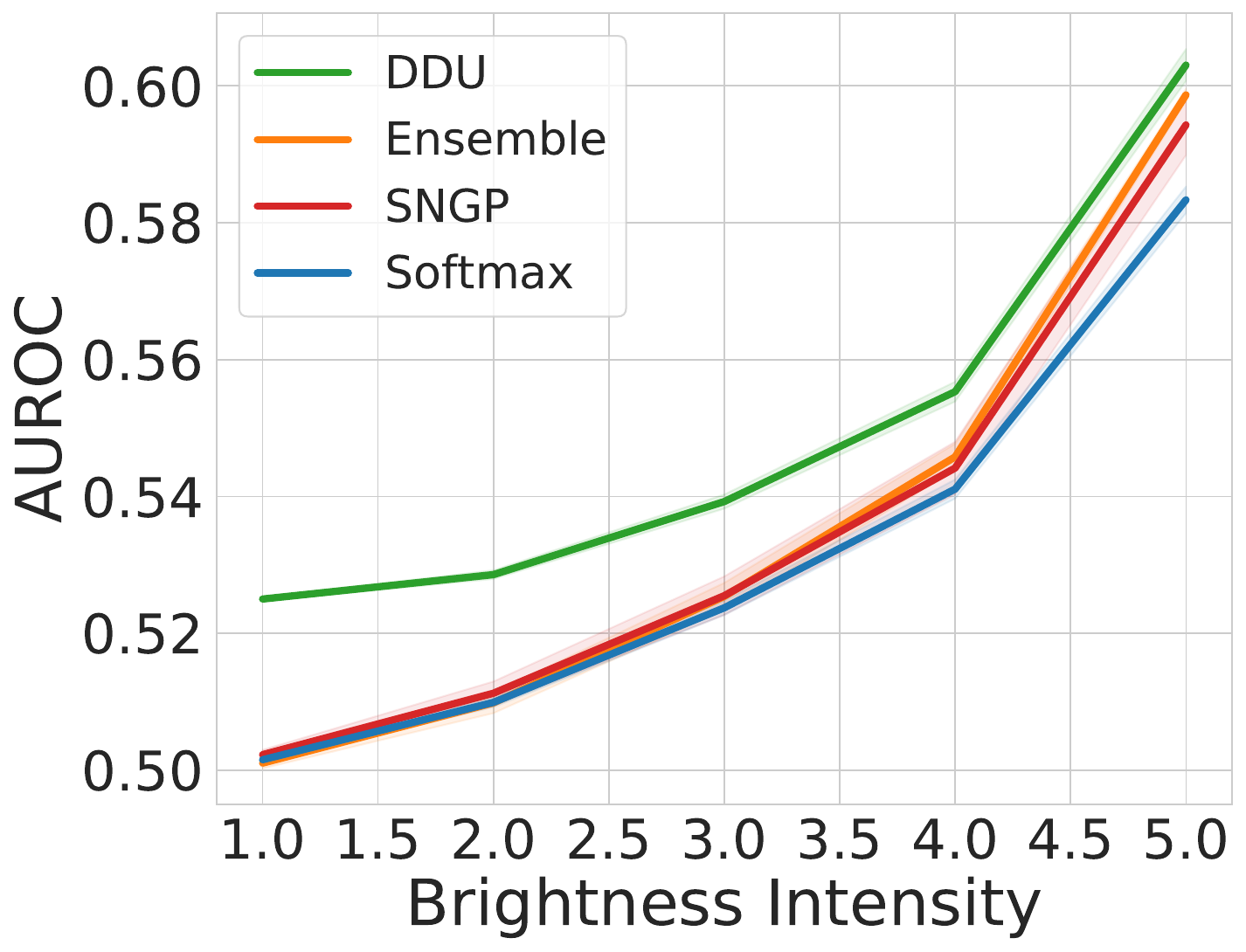}
    \end{subfigure}\hfill
    \begin{subfigure}{0.25\linewidth}
        \centering
        \includegraphics[width=\linewidth]{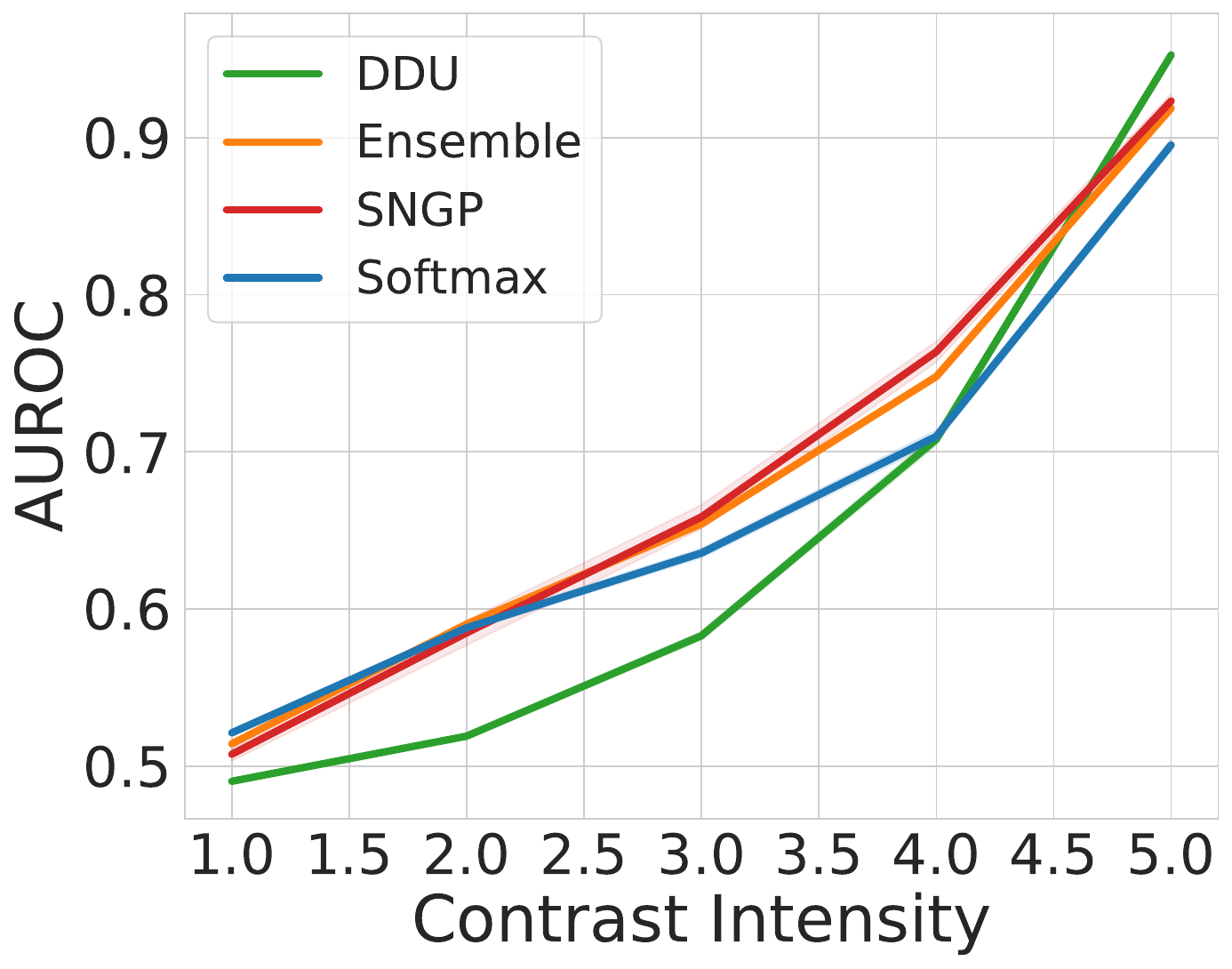}
    \end{subfigure}\hfill 
    \begin{subfigure}{0.25\linewidth}
        \centering
        \includegraphics[width=\linewidth]{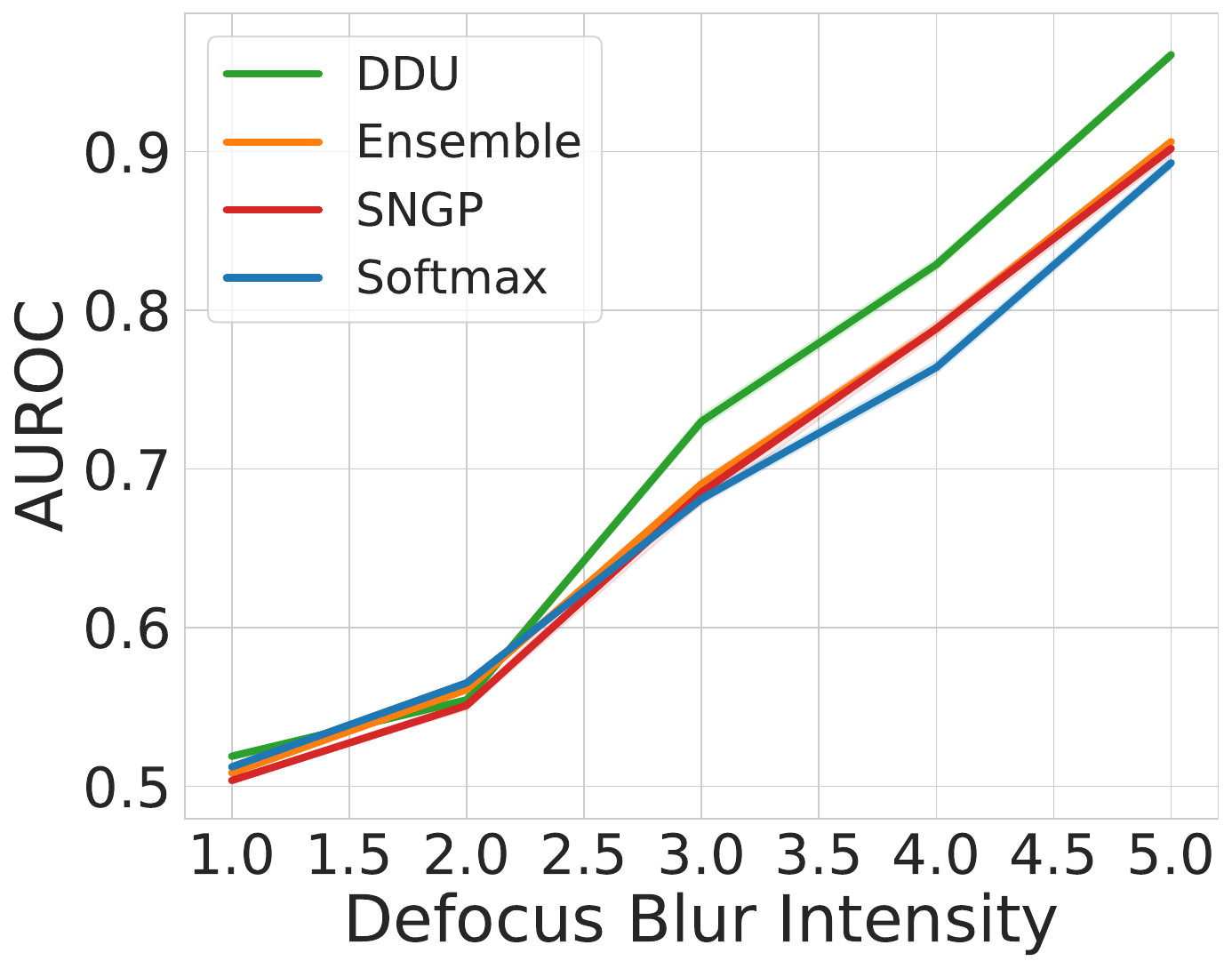}
    \end{subfigure}\hfill 
    \begin{subfigure}{0.25\linewidth}
        \centering
        \includegraphics[width=\linewidth]{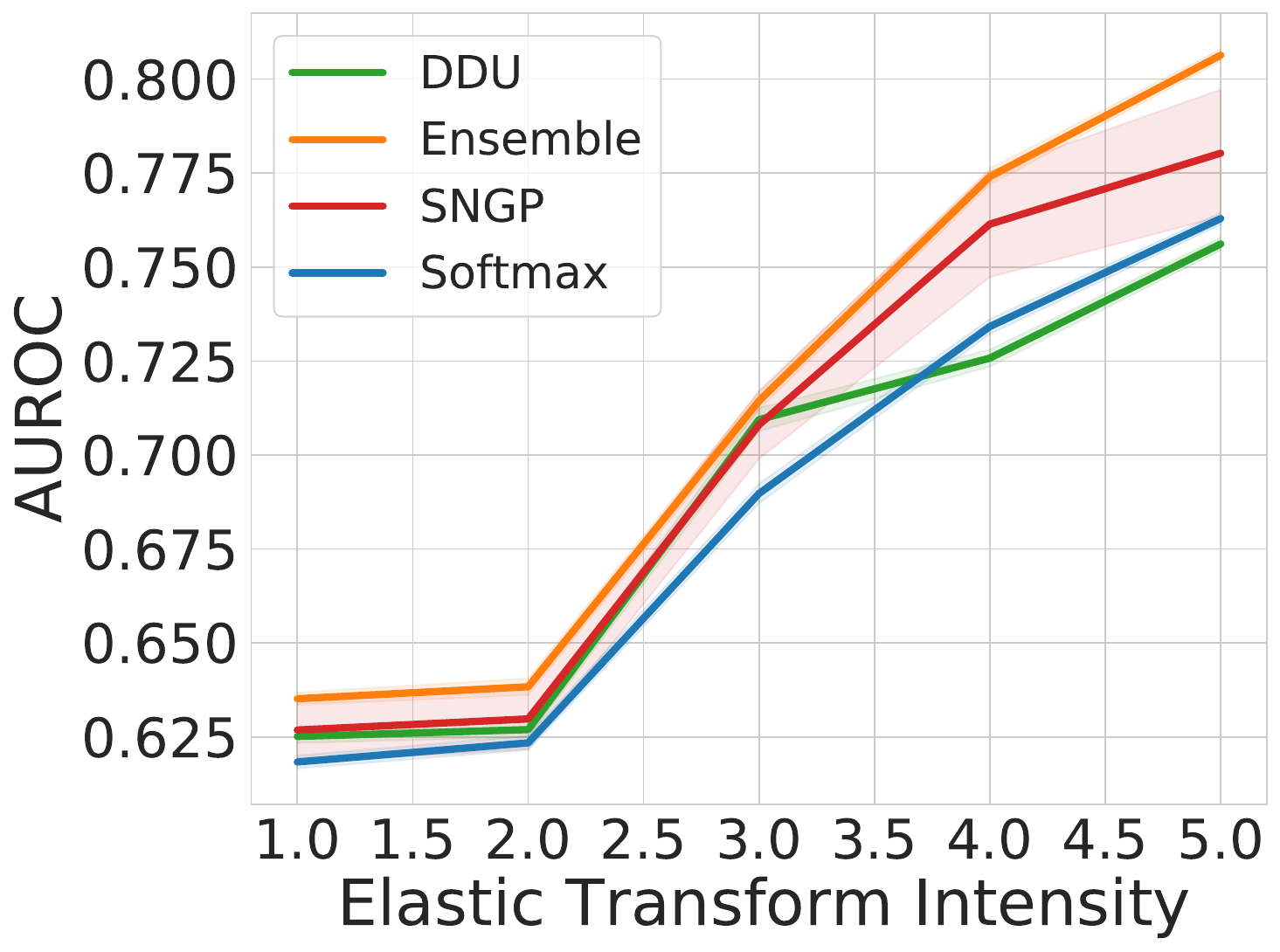}
    \end{subfigure}\hfill
    \begin{subfigure}{0.25\linewidth}
        \centering
        \includegraphics[width=\linewidth]{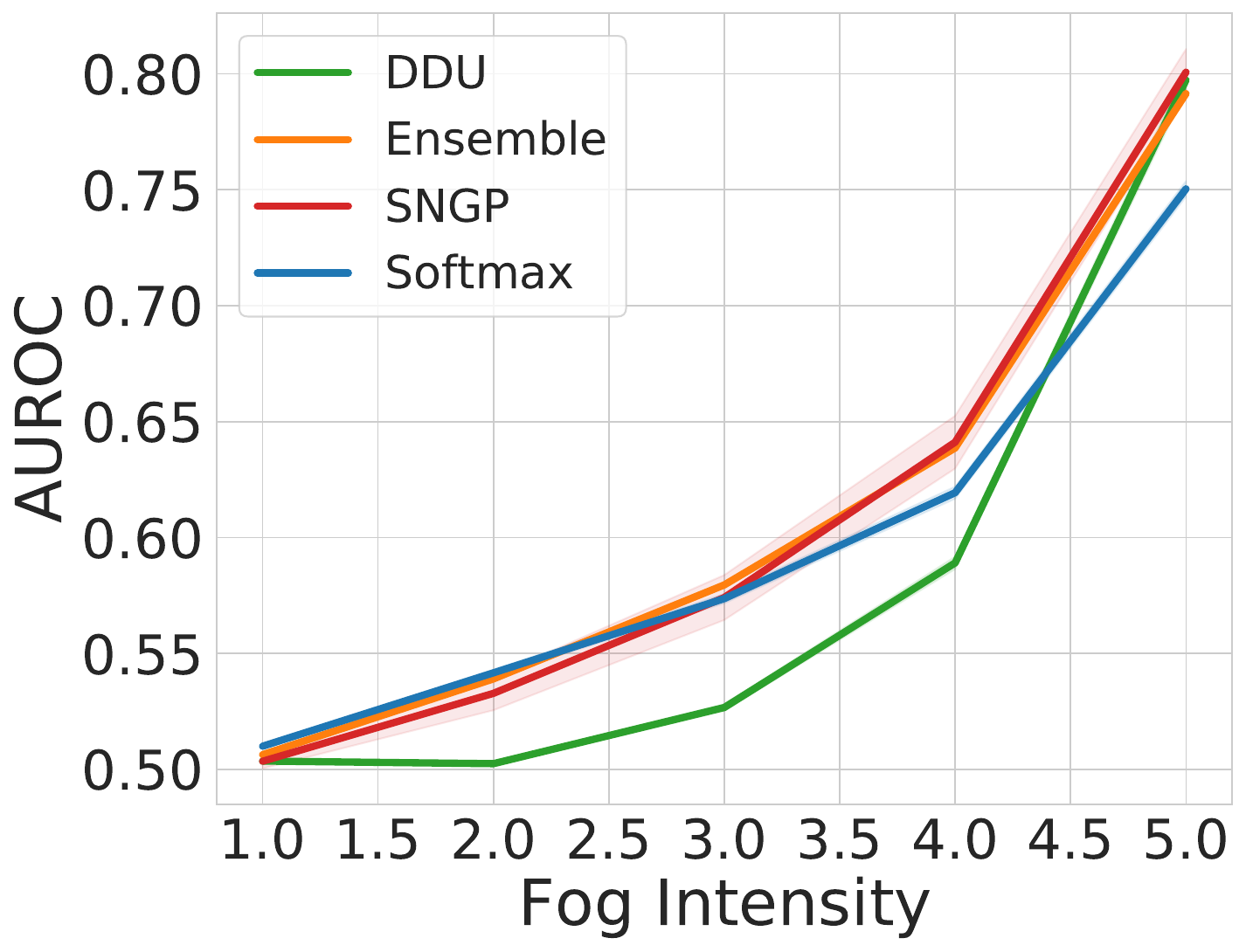}
    \end{subfigure}\hfill
    \begin{subfigure}{0.25\linewidth}
        \centering
        \includegraphics[width=\linewidth]{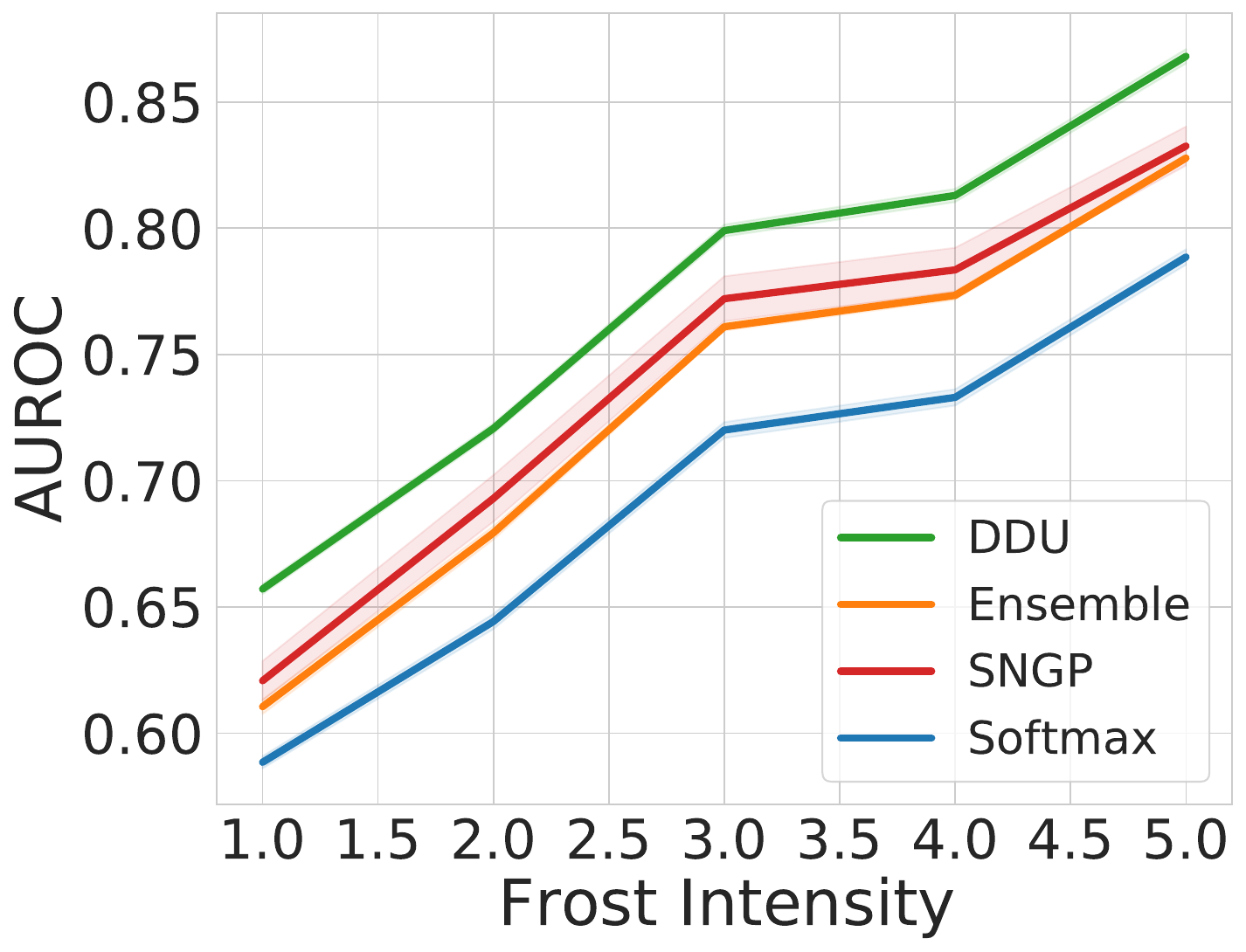}
    \end{subfigure}\hfill
    \begin{subfigure}{0.25\linewidth}
        \centering
        \includegraphics[width=\linewidth]{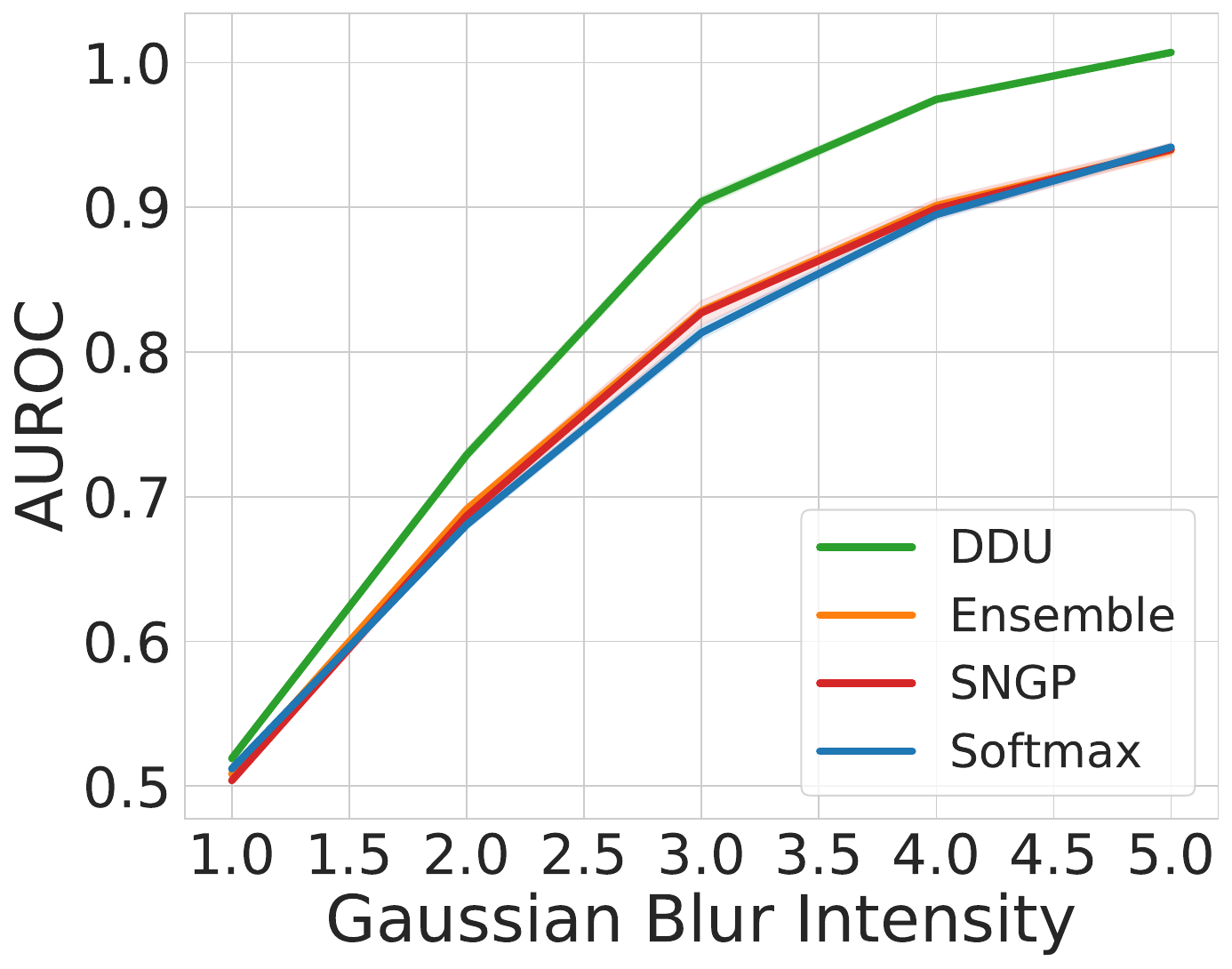}
    \end{subfigure}\hfill
    \begin{subfigure}{0.25\linewidth}
        \centering
        \includegraphics[width=\linewidth]{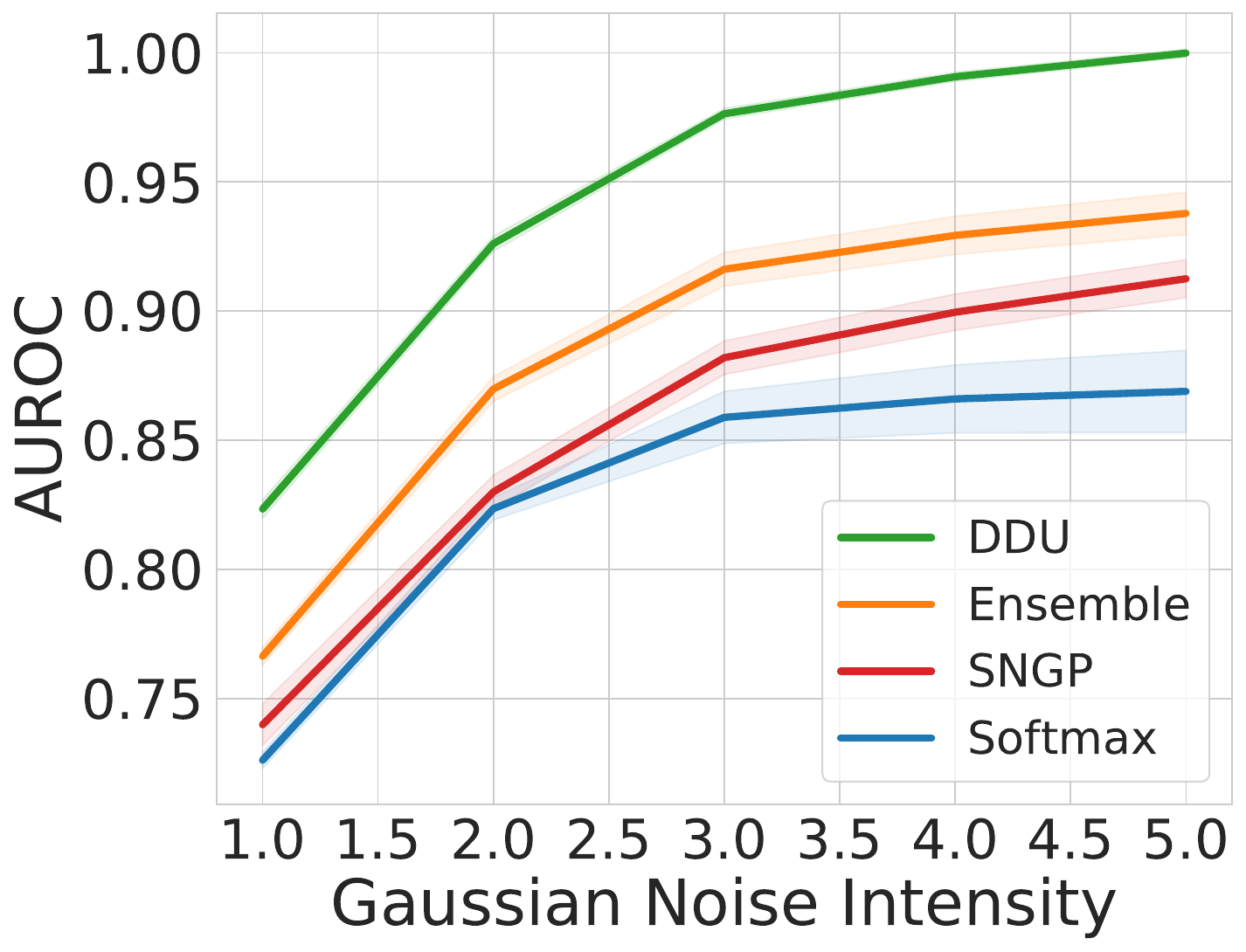}
    \end{subfigure}\hfill
    \begin{subfigure}{0.25\linewidth}
        \centering
        \includegraphics[width=\linewidth]{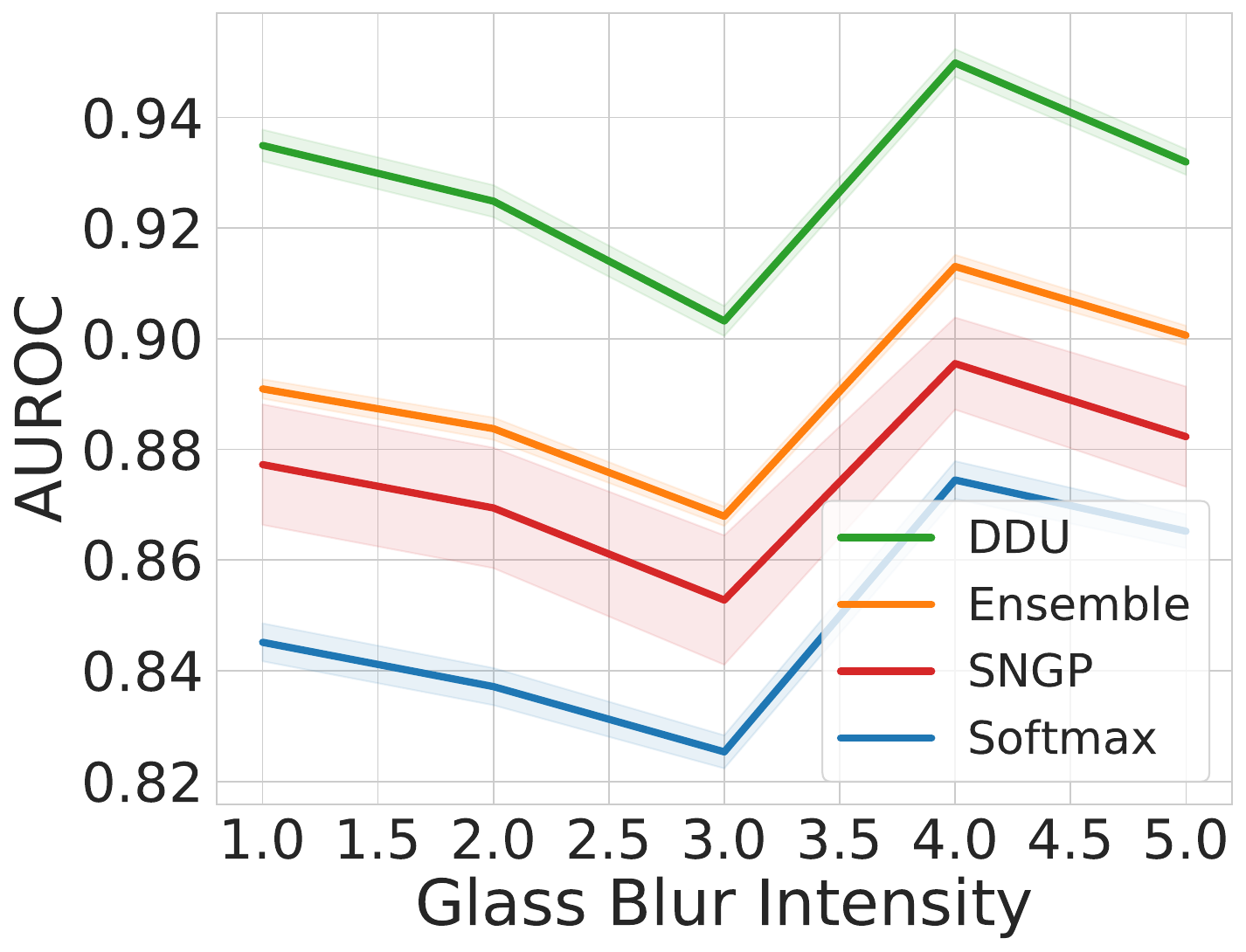}
    \end{subfigure}\hfill
    \begin{subfigure}{0.25\linewidth}
        \centering
        \includegraphics[width=\linewidth]{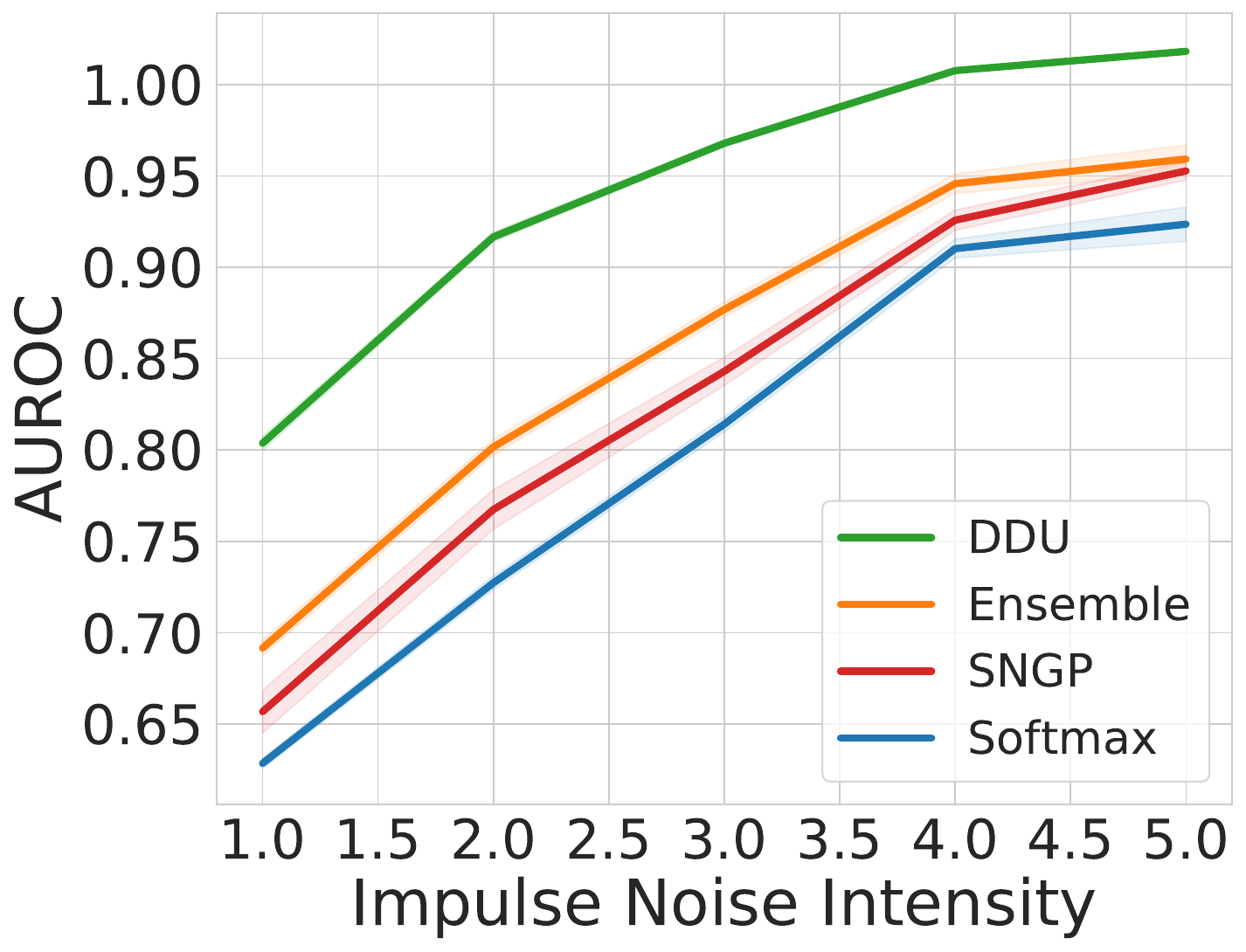}
    \end{subfigure}\hfill
    \begin{subfigure}{0.25\linewidth}
        \centering
        \includegraphics[width=\linewidth]{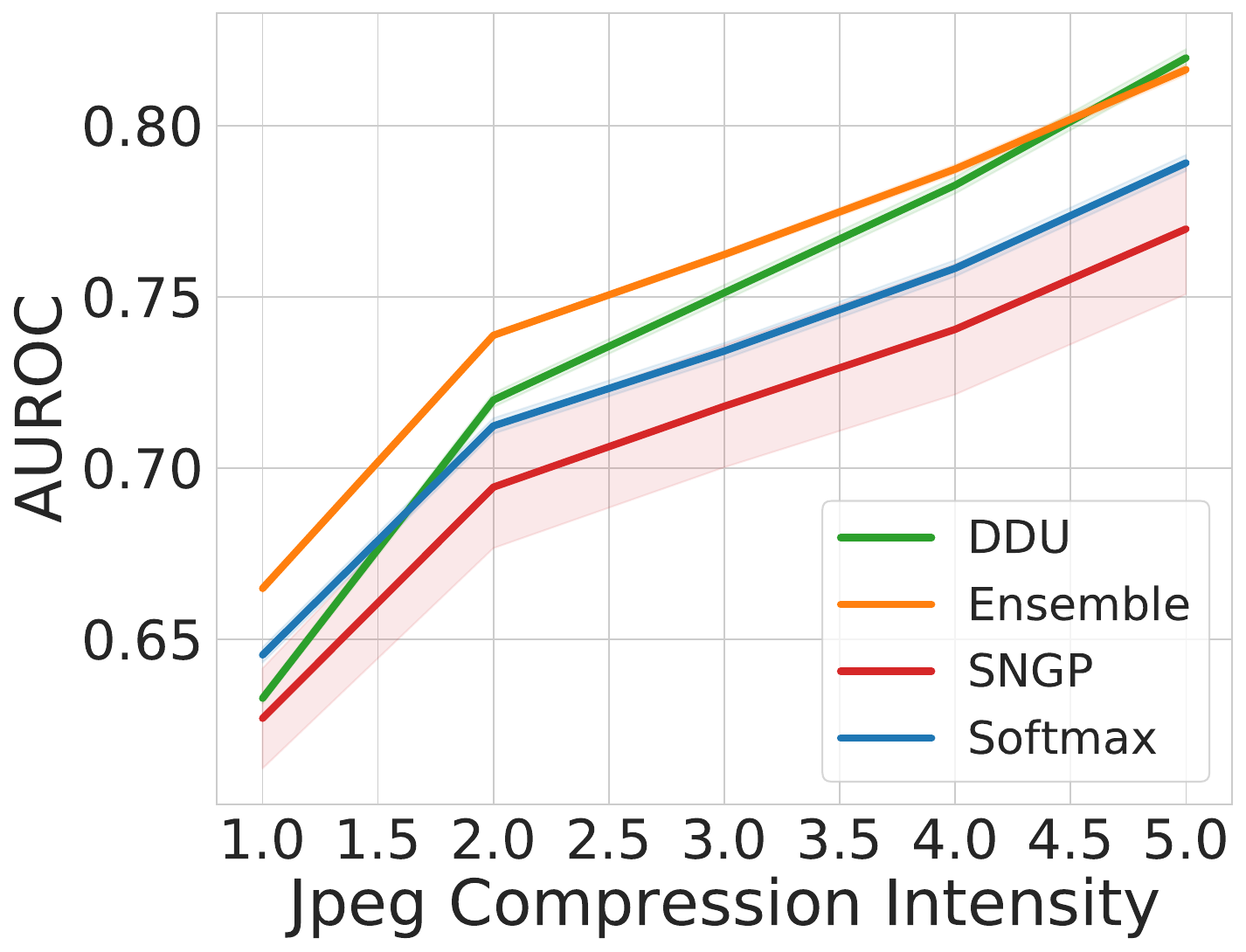}
    \end{subfigure}\hfill
    \begin{subfigure}{0.25\linewidth}
        \centering
        \includegraphics[width=\linewidth]{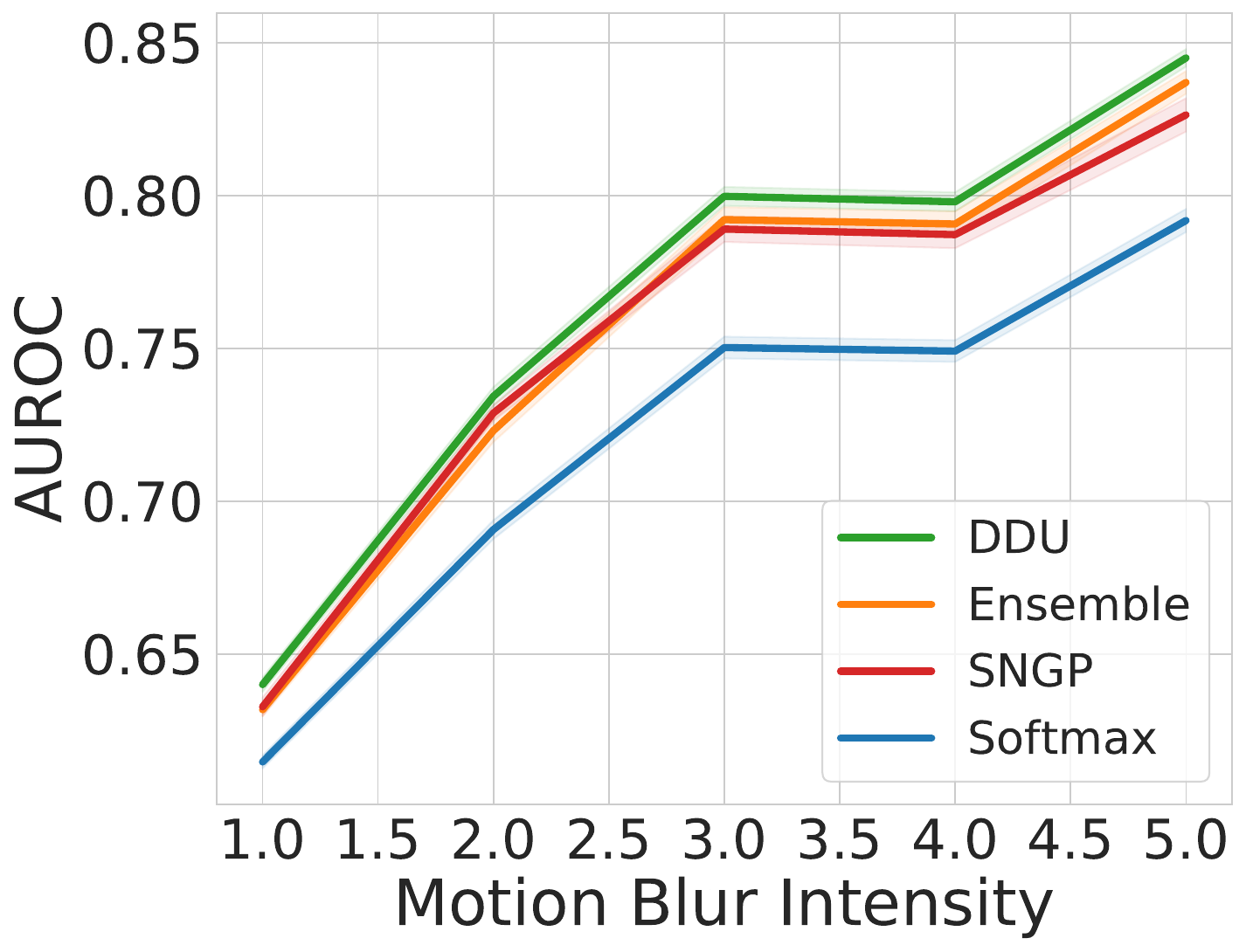}
    \end{subfigure}\hfill
    \begin{subfigure}{0.25\linewidth}
        \centering
        \includegraphics[width=\linewidth]{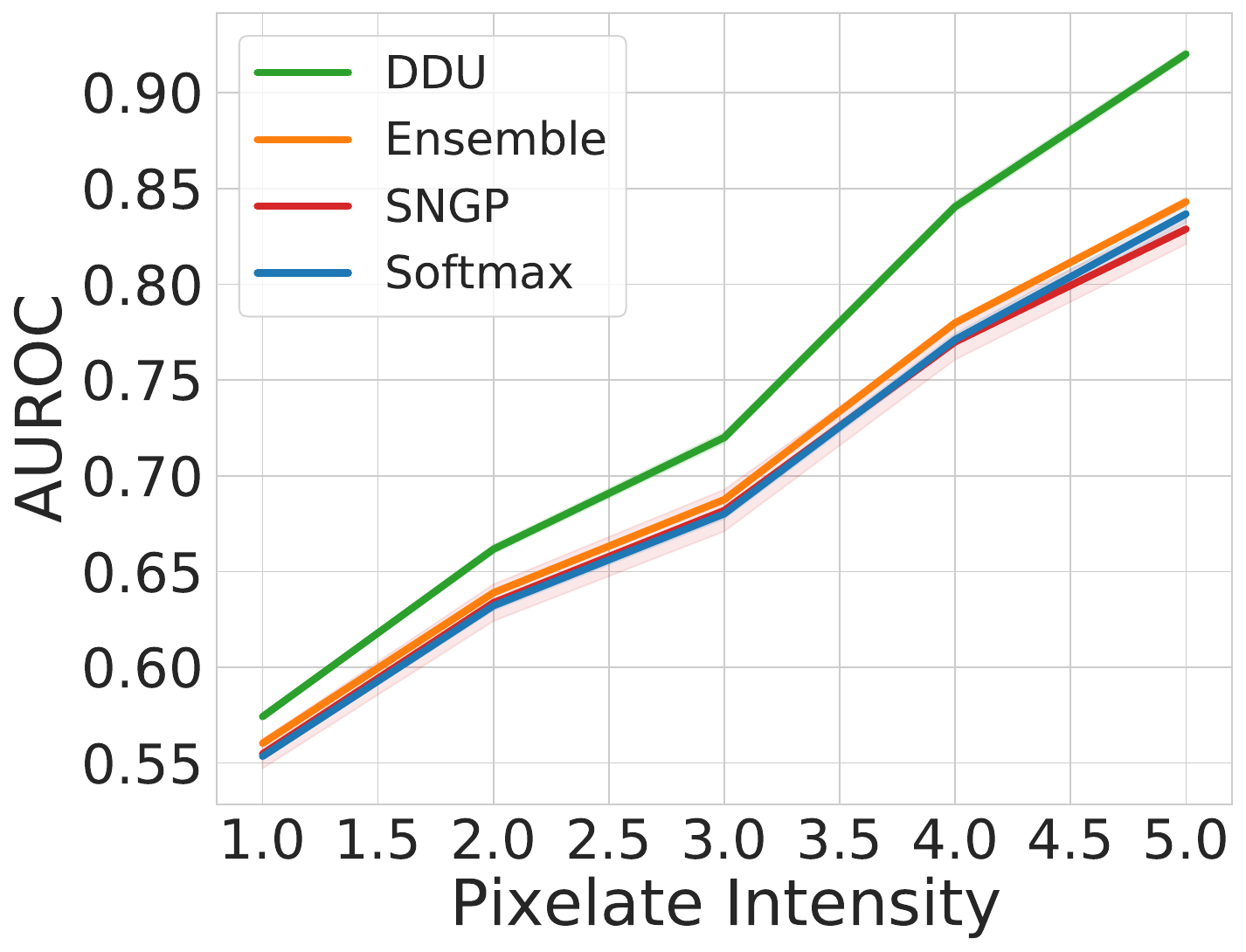}
    \end{subfigure}\hfill
    \begin{subfigure}{0.25\linewidth}
        \centering
        \includegraphics[width=\linewidth]{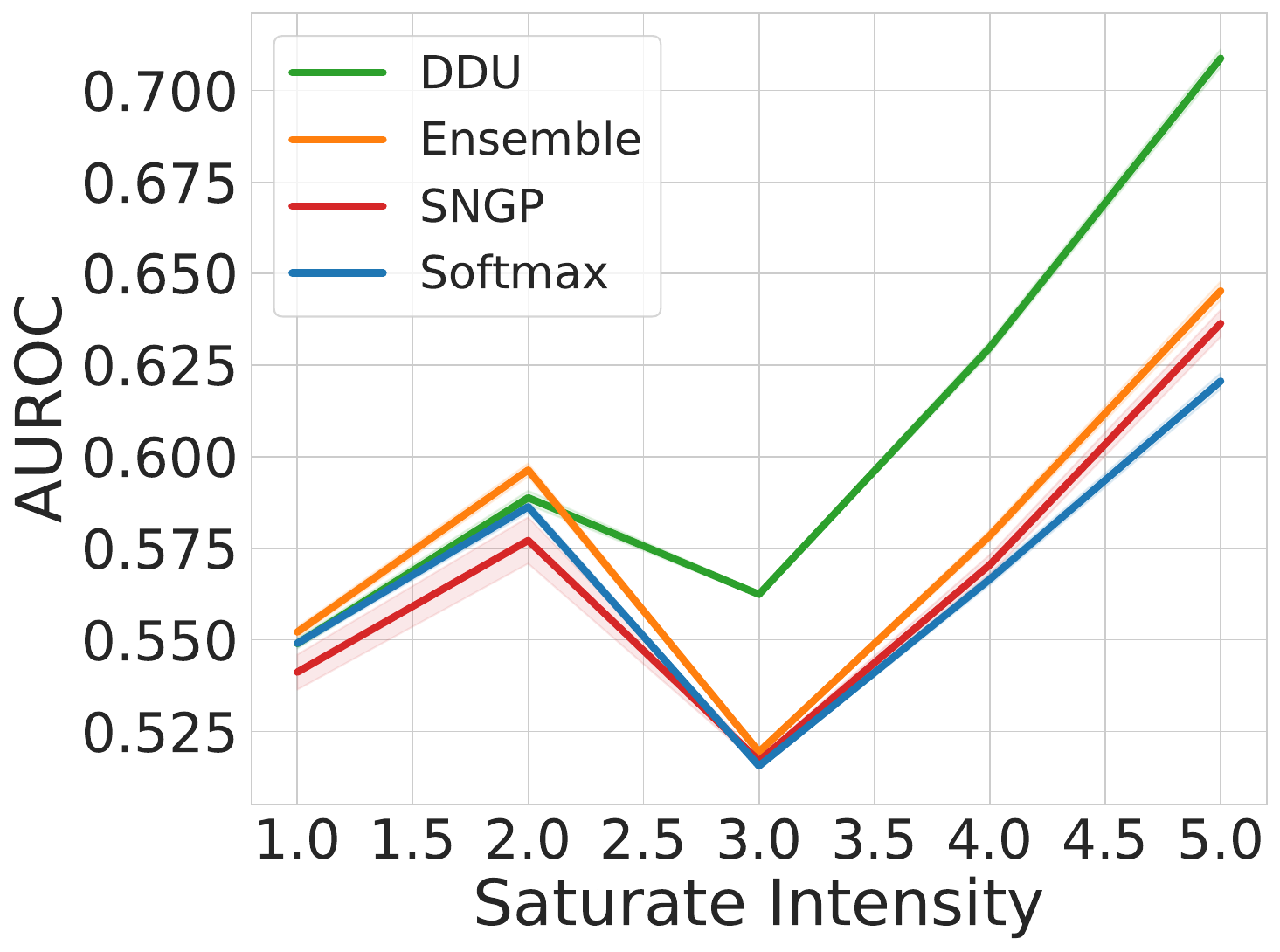}
    \end{subfigure}\hfill
    \begin{subfigure}{0.25\linewidth}
        \centering
        \includegraphics[width=\linewidth]{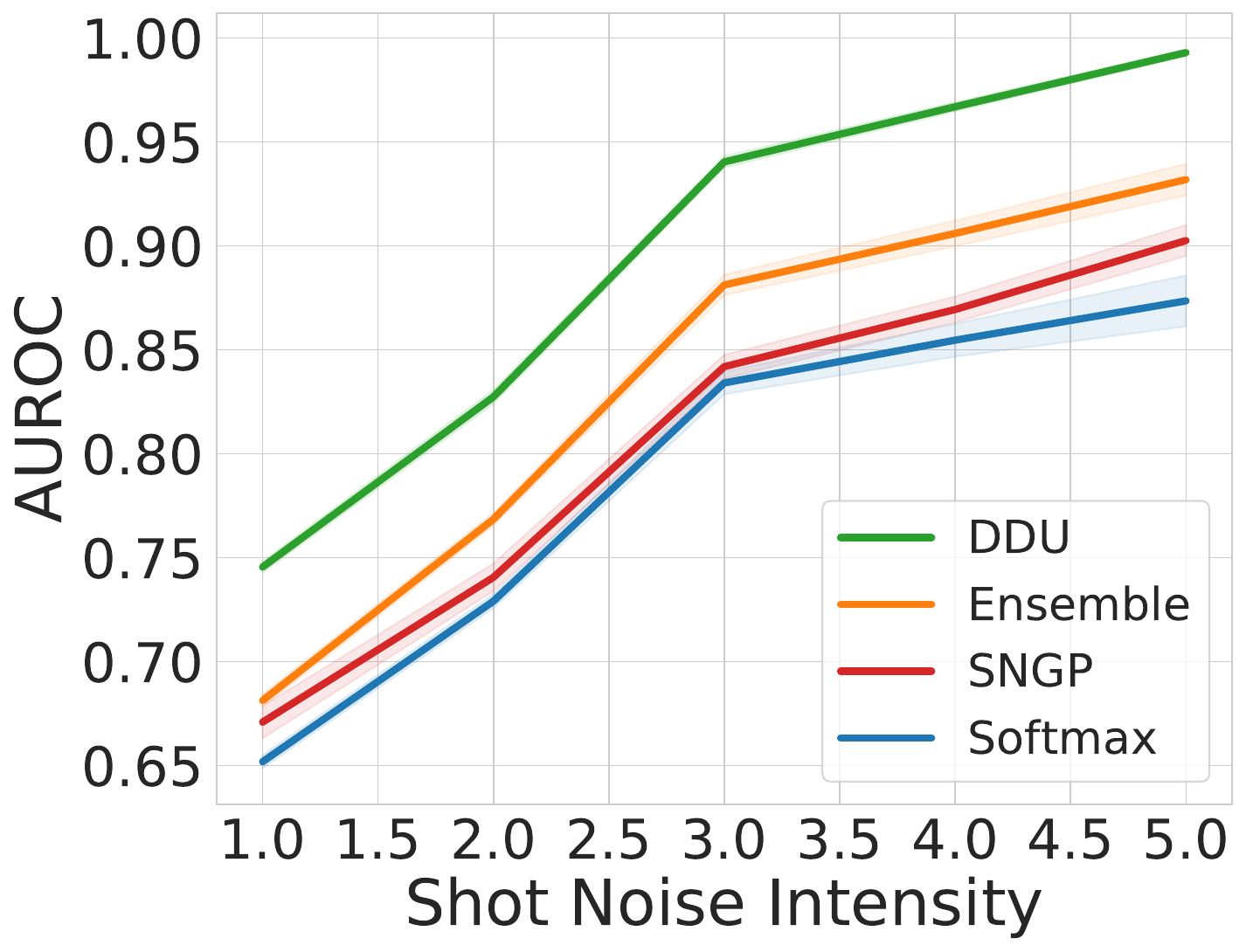}
    \end{subfigure}\hfill
    \begin{subfigure}{0.25\linewidth}
        \centering
        \includegraphics[width=\linewidth]{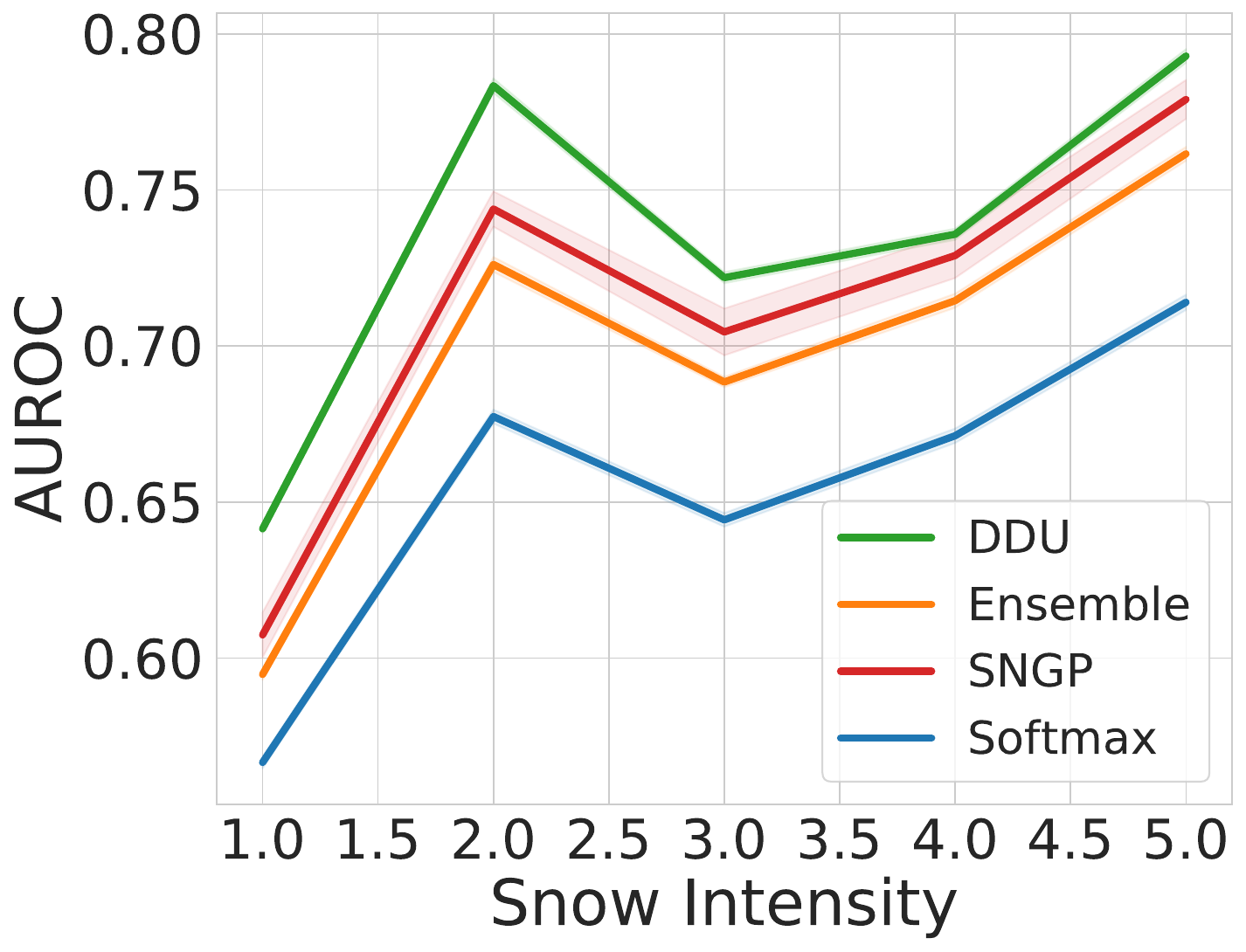}
    \end{subfigure}\hfill
    \begin{subfigure}{0.25\linewidth}
        \centering
        \includegraphics[width=\linewidth]{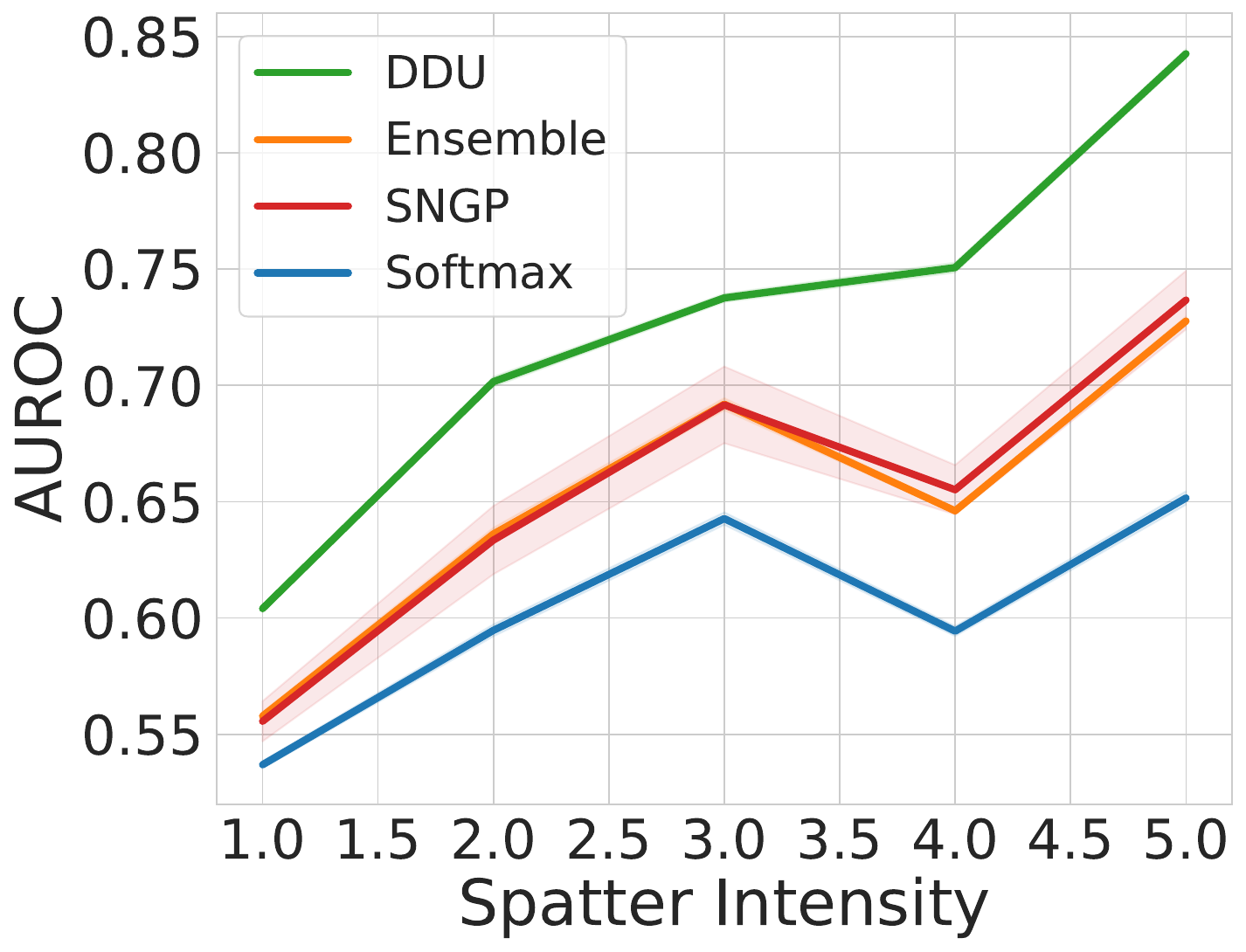}
    \end{subfigure}
    \begin{subfigure}{0.25\linewidth}
        \centering
        \includegraphics[width=\linewidth]{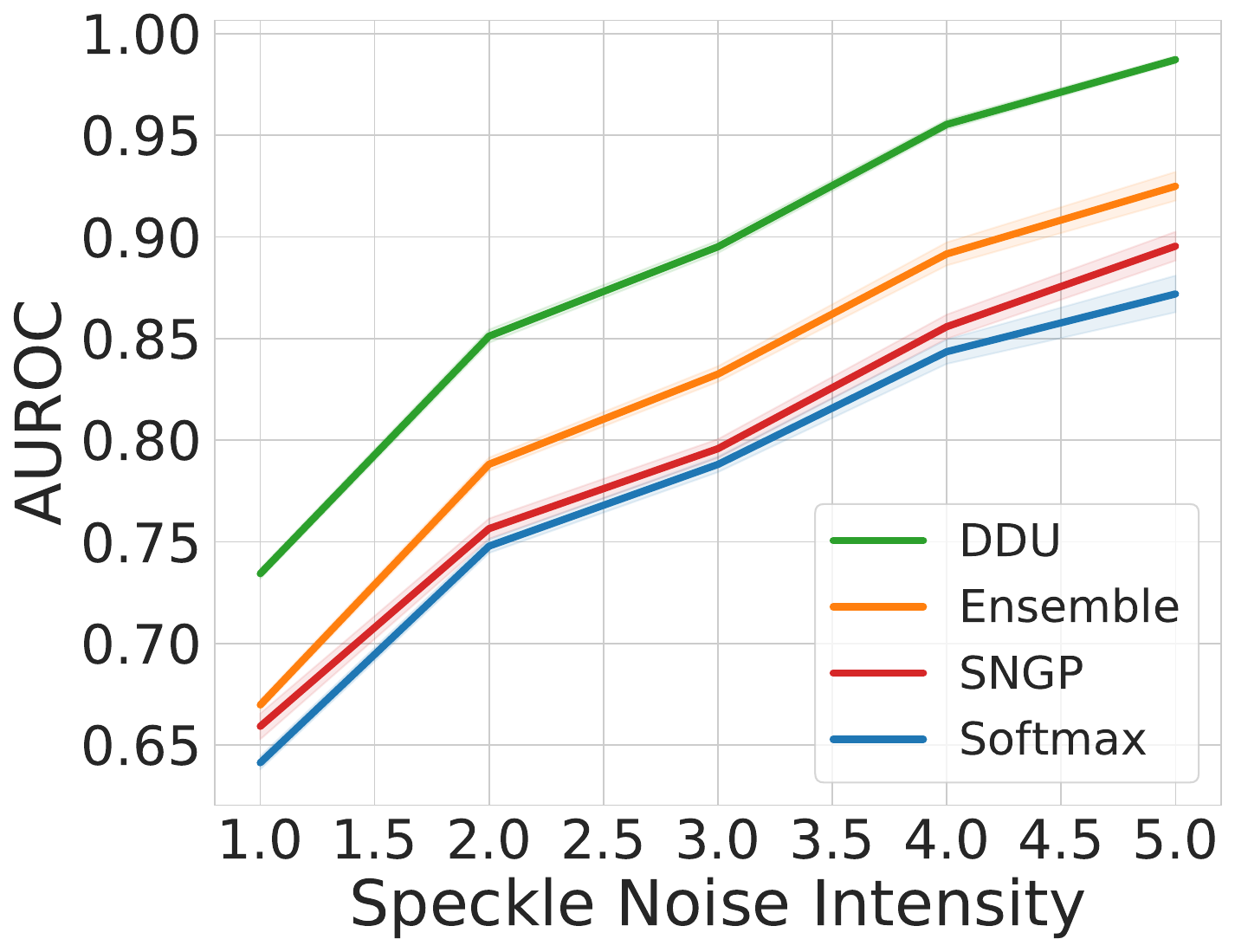}
    \end{subfigure}
    \begin{subfigure}{0.25\linewidth}
        \centering
        \includegraphics[width=\linewidth]{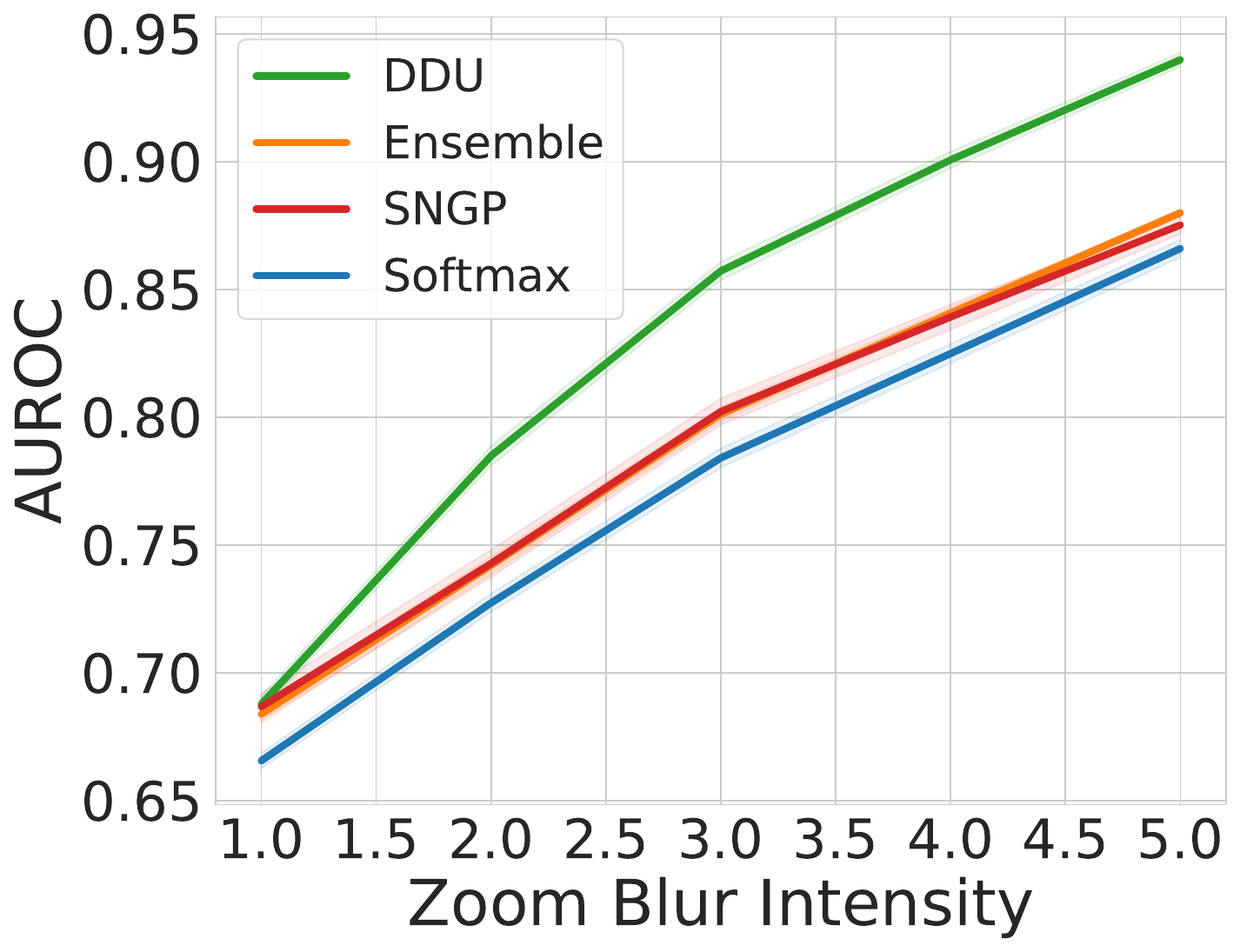}
    \end{subfigure}
    \caption{
    AUROC vs corruption intensity for all corruption types in CIFAR-10-C with DenseNet-121 as the architecture and baselines: Softmax Entropy, Ensemble (using Predictive Entropy as uncertainty), SNGP and DDU feature density.
    }
    \label{ddu:fig:cifar10_c_results_densenet121}
\end{figure}

\section{Additional Results}
\label{ddu:app:additional_exp_results}

In this section, we provide details of additional results on the OoD detection task using CIFAR-10 vs SVHN/CIFAR-100/Tiny-ImageNet/CIFAR-10-C and CIFAR-100 vs SVHN/Tiny-ImageNet for ResNet-50, ResNet-110 and DenseNet-121 architectures. We present results on ResNet-50, ResNet-110 and DenseNet-121 for CIFAR-10 vs SVHN/CIFAR-100/Tiny-ImageNet and CIFAR-100 vs SVHN/Tiny-ImageNet in \Cref{ddu:table:ood_resnet50}, \Cref{ddu:table:ood_resnet110} and \Cref{ddu:table:ood_densenet121} respectively. We also present results on individual corruption types for CIFAR-10-C for Wide-ResNet-28-10, ResNet-50, ResNet-110 and DenseNet-121 in \Cref{ddu:fig:cifar10_c_results_wide_resnet}, \Cref{ddu:fig:cifar10_c_results_resnet50}, \Cref{ddu:fig:cifar10_c_results_resnet110} and \Cref{ddu:fig:cifar10_c_results_densenet121} respectively.

Finally, we provide results for various ablations on DDU. As mentioned in \S\ref{ddu:section:methods}, DDU consists of a deterministic softmax model trained with appropriate inductive biases. It uses softmax entropy to quantify aleatoric uncertainty and feature-space density to quantify epistemic uncertainty. In the ablation, we try to experimentally evaluate the following scenarios:
\begin{enumerate}[leftmargin=*]
    \item \textbf{Effect of inductive biases (sensitivity + smoothness):} We want to see the effect of removing the proposed inductive biases (i.e.\ no sensitivity and smoothness constraints) on the OoD detection performance of a model. To do this, we train a VGG-16 with and without spectral normalization. Note that VGG-16 does not have residual connections and hence, a VGG-16 does not follow the sensitivity and smoothness (bi-Lipschitz) constraints.
    
    \item \textbf{Effect of sensitivity alone:} Since residual connections make a model sensitive to changes in the input space by lower bounding its Lipschitz constant, we also want to see how a network performs with just the sensitivity constraint alone. To observe this, we train a Wide-ResNet-28-10 without spectral normalization (i.e.\ no explicit upper bound on the Lipschitz constant of the model).
    
    \item \textbf{Metrics for aleatoric and epistemic uncertainty:} With the above combinations, we try to observe how different metrics for aleatoric and epistemic uncertainty perform. To quantify aleatoric uncertainty, we use the softmax entropy of the model.
    On the other hand, to quantify the epistemic uncertainty, we use \textbf{i)} the softmax entropy, \textbf{ii)} the softmax density \citep{liu2020energy} or \textbf{iii)} the GMM feature density (as described in \S\ref{ddu:section:methods}). 
\end{enumerate}

For the purposes of comparison, we also present scores obtained by a 5-Ensemble of the respective architectures (i.e.\ Wide-ResNet-28-10 and VGG-16) in \Cref{ddu:table:ood_2} for CIFAR-10 vs SVHN/CIFAR-100 and in \Cref{ddu:table:ood_3} for CIFAR-100 vs SVHN. Based on these results, we can make the following observations (in addition to the ones we make in \S\ref{ddu:sec:experiments_ood_detection}):

\textbf{Inductive biases are important for feature density.} From the AUROC scores in \Cref{ddu:table:ood_2}, we can see that using the feature density of a GMM in VGG-16 without the proposed inductive biases yields significantly lower AUROC scores as compared to Wide-ResNet-28-10 with inductive biases. In fact, in none of the datasets is the feature density of a VGG able to outperform its corresponding ensemble. This provides yet more evidence (in addition to \Cref{ddu:fig:intro_histograms}) to show that the GMM feature density alone cannot estimate epistemic uncertainty in a model that suffers from feature collapse. We need sensitivity and smoothness conditions (see \S\ref{ddu:sec:motivation}) on the feature space of the model to obtain feature densities that capture epistemic uncertainty.

\textbf{Sensitivity creates a bigger difference than smoothness.} We note that the difference between AUROC obtained from feature density between Wide-ResNet-28-10 models with and without spectral normalization is minimal. Although Wide-ResNet-28-10 with spectral normalization (i.e.\ smoothness constraints) still outperforms its counterpart without spectral normalization, the small difference between the AUROC scores indicates that it might be the residual connections (i.e.\ sensitivity constraints) that make the model detect OoD samples better. This observation is also intuitive as a sensitive feature extractor should map OoD samples farther from iD ones.

\textbf{DDU as a simple baseline.} In DDU, we use the softmax output of a model to get aleatoric uncertainty. We use the GMM's feature-density to estimate the epistemic uncertainty. Hence, DDU does not suffer from miscalibration as the softmax outputs can be calibrated using post-hoc methods like temperature scaling. At the same time, the feature-densities of the model are not affected by temperature scaling and capture epistemic uncertainty well.

\section{Additional Ablations \&\ Toy Experiments}
Here, we provide details for the toy experiments mentioned in the main part of the thesis 
which are visualized in \Cref{ddu:fig:intro_histograms}, \Cref{ddu:fig:lewis_vis} and \Cref{ddu:fig:kendall_viz_2}.

\subsection{QUBIQ Challenge}
\label{ddu:app:qubiq}

In this section, we evaluate DDU's performance on the real-world QUBIQ challenge related to biomedical imaging. QUBIQ has a total of 7 binary segmentation tasks in 4 biomedical imaging datasets with multiple annotations per image. The task is to predict the distribution of source labels with a mask of values between 0 and 1. For evaluation, the annotations are averaged to provide a continuous ground-truth. The prediction mask and continuous ground-truth are binarized by thresholding between $[0,1]$ and a Dice score is computed between the resulting binary masks. The average dice score across thresholds, images and tasks is reported. Note that in the continuous ground-truth, 0.5 indicates maximum uncertainty and values above or below indicate lower uncertainty. Thus, for our comparison, we scale all uncertainty values to the range $u \in [0, 0.5]$ and use  $p+u$ if $p=0$ and $p-u$ if $p=1$, where $p$ is the binary prediction. We use a UNet model with a ResNet encoder and report the Dice scores averaged over 5 runs in \Cref{ddu:table:qubiq}. Even on this real-world dataset, DDU performs as well as ensembles and outperforms Softmax and Energy baselines.

\begin{table}[t]
    \caption{Dice scores for the QUBIQ 2021 challenge.}
    \label{ddu:table:qubiq}
    \centering
    \begin{tabular}{@{}lcccc@{}}
    \toprule
    \textbf{Method} & \textbf{Softmax} & \textbf{Energy} & \textbf{3-Ensemble PE} & \textbf{DDU} \\
    \midrule
    Dice Score (\textuparrow) & $78.4 \pm 1.31$ & $77.31 \pm 1.5$ & $82.25 \pm 0.83$ & $\mathbf{82.63 \pm 1.08}$ \\
    \bottomrule
    \end{tabular}
\end{table}

\subsection{Additional Calibration Metrics}
\label{ddu:app:additional_calibration_metrics}

In \Cref{ddu:table:tace_nll}, in addition to ECE\%, we provide additional calibration error scores: temperature scaled Thresholded Adaptive Calibration Error (TACE) \% and Negative Log Likelihood (NLL) for Wide-ResNet-28-10 trained on CIFAR-10/100---the main results for this can be found in \Cref{ddu:table:ood_wrn}. Here, we see that the results for TACE and NLL are consistent with what we see for ECE. Ensembles produce the most calibrated models and among deterministic baselines, DDU is the best calibrated.

\begin{table}[!t]
    \centering
    \scriptsize
    \resizebox{\linewidth}{!}
    {
    \begin{tabular}{llrrrrr}
    \textbf{Dataset} & \textbf{Metric} & \textbf{Softmax \&\ Energy} & \textbf{DUQ} & \textbf{SNGP} & \textbf{DDU} & \textbf{5-Ensemble} \\
    \midrule
    \multirow{3}{*}{CIFAR-10} & ECE & $\mathbf{0.85 \pm 0.02}$ & $1.55 \pm 0.08$ & $1.8 \pm 0.1$ & $\mathbf{0.85 \pm 0.04}$ & $\mathbf{0.76 \pm 0.03}$ \\
                         & TACE & $0.63\pm0.01$ & $0.84\pm0.03$ & $0.9\pm0.04$ & $\mathbf{0.61\pm0.01}$ & $\mathbf{0.48\pm0.01}$ \\
                         & NLL & $0.18 \pm 0.06$ & $0.23 \pm 0.07$ & $0.27 \pm 0.08$ & $\mathbf{0.16 \pm 0.06}$ & $\mathbf{0.11 \pm 0.02}$\\
    \midrule
    \multirow{3}{*}{CIFAR-100} & ECE & $4.62 \pm 0.06$ & - & $4.33 \pm 0.01$ & $\mathbf{4.1 \pm 0.08}$ & $\mathbf{3.32 \pm 0.09}$ \\
                          & TACE & $1.31\pm0.02$ & - & $1.23\pm0.04$ & $\mathbf{1.06\pm0.03}$ & $\mathbf{0.58\pm0.03}$ \\
                          & NLL & $1.17 \pm 0.13$ & - & $0.92 \pm 0.16$ & $\mathbf{0.86 \pm 0.14}$ & $\mathbf{0.73 \pm 0.09}$\\
    \bottomrule
    \end{tabular}
    }
    \caption{Calibration error scores ECE\%, TACE\% and NLL for WRN-28-10.}
    \label{ddu:table:tace_nll}
\end{table}

\section{Big Figure 1}
\begin{sidewaysfigure}[t!]
    \centering
    \begin{minipage}[t]{0.7\linewidth}
    \begin{subfigure}{\linewidth}
        \includegraphics[width=\linewidth]{figs/dirty_ambiguous_fashion_mnist.pdf}%
        \caption{{\color{sns-blue} Dirty-MNIST (iD)} and {\color{sns-orange} Fashion-MNIST (OoD)}}
        \label{ddu:app_fig:intro_sample_viz}
    \end{subfigure}
    \begin{subfigure}{\linewidth}
        \begin{subfigure}{0.33\linewidth}
            \centering
            \includegraphics[width=\linewidth]{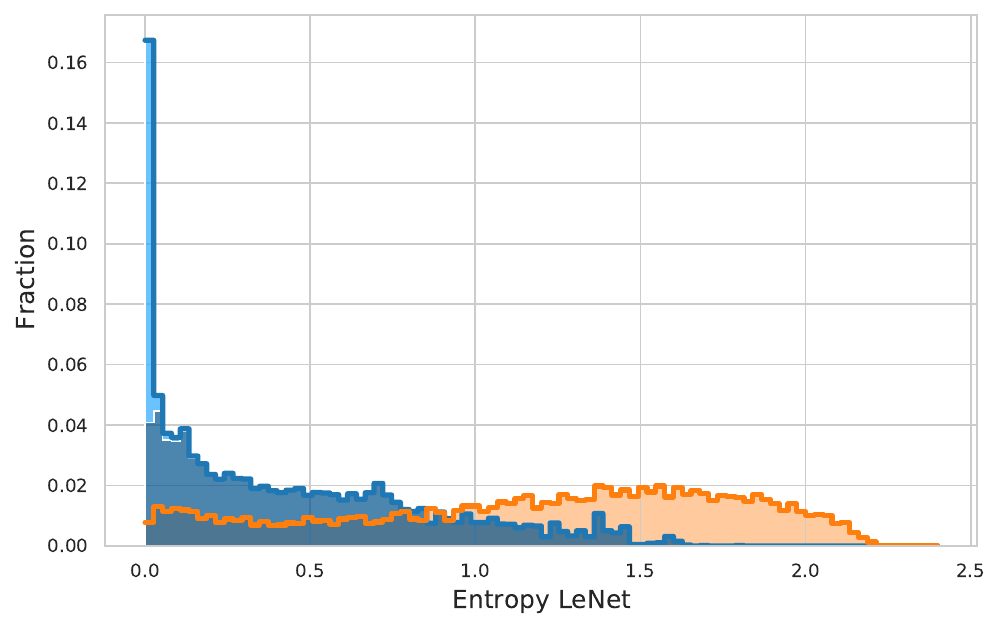}%
        \end{subfigure}\hfill
        \begin{subfigure}{0.33\linewidth}
            \centering
            \includegraphics[width=\linewidth]{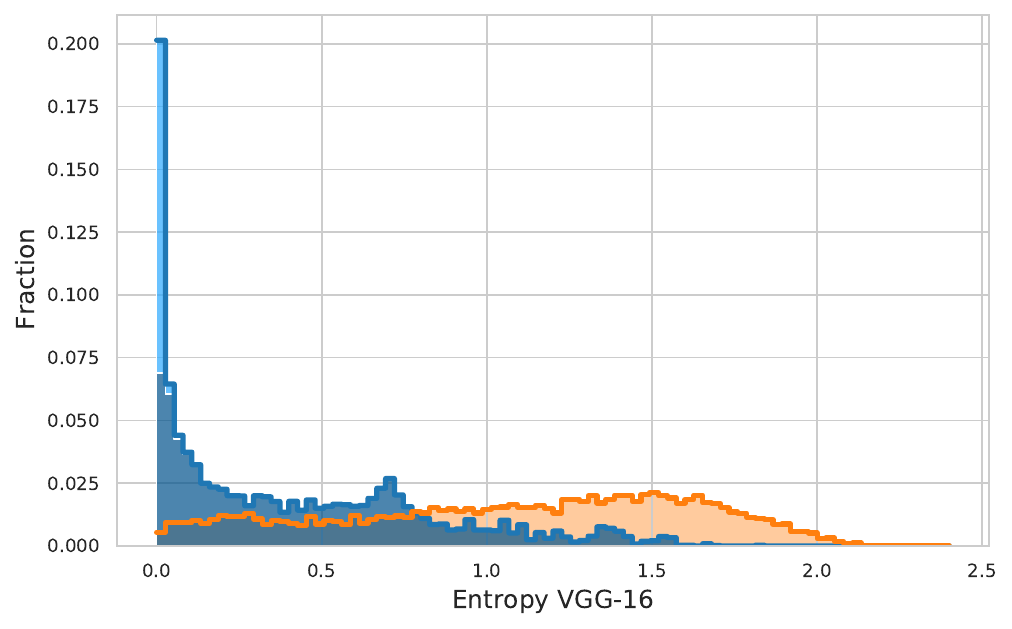}%
        \end{subfigure}\hfill
        \begin{subfigure}{0.33\linewidth}
            \centering
            \includegraphics[width=\linewidth]{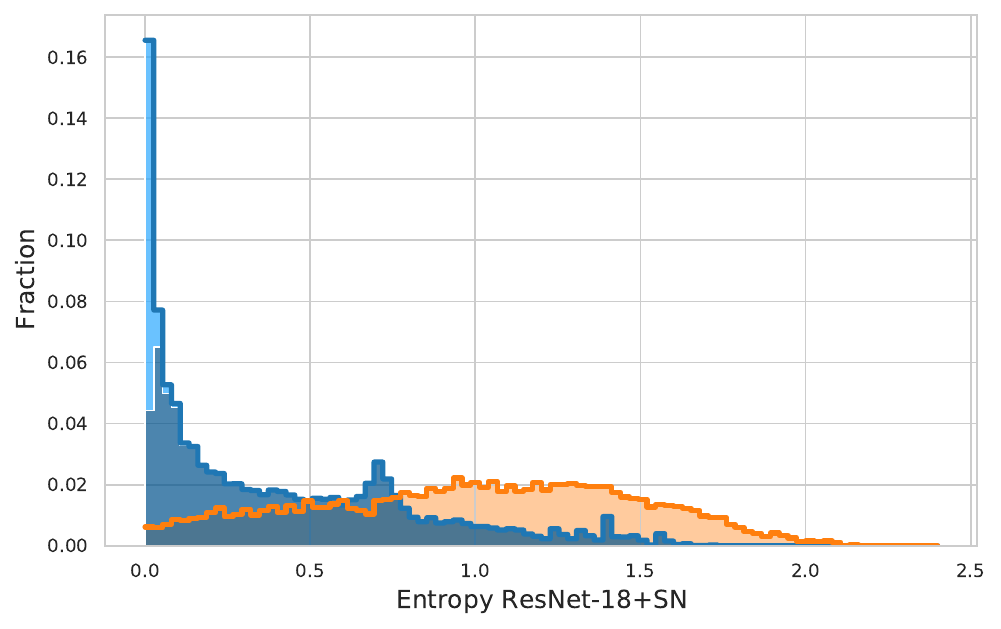}%
        \end{subfigure}%
        \subcaption{{Softmax entropy}}
        \label{ddu:app_fig:intro_softmax_ent}
    \end{subfigure}
    \begin{subfigure}{\linewidth}
        \begin{subfigure}{0.33\linewidth}
            \centering
            \includegraphics[width=\linewidth]{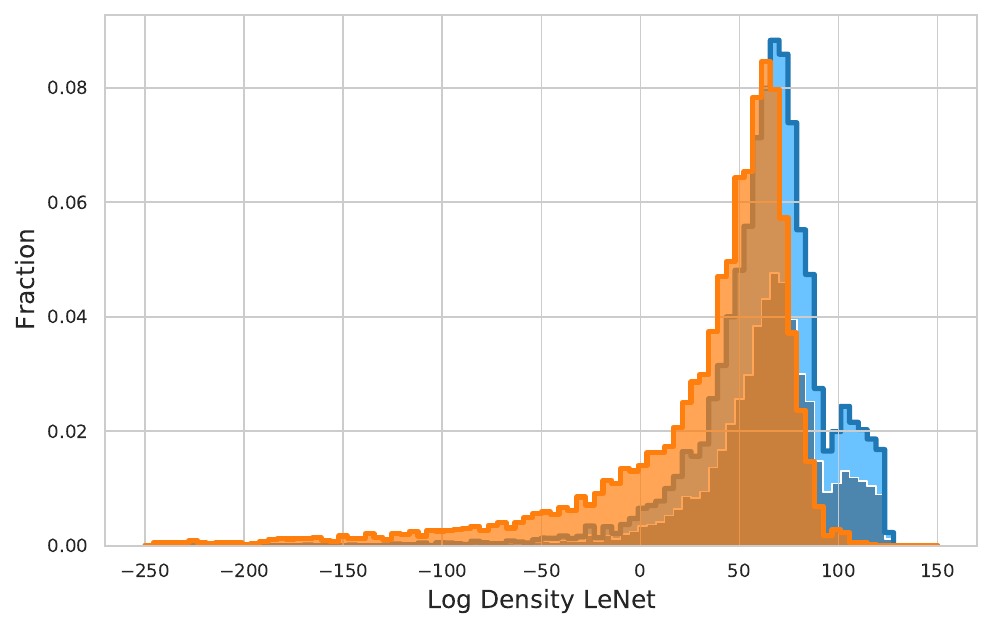}%
        \end{subfigure}\hfill
        \begin{subfigure}{0.33\linewidth}
            \centering
            \includegraphics[width=\linewidth]{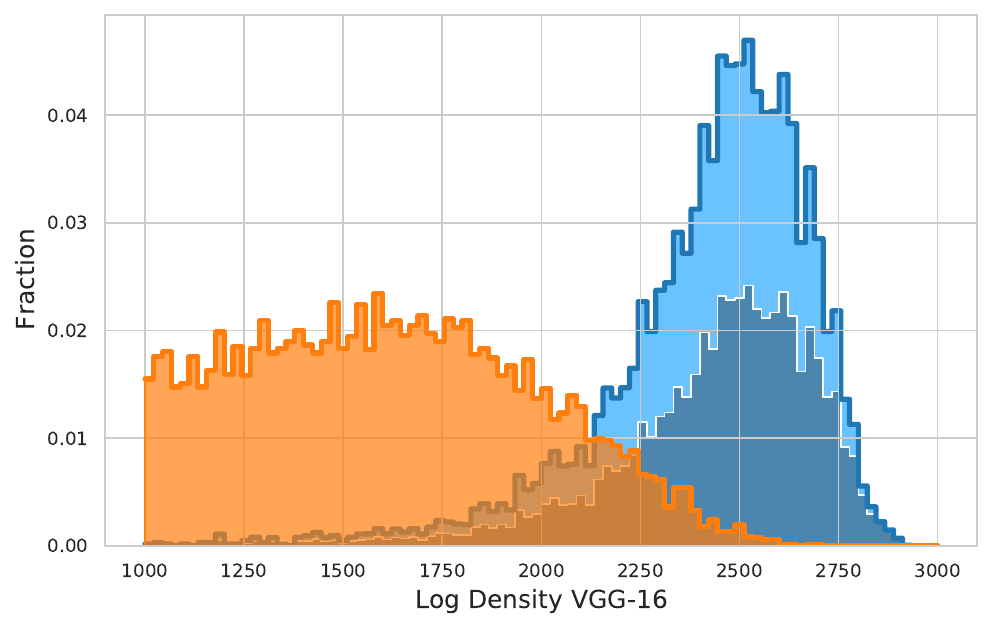}%
        \end{subfigure}\hfill
        \begin{subfigure}{0.33\linewidth}
            \centering
            \includegraphics[width=\linewidth]{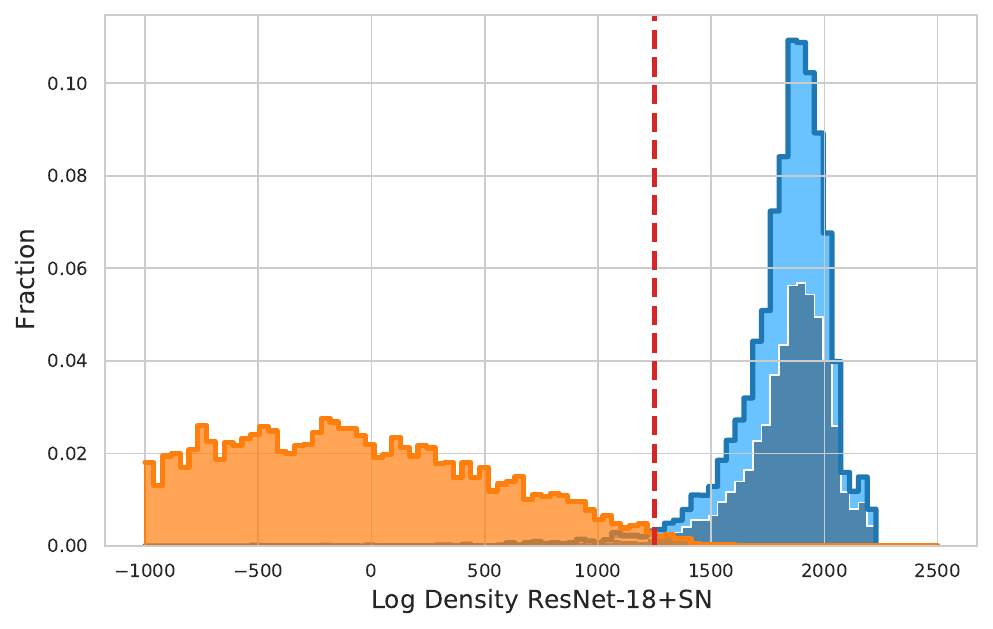}%
        \end{subfigure}%
        \subcaption{{Feature-space density}}
        \label{ddu:app_fig:intro_gmm}
    \end{subfigure}%
    \end{minipage}
    \caption{
    \emph{
    Disentangling aleatoric and epistemic uncertainty on {\color{sns-blue}Dirty-MNIST (iD)} and {\color{sns-orange} Fashion-MNIST (OoD)} \textbf{\subref{ddu:app_fig:intro_sample_viz}} requires using \emph{softmax entropy} \textbf{\subref{ddu:app_fig:intro_softmax_ent}} and \emph{feature-space density (GMM)} \textbf{\subref{ddu:app_fig:intro_gmm}} with appropriate inductive biases (\emph{ResNet-18+SN} vs \emph{LeNet} \&\ \emph{VGG-16} without them). Enlarged version.
    }
    \textbf{\subref{ddu:app_fig:intro_softmax_ent}:}
    Softmax entropy captures aleatoric uncertainty for iD data (Dirty-MNIST), thereby separating {\color{sns-nonambiguous}unambiguous MNIST samples} and {\color{sns-ambiguous}Ambiguous-MNIST samples}. However, iD and OoD are confounded: softmax entropy has arbitrary values for OoD, indistinguishable from iD.
    \textbf{\subref{ddu:app_fig:intro_gmm}:}
    With appropriate inductive biases (DDU with ResNet-18+SN), iD and OoD densities do not overlap, capturing epistemic uncertainty. However, without appropriate inductive biases (LeNet \&\ VGG-16), feature density suffers from \emph{feature collapse}: iD and OoD densities overlap. Generally, feature-space density confounds unambiguous and ambiguous iD samples as their densities overlap.
    \textbf{Note:} {\color{sns-nonambiguous}Unambiguous MNIST samples} and {\color{sns-ambiguous}Ambiguous-MNIST samples} are shown as stacked histograms with the total fractions adding up to 1 for {\color{sns-blue}Dirty-MNIST}. 
    }
    \label{ddu:app_fig:intro_histograms}
\end{sidewaysfigure}

}

\chapter{\baldtitle}
{	
	
	\section{Proof of Submodularity}
\label{batchbald:submodular_proof}

\citet{nemhauser1978analysis} show that if a function is submodular, then a
greedy algorithm like algorithm \ref{batchbald:algo:greedy_batchbald} is $1-\nicefrac{1}{e}$-approximate. Here, we show that $\BatchBALDAF$ is submodular.

We will show that $\BatchBALDAF$ satisfies the following equivalent definition of submodularity:

\newcommand{\pa}{X}
\newcommand{\pb}{Y}

\begin{definition}
	A function $f$ defined on subsets of $\Omega$ is called \emph{submodular} if for every set $A \subset \Omega$ and two non-identical points $\pa, \pb \in \Omega \setminus A$:
	\begin{equation}
		f(A \cup \{\pa\}) + f(A \cup \{\pb\}) \ge f(A \cup \{\pa, \pb\}) + f(A) \label{batchbald:eq:submod_ie}
	\end{equation}
\end{definition}

Submodularity expresses that there are "diminishing returns" for adding additional points to $f$.

\begin{lemma}
	$\BatchBALDAF(A, \pw) := \MIof{A; \W}$ is submodular for $A \subset \Dpool$.
\end{lemma}
\begin{proof}
	\newcommand{\MIw}[1]{\MIof{#1; \W}}
	\newcommand{\MIcw}[2]{\MIof{#1; \W \given #2}}
	Let $\pa, \pb \in \Dpool, \pa \ne \pb$.
	We start by substituting the definition of $\BatchBALDAF$ into \eqref{batchbald:eq:submod_ie} and subtracting $\MIof{A; \W}$ twice on both sides, using that $\MIof{A \cup B; \W} - \MIof{B; \W} = \MIof{A ; \W \given B}$:
	\begin{align}
		&\MIw{A \cup \{Y\}} + \MIw{A \cup \{X\}} \ge \MIw{A \cup \{X, Y\}} + \MIw{A} \\
	\Leftrightarrow \; &\MIcw{Y}{A} + \MIcw{X}{A} \ge \MIcw{X, y}{A}.
	\end{align}
	We rewrite the left-hand side using the definition of the mutual information $\MIof{A; B} = \Hof{A} - \Hof{A \given B}$ and reorder:
	\begin{align}
		&\MIcw{y}{A} + \MIcw{x}{A} \\
		=
		&\underbrace{\Hof{\pa \given A} + \Hof{\pa \given A}}_{\ge \Hof{\pa, \pb \given A}}
		-
		\underbrace{\left(\Hof{\pa \given A, \W} + \Hof{\pb \given A, \W}\right)}_{= \Hof{\pa, \pb \given A, \W}}
		\\
		\ge &\Hof{\pa, \pb \given A} - \Hof{\pa, \pb \given A, \W}\\
		=
		&\MIcw{\pa, \pb}{A}
		,
	\end{align}
	where we have used that entropies are subadditive in general and additive given $\pa \independent \pb \; \vert \; \W$.
\end{proof}

Following \citet{nemhauser1978analysis}, we can conclude that algorithm \ref{batchbald:algo:greedy_batchbald} is $1 - \nicefrac{1}{e}$-approximate.

\section{BALD as an Upper-Bound of BatchBALD}
\label{batchbald:bald_approximates_batchbald}
In the following section, we show that BALD approximates BatchBALD. The BALD score is an upper
bound of the BatchBALD score for any candidate batch.

Using the subadditivity of information entropy and the independence of the
$\yi$ given $\w$, we show that BALD is an approximation of BatchBALD
and is always an upper bound on the respective BatchBALD score:
\begin{align}
	&\BatchBALDAF \left ( \left \{ \xbs \right \}, \pof{\w \given \Dtrain} \right ) \\
	= &\MIof{\Yacqset ; \W \given \xacqset, \Dtrain} \\
	= &\Hof{\Yacqset \given \xacqset, \Dtrain} - \Hof{\Yacqset \given \xacqset, \W, \Dtrain} \\
	\le &\sum_{\ii=1}^{\numB} \Hof{\Yi \given \x_\ii, \Dtrain} - \sum_{\ii=1}^{\numB} \Hof{\Yi \given \x_\ii, \W, \Dtrain} \\
	= &\sum_{\ii=1}^{\numB} \MIof{\Yi; \W \given \x_\ii, \Dtrain} = \BALDAF \left ( \left \{ \xacqset \right \}, \pof{\w \given \Dtrain} \right )
\end{align}

\textbf{Relevance for the active training loop.}
We see that the active training loop as a whole is computing a greedy $1 - \nicefrac{1}{e}$-approximation of
the mutual information of all acquired data points over all acquisitions with the model parameters.

\section{Sampling of Configurations}
\label{batchbald:batchbald_mc_approx}
{
	\renewcommand{\pw}{\pwfuchsia}
	We are using the same notation as in section \ref{batchbald:batchbald_derivation}.
We factor $\pof{\yset \given \w}$ to avoid recomputations and rewrite
$\Hof{\Yset}$ as:
\begin{align}
	\Hof{\Yset} &=
		\simpleE{\pw}{
			\E{\pof{\yset \given \w}}{
				-\log{\pof{\yset}}
			}
		} \\
	&=
		\simpleE{\pw}{
			\E{\pof{\yndv \given \w} \pof{\yn \given \w} }{
				-\log{\pof{\yset}}
			}
		} \\
	&=
		\simpleE{\pw}{
			\simpleE{\pof{\yndv \given \w}}{
				\E{\pof{\yn \given \w}}{
					-\log{\pof{\yset}}
				}
			}
		}
\end{align}
To be flexible in the way we sample $\yndv$, we perform importance sampling of
$\pof{\yndv \given \w}$ using $\pof{\yndv}$, and, assuming we also have $\numM$
samples $\yndvr$ from $\pof{\yndv}$, we can approximate:
\begin{align}
	& \Hof{\Yset} =
		\simpleE{\pw}{
			\E{\pof{\yndv}}{
				\frac{\pof{\yndv \given \w}}{\pof{\yndv}}
				\E{\pof{\yn \given \w}}{
					-\log{\pof{\ynv}}
				}
			}
		} \\
	&=
		\simpleE{\pof{\yndv}}{
			\simpleE{\pw}{
				\E{\pof{\yn \given \w}}{
					-\frac{\pof{\yndv \given \w}}{\pof{\yndv}} \log{
						\E{\pw}{
							\pof{\yndv \given \w} \pof{\ynv \given \w}
						}
					}
				}
			}
		} \\
	& \approx
		-\frac{1}{\numM} \sum_{\yndvr}^{\numM}{
			\sum_{\ynr}{
				\frac{
					\frac{1}{\numK} \sum_{\wj}{
						\pof{\yndvr \given \wj} \pof{\ynr \given \wj}
					}
				}{
						\pof{\yndvr}
				}
				\log \left ( {
					\frac{1}{\numK} \sum_{\wj} \pof{\yndvr \given \wj} \pof{\ynr \given \wj}
				} \right )
			}
		} \\
	& =
		-\frac{1}{\numM} \sum_{\yndvr}^{\numM}{
			\sum_{\ynr}{
				\frac{
					\left ( \hat{P}_{1:{\numN-1}} \hat{P}_{\numN}^T \right )_{\yndvr, \ynr}
				}{
					\left ( \hat{P}_{1:{\numN-1}} \mathbb{1}_{\numK,1} \right )_{\yndvr}
				}
				\log \left ( {
					\frac{1}{\numK} \left ( \hat{P}_{1:{\numN-1}} \hat{P}_{\numN}^T \right )_{\yndvr, \ynr}
				} \right )
			}
		}, \label{batchbald:eq:is_matrix}
\end{align}
where we store $\pof{\yndvr \given \wj}$ in a matrix $\hat{P}_{1:\numN-1}$ of shape
$\numM \times \numK$ and $\pof{\ynr \given \wj}$ in a matrix $\hat{P}_{\numN}$ of
shape $\numC \times \numK$ and $\mathbb{1}_{\numK,1}$ is a $\numK \times 1 $
matrix of $1$s. Equation \eqref{batchbald:eq:is_matrix} allows us to cache $\hat{P}_{1:\numN-1}$
inside the inner loop of algorithm \ref{batchbald:algo:greedy_batchbald} and use batch
matrix multiplication for efficient computation.
}

\section{Ablation Study on Repeated-MNIST}
\label{batchbald:ablation_rmnist}
To better understand the effect of redundant data points on BALD and BatchBALD,
we run the RMNIST experiment with an increasing number of repetitions. The
results can be seen in figure \ref{batchbald:ablation_rmnist_figure}. We use the same setup as in
section \ref{batchbald:repeated_mnist}. BatchBALD performs the same on all repetition
numbers (100 data points till 90\%). BALD achieves 90\% accuracy at 120 data
points (0 repetitions), 160 data points (1 repetition), 280 data points (2
repetitions), 300 data points (4 repetitions). This shows that BALD and
BatchBALD behave as expected.

\begin{figure}[h]
	\centering
	\includegraphics[height=5cm]{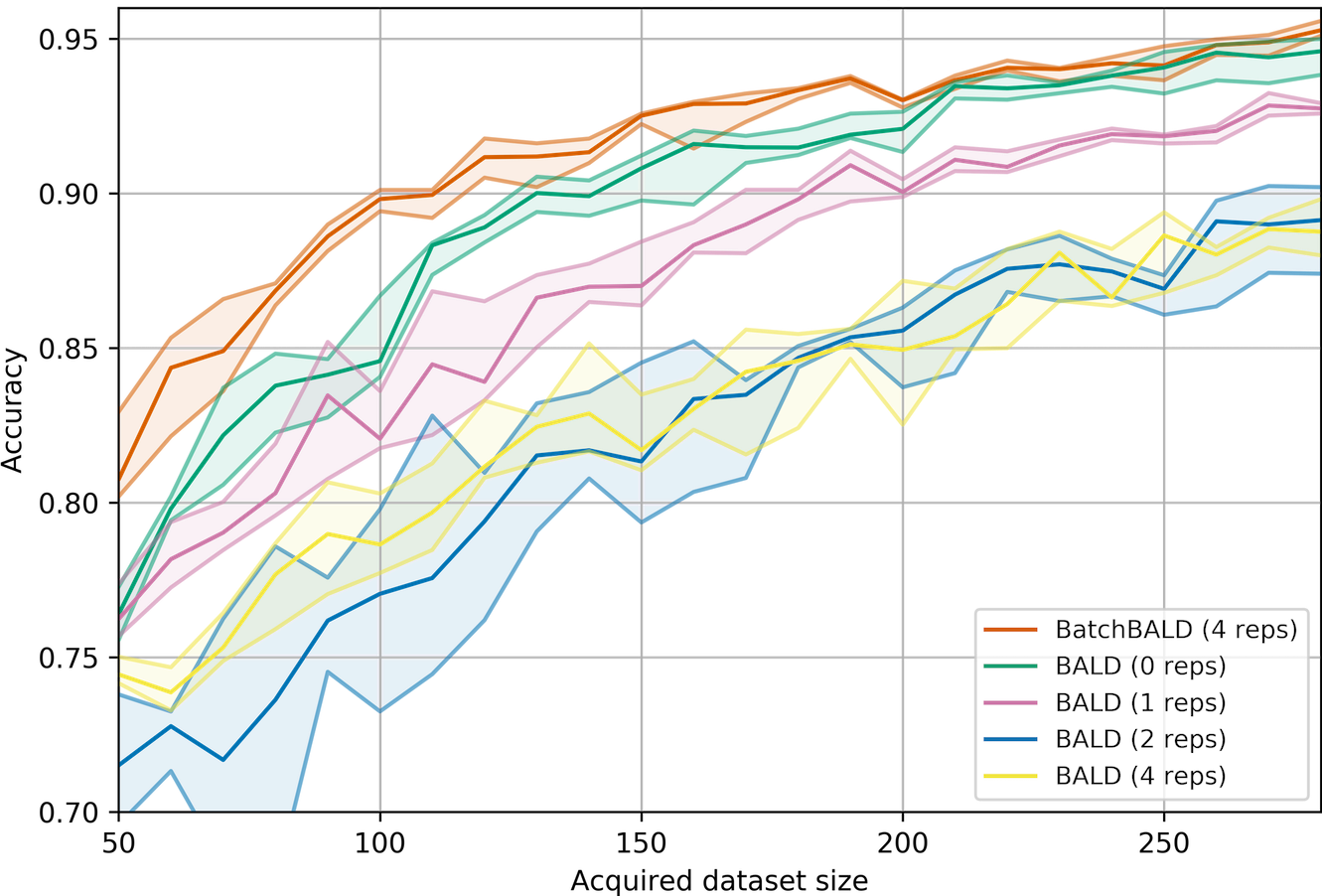}
	\caption{\emph{Performance of \emph{BALD} on \emph{Repeated MNIST} for increasing amount of repetitions.}
	We see that BALD performs worse as the number of repetitions is increased,
	while BatchBALD outperforms BALD with zero repetitions.}
	\label{batchbald:ablation_rmnist_figure}
\end{figure}

\section{Additional Results for Repeated-MNIST}
\label{batchbald:rmnist_comparison_graph}

We show that BatchBALD also outperforms Var Ratios \citep{freeman1965elementary}
and Mean STD \citep{kendall2015bayesian}.

\begin{figure}[h]
	\centering
	\includegraphics[height=5cm]{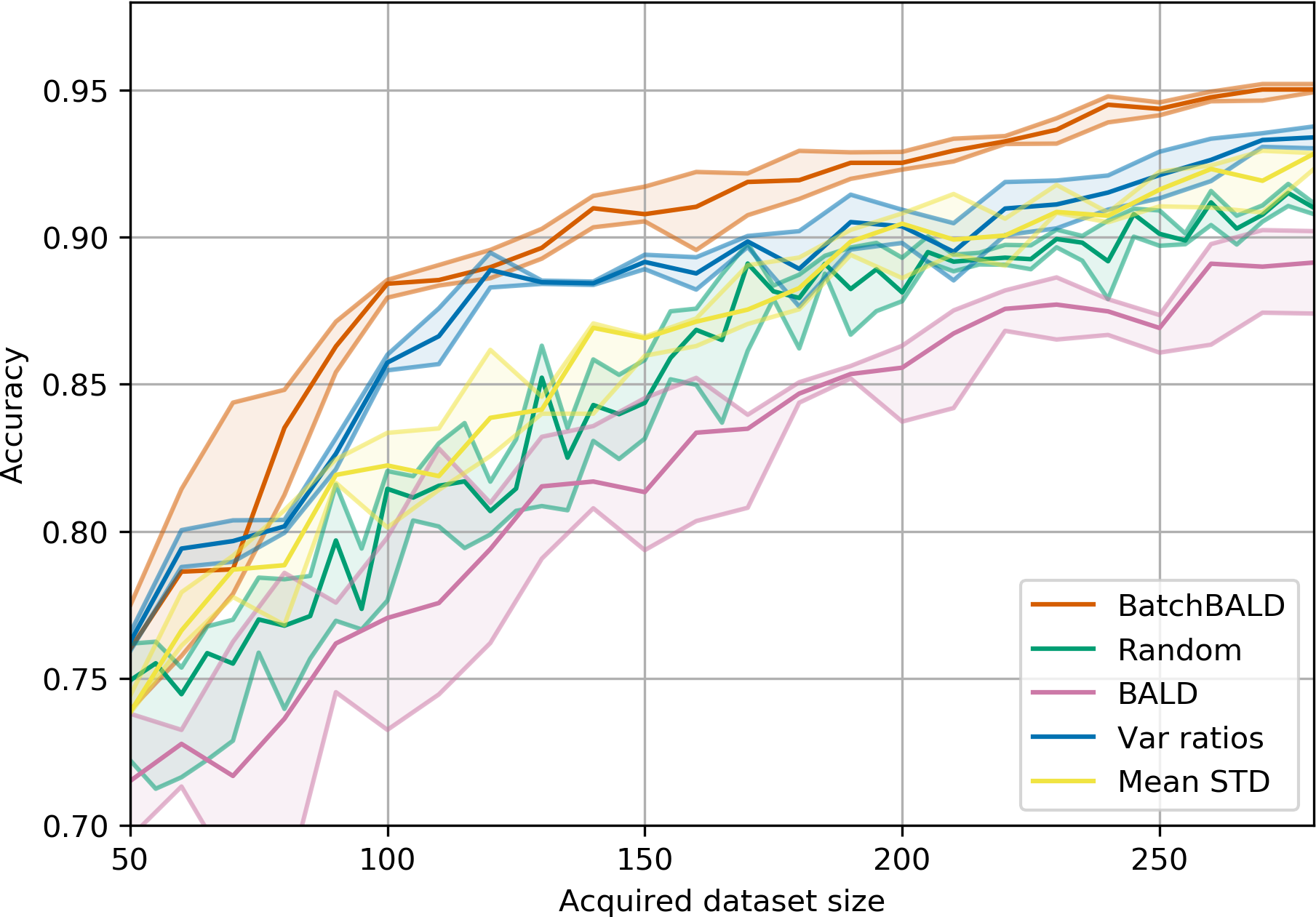}
	\caption{\emph{Performance on \emph{Repeated MNIST}.}
	BALD, BatchBALD, Var Ratios, Mean STD and random acquisition with acquisition size 10 and 10 MC dropout samples.}
\end{figure}

\newpage
\section{Example Visualisation of EMNIST}
\label{batchbald:emnist_visualisation}

\begin{figure}[h]
	\centering
	\includegraphics[height=5cm]{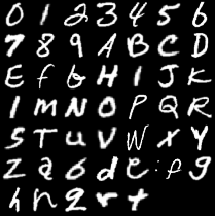}
	\caption{\emph{Examples of all 47 classes of EMNIST}}
\end{figure}

\section{Entropy and Per-Class Acquisitions (including Random Acquisition)}
\label{batchbald:emnist_random_diversity}
\begin{figure}[h]
	\begin{minipage}[t]{0.49\textwidth}
		\centering
		\includegraphics[width=0.95\linewidth]{images/EMNIST_zero_initial_data.png}
		\caption{\emph{Performance on \emph{EMNIST}.}
		BatchBALD consistently outperforms both random acquisition and BALD
		while BALD is unable to beat random acquisition.
		}
	\end{minipage}
	\hfill
	\begin{minipage}[t]{0.49\textwidth}
		\centering
        \includegraphics[width=0.99\linewidth]{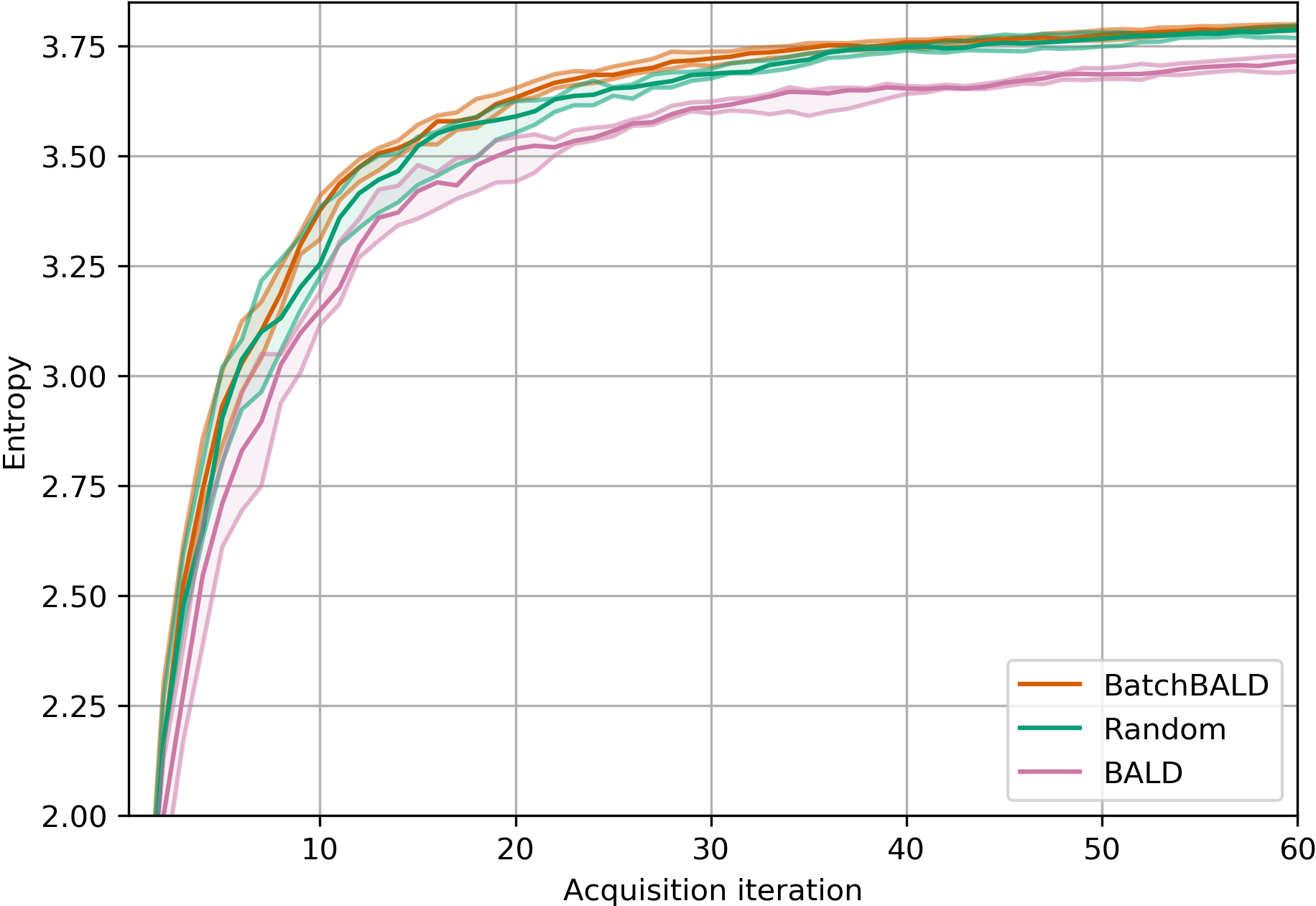}
		\caption{\emph{Entropy of acquired class labels over acquisition steps on \emph{EMNIST}.}
		BatchBALD steadily acquires a more diverse set of data points than BALD.
		}
	\end{minipage}
\end{figure}

}

\chapter{\stochacqtitle}
{
	
	\newcommand{\miotcd}{MIO-TCD\xspace}
\newcommand{\synminority}{Synbols Minority Groups\xspace}
\newcommand{\synmissing}{Synbols Missing Characters\xspace}
\newcommand{\synspurious}{Synbols Spurious Correlations\xspace}

\section{Proof of \Cref{stochacq:prop:gumbel_top_k}}
\label{stochacq:app:proof_gumbel_top_k}

First, we remind the reader that a random variable $G$ is Gumble distributed $G \sim \gumbel{\mu}{\beta}$ when its cumulative distribution function follows $\pof{G \le g} = \exp(-\exp(-\frac{g - \mu}{\beta}))$.

Furthermore, the Gumbel distribution is closed under translation and positive scaling:
\begin{lemma}
Let $G \sim \gumbel{\mu}{\beta}$ be a Gumbel distributed random variable, then:
\begin{align}
    \alpha G + d \sim \gumbel{d + \alpha \mu}{\alpha \beta}.
\end{align}
\end{lemma}
\begin{proof}
We have $\pof{\alpha G + d \le x} = \pof{G \le \frac{x - d}{\alpha}}$. Thus, we have:
\begin{align}
    \pof{\alpha G + d \le x} &= \exp(-\exp(-\frac{\frac{x - d}{\alpha} - \mu}{\beta})) \\
    &= \exp(-\exp(-\frac{x - (d + \alpha \mu)}{\alpha \beta})) \\
    \Leftrightarrow & \alpha G + d \sim \gumbel{d + \alpha \mu}{\alpha \beta}.
\end{align}
\end{proof}

We can then easily prove \Cref{stochacq:prop:gumbel_top_k} using Theorem 1 from \citet{kool2019stochastic}, which we present it here slightly reformulated to fit our notation:
\begin{lemma}
\label{stochacq:lem:koolthm}
For $k \leq n$, let $I_{1}^{*}, \ldots, I_{k}^{*}=\arg \operatorname{top}_k \{s_i + \epsilon_i\}_i$ with $\epsilon_i \sim \gumbel{0}{1}$, i.i.d.. Then $I_{1}^{*}, \ldots, I_{k}^{*}$ is an (ordered) sample without replacement from the categorical distribution
\begin{align}
    \mathrm{Categorical} \left(\frac{\exp s_{i}}{\sum_{j \in n} \exp s_{j}}, i \in \{1, \ldots, n\}\right),
\end{align} 
e.g. for a realization $i_{1}^{*}, \ldots, i_{k}^{*}$ it holds that
\begin{align}
P\left(I_{1}^{*}=i_{1}^{*}, \ldots, I_{k}^{*}=i_{k}^{*}\right)=\prod_{j=1}^{k} \frac{\exp s_{i_{j}^{*}}}{\sum_{\ell \in N_{j}^{*}} \exp s_{\ell}}
\end{align} 
where $N_{j}^{*}=N \backslash\left\{i_{1}^{*}, \ldots, i_{j-1}^{*}\right\}$ is the domain (without replacement) for the $j$-th sampled element.
\end{lemma}
Now, it is easy to prove the proposition:
\gumbeltopk*
\begin{proof}
As $\epsilon_i \sim \gumbel{0}{\beta^{-1}}$, define $\epsilon'_i \defeq \beta \epsilon_i \sim \gumbel{0}{1}$. Further, let $s'_i \defeq \beta s_i$. Applying \Cref{stochacq:lem:koolthm} on $s'_i$ and $\epsilon'_i$, $\arg \operatorname{top}_k \{s'_i + \epsilon'_i\}_i$ yields (ordered) samples without replacement from the categorical distribution $\text{Categorical}(\frac{\exp(\beta \, s_i)}{\sum_j \exp(\beta \, s_j)}, i \in \{1, \ldots, n\})$.
However, multiplication by $\beta$ does not change the resulting indices of $\arg \operatorname{top}_k$:
\begin{align}
    \arg \operatorname{top}_k \{s'_i + \epsilon'_i\}_i = \arg \operatorname{top}_k \{s_i + \epsilon_i\}_i,
\end{align}
concluding the proof.
\end{proof}
\begin{figure}
    \centering
    \begin{subfigure}[b]{0.49\linewidth}
        \centering
        \includegraphics[width=\linewidth]{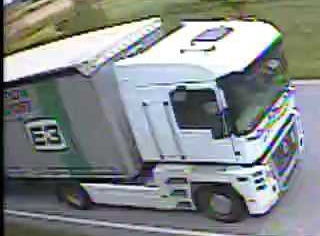}
        \hspace*{\fill}\caption{A good example in MIOTCD dataset.}\hspace*{\fill}
    \label{stochacq:fig:good}
    \end{subfigure}
    \begin{minipage}[b]{0.49\linewidth}
            \begin{subfigure}[b]{0.59\linewidth}
            \centering
            \hspace*{\fill}\includegraphics[width=\linewidth,height=2cm]{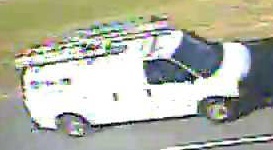}\hspace*{\fill}\par
            \hspace*{\fill}\caption{An example of duplicated samples in the dataset.}\hspace*{\fill}
        \label{stochacq:fig:duplicat}
        \end{subfigure}
        \begin{subfigure}[b]{0.39\linewidth}
            \centering
            \hspace*{\fill}\includegraphics[width=\linewidth,height=3cm]{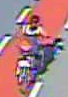}\hspace*{\fill}\par
            \caption{An example of class confusion between motorcycle and bicycle.}
        \label{stochacq:fig:confusion}
        \end{subfigure}
        \begin{subfigure}[b]{0.49\linewidth}
            \centering
            \hspace*{\fill}\includegraphics[width=\linewidth,height=2cm]{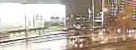}\hspace*{\fill}\par
            \hspace*{\fill}\caption{An example of heavy compression artifact.}\hspace*{\fill}
        \label{stochacq:fig:compression}
        \end{subfigure}
        \begin{subfigure}[b]{0.49\linewidth}
            \centering
            \hspace*{\fill}\includegraphics[width=\linewidth,height=2cm]{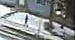}\hspace*{\fill}\par
            \hspace*{\fill}\caption{An example of low resolution samples.}\hspace*{\fill}
        \label{stochacq:fig:lowres}
        \end{subfigure}
    \end{minipage}
    \caption{\emph{MIO-TCD Dataset} is designed to include common artifacts from production data. The size and quality of the images vary greatly between crops; from high-quality cameras on sunny days to low-quality cameras at night. (a) shows an example of clean samples that can be clearly assigned to a class. (b)(c)(d) and (e) show the different categories of noise. (b) shows an example of many near-duplicates that exist in the dataset. (c) is a good example where the assigned class is subject to interpretation (d) is a sample with heavy compression artifacts and (e) is an example of samples with low resolution which again is considered a hard example to learn for the model.}
    \label{stochacq:fig:mio_tcd_examples}
\end{figure}

\section{Empirical Validation}
\label{stochacq:appsec:experiments}
\subsection{Experimental Setup \&\ Compute}
\label{stochacq:sec:exp_setup}

Full code for all experiments will be available at \shorturl{anonymized_github_repo}. 

\textbf{Frameworks.} %
We use PyTorch. Repeated-MNIST and EMNIST experiments use PyTorch Ignite. Synbols and MIO-TCD experiments use the BaaL library \shorturl{https://github.com/baal-org/baal} \citep{atighehchian2020bayesian}. Predictive parity is calculated using FairLearn \citep{bird2020fairlearn}. The CausalBALD experiments use \shorturl{https://github.com/anndvision/causal-bald} \citep{jesson2021causal}. 

\textbf{Compute.} %
Results shown in Table \ref{stochacq:tab:runtime} were run inside Docker containers with 8 CPUs (2.2Ghz) and 32 Gb of RAM. Other experiments were run on similar machines with Titan RTX GPUs. The Repeated-MNIST and EMNIST experiments take about 5000 GPU hours. The MIO, Synbols and CLINC-150 experiments take about 19000 GPU hours. The CausalBALD experiments take about 1000 GPU hours.

\textbf{Dataset Licenses.} Repeated-MNIST is based on MNIST which is made available under the terms of the Creative Commons Attribution-Share Alike 3.0 license. The EMNIST dataset is made available as CC0 1.0 Universal Public Domain Dedication. Synbols is a dataset generator. MIO-TCD is made available under the terms of the Creative Commons Attribution-NonCommercial-ShareAlike 4.0 International License. CLINC-150 is made available under the terms of Creative Commons Attribution 3.0 Unported License.

\subsubsection{Runtime Measurements}

The synthetic dataset used for benchmarking has 4,096 features, 10 classes, and 10,000 pool points.
VGG-16 models \citep{simonyan2014very} were used to sample predictions and latent embeddings.

\subsubsection{Repeated-MNIST}

The Repeated-MNIST dataset (introduced in \S\ref{chapter:batchbald}) with duplicated examples from MNIST with isotropic Gaussian noise added to the input images (standard deviation 0.1).

We use the same setup as in \S\ref{chapter:batchbald}: a LeNet-5-like architecture with ReLU activations instead of tanh and added dropout. The model obtains 99\% test accuracy when trained on the full MNIST dataset. 
Specifically, 
the model is made up of two blocks of a convolution, dropout, max-pooling, ReLU with 32 and 64 channels and 5x5 kernel size, respectively. As classifier head, a two-layer MLP with 128 hidden units (and 10 output units) is used that includes dropout between the layers. We use a dropout probability of 0.5 everywhere. 
The model is trained with early stopping using the Adam optimizer and a learning rate of 0.001. We sample predictions using 100 MC-Dropout samples for BALD. Weights are reinitialized after each acquisition step.

\subsubsection{EMNIST}

We follow the setup from \S\ref{chapter:batchbald} with 20 MC dropout samples. We use a similar model as for Repeated-MNIST but with three blocks instead of two.
Specifically, we use 32, 64, and 128 channels and 3x3 kernel size. This is followed by a 2x2 max pooling layer before the classifier head. The classifier head is a two-layer MLP but with 512 hidden units instead of 128. Again, we use dropout probability 0.5 everywhere.

\subsubsection{Synbols \&\ MIO-TCD} The full list of hyperparameters for the Synbols and MIO-TCD experiments is presented in Table \ref{stochacq:tab:hp_al}.
Our experiments are built using the BaaL library \citep{atighehchian2020bayesian}.
We compute the predictive parity using FairLearn \citep{bird2020fairlearn}.
We use VGG-16 model \citep{simonyan2014very} trained for 10 epochs using Monte Carlo dropout for acquisition \citep{gal2017deep} with 20 dropout samples.

\begin{table}[t]
    \centering
        \caption{Hyperparameters used in Section \ref{stochacq:sec:miotcd} and \ref{stochacq:sec:synbols}}
    \label{stochacq:tab:hp_al}
    \scriptsize
    \begin{tabular}{cc}
    \toprule
    Hyperparameter & Value \\
    \midrule
        Learning Rate & 0.001 \\
        Optimizer & SGD \\
        Weight Decay & 0 \\
        Momentum & 0.9 \\
        Loss Function & Cross-Entropy \\
        Training Duration & 10 \\
        Batch Size & 32 \\
        Dropout $p$ & 0.5 \\
        MC Iterations & 20 \\
        Query Size & 100 \\
        Initial Set & 500 \\
        \bottomrule
    \end{tabular}
\end{table}

In \Cref{stochacq:fig:mio_tcd_examples}, we show a set of images with common problems that can be found in MIO-TCD.

\subsubsection{CLINC-150}

We fine-tune a pretrained DistilBERT model from HuggingFace~\citep{dosovitskiy2020image} on CLINC-150 for 5 epochs with Adam as optimizer. Estimating epistemic uncertainty in transformer models is an open research question, and hence, we do not report results using BALD and focus on entropy instead.

\subsubsection{CausalBALD}
Using the Neyman-Rubin framework \citep{neyman1923applications, rubin1974estimating, sekhon2008neyman}, the CATE is formulated in terms of the potential outcomes, $\mathrm{Y}_\mathrm{t}$, of treatment levels $\mathrm{t} \in \{0, 1\}$.
Given observable covariates, $\mathbf{X}$, the CATE is defined as the expected difference between the potential outcomes at the measured value $\mathbf{X} = \mathbf{x}$: $\tau(\mathbf{x}) = \mathbb{E}[\mathrm{Y}_1 - \mathrm{Y}_0 \mid \mathbf{X}=\mathbf{x}]$.
This causal quantity is fundamentally unidentifiable from observational data without further assumptions because it is not possible to observe both $\mathrm{Y}_1$ and $\mathrm{Y}_0$ for a given unit. 
However, under the assumptions of consistency, non-interference, ignoreability, and positivity, the CATE is identifiable as the statistical quantity $\widetilde{\tau}(\mathbf{x}) = \mathbb{E}[\mathrm{Y} \mid \mathrm{T}=1, \mathbf{X}=\mathbf{x}] - \mathbb{E}[\mathrm{Y} \mid \mathrm{T}=0, \mathbf{X}=\mathbf{x}]$ \citep{rubin1980randomization}.

\citet{jesson2021causal} define BALD acquisition functions for active learning CATE functions from observational data when the cost of acquiring an outcome, $\mathrm{y}$, for a given covariate and treatment pair, $(\mathbf{x}, \mathrm{t})$, is high.
Because we do not have labels for $\mathrm{Y}_1$ and $\mathrm{Y}_0$ for each $(\mathbf{x}, \mathrm{t})$ pair in the dataset, their acquisition function focuses on acquiring data points $(\mathbf{x}, \mathrm{t})$ for which it is likely that a matched pair $(\mathbf{x}, 1 - \mathrm{t})$ exists in the pool data or has already been acquired at a previous step.
We follow their experiments on their synthetic dataset with limited positivity and the semisynthetic IHDP dataset \citep{hill2011bayesian}.
Details of the experimental setup are given in \citep{jesson2021causal}, we use their provided code, and implement the power acquisition function. 

The settings for causal inference experiments are identical to those used in \citet{jesson2021causal}, using the IHDP dataset \citep{hill2011bayesian}.
Like them, we use a Deterministic Uncertainty Estimation Model \citep{van2021improving}, which is initialized with 100 data points and acquire 10 data points per acquisition batch for 38 steps.
The dataset has 471 pool points and a 201 point validation set.

\subsection{Repeated-MNIST}
\label{stochacq:appsec:exp_repeated_mnist}

\begin{figure}[H]
    \centering
    \includegraphics[width=\linewidth]{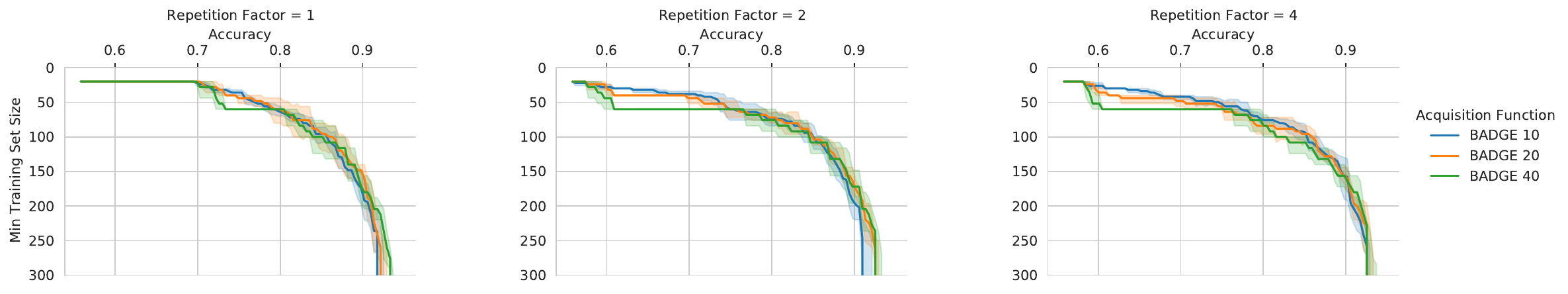}
    \caption{
    \emph{Repeated-MNIST x4 (5 trials): acquisition size ablation for BADGE.}
    Acquisition size 20 performs best out of $\{10,20,40\}$. Hence, we use that for \Cref{stochacq:fig:rmnist}.
    }
    \label{stochacq:appfig:rmnist_badge_ablation}
\end{figure}

\textbf{BADGE Ablation.} In \Cref{stochacq:appfig:rmnist_badge_ablation}, we see that BADGE performs best with acquisition size 20 on Repeated-MNISTx4 overall. BADGE 40 and BADGE 20 have the highest final accuracy, cf.\ BADGE 10 while BADGE 20 performs better than BADGE 40 for small training set sizes.

\subsubsection{Other Scoring Functions}
\label{stochacq:appsec:exp_repeated_mnist_other_acq_functions}

\begin{figure}[H]
    \centering
    \includegraphics[width=0.75\linewidth]{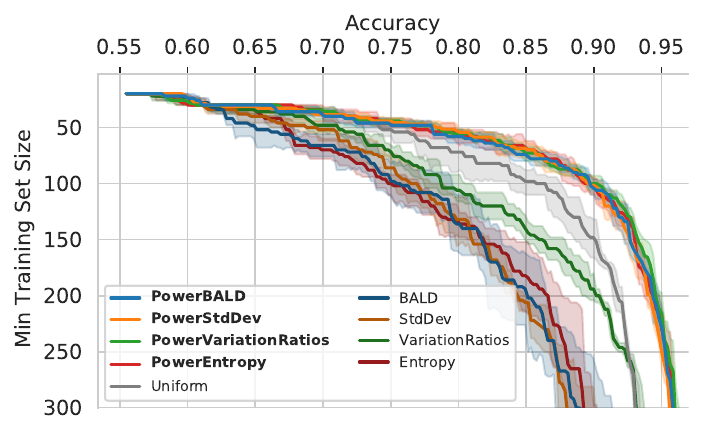}
    \caption{\emph{Repeated-MNIST x4 (5 trials): Performance for other scoring functions.} Entropy, std dev, variation ratios behave like BALD when applying our stochastic sampling scheme.}
    \label{stochacq:fig:rmnist_other_acquisition_functions}
\end{figure}

In \Cref{stochacq:fig:rmnist_other_acquisition_functions} shows the performance of other scoring functions than BALD on Repeated-MNIST x4.

\subsubsection{Redundancy Ablation}

\begin{figure}[H]
    \centering
    \includegraphics[width=\linewidth]{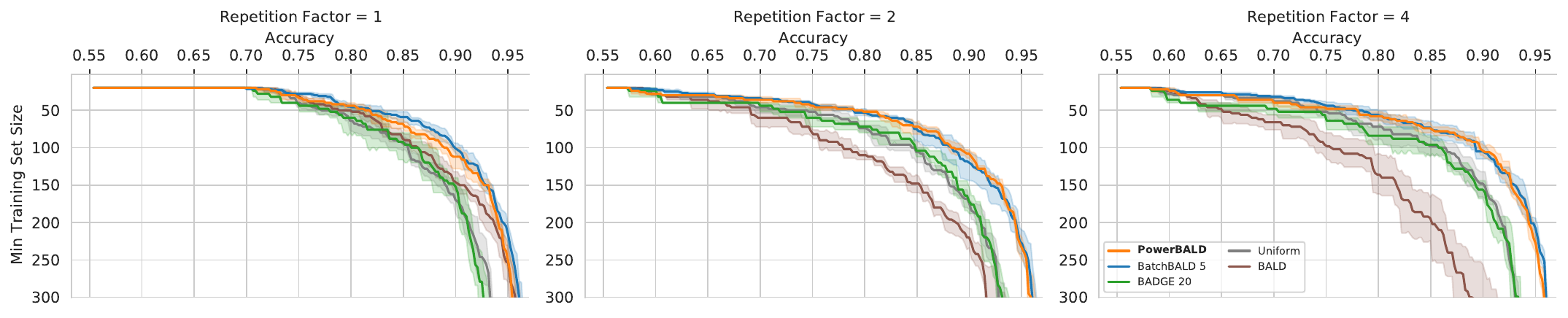}
    \caption{\emph{Repeated-MNIST (5 trials): Performance ablation for different repetition counts.}}
    \label{stochacq:appfig:rmnist_repetition_ablation}
\end{figure}

In \Cref{stochacq:appfig:rmnist_repetition_ablation}, we see the same behavior in an ablation for different repetition sizes of Repeated-MNIST.

\subsection{MIO-TCD}
\label{stochacq:appsec:miotcd}

\begin{figure}[H]
    \begin{subfigure}[t]{0.45\linewidth}
        \centering
        \includegraphics[width=\linewidth]{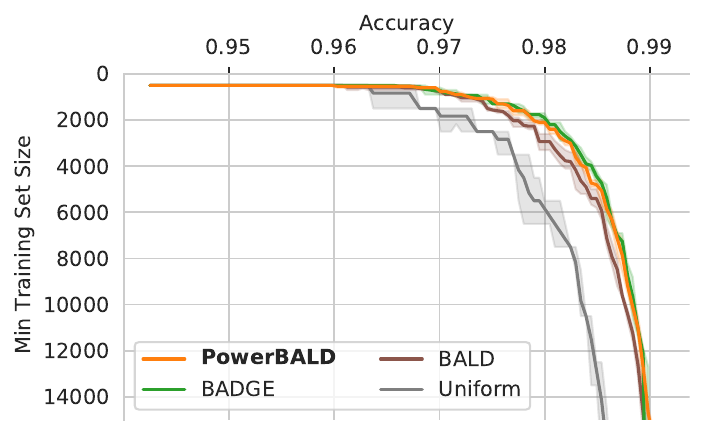}
        \caption{ %
        BALD
        }
        \label{stochacq:appfig:miotcd_bald}
    \end{subfigure}
    \hfill
    \begin{subfigure}[t]{0.45\linewidth}
        \centering
        \includegraphics[width=\linewidth]{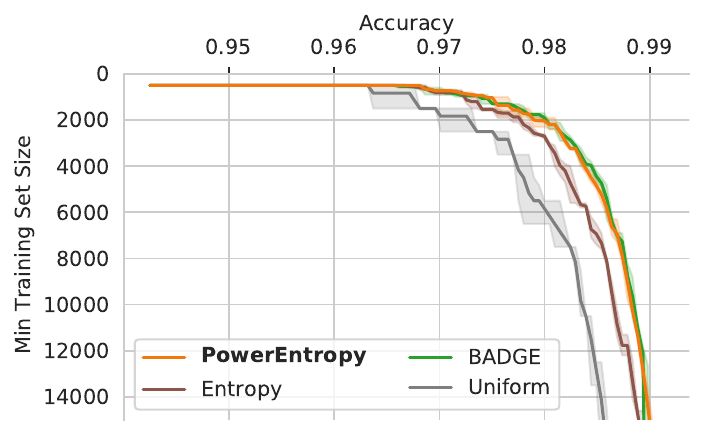}
        \caption{ %
        Entropy
        }
        \label{stochacq:appfig:miotcd_entropy}
    \end{subfigure}
    \caption{\emph{\miotcd (5 trials).}}
    \label{stochacq:appfig:miotcd}
\end{figure}

In \Cref{stochacq:appfig:miotcd}, we see that power acquisition performs on par with BADGE with both BALD and entropy as underlying score functions.

\subsection{EMNIST}
\label{stochacq:appsec:emnist}

\begin{figure}[H]
    \begin{minipage}[t]{0.45\linewidth}
        \centering
        \includegraphics[width=\linewidth]{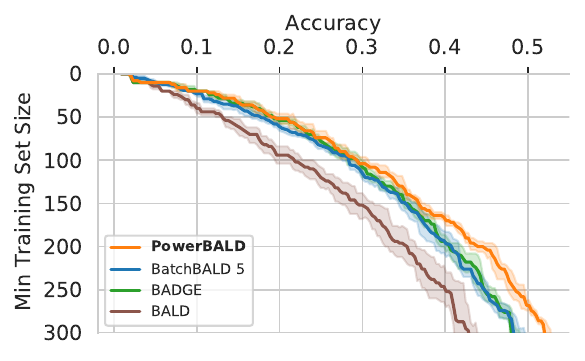}
        \caption{ %
        \emph{EMNIST (Balanced) (5 trials): Performance with BALD.}
        }
        \label{stochacq:appfig:emnist_balanced_main}
    \end{minipage}
    \hfill
    \begin{minipage}[t]{0.45\linewidth}
        \centering
        \includegraphics[width=\linewidth]{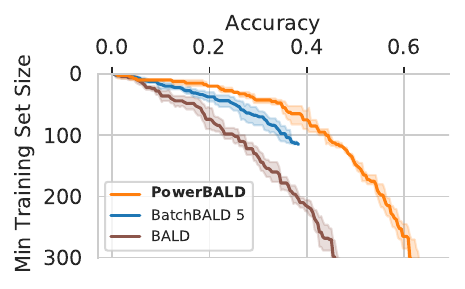}
        \caption{ %
        \emph{EMNIST (ByMerge) (5 trials): Performance with BALD.} 
        }
        \label{stochacq:appfig:emnist_main}
    \end{minipage}
\end{figure}

In \Cref{stochacq:appfig:emnist_balanced_main} and \ref{stochacq:appfig:emnist_main}, we see that PowerBALD outperforms BALD, BatchBALD, and BADGE.

\begin{figure}[H]
    \centering
    \begin{minipage}[t]{0.75\linewidth}
        \centering
        \includegraphics[width=\linewidth]{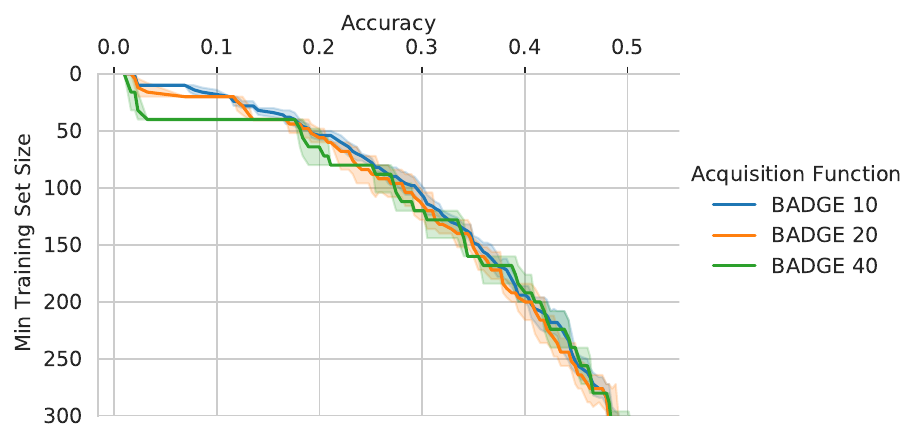}
        \caption{ %
        \emph{EMNIST (Balanced) (5 trials): acquisition size ablation for BADGE.}
        }
        \label{stochacq:appfig:emnist_balanced_badge_ablation}
    \end{minipage}
\end{figure}

\textbf{BADGE Ablation.} In \Cref{stochacq:appfig:emnist_balanced_badge_ablation}, we see that BADGE performs similarly with all three acquisition sizes. Acquisition size 10 is the smoothest.

\subsection{Edge Cases in Synbols} \label{stochacq:sec:synbols}
We use Synbols~\citep{lacoste2020synbols} to demonstrate the behavior of batch active learning in artificially constructed edge cases.
Synbols is a character dataset generator for classification where a user can specify the type and proportion of bias and insert artifacts, backgrounds, masking shapes, and so on.
We selected three datasets with strong biases supplied by \citet{lacoste2020synbols, branchaud2021can} to evaluate our method.
The experimental settings are described in appendix \ref{stochacq:sec:exp_setup}.

For these tasks, performance evaluation includes `predictive parity', also known as `accuracy difference', which is the maximum difference in accuracy between subgroups---which are, in this case, different colored characters.
This measure is used most widely in domain adaptation and ethics~\citep{verma2018fairness}.
We want to maximize the accuracy while minimizing the predictive parity.

\begin{figure}[H]
    \centering
    \begin{subfigure}[t]{0.49\linewidth}
        \centering
        \includegraphics[width=\linewidth]{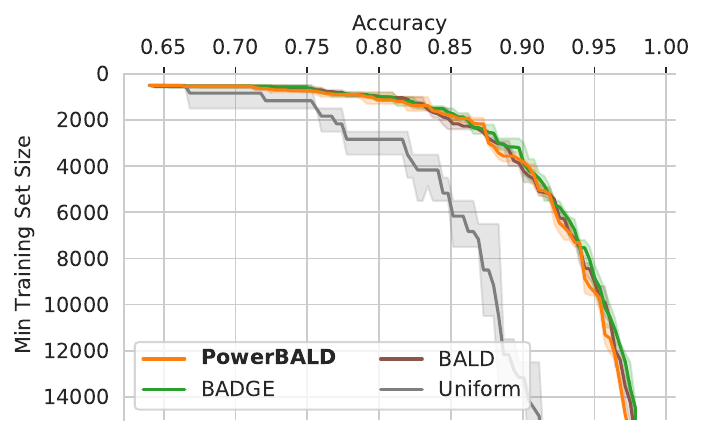}
        \caption{Accuracy}
        \label{stochacq:fig:spurious_acc}
    \end{subfigure}
    \hfill
    \begin{subfigure}[t]{0.49\linewidth}
        \centering
        \includegraphics[width=\linewidth]{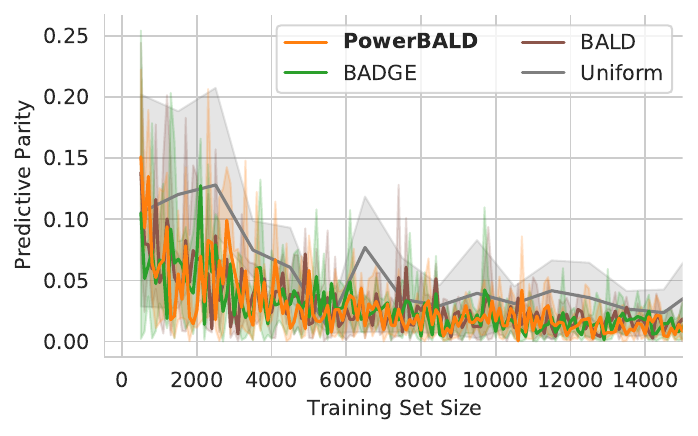}
        \caption{Predictive parity (\textbf{Down and left is better.})}
        \label{stochacq:fig:synbols_spurious}
    \end{subfigure}
    \caption{ %
        \emph{Performance on \synspurious (3 trials) with BALD.}
        Stochastic acquisition matches BADGE and BALD's predictive parity and performance, which is reassuring as stochastic acquisition functions might be affected by spurious correlations.%
        }
\end{figure}

\textbf{Spurious Correlations.} This dataset includes spurious correlations between character color and class. As shown in \citet{branchaud2021can}, active learning is especially strong here as characters that do not follow the correlation will be informative and thus selected.  %

We compare the predictive parity between methods in \figref{stochacq:fig:synbols_spurious}. We do not see any significant difference between our method and BADGE or BALD. This is encouraging, as stochastic approaches might select more examples following the spurious correlation and thus have higher predictive parity, but this is not the case.

\begin{figure}[H]
    \centering
    \begin{subfigure}[t]{0.49\linewidth}
        \centering %
        \includegraphics[width=\linewidth]{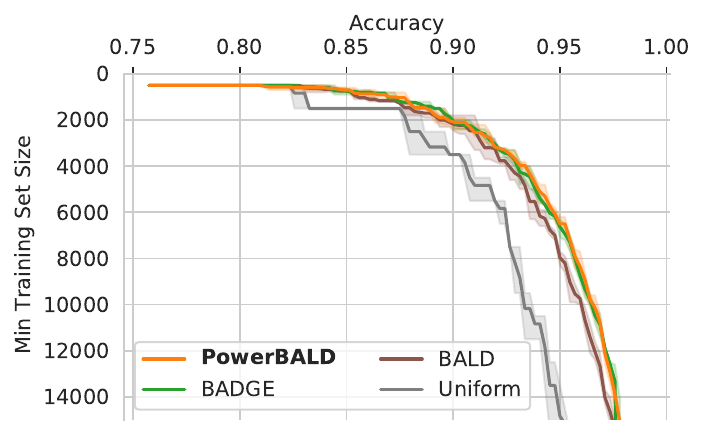} %
        \caption{
            Accuracy
        } %
        \label{stochacq:appsubfig:synbols_minority_accuracy}
    \end{subfigure} %
    \hfill %
    \begin{subfigure}[t]{0.49\linewidth} %
        \centering %
        \includegraphics[width=\linewidth]{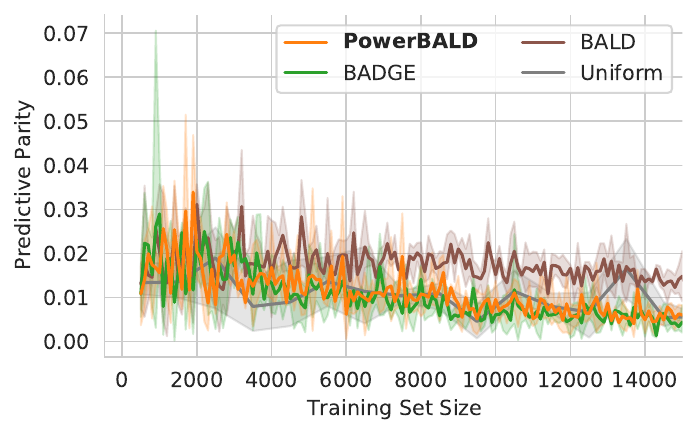}
        \caption{
            Predictive parity
        } %
        \label{stochacq:appsubfig:synbols_minority_diff_accuracy}
    \end{subfigure}
    \caption{
        \emph{\synminority (3 trials): Performance on BALD}.
        PowerBALD outperforms BALD and matches BADGE for both accuracy and predictive parity.
    }
    \label{stochacq:appfig:minority_bald} %
\end{figure}

\textbf{Minority Groups.} This dataset includes a subgroup of the data that is under-represented; specifically, most characters are red while few are blue.
As \citet{branchaud2021can} shows, active learning can improve the accuracy for these groups.

Our stochastic approach lets batch acquisition better capture under-represented subgroups.
In \Cref{stochacq:appsubfig:synbols_minority_accuracy}, PowerBALD has an accuracy almost identical to that of BADGE, despite being much cheaper, and outperforms BALD.
At the same time, we see in \Cref{stochacq:appsubfig:synbols_minority_diff_accuracy} that PowerBALD has a lower predictive parity than BALD, demonstrating a fairer predictive distribution given the unbalanced dataset.
\seb{This actually doesn't quite make sense - it should only be the case if for some reason the BALD score is not succeeding in finding novel situations}

\begin{figure}[H]
    \centering
    \begin{minipage}[t]{0.49\linewidth}
        \centering
        \includegraphics[width=\textwidth]{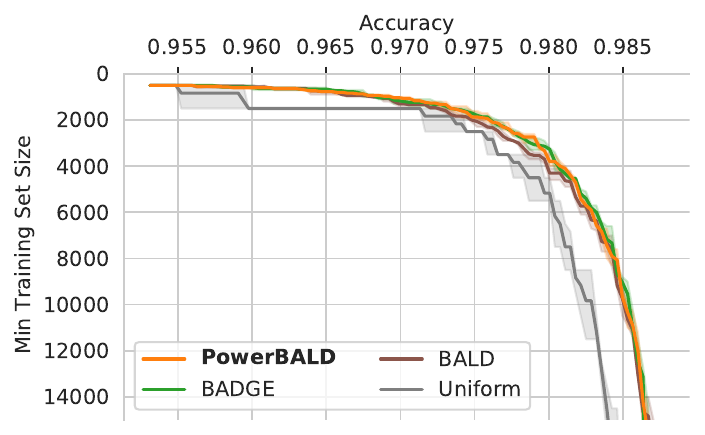}
        \caption{BALD}
        \label{stochacq:appfig:missing_bald}
    \end{minipage}
    \hfill
    \begin{minipage}[t]{0.49\linewidth}
        \centering
        \includegraphics[width=\textwidth]{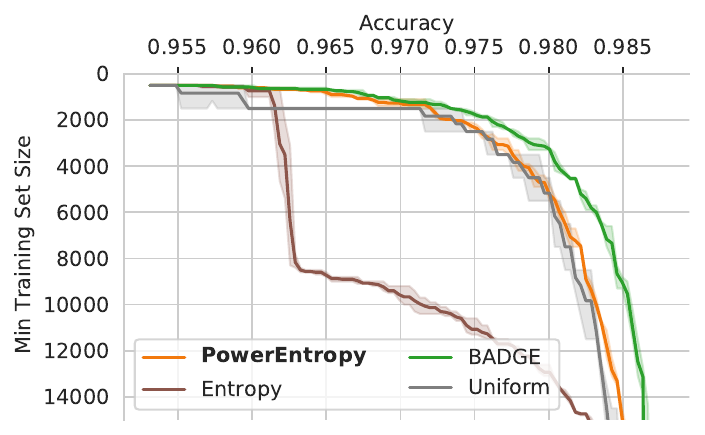}
        \caption{Entropy}
        \label{stochacq:fig:missing_entropy}
    \end{minipage}
    \caption{
        \emph{Performance on \synmissing (3 trials).}
        In this dataset with high aleatoric uncertainty, PowerBALD matches BADGE and BALD performance. PowerEntropy significantly outperforms Entropy which confounds aleatoric and epistemic uncertainty.
    }
    \label{stochacq:appfig:synbols_missing}
\end{figure}

\textbf{Missing Synbols.} This dataset has high aleatoric uncertainty.
Some images are missing information required to make high-probability predictions---these images have shapes randomly occluding the character---so even a perfect model would remain uncertain.
\citet{lacoste2020synbols} demonstrated that entropy is ineffective on this data as it cannot distinguish between aleatoric and epistemic uncertainty, while BALD can do so. As a consequence, entropy will unfortunately prefer samples with occluded characters, resulting in degraded active learning performance.
For predictive entropy, stochastic acquisition largely corrects the failure of entropy acquisition to account for missing data (\Cref{stochacq:appfig:synbols_missing}) although PowerEntropy still underperforms BADGE here. For BALD, we show in \Cref{stochacq:appfig:missing_bald} in the appendix that, as before, our stochastic method performs on par with BADGE and marginally better than BALD.

\subsection{CLINC-150}
\label{stochacq:appsec:clinc}

\begin{figure}[H]
    \centering %
    \includegraphics[width=0.5\linewidth]{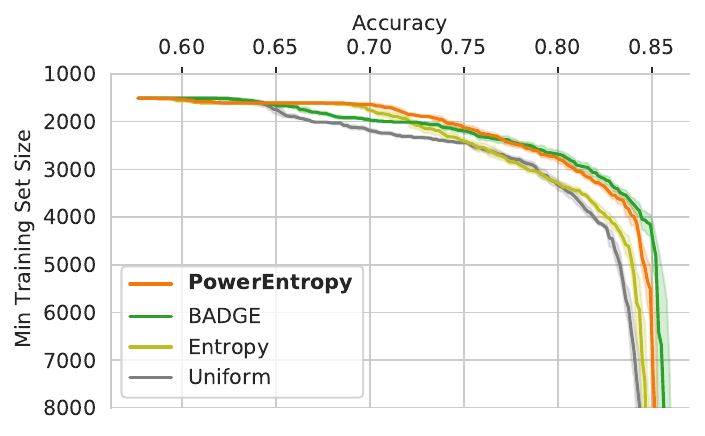} %
    \caption{\emph{Performance on CLINC-150 (10 trials).} PowerEntropy performs much better than entropy, which only performs marginally better than uniform, and almost on par with BADGE.}%
    \label{stochacq:appfig:clinc150} %
\end{figure}
    
In \Cref{stochacq:appfig:clinc150}, we see that PowerEntropy performs much better than entropy which only performs marginally better than the uniform baseline. PowerEntropy also performs better than BADGE at low training set sizes, but BADGE performs better in the second half. Between $\approx2300$ and $4000$ samples, BADGE and PowerEntropy perform the same. 

\section{Comparing Power, Softmax and Soft-Rank}
\label{stochacq:sec:power_softmax_softrank_comparison}

\subsection{Empirical Evidence}

\begin{figure}[H]
    \begin{subfigure}[t]{\linewidth}
        \centering
        \includegraphics[width=\linewidth]{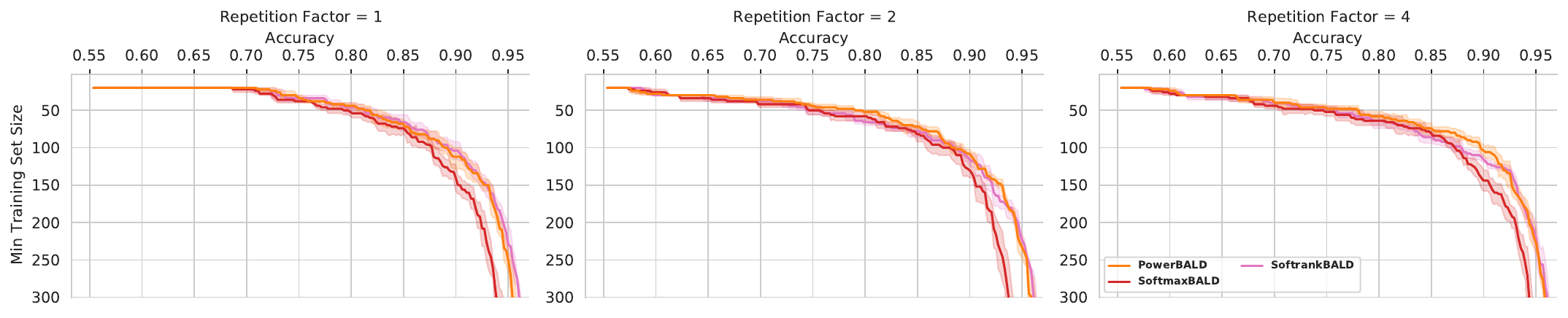}
        \caption{ %
            BALD
        }
    \end{subfigure}
    \begin{subfigure}[t]{\linewidth}
        \centering
        \includegraphics[width=\linewidth]{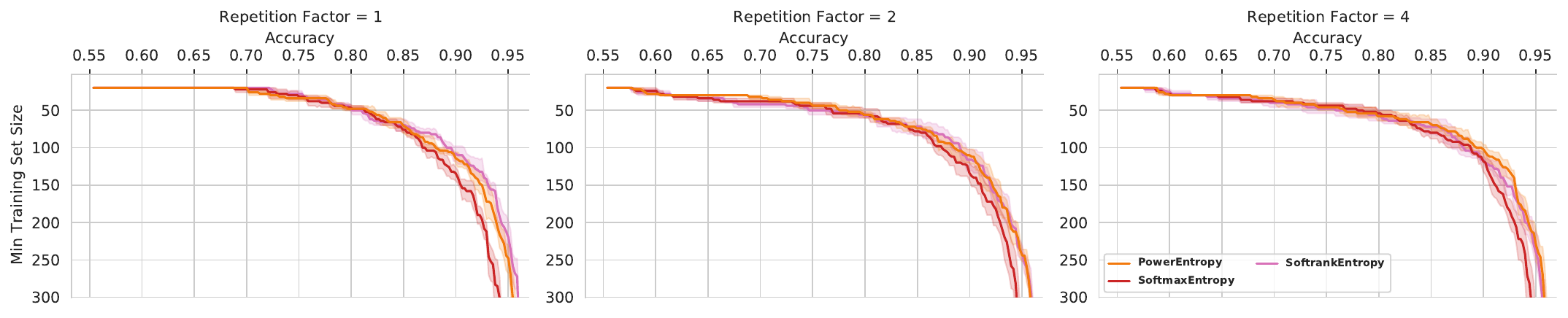}
        \caption{ %
        Entropy
        }
    \end{subfigure}
    \caption{ %
    \emph{Repeated-MNIST (5 trials): Performance with all three stochastic strategies.}
    }
    \label{stochacq:appfig:rmnist_all_three}
\end{figure}

\textbf{Repeated-MNIST.} In \Cref{stochacq:appfig:rmnist_all_three}, power acquisition performs the best overall, followed by soft-rank and then softmax.

\begin{figure}[H]
    \begin{minipage}[t]{0.45\linewidth}
        \centering
        \includegraphics[width=\linewidth]{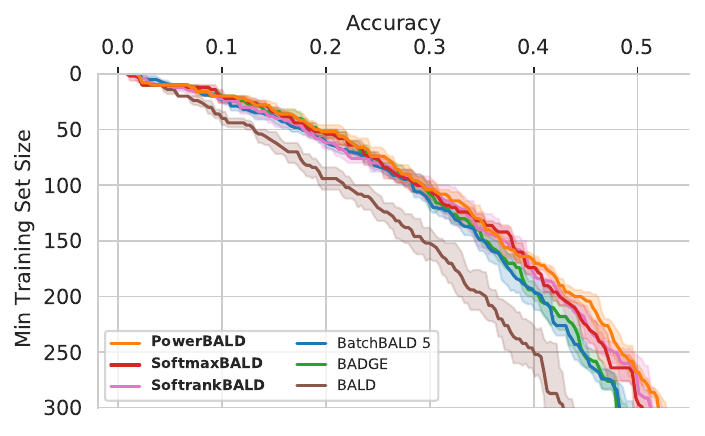}
        \caption{ %
        \emph{EMNIST (Balanced) (5 trials): Performance with all three stochastic strategies with BALD.} PowerBALD performs best.
        }
        \label{stochacq:appfig:emnist_balanced_bald}
    \end{minipage}
    \hfill
    \begin{minipage}[t]{0.45\linewidth}
        \centering
        \includegraphics[width=\linewidth]{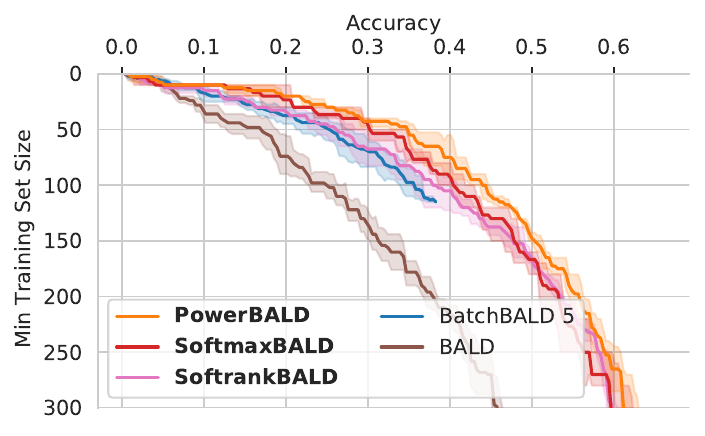}
        \caption{ %
        \emph{EMNIST (ByMerge) (5 trials): Performance with all three stochastic strategies with BALD.} PowerBALD performs best.
        }
        \label{stochacq:appfig:emnist_bald}
    \end{minipage}
\end{figure}

\textbf{EMNIST.} In \Cref{stochacq:appfig:emnist_balanced_bald} and \ref{stochacq:appfig:emnist_bald}, we see that PowerBALD performs best, but Softmax- and SoftrankBALD also outperform other methods. BADGE did not run on EMNIST (ByMerge) due to out-of-memory issues and BatchBALD took very long as EMNIST (ByMerge) has more than 800,000 samples. 

\begin{figure}[H]
    \begin{subfigure}[t]{0.45\linewidth}
        \centering
        \includegraphics[width=\linewidth]{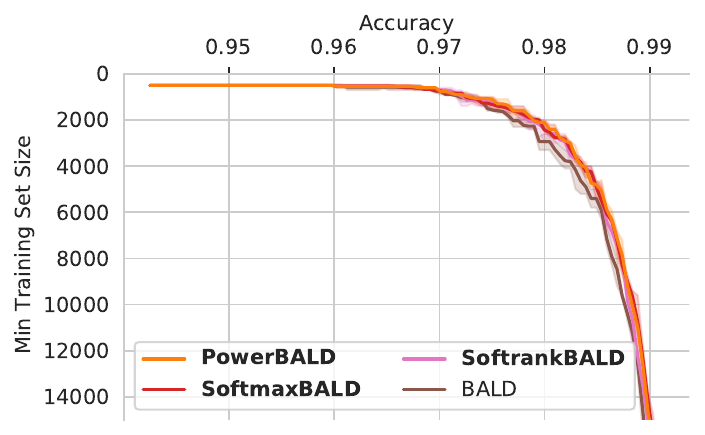}
        \caption{ %
            BALD
        }
    \end{subfigure}
    \hfill
    \begin{subfigure}[t]{0.45\linewidth}
        \centering
        \includegraphics[width=\linewidth]{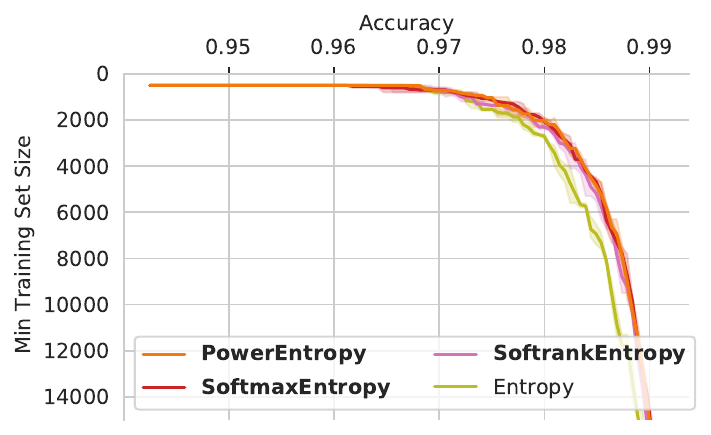}
        \caption{ %
        Entropy
        }
    \end{subfigure}
    \caption{ %
    \emph{MIO-TCD (3 trials): Performance with all three stochastic strategies.}
    }
    \label{stochacq:appfig:miotcd_all_three}
\end{figure}

\textbf{MIO-TCD.} In \Cref{stochacq:appfig:miotcd_all_three}, we see that all three stochastic acquisition methods perform about equally well.

\begin{figure}[H]
    \begin{subfigure}[t]{\linewidth}
        \centering
        \includegraphics[width=\linewidth]{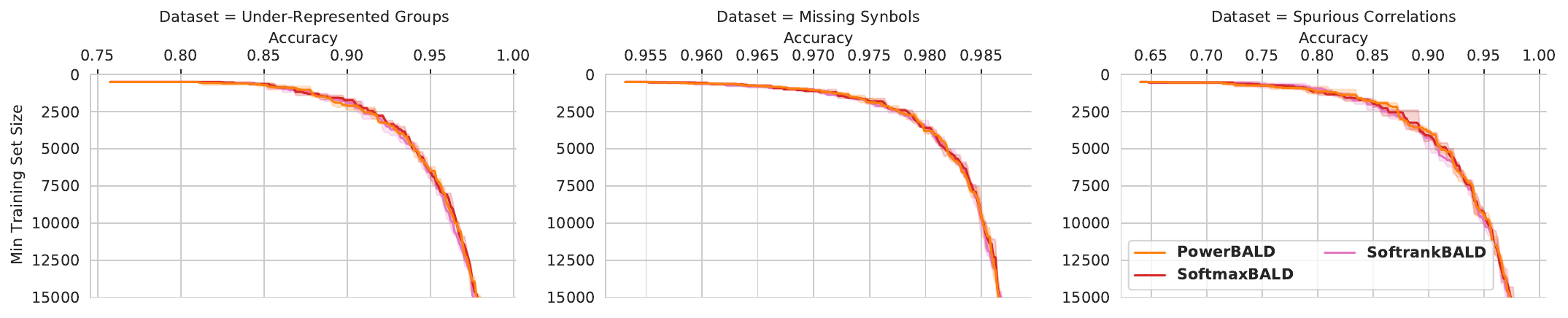}
        \caption{ %
            BALD
        }
    \end{subfigure}
    \begin{subfigure}[t]{\linewidth}
        \centering
        \includegraphics[width=\linewidth]{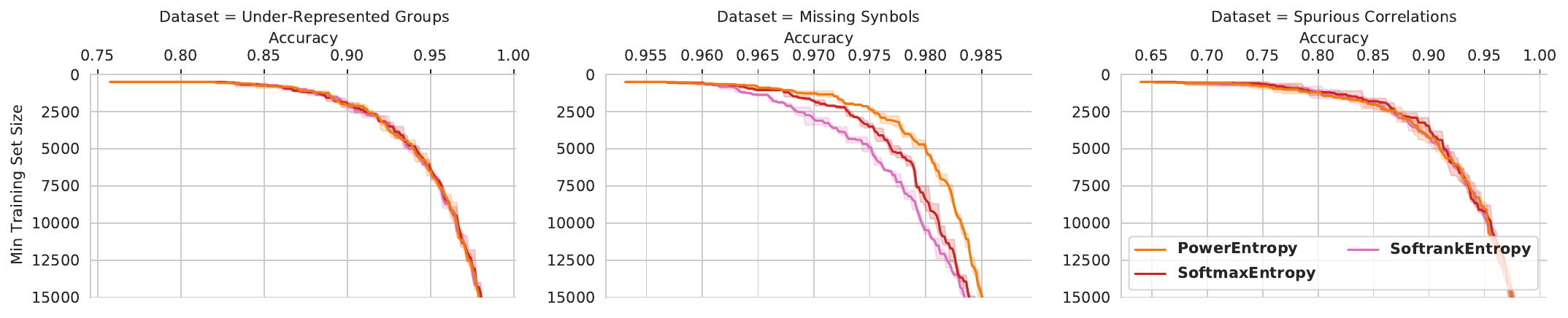}
        \caption{ %
        Entropy
        }
    \end{subfigure}
    \caption{ %
    \emph{Synbols edge cases (3 trials): Performance with all three stochastic strategies.}
    }
    \label{stochacq:appfig:synbols_all_three}
\end{figure}

\textbf{Synbols.} In \Cref{stochacq:appfig:synbols_all_three}, power acquisition seems to perform better overall---mainly due to the performance in \synmissing. 

\begin{figure}[H]
    \centering %
    \begin{minipage}[t]{0.50\linewidth}
        \centering
        \includegraphics[width=\linewidth]{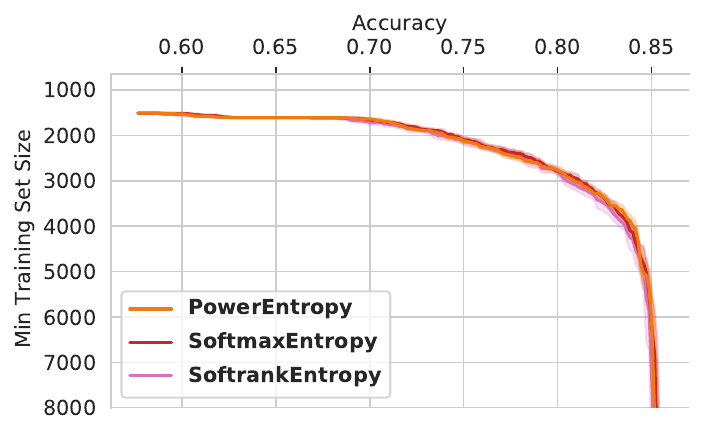}
        \caption{ %
            \emph{CLINC-150 (10 trials): Performance with all three stochastic strategies.}
        }
    \label{stochacq:appfig:clinc_all_three}
    \end{minipage}
\end{figure}

\textbf{CLINC-150.} In \Cref{stochacq:appfig:clinc_all_three}, all three stochastic methods perform similarly.

\subsection{Investigation}
\label{stochacq:appsec:sampling_distribution_investigation}

\begin{figure}[t]
    \begin{minipage}[t]{\linewidth} %
        \centering %
        \includegraphics[width=\linewidth]{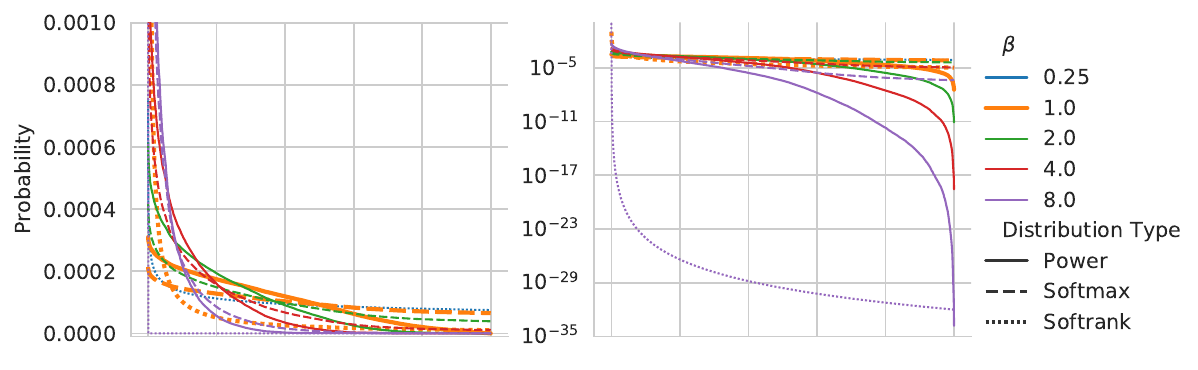} %
        \caption{
            \emph{Score distribution for power and softmax acquisition of BALD scores on MNIST for varying Coldness $\beta$ at $t=0$.} Linear and log plot over samples sorted by their BALD score. At $\beta=8$ both softmax and power acquisition have essentially the same distribution for high scoring points (closely followed by the power distribution for $\beta=4$). This might explain why the coldness ablation shows that these $\beta$ to have very similar AL trajectories on MNIST. Yet, while softmax and power acquisition seem transfer to RMNIST, this is not the case for softrank which is much more sensitive to $\beta$. 
            At the same time, power acquisition avoids low-scoring points more than softmax acquisition. %
        } %
        \label{stochacq:fig:score_distribution} %
    \end{minipage}
\end{figure}

To further examine the three stochastic acquisition variants, we plot their score distributions, extracted from the same MNIST toy example, in \Cref{stochacq:fig:score_distribution}. 
Power and softmax acquisition distributions are similar for $\beta=8$ (power, softmax) and $\beta=4$ (softmax).
This might explain why active learning with these $\beta$ shows similar accuracy trajectories.

We find that power and softmax acquisition are quite insensitive to $\beta$ and thus selecting $\beta=1$ might generally work quite well.

\section{Effect of Changing $\beta$}
\label{stochacq:appsec:tuning_beta}

\subsection{Repeated-MNIST}
\label{stochacq:appsec:rmnist_beta}

\begin{figure}[H]
    \centering
    \includegraphics[height=0.85\textheight]{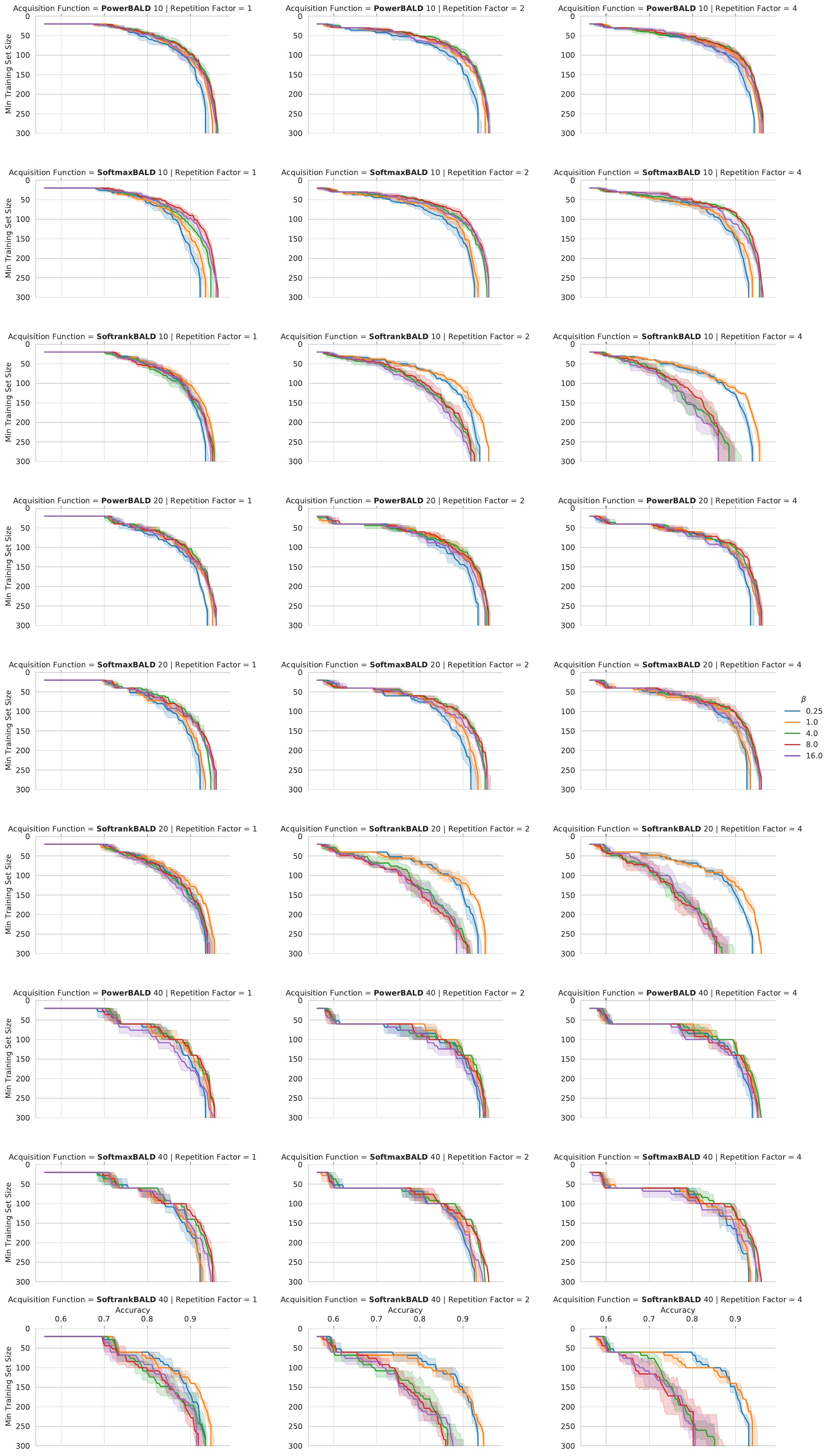}
    \caption{
    \emph{Repeated-MNIST: $\beta$ ablation for *BALD.}
    }
    \label{stochacq:appfig:rmnist_temperature_ablation}
\end{figure}

\subsubsection{\miotcd and Synbols}
\label{stochacq:appsec:eai_temperature_ablations}

\begin{figure}[H]
    \centering
    \includegraphics[width=\linewidth]{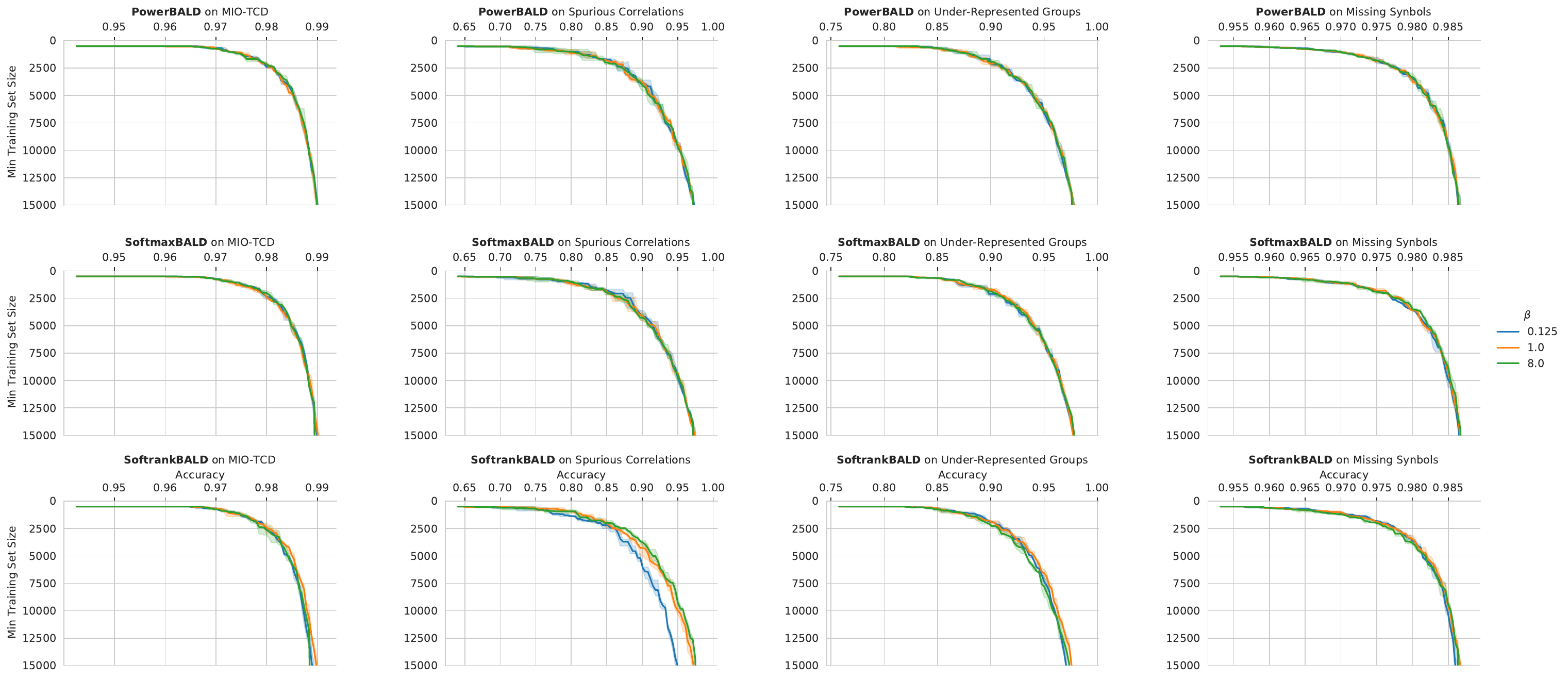}
    \caption{
    \emph{\miotcd and Synbols: $\beta$ ablation for *BALD.}
    }
\end{figure}

\begin{figure}[H]
    \centering
    \includegraphics[width=\linewidth]{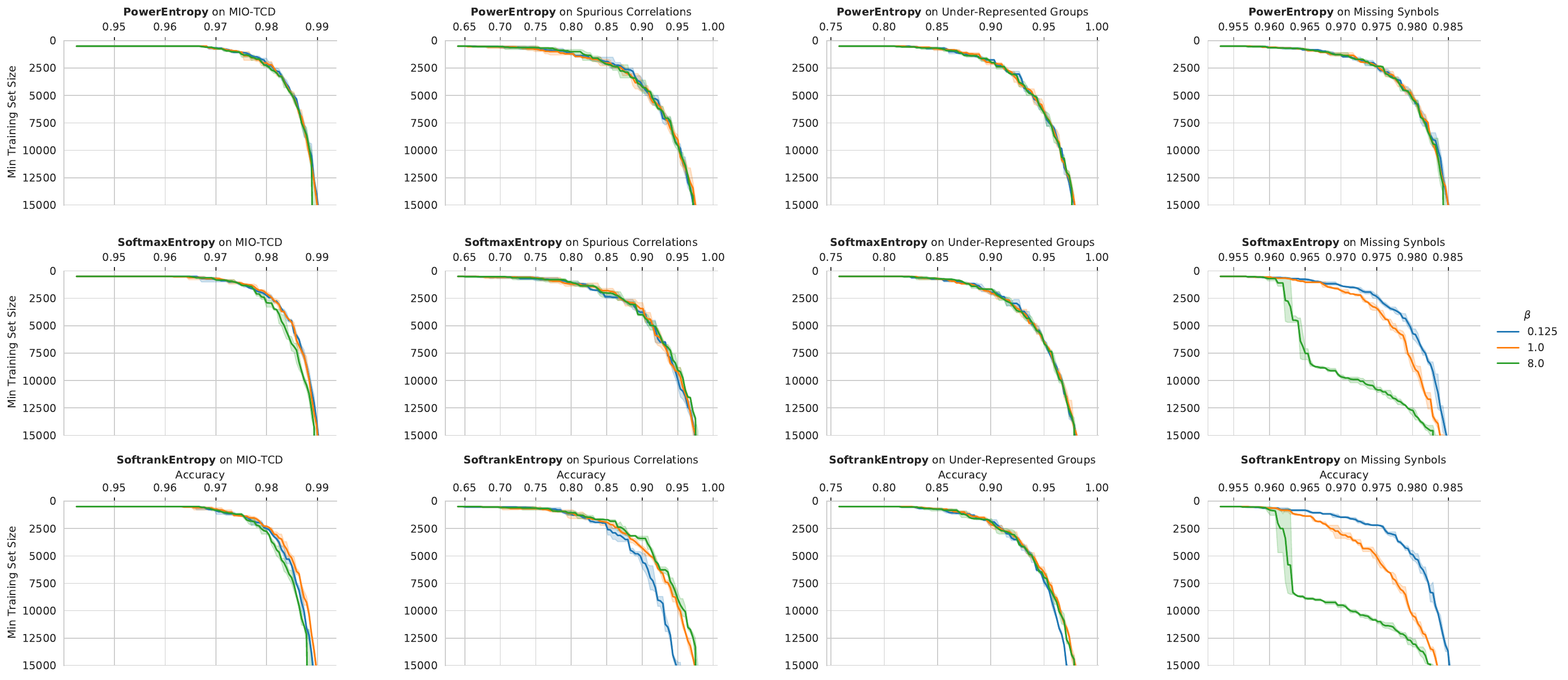}
    \caption{
    \emph{\miotcd and Synbols: $\beta$ ablation for *Entropy.}
    }
\end{figure}

\subsection{CausalBALD: Synthetic Dataset}
\label{stochacq:app:causal_synth_temp}

\begin{figure}[H]
    \centering
    \begin{subfigure}[b]{0.64\linewidth}
        \centering
        \includegraphics[width=\linewidth]{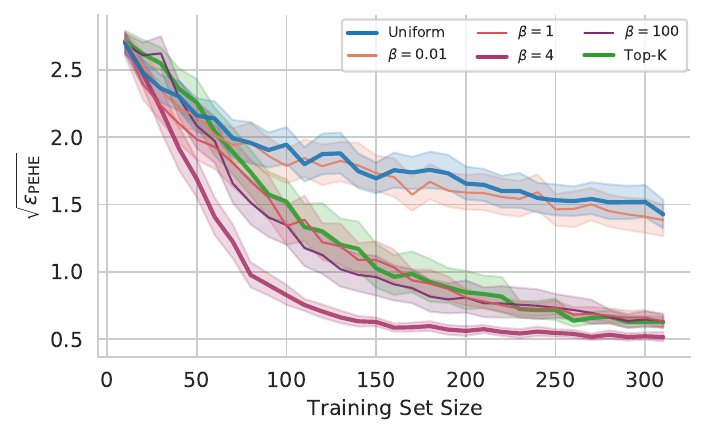}
        \caption{Overall Ablation (Subset)}
    \label{stochacq:fig:synth}
    \end{subfigure}
    \begin{minipage}[b]{0.30\linewidth}
            \begin{subfigure}[b]{\linewidth}
            \centering
            \includegraphics[width=\linewidth]{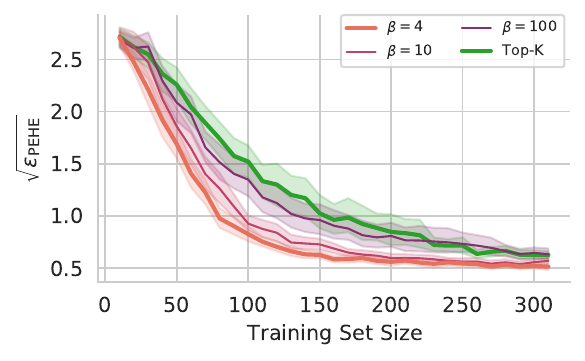}
            \caption{Low Temperature Only}
        \label{stochacq:fig:low_temp_synth}
        \end{subfigure}
        \begin{subfigure}[b]{\linewidth}
            \centering
            \includegraphics[width=\linewidth]{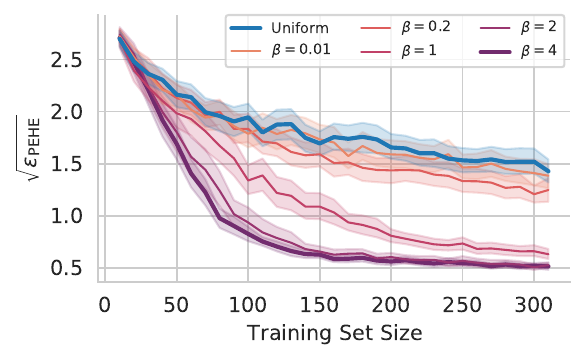}
            \caption{High Temperature Only}
            
        \label{stochacq:fig:high_temp_synth}
        \end{subfigure}
    \end{minipage}
    \caption{\emph{CausalBALD: Synthetic Dataset.} (a) At a very high temperature ($\beta = 0.1$), PowerBALD behaves very much like random acquisition, and as the temperature decreases the performance of the acquisition function improves (lower $\sqrt{\epsilon_{\mathrm{PEHE}}}$). (b) Eventually, the performance reaches an inflection point ($\beta = 4.0$) and any further decrease in temperature results in the acquisition strategy performing more like top-\batchvar. We see that under the optimal temperature, power acquisition significantly outperforms both random acquisition and top-\batchvar over a wide range of temperature settings.}
    \label{stochacq:fig:synth_causal}
\end{figure}

We provide further $\beta$ ablations for CausalBALD on the entirely synthetic dataset which is used by \citet{jesson2021causal}. This demonstrates the ways in which $\beta$ interpolates between uniform and top-\batchvar acquisition.

\subsection{CLINC-150}

\begin{figure}[H]
    \centering
    \includegraphics[width=\linewidth]{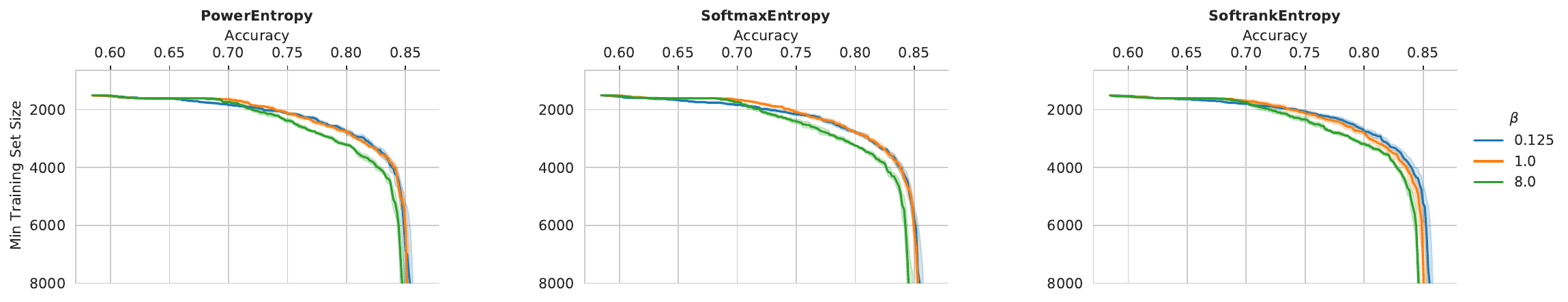}
    \caption{Performance CLINC-150: $\beta$ ablation for *Entropy.}
\end{figure}

}

\chapter{\pobaltitle}
{
	
	\graphicspath{{./epig/figures/}}

\newcommand{\ptest}[1]{\pcof{\mathrm{test}}{#1}}
\newcommand{\petest}[1]{\hpcof{\mathrm{test}}{#1}}
\newcommand{\ppool}[1]{\pcof{\mathrm{pool}}{#1}}
\newcommand{\peval}[1]{\pcof{\mathrm{eval}}{#1}}

\newcommand{\Dbatch}[1]{\Dany^\mathrm{acq}(#1)}
\newcommand{\yDeval}{\Dany^\mathrm{eval}(\yevalset)}

	\graphicspath{{./pobal/}{./epig/figures/}}
\makeatletter
\def\input@path{{./pobal/}}
\makeatother

\setlength{\figureheight}{4cm}

\newcommand{\basedata}[0]{\mathcal{D}_{\mathrm{base}}}
\newcommand{\pooldata}[0]{\mathcal{D}_{\mathrm{pool}}}
\newcommand{\traindata}[0]{\mathcal{D}_{\mathrm{train}}}
\newcommand{\testdata}[0]{\mathcal{D}_{\mathrm{test}}}

\section{BALD Estimation}
\label{pobal:sec:bald_estimation}

In general, we can estimate BALD using nested Monte Carlo \citep{rainforth2018nesting}:
\begin{align}
\baldof{\x}
&=
\E{\pof{\theta}}{
    -\E{\pof{\y \given \x}}{\log \pof{\y \given \x}}
    +\E{\pof{\y \given \x,\theta}}{\log \pof{\y \given \x,\theta}}
}
\\
&\approx
\frac{1}{M} \sum_{j=1}^M
-\log\left(\frac{1}{K}\sum_{i=1}^K \pof{y_j \given \x,\theta_i}\right)
+ \log \pof{y_j \given \x,\theta_j}
,
\end{align}
where $\theta_i \sim \pof{\theta}$, $(\theta_j, y_j)\sim \pof{\theta}\pof{\y \given \x,\theta}$.
Special cases allow us to use computationally cheaper estimators.

\subsection{Categorical Predictive Distribution}

When $\y$ and $\yeval$ are discrete, we can write
\begin{align}
\baldof{\x}
&=
\E{\pof{\theta}}{
    -\E{\pof{\y \given \x}}{\log \pof{\y \given \x}}
    +\E{\pof{\y \given \x,\theta}}{\log \pof{\y \given \x,\theta}}
}
\\
&=
-\E{\pof{\y \given \x}}{\log \pof{\y \given \x}}
+\E{\pof{\theta}\pof{\y \given \x,\theta}}{\log \pof{\y \given \x,\theta}}
\\
&=
-\sum_{\y\in\mathcal{Y}} \pof{\y \given \x} \log \pof{\y \given \x}
+\E{\pof{\theta}}{\sum_{\y\in\mathcal{Y}} \pof{\y \given \x,\theta} \log \pof{\y \given \x,\theta}}
.
\end{align}
This can be estimated using samples, $\theta_i\sim \pof{\theta}$ \citep{houlsby2014efficient}:
\begin{align}
\label{pobal:eq:bald_classification_estimator}
\baldof{\x}
&\approx
-\sum_{\y\in\mathcal{Y}} \hpof{\y \given \x} \log \hpof{\y \given \x}
+\frac{1}{K} \sum_{i=1}^K \sum_{\y\in\mathcal{Y}} \pof{\y \given \x,\theta_i} \log \pof{\y \given \x,\theta_i}
,
\end{align}
where
\begin{align}
\hpof{\y \given \x} = \frac{1}{K}\sum_{i=1}^K \pof{\y \given \x,\theta_i}
.
\end{align}

\subsection{Gaussian Predictive Distribution}

Suppose we have a model whose likelihood function, $\pof{\y \given \x,\theta}$, and predictive distribution, $\pof{\y \given \x}$, are Gaussian.
Then, using the symmetry of the mutual information along with knowledge of the entropy of a Gaussian \citep{cover2005elements}, we have
\begin{align}
\baldof{\x} &= \frac{1}{2} \log 2 \pi e \implicitVar{{\Y \given \x}} - \frac{1}{2} \E{\pof{\theta}}{\log 2 \pi e \implicitVar{{\Y \given \x,\theta}}}
\\
&= \frac{1}{2} \left( \log \implicitVar{{\Y \given \x}} - \E{\pof{\theta}}{\log \implicitVar{{\Y \given \x,\theta}}} \right)
.
\end{align}
Relatedly, \citet{houlsby2011bayesian} identified a closed-form approximation of BALD for the particular case of using a probit likelihood function, a Gaussian-process prior and a Gaussian approximation to the predictive distribution.
\section{EPIG Derivation}
\label{pobal:sec:epig_derivation}

Computing an expectation over both $\y$ and $\xeval$ gives the expected predictive information gain:
\begin{align}
\epigof{\x}
&=
\E{\peval{\xeval} \pof{\y \given \x}}{\xHof{\pof{\yeval \given \xeval}} - \xHof{\pof{\yeval \given \y,\x,\xeval}}}
\\
&=
\E{\peval{\xeval} \pof{\y \given \x}}{
    -\E{\pof{\yeval \given \xeval}}{\log \pof{\yeval \given \xeval}}
} \notag \\
&\quad +\E{\peval{\xeval} \pof{\y \given \x}}{\E{\pof{\yeval \given \y,\x,\xeval}}{\log \pof{\yeval \given \y,\x,\xeval}}
}
\\
&=
\E{\peval{\xeval} \pof{\y,\yeval \given \x,\xeval}}{
    \log\frac{\pof{\yeval \given \y,\x,\xeval}}{\pof{\yeval \given \xeval}}
}
\\
&=
\E{\peval{\xeval} \pof{\y,\yeval \given \x,\xeval}}{
    \log\frac{\pof{\y \given \x}\pof{\yeval \given \y,\x,\xeval}}{\pof{\y \given \x}\pof{\yeval \given \xeval}}
}
\\
&=
\E{\peval{\xeval} \pof{\y,\yeval \given \x,\xeval}}{
    \log\frac{\pof{\y,\yeval \given \x,\xeval}}{\pof{\y \given \x}\pof{\yeval \given \xeval}}
}
\\
&=
\E{\peval{\xeval}}{
    \MIof{y;\yeval \given \x,\xeval}
}
\\
&=
\E{\peval{\xeval}}{
    \Kale{\pof{\y,\yeval \given \x,\xeval}}{\pof{\y \given \x}\pof{\yeval \given \xeval}}
}.
\end{align}
\section{EPIG Estimation}
\label{pobal:sec:epig_estimation}

While in general we can use \Cref{pobal:eq:epig_sample_estimator} to estimate EPIG, special cases allow computationally cheaper estimators.

\subsection{Categorical Predictive Distribution}

When $\y$ and $\yeval$ are discrete, we can write
\begin{align}
\epigof{\x}
&=
\E{\peval{\xeval}}{
    \Kale{\pof{\y,\yeval \given \x,\xeval}}{\pof{\y \given \x}\pof{\yeval \given \xeval}}
}
\\
&=
\E{\peval{\xeval}}{
    \sum_{y\in\mathcal{Y}} \sum_{\yeval\in\mathcal{Y}} \pof{\y,\yeval \given \x,\xeval} \log \frac{\pof{\y,\yeval \given \x,\xeval}}{\pof{\y \given \x}\pof{\yeval \given \xeval}}
}
.
\end{align}
This can be estimated using samples, $\theta_i\sim \pof{\theta}$ and $\xeval_j\sim \peval{\xeval}$:
\begin{align}
\epigof{\x}
&\approx
\frac{1}{M} \sum_{j=1}^M \sum_{y\in\mathcal{Y}} \sum_{\yeval\in\mathcal{Y}}
\hpof{y,\yeval \given \x,\xeval_j} \log \frac{\hpof{y,\yeval \given \x,\xeval_j}}{\hpof{y \given \x}\hpof{\yeval \given \xeval_j}}
,
\end{align}
where
\begin{align}
\hpof{y,\yeval \given \x,\xeval_j}
&=
\frac{1}{K}\sum_{i=1}^K \pof{\y \given \x,\theta_i}\pof{\yeval \given \xeval_j,\theta_i}
\\
\hpof{y \given \x}
&=
\frac{1}{K}\sum_{i=1}^K \pof{\y \given \x,\theta_i}
\\
\hpof{\yeval \given \xeval_j}
&=
\frac{1}{K}\sum_{i=1}^K \pof{\yeval \given \xeval_j,\theta_i}
.
\end{align}

\subsection{Gaussian Predictive Distribution}
\label{pobal:sec:epig_estimation_gaussian}

Consider a joint predictive distribution that is multivariate Gaussian with mean vector $\mu$ and covariance matrix $\Sigma$:
\begin{align}
\pof{\y,\yeval \given \x,\xeval}
=
\mathcal{N}(\mu, \Sigma)
=
\mathcal{N}
\left(
\mu,
\begin{bmatrix}
\implicitCov{x; x} & \implicitCov{x; \xeval} \\
\implicitCov{x; \xeval} & \implicitCov{\xeval; \xeval}
\end{bmatrix}
\right)
.
\end{align}
In this setting the mutual information between $\y$ and $\yeval$ given $\x$ and $\xeval$ is a closed-form function of $\Sigma$:
\begin{align}
\MIof{Y;\Yeval \given \x,\Xeval}
&=
\xHof{\pof{\y \given \x}} + \xHof{\pof{\yeval \given \xeval}} - \xHof{\pof{\y,\yeval \given \x,\xeval}}
\\
&=
\frac{1}{2} \log 2 \pi e \implicitVar{\pof{\y \given \x}} + \frac{1}{2} \log 2 \pi e \implicitVar{\pof{\yeval \given \xeval}}
\notag \\
&\quad = - \frac{1}{2} \log \det 2 \pi e \Sigma
\\
&=
\frac{1}{2} \log \frac{
    \implicitVar{\pof{\y \given \x}} \implicitVar{\pof{\yeval \given \xeval}}
}{
    \det\Sigma
} \\
&=
\frac{1}{2} \log \frac{
    \implicitCov{x; x} \implicitCov{\xeval; \xeval}
}{
    \det\Sigma
} \\
&=
\frac{1}{2} \log \frac{
    \implicitCov{x; x} \implicitCov{\xeval; \xeval}
}{
    \implicitCov{x; x} \implicitCov{\xeval; \xeval} - \implicitCov{x; \xeval}^2
}
.
\end{align}
We can estimate EPIG using samples, $\xeval_j\sim \peval{\xeval}$:
\begin{align}
\epigof{\x}
&=\E{\peval{\xeval}}{\MIof{y;\yeval \given \x,\xeval}}
\approx
\frac{1}{M} \sum_{j=1}^M \MIof{y;\yeval \given \x,\xeval_j}
\\
&=
\frac{1}{2M} \sum_{j=1}^M \log \frac{
    \implicitCov{x; x} \implicitCov{\xeval_j; \xeval_j}
}{
    \implicitCov{x; x} \implicitCov{\xeval_j; \xeval_j} - \implicitCov{x; \xeval_j}^2
}
.
\end{align}

\subsection{Connection to \citet{foster2019variational}}
\label{pobal:sec:app:vboed_connection}

\citet{foster2019variational} primarily considered variational estimation of the expected information gain.
Since the joint density, $\pof{\y,\yeval \given \x,\xeval}$, that appears in EPIG is often not known in closed form, EPIG estimation broadly falls under the ``implicit likelihood'' category of methods considered in that paper.
Here, we focus on showing how the ``posterior'' or Barber-Agakov bound \citep{barber2003im} from this earlier work applies to EPIG estimation.
We first recall \Cref{pobal:eq:gen_loss},
\begin{align}
    \epigof{\x} = \E{\peval{\xeval} \pof{\y,\yeval \given \x,\xeval}}{\log \pof{\yeval \given \xeval,\y,\x}} + \xHof{\pof{\yeval \given \xeval}}
,
\end{align}
and the observation that $c = \xHof{\pof{\yeval \given \xeval}}$ does not depend upon $\x$ and hence can be neglected when choosing between designs.
By Gibbs's inequality, we must have
\begin{align}
    \epigof{\x} \ge \E{\peval{\xeval} \pof{\y,\yeval \given \x,\xeval}}{\log q(\yeval|\xeval,\y,\x)} + \xHof{\pof{\yeval \given \xeval}}
\end{align}
for any distribution $q$.
We can now consider a variational family, $q_\psi(\yeval|\xeval,\y,\x)$, and a maximization over the variational parameter, $\psi$:
\begin{align}
    \epigof{\x} \ge \sup_\psi \E{\peval{\xeval} \pof{\y,\yeval \given \x,\xeval}}{\log q_\psi(\yeval|\xeval,\y,\x)} + \xHof{\pof{\yeval \given \xeval}}
    .
\end{align}
A practical implication of this bound is that we could estimate EPIG by learning an auxiliary network, $q_\psi(\yeval|\xeval,\y,\x)$, using data simulated from the model to make one-step-ahead predictions.
That is, $q_\psi$ is trained to make predictions at $\xeval$, incorporating the knowledge of the hypothetical acquisition $(\x,
\y)$.
For our purposes, training such an auxiliary network at each acquisition is prohibitively expensive.
But this approach might be valuable in other applications of EPIG.

\section{Dataset Construction}

\subsection{UCI Data}
\label{pobal:sec:uci_appendix}

For each dataset we start by taking the base dataset, $\basedata$, from the UCI repository.
Satellite and Vowels have predefined test datasets, $\testdata$.
In contrast, Magic does not have a predefined train-test split.
It is stated in Magic's documentation that one of the classes is underrepresented in the dataset relative to real-world data (Magic is a simulated dataset).
Whereas classes 0 and 1 respectively constitute 65\% and 35\% of the dataset, it is stated that class 1 constitutes the majority of cases in reality (the exact split is not stated; we assume 75\% for class 1).
We therefore uniformly sample 30\% of $\basedata$ to form a test base dataset, $\basedata'$; then we set $\basedata\gets\basedata\setminus\basedata'$; then we make $\testdata$ by removing input-label pairs from $\basedata'$ such that class 1 constitutes 75\% of the subset.
With the test set defined, we proceed to sample two disjoint subsets of $\basedata$ such that their class proportions match those of $\basedata$: a pool set, $\pooldata$, whose size varies between datasets, and a validation set, $\mathcal{D}_\mathrm{val}$, of 60 input-label pairs.
Regardless of the class proportions of $\basedata$, we always use an initial training dataset, $\mathcal{D}_\mathrm{init}$, of 2 input-label pairs per class, sampled from $\basedata$.
Finally, we sample a representative set of inputs, $\mathcal{D}_*$, whose class proportions match those of $\testdata$.

\subsection{MNIST Data}
\label{pobal:sec:mnist_appendix}

Implementing each setting starts by using the standard MNIST training data (60,000 input-label pairs) as the base dataset, $\basedata$, and the standard MNIST testing data (10,000 input-label pairs) as the test base dataset, $\basedata'$.
For Redundant MNIST we make $\testdata$ by removing input-label pairs from $\basedata'$ such that only classes 1 and 7 remain.
Otherwise, we set $\testdata=\basedata'$.
Next we construct the pool set, $\pooldata$.
For Curated MNIST and Redundant MNIST we sample 4,000 inputs per class from $\basedata'$.
For Unbalanced MNIST we sample 400 inputs per class for classes 0-4 and 4,000 inputs per class for classes 5-9.
After this we make the initial training dataset, $\mathcal{D}_\mathrm{init}$.
For Curated MNIST and Unbalanced MNIST we sample 2 input-label pairs per class from $\basedata$.
For Redundant MNIST we sample 2 input-label pairs from class 1, 2 input-label pairs from class 7 and 1 input-label pair per class from 2 randomly selected classes other than 1 and 7.
Next, the validation set, $\mathcal{D}_\mathrm{val}$.
For all settings this comprises 60 input-label pairs such that the class proportions match those used to form $\pooldata$.
Finally, we sample a representative set of inputs, $\mathcal{D}_*$, whose class proportions match those of $\testdata$.
	{

\section{EPIG \&\ JEPIG}
\label{EPIG:appsec:epig_and_jepig}
\relationjpigepig*
\begin{proof}
    We drop conditioning on $\Dtrain$ in this proof.
    First, we remind ourselves of the total correlation:
    \begin{align}
        \TCof{A_1;\ldots; A_n} = \sum_{i=1}^n \Hof{A_i} - \Hof{A_{1, \ldots,n}} \ge 0.
    \end{align}
    We expand EPIG to the right:
    \begin{align}
        &\MIof{\Yeval; \Yacq \given \Xeval, \xacq} = \Hof{\Yeval \given \Xeval} - \Hof{\Yeval \given \Xeval, \Yacq, \xacq}.
    \end{align}
    The first term on the right-hand side is constant given a fixed evaluation set.
    We can express both terms in terms of the total correlation:
    \begin{align}
        &\evalsize \, \Hof{\Yeval \given \Xeval} \\
        &\quad = \TCof{\Yeval_1;\ldots;\Yeval_\evalsize \given \xevalset} + \Hof{\Yevalset \given \xevalset, \Yacq, \xacq},
    \intertext{and}
        &\evalsize\, \Hof{\Yeval \given \Xeval, \Yacq, \xacq} \\
        &\quad = \TCof{\Yeval_1;\ldots;\Yeval_\evalsize \given \xevalset, \Yacq, \xacq} + \Hof{\Yevalset \given \xevalset, \Yacq, \xacq}.
    \end{align}
    $c_\typeeval \defeq \nicefrac{1}{\evalsize} \TCof{\Yeval_1;\ldots;\Yeval_\evalsize \given \Xeval}$ is independent of the acquisition set.
    Hence, we can conclude:
    \begin{align}
        &\MIof{\Yeval; \Yacq \given \Xeval, \xacq} \\
        &\quad =
        \nicefrac{1}{\evalsize} (\Hof{\Yevalset \given \xevalset, \Yacq, \xacq} - \Hof{\Yevalset \given \xevalset, \Yacq, \xacq} \\
        &\quad \quad - \TCof{\Yeval_1;\ldots;\Yeval_\evalsize \given \Xeval, \Yacq, \xacq}) + c_\typeeval \\
        &\quad = \nicefrac{1}{\evalsize} \MIof{\Yevalset;\Yacq \given \xevalset, \xacq} + c_\typeeval\\
        &\quad \quad - \nicefrac{1}{\evalsize} \underbrace{\TCof{\Yeval_1;\ldots;\Yeval_\evalsize \given \xevalset, \Yacq, \xacq}}_{\ge 0} \\
        &\quad \le \nicefrac{1}{\evalsize} \MIof{\Yevalset;\Yacq \given \xevalset, \xacq} + c_\typeeval.
    \end{align}
    Thus, we see that the inequality gap is $\nicefrac{1}{\evalsize} \TCof{\Yeval_1;\ldots;\Yeval_\evalsize \given \xevalset, \Yacq, \xacq}$, which is zero, if and only if the random variables $\Yeval_i \given \xeval_i$ are independent given the acquisition samples.
\end{proof}

\section{A Practical Approximation of JEPIG}
\label{EPIG:sec:practical_approximation}
We want to find an approximation $\hat{\W}$ with distribution $\qof{\hat{\w}}$, such that for all possible acquisition sets, we have:
\begin{align}
    \MIof{\Yacqset ; \W \given \xacqset, \Yevalset, \xevalset, \Dtrain}  \approx \MIof{\Yacqset ; \hat{\W} \given \xacqset}.
\end{align}
We note two properties of this conditional mutual information and the underlying models $\pof{\W \given \yevalset, \xevalset, \Dtrain}$ for different $\yevalset \sim \pof{\yevalset \given \Dtrain}$:
\begin{enumerate}[leftmargin=*]
    \item marginalizing $\pof{\W \given \yevalset, \xevalset, \Dtrain}$ over all possible $\yevalset$ yields the predictions of the original posterior $\pof{\W \given \Dtrain}$, so we would like to have
    \begin{align*}
        \simpleE{\qof{\hat{\w}}}\pof{y \given x, \hat{\w}} = \pof{y \given x, \Dtrain}; 
    \end{align*}
    \item $\MIof{Y ; \W \given \x, \Yevalset, \xevalset, \Dtrain} \le \MIof{Y ; \W \given \x, \Dtrain}$, and when $\x \in \{\xeval_i\}_i$, we expect $\MIof{Y ; \W \given \x, \Yevalset, \xevalset, \Dtrain} \ll \MIof{Y ; \W \given \x, \Dtrain}$. In other words, the epistemic uncertainty of evaluation samples $\xeval$ ought to decrease when we also train on the evaluation set (using pseudo-labels $\yevalset$), and we would like the same for $\hat{\W}$: 
    \begin{align*}
        \MIof{Y ; \hat{\W} \given x} \le \MIof{Y ; \W \given x, \Dtrain}.
    \end{align*}
\end{enumerate}
Note, that the second property follows from EPIG being non-negative as mutual information in two terms and thus so is JEPIG, which means the difference between the two BALD terms is non-negative and the second property is obtained from that.

Hence, as a tractable approximation $\hat{\W}$, we choose to use a form of \emph{self-distillation}, where we train a model with $\Dtrain$ and the predictions of the original model $\pof{\w \given \Dtrain}$ on $\xevalset$ using a KL-divergence loss, inspired by \citet{hinton2015distilling} and \citet{zhang2019teacher}\footnote{Essentially using \(\alpha=1, \lambda=0\)}. The loss function this is:
\begin{align}
    &L(\xtrainset, \pof{\w}, \qof{\hat{w} \given \Dtrain}) = \notag \\
    & \quad = \frac{1}{\lvert \Dtrain \rvert} \sum_i \Kale{\pof{Y\given \xtrain_i, \Dtrain}}{\qof{Y \given \xtrain_i}} \notag \\
    & \quad \quad + \Kale{\qof{\w}}{\pof{\w}},
\end{align}
with \(\pof{Y\given \xtrain_i} = \simpleE{\pof{\w \given\Dtrain}}{\pof{Y\given \xtrain_i, \w}}\) and \(\qof{Y\given \xtrain_i} = \simpleE{\qof{\hat{\w}}}{\pof{Y\given \xtrain_i, \hat{\w}}}\).
The resulting model posterior $\hat{\W}$ fulfills both properties described above. This is similar to self-distillation in that we train a new model on the predictions of the original model. However, self-distillation does not use predictions on otherwise unlabeled data.
It is also similar to semi-supervised learning \citep{Lee2013PseudoLabelT, yarowsky1995unsupervised} in that we use the predictions of the model on unlabeled data to train a new model. However, semi-supervised learning only uses the samples for which the model is most confident via confidence thresholding and either temperature-scales them for training (soft pseudo-labels) or takes the argmax (hard pseudo-labels) whereas we use the predictions without change for all evaluation samples.

\textbf{Advantages of JEPIG.} Compared to JEPIG, when evaluating
\begin{align}
    &\MIof{\Yevalset; \Yacqset \given \xacqset, \xevalset, \Dtrain} \\
    & \quad = \Hof{\Yacqset \given \xacqset, \Dtrain} \\
    & \quad \quad - \Hof{\Yacqset \given \xacqset, \Yevalset, \xevalset, \Dtrain} \notag \\
    & \quad \approx \varHof{\opq_\emptyset}{\Yacqset \given \xacqset} - \frac{1}{M} \sum_i  \varHof{\opq_{i}}{\Yacqset \given \xacqset}    
\end{align}
with separate approximate Bayesian models \(\opq_\emptyset\) and \(\opq_{i}\) where $\qcof{\emptyset}{\w} \approx \pof{\w \given \Dtrain}$ and $\qcof{i}{\w} \approx \pof{\w \given \yset^i, \xevalset, \Dtrain}$ for $\yset^{i} \sim \pof{\yevalset \given \xevalset, \Dtrain}, i \in 1..M$ for $M$ draws of pseudo-labels for \(\yevalset\), we found that:
\begin{equation*}
    \xHof{\qcof{\emptyset}{Y \given x, \W}} {\color{red}\not=} \frac{1}{M} \sum_i \xHof{\qcof{i}{Y \given x, \W}},
\end{equation*}
which violates the modelling assumption
\begin{equation*}
    \Hof{Y \given x, \W, \Dtrain} = \Hof{Y \given x, \W, \Yevalset, \xevalset, \Dtrain}
\end{equation*}
as $Y \independent \Yevalset \given \x, \xevalset, \W$. This is even more of an issue when using a single approximate model with the self-distillation described in the previous section because the two properties we wish for will force \(\Hof{Y \given x, \hat{\W}} \not= \Hof{Y \given x, \Dtrain}\).
This follows immediately from the expansion of the mutual information:
\begin{math}
    \MIof{Y ; \W \given x, \Dtrain}  = \Hof{Y \given x, \Dtrain} - \Hof{Y \given x, \W, \Dtrain}
\end{math}
as the first property will fix \(\Hof{Y \given x, \Dtrain}\) and the second will force the \(\Hof{Y \given x, \W, \Dtrain}\) terms apart to achieve the inequality.
However, JEPIG does not need this assumption as it explicitly estimates the epistemic uncertainty, and thus performs better when using our approximation with self-distillation. Moreover, it has its own strong intuitive motivation.

}
}

\chapter{\rholosstitle}
{
	
	\graphicspath{{./epig/figures/}}

\newcommand{\ptest}[1]{\pcof{\mathrm{test}}{#1}}
\newcommand{\petest}[1]{\hpcof{\mathrm{test}}{#1}}
\newcommand{\ppool}[1]{\pcof{\mathrm{pool}}{#1}}
\newcommand{\peval}[1]{\pcof{\mathrm{eval}}{#1}}

\newcommand{\Dbatch}[1]{\Dany^\mathrm{acq}(#1)}
\newcommand{\yDeval}{\Dany^\mathrm{eval}(\yevalset)}

	\graphicspath{{./rholoss/}}

\newcommand{\xval}{\xevalset}
\newcommand{\yval}{\yevalset}

\newcommand{\redloss}{\mathcal{L}_\mathrm{red}}
\newcommand{\irrloss}{\mathcal{L}_\mathrm{irr}}
\newcommand{\loss}{\mathcal{L}_\mathrm{t}}

\newcommand{\xsubset}{\x^\mathrm{subset}}
\newcommand{\ysubset}{\y^\mathrm{subset}}

\newcommand{\xevali}{\x^\mathrm{eval}}
\newcommand{\Xevali}{\X^\mathrm{eval}}
\newcommand{\yevali}{\y^\mathrm{eval}}
\newcommand{\Yevali}{\Y^\mathrm{eval}}
\newcommand{\xbatch}{\x^\mathrm{batch}}
\newcommand{\Xbatch}{\X^\mathrm{batch}}
\newcommand{\ybatch}{\y^\mathrm{batch}}
\newcommand{\Ybatch}{\Y^\mathrm{batch}}

\newcommand{\fix}{\marginpar{FIX}}
\newcommand{\new}{\marginpar{NEW}}

\newtcolorbox{mybox}[1]{%
    tikznode boxed title,
    enhanced,
    arc=0mm,
    interior style={white},
    attach boxed title to top center= {yshift=-\tcboxedtitleheight/2},
    fonttitle=\bfseries,
    colbacktitle=white,coltitle=black,
    boxed title style={size=normal,colframe=white,boxrule=0pt},
    title={#1}}
	\FloatBarrier
\section{Steps Required for a Given Test Accuracy} \label{rholoss:app:steps_needed}

Figs.~\ref{rholoss:fig:walltime}~(vision)~and~\ref{rholoss:fig:walltime_nlp}~(NLP) show the number of steps required to reach a given test accuracy across several datasets for different selection methods. Interestingly, on CoLA (unbalanced and noisy), the uniform sampling baseline shows high variance across seeds, while RHO-LOSS works robustly across seeds. 

\Cref{rholoss:tab:double_irlomo} shows results for RHO-LOSS training without holdout data. Results are similar to \Cref{rholoss:tab:speedups}. Here, we train the IL model without any holdout data. We split the training set $\Dpool$ into two halves and train an IL model on each half.  Each model computes the IL for the half of $\Dpool$ that it was not trained on. (This is as in \Cref{rholoss:fig5}, row 3, except that previously we only used half of $\Dpool$ and further split it into halves of the half.) Training two IL models costs no additional compute since each model is trained on half as much data compared to the default settings.

\begin{figure*}[t]
    \centering
    \includegraphics[width=\textwidth]{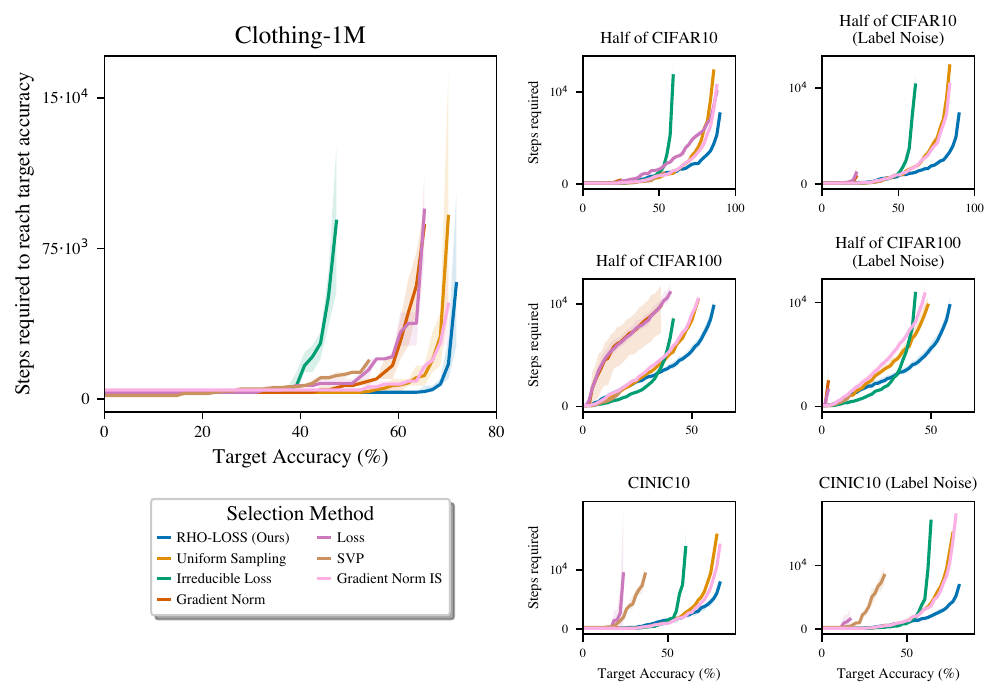}
    \caption{Vision datasets---gradient steps required to achieve a given test accuracy \textbf{(lower is better)}. \textbf{Left column:} The speedup of RHO-LOSS over uniform sampling is the greatest on a large-scale web-scraped dataset with noisy labels.  \textbf{Middle column:}  Speedups are still substantial on clean datasets and RHO-LOSS still achieves higher final accuracy than all prior art. \textbf{Right column:} Applying $10\%$ uniform label noise to training data degrades other methods but increases the speedup of our method. A step corresponds to lines $5-10$ in Algorithm~1. Lines correspond to means and shaded areas to minima and maxima across 3 random seeds. On CIFAR10/100, only half of the data is used for training (see text).}
    \label{rholoss:fig:walltime}
\end{figure*}

\begin{figure*}[t]
    \centering
    \includegraphics[width=\textwidth]{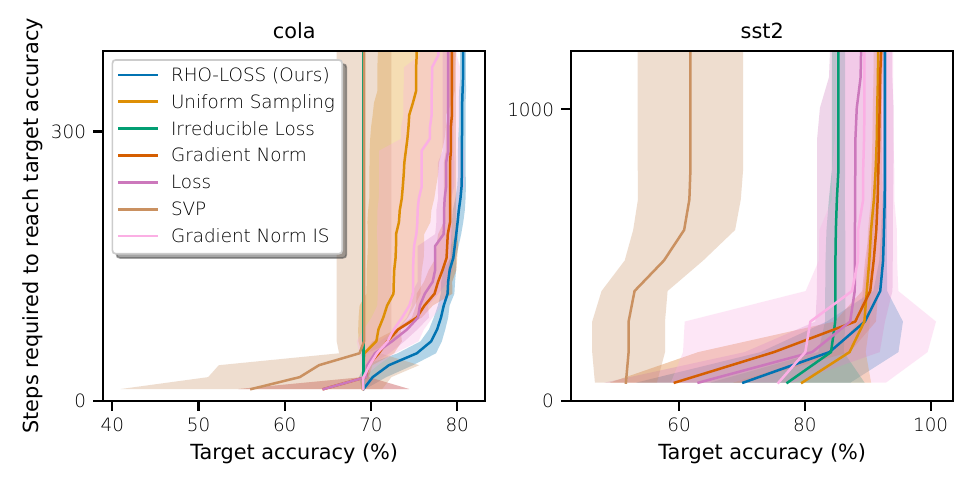}
    \caption{NLP datasets---gradient steps required to achieve a given test accuracy \textbf{(lower is better)}. \textbf{Left:} CoLA grammatical acceptability classification. \textbf{Right:} SST2 sentiment classification. A step corresponds to lines $5-10$ in Algorithm~1. Lines correspond to means and shaded areas to standard deviations across 4 or more random seeds. Only half of the data is used for training (see text).}
    \label{rholoss:fig:walltime_nlp}
\end{figure*}

\begin{table*}[t]
\caption{Epochs required to reach a given target test accuracy when using no holdout data (lower is better). Final accuracy in parentheses. Results averaged across 2-3 seeds. Best performance in bold. RHO-LOSS performs best in both epochs required and final accuracy.}
\scriptsize
\centering
\begin{tabular}{@{}lrrr@{}}
    \toprule
\textbf{Dataset}               & \textbf{Target Accuracy} & \textbf{Uniform Sampling} & \textbf{RHO-LOSS}              \\
CIFAR10               & 80\%       & 39           & \textbf{17}           \\
 & 90\%       & 177 (90.8\%) & \textbf{47 (92.2\%)}  \\
CIFAR100              & 50\%       & 47           & \textbf{22}           \\
 & 65\%       & 142 (67.8\%) & \textbf{87 (68.1\%)}  \\
CINIC10               & 70\%       & 37           & \textbf{26}           \\
 & 80\%       & 146 (80.1\%)  & \textbf{70 (82.1\%)} \\
\bottomrule
\end{tabular}
\label{rholoss:tab:double_irlomo}
\end{table*}

\section{Experiment Details}
\label{rholoss:app:exp_details}
\paragraph{Architectures.} We experiment with various architectures in Figs.~\ref{rholoss:fig:clothing-1M} and \ref{rholoss:fig5} (row 4). In all other figures and tables, we use the following architectures: For experiments on QMNIST, we use a multi-layer perceptron with 2 hidden layers and 512 units in each hidden layer. For experiments on CIFAR-10, CIFAR-100 and CINIC-10, we use a variant of ResNet-18 \citep{he2016deep}. We adapted the ResNet18 to 32x32 images by modifying the architecture to remove the downsampling effect. We replaced the spatial downsampling of a strided convolution and max pooling in the original ResNet18, with a convolutional layer with 64 filters and a kernel size of 3x3. We also removed the average pooling at the end of the ResNet18. This ResNet18 variant is similar to Resnet20, just with more filters. For experiments on Clothing-1M, following the experimental set-up of \citet{Yi_2019_CVPR}, the target model is a ResNet-50 pre-trained on ImageNet. The irreducible loss model is a ResNet-18 with random initialization. The multiple target architectures in Fig~\ref{rholoss:fig5} were adapted from \cite{huyphan_2021}. For NLP datasets, we use a pre-trained ALBERT v2 \citep{lan2019albert}.

\paragraph{Hyperparameters.} \textit{Vision}: All models are trained using the AdamW optimizer with default PyTorch hyperparameters ($\beta_1$=0.9, $\beta_2$=0.999, and weight decay of $0.01$, learning rate $0.001$), a $\batchvar=32$ ($64$ for CINIC-10) $\batchvar'=320$ ($640$ for CINIC-10), meaning we select $\frac{\batchvar}{\batchvar'}=10\%$ of points. \textit{NLP}: ALBERT v2 was trained using the AdamW optimizer with a learning rate as indicated in the original paper ($2 \cdot 10^{-5}$) and weight decay of 0.02. We fine-tuned all weights, not just the final layer. The batch size $\batchvar$ was $32$, $\batchvar'=320$, meaning we select $\frac{\batchvar}{\batchvar'}=10\%$ of points. We use between 2 and 10 seeds for each experiment.

\paragraph{Data Augmentation.} On CIFAR-10, CIFAR-100, and CINIC-10, we train using data augmentation (random crop and horizontal flip), both for training the IL model, and in the main training runs. Remember that we only compute the irreducible losses once at the start of training, to save compute (Algorithm~\ref{rholoss:alg:RHOLOSS}). We use the unaugmented images for this as we found that using augmented images makes little difference to performance but costs more compute.

\paragraph{Irreducible Loss Model Training.} The irreducible loss models are trained on holdout sets, i.e.\ labeled evaluation sets (not test sets, see dataset description in main text). For each dataset, we select the irreducible loss model checkpoint from the epoch with the \textit{lowest holdout loss} on $\Dpool$ (as opposed to the highest accuracy); we find that this improves performance while also saving compute as the holdout loss typically reaches its minimum early in training.

\paragraph{BatchNorm.} Like many deep-learning methods, RHO-LOSS interacts with BatchNorm \cite{ioffe2015batch} since the loss of a given point is affected by other points in the same batch. \textbf{Important:} We compute the BatchNorm statistics for selection and model update separately. For selection (line 5-8 in Algorithm~\ref{rholoss:alg:RHOLOSS}), the statistics are computed across the large batch $B_t$. For training (line 9-10), the statistics are computed across the small batch $b_t$. These choices can affect performance a lot. For new datasets, we recommend varying how the batch-norm statistics are computed during selection (trying both train mode and evaluation mode) and choose the option that works best. 

\section{Robustness to Noise} \label{rholoss:app:noise}
\begin{figure*}[t]
    \centering
    \includegraphics[width=\textwidth]{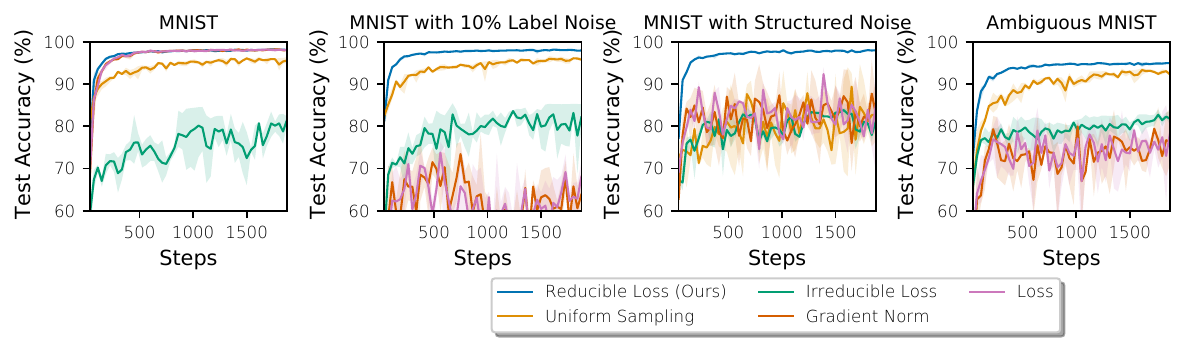}
    \caption{RHO-LOSS is robust to a variety of label noise patterns, while other selection methods degrade. A step corresponds to lines $6-11$ in Algorithm~\ref{rholoss:alg:RHOLOSS}. Lines correspond to means and shaded areas to minima and maxima across 3 random seeds.}
    \label{rholoss:fig:mnist}
\end{figure*}

In this set of experiments, we evaluate the performance of different selection methods under a variety of noise patterns on QMNIST (MNIST with extra holdout data) and variations thereof. We use this dataset because it has little label noise in its original form, allowing us to test the effect of adding noise. Firstly, we add uniform label noise to 10\% of training points. Secondly, we add structured label noise that affects easily confused classes. We follow \cite{rolnick2017deep} and flip the labels of the four most frequently confused classes (in the confusion matrix of a trained model) with 50\% probability. For example, a 2 is often confused with a 5; thus we change the label of all 2s to 5s with 50\% probability. Thirdly, we leverage the natural noise distribution of MNIST by using Ambiguous-MNIST \citep{mukhoti2021deterministic} as the training set. Ambiguous-MNIST contains a training set with 60k generated ambiguous digits that have more than one plausible label. While selecting with loss and gradient norm trains accelerates training on the MNIST training set, their performance degrades on all three types of noise distributions (Figure \ref{rholoss:fig:mnist}).

\section{Irreducible Holdout Loss Approximation}\label{rholoss:app:update_irreducible_2}

In this appendix section, we examine one of the key approximations made in the theory section. To arrive at Eq.~(\ref{rholoss:eq:rholoss}), we used the approximation $\Lof{y \given \x, \Deval} \approx \Lof{y \given \x, \Deval, \Dtrain}$. In words, we approximated the cross-entropy loss of a model trained on the data points acquired so far $\Dtrain$ and the holdout dataset $\Deval$, with the cross-entropy loss of a model trained only on the holdout set (i.e.\ the \emph{labeled} evaluation set). This approximation saves a lot of compute: rather than having to recompute the term with every change of $\mathcal{D}_t$, it is now sufficient to compute it once at the start of training.

We have already highlighted the impact of the approximation on points selected when training on QMNIST in \Cref{rholoss:sec:impact_of_approximations}. In our main experiment setting---using neural networks trained with gradient descent---we empirically find that the approximation does not reduce speed of target model training or final target model accuracy (\Cref{rholoss:tab:update_irreducible_approximation}). This finding holds across a range of datasets (CIFAR-10, CIFAR-100, CINIC-10). Updating the irreducible loss model on $\Dtrain$ seems empirically not necessary.

\begin{table}[t]
\caption{Number of epochs required to reach a given target test accuracy across several datasets. Results averaged across 2-3 random seeds. NR indicates that the target accuracy was not reached.}
\centering
\scriptsize
\begin{tabular}{@{}lrrr@{}}
\toprule
Dataset  & Target accuracy                & RHO-LOSS: Approximated JPIG & JPIG        \\
& & $\Lof{y \given \x, \Dtrain} - \Lof{y \given \x, \Deval}$ & $\Lof{y \given \x, \Dtrain} - \Lof{y \given \x, \Deval, \Dtrain}$ \\ \midrule
  & 60\% & 18                              & 13                                 \\
       CIFAR10 & 75\% & 30                              & 24                                 \\
        & 90\% & 102                             & NR, but reaches 88\% in 157 epochs \\ \midrule
 & 30\% & 35                              & 21                                 \\
    CIFAR100    & 45\% & 58                              & NR, but reaches 43\% in 61 epochs  \\
        & 60\% & 123                             & NR                                 \\ \midrule
       & 55\% & 12                              & 12                                 \\
     CINIC10    & 65\% & 19                              & 21                                 \\
        & 75\% & 32                              & NR, but reaches 74\% in 68 epochs  \\ \bottomrule
\end{tabular}

\label{rholoss:tab:update_irreducible_approximation}
\end{table}

Indeed, the approximation actually has two desirable properties when used for neural networks trained with gradient descent. We will first describe why we expect these desirable properties, and then show that they indeed appear. 
First, let us restate both selection functions:
\begin{itemize}
    \item JPIG:
\begin{align}
    \argmax_{(x,y) \in B_t}\ &\Lof{y \given x, \Dtrain} - \Lof{y \given x, \Deval, \Dtrain}; and
\end{align}
\item
approximated JPIG:
\begin{align}
\argmax_{(x,y) \in B_t}\ &\Lof{y \given x, \Dtrain} - \Lof{y \given x, \Deval}.
\end{align}
\end{itemize}

\paragraph{Desirable property 1.} \emph{The approximation prevents repeated selection of undesirable points.} When using SGD instead of Bayesian updating, the original selection function can acquire undesired points repeatedly. Let's say that we acquire, for whatever reason, a noisy, redundant, or irrelevant point. We only take one gradient step each time we acquire a (batch of) point(s), meaning the training loss (first term in the selection function) will on each only decrease somewhat. In the original selection function, the second term will also decrease somewhat, meaning that the difference between the first and second term may remain large. In the approximated selection function, the second term is constant, the difference between first and second term will thus likely decrease more than under the original selection function. Under the approximated selection function, we are thus less likely to acquire undesired points again, if we have acquired them in earlier epochs.

\paragraph{Desirable property 2.} \emph{The approximation prevents deterioration of the irreducible loss model over time.} With both selection functions, we compute the second term of the selection function with an "irreducible loss model", which we train on a holdout set (i.e.\ a \emph{labeled} evaluation set) before we start target model training. In the target model training, we (greedily) acquire the points that most improve the loss of the target model (on the holdout set, the \emph{labeled} evaluation set). We thus deliberately introduce bias into the data selection. However, this bias is tailored to the target model and may not be suitable for the irreducible loss model. As a simplifying example, consider a target model early in training, which has not yet learned a certain class, and an irreducible loss model, which has learned that class. Data points in that class will have high training loss, low irreducible loss, and will be acquired often. This, however, is not useful for the irreducible loss model, and might lead to decreased accuracy on data points from other classes. With the approximation, this can't happen. The described failure mode could likely also be alleviated by more sophisticated training schemes for the irreducible loss model, such as periodically mixing in data points from the holdout set (i.e.\ the \emph{labeled} evaluation set). However, such training schemes would require even more compute and/or overhead.

\begin{figure}[t]
    \centering
    \includegraphics[width=\linewidth]{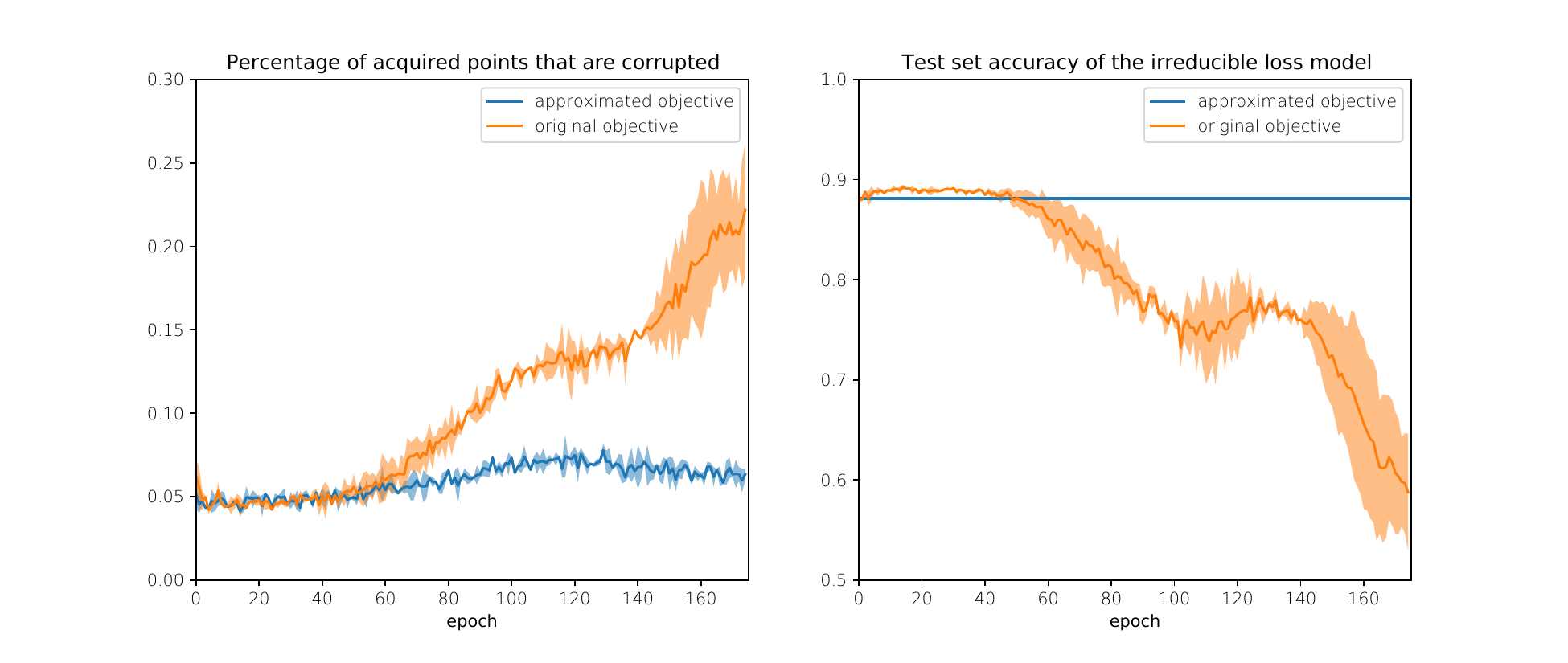}
    \caption{Desired properties of the irreducible loss model approximation. \textbf{Left.} The approximated selection function selects fewer corrupted points later on in training. \textbf{Right.} The test set accuracy of the irreducible loss model deteriorates over time if it is updated on $\mathcal{D}_t$. With the approximation, the irreducible loss is not updated during target model training. Results on CIFAR-10 with 20\% of data points corrupted with uniform label noise. Shaded areas represent standard deviation across three different random seeds. }
    \label{rholoss:fig:update_irlomo_failure_modes}
\end{figure}

We find empirically that both desired properties of the approximation indeed manifest themselves. In \Cref{rholoss:fig:update_irlomo_failure_modes}, we train a target model (Resnet-18) on CIFAR-10, with 20\% of the data points corrupted by uniform label noise. The approximated selection function leads to faster target model training (the approximated selection function needs 80 epochs to reach the same target model accuracy that the original selection function reaches in 100 epochs) and higher final accuracy than the original selection function (88.6\% vs 86.1\%). Indeed, the original selection function leads to acquiring more corrupted points, especially later in training (\Cref{rholoss:fig:update_irlomo_failure_modes}, left), and the accuracy \textit{of the irreducible loss model} deteriorates over time (\Cref{rholoss:fig:update_irlomo_failure_modes}, right). We tuned the learning rate of the irreducible loss model to 0.01 times that of the target model. Without this adjustment, the results look similar but the original selection function performs worse.

\section{Experimental Details for Assessing Impact of Approximations}\label{rholoss:sec:approximation_appendix}
\paragraph{Dataset.} QMNIST, with uniform label noise applied to 10\% of the dataset. Batch size of 1000 is used.

\paragraph{Models.} Deep Ensemble contains 5 3-layer MLP's with 512 hidden units. The weaker irreducible loss model is an MLP with 256 hidden units.

\paragraph{Training.} 
For Approximation 0, we use a deep ensemble for both models. The irreducible loss model is trained to convergence on $\Deval$. Then the target model and the irreducible model are used to acquire 10\% of points each batch using the selection function. They are then trained to convergence on each batch of points acquired. The irreducible loss model is trained on $\Deval \cup \Dtrain$, while the target model is only trained on $\Dtrain$. We train for a maximum of 5 epochs, which often is to convergence, to enable a fair comparison to further approximations.
For Approximation 1a, the deep ensembles are replaced with single MLPs. The training regime remains the same. We compare the approximations over the first epoch.
To compare Approximation 1b to 0, and for all further approximations, we increase the size of the dataset five-fold, by duplicating samples in QMNIST. This means for approximation 1b, we have 5x the data that we have for Approximation 1a, but with increased redundancy. We train the model in Approximation 1b by taking a single gradient step per data point, with the larger dataset. On the other hand, we train the model for Approximation 0 (still to convergence or 5 epochs) on the standard dataset size. By doing this, Approximation 0 and 1b have taken the equivalent number of gradient steps, at the time-steps where we are tracking the reducible loss of points selected, enabling a fair comparison between the approximations. The irreducible loss models are trained on $\Deval \cup \Dtrain$ in their respective set-ups.  
To compare Approximation 2 to Approximation 0, we compare updating the irreducible loss model with a single gradient on each set of acquired points, to not updating the irreducible loss model on $\Dtrain$ at all. To isolate effect of not updating, we utilize the same initial irreducible loss model.
To compare Approximation 3, we simply train a small irreducible model (one with 256 hidden units) and follow the same training regime as Approximation 2.

\section{Ablation of Percentage Selected}
\label{rholoss:app:selectivity}
Our method has a hyperparameter, the percentage $\frac{\batchvar}{\batchvar'}$ of evaluated points which are selected for training. In the experiments above, this parameter was set to $0.1$. We have not tuned this parameter, as we aim to analyze how well our method works ``out of the box''. In fact, on 2/3 datasets, performance further improves with other values of this parameter. Adjusting this percentage should allow practitioners to specify their preferred trade-off between training time and computation, where a low percentage typically corresponds to a lower training time and greater compute cost. For these experiments, we kept $\batchvar=32$ and adapt $\batchvar'$ accordingly. The percentage  $\frac{\batchvar}{\batchvar'}$ of data points selected per batch has different effects across datasets as shown in \Cref{rholoss:fig:percent_train}.

\begin{figure}[t]
    \centering
    \includegraphics[width=0.5\linewidth]{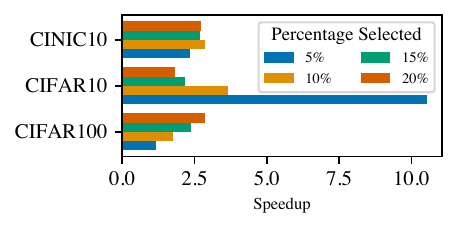}
    \caption{Varying the percent of data points selected in each training batch. Average over 3 random seeds.}
    \label{rholoss:fig:percent_train}
\end{figure}

\section{Active Learning Baselines} 
\label{rholoss:app:active_learning}
We compare our method to typical methods used in the Active Learning (AL) literature. Note that our method is label-aware, while active learning acquires data points without using labeled information. We consider the following baselines, which select the top-$k$ points using an acquisition function, $\alpha(\x)$:

\begin{itemize}%
    \item Bayesian Active Learning by Disagreement \citep{houlsby2011bayesian} with $\alpha(\x) = \Hof{y \given \x, \mathcal{D}_t} - \E{\pof{\theta \given \mathcal{D}_t}}{ \Hof{y \given \x, \theta}}$.
    \item (Average) conditional entropy, $\alpha(\x) = \E{\pof{\theta \given \mathcal{D}_t}}{ \Hof{y \given \x, \theta}}$, where the average is taken over the model parameter posterior. 
    \item (Average predictive) entropy, $\alpha(\x) = \Hof{y \given \x, \mathcal{D}_t}$.
    \item Loss minus conditional entropy $\alpha(\x) = \Lof{y \given \x, \theta} - \E{\pof{\theta \given \mathcal{D}_t}}{ \Hof{y \given \x, \theta}}$. This uses the (average) conditional entropy as an estimate of how noisy data point $\x$ is---points with high noise are deprioritized. Compared to RHO-LOSS, it replaces the IL with the conditional entropy. This acquisition function uses the label and therefore cannot be used for active learning.
\end{itemize}
We additionally compare our method to uniform sampling. We run all baselines on MNIST and CIFAR10. Note that several of these active learning baselines consider epistemic uncertainty; that is, uncertainty in predictions driven by uncertainty in the model parameters. This mandates performing (approximate) Bayesian inference. We use Monte-Carlo Dropout\citep{gal2015dropout} to perform approximate inference. For MNIST, we use an 2 hidden layer MLP with 512 hidden units per hidden layer, and a dropout probability of a 0.5. For experiments on CIFAR10, we use a small-scale CNN with dropout probability 0.05 (the dropout probability follows \citep{osawa2019practical}).

\Cref{rholoss:fig:active_learning_baselines} shows training curves for our method, uniform sampling, and the active learning baselines. Our method accelerates training across both datasets. The active learning methods accelerate training for MNIST but not for CIFAR10. This highlights that active learning methods, if naively applied to online batch selection, may not accelerate model training.   

\begin{figure}[t]
    \centering
    \includegraphics[width=\linewidth]{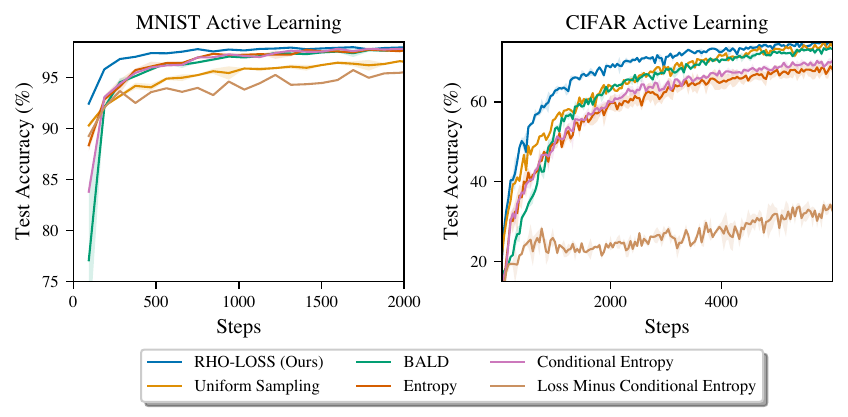}
    \caption{Training curves for several active learning baselines on the MNIST and CIFAR10 datasets.}
    \label{rholoss:fig:active_learning_baselines}
\end{figure}

\FloatBarrier
}

\chapter{\fimtitle}
{
	\graphicspath{{./fim/figs/}}

\newcommand{\Yhat}{\hat{\Y}}
\newcommand{\yhat}{\hat{\y}}

\newcommand{\Ypred}{\Y}
\newcommand{\ypred}{\y}
\newcommand{\Ytrue}{\Y}
\newcommand{\ytrue}{\y}

\newcommand{\logits}{{\hat \z}}

\newcommand{\xacq}{{\x^\text{acq}}}
\newcommand{\Yacq}{{\Y^\text{acq}}}
\newcommand{\yacq}{{\y^\text{acq}}}
\newcommand{\xacqs}{{\{\x^\text{acq}_i\}}}
\newcommand{\yacqs}{{\{\y^\text{acq}_i\}}}
\newcommand{\Yacqs}{{\{\Y^\text{acq}_i\}}}

\newcommand{\yacqsstar}{{\{\y^\text{acq,*}_i\}}}

\newcommand{\xs}{{\{\x_i\}}}
\newcommand{\ys}{{\{\y_i\}}}
\newcommand{\Ys}{{\{\Y_i\}}}

\newcommand{\Xeval}{{X^\text{eval}}}
\newcommand{\xeval}{{\x^\text{eval}}}
\newcommand{\Yeval}{{\Y^\text{eval}}}
\newcommand{\yeval}{{\y^\text{eval}}}

\newcommand{\xevals}{{\{\x^\text{eval}_i\}}}
\newcommand{\yevals}{{\{\y^\text{eval}_i\}}}
\newcommand{\Yevals}{{\{\Y^\text{eval}_i\}}}

\newcommand{\xtrain}{{\x^\text{train}}}
\newcommand{\ytrain}{{\y^\text{train}}}
\newcommand{\Ytrain}{{\Y^\text{train}}}

\newcommand{\xtrains}{{\{\x^\text{train}_i\}}}

\newcommand{\w}{\omega}
\newcommand{\W}{\Omega}
\newcommand{\wstar}{\w^*}

\newcommand{\realnum}{\mathbb{R}}

\newcommand{\verteq}{\rotatebox{90}{$\,=$}}
\newcommand{\equalto}[2]{\underset{\displaystyle\overset{\mkern4mu\verteq}{#2}}{#1}}

\newcommand{\encoder}[1]{{\hat f(#1)}}
\newcommand{\realencoder}[1]{{f (#1)}}
\newcommand{\embedding}{\z}

\newcommand{\similarityMatrix}[2]{{S_{#1}[#2]}}
\newcommand{\HofJacobianData}[1]{{\hat{\opEntropy}'[#1]}}
\newcommand{\HofJacobianDataShort}{\hat{\opEntropy}'}
	\section{Fisher Information: Additional Derivations \&\ Proofs}
\label{fim:appsec:fisher_information}

\fimadditive*
\begin{proof}
    This follows immediately from $Y_i \independent{} Y_j \given \x_i, \x_j, \wstar$ for $i \not= j$ and the additivity of the observed information:
    \begin{align}
        \FisherInfo{\Ys \given \xs, \wstar} &= \E{\pof{\ys \given \xs, \wstar}}{\HofHessian{\ys \given \xs, \wstar}}
        = \E{\pof{\ys \given \xs, \wstar}}{ \sum_i \HofHessian{y_i \given x_i, \wstar}} \\
        &= \sum_i \E{\pof{y_i \given x_i, \wstar}}{ \HofHessian{y_i \given x_i, \wstar}} 
        = \sum_i \FisherInfo{y_i \given x_i, \wstar}.
    \end{align}
\end{proof}

\fisherinformationequivalences*
To prove \Cref{fim:prop:fisher_information_equivalences}, we use the two generally useful lemmas below:
\begin{restatable}{lemma}{helperjacobianhessianequivalent}
    \label{fim:lemma:helper_jacobian_hessian_equivalent}
    For the Jacobian $\HofJacobian{\y \given \x, \wstar}$, we have:
    \begin{align}
        \HofJacobian{y \given x, \wstar} = \nabla_\w[-\log \pof{y \given x, \wstar}] = -\frac{\nabla_\w \pof{y \given x, \wstar}}{\pof{y \given x,\wstar}},
    \end{align}
    and for the Hessian $\HofHessian{\y \given \x, \wstar}$, we have:
    \begin{align}
        \HofHessian{y \given x, \wstar} = \HofJacobian{y \given x, \wstar}^T \, \HofJacobian{y \given x, \wstar} - \frac{\nabla_\w^2 \pof{y \given x, \wstar}}{\pof{y \given x, \wstar}}.
    \end{align}
\end{restatable}
\begin{proof}
    The result follows immediately from the application of the rules of multivariate calculus.
\end{proof}
\begin{restatable}{lemma}{helpervanishingexpectations}
    \label{fim:lemma:helper_vanishing_expectations}
    The following expectations over the model's own predictions vanish:
    \begin{align}
        \E{\pof{y\given x, \wstar}}{\HofJacobian{y \given x, \wstar}} = 0, \\
        \E*{\pof{y\given x, \wstar}}{\frac{\nabla_\w^2 \pof{y \given x, \wstar}}{\pof{y \given x,\wstar}}} = 0.
    \end{align}
\end{restatable}
\begin{proof}
    We use the previous equivalences and rewrite the expectations as integral; the results follows:
    \begin{align}
        &\E{\pof{y\given x, \wstar}}{\HofJacobian{y \given x, \wstar}} = \E{\pof{y\given x, \wstar}}{-\nabla_\w \log \pof{y \given x, \wstar}} 
        \\
        &\quad = -\E*{\pof{y\given x, \wstar}}{\frac{\nabla_\w \pof{y \given x, \wstar}}{\pof{y \given x,\wstar}}} \notag\\
        &\quad = -\int \nabla_\w \pof{y \given x, \wstar} \, dy 
        = -\nabla_\w  \int \pof{y \given x, \wstar} \, dy  
        = -\nabla_\w 1 = 0,\\
        &\E*{\pof{y\given x, \wstar}}{\frac{\nabla_\w^2 \pof{y \given x, \wstar}}{\pof{y \given x,\wstar}}} 
        = \int \nabla_\w^2 \pof{y \given x, \wstar} \, dy \\
        &\quad = \nabla_\w^2 \int \pof{y \given x, \wstar} \, dy \\
        &\quad = \nabla_\w^2 1 = 0.
    \end{align}
\end{proof}

\begin{proof}[Proof of \Cref{fim:prop:fisher_information_equivalences}]
    With the previous lemma, we have the following.
    \begin{align}
        \implicitCov{\HofJacobian{Y \given x, \wstar}} &= \implicitE{\HofJacobian{Y \given x, \wstar}^T \, \HofJacobian{Y \given x, \wstar}} x \\
        &\quad - \underbrace{\implicitE{\HofJacobian{Y \given x, \wstar}^T}}_{=0} \, \underbrace{\implicitE{\HofJacobian{Y \given x, \wstar}}}_{=0} \notag \\
        &= \implicitE{\HofJacobian{Y \given x, \wstar}^T \, \HofJacobian{Y \given x, \wstar}}.
    \end{align}
    For the expectation over the Hessian, we plug \Cref{fim:lemma:helper_jacobian_hessian_equivalent} into \Cref{fim:lemma:helper_vanishing_expectations} and obtain:
    \begin{align}
        \FisherInfo{Y \given x, \wstar} &= \E{\pof{y \given x, \wstar}}{ \HofHessian{y \given x, \wstar}} \\
        &= \E*{\pof{y \given x, \wstar}}{\HofJacobian{y \given x, \wstar}^T \, \HofJacobian{y \given x, \wstar} - \frac{\nabla_\w^2 \pof{y \given x, \wstar}}{\pof{y \given x, \wstar}}} \\
        &= \E{\pof{y \given x, \wstar}}{\HofJacobian{y \given x, \wstar}^T \, \HofJacobian{y \given x, \wstar}} - 0 \\
        &= \implicitCov{\HofJacobian{Y \given x, \wstar}}.        
    \end{align}
\end{proof}

\subsection{Special Case: Exponential Family}

\fimexpfamily*
\begin{proof}
    We apply the second equivalence in \Cref{fim:prop:fisher_information_equivalences} twice:
    \begin{align}
        \FisherInfo{Y \given x, \wstar} &=\implicitCov{\HofJacobian{Y \given x, \wstar}}
        \\
        &=\implicitCov{\nabla_\w \encoder{x ; \wstar}^T \, \nabla_\logits \Hof{y \given \logits=\encoder{x ; \wstar}} \, \nabla_\w \encoder{x ; \wstar}} \\
        &=\nabla_\w \encoder{x ; \wstar}^T \, \implicitCov{\nabla_\logits \Hof{y \given \logits=\encoder{x ; \wstar}}} \, \nabla_\w \encoder{x ; \wstar} \\
        &=\nabla_\w \encoder{x ; \wstar}^T \, \E{\pof{y \given x, \wstar}}{\nabla_\logits^2 \Hof{y \given \logits=\encoder{x ; \wstar}}} \, \nabla_\w \encoder{x ; \wstar}
    \end{align}
\end{proof}

\subsection{Special Case: Generalized Linear Models}

\glmhessian*
\begin{proof}
    \begin{align}
        \HofHessian{y \given x, \wstar} \\
        &= \nabla_\w [ \HofJacobian{y \given x, \wstar} ] \\
        &= \nabla_\w [ \nabla_\logits \Hof{y \given \logits=\encoder{x ; \wstar}} \, \nabla_\w \encoder{x ; \wstar} ] \\
        &= \nabla_\logits \Hof{y \given \logits=\encoder{x ; \wstar}} \, \underbrace{\nabla_\w^2 \encoder{x ; \wstar}}_{=\nabla_\w^2 [w^T x]=0} \\
        &\quad + \nabla_\w \encoder{x ; \wstar}^T \, \nabla_\logits^2 \Hof{y \given \logits=\encoder{x ; \wstar}} \, \nabla_\w \encoder{x ; \wstar} \notag \\
        &= \nabla_\w \encoder{x ; \wstar}^T \, \nabla_\logits^2 A(w^T x) \, \nabla_\w \encoder{x ; \wstar}.
    \end{align}
\end{proof}

\propkroneckerproductfisher*
\begin{proof}
    We begin with a few statements that lead to the conclusion step by step, where $\x \in \realnum^D, A \in \realnum^{C \times C}, G \in \realnum^{C \times (C \cdot D)}$:
    \begin{align}
        (x \, x^T)_{ij} &= x_i \, x_j\\ %
        (Id_C \otimes x^T)_{c, d \, D + i} &= x_i \cdot \indicator{c=d} \\ %
        (G^T \, A \, G)_{ij} &= \sum_{k,l} G_{ki} \, A_{kl} \, G_{lj} \\ %
        (A \otimes x\, x^T)_{c\,D + i, d\,D + j} &= A_{cd} \, x_i \, x_j, \\
        ((Id_C \otimes x^T)^T \, A \, (Id_C \otimes x^T))_{c\,D+i,d\,D+j} &= 
        \sum_{k,l} (Id_C \otimes x^T)_{k,c\,D + i} \, A_{kl} \, (Id_C \otimes x^T)_{l, d\,D + j} \\
        &= \sum_{k,l} x_i \cdot \indicator{k=c} \, A_{kl} \, x_j \cdot \indicator{l=d} \\
        &= x_i \, A_{cd} \, x_j \\
        &= (A \otimes x\, x^T)_{c\,D + i, d\,D + j}. \\
    \implies \nabla_\w \encoder{x ; \wstar}^T \, \nabla_\logits^2 A(\w^T x) \, \nabla_\w \encoder{x ; \wstar} &= \nabla_\logits^2 A(w^T x) \otimes x \, x^T.
    \end{align}
\end{proof}

\glmhessianfim*
\begin{proof}
    This follows directly from \Cref{fim:prop:special_case_exp_family_fim}. In particular, we have:
    \begin{align}
        \FisherInfo{Y \given x, \wstar} &= \E{\pof{y \given x, \wstar}} {\HofHessian{y\given x, \wstar}} = \HofHessian{y^*\given x, \wstar},
    \end{align}
    where we have fixed $\y^*$ to an arbitrary value.
\end{proof}

\section{Approximating Information Quantities}
\label{fim:appsec:approximate_iq}

\subsection{Approximate Expected Information Gain}
\label{fim:appsec:eig_general_case}

\logdettrinequality*
\begin{proof}
    When $A$ is positive semi-definite and symmetric, its eigenvalues $(\lambda_i)_i$ are real and non-negative.
    Moreover, $A+Id$ has eigenvalues $(\lambda_i + 1)_i$; $\det (A + Id) = \prod_i (\lambda_i + 1)$; and $\tr A = \sum_i \lambda_i$. These properties easily follow from the respective eigenvalue decomposition.
    Thus, we have:
    \begin{align}
        \log \det (A + Id) \le \log \prod_i (\lambda_i + 1) = \sum_i \log (\lambda_i + 1) 
        \le \sum_i \lambda_i = \tr(A),
    \end{align}
    where we have used $\log (x + 1) \le \x$ iff equality for $\x=0$.
\end{proof}

\textbf{General Case.} In the main text, we only skimmed the general case and mentioned the main assumption. Here, we look at the general case in detail.

For the general case, we need to make strong approximations to be able to pursue a similar derivation. 
First, we cannot drop the expectation; instead, we note that the log determinant is a concave function on the positive semi-definite symmetric cone \citep{cover1988determinant}, and we can use Jensen's inequality on the log determinant term from \Cref{fim:eq:eig_log_det} as follows:
\begin{align}
    &\E{\pof{\yacqs \given \xacqs}}{\log \det \left ( \HofHessian{\yacqs \given \xacqs, \wstar} \, \HofHessian{\wstar}^{-1} + Id \right )} 
    \\&\quad \le \log \det \left ( \E{\pof{\yacqs \given \xacqs}}{\HofHessian{\yacqs \given \xacqs, \wstar}} \, \HofHessian{\wstar}^{-1} + Id\right ) \label{fim:eq:log_det_concavity}.
\end{align}

Second, we need to use the following approximation:
\begin{align}
    \pof{\yacqs \given \xacqs} \approx \pof{\yacqs \given \xacqs, \wstar}
\end{align}
to obtain a Fisher information and use its additivity. That is, we obtain:
\begin{align}
    \E{\pof{\yacqs \given \xacqs, \wstar}}{\HofHessian{\yacqs \given \xacqs, \wstar}} &= \FisherInfo{\Yacqs \given \xacqs, \wstar} \\
    &= \sum_i \FisherInfo{\yacq_i \given \xacq_i, \wstar}.
\end{align}

Plugging all of this together and applying \Cref{fim:lemma:log_det_tr_inequality}, we obtain the same final approximation:
\begin{align}
    &\MIof{\W; \Yacqs \given \xacqs} = \ldots \\
    &\quad \approx \tfrac{1}{2}\E{\pof{\yacqs \given \xacqs}}{\log \det \left ( \HofHessian{\yacqs \given \xacqs, \wstar} \, \HofHessian{\wstar}^{-1} + Id \right )} \notag \\
    &\quad \le \tfrac{1}{2} \log \det \left ( \E{\pof{\yacqs \given \xacqs}}{\HofHessian{\yacqs \given \xacqs, \wstar}} \, \HofHessian{\wstar}^{-1} + Id\right ) \\
    &\quad \approx \tfrac{1}{2} \log \det \left ( \E{\pof{\yacqs \given \xacqs, \wstar}}{\HofHessian{\yacqs \given \xacqs, \wstar}} \, \HofHessian{\wstar}^{-1} + Id\right ) \\
    &\quad = \tfrac{1}{2} \log \det \left (  \sum_i \FisherInfo{\Yacq_i \given \xacq_i, \wstar} \, \HofHessian{\wstar}^{-1} + Id\right ) \\
    &\quad \le \tfrac{1}{2} \sum_i \tr \left ( \FisherInfo{\Yacq_i \given \xacq_i, \wstar} \, \HofHessian{\wstar}^{-1} \right ).
\end{align}
Unlike in the case of generalized linear models, a stronger assumption was necessary to reach the same result. Alternatively, we could use the GGN approximation, which leads to the same result.

\subsection{Approximate Expected Predicted Information Gain}
\label{fim:appsec:epig_general_case}

In the main text, we only briefly referred to not knowing a principled way to arrive at the same result of \Cref{fim:prop:epig_fisher_approximation} for the general case. This is because unlike the expected information gain, the Fisher information for an acquisition candidate now lies within a matrix inversion. 
Even if we used the fact that $\log \det (Id + X \, Y^{-1})$ is concave in $X$ and convex in $Y$, we would end up with:
\begin{align}
    &\cdots \notag \\
    &\quad \approx \tfrac{1}{2} \simpleE{\pof{\yeval, \yacq \given \xeval, \xacq} \, \pof{\xeval}}
    \begin{aligned}[t]
        [\log \det
        (
        &\HofHessian{\yeval \given \xeval, \wstar} \, ( \HofHessian{\yacq \given \xacq, \wstar} \\
        &+ \HofHessian{\wstar} )^{-1} + Id
        )]
    \end{aligned} \\
    &\quad \le 
    \begin{aligned}[t]
        \tfrac{1}{2} \simpleE{\pof{\yacq \given \xacq}} [\log \det
        (
        \simpleE{\pof{\yeval, \xeval}}[\HofHessian{\yeval \given \xeval, \wstar}] \, (
            &\HofHessian{\yacq \given \xacq, \wstar} + \HofHessian{\wstar} )^{-1} \\
            & + Id
        ) ]
    \end{aligned} \\
    &\quad \ge 
    \begin{aligned}[t]
        \tfrac{1}{2} \log \det
        (
        \simpleE{\pof{\yeval, \xeval}} [\HofHessian{\yeval \given \xeval, \wstar}] \, (
            & \simpleE{\pof{\yacq \given \xacq}}[\HofHessian{\yacq \given \xacq, \wstar}] + \HofHessian{\wstar} )^{-1} \\
            & + Id
        )
    \end{aligned} \\
    &\quad = \tfrac{1}{2} {\log \det
    (
    \E{\pof{\xeval}}{\FisherInfo{\Yeval \given \xeval, \wstar}} \, ( \FisherInfo{\Yacq \given \xacq, \wstar} + \HofHessian{\wstar} )^{-1} + Id
    )} \\
    &\quad \le \tfrac{1}{2} {\tr
    (
    \E{\pof{\xeval}}{\FisherInfo{\Yeval \given \xeval, \wstar}} \, ( \FisherInfo{\Yacq \given \xacq, \wstar} + \HofHessian{\wstar} )^{-1}
    )}.
\end{align}
Note the $\le \ldots \ge$, which invalidates the chain. The errors could cancel out, but a principled statement hardly seems possible using this deduction.

\subsection{Approximate Predictive Information Gain}
\label{fim:appsubsec:pig}

Similarly to \Cref{fim:prop:approximate_ig}, we can approximate the predictive information gain. We assume that we have access to an (empirical) distribution $\pdataof{\xeval, \yeval}$:
\begin{importantresult}
\begin{proposition}
    \label{fim:prop:pig_fisher_approximation}
    For a generalized linear model (or with the GGN approximation), when we take the expectation over $\pdataof{\xeval, \yeval}$, we have:
    \begin{align}
        \argmax_\xacq \MIof{\Yeval; \yacq \given \Xeval, \xacq, \Dtrain} = 
        \argmin_\xacq \MIof{\W ; \Yeval \given \Xeval, \yacq, \xacq, \Dtrain}
    \end{align}
    with
    \begin{align}
        &\MIof{\W ; \Yeval \given \Xeval, \yacq, \xacq, \Dtrain} \\
        &\quad = \simpleE{\pdataof{\xeval, \yeval}} \MIof{\W ; \yeval \given \xeval, \yacq, \xacq, \Dtrain} \notag \\
        &\quad 
        \approx
        \begin{aligned}[t]
            &\simpleE{\pdataof{\xeval, \yeval}}  [\tfrac{1}{2} \log \det
            ( \\
            &\quad
            \HofHessian{\yeval \given \xeval, \wstar} \, ( \HofHessian{\yacq \given \xacq, \wstar} + \HofHessian{\wstar \given \Dtrain} )^{-1} + Id
            )]
        \end{aligned}
        \\
        &\quad \le 
        \begin{aligned}[t]
            \tfrac{1}{2} \log \det
            (
            &\simpleE{\pdataof{\xeval, \yeval}} [
            \HofHessian{\yeval \given \xeval, \wstar}] \\
            & \quad ( \HofHessian{\yacq \given \xacq, \wstar} + \HofHessian{\wstar \given \Dtrain} )^{-1} + Id
            )
        \end{aligned} \\
        &\quad \le \tfrac{1}{2} {\tr
        \left (
        \E{\pdataof{\xeval, \yeval}} {\HofHessian{\yeval \given \xeval, \wstar}} \, ( \HofHessian{\yacq \given \xacq, \wstar} + \HofHessian{\wstar \given \Dtrain} )^{-1}
        \right )}.
    \end{align}
\end{proposition}
\end{importantresult}
All of this follows immediately. Only for the second inequality, we need to use Jensen's inequality and that the log determinant is on the positive semi-definite symmetric cone \citep{cover1988determinant}. Like for the information gain, there is no difference between having access to labels or not when we have a GLM or use the GGN approximation.

\subsection{Approximate Joint (Expected) Predictive Information Gain}
\label{fim:appsubsec:jepig_jpig}

A comparison of EPIG and JEPIG shows that JEPIG does not require an expectation over $\pdataof{\xeval}$ but uses a set of \emph{evaluation samples} $\xevals$. As such, we can easily adapt \Cref{fim:prop:epig_fisher_approximation} to JEPIG and obtain:

\begin{importantresult}
\begin{proposition}[JEPIG]
    \label{fim:prop:jepig_fisher_approximation}
    For a generalized linear model (or with the GGN approximation), we have:
    \begin{align}
        &\argmax_\xacqs \MIof{\Yevals; \Yacqs \given \xevals, \xacqs, \Dtrain} \\
        &\quad= 
        \argmin_\xacqs \MIof{\W ; \Yevals \given \xevals, \Yacqs, \xacqs, \Dtrain}
    \end{align}
    with
    \begin{align}
        &\MIof{\W ; \Yevals \given \xevals, \Yacqs, \xacqs, \Dtrain} \notag \\
        &\quad \medmath{
        \approx
        \tfrac{1}{2} {\log \det
        \left (
        {\FisherInfo{\Yevals \given \xevals, \wstar}} \, ( \FisherInfo{\Yacqs \given \xacqs, \wstar} + \HofHessian{\wstar \given \Dtrain} )^{-1} + Id
        \right )}} \label{fim:eq:jepig_log_det_approx} \\
        &\quad \le \tfrac{1}{2} {\tr
        \left (
        {\FisherInfo{\Yevals \given \xevals, \wstar}} \, ( \FisherInfo{\Yacqs \given \xacqs, \wstar} + \HofHessian{\wstar \given \Dtrain} )^{-1}
        \right )}.
    \end{align}
\end{proposition}
\end{importantresult}

Similarly, for JPIG, we obtain without relying on the GGN approximation or GLMs:
\begin{importantresult}
\begin{proposition}[JPIG]
    \label{fim:prop:jpig_fisher_approximation}
    We have:
    \begin{align}
        &\argmax_\xacqs \MIof{\yevals; \yacqs \given \xevals, \xacqs, \Dtrain} \\
        &\quad =
        \argmin_\xacqs \MIof{\W ; \yevals \given \xevals, \yacqs, \xacqs, \Dtrain}
    \end{align}
    with
    \begin{align}
        &\MIof{\W ; \yevals \given \xevals, \Yacqs, \xacqs, \Dtrain} \notag \\
        &\quad
        \approx \medmath{
        \tfrac{1}{2} {\log \det
        \left (
        {\HofHessian{\yevals \given \xevals, \wstar}} \, ( \HofHessian{\yacqs \given \xacqs, \wstar} + \HofHessian{\wstar \given \Dtrain} )^{-1} + Id
        \right )}} \\
        &\quad \le \tfrac{1}{2} {\tr
        \left (
        {\HofHessian{\yevals \given \xevals, \wstar}} \, ( \HofHessian{\yacqs \given \xacqs, \wstar} + \HofHessian{\wstar \given \Dtrain} )^{-1}
        \right )}.
    \end{align}
\end{proposition}
\end{importantresult}

\textbf{Comparison between (E)PIG and J(E)PIG approximations.} As observed information and Fisher information are additive, the difference between the approximations when we have an empirical, that is finite, evaluation distribution $\pdataof{\xeval}$ with $M$ samples is a factor of $\evalsize$ inside the log determinant or trace:
\begin{align}
    \E{\pdataof{\xeval, \yeval}} {\HofHessian{\yeval \given \xeval, \wstar}} &= \frac{1}{\evalsize} \sum_i {\HofHessian{\yeval_i \given \xeval_i, \wstar}} \\
    &= \frac{1}{\evalsize} \HofHessian{\yevals \given \xevals, \wstar}.
\end{align}
For the log determinant, $\frac{1}{\evalsize} \log \det (A + Id) \not= \log \det (\frac{1}{\evalsize} A + Id)$, but for the trace approximation, we see that both approximations are equal up to a constant factor. For example:
\begin{align}
    &\tfrac{1}{2} {\tr
        (
        \E*{\pdataof{\xeval}} {\FisherInfo{\Yeval \given \xeval, \wstar}} \, ( \FisherInfo{\Yacqs \given \xacqs, \wstar} + \HofHessian{\wstar \given \Dtrain} )^{-1}
        )
    } \\
    &\quad = 
    \tfrac{1}{2} {\tr
        (
        \frac{1}{\evalsize} {\FisherInfo{\Yevals \given \xevals, \wstar}} \, ( \FisherInfo{\Yacqs \given \xacqs, \wstar} + \HofHessian{\wstar \given \Dtrain} )^{-1}
        )
    } \\
    &\quad =
    \tfrac{1}{2\,\evalsize} {\tr
        (
        {\FisherInfo{\Yevals \given \xevals, \wstar}} \, ( \FisherInfo{\Yacqs \given \xacqs, \wstar} + \HofHessian{\wstar \given \Dtrain} )^{-1}
        )
    }.
\end{align}

\section{Similarity Matrices and One-Sample Approximations}
\label{fim:appsec:similarity_matrices}

\propsimilaritybasedIG*
\begin{proof}
    \begin{align}
        &\MIof{\W; \Yacqs \given \xacqs, \Dtrain} \overset{\approx}{\le} \tfrac{1}{2} {\log \det \left ( \FisherInfo{\Yacqs \given \xacqs, \wstar} \, \HofHessian{\wstar \given \Dtrain}^{-1} + Id \right )} \\
        &\quad= \tfrac{1}{2} {\log \det \left ( (\FisherInfo{\Yacqs \given \xacqs, \wstar} + \HofHessian{\wstar \given \Dtrain}) \, \HofHessian{\wstar \given \Dtrain}^{-1} \right )} \\
        &\quad\approx \tfrac{1}{2} {\log \det \left ( ( \HofJacobianData{\Dacq \given \wstar}^T \HofJacobianData{\Dacq \given \wstar} + \HofHessian{\wstar \given \Dtrain}) \, \HofHessian{\wstar \given \Dtrain}^{-1} \right )} \\
        &\quad= \tfrac{1}{2} {\log \det \left ( \HofJacobianData{\Dacq \given \wstar} \HofHessian{\wstar \given \Dtrain}^{-1} \HofJacobianData{\Dacq \given \wstar}^T  + Id \right )} \\
        &\quad= \tfrac{1}{2} {\log \det \left ( \similarityMatrix{\HofHessian{\wstar \given \Dtrain}}{ \Dacq \given \wstar}  + Id \right )},
    \intertext{
    where we have used the matrix determinant lemma:
    }
        \det (A B + M) &= \det (B M^{-1} A + Id) \, \det M. 
    \end{align}
\end{proof}

\textbf{Connection to the Joint (Expected) Predictive Information Gain.} %
Following \cref{fim:eq:epig_bald_decomposition}, JEPIG can be decomposed as the difference between two EIG terms, which we can further divide into three terms that are only conditioned on $\Dtrain$:
\begin{align}
    & \MIof{\Yevals; \Yacq \given \xevals, \xacq, \Dtrain} \\
    &\quad = \MIof{\W; \Yevals \given \xevals, \Dtrain} - \MIof{\W ; \Yevals \given \xevals, \Yacq, \xacq, \Dtrain} \notag \\
    &\quad = 
    \begin{aligned}[t]
        &\MIof{\W; \Yevals \given \xevals, \Dtrain} - \MIof{\W ; \Yevals, \Yacq \given \xevals, \xacq, \Dtrain} 
        \\
        &+ 
        \MIof{\W ; \Yacq \given \xacq, \Dtrain}
    \end{aligned}
\end{align}
Using \Cref{fim:prop:similarity_based_IG}, we can approximate this as:
\begin{align}
    & \MIof{\Yevals; \Yacq \given \xevals, \xacq, \Dtrain} \\
    & \approx \tfrac{1}{2} {\log \det \left ( \similarityMatrix{\HofHessian{\wstar \given \Dtrain}}{ \Deval \given \wstar}  + Id \right )}
    -
    \tfrac{1}{2} {\log \det \left ( \similarityMatrix{\HofHessian{\wstar \given \Dtrain}}{ \Dacq, \Deval \given \wstar} + Id \right )} \\
    &\quad \quad 
    + \tfrac{1}{2} {\log \det \left ( \similarityMatrix{\HofHessian{\wstar \given \Dtrain}}{ \Dacq \given \wstar}  + Id \right )}
\end{align}

Furthermore, we can apply the approximation in \Cref{fim:prop:similar_uninformative_posterior} and find that the $\log \lambda$ terms cancel because $\lvert \Dacq \rvert + \lvert \Dtrain \rvert = \lvert \Dacq \cup \Dtrain \rvert$. Taking the limit $\lambda \to 0$, we obtain:
\begin{align}
    & \MIof{\Yevals; \Yacq \given \xevals, \xacq, \Dtrain} \\
    &\quad \approx 
    \begin{aligned}[t]
        &\tfrac{1}{2} \log \det \left (  \similarityMatrix{}{ \Deval \given \wstar} + \lambda Id \right )
        -
        \tfrac{1}{2} \log \det \left (  \similarityMatrix{}{ \Dacq, \Deval \given \wstar} + \lambda Id \right ) \\
        &+
        \tfrac{1}{2} \log \det \left (  \similarityMatrix{}{ \Dacq \given \wstar}  + \lambda Id \right )
    \end{aligned} \\
    &\quad \to 
    \tfrac{1}{2} \log \det \left (  \similarityMatrix{}{ \Deval \given \wstar} \right )
    -
    \tfrac{1}{2} \log \det \left (  \similarityMatrix{}{ \Dacq, \Deval \given \wstar} \right )
    +
    \tfrac{1}{2} \log \det \left (  \similarityMatrix{}{ \Dacq \given \wstar} \right ).
\end{align}
Finally, the first term is independent of $\Dacq$, and if we are interested in approximately maximizing JEPIG, we can maximize as proxy objective:
\begin{align}
    {\log \det \left ( \similarityMatrix{\HofHessian{\wstar \given \Dtrain}}{ \Dacq \given \wstar}  + Id \right )}
    &-
    {\log \det \left ( \similarityMatrix{\HofHessian{\wstar \given \Dtrain}}{ \Dacq, \Deval \given \wstar} + Id \right )},
    \intertext{or}
    \log \det \left (  \similarityMatrix{}{ \Dacq \given \wstar} \right ) 
    &-
    \log \det \left (  \similarityMatrix{}{ \Dacq, \Deval \given \wstar} \right ). \label{fim:eq:similar_jepig_approx_strong}
\end{align}

\section{Connection to Other Acquisition Functions in the Literature}
\label{fim:appsec:unification}

\subsection{SIMILAR \citep{kothawade2021similar} and PRISM \citep{kothawade2022prism}}
\label{fim:appsec:unify_similar}

\textbf{Connection} \textbf{to LogDetMI.} If we apply the Schur decomposition to $\log \det \linebreak[0] \similarityMatrix{}{\Dacq, \Deval \given \wstar}$ from \cref{fim:eq:similar_jepig_approx_strong}, we obtain the following:
\begin{align}
    &\log \det \similarityMatrix{}{\Dacq \given \wstar} - \log \det \similarityMatrix{}{\Dacq, \Deval \given \wstar} \\
    &\quad =\log \det \similarityMatrix{}{\Dacq \given \wstar} -  \log \det \similarityMatrix{}{\Deval \given \wstar} \\
    &\quad \quad - \log \det (
        \similarityMatrix{}{\Dacq \given \wstar} 
        -
        \similarityMatrix{}{\Dacq ; \Deval \given \wstar} 
        \similarityMatrix{}{\Deval \given \wstar} ^{-1}
        \similarityMatrix{}{\Deval ; \Dacq \given \wstar} 
    ), \notag
\end{align}
where $\similarityMatrix{}{\Dacq ; \Deval \given \wstar}$ is the non-symmetric similarity matrix between $\Dacq$ and $\Deval$ etc.

Dropping $\log \det \similarityMatrix{}{\Deval \given \wstar}$ which is independent of $\Dacq$, we can instead maximize:
\begin{align}
    \log \det \similarityMatrix{}{\Dacq \given \wstar} - \log \det (
        &\similarityMatrix{}{\Dacq \given \wstar} \\
        &-
        \similarityMatrix{}{\Dacq ; \Deval \given \wstar} 
        \similarityMatrix{}{\Deval \given \wstar} ^{-1}
        \similarityMatrix{}{\Deval ; \Dacq \given \wstar},
    ) \notag
\end{align}
which is exactly the LogDetMI objective of SIMILAR \citep{kothawade2021similar} and PRISM \citep{kothawade2022prism}.

We can further rewrite this objective by extracting $\similarityMatrix{}{\Dacq \given \wstar}$ from the second term, obtaining:
\begin{align}
    & 
    \begin{aligned}[t]
        \log \det \similarityMatrix{}{\Dacq \given \wstar} - \log \det (
        &\similarityMatrix{}{\Dacq \given \wstar} \\
        &-
        \similarityMatrix{}{\Dacq ; \Deval \given \wstar} 
        \similarityMatrix{}{\Deval \given \wstar} ^{-1}
        \similarityMatrix{}{\Deval ; \Dacq \given \wstar} 
        ) 
    \end{aligned}
    \\
    & \quad = -\log \det (
        Id
        -
        \similarityMatrix{}{\Dacq \given \wstar} ^{-1}
        \similarityMatrix{}{\Dacq ; \Deval \given \wstar} 
        \similarityMatrix{}{\Deval \given \wstar} ^{-1}
        \similarityMatrix{}{\Deval ; \Dacq \given \wstar} 
    ). \label{fim:eq:extract_acq_det_identity}
\end{align}

\textbf{Connection to LogDetCMI.} Using information-theoretic decompositions, it is easy to show that:
\begin{align}
    &\MIof{\Yevals; \Yacq \given \xevals, \xacq, \Ys, \xs, \Dtrain} \\
    &\quad = 
    \MIof{\Yevals; \Yacq, \Ys \given \xevals, \xacq, \xs, \Dtrain}
    - \MIof{\Yevals; \Ys \given \xs, \Dtrain}.
\end{align}
These are two JEPIG terms, and using above approximations, including \eqref{fim:eq:extract_acq_det_identity}, leads to the LogDetCMI objective:
\begin{align}
    \log \frac
    {
        \det (
            Id
            -
            \similarityMatrix{}{\Dacq \given \wstar} ^{-1}
            \similarityMatrix{}{\Dacq ; \Deval \given \wstar} 
            \similarityMatrix{}{\Deval \given \wstar} ^{-1}
            \similarityMatrix{}{\Deval ; \Dacq \given \wstar} 
        )
    }
    {
        \det (
        Id
        -
        \similarityMatrix{}{\Dacq, \Dany \given \wstar} ^{-1}
        \similarityMatrix{}{\Dacq, \Dany ; \Deval \given \wstar} 
        \similarityMatrix{}{\Deval \given \wstar} ^{-1}
        \similarityMatrix{}{\Deval ; \Dacq, \Dany \given \wstar} 
    )
    }.
\end{align}

\subsection{Expected Gradient Length}
\label{fim:appsec:unify_egl}
\propegleigconnection*
\begin{proof}
    The EIG is equal to the conditional entropy up to a constant term, via \cref{fim:eq:approx_conditional_entropy} in \Cref{fim:prop:approximate_eig}:
    \begin{align}
        \MIof{\W; \Yacq \given \xacq} &\overset{\approx}{\le}
        \tfrac{1}{2} \log \det \left ( \FisherInfo{\Yacq \given \xacq, \wstar} + \HofHessian{\wstar \given \Dtrain} \right ) + \text{const.}
    \intertext{We apply a diagonal approximation for the Fisher information and Hessian, noting that the determinant of the diagonal matrix upper-bounds the determinant of the full matrix:}
        &\le \tfrac{1}{2} \log \det \left ( \specialFisherInfo{_{diag}}{\Yacq \given \xacq, \wstar} + \specialHofHessian{_{diag}}{\wstar \given \Dtrain} \right ) + \text{const.} \\
        &= \tfrac{1}{2} \sum_k \log \left ( \specialFisherInfo{_{diag, kk}}{\Yacq \given \xacq, \wstar} + \specialHofHessian{_{diag, kk}}{\wstar \given \Dtrain} \right ) + \text{const.}
    \intertext{We use $\log x \le x - 1$ and that $\HofHessian{\wstar \given \Dtrain}$ is constant:}
        &\le \tfrac{1}{2} \sum_k \left ( \specialFisherInfo{_{diag, kk}}{\Yacq \given \xacq, \wstar} + \specialHofHessian{_{diag, kk}}{\wstar \given \Dtrain} \right ) + \text{const.} \\
        &\le \tfrac{1}{2} \sum_k \specialFisherInfo{_{diag, kk}}{\Yacq \given \xacq, \wstar} + \text{const.}
    \intertext{From \Cref{fim:prop:fisher_information_equivalences}, we know that the Fisher information is equivalent to the outer product of the Jacobians: $\FisherInfo{\Yacq \given \xacq, \wstar} = \E{\pof{\yacq \given \xacq, \wstar}}{\HofJacobian{\yacq \given \xacq, \wstar} \, \HofJacobian{\yacq \given \xacq, \wstar}^T}$, and we finally obtain for the diagonal elements:}
        &= \tfrac{1}{2} \sum_k \E*{\pof{\yacq \given \xacq, \wstar}}{\specialHofJacobian{_k}{\yacq \given \xacq, \wstar}^ 2} + \text{const.} \\
        &= \tfrac{1}{2} \E*{\pof{\yacq \given \xacq, \wstar}}{\left \Vert \HofJacobian{\yacq \given \xacq, \wstar} \right \Vert^2} + \text{const.}
    \end{align}
\end{proof}

\subsection{Deep Learning on a Data Diet}
\label{fim:appsec:unify_data_diet}

\propdatadietigconnection*
\begin{proof}
    For any fixed $\wstar$, the IG is equal to the conditional entropy up to a constant term, via \Cref{fim:prop:approximate_ig}:
    \begin{align}
        \MIof{\W; \Yacq \given \xacq} &\overset{\approx}{\le}
        \tfrac{1}{2} \log \det \left ( \HofHessian{\yacq \given \xacq, \wstar} + \HofHessian{\wstar \given \Dtrain} \right ) + \text{const.}
    \intertext{As in the previous proof, we apply a diagonal approximation for the Hessian, noting that the determinant of the diagonal matrix upper-bounds the determinant of the full matrix:}
        &\le \tfrac{1}{2} \log \det \left ( \specialHofHessian{_{diag}}{\yacq \given \xacq, \wstar} + \specialHofHessian{_{diag}}{\wstar \given \Dtrain} \right ) + \text{const.} \\
        &= \tfrac{1}{2} \sum_k \log \left ( \specialHofHessian{_{diag, kk}}{\yacq \given \xacq, \wstar} + \specialHofHessian{_{diag, kk}}{\wstar \given \Dtrain} \right ) + \text{const.}
    \intertext{Again, we use $\log x \le x - 1$ and that $\HofHessian{\wstar \given \Dtrain}$ is constant:}
        &\le \tfrac{1}{2} \sum_k \left ( \specialHofHessian{_{diag, kk}}{\yacq \given \xacq, \wstar} + \specialHofHessian{_{diag, kk}}{\wstar \given \Dtrain} \right ) + \text{const.} \\
        &\le \tfrac{1}{2} \sum_k \specialHofHessian{_{diag, kk}}{\yacq \given \xacq, \wstar} + \text{const.}
    \intertext{From \Cref{fim:lemma:helper_jacobian_hessian_equivalent}, we know that the Hessian is equivalent to the outer product of the Jacobians plus a second-order term: $\HofHessian{\yacq \given \xacq, \wstar} = \HofJacobian{\yacq \given \xacq, \wstar} \, \HofJacobian{\yacq \given \xacq, \wstar}^T - \frac{\nabla_\w^2 \pof{\yacq \given \xacq, \wstar}}{\pof{\yacq \given \xacq, \wstar}}$, and we finally obtain for the diagonal elements:}
        &= \tfrac{1}{2} \sum_k {\specialHofJacobian{_k}{\yacq \given \xacq, \wstar}^ 2} - \tfrac{1}{2} \tr \left ( \frac{\nabla_\w^2 \pof{\yacq \given \xacq, \wstar}}{\pof{\yacq \given \xacq, \wstar}} \right ) + \text{const.} \\
        &= \tfrac{1}{2} {\left \Vert \HofJacobian{\yacq \given \xacq, \wstar} \right \Vert^2} - \tfrac{1}{2} \tr \left ( \frac{\nabla_\w^2 \pof{\yacq \given \xacq, \wstar}}{\pof{\yacq \given \xacq, \wstar}} \right ) + \text{const.}
    \end{align}
    Taking an expectation over $\wstar \sim \qof{\w}$ yields the statement.
\end{proof}

\section{Preliminary Empirical Comparison of Information Quantity Approximations}
\label{fim:appsec:empirical_comparison}

\begin{table}
    \caption{\emph{Spearman Rank Correlation of Prediction-Space and Weight-Space Estimates.}
    BALD and EPIG are both strongly positively rank-correlated with each other. The weight-space approximations are strongly rank-correlated with the prediction-space approximations, but the weight-space approximations are less accurate than the prediction-space approximations. Note that we have reversed the ordering for the proxy objectives for JEPIG and EPIG as they are minimized while EPIG is maximized.}
    \label{fim:table:spearman_rank_correlation}
    \centering
    \renewcommand{\arraystretch}{1.2}
    \resizebox{\linewidth}{!}{%
    \begin{tabular}{lrrrrrrr}
        \hline
        {} & BALD & EIG & EIG & EPIG & EPIG & JEPIG & (J)EPIG \\
         & \multicolumn{1}{l}{(Prediction)} & \multicolumn{1}{l}{(LogDet)} & \multicolumn{1}{l}{(Trace)} & \multicolumn{1}{l}{(Prediction)} & \multicolumn{1}{l}{(LogDet)} & \multicolumn{1}{l}{(LogDet)} & \multicolumn{1}{l}{(Trace)} \\ \hline
        BALD (Prediction) & 1.000 & 0.955 & 0.940 & 0.984 & 0.948 & 0.955 & 0.927 \\
        EPIG (Prediction) & 0.984 & 0.918 & 0.897 & 1.000 & 0.918 & 0.918 & 0.903 \\ \hline
        \end{tabular}}
\end{table}

\begin{figure}
    \begin{minipage}{0.55\linewidth}
        \centering
        \includegraphics[width=\linewidth]{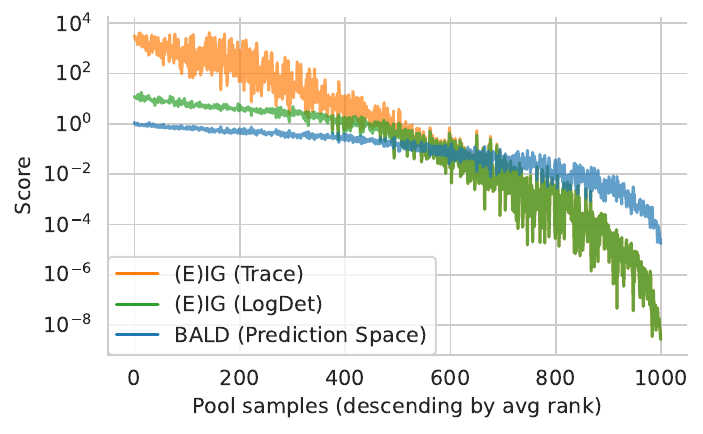}
        \caption{\emph{EIG Approximations.} Trace and $\log \det$ approximations match for small scores (because of \Cref{fim:lemma:log_det_tr_inequality}). They diverge for large scores. Qualitatively, the order matches the prediction-space approximation using BALD with MC dropout.}
        \label{fim:fig:bald_eig_log_det_trace_comparison}
    \end{minipage}
    \begin{minipage}{0.39\linewidth}
        \centering
        \includegraphics[width=\linewidth]{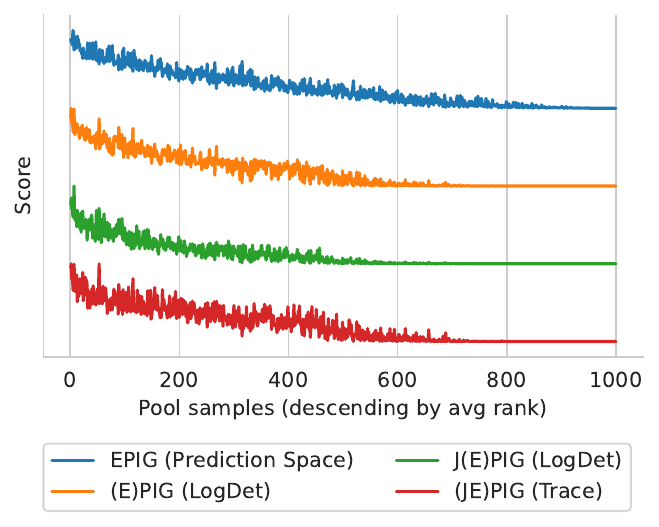}
        \caption{\emph{(J)EPIG Approximations (Normalized).} The scores match qualitatively. Note we have reversed the ordering for the proxy objectives for JEPIG and EPIG as they are minimized while EPIG is maximized.}
        \label{fim:fig:epig_log_det_trace_jepig_qualitative_comparison}
    \end{minipage}
\end{figure}
\begin{figure}
    \centering
    \includegraphics[width=\linewidth]{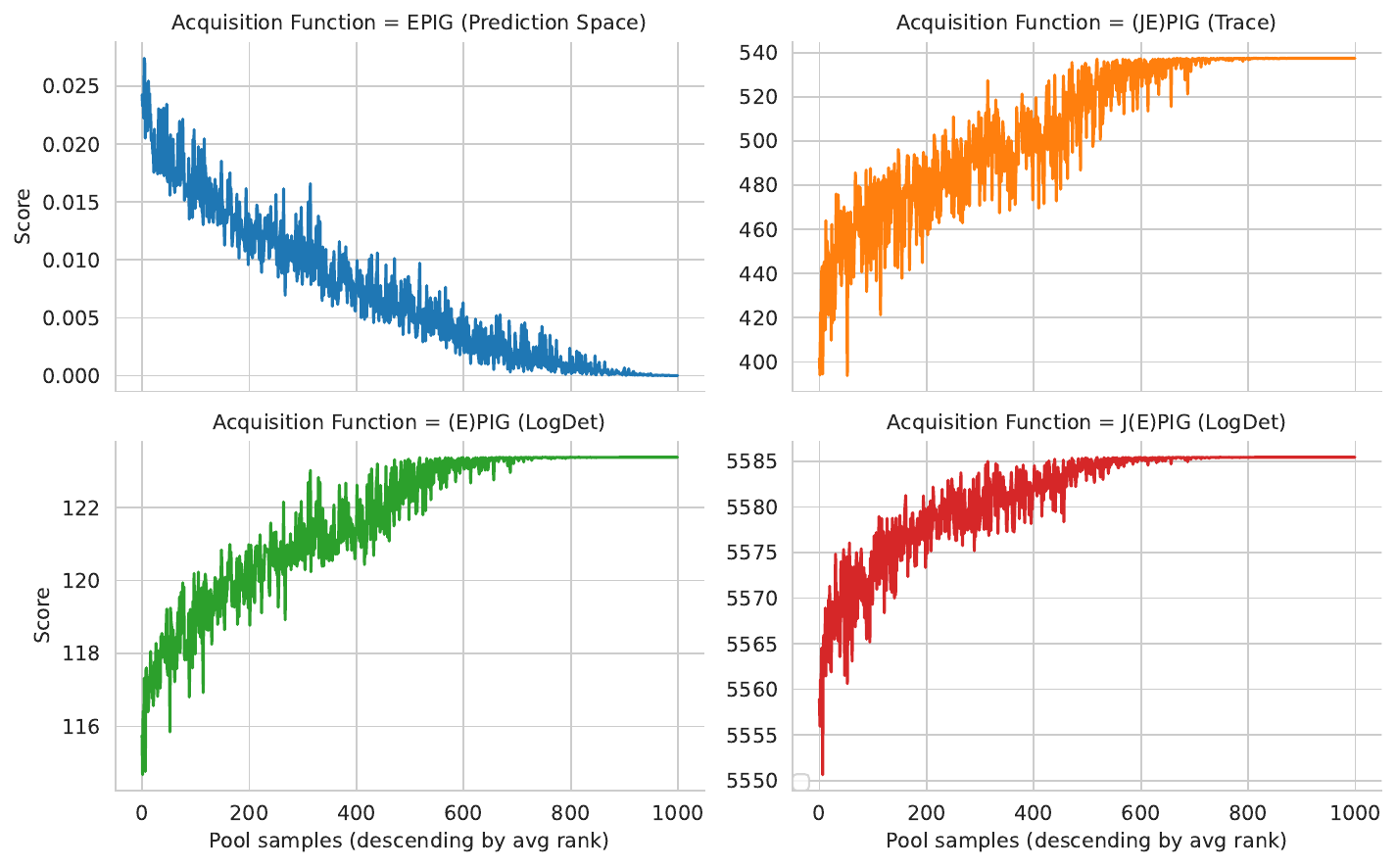}
    \caption{\emph{(J)EPIG Approximations.} The scores match quantitatively. Note the proxy objectives for JEPIG and EPIG are minimized while EPIG is maximized. The value ranges are off by a lot: the true EPIG score is upper bounded by $\log \numclasses \approx 2.3 \, nats$.}
    \label{fim:fig:epig_log_det_trace_jepig_comparison}
\end{figure}

The following section describes an initial empirical evaluation\footnote{Code at: \shorturl{https://github.com/BlackHC/2208.00549}} of the bounds from \S\ref{fim:sec:iq_approximations} on MNIST. We train a model on a subset of MNIST and compare the approximations of EIG and EPIG in weight space to BALD and EPIG computed in prediction space. We use a last-layer approach (GLM) which means that active sampling and active learning approximations are equivalent. We do not attempt to estimate JEPIG in prediction space.

\textbf{Setup.} %
We train a BNN using MC dropout on 80 randomly selected training samples from MNIST, achieving 83\% accuracy. The model architecture follows the one described in \Cref{chapter:batchbald}. We use 100 Monte-Carlo dropout samples \citep{gal2015dropout} to compute the prediction-space estimates.
For EPIG, we sample the evaluation set from the remaining training set (20000 samples). We randomly select 1000 samples from the training set as pool set.
We compute BALD and EPIG in prediction space as described in \citet{gal2017deep} and \Cref{chapter:pobal}. 
For the weight-space approximations, we use a last-layer approximation---we thus have a GLM.
For the implementation, we use \href{fim:https://pytorch.org/}{PyTorch} \citep{pytorchgithub} and the \href{fim:https://aleximmer.github.io/Laplace/}{laplace-torch} library \citep{daxberger2021laplace}.

We chose 80 samples and 83\% accuracy as the accuracy trajectory of BALD and EPIG is steep at this point, see e.g.\ \Cref{chapter:batchbald}, and thus we expect a wider range of scores.

\textbf{Results.} %
In \Cref{fim:fig:bald_eig_log_det_trace_comparison}, we see a comparison of BALD with the approximations in \cref{fim:eq:eig_log_det_approx} and \cref{fim:eq:eig_trace_approx}. Not shown is \cref{fim:eq:approx_conditional_entropy}, which performs like \cref{fim:eq:eig_log_det_approx} (up to a constant).
In \Cref{fim:fig:epig_log_det_trace_jepig_qualitative_comparison} and \Cref{fim:fig:epig_log_det_trace_jepig_comparison}, we show a comparison of EPIG with the approximations in \Cref{fim:eq:epig_fisher_approximation_logdet}, \Cref{fim:eq:epig_fisher_approximation_trace}, and \Cref{fim:eq:jepig_log_det_approx}. \Cref{fim:fig:epig_log_det_trace_jepig_qualitative_comparison} shows normalized scores individually as the score ranges are very different.

Importantly, while the prediction-space scores (BALD and EPIG) have valid scores as the EIG/BALD and EPIG scores of a sample are bounded by the $\log \numclasses$, the weight-space scores are not valid. As such, they only provide rough estimates of the information quantities.

However, as we see in \Cref{fim:table:spearman_rank_correlation}, the Spearman rank correlation coefficients between the weight-space and prediction-space scores are very high. Thus, while the scores themselves are not good estimates, their order seems informative, and this is what matters for selecting acquisition samples in data subset selection.

\textbf{Future Work.} %
The quality of these approximations needs further verification using more complex models and datasets. Comparisons in active learning and active sampling experiments are also necessary to validate the usefulness of these approximations. However, given the connections shown in \S\ref{fim:sec:unification}, we expect that the approximations will be useful in these settings as well.

}

\chapter{\bbbaltitle}
{

\begin{figure}[h]
    \centering
    \includegraphics[width=\linewidth,trim=0 0 0 0.2cm,clip]{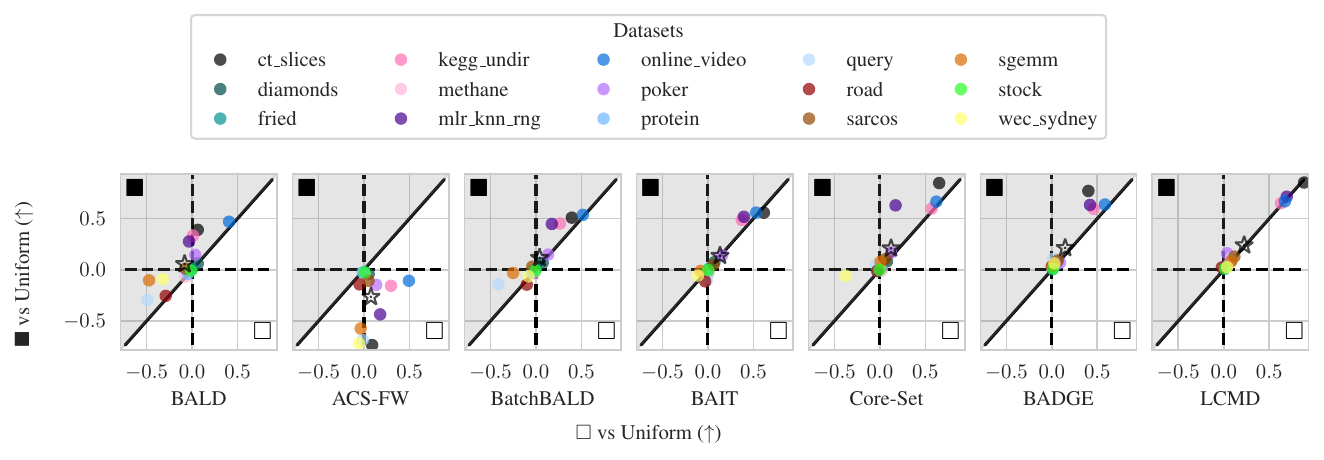}
    \caption{\emph{Final Logarithmic RMSE by regression datasets for DNNs: $\blacksquare$ vs $\square$ (vs Uniform).}
    Across acquisition functions, the performance of black-box methods is highly correlated with the performance of white-box methods, even though black-box methods make fewer assumptions about the model. 
    We plot the improvement of the white-box method ($\square$) over the uniform baseline on the x-axis, so for datasets with markers left of the dashed vertical lines, the white-box method performs better than uniform, and the improvement of the black-box method ($\blacksquare$) over the uniform baseline on the y-axis, so for datasets with markers above the dashed horizontal lines, the black-box method performs better than uniform.
    Similarly, for datasets with markers in the $\blacksquare$ region, the black-box method performs better than the white-box method.
    The average over all datasets is marked with a star $\star$.
    }
    \label{b3al:fig:correlation_between_bb_vs_wb_methods_final}
\end{figure}

\begin{sidewaysfigure}
    \centering
    \begin{subfigure}[b]{0.5\linewidth}
        \centering
        \includegraphics[width=0.9\linewidth,trim=0 0 0 0.2cm,clip]{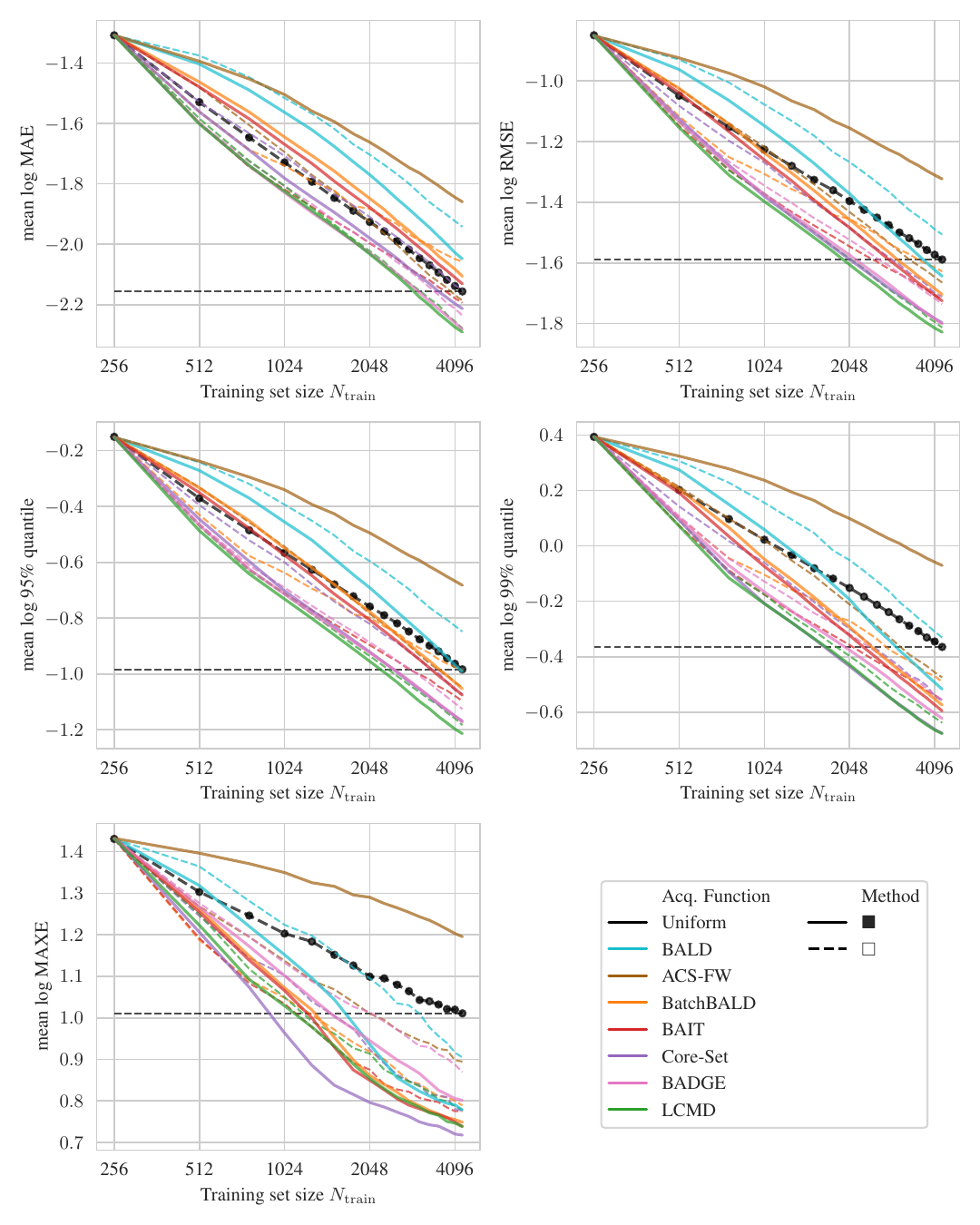}
        \caption{Learning Curves}
        \label{b3al:fig:details:relu:learning_curves}
    \end{subfigure}\hfill%
    \begin{subfigure}[b]{0.5\linewidth}
        \centering
        \includegraphics[width=0.9\linewidth,trim=0 0 0 0.2cm,clip]{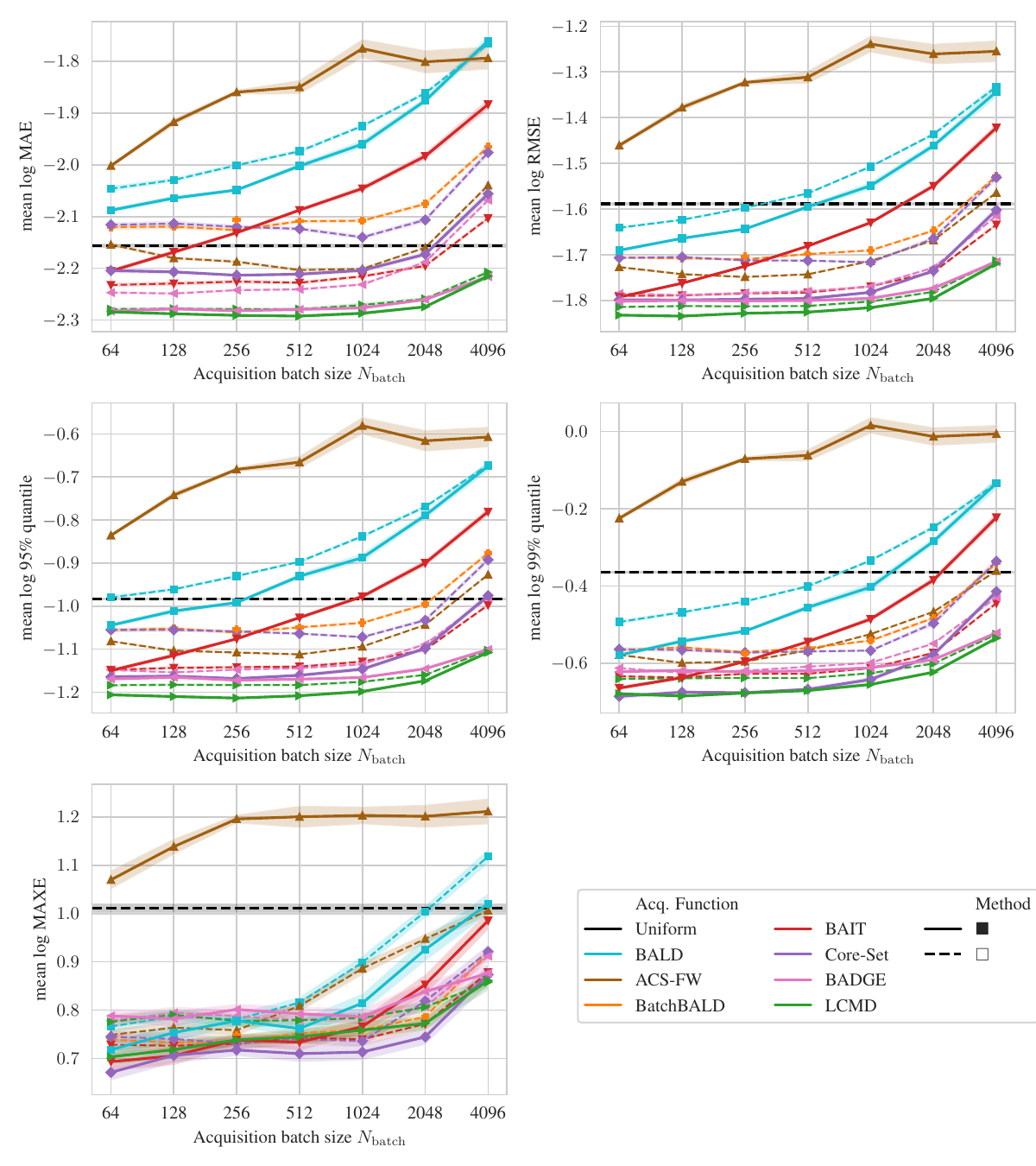}
        \caption{Acquisition Batch Sizes}
        \label{b3al:fig:details:relu:batch_size}
    \end{subfigure}
    \caption{\emph{DNNs: Error Metrics over 15 regression datasets.}
    We report mean absolute error (MAE), root mean squared error (RMSE), 95\% and 99\% quantiles, and the maximum error (MAXE). Averaged over 20 trials.}%
    \label{b3al:fig:details:relu}%
\end{sidewaysfigure}

\begin{sidewaysfigure}
    \centering
    \begin{subfigure}[b]{0.5\linewidth}
        \centering
        \includegraphics[width=0.9\linewidth,trim=0 0 0 0.2cm,clip]{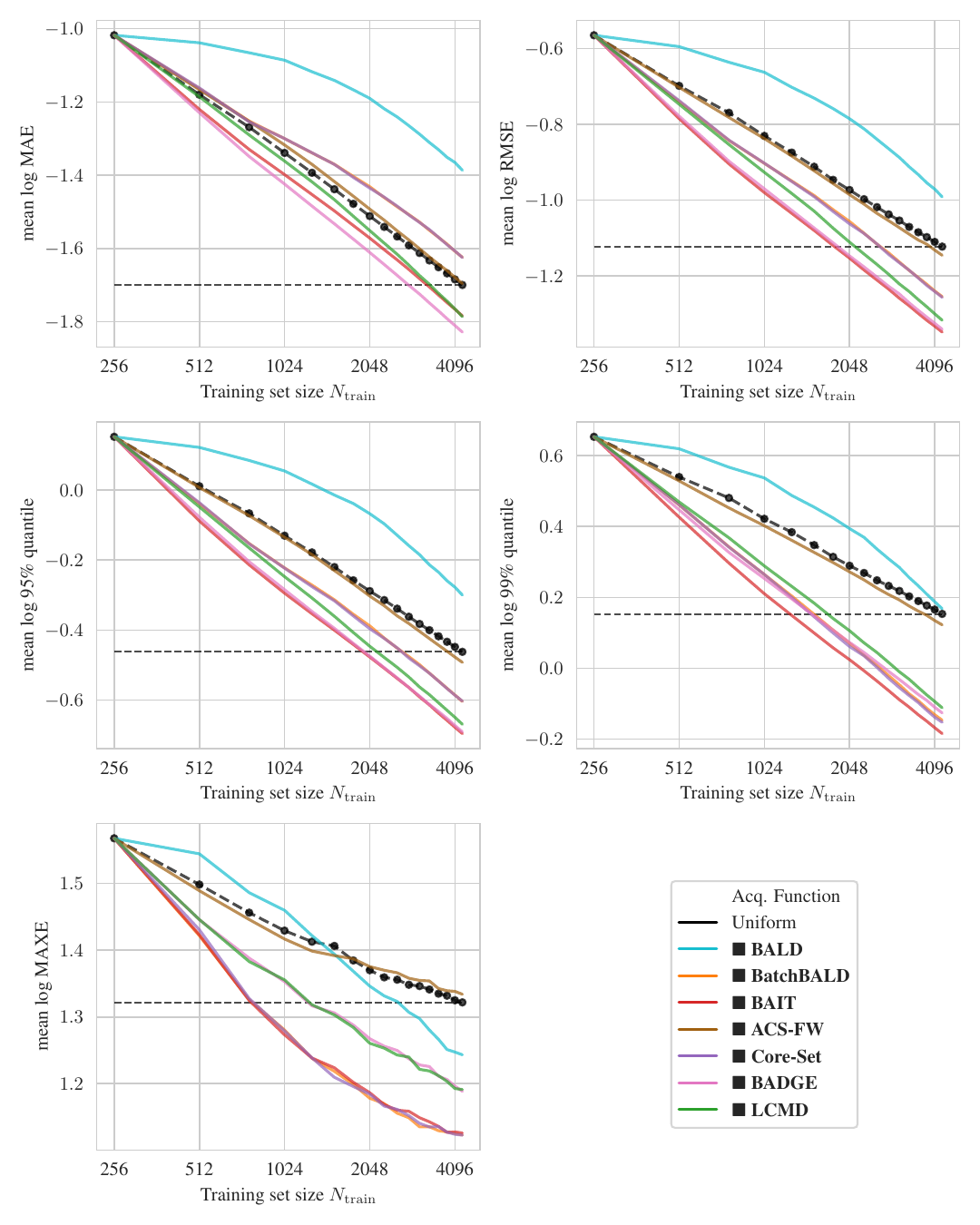}
        \caption{Learning Curves}
        \label{b3al:fig:details:rf:learning_curves}
    \end{subfigure}\hfill
    \begin{subfigure}[b]{0.50\linewidth}
        \centering
        \includegraphics[width=0.9\linewidth,trim=0 0 0 0.2cm,clip]{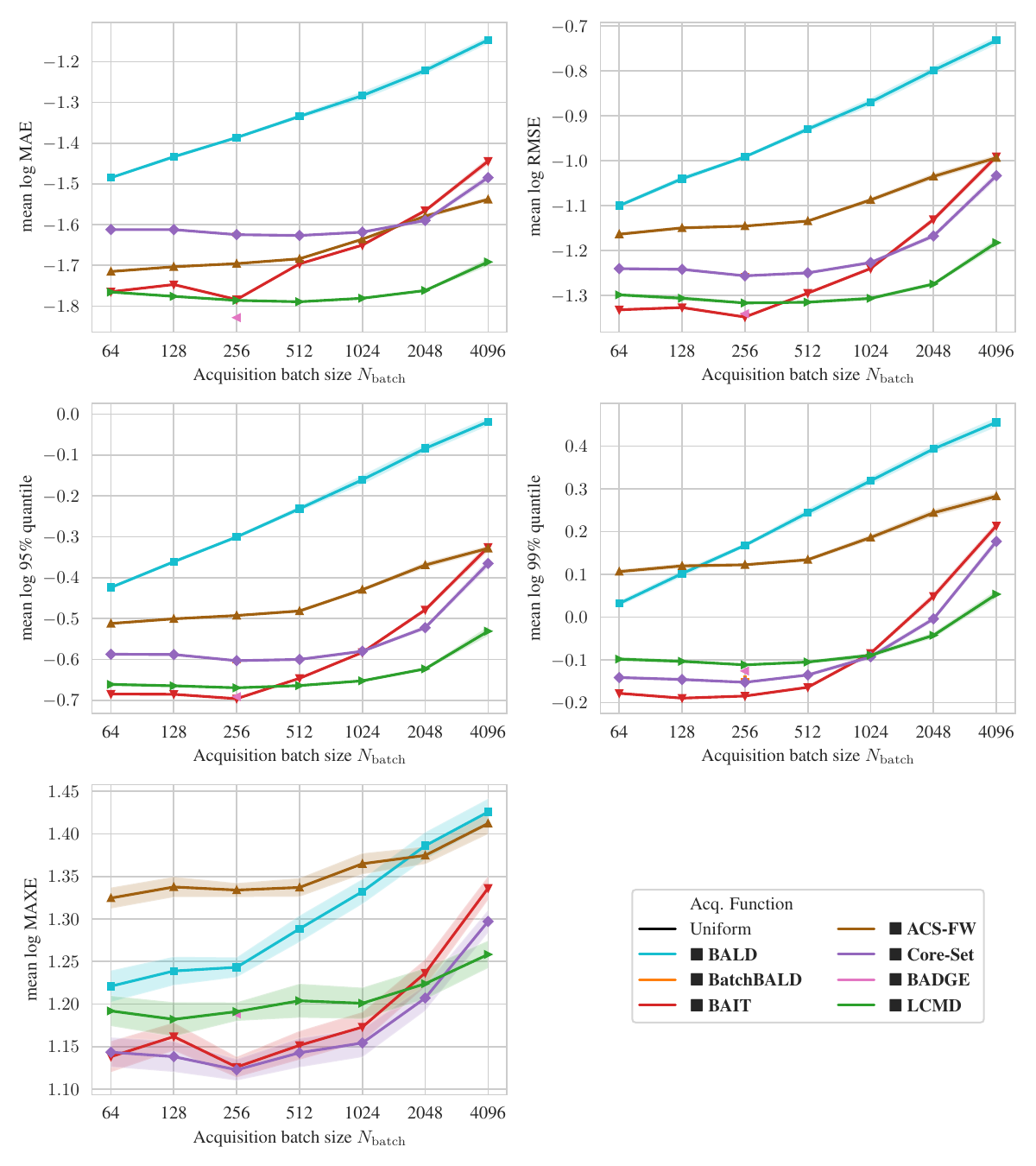}
        \caption{Acquisition Batch Size}
        \label{b3al:fig:details:rf:batch_size}
    \end{subfigure}
    \caption{\emph{Random Forests: Error Metrics over 15 regression datasets (cont'd).}
    We report mean absolute error (MAE), root mean squared error (RMSE), 95\% and 99\% quantiles, and the maximum error (MAXE). Averaged over 20 trials.}%
    \label{b3al:fig:details:rf}%
\end{sidewaysfigure}

\begin{sidewaysfigure}
    \centering
    \begin{subfigure}[b]{0.5\linewidth}
        \centering
        \includegraphics[width=0.9\linewidth,trim=0 0 0 0.2cm,clip]{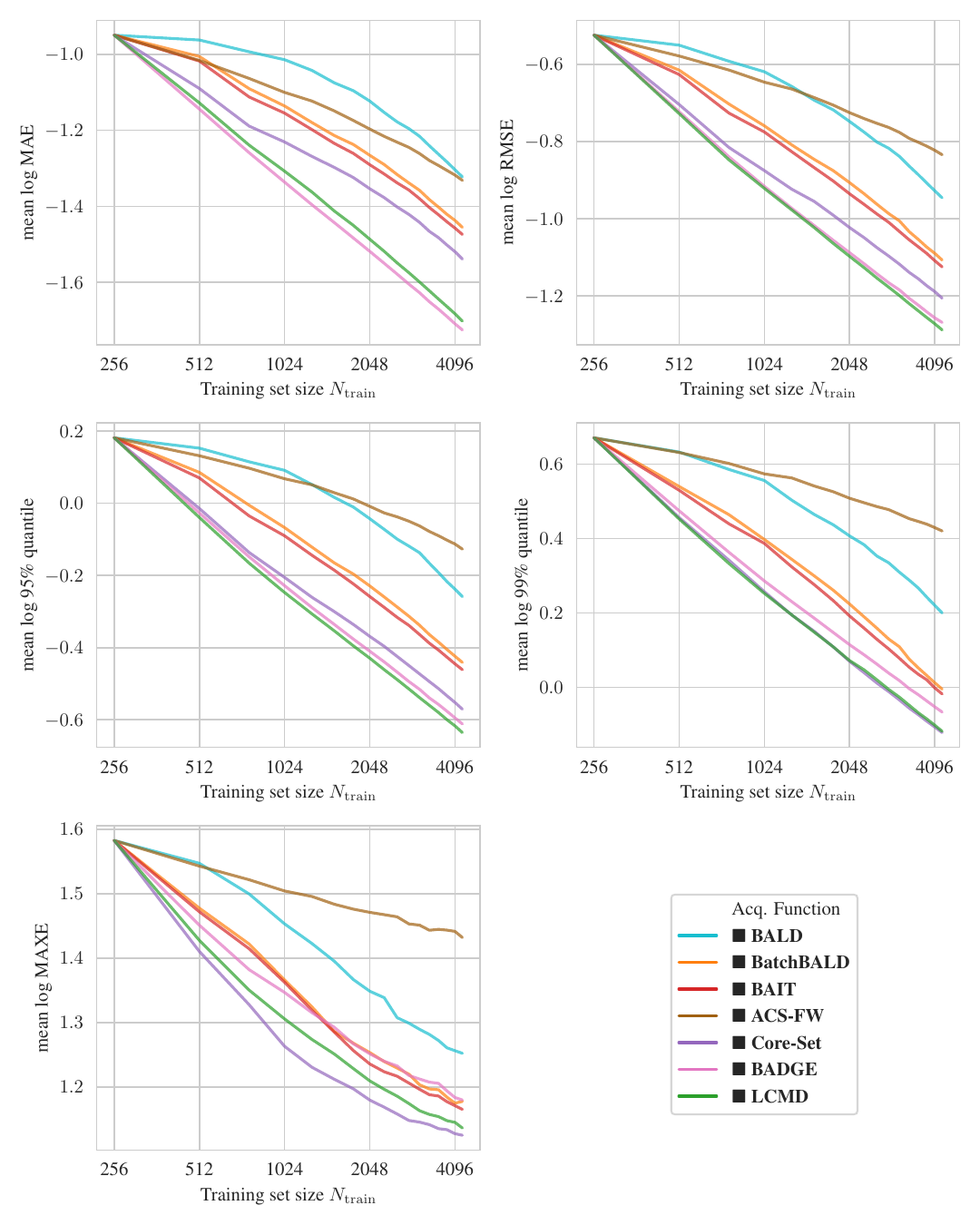}
        \caption{Learning Curves}
        \label{b3al:fig:details:bagging_rf:learning_curves}
    \end{subfigure}\hfill
    \begin{subfigure}[b]{0.50\linewidth}
        \centering
        \includegraphics[width=0.9\linewidth,trim=0 0 0 0.2cm,clip]{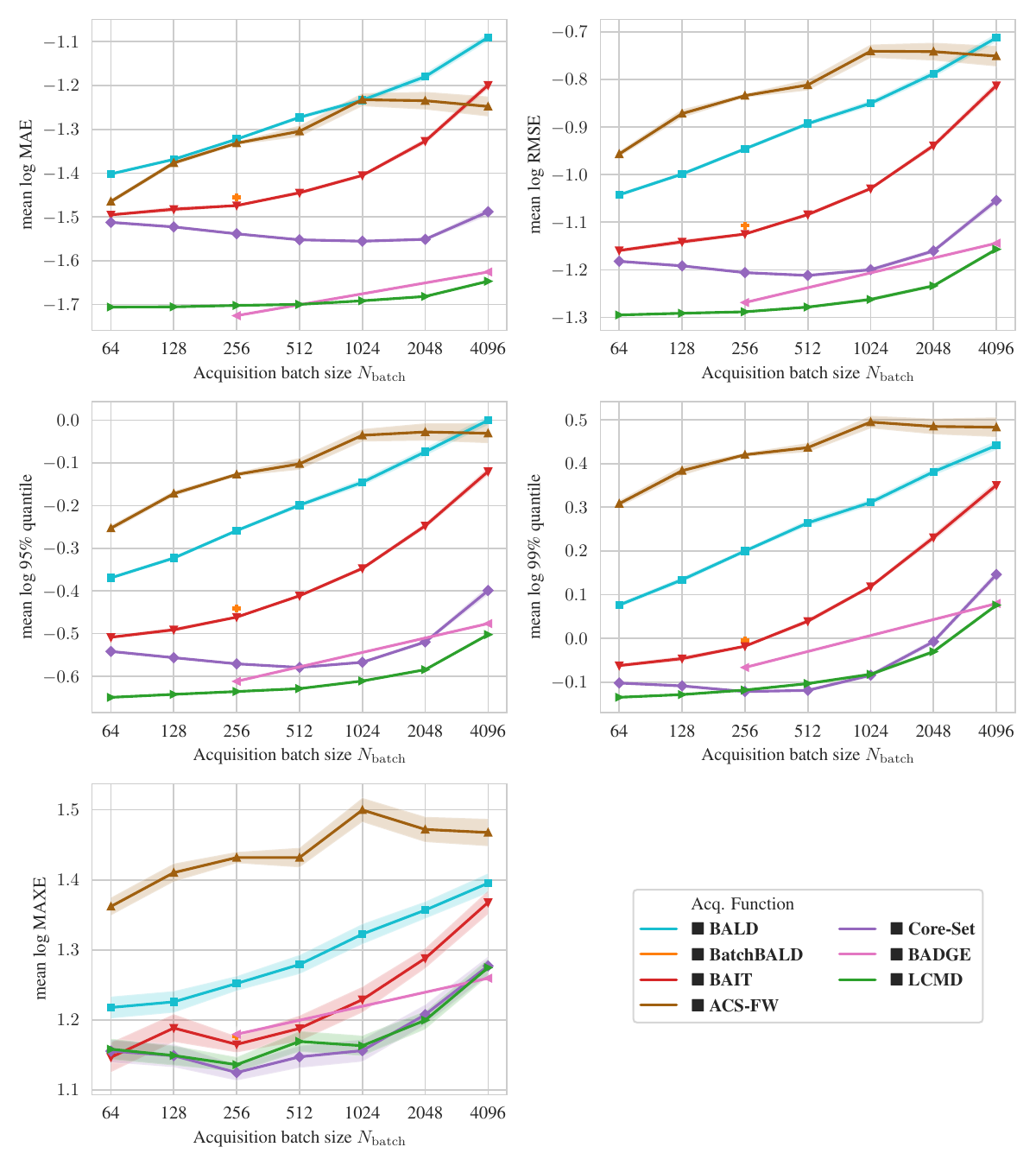}
        \caption{Acquisition Batch Size}
        \label{b3al:fig:details:bagging_rf:batch_size}
    \end{subfigure}
    \caption{\emph{Random Forests (Bagging): Error Metrics over 15 regression datasets (cont'd).}
    We report mean absolute error (MAE), root mean squared error (RMSE), 95\% and 99\% quantiles, and the maximum error (MAXE). Averaged over 20 trials.}%
    \label{b3al:fig:details:bagging_rf}%
\end{sidewaysfigure}

\begin{sidewaysfigure}
    \centering
    \begin{subfigure}[b]{0.5\linewidth}
        \centering
        \includegraphics[width=0.9\linewidth,trim=0 0 0 0.2cm,clip]{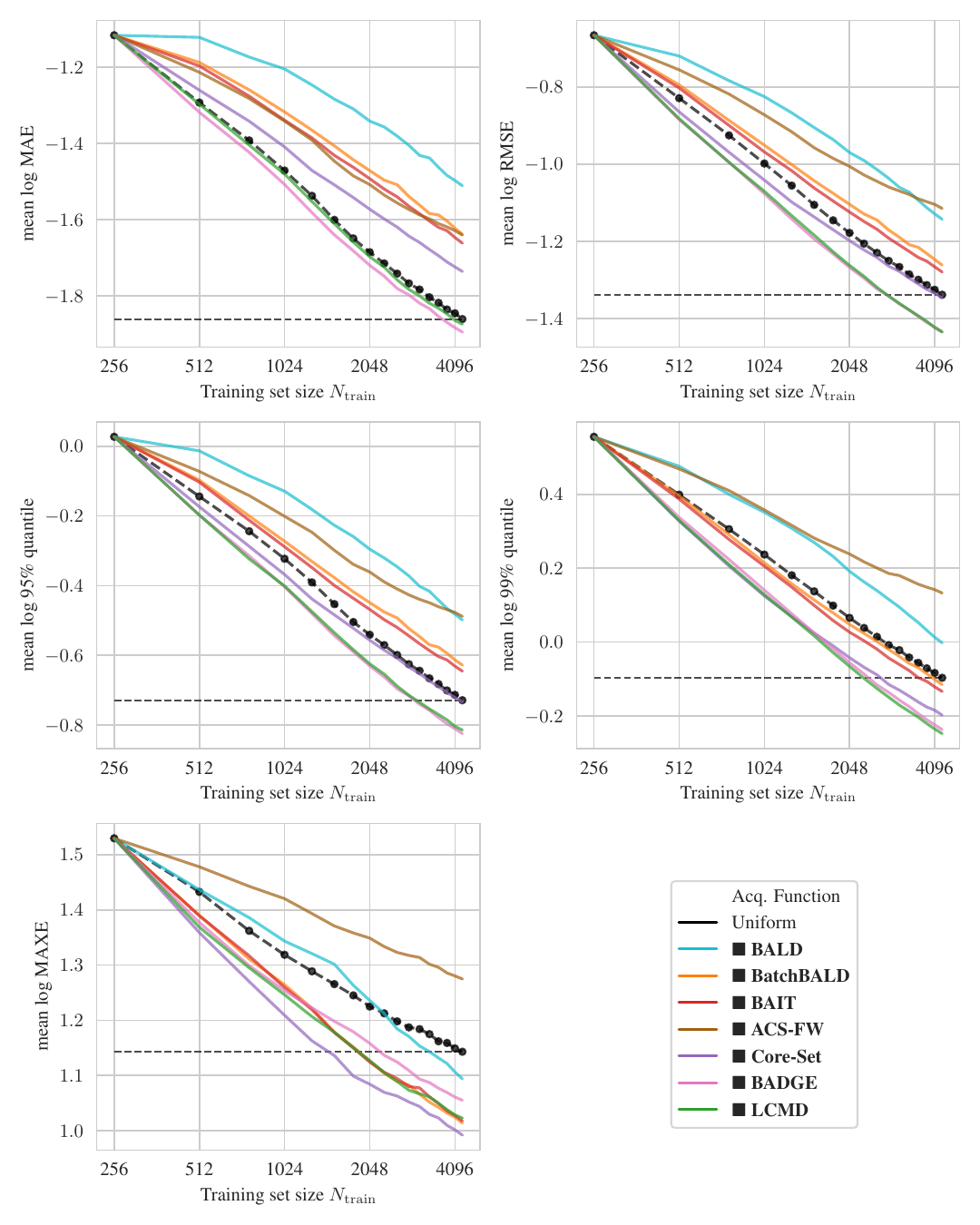}
        \caption{Learning Curves}
        \label{b3al:fig:details:ve-cat:learning_curves}
    \end{subfigure}\hfill
    \begin{subfigure}[b]{0.50\linewidth}
        \centering
        \includegraphics[width=0.9\linewidth,trim=0 0 0 0.2cm,clip]{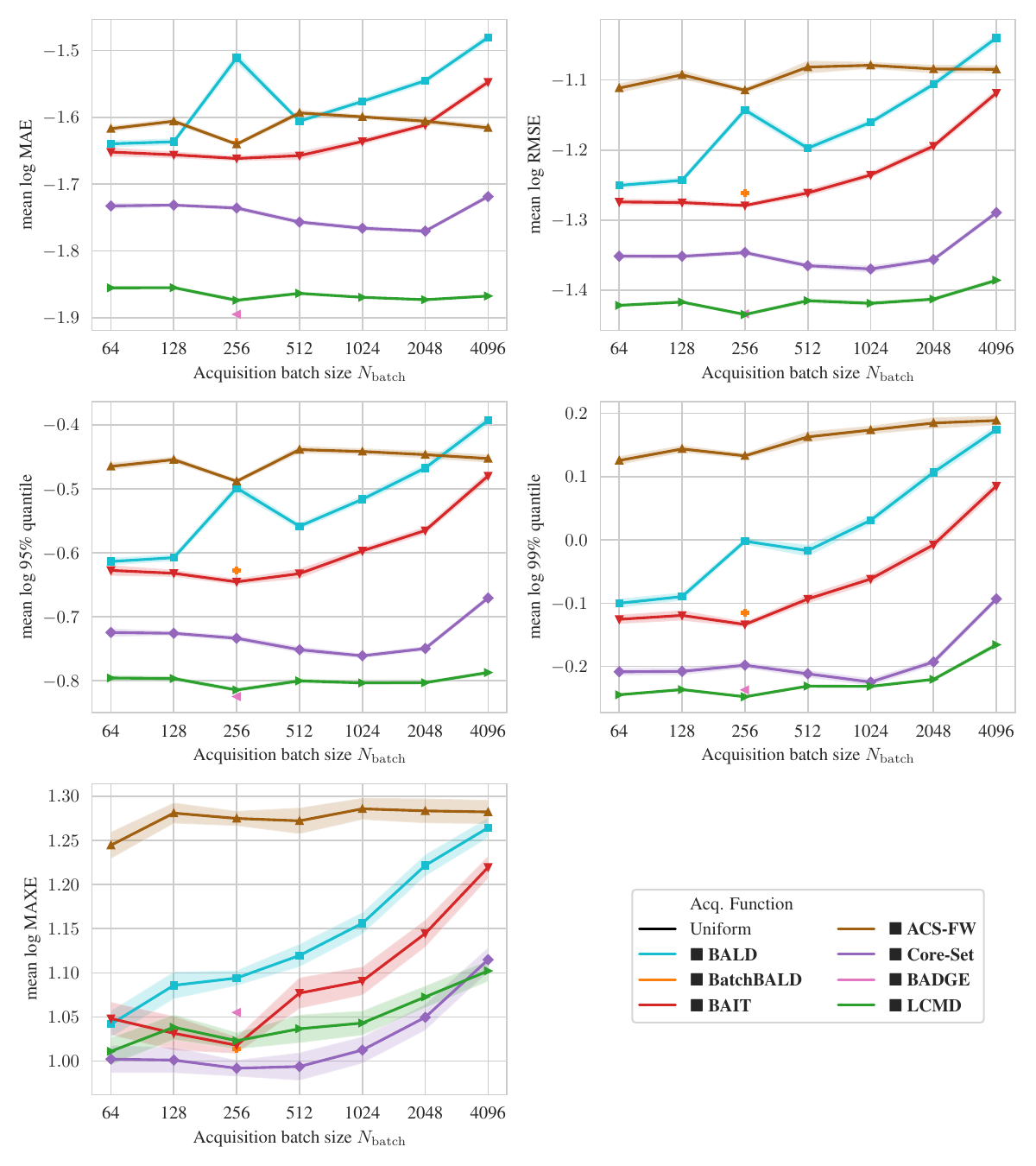}
        \caption{Acquisition Batch Size}
        \label{b3al:fig:details:ve-cat:batch_size}
    \end{subfigure}
    \caption{\emph{Gradient-Boosted Trees with Virtual Ensemble: Error Metrics over 15 regression datasets (cont'd).}
    We report mean absolute error (MAE), root mean squared error (RMSE), 95\% and 99\% quantiles, and the maximum error (MAXE). Averaged over 20 trials.}%
    \label{b3al:fig:details:ve-cat}%
\end{sidewaysfigure}

\begin{table}[h!]
    \begin{minipage}[t]{0.55\linewidth}
        \centering
        \caption{\emph{Average performance of black-box $\blacksquare$ and white-box $\square$ batch active learning  acquisition functions using DNNs.} On average, for five acquisition methods, the black-box method performs better than the white-box method. Cf. \Cref{b3al:fig:correlation_between_bb_vs_wb_methods}, which analyzes the final epoch.}
        \label{b3al:tab:avg_performance_bb_vs_wb_relu}
        \resizebox{\linewidth}{!}{%
            \begin{tabular}{lrrrrr}
\toprule
Acquisition function & MAE & RMSE & 95\% & 99\% & MAXE\\
\midrule
{Uniform} & -1.934 & -1.401 & -0.766 & -0.163 & 1.107 \\
\midrule
\textbf{$\blacksquare$ {BALD}} & -1.794 & -1.389 & -0.713 & -0.221 & 0.946\\
$\square$ {BALD} & -1.722 & -1.285 & -0.614 & -0.077 & 1.080 \\
\midrule
\textbf{$\blacksquare$ {BatchBALD}} & -1.865 & -1.465 & -0.792 & -0.303 & 0.892\\
$\square$ {BatchBALD} & -1.895 & -1.463 & -0.808 & -0.288 & 0.916 \\
\midrule
$\square$ {BAIT} & -1.998 & -1.541 & -0.895 & -0.357 & 0.888\\
\textbf{$\blacksquare$ {BAIT}} & -1.892 & -1.489 & -0.817 & -0.328 & 0.881 \\
\midrule
$\square$ {ACS-FW} & -1.937 & -1.439 & -0.793 & -0.225 & 1.016\\
\textbf{$\blacksquare$ {ACS-FW}} & -1.678 & -1.168 & -0.509 & 0.085 & 1.278 \\
\midrule
\textbf{$\blacksquare$ {Core-Set}} & -1.988 & -1.585 & -0.926 & -0.435 & \textbf{0.831}\\
$\square$ {Core-Set} & -1.923 & -1.490 & -0.831 & -0.307 & 0.929 \\
\midrule
\textbf{$\blacksquare$ {BADGE}} & -2.042 & -1.579 & -0.931 & -0.383 & 0.948\\
$\square$ {BADGE} & -2.007 & -1.530 & -0.895 & -0.329 & 1.008 \\
\midrule
\textbf{$\blacksquare$ {LCMD}} & \textbf{-2.048} & \textbf{-1.609} & \textbf{-0.965} & \textbf{-0.437} & 0.874\\
$\square$ {LCMD} & -2.033 & -1.589 & -0.940 & -0.402 & 0.914\\
\bottomrule
\end{tabular}
        }
    \end{minipage}\hfill
    \begin{minipage}[t]{0.42\linewidth}
        \centering
        \caption{\emph{Average performance of black-box $\blacksquare$ batch active learning  acquisition functions on non-differentiable models.}}
        \begin{subtable}{\linewidth}
            \caption{Random Forests}
            \resizebox{\linewidth}{!}{%
                \begin{tabular}{lrrrrr}
\toprule
Acquisition function & MAE & RMSE & 95\% & 99\% & MAXE\\
\midrule
{Uniform} & -1.516 & -0.975 & -0.293 & 0.290 & 1.376 \\
\midrule
\textbf{$\blacksquare$ {BALD}} & -1.222 & -0.815 & -0.106 & 0.365 & 1.348 \\
\midrule
\textbf{$\blacksquare$ {BatchBALD}} & -1.449 & -1.070 & -0.401 & 0.064 & \textbf{1.195} \\
\midrule
\textbf{$\blacksquare$ {BAIT}} & -1.582 & \textbf{-1.158} & \textbf{-0.485} & \textbf{0.021} & 1.198 \\
\midrule
\textbf{$\blacksquare$ {ACS-FW}} & -1.502 & -0.989 & -0.310 & 0.266 & 1.379 \\
\midrule
\textbf{$\blacksquare$ {Core-Set}} & -1.449 & -1.071 & -0.403 & 0.059 & 1.196 \\
\midrule
\textbf{$\blacksquare$ {BADGE}} & \textbf{-1.618} & -1.150 & -0.480 & 0.070 & 1.273 \\
\midrule
\textbf{$\blacksquare$ {LCMD}} & -1.564 & -1.116 & -0.450 & 0.097 & 1.270\\
\bottomrule
\end{tabular}
            }
        \end{subtable}
        \begin{subtable}{\linewidth}
            \caption{Gradient-Boosted Decision Trees}
            \resizebox{\linewidth}{!}{%
                \begin{tabular}{lrrrrr}
\toprule
Acquisition function & MAE & RMSE & 95\% & 99\% & MAXE\\
\midrule
{Uniform} & -1.675 & -1.172 & -0.533 & 0.068 & 1.232 \\
\midrule
\textbf{$\blacksquare$ {BALD}} & -1.353 & -0.979 & -0.308 & 0.182 & 1.225 \\
\midrule
\textbf{$\blacksquare$ {BatchBALD}} & -1.474 & -1.103 & -0.448 & 0.052 & 1.134 \\
\midrule
\textbf{$\blacksquare$ {BAIT}} & -1.497 & -1.122 & -0.467 & 0.034 & 1.137 \\
\midrule
\textbf{$\blacksquare$ {ACS-FW}} & -1.501 & -1.000 & -0.354 & 0.241 & 1.346 \\
\midrule
\textbf{$\blacksquare$ {Core-Set}} & -1.573 & -1.193 & -0.552 & -0.036 & \textbf{1.100} \\
\midrule
\textbf{$\blacksquare$ {BADGE}} & \textbf{-1.709} & \textbf{-1.259} & \textbf{-0.621} & -0.050 & 1.156 \\
\midrule
\textbf{$\blacksquare$ {LCMD}} & -1.688 & -1.256 & -0.616 & \textbf{-0.061} & 1.132\\
\bottomrule
\end{tabular}
            }
        \end{subtable}
    \end{minipage}
\end{table}

\begin{table}[tb]
    \centering
    \caption{\emph{Overview over the used datasets.} See \citet{ballester-ripoll_sobol_2019, neshat_detailed_2018, graf_2d_2011, shannon_cytoscape_2003, deneke_video_2014, anagnostopoulos_scalable_2018, savva_explaining_2018, friedman_multivariate_1991, slezak_framework_2018}. The second column entries are hyperlinks to the respective web pages. Taken from \citet{holzmuller2022framework}.}
    \label{b3al:table:data_sets_source}
    \resizebox{\linewidth}{!}{%
    \begin{tabular}{lrrrrrrR{0.3\textwidth}}
      \toprule
      Short name & Initial pool set size & Test set size & Number of features & Source & OpenML ID & Full name \\
      \midrule
      sgemm & 192000 & 48320 & 14 & \href{https://archive.ics.uci.edu/ml/datasets/SGEMM+GPU+kernel+performance}{UCI} & & SGEMM GPU kernel performance \\
      wec\_sydney & 56320 & 14400 & 48 & \href{https://archive.ics.uci.edu/ml/datasets/Wave+Energy+Converters}{UCI} & & Wave Energy Converters \\
      ct\_slices & 41520 & 10700 & 379 & \href{https://archive.ics.uci.edu/ml/datasets/Relative+location+of+CT+slices+on+axial+axis}{UCI} & & Relative location of CT slices on axial axis \\
      kegg\_undir & 50407 & 12921 & 27 & \href{https://archive.ics.uci.edu/ml/datasets/KEGG+Metabolic+Reaction+Network+\%28Undirected\%29}{UCI} & & KEGG Metabolic Reaction Network (Undirected) \\
      online\_video & 53748 & 13756 & 26 & \href{https://archive.ics.uci.edu/ml/datasets/Online+Video+Characteristics+and+Transcoding+Time+Dataset}{UCI} & & Online Video Characteristics and Transcoding Time \\
      query & 158720 & 40000 & 4 & \href{https://archive.ics.uci.edu/ml/datasets/Query+Analytics+Workloads+Dataset}{UCI} & & Query Analytics Workloads \\
      poker & 198720 & 300000 & 95 & \href{https://archive.ics.uci.edu/ml/datasets/Poker+Hand}{UCI} & & Poker Hand \\
      road & 198720 & 234874 & 2 & \href{https://archive.ics.uci.edu/ml/datasets/3D+Road+Network+\%28North+Jutland\%2C+Denmark\%29}{UCI} & & 3D Road Network (North Jutland, Denmark) \\
      mlr\_knn\_rng & 88123 & 22350 & 132 & \href{https://www.openml.org/d/42454}{OpenML} & 42454 & mlr\_knn\_rng \\
      fried & 31335 & 8153 & 10 & \href{https://www.openml.org/d/564}{OpenML} & 564 & fried \\
      diamonds & 41872 & 10788 & 29 & \href{https://www.openml.org/d/42225}{OpenML} & 42225 & diamonds \\
      methane & 198720 & 300000 & 33 & \href{https://www.openml.org/d/42701}{OpenML} & 42701 & Methane \\
      stock & 45960 & 11809 & 9 & \href{https://www.openml.org/d/1200}{OpenML} & 1200 & BNG(stock) \\
      protein & 35304 & 9146 & 9 & \href{https://www.openml.org/d/42903}{OpenML} & 42903 & physicochemical-protein \\
      sarcos & 34308 & 8896 & 21 & \href{http://www.gaussianprocess.org/gpml/data/}{GPML} & & SARCOS data \\
      \bottomrule
    \end{tabular}
    }
\end{table}

}

\clearpage

\renewcommand{\chapterheadstartvskip}{\vspace*{-30pt}}
\bibliography{batchbald,epig,stochacq,rholoss,causalbald,ddu,practical_it,fim,pobal,dldd,obi,b3al,active_sampling,thesis,jiang,bayesopt}

\end{document}